\documentclass[onecolumn, 12 pt, singlespace, fullpage, a4paper]{report}

\makeatletter
\AtBeginDocument{%
  \let\@biblabel\NAT@biblabelnum
  \let\@bibsetup\NAT@bibsetnum
}
\makeatother

\usepackage{amssymb}
\usepackage{amsthm}
\usepackage[cmex10]{amsmath}
\usepackage{adjustbox} 
\usepackage{academicons}
\usepackage{breakcites} 
\usepackage[round,authoryear]{natbib}
\usepackage{color}
\usepackage{epsfig}
\usepackage{epstopdf}
\usepackage{etoolbox}
\usepackage{enumitem}
\usepackage{float}
\usepackage{fancyhdr}
\usepackage{graphicx}
\usepackage{graphics}
\usepackage{latexsym,amsfonts}
\usepackage{longtable}
\usepackage{listings}
\usepackage{lscape} 
\usepackage{lipsum}
\usepackage{multirow}             
\usepackage{pdfpages}
\usepackage[figuresright]{rotating} 
\usepackage{sectsty}
\usepackage{setspace}
\usepackage{subcaption}
\usepackage{textcomp}
\usepackage{url} 
\usepackage{wrapfig} 
\usepackage{wasysym} 
\usepackage{xcolor}

\usepackage{algorithm}
\usepackage{algorithmic}

\usepackage{tcolorbox}
\usepackage{pifont}
\definecolor{mydarkgreen}{RGB}{39,130,67}
\definecolor{mydarkred}{RGB}{192,25,25}
\newcommand{\green}{\color{mydarkgreen}}
\newcommand{\red}{\color{mydarkred}}
\newcommand{\cmark}{\green\ding{51}}%
\newcommand{\xmark}{\red\ding{55}}%

\usepackage{xspace}
\usepackage{bm}

\usepackage{relsize}
\newcommand{\algname}[1]{{\sf\green\relscale{1.}#1}\xspace}

\DeclareMathOperator{\prox}{prox}
\DeclareMathOperator*{\argmin}{argmin}

\DeclareMathOperator{\range}{range}
\def\<#1,#2>{\langle #1,#2\rangle}

\usepackage[colorinlistoftodos,bordercolor=orange,backgroundcolor=orange!20,linecolor=orange,textsize=scriptsize]{todonotes}

 \newcommand{\squeeze}{} 

\newcommand{\E}[1]{\mathbb{E}\left[#1\right]}

\newcommand{\cL}{\mathcal{L}}
\newcommand{\cO}{\mathcal O}
\newcommand{\cC}{\mathcal C}

\newcommand{\mI}{\mathbf{I}}

\newcommand{\mL}{\mathbf{L}}
\newcommand{\mW}{\mathbf{W}}

\newcommand{\cD}{\mathcal D}
\newcommand{\cQ}{\mathcal Q}
\newcommand{\cS}{\mathcal S}

\usepackage{mathtools}
\newcommand{\eqdef}{\coloneqq}

\usepackage{units}

\newcommand{\MM}{W}
\newcommand{\Exp}[1]{\mathbb{E}\left[#1\right]}
\newcommand{\norm}[1]{\left\|#1\right\|}

\usepackage{threeparttable}
\usepackage{booktabs}
\usepackage{makecell}
\usepackage{tabulary}
\usepackage{colortbl}
\definecolor{bgcolor}{rgb}{0.94,0.97,1}
\definecolor{bgcolor2}{rgb}{0.8,1,0.8}
\definecolor{bgcolor3}{rgb}{0.50,0.90,0.50}

\usepackage{backref}

\usepackage[pagebackref=true,backref=page, hidelinks]{hyperref}
\renewcommand*{\backrefalt}[4]{%
    \ifcase #1 \footnotesize{(Not cited.)}%
    \or        \footnotesize{(Cited on page~#2)}%
    \else      \footnotesize{(Cited on pages~#2)}%
    \fi}

\usepackage{cleveref}

\newtheorem{assumption}{Assumption}
\newtheorem{lemma}{Lemma}
\newtheorem{theorem}{Theorem}
\newtheorem{corollary}{Corollary}
\newtheorem{definition}{Definition}
\newtheorem{remark}{Remark}
\newtheorem*{lemma_empt}{Lemma} 
\newtheorem*{theorem_empt}{Theorem} 

\newcommand{\R}{\mathbb{R}}

\definecolor{algcolor}{RGB}{150,80,0}
\definecolor{mydarkgreen}{RGB}{0,160,0}
\definecolor{mydarkred}{RGB}{170,20,20}
\definecolor{mydarkblue}{RGB}{20,20,170}
\newcommand{\mygreen}{\color{mydarkgreen}}
\newcommand{\myred}{\color{mydarkred}}
\newcommand{\myblue}{\color{mydarkblue}}
\usepackage{multirow}
\usepackage{colortbl}
\definecolor{bgcolor}{rgb}{0.8,1,1}
\definecolor{bgcolor2}{rgb}{0.8,1,0.8}

\usepackage{arydshln}
\newcommand{\cstep}{{\myred\gamma}} 
\newcommand{\cstepsquared}{{\myred\gamma^2}} 
\newcommand{\cstepcubed}{{\myred\gamma^3}} 
\newcommand{\sstep}{{\mygreen\eta}} 
\newcommand{\sstepsquared}{{\mygreen\eta^2}} 
\newcommand\ec[2][]{\ensuremath{\mathbb{E}_{#1} \left[#2\right]}}
\newcommand\ecn[2][]{\ec[#1]{\norm{#2}^2}}

\newcommand{\Kx}{\localsteps}
\newcommand{\Cx}{\kcohort}
\newcommand{\Mx}{\nclients}
\newcommand{\Tx}{\comm}

\newcommand{\gammaM}{{\gamma}}
\newcommand{\tauM}{{\tau}}

\newcommand{\cG}{{\cal G}}

\newcommand{\nclients}{{M}}
\newcommand{\iclient}{m}
\newcommand{\kstep}{t}
\newcommand{\kcohort}{{C}}

\newcommand{\Koper}{{{H}}}
\newcommand{\localsteps}{K}
\newcommand{\comm}{T}
\newcommand{\betaM}{\delta}
\newcommand{\deltaM}{\alpha}

\newcommand{\localsolver}{\mathcal{A}}

\newcommand{\localsolmk}{y_{\iclient}^{\star,\kstep}}
\newcommand{\lastlocittermk}{y_{\iclient}^{\localsteps,\kstep}}
\newcommand{\lastlocitterk}{y^{\localsteps,\kstep}}
\newcommand{\localsolk}{y^{\star,\kstep}}

\newcommand{\set}{S^{t}}
\newcommand{\Exps}[1]{\mathbb{E}_S\!\left[ #1 \right]}

\newcommand{\localfun}{\psi^t}
\newcommand{\localfuni}{\localfun_\iclient}

\newcommand{\cN}{{\cal N}}
\newcommand{\cP}{{\cal P}}

\def\la{\langle}
\def\ra{\rangle}

\newcommand{\PP}{\operatorname{Prob}}

\def\clip{\texttt{clip}}

\def\deltar{\delta_{\text{real}}}

\newcommand{\mA}{{\bf A}}

\newcommand{\rb}[1]{\left(#1\right)}

\newcommand{\sqn}[1]{{\left\lVert#1\right\rVert}^2}

\newcommand\br[1]{\left ( #1 \right )}
\newcommand\ev[1]{\left \langle #1 \right \rangle}

\newcommand{\sqnorm}[1]{\left\| #1 \right\|^2}
\newcommand{\EE}[2]{\mathbb{E}_{#1}\!\left[ #2 \right]}
\newcommand{\Rop}{\mathcal{P}}

\newcommand{\oma}{\omega_{\mathrm{ran}}}

\newcommand{\myref}[1]{$\left(\ref{#1}\right)$} 
\def\cB{{\mathcal{B}}}
\def\cA{{\mathcal{A}}}

\DeclareMathOperator*{\argmax}{arg\,max}

\def\2{$^2$}			 
\def\3{$^3$}			 
\def\-2{$^{-2}$}		 
\def\-3{$^{-3}$}		 
\def\-1{$^{-1}$}		 

\def\la{\langle}
\def\ra{\rangle}

\chapterfont{\fontsize{14}{15}\selectfont}   
\sectionfont{\fontsize{14}{15}\selectfont}
\subsectionfont{\fontsize{14}{15}\selectfont}
\subsubsectionfont{\fontsize{14}{15}\selectfont}

\fancyheadoffset[L]{0.01mm}

\makeatletter
\patchcmd{\@makechapterhead}{50\p@}{20pt}{}{}
\patchcmd{\@makeschapterhead}{50\p@}{20pt}{}{}
\makeatother

\fancypagestyle{plain}{
\fancyhf{} 
\fancyhead[C]{\thepage} 

}

   \usepackage{ifthen,xkeyval,xfor,amsgen}
   \usepackage[acronym,toc, nogroupskip]{glossaries}
   \newglossary[slg]{symbols}{syi}{sbl}{List of Symbols}
   \newglossary[alg-glg]{algorithms}{alg-syi}{alg-sbl}{List of Algorithms}
 
   \makeglossaries

\newglossaryentry{symb:Pi}{
name=$\pi$, type=symbols,
description=A mathematical constant whose value is the ratio of any circle's circumference to its diameter,
sort=symbolpi
}

\newglossaryentry{symb:Phi}{
name=$\varphi$, type=symbols,
description=An angle,
sort=symbolphi
}

\newglossaryentry{symb:Lambda}{
name=$\lambda$, type=symbols,
description=Lambda indicates usually an eigenvalue in linear algebra,
sort=symbollambda
}

\newacronym{toc}{ToC}{Table of Contents}
\newacronym{los}{LoS}{List of Symbols}
\newacronym{loa}{LoA}{List of Abbreviations}
\newacronym{phd}{PhD}{Doctoral}
\newacronym{MS}{MS}{Masters}
\newacronym{CD}{CD}{Compact Disc}
\newacronym{kaust}{KAUST}{King Abdullah University of Science and Technology}

\newacronym{AD}{AD}{Active Directory\protect\glsadd{glos:AD}}

\newacronym{ALIE}{ALIE}{A Little is Enough (a type of Byzantine attack)}
\newacronym{ARAgg}{ARAgg}{Agnostic Robust Aggregator}
\newacronym{BF}{BF}{Bit Flipping (a type of Byzantine attack)}
\newacronym{LF}{LF}{Label Flipping (a type of Byzantine attack)}
\newacronym{CC}{CC}{Compressed Communication}
\newacronym{CM}{CM}{Coordinate Median}
\newacronym{PP}{PP}{Partial Participation}
\newacronym{DL}{DL}{deep learning}
\newacronym{ML}{ML}{machine learning}
\newacronym{DS}{DS}{Data Sampling}
\newacronym{ERM}{ERM}{Empirical Risk Minimization}
\newacronym{FL}{FL}{federated learning}
\newacronym{FPFT}{FPFT}{Full-Parameter Fine-Tuning}

\newacronym{LT}{LT}{Local Training}
\newacronym{MLP}{MLP}{Multilayer Perceptron}
\newacronym{PEFT}{PEFT}{Parameter-Efficient Fine-Tuning}
\newacronym{PL}{PL}{Polyak-{\L}ojasiewicz}

\newacronym{RAgg}{RAgg}{Robust Aggregator}
\newacronym{RFA}{RFA}{Robust Federated Averaging}

\newacronym{SHB}{SHB}{Shift-Back (a type of Byzantine attack)}
\newacronym{VR}{VR}{Variance Reduction}

\newacronym[type=algorithms]{GD}{\algname{GD}}{Gradient Descent}
\newacronym[type=algorithms]{LGD}{\algname{LGD}}{Local Gradient Descent}
\newacronym[type=algorithms]{LoRA}{\algname{LoRA}}{Low-Rank Adaptation}
\newacronym[type=algorithms]{LSGD}{\algname{LSGD}}{Local Stochastic Gradient Descent}
\newacronym[type=algorithms]{SVRG}{\algname{SVRG}}{Stochastic Variance Reduced Gradient}
\newacronym[type=algorithms]{L-SVRG}{\algname{L-SVRG}}{Loopless Stochastic Variance Reduced Gradient}
\newacronym[type=algorithms]{RR}{\algname{RR}}{Random Reshuffling}
\newacronym[type=algorithms]{SGD}{\algname{SGD}}{Stochastic Gradient Descent}
\newacronym[type=algorithms]{ProxGD}{\algname{ProxGD}}{Proximal Gradient Descent}
\newacronym[type=algorithms]{RAC-LoRA}{\algname{RAC-LoRA}}{Randomized Asymmetric Chain of LoRA}
\newacronym[type=algorithms]{ProxSkip}{\algname{ProxSkip}}{Proximal Skipping}
\newacronym[type=algorithms]{SProxSkip}{\algname{SProxSkip}}{Stochastic Proximal Skipping}
\newacronym[type=algorithms]{ProxSkip-VR}{\algname{ProxSkip-VR}}{Variance Reduced Proximal Skipping}
\newacronym[type=algorithms]{SplitSkip}{\algname{SplitSkip}}{Primal Dual Splitting Version of ProxSkip}
\newacronym[type=algorithms]{5GCS}{\algname{5GCS}}{5th Generation Client Sampling}
\newacronym[type=algorithms]{DIANA}{\algname{DIANA}}{Distributed Optimization with Compressed Gradient Differences}
\newacronym[type=algorithms]{DIANA-RR}{\algname{DIANA-RR}}{DIANA with Random Reshuffling}
\newacronym[type=algorithms]{DIANA-NASTYA}{\algname{DIANA-NASTYA}}{NASTYA with Compressed Gradient Differences}
\newacronym[type=algorithms]{Q-RR}{\algname{Q-RR}}{Quantized Random Reshuffling}
\newacronym[type=algorithms]{Q-NASTYA}{\algname{Q-NASTYA}}{Quantized NASTYA}
\newacronym[type=algorithms]{NASTYA}{\algname{NASTYA}}{Federated Optimization with Server Stepsize, Random Shuffling and Partial Participation}
\newacronym[type=algorithms]{Byz-VR-MARINA}{\algname{Byz-VR-MARINA}}{Byzantine Robust Variance Reduced MARINA}
\newacronym[type=algorithms]{Byz-VR-MARINA-PP}{\algname{Byz-VR-MARINA-PP}}{Byzantine Robust Variance Reduced MARINA with Partial Participation}
\newacronym[type=algorithms]{ProxSkip-HUB}{\algname{ProxSkip-HUB}}{ProxSkip for Hierarchical Architecture}
\newacronym[type=algorithms]{ProxSkip-LSVRG}{\algname{ProxSkip-LSVRG}}{ProxSkip with Loopless SVRG}
\newacronym[type=algorithms]{Fed-RAC-LoRA}{\algname{Fed-RAC-LoRA}}{Federated Randomized Asymmetric Chain of LoRA}

\newglossaryentry{glos:AD}{
name=Active Directory,
description={Active Directory is the directory service for Windows based networks, that allows central organization and administration of any network resource. It allows a single-sign-on concept independent from network topologies or network protocols. As a prerequisite you need a Windows Server acting as Domain Controller. This computer stores all necessary data, e.g.~usernames and corresponding passwords}
}

\newglossaryentry{glos:RespF}{name=response file, description={A file 
that allows unattended software installation}}

\usepackage[lmargin=40mm, rmargin=25mm, vmargin=25mm, headsep=2.5mm]{geometry}
\newcommand{\mathsym}[1]{{}}
\newcommand{\unicode}[1]{{}}
\renewcommand{\thechapter}{\arabic{chapter}}
\renewcommand\bibname{\centering BIBLIOGRAPHY}
\newcommand{\orcid}{\includegraphics[width=8pt]{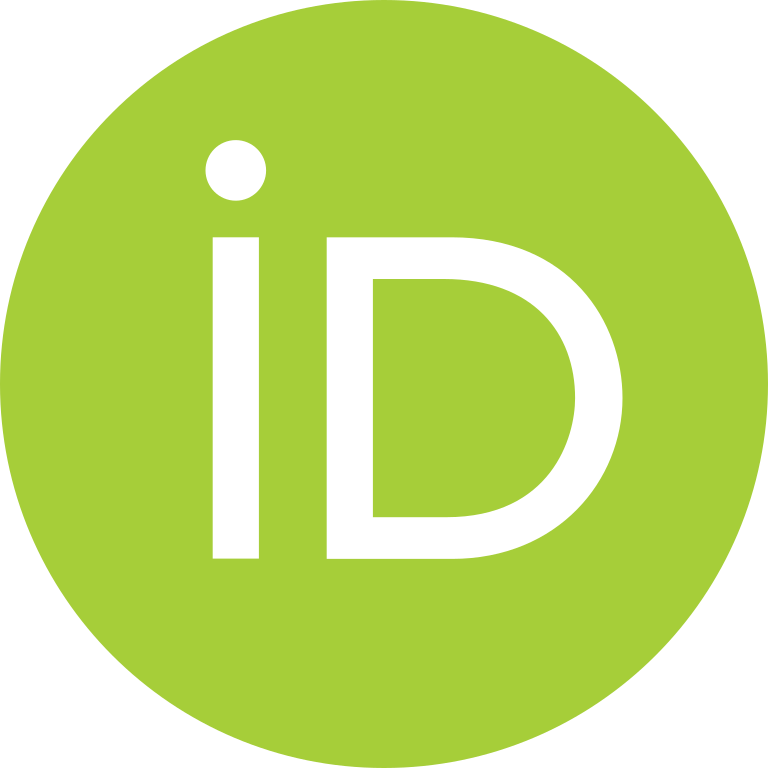}} 

\begin{document}


\thispagestyle{empty}
\addvspace{5mm}  


\begin{center}
\begin{doublespace}
{\textbf{{\large Theoretical Foundations of Communication-Efficient, Robust, and Practical Distributed and Federated Optimization}}}
\end{doublespace}

\vspace{10mm}
{Dissertation by}\\
{Grigorii Malinovskii} 

\vspace{30mm}

{ In Partial Fulfillment of the Requirements}\\[12pt]
{ For the Degree of}\\[12pt]
{Doctor of Philosophy} \vfill
{King Abdullah University of Science and Technology }\\
{Thuwal, Kingdom of Saudi Arabia}
\vfill

\begin{onehalfspace}
{\copyright June, 2026}\\
Grigorii Malinovskii\\               
All rights reserved\\

\orcid{} \small \href{https://orcid.org/0000-0001-2345-6789}{https://orcid.org/0000-0001-6428-1866}\\
\end{onehalfspace}

\end{center}
\newpage


%
\chaptertitlefont{\fontsize{14}{15}\selectfont\centering}

\begin{center}

\end{center}

\begin{center}
{{\bf\fontsize{14pt}{14.5pt}\selectfont \uppercase{ABSTRACT}}}
\end{center}

\addcontentsline{toc}{chapter}{Abstract}

\begin{center}
	\begin{doublespace}
{\fontsize{14pt}{14.5pt}\selectfont {Theoretical Foundations of Communication-Efficient, Robust, and Practical Distributed and Federated Optimization}}\\
		{\fontsize{14pt}{14.5pt}\selectfont {Grigorii Malinovskii}}\\
	\end{doublespace}
\end{center}

The rapid advancement of Machine Learning and the evolution of optimization have consistently evolved together, as practical needs and theoretical breakthroughs push both areas to progress. While modern large-scale training relies on classical optimization, the unique constraints of distributed systems have necessitated a complete rethink of foundational theory. In this thesis, we address seven critical challenges at the intersection of theory and practice, specifically focusing on the fundamental bottlenecks of Federated Learning and distributed optimization. For each of these challenges, we design novel algorithmic frameworks and provide sharp theoretical guarantees that bridge the gap between heuristic success and mathematical rigor.

Our first contribution resolves a fundamental question in communication efficiency within federated environments by introducing ProxSkip and proving that local gradient steps can lead to provable acceleration, finally establishing a formal foundation for this widely used heuristic. We extend this framework with Variance Reduced ProxSkip, which integrates variance reduction to eliminate the neighborhood error typically observed in stochastic local updates while effectively balancing communication and local computation costs. Our third contribution addresses the scalability and efficiency of these methods under partial participation, demonstrating that local steps still achieve provable communication acceleration even when only a subset of clients is active in each round. In the fourth challenge, we investigate the interplay of server-side dynamics and data selection, proving that server-side stepsizes and sampling without replacement significantly improve convergence in heterogeneous settings.

Building on these insights regarding data selection, our fifth contribution shows that in the context of Random Reshuffling, compressing gradient differences rather than raw gradients leads to superior theoretical and practical performance. We then address system security by proving that Byzantine robustness and partial participation can be achieved simultaneously through a gradient difference clipping mechanism. Finally, we move beyond classical optimization to establish the first theoretical framework for low-rank adaptation via randomized asymmetric chains, offering new insights into the fine-tuning of massive-scale models. Across all contributions, our results provide tighter analyses and more realistic assumptions than prior work, supported by numerical experiments that confirm the sharpness of our theoretical findings.


\begin{center}

\end{center}
\begin{center}

{\bf\fontsize{14pt}{14.5pt}\selectfont \uppercase{Acknowledgements}}\\\vspace{1cm}
\end{center}

\addcontentsline{toc}{chapter}{Acknowledgements} 


I would like to express my sincere gratitude to my supervisor, Peter Richtárik, for his constant support and guidance throughout my PhD. My journey started when Peter invited me for an internship, which eventually led to my joining the MS/PhD program under his supervision.

Our discussions and projects have deeply shaped my research, teaching me to value the combination of mathematical rigor, simplicity, and elegance. This approach led to several papers that I am truly proud of. Most importantly, Peter’s mentorship helped me become an independent researcher and gave me the opportunity to collaborate with colleagues from all over the world. I am incredibly grateful for the solid foundation he helped me build for my future career.

I am also grateful for several research visits and internships that led to many inspiring discussions. I would like to specifically thank Ilia Zharikov for hosting me at the Moscow Institute of Physics and Technology; Sebastian Stich for hosting my research visit at the CISPA Helmholtz Center for Information Security; Yurii Nesterov for the opportunity to visit the Université catholique de Louvain; Martin Jaggi for the opportunity to visit the École Polytechnique Fédérale de Lausanne; Eduard Gorbunov and Samuel Horváth for hosting my research visit and collaboration at the Mohamed bin Zayed University of Artificial Intelligence; Mete Ozay and Umberto Michieli for hosting my research internship at Samsung Research UK in Staines-upon-Thames; Christos Thrampoulidis for inviting me to visit the University of British Columbia in Vancouver; Praneeth Karimireddy for hosting my visit to the University of Southern California; and Angelia Nedić for hosting my visit to Arizona State University in Tempe.

Looking back on my PhD, one of the greatest rewards has been the chance to meet and collaborate with so many remarkable people. I simply could not have succeeded in these projects, nor completed this journey, without the unwavering help of Dmitry Kovalev, Elnur Gasanov, Laurent Condat, Konstantin Mishchenko, Kai Yi, Samuel Horváth, Konstantin Burlachenko, Ivan Agarský, Michał Grudzień, Alibek Sailanbayev, Yury Demidovich, Igor Sokolov, Soumia Boucherouite, El Houcine Bergou, Hasan Abed Al Kader Hammoud, Taha Ceritli, Hayder Elesedy, Ionut-Vlad Modoranu, Mher Safaryan, Eldar Kurtic, Thomas Robert, Dan Alistarh, Abdurakhmon Sadiev, Eduard Gorbunov, Ahmed Khaled, Zhirayr Tovmasyan, Egor Shulgin, Petr Ostroukhov, Martin Takáč, Sarit Khirirat, Rustem Islamov, Alexander Gaponov, Aurelien Lucchi, and Xun Qian. I was also fortunate to engage in insightful discussions with many other exceptional researchers, including Ahmet Alacaoğlu, Pavel Dvurechensky, Alexander Gasnikov, Sharan Vasvani, Boris Mordukhovich, Volkan Cevher, Robert Mansel Gower, Aritra Dutta, Nicolas Loizou, Michael Crawshaw, Anastasia Koloskova, Nikita Doikov, Anton Rodomanov, Hadrien Hendrikx, El Mahdi Chayti, Abhishek Chakraborty, and Ilyas Fathullin.

Being part of this research group has been a truly rewarding experience. While several colleagues have been acknowledged above, I am equally grateful to the other past and current members of our group for their inspiring discussions: Artavazd Maranjyan, Majied Ammar Mahran, Yassine Maziane, Ivan Ilin, Kaja Gruntkowska, Hanmin Li, Alina Abdikarimova, Igor Klimczak, Zhize Li, Avetik Karagulyan, Alexander Tyurin, Filip Hanzely, Adil Salim, Lukang Sun, Slavomír Hanzely, Omar Shaikh Omar, Artem Riabinin.

None of this would have been possible without the love and support of my parents, family, and friends. I am deeply grateful to my wife, Anastasia, for standing by my side, managing so much behind the scenes, and for the sacrifices she made so I could complete this PhD. I also want to thank my daughter, Maria, for being my greatest motivation and my daily reminder of what truly matters.

\begin{onehalfspacing}

\tableofcontents
\cleardoublepage

\printglossary[type=\acronymtype,style=long3col, title=\centerline{LIST OF ABBREVIATIONS}, toctitle=List of Abbreviations, nonumberlist=true] 

{
\let\oldglossentrydesc\glossentrydesc
\renewcommand{\glossentrydesc}[1]{\raggedright\oldglossentrydesc{#1}}
\printglossary[type=algorithms,style=long3col, title=\centerline{LIST OF ALGORITHMS}, toctitle=List of Algorithms, nonumberlist=true]
}

\printglossary[type=symbols,style=long3col, title=\centerline{LIST OF SYMBOLS}, toctitle=List of Symbols, nonumberlist=true]

\cleardoublepage
\phantomsection
\addcontentsline{toc}{chapter}{\listfigurename} 
\renewcommand*\listfigurename{LIST OF FIGURES}
\listoffigures

\cleardoublepage
\phantomsection
\addcontentsline{toc}{chapter}{\listtablename}  
\renewcommand*\listtablename{LIST OF TABLES}
\listoftables

\end{onehalfspacing}


\chaptertitlefont{\fontsize{14}{15}\selectfont}  


\chapter{Introduction}
\label{chapter1}
\thispagestyle{empty}

Over the past decade, advances in \gls{ML} in general, and \gls{DL} in particular, have fundamentally reshaped the landscape of artificial intelligence and transformed our daily interactions with digital infrastructure. Tasks that once seemed confined to the realm of science fiction are now ubiquitous: sophisticated reasoning is routinely performed by massive large language models, photorealistic imagery is synthesized seamlessly from text prompts by generative architectures, and highly specialized neural networks are driving breakthroughs in personalized medicine, autonomous vehicle navigation, and drug discovery. The engine behind these remarkable successes relies heavily on a triad of advancements: increasingly complex model architectures, massive leaps in computational hardware (primarily GPUs), and the unprecedented availability of vast, high-quality datasets.

In this thesis, our objective is to develop, analyze, and optimize more efficient techniques capable of training these massive models at scale. To ground our work in realistic deployment environments and their associated physical, network, or legal constraints, we systematically study two distinct but related paradigms: distributed optimization and \gls{FL} \citep{FedLearn2016, FedOpt2016, FLblog2017, mcmahan2017communication}. While both frameworks are designed to collaboratively train a single global \gls{ML} model across multiple computing nodes, the fundamental bottlenecks and underlying motivations for utilizing them differ drastically.

The push toward distributed optimization is primarily driven by the sheer physical scale of modern \gls{ML}. As the parameter counts of state-of-the-art models swell into the billions or trillions \citep{touvron2023llama}, and the datasets required to train them grow exponentially, the memory and computational demands quickly exceed the capacity of even the most advanced single-machine accelerators. To achieve feasible training times, engineers are forced to parallelize the workload \citep{dean2012large}. This involves distributing the computation across a massive cluster of interconnected nodes, where each machine processes shards of the full dataset simultaneously. In this regime, the primary bottleneck is not a lack of access to data, but rather insufficient localized computational resources \citep{mcmahan2017communication}. The fundamental challenge lies in designing techniques that can orchestrate communication and computation efficiently across high-speed network topologies without suffering from crippling synchronization delays \citep{ben2019demystifying}.

Conversely, there are countless scenarios where centralizing data onto a single machine or cloud cluster is either physically infeasible or strictly prohibited, regardless of how much computing power is available \citep{kairouz2019advances}. The increasingly data-hungry nature of modern \gls{ML} has rightfully triggered severe privacy concerns among users \citep{dwork2014algorithmic}, leading to stringent government regulations and a broad push toward privacy-preserving digital ecosystems. Furthermore, the physical limitations of network bandwidth and the energy costs associated with transmitting raw data to centralized servers have made the \textit{data locality paradigm} essential \citep{abreha2022federated}. In this paradigm, data must be processed at the exact edge location where it is captured and stored.

Federated learning \citep{konevcny2015federated, FedLearn2016, FedOpt2016, mcmahan2017communication, bonawitz2019towards} is an emerging, transformative subfield of \gls{ML} explicitly designed to solve this problem. \gls{FL} enables a network of clients to collaboratively train a shared, global model without their raw, sensitive data ever leaving their local devices. In its standard operational form, \gls{FL} orchestrates training through a centralized server. In each communication round, a subset of participating devices is selected. These devices download the current global model, execute local training steps using their proprietary data, and then transmit only the computed model updates (such as gradients or weight differences) back to the server. These ephemeral updates are then securely aggregated to refine the global model, ensuring that the raw data remains decentralized and private.

Practical \gls{FL} deployments generally materialize in two distinct regimes, each presenting its own unique mathematical and infrastructural challenges. In the {\em cross-silo} setting \citep{kairouz2019advances}, a relatively small number of identifiable participants with substantial computational resources collaborate to train a model. These participants are typically large organizations, such as competing hospitals aiming to improve medical imaging diagnostics, or financial institutions looking to detect sophisticated money laundering patterns. While they share a common objective, they are strictly barred from pooling their raw data due to rigorous regulatory, security, or proprietary business constraints.

In contrast, the {\em cross-device} setting involves networks consisting of millions, or even hundreds of millions, of edge devices, such as mobile smartphones or Internet of Things sensors. Each device holds a relatively tiny amount of highly personal data. In this scenario, devices are often anonymous, have high dropout rates due to battery or connectivity issues, and may participate in the training process a single time only. Major industry providers already deploy this to preserve privacy in applications like next-word prediction on mobile keyboards or real-time traffic condition modeling for autonomous vehicles \citep{FLblog2017, abadi2016deep, bonawitz2019towards}.

While \gls{FL} and distributed systems have seen massive empirical adoption by major industry players, much of this success has historically relied on heuristic technique design. Instead of merely proposing new techniques and validating them solely through empirical benchmarks, this thesis seeks to rigorously understand the fundamental mathematical properties of these training paradigms. Mathematically, the training process is equivalent to solving a highly complex optimization problem. By defining a loss function $f$ that measures the error of our model, the goal is to minimize this loss over an immense parameter space of dimension $d$:
\begin{equation}
\min_{x \in \mathbb{R}^d} f(x).
\label{eq:main_optimization}
\end{equation}

Historically, \gls{ML} has advanced in lockstep with breakthroughs in mathematical optimization. As established in classical optimization literature \citep{NesterovBook}, the theoretical prediction of an optimization technique's performance (measured through computational complexity) usually matches its practical behavior. This makes abstract complexity analysis a ubiquitous and powerful tool: it allows us to understand the deep structure of learning problems, design structurally sound techniques, and predict their behavior before deployment.

However, classical abstract optimization theory frequently fails to capture the intricate realities and essential constraints of modern distributed and federated environments. Theoretical guarantees designed for idealized settings inevitably break down when subjected to the practical limitations of edge computing \citep{FieldGuide2021}. Therefore, to bridge the gap between the empirical success of these distributed techniques and their rigorous theoretical foundations, our first step must be to formally define the learning objective in a way that accurately reflects the decentralized nature of the data. By establishing a precise mathematical formulation of the distributed optimization problem, we construct the necessary scaffolding for our theoretical analysis. With this foundational framework in place, we can then systematically articulate the fundamental, real-world challenges inherent to federated systems that our proposed techniques are designed to overcome.

\section{Problem Formulation}

The most common approach to \gls{FL} is to learn a single global model $x \in \mathbb{R}^d$ from decentralized data $\{D_m\}_{m=1}^M$ stored across $M$ remote clients. To account for varying dataset sizes across clients, let $n_m = |D_m|$ be the number of training data points on client $m$, and $N := \sum_{m=1}^M n_m$ be the total number of data points collectively owned by the federation. The objective is to minimize the empirical risk over the entire federated dataset, which is formulated as a weighted average of the local losses,
\begin{equation}
\label{eq:intro_main}
    \min_{x\in \mathbb{R}^d} \left[ f(x) := \sum_{m=1}^M \frac{n_m}{N} f_m(x) \right],
\end{equation}
where $f_m(x)$ is the local loss on the $m$-th device.

We assume that each local objective admits the stochastic representation
\begin{equation*}
    f_m(x) := \mathbb{E}_{\xi \sim \mathcal{D}_m} [f_m(x, \xi)],
\end{equation*}
where $\mathcal{D}_m$ denotes the data distribution on device $m$ and $\xi$ is a sample drawn from it. Although all clients share the same loss function structure, the resulting objectives $\{f_m\}$ generally differ due to variations in the underlying data distributions $\{\mathcal{D}_m\}$, a phenomenon known as data heterogeneity.

In the common empirical setting where the local datasets $D_m$ consist of a finite number of samples, each local objective can be expressed as an empirical average over its respective local dataset. Letting $\xi_{m,i}$ denote the $i$-th data point on device $m$, the local objective is written as
\begin{equation*}
    f_m(x) = \frac{1}{n_m} \sum_{i=1}^{n_m} f_m(x, \xi_{m,i}) \equiv \frac{1}{n_m} \sum_{i=1}^{n_m} f_{m,i}(x),
\end{equation*}
where $f_{m,i}(x)$ serves as a concise notation for the loss evaluated on that specific data point. Substituting this back into the global objective demonstrates that $f(x)$ is indeed the exact average loss over all $N$ training points in the federation:
\begin{equation*}
    f(x) = \frac{1}{N} \sum_{m=1}^M \sum_{i=1}^{n_m} f_{m,i}(x).
\end{equation*}

While this general formulation accounts for varying dataset sizes, for simplicity, we may sometimes assume that all devices hold an equal number of data points $n$ (i.e., $n_1 = \dots = n_M = n$). Whenever this simplifying assumption is applied, it will be explicitly stated in the respective chapters.

Having described the structure of the optimization problem addressed in federated learning and distributed optimization, we now turn to the key practical challenges faced by these methods.

\section{Fundamental Challenges in Federated Learning}
\label{sec:chall_intro}
As established in our discussion, federated learning introduces a unique set of constraints that distinguish it from the broader field of distributed learning. While \gls{FL} shares mathematical roots with classical distributed optimization \citep{nedic2009distributed} and privacy-preserving data analytics \citep{dwork2014algorithmic}, its operational reality is vastly more complex \citep{FieldGuide2021}. Below, we detail the core systemic challenges that separate the federated paradigm from traditional, centralized setups and necessitate the development of specialized algorithmic frameworks.

\subsection{The communication bottleneck}

It is widely recognized that communication overhead is the primary bottleneck in federated learning systems \citep{mcmahan2017communication,konevcny2015federated}. Unlike the high-speed, dedicated, and symmetric interconnects available in modern data centers, the connections between edge devices and the coordinating server, such as wireless or mobile networks, are typically slow, highly asymmetric, and unreliable. As a consequence of this limited bandwidth and high latency, the time and energy required to transmit high-dimensional model updates frequently exceed the costs associated with local computation by several orders of magnitude.

Training modern large-scale models under such highly constrained distributed conditions therefore requires algorithms that are fundamentally efficient in terms of communication. Broadly speaking, the literature addresses this challenge through two main, often complementary, approaches. The first relies on communication compression techniques, such as quantization \citep{alistarh2017qsgd} and sparsification \citep{stich2018sparsified}. By mapping dense vectors to lower-precision or sparse representations, these methods drastically reduce the payload size exchanged during each communication round. The second approach focuses on periodic aggregation methods utilizing local updates, with algorithms such as \algname{Local SGD} (\gls{LSGD}) \citep{mangasarian1995parallel, zinkevich2010parallelized,  stich2018,woodworth2020minibatch} serving as a foundational baseline. In this paradigm, clients perform multiple local optimization steps on their respective datasets before communicating with the server, thereby significantly reducing the total number of communication rounds required to achieve convergence.

\subsection{System heterogeneity and partial participation}

In addition to the communication bottleneck, federated networks, particularly in the cross-device regime \citep{kairouz2019advances}, routinely encompass millions of potential participants. However, the bandwidth capacity of the central orchestrating server imposes a strict upper limit on the number of clients that can actively communicate in any given training round. To address this massive scale, practical \gls{FL} systems fundamentally rely on \acrfull{PP}. By employing \gls{PP} methodologies (also known as Client Sampling (CS)) \citep{gower2019sgd, OptClientSampling2020}, only a small, manageable subset of available devices is selected to compute and transmit updates per round, strategically reducing the network load without sacrificing the rigorous mathematical guarantees of the global model.

Beyond the limitations of the central server, the devices themselves introduce severe systemic challenges. Unlike traditional distributed training, which relies on powerful and uniform high-end compute architectures, \gls{FL} clients are typically resource-constrained, highly heterogeneous, and unreliable. This profound hardware heterogeneity manifests as significant variability in local processing speeds, network connection reliability, storage capacity, and battery levels. Consequently, if a training algorithm requires strict synchronization, the entire aggregation process is forced to wait for the slowest participants, commonly referred to as stragglers \citep{damaskinos2018asynchronous}. 

Furthermore, mid-tier and low-tier devices may lack the memory or processing capabilities required to train complex models efficiently. As a result, they are frequently dropped due to timeouts or excluded from the training process altogether. Excluding these constrained devices introduces profound issues beyond simple algorithmic inefficiency. The hardware capabilities of a user's device often correlate strongly with their demographic and socioeconomic status. Systematically dropping these clients leads to an inherently biased and unfair training process \citep{xu2020towards}, discarding unique and highly valuable local data. Therefore, the design of modern federated optimization methods must be theoretically robust not only to the orchestrated partial participation of clients, but also to the unpredictable realities of extreme device heterogeneity, asynchronous stragglers, and sudden dropouts \citep{kairouz2019advances}.

\subsection{Data heterogeneity}

Unlike traditional distributed training setups where data is uniformly shuffled and randomly partitioned across compute nodes to ensure identical distributions, federated learning operates strictly on localized data. Clients generate and store data based on their unique environments, personal preferences, and individual usage patterns. This inherent statistical disparity, widely referred to as data heterogeneity \citep{mcmahan2017communication, kairouz2019advances, FL_survey_2020}, implies that the local dataset $\mathcal{D}_m$ on any given client is fundamentally unrepresentative of the global distribution. Consequently, the local objective functions $\{f_m\}$ can differ significantly from one another as local models optimize toward entirely different local minima \citep{karimireddy2020scaffold}.

In theory, this heterogeneity across local objectives constitutes a rigorous mathematical barrier, rather than simply a practical issue. The specific nature of heterogeneity fundamentally restricts the achievable convergence guarantees of distributed algorithms. Dictated by established theoretical limits in optimization, the variance introduced by heterogeneous local objectives inherently bottlenecks the training process \citep{khaled2019first, woodworth2020minibatch}. Algorithms that perform optimally under the assumption of perfectly uniform data distributions frequently experience severe performance degradation or even complete failure when forced to reconcile conflicting local updates. Therefore, designing robust algorithms that can maintain tight convergence guarantees despite the intrinsic mathematical barriers imposed by statistically disjoint local datasets \citep{koloskova2020unified, localSGD-AISTATS2020, allouah2024robust} remains one of the most critical theoretical challenges in federated optimization.

\subsection{Privacy and Byzantine-robustness}

While data heterogeneity addresses the statistical challenges introduced by honest clients, federated systems must also contend with severe security vulnerabilities inherent to decentralized networks. The fundamental premise of federated learning is to protect user privacy by ensuring that raw data never leaves the local device. However, recent research has demonstrated that simply transmitting model updates, such as gradients or weight differentials, is insufficient to guarantee strict privacy. Adversaries can employ sophisticated gradient leakage or model inversion attacks to reconstruct sensitive local data points directly from the shared updates \citep{lyu2020privacy}. Consequently, practical federated optimization algorithms must often integrate formal privacy-preserving mechanisms, such as Differential Privacy, which inject calibrated noise into the training process to mathematically bound the maximum information leakage \citep{dwork2006calibrating, dwork2014algorithmic, abadi2016deep}. 

Furthermore, because the orchestrating server has no visibility into the local data or the computational integrity of the participating edge devices, the network is highly susceptible to both systemic faults and malicious interference. In traditional distributed settings, all computing nodes are controlled by a single trusted entity. In contrast, federated networks are inherently untrustworthy. Clients may transmit severely corrupted updates due to malfunctioning hardware, degraded network connections, or compromised local datasets. More critically, the system may face intentional data poisoning or adversarial attacks orchestrated by malicious participants aiming to subvert the global model \citep{blanchard2017machine}. 

This vulnerability necessitates the design of algorithms that exhibit Byzantine-robustness. A system is considered Byzantine-robust if it can successfully converge to an accurate global model even when a fraction of the participating clients act maliciously and send arbitrary, worst-case updates to the server \citep{yin2018byzantine, alistarh2018byzantine}. Classical aggregation techniques, such as simple averaging, are catastrophically fragile in this threat model, as a single infinitely large malicious update can instantly destroy the global model. Therefore, developing sophisticated, robust aggregation frameworks that can mathematically guarantee convergence in the presence of Byzantine adversaries, while simultaneously maintaining communication efficiency and respecting privacy constraints, is a paramount challenge in modern federated learning \citep{wu2020federated, el2021collaborative}.

\section{Baseline Method in Federated Optimization}

Assuming access to exact local gradients, the global objective presented in problem (1.1) can theoretically be minimized using the canonical first-order method: Gradient Descent (\gls{GD}). A standard \gls{GD} update takes the form
\begin{equation*}
    x^{t+1} = x^t - \gamma^{(t)} \nabla f(x^t), \quad t=0,1,2,\dots
\end{equation*}
where $\gamma^{(t)} > 0$ denotes the learning rate (step size). By decomposing the global gradient, we can express this update in terms of the local client objectives:
\begin{equation*}
    \nabla f(x^t) = \frac{1}{M}\sum_{m=1}^M \nabla f_m(x^t).
\end{equation*}
This demonstrates that the global gradient is simply the average of the local gradients. Because privacy constraints prohibit direct peer-to-peer communication among clients, this gradient aggregation is managed by a central orchestrating server, which subsequently broadcasts the updated model back to the network.

Under standard convexity and smoothness assumptions, \gls{GD} is a thoroughly analyzed method. However, while theoretically sound, naive \gls{GD} is fundamentally ill-suited for practice in federated learning due to the severe systemic constraints discussed in Section \ref{sec:chall_intro}. Consequently, making gradient-based optimization viable in \gls{FL} requires targeted algorithmic modifications. In many ways, the core contributions of this thesis revolve around enhancing these foundational methods to achieve both theoretical rigor and practical efficiency in the federated setting.

Below, we detail the primary techniques and considerations required to adapt optimization algorithms for \gls{FL}:

\begin{itemize}
    \item \textbf{Local Training:} To directly combat the communication bottleneck, algorithms frequently allow clients to perform multiple local descent steps before synchronizing with the central server \citep{stich2018}. While local training is a cornerstone heuristic in practical \gls{FL} deployments, fully characterizing its theoretical superiority over standard methods remains a challenging and active area of research \citep{khaled2019first, koloskova2020unified, woodworth2020minibatch, glasgow2022sharp}.

    \item \textbf{Client Drift Reduction:} A direct consequence of combining local training with data heterogeneity is the phenomenon of client drift \citep{karimireddy2020scaffold}. Because local objective functions differ, taking multiple local gradient steps causes individual client models to diverge toward their respective local minima rather than the true global optimum, severely degrading convergence. To mitigate this, advanced federated algorithms incorporate drift reduction mechanisms \citep{karimireddy2020scaffold,gorbunov2021local, mitra2021linear}. These techniques dynamically correct the local update directions, realigning the client trajectories with the global objective and stabilizing the optimization process without sacrificing the benefits of local steps.

    \item \textbf{Communication Compression:} To alleviate the strict bandwidth limitations of edge networks, algorithms often employ compression operators (such as quantization or sparsification) to reduce the bit-width or payload size of the messages transmitted between clients and the server \citep{alistarh2017qsgd, wen2017terngrad, stich2018sparsified, wangni18_sparsification, mishchenko2019distributed}.

\item \textbf{Partial Participation:} In any given communication round $t$, it is unrealistic to expect all clients to be available \citep{mcmahan2017communication}. Instead, only a subset $\set \subset [M]$ actively communicates with the server. The model update rule is therefore restricted to this active cohort:
    \begin{equation*}
        x^{t+1} = x^t - \gamma^{(t)} \frac{1}{C} \sum_{m \in \set} \nabla f_m(x^t),
    \end{equation*}
    where $C = |\set|$ denotes the number of active clients in round $t$. Client availability is notoriously unpredictable, often depending on exogenous factors like network access or device battery levels. Consequently, guaranteeing theoretical convergence under partial participation requires specific probabilistic assumptions regarding the Partial Participation mechanism \citep{ClientSelection-Gauri, OptClientSampling2020, Artemis2020}. Unsurprisingly, relying on partial updates typically inflates the total number of communication rounds required to reach a target accuracy.

    \item \textbf{Data Sampling:} In modern machine learning, replacing the exact full-batch gradient with a stochastic estimator is standard practice \citep{robbins1951stochastic, gower2019sgd}. Theoretically, this relies on access to an unbiased stochastic gradient $g_m(x)$ such that $\mathbb{E}_{\mathcal{D}_m}[g_m(x)] = \nabla f_m(x)$. In practical finite-sum settings, however, this expectation is almost universally approximated using mini-batches constructed from the local dataset $D_m$ \citep{Dekel2012:minibatch, Li2014}. In addition, practical implementations commonly employ \gls{RR}, i.e., sampling without replacement across consecutive epochs, a technique that naturally aligns with standard software pipelines and often yields faster convergence than uniform sampling with replacement \citep{bottou2009curiously, MKR2020rr, rajput20_closin_conver_gap_sgd_without_replac}. Data sampling is inherently required when the true underlying data distribution is inaccessible and becomes computationally indispensable when local datasets are prohibitively large.

\item \textbf{Variance Reduction:} The reliance on Data Sampling and the randomness inherent in partial client participation inject considerable noise into the optimization trajectory \citep{johnson2013accelerating, defazio2014saga, schmidt2017minimizing, gower2020variance, gorbunov2020unified}. To prevent this compounded variance from dominating the update steps and stalling convergence, particularly when seeking high-precision solutions, federated algorithms frequently adapt classical variance reduction techniques \citep{gorbunov2021local}. Crucially, the theoretical utility of these techniques extends well beyond data stochasticity; analogous variance reduction mechanisms are fundamentally required to mitigate the quantization error and variance introduced by lossy communication compression operators \citep{mishchenko2019distributed, DIANA2, gorbunov2021marina}. By maintaining local state variables or leveraging historical gradient information, these methods systematically diminish the variance of the stochastic estimators over time, ensuring faster and more stable asymptotic convergence despite the severe stochasticity of the federated setting.
\end{itemize}

A detailed summary of the representative algorithms proposed in each chapter, along with the topics covered and techniques considered throughout the thesis, is presented in Table~\ref{tab:chapter_summary}.

\begin{table}[t]
    \centering
    \caption{Summary of the primary theoretical techniques utilized and specific settings addressed per chapter. Columns (chapter topics): LT = Local Training, Acc = acceleration via local steps, CS = Partial Participation, CC = Communication Compression, RR = Random Reshuffling, Byz = Byzantine Robustness, VR = Variance Reduction.}
    \label{tab:chapter_summary} 
    
    \renewcommand{\arraystretch}{1.5} 
    \begin{tabular}{|c|c|c|c|c|c|c|c|c|c|c|}
        \hline
        Ch & Ref & Algorithm &  New & LT & Acc & CS & CC & RR & Byz & VR \\
        \hline
        \ref{chapter2} & [\citenum{ProxSkip}] & {\scriptsize\glsentryshort{ProxSkip}}& \cmark & \cmark & \cmark & \xmark & \xmark & \xmark & \xmark & \xmark \\
        \hline
        \ref{chapter3} & [\citenum{ProxSkip-VR}] & {\scriptsize\glsentryshort{ProxSkip-VR}}& \cmark &\cmark  & \cmark & \xmark & \xmark & \xmark & \xmark & \cmark \\
        \hline
        \ref{chapter4} & [\citenum{grudzien2023can}] & {\scriptsize\glsentryshort{5GCS}} &\cmark& \cmark  & \cmark & \cmark & \xmark & \xmark & \xmark & \cmark  \\
        \hline
        \ref{chapter5} & [\citenum{malinovsky2023server}] & {\scriptsize\glsentryshort{NASTYA}}&\cmark & \cmark  & \xmark & \cmark & \xmark & \cmark & \xmark & \xmark \\
        \hline
        \ref{chapter6} & [\citenum{sadiev2022federated}] & {\scriptsize\glsentryshort{DIANA-NASTYA}}&\cmark & \cmark  & \xmark & \cmark & \cmark & \cmark & \xmark & \cmark \\
        \hline
        \ref{chapter7} & [\citenum{malinovsky2024byzantine}] & {\scriptsize\glsentryshort{Byz-VR-MARINA-PP}}&\cmark & \xmark  & \xmark & \cmark & \cmark & \xmark & \cmark & \cmark \\
        \hline
        \ref{chapter8} & [\citenum{malinovsky2024randomized}] & {\scriptsize\glsentryshort{Fed-RAC-LoRA}}&\cmark & \cmark  & \xmark & \cmark & \xmark & \cmark & \xmark & \xmark \\
        \hline
    \end{tabular}
\end{table}

\section{Thesis Outline and Focus}

We directly focus on these core challenges throughout this thesis, with a universal emphasis on alleviating the communication bottleneck in all chapters. In Chapters \ref{chapter2}, \ref{chapter3}, and \ref{chapter4}, we tackle the mathematical and algorithmic hurdles introduced by data heterogeneity, designing methods that converge reliably despite statistically mismatched local datasets. In Chapters 4 through 8, we shift our attention to the massive scale of federated networks by explicitly integrating partial client participation into our theoretical frameworks. Concurrently, in Chapters \ref{chapter5}, \ref{chapter6}, and \ref{chapter8}, we delve into the mechanics of local computation by analyzing algorithms that utilize sampling without replacement, specifically, the random reshuffling of local data, which is a highly practical and empirically successful technique for enhancing the efficiency of local training. Finally, in Chapter \ref{chapter7}, we extend our analysis to address network security and unreliability, focusing on Byzantine-robustness to ensure global model convergence even in the presence of malicious clients or arbitrarily corrupted updates. While we do not directly develop new formal privacy mechanisms, the distributed optimization schemes presented in this thesis are designed to be broadly compatible with standard privacy-preserving techniques.

For a concise summary of our contributions across different settings, Table \ref{tab:chapter_summary} presents a representative algorithm for each chapter of this thesis, as well as the covered topics.

\section{Challenge 1: Achieving Accelerated Communication Complexity Via Local Gradient Updates}

\paragraph{Motivation.} Revisiting our introduction, the communication bottleneck stands out as a principal constraint in federated learning. Since communication channels are slow and model sizes are huge, realistic algorithms operate under delayed communication. To further decouple computation from communication, existing approaches locally take multiple steps of gradient descent before averaging at the server. However, highly data heterogeneous clients suffer from dramatic ``client drift''. In practice, client drift is a major issue that degrades performance in experiments, and it also presents significant difficulties for the theoretical understanding of local algorithms. Algorithms such as \algname{SCAFFOLD} \cite{karimireddy2020scaffold}, \algname{S-Local-GD}\cite{gorbunov2021local} and \algname{FedLin}\cite{mitra2021linear} overcame client drift, but led to the discovery of another deeper theoretical disappointment. Under typical strong convexity and smoothness assumptions, their communication complexity was $\mathcal{O}(\kappa \log 1/\varepsilon)$ where $\kappa$ denotes the condition number of the objective. This matches with vanilla \gls{GD}, which only takes one local gradient step before communicating at each round. Despite the empirical success of many such local algorithms, for years their theoretical communication complexity refused to beat standard \gls{GD}. This gap raised an urgent question: can local gradient descent algorithms offer provable improvements in communication complexity, without explicitly invoking Nesterov-style acceleration? 

\paragraph{Contributions.} In Chapter \ref{chapter2}, we answer this open question affirmatively. We introduce \gls{ProxSkip}, a simple and provably efficient optimization method that formally establishes communication acceleration through local gradient steps.

We approach the federated optimization problem by minimizing the sum of a smooth function and an expensive, nonsmooth proximable regularizer. In distributed learning, the gradient operator corresponds to a cheap local \gls{GD} step. The proximity operator encodes a consensus constraint, representing the highly expensive communication required for averaging the local iterates.

\gls{ProxSkip} allows skipping this expensive proximity operator in most iterations. By combining randomized prox skipping with a novel control variate to stabilize forward gradient steps, \gls{ProxSkip} shifts local gradients to correct for drift. While total iteration complexity remains $\mathcal{O}(\kappa \log 1/\varepsilon)$, the expected proximity operator evaluations, and thus communication complexity, drastically reduce to $\mathcal{O}(\sqrt{\kappa} \log 1/\varepsilon)$. Unlike previous methods, our algorithm guarantees strictly superior communication complexity. This provides rigorous theoretical justification for local training without requiring heterogeneity bounding assumptions.

Finally, we extend \gls{ProxSkip} to decentralized optimization over graphs and the stochastic regime. While communication acceleration is preserved, the derived convergence rates do not exhibit a linear speedup (division of variance by the number of clients), leaving room for further theoretical refinement to fully characterize the algorithm's capabilities in this regime.

\noindent\paragraph{Paper.} The chapter is based on the paper:

\begin{itemize}
    \item[{[\citenum{ProxSkip}]}] Konstantin Mishchenko, Grigory Malinovsky, Sebastian Stich, and Peter Richt\'{a}rik. ProxSkip: Yes! Local gradient steps provably lead to communication acceleration! Finally! In \textit{International Conference on Machine Learning}, pp. 15750--15769. PMLR, 2022.
\end{itemize}

\section{Challenge 2: Balancing Resource Costs and Reducing Variance in Federated Optimization}

\paragraph{Motivation.} As established in Challenge 1, \algname{ProxSkip} demonstrated that local gradient steps lead to communication acceleration, initiating a new class of fifth-generation local training algorithms. However, a theoretical gap remains when extending these methods to stochastic settings. In standard stochastic formulations, the inherent variance of the gradient estimates introduces a noise term that slows down convergence. As noted previously, the stochastic extension of \algname{ProxSkip} does not exhibit a linear speedup, meaning it does not benefit from dividing this variance by the number of participating clients. Furthermore, prior theoretical frameworks for local training primarily measured efficiency in terms of communication rounds, largely overlooking the overall computational cost of the local gradient steps.

\paragraph{Contributions.} In Chapter \ref{chapter3}, we address these challenges by proposing a variance-reduced extension of \algname{ProxSkip} (\gls{ProxSkip-VR}). Rather than attempting to recover the division by the number of clients for the variance term, we bypass the issue by neutralizing the stochastic noise directly. By incorporating variance reduction mechanisms into the \gls{ProxSkip} framework, our method progressively drives the variance of the stochastic gradients to zero. This modification enables \gls{ProxSkip-VR} to achieve linear convergence even in the stochastic regime, effectively rendering the lack of a linear speedup inconsequential as the additive noise term vanishes. 

Beyond resolving the stochastic bottleneck, our theoretical analysis extends to evaluate the complete optimization process. We demonstrate that \gls{ProxSkip-VR} reduces overall training cost, including both local computation and server communication, relative to the baseline \gls{ProxSkip} algorithm. Lastly, we generalize the typical federated topology by proposing a new hierarchical structure with regional hubs. By positioning these hubs as intermediate nodes that connect the clients to the central server, this framework naturally accommodates more complex network structures and provides an organized approach to managing communication across multiple tiers.

\noindent\paragraph{Paper.} The chapter is based on the paper:

\begin{itemize}
    \item[{[\citenum{ProxSkip-VR}]}] Grigory Malinovsky, Kai Yi, and Peter Richt\'{a}rik. Variance reduced ProxSkip: Algorithm, theory and application to federated learning. In \textit{Advances in Neural Information Processing Systems 35} (2022), pp. 15176--15189.
\end{itemize}

\section{Challenge 3: Integrating Partial Participation Into Accelerated Local Training}

\paragraph{Motivation.} As demonstrated in the previous chapters, the \gls{ProxSkip} framework established a new class of fifth-generation local training algorithms by achieving provable communication acceleration. While these methods successfully handle heterogeneous data and stochastic gradients, they share a critical limitation. Specifically, their theoretical analysis assumes full client participation in every communication round. In massive federated networks, communicating with all devices simultaneously is infeasible. Therefore, Partial Participation, where only a fraction of clients participates in each round, is a practical necessity. However, integrating Partial Participation into fifth-generation local training remained an open problem, as the underlying drift correction mechanisms previously relied on continuous updates from the entire network.

\paragraph{Contributions.} In Chapter \ref{chapter4}, we address this gap by proposing \gls{5GCS}, the first fifth-generation local training method that explicitly achieves accelerated communication complexity via local gradient steps while supporting Partial Participation. Furthermore, we show that the communication complexity of \gls{5GCS} matches previously established theoretical lower bounds, confirming its optimality.

The core of our contribution lies in demonstrating how local computation directly drives this communication acceleration under partial participation. \gls{5GCS} is highly flexible and supports arbitrary local training subroutines under standard technical assumptions. To highlight the precise role of local steps, we analyze the theoretical extremes of this framework. If only a single local step is taken, the method achieves only a linear, non-accelerated communication rate. Conversely, taking an infinite number of local steps equates to exactly evaluating the proximity operator of the local functions, reducing \gls{5GCS} to a minibatch version of \algname{PointSAGA} \citep{defazio2016simple}. While this exact evaluation achieves optimal communication complexity, it places a heavy computational burden on the clients.

Our primary theoretical result resolves this tension. We prove that applying only a relatively small number of local gradient steps is entirely sufficient to preserve the accelerated communication complexity of the exact proximity evaluation. This firmly establishes that inexpensive local steps are the key mechanism for achieving optimal communication rates, successfully unifying the core benefits of fifth-generation local training with practical Partial Participation.

\noindent\paragraph{Paper.} The chapter is based on the paper:

\begin{itemize}
    \item[{[\citenum{grudzien2023can}]}] Micha\l{} Grudzie\'{n}, Grigory Malinovsky, and Peter Richt\'{a}rik. Can 5th generation local training methods support Partial Participation? Yes! In \textit{International Conference on Artificial Intelligence and Statistics}, pp. 1055--1092. PMLR, 2023.
\end{itemize}

\section{Challenge 4: Investigating Server-Side Stepsizes and Random Reshuffling in Federated Optimization}

\paragraph{Motivation.} In practical federated optimization, algorithms like Federated Averaging often rely on heuristic techniques to improve convergence and overcome the communication bottleneck. One common heuristic is the application of a server-side stepsize, where the aggregated client updates are scaled by an extra parameter before updating the global model. Simultaneously, practitioners frequently perform local training passes over the data using sampling without replacement, known as Random Reshuffling (\gls{RR}). Despite their empirical success, a clear theoretical justification for combining these two techniques was lacking. Prior analysis did not fully explain how server-side optimization and without-replacement sampling jointly affect the communication complexity of local methods.

\paragraph{Contributions.} In Chapter \ref{chapter5}, we address this by analyzing how server-side stepsizes and sampling without replacement influence the performance of federated optimization. We propose a new algorithm, \algname{Nastya}, that integrates server-side stepsizes and sampling without replacement. We show that when local passes over client data are performed using \gls{RR}, scaling the aggregated updates via a server-side stepsize yields improved convergence rates and complexities.

Our analysis identifies two main regimes depending on the local stepsize choice. When local stepsizes are small, applying Random Reshuffling across all clients produces an update direction that allows for a larger update at the server level. Taking a large server-side stepsize in this direction improves convergence rates across convex, strongly convex, and nonconvex objectives. In the nonconvex regime, our analysis reduces the convergence complexity from $\mathcal{O}(1/\varepsilon^3)$ to $\mathcal{O}(1/\varepsilon^2)$. This improvement also applies to Random Reshuffling executed on a single machine.

When local stepsizes are large, we show that the noise introduced by Partial Participation can be mitigated by utilizing a small server-side stepsize. This observation provides a theoretical basis for the common practice of tuning server-side stepsizes. By showing that local steps help overcome the communication bottleneck, and by extending our analysis to support partial client participation, this work provides a clearer theoretical understanding of these optimization techniques.

\paragraph{Paper.} The chapter is based on the paper:
\begin{itemize}
    \item[{[\citenum{malinovsky2023server}]}] Grigory Malinovsky, Konstantin Mishchenko, and Peter Richt\'{a}rik. Server-side stepsizes and sampling without replacement provably help in federated optimization. In \textit{Proceedings of the 4th International Workshop on Distributed Machine Learning}, pp. 85--104. 2023.
\end{itemize}

\section{Challenge 5: Integrating Gradient Compression with Random Reshuffling and Local Computation}

\paragraph{Motivation.} In distributed and federated optimization, gradient compression is a fundamental technique used to reduce the size of transmitted messages and overcome communication bottlenecks. Concurrently, practitioners overwhelmingly prefer sampling data without replacement, commonly known as \gls{RR}, because it yields superior convergence compared to sampling with replacement. While each technique is highly effective individually, their combined application remained largely underexplored. A major theoretical obstacle is that naive compression injects an additional layer of noise into the optimization process. This noise can easily dominate the inherent sampling variance, potentially nullifying the superior convergence properties of without-replacement sampling. Furthermore, designing practical federated algorithms requires supporting multiple local computational steps and distinct stepsizes, which complicates the integration of these techniques even further.

\paragraph{Contributions.} In Chapter \ref{chapter6}, we address these challenges by systematically developing algorithms that successfully combine gradient compression, sampling without replacement (\gls{RR}), and local training. We structure our contributions around identifying the limitations of naive integration and subsequently designing variance reduction mechanisms.

First, we isolate the fundamental issue by proposing and analyzing a quantized version of \gls{RR}, namely \gls{Q-RR}, a method that naively applies gradient compression at every communication round within an \gls{RR} framework. Our theoretical and empirical results confirm that this naive combination fails to improve upon standard quantized methods. The noise introduced by communication compression overwhelms the algorithmic benefits of \gls{RR}.

To resolve this bottleneck, we propose \gls{DIANA-RR}. The method integrates the \algname{DIANA} variance reduction mechanism to mitigate the noise introduced by gradient compression. We establish that, unlike standard \algname{DIANA}, achieving optimal theoretical guarantees in conjunction with \gls{RR} requires maintaining multiple shift vectors per worker rather than a single shift vector. We prove that \gls{DIANA-RR} successfully unlocks the benefits of \gls{RR} in compressed distributed learning, yielding superior convergence rates.

Building directly upon our analytical framework from Chapter \ref{chapter5}, we then extend these insights to federated environments. We introduce \gls{Q-NASTYA}, an advanced variant tailored for federated applications that naively mixes quantization, local steps via \gls{RR}, and separate local and server stepsizes. While this method successfully improves the per-round communication cost, our analysis reveals it still suffers from the dominating quantization variance. To overcome this final limitation, we propose \gls{DIANA-NASTYA}. By integrating our targeted variance reduction strategy into the local training environment, this algorithm successfully removes the additional variance, providing a provably efficient and theoretically sound method for federated learning.

\noindent\paragraph{Paper.} The chapter is based on the paper:

\begin{itemize}
    \item[{[\citenum{sadiev2022federated}]}] Abdurakhmon Sadiev, Grigory Malinovsky, Eduard Gorbunov, Igor Sokolov, Ahmed Khaled, Konstantin Burlachenko, and Peter Richt\'{a}rik. Don't compress gradients in random reshuffling: compress gradient differences. In \textit{Advances in Neural Information Processing Systems 37} (2024), pp. 84523--84605.
\end{itemize}

\section{Challenge 6: Adaptive Gradient Difference Clipping for Byzantine Robustness with Partial Participation}

\paragraph{Motivation.} In distributed learning, systems frequently encounter unreliable or malicious participants, commonly referred to as Byzantine workers. While various Byzantine fault tolerance mechanisms exist to protect model integrity, they traditionally rely on a highly restrictive assumption: full participation from all clients in every communication round. In practical federated environments, communication constraints and client unavailability make partial participation essential. However, integrating Partial Participation into Byzantine-robust algorithms introduces a critical vulnerability. If malicious workers dominate the subset of sampled clients in a given round, they can compromise the entire optimization process.

\paragraph{Contributions.} In Chapter \ref{chapter7}, we overcome this vulnerability by proposing \algname{Byz-VR-MARINA-PP}, the first distributed method that simultaneously accommodates partial participation, Byzantine-robustness, and communication compression, while providing provable guarantees, even in settings where malicious clients may form a majority among the sampled participants. The key innovation of \algname{Byz-VR-MARINA-PP} is the application of gradient clipping directly to stochastic gradient differences within a recursive variance reduction framework. This specific clipping mechanism strictly bounds the potential harm inflicted by malicious participants, ensuring algorithmic stability even during worst-case iterations in which every sampled client happens to be Byzantine.

Furthermore, we enhance the overall efficiency of \algname{Byz-VR-MARINA-PP} by incorporating communication compression in a way that preserves both robustness and convergence guarantees. Through rigorous analysis, we prove that the method achieves convergence rates matching current state-of-the-art theoretical results under general assumptions. To extend the impact of this work beyond our specific algorithm, we also introduce a general heuristic demonstrating how to adapt existing Byzantine-robust methods to partial participation settings using our clipping strategy.

\noindent\paragraph{Paper.} The chapter is based on the paper:

\begin{itemize}
    \item[{[\citenum{malinovsky2024byzantine}]}] Grigory Malinovsky, Peter Richt\'{a}rik, Samuel Horv\'{a}th, and Eduard Gorbunov. Byzantine-robustness and partial participation can be achieved at once: Just clip gradient differences. In \textit{Advances in Neural Information Processing Systems 37} (2024), pp. 34900--34979.
\end{itemize}

\section{Challenge 7: Provable Convergence for Low-Rank Adaptation Via Randomized Asymmetric Chains}

\paragraph{Motivation.} Fine-tuning remains the standard approach for adapting large foundational models to specific tasks. To manage the immense computational overhead of these models, parameter-efficient techniques like Low-Rank Adaptation have become essential. However, despite their empirical use, the theoretical optimization properties of these methods remain problematic. Specifically, standard low-rank adaptation, standalone asymmetric variants, and chains of standard low-rank adaptation all inherently suffer from severe convergence issues. These existing approaches frequently destabilize the optimization process, causing models to diverge or significantly underperform compared to the optimal baseline of full parameter fine tuning. This fundamental lack of theoretical reliability prevents their guaranteed success in rigorous deployments.

\paragraph{Contributions.} In Chapter \ref{chapter8}, we resolve these issues by introducing a novel framework built upon randomized asymmetric chains, resulting in the algorithm \algname{RAC-LoRA}. To explicitly illustrate the fundamental instability of existing approaches, we first construct a precise counterexample on a standard quadratic function. We then demonstrate a crucial theoretical distinction: while standard low-rank adaptation, asymmetric structures, and chains of standard adaptations fail independently, integrating them into the asymmetric chain underlying \algname{RAC-LoRA} perfectly stabilizes the optimization trajectory. Through rigorous mathematical analysis, we provide provable convergence guarantees for this combined approach, establishing that it reliably reaches the exact same optimal solution as full-parameter fine-tuning. Furthermore, we derive explicit convergence rates for smooth non-convex loss functions and successfully extend these rigorous theoretical guarantees to a broad range of optimization algorithms, including the stochastic and federated methods that form the core of this research. In conclusion, this shows that parameter-efficient methods can be applied in large-scale distributed settings while still maintaining the guarantees of analytically convergent gradient updates enjoyed by classical full-parameter methods. Through the development and analysis of \algname{Fed-RAC-LoRA}, we close this theoretical gap, making scalable and provably correct base-model adaptation more broadly accessible.

\noindent\paragraph{Paper.} The chapter is based on the paper:

\begin{itemize}
    \item[{[\citenum{malinovsky2024randomized}]}] Grigory Malinovsky, Umberto Michieli, Hasan Abed Al Kader Hammoud, Taha Ceritli, Hayder Elesedy, Mete Ozay, and Peter Richt\'{a}rik. Randomized asymmetric chain of LoRA: The first meaningful theoretical framework for low-rank adaptation. \textit{arXiv preprint arXiv:2410.08305} (2024).
\end{itemize}

\section{Excluded Papers}

All remaining publications authored during my doctoral studies have been excluded from this manuscript. Each of these exclusions was made for at least one of three reasons: to preserve a concise narrative, to prevent the thesis from becoming excessively long, or because my contribution to the collaborative project was secondary. Accordingly, the following works are not discussed:
\begin{itemize}
    \item Communication-efficient Gluon in federated learning [\citenum{qian2026communication}]
    \item Byzantine-robust and differentially private federated optimization under weaker assumptions [\citenum{islamov2026byzantine}]
    \item Improved convergence in parameter-agnostic error feedback through momentum [\citenum{sadiev2025improved}]
    \item First provable guarantees for practical private \gls{FL}: beyond restrictive assumptions [\citenum{shulgin2025first}]
    \item An optimal algorithm for strongly convex min-min optimization [\citenum{kovalev2025optimal}]
    \item Methods with local steps and random reshuffling for generally smooth non-convex federated optimization [\citenum{demidovich2024methods}]
    \item MAST: Model-agnostic sparsified training [\citenum{demidovich2023mast}]
    \item Revisiting stochastic proximal point methods: generalized smoothness and similarity [\citenum{tovmasyan2025revisiting}]
    \item MicroAdam: accurate adaptive optimization with low space overhead and provable convergence [\citenum{modoranu2024microadam}]
    \item Minibatch stochastic three points method for unconstrained smooth minimization [\citenum{boucherouite2024minibatch}]
    \item Streamlining in the riemannian realm: efficient Riemannian optimization with loopless variance reduction [\citenum{demidovich2024streamlining}]
    \item A guide through the zoo of biased SGD [\citenum{demidovich2023guide}]
    \item Random reshuffling with variance reduction: new analysis and better rates [\citenum{malinovsky2023random}]
    \item Improving accelerated federated learning with compression and importance sampling [\citenum{grudzien2023improving}]
    \item TAMUNA: doubly accelerated distributed optimization with local training, compression, and partial participation [\citenum{condat2023tamuna}]
    \item Federated learning with regularized client participation [\citenum{malinovsky2023federated}]
    \item Federated random reshuffling with compression and variance reduction [\citenum{malinovsky2022federated}].
\end{itemize}

\section{Basic Facts and Notation}

Before proceeding with the main results of this thesis, we detail the core notation and theoretical background utilized throughout the subsequent chapters.

\subsection{General notation}
Apart from the notation introduced in Section 1.1, we denote by $x^\star$ an optimal solution to problem \ref{eq:intro_main}. We operate under the standard assumption that the objective function is bounded from below, i.e., $f(x) \geq f^\star > -\infty$. Furthermore, we assume access to the gradients of $f_m(x)$ and $f_{m,i}(x)$, denoted by $\nabla f_m(x)$ and $\nabla f_{m,i}(x)$, respectively. In settings where the exact local gradient $\nabla f_m(x)$ is unavailable, we rely on a stochastic estimator $g_m(x)$, which is typically assumed to be unbiased (i.e., $\mathbb{E}[g_m(x)| x] = \nabla f_m(x)$) and to possess bounded variance. Explicit assumptions regarding variance bounds are detailed in the respective chapters.

Regarding vector operations, $\langle x, y \rangle = \sum_{j=1}^d x_j y_j$ denotes the standard Euclidean inner product of two vectors $x, y \in \mathbb{R}^d$. This inner product induces the $\ell_2$-norm in $\mathbb{R}^d$, defined as $\|x\| = \sqrt{\langle x, x \rangle}$. When discussing asymptotic convergence rates or complexity bounds, we rely on the standard big-$\mathcal{O}$ notation, denoted by $\mathcal{O}(\cdot)$.

\subsection{Smoothness, convexity, and Bregman divergence}
To facilitate rigorous convergence and complexity analyses throughout this thesis, we frequently assume that the global loss $f(x)$ and the local losses $f_m(x)$ satisfy certain structural properties, most notably smoothness and strong convexity.

\begin{definition}[Strong Convexity]
    A differentiable function $h\colon \mathbb{R}^d \to \mathbb{R}$ is $\mu$-strongly convex with constant $\mu \ge 0$ if, for all $x, y \in \mathbb{R}^d$:
    \begin{equation*}
        h(y) \ge h(x) + \langle \nabla h(x), y-x \rangle + \frac{\mu}{2}\|y-x\|^2.
    \end{equation*}
    If this inequality holds for $\mu = 0$, we say that $h$ is strictly convex.
\end{definition}

\begin{definition}[Smoothness]
    A differentiable function $h\colon \mathbb{R}^d \to \mathbb{R}$ is $L$-smooth with constant $L > 0$ if its gradient is Lipschitz continuous, meaning for all $x, y \in \mathbb{R}^d$:
    \begin{equation*}
        \|\nabla h(x) - \nabla h(y)\| \le L\|x-y\|.
    \end{equation*}
\end{definition}

To rigorously quantify the distance between points in the optimization landscape, we frequently utilize the Bregman divergence. For a differentiable function $f\colon \mathbb{R}^d \to \mathbb{R}$, the Bregman divergence is defined as:
\begin{equation*}
    D_f(x,y) \eqdef f(x) - f(y) - \langle \nabla f(y), x-y \rangle.
\end{equation*}
From this definition, it is straightforward to verify the following symmetric property for any $x, y \in \mathbb{R}^d$:
\begin{equation} \label{eq:sym_Bregman_} 
    \langle \nabla f(x) - \nabla f(y), x-y \rangle = D_f(x,y) + D_f(y,x).
\end{equation}

When the function $f$ is both $L$-smooth and $\mu$-strongly convex, the Bregman divergence is strictly bounded from above and below by quadratic distance metrics. Specifically, for all $x,y \in \mathbb{R}^d$, it holds that:
\begin{equation}\label{eq:Bregman_distance_bounds}
    \frac{\mu}{2} \|x-y\|^2 \le D_f(x,y) \le \frac{L}{2} \|x-y\|^2,
\end{equation}
and correspondingly, in terms of gradient distances:
\begin{equation}\label{eq:Bregman_gradient_bounds}
    \frac{1}{2L} \|\nabla f(x)-\nabla f(y)\|^2 \le D_f(x,y) \le \frac{1}{2\mu} \|\nabla f(x)-\nabla f(y)\|^2.
\end{equation}

\subsection{Regularization and proximity operators}
In many scenarios throughout this thesis, we explicitly minimize a composite objective of the form $f(x) + \psi(x)$, where $\psi\colon \mathbb{R}^d \to \mathbb{R} \cup \{\infty\}$ acts as a proper, closed, and convex regularizer. To handle this non-smooth component, optimization algorithms frequently leverage proximal steps based on the proximity operator.

\begin{definition}[Proximity Operator]
    For a proper, closed, and convex function $\psi\colon \mathbb{R}^d \to \mathbb{R} \cup \{+\infty\}$ and a parameter $\gamma > 0$, the proximity operator $\prox_{\gamma \psi}\colon \mathbb{R}^d \to \mathbb{R}^d$ is defined as:
\begin{equation*}
    \prox_{\gamma \psi}(x) \eqdef \argmin\limits_{y \in \mathbb{R}^d} \left\{ \gamma \psi(y) + \frac{1}{2} \|y-x\|^2 \right\}.
\end{equation*}
\end{definition}

Furthermore, the proximity operator exhibits a deep connection with convex duality. For the regularizer $\psi$, its Fenchel conjugate (or convex conjugate) is defined as:
\begin{equation*}
    \psi^*(y) \eqdef \sup_{x\in\mathbb{R}^d}\{\langle x, y \rangle - \psi(x)\}.
\end{equation*}
The proximity operator associated with this conjugate function satisfies a crucial implicit relationship. Specifically, for any step size $\tau > 0$, it holds that:
\begin{equation} \label{eq:prox_implicit_}
    \mathrm{if} \quad u = \prox_{\tau \psi^*}(y), \quad \mathrm{then} \quad u \in y - \tau\partial \psi^*(u),
\end{equation}
where $\partial \psi^*(u)$ denotes the subdifferential of the conjugate function at $u$. This property is instrumental in the design and analysis of proximal algorithms presented later in this thesis.

\begin{table}[t]
    \centering
    \small
    \caption{A summary of the results obtained in this thesis.}
    \label{tab:thesis_summary}
    \renewcommand{\arraystretch}{1.5} 
    \begin{tabular}{>{\raggedright\arraybackslash}p{0.25\textwidth} 
                    >{\raggedright\arraybackslash}p{0.55\textwidth} 
                    l}
        \toprule
        \multicolumn{1}{c}{\textbf{Challenge}} & \multicolumn{1}{c}{\textbf{Summary}} & \textbf{Chapter} \\
        \midrule
        \hspace{0pt}\algname{ProxSkip}: Achieving Accelerated Communication Complexity via Local Gradient Updates & 
We propose \gls{ProxSkip}, the first method proving acceleration via local training under heterogeneous data, achieving
$\mathcal{O}\left(\sqrt{\kappa}\log(1/\epsilon)\right)$
communication complexity and breaking the
$\mathcal{O}\left(\kappa\log(1/\epsilon)\right)$
barrier.
  & 
        Chapter \ref{chapter2} \\
        \hspace{0pt}\algname{ProxSkip-VR}: Balancing Resource Costs and Reducing Variance in Federated Optimization & 
We propose \gls{ProxSkip-VR}, a variance-reduced variant of \algname{ProxSkip} with lower total training cost. Our analysis is the first to combine the communication acceleration of \algname{ProxSkip} with the variance reduction induced by data sampling, within a general framework.
 & 
        Chapter \ref{chapter3} \\ 
        \hspace{0pt}\algname{5GCS}: Integrating Partial Participation into Accelerated Local Training & 
We propose \gls{5GCS} (\emph{5th Generation Accelerated Local Method with Client Sampling}), highlighting a new generation of methods achieving acceleration via local steps, with optimal communication complexity under partial participation (client sampling).
& 
        Chapter \ref{chapter4} \\
\hspace{0pt}\algname{NASTYA}: Investigating Server-Side Stepsizes and Random Reshuffling in Federated Optimization & 
We develop a federated learning method combining partial participation, local training, data shuffling, and server-side stepsizes, providing the first theoretical explanation of server-side stepsizes in FL together with a rigorous analysis of their interplay. 
& 
        Chapter \ref{chapter5} \\
\hspace{0pt}\algname{DIANA-RR}: Integrating Gradient Compression with Random Reshuffling and Local Computation & 
This paper studies gradient compression with Random Reshuffling and proposes improved methods using variance reduction for compression and local updates in distributed and federated settings.
& 
        Chapter \ref{chapter6} \\
\hspace{0pt}\algname{Byz-VR-MARINA-PP}: Adaptive Gradient Difference Clipping for Byzantine Robustness with Partial Participation & 
We propose a distributed method combining gradient clipping and variance reduction for Byzantine-robust learning under partial participation, allowing temporary majority of malicious clients while controlling their impact.
& 
        Chapter \ref{chapter7} \\
\hspace{0pt}\algname{RAC-LoRA}: Provable Convergence for Low-Rank Adaptation via Randomized Asymmetric Chains & 
This paper develops the first theoretical framework for \algname{LoRA}-type methods, showing that standard \algname{LoRA} is heuristic and may diverge. We propose \algname{RAC-LoRA}, a randomized asymmetric framework bridging full fine-tuning and low-rank adaptation, with convergence guarantees and federated extensions.
& 
        Chapter \ref{chapter8} \\        
        \bottomrule
    \end{tabular}
\end{table}


\chapter{ProxSkip: Achieving Accelerated Communication Complexity via Local Gradient Updates}
\label{chapter2}

\thispagestyle{empty}

\section{Introduction}
\label{sec:intro_proxskip}

We  study optimization problems of the  form
\begin{equation}
\min_{x\in \mathbb{R}^d} f(x) + \psi(x), \label{eq:main_proxskip}
\end{equation}
where $f \colon \mathbb{R}^d\to \mathbb{R}$ is a smooth  function, and $\psi \colon \mathbb{R}^d\to \mathbb{R} \cup \{+\infty\}$ is a proper, closed and convex regularizer. 

Such problems are ubiquitous, and appear in numerous applications associated with virtually all areas of science and engineering, including signal processing \citep{ProxSplit2009}, image processing \citep{Luke2020}, data science \citep{Parikh2014:proxbook} and machine learning~\citep{shai_book}. 

\subsection{Proximal gradient descent}

One of the most canonical methods for solving \eqref{eq:main_proxskip}, often used as the basis for further extensions and improvements, is \gls{ProxGD}, also known as the forward-backward algorithm~\citep{ProxSplit2009,Nesterov_composite2013}.  This method solves \eqref{eq:main_proxskip} via the iterative process defined by 
\begin{equation}\label{eq:ProxGD_proxskip} x^{t+1} = \prox_{\gamma^{(t)} \psi} (x^t - \gamma^{(t)} \nabla f(x^t)),
\end{equation}
where $\gamma^{(t)}>0$ is a suitably chosen stepsize at time $t$, and $\prox_{\gamma \psi}(\cdot)\colon\mathbb{R}^d\to \mathbb{R}^d$ is the proximity operator of $\psi$, defined via
\begin{equation}\label{eq:prox}
\squeeze \prox_{\gamma \psi}(x) \eqdef \arg \min \limits_{y \in \mathbb{R}^d} \left[\frac{1}{2}\|y-x\|^2+ \gamma \psi(y)\right]. \end{equation}


It is typically assumed that the proximity operator \eqref{eq:prox} can be evaluated in closed form, which means that the iteration \eqref{eq:ProxGD_proxskip} defining \algname{ \gls{ProxGD}} can be performed exactly.  \algname{ \gls{ProxGD}} is most suited to situations when the proximity operator is relatively cheap to evaluate, so that the bottleneck of \eqref{eq:ProxGD_proxskip} is in the forward step (i.e., computation of the gradient $\nabla f$) rather than in the backward step (i.e., computation of $\prox_{\gamma \psi}$). This is the case for many regularizers, including the $L_1$ norm ($\psi(x)=\|x\|_1$), the $L_2$ norm ($\psi(x) = \|x\|^2_2$), and elastic net \citep{ZouHastie:elastic-net:2005}. For many further examples, we refer the reader to the books~\citep{Parikh2014:proxbook,beck-book-first-order}.

\subsection{Expensive proximity operators}
However, in this work we are interested in the situation when the  evaluation of the {\em proximity operator is expensive}. That is, we assume that the computation of $\prox_{\gamma \psi}$ (the backward step)  is costly relative to the evaluation of the gradient of $f$ (the forward step). 

A conceptually simple yet rich class of expensive proximity operators arises from regularizers $\psi$ encoding a ``complicated-enough'' nonempty constraint set $\cC\subset \mathbb{R}^d$ via \begin{equation}\label{eq:indicator_proxskip}\psi(x)=\begin{cases}0 & x \in \cC \\ +\infty & x\notin \cC\end{cases}.\end{equation}
The evaluation of the proximity operator of $\psi$ given by \eqref{eq:indicator_proxskip} reduces to Euclidean projection onto $\cC$,
\[ \prox_{\gamma \psi}(x) = \arg \min_{y\in \cC} \|y-x\|, \]
 which can be a difficult optimization problem on its own. For instance, this is the case when $\cC$ is a polyhedral or a spectral set~\citep{Parikh2014:proxbook}. Other examples of expensive proximity operators include Schatten-$p$ norms of matrices (e.g., the nuclear norm), and certain variants of quadratic support functions~\citep{QSF2016}.

 \subsection{Distributed machine learning and consensus constraints}

An important example of  expensive proximity operators associated with indicator functions \eqref{eq:indicator_proxskip} arises in the {\em consensus} formulation of distributed optimization problems. In particular, consider the problem of minimizing the average of  $n$ functions using a cluster of  $n$ compute nodes/clients,
\begin{equation}\label{eq:finite-sum_097_proxskip} \min_{x\in \mathbb{R}^d} \left\{f(x)\eqdef \frac{1}{M} \sum_{m=1}^M f_m(x) \right\}, \end{equation}
where function $f_m\colon\mathbb{R}^d \to \mathbb{R}$, and the data describing it, is owned by and stored on client $m \in [M]\eqdef \{1, 2, \dots, M\}$. This problem is of key importance in Machine Learning as it is an abstraction of the {\em empirical risk minimization} \citep{shai_book}, which is currently the dominant paradigm for training supervised Machine Learning models.

 By cloning the model $x\in \mathbb{R}^d$ into $M$ independent copies $x_1,\dots, x_M \in \mathbb{R}^d$, problem \eqref{eq:finite-sum_097_proxskip} can be reformulated into the  {\em consensus form}~\citep{Parikh2014:proxbook}
\begin{equation}\label{eq:mainFL}
 \min_{x_1,\dots, x_M \in \mathbb{R}^d}  \frac{1}{M} \sum_{m=1}^M f_m(x_m) + \psi(x_1,\dots, x_M)\,,
\end{equation}
where the regularizer $\psi\colon\mathbb{R}^{Md}\to \mathbb{R}$ given by
\begin{equation}\label{eq:constraint_consensus_0980_proxskip}
\psi(x_1,\dots, x_M) \eqdef 
	\begin{cases}
		0, & \textrm{if } x_1=\dotsb=x_M\,, \\
		+\infty, & \textrm{otherwise},
	\end{cases}
\end{equation}
encodes the consensus constraint \[\cC\eqdef \{(x_1,\dots,x_M) \in \mathbb{R}^{Md} \;:\; x_1=\dotsb=x_M\}.\]

Evaluating the proximity operator of \eqref{eq:constraint_consensus_0980_proxskip} is not computationally expensive as it simply amounts to taking the average of the variables \citep{Parikh2014:proxbook}:
\begin{equation}\label{eq:prox-avg_proxskip} \prox_{\gamma \psi}(x_1,\dots,x_M) = (\bar{x},\dots,\bar{x})\in \mathbb{R}^{Md}, \end{equation}
where
\begin{equation}\label{eq:avg_proxskip}\bar{x} \eqdef \frac{1}{M}\sum_{m=1}^M x_m.\end{equation}
 However, it often involves \emph{high communication cost} since the vectors $x_1, \dots, x_M$  are stored on different compute nodes. Indeed, even simple averaging can be very time consuming if the communication links connecting the clients (e.g., through an orchestrating server) are slow and the dimension $d$ of the aggregated vectors/models high, which is the case in {\em Federated Learning (\gls{FL})}~\citep{FedLearn2016, kairouz2019advances}. 

\begin{table*}[!t]
	\centering
	\caption{The performance of Federated Learning methods employing multiple local gradient steps in the strongly convex regime.}\label{tbl:main-2}
	\begin{threeparttable}
		\tiny\setlength\tabcolsep{4.pt} 
		\begin{tabular}{ c  c  c  c  c  c  c }
			\toprule[.1em]
			 \begin{tabular}{c}\bf method \end{tabular} &  \begin{tabular}{c}\bf \# local steps \\  \bf per round \end{tabular} &\begin{tabular}{c}\bf \# floats sent  \\  \bf per round \end{tabular}   & \begin{tabular}{c}\bf stepsize \\ \bf on client $i$  \end{tabular}  & 
			\begin{tabular}{c} \bf linear \\ \bf rate? \end{tabular} & \begin{tabular}{c} \bf \# rounds \tnote{\color{blue}(c) }  \end{tabular}   & \begin{tabular}{c}\bf rate better  \\ \bf than \algname{GD}? \end{tabular} \\
			\midrule
			\begin{tabular}{c}\algname{GD}  {\tiny[\citenum{NesterovBook}]}\end{tabular}  & $1$ & $d$&$\frac{1}{L}$ & \cmark & $\tilde{\cO}(\kappa)$  & \xmark \\ \cmidrule{1-1}
			\makecell{\algname{LocalGD} {\tiny[\citenum{khaled2019first, localSGD-AISTATS2020}]}} & $\tau$  &$d$ & $\frac{1}{\tau L}$ & \xmark  & $\cO\left(\frac{G^2}{\mu n \tau \varepsilon}\right) $\tnote{\color{blue}(d)} & \xmark \\ \cmidrule{1-1}
			\makecell{\algname{Scaffold} {\tiny[\citenum{karimireddy2020scaffold}]}} & $\tau$ & $2d$~\tnote{\color{blue}(e)}& $\frac{1}{\tau L}$ \tnote{\color{blue}(e)}& \cmark & $\tilde{\cO}(\kappa)$ & \xmark \\ \cmidrule{1-1}
			\makecell{\algname{S-Local-GD} \tnote{\color{blue}(a)}\quad {\tiny[\citenum{gorbunov2021local}]}  }  & $\tau$ &$d<\#<2d$ \tnote{\color{blue}(f)}& $\frac{1}{\tau L}$  & \cmark & $\tilde{\cO}(\kappa)$  & \xmark \\ \cmidrule{1-1}
			\makecell{\algname{FedLin} \tnote{\color{blue}(b)}\quad {\tiny[\citenum{mitra2021linear}]}} & $\tau_i$&$2d$  & $\frac{1}{\tau_i L} $ & \cmark &$\tilde{\cO}(\kappa)$  & \xmark \\
			\bottomrule[.1em] 
			\cellcolor{bgcolor} \begin{tabular}{c}\algname{Scaffnew}\tnote{\color{blue}(g)}\quad (this work) \\ for any $p\in (0,1]$ \end{tabular} & \cellcolor{bgcolor}$\frac{1}{p}$~\tnote{\color{blue}(h)} &\cellcolor{bgcolor}$d$&\cellcolor{bgcolor} $\frac{1}{L}$ & \cellcolor{bgcolor}\cmark & \cellcolor{bgcolor}$\tilde{\cO}\left(p \kappa + \frac{1}{p}\right)$ &\cellcolor{bgcolor} \begin{tabular}{c}{\cmark}\\  $p \in \left(\frac{1}{\kappa},1\right)$\end{tabular} \\			
			\hline
			\cellcolor{bgcolor} \begin{tabular}{c}\algname{Scaffnew}\tnote{\color{blue}(g)}\quad  (this work) \\ for optimal $p=\frac{1}{\sqrt{\kappa}}$ \end{tabular} & \cellcolor{bgcolor}$\sqrt{\kappa}$~\tnote{\color{blue}(h)} &\cellcolor{bgcolor}$d$&\cellcolor{bgcolor} $\frac{1}{L}$ & \cellcolor{bgcolor}\cmark & \cellcolor{bgcolor}$\tilde{\cO}(\sqrt{\kappa})$ &\cellcolor{bgcolor} \cmark \\			
			\hline			
		\end{tabular} 
	\begin{tablenotes}
		{\scriptsize      
			\item [{\color{blue}(a)}] This is a special case of \algname{S-Local-SVRG}, which is a more general method presented in \citep{gorbunov2021local}. \algname{S-Local-GD} arises as a special case when full gradient is computed on each client.      
			\item [{\color{blue}(b)}]  \algname{FedLin} is a variant with a fixed but different number of local steps for each client. Earlier method \algname{S-Local-GD} has the same update but random loop length. 
			\item[{\color{blue}(c)}] The $\tilde{\cO}$ notation used in this column hides logarithmic factors.
			\item[{\color{blue}(d)}] $G$ is the level of dissimilarity at the solution $x_*$: $G^2 = \frac{1}{M}\sum_{m=1}^M \|\nabla f_m(x_*)\|^2$.
			\item[{\color{blue}(e)}] The number of floats used by Scaffold is $d$ when using Option II without Partial Participation. For the stepsize on client $i$, we use  \algname{Scaffold}'s cumulative local-global stepsize $\eta_l \eta_g$ for a  fair comparison.  
			\item[{\color{blue}(f)}] The number of sent vectors depends on hyper-parameters, and it is randomized.
			\item[{\color{blue}(g)}] \algname{Scaffnew} (Algorithm~\ref{alg:fl}) = \gls{ProxSkip} (Algorithm~\ref{alg:ProxSkip}) applied to the consensus formulation \eqref{eq:mainFL} + \eqref{eq:constraint_consensus_0980_proxskip} of the finite-sum problem \eqref{eq:finite-sum_097_proxskip}.
			\item[{\color{blue}(h)}] \gls{ProxSkip} (resp.\ \algname{Scaffnew}) takes a {\em random} number of gradient (resp.\ local) steps before  prox (resp.\ communication) is computed (resp.\ performed). What is shown in the table is the {\em expected} number of gradient (resp.\ local) steps.	}
	\end{tablenotes}  		
	\end{threeparttable}

\end{table*}  

\subsection{Federated learning}

For the above reasons, practical \gls{FL} methods use various communication reduction mechanisms to achieve a useful computation-to-communication ratio, such as delayed communication. That is, the methods perform multiple local steps independently, based on their local objective~\citep{Mangasarian1994,mcdonald2010distributed,zhang2016parallel,
mcmahan2017communication,stich2018,lin2018don}.

However, when all the local functions $f_m$ are \emph{different} (i.e., when each individual machine has data drawn from a different distribution), local steps introduce
a {\em drift} in the updates of each client, which results in convergence issues. Indeed, even in the case of the simplest local gradient-type method, \algname{LocalGD}, a theoretical understanding that would not require any data similarity/homogeneity assumptions eluded the community for a long time. A resolution was found only recently~\citep{khaled2019first, localSGD-AISTATS2020, koloskova2020unified}. However, the rates obtained in these works paint a pessimistic picture for \algname{LocalGD}; for example, due to client drift, they are sublinear even for smooth and  strongly convex problems.

The next task for the \gls{FL} community was to propose algorithmic adjustments that could provably mitigate the client drift issue. A   handful of recent methods, including
 \algname{Scaffold}~\citep{karimireddy2020scaffold}, \algname{S-Local-GD} \citep{gorbunov2021local} and \algname{FedLin}~\citep{mitra2021linear}, managed to do that.  For instance, under the assumption that $f$ is $L$-smooth and $\mu$-strongly convex, with condition number $\kappa = \nicefrac{L}{\mu}$, \algname{Scaffold}, \algname{S-Local-GD} and \algname{FedLin}  obtain a $\cO(\kappa \log \nicefrac{1}{\varepsilon})$ communication complexity, which matches the communication complexity of \algname{\gls{GD}} (that computes a single gradient on every client per round of communication).  However, and despite the empirical superiority of these methods over vanilla \algname{\gls{GD}}, their theoretical communication complexity does {\em not} improve upon \algname{\gls{GD}}. This reveals a fundamental gap in our understanding of local methods.
 
Due to the enormous effort that was exerted over the last several years  by the \gls{FL} community in this direction without it bearing the desired fruit~\citep{kairouz2019advances},  it seems very challenging to establish theoretically that performing independent local updates improves upon the communication complexity of \algname{\gls{GD}}.  In contrast, accelerated gradient descent (without local steps)
 can reach the optimal  $\cO(\sqrt{\kappa} \log \nicefrac{1}{\varepsilon})$ communication complexity~\citep{Lan2012:ac-sc,woodworth2020minibatch,woodworth21_min_max_compl_distr_stoch}.
 
 This raises the question of whether this is a fundamental limitation of local methods. Is it possible to prove a better communication complexity than $\cO(\kappa \log \nicefrac{1}{\varepsilon})$ for {\em simple} local gradient-type methods, without resorting to any explicit acceleration mechanisms?

\section{Contributions}
\label{sec:contributions}

We now summarize the main contributions of this work.

\subsection{ProxSkip: A general prox skipping Algorithm}
We develop a new \algname{ \gls{ProxGD}}-like algorithm for solving the general regularized problem \eqref{eq:main_proxskip}. Our method,  which we call \gls{ProxSkip} (see Algorithm~\ref{alg:ProxSkip}), is designed to handle expensive proximal operators.  

A key ingredient in its design is a {\em randomized prox-skipping procedure}: in each iteration of \gls{ProxSkip}, we evaluate the proximity operator with probability $p \in (0,1]$. If $p=1$, several steps in our method are vacuous, and we recover \algname{ \gls{ProxGD}} as a special case (and the associated standard theory). 
Of course, the interesting choice  is $0<p<1$.  In expectation, the proximity operator is evaluated every $\nicefrac{1}{p}$ iterations, which can be very rare if $p$ is small.

{\bf Control variates stabilizing prox skipping.} We had  to introduce several new algorithmic design adjustments for such a  method to provably work.  In particular, \gls{ProxSkip} uses a control  variate $\red h^t$ on line 3 to shift the gradient $\nabla f(x^t)$ when the forward step is performed. 

Note that $\red h^t$ stays constant in between two consecutive prox calls. Indeed, this is because in that case we have $x^{t+1} = \hat x^{t+1}$  from line 8, and line 10 therefore simplifies to $\red h^{t+1} = \red h^t$.  So, when operating in between two prox calls, our method performs iterations of the form
\[x^{t+1} = x^t - \gamma (\nabla f (x^t) - {\red h^t}),\] where $\gamma>0$ is a stepsize parameter. 
When a prox step is executed, both the iterate $x^t$ and the control variate $\red h^t$ are adjusted, and the process is repeated.

This control mechanism  is necessary to allow for prox-skipping  to work.
To illustrate this, consider an optimal point $x^{\star} =\argmin_{x} f(x)+ \psi(x)$. In general, it does not hold $\nabla f(x^{\star})=0$, so skipping the prox (without control variate adjustment) would imply a  \emph{drift away} from $x^{\star}$. 
We show below that the control variate converges to 
\[{\red h^t} \to \nabla f(x^{\star}),
\]
which means that $x^{\star}$ is a  fixed point. This allows skipping the prox for a significant amount of steps without impacting the convergence.

\begin{algorithm*}[t]
	\caption{\gls{ProxSkip}}
	\label{alg:ProxSkip}
	\begin{algorithmic}[1]
		\STATE \textbf{Input:} stepsize $\gamma > 0$, probability $p>0$, initial iterate $x^0\in \mathbb{R}^d$, initial control variate ${\red h^0} \in \mathbb{R}^d$, number of iterations $T\geq 1$
		\FOR{$t=0,1,\dotsc,T-1$}
		\STATE $\hat x^{t+1} = x^t - \gamma (\nabla f (x^t) - {\red h^t})$ \hfill $\diamond$ Take a gradient-type step\\ adjusted via the control variate ${\red h^t}$
		\STATE Flip a coin $\theta_t \in \{0,1\}$ where $\mathop{\rm Prob}(\theta_t =1) = p$ \hfill $\diamond$ Flip a coin that decides\\  whether to skip the prox or not
		\IF{$\theta_t=1$} 
		\STATE  $x^{t+1} = \prox_{\frac{\gamma}{p}\psi}\bigl(\hat x^{t+1} - \frac{\gamma}{p}{\red h^t} \bigr)$ \hfill $\diamond$ Apply prox, but only very rarely!\\  (with small probability $p$)
		\ELSE
		\STATE $x^{t+1} = \hat x^{t+1}$ \hfill $\diamond$ Skip the prox!
		\ENDIF
		\STATE ${\red h^{t+1}} = {\red h^t} + \frac{p}{\gamma}(x^{t+1} - \hat x^{t+1})$ \hfill $\diamond$ Update the control variate ${\red h^t}$
		\ENDFOR
	\end{algorithmic}
\end{algorithm*}

{\bf Theory.} If $f$ is $L$-smooth and $\mu$-strongly convex, we prove that \gls{ProxSkip} converges at a linear rate. In particular, we show that after $T$ iterations, 
\[\E{\Psi^{(T)}}
		\le (1 - \min\{\gamma\mu, p^2\})^T \Psi^{(0)},\]
where $\Psi^{(t)}$ is a certain Lyapunov function  (see \eqref{eq:Lyapunov_def}) involving both $x^{t}$ and $\red h^{t}$. If we choose $\gamma = \nicefrac{1}{L}$ and $p=\nicefrac{1}{\sqrt{\kappa}}$, where $\kappa = \nicefrac{L}{\mu}$ is the condition number, then the iteration complexity of \gls{ProxSkip}  is $\cO(\kappa\log\nicefrac{1}{\varepsilon})$, whereas the number of prox evaluations (in expectation) is $\cO(\sqrt{\kappa}\log\nicefrac{1}{\varepsilon})$ only! 
 For more details related to theory, see Section~\ref{sec:theory}.

 \subsection{Scaffnew: ProxSkip applied to federated learning}

When applied to the consensus reformulation \eqref{eq:mainFL}--\eqref{eq:constraint_consensus_0980_proxskip} of problem \eqref{eq:finite-sum_097_proxskip}, \gls{ProxSkip} can be interpreted as a new distributed gradient-type method performing local steps, adding to the existing rich literature on local methods. 
In this context, we decided to call our method \algname{Scaffnew} (Algorithm~\ref{alg:fl}).\footnote{\scriptsize This is a homage to the influential \algname{Scaffold} method \citep{karimireddy2020scaffold}, which in our experiments performs very similarly to \algname{Scaffnew} if the former method is used with fine-tuned stepsizes. } 
Since prox evaluation now means communication via averaging across the nodes (see \eqref{eq:prox-avg_proxskip} and \eqref{eq:avg_proxskip}), and since \algname{Scaffnew} inherits the strong theoretical prox-skipping properties of its parent method \gls{ProxSkip}:
\begin{quote}\em We resolve one of the most important open problems in the \gls{FL} literature: breaking the $\cO(\kappa \log \nicefrac{1}{\varepsilon})$ communication complexity barrier with a simple local method. In particular,  \algname{Scaffnew} reaches an $\cO(\sqrt{\kappa} \log \nicefrac{1}{\varepsilon})$ communication complexity without imposing any additional assumptions (e.g., data similarity or stronger smoothness assumptions). 
\end{quote}

Note that since the iteration complexity of \algname{Scaffnew}  is $\cO(\kappa \log \nicefrac{1}{\varepsilon})$, the number of local steps per communication round is (on average)  $\cO(\sqrt{\kappa})$. According to Arjevani and Shamir (2015)~\citep{Arjevani2015:communication}, the communication lower bound for first order distributed algorithms is $\cO(\sqrt{\kappa} \log \nicefrac{1}{\varepsilon})$. This means that \algname{Scaffnew} is optimal in terms of communication rounds. \looseness=-1

Please refer to Table~\ref{tbl:main-2} in which we compare our results with the results obtained by existing state-of-the-art methods.

\subsection{Extensions}

We develop two extensions of the vanilla \gls{ProxSkip} method; see Section~\ref{sec:extensions}. We are not attempting to be exhaustive: these extensions are meant to illustrate that our method and proof technique combine well with other tricks and techniques often used in the literature. 

{\bf From deterministic to stochastic gradients.} First, in Section~\ref{sec:stochastic} we  perform an extension enabling us to use a {\em stochastic gradient} $g^{t}(x^t)\approx \nabla f(x^t)$ in  \gls{ProxSkip}  instead of the true gradient $\nabla f(x^t)$.  This is of importance in many applications, and is of particular importance for our method since now that the cost of the prox step was reduced, the cost of the gradient steps becomes more important. We operate under the modern {\em expected smoothness} assumption introduced in \citep{gower2019sgd,gower2021stochastic}, which is less restrictive than the standard bounded variance assumption.

{\bf From a central server to fully decentralized training.} Second, in Section~\ref{sec:decentralized} we present and analyze  \gls{ProxSkip} in a fully {\em decentralized} optimization setting, where the communication between nodes is restricted to a communication graph.
Our decentralized algorithm inherits the  property that it is not affected by data-heterogeneity. 
The control variate technique in \gls{ProxSkip} resembles, to some extent, some of the existing \emph{gradient tracking} mechanisms~\citep{Lorenzo2016GT-first-paper,Nedic2016DIGing}.  However, while gradient tracking provably addresses data-heterogeneity, its communication complexity scales proportional to the iteration complexity, $\cO(\kappa)$~\citep{Yuan2021d2-exact-diff-rates,koloskova2021improved}.
The same holds for almost all other schemes that have been designed to address data-heterogeneity in decentralized optimization \citep{Tang2018:d2,Vogels2021:relay}. Notable exceptions include the optimal methods developed in \citep{OPAPC,ADOM,ADOM+}; see also the references therein. However, these methods are based on classical acceleration schemes, and do not perform multiple local steps. 

\section{Theory}
\label{sec:theory}

We are now ready to describe our key theoretical development: the convergence analysis of \gls{ProxSkip}.

\subsection{Assumptions}
We rely on several standard  assumptions to establish our results. First, we need $f$ to be smooth and strongly convex (see Appendix~\ref{appendix:facts} for complementary details).
\begin{assumption}\label{as:f}
	$f$ is $L$-smooth and $\mu$-strongly convex.
	\end{assumption}

We also need the following standard assumption\footnote{\scriptsize Note that this assumption is automatically satisfied for $\psi$ defined in~\eqref{eq:constraint_consensus_0980_proxskip}.} on the regularizer $\psi$.

\begin{assumption}\label{as:proper_psi}
$\psi$ is proper, closed and convex.
\end{assumption}

These assumptions imply that problem \eqref{eq:main_proxskip} has a unique minimizer, which we denote $x^{\star}\eqdef \argmin f(x)+\psi(x)$.

\subsection{Firm nonexpansiveness}
In one step of our analysis we will rely on firm nonexpansiveness of the proximity operator \citep{bauschke2021generalized}: \begin{lemma}\label{lem:prox-contraction} 
	Let \Cref{as:proper_psi} be satisfied. Let $P(x)\eqdef \prox_{\frac{\gamma}{p}\psi}(x)$ and $Q(x) \eqdef x - P(x)$. Then 
	\begin{equation}
		\|P(x)-P(y)\|^2 + \|Q(x) - Q(y)\|^2
		\le \|x - y\|^2, \label{eq:prox_firm_non_exp}
	\end{equation}
	for all $x, y \in \mathbb{R}^d$ and any $\gamma,p>0$.
\end{lemma}

\subsection{Two technical lemmas}

The strength of our method comes from the role the control variates $\red h^t$ play in stabilizing the effect of skipping prox evaluations. Our analysis captures this effect. In particular, a by-product of our analysis is a proof that the control variates converge to  
$h^{\star}\eqdef \nabla f(x^{\star})$, where $x^{\star}$ is the solution. In order to show this, we work with the following natural candidate for a Lyapunov function: \begin{equation} \label{eq:Lyapunov_def} \Psi^{(t)} \eqdef \|x^{t} - x^{\star}\|^2 + \frac{\gamma^2}{p^2}\|h^{t} - h^{\star}\|^2\,.\end{equation}
We  further define
\begin{equation}\label{eq:98g9gbjfd8d}\squeeze w^t \eqdef x^t - \gamma \nabla f(x^t), \quad \text{and} \quad w^{\star} \eqdef x^{\star} - \gamma \nabla f(x^{\star}).\end{equation} 
Note that if our method works, i.e., if $x^t\to x^{\star}$, then gradient smoothness  implies that $w^t \to w^{\star}$. In our first technical lemma, we show that after one step of \gls{ProxSkip}, the Lyapunov function can be bounded in terms of the distance $\|w^t - w^{\star}\|^2$ and the control variate error $\|h^{t} - h^{\star}\|^2$. It is this lemma in whose proof we rely on firm nonexpansiveness. We do not use it anywhere else.
	
\begin{lemma}\label{lem:A}
If Assumptions \ref{as:f} and \ref{as:proper_psi} hold, $\gamma >0$ and $0<p\leq 1$, then
\begin{equation} \label{eq:b8f9d89fd8df_09}\squeeze \E{ \Psi^{(t+1)}  } \leq  \|w^t - w^{\star}\|^2 +  (1-p^2)\frac{\gamma^2}{p^2}\|h^{t} - h^{\star}\|^2 \,, \end{equation}
where the expectation is taken over the $\theta_t$ in Algorithm~\ref{alg:ProxSkip}.
\end{lemma}

 Our next lemma bounds the first term in the right-hand side of \eqref{eq:b8f9d89fd8df_09} by a multiple of $\|x^t-x^{\star}\|^2$.
\begin{lemma} \label{lem:B} Let \Cref{as:f} hold with any $\mu\geq 0$. If $0< \gamma \leq \frac{1}{L}$, then \begin{equation}\label{eq:nbo98fd8f_09uf}\|w^{t} - w^{\star}\|^2 \le (1-\gamma\mu)\|x^t - x^{\star}\|^2.\end{equation}
\end{lemma}

\subsection{Main Theorem}

As we shall now see, our main theorem follows simply by combining the last two lemmas.

\label{sec:maintheorem}
\begin{theorem}
 \label{thm:main}
	Let \Cref{as:f} and \Cref{as:proper_psi} hold, and let $0<\gamma \le \frac{1}{L}$ and $0<p\leq 1$. Then, the iterates of \gls{ProxSkip} (\Cref{alg:ProxSkip}) satisfy
	\begin{equation}\label{eq:lyapunov_decrease}
		\E{\Psi^{(T)}}
		\le (1 - \zeta)^T \Psi^{(0)} ,
	\end{equation}
	where $\zeta\eqdef \min\{\gamma\mu, p^2\}$.
\end{theorem}

\begin{proof}
By combining Lemmas~\ref{lem:A} and \ref{lem:B}, we get
	\begin{align*}
		\E{\Psi^{(t+1)} }
				&\le (1 - \gamma\mu)\|x^t - x^{\star}\|^2 + (1-p^2)\frac{\gamma^2}{p^2}\|h^{t} - h^{\star}\|^2 \\
		&\le (1-\zeta)\left(\|x^t - x^{\star}\|^2 + \frac{\gamma^2}{p^2}\|h^{t} - h^{\star}\|^2 \right) \\
		& = (1-\zeta) \Psi^{(t)}.
	\end{align*}
	We get the theorem's claim by unrolling the recurrence.
\end{proof}

\subsection{How often should one skip the prox?}

Note that by choosing $p=1$ (no prox skipping) and $\gamma=\nicefrac{1}{L}$, we get $\zeta=\nicefrac{1}{\kappa}$, which leads to  the rate $\cO(\kappa \log \nicefrac{1}{\varepsilon})$ of \algname{\gls{ProxGD}}. This is not a surprise since when $p=1$, \gls{ProxSkip} {\em is} identical to \algname{\gls{ProxGD}}. 

More importantly, note that for any fixed stepsize $\gamma>0$, the reduction factor $\zeta\eqdef \min\{\gamma\mu, p^2\}$ in \eqref{eq:lyapunov_decrease} remains unchanged as we decrease $p$ from $1$ down to $p=\nicefrac{1}{\sqrt{\gamma \mu}}$. This is the reason why we can often {\em skip the prox}, and get away with it {\em for free}, i.e., without any deterioration of the convergence rate! 

 By inspecting \eqref{eq:lyapunov_decrease} it is easy to see that \begin{equation}\label{eq:nbi9fgd9gfd9_90}T\geq \max \left\{\frac{1}{\gamma \mu}, \frac{1}{p^2}\right\} \log \frac{1}{\varepsilon} \; \Longrightarrow \; \E{\Psi^{(T)}} \leq \varepsilon \Psi^{(0)} .\end{equation} Since in each iteration we evaluate the prox with probability $p$, the {\em expected number of 
prox evaluations} is \begin{equation}\label{eq:pT}pT \overset{\eqref{eq:nbi9fgd9gfd9_90}}{\approx}  \max \left\{\frac{p}{\gamma \mu},\frac{1}{p} \right\}\log \frac{1}{\varepsilon}.\end{equation}
Clearly, the best result is obtained if we use the largest stepsize allowed by \Cref{thm:main}: \begin{equation}\label{eq:niubfd_0909}\gamma =\frac{1}{L}.\end{equation} 
Next, the value of $p$ that minimizes expression \eqref{eq:pT} satisfies $\frac{pL}{ \mu}=\frac{1}{p}$, which gives the {\em optimal probability} \begin{equation}\label{eq:iuufd7-72332}p=\sqrt{\frac{\mu}{L}}=\frac{1}{\sqrt{\kappa}},\end{equation}
where $\kappa\eqdef \nicefrac{L}{\mu}$ is the condition number. With these optimal choices of the parameters $\gamma$ and $p$, the number of iterations of \gls{ProxSkip} is 
\[T\overset{\eqref{eq:nbi9fgd9gfd9_90}}{\approx}  \max\left\{\frac{1}{\gamma \mu} , \frac{1}{p^2}\right\}\log \frac{1}{\varepsilon} \overset{\eqref{eq:niubfd_0909}+\eqref{eq:iuufd7-72332}}{=}  \kappa \log \frac{1}{\varepsilon},
\]
and the expected number of prox evaluations performed in the process is 
\[
pT \overset{\eqref{eq:pT}}{\approx}   \max \left\{\frac{p}{\gamma \mu},\frac{1}{p} \right\}\log \frac{1}{\varepsilon} \overset{\eqref{eq:niubfd_0909}+\eqref{eq:iuufd7-72332}}{=} 
\sqrt{\kappa}\log \frac{1}{\varepsilon}.
\]

 Let us summarize the above findings.

\begin{corollary}
	If we choose $\gamma = \nicefrac{1}{L}$ and $p=\nicefrac{1}{\sqrt{\kappa}}$, then the iteration complexity of \gls{ProxSkip} (\Cref{alg:ProxSkip}) is $\cO(\kappa\log\nicefrac{1}{\varepsilon})$ and its prox calculation complexity is $\cO(\sqrt{\kappa}\log\nicefrac{1}{\varepsilon})$.
\end{corollary}

\begin{proof} Clearly, if $T\geq \kappa \log \nicefrac{1}{\varepsilon}$, then 
$\E{\Psi^{(T)}} \leq \varepsilon \Psi^{(0)} $. Each of these iterations performs a prox independently with probability $p$, and hence the expected number of prox evaluations is $p \kappa \log \nicefrac{1}{\varepsilon} =  \sqrt{\kappa} \log \nicefrac{1}{\varepsilon}$.
\end{proof}

\section{Application to Federated Learning}
Let us now consider the problem of minimizing 
the average of $M$ functions stored on $M$ devices, as formulated in~\eqref{eq:finite-sum_097_proxskip}.
This is the canonical problem in Federated Learning~\citep{mcmahan2017communication,kairouz2019advances}.\footnote{\scriptsize
As our focus is on a new communication-efficient scheme, we  disregard here other important aspects such as Partial Participation.} 
In this setting the functions $f_m \colon \mathbb{R}^d \to \mathbb{R}$ denote the local loss function of client $m$ defined over its own private data. For simplicity, we assume in this section that every client can compute the gradient $\nabla f_m(x)$ exactly (i.e., a full pass over the local data), see Section~\ref{sec:stochastic} for the discussion of the stochastic setting. When applied to the consensus reformulation \eqref{eq:mainFL}--\eqref{eq:constraint_consensus_0980_proxskip} of problem \eqref{eq:finite-sum_097_proxskip}, \gls{ProxSkip} reduces to \algname{Scaffnew} (Algorithm~\ref{alg:fl}).

\begin{algorithm*}[t]
	\caption{\algname{Scaffnew}: Application of \gls{ProxSkip} to Federated Learning}
	\label{alg:fl}
	
	\let\oldwhile\algorithmicwhile
	\renewcommand{\algorithmicwhile}{\textbf{in parallel on all workers $m \in [M]$}}
	\let\oldendwhile\algorithmicendwhile
	\renewcommand{\algorithmicendwhile}{\algorithmicend\ \textbf{local updates}}
	
	\begin{algorithmic}[1]
		\STATE stepsize $\gamma > 0$, probability $p>0$, initial iterate $x^0_{1}=\dots = x^0_{M}\in \mathbb{R}^d$, initial control variates ${\red h_{1}^{0}, \dots, h_{M}^{0}}  \in \mathbb{R}^d$  on each client such that  $\sum_{m=1}^{M}{\red h_{m}^{0}} = 0$, number of iterations $T\geq 1$
		\STATE \textbf{server:} flip a coin, $\theta_t \in \{0,1\}$, $T$ times,  where $\mathop{\rm Prob}(\theta_t =1) = p$\hfill $\diamond$ Decide when to skip communication
		\STATE send the sequence  $\theta_0, \dots, \theta_{T-1}$ to all workers 
		\FOR{$t=0,1,\dotsc,T-1$}
		\WHILE{}
		\STATE $\hat x_{m}^{t+1} = x_{m}^{t} - \gamma (g_{m}^{t} (x_{m}^{t}) - {\red h_{m}^{t}})$ \hfill $\diamond$ Local gradient-type step  adjusted via the local control variate ${\red h_{m}^{t}}$
		\IF{$\theta_t=1$} 
		\STATE  $x_{m}^{t+1} = \frac{1}{M}\sum \limits_{m=1}^M \hat x_{m}^{t+1}$ \hfill  $\diamond$ Average the iterates, but only very rarely! \\ (with small probability $p$)
		\ELSE
		\STATE $x_{m}^{t+1} = \hat x_{m}^{t+1}$ \hfill $\diamond$ Skip communication!
		\ENDIF
		\STATE ${\red h_{m}^{t+1}} = {\red h_{m}^{t}} + \frac{p}{\gamma}(x_{m}^{t+1} - \hat x_{m}^{t+1})$ \hfill $\diamond$ Update the local control variate ${\red h_{m}^{t}}$
		\ENDWHILE
		\ENDFOR
	\end{algorithmic}
\end{algorithm*}

{\bf Method description.} 
Algorithm~\ref{alg:fl} has three main steps:\ local updates  to the client model $x_{m}^{t} \in \mathbb{R}^d$, local updates to the client control variate ${\red h_{m}^{t}}\in \mathbb{R}^d$, and averaging the client models with probability $p$ in every iteration.

When $g_{m}^{t}(x_{m}^{t}) = \nabla f_m(x_{m}^{t})$, then each local update on client $m$ takes the form
\begin{align*}
 \squeeze\hat x_{m}^{t+1} = x_{m}^{t} - \gamma(\nabla f_m(x_{m}^{t}) - {\red h_{m}^{t}})\,.
\end{align*}
We will show below that ${\red h_{m}^{t} } \stackrel{t \to \infty}{\to} \nabla f_m(x^{\star}),$ so that it becomes evident that the optimal solution $x^\star$ is a fixed point of the algorithm (this is a key difference from, e.g., \algname{LocalGD} \citep{localSGD-AISTATS2020, koloskova2020unified,malinovskiy2020local}). The local control variates ${\red h_{m}^{t}}$ are updated after each communication round, i.e., when $\theta_t=1$, as
\begin{align*}
 {\red h_{m}^{t+1}} = {\red h_{m}^{t}} + \frac{p}{\gamma} \underbrace{\left(\frac{1}{M}\sum_{m^\prime=1}^M \hat x_{m^\prime}^{t+1}  - \hat x_{m}^{t+1} \right)}_{\text{accumulated `client drift'}}\,.
\end{align*}
The local drift (i.e., deviation from the client mean) is divided by the stepsize and the expected length (i.e., $\nicefrac{1}{p}$) of the local phase during which the drift has been accumulated. 
This drift correction shares similarities with option II in \algname{Scaffold}~\citep{karimireddy2020scaffold} and \algname{QG-DSGD}~\citep{lin2021quasiglobal}, yet differs from option I in \algname{Scaffold} and \algname{FedLin}~\citep{mitra2021linear}, that both propose to compute an additional gradient at the client average.

{\bf Implementation Details}.
Similar to Algorithm~\ref{alg:ProxSkip}, the prox operator (here averaging executed on line 7) is only invoked with small probability. It is important to note that flipping the coin on line 3 can be implemented without additional communication overhead:\ the server can sample the length of each local update phase from a geometric distribution and communicate to each worker how many local updates to perform until the next communication round. We chose here this presentation to make the connection to Algorithm~\ref{alg:ProxSkip} more prominent.

\subsection{Convergence}

We will need an assumption on the individual functions $f_m$:

\begin{assumption}\label{as:f_m}
 Each $f_m$ is $L$-smooth and $\mu$-strongly convex.
\end{assumption}

Note though that we do not need to make any assumption on the similarity of the functions $f_m$. Convergence of  \algname{Scaffnew} (Algorithm~\ref{alg:fl}) in the deterministic case follows as a corollary of Theorem~\ref{thm:main}. 

\begin{corollary}[Federated learning]
Let Assumption~\ref{as:f_m} hold and let $\gamma = \nicefrac{1}{L}$, $p=\nicefrac{1}{\sqrt{\kappa}}$ and $g_{m}^{t}(x_{m}^{t}) = \nabla f_m(x_{m}^{t})$. Then the iteration complexity of Algorithm~\ref{alg:fl} is $\cO(\kappa \log \nicefrac{1}{\varepsilon}$) and its communication complexity is $\cO(\sqrt{\kappa}\log \nicefrac{1}{\varepsilon})$.
\end{corollary}

{\bf Usefulness of local steps.}
Our result shows for the first time a real advantage of local update methods \emph{without imposing any similarity assumptions}. For instance, the work in \citep{woodworth2020local} assumes quadratic functions, the works in \citep{karimireddy2020scaffold,karimireddy2020:mime} assume bounded Hessian dissimilarity, and the work in \citep{Yuan2020:accelerated} assumes bounded Hessian. Without any such assumption, we show here that local methods can converge in significantly fewer update rounds than large-batch methods without local steps~\citep{Dekel2012:minibatch}. The method matches the communication-complexity lower bound derived in \citep{Arjevani2015:communication} and is optimal in this regard. Moreover, and unlike the approaches adopted in \citep{FL-personal-mixture2020,personalized-optimal-2020}, our improvements do not rely on interpreting local methods as methods for solving {\em personalized formulations of \gls{FL}}.

\section{Extensions} \label{sec:extensions}

\subsection{Stochastic gradients}

\label{sec:stochastic}
In Machine Learning, calculating full gradients may be extremely expensive and in some cases not possible. In this section, we are going to make an extension of the basic \gls{ProxSkip} (\Cref{alg:ProxSkip}) to allow stochastic updates:
\begin{align}
\hat x^{t+1} = x^{t} - \gamma ({\color{blue} g^{t} (x^{t})} - h^{t}).
\end{align}
In a generic \algname{\gls{SGD}} method, we  work with unbiased estimators of gradients only.
\begin{assumption}[Unbiasedness]\label{as:unbias}
For all $t \geq 0$, $g^t(x^t)$ is an unbiased estimator of the gradient $\nabla f(x^{t})$. That is,
\begin{align} 
\E{g^{t}(x^{t}) \mid x^{t}} =\nabla f(x^{t}).
\end{align}
\end{assumption}

In our analysis of \gls{ProxSkip} in the stochastic case, we rely on the {\em expected smoothness} assumption introduced in~\citep{gower2019sgd} in the context of variance reduction, and later extended in \citep{gower2021stochastic} in the context of \algname{\gls{SGD}} analysis.

\begin{algorithm*}[t]
	\caption{\gls{SProxSkip} (Stochastic gradient version of \gls{ProxSkip})}
	\label{alg:stoch_rand_prox}
	\begin{algorithmic}[1]
		\STATE stepsize $\gamma > 0$, probability $p>0$, initial iterate $x_0\in \mathbb{R}^d$, initial control variate ${\red h^0} \in \mathbb{R}^d$, number of iterations $T\geq 1$
		\FOR{$t=0,1,\dotsc,T-1$}
		\STATE $\hat x^{t+1} = x^t - \gamma ({\color{blue}g^t(x^t)} - {\red h^t})$ \hfill $\diamond$ Take a {\color{blue}stochastic} gradient-type step adjusted via the control variate ${\red h^t}$
		\STATE Flip a coin $\theta_t \in \{0,1\}$ where $\mathop{\rm Prob}(\theta_t =1) = p$ \hfill $\diamond$ Flip a coin that decides whether to skip the prox or not
		\IF{$\theta_t=1$} 
		\STATE  $x^{t+1} = \prox_{\frac{\gamma}{p}\psi}\bigl(\hat x^{t+1} - \frac{\gamma}{p}{\red h^t} \bigr)$ \hfill $\diamond$ Apply prox, but only very rarely!\\  (with small probability $p$)
		\ELSE
		\STATE $x^{t+1} = \hat x^{t+1}$ \hfill $\diamond$ Skip the prox!
		\ENDIF
		\STATE ${\red h^{t+1}} = {\red h^t} + \frac{p}{\gamma}(x^{t+1} - \hat x^{t+1})$ \hfill $\diamond$ Update the control variate ${\red h^t}$
		\ENDFOR
	\end{algorithmic}
\end{algorithm*}

\begin{assumption}[Expected smoothness]\label{as:exp_smooth}
	There exist constants $A \geq 0$ and $C \geq 0$ such that for all $t \geq 0$,
	\begin{align}
	 \mathrm{E}\left[\left\|g^{t}(x^{t})-\nabla f\left(x^{\star}\right)\right\|^{2} \mid x^{t}\right] \leq 2 A D_{f}\left(x^{t}, x^{\star}\right)+C.
\end{align}
\end{assumption}
This assumption is satisfied in many practical settings, including when the randomness in $g^{t}$ arises from subsampling (i.e., minibatching) and communication compression~\citep{gorbunov2021local}. It is also satisfied in the popular but artificial setting when an additive zero mean and bounded variance noise is added to the gradient, formalized next.
\begin{assumption}[Bounded variance]\label{as:bounded}
For all $t \geq 0$, the stochastic estimator $g^t(x^t)$ has bounded variance:
	\begin{align}
\operatorname{Var}[g^t(x^t) \;|\; x^t] \leq \sigma^{2}.
	\end{align}
\end{assumption}
The next lemma, due to the work of \citep{gower2019sgd}, shows that this is indeed the case. \begin{lemma}
	\label{lemma:exp_smooth}
	Let~\Cref{as:unbias} and~\Cref{as:bounded} hold and let $f$ be convex and $L$-smooth, then expected smoothness (i.e.,~\Cref{as:exp_smooth}) holds with $A = L$ and $C = \sigma^2$.
\end{lemma}

The main result of this section is formulated next.
\begin{theorem}
	\label{thm:main-stoch}
	Let Assumptions~\ref{as:f}, \ref{as:proper_psi}, \ref{as:exp_smooth} and \ref{as:unbias} hold. Let $0<\gamma \le \nicefrac{1}{A}$ and $0<p\leq 1$. Then, the iterates of \gls{SProxSkip} (\Cref{alg:stoch_rand_prox}) satisfy
	\begin{align*}
		\squeeze\E{\Psi^{(T)}}
		\le (1 - \zeta)^T \Psi^{(0)}  + \frac{\gamma^2 C}{\zeta},
	\end{align*}
	where $\zeta\eqdef \min\{\gamma\mu, p^2\}$.
\end{theorem}
This result also gives us rates for~\algname{Scaffnew} (\Cref{alg:fl}).
\begin{corollary}\label{cor:0099887766}
 Consider  \algname{Scaffnew} (\Cref{alg:fl}) or \gls{SProxSkip} (\Cref{alg:stoch_rand_prox}). Set any $0<\varepsilon <1$. If we choose $\gamma = \min\left\{  \frac{1}{A}, \frac{ \varepsilon \mu}{2 C} \right\}$ and $p=\sqrt{\gamma \mu}$,  then in order to guarantee $\E{ \Psi^{(0)} } \leq \varepsilon   $, it suffices to take
	$T \geq \max \left\{ \frac{A}{\mu}, \frac{2C}{\varepsilon \mu^2}\right\}\log \left(\frac{2 \Psi^{(0)}}{\varepsilon}\right)$ iterations, which results in $\max \left\{ \sqrt{\frac{A}{\mu}}, \sqrt{\frac{2C}{\varepsilon \mu^2}}\right\}\log \left(\frac{2 \Psi^{(0)}}{\varepsilon}\right)$ communications (in case of \algname{Scaffnew}) resp.\ prox evaluations (in case of  \gls{SProxSkip}) on average.
\end{corollary}

{\bf Limitations}. The main limitation of applying analysis of \gls{SProxSkip} in the \gls{FL} setting (\Cref{alg:fl}) is that we do not achieve linear speedup in terms of the number of clients. This issue comes from the analysis technique and it needs deeper investigation. The same problem appears in the analysis of \algname{FedLin}, but it does not in the analysis of \algname{Scaffold}.

\subsection{Decentralized training}\label{sec:decentralized}

Let us now discuss the minimization problem with decentralized communication. Given a graph $G=(V, E)$ with $M$ nodes $V$ and edges $E$, we assume that every communication node $m^\prime$ receives a weighted average of its neighbors' vectors with weights $W_{m^\prime,1},\dotsc, W_{m^\prime, M}\in [0, 1]$. Besides, nodes $i$ and $j$ communicate if and only if $W_{m^\prime,m}\neq 0$, which is also equivalent to $(m^\prime,m)\in E$. The weights $W_{m^\prime,m}$ define the \emph{mixing matrix} $\mW$ that we assume to be symmetric, doubly stochastic, and positive semi-definite. Then, the problem is equivalent to
\vspace{-1mm}
\[
\squeeze	
\min_{x\in\mathbb{R}^{d\cdot M}} f(x)\quad \text{subject to}\quad (\mI-\mW)x=0,
\]
where $\mI$ is the identity matrix. Let us set $\mL$ to be the square-root of $\mI-\mW$ and define the indicator function $\psi(y)$ by setting $\psi(0)=0$ and $\psi(y)=+\infty$ for any $y\neq 0$, which is similar to our previous definition in equation~\eqref{eq:constraint_consensus_0980_proxskip}. Then, the constraint $(\mI-\mW)x=0$ is equivalent to $\mL x=0$, so the problem can be rewritten as
\begin{equation}
\squeeze	\min_{x\in\mathbb{R}^{d'}} f(x) + \psi(\mL x), \label{eq:problem_with_matrix}
\end{equation}
where $d'=d\cdot M$. The reason we define $\mL$ this way is that splitting algorithms require computation of $\mL\mL^\top u$ for some vector $u$, which in our case is exactly $(\mI-\mW)u$. Algorithmically, computing this product corresponds to communicating over the graph. For convenience, we provide the algorithm formulation in the graph notation in \Cref{alg:dist_gd}. The convergence of our decentralized algorithm is stated below.

\begin{algorithm*}[t]
	\caption{\algname{Decentralized Scaffnew}}
	\label{alg:dist_gd}
	\begin{algorithmic}[1]
		\STATE stepsizes $\gamma > 0$ and $\tau>0$, initial iterates $x_{1}^{0} = \ldots = x_{M}^{0} = x^0 \in \mathbb{R}^d$, initial control variables ${\red h_{1}^{0}} = \ldots = {\red h_{M}^{0}} = 0\in\mathbb{R}^d$, weights for averaging $\mathbf{W}=(W_{m^\prime,m})_{m^\prime,m=1}^M$
		\FOR{$t=0,1,\dotsc,T-1$}
		\STATE Flip a coin $\theta_t \in \{0,1\}$ where $\mathop{\rm Prob}(\theta_t =1) = p$ \hfill $\diamond$ Flip a coin that decides whether to skip the prox or not
		\FOR{$m=1,\dotsc, M$}
		\STATE $\hat x_{m}^{t+1} = x_{m}^{t} - \gamma (\nabla f_m (x_{m}^{t}) - {\red h_{m}^{t}})$\hfill $\diamond$ Take a gradient-type step adjusted via the control variate ${\red h_{m}^{t}}$
		\IF{$\theta_t=1$}
		\STATE $x_{m}^{t+1} = \left(1 - \frac{\gamma\tau}{p}\right)\hat x_{m}^{t+1} + \frac{\gamma\tau}{p}\sum_{m^\prime=1}^M W_{m,m^\prime}\hat x_{m^\prime}^{t+1}$ \hfill $\diamond$  Communicate, but only very rarely! (with small prob.\ $p$)
		\STATE ${\red h_{m}^{t+1}} = {\red h_{m}^{t}} + \frac{p}{\gamma}(x_{m}^{t+1} -  \hat x_{m}^{t+1})$  \hfill $\diamond$ Update the control variate ${\red h_{m}^{t}}$
		\ELSE
		\STATE $x_{m}^{t+1} = \hat x_{m}^{t+1}$ \hfill$\diamond$ Skip communication!
		\STATE ${\red h_{m}^{t+1}} =  {\red h_{m}^{t}}$
		\ENDIF
		\ENDFOR
		\ENDFOR
	\end{algorithmic}
\end{algorithm*}

\begin{theorem}\label{th:decentralized}
	Let $f$ satisfy Assumption \ref{as:f_m} and define the spectral gap of $\mW$ as $\delta=1-\lambda_2(\mW)\in(0, 1)$, where $\lambda_2(\mW)$ is the second largest eigenvalue of $\mW$. If we set $p\in(0, 1]$, $\gamma\le \nicefrac{1}{L}$, $\tau\le \nicefrac{p}{\gamma}$, then the average iterate $\overline x^T$ satisfies
	\[
		\E{\| \overline x^T-x^\star \|^2 } \le (1-\min(\gamma\mu, p\gamma\tau\delta))^T \Phi^{(0)} ,
	\]
	where $\Phi^{(0)} \le \|x^0-x^\star\|^2 + \frac{\gamma}{p\tau\delta n}\sum_{m=1}^M\|\nabla f_m(x^\star)\|^2$.
\end{theorem}
If we plug-in $\tau=\nicefrac{p}{\gamma}$, the theorem implies that the new rate is $\tilde{\cO}(\kappa + \frac{1}{p^2\delta})$. Thus, it is optimal to choose $p=\sqrt{\nicefrac{1}{(\delta\kappa)}}$ whenever the network is sufficiently well-connected. If passing a message is challenging, which happens when $\delta\le \nicefrac{1}{\kappa}$, then it is optimal to communicate every iteration by setting $p=1$. This trade-off is to be expected as our algorithm for \eqref{eq:problem_with_matrix} matches the lower bound in \citep{OPAPC} in terms of number of matrix-vector multiplications. 

\section[Experiments]{Experiments}

\begin{figure*}[t]
	\centering
	\begin{tabular}{ccc}
		$\!\!$\includegraphics[scale=0.18]{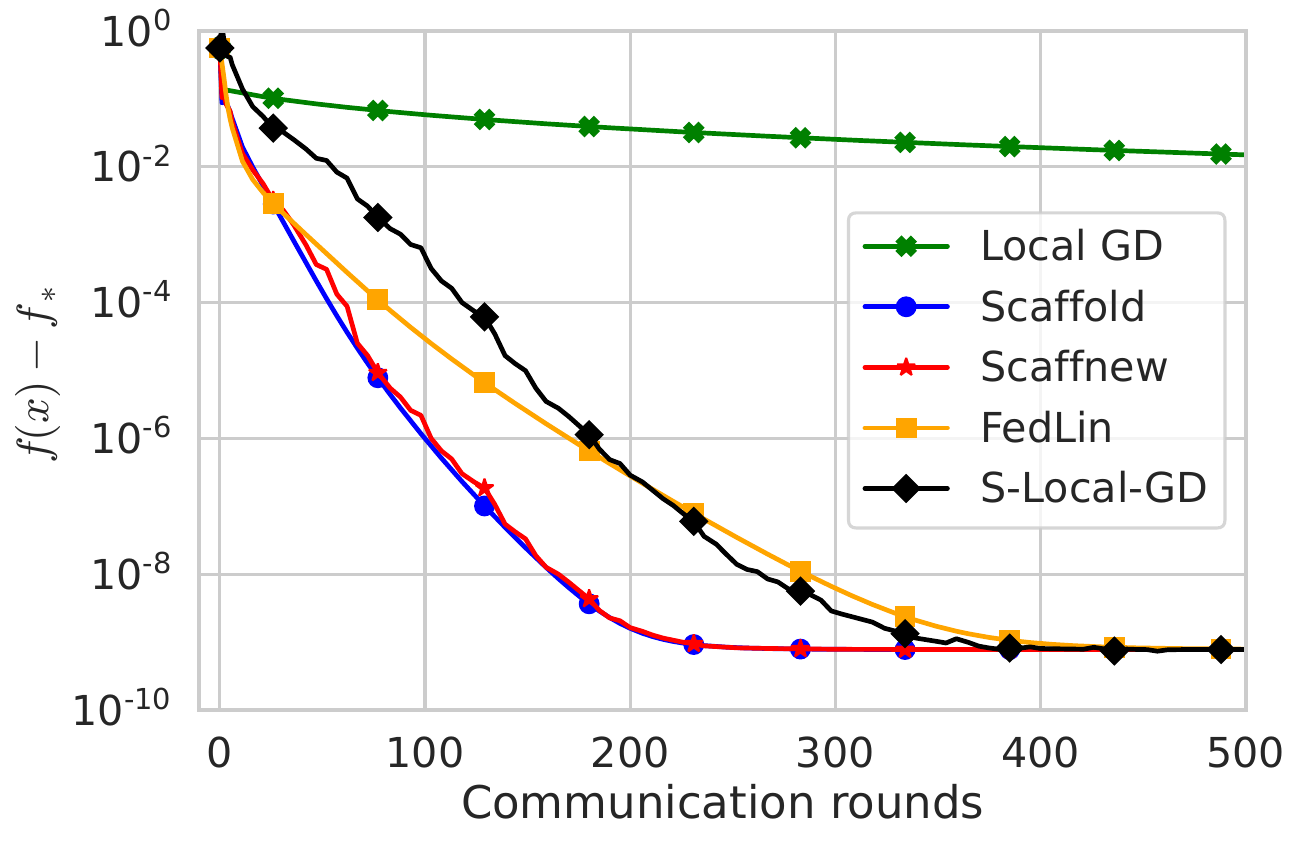}&

		$\!\!$\includegraphics[scale=0.18]{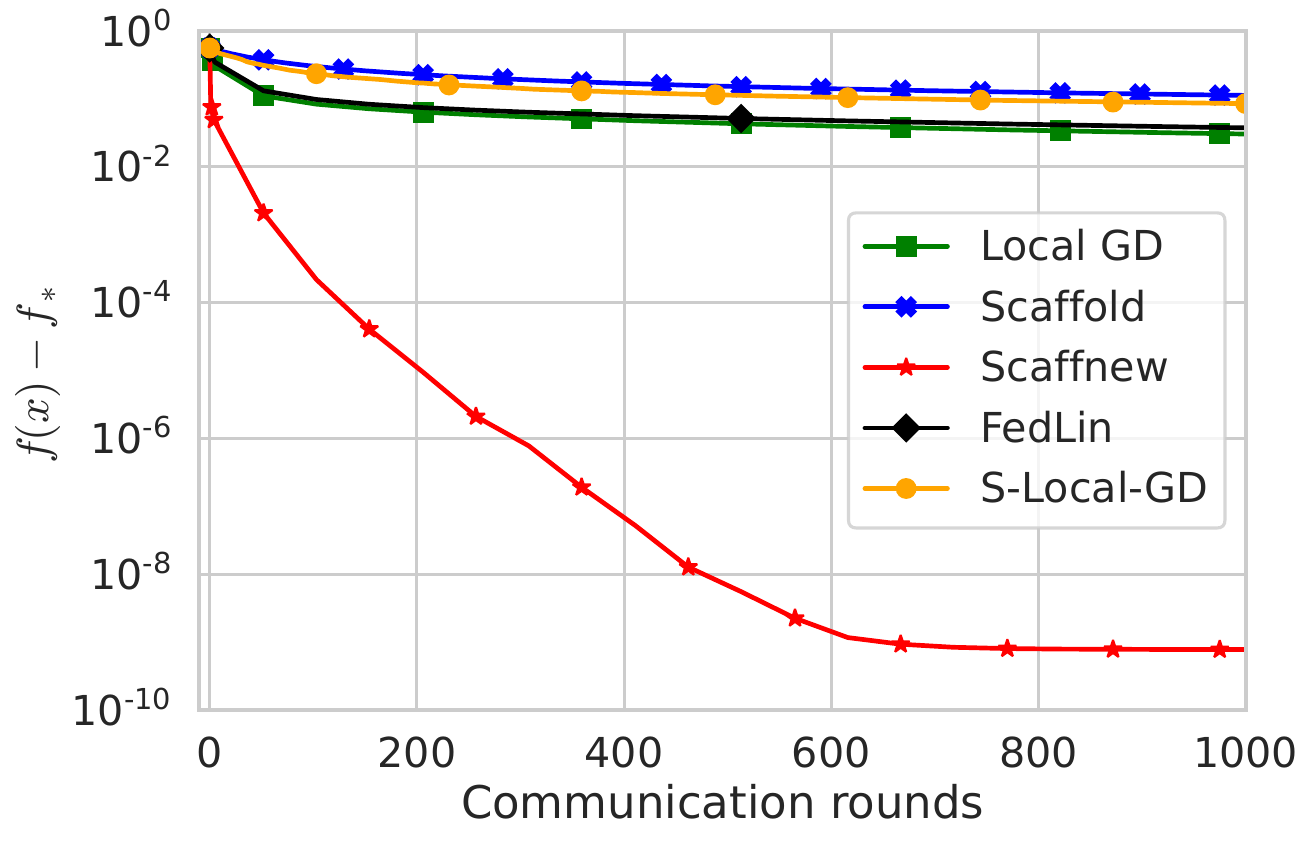}&
						$\!\!$\includegraphics[scale=0.18]{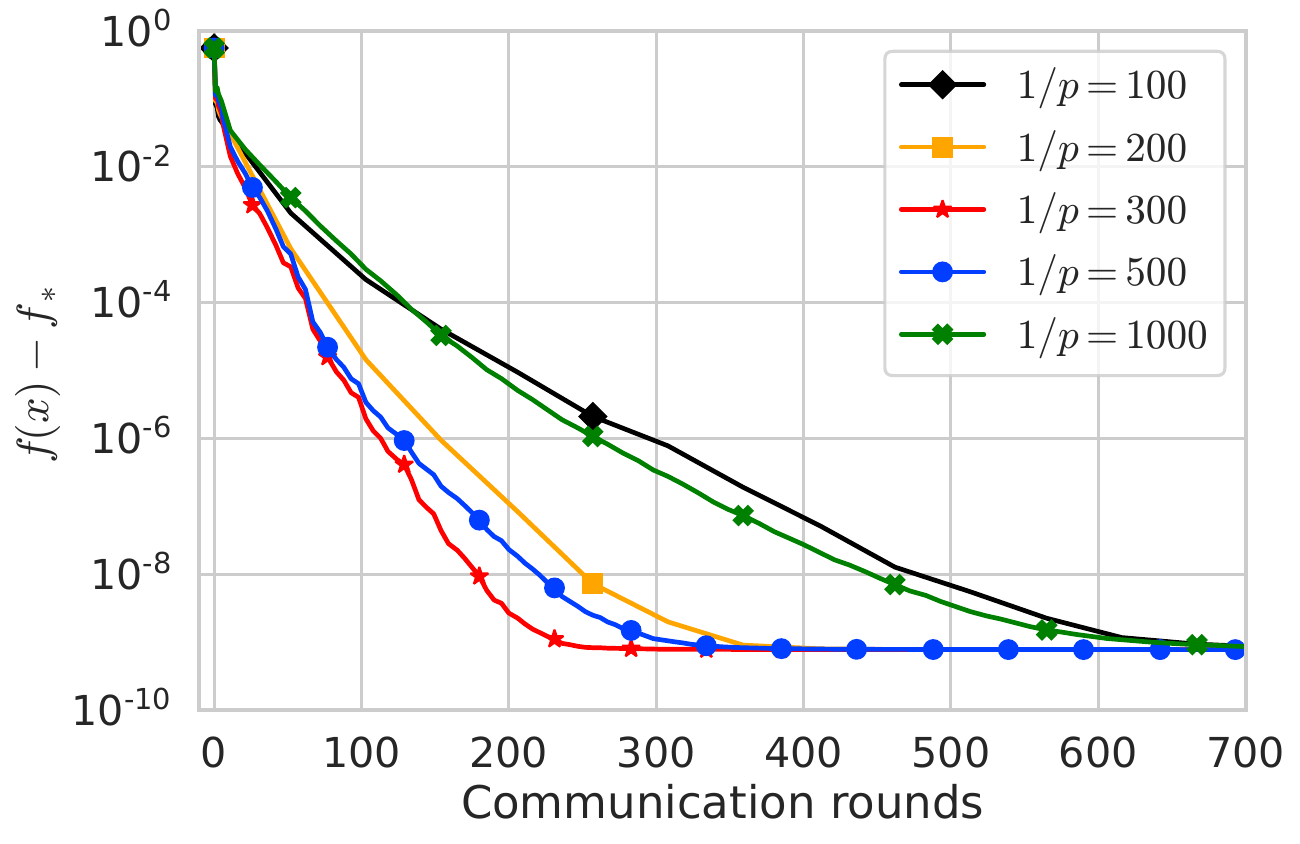}\\
		\footnotesize (a) tuned hyper-parameters & 
		\footnotesize (b) theoretical hyper-parameters&
		\footnotesize (c) different options of $p$
	\end{tabular}

	\caption{\textbf{Deterministic Case}. Comparison of \algname{Scaffnew} to other local update methods that tackle data-heterogeneity and to \algname{LocalGD}. In (a) we compare communication rounds with optimally tuned hyper-parameters. 
	In (b), we compare communication rounds with the algorithm parameters set to the best theoretical stepsizes used in the convergence proofs.
	In (c), we compare communication rounds with the algorithm stepsize set to the best theoretical stepsize and different options of parameter $p$.
	}
	\label{fig:image1}
\end{figure*}
\begin{figure*}[t]
	\centering
	\begin{tabular}{ccc}
		$\!\!$\includegraphics[scale=0.18]{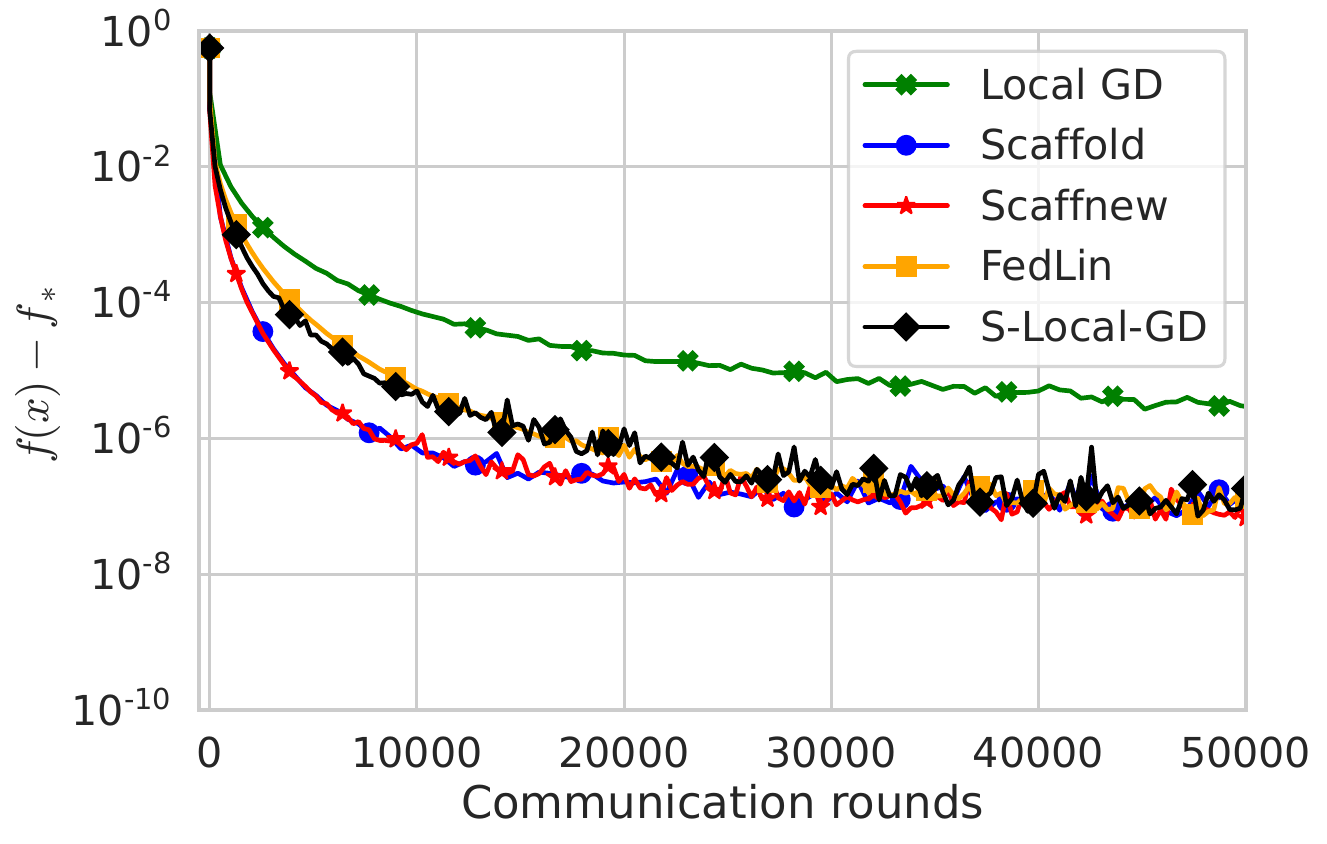}&
		$\!\!$\includegraphics[scale=0.18]{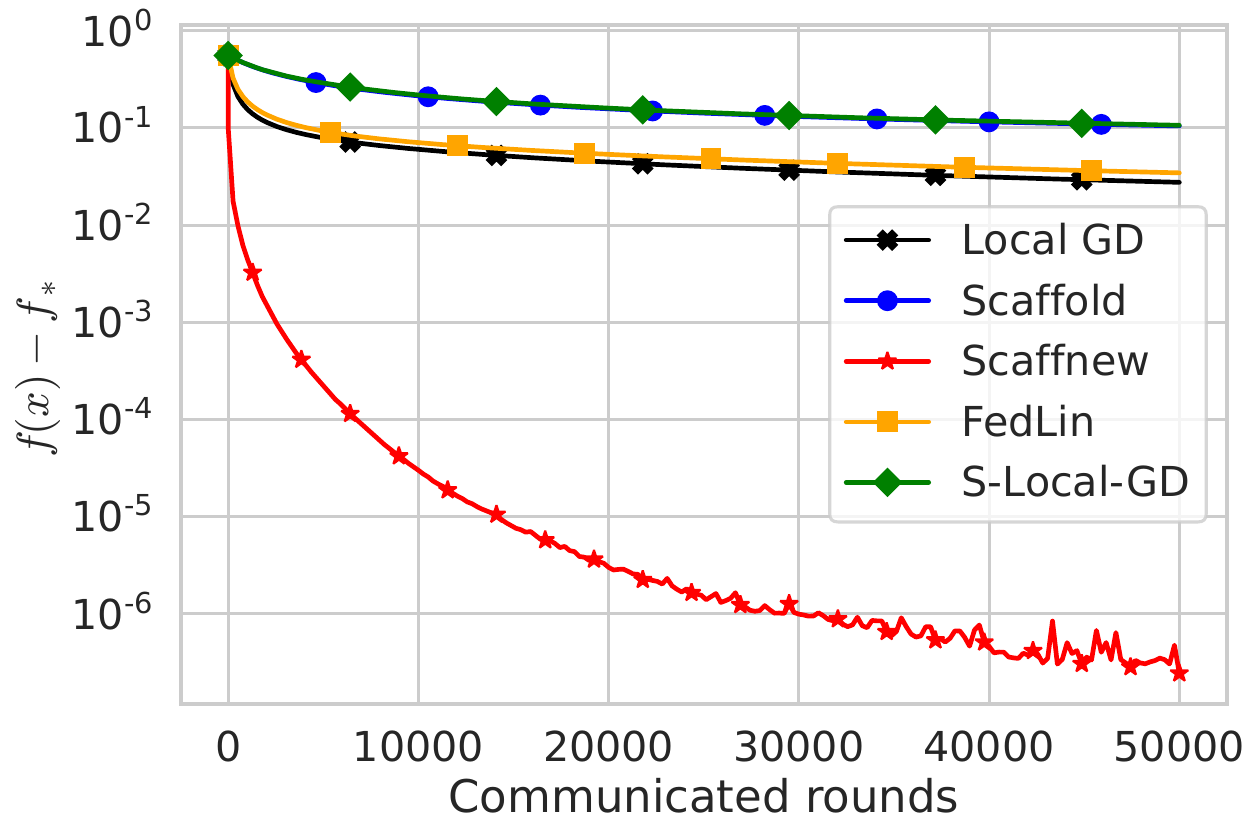}&
		$\!\!$\includegraphics[scale=0.18]{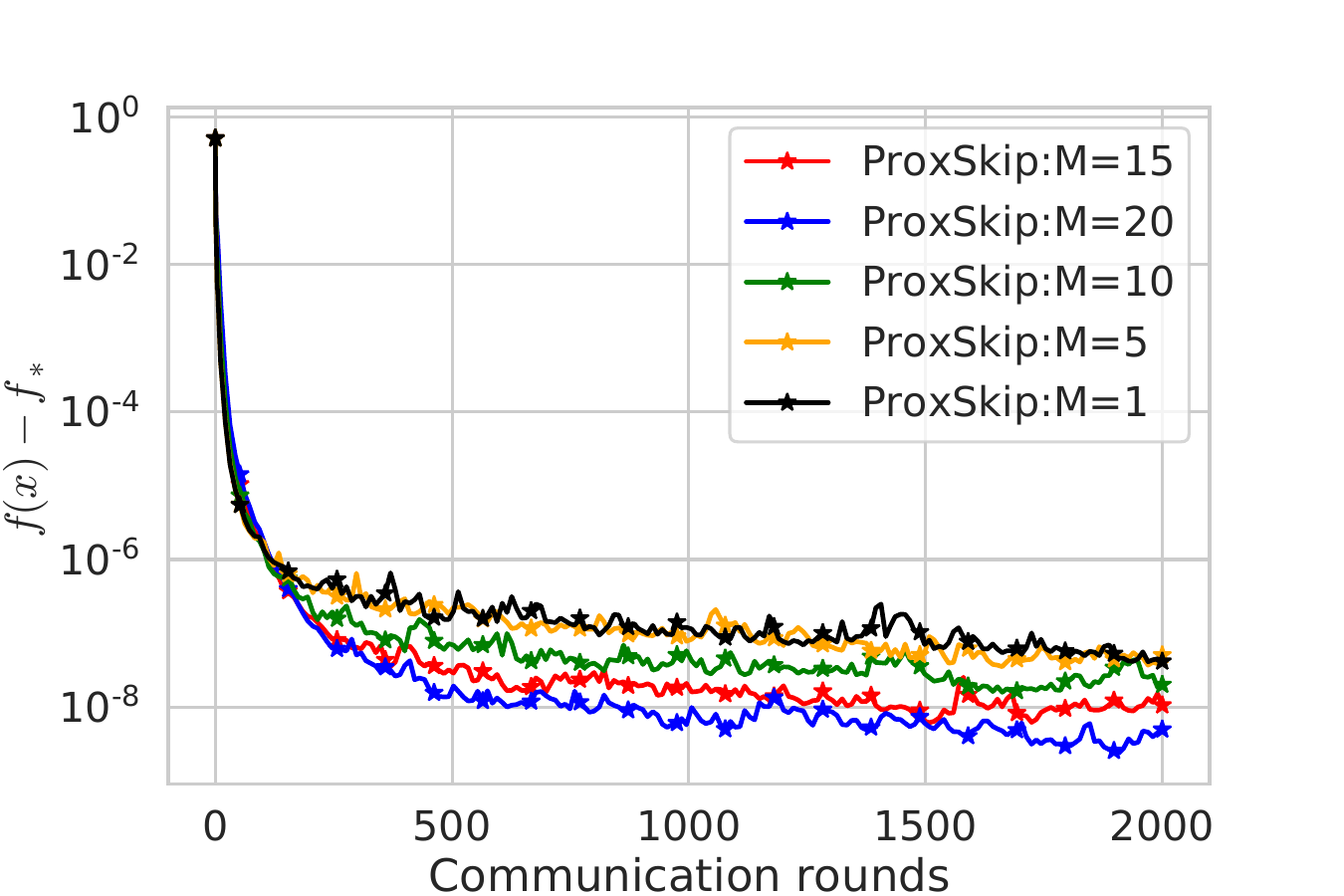}\\
		\footnotesize (a) tuned hyper-parameters & 
		\footnotesize (b) theoretical hyper-parameters&
			\footnotesize (c) different $\#$ of clients
	\end{tabular}
	
	\caption{\textbf{Stochastic Case}. Comparison of \algname{Scaffnew} to other local update methods that tackle data-heterogeneity  and to \algname{LSGD}. In (a) we compare communication rounds with optimally tuned hyper-parameters. 
	In (b), we compare communication rounds with the algorithm parameters set to the best theoretical stepsizes used in the convergence proofs.
	In (c), we compare communication rounds with the algorithm parameters set to the best theoretical stepsizes used in the convergence proofs and a different number of clients.
	}
	\label{fig:image2}
\end{figure*}

To test the performance of algorithms and illustrate theoretical results, we use the classical logistic regression problem. The loss function for this model has the following form:
\begin{align*}
	\squeeze 	f(x)=\frac{1}{N} \sum_{i=1}^{N} \log \left(1+\exp \left(-b_{i} a_{i}^{\top} x\right)\right)+\frac{\lambda}{2}\|x\|^{2},
\end{align*}
where $a_{i} \in \mathbb{R}^{d} \text { and } b_{i} \in\{-1,+1\}$ are the data samples and $N$ is their total number. We set the regularizer $\lambda = 10^{-4}L$, where $L$ is the smoothness constant.

 We implemented all algorithms in Python using the package RAY~\citep{moritz2018ray} to utilize parallelization. All methods were evaluated on a workstation with an Intel(R) Xeon(R) Gold 6146 CPU at 3.20GHz with 24 cores. We use the `w8a' dataset from LIBSVM library~\citep{chang2011libsvm}.
 
  In our experiments, we have two settings: deterministic (\Cref{fig:image1}) and stochastic problems (\Cref{fig:image2}). First, we provide results with tuned hyper-parameters (subplot (a)). \algname{Local GD} converges to the neighborhood of the solution due to data-heterogeneity. \algname{Scaffold} and \algname{Scaffnew} have the same convergence rate in terms of communication rounds and this rate is better than others. 
Second, we test algorithms with theoretical hyper-parameters (subplot (b)). In this setting, \algname{Scaffnew} outperforms other methods since our theory guarantees that we can use large stepsizes. The number of local steps is set to be $\sqrt{\hat{\kappa}}$, where $\hat{\kappa} = \frac{L}{\lambda}$.

 As we can see, if our method communicates either too often ($\nicefrac{1}{p} = 100$) or too rarely ($\nicefrac{1}{p} = 1000$), convergence suffers. The optimal number of local steps in this experiment is $\nicefrac{1}{p} = 300$. Our theory predicted that the choice $p=\frac{1}{\sqrt{\kappa}}$ is close to the experiment's results. Moreover, we compared Scaffnew in the stochastic case with a different number of clients $M$. As we can see, we can obtain the linear speedup, which is more optimistic than we have in theory.


\chapter{ProxSkip-VR: Balancing Resource Costs and Reducing Variance in Federated Optimization}
\label{chapter3}
\thispagestyle{empty}

\section{Introduction}

Announced in April 2017 in a Google AI blog \citep{FLblog2017}, and citing four foundational papers \citep{FedLearn2016, FedOpt2016, mcmahan2017communication, FL-secure_aggreg} of what was to become a new and rapidly growing interdisciplinary field, {\em Federated Learning} (\gls{FL}) constitutes a novel paradigm for training supervised Machine Learning models. The key idea  is the acknowledgement that increasing amounts of data are being captured and stored on edge devices, such as mobile phones, sensors and hospital workstations, and that moving the data to a  datacenter for centralized processing may be infeasible or undesirable for various reasons, including high energy costs and  data privacy concerns~\citep{kairouz2019advances,FL_survey_2020}. \gls{FL} faces a multitude of challenges which are being actively addressed by the research community.

\subsection{Formalism}
We study the standard optimization formulation of Federated Learning \citep{FedOpt2016,mcmahan2017communication,kairouz2019advances,FieldGuide2021} given by
\begin{equation}\label{eq:P}
 \min\limits_{x\in \mathbb{R}^{d}} f(x), \qquad  f(x)\eqdef  \sum\limits_{m=1}^M \frac{n_m}{N} f_m(x),  \qquad f_m(x) \eqdef 
\frac{1}{n_m} \sum\limits_{i=1}^{n_m} f_{m,i}(x),
\end{equation}
where $M$ is the number of clients (devices, machines, workers), $n_m$ is the number of training data points on client $m\in \{1,2,\dots, M\}$, and  $N\eqdef \sum_{m=1}^M n_m$ is the total number of training data points collectively owned by this federation of $M$ clients. Note that $f$ is the empirical risk over the federated dataset. The purpose of the weights is to make sure that $f$ is  the average loss over all $N$ training points owned by this federation of $M$ clients. Indeed, notice that
\begin{equation}
f(x) = \frac{1}{N} \sum_{m=1}^M \sum_{i=1}^{n_m} f_{m,i}(x).
\end{equation}
In other words, solving (\ref{eq:P}) amounts to finding the empirical risk minimizer over the federated dataset. 
Perhaps conceptually the simplest method for solving (\ref{eq:P}) is {\em gradient descent (\algname{\gls{GD}})},
\begin{equation} \label{eq:GD}
x^{t+1} = x^{t} - \gamma \nabla f(x^{t}) = x^{t} - \gamma \sum\limits_{m=1}^M \frac{ n_m}{N} \nabla f_m(x^{t})  = \sum\limits_{m=1}^M \frac{ n_m}{N}  \left(x^{t} - \gamma  \nabla f_{m}(x^{t}) \right),
\end{equation}
where $\gamma>0$ is the stepsize. It will be useful to describe how \algname{\gls{GD}}  would be implemented in a federated environment. First, all clients $m\in \{1,\dots, M\}$ in parallel perform a single local gradient step starting from the current global model $x^{t}$, arriving at the local models $x_{m}^{t+1} \eqdef x^{t} - \gamma  \nabla f_{m}(x^{t})$, $m\in \{1,\dots,M\}$. These local models are then communicated to the {\em orchestrating server}, which aggregates them via weighted averaging, arriving at the new global model $x^{t+1} =\sum_{m=1}^M \frac{n_m}{N} x_{m}^{t+1}$. This new model is then broadcast back to all clients, and the process is repeated until a model of sufficient quality is found.

\subsection{Federated averaging}\label{sec:FedAvg}

Proposed in the works \citep{Povey2015,SparkNet2016,mcmahan2017communication}, federated averaging (\algname{FedAvg}) is arguably the most  popular method for solving the standard \gls{FL} formulation (\ref{eq:P}). Motivated by the specific constraints of federated environments, \algname{FedAvg} can be seen as a practical enhancement of \algname{GD} via the simultaneous application of three techniques: a) \gls{DS}, b) Partial Participation (\gls{PP}), and c) \gls{LT}. That is, $$\text{\algname{FedAvg} = \algname{\gls{GD}}  + (\gls{DS} + \gls{PP} + \gls{LT})}.$$ We will now briefly describe each of these three \algname{\gls{GD}}-enhancing techniques separately. 


\begin{itemize} 
\item [(a)] {\bf \algname{\gls{GD}}  + Data Sampling.} In situations when the local datasets are so large that the computation of the exact local gradients  becomes a bottleneck, it makes sense to approximate them via data sampling. That is, instead of passing through all local data  to compute the local gradient $\nabla f_m(x^{t})$, each client $i$ computes the gradients $\nabla f_{m,i}(x^{t})$ for $i\in \cD_{m}$ only, where $\cD_{m}$ is a suitably chosen  small-enough subset of the local dataset $\{1,\dots, n_m\}$. These gradients are then used to form gradient estimators $g_{m}(x^{t})\approx \nabla f_m(x^{t})$ which are used to perform  a local \algname{\gls{SGD}} step on all clients. The rest of the procedure is the same as in the case of \algname{\gls{GD}}. That is, the local models obtained in this way are sent to the orchestrating server, the server aggregates them via weighted averaging and broadcasts the resulting model back to all clients. A combination of \algname{\gls{GD}} and \gls{DS} can be interpreted as a particular variant of \algname{\gls{SGD}}, where the stochastic gradient estimator is constructed from gradients associated with individual data points across all local datasets. While \gls{DS} is still an active area of research, it has been studied for a long time, and is in general well  understood \citep{pegasos2,Li2014,csiba2016importance,gower2019sgd,nonconvex_arbitrary,khaled2023better}.

\item [(b)] {\bf \algname{GD} + Partial Participation.} 
 In practical federated environments, and especially in cross-device \gls{FL} \citep{kairouz2019advances}, the number of clients  is enormous, they are not all available at all times, and the orchestrating server has limited compute and memory capacity. For these and other reasons, practical \gls{FL} methods need to work in an environment in which a small subset $\cS^t\subseteq \{1,\dots,M\}$ of the clients is selected (``participates'') in each communication/aggregation/training round only. Since only the participating clients $i\in \cS^t$ perform a local \algname{\gls{GD}} step and  communicate the resulting local model to the orchestrating server for aggregation, this induces an error compared to \algname{\gls{GD}}, which has an adverse effect on the convergence rate. A combination of \algname{GD} and \gls{PP} can be interpreted as a particular variant of \algname{\gls{SGD}}, where the stochastic gradient estimator is constructed from gradients associated with data points sampled from the selected set of clients at each iteration. While \gls{PP} is still an active area of research, since \gls{PP} is a special type of \gls{DS}, much was known about \gls{PP} long before \algname{FedAvg} was proposed~\citep{gower2019sgd,nonconvex_arbitrary}.  Still, \gls{PP} poses new challenges tackled by the community \citep{Eichner2019semi-cyclicSGD_for_FL,OptClientSampling2020,gower2019sgd,ClientSelection-Gauri,Cohort2021}.

\item [(c)] {\bf \algname{GD} + Local Training.} 
In Federated Learning, the cost of communication between the clients and the orchestrating server forms the key bottleneck. Indeed, in their \algname{FedAvg} paper,  which introduced \gls{LT} to the world of Federated Learning, McMahan et al. \citep{mcmahan2017communication} wrote: \begin{quote}{\em \footnotesize ``In contrast\footnote{to datacenter optimization}, in federated optimization communication costs dominate''.}\end{quote} \gls{LT} is a conceptually simple and surprisingly powerful communication acceleration technique. The basic idea behind \gls{LT} is for the clients to perform {\em multiple} local \algname{\gls{GD}} steps instead of a single step (which is how \algname{\gls{GD}} operates) before communication and aggregation take place. The intuitive reasoning used in virtually all papers on this topic is: performing  multiple local \algname{\gls{GD}} steps results in ``richer'' and ultimately more useful local training in  the sense that fewer communication rounds will {\em hopefully} suffice to finish the training. The work \citep{mcmahan2017communication} supported this intuition with ample empirical evidence, and credited \gls{LT} as the critical component behind the success of \algname{FedAvg}: \begin{quote}{\em\footnotesize ``Thus, our goal is to use additional computation in order to decrease the number of rounds of communication needed to train a model\dots'' ``Communication costs are the principal constraint, and we show a reduction in required communication rounds by 10--100$\times$ as compared to synchronized stochastic gradient descent.'' ``\dots the speedups we achieve are due primarily to adding more computation on each client''. }\end{quote}


 \end{itemize}

 \section{Five Generations of Local Training Methods}

\begin{table*}[t]
    \centering
    \scriptsize
    \caption{ Five generations of local training (LT) methods summarizing the progress made by the ML/FL community over the span of 7+ years in the understanding of the {\em communication acceleration properties of LT}. }
    \label{tab:comparison}
    \begin{threeparttable}
\begin{tabular}{lclll}
{\bf Generation}\tnote{\color{blue}(a)}  & \bf Theory & \bf Assumptions & {\bf Comm.\ Complexity}\tnote{\color{blue}(b)} & \bf  Key References \\
\hline
\multirow{3}{*}{1. Heuristic} 
& \xmark & --- & empirical results only & \algname{LSGD} [\citenum{Povey2015}]   \\ 
& \xmark & --- & empirical results only & \algname{SparkNet} [\citenum{SparkNet2016}]  \\ 
& \xmark & --- & empirical results only & \algname{FedAvg} [\citenum{mcmahan2017communication}] \\
\hline
\multirow{2}{*}{2. Homogeneous} 
& \cmark & bounded gradients & sublinear &   \algname{FedAvg} [\citenum{FedAvg-nonIID}]\\
& \cmark & bounded grad.\ diversity\tnote{\color{blue}(c)}  & linear \& worse than \algname{GD} & \algname{LFGD} [\citenum{LocalDescent2019}]  \\
\hline
\multirow{2}{*}{3. Sublinear} 
& \cmark & standard\tnote{\color{blue}(d)}  & sublinear &  \algname{LGD} [\citenum{khaled2019first}] \\
& \cmark & standard                                    & sublinear &  \algname{LSGD} [\citenum{localSGD-AISTATS2020}]  \\
\hline
\multirow{3}{*}{4. Linear} 
& \cmark & standard & linear \& worse than \algname{GD}  & \algname{Scaffold} [\citenum{karimireddy2020scaffold}] \\ 
& \cmark & standard & linear \& worse than \algname{GD}  & \algname{S-Local-GD} [\citenum{gorbunov2021local}] \\ 
& \cmark & standard & linear \& worse than \algname{GD}  & \algname{FedLin} [\citenum{mitra2021linear}] \\
\hline
\multirow{2}{*}{5. Accelerated} 
& \cmark & standard & linear \& better than \algname{GD} &  \gls{ProxSkip} [\citenum{ProxSkip}] \\ 
&\cellcolor{bgcolor2}\cmark & \cellcolor{bgcolor2}standard & \cellcolor{bgcolor2}linear \& \cellcolor{bgcolor2}better than \algname{GD} &  \cellcolor{bgcolor2}\gls{ProxSkip-VR}  \\
\hline
\end{tabular}
  \begin{tablenotes}
        {\tiny
        \item [{\color{blue}(a)}]  Since Partial Participation (\gls{PP}) and data sampling (\gls{DS}) can only {\em worsen}  theoretical communication complexity, our historical breakdown of the literature into 5 generations of LT methods focuses on the full client participation (i.e., no \gls{PP}) and exact local gradient (i.e., no \gls{DS}) setting. While some of the referenced methods incorporate \gls{PP} and \gls{DS} techniques, these are irrelevant for our purposes. Indeed, from the viewpoint of communication complexity, all these algorithms enjoy best theoretical performance  in the no-\gls{PP} and no-\gls{DS} regime.        
        \item [{\color{blue}(b)}] For the purposes of this table, we consider problem (\ref{eq:P}) in the {\em smooth} and {\em strongly convex} regime only. This is because the literature on LT methods struggles to understand LT even in this simplest (from the point of view of optimization) regime.  
        \item [{\color{blue}(c)}] {\em Bounded gradient diversity} is a uniform bound on a specific notion of gradient variance depending on Partial Participation probabilities. However, this assumption (as all homogeneity assumptions) is very restrictive. For example, it is not satisfied for the standard class of smooth and strongly convex functions. 
        \item [{\color{blue}(d)}]  The notorious \gls{FL} challenge of handling non-i.i.d.\ data by LT methods was solved  in \citep{khaled2019first} (from the viewpoint of {\em optimization}). From generation 3 onwards, there was no need to invoke any data/gradient homogeneity assumptions.  Handling non-i.i.d.\ data remains a challenge from the point of view of {\em generalization}, typically by considering {\em personalized} \gls{FL} models.         
        }
    \end{tablenotes}
    \end{threeparttable}
\end{table*}

We now offer several historical comments on the most important developments related to the {\em theoretical} understanding of \gls{LT}.
To this end,  we have identified 5 distinct generations of \gls{LT} methods, each with its unique challenges and characteristics. To make the narrative simple, and since we focus on this regime in our paper, we limit our overview to loss functions $f_m$ that are $\mu$-strongly convex and $L$-smooth. This is arguably the most  studied class of functions in continuous optimization \citep{NesterovBook}, and for this reason, it presents a valuable litmus test for any theory of \gls{LT}.


\subsection{Generation 1: Heuristic age} 
While \gls{LT} ideas were used in several Machine Learning domains before \citep{Povey2015,SparkNet2016}, LT truly rose to prominence as a practically potent communication acceleration technique due to the seminal paper \citep{mcmahan2017communication}  which introduced the \algname{FedAvg} algorithm. However, no theory was provided in their work, nor in any prior work. LT-based heuristics, i.e., methods without any theoretical guarantees, dominated the initial development of the field up to, and including, the \algname{FedAvg} paper. 

\subsection{Generation 2: Homogeneous age} 
The first theoretical results for \gls{LT} methods offering explicit convergence rates relied on various data/gradient {\em homogeneity}\footnote{We use the term {\em homogeneity} to refer to various related assumptions used in the literature, including  {\em bounded gradient norms} \citep{FedAvg-nonIID}, {\em bounded gradient variance} \citep{Li2019-local-homogeneous,Yu-local-homogeneous-2019} and {\em bounded gradient diversity} \citep{LocalDescent2019}.} 
assumptions.
The intuitive rationale behind such assumptions comes from the following thought process. In the extreme case when all the local functions $f_m$ are {\em identical} (this is often referred to as the {\em homogeneous} or {\em i.i.d.\ data} regime), there is a very simple approach to making \algname{\gls{GD}}  communication-efficient: push the idea of \gls{LT} to its extreme by running \algname{\gls{GD}} on all clients, independently and in parallel, without any communication/synchronization/averaging whatsoever.
Extrapolating from this, it is reasonable to assume that as we increase heterogeneity, taking multiple local steps should still be beneficial as long as we do not take too many steps. Several authors analyzed various \gls{LT} methods under such assumptions, and obtained rates \citep{LocalDescent2019,Yu-local-homogeneous-2019,Li2019-local-homogeneous,FedAvg-nonIID}.  However, bounded dissimilarity assumptions are highly problematic. First, they do not seem to be satisfied even for some of the simplest function classes, such as strongly convex quadratics \citep{khaled2019first,localSGD-AISTATS2020}, and moreover, it is well known that practical \gls{FL} datasets are highly heterogeneous/non-i.i.d. \citep{mcmahan2017communication,kairouz2019advances}. So, analyses relying on such strong assumptions are both mathematically questionable, and practically irrelevant.

\subsection{Generation 3: Sublinear age}
The third generation of \gls{LT} methods is characterized by the successful removal of the bounded dissimilarity assumptions from the convergence theory. The work \citep{khaled2019first} is the first that achieved this breakthrough by studying the simplest \gls{LT} method: \gls{LGD} (i.e., a simple combination of \algname{\gls{GD}}  and \gls{LT}). While works belonging to this generation elevated LT to the same theoretical footing as \algname{\gls{GD}}  in terms of the assumptions, which marked an important milestone in our understanding of \gls{LT},  unfortunately, the obtained communication complexity theory of \algname{LGD}  is pessimistic when compared to vanilla \algname{\gls{GD}}. Indeed, the inclusion of LT did {\em not} lead to an improvement upon the communication complexity of vanilla \algname{\gls{GD}}. Moreover, while \algname{\gls{GD}} enjoys a linear communication complexity (in the smooth and strongly convex regime), the communication complexity of \algname{\gls{LGD}} is {\em sublinear}. In a follow-up work \citep{localSGD-AISTATS2020} \algname{LGD} was later analyzed in combination with \gls{DS} as well. Works \citep{woodworth2020minibatch} and \citep{glasgow2022sharp} provided lower bounds for \algname{\gls{LGD}} with \gls{DS} showing that it is not better than minibatch \algname{SGD} in the heterogeneous setting. See the work  \citep{malinovskiy2020local} for a fixed-point theory viewpoint.

\subsection{Generation 4: Linear age} 
The fourth generation of \gls{LT} methods is characterized by the effort to design {\em linearly} converging variants of \gls{LT} algorithms. In order to achieve this, it was important to tame the adverse effect of the so-called {\em client drift} \citep{karimireddy2020scaffold}, which was identified as the culprit of the worse-than-\algname{GD} theoretical performance of the previous generation of \gls{LT} methods. The first \gls{LT}-based method that successfully tamed  client drift, and as a result obtained a linear convergence rate, was \algname{Scaffold} \citep{karimireddy2020scaffold}. Several alternative approaches to obtaining the same effect were later proposed in \citep{gorbunov2021local} and  \citep{mitra2021linear}. While obtaining a linear rate for \gls{LT} methods under standard assumptions  was a major achievement, the communication complexity of these methods is still somewhat worse\footnote{Both \algname{\gls{GD}}, and \gls{LT} methods such as \algname{Scaffold} \citep{karimireddy2020scaffold},  \algname{S-Local-GD} \citep{gorbunov2021local} and  \algname{FedLin} \citep{mitra2021linear} enjoy the linear rate $\cO(\kappa \log \frac{1}{\varepsilon})$, where $\kappa$ is a condition number. However, this condition number is in general slightly worse for the LT methods.} than that of vanilla \algname{\gls{GD}}, and is at best equal to that of \algname{GD}.

\subsection{Generation 5: Accelerated age} Finally, the fifth generation of \gls{LT} methods was initiated recently in \citep{ProxSkip} with their \gls{ProxSkip} method which enjoys {\em accelerated  communication complexity}. Acceleration comes from the \gls{LT} steps  coupled with a new client drift reduction technique and a probabilistic approach to deciding whether communication takes place or not.  \citep{ProxSkip} first reformulates  \eqref{eq:P}  into the equivalent consensus form 
\begin{equation} \label{eq:Composite}\min_{x\in \mathbb{R}^{d}} f(x) + r(x),\end{equation}
where $d^\prime=Md$, $x=(x_1,\dots,x_M)\in \mathbb{R}^{d^\prime}$, and 
\begin{equation}\label{eq:consensus} 
\squeeze f(x) \eqdef \sum\limits_{m=1}^M \frac{n}{n_m} f_m(x_m), \quad r(x) = \begin{cases} 0 & \text{ if } x_1=\dots=x_M, \\ +\infty & \text{ otherwise.} \end{cases}\end{equation}
The \gls{ProxSkip} method is a randomized variant of the proximal gradient descent algorithm (\algname{ \gls{ProxGD}})~\citep{Nesterov_composite2013,beck-book-first-order}  for solving \eqref{eq:Composite}, 
with the proximity operator of $r$, given by $$\prox_{r}(x) \eqdef \arg \min_{y} \left(r(y) + \frac{1}{2}\|y-x\|^2\right),$$ being	 evaluated in each iteration with probability $p\in (0,1]$ only. Remarkably, in work \citep{ProxSkip} it is shown that it is possible to choose $p$ as low as $1/\sqrt{\kappa}$, where $\kappa$ is the condition number of $f$, without this worsening the rate of its parent method \algname{ \gls{ProxGD}}. In summary, \gls{ProxSkip} lets the $M$ clients perform $\sqrt{\kappa}$ local gradient steps in expectation, followed by the evaluation of the prox of $r$, which in the case of the  consensus reformulation of \eqref{eq:P}  means averaging across all $M$ nodes, i.e., communication.

\section{ProxSkip-VR: A General Variance Reduction Framework for ProxSkip}\label{sec:main_theory}
\begin{algorithm*}[t]
	\caption{\gls{ProxSkip-VR}}
	\label{alg:ProxSkip-VR}
	\begin{algorithmic}[1]
		\STATE {\bf Parameters:} stepsize $\gamma > 0$, probability $p\in (0,1]$, initial iterate $x^0\in \mathbb{R}^d$, {\red initial control vector $y^{0}\in \mathbb{R}^d$}, {\color{blue} initial gradient shift $h^0 \in \mathbb{R}^d$}, number of iterations $T\geq 1$
		\FOR{$t=0,1,\dotsc,T-1$}
		\STATE {\red$g^t = g(x^{t},y^{t},\xi^t)$}\hfill $\diamond$  Sample  $\xi^t$ and construct an unbiased estimator of $\nabla f(x^{t})$ 
		\STATE $\hat x^{t+1} = x^{t} - \gamma ( {\red g^t} - {\color{blue} h^t})$ \hfill $\diamond$ Take a gradient-type step adjusted via the {\color{blue} shift $h^t$}
		\STATE {\red Construct new control vector  $y^{t+1}$}
		\STATE Flip a coin $\theta_t \in \{0,1\}$ where $\mathop{\rm Prob}(\theta_t =1) = p$ \hfill $\diamond$ Decides whether to skip the prox or not
		\IF{$\theta_t=1$} 
		\STATE  $x^{t+1} = \prox_{\frac{\gamma}{p}r}\bigl(\hat x^{t+1} - \frac{\gamma}{p}{\color{blue}  h^t} \bigr)$ \hfill $\diamond$ Apply prox, but only with probability $p$
		\ELSE
		\STATE $x^{t+1} = \hat x^{t+1}$ \hfill $\diamond$ Skip the prox!
		\ENDIF
		\STATE ${\color{blue} h^{t+1}} = {\color{blue} h^t} + \frac{p}{\gamma}(x^{t+1} - \hat x^{t+1})$ \hfill $\diamond$ Update the {\color{blue} shift $h^t$}
		\ENDFOR
	\end{algorithmic}
\end{algorithm*}

In this work we contribute to the fifth generation of \gls{LT} methods by extending the work \citep{ProxSkip} to allow for a very large family    of gradient estimators, including variance reduced (VR) ones \citep{johnson2013accelerating, defazio2014saga, kovalev2020don, mishchenko2019distributed}. 

Like \gls{ProxSkip}, our method  \gls{ProxSkip-VR} (Algorithm~\ref{alg:ProxSkip-VR}) aims to solve the composite problem \eqref{eq:Composite} in a more general setting (see Assumptions~\ref{ass:L-smoothness}--\ref{ass:Reg}), with the special structure \eqref{eq:consensus} coming from the consensus reformulation  being a special case only. Our method differs from \gls{ProxSkip} in that we replace the gradient $\nabla f(x^t)$ by an unbiased estimator $g^t = g(x^t,{\red y^t},\xi^t)$, where $\xi^t$ is the source of randomness controlling unbiasedness and  {\red $y^{t}$ is a control vector} whose role is to progressively reduce the variance of the estimator, so that $$\Exp{g^t \;|\; x^{t}, {\red y^{t}}} = \nabla f(x^{t}).$$

There are several motivations behind this endeavor. First, it is a-priori not clear whether the novel proof technique employed in \citep{ProxSkip} can be combined with the proof techniques used in the analysis of \gls{VR} methods, and hence it is scientifically significant to investigate the possibility of such a merger of two strands of the literature. We show  that this is possible. Second, marrying \gls{VR} estimators with \gls{ProxSkip} can lead to novel system architectures  which are more elaborate than the simplistic client-server architecture (see Section~\ref{sec:tree}).  Lastly, while researchers contributing to generations 1--4 of \gls{LT} methods were preoccupied with trying to close the gap on \algname{\gls{GD}} in terms of communication efficiency, they {\em ignored} the number of the local steps appearing in their algorithms, and reported their bounds primarily in terms of the number of communication rounds. Bounds reported this way make complete sense in the scenario when the cost of local work (e.g., one \algname{\gls{SGD}} step w.r.t.\ a single data point), say $\delta$, is negligible compared to the cost of communication, which we can w.l.o.g.\ assume to be 1, and when the number of local steps is small. With the advent of the fifth generation of \gls{LT} methods, we can (to a large degree) stop worrying about communication efficiency, and can now ask more refined questions, such as: \begin{quote}{\em Are there gradient estimators which, when combined with \gls{ProxSkip}, lead to faster algorithms in terms of the total cost, which includes the communication cost as well as the cost of local training?} \end{quote} We give an affirmative answer to the question in Sections~\ref{sec:tree} and \ref{sec:experiments}. 


\subsection{Standard assumptions}

We assume throughout that $f$ is differentiable, and let $$D_f(x,y) \eqdef f(x) - f(y) - \langle \nabla f(y), x-y \rangle $$ denote the Bregman divergence of $f$. Throughout the work we make the following assumptions:

\begin{assumption}[$L$-smoothness]
	\label{ass:L-smoothness}
	There exists $L > 0$ such that
    $$2D_f(x,y) \leq L \left\|x-y\right\|^{2}$$
    for all $x,y \in \mathbb{R}^{d}
$.  \end{assumption}

\begin{assumption}[$\mu$-convexity]
	\label{ass:mu-strongly-convex}
	There exists $\mu > 0$ such that 
    $$\mu \left\|x-y\right\|^{2} \leq 2D_f(x,y)$$
    for all $x,y \in \mathbb{R}^{d}$. 	
  \end{assumption}

\begin{assumption}
	\label{ass:Reg}
	The regularizer $r:\mathbb{R}^d\to \mathbb{R}\cup \{+\infty\}$ is proper, closed and convex. \end{assumption}

Under the above assumptions, \eqref{eq:Composite} has a unique minimizer $x^{\star}$. Let $h^{\star} \eqdef \nabla f(x^{\star})$.


\subsection{Modelling variance reduced gradient estimators}

Our next assumption, initially introduced in~\citep{gorbunov2020unified}, postulates several parametric inequalities characterizing the behavior and ultimately the quality of a gradient estimator.  Similar assumptions appeared later in~\citep{gorbunov2021local, gorbunov2020linearly}.
\begin{assumption}
	\label{sigma_t}
	Let $\{x^{t}\}$ be iterates produced by \gls{ProxSkip-VR}. First, we assume that the stochastic gradients $g^{t}=g(x^{t},y^{t},\xi^t)$ are unbiased for all $t\geq 0$, namely \begin{equation}\label{eq:unbiased}\Exp{g^t \;|\; x^{t}, y^{t}} =\nabla f(x^{t}) .\end{equation}
Second, we assume that there exist non-negative constants $A, B, C, \tilde{A}, \tilde{B}, \tilde{C}$, with $\tilde{B}<1$, and a nonnegative mapping $y^{t} \mapsto \sigma(y^{t}) \eqdef \sigma^{(t)}$  such that the following two relations hold for all $t\geq 0$,
\begin{eqnarray}
	\Exp{ \left\|g^{t}-\nabla f(x_{*})\right\|^{2} \;|\; x^{t}, y^{t}} &\leq& 2 A D_{f} (x^{t}, x_{*})+B \sigma^{(t)} +C,\label{eq:sigma-1}\\
	\Exp{ \sigma^{(t+1)} \;|\; x^{t}, y^{t} } &\leq & 2 \tilde{A} D_{f}(x^{t}, x_{*}) +\tilde{B} \sigma^{(t)} +\tilde{C}.\label{eq:sigma_2}
\end{eqnarray}
\end{assumption}

Assumption~\ref{sigma_t} covers a very large collection of  gradient estimators, including an infinite variety of subsampling/minibatch estimators, gradient sparsification and quantization estimators, and their combinations; see \citep{gorbunov2020unified} for examples. \gls{VR} estimators are characterized by $C=\tilde{C}=0$; most non-\gls{VR} estimators by $\tilde{A}=\tilde{B}=\tilde{C}=B=0$ and $C>0$~\citep{gower2019sgd}. 

\subsection{Main result}

We are now ready to formulate our main result.

\begin{theorem}\label{thm:main_vr_proxskip}
Let Assumptions~\ref{ass:mu-strongly-convex} and \ref{ass:Reg} hold, and let $g^t$ be a gradient estimator satisfying Assumption~\ref{sigma_t}. If $B>0$, choose any  $\MM >\nicefrac{B}{(1-\tilde{B})}$ and $\beta= \nicefrac{(B + \MM\tilde{B})}{\MM}$. If $B=0$, let $\MM=0$ and $\beta=\tilde{B}$. Choose stepsize  $
	0<\gamma \leq \min \left\{\nicefrac{1}{\mu}, \nicefrac{1}{(A+\MM \tilde{A}) }\right\}.$
	Then the iterates of \gls{ProxSkip-VR} for any $p\in (0,1]$ satisfy
	\begin{equation}\label{eq:main_thm}
	\squeeze	\Exp{ \Psi^{(T)} } \leq \max \left\{(1-\gamma \mu)^{T},\beta^{T},(1-p^2)^T\right\} \Psi^{(0)}+\frac{\left(C+\MM \tilde{C}\right) \gamma^{2}}{\min \left\{\gamma \mu,p^2, 1 - \beta \right\}},
	\end{equation}
	where the Lyapunov function  is defined by $$\squeeze	\Psi^{(T)} \eqdef \|x^{t} - x^{\star}\|^2 + \frac{\gamma^2}{p^2}\|h^t - h^{\star}\|^2 +\gamma^{2}  \MM \sigma^{(t)}.$$
\end{theorem}

\begin{table*}[t]
    \centering
    \scriptsize
    \caption{Special cases of \gls{ProxSkip-VR}, depending on the choice of the gradient estimator $g^t$.}
    \label{tab:comparison2}
    \begin{threeparttable}
\begin{tabular}{lllc}
 \bf Estimator & \begin{tabular}{c}\bf Communication Complexity\\\bf of \gls{ProxSkip-VR} \end{tabular}&\begin{tabular}{c}\bf Iteration Complexity\\\bf of \gls{ProxSkip-VR} \end{tabular}&\bf \begin{tabular}{c}\bf Corollaries\\\bf of Theorem~\ref{thm:main_vr_proxskip} \end{tabular} \\
\hline
 \algname{GD}\tnote{\color{blue}(b)} & $\mathcal{O}\left(\sqrt{\nicefrac{L}{\mu}}\log\nicefrac{1}{\varepsilon}\right)$ &$\mathcal{O}\left(\nicefrac{L}{\mu}\log\nicefrac{1}{\varepsilon}\right)$&Theorem~\ref{thm:proxskip}  \\ 
\hline
 \algname{\gls{SGD}}\tnote{\color{blue}(c)} & $\mathcal{O} \left(\left(\sqrt{\nicefrac{A}{\mu}}+\sqrt{\nicefrac{2 C}{\varepsilon \mu^{2}}}\right) \log \nicefrac{1}{\varepsilon}\right)$  &$\mathcal{O} \left(\left(\nicefrac{A}{\mu}+\nicefrac{2 C}{\varepsilon \mu^{2}}\right) \log \nicefrac{1}{\varepsilon}\right)$&Theorem~\ref{thm:sproxskip}  \\ 
 \hline
\cellcolor{bgcolor2}\algname{HUB} {\bf [NEW]} &\cellcolor{bgcolor2}$\mathcal{O}\left(\sqrt{\nicefrac{L_{\max}}{\mu}\left(1+\nicefrac{\omega}{\tau}\right)}\log\nicefrac{1}{\varepsilon}\right)$ &\cellcolor{bgcolor2}$\mathcal{O}\left(\nicefrac{L_{\max}}{\mu}\left(1+\nicefrac{\omega}{\tau}\right)\log\nicefrac{1}{\varepsilon}\right)$ &\cellcolor{bgcolor2}Theorem~\ref{thm:QLSVRG}  \\ 
\cellcolor{bgcolor2}\algname{LSVRG} {\bf [NEW]} &\cellcolor{bgcolor2}$\mathcal{O}\left(\sqrt{\nicefrac{L(\tau)}{\mu}}\log\nicefrac{1}{\varepsilon}\right)$ &\cellcolor{bgcolor2}$\mathcal{O}\left(\nicefrac{L(\tau)}{\mu}\log\nicefrac{1}{\varepsilon}\right)$&\cellcolor{bgcolor2}Corollary~\ref{thm:lsvrg-proxskip}  \\ 
 \cellcolor{bgcolor2}\algname{Q} {\bf [NEW]} & \cellcolor{bgcolor2}$\mathcal{O}\left(\sqrt{\nicefrac{L_{\max}}{\mu}\left(1+\nicefrac{\omega}{M}\right)}\log\nicefrac{1}{\varepsilon}\right)$ &\cellcolor{bgcolor2}$\mathcal{O}\left(\nicefrac{L_{\max}}{\mu}\left(1+\nicefrac{\omega}{M}\right)\log\nicefrac{1}{\varepsilon}\right)$ &\cellcolor{bgcolor2}Corollary~\ref{thm:rand-diana-proxskip}  \\ 
\hline
\end{tabular}
  \begin{tablenotes}
        {\tiny
        \item [{\color{blue}(a)}]  Any estimator satisfying Assumption~\ref{sigma_t}
        \item [{\color{blue}(b)}]   \gls{ProxSkip-VR} with the \algname{GD} estimator reduces to the \gls{ProxSkip} method in \citep{ProxSkip}
         \item [{\color{blue}(c)}]     \gls{ProxSkip-VR} with the \algname{\gls{SGD}} estimator satisfying Assumption~\ref{Expected_smoothness} reduces to the \gls{SProxSkip} method in \citep{ProxSkip}
         \item [{\color{blue}(d)}]   $L(\tau) \eqdef \frac{n-\tau}{\tau(n-1)} L_{\max}+\frac{n(\tau-1)}{\tau(n-1)} L$, where $\tau$ is the mini-batch size and $m$ is the number of clients belonging to one hub 
        }
    \end{tablenotes}
    \end{threeparttable}
\end{table*}

\subsection{Two examples of gradient estimators}
Here we give two illustrative examples of  estimators satisfying 
Assumption~\ref{sigma_t}. 
\begin{theorem}[\algname{GD} estimator]
	\label{thm:proxskip}
Let  Assumption~\ref{ass:L-smoothness}, \ref{ass:mu-strongly-convex} and \ref{ass:Reg} hold. Then for the trivial estimator $g^t=\nabla f(x^{t})$,  Assumption~\ref{sigma_t} holds with the following parameters:
\begin{align*}
	A = L, \quad B = 0, \quad C = 0, \quad \tilde{A} = 0, \quad \tilde{B} = 0,  \quad \tilde{C} = 0, \quad \sigma^{(t)} \equiv 0.
\end{align*}
Choose a stepsize satisfying $	0<\gamma \leq  \nicefrac{1}{L}.$ Then the iterates of \gls{ProxSkip-VR} for any  $p\in (0,1]$ satisfy 
	\begin{equation}\label{eq:GD-xx}
	\Exp{\Psi^{(T)}} \leq \max \left\{(1-\gamma \mu)^{T},(1-p^2)^T\right\} \Psi^{(0)},
\end{equation}
where $$\squeeze	\Psi^{(T)} \eqdef \|x^{t} - x^{\star}\|^2 + \frac{\gamma^2}{p^2}\|h^t - h^{\star}\|^2.$$ Let $\gamma = \nicefrac{1}{L}$ and $p = \sqrt{\nicefrac{\mu}{L}}$ then the communication and iteration complexities of \gls{ProxSkip-VR} are 
$$\squeeze \# comms = \mathcal{O}\left(\sqrt{\frac{L}{\mu}}\log\frac{1}{\varepsilon}\right), \qquad \# iters = \mathcal{O}\left(\frac{L}{\mu}\log\frac{1}{\varepsilon}\right),$$ 
respectively.
\end{theorem}

This recovers the result obtained in~\citep{ProxSkip} for their \gls{ProxSkip} method.  The next assumption holds for virtually all (non-\gls{VR}) estimators based on subsampling~\citep{gower2019sgd}. 
   \begin{assumption}[Expected smoothness]
   	\label{Expected_smoothness}
   	We say that an unbiased estimator $g(x;\xi):\mathbb{R}^d\to\mathbb{R}^d$ of the gradient $\nabla f(x)$ satisfies the expected smoothness inequality if there exists $A^{\prime\prime} >0$ such that
   	\begin{align*}
   		\Exp{\|g(x;\xi) - g(x^{\star};\xi)\|^2} \leq 2A^{\prime\prime}D_f(x,x^{\star}),\quad \forall x\in \mathbb{R}^d.
   	\end{align*}	   	
   \end{assumption}

\begin{theorem}[\algname{\gls{SGD}} estimator]
	\label{thm:sproxskip}
	Let	$g(x,\xi)$ satisfy Assumption~\ref{Expected_smoothness} and define estimator $g^t \eqdef g(x^{t},\xi^t)$, where $\xi^t$ is chosen independently at time $t$.  Then Assumption~\ref{sigma_t} holds with the following parameters:
	\begin{align*}
		&A = 2A^{\prime\prime}, \quad B = 0, \quad C = 2{\rm Var}(g(x^{\star},\xi)),\\
        &\tilde{A} = 0, \quad \tilde{B} = 0,  \quad \tilde{C} = 0, \quad \sigma^{(t)} \equiv 0.
	\end{align*}
 Moreover, assume that Assumption~\ref{ass:mu-strongly-convex} holds. Set stepsize $	0<\gamma \leq \min \left\{\nicefrac{1}{\mu}, \nicefrac{1}{A}\right\}.$ Then the iterates of \gls{ProxSkip-VR} for any  probability $p\in (0,1]$ satisfy 
	\begin{equation}
		\squeeze
		\Exp{\Psi^{(T)}} \leq \max \left\{(1-\gamma \mu)^{T},(1-p^2)^T\right\} \Psi^{(0)} + \gamma^2 \frac{2{\rm Var}(g(x^{\star},\xi))}{\min\left\lbrace \gamma\mu,p^2 \right\rbrace},\label{eq:GD-yy}
	\end{equation}
	where the Lyapunov function is defined by $$\squeeze	\Psi^{t} \eqdef \|x^{t} - x^{\star}\|^2 + \frac{\gamma^2}{p^2}\|h^t - h^{\star}\|^2.$$ If we choose $\gamma=\min \left\{\nicefrac{1}{A}, \nicefrac{\varepsilon \mu}{2 C}\right\}$ and $p=\sqrt{\gamma \mu}$ then the communication and iteration complexities of  \gls{ProxSkip-VR}  are 
	$$\squeeze \# comms = \mathcal{O} \left(\left(\sqrt{\frac{A}{\mu}}+\sqrt{\frac{C}{\varepsilon \mu^{2}}}\right) \log \frac{1}{\varepsilon}\right), \quad \# iters = \mathcal{O} \left(\left(\frac{A}{\mu}+\frac{C}{\varepsilon \mu^{2}}\right) \log \frac{1}{\varepsilon}\right),$$ 
	respectively.
\end{theorem}
This recovers the result obtained in~\citep{ProxSkip} for their stochastic variant of \gls{ProxSkip}, which they call \gls{SProxSkip}.

\section{New FL Architecture: Regional Hubs Connecting the Clients to the Server} \label{sec:tree}

We  now illustrate the versatility of our \gls{ProxSkip-VR} framework by designing a new ``\gls{FL} architecture'' and proposing an algorithm that can efficiently operate in this setting.

\begin{figure}[!h]
	\centering
	\includegraphics[width=3in]{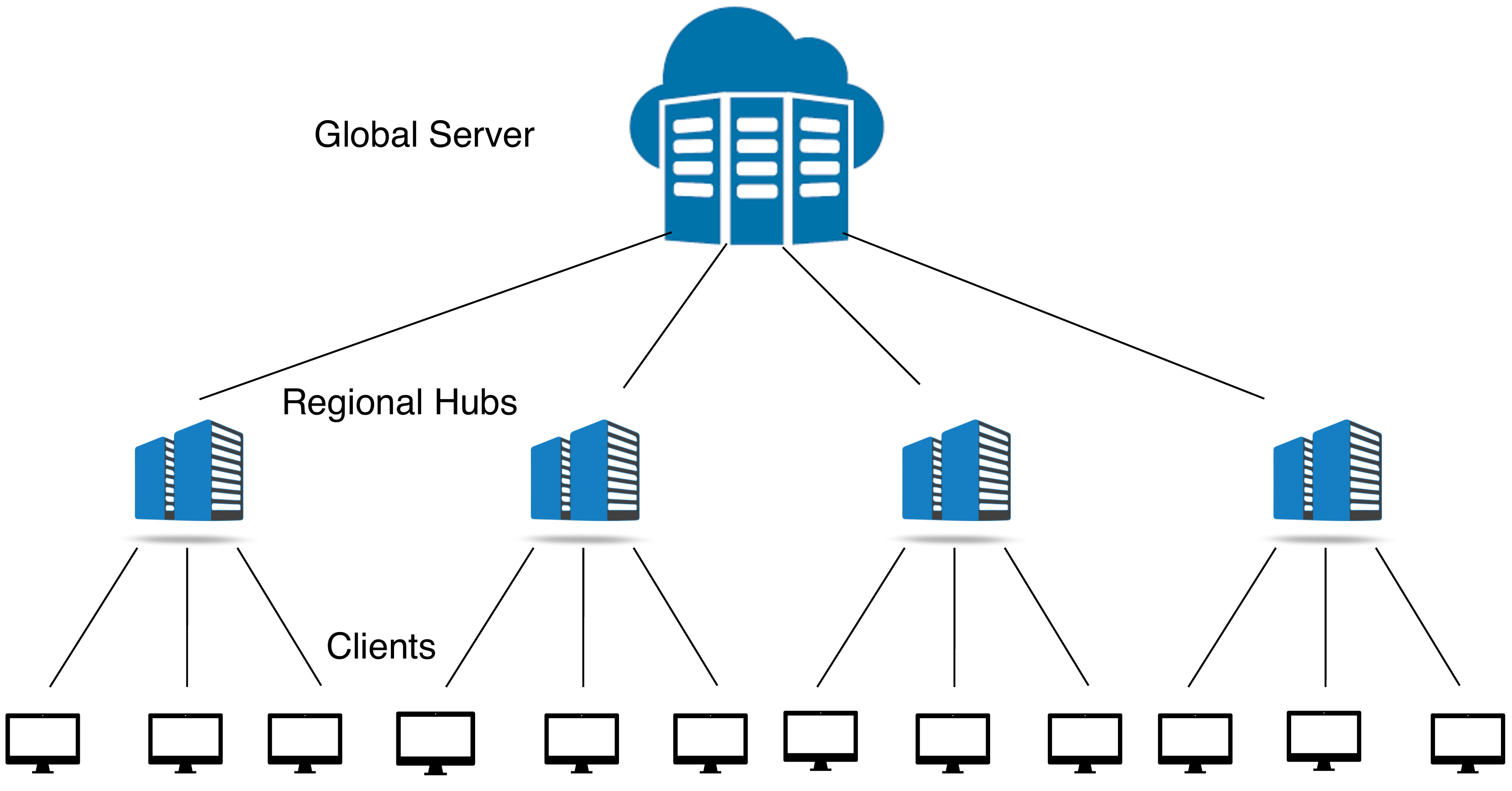}
	\caption{Server-hubs-clients FL architecture with 4 hubs and 12 clients.}
	\label{logo}
\end{figure}

In particular, we consider the situation where the clients are clustered (e.g., based on region), and where {\em a hub} is placed in between each cluster and the central server. Clients communicate with their regional hub only, which can communicate with the central server (see Figure~\ref{logo}). There are $M$ hubs, hub $m$ handles $n_m$ clients, and client $i$ associated with hub $m$ owns loss function $f_{m,i}$. 

Mathematically, this can be modeled by problem \eqref{eq:P}. In this situation, we care about two sources of communication cost: between the server and the hubs, and between the hubs and the clients. We propose to handle this via {\em local training (\gls{LT})}  between the server and the hubs, and via {\em Partial Participation (\gls{PP})} and  {\em \gls{CC}}  between the hubs and the clients. Algorithmically, from the server-hubs perspective, we are applying a particular variant of \gls{ProxSkip-VR}  to \eqref{eq:Composite}--\eqref{eq:consensus}, where $f_m$ is the aggregate loss handled by hub $m$. This takes care of communication efficiency between the server and the hub. Note also that we need not worry about partial participation of hubs, as these are designed to be always available. However, in this situation, it is costly for hub $i$ to compute the gradient of $f_m$  as this involves communication with all the clients it handles. 

\subsection{Handling of  partial participation and compressed communication}
In order to alleviate this burden, we propose a combination of \gls{PP} and \gls{CC}. However, we need to be very careful about how to do this. Indeed, both \gls{PP} and \gls{CC}, even when applied in isolation, and without \gls{ProxSkip} in the mix, can lead to a substantial slowdown in convergence. For example, one will typically lose linear convergence in the strongly convex regime. However, techniques for preserving linear convergence in the presence of \gls{PP} and \gls{CC} exist: this is what variance reduction strategies are designed to do. For example,  \algname{LSVRG} \citep{hofmann2015variance,kovalev2020don}
is a \gls{VR} technique for reducing the variance due to \gls{PP}, and \gls{DIANA} \citep{mishchenko2019distributed} is a \gls{VR} technique for reducing the variance due to \gls{CC}. However, we are not aware of any \gls{VR} method that combines \gls{PP} (applied first) and \gls{CC} (applied second). 

We now propose such a technique. In iteration $t$, every hub $m\in \{1,2,\dots,M\}$ selects a random subset $\set_m\subseteq \{1,2,\dots,n_m\}$ of the clients it handles of cardinality $\tau_m$, chosen uniformly at random, and estimates the hub gradient via
 \begin{equation}
 	\label{eq:hub_grad}
 \squeeze	\nabla f_m(x^{t}) \approx g^t_m \eqdef \frac{1}{|\set_m|} \sum \limits_{j \in \set_m} \cQ^t_{m,i} \left( \nabla f_{m,j}(x^{t}) - \nabla f_{m,j}(y^{t}) \right) + \nabla f_m(y^{t}),
 \end{equation}
where $\cQ^t_{m,i}:\mathbb{R}^{d'}\to\mathbb{R}^{d'}$ is a randomized compression (e.g., sparsification or quantization) operator~\citep{alistarh2017qsgd,khirirat2018distributed,DIANA2,Cnat,Artemis2020}, i.e., a mapping satisfying 	
 \begin{align}
 	\label{compress}
		\Exp{\cQ^t_{m,i}(x)}=x, \quad \Exp{ \|\cQ^t_{m,i}(x)-x\|^{2} } \leq \omega\left\|x\right\|^{2}, \quad \forall x\in \mathbb{R}^{d'},
	\end{align}  
and the control vector $y^{t}$ is updated probabilistically as follows:
 \begin{equation}\label{eq:LSVRG-step}
	y^{t+1}=\left\{\begin{array}{lll}
		x^{t} & \text { with probability } & q \\
		y^{t} & \text { with probability } & 1-q
	\end{array}\right..
\end{equation}
The global gradient estimator (a vector in $\mathbb{R}^{Md'}$), which we call \algname{HUB}, is constructed as a concatenation of the above hub estimators:
\begin{equation}\squeeze \nabla f(x^{t}) \eqdef \left(\frac{n_m}{n} \nabla f_m(x^{t})\right)_{m=1}^M \approx g^t \eqdef g(x^{t},y^{t},\xi^t)\eqdef \left(\frac{n_m}{n} g^t_m\right)_{m=1}^M,\label{eq:U*G(SY*(S*Yddd}\end{equation}
where $\xi^t$ represents the combined randomness from the compressors $\{\cQ^t_{m,i}\}$ and random sets $\{\set_m\}$. 

In order to analyze \gls{ProxSkip-VR} in the consensus form, from now on we assume that $n_m=n = \nicefrac{N}{M}$ and $\tau_i=\tau \in \{1,2,\dots,n\}$ for all $i$, and rely on a slightly different, more general reformulation: 
\begin{align*}
\squeeze	\min \limits_{x \in \mathbb{R}^{d}} \frac{1}{n}\sum\limits_{j=1}^{n}\widetilde{f}_j(x)+r(x), \quad \widetilde{f}_j(x)\eqdef\frac{1}{M}\sum\limits_{m=1}^{M}  f_{m,j}\left(x_{m}\right),
\end{align*}
where regularizer $r(x)$ has the following form:
\begin{align*}
r(x)\eqdef \begin{cases}0 & \text { if } x_{1}=\cdots=x_{M}, \\ +\infty & \text { otherwise.}\end{cases}
\end{align*}

 Our proposed method \gls{ProxSkip-HUB} is \gls{ProxSkip-VR} combined with the novel \algname{HUB} estimator \eqref{eq:U*G(SY*(S*Yddd}, applied to the above reformulation; see Algorithm~\ref{alg:ProxSkip-HUB}.

 \begin{algorithm*}[!th]
	\caption{\gls{ProxSkip-HUB}}
	\label{alg:ProxSkip-HUB}
	\begin{algorithmic}[1]
		\STATE {\bf Input}: stepsize $\gamma > 0$, probabilities $p>0$, $q>0$, initial iterate $x^0\in \mathbb{R}^d$, initial shift ${\red y^{0}}\in \mathbb{R}^d$, initial control variate ${\color{blue} h^0} \in \mathbb{R}^d$, number of iterations $T\geq 1$
		\FOR{$t=0,1,\dotsc,T-1$}
		\STATE broadcast $x^{t}$ to all clients
		\FOR{$m \in \set$}
		\STATE $ \hat{\Delta}^{t}_{m}=Q\left(\nabla f_m(x^{t}) - \nabla f_m({\red y^{t}})\right) $ \hfill $\diamond$ Apply compression operator
		\ENDFOR
		\STATE $\hat{\Delta}^{t}=\frac{1}{\tau} \sum_{m \in \set} \hat{\Delta}^{t}_{m}$
		\STATE $\hat{g}=\hat{\Delta}^{t} + \nabla f({\red y^{t}})$
		
		\STATE $\hat x^{t+1} = x^{t} - \gamma (\hat{g}^t - {\color{blue} h^t})$ \hfill $\diamond$ Take a gradient-type step adjusted via the control variate ${\red h^t}$
		\STATE Flip a coin $\theta_t \in \{0,1\}$ where $\mathop{\rm Prob}(\theta_t =1) = p$ \hfill $\diamond$ To decide whether to skip the prox or not
		\IF{$\theta_t=1$} 
		\STATE  $x^{t+1} = \prox_{\frac{\gamma}{p}\psi}\bigl(\hat x^{t+1} - \frac{\gamma}{p}{\color{blue} h^t} \bigr)$ \hfill $\diamond$ Apply prox, but only very rarely! (with probability $p$)
		\ELSE
		\STATE $x^{t+1} = \hat x^{t+1}$ \hfill $\diamond$ Skip the prox!
		\ENDIF
		\STATE ${\red h^{t+1}} = {\red h^t} + \frac{p}{\gamma}(x^{t+1} - \hat x^{t+1})$ \hfill $\diamond$ Update the control variate ${\color{blue} h^t}$
		
		\STATE $y^{t+1}=\left\{\begin{array}{lll}
			x^{t} & \text { with probability } & q \\
			y^{t} & \text { with probability } & 1-q
		\end{array}\right.$ \hfill $\diamond$ Update the shift ${\red y^{t}}$
		\ENDFOR
	\end{algorithmic}
\end{algorithm*}

\subsection{Theory for ProxSkip-HUB}
In the following result we first claim that the above estimator satisfies Assumption~\ref{sigma_t} with $C=\tilde{C}=0$ (i.e., it is variance-reduced), and the rest of the claim follows by application of our general theorem, Theorem~\ref{thm:main_vr_proxskip}.

\begin{theorem}\label{thm:QLSVRG}
Assume that $\nabla \widetilde{f}_i$ is $L_i$-smooth for all $i$ and let Assumptions~\ref{ass:mu-strongly-convex} and \ref{ass:Reg} hold. Then for the gradient estimator \eqref{eq:U*G(SY*(S*Yddd}, Assumption~\ref{sigma_t} holds with the following constants:
\begin{align*}
&A = 4\left(L(\tau)+\frac{\omega}{\tau}L_{\max}\right) , 
\quad B = 4\left(1+\frac{\omega}{\tau}\right), \quad C = 0,\\
 &\tilde{A} = q L_{\max},\quad \tilde{B} = 1-q, \quad \tilde{C} = 0,
\end{align*}
and $\sigma^{(t)} \eqdef \sigma(y^{t})$, $\sigma(y) \eqdef \frac{1}{n}\sum_{i=1}^{n} \|\nabla \widetilde{f}_i(y) - \nabla \widetilde{f}_i(x^{\star})\|^2$,	
 $L_{\max} \eqdef \max_i L_i $.  Set $\MM=\nicefrac{2B}{(1-\tilde{B})}$ and $
	0<\gamma \leq \min \left\{\nicefrac{1}{\mu}, \nicefrac{1}{(A+\MM\tilde{A}) }\right\}.$
Then the iterates of \gls{ProxSkip-VR} for any $p\in (0,1]$ satisfy 
		\begin{align*}
	\squeeze 	\Exp{\Psi^{(T)}} \leq \max \left\{(1-\gamma \mu)^{T},(1-p^2)^T,\left(1-\frac{q}{2}\right)^{T}\right\} \Psi^{(0)},
	\end{align*}
	where the Lyapunov function is defined by $$\squeeze	\Psi^{(T)} \eqdef \|x^{t} - x^{\star}\|^2 + \frac{\gamma^2}{p^2}\|h^t - h^{\star}\|^2 + \gamma^{2} \frac{8}{q}\left(1+\frac{\omega}{\tau}\right) \sigma^{(t)}.$$
\end{theorem}

We now consider two special cases. In the first, we specialize to the no compression regime, and in the second, to the full participation regime.

\begin{corollary}[No compression]
			\label{thm:lsvrg-proxskip}
	If we do not use compression (i.e., $\omega = 0$), then the communication and iteration complexities are $$\squeeze \# comms = \mathcal{O}\left(\sqrt{\frac{L_{\max}}{\mu}}\log \frac{1}{\varepsilon}\right), \qquad \# iters = \mathcal{O}\left(\frac{L_{\max}}{\mu}\log \frac{1}{\varepsilon}\right),$$ respectively. However, if we use the estimator~\eqref{eq:hub_grad} in Theorem~\ref{thm:main_vr_proxskip} directly, then the communication and iteration complexities are
	$$\squeeze \# comms = \mathcal{O}\left(\sqrt{\frac{L(\tau)}{\mu}}\log \frac{1}{\varepsilon}\right), \qquad \# iters = \mathcal{O}\left(\frac{L(\tau)}{\mu}\log \frac{1}{\varepsilon}\right),$$ where $L(\tau)\eqdef \frac{n-\tau}{\tau(n-1)} L_{\max}+\frac{n(\tau-1)}{\tau(n-1)} L$.
\end{corollary}
Notice that $L_{\max}\geq L$, and that $L(\tau)=L_{\max}$ for $\tau=1$ and $L(\tau) = L$ for $\tau=n$. Moreover, $L(\tau)$ decreases as the minibatch size $\tau$ increases. 

\begin{corollary}[No Partial Participation]
		\label{thm:rand-diana-proxskip}
	If we do not use Partial Participation (i.e., $\tau = n$), then the communication and iteration complexities are 
 $$\squeeze \# comms = \mathcal{O}\left(\sqrt{\frac{L_{\max}}{\mu}\left(1+\frac{\omega}{M}\right)}\log \frac{1}{\varepsilon}\right), \text{ } \#iters = \mathcal{O}\left(\frac{L_{\max}}{\mu}\left(1+\frac{\omega}{M}\right)\log \frac{1}{\varepsilon}\right)$$ respectively.
\end{corollary}

Assume $r(x)\equiv 0$. If $\cQ^t_{m,i}(x)\equiv x$, then $\omega = 0$, and we restore  the well-known \algname{LSVRG} method~\citep{hofmann2015variance,kovalev2020don}, assuming that the functions $f_{ij}$ have the same smoothness constant. We can recover the same rate exactly as well, but with a slightly more refined analysis, one in which we do not need to work with compressors (\algname{LSVRG} does not involve any), which makes for a tighter analysis. On the other hand, if $\tau = n$, we restore the rate of the well-known \gls{DIANA}~\citep{mishchenko2019distributed,DIANA2} method and its sibling \algname{Rand-DIANA}~\citep{Shifted}.

\section{Experiments}\label{sec:experiments}
\begin{figure}[t]
	\centering
	\begin{subfigure}[b]{0.3\textwidth}
		\centering
		\includegraphics[trim=20 10 40 40, clip, width=\textwidth]{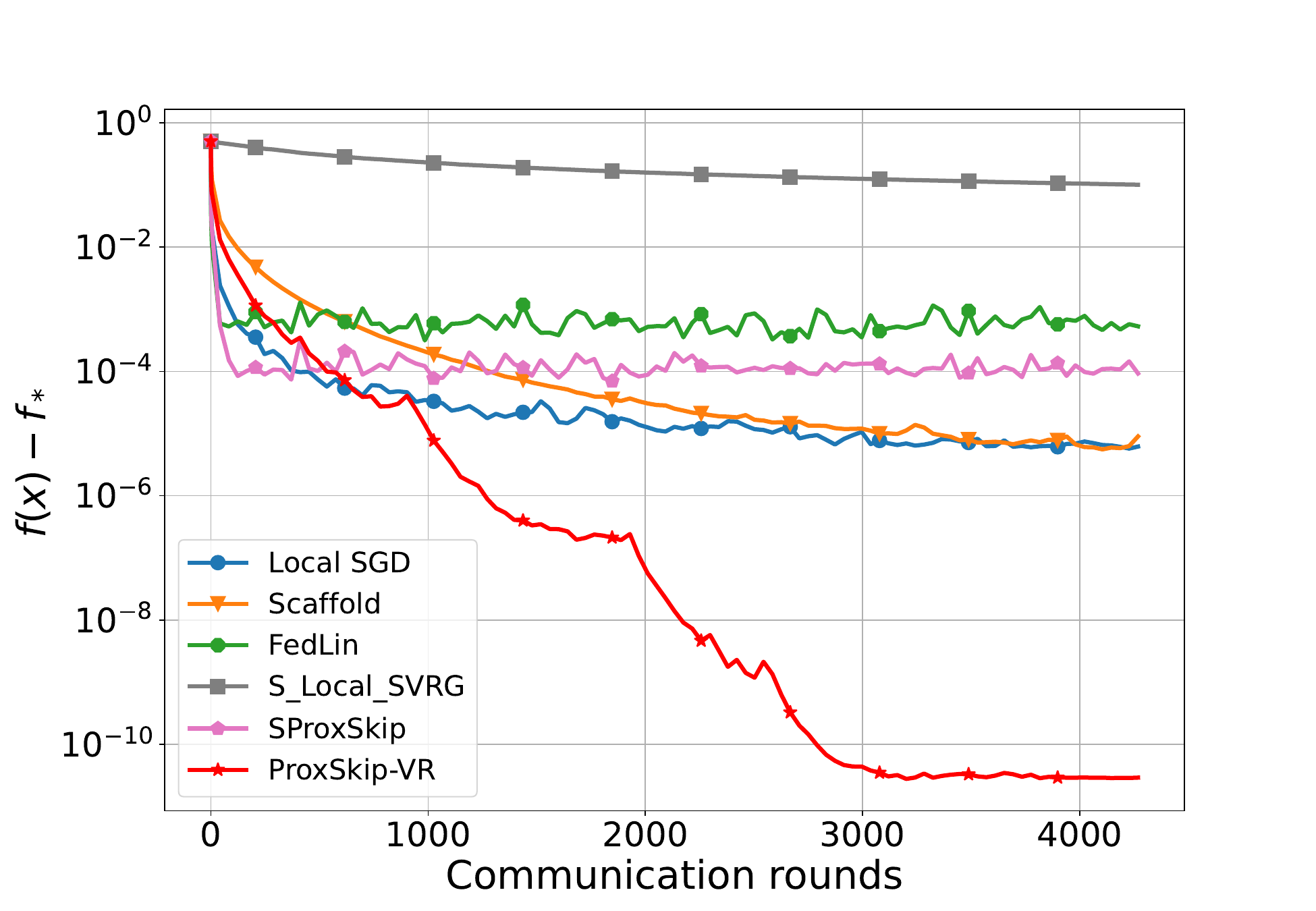}
		\caption{$\tau=16$}
	\end{subfigure}
	\hfill 
	\begin{subfigure}[b]{0.3\textwidth}
		\centering
		\includegraphics[trim=20 10 40 40, clip, width=\textwidth]{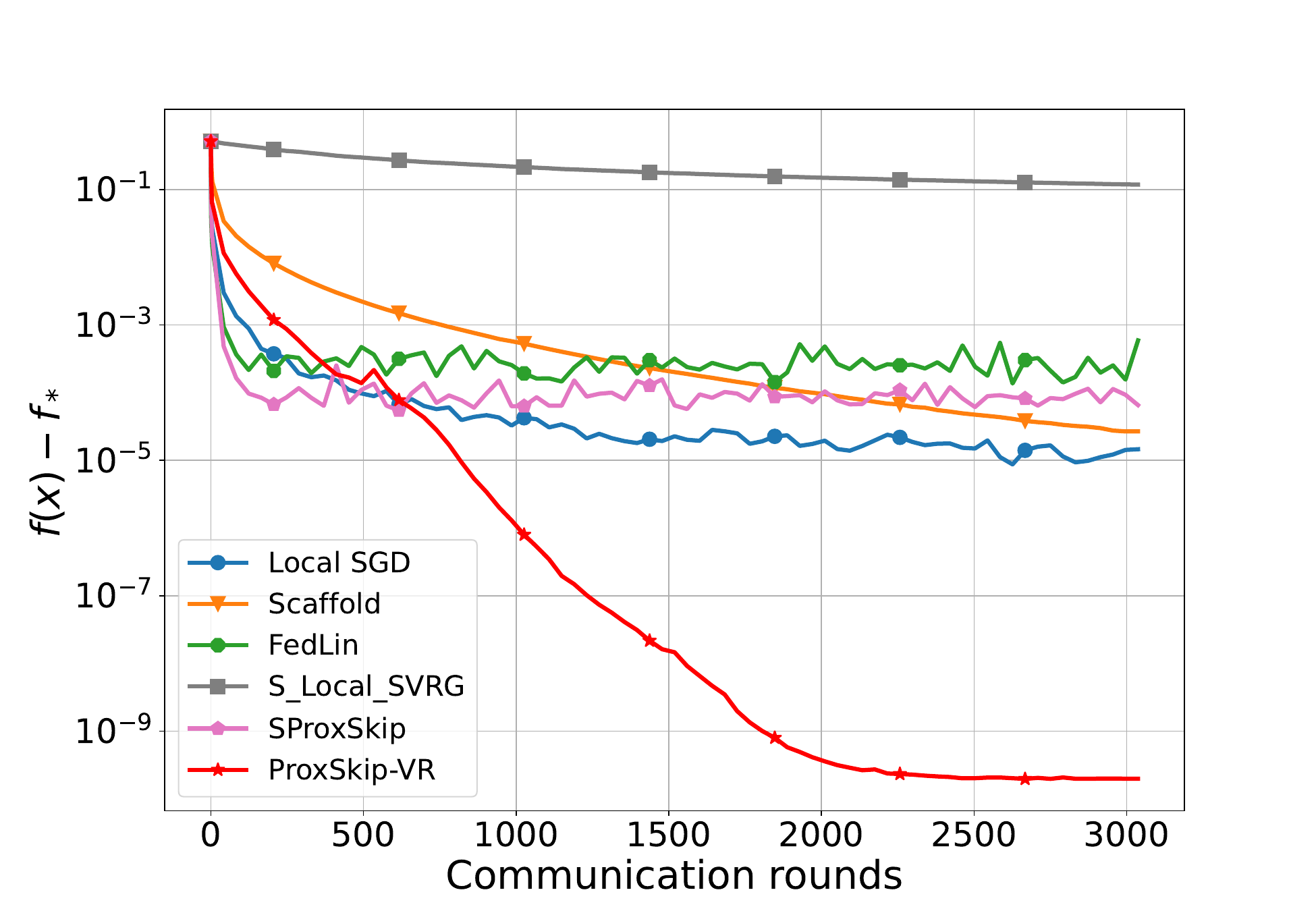}
		\caption{$\tau=32$}
		\end{subfigure}
		\hfill
	\begin{subfigure}[b]{0.3\textwidth}
		\centering
		\includegraphics[trim=20 10 40 40, clip, width=\textwidth]{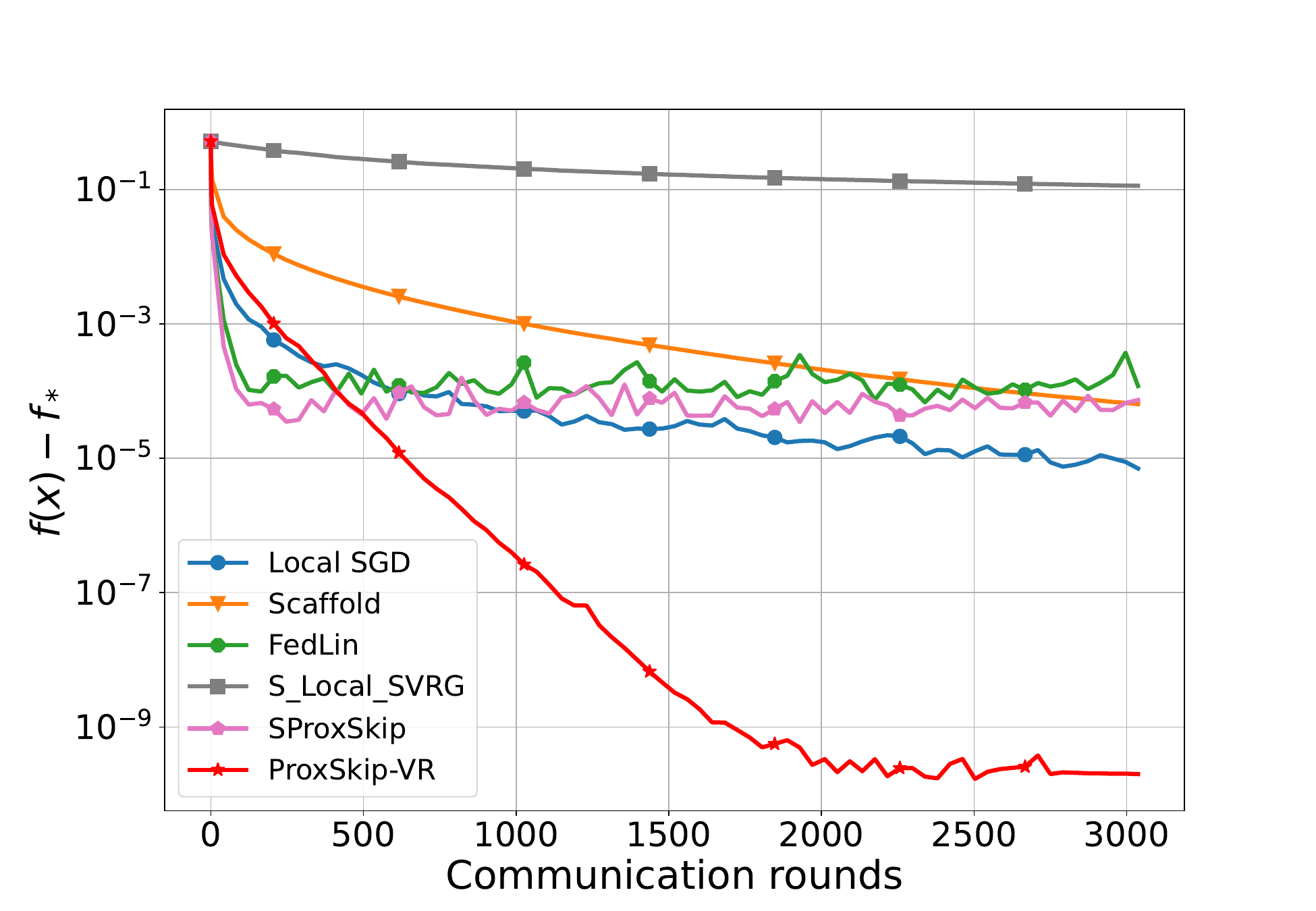}
		\caption{$\tau=64$}
	\end{subfigure}
	\begin{subfigure}[b]{0.3\textwidth}   
		\centering
		\includegraphics[trim=0 10 40 20, clip, width=\textwidth]{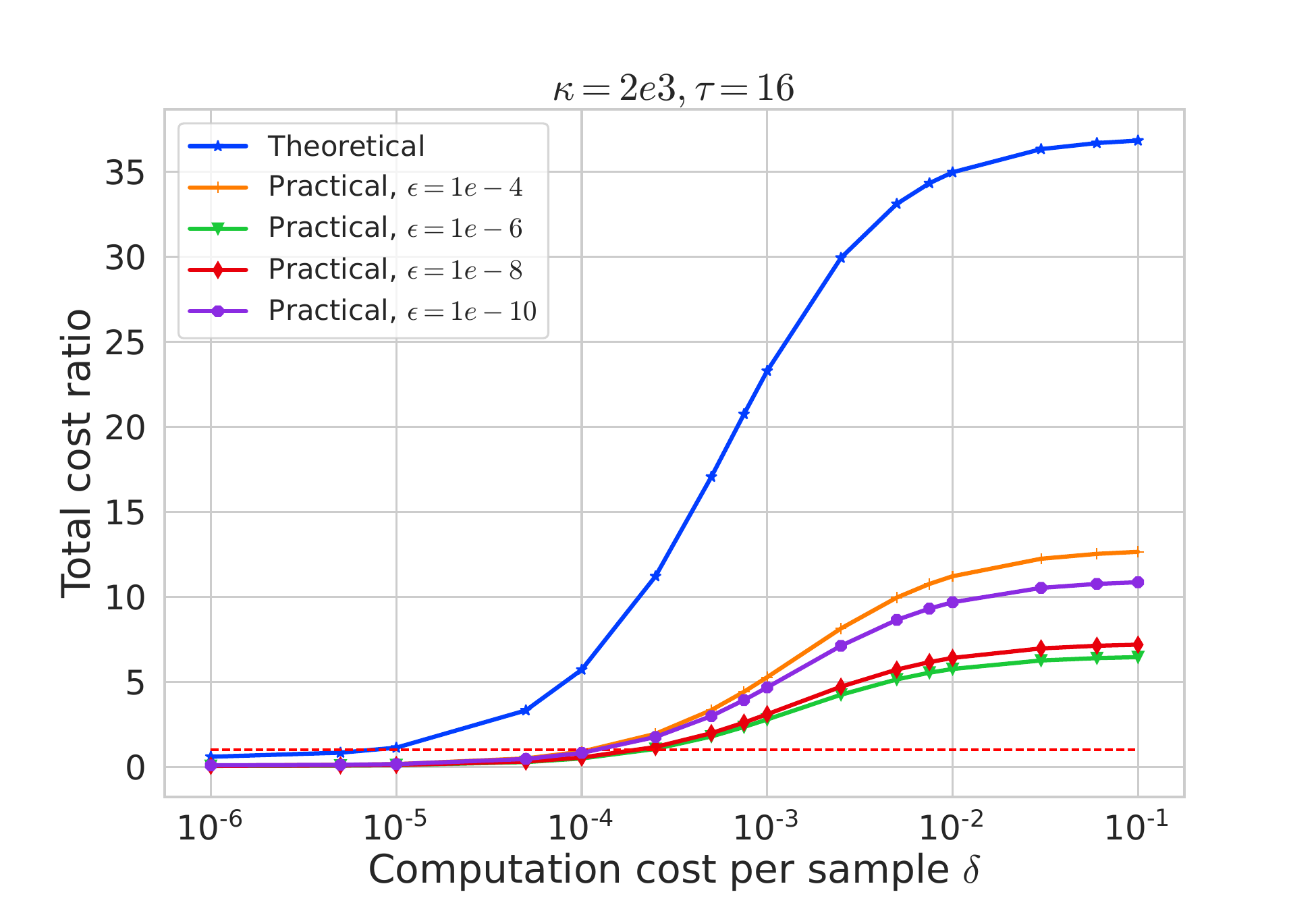}
	\end{subfigure}
	\hfill  
	\begin{subfigure}[b]{0.3\textwidth}
		\centering
		\includegraphics[trim=0 10 40 20, clip, width=\textwidth]{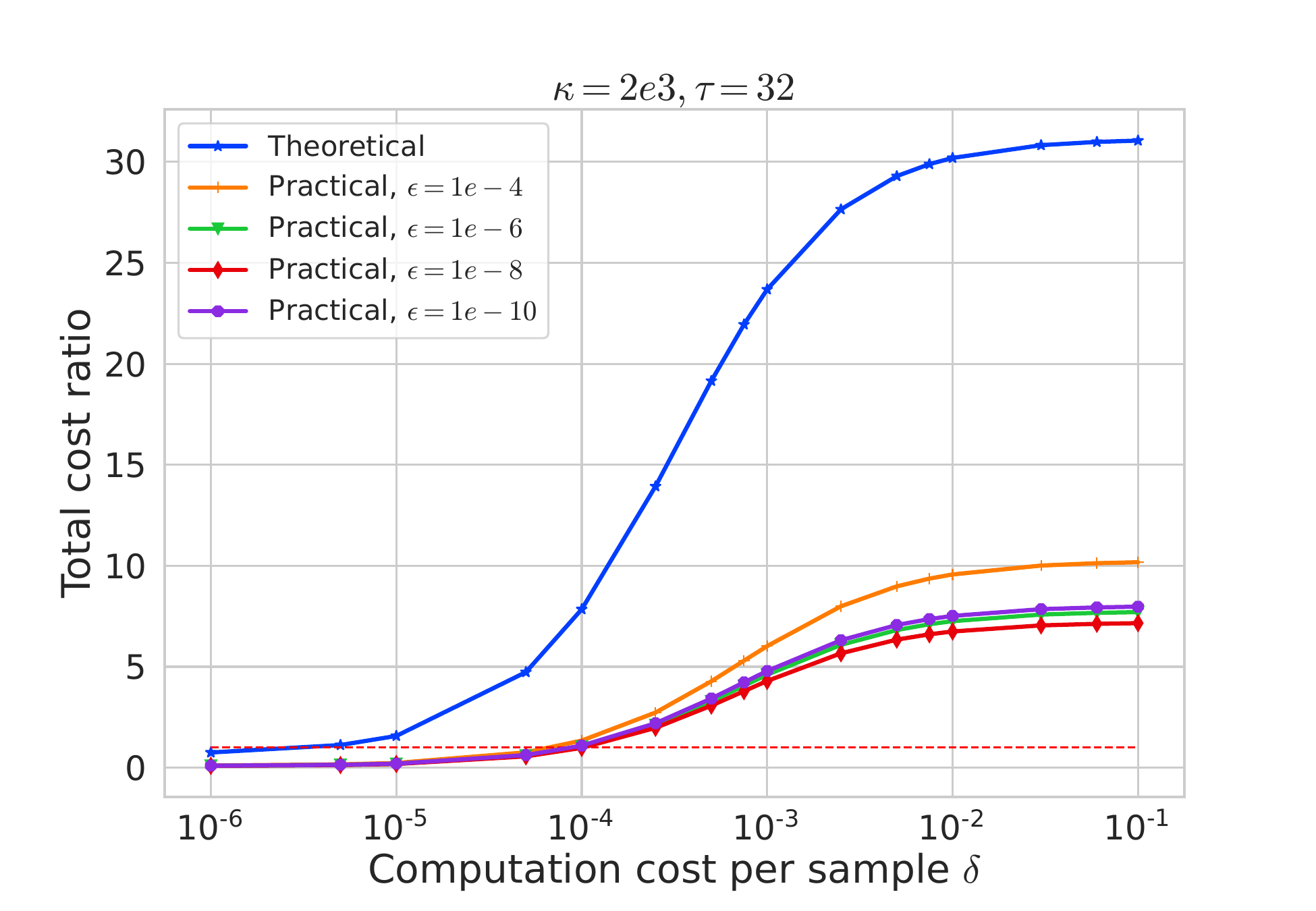}
	\end{subfigure}   
	\hfill  
	\begin{subfigure}[b]{0.3\textwidth}
		\centering
		\includegraphics[trim=0 10 40 20, clip, width=\textwidth]{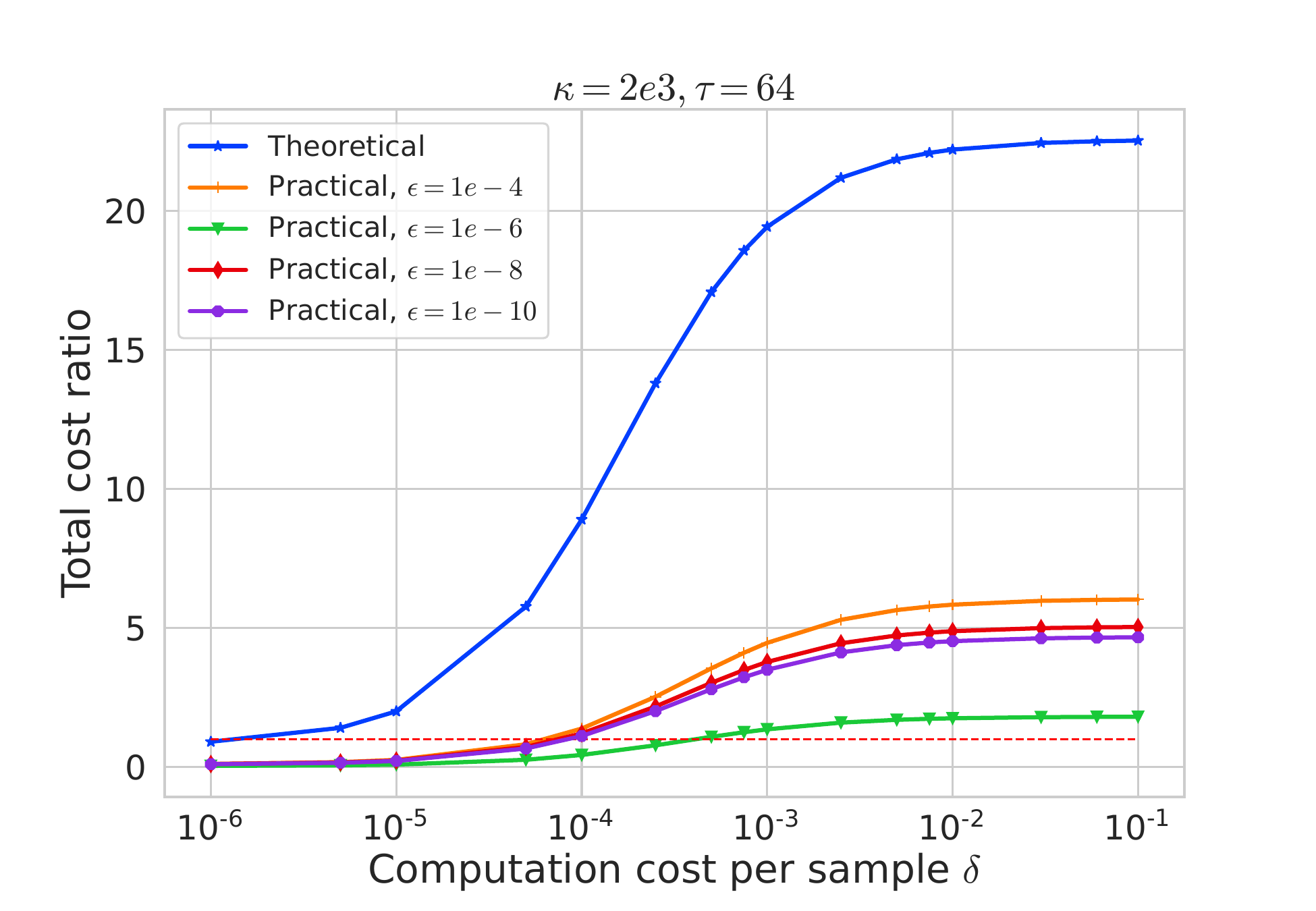}
	\end{subfigure} 
	\caption{The top row shows the convergence results compared with baselines and the second row is the total cost ratio of \gls{ProxSkip} over our \gls{ProxSkip-VR}.}
	\label{fig:exp_comp_total_cost}
\end{figure} 

To illustrate the predictive power of our theory, it suffices to  consider logistic regression with an $L_2$ regularizer in the distributed setting \eqref{eq:P},

with 
$$ f_m(x)=\frac{1}{n_m} \sum \limits_{i=1}^{n_m} \log \left(1+\exp \left(-b_{m,i} a_{m,i}^{\top} x\right)\right)+\frac{\lambda}{2}\|x\|^{2},$$ where $n_m$ is the number of data points per worker, where $a_{m,i}\in \mathbb{R}^{d'}$ and $b_{m,i} \in \{-1, +1\}$ are the data samples and labels. We choose $n_m=n$ for all $m$. We set the regularization parameter $\lambda = 5\cdot10^{-4}L$ by default, where $L$ is the smoothness constant of $f$. We conduct several experiments on the \textbf{w8a} dataset from the LibSVM library~\citep{chang2011libsvm}.

\subsection{ProxSkip-LSVRG vs baselines}
In Figure~\ref{fig:exp_comp_total_cost} (first row), we compare various \gls{LT} baselines\footnote{With the exception of \algname{\gls{LSGD}}, all  use client drift correction.}  for three choices of mini-batch sizes ($\tau=16, 32, 64$) with our method \gls{ProxSkip-VR} combined with the \algname{LSVRG} estimator, which is a special case of the \algname{HUB} estimator when $\cQ^t_{m,i}(x)\equiv x$ for all $m, i$. We see that our method outperforms all other methods significantly due to its communication-acceleration properties. 

\subsection{ProxSkip-LSVRG can significantly reduce total cost compared to ProxSkip}

Next, we derive the total cost, which includes communication cost (assumed to be $1$, for normalization purposes), and computational cost (assumed to be $\delta$, and equal to the cost of performing one \algname{\gls{SGD}} step with a single data point). Let us consider the total cost of \gls{ProxSkip-VR} in the case of the \algname{LSVRG} estimator: in each iteration we compute 1 stochastic gradient, and with probability $q$ we compute the exact gradient. We do not need to compute a second stochastic gradient since we can use memory and the relation $y^{t+1} = x^{t}$. The total cost for \gls{ProxSkip-VR} is equal to 
\begin{align*}
\text{Cost}(\text{\gls{ProxSkip-VR}}) &\eqdef T_{\text{comm.}}(\text{\gls{ProxSkip-VR}})\\
&+ \delta   \left(q n + (1-q)\tau+\tau\right)T_{\text{iter}}({\text{\gls{ProxSkip-VR}}}).
\end{align*}
\gls{ProxSkip} requires full/exact gradient computation at each iteration, so the total cost of \gls{ProxSkip} is 
$$\text{Cost}(\text{\gls{ProxSkip}}) \eqdef T_{\text{comm.}}(\text{\gls{ProxSkip}}) + \delta n  T_{\text{iter}}({\text{\gls{ProxSkip}}}).$$ 
Using Theorems~\ref{thm:proxskip} and \ref{thm:QLSVRG} and the value  $\squeeze  L(\tau) \eqdef \frac{n-\tau}{\tau (n-1)}L_{\max} + \frac{n(\tau - 1)}{\tau (n-1)} L$
of the expected smoothness constant for sampling with minibatch size $\tau$ uniformly at random,
 we get the following expression for the cost ratio, expressed as a function of $\delta$: 
\begin{align}\label{eq:09uf09ufdff} 
\notag		    \text{Cost ratio}(\delta) &\eqdef \frac{\text{Cost}(\text{\gls{ProxSkip}})}{\text{Cost}(\text{\gls{ProxSkip-VR}})}\\
            &= \frac{\sqrt{\mu L} + nL \delta}{\sqrt{\mu L(\tau)} + \left( 2n\mu + (2L(\tau)-2\mu)\tau\right)\delta}.
\end{align}
We can easily calculate the limits of this expression: 
\begin{align*}
	\squeeze
	   \text{Cost ratio} (\delta=0) = \sqrt{\frac{L}{L(\tau)}},      \quad    \text{Cost ratio} (\delta\to \infty)= \frac{n L}{2 (n \mu+\left(L(\tau)- \mu\right) \tau)}.
\end{align*}
Since $L(\tau) \geq L$, the cost ratio is below 1 when $\delta=0$ and $\tau<n$, which means that \gls{ProxSkip} is better than \gls{ProxSkip-VR}. This is to be expected since $\delta=0$ means that we {\em ignore} the cost of local computation entirely, which offers an advantage to the former method. On the other hand, the cost ratio is an increasing function of $\delta$, and often reaches the threshold of 1 with a small value of $\delta$; in Figure~\ref{fig:exp_comp_total_cost} (second row), this threshold is reached for $\delta \in [10^{-6},10^{-5}]$ in all three plots. 

In Figure~\ref{fig:exp_comp_total_cost} (second row), we depict the theoretical cost ratio according to \eqref{eq:09uf09ufdff} and the corresponding experimental cost ratio obtained by an actual run of both methods to achieve $\varepsilon$-accuracy, with $\varepsilon = 10^{-6} $ and  $\varepsilon = 10^{-8}$. Remarkably, the experimental results follow the same pattern as our theoretical prediction. The empirical curves appear lower because we use approximations for $L_{\max}$, $L$ and $\mu$. As we can see, starting from $\delta = 10^{-4}$, \gls{ProxSkip-VR} starts to outperform \gls{ProxSkip}. As $\delta$ grows, the advantage of variance reduction embedded in \gls{ProxSkip-VR} over vanilla \gls{ProxSkip} grows. These results suggest that variance reduction is of practical utility in terms of the total cost, even for small values of $\delta$, and its effectiveness grows with $\delta$.

\chapter{5GCS: Integrating Partial Participation into Accelerated Local Training}
\label{chapter4}
\thispagestyle{empty}

\section{Introduction} 

 {\em Federated learning} (\gls{FL}) is an emerging paradigm for the training of supervised Machine Learning models over distributed and often private datasets stored across a potentially very large number of clients' devices, such as mobile phones, edge devices and hospital servers. 
 
 The roots of this young field can be traced to four foundational papers dealing with federated optimization \citep{FedOpt2016}, communication compression \citep{FedLearn2016},  federated averaging \citep{mcmahan2017communication} and secure aggregation \citep{FL-secure_aggreg}\footnote{These four works are cited in the Google AI blog \citep{FLblog2017} which  originally announced \gls{FL} to the general public.}.  
 
 Federated learning has grown massively
 since its inception---in volume, depth and breadth alike---with many advances in theory, algorithms, systems and practical applications~\citep{kairouz2019advances,FL_survey_2020,FieldGuide2021}. 

In this work we study the standard
optimization formulation of  Federated Learning, which has  the form
\begin{equation}
	\label{eq:main}
	\squeeze 
		\min \limits_{x \in \mathbb{R}^d}\left[f(x) \eqdef \frac{1}{\Mx} \sum \limits_{\iclient=1}^{\Mx}  f_\iclient(x)\right],
\end{equation}
where $\Mx$ is the number of clients and each function 
$f_\iclient(x)\eqdef \EE{\xi\sim \cD_\iclient}{\ell(x,\xi)}$ 
represents the average loss, measured via the loss function $\ell$, of the model parameterized by $x\in \mathbb{R}^d$ over the training data $\cD_\iclient$ owned by client $\iclient \in [\Mx]\eqdef \{1,\dots,\Mx\}$.

\subsection{Federated averaging}

Proposed in \citep{mcmahan2017communication}, federated averaging (\algname{FedAvg}) is an immensely popular method   specifically  designed to  solve problem \myref{eq:main} while being mindful of several constraints characteristic of practical federated environments. In particular, \algname{FedAvg} is based on gradient descent (\algname{\gls{GD}}), but introduces three modifications: a) \gls{PP}, a.k.a.\ partial  participation,   b) data sampling (\gls{DS}),  a.k.a.\ mini-batching, and c) local training (\gls{LT}). 

Training via \algname{FedAvg} proceeds in a number of communication rounds. Each round $t$ starts with the selection of a subset/cohort  $\set \subseteq [\Mx]$ of the clients of size $\Cx^t = |\set|$; these will  participate in the training in this round. The aggregating server then broadcasts the current version of the model, $x^t$,  to all clients $\iclient \in \set$ in the current cohort. Subsequently, each  client $\iclient \in \set$ performs $\Kx$ iterations  of    \algname{\gls{SGD}} on its local loss function $f_\iclient$,  initiated with $x^t$, using minibatches $\cB_\iclient^{k,t}\subseteq \cD_\iclient$ of size $b_\iclient=|\cB_\iclient^{k,t}|$  for $k=0,\dots,\Kx-1$. Finally, all participating devices send their updated models to the server for aggregation into a new model $x^{t+1}$, and the process is repeated. 

All three modifications can be turned on or off, individually, or in any combination. For example, if we set $\Cx^t=M$ for all $t$, then {\em all} clients are participating in all rounds, i.e., \gls{PP} is turned off. 
Further, if we set $b_\iclient=|\cD_\iclient|$ for each client $\iclient \in [M]$, then all clients  use {\em all} their data to compute the local gradient estimator needed to perform each \algname{\gls{SGD}} step, i.e., \gls{DS} is turned off.  
Finally, if we set $\Kx=1$, then only a {\em single} \algname{\gls{SGD}} step is taken by each participating client, i.e., \gls{LT} is turned off. 
If all of these modifications are turned off, \algname{FedAvg} reduces to vanilla \algname{\gls{GD}}.

\subsection{Partial participation and data sampling}
While work \citep{mcmahan2017communication} provided convincing empirical evidence for the efficacy of \algname{FedAvg}, their work did not contain any theoretical results. Much progress in \gls{FL} in the last five years can be attributed to the efforts by the \gls{FL} community to understand, analyze, and improve upon these mechanisms, often first in isolation, as this is easier when deep understanding is desired. 

Since {\em unbiased} partial participation and data sampling mechanisms  are  intimately linked to the stochastic approximation literature dating back to the work \citep{robbins1951stochastic}, it is not surprising that \gls{PP} and \gls{DS} are relatively well understood. For example,  variants of \algname{\gls{SGD}} supporting  virtually arbitrary unbiased \gls{PP} and \gls{DS} mechanisms have been analyzed in \citep{gower2019sgd}  in the smooth strongly convex regime and in \citep{khaled2023better,OptClientSampling2020} in the smooth nonconvex regime. Oracle optimal\footnote{See also the earlier work \citep{nonconvex_arbitrary}, who analyzed arbitrary sampling mechanisms in the smooth nonconvex regime with suboptimal variance-reduced methods.} (in the smooth  nonconvex regime) variants of \algname{\gls{SGD}} supporting  virtually arbitrary unbiased \gls{PP} and \gls{DS} mechanisms were proposed and analyzed in \citep{PAGE-AS}, who built upon the previous works \citep{li2021page,fang2018spider} and \citep{nguyen2017sarah}.  

However, all the works mentioned above  analyze  \algname{\gls{GD}} + \gls{PP}/\gls{DS} only, with \gls{LT} turned off. If \gls{LT} is included in the mix as well, or even considered in isolation as a single add-on to vanilla \algname{\gls{GD}}, significant technical issues arise. These issues have kept the \gls{FL} community uneasy and therefore busy and immensely productive for many years. Since, as we shall see, this will be of crucial importance for us to motivate the contributions of this paper, we will now   outline the  development of the theoretical understanding of the \gls{LT} mechanism by the \gls{FL} community over the last seven years. 

\subsection{Local training}
Local training---the practice of requiring each participating client to perform {\em multiple}  local optimization steps (as opposed to performing a {\em single} step only) based on their local data before communication-expensive parameter synchronization is allowed to take place---is one of the most practically useful algorithmic ingredients in the training of \gls{FL} models. In fact, \gls{LT} is so central to the practical success of \gls{FL}, and so unique and novel within the trio (\gls{PP}, \gls{DS} and \gls{LT}) of techniques forming the \algname{FedAvg} method, that many authors attach the prefix ``Fed'' (meaning ``federated'') to any optimization method performing some version of \gls{LT}, whether \gls{PP} and \gls{DS} are present as well or not.

While \gls{LT} was popularized in \citep{mcmahan2017communication}, it was proposed in the same form before  \citep{Povey2015,SparkNet2016},  also without any theoretical justification\footnote{However, the even earlier and closely related line of work on the \algname{CoCoA} framework,  which is based on solving the dual problem using arbitrary local solvers, comes with solid theoretical justification \citep{cocoa,COCOA+,COCOA+journal}. Finally, we would be remiss if we did not mention that another related method was proposed and studied more than 25 years ago in \citep{Mang1995}.}. 
However, until recently, the empirically observed and often very significant communication-saving potential of \gls{LT} remained elusive, escaping all attempts at a satisfying theoretical justification.  

\subsection{Five generations of local training methods}

We shall now briefly review the development of the theoretical understanding of \gls{LT}  in the smooth strongly convex regime. We  follow the classification proposed in \citep{ProxSkip-VR}, who identified five distinct generations of \gls{LT} methods---1) heuristic, 2) homogeneous, 3) sublinear, 4) linear, and 5) accelerated---each new one improving upon the previous one in a certain important way.  

{\bf  1${}^{\rm st}$ generation of \gls{LT} methods (heuristic).}  The 1${}^{\rm st}$ generation methods offer ample empirical evidence, but do not come with any convergence rates~\citep{Povey2015,SparkNet2016,mcmahan2017communication}. 

{\bf  2${}^{\rm nd}$ generation of \gls{LT} methods (homogeneous).}  The 2${}^{\rm nd}$ generation \gls{LT} methods do provide guarantees, but their analysis crucially depends on one or another of the many incarnations of  data homogeneity assumptions, such as i) bounded gradients, i.e., requiring $\| \nabla f_\iclient(x)\|\leq c$ for all $\iclient\in [\Mx]$ and $x\in \mathbb{R}^d$ \citep{FedAvg-nonIID}, or ii) bounded gradient dissimilarity (a.k.a.\ strong growth), i.e., requiring $\frac{1}{\Mx}\sum_{\iclient=1}^\Mx \| \nabla f_\iclient(x) \|^2\leq c\|\nabla f(x)\|^2$ for all  $x\in \mathbb{R}^d$ \citep{LocalDescent2019}. This is  problematic since such assumptions  are prohibitively restrictive; indeed, they are typically not satisfied in real \gls{FL} environments \citep{kairouz2019advances,FieldGuide2021}. 

{\bf  3${}^{\rm rd}$ generation of \gls{LT} methods (sublinear).}  The 3${}^{\rm rd}$ generation \gls{LT} theory managed to succeed in disposing of the problematic data homogeneity assumptions \citep{khaled2019first,localSGD-AISTATS2020}. Works \citep{woodworth2020minibatch} and \citep{glasgow2022sharp} subsequently provided lower bounds for \algname{LocalGD} with \gls{DS}, showing that its communication complexity is not better than that of minibatch \algname{\gls{SGD}} in the heterogeneous data setting.  Additionally, paper \citep{malinovskiy2020local} analyzed \gls{LT} methods for general fixed point problems. 

Unfortunately, these results suggest that \gls{LT}-enhanced \algname{\gls{GD}}, often called \algname{LocalGD}, suffers from a  sublinear convergence rate, which is clearly inferior to the linear convergence rate of vanilla \algname{\gls{GD}}. While removing the reliance on data homogeneity assumptions was clearly an important step forward, this rather pessimistic theoretical result seems to suggest that \gls{LT} makes \algname{\gls{GD}} worse. However, this is at odds with the empirical evidence, which maintains that \gls{LT} enhances \algname{\gls{GD}}, and often significantly so. For these reasons, theoreticians continued to soldier on, with the quest to at least close the theoretical gap between \gls{LT}-based methods and vanilla \algname{\gls{GD}}.

{\bf  4${}^{\rm th}$ generation of \gls{LT} methods (linear).} These efforts led to the identification of the {\em client drift} phenomenon as the culprit responsible for the  gap, and to a solution based on various techniques for the reduction of client drift. This development  marks the start of the  4${}^{\rm th}$  generation of \gls{LT} methods. The first\footnote{If we do not count the closely related works belonging to the \algname{CoCoA} framework~\citep{cocoa,COCOA+,COCOA+journal}.} method belonging to this generation, called \algname{Scaffold} \citep{karimireddy2020scaffold}, employs a \algname{SAGA}-like variance reduction technique \citep{defazio2014saga} to tame the client drift caused by \gls{LT}. As a result, \algname{Scaffold} has the same communication complexity as \algname{\gls{GD}}. 
Paper \citep{gorbunov2021local} subsequently proposed a unified framework for designing and analyzing 3${}^{\rm rd}$ and 4${}^{\rm th}$ generation methods in a single theorem, including new 4${}^{\rm th}$ generation  \gls{LT} methods such as \algname{S-Local-GD} and \algname{S-Local-SVRG}. Finally, work \citep{mitra2021linear} proposed the \algname{FedLin} method, which can be seen as a variant of one of the methods from paper \citep{gorbunov2021local} allowing for the clients to take a different number of local steps (without this leading to any theoretical benefit).

{\bf  5${}^{\rm th}$ generation of \gls{LT} methods (accelerated).}  In a recent breakthrough \citep{ProxSkip}, it was proved that a certain new and simple form of local training, embodied in their \gls{ProxSkip} method, leads to {\em provable communication acceleration} in the smooth strongly convex regime, even in the notoriously difficult heterogeneous data setting in which the client data $\{\cD_\iclient\}_{\iclient=1}^{\Mx}$ is allowed to be arbitrarily different. In particular, if each $f_\iclient$ is $L$-smooth and $\mu$-strongly convex, then \gls{ProxSkip} solves \myref{eq:main} in $\cO(\sqrt{\nicefrac{L}{\mu}} \log \nicefrac{1}{\varepsilon})$ communication rounds, which is a significant acceleration when compared with the $\cO(\nicefrac{L}{\mu} \log \nicefrac{1}{\varepsilon})$ complexity of \algname{GD}. 
According to work \citep{scaman2019optimal}, this accelerated communication complexity is optimal.
In \citep{ProxSkip} several extensions are provided of  their method. In particular,  \gls{ProxSkip} was enhanced with a very flexible  \gls{DS} mechanism which can capture virtually any form of (unbiased and non-variance-reduced) data sampling scheme\footnote{The \gls{ProxSkip} method in \citep{ProxSkip}  can incorporate all forms of \gls{DS} strategies captured by the {\em arbitrary sampling} approach \citep{gower2019sgd} which is enabled by their {\em expected smoothness} inequality.}. 
Motivated by this progress, several other methods belonging to the  5${}^{\rm th}$ generation of \gls{LT} methods were recently proposed. 

First, work \citep{ProxSkip-VR} extended the  \gls{ProxSkip} method via the inclusion of virtually arbitrary {\em variance-reduced} \algname{\gls{SGD}} methods~\citep{gorbunov2020unified} in lieu of simple \algname{\gls{SGD}}, including \gls{SVRG}~\citep{johnson2013accelerating,S2GD}, \algname{SAGA}~\citep{defazio2014saga}, \algname{JacSketch}~\citep{gower2021stochastic}, \gls{L-SVRG}~\citep{hofmann2015variance,kovalev2020don} or \gls{DIANA}~\citep{mishchenko2019distributed,DIANA2}. 

Second, in \citep{RandProx} it was observed that the Bernoulli-type randomness employed in the \gls{ProxSkip} method whose role is to avoid the computation of an expensive proximity operator  is a special case of a more general principle: the application of an unbiased compressor to the proximity operator, combined with a bespoke variance reduction mechanism to tame the variance introduced by the compressor. Work \citep{RandProx} further generalized the forward-backward setting used in \citep{ProxSkip} to more complex splitting schemes involving the sum of three operators (e.g., \algname{ADMM}~\citep{ADMM-Hestenes-1969,ADMM-Powel-1969} and \algname{PDDY}~\citep{DY,PDDY2020}), and besides analyzing the smooth strongly convex regime, provided results in the  convex regime as well. 

Finally, the work \citep{sadiev2022communication} pioneered an alternative approach, based on an \gls{LT}-friendly modification of the celebrated \algname{Chambolle-Pock} method \citep{CP2011}. In their \algname{APDA-Inexact} method, the accelerated communication complexity is preserved, but compared to \gls{ProxSkip}, the \# of gradient-type \gls{LT} steps in each communication round is improved from $\cO(\kappa^{1/2})$ to $\cO(\kappa^{1/3})$ and $\cO(\kappa^{1/4})$, where $\kappa=\nicefrac{L}{\mu}$ is the condition number. They further improve on some results of work \citep{ProxSkip} related to the decentralized regime where communication happens along the edges of a connected network.

\section{Contributions}

\begin{table*}[!t]
	\centering
	\caption{Comparison of all 5${}^{\rm th}$ generation local training (LT) methods. Our \gls{5GCS} method is the first that supports Partial Participation (\gls{PP}). Moreover, similarly to \algname{APDA-Inexact}, our theory allows for the LT solver to be chosen virtually arbitrarily.}
	\label{tbl:main}
	\begin{threeparttable}
		\begin{tabular}{ccccc}
&&&&  \\			
			{\bf 5${}^{\rm th}$ gen. LT Method}  &  \bf \shortstack{LT Solver}&    {\bf \shortstack{DS}}&  {\bf \shortstack{CS}} & \bf  \shortstack{Reference} \\			
			\hline
			\gls{ProxSkip} 
			& \algname{GD}, \algname{\gls{SGD}}	
			& \cmark \tnote{\color{blue}(a)}					
			& \xmark
			& [\citenum{ProxSkip}]\\ 
			\hline
			\gls{ProxSkip-VR} 
			& \algname{GD}, \algname{\gls{SGD}}, \algname{VR-SGD}			
			& \cmark \tnote{\color{blue}(b)}						
			& \xmark
			& [\citenum{ProxSkip-VR}]\\ 
			\hline
			\algname{APDA-Inexact} 
			& any				
			& \xmark
			& \xmark 
			& [\citenum{sadiev2022communication}]\\ 
			\hline
			\algname{RandProx} 
			& \algname{GD} 						
			& \xmark
			& \xmark 
			& [\citenum{RandProx}]\\ 
			\hline
			\gls{5GCS} 
			& any						
			& \cmark
			& \cmark 
			& this work \\ 
		\end{tabular}
		\begin{tablenotes}
			{\scriptsize
				\item [{\color{blue}(a)}]  Only supports non-variance reduced DS on clients.
				\item [{\color{blue}(b)}]  Supports non-variance reduced {\em and} variance-reduced DS on clients.
				}
		\end{tablenotes}
	\end{threeparttable}
\end{table*}

Now that the \gls{FL} community finally managed to show that (appropriately designed) \gls{LT} techniques, which as we have seen are key behind the success of modern federated optimization methods for solving \myref{eq:main}, 
lead to provable communication acceleration guarantees (in the smooth strongly convex regime), we adopt the stance that further algorithmic and theoretical progress in \gls{FL} should be focused on advancing the 5${}^{\rm th}$ generation of \gls{LT} methods. 

To the best of our knowledge, there are only a handful of papers providing methods and results that belong to this latest generation of \gls{LT} methods \citep{ProxSkip, ProxSkip-VR, sadiev2022communication, RandProx}. A close examination of these works reveals that much is yet to be discovered.

\subsection{The open problem we address in this work}

\begin{quote}\em \footnotesize The starting point of our work is the observation that  none of the 5${}^{\rm th}$ generation local training (\gls{LT}) methods support Partial Participation (\gls{PP}). In other words, it is not  known whether it is possible to design a method that would  enjoy communication acceleration via \gls{LT} and at the same time also support \gls{PP}.
\end{quote}

 The problem is harder than one may initially think. We have talked to several people about this, including the authors of the \gls{ProxSkip} method. It turns out that they have tried---``very hard'' in their own words---but their efforts did not bear any fruit. We have tried as well, and failed. The analysis of \gls{ProxSkip} is remarkably tight, and every adaptation towards supporting \gls{PP} seems to either lead to technical problems during the proof construction, or to a loss of communication acceleration. In fact, it is not even clear what a \gls{PP} variant of \gls{ProxSkip} should look like. Our attempts at guessing what such a method could look like failed as well, and the variants we brainstormed diverged in our numerical experiments as soon as \gls{PP} was enabled. 
 
Fortunately, it turns out that these negative results were helpful to us after all. Indeed, they led us to the idea that we should try to develop an entirely different method; one that is not based on either \gls{ProxSkip} or \algname{APDA-Inexact}. Once we started to think outside the box created by our pre-conceived solution path, we eventually managed to succeed. 

\subsection{Contributions}

\begin{table*}[!t]
	\centering
	\footnotesize
	\caption{Variants of \gls{5GCS} (Algorithm~\ref{alg:5GCS}) depending on the choice of the \gls{LT} procedure  run by clients $\iclient \in \set$ in the current cohort. $\Mx = $ number of clients; $\Cx = $ cohort size. }
	\label{tbl:variants}
	\begin{threeparttable}
		\begin{tabular}{cccc}
			{\bf Algorithm}  &  \bf \shortstack{\gls{LT} via Subroutine $\cA$} & \bf  \shortstack{Comm. Complexity}  &   \bf  \shortstack{Theorem} \\			
			\hline
			\algname{5GCS${}_\infty$} \tnote{\color{blue}(a)} 
			& $K=\infty$ steps of \algname{GD}				& $\cO\left(\left(\frac{\Mx}{\Cx}+\sqrt{\frac{\Mx}{\Cx}\frac{L}{\mu }}\right)\log \frac{1}{\varepsilon}\right)$
			&  \ref{thm:5GCS-infty} \\				
			\hline
			\algname{5GCS${}_K$} 
			& $K=\cO(\sqrt{\frac{\Cx}{\Mx}\frac{L}{\mu}})$ steps of \algname{GD}				& $\cO\left(\left(\frac{\Mx}{\Cx}+\sqrt{\frac{\Mx}{\Cx}\frac{L}{\mu }}\right)\log \frac{1}{\varepsilon}\right)$
			&  \ref{thm:5GCS} \\
			\hline
			\algname{5GCS${}_0$} \tnote{\color{blue}(b)} 
			& $K=0$ steps of \algname{GD}						
			& $ \cO\left(\frac{\Mx}{\Cx}\frac{L}{\mu} \log \frac{1}{\varepsilon}\right)$ \tnote{\color{blue}(c)}	
			& \ref{thm:5GCS-0} 	
			\\			
			\hline
			\algname{5GCS${}_\cA$} 
			& any method $\cA$ (Assumption~\ref{ass:GTPS})			& $\cO\left(\left(\frac{\Mx}{\Cx}+\sqrt{\frac{\Mx}{\Cx}\frac{L}{\mu }}\right)\log \frac{1}{\varepsilon}\right)$
			&  \ref{thm:INEXACTPPanyM}					
		\end{tabular}
		\begin{tablenotes}
			{\scriptsize
				\item [{\color{blue}(a)}] This method can be found in the appendix as Algorithm~\ref{alg:5GCS-infty}.
				\item [{\color{blue}(b)}] This method can be found in the appendix as Algorithm~\ref{alg:5GCS-0}.				
				\item [{\color{blue}(c)}]  
			Does not have accelerated communication complexity. Indeed, the communication complexity is $\cO\left(\nicefrac{L}{\mu} \log \nicefrac{1}{\varepsilon}\right)$ instead of $\cO\left(\sqrt{\nicefrac{L}{\mu}} \log \nicefrac{1}{\varepsilon}\right)$ in the $\Cx=\Mx$ regime.
			
				}
		\end{tablenotes}
	\end{threeparttable}
\end{table*}

We are now ready to outline the key insights and contributions of our work.  Our main idea is to start our development with the remarkable \algname{Point-SAGA} method \citep{defazio2016simple}. The key appealing property of this method is that it can solve \myref{eq:main} with an accelerated rate in the smooth strongly convex regime. However, \algname{Point-SAGA} has two critical drawbacks: 

\phantom{X} (i) In each communication round, \algname{Point-SAGA}  samples  a single client only, uniformly at random, which means it supports a very rudimentary and hence not practically interesting form of \gls{PP} only.

\phantom{X} (ii) \algname{Point-SAGA} requires a prox-oracle for each $f_\iclient$, where $\iclient$ is the active client, i.e., 
\[ 
\mathrm{prox}_{\frac{1}{\tau} f_\iclient} (x) \eqdef \arg \min \limits_{u\in \mathbb{R}^d} \left \{  f_\iclient(u) + \frac{\tau}{2}\|x-u\|^2 \right\}\]
for some $x\in \mathbb{R}^d$ and $\tau>0$ in each communication round, and evaluate it exactly. This is problematic, since exact evaluation of the proximity operator is rarely possible, and inexact evaluation (with a small error) may be overly expensive, imparting an excessive computational burden on the clients. 

Our main contributions  can be summarized as follows.

$\diamond$ We propose a new \gls{LT} method for \gls{FL}, which we call \gls{5GCS} (Algorithm~\ref{alg:5GCS}), which achieves accelerated communication complexity, and also supports Partial Participation. To the best of our knowledge, this is the first 5${}^{\rm th}$ generation \gls{LT} method which works with Partial Participation (see Table~\ref{tbl:main}). Moreover, the communication complexity of \gls{5GCS} is optimal \citep{NIPS2016_645098b0}.

$\diamond$ Our method supports arbitrary \gls{LT} subroutines as long as they satisfy a certain technical assumption (Assumption~\ref{ass:GTPS}). See Table~\ref{tbl:variants} for a list of four variants of \gls{5GCS} depending on what \gls{LT} subroutine is applied, and the associated communication complexities.

$\diamond$ When an infinity of \algname{\gls{GD}} steps is used as the \gls{LT} subroutine, our method  \gls{5GCS} in each communication round evaluates the prox of  $f_\iclient$ for all clients $\iclient$ in the cohort, and  reduces to  a minibatch version of \algname{PointSAGA}, which is  new\footnote{There is one exception: this method was recently analyzed in \citep{RandProx}.}. While this method enjoys accelerated communication complexity, its reliance on a prox oracle puts a heavy computation burden on the clients. On the other hand, when zero \algname{\gls{GD}} steps are used as a subroutine, our method achieves linear but nonaccelerated communication complexity only. Fortunately, it is sufficient to apply a relatively small number of \algname{\gls{GD}} steps as the \gls{LT} subroutine while preserving the accelerated communication complexity of minibatch \algname{PointSAGA}.

$\diamond$ Several further contributions are mentioned in the remaining text.

\section{Main Results}

\begin{algorithm*}[!t]
	\caption{\gls{5GCS}}
	\footnotesize
	\begin{algorithmic}[1]\label{alg:5GCS}
		\STATE  \textbf{Input:} initial primal iterates $x^0\in\mathbb{R}^d$; initial dual iterates $u_1^0, \dots,u_{\Mx}^0 \in\mathbb{R}^d$; primal stepsize $\gammaM>0$; dual stepsize $\tauM>0$; cohort size $\Cx\in \{1,\dots,\Mx\}$
		\STATE  \textbf{Initialization:}  $v^0\eqdef \sum_{\iclient=1}^\Mx u_\iclient^0$  \hfill {\color{gray} \footnotesize $\diamond$ The server initiates $v^{0}$ as the sum of the initial dual iterates} 
		\FOR{communication round $t=0, 1, \ldots$} 
		\STATE Choose a cohort $\set\subset \{1,\ldots,\Mx\}$ of clients of cardinality $\Cx$, uniformly at random \hfill {\color{gray} \footnotesize $\diamond$  \gls{PP} step} 
		\STATE Compute $\hat{x}^{t} = \frac{1}{1+\gammaM\mu} \left(x^t - \gammaM v^t\right)$ and broadcast it to the clients  in the cohort 
		\FOR{$\iclient\in \set$}
		\STATE Find $\lastlocittermk$ as the final point after $\Kx$ iterations of some local optimization algorithm $\mathcal{A}_\iclient$, initiated with $y_\iclient^0=\hat{x}^t$, for solving the optimization problem \hfill {\color{gray} \footnotesize $\diamond$ Client $\iclient$ performs $\Kx$ \gls{LT} steps} 
		\begin{eqnarray}
			\squeeze 
			\lastlocittermk \approx \argmin \limits_{y\in\mathbb{R}^d}\left\{\localfuni(y) \eqdef  F_\iclient(y)+\frac{\tauM}{2} \sqnorm{y-\left(\hat{x}^\kstep+\frac{1}{\tauM}u_\iclient^t\right)}\right\}\label{localprob}
		\end{eqnarray}
		\STATE Compute $u_\iclient^{t+1}=\nabla F_\iclient(\lastlocittermk)$ and send it to the server \hfill {\color{gray} \footnotesize $\diamond$ Client $\iclient$ updates its dual iterate} 
		\ENDFOR
		\FOR{$\iclient\in\{1, \ldots,\Mx\}\backslash \set$}
		\STATE $u_{\iclient}^{t+1}\eqdef u_{\iclient}^t $ \hfill {\color{gray} \footnotesize $\diamond$ Non-participating clients do nothing} 
		\ENDFOR
		\STATE $v^{t+1} \eqdef \sum_{\iclient=1}^{\Mx} u_\iclient^{t+1} $ \hfill {\color{gray} \footnotesize $\diamond$ The server maintains $v^{t+1}$ as the sum of the dual iterates} 
		\STATE  $x^{t+1} \eqdef \hat{x}^{t}- \gammaM \frac{\Mx}{\Cx} (v^{t+1}-v^t)$ \hfill {\color{gray} \footnotesize $\diamond$ The server updates the primal iterate} 
		\ENDFOR
	\end{algorithmic}
\end{algorithm*}
In this section we describe our new method, \gls{5GCS} (Algorithm~\ref{alg:5GCS}) for solving \myref{eq:main}, and formulate our main convergence results (see Table~\ref{tbl:variants} for a summary).

\subsection{Convexity and smoothness}
In our analysis  we focus on the regime when each $f_\iclient$ is $L$-smooth and $\mu$-strongly convex, which are standard assumptions in the convex optimization literature\footnote{While many practical \gls{FL} models  involve neural networks which lead to nonconvex problems instead, in our work we focus on resolving a certain key open problem in the foundations of \gls{FL} for which there is no answer even in the regime we consider. }. 

\begin{assumption}\label{ass:main} The functions $f_\iclient$ are $L$-smooth and $\mu$-strongly convex for all $\iclient \in \{1,\dots,\Mx\}$.
\end{assumption}

We shall use this assumption in what follows without explicitly mentioning this.  Recall that a continuously differentiable function $\phi :\mathbb{R}^{d}\to\mathbb{R}$ is $L$-smooth if
$\phi(x)-\phi(y)-\langle\nabla \phi(y),x-y\rangle \leq \frac{L}{2}\|x-y\|^2$ for all $x,y\in \mathbb{R}^d$, 
and $\mu$-strongly convex if
$\phi(x)-\phi(y)-\langle\nabla \phi(y),x-y\rangle \geq \frac{\mu}{2}\|x-y\|^2$ for all $x,y\in \mathbb{R}^d$.

\subsection{Problem reformulation and its dual}\label{sec:H}

Our method applies to a certain reformulation of \myref{eq:main} which we shall now describe. Let $\Koper\colon\mathbb{R}^d \to \mathbb{R}^{\Mx d}$ be the linear operator which maps $x\in \mathbb{R}^{d}$ into the vector $(x,\dots,x)\in \mathbb{R}^{\Mx d}$ consisting of $\Mx$ copies of $x$. First, notice that $F_\iclient(x)\eqdef \frac{1}{\Mx}(f_\iclient(x)-\frac{\mu}{2}\sqnorm{x})$ is convex and $L_F$-smooth with  $L_F \eqdef \frac{1}{\Mx}(L-\mu)$. Further,  define $F:\mathbb{R}^{\Mx d}\to \mathbb{R}$ via $F(x_1, \dots,x_\Mx) \eqdef \sum_{\iclient=1}^{\Mx} F_\iclient(x_\iclient).$ 

Having established the necessary notation, we consider the following reformulation of problem~\myref{eq:main}: 
\begin{equation}
	\squeeze 
	x^\star = \argmin \limits_{x\in\mathbb{R}^d}  \left[f(x) \eqdef F(\Koper x) + \frac{\mu}{2}\sqnorm{x}\right].\label{eq:main-new}
\end{equation}
It is straightforward to see that $f$ from \myref{eq:main} and \myref{eq:main-new} are identical functions. The  dual problem to \myref{eq:main-new} is
\begin{align*}\label{dualnew} 
	u^\star = \argmax\limits_{u\in\mathbb{R}^{\Mx d}} \, \left( \frac{1}{2\mu}\sqnorm{\sum \limits_{\iclient=1}^{\Mx} u_\iclient}+\sum \limits_{\iclient=1}^{\Mx} F_\iclient^*(u_\iclient)\right),\notag
\end{align*}
where $F_\iclient^*$ is the Fenchel conjugate of $F_\iclient$, defined by $F_\iclient^*(y)\eqdef\sup_{x\in\mathbb{R}^{d}}\{\langle x,y \rangle-F_\iclient(x) \}.$
Under Assumption~\ref{ass:main}, the primal and dual problems have unique optimal solutions $x^\star$ and $u^\star$, respectively.

\subsection{The 5GCS Algorithm}

Our proposed algorithm, \gls{5GCS}, is formalized as Algorithm~\ref{alg:5GCS}. The method produces a sequence of primal iterates $x^t$, and a sequence of dual iterates
$u^t = (u_1^t, \dots, u_\Mx^t)$. We have added several comments explaining the steps, and believe that the method should be easy to parse without additional commentary.  
 In each communication round $t$, the participating clients $\iclient \in \set$ in parallel perform \gls{LT} via $K$ steps of \algname{\gls{GD}} applied to minimizing the function $\localfuni$; see  \myref{localprob}. Below we outline four special variants of \gls{5GCS}, depending on the choice of the \gls{LT} subroutines $\{\cA_\iclient\}_{\iclient=1}^{\Mx}$.

\subsection{LT subroutine: GD with \texorpdfstring{$\Kx = +\infty$}{kx = +infty} steps (i.e., prox)}

The choice $\Kx = +\infty$ corresponds to exact minimization of the function $\localfuni$ defined in \myref{localprob}, i.e., to the evaluation of the prox operator of $F_\iclient$ for all $\iclient \in \set$. In this case, \gls{5GCS} reduces to \algname{Minibatch-Point-SAGA} (see  Algorithm~\ref{alg:5GCS-infty}), and its convergence properties are  described by the next result.

\begin{theorem}\label{thm:5GCS-infty} Consider Algorithm~\ref{alg:5GCS} (\gls{5GCS}) with the \gls{LT} solver being \algname{\gls{GD}} run for $K=+\infty$ iterations (this is equivalent to Algorithm~\ref{alg:5GCS-infty}; we shall also call the method \algname{5GCS${}_\infty$}).
	Let $\gammaM>0$, $\tauM>0$ and $\gammaM \tauM \leq \frac{1}{\Mx}$. Then for the Lyapunov function
	\begin{equation*}
	\squeeze	
	\Psi^{(\kstep)}\eqdef  \frac{1}{\gammaM}\sqnorm{x^{\kstep}-x^\star}+\frac{\Mx}{\Cx}\left(\frac{1}{\tauM}+2\frac{1}{L_F}\right)\sqnorm{u^{\kstep}-u^\star},
	\end{equation*}
 the iterates of the method satisfy
	$$
		\Exp{\Psi^{(\Tx)}}\leq (1-\rho)^\Tx \Psi^{(0)},
$$
	where 
$ 	\rho \eqdef  \min\left(\frac{\gammaM\mu}{1+\gammaM\mu}, \frac{\Cx}{\Mx}\frac{2\tauM }{L_F+2\tauM }\right)<1.
$

\end{theorem}

The following corollary gives a bound on the number of communication rounds needed to solve the problem.
\begin{corollary}\label{cor:5GCS-infty}
Choose any $0<\varepsilon<1$. If
	we choose $\gammaM=\sqrt{\frac{2\Cx}{L_F\mu \Mx^2}}$ and $\tauM=\sqrt{\frac{L_F\mu}{2\Cx}}$, then  in order to guarantee $\Exp{\Psi^{(\Tx)}}\leq \varepsilon \Psi^{(0)}$, it suffices to take
\[
	\squeeze	
	\Tx \geq	\left(\frac{\Mx}{\Cx}+\sqrt{\frac{\Mx}{ \Cx} \frac{L-\mu}{2\mu}}\right)\log \frac{1}{\varepsilon} =\tilde{\cO}\left(\frac{\Mx}{\Cx}+\sqrt{\frac{\Mx}{\Cx}\frac{L}{\mu}}\right)
\]
	communication rounds.
\end{corollary}

Note that the communication complexity improves as the cohort size $\Cx$ increases, and becomes $\tilde{\cO}(\sqrt{\nicefrac{L}{\mu }}) $ for $\Cx=\Mx$. This recovers the accelerated communication complexity of existing 5${}^{\rm th}$ generation local training methods \gls{ProxSkip}, \gls{ProxSkip-VR}  and \algname{APDA-Inexact} in the regime when \algname{\gls{GD}} is used as the \gls{LT} method. However, unlike these methods, \algname{5GCS${}_\infty$} supports Partial Participation (\gls{PP}). In the opposite extreme, i.e., when the cohort size is minimal ($\Cx=1$), the communication complexity of \algname{5GCS${}_\infty$} becomes $ \tilde{\cO}(\Mx+\sqrt{\nicefrac{\Mx L}{\mu}})$. If $\nicefrac{L}{\mu}\leq \Mx$, which will typically be the case in \gls{FL} settings with a very large number of clients (e.g., cross-device \gls{FL}), the complexity simplifies to $ \tilde{\cO}(\Mx)$, which says that we need as many communication rounds as there are clients, which makes sense, since we do not assume any form of data homogeneity, and this means that all clients may contain valuable data. In general, as the cohort size $\Cx$ increases, the communication complexity improves, and interpolates between these two extreme cases.

\begin{figure}[!t]
	\centering
	\includegraphics[width=0.99\linewidth]{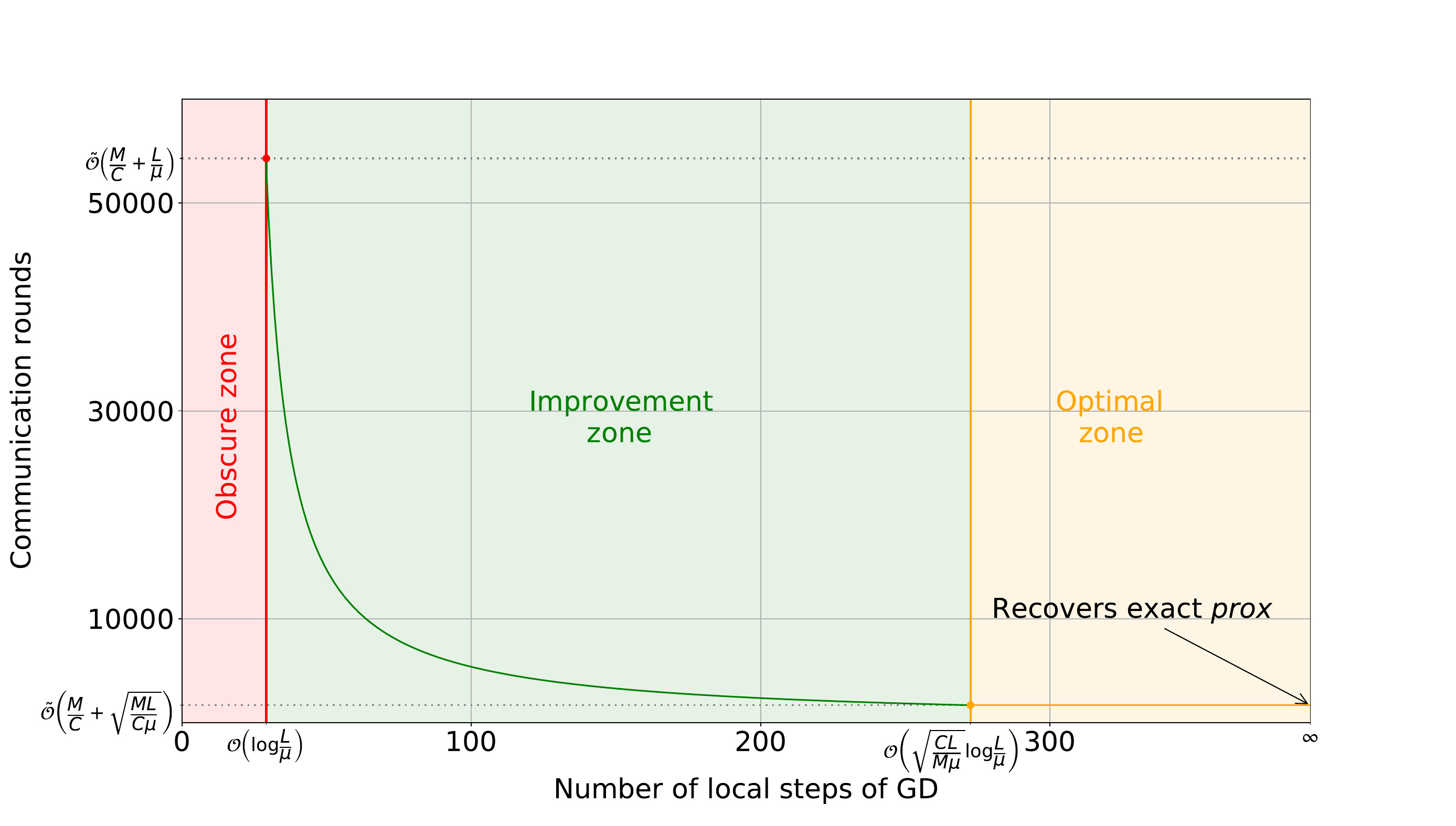}
	\caption{The number of communication rounds of \gls{5GCS} as a function of the number of \gls{GD} steps forming the \gls{LT} subroutine $\cA$ with $\nicefrac{L}{\mu}=10^4$ and $\nicefrac{\Cx}{\Mx}=0.1$. The key observation is that it is enough to choose $\Kx = \cO(\sqrt{\frac{\Mx}{\Cx}\frac{L}{\mu}})$, which is at the left end-point of the ``optimal zone''. More steps do {\em not} lead to better communication complexity. }
	\label{ris:image}	
\end{figure}

\subsection{LT subroutine: GD with \texorpdfstring{\ensuremath{\Kx = \cO\left(\sqrt{\frac{\Cx}{\Mx}\frac{L}{\mu}}\right)}}{kx = o(sqrt{cx/mx l/mu})} steps}

The key drawback of \algname{5GCS${}_\infty$} is that the \gls{LT} subroutine needs to take an infinite number of \algname{\gls{GD}} steps, or equivalently, the method requires the exact evaluation of the prox of $F_\iclient$.  We now show that it is possible to obtain the same accelerated communication complexity as in the $\Kx=+\infty$ case with a finite, and in fact surprisingly small, number of \algname{\gls{GD}} iterations.
 
\begin{theorem}\label{thm:5GCS} Consider Algorithm~\ref{alg:5GCS} (\gls{5GCS}) with the \gls{LT} solver being \algname{\gls{GD}} run for $\Kx \geq \left(\frac{3}{4}\sqrt{\frac{\Cx}{\Mx}\frac{L}{\mu}}+2\right)\log\left(4\frac{L}{\mu}\right)$ iterations.
Let $0<\gammaM\leq \frac{3}{16}\sqrt{\frac{\Cx}{L\mu \Mx}}$ and $\tauM=\frac{1}{2\gammaM \Mx}$. Then for the Lyapunov function
	\begin{equation*}
		\squeeze
		 	\Psi^{(\kstep)}\eqdef \frac{1}{\gammaM}\sqnorm{x^{\kstep}-x^\star}+\frac{\Mx}{\Cx}\left(\frac{1}{\tauM}+\frac{1}{L_F}\right)\sqnorm{u^{\kstep}-u^\star},
	\end{equation*}
 the iterates of  the method satisfy
$$		\Exp{\Psi^{(\Tx)}}\leq (1-\rho)^\Tx \Psi^{(0)},
$$	where 
$
	\rho\eqdef  \min\left\{\frac{\gammaM\mu}{1+\gammaM\mu},\frac{\Cx}{\Mx}\frac{\tauM}{(L_F+\tauM)}\right\}<1.
$

\end{theorem}

Note that \algname{\gls{GD}} needs to be run for $\Kx = \cO\left(\sqrt{\frac{\Cx}{\Mx}\frac{L}{\mu}}\right)$ local steps on each client in the cohort. This quantity depends on the square root of the condition number only, and is smaller for smaller cohort size $\Cx$. 

It turns out that this result can be improved using a finer analysis. In particular, we can show that some clients can get away with fewer \gls{LT} steps than this, provided that their local datasets are favorable\footnote{To the best of our knowledge, a result of this type does not exist in the \gls{FL} literature.}.  To see this, assume that each $f_\iclient$ is $L_\iclient$-smooth. Clearly, this implies that each $f_\iclient$ is $L$-smooth with $L=\max_{\iclient} L_\iclient$, and Theorem~\ref{thm:5GCS} holds with this $L$. However, recall that client $\iclient$ applies \algname{\gls{GD}}  to (approximately) minimize $\localfuni$ from \myref{localprob}, and this function happens to be $\left(\frac{1}{\Mx}\left(L_\iclient-\mu\right)+\tauM\right)$-smooth and $\tauM$-strongly convex.
It can be easily seen that $\tauM \geq \frac{8}{3}\sqrt{\frac{\mu L}{\Mx\Cx}}$, and hence the condition number of $\localfuni$ is
$\frac{1}{\Mx}\left(L_\iclient-\mu\right)\frac{1}{\tauM}+1 \leq \frac{3}{8}\sqrt{\frac{\Cx}{\Mx } \frac{\nicefrac{L_\iclient^2}{L}}{\mu}} +1
$. So,  \algname{\gls{GD}} only needs  $\Kx_\iclient = \cO\left( \sqrt{\frac{\Cx}{\Mx} \frac{\nicefrac{L_\iclient^2}{L} }{\mu}}\right)$ iterations on client $\iclient$, which can be much smaller than the worst-case bound $\Kx= \cO\left( \sqrt{\frac{\Cx}{\Mx} \frac{L}{\mu}}\right)$.

The following corollary gives a bound on the number of communication rounds needed to solve the problem.

\begin{corollary}\label{cor:5GCS}
	Choose any $0<\varepsilon<1$ and $\gammaM = \frac{3}{16}\sqrt{\frac{\Cx}{L\mu \Mx}}$. In order to guarantee $\Exp{\Psi^{(\Tx)}}\leq \varepsilon \Psi^{(0)}$, it suffices to take 
	\begin{eqnarray}
		\squeeze 
			T &\geq	& 
			\squeeze
			 \max\left\{1+\frac{16}{3}\sqrt{\frac{\Mx}{\Cx}\frac{L}{\mu}},\frac{\Mx}{\Cx}+\frac{3}{8}\sqrt{\frac{\Mx}{\Cx}\frac{L}{\mu} }\right\}\log\frac{1}{\varepsilon} \notag \\
		&=& \squeeze 
		\tilde{\cO}\left(\frac{\Mx}{\Cx}+\sqrt{\frac{\Mx}{\Cx}\frac{L}{\mu }}\right) \notag
	\end{eqnarray}
	communication rounds.
\end{corollary}

This is the same expression as that from Corollary~\ref{cor:5GCS-infty}, and hence the same comments we've made there apply here, too.

\subsection{LT subroutine: GD with \texorpdfstring{$\Kx = 0$}{kx = 0} steps}

\begin{theorem}\label{thm:5GCS-0}
Consider Algorithm~\ref{alg:5GCS} (\gls{5GCS}) with the \gls{LT} solver being \algname{\gls{GD}} run for $\Kx=0$ iterations (this is equivalent to Algorithm~\ref{alg:5GCS-0}; we shall also call the method \algname{5GCS${}_0$}).	Let $0<\gammaM\leq \frac{\Cx}{4L\Mx}$. Then for the Lyapunov function
	\begin{equation*}
		\squeeze 	\Psi^{(\kstep)}\eqdef  \frac{\Cx}{\Mx^2\gammaM^2}\left(1-\sqrt{\frac{\gammaM \Mx L_F}{2}}\right)\sqnorm{x^{\kstep}-x^\star}+\sqnorm{u^{\kstep}-u^\star},
	\end{equation*}
the iterates of the method satisfy
$$		\Exp{\Psi^{(\Tx)}}\leq \left(1-\rho\right)^\Tx \Psi^{(0)},
$$	
where 
$
	\rho\eqdef  \min\left(\frac{\gammaM\mu}{1+\gammaM\mu},\frac{\Cx}{\Mx+2\gammaM  L_F\Mx^2}\right)<1.
$

\end{theorem}

The following corollary gives a bound on the number of communication rounds needed to solve the problem.

\begin{corollary}\label{cor:5GCS-0}
	Choose any $0<\varepsilon<1$ and
	 $\gammaM=\frac{\Cx}{4 L \Mx}$. In order to guarantee $\Exp{\Psi^{(\Tx)}}\leq \varepsilon \Psi^{(0)}$, it suffices to take
\[
		\squeeze 	
		\Tx \geq 	\max\left\{1+\frac{4\Mx}{\Cx}\frac{L}{\mu},\frac{\Mx}{\Cx}+\frac{L_F\Mx}{L}\right\}\log \frac{1}{\varepsilon}= \tilde{O}\left(\frac{\Mx}{\Cx}\frac{L}{\mu}\right)
\]
	communication rounds.
\end{corollary}

In this case, we do {\em not} obtain communication acceleration. This is because \gls{LT}  with $\Kx=0$ is not extensive enough.

\subsection{LT subroutine: Any method \texorpdfstring{$\cA$}{a}}

Finally, we now show that \gls{5GCS} is not limited to exclusively using \algname{\gls{GD}} as the \gls{LT} solver. To the contrary, \gls{5GCS} works with any subroutine $\cA$ as long as it is possible to guarantee that, after a sufficiently large number $\Kx$ of iterations, a certain inequality  holds. 

\begin{assumption}\label{ass:GTPS}
	Let $\{\mathcal{A}_1,\dots,\mathcal{A}_\Mx\}$ be any \gls{LT} subroutines for minimizing functions $\{\psi_1^t, \dots, \psi_\Mx^t\}$ defined in \myref{localprob}, capable of finding points $\{y_{1}^{\Kx,\kstep},\dots,y_{\Mx}^{\Kx,\kstep}\}$ in $\Kx$ steps, from the starting point $y_{\iclient}^{0,\kstep}=\hat{x}^\kstep$ for all $\iclient \in \{1,\dots,\Mx\}$, which satisfy the inequality
	\begin{align*}
		\squeeze 	\sum\limits_{\iclient=1}^{\Mx}\frac{4}{\tauM^2}\frac{\mu L_F^2}{3\Mx}\sqnorm{\lastlocittermk-\localsolmk}&	\squeeze +\sum\limits_{\iclient=1}^{\Mx}\frac{L_F}{\tauM^2}\sqnorm{\nabla\localfuni(\lastlocittermk)}\\
		&	\squeeze  \leq\sum\limits_{\iclient=1}^{\Mx}\frac{\mu}{6\Mx}\sqnorm{\hat{x}^\kstep-\localsolmk},
	\end{align*}
	where $\localsolmk$ is the unique minimizer of $\localfuni$, and $\tauM\geq \frac{8}{3}\sqrt{\frac{L\mu}{\Mx \Cx}}.$
\end{assumption}

Our most general result follows:

\begin{theorem}\label{thm:INEXACTPPanyM}
Consider Algorithm~\ref{alg:5GCS} (\gls{5GCS}) with the \gls{LT} solvers $\{\cA_1,\dots,\cA_\Mx\}$ satisfying Assumption~\ref{ass:GTPS}. Let $0<\gammaM$ and $0<\tauM$ satisfy $\gammaM\leq\frac{1}{\tauM\Mx}\left(1-\frac{4\mu}{3\Mx\tauM}\right)$. 
Then for the Lyapunov function
	\begin{equation*}
		\squeeze 	
		\Psi^{(\kstep)}\eqdef \frac{1}{\gammaM}\sqnorm{x^{\kstep}-x^\star}+\frac{\Mx}{\Cx}\left(\frac{1}{\tauM}+\frac{1}{L_F}\right)\sqnorm{u^{\kstep}-u^\star}\label{PPINEXACTpsianyM},
	\end{equation*}
	the iterates of the method satisfy
$$
		\Exp{\Psi^{(\Tx)}}\leq (1-\rho)^\Tx \Psi^{(0)},
$$
	where 
$	\rho\eqdef  \min\left\{\frac{\gammaM\mu}{1+\gammaM\mu},\frac{\Cx}{\Mx}\frac{\tauM}{(L_F+\tauM)}\right\}<1.
$

\end{theorem}

Note that the convergence rate in this result is identical to the convergence rate from Theorem~\ref{thm:5GCS}. Therefore, the same conclusions apply here as well.

\subsection{Relation between the number of rounds \texorpdfstring{$\Tx$}{tx} and the number of local steps \texorpdfstring{$\Kx$}{kx}}
We now study the dependence of the \# of communication rounds $\Tx$ on the \# of local steps $\Kx$ used by \algname{\gls{GD}} as the \gls{LT} subroutine. We first show in Theorem~\ref{thm:relation2} that with merely  $\Kx = \mathcal{O}\left(\log \frac{L}{\mu}\right)$ local \algname{\gls{GD}} steps  we can improve the communication complexity from $\Tx = \tilde{O}\left(\frac{\Mx}{\Cx}\frac{L}{\mu}\right)$ (provided in Theorem~\ref{thm:5GCS-0}) to $\Tx=\tilde{O}\left(\frac{\Mx}{\Cx}+\frac{L}{\mu}\right)$. 
\begin{theorem}\label{thm:relation2}
	Consider Algorithm~\ref{alg:5GCS} (\gls{5GCS}) with the \gls{LT} solver being \algname{\gls{GD}}. Let $\gammaM=\frac{3}{16L}$  and $\tauM=\frac{8L}{3\Mx}.$
	With these stepsizes, if  \gls{LT} is performed via  
	\begin{eqnarray*}
		\squeeze		\Kx\geq \left(2+\frac{3\Mx L_F}{4L}\right)\log\left(4\frac{L}{\mu}\right)=\mathcal{O}\left(\log\frac{L}{\mu}\right)
	\end{eqnarray*}
	steps of \algname{\gls{GD}}, then  
	\begin{align*}
		\squeeze	T&\squeeze \geq\max\left\{1+\frac{16}{3}\frac{L}{\mu}, \frac{\Mx}{\Cx}+\frac{3\Mx}{8\Cx}\frac{\Mx L_F}{L}\right\}\log\frac{1}{\epsilon} =\tilde{\mathcal{O}}\left(\frac{\Mx}{\Cx}+\frac{L}{\mu}\right)
	\end{align*}
communication rounds suffice to find an $\varepsilon$-solution.	
\end{theorem}
In Theorem~\ref{thm:5GCS} we showed that an accelerated communication complexity can be achieved with merely $\Kx = \cO\left(\sqrt{\frac{\Cx}{\Mx}\frac{L}{\mu}}\log\frac{L}{\mu}\right)$ local \algname{\gls{GD}} steps. However, the behavior of $\Tx$  on the interval between $\Kx = \mathcal{O}\left(\log \frac{L}{\mu}\right)$ (studied in Theorem~\ref{thm:relation2}) and $\Kx = \cO\left(\sqrt{\frac{\Cx}{\Mx}\frac{L}{\mu}}\log\frac{L}{\mu}\right)$ was not studied there.  We shall do so now.
\begin{theorem}\label{thm:relation}
	Consider Algorithm~\ref{alg:5GCS} (\gls{5GCS}) with the \gls{LT} solver being \algname{\gls{GD}}, which we run for \begin{equation}\label{eq:h9y98fd8hfd}\squeeze \Kx \geq \Kx(\deltaM)\eqdef 2\deltaM \log\left(\frac{4L}{\mu}\right)\end{equation} iterations, where $\deltaM$ is any constant satisfying $$\squeeze 1<\deltaM< 1+\frac{3}{8}\sqrt{\frac{\Cx}{\Mx} \frac{L}{\mu}}.$$  Let $\gammaM=\frac{1}{2\Mx\tauM}$  and $\tauM=\max\left\{\frac{L}{\Mx (\deltaM-1)},\frac{8}{3}\sqrt{\frac{L\mu}{\Mx \Cx}}\right\}.$
	Then for the Lyapunov function
	\begin{equation*}
		\squeeze 	\Psi^{(\kstep)}\eqdef \frac{1}{\gammaM}\sqnorm{x^{\kstep}-x^\star}+\frac{\Mx}{\Cx}\left(\frac{1}{\tauM}+\frac{1}{L_F}\right)\sqnorm{u^{\kstep}-u^\star}\label{PPINEXACTpsianyM_},
	\end{equation*}
	the iterates of the method satisfy
	$$
	\Exp{\Psi^{(\Tx)}}\leq (1-\rho)^\Tx \Psi^{(0)},
	$$
	where 
	$	\rho\eqdef  \min\left\{\frac{\gammaM\mu}{1+\gammaM\mu},\frac{\Cx}{\Mx}\frac{\tauM}{(L_F+\tauM)}\right\}<1.
	$
\end{theorem}

\begin{corollary}\label{cor:rela}
	Choose any $0<\varepsilon<1$. In order to guarantee $\Exp{\Psi^{(\Tx)}}\leq \varepsilon \Psi^{(0)}$, it suffices to take 
	\begin{equation*}
	\squeeze	T\geq	\max\left\{1+\frac{2L}{\left(\deltaM-1\right)\mu},\frac{\Mx}{\Cx}\deltaM\right\}\log\frac{1}{\epsilon}.
	\end{equation*}
	Note that if $\deltaM\leq \frac{\Mx+\Cx}{2\Mx}+\sqrt{\frac{2L\Cx}{\mu \Mx}+\left(\frac{\Mx-\Cx}{2\Mx}\right)^2}$, then
	\begin{equation*}
		\squeeze 	\Tx \geq	T(\alpha)\eqdef \left(1+\frac{2}{\deltaM-1} \frac{L}{\mu}\right)\log \frac{1}{\varepsilon}.
	\end{equation*}
\end{corollary}

Theorem~\ref{thm:relation}
and Corollary~\ref{cor:rela} imply that as long as $K\geq  K(\alpha)$ and $T\geq T(\alpha)$, then ${\mathbb{E}}[\Psi^{(T)}] \leq \varepsilon \Psi^{(0)}$. By substituting  $\alpha = \frac{K(\alpha)}{2\log \frac{4L}{\mu}}$ (see \myref{eq:h9y98fd8hfd}) into the expression for $T(\alpha)$, we get $$\squeeze T(\alpha) = \left(1+ \frac{4\log \frac{4L}{\mu}}{K(\alpha)-2\log \frac{4L}{\mu}} \frac{L}{\mu}\right) \log \frac{1}{\varepsilon} = \cO\left(\frac{1}{K(\alpha)}\right) \log \frac{1}{\varepsilon} .$$  This inverse dependence of $T(\alpha)$ on $K(\alpha)$ can  be observed empirically; see Figure~\ref{ris:image} (right). 


\section{Experiments}

We consider $\ell_2$-regularized logistic regression,
\begin{equation*}
	\squeeze f(x) =	\frac{1}{\Mx n} \sum \limits_{m=1}^{\Mx}\sum \limits_{i=1}^{n}\log  \left(1+ e^{\left(-b_{m,i} a_{m,i}^{\top} x\right)}\right)+\frac{\lambda}{2}\|x\|^2,
\end{equation*}
where $a_{m,i}\in \mathbb{R}^d$ and $b_{m,i}\in \{-1, +1\}$ are the data samples and labels, $\Mx$ is the number of clients and $n$ is the number of data points per client.  Following the work \citep{ProxSkip-VR}, we set  $\lambda = 10^{-3}L$, where $L$ is as in  Assumption~\ref{ass:main}. We chose to highlight an experiment on the \texttt{a1a} dataset from the LibSVM library~\citep{chang2011libsvm}. All methods were implemented in Python using the RAY package to simulate parallelization.

\begin{figure*}[t]
	\centering
	\begin{tabular}{ccc}
		\includegraphics[width=0.38\linewidth]{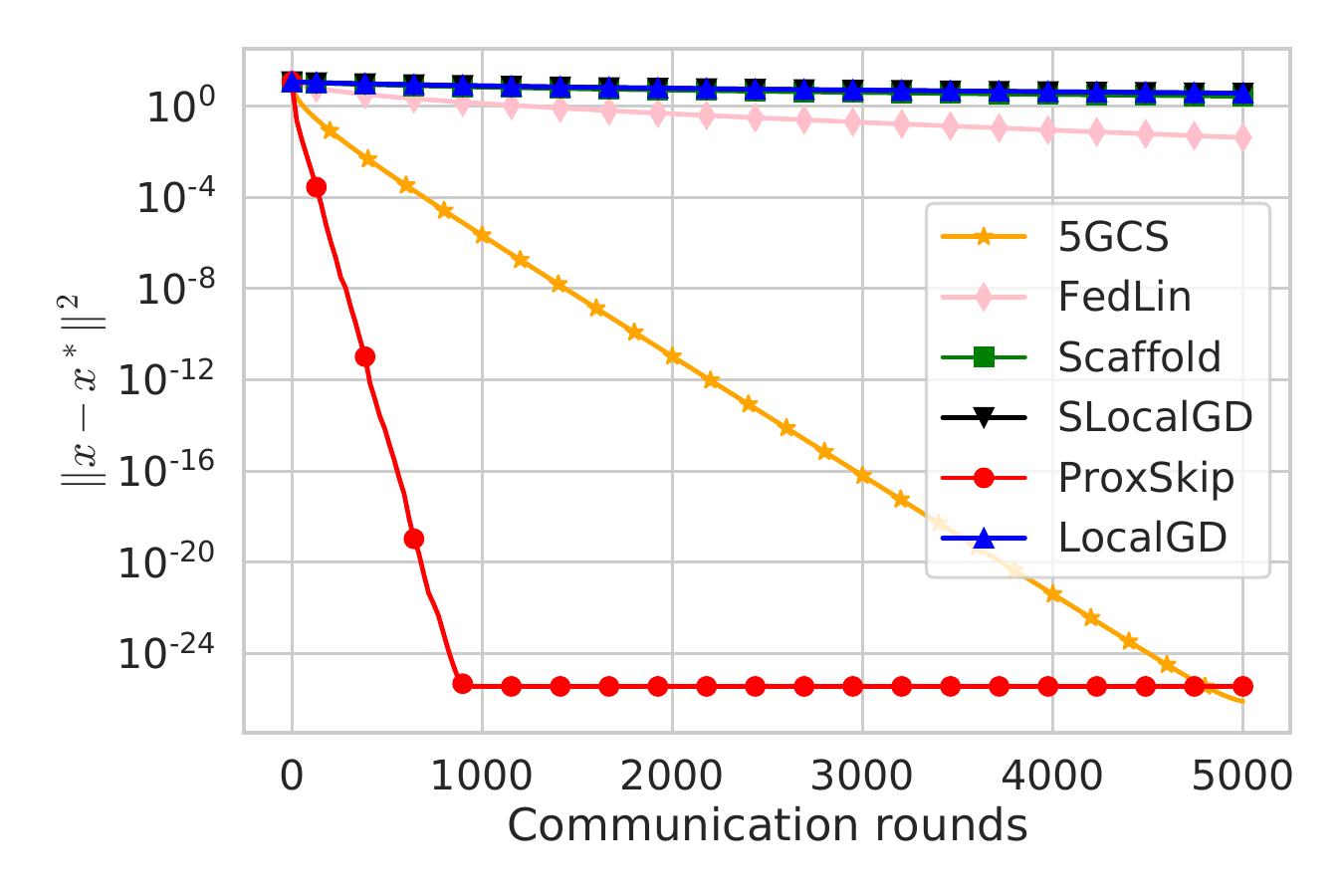}&	\includegraphics[width=0.38\linewidth]{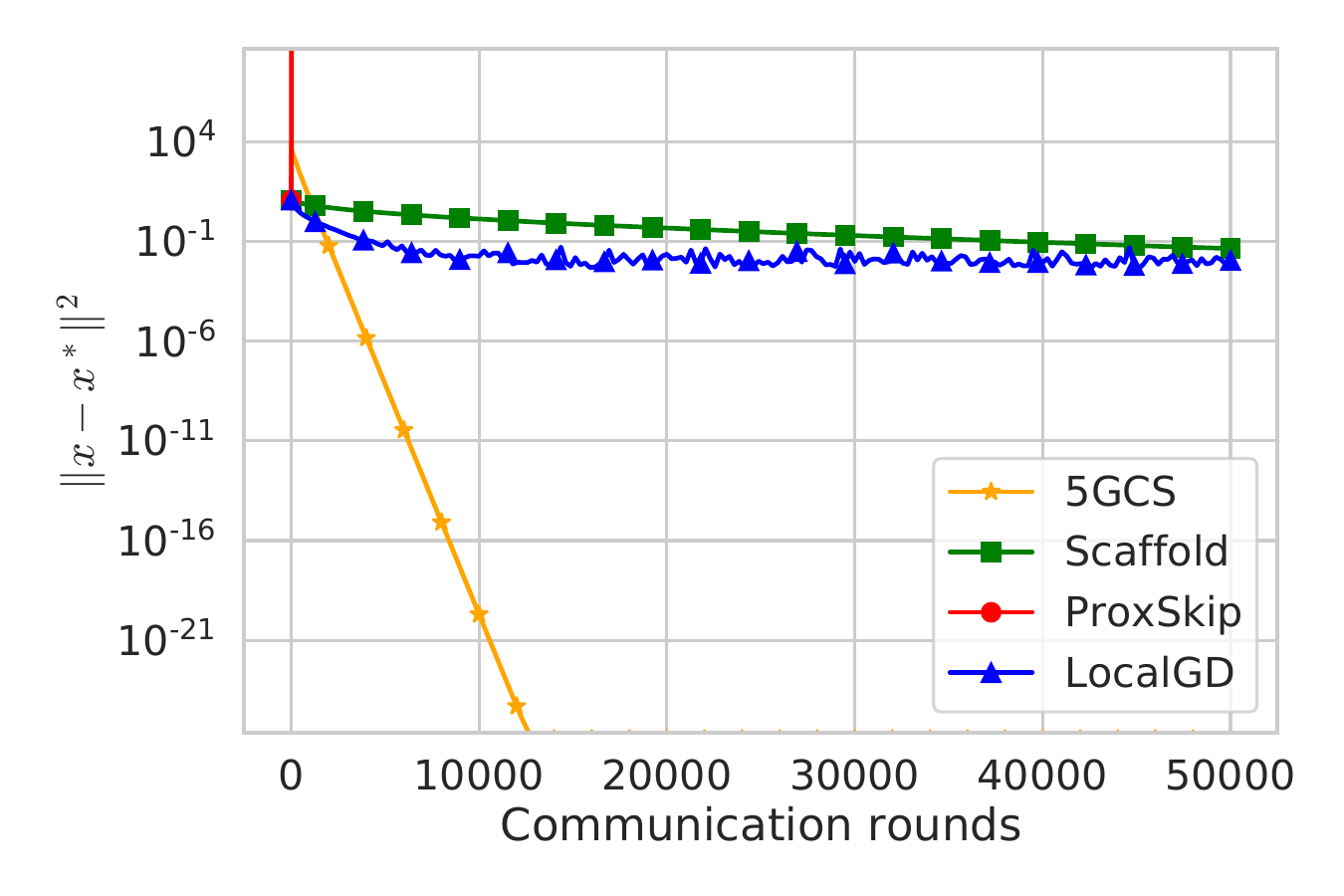}&

	\end{tabular}
	\centering
	\caption{Performance of our \gls{5GCS} method without (left) and with (middle) \gls{PP}.}
	\label{ris:image_2}
\end{figure*}

\subsection{Full participation (\texorpdfstring{$\Cx = \Mx$}{cx = mx})}

As a sanity check, we first perform an experiment in the full participation regime $\Cx = \Mx =5$, comparing our method \gls{5GCS}  with \algname{LocalGD} (3${}^{\rm rd}$ generation), \algname{Scaffold}, \algname{SLocalGD} and \algname{FedLin} (4${}^{\rm th}$ generation) and \gls{ProxSkip} (5${}^{\rm th}$ generation). We used theoretical stepsizes. For \gls{ProxSkip} we used the optimal communication probability parameter $p = \nicefrac{1}{\sqrt{\kappa}}$, where $\kappa=\nicefrac{L}{\mu}$. In the case of all 4${}^{\rm th}$ generation \gls{LT} methods and \algname{LocalGD}, the theoretical rate does {\em not} depend on the number of local steps $\Kx$. In our experiments we used the same number of local steps $\Kx = \nicefrac{1}{p} = \sqrt{\kappa}$ for all competing methods.  Figure~\ref{ris:image_2} (left) clearly shows that \gls{5GCS} has accelerated communication complexity, outperforming all 4${}^{\rm th}$ and 3${}^{\rm rd}$ generation \gls{LT} methods by a large margin. However, due to a small numerical constant for the stepsize in our theory $\left(\nicefrac{3}{16}\right)$, \gls{5GCS} converges more slowly than \gls{ProxSkip}, which shows excellent performance.

\subsection{Partial participation (\texorpdfstring{$\Cx < \Mx$}{cx < mx})}
Our key contribution is to bring Partial Participation (\gls{PP}) to the world of 5${}^{\rm th}$ generation \gls{LT} methods. Once \gls{PP} is required, \gls{ProxSkip} and \algname{APDA-Inexact} fall out of the competition as they do not support \gls{PP}. We therefore compare our method \gls{5GCS} with 4${}^{\rm th}$ and 3${}^{\rm rd}$ generation \gls{LT} methods supporting \gls{PP}: we have chosen \algname{Scaffold} and \algname{LocalGD}. We set $\Mx=15$ and $\Cx=3$ and used theoretical parameters. Figure~\ref{ris:image_2} (middle) shows that \gls{ProxSkip} diverges in the \gls{PP} regime, as expected. Moreover,  \gls{5GCS}  significantly outperforms the competing methods.


\chapter{Investigating Server-Side Stepsizes and Random Reshuffling in Federated Optimization}
\label{chapter5}
\thispagestyle{empty}

\section{Introduction} 

The unprecedented industrial success of modern Machine Learning techniques, tools and  models can to a large degree be attributed to the abundance of data available for training. Indeed, the most popular and best performing  deep learning models rely on a very large number of parameters, and in order to generalize well, need to be trained using optimization algorithms over very large training datasets. All other things being equal, the more data we have, the better. A key driving force behind the proliferation of such data is the massive digitization of society over the last few decades. People have access to increasingly more elaborate personal and home smart devices capable of generating, capturing and processing data such as text, images and videos. Similarly, in the sphere of governments and  corporations, much of what used to be done through a physical exchange (e.g., via paper/fax/letter) is now performed in a digital form, generating treasure troves of potentially useful data. For example, hospitals  collect, store and make use of a variety of patient data, ranging from routine bodily functions to PET scans and genome sequencing.

\subsection{Federated learning}

The traditional way of learning from this data is to collect it in a single (and often proprietary) data center, where it is subsequently  processed  using  modern Machine Learning algorithms. However, due to several considerations which keep gaining in importance, such as energy efficiency and privacy, it is often desirable to avoid centralized training altogether, and instead perform the training without the data ever leaving the clients' secure sites. Introduced in 2016 in the works \citep{FedOpt2016, FedLearn2016,  mcmahan2017communication}, this is precisely the promise and subject of study of {\em Federated Learning (\gls{FL})}.
In other words, Federated Learning means efficient Machine Learning over data stored in a distributed fashion across a network of heterogeneous clients (e.g., mobile phones, smart devices, companies) that capture and own the data, using these clients' machines/devices not only as data sources, but also as computers that  contribute to the training. 

\subsection{Problem formulation}

We consider the  standard optimization formulation of Federated Learning
\begin{equation}\label{eq:main_chapter_5}
\squeeze	\min \limits_{x \in \mathbb{R}^{d}} \left[f(x)\eqdef\frac{1}{M} \sum\limits_{m=1}^{M} f_{m}(x)\right],
\end{equation}
where $M$ is the total number of clients, $x\in \mathbb{R}^d$ represents the parameters of the model we wish to train, and $f_m:\R^d \to \R$ is the loss of model $x$ on the training data owned by client $m \in [M]\eqdef \{1,2,\dots,M\}$. Typically, $M$ is very large. 

Since the training dataset on each client is necessarily finite, we assume that $f_m$ has the finite-sum structure
\begin{equation}\label{eq:local_sum_ch5}
\squeeze f_m(x) \eqdef \frac{1}{n}\sum\limits_{i=1}^{n}f_{m,i}(x),
\end{equation}
where $f_{m,i}:\R^d \to \R$ is the loss of model $x$ on training example $i \in [n]\eqdef \{1,2,\dots,n\}$ stored on client $m$. We assume that the functions $f_{m,i}$ are differentiable, and consider the strongly convex, convex and non-convex regimes. 

\subsection{Ingredients of successful federated learning methods}

Practical considerations of Federated Learning systems and vast experimental evidence accrued over the last few years point to several design constraints and algorithmic ingredients which have proved useful in the context of Federated Learning methods for solving \eqref{eq:main_chapter_5}-\eqref{eq:local_sum_ch5}. We now very briefly outline some of them. More details can be found in the appendix where we review related work.

{\bf Partial participation.} In Federated Learning, training is performed through several communication rounds in each of which an orchestrating server chooses a {\em cohort} of clients that will be participating in the training process in that round. This practice is known as {\em partial participation}, and is necessary due to practical considerations and limitations, such as limited server capacity, and limited client availability \citep{kairouz2019advances}. However, partial participation can be useful also due to the diminishing returns one gets as the number of participating clients grows \citep{Cohort2021}.
Partial participation is a necessity  in the cross-device regime where the training is  performed over a very large number of clients (i.e., $M$ is very large) most  of which will only participate in the entire training procedure at most once. Partial participation of clients to form a cohort can be done adaptively so as to choose the most informative  clients~\citep{OptClientSampling2020}.

{\bf Local training.} At the beginning of each communication round, each client in the cohort is provided with the latest model by the orchestrating server, which is used as a starting point for {\em local training}. Local training refers to the common practice in \gls{FL} of  performing several steps of a suitably chosen local optimization procedure, such as one of the many variants of \algname{\gls{SGD}}, using the client's own local training data. Perhaps the simplest approach is to perform a single local \algname{\gls{GD}} iteration. If the model updates are simply aggregated by the server, then the resulting method can be seen as \algname{Minibatch \gls{SGD}}, where the minibatches correspond to the cohorts. However, it is typically more efficient to perform {\em multiple} local steps \citep{mcmahan2017communication}, and to use local optimizers that rely on {\em incremental}  data processing, such as \algname{\gls{SGD}}.

{\bf Data shuffling.}
 Typically, the local training dataset is processed once or several times in an incremental fashion; that is, one data point (or one small minibatch) at a time. However, experimental evidence shows that processing the local data
{\em without replacement} can lead to substantially better results than processing the data {\em with replacement}. In particular, processing the local training data in an order dictated by a random permutation---a technique known as  Random Reshuffling (\algname{\gls{RR}})---is often set as default in modern deep learning and Federated Learning software ~\citep{bottou2009curiously, bengio2012practical,sun2020optimization}. 
This is in sharp contrast with the {\em with-replacement} sampling of data employed by \algname{\gls{SGD}}. With-replacement sampling ensures that the gradient updates are unbiased, and this simplifies the analysis. For this reason, \algname{\gls{SGD}} is significantly better understood in theory than its better performing but much more poorly understood  cousin \algname{\gls{RR}}. However, recent results \citep{MKR2020rr}, and extensions \citep{mishchenko2022proximal, yun2021minibatch} to distributed training, show that \algname{\gls{RR}} can have clear theoretical advantages over \algname{\gls{SGD}}.

{\bf Server stepsizes.} Once local training is finished, the clients in the cohort send their models or model updates to the orchestrating server, which typically aggregates them via averaging. This information is then used to perform {\em server side} optimization. The simplest approach is to do nothing; that is, to  treat the aggregated models as the next global model that is broadcast to the new cohort in the next communication round. However, empirical evidence suggests that it is better to aggregate {\em model updates}, and treat them as gradient-type information which can be injected into a suitably chosen server side optimization routine~\citep{karimireddy2020scaffold}. For example, the server may run one step of \algname{\gls{GD}} using the aggregated model update as a proxy for the gradient which is not available,  with its own server-side stepsize.  

\begin{table*}[t!]
\begin{center}
  \begin{threeparttable}
    \centering
    \small
    {
	\caption{Conceptual comparison of results for FedAvg from prior work with our results. }
  \label{tab:compare_with_others}
	\centering 
	\begin{tabular}{cccccccc}\toprule
		 \makecell{Partial\\ participation}  & \makecell{Local \\ training} & \makecell{Data\\ shuffling} & \makecell{\textbf{Large} server\\ stepsizes help} & \makecell{\textbf{Small} server\\ stepsizes help} &   Reference \\
		\midrule
	\cmark & 	\cmark & \xmark  & \cmark & \xmark &  [\citenum{karimireddy2020scaffold}] \\
		\cmark &	\cmark & \xmark  & \cmark & \xmark &  [\citenum{woodworth2020minibatch}] \\
		\xmark  &	\cmark & \xmark  & \xmark & \xmark  &  [\citenum{koloskova2020unified}] \\
		\xmark &	\cmark & \xmark  & \xmark & \xmark  &  [\citenum{localSGD-AISTATS2020}] \\
		\xmark & \cmark & \cmark  & \xmark & \xmark  &  [\citenum{mishchenko2022proximal}]\\
\midrule		
		\cmark & \cmark & \cmark  & \cmark & \cmark  &  \textbf{This paper} \\
		\bottomrule
    \end{tabular}
    }
  \end{threeparttable}
  \end{center}
\end{table*}

{\bf Further useful tricks.} Additional tricks that are often employed in the context of Federated Learning include the use of compressed communication \citep{Alistarh-EF2018, gorbunov2021marina}, drift reduction \citep{karimireddy2020scaffold,gorbunov2021local}, error compensation~\citep{stich2019, richtarik2021ef21}, server side momentum~\citep{hsu2019measuring}, and adaptive stepsize selection~\citep{reddi2020adaptive}. These techniques are beyond the scope of this paper.

\section{Contributions}\label{sec:contributions_ch5}

Despite the fact that {\em partial participation}, {\em local training}, {\em data shuffling} and {\em server stepsizes} have all been empirically found  to be very useful building blocks of \gls{FL} methods, most of these techniques are not very well understood in theory even in isolation. Informally speaking, and at the risk of oversimplifying the current state of affairs, we know virtually nothing about {\em server stepsizes}, very little about  {\em data shuffling}, relatively much more about 
{\em local training}, and quite a bit, but still ``not enough'', about {\em partial participation}.

\begin{quote}  The key focus of this paper is to make a substantial advance in the current theoretical understanding of  {\em server stepsizes} in the context of {\em realistic} Federated Learning.  
\end{quote}

In order to theoretically understand the server stepsize phenomenon in a realistic context of techniques commonly used in \gls{FL}, we study this phenomenon {\em together} with data shuffling, local training and partial participation. While this makes the analysis substantially harder and different from all\footnote{Except for the recent work \citep{mishchenko2022proximal} which we used as an inspiration.} existing analyses of \algname{FedAvg}, we believe it is important to do so as this will highlight the {\em  interplay} between these algorithmic techniques and their {\em combined} impact on training.

A brief visual summary of this in the context of selected existing methods is provided in \Cref{tab:compare_with_others}.  We summarize our contributions as follows:

$\bullet$ {\bf New algorithm.} We design a new algorithm, for which we coin the name \algname{Nastya} (Algorithm~\ref{alg:pp-jumping}; see Section~\ref{sec:algorithm}), which combines all of the aforementioned practical tricks and techniques in a single method: 
{\em partial participation}, {\em local training}, {\em data shuffling} and, most importantly,  {\em server stepsizes}.
In our method, in each communication round $t$, the cohort is chosen as a random subset $\set$ of the set  $\{1,2,\dots,M\}$ of clients of cardinality $1\leq C \leq M$, chosen uniformly from all subsets of cardinality $C$. Each device performs local training  via a  single pass of incremental \algname{\gls{GD}} with {\em client stepsize} $\cstep > 0$ over the local training data points in an order dictated by a {\em random permutation}. We allow for two options: i) either the random permutation for all clients is sampled just once and used in all communication rounds ({\em Shuffle-Once} option), or ii) the random permutation is sampled afresh at the start of each communication round ({\em Random-Reshuffling} option). At the end of local training, the updated models are communicated back to the server, which uses these updates to form a {\em gradient estimator}, and applies one step of \algname{GD} using a server stepsize $\sstep > 0$ with this estimator in lieu of the true gradient. The new model is then broadcast to a new cohort in the next communication round, and the process is repeated.

\begin{table*}[t]
	\begin{center}
		\begin{threeparttable}
			\centering
			{\footnotesize
				\caption{Comparison of convergence results for FedAvg from prior work with our results. }
				\label{tab:compare_with_others_second}
				\centering 
				\begin{tabular}{ccccccc}\toprule
					Method & \makecell{Strongly convex\tnote{(2)}}  & \makecell{Non-convex } &   Reference \\
					\midrule
					\algname{SCAFFOLD} \tnote{(1)}& 	$\mathcal{\tilde{O}}\left(\frac{\sigma^{2}}{\mu M n \epsilon}+\frac{1}{\mu}\right)$   & $\mathcal{O}\left(\frac{\sigma^{2}}{M n \epsilon^{2}}+\frac{1}{\epsilon}\right)$ &  [\citenum{karimireddy2020scaffold}] \\
					\algname{\gls{LSGD}} \tnote{(1)} &	$\mathcal{\tilde{O}}\left(\frac{L}{\mu}+\frac{\sigma^{2}}{M \mu \varepsilon}+\sqrt{\frac{L n\left(\sigma^{2}+n \zeta^{2}\right)}{\mu^{2} \varepsilon}}\right)$ \tnote{(3)}    & \xmark  &  [\citenum{woodworth2020minibatch}] \\
					\algname{\gls{LSGD}}&	$\tilde{\mathcal{O}}\left(\frac{\sigma_\star^{2}}{M \mu \epsilon}+\frac{\sqrt{L}(n \zeta+\sqrt{n} \sigma)}{\mu \sqrt{\epsilon}}+\kappa n\right) $ \tnote{(3)}    & $\mathcal{O}\left(\frac{L \sigma_\star^{2}}{M \epsilon^{2}}+\frac{L(n \zeta+\sqrt{n} \sigma)}{\epsilon^{3 / 2}}+\frac{L n}{\epsilon}\right)$\tnote{(3)}    &  [\citenum{koloskova2020unified}] \\
					\algname{FedRR} &	$\mathcal{\tilde{O}}\left(\frac{L}{\mu}+\frac{\sqrt{\kappa n}\left( \sigma_{*}+\sqrt{n}\zeta\right)}{\mu \sqrt{\varepsilon}}\right)$ \tnote{(3)}     & \xmark  &  [\citenum{mishchenko2022proximal}]\\
					\midrule					
					\algname{Nastya} &	$\mathcal{\tilde{O}}\left(\frac{Ln}{\mu}\right)$ & $\mathcal{O}\left(\frac{Ln}{\varepsilon}\right)$     &  \textbf{New} \\
					\bottomrule
				\end{tabular}
			}
			\begin{tablenotes}
				{\scriptsize
					\item [(1)] The analysis is done under the bounded variance assumption: $g_{i}(x):=\nabla f_{i}\left(x; \zeta_{i}\right)$ is unbiased stochastic gradient of $f_{i}$ with bounded variance
					$
					\mathbb{E}_{\zeta_{i}}\left[\left\|g_{i}(x)-\nabla f_{i}(x)\right\|^{2}\right] \leq \sigma^{2}, \text { for any } i, x.
					$			        
					\item [(2)] The $\tilde{O}$ notation omits $\log \frac{1}{\varepsilon}$ factors 
					\item[(3)]Here we use $\zeta^2 \eqdef \frac{1}{M}\sum_{m=1}^{M}\|\nabla f_m(x^{\star})\|^2$.
				}
			\end{tablenotes}
		\end{threeparttable}
	\end{center}
\end{table*}

$\bullet$ {\bf Complexity analysis.} We provide strong complexity analysis of our new algorithm for strongly convex (Theorem~\ref{thm:PP-SC}), convex (Theorem~\ref{thm:PP-C}) and non-convex (Theorem~\ref{thm:PP-NC}) functions; see Table~\ref{tab:our_results}. This is the first theory for a variant of \algname{FedAvg} that combines the benefits of partial participation, data shuffling, local training and, most importantly, also {\em server stepsizes}. Most importantly, with only a couple of exceptions~\citep{karimireddy2020scaffold,woodworth2020minibatch}, there are no prior theoretical works analyzing the effect of server stepsizes in \gls{FL}. The methods in the aforementioned works use local training and partial participation, but do not use data shuffling, and are significantly different from ours. 

$\bullet$ {\bf Small client stepsizes, large server stepsizes, and no need for drift reduction.} In particular, Theorems~\ref{thm:PP-SC}, \ref{thm:PP-C} and \ref{thm:PP-NC}, covering the strongly convex, convex and non-convex regimes, respectively,  suggest that the server can use the {\em large} $\cO(1/L)$ stepsize, where $L$ is the Lipschitz constant of the gradient of $f$. In the strongly convex and convex regimes, based on our theory, it is optimal for the client stepsize $\cstep$ to be {\em small}, which completely eliminates the second of the three terms in the complexity bounds (see the third column of Table~\ref{tab:our_results})  which controls the price one pays due to {\em data heterogeneity}. Indeed, our theory allows for the client stepsize $\cstep$ to be small while the server stepsize $\sstep$ can be large (see the second column of  Table~\ref{tab:our_results}).

Note that  in all three regimes, and thanks to the fact that we employ a {\em data shuffling} strategy, this second term depends on the square $\cstepsquared$ of the client stepsize, which means that we can make this term small without making the client stepsizes infinitesimal. So, thanks to \algname{Nastya}'s use of data shuffling strategies, it does {\em not} require any explicit drift reduction technique such as \algname{SCAFFOLD} to handle data heterogeneity~\citep{karimireddy2020scaffold}. 

$\bullet$ {\bf Small server stepsizes can be beneficial.} To the best of our knowledge, no prior theoretical work suggests that it might be beneficial to use {\em small} server stepsizes. Our results (see Theorem~\ref{th:small_alpha}) suggest that this can be the case  when  each $f_{m,i}$ is strongly convex and smooth, and when the strong convexity parameter is very small.

$\bullet$ {\bf Experimental validation of our theoretical predictions.} We provide experimental examination of \algname{Nastya} and compare it with selected benchmarks. Our goal is not to perform large scale experiments and claim empirical superiority because the algorithmic ingredients embedded in \algname{Nastya} already {\em are} being used in practical \gls{FL} methods precisely because they have already been empirically found to be useful. This allows us to focus on simple experiments which test the theoretical predictions of our theory. 

Our experimental results confirm our theory, and illustrate the behavior of the methods we test in various settings. Moreover, we go beyond the theory and conduct additional experiments with the adaptive stepsize strategy introduced in~\citep{malitsky2019adaptive}. Inspired by the work of \citep{reddi2020adaptive}, we additionally utilize several server-side optimization subroutines on top of  the local updates.

\begin{figure*}[t]
	\centering
	\begin{tabular}{cc}
		$\!\!\!\!$\includegraphics[scale=0.12]{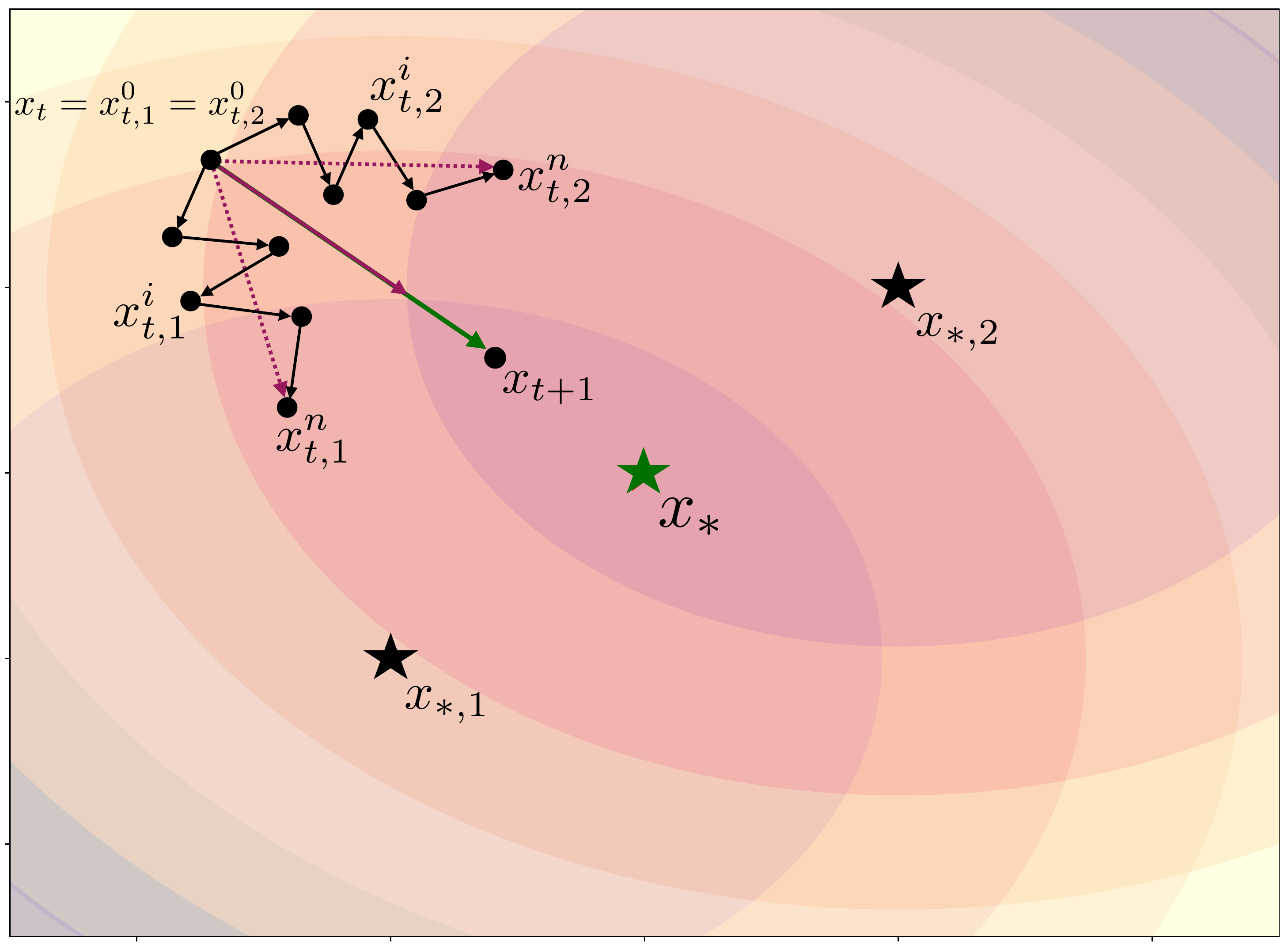}&
		$\!\!\!\!$\includegraphics[scale=0.12]{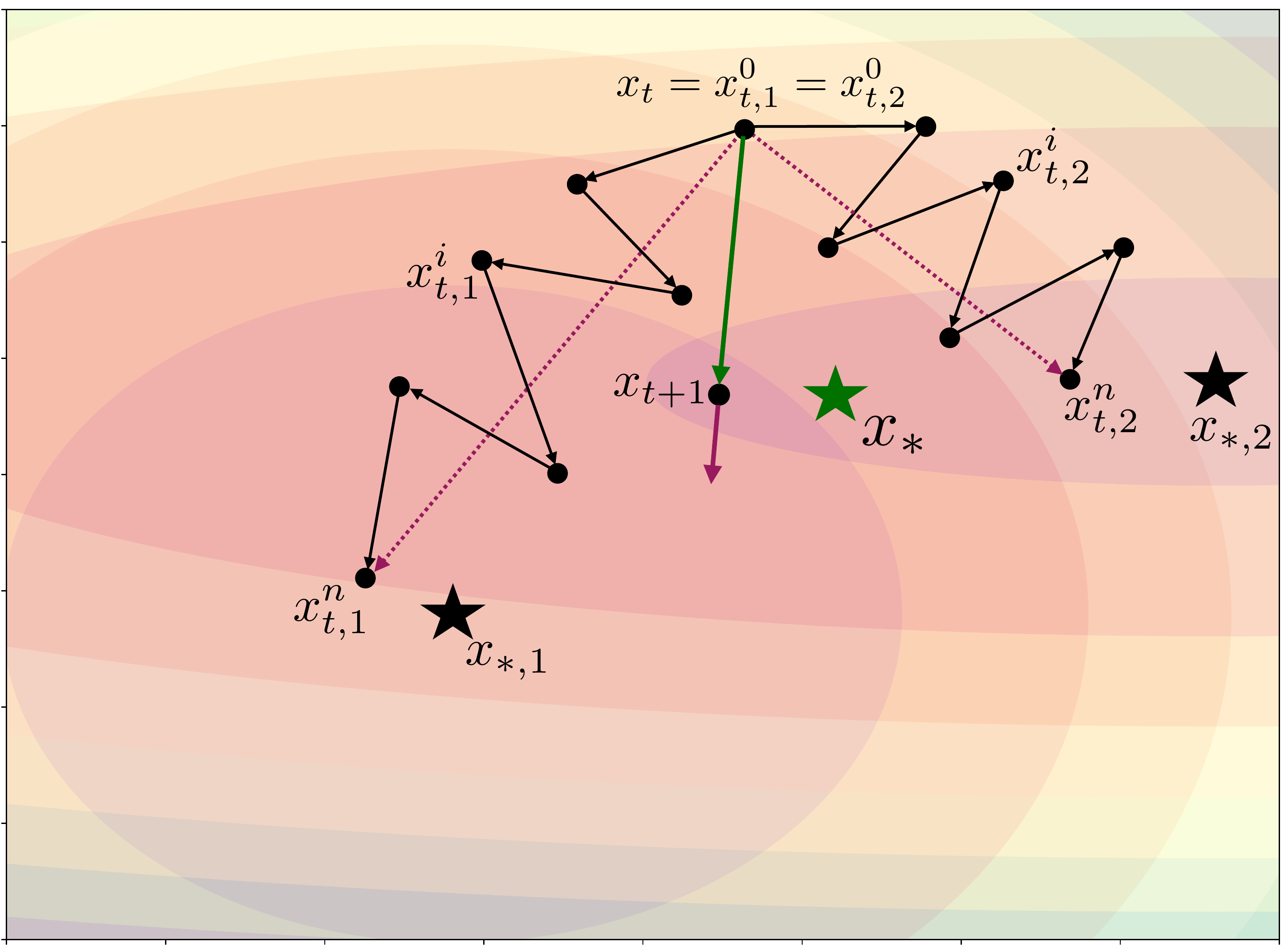}\\
		(a)&(b)
	\end{tabular}	
	\caption{Illustration of the dependence between server and client stepsizes on a simple example with $M=2$ clients. $x_{*,1}$ and $x_{*,2}$ are the minimizers of the local functions $f_{1}$ and $f_2$, respectively, and $x^{\star}$ is the minimizer of the global function $f = \frac{1}{2}f_1 + \frac{1}{2}f_2$. {\bf (a)} In the case of small client stepsizes $\cstep$, the average of local steps is not large, but at the same time the variance is small and the direction is close to direction of the full gradient, which allows us to go further towards this direction by employing a large server stepsize $\sstep$. {\bf (b)} In the case of large client stepsizes $\cstep$, each client step contributes to the global step, but the variance grows as well, so it is useful to use smaller server stepsize $\sstep$ to reduce this variance. These intuitions are confirmed by our theory. }
	\label{fig:image4}
\end{figure*}

\section{Preliminaries}\label{sec:prelim}

In this section we introduce several key concepts that will help us to formulate our theoretical results.

\subsection{Convexity and smoothness}

In all our theoretical results we rely on smoothness, and in some we require convexity or strong convexity.

\begin{definition}[$L$-smoothness]
Function $\phi\colon \R^d \rightarrow \R$ is $L$-smooth if it has $L$-Lipschitz continuous gradient for some $L>0$  
\begin{align}
	\squeeze 
	\label{eq:lipsc}
	\|\nabla \phi(x) - \nabla \phi(y)\| \leq L\|x-y\| \quad \forall x, y \in \mathbb{R}^{d}.
\end{align}
\end{definition}

\begin{definition}[Convexity and strong convexity]
Function $\phi\colon \R^d \rightarrow \R$ is convex if $\forall x, y \in \mathbb{R}^{d}$
\begin{align}
	\squeeze 
		\label{eq:convex}
	\phi(y) \geq \phi(x)+\langle\nabla \phi(x), y-x\rangle,
\end{align}
and $\mu$-strongly convex if $\forall x, y \in \mathbb{R}^{d}$
\begin{align}
	\label{eq:mu-convex}
	\squeeze 
	\phi(y) \geq \phi(x)+\langle\nabla \phi(x), y-x\rangle+\frac{\mu}{2}\|y-x\|^{2}.
\end{align}
\end{definition}

In our analysis we use the following assumption.
\begin{assumption}\label{assump: L-smooth_1}
	The objective $f$ and the individual losses $f_{m,1}, \ldots, f_{m,n}$ are all $L$-smooth. Further, for all $i$ and $m\in \{1,2,\dots,M\}$ and $i\in \{1,2,\dots,n\}$, (i) $f^\star  \eqdef \inf_{x} f(x) > -\infty $, 
	(ii) $f^{\star}_{m} \eqdef \inf_{x} f_m(x)> -\infty$, and
(iii) $f^{\star}_{m,i} \eqdef \inf_{x} f_{m,i}(x) > -\infty$.
If $f_{m,i}$ is convex, we further assume the existence of minimizers $x^{\star} = \arg\min_{x\in \mathbb{R}^{d}} f(x)$ and $x^\star_{m,i} =\arg \min_{x\in \mathbb{R}^{d}} f_{m,i}(x) $.
\end{assumption}

\subsection{Measures of data heterogeneity}

While our theory does not require any {\em assumptions} on data homogeneity,  our {\em results} will reflect the degree to which the data are heterogeneous, and are better for data that are ``more'' homogeneous. In particular, in the strongly convex and convex regimes we rely on the following notions.

\begin{definition}[Variance at the optimum]
	\label{def:variance}
 The variance of the local gradients $\{\nabla f_m\}_{m=1}^M$ at $x^{\star}$  is defined as 
\[
	\squeeze 
	\sigma_{*}^{2} \stackrel{\text { def }}{=} \frac{1}{M} \sum\limits_{m=1}^{M}\left\|\nabla f_m\left(x^{\star}\right)\right\|^{2},
\]
where $x^{\star}$ is a minimizer of $f$. The  variance of the gradients $\{\nabla f_{m,i}\}_{i=1}^n$ at $x^{\star}$ is 
\[
	\squeeze 
	\sigma_{*,m}^{2} \stackrel{\text { def }}{=} \frac{1}{n} \sum\limits_{i=1}^{n}\left\|\nabla f_{m,i}\left(x^{\star}\right) \right\|^{2}.
\]
\end{definition}

An important lemma that allows us to obtain a strong upper bound for variance in the case of sampling without replacement, which our data shuffling methods rely on, was formulated in~\citep{MKR2020rr}. We include it here for completeness.
\begin{lemma}[Sampling without replacement]\label{lem:sampling_wo_replacement}
	Let $X_1,\dotsc, X_n\in \R^d$ be fixed vectors, $\overline X\eqdef \frac{1}{n}\sum_{i=1}^n X_i$ be their average and $ \sigma^2 \eqdef \frac{1}{n}\sum\limits_{i=1}^n \norm{X_i-\overline X}^2$ be the population variance. Fix any $k\in\{1,\dotsc, n\}$, let $X_{\pi_1}, \dotsc X_{\pi_k}$ be sampled uniformly without replacement from $\{X_1,\dotsc, X_n\}$ and $\overline X_\pi$ be their average. Then, it holds
	\begin{align}
\squeeze
		\ec{\overline X_\pi}=\overline X,  \quad \ec{\norm{\overline X_{\pi} - \overline X}^2}= \frac{n-k}{k(n-1)}\sigma^2. \label{eq:sampling_wo_replacement}
	\end{align}
\end{lemma}

For non-convex functions, we use a different notion of data heterogeneity. 

\begin{definition}[Functional dissimilarity]
	\label{def:variance-non-convex}
	The variance at the optimum in the non-convex regime is defined as 
	\[
		\Delta^\star \eqdef  f^\star -  \frac{1}{M} \sum \limits_{m=1}^{M}f^{\star}_{m},
	\]
	where $f^{\star}_{m} = \inf_{x} f_m(x)$ and $f^\star=\inf_x f(x)$. 
	For each device $m$, the variance at the optimum is defined as
	\[
		\Delta^{\star}_{m} \eqdef f^\star - \frac{1}{n} \sum \limits_{i=1}^{n}f^{\star}_{m,i},  
	\]
	where $f^{\star}_{m,i} = \inf_x f_{m,i}(x)$.
\end{definition}

Again, the above is a definition and not an assumption. The concepts are well defined as long as Assumption~\ref{assump: L-smooth_1} is satisfied.

\section{The Nastya Algorithm} \label{sec:algorithm}

\begin{algorithm*}[!t]
	\caption{\algname{Nastya}: Federated optimization with server stepsize, random shuffling and partial participation}
	\label{alg:pp-jumping}
	\begin{algorithmic}[1]
		\STATE {\bf Input:}  {\myred client stepsize $\cstep > 0$}; {\mygreen server stepsize $\sstep \geq 0$};  cohort size $C \in \{1,2,\dots,M\}$; initial iterate/model $x^{0} \in \mathbb{R}^d$; number of communication rounds $T\geq 1$
		\STATE {\myblue \textbf{Shuffle-Once option:} For each client $m$, sample a permutation $\pi_m=(\pi^0_{m}, \pi^1_{m}, \ldots, \pi^{n-1}_{m})$ of  $\{1,2,\dots,n\}$}

		\FOR{communication round $t = 0,1,\ldots, T-1$}
		\STATE Sample a cohort $\set$ of $C$ clients \hfill {\scriptsize (server chooses a random set $\set\subseteq \{1,2,\dots,M\}$ of size $|\set| = C$, uniformly at random)}
		\STATE Send model $x^t$ to all participating clients $m\in \set$  \hfill {\scriptsize (server broadcasts  $x^t$ to all clients $m\in \set$ )}
		\FOR{all clients $m\in \set$, locally in parallel}
		\STATE $x^{t}_{m,0}= x^t$ \hfill {\scriptsize (client $m$ initializes local training  using the latest global model $x^t$)}
		\STATE {\myblue \textbf{Random-Reshuffling option:} Sample a permutation for current client $\pi_m=(\pi^0_{m}, \pi^1_{m}, \ldots, \pi^{n-1}_{m})$   of $\{1,2,\dots,n\}$}
		\FOR{all local training data points $i=0, 1, \ldots, n-1$}
		\STATE $x^{t}_{m,i+1} = x^{t}_{m,i} - \cstep \nabla f_{m,\pi^{i}_m} (x^{t}_{m,i})$ \hfill {\scriptsize (client $m$ makes one  pass over its local training data in the order dictated by $\pi_m$)}
		\ENDFOR
		\STATE $g^{t}_{m} = \frac{1}{{\myblue\cstep} n}(x^t - x^{t}_{m,n})$ \hfill {\scriptsize (client $m$ computes local update direction $g^{t}_{m}$)}
		\ENDFOR
		\STATE $g^{t} = \frac{1}{C}\sum \limits_{m\in \set}g^{t}_{m} $ \hfill {\scriptsize (server aggregates the local updates $g^{t}_{m}$ discovered by the cohort $\set$ of clients)}

		\STATE $x^{t+1} = x^t - {\myred\sstep} g^t $ \hfill {\scriptsize (server updates the model using the direction $g^t$ and applying server stepsize $\sstep$)}
		\ENDFOR
	\end{algorithmic}
\end{algorithm*}

We now formally describe our \algname{Nastya} algorithm (see Algorithm~\ref{alg:pp-jumping}). \algname{Nastya} combines several techniques that were empirically found to be useful in \gls{FL}:
{\em partial participation}, {\em local training}, {\em data shuffling} and {\em server stepsizes}.

In each communication round $t\geq 0$ of \algname{Nastya}, the cohort $\set$ is chosen as a random subset of the set  $\{1,2,\dots,M\}$ of all clients. In particular, we choose a random subset of cardinality $C$ (the cohort size), where $1\leq C \leq M$, uniformly at random. The server then sends the global model $x^t$ to all clients in the cohort. Setting $C=M$ models the full participation regime.

Each participating client $m\in \set$ then performs local training using a single pass of incremental \algname{\gls{GD}} with {\em client stepsize} $\cstep > 0$ over the local training data points in an order dictated by a {\em random permutation} \[\pi_m = (\pi_m^1,\pi_m^2,\dots,\pi_m^n)\] of the indices of the local training dataset $\{1,2,\dots,n\}$. In particular, the following update is iterated for $i=0,\dots,n-1$:
\[
	x^{t}_{m,i+1} = x^{t}_{m,i} - \cstep \nabla f_{m,\pi^{i}_m} (x^{t}_{m,i}),
\]
where $x_{t,m}^0$ is initialized to $x^t$, and $\cstep > 0$ is the client stepsize. That is, we run one pass over the local data using the \algname{\gls{RR}} method~\citep{MKR2020rr}. This differs from one pass over the data via \algname{\gls{SGD}} in that each data point is sampled exactly once.

Note that we allow for two options for how the permutation is formed: i) either the random permutation  is sampled just once for all clients, and used in all communication rounds ({\em Shuffle-Once} option), or ii) the random permutation is sampled afresh at the start of each communication round ({\em Random-Reshuffling} option). Both have the same theoretical properties in our analysis.

At the end of local training, the updated models $x_{t,m}^{n}$ are communicated back to the server, which uses these updates to form a {\em gradient-type estimator} $g^t$, and applies one step of \algname{\gls{GD}} using a server stepsize $\sstep > 0$ with this estimator in lieu of the true gradient. Equivalently — and this is how we decided to formally state the method — each client $m\in \set$ sends the following scaled model difference to the server:
\[
	\squeeze g^{t}_{m} = \frac{1}{{\myblue\cstep} n}(x^t - x^{t}_{m,n}),
\]
where $x^{t}_{m,n}$ is the model found by the client after one pass over the data via \algname{\gls{RR}}. The server then aggregates these vectors from all clients in the cohort to form
$g^{t} = \frac{1}{C}\sum_{m\in \set}g^{t}_{m},$ and then takes a gradient-type step using this quantity in lieu of the gradient, using server stepsize $\sstep>0$: 
\[
	x^{t+1} = x^t - {\myred\sstep} g^t .
\]

The new model is then broadcast to a new cohort in the next communication round, and the process is repeated.

\section{Warm-Up: How to Improve Random Reshuffling}
In this section, we provide the intuition behind our complexity improvements through the lens of single-node Random Reshuffling (\algname{\gls{RR}}). In particular, when $M=1$, objective~\eqref{eq:main_chapter_5} recovers the standard \gls{ERM} problem:
\[
 \min \limits_{x\in\R^d} \frac{1}{n}\sum \limits_{i=1}^n f_{i}(x).
\]
The update of  \algname{\gls{RR}}  for this problem has the form
\begin{align*}
	x^{t}_{i+1} = x^{t}_{i} - \cstep \nabla f_{\pi^{i}} (x^{t}_{i}),
\end{align*}
where we use a permutation $\pi=(\pi^0, \dotsc, \pi^{n-1})$ that is randomly sampled at the beginning of epoch $t$.
Unrolling this recursion, we get 
\begin{align*}
	\squeeze
	x^{t}_{n} = x^t - \cstep \sum\limits_{i=0}^{n-1} \nabla f_{\pi^{i}} (x^{t}_{i}).
\end{align*}
The key insight is that the gradients evaluated at points $x^t_i$ can be viewed as approximations of the gradients at point $x^t$. If we denote, for simplicity,
\begin{align*}
	\squeeze	g^t = \frac{1}{n}\sum\limits_{i=0}^{n-1}\nabla f_{\pi^i}(x^{t}_{i}) = \frac{x^t - x^{t}_{n}}{\cstep n},
\end{align*}
then one can show that $g^t\approx \nabla f(x^t)$ whenever $\cstep$ is small. The update of \Cref{alg:pp-jumping} becomes much simpler and reduces to 
\begin{align*}
	x^{t+1} &= x^{t}_{n}+\beta(x^{t}_{n} - x^t) = x^t + (1+\beta)(x^{t}_{n} - x^t)\\
	&=  x^t - (1+\beta)\cstep \sum\limits_{i=0}^{n-1}\nabla f_{\pi^i}(x^{t}_{i}) = x^t - \sstep g^t,
\end{align*}
where $\sstep = (1+\beta)\cstep n $. If we imagine for a moment that $g^t$ is indeed a very good approximation of $\nabla f(x^t)$, then the theory of gradient descent suggests that one should use $\sstep \sim \frac{1}{L}$, regardless of the value of $\cstep$. 

{\bf Complexity improvements.} By following this intuition, we establish, as special cases of our general theory, several improved complexity bounds. In the strongly convex setting, we obtain
\[
\mathcal{O}\!\left(\kappa n \log \frac{1}{\varepsilon}\right),
\]
for the modified Random Reshuffling method, improving upon the standard Random Reshuffling complexity
\[
\mathcal{O}\!\left(
\kappa n
+
\frac{\sqrt{\kappa n}\,\sigma_*}{\mu \sqrt{\varepsilon}}
\right)
\log \frac{1}{\varepsilon}.
\]

In the convex setting, our result achieves the complexity
\[
\mathcal{O}\!\left(\frac{Ln}{\varepsilon}\right),
\]
which improves over the slower bound
\[
\mathcal{O}\!\left(
\frac{Ln}{\varepsilon}
+
\frac{\sqrt{Ln}\,\sigma_*}{\varepsilon^{3/2}}
\right).
\]

Finally, in the general non-convex setting, we derive the bound
\[
\mathcal{O}\!\left(\frac{Ln}{\varepsilon^2}\right),
\]
which improves upon
\[
\mathcal{O}\!\left(
\frac{Ln}{\varepsilon^2}
+
\frac{L\sqrt{n}(B+\sqrt{A})}{\varepsilon^3}
\right),
\]
where the constants \(A\) and \(B\) are defined, following \citep{MKR2020rr}, through the assumption
\[
\frac{1}{n}\sum_{i=1}^{n}
\left\|
\nabla f_i(x)-\nabla f(x)
\right\|^2
\leq
2A\bigl(f(x)-f^\star\bigr)+B^2.
\]
\subsection{Extending the intuition to multiple nodes}
Motivated by the example of Random Reshuffling, we can extend the complexity improvements to the case of multiple nodes. To achieve this, we utilize a large server stepsize and a small client stepsize. The main idea of this approach is again to approximate the full gradient using local passes over the clients' datasets. During each round, a worker node $m$ computes $n$ steps of permutation-based algorithm and obtains local model parameters $x^{t}_{m,n}$:
\begin{align*}
	x^{t}_{m,n} =  x^t - \cstep \sum\limits_{i=0}^{n-1}\nabla f_{m, \pi^i_m} (x^{t}_{m,i}).
\end{align*}
If $\cstep$ is not large, the sum of local steps serves as a good approximation of the full client gradient, so we define
\begin{align*}
	g^{t}_{m} &  = \frac{1}{\cstep n }(x^t - x^{t}_{m,n})= \frac{1}{n}\sum\limits_{i=0}^{n-1} \nabla f_{m, \pi^i_m}
	(x^{t}_{m,i}).
\end{align*} 
When the epoch ends, the server aggregates local approximations and then computes a step with the larger stepsize, which is equivalent to averaging the final local iterates and then extrapolating in the obtained direction:
\begin{align}
\notag	x^{t+1} &= x^t - \sstep  \frac{1}{C}\sum\limits_{m\in \set}g^{t}_{m} \\
\notag	&=x^t - \frac{\sstep}{\cstep n}  \frac{1}{C}\sum\limits_{m\in \set}(x^t - x^{t}_{m,n})\\
	&= \frac{1}{C}\sum\limits_{m\in \set}\left( x^{t}_{m,n}+\beta\left(x^{t}_{m,n} - x^t\right)\right). \label{eq:iteration}
\end{align}
Above, $\beta =  \nicefrac{\sstep}{\cstep n} - 1$ is the extrapolation coefficient. Small client stepsizes allow us to get a better approximation of the full gradient, hence we obtain significantly smaller variance of stochastic steps. In the extreme case when client stepsize goes to zero, $\cstep \rightarrow 0$, the gradient estimator converges to the exact gradient: $g^{t}_{m} \rightarrow \nabla f_{m}(x^t)$ and we obtain distributed gradient descent method. 

{\em Large} client stepsizes, on the other hand, combine better with {\em small} server stepsize. In that case, each local step has a big impact, and full-gradient approximation breaks down. Since $g^t$ no longer stays close to $\nabla f(x^t)$, we use a different analysis for this case, which shows a benefit whenever the Partial Participation noise is significant. This is particularly relevant to the cross-device Federated Learning, where only a tiny percentage of clients can participate at each round. 

\section{Theory}\label{sec:theory_ch_5}

\begin{table*}[t]
	\begin{center}
		\begin{threeparttable}
			\centering
			{\tiny
				\caption{The main convergence results obtained in this paper. }
				\label{tab:our_results}
				\centering 
				\begin{tabular}{c | c c}\toprule
					Regime & Stepsizes & Result\tnote{(1)}  \\
					\midrule
					\makecell{$\mu$-Convex  \\ (Theorem~\ref{thm:PP-SC})}	& $\cstep n \leq \sstep\leq \frac{1}{16L}$ & $\mathbb{E}\| x^{T} - x^{\star} \|^2  \leq \left(1 - \frac{\sstep \mu}{2}\right)^T  \left\| x^{0} - x^{\star} \right\|^2  + \frac{5\cstepsquared nL}{\mu} \Sigma_\star^2
					+\frac{8\sstep }{\mu} \frac{(M-C)\sigma_{*}^2}{C\max\{1,M-1\}} $ \\
					\makecell{Convex \\ (Theorem~\ref{thm:PP-C})} 
					& $\cstep n \leq \sstep\leq \frac{1}{16L}$	
					&  $\ec{ f(\hat{x}_T) - f(x^{\star})} \leq  \frac{ 5\left\| x^{0} - x^{\star} \right\|^2}{2\sstep  T}  + 7\cstepsquared nL \Sigma_\star^2
					+10\sstep \frac{(M-C)\sigma_{*}^2}{C(\max\{1,M-1\}} $ \\	
					\makecell{Non-convex \\ (Theorem~\ref{thm:PP-NC})}	
					& $\cstep \leq \frac{1}{2nL}$ \& $\sstep \leq \frac{1}{L}$ 
					& $\min \limits_{t} \mathbb{E}\left[\left\|\nabla f\left(x^{t}\right)\right\|^{2}\right] \leq \frac{2\left(1+4\sstep  \cstepsquared  n^2 L^3\right)^T}{\sstep  T} \delta^0
					+ 2 \cstepsquared  n L^3  D_\star^2
					+ 4L^2\sstep  \frac{(M-C)\Delta^\star}{C\max\{1,M-1\}}  $ \\
					\bottomrule
				\end{tabular}
			}
			\begin{tablenotes}
				{\scriptsize
					\item [(1)] $\cstep$ = client stepsize; $\sstep$ = server stepsize; $M$ = total \# of clients;  $C$ = \# of participating clients (cohort size); $n$ = \# of training data points per client; $L$ = Lipschitz constant of the gradient of $f$; $\mu$ = strong convexity constant of $f$; $T$ = total \# of communication rounds; $x^{0}$ = initial model; $x^{\star}$ = optimal model; $\delta^0$ =  $f(x^{0}) - f^\star$; $\Sigma_\star^2 = \left(\frac{1}{M}\sum \limits_{m=1}^{M}\sigma^2_{*,m} + n \sigma_{*}^2 \right)$; $\sigma_{*}^{2} = \frac{1}{M} \sum\limits_{m=1}^{M}\left\| \nabla f_m\left(x^{\star}\right)\right\|^{2}$;  $\sigma_{*,m}^{2} = \frac{1}{n} \sum\limits_{i=1}^{n}\left\|\nabla f_{m,i}\left(x^{\star}\right) \right\|^{2}$; $D_\star^2 = \left(\frac{1}{M}\sum \limits_{m=1}^{M}\Delta^{\star}_{m}+n\Delta^\star\right)$; $\Delta^\star =\frac{1}{M} \sum \limits_{m=1}^{M}(f^{\star}_{m} - f^\star)\ge 0$; $\Delta^{\star}_{m} = \frac{1}{n}\sum\limits_{i=1}^{n}(f^\star - f^{\star}_{m,i})\ge 0$, where $f^\star=\inf f$, $f^{\star}_{m} = \inf f_m$ and $f^{\star}_{m,i} = \inf f_{m,i}$ are all assumed to be finite (i.e., not $-\infty$).}
			\end{tablenotes}
		\end{threeparttable}
	\end{center}
\end{table*}

We now formulate our three main results.

\begin{theorem}[Strongly convex regime]\label{thm:PP-SC}
	Let Assumption~\ref{assump: L-smooth_1} hold, let each $f_{m,i}$ be convex, and let $f$ be $\mu$-strongly convex. Let $\cstep n \leq \sstep\leq \frac{1}{16L}$. Then for iterates $x^t$ generated by Algorithm~\ref{alg:pp-jumping}, we have 
	\begin{align*}
		\squeeze 
		\mathbb{E}\left[\| x^{T} - x^{\star} \|^2\right] &\squeeze \leq \left(1 - \frac{\sstep \mu}{2}\right)^T  \left\| x^{0} - x^{\star} \right\|^2\\
		& \squeeze + \frac{5\cstepsquared nL}{\mu}\left(\frac{1}{M}\sum\limits_{m=1}^{M}\sigma^2_{*,m} + n \sigma_{*}^2 \right)\\
		&\squeeze +\frac{8\sstep }{\mu} \frac{M-C}{C\max\{M-1,1\}} \sigma_{*}^2.
	\end{align*}
\end{theorem}

	In the full participation regime, the server stepsize restriction can be relaxed to $\sstep\leq \frac{1}{8 L}$.

\subsection{Convex regime}
Next, we cover the convex regime. 

\begin{theorem}\label{thm:PP-C}
Let Assumption~\ref{assump: L-smooth_1} hold, and let each $f_{m,i}$ be a convex function. Let $\cstep n \leq \sstep\leq \frac{1}{16L}$. Let $\hat{x}_{T} \eqdef \frac{1}{T} \sum_{t=1}^{T} x^{t}$. Then for iterates $x^t$ of Algorithm~\ref{alg:pp-jumping}, we have 
	\begin{align*}
		\mathbb{E} [ f(\hat{x}_T) - f(x^{\star})]& \leq  \frac{ 5\left\| x^{0} - x^{\star} \right\|^2}{2\sstep  T} \squeeze+10\sstep \frac{M-C}{C\max\{M-1,1\}} \sigma_{*}^2\\
		&+ 7\cstepsquared nL \left(\frac{1}{M}\sum\limits_{m=1}^{M}\sigma^2_{*,m} + n \sigma_{*}^2 \right)
	\end{align*}
\end{theorem}
As can be seen, we get an additional source of variance which is proportional to $\sstep$ and $\sigma_{*}^2$. This term represents the variance of Partial Participation. Since partial participation has an \gls{SGD}-type structure, we have that variance is proportional to the first order of the server-side stepsize. 

\subsection{Non-convex regime}
Finally, we provide  guarantees in the non-convex case.  

\begin{theorem}\label{thm:PP-NC}
Let the assumption of smoothness hold. Let $\delta^0 = f(x^{0}) - f^\star$ and $\Delta^{\star}_{m} = \frac{1}{n}\sum\limits_{i=1}^{n}(f^\star - f^{\star}_{m,i})$. Let $\cstep \leq \frac{1}{2nL}$ and $\sstep \leq \frac{1}{4L}$. Then for iterates $x^t$ of Algorithm~\ref{alg:pp-jumping}, we have
\begin{align*}
	\min _{t = 0, \ldots, T-1}  \mathbb{E}\left[\left\|\nabla f\left(x^{t}\right)\right\|^{2}\right] &\leq  \squeeze 8L^2\sstep  \frac{M-C}{C\max\{M-1,1\}} \Delta^\star\\
	& \squeeze +6\cstepsquared  n L^3 \left(\frac{1}{M}\sum\limits_{m=1}^{M}\Delta^{\star}_{m}+n\Delta^\star\right) \\
	&\squeeze+\frac{4\left(1+\frac{2L^2\sstepsquared(M-C)}{C\max\left\lbrace M-1,1 \right\rbrace }+\frac{3}{2}\sstep\cstepsquared n^2L^3\right)^T}{\sstep  T} \delta^0.
\end{align*}

\end{theorem}

Similarly to the analysis in the full participation case, we use $\Delta^{\star}_{m}$ and $\Delta^\star$ instead of $\sigma_{*,m}^2$ and $\sigma_{*}^2$, since the point of the minimizer cannot be defined. 

{\bf Client and server stepsizes.}
Theorems~\ref{thm:PP-SC}, \ref{thm:PP-C} and \ref{thm:PP-NC} suggest that the server can use the large $\cO(1/L)$ stepsize, where $L$ is the Lipschitz constant of the gradient of $f$. In all regimes, it is optimal for the client stepsize $\cstep$ to be small, which completely eliminates the second of the three terms in the complexity bounds, which controls the price one pays due to {\em data heterogeneity}. 

{\bf Partial participation.}
Notice that if the cohort size is equal to $M$, then $ \frac{M-C}{C\max\{1,M-1\}}$ is equal to $0$, and this means that the last (third) term in all our complexity results disappears. The last term can thus be interpreted  as the price we pay for partial participation.  While we can reduce the variance of \algname{\gls{RR}} and the client drift by decreasing $\cstep$, we cannot make the variance due to Partial Participation  arbitrarily small, since it depends on $\sstep$. 

{\bf Comparison with existing rates.} In Table~\ref{tab:compare_with_others} we compare our results in the strongly convex and non-convex regimes with selected existing results. 

\section{Benefits of Small Server Stepsize}

Our analysis shows that small client stepsizes can control variance. It turns out that using small client stepsizes means that we do not have any benefits from local steps. However, in some cases, our analysis shows that using a small server stepsize and large client stepsizes can be beneficial and it means that we gain from using local steps. The advantage of local steps is obtained in the case of data reshuffling \citep{mishchenko2022proximal}.  Moreover, the goal of learning is not obtaining the best value of the loss function, but the performance of the model. In recent papers, it was shown that large stepsizes are a better option in terms of generalization~\citep{smith2020origin}. 

Next, we introduce analysis for the case when each $f_i$ is strongly convex.

\begin{theorem}\label{th:small_alpha}
	Assume that all losses $f_{m,i}$ are $L$-smooth and $\mu$-strongly convex. Define $\alpha  = \frac{\sstep}{\cstep n}$. Let $\cstep\leq \frac{1}{L}$ and $0\leq \alpha <1$. Then, for iterates $x^t$ generated by Algorithm~\ref{alg:pp-jumping}, we have 
	\begin{align*}
		\mathbb{E}\left[\left\|x^{T}-x^{\star}\right\|^{2}\right] &\squeeze \leq\left(1-\alpha+\alpha(1-\cstep \mu)^{n}\right)^{T}\left\|x^{0}-x^{\star}\right\|^{2}\\
		&\squeeze +\frac{\alpha}{(1-\alpha)\left(1-(1-\cstep \mu)^{n}\right)} \cstepsquared  \frac{M-C}{C\max \left\lbrace M-1,1 \right\rbrace}\sigma_{*}^{2} \\
		&\squeeze +2 \cstepcubed \sigma_{\operatorname{rad}}^{2} \frac{1}{1-(1-\cstep \mu)^{n}} \sum\limits_{i=0}^{n-1}(1-\cstep \mu)^{i},
	\end{align*}
	\end{theorem}
where $\sigma^2_{\text{rad}}$ is introduced in \citep{mishchenko2022proximal} and it corresponds to the variance of the Random Reshuffling method. The upper bound depends on $\alpha$ in a nonlinear way, so the optimal value of $\alpha$ would often lie somewhere in the interval $(0, 1)$. Furthermore, the last term does not change with $\alpha$, so the optimal value $\alpha^\star$ of $\alpha$ is completely determined by the first two terms.

Let us derive optimal $\alpha^\star$ under some approximations. In particular, for ill-conditioned problems where $\mu$ is sufficiently small, it holds $(1-\cstep \mu)^n \approx 1 - \cstep \mu n$. Ignoring the last term in the upper bound of \Cref{th:small_alpha}, which does not affect the value $\alpha^\star$, and using $\frac{1}{1-\alpha}\le 2$ for $\alpha\le \frac{1}{2}$, we simplify the upper bound to
\begin{align*}	
		\mathbb{E}\left[\left\|x^{T}-x^{\star}\right\|^{2}\right] 	&\leq (1-\alpha+\alpha(1-\cstep\mu n))^T\|x^{0} - x^{\star}\|^2\\
	& + \frac{2\alpha\cstepsquared }{1-(1-\cstep\mu n)} \frac{M-C}{C\max \left\lbrace M-1,1 \right\rbrace}\sigma_{*}^{2}  \\
	& \leq (1-\alpha\cstep \mu n)^T\|x^{0}-x^{\star}\|^2 + \frac{2\alpha\cstep}{\mu n} \frac{M-C}{C\max \left\lbrace M-1,1 \right\rbrace}\sigma_{*}^{2} .
\end{align*}
To have this upper bound smaller than some $\varepsilon\ge 0$, we need to use $\alpha = \cO\left(\frac{n \varepsilon C}{\cstep \sigma_\star^2}\right)$ and $T= \cO(\frac{1}{\alpha\cstep\mu n}\log\frac{1}{\varepsilon})$, where we ignore constants unrelated to $\alpha, \cstep, \varepsilon, \mu$ and $n$. Thus, the server stepsize $\sstep=\alpha \cstep n$ should ideally be $\sstep=\cO\left(\frac{C \varepsilon}{\sigma_\star^2} \right)$. In other words, it is better to decrease $\sstep$ if only a small subset of clients is used and the variance of Partial Participation $ \frac{M-C}{C\max \left\lbrace M-1,1 \right\rbrace}\sigma_{*}^{2} $ is large.

\section{Experiments} \label{sec:experiments_ch_5}

\begin{figure*}[t!]
	\centering
	\begin{tabular}{cc}
		\includegraphics[scale=0.22]{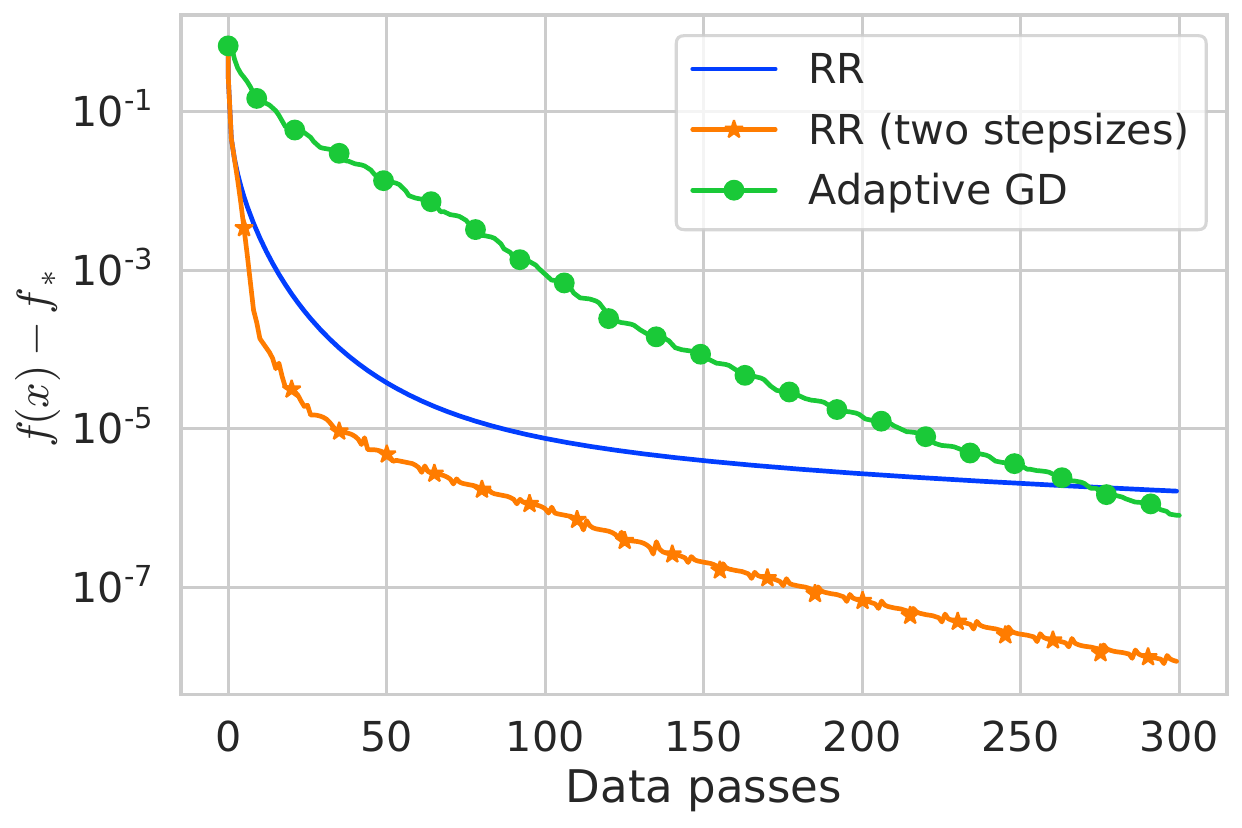}\includegraphics[scale=0.22]{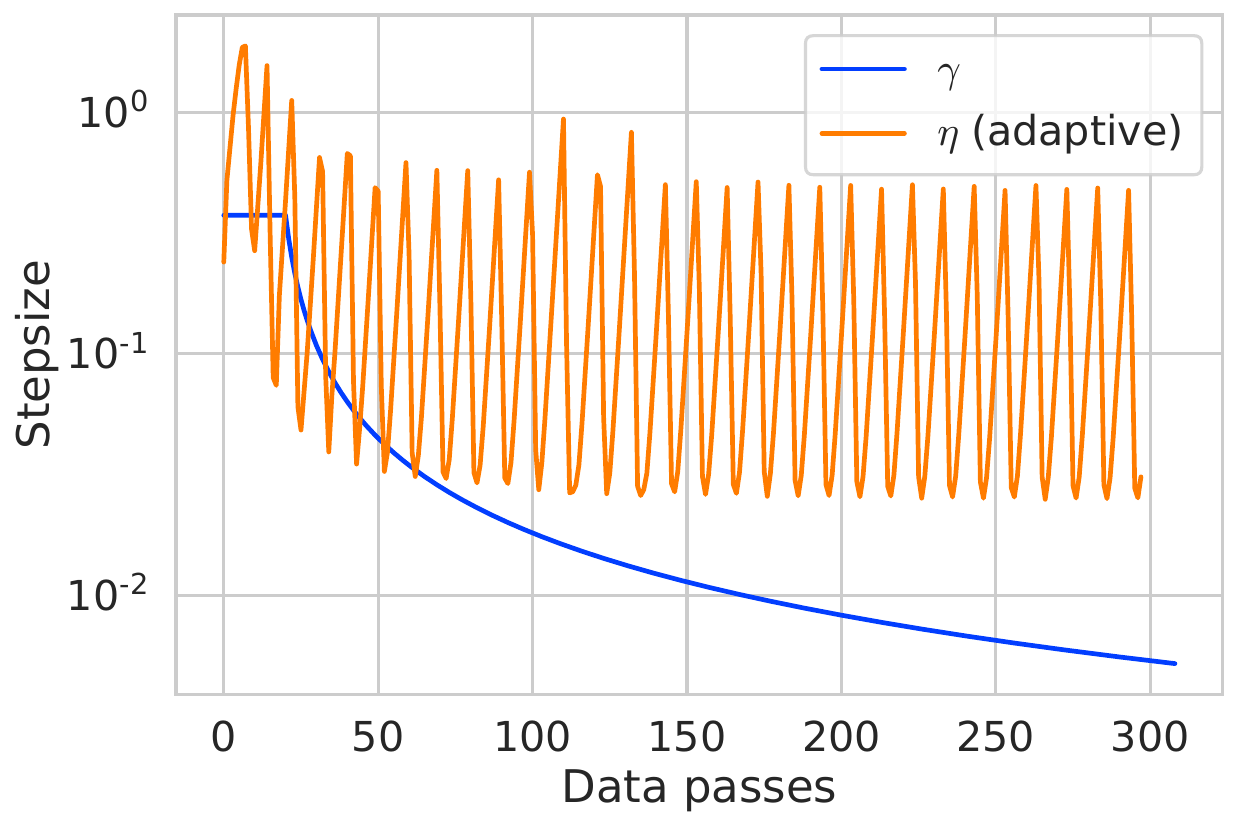}
		\includegraphics[scale=0.22]{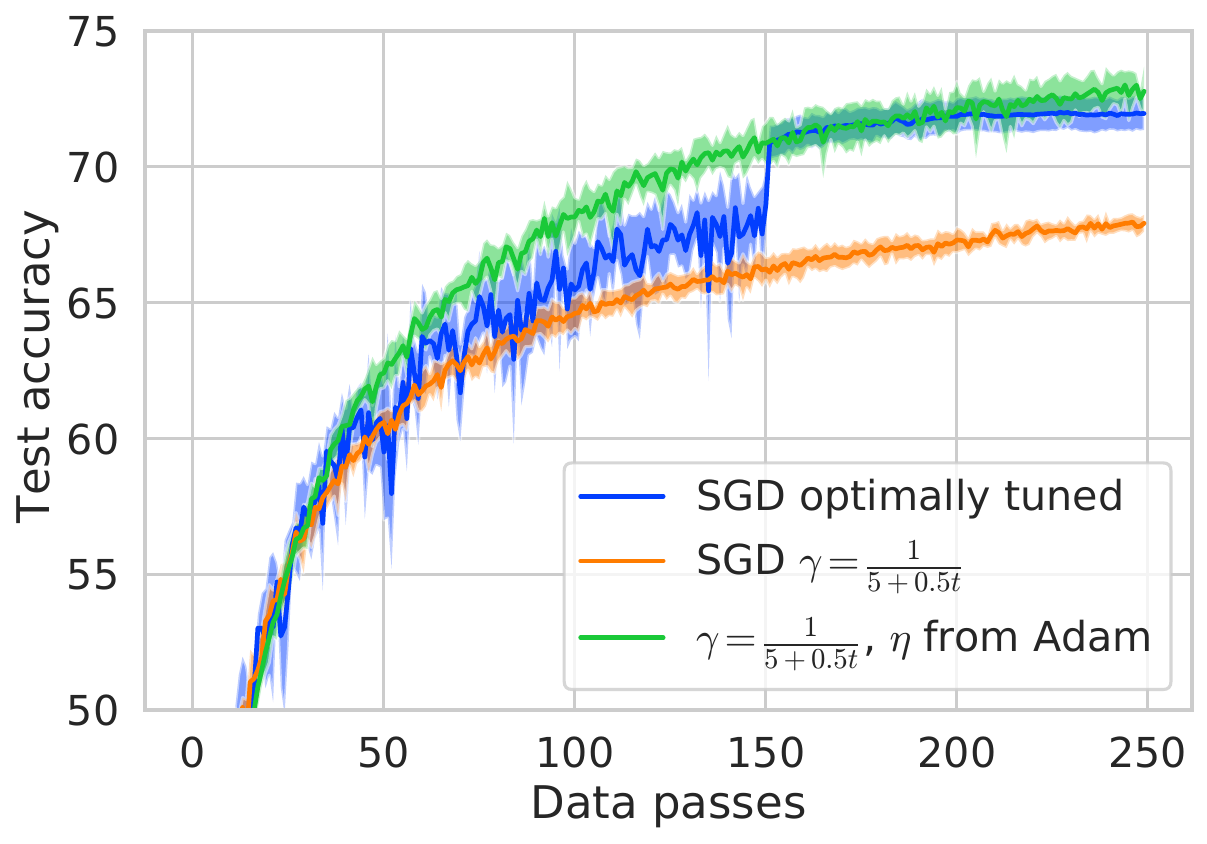}
	\end{tabular}	
	\caption{{\bf Left and middle:} We compare running standard Random Reshuffling (\algname{RR}), adaptive gradient descent (\algname{Adaptive GD}), and the combination of \algname{RR} with outer adaptive stepsize (\algname{Nastya}) (\algname{RR} (two stepsizes)) on logistic regression. As one can see, the variant with two stepsizes outperforms both of them and does not require more hyper-parameters than \algname{RR}, and the middle plot shows the exact values of $\cstep$ and $\sstep$. {\bf Right:} The right plot shows the training curves of LeNet on CIFAR-10 with minibatch size 1024, where we compare carefully tuned \algname{SGD} (blue) to poorly tuned \algname{SGD} (orange) and show that using \algname{Adam} optimizer with stepsize $10^{-2}$ after each data pass can significantly improve the poorly tuned version.} 
	\label{fig:logistic_and_lenet}
\end{figure*}

To showcase the speed-up that can be obtained from the server-side stepsizes, we run a toy experiment in the single-node setup, i.e., we consider standard minimization of a finite-sum. We combine the local passes over the data with the adaptive estimation of smoothness proposed by~\citep{malitsky2019adaptive}. We run our experiment on $\ell_2$-regularized logistic regression with the `mushrooms' dataset from LibSVM~\citep{chang2011libsvm}. The results are reported in \Cref{fig:logistic_and_lenet}.

We use the standard LeNet architecture, which is a 5-layer convolutional neural network, implemented in PyTorch~\citep{paszke2017automatic} and train them to classify images from the CIFAR-10 dataset~\citep{cifar} with cross-entropy loss. At each iteration, we use a minibatch of size 1024. For the tuned \algname{\gls{SGD}}, we start with stepsize 0.2 and divide by 10 at epochs 150 and 200. For the other version, we take \algname{\gls{SGD}} with stepsize 0.2 and decrease as $\cO(\frac{1}{t})$, where $t$ is the epoch number.

For our method, we treat the full sum of gradients over epoch as an approximation of full gradient and use \algname{Adam} with stepsize 0.01 to improve this update. We can see from \Cref{fig:logistic_and_lenet} that by applying \algname{Adam}, we can improve the performance of \algname{\gls{SGD}} with decreasing stepsize. At the same time, applying it to the tuned stepsize schedule only made the results much worse, so we do not report that line. This highlights that adaptive outer stepsizes are helpful when the base stepsize $\cstep$ is not chosen well, which is in line with our theory.

\chapter{Integrating Gradient Compression with Random Reshuffling and Local Computation}
\label{chapter6}
\thispagestyle{empty}

\section{Introduction}

Distributed learning plays a crucial role in the training of modern \gls{DL} models since distributed approaches are able to significantly reduce training time \citep{goyal2017accurate, you2019large}. Moreover, distributed methods are mandatory for such applications as Federated learning (\gls{FL}) \citep{FedLearn2016, mcmahan2017communication}, where multiple nodes connected over a network collaborate on a learning task. Each node possesses its own dataset and cannot share this data with other nodes or a central server. As a result, algorithms for Federated Learning often rely on local computation and lack access to the entire dataset of training examples. Federated learning finds applications in diverse fields, including language modeling for mobile keyboards~\citep{liu21_feder_learn_meets_natur_languag_proces}, healthcare~\citep{antunes2022federated}, and wireless communications~\citep{yang22_fed_learn_6g}. Its applications extend to various other domains~\citep{kairouz2019advances}.

Distributed learning tasks are often solved through \emph{empirical-risk minimization} (\gls{ERM}), where the \(m\)-th device contributes an empirical loss function \(f_m (x)\) representing the average loss of model \(x\) on its local dataset, and our goal is to then minimize the average loss over all the nodes:
\begin{align}
\label{eq:erm-opt-orig}
\min_{x \in \mathbb{R}^d} \left[ f(x) \eqdef \frac{1}{M} \sum_{m=1}^M f_m (x) \right],
\end{align}
where the function \(f\) represents the average loss. Every \(f_m\) is an average of sample loss functions \(f_{m,i}\) each representing the loss of model \(x\) on the \(i\)-th datapoint on the \(m\)-th clients' dataset: that is for each \(m \in \{1, 2, \ldots, M\}\) we have
\begin{align*}
f_m (x) \eqdef \frac{1}{n_m} \sum_{i=1}^{n_m} f_{m,i} (x).
\end{align*}
For simplicity we shall assume that the datasets on all clients are of equal size: \(n_1 = n_2 = \ldots = n_M = n\), though this assumption is only for convenience and our results easily extend to the case when clients have datasets of unequal sizes. Thus, our optimization problem is
\begin{align}\label{eq:erm-opt}
\min_{x \in \mathbb{R}^d} \left[ f(x) = \frac{1}{nM} \sum_{m=1}^M \sum_{i=1}^n f_{m,i} (x) \right].
\end{align}
Because \(d\) is often very large in practice, the dominant paradigm for solving~\eqref{eq:erm-opt} relies on first-order (gradient) information. Federated learning algorithms have access to two key primitives: (a) local computation, where for a given model \(x \in \mathbb{R}^d\) we can compute stochastic gradients \(\nabla f_{m,i} (x)\) locally on client \(m\), and (b) communication, where the different clients can exchange their gradients or models with a central server.

\subsection{Communication compression}

In practice, communication is more expensive than local computation~\citep{kairouz2019advances}, and as such one of the chief concerns of algorithms for distributed learning is communication efficiency. Algorithms for distributed/Federated Learning have thus focused on achieving communication efficiency, with one common ingredient being the use of \emph{gradient compression}, where each client sends a compressed or quantized version of their update instead of the full update vector, potentially saving communication bandwidth by sending fewer bits over the network. There are many operators that can be used for compressing the update vectors: stochastic quantization~\citep{alistarh2017qsgd}, random sparsification~\citep{wangni18_sparsification, stich2018sparsified}, and others~\citep{tang20_commun_effic_distr_deep_learn}. In this work we consider compression operators satisfying the following assumption:

\begin{assumption}\label{asm:quantization_operators}
A compression operator is an operator \(\cQ: \mathbb{R}^d \rightarrow \mathbb{R}^d\) such that for some \(\omega > 0\), the relations
\begin{align*}
\ec{\cQ(x)} = x \qquad \text {and} \qquad \ecn{\cQ(x) - x} \leq \omega \sqn{x} \quad \text{hold for \(x \in \mathbb{R}^d\).}
\end{align*}

\end{assumption}

Unbiased compressors can reduce the number of bits clients communicate per round, but also increase the variance of the stochastic gradients used, slowing down overall convergence, see e.g.~\cite[Theorem 5.2]{khirirat2018distributed} and \cite[Theorem 1]{stich20_commun_compr_distr_optim_heter_data}. By using control iterates, in work \citep{mishchenko2019distributed} \gls{DIANA} was developed---an algorithm that can reduce the variance  due to gradient compression with unbiased compression operators, and thus ensure fast convergence. \gls{DIANA} has been extended and analyzed in many settings~\citep{DIANA2, stich20_commun_compr_distr_optim_heter_data, safaryan21_smoot_matric_beat_smoot_const} and forms an important tool in our arsenal for using gradient compression.

\subsection{Random reshuffling}

Despite the importance of addressing the communication bottleneck, local computations also significantly affect the training. For simplicity, consider the $1$-node scenario. In this case, the update rule of the standard work-horse method in stochastic optimization -- stochastic gradient descent (\algname{\gls{SGD}}) \citep{robbins1951stochastic} -- can be written as follows: $x^{t+1} = x^{t} - \gamma \nabla f_i (x^{t})$,  
where \(i\) is sampled from \(\{1, \ldots, n\}\) uniformly at random. This procedure thus uses \emph{with-replacement sampling} in order to select the stochastic gradient used at each step from the dataset. However, in the training of \gls{DL} models, {\em without-replacement sampling} is used much more often: that is, at the beginning of each \emph{epoch} we choose a permutation \(\pi_1, \pi_2, \ldots, \pi_n\) of \(\{1, 2, \ldots, n\}\) and do the \(i\)-th update using the \(\pi_i\)-ith gradient: $x_{i+1}^{t} = x^{t}_i - \gamma \nabla f^{\pi_i} (x^{t}_i)$. 
Without-replacement sampling \algname{\gls{SGD}}, also known as Random Reshuffling (\algname{\gls{RR}}) \citep{bottou2009curiously}, typically achieves better asymptotic convergence rates compared to with-replacement \gls{SGD} and can improve upon it in many settings as shown by recent theoretical progress~\citep{MKR2020rr, ahn2020sgd, rajput20_closin_conver_gap_sgd_without_replac, safran2021random}. While with-replacement \algname{\gls{SGD}} achieves an error proportional to \(\mathcal{O} \br{\frac{1}{T}}\) after \(T\) steps~\citep{stich19_unified_optim_analy_stoch_gradien_method}, Random Reshuffling achieves an error of \(\mathcal{O} \br{\frac{n}{T^2}}\) after \(T\) steps, faster than \gls{SGD} when the number of steps \(T\) is large.

\subsection{Can compression and reshuffling be friends?}

As we described earlier, Random Reshuffling and communication compression are two important tools for training modern \gls{DL} models, and both techniques are relatively well understood. However, there are no papers that study Random Reshuffling and communication compression in combination. This leads us to the natural question: \emph{how to combine these techniques to improve the convergence speed of existing distributed methods.}

\subsection{Contributions}

\begin{table*}[t]
    \centering
    \small
    \caption{ Summary of known and new complexity results for solving distributed finite-sum optimization problem~\eqref{eq:erm-opt}. Column ``Complexity" indicates the number of communication rounds to find a solution with accuracy $\varepsilon > 0$. Column ``\gls{RR}?" shows whether an algorithm uses {\it Random Reshuffling}, ``C?" indicates whether a method applies the compression of gradients or differences between the gradients and also whether a method applies compression of gradients or gradient differences for communication, ``H?" means independence from the constant of data heterogeneity in the complexity, ``CVX?" indicates whether each loss on the $i$-th datapoint on the $m$-th client is convex, but not strongly convex.}
    \label{tab:comparison_of_rates}
    \begin{threeparttable}
        \begin{tabular}{c c c c c c c}
        \hline
         Method & Complexity \tnote{\color{blue}(4)}  & \gls{RR}? &C? & H?& CVX?\\
        \hline
        \begin{tabular}{c}
             \algname{\gls{SGD}} \\ {[\citenum{gower2019sgd}]}
        \end{tabular}  & $\kappa + \frac{\sigma_{\star,n}^2}{\mu^2 \varepsilon}$& \xmark &\xmark & \xmark\tnote{\color{blue}(1)} & \cmark  \\
        
        \begin{tabular}{c}
             \algname{\gls{RR}} \\ {[\citenum{MKR2020rr}]}
        \end{tabular}  & $\widetilde{\kappa} + \frac{\sigma_{\star,n}}{\widetilde{\mu}}\sqrt{\frac{\widetilde{\kappa} n}{\varepsilon}}$& \cmark &\xmark & \xmark\tnote{\color{blue}(1)} & \xmark \\
        \begin{tabular}{c}
             \algname{\gls{RR}} \\ {[\citenum{MKR2020rr}]}
        \end{tabular}  & $n\kappa + \frac{\sigma_{\star,n}}{\widetilde{\mu}}\sqrt{\frac{\kappa n}{\varepsilon}}$& \cmark &\xmark & \xmark\tnote{\color{blue}(1)} & \cmark \\
        \hline
        \begin{tabular}{c}
             \algname{QSGD} \\ {[\citenum{gorbunov2020unified}]}
        \end{tabular}  & $\left(1+\frac{\omega}{M}\right)\kappa + \frac{\omega}{M}\frac{\sigma_{\star}^2 + \zeta^2_{\star}}{\mu^2 \varepsilon} + \frac{\sigma_{\star}^2}{M \mu^2 \varepsilon}$\tnote{\color{blue}(2)}  & \xmark &\cmark  & \xmark &\cmark  \\
         \cellcolor{bgcolor} \begin{tabular}{c}
             \gls{Q-RR} \\ Corollary~\ref{cor:complexity_q_rr}
        \end{tabular}  &  \cellcolor{bgcolor} $\left(1+\frac{\omega}{M}\right)\widetilde\kappa + \frac{\omega}{M}\frac{\sigma_{\star}^2 + \zeta^2_{\star}}{\widetilde\mu^2 \varepsilon} + \frac{\sigma_{\star,n}}{\widetilde{\mu}}\sqrt{\frac{\widetilde{\kappa} n}{\varepsilon}}$  &  \cellcolor{bgcolor}\cmark & \cellcolor{bgcolor} \cmark & \cellcolor{bgcolor} \xmark & \cellcolor{bgcolor} \xmark \\
        \cellcolor{bgcolor} \begin{tabular}{c}
             \gls{Q-RR} \\ Corollary~\ref{cor_convergence_new_Q_RR_cvx_case}
        \end{tabular}  &  \cellcolor{bgcolor} $\left(n+\frac{\omega}{M}\right)\kappa + \frac{\omega}{M}\frac{\sigma_{\star}^2 + \zeta^2_{\star}}{\mu^2 \varepsilon} + \frac{\rho_{\star}}{\mu}\sqrt{\frac{\kappa n}{\varepsilon}}$\tnote{\color{blue}(3)}  &  \cellcolor{bgcolor}\cmark & \cellcolor{bgcolor} \cmark &   \cellcolor{bgcolor} \xmark & \cellcolor{bgcolor} \cmark \\
        \begin{tabular}{c}
            \gls{DIANA} \\ {[\citenum{mishchenko2019distributed}]}
        \end{tabular}  & $\left(1+\frac{\omega}{M}\right)\kappa + \frac{\omega}{M}\frac{\sigma_{\star}^2}{\mu^2 \varepsilon} + \frac{\sigma_{\star}^2}{M \mu^2 \varepsilon}$  & \xmark &\cmark & \cmark & \cmark \\
         \cellcolor{bgcolor} \begin{tabular}{c}
            \gls{DIANA-RR} \\ Corollary~\ref{cor:complexity_diana_rr}
        \end{tabular}  & \cellcolor{bgcolor} $n(1+\omega)+\left(1+\frac{\omega}{M}\right)\widetilde\kappa + \frac{\sigma_{\star,n}}{\widetilde{\mu}}\sqrt{\frac{\widetilde{\kappa} n}{\varepsilon}}$  & \cellcolor{bgcolor} \cmark &  \cellcolor{bgcolor}\cmark & \cellcolor{bgcolor} \cmark& \cellcolor{bgcolor} \xmark  \\
        \cellcolor{bgcolor} \begin{tabular}{c}
            \gls{DIANA-RR} \\ Corollary~\ref{cor_convergence_DIANA_RR_cvx_case}
        \end{tabular}  & \cellcolor{bgcolor} $n(1+\omega)+\left(n+\frac{\omega}{M}\right)\kappa + \frac{\sigma_{\star,n}}{\mu}\sqrt{\frac{\kappa n}{\varepsilon}}$  & \cellcolor{bgcolor} \cmark  & \cellcolor{bgcolor}\cmark & \cellcolor{bgcolor} \cmark& \cellcolor{bgcolor} \cmark  \\
        \hline
    \end{tabular}
    \begin{tablenotes}
        {\scriptsize 
        \item [{\color{blue}(1)}] In the case of \algname{\gls{SGD}}, \algname{\gls{RR}} we use ~\xmark~ in ``H?" to show that the complexity of these methods is provided in the non-distributed setup.
        \item [{\color{blue}(2)}] The following inequality is useful for the comparison of complexities: $\sigma^2_{\star,n} \leq \sigma^2_{\star}$.
        \item [{\color{blue}(3)}] We denote $\rho^2_{\star} = \frac{\omega}{M}(\sigma_{\star}^2 + \zeta^2_{\star})+\sigma_{\star,n}^2 $.
        \item [{\color{blue}(4)}] Notation: $\kappa = \nicefrac{L_{\max}}{\mu} $ and $\widetilde\kappa = \nicefrac{L_{\max}}{\widetilde\mu}$ are the condition number of the problem ~\eqref{eq:erm-opt}, where $L_{\max} = $ Lipschitz constant, $\mu$ and $\widetilde \mu$ are the strong convexity constants of $f$ and $f_{m,i}$ respectively; variances at the solution point $x^{\star}$:  $\sigma^2_{\star} = \frac{1}{Mn}\sum\limits^M_{m=1}\sum\limits^n_{i=1}\|\nabla f_{m,i}(x^{\star}) - \nabla f_m(x^{\star}) \|^2$ and $\sigma^2_{\star,n} = \frac{1}{n}\sum\limits^n_{i=1}\|\nabla f_i(x^{\star})\|^2$; heterogeneity constant $\zeta^2_{\star} = \frac{1}{M}\sum\limits^M_{m=1}\|\nabla f_m(x^{\star})\|^2$. The results of this paper are highlighted in blue.
        }
    \end{tablenotes}
    \end{threeparttable}
\end{table*}

In this paper, we aim to develop methods for distributed and Federated Learning that combine gradient compression and random reshuffling. While each of these techniques can aid in reducing the communication complexity of distributed optimization, their combination is under-explored. Thus, our goal is to design methods that improve upon existing algorithms in convergence rates and in practice. We summarize our contributions as follows.

\begin{itemize}[leftmargin=*]

\item[$\diamond$]\textbf{The issue: na\"ive combination has no improvements.} As a natural step towards our goal, we  propose and study a new algorithm, \gls{Q-RR} (Algorithm~\ref{alg_new_Q_RR}), that combines random reshuffling with gradient compression at every communication round. However, for \gls{Q-RR} our theoretical results do not show any improvement upon \algname{QSGD} when the compression level is reasonable (see Table~\ref{tab:comparison_of_rates}). Moreover, we observe similar performance of \gls{Q-RR} and \algname{QSGD} in various numerical experiments. Therefore, we conclude that this phenomenon is not an artifact of our analysis but rather an issue of \gls{Q-RR}: communication compression adds an additional noise that dominates the one coming from the stochastic gradients sampling.

\item[$\diamond$]\textbf{The remedy: reduction of compression variance.} To remove the additional variance added by the compression and unleash the potential of Random Reshuffling in distributed learning with compression, we propose \gls{DIANA-RR} (Algorithm~\ref{alg_new_RR_DIANA}), a combination of \gls{Q-RR} and the \gls{DIANA} algorithm. We derive the convergence rates of the new method and show that it improves upon the convergence rates of \gls{Q-RR}, \algname{QSGD}, and \gls{DIANA} (see Table~\ref{tab:comparison_of_rates}). We point out that to achieve such results we use $n$ shift-vectors per worker in \gls{DIANA-RR} unlike \gls{DIANA} that uses only $1$ shift-vector.

\item[$\diamond$]\textbf{Extensions to the local steps.} Inspired by the \gls{NASTYA} algorithm \citep{malinovsky2023server}, we propose a variant of \gls{NASTYA}, \gls{Q-NASTYA} (Algorithm~\ref{alg:Q_NASTYA}), that na\"ively mixes quantization, local steps with random reshuffling, and uses different local and server stepsizes. Although it improves in per-round communication cost over \gls{NASTYA}, similar to \gls{Q-RR}, we show that \gls{Q-NASTYA} suffers from added variance due to gradient quantization. To overcome this issue, we propose another algorithm, \gls{DIANA-NASTYA} (Algorithm~\ref{alg:diana-nastya}), that adds \gls{DIANA}-style variance reduction to \gls{Q-NASTYA} and removes the additional variance.
\end{itemize}

Finally, to illustrate our theoretical findings we conduct experiments on federated logistic regression tasks and on distributed training of neural networks.

\section{Algorithms and Convergence Theory}

We will primarily consider the setting of strongly-convex and smooth optimization. We assume that the average function \(f\) is strongly convex:

\begin{assumption}\label{asm:sc_general_f}
Function $f: \R^d \rightarrow \R$ is $\mu$-strongly convex, i.e., for all \(x, y \in \mathbb{R}^d\),
\begin{equation}
\label{convexity}
    f(x) - f(y) - \la\nabla f(y), x - y\ra \geq \frac{\mu}{2}\|x-y\|^2,
\end{equation}
and functions $f_{1,i}, f_{2,i}, \dots, f_{M,i}: \R^d \rightarrow \R$  are convex for all $i = 1,\dots, n$.
\end{assumption}

Examples of objectives satisfying Assumption~\ref{asm:sc_general_f} include \(\ell_2\)-regularized linear and logistic regression. Throughout the paper, we assume that \(f\) has the unique minimizer \(x^{\star} \in \mathbb{R}^{d}\). We also use the assumption that each individual loss \(f_{m,i}\) is smooth, i.e.\ has Lipschitz-continuous first-order derivatives:

\begin{assumption}\label{asm:lip_max_f_m}
Function $f_{m,i}: \R^d \rightarrow \R$ is $L_{m,i}$-smooth for every $i \in [n]$ and $m \in [M]$, i.e., for all \(x, y \in \mathbb{R}^d\) and for all \(m \in [M]\) and \(i \in [n]\),
\begin{equation}
    \|\nabla f_{m,i}(x) - \nabla f_{m,i}(y)\|\leq L_{m,i}\|x - y\|.
\end{equation}
We denote the maximal smoothness constant as $L_{\max} \eqdef \max_{m,i}L_{m,i}$.
\end{assumption}

For some methods, we shall additionally impose the assumption that \emph{each} function is strongly convex:

\begin{assumption}
\label{asm:sc_each_f_m}
Each function $f_{m,i}: \R^d \rightarrow \R$ is $\widetilde{\mu}$-strongly convex.
\end{assumption}

The \emph{Bregman divergence} associated with a convex function \(h\) is defined for all \(x, y \in \mathbb{R}^d\) as $$D_h (x, y) \eqdef h(x) - h(y) - \ev{\nabla h(y), x-y}.$$ Note that the inequality \eqref{convexity} can be written as $D_f(x,y) \geq \frac{\mu}{2}\|x - y\|^2$.

\subsection{Algorithm Q-RR}

The first method we introduce is \gls{Q-RR} (Algorithm~\ref{alg_new_Q_RR}). \gls{Q-RR} is a straightforward combination of distributed random reshuffling and gradient quantization. This method can be seen as the stochastic without-replacement analogue of the distributed quantized gradient method \citep{khirirat2018distributed}.

\begin{algorithm}[t]
\caption{\algname{Q-RR:} Distributed Random Reshuffling with Quantization}
\label{alg_new_Q_RR}
	\begin{algorithmic}[1]
		\REQUIRE $x^0$ -- starting point, $\gamma > 0$ -- stepsize
	    \FOR{$t =0,1,\dots, T-1$}
		    \STATE Receive $x^t$ from the server and set $x_0^{t} = x^{t}$
		    \STATE Sample random permutation of $[n]$: $\pi_m = (\pi^0_m, \dots, \pi^{n-1}_m)$
		    \FOR{$i = 0, 1,\dots, n-1$}
		    	\FOR{$m = 1,\dots, M$ in parallel}
		    	\STATE Receive $x^{t}_{i}$ from the server, compute and send $\cQ\left(\nabla f_{m, \pi^{i}_m}(x_{i}^{t})\right)$ back
		    	\ENDFOR
		        \STATE Compute and send $x_{i+1}^{t} = x_{i}^{t} - \gamma\frac{1}{M}\sum^M_{m=1} \cQ\left(\nabla f_{m, \pi^{i}_m}(x_{i}^{t})\right)$ to the workers
		    \ENDFOR
		    \STATE $x^{t+1} = x_n^t$
    	\ENDFOR
    	\ENSURE $x^T$
	\end{algorithmic}
\end{algorithm}

We shall use the notion of \emph{shuffling radius} defined in \citep{mishchenko2022proximal} for the analysis of distributed methods with random reshuffling:

\begin{definition}
\label{def:shuffling_radius}
Define the iterate sequence \(x_{i+1}^{\star} = x_{i}^{\star} -\frac{\gamma}{M}\sum^M_{m=1}\nabla f^{\pi^i_m}_m(x^{\star})\). Then the shuffling radius is the quantity
$$\sigma^2_{\text{rad}} \eqdef \max_{i}\left\{\frac{1}{\gamma^2 M}\sum^M_{m=1} \mathbb{E}[ D_{f_{m,\pi^i}}(x^{\star}_{i}, x^{\star})]\right\}.$$
\end{definition}

We provide clarifications regarding this term in Appendix~\ref{appendix:shuffling_radius}. To compare our subsequent results with known ones, we introduce bounds on the shuffling radius. The following lemma demonstrates that these bounds are independent of the stepsize \(\gamma\), even though \(\gamma\) is used in Definition~\ref{def:shuffling_radius}.

\begin{lemma}[\citep{MKR2020rr}]\label{lem:shuffling_radius}
    \label{lem:bounds_on_shuffling_radius}
    Let Assumptions \ref{asm:lip_max_f_m}, \ref{asm:sc_each_f_m} hold. Then the shuffling radius $\sigma^2_{\text{rad}}$ satisfies the following inequality 
    \begin{equation*}
        \frac{\widetilde\mu n}{8}\sigma^2_{\star,n} \leq \sigma^2_{\text{rad}} \leq \frac{L_{\max}n}{4} \sigma^2_{\star,n},
    \end{equation*}
    where $\sigma^2_{\star,n} \eqdef \frac{1}{n}\sum\limits^n_{i=1}\|\nabla f_i(x^{\star})\|^2$, and $f_i = \frac{1}{M}\sum\limits^M_{m=1}f_{m,i}$.
\end{lemma}

We now state the main convergence theorem for Algorithm~\ref{alg_new_Q_RR}:

\begin{theorem}
    \label{th_conv_new_rr_q}
    Let Assumptions~\ref{asm:quantization_operators}, \ref{asm:lip_max_f_m}, \ref{asm:sc_each_f_m} hold and let the stepsize satisfy  $
    	0 < \gamma \leq \frac{1}{\left( 1+2\frac{\omega}{M} \right)L_{\max}}.$
    Then, for all $T \geq 0$ the iterates produced by \gls{Q-RR} (Algorithm~\ref{alg_new_Q_RR}) satisfy
    \begin{equation}
    	\mathbb{E}\|x^T-x^{\star}\|^2 \leq \left(1-\gamma\widetilde{\mu}\right)^{nT}\|x^0-x^{\star}\|^2 + \frac{2\gamma^2\sigma^2_{\text{rad}}}{\widetilde{\mu}} + \frac{2\gamma\omega}{\widetilde{\mu} M}(\zeta_{\star}^2 + \sigma_{\star}^2),\label{eq:QRR_bad_term}
    \end{equation}
    where $\zeta^2_{\star} \eqdef \frac{1}{M}\sum\limits_{m=1}^M\|\nabla f_m (x^{\star})\|^2,$ and $\sigma_{\star}^2 \eqdef \frac{1}{Mn}\sum\limits_{m=1}^M\sum\limits_{i=1}^n \|\nabla f_{m,i}(x^{\star}) - \nabla f_m(x^{\star})\|^2.$
\end{theorem}

All proofs are relegated to the appendix. By choosing the stepsize \(\gamma\) properly, we can obtain the communication complexity (number of communication rounds) needed to find an \(\varepsilon\)-approximate solution as follows:

\begin{corollary}
\label{cor:complexity_q_rr}
	Under the same conditions as Theorem~\ref{th_conv_new_rr_q} and for Algorithm~\ref{alg_new_Q_RR}, there exists a stepsize \(\gamma > 0\) such that the number of communication rounds $nT$ to find a solution with accuracy $\varepsilon > 0$ (i.e. \(\mathbb{E}\left\|x^T - x^{\star}\right\|^2 \leq \epsilon\)) is equal to $\widetilde{\cO}\left( \left( 1+ \frac{\omega}{M} \right) \frac{L_{\max}}{\widetilde{\mu}} + \frac{\omega(\zeta^2_{\star} + \sigma_{\star}^2)}{M\widetilde{\mu}^2\varepsilon}+ \frac{\sigma_{\text{rad}}}{\sqrt{\widetilde{\mu}^3\varepsilon}}\right),\label{eq:qrr-complexity}$
where $\widetilde{\cO}(\cdot)$ hides constants and logarithmic factors.
\end{corollary}

The complexity of Quantized SGD (\algname{QSGD}) is~\citep{gorbunov2020unified}: $$
	\widetilde{\cO}\left(\left(1 + \frac{\omega}{M}\right)\frac{L_{\max}}{\mu} + \frac{\left(\omega\zeta_{\star}^2 + (1+\omega)\sigma_{\star}^2\right)}{M\mu^2 \varepsilon}\right).$$
For simplicity, let us neglect the differences between $\mu$ and $\widetilde{\mu}$. First, when $\omega = 0$ we recover the complexity of \algname{FedRR} \citep{mishchenko2022proximal} which is known to be better than the one of \algname{\gls{SGD}} as long as $\varepsilon$ is sufficiently small as we have $\nicefrac{n\mu\sigma_{\star,n}^2}{8}\leq\sigma_{\text{rad}}^2 \leq \nicefrac{nL\sigma_{\star,n}^2}{4}$ from Lemma~\ref{lem:bounds_on_shuffling_radius}. Next, when $M = 1$ and $\omega = 0$ (single node, no compression) our results recover the rate of \algname{\gls{RR}} \citep{MKR2020rr}.

However, it is more interesting to compare \gls{Q-RR} and \algname{QSGD} when $M > 1$ and $\omega > 1$, which is typically the case. In these settings, \gls{Q-RR} and \algname{QSGD} have \emph{the same complexity} since the $\cO(\nicefrac{1}{\varepsilon})$ term dominates the $\cO(\nicefrac{1}{\sqrt{\varepsilon}})$ one if $\varepsilon$ is sufficiently small. That is, the derived result for \gls{Q-RR} has no advantages over the known one for \algname{QSGD} unless $\omega$ is very small, which means that there is almost no compression at all. We also observe this phenomenon in the experiments.

The main reason for that is the variance appearing due to compression. Indeed, even if the current point is the solution of the problem ($x^{t}_{i} = x^\star$), the update direction $-\gamma\frac{1}{M}\sum^M_{m=1} \cQ\left(\nabla f_{m, \pi^{i}_m}(x_{i}^{t})\right)$  has the compression variance
\begin{align*}
	\mathbb{E}_{\cQ}\Bigg[\Bigg\|\frac{\gamma}{M}\sum^M_{m=1} \Big(&\cQ(\nabla f_{m, \pi^{i}_m}(x^{\star})) - \nabla f_{m, \pi^{i}_m}(x^{\star})\Big)\Bigg\|^2\Bigg] \leq \frac{\gamma^2\omega}{M^2}\sum^M_{m=1}\| \nabla f_{m, \pi^{i}_m}(x^{\star})\|^2.
\end{align*}
This upper bound is tight and non-zero in general. Moreover, it is proportional to $\gamma^2$ that creates the term proportional to $\gamma$ in \eqref{eq:QRR_bad_term} like in the convergence results for \algname{QSGD}/\algname{\gls{SGD}}, while the \algname{\gls{RR}}-variance is proportional to $\gamma^2$ in the same bound. Therefore, during the later stages of the convergence \gls{Q-RR} behaves similarly to \algname{QSGD} when we decrease the stepsize.

\subsection{Algorithm DIANA-RR}

To reduce the additional variance caused by compression, we apply \gls{DIANA}-style shift sequences \citep{mishchenko2019distributed,DIANA2}. Thus we obtain \gls{DIANA-RR} (Algorithm~\ref{alg_new_RR_DIANA}), which applies compression to the differences between the gradients and learnable shifts. Since the shifts are updated using the past gradients information, one can see \gls{DIANA-RR} as a method with compression of gradient differences. We notice that unlike \gls{DIANA}, \gls{DIANA-RR} has $n$ shift-vectors on each node.

\begin{algorithm}[t]
\caption{\gls{DIANA-RR}}
\label{alg_new_RR_DIANA}
	\begin{algorithmic}[1]
		\REQUIRE $x^0$ -- starting point, $\{h_{m,i}^0\}_{m,i=1,1}^{M,n}$ -- initial shift-vectors, $\gamma > 0$ -- stepsize,  $\alpha > 0$ -- stepsize for learning the shifts
	    \FOR{$t =0,1,\dots, T-1$}
	    		    \STATE Receive $x^{t}$ from the server and set $x^t_{m,0} = x^{t}$
            \STATE Sample random permutation of $[n]$: $\pi_m = (\pi^0_m, \dots, \pi^{n-1}_m)$
		    \FOR{$i = 0, 1,\dots, n-1$}
		        \FOR{$m = 1, 2,\dots, M$ in parallel}
					\STATE Receive $x^{t}_{i}$ from the server, compute and send $\cQ\left(\nabla f_{m, \pi^{i}_m}(x_{i}^{t}) - h_{m, \pi^i_m}^{t}\right)$ back
    		        \STATE Set $\hat{g}_{m,\pi^i_m}^{t} = h_{m,\pi^i_m}^{t} + \cQ\left(\nabla f^{\pi^i_m}_m(x^t_{m,i}) - h_{m, \pi^i_m}^{t}\right)$
    		        \STATE Set $h_{m,\pi^i_m}^{t+1} = h_{m,\pi^i_m}^{t} + \alpha\cQ\left(\nabla f^{\pi^i_m}_m(x^t_{m,i}) - h_{m, \pi^i_m}^{t}\right)$
    		    \ENDFOR
    		    \STATE Compute $x_{i+1}^{t} = x_i^{t} - \gamma \frac{1}{M}\sum^M_{m=1}\hat{g}_{m,\pi^i_m}^{t}$ and send $x^t_{i+1}$ to the workers
    		\ENDFOR
    		\STATE $x^{t+1} = x_n^t$
    	\ENDFOR
    	\ENSURE $x^T$
	\end{algorithmic}
\end{algorithm}

\begin{theorem}
    \label{th_conv_rr_diana}
    Let Assumptions~\ref{asm:quantization_operators}, \ref{asm:lip_max_f_m}, \ref{asm:sc_each_f_m} hold and suppose that the stepsizes satisfy $
    	\gamma \leq \min\left\{\frac{\alpha}{2n\widetilde\mu}, \frac{1}{\left( 1+\frac{6\omega}{M} \right)L_{\max}}\right\},$ and $\alpha \leq \frac{1}{1+\omega}.$
Define the following Lyapunov function for every $t \geq 0$
\begin{equation}
  \Psi^{t+1} \eqdef \|x^{t+1}-x^{\star}\|^2 +\frac{4\omega\gamma^2}{\alpha M^2}\sum^M_{m=1}\sum^{n-1}_{j=0}(1-\gamma\mu)^j\|\Delta^{t+1}_{m,j}\|^2,\label{lyapunov_func_rr_diana}
\end{equation}
where $\Delta^{t+1}_{m,j} = h_{m,\pi^j_m}^{t+1} - \nabla f_{m,\pi^j_m}(x^{\star})$
	Then, for all $T \geq 0$ the iterates produced by \gls{DIANA-RR} (Algorithm~\ref{alg_new_RR_DIANA}) satisfy
    \begin{equation*}
        \mathbb{E}\left[\Psi^{(T)}\right] \leq \left(1-\gamma\widetilde{\mu}\right)^{nT}\Psi^{(0)}  +\frac{2\gamma^2\sigma^2_{\text{rad}}}{\widetilde\mu}
    \end{equation*}
\end{theorem}
\begin{corollary}
\label{cor:complexity_diana_rr}
Under the same conditions as Theorem~\ref{th_conv_rr_diana} and for Algorithm~\ref{alg_new_RR_DIANA}, there exists stepsizes \(\gamma, \alpha > 0\) such that the number of communication rounds $nT$ to find a solution with accuracy $\varepsilon > 0$ is $	\widetilde{\cO}\left(n(1+\omega)+ \br{1 + \frac{\omega}{M}} \frac{L_{\max}}{\widetilde \mu}+\frac{\sigma_{\text{rad}}}{\sqrt{\varepsilon\widetilde{\mu}^3}}\right).$
\end{corollary}

Unlike \gls{Q-RR}/\algname{QSGD}/\gls{DIANA}, \gls{DIANA-RR} does not have a $\widetilde\cO(\nicefrac{1}{\varepsilon})$-term, which makes it superior to \gls{Q-RR}/\algname{QSGD}/\gls{DIANA} for small enough $\varepsilon$. However, the complexity of \gls{DIANA-RR} has an additive $\widetilde{\cO}(n(1+\omega))$ term arising due to learning the shifts $\{h^{t}_{m,i}\}_{m\in [M], i \in [n]}$. Nevertheless, this additional term is not the dominating one when $\varepsilon$ is small enough. Next, we elaborate a bit more on the comparison between \gls{DIANA} and \gls{DIANA-RR}. That is, \gls{DIANA}  has $
	\widetilde{\cO}\left(\left(1 + \frac{\omega}{M}\right)\frac{L_{\max}}{\mu} + \frac{(1+\omega)\sigma_{\star}^2}{M\mu^2 \varepsilon}\right)$
complexity \citep{gorbunov2020unified}. Neglecting the differences between $\mu$ and $\widetilde{\mu}$, we observe a similar relation between \gls{DIANA-RR} and \gls{DIANA} as between \algname{\gls{RR}} and \algname{\gls{SGD}}: instead of the term $\cO(\nicefrac{(1+\omega)\sigma_{\star}^2}{(M\mu^2 \varepsilon)})$ appearing in the complexity of \gls{DIANA}, \gls{DIANA-RR} has $\cO(\nicefrac{\sigma_{\text{rad}}}{\sqrt{\varepsilon\widetilde{\mu}^3}})$ term much better depending on $\varepsilon$. To the best of our knowledge, our result is the only known one establishing the theoretical superiority of \algname{\gls{RR}} to regular \algname{\gls{SGD}} in the context of distributed learning with gradient compression. Moreover, when $\omega = 0$ (no compression) we recover the rate of \algname{FedRR} and when additionally $M = 1$ (single worker) we recover the rate of \algname{\gls{RR}}.

\subsection{Algorithms with local steps}
\begin{algorithm}[t]
\caption{\gls{Q-NASTYA}}
\label{alg:Q_NASTYA}
	\begin{algorithmic}[1]
		\REQUIRE $x^0$ -- starting point, $\gamma > 0$ -- local stepsize, $\eta > 0$ -- global stepsize
	    \FOR{$t =0,1,\dots, T-1$}
    		\FOR{$m \in [M]$ in parallel}
    		    \STATE Receive $x^t$ from the server and set $x^t_{m,0} = x^t$
    		    \STATE Sample random permutation of $[n]$: $\pi_m = (\pi^0_m, \dots, \pi^{n-1}_m)$
    		    \FOR{$i = 0, 1,\dots, n-1$}
    		        \STATE Set $x_{m,i+1}^{t} = x^{t}_{m,i} - \gamma \nabla f_{m, \pi^{i}_m}(x^{t}_{m,i})$
    		    \ENDFOR
    		    \STATE Compute $g_{m}^{t} = \frac{1}{\gamma n}\left(x^t - x^t_{m,n}\right)$ and send $\cQ^{t}(g_{m}^{t})$ to the server
    		\ENDFOR
    		\STATE Compute $g^{t} = \frac{1}{M}\sum_{m=1}^M\mathcal{Q}^{t}(g_{m}^{t})$
    		\STATE Compute $x^{t+1} = x^{t} - \eta g^{t}$ and send $x^{t+1}$ to the workers
    	\ENDFOR
    	\ENSURE $x^T$
	\end{algorithmic}
\end{algorithm}

In this subsection, we study a new variant of \gls{NASTYA}, \gls{Q-NASTYA} (Algorithm~\ref{alg:Q_NASTYA}), that unifies quantization, local steps with random reshuffling, and uses different local and server stepsizes. Although it improves in per-round communication cost over \gls{NASTYA}, similar to \gls{Q-RR}, we show that \gls{Q-NASTYA} suffers from added variance due to gradient quantization. To overcome this issue, we propose another algorithm, \gls{DIANA-NASTYA} (Algorithm~\ref{alg:diana-nastya}), that adds \gls{DIANA}-style variance reduction to \gls{Q-NASTYA} and removes the additional variance.

\begin{theorem}
\label{thm:convergence_Q_NASTYA}
Let Assumptions~\ref{asm:quantization_operators}, \ref{asm:sc_general_f}, \ref{asm:lip_max_f_m} hold. Let the stepsizes $\gamma$, $\eta$ satisfy $0 < \eta \leq \frac{1}{16L_{\max}\left(1+\frac{\omega}{M}\right)},$ $0 < \gamma \leq \frac{1}{5nL_{\max}}.$
Then, for all $T \geq 0$ the iterates produced by \gls{Q-NASTYA} (Algorithm~\ref{alg:Q_NASTYA}) satisfy
\begin{align*}
	\mathbb{E}\left[\|x^T-x^{\star}\|^2\right] &\leq \left(1-\frac{\eta\mu}{2}\right)^T\|x^0 - x^{\star}\|^2+  8\frac{\eta\omega}{\mu M}\zeta^2_{\star}\\
    &+\frac{9}{2}\frac{\gamma^2nL_{\max}}{\mu}\left((n+1)\zeta^2_{\star} + \sigma_{\star}^2\right).
\end{align*}
\end{theorem}

\begin{corollary}
Under the same conditions as Theorem~\ref{thm:convergence_Q_NASTYA} and for Algorithm~\ref{alg:Q_NASTYA}, there exist stepsizes \(\gamma = \nicefrac{\eta}{n}\) and \(\eta > 0\) such that the number of communication rounds $T$ to find a solution with accuracy $\varepsilon > 0$ is $$			\widetilde{\cO}\left(\frac{L_{\max}}{\mu}\left(1+\frac{\omega}{M}\right)+ \frac{\omega}{M}\frac{\zeta^2_{\star}}{\varepsilon\mu^3}+\sqrt{\frac{ L_{\max} }{\varepsilon\mu^3}} \sqrt{\zeta^2_{\star}+\frac{\sigma_{\star}^2}{n}}\right).$$
If $\gamma \rightarrow 0$, one can choose $\eta > 0$ such that the above complexity bound improves to $			\widetilde{\cO}\left( \frac{L_{\max}}{\mu}\left(1+\frac{\omega}{M}\right)+\frac{\omega}{M}\frac{\zeta^2_{\star}}{\varepsilon\mu^3}\right).$
\end{corollary}
We emphasize several differences with the known theoretical  results. First, the \algname{FedCOM} method \citep{haddadpour2021federated} was analyzed in the homogeneous setting only, i.e., $f_m(x) = f(x)$ for all $m \in [M]$, which is an unrealistic assumption for \gls{FL} applications. In contrast, our result holds in the fully heterogeneous case. Next, the analysis of \algname{FedPAQ} ~\citep{reisizadeh19_fedpaq} uses a bounded variance assumption, which is also known to be restrictive. Nevertheless, let us compare to their result. In \citep{reisizadeh19_fedpaq} the following complexity is derived for their method: $
	\widetilde{\cO}\left( \frac{L_{\max}}{\mu}\left(1+\frac{\omega}{M}\right)+\frac{\omega}{M}\frac{\sigma^2}{\mu^2\varepsilon} + \frac{\sigma^2}{M\mu^2\varepsilon}\right).
$  This result is inferior to the one we show for \gls{Q-NASTYA}: when $\omega$ is small, the main term in the complexity bound of \algname{FedPAQ} is $\widetilde{\cO}\left(\nicefrac{1}{\varepsilon}\right)$, while for \gls{Q-NASTYA} the dominating term is of the order  $\widetilde{\cO}\left(\nicefrac{1}{\sqrt{\varepsilon}}\right)$ (when $\omega$ and $\varepsilon$ are sufficiently small). 
We also highlight that \algname{FedCRR}~\citep{malinovsky2022federated} does not converge if $\omega > \nicefrac{M^2 \gamma \mu \varepsilon}{(2\left\|x_{*, m}^{n}\right\|^{2})}$, while \gls{Q-NASTYA} does for any $\omega \geq 0$. Finally, when $\omega = 0$ (no compression) we recover \gls{NASTYA} as a special case, and using $\gamma = \nicefrac{\eta}{n}$, we recover the rate of \algname{FedRR}~\citep{mishchenko2022proximal}.

\begin{algorithm}[t]
\caption{\gls{DIANA-NASTYA}}
\label{alg:diana-nastya}
	\begin{algorithmic}[1]
		\REQUIRE $x^0$ -- starting point, $\{h_{0,m}\}_{m=1}^M$ -- initial shift-vectors, $\gamma > 0$ -- local stepsize, $\eta > 0$ -- global stepsize, $\alpha > 0$ -- stepsize for learning the shifts
	    \FOR{$t =0,1,\dots, T-1$}
    	   \FOR{$m = 1,\dots, M$ in parallel}
    		    \STATE Receive $x^t$ from the server and set $x^t_{m,0} = x^t$
    		    \STATE Sample random permutation of $[n]$: $\pi_m = (\pi^0_m, \dots, \pi^{n-1}_m)$
    		    \FOR{$i = 0, 1,\dots, n-1$}
    		        \STATE Set $x_{m,i+1}^{t} = x^{t}_{m,i} - \gamma \nabla f_{m, \pi^{i}_m}(x^{t}_{m,i})$
    		    \ENDFOR
    		    \STATE Compute $g_{m}^{t} = \frac{1}{\gamma n}\left(x^t - x^t_{m,n}\right)$ and send $\cQ^{t}\left(g_{m}^{t} - h^{t}_{m}\right)$ to the server
    		    \STATE Set $h^{t+1}_{m} = h^{t}_{m} + \alpha\cQ^{t}\left(g_{m}^{t} - h^{t}_{m}\right)$
    		    \STATE Set $\hat{g}^{t}_{m} = h^{t}_{m} + \cQ^{t}\left(g_{m}^{t} - h^{t}_{m}\right)$
    		\ENDFOR
			\STATE $h_{t+1} = \frac{1}{M}\sum^M_{m=1}h^{t+1}_{m} = h^{t} + \frac{\alpha}{M}\sum^M_{m=1} \cQ^{t}\left(g_{m}^{t} - h^{t}_{m}\right)$
    		\STATE $\hat{g}^t = \frac{1}{M}\sum^M_{m=1}\hat{g}^{t}_{m} = h^{t} + \frac{1}{M}\sum^M_{m=1}\cQ^{t}\left(g_{m}^{t} - h^{t}_{m}\right)$
    		\STATE $x^{t+1} = x^{t} - \eta \hat{g}^t$
    	\ENDFOR
    	\ENSURE $x^T$
	\end{algorithmic}
\end{algorithm}

\begin{theorem}
    \label{thm:diana-nastya-conv}
    Let Assumptions~\ref{asm:quantization_operators}, \ref{asm:sc_general_f}, \ref{asm:lip_max_f_m} hold. Suppose the stepsizes $\gamma$, $\eta, \alpha$ satisfy $
        0 < \gamma \leq  \frac{1}{16L_{\max}n}$, $0 < \eta \leq \min\left\{\frac{\alpha}{2\mu}, \frac{1}{16L_{\max}\left(1+\frac{9\omega}{M}\right)}\right\},$ and $\alpha \leq \frac{1}{1+\omega}.$
    Define the following Lyapunov function:
    \begin{equation}
    \label{eq:diana-nastya-lyapunov-function}
    \Psi^{t+1} \eqdef \|x^{t+1}-x^{\star}\|^2 +\frac{8\omega\eta^2}{\alpha M^2}\sum^M_{m=1}\|h^{t+1}_{m}-h^{\star}_m\|^2.
    \end{equation}
    Then, for all $T \geq 0$ the iterates produced by \gls{DIANA-NASTYA} (Algorithm~\ref{alg:diana-nastya}) satisfy
    \begin{equation}
    \mathbb{E}\left[\Psi^{(T)}\right] \leq \left(1-\frac{\eta\mu}{2}\right)^T\Psi^{(0)}  + \frac{9}{2}\frac{\gamma^2 nL}{\mu}\left((n+1)\zeta^2_{\star} + \sigma_{\star}^2\right).
    \end{equation}
\end{theorem}

\begin{corollary}
Under the same conditions as Theorem~\ref{thm:diana-nastya-conv} and for Algorithm~\ref{alg:diana-nastya}, there exist stepsizes \(\gamma = \nicefrac{\eta}{n}\), \(\eta > 0\), \(\alpha > 0\) such that the number of communication rounds $T$ to find a solution with accuracy $\varepsilon > 0$ is $$	\widetilde{\cO}\left(\omega + \frac{L_{\max}}{\mu}\left(1+\frac{\omega}{M}\right)+\sqrt{\frac{ L_{\max} }{\varepsilon\mu^3}} \sqrt{\zeta^2_{\star}+\frac{\sigma_{\star}^2}{n}}\right).$$
If $\gamma \rightarrow 0$, one can choose $\eta > 0$ such that the above complexity bound improves to $		\widetilde{\cO}\left(\omega + \frac{L_{\max}}{\mu}\left(1+\frac{\omega}{M}\right)\right).$
\end{corollary}
In contrast to \gls{Q-NASTYA}, \gls{DIANA-NASTYA} does not suffer from the $\widetilde{\cO}(\nicefrac{1}{\varepsilon})$ term in the complexity bound. This shows the superiority of \gls{DIANA-NASTYA} to \gls{Q-NASTYA}. Next, \algname{FedCRR-VR}~\citep{malinovsky2022federated} has the rate $$
	\widetilde{\mathcal{O}}\left(\frac{(\omega+1)\left(1-\frac{1}{\kappa}\right)^{n}}{\left(1-\left(1-\frac{1}{\kappa}\right)^{n}\right)^{2}}+\frac{\sqrt{\kappa}\left(\zeta_{\star}+ \sigma_{\star}\right)}{\mu \sqrt{\varepsilon}}\right),$$ which depends on $\widetilde{\cO}\left(\nicefrac{1}{\sqrt{\varepsilon}}\right)$. However, the first term is close to $\widetilde{\cO}\left((\omega+1)\kappa^2\right)$ for a large condition number. \algname{FedCRR-VR-2} utilizes a variance reduction technique from the work of~\citep{malinovsky2023random} and it allows to get rid of permutation variance. This method has $$\widetilde{\cO}\left(\frac{(\omega+1)\left(1-\frac{1}{\kappa \sqrt{\kappa n}}\right)^{\frac{n}{2}}}{\left(1-\left(1-\frac{1}{\kappa \sqrt{\kappa n}}\right)^{\frac{n}{2}}\right)^{2}}+\frac{\sqrt{\kappa}\zeta_{\star}}{\mu \sqrt{\varepsilon}}\right) $$ complexity, but it requires additional assumption on number of functions $n$ and is thus not directly comparable with our result. Note that if we have no compression $(\omega = 0)$, \gls{DIANA-NASTYA} recovers the rate of \gls{NASTYA}. 

In Appendix~\ref{appendix:PP}, we provide versions of \gls{Q-NASTYA} and \gls{DIANA-NASTYA} with partial participation of clients, which is another important aspect of \gls{FL}, and derive the convergence results for them.

\section{Experiments}\label{sec:experiments_ch_6}

We evaluated our methods for solving logistic regression problems and training neural networks in three parts: (i) Comparison of the proposed non-local methods with existing baselines;  (ii) Comparison of the proposed local methods with existing baselines; (iii) Comparison of the proposed non-local methods in training \texttt{ResNet-18} on \texttt{CIFAR10}.

\begin{figure*}[t]
	\centering
		\includegraphics[width=0.9\linewidth]{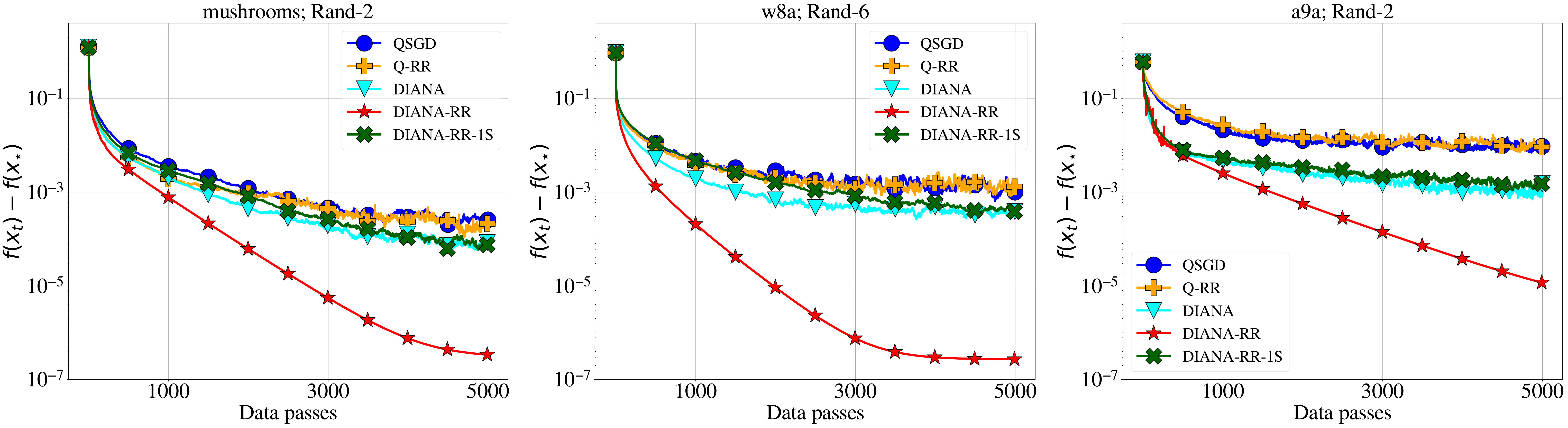} 
		\caption{Non-local methods}\label{fig:non_local}
		\includegraphics[width=0.9\linewidth]{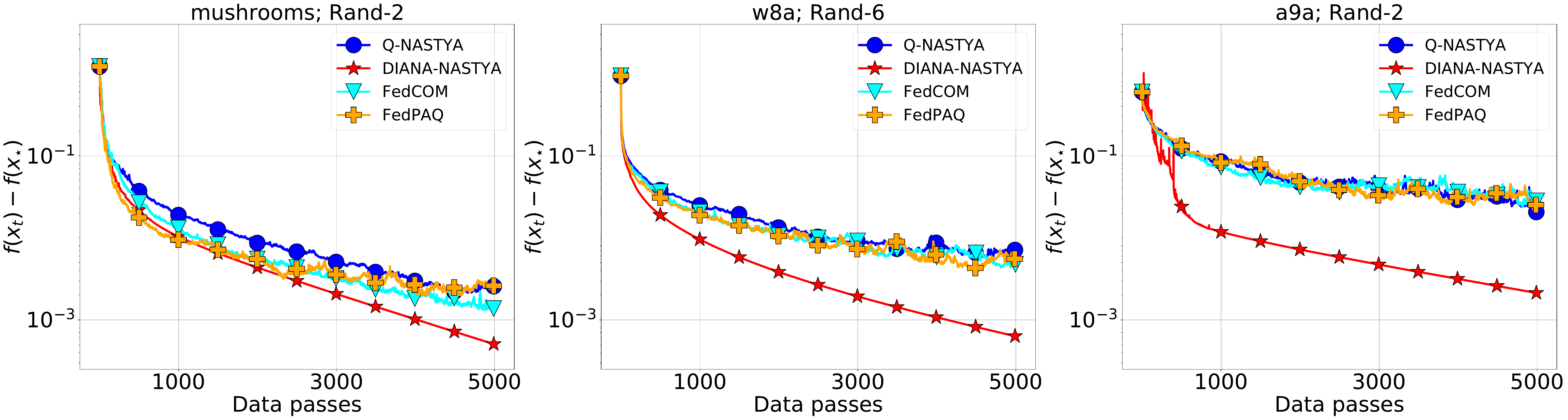} 
		\caption{Local methods}\label{fig:local}
 \label{fig:tuned_experiments}
 \caption{The comparison of the proposed methods (\gls{Q-NASTYA}, \gls{DIANA-NASTYA}, \gls{Q-RR}, \gls{DIANA-RR}), \algname{DIANA-RR-1S} (a variant of \gls{DIANA-RR}), and existing baselines (\algname{QSGD}, \gls{DIANA}, \algname{FedCOM}, \algname{FedPAQ}). All methods use tuned stepsizes and the Rand-$k$ compressor.}
\end{figure*}


\paragraph{Logistic Regression.} To confirm our theoretical results we conducted several numerical experiments on a binary classification problem with $\ell_2$-regularized logistic regression of the form
\begin{eqnarray}\label{eq:log-reg_1}
\min _{x \in \mathbb{R}^{d}}\left[f(x) \eqdef \frac{1}{M} \sum_{m=1}^{M} f_m(x) \right] \text {, }
\end{eqnarray}
where $f_m \eqdef\frac{1}{n_m} \sum_{i=1}^{n_m} f_{m,i}$ and $f_{m,i}\eqdef \log \left(1+\exp({-y_{mi} a_{mi}^{\top} x})\right)+\lambda \|x\|^2_2,$ where  $(a_{mi},  y_{mi}) \in \mathbb{R}^{d} \times \{-1,1\}, i =1,\dots,n_m$ are the training data samples stored on machines $m =1,\dots,M$, and $\lambda>0$ is a regularization parameter. In all experiments, for each method, we used the largest stepsize allowed by its theory multiplied by some individually tuned constant multiplier. For better parallelism, each worker $m$ uses mini-batches of size $\approx 0.1 n_m$. In all algorithms, as a compression operator $\cQ$, we use Rand-$k$ \citep{beznosikov20_biased_compr_distr_learn}  with fixed compression ratio $\nicefrac{k}{d} \approx 0.02$, where $d$ is the number of features in the dataset. 

In our first experiment (see Figure~\ref{fig:non_local}), we compare \gls{Q-RR}, \gls{DIANA-RR}, and \algname{DIANA-RR-1S} with baselines (\algname{QSGD} \citep{alistarh2017qsgd}, \gls{DIANA} \citep{mishchenko2019distributed}) that use a with-replacement mini-batch \algname{\gls{SGD}} estimator. \algname{DIANA-RR-1S} is a memory-friendly version of \gls{DIANA-RR} that stores and uses a single shift $h^{t}_{m}$ on the worker side rather than $n$ individual shifts $h^{t}_{m, \pi_m^i}$. Figure~\ref{fig:non_local} illustrates that \gls{Q-RR} exhibits similar behavior to \algname{QSGD}, with both methods being slower than \gls{DIANA} methods across all considered datasets.
\algname{DIANA-RR-1S} and \gls{DIANA} show comparable convergence rates, indicating that random reshuffling alone, without introducing additional shifts, does not make a significant difference.
Finally, \gls{DIANA-RR} achieves the best rate among all considered non-local methods, efficiently reducing the variance and reaching the lowest functional sub-optimality tolerance. These experimental results align perfectly with our theoretical analysis.

The second experiment shows that \gls{DIANA}-based methods can lead to significantly better performance in practice when applied to local methods as well. In particular, whereas \gls{Q-NASTYA} shows behavior comparable to that of existing methods \algname{FedCOM} \citep{haddadpour2021federated}, \algname{FedPAQ} \citep{reisizadeh19_fedpaq} in all considered datasets, \gls{DIANA-NASTYA} noticeably outperforms other methods.


\begin{figure*}[h!]
	\centering
	\captionsetup[sub]{font=small,labelfont={}}	
		\includegraphics[width=0.49\textwidth]{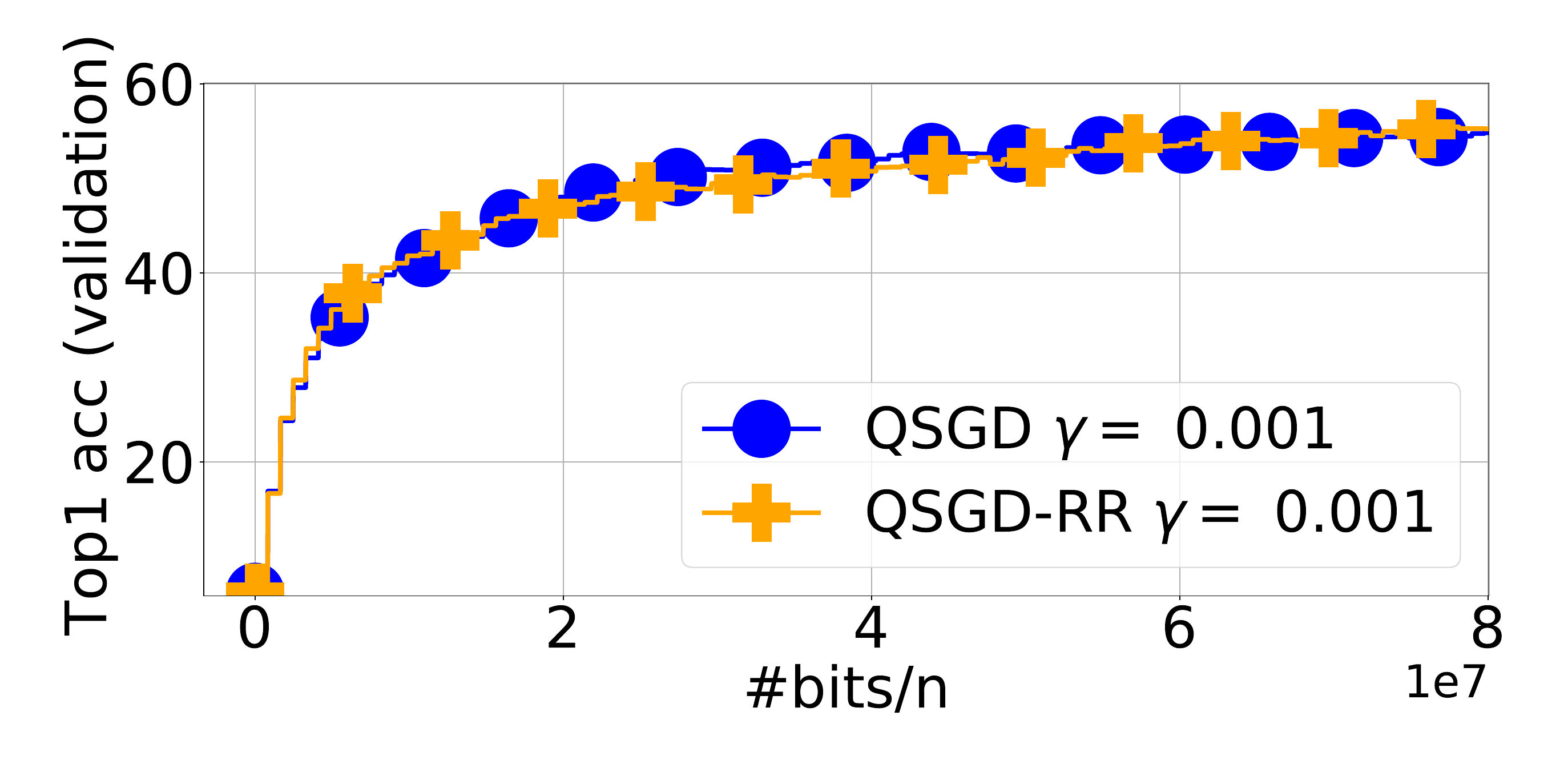} 
		\includegraphics[width=0.49\textwidth]{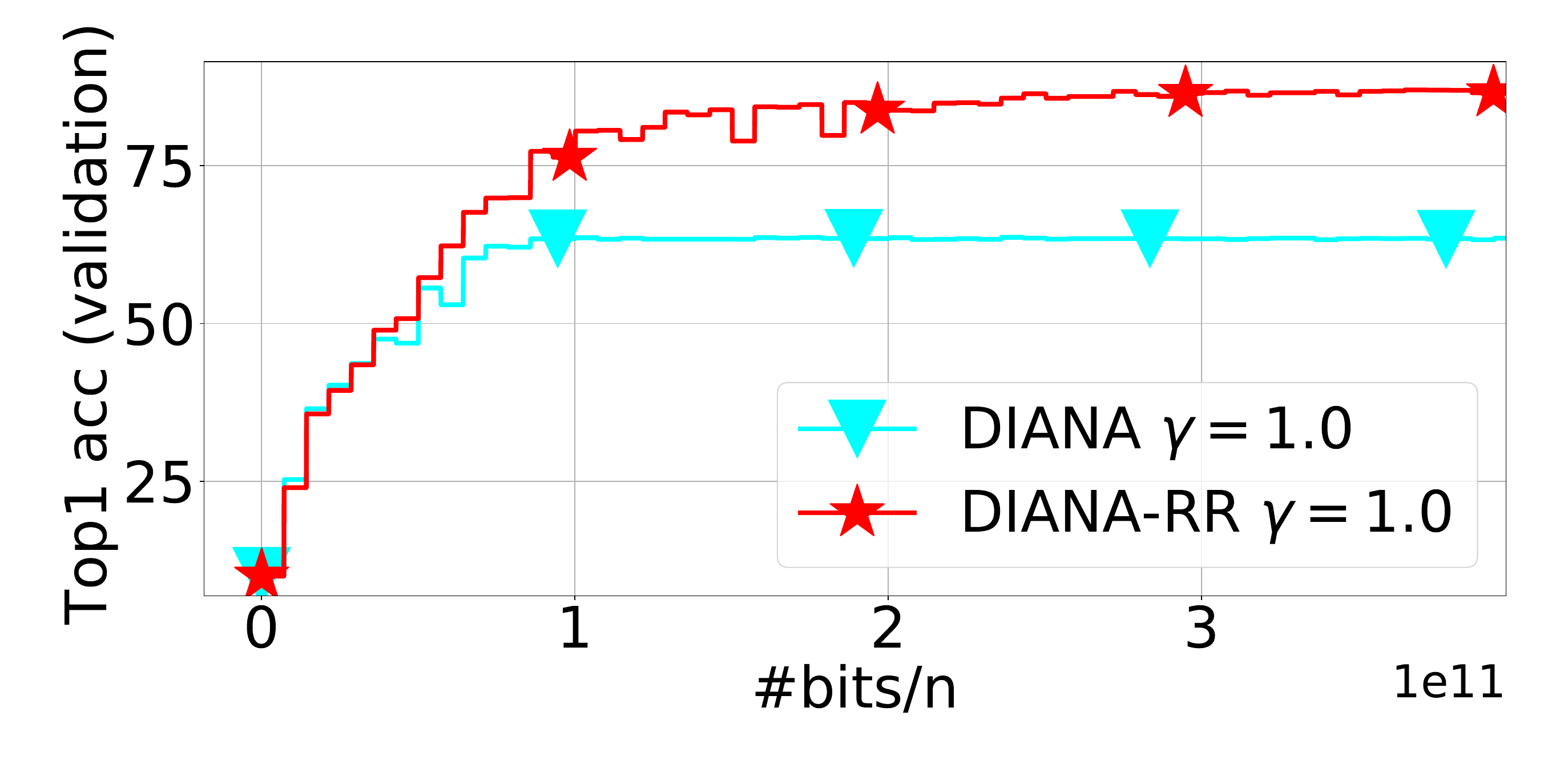} 
	\caption{The comparison of \gls{Q-RR}, \algname{QSGD}, \gls{DIANA}, and \gls{DIANA-RR} on the task of training \texttt{ResNet-18} on \texttt{CIFAR-10} with $M=10$ workers. Top-1 accuracy on test set is reported. Stepsizes were tuned and workers used Rand-$k$ compressor with $\nicefrac{k}{d} \approx 0.05$. 
 }
	\label{fig:NN_plots_main_1}
\end{figure*}

\paragraph{Training Deep Neural Network model: ResNet-18 on CIFAR-10.} Since random reshuffling is a very popular technique in training neural networks, it is natural to test the proposed methods on such problems. Therefore, in the second set of experiments, we consider training \texttt{ResNet-18}~\citep{resnet18} model on the \texttt{CIFAR10} dataset \citep{cifar}. To conduct these experiments we use \texttt{FL\_PyTorch} simulator \citep{burlachenko2021fl_pytorch}. 

The main goal of this experiment is to verify the phenomenon observed in Experiment 1 on the training of a deep neural network. That is, we tested \gls{Q-RR}, \algname{QSGD}, \gls{DIANA}, and \gls{DIANA-RR} in the distributed training of \texttt{ResNet-18} on \texttt{CIFAR10}, see Figure~\ref{fig:NN_plots_main_1}. As in the logistic regression experiments, we observe that (i) \gls{Q-RR} and \algname{QSGD} behave similarly and (ii) \gls{DIANA-RR} outperforms \gls{DIANA}. For further experimental results and details, we refer to Appendix~\ref{subsec:extra_exp}.


\chapter{Byz-VR-MARINA-PP: Adaptive Gradient Difference Clipping for Byzantine Robustness with Partial Participation
}
\label{chapter7}
\thispagestyle{empty}

\section{Introduction}
\label{introduction}

Distributed optimization problems are a cornerstone of modern Machine Learning research. They naturally arise in scenarios where data is distributed across multiple clients; for instance, this is typical in Federated Learning (\gls{FL})~\citep{FedLearn2016, kairouz2019advances}. Such problems require specialized algorithms adapted to the distributed setup. Additionally, the adoption of distributed optimization methods is motivated by the sheer computational complexity involved in training modern Machine Learning models. Many models deal with massive datasets and intricate architectures, rendering training infeasible on a single machine \citep{gpt3costlambda}. Distributed methods, by parallelizing computations across multiple machines, offer a pragmatic solution to accelerate training and address these computational challenges, thus pushing the boundaries of Machine Learning capabilities.

To make distributed training accessible to the broader community, collaborative learning approaches have been actively studied in recent years \citep{kijsipongse2018hybrid, hivemind_dmoe, atre2021distributed, diskin2021distributed}. In such applications, there is a high risk of the occurrence of so-called \emph{Byzantine workers} \citep{lamport1982byzantine, su2016fault}—participants who can violate the prescribed distributed algorithm/protocol either intentionally or simply because they are faulty. In general, such workers may even have access to some private data of certain participants and may collude to increase their impact on the training. Since the ultimate goal is to achieve robustness in the worst case, many papers in the field make no assumptions limiting the power of Byzantine workers. Clearly, in this scenario, standard distributed methods based on the averaging of received information (e.g., stochastic gradients) are not robust, even to a single Byzantine worker. Such a worker can send an arbitrarily large vector that can shift the method arbitrarily far from the solution. This aspect makes it non-trivial to design methods with provable robustness to Byzantines \citep{baruch2019little, xie2020fall}. Despite all the challenges, multiple methods are developed/analyzed in the literature \citep{alistarh2018byzantine, allen2020byzantine, wu2020federated, zhu2021broadcast, karimireddy2021learning, karimireddy2020byzantine, gorbunov2021secure, gorbunov2023variance, allouah2023fixing}.

However, literally all existing methods with provable Byzantine robustness require \emph{the full (or close to full) participation of clients or rely on extra assumptions}. The requirement of full participation is impractical for modern distributed learning problems since they can have millions of clients \citep{FL-secure_aggreg, niu2020billion}. In such scenarios, it is more natural to use partial participation to speed up the training. Moreover, some clients can be unavailable at certain moments, e.g., due to a poor connection, low battery, or simply because of the need to use the computing power for some other tasks. Although \emph{partial participation of clients} is a natural attribute of large-scale collaborative training, it is not studied under the presence of Byzantine workers. Moreover, this question is highly non-trivial: the existing methods can fail to converge if combined na\"ively with partial participation since Byzantine workers can form a majority during particular rounds and thus destroy the whole training with just one round of communication. \emph{Therefore, the field requires the development of new distributed methods that are provably robust to Byzantine attacks and can work with partial participation even when Byzantine workers form a majority during some rounds.}

\subsection{Contributions}\label{section:contribution}

We develop Byzantine-tolerant Variance-Reduced \algname{MARINA} with Partial Participation (\gls{Byz-VR-MARINA-PP}, Algorithm~\ref{alg:byz_vr_marina}) -- the first distributed method having Byzantine robustness and allowing partial participation of clients without strong additional assumptions. Our method uses variance reduction to handle Byzantine workers and clipping of stochastic gradient differences to bound the potential harm of Byzantine workers even when they form a majority during particular rounds of communication. To make the method even more communication efficient, we add communication compression. We prove the convergence of \gls{Byz-VR-MARINA-PP} for general smooth non-convex functions and \gls{PL} functions. In the special case of full participation, our complexity bounds recover the ones for \gls{Byz-VR-MARINA} \citep{gorbunov2023variance} that are the current SOTA convergence results. Moreover, we prove that in some cases, partial participation is theoretically beneficial for \gls{Byz-VR-MARINA-PP}. We also propose a simplified version of \gls{Byz-VR-MARINA-PP} with a better neighborhood term in the convergence bounds (\algname{Byz-VR-MARINA-PP+}, Algorithm~\ref{alg:byz_vr_marina+}) and a heuristic on how to use clipping to adapt any Byzantine-robust method to the partial participation setup and illustrate its performance in experiments.

\subsection{Related work}

Below, we overview closely related works. Additional discussion is deferred to Appendix~\ref{appendix:extra_related_work}.

\textbf{Byzantine robustness.} The primary vulnerability of standard distributed methods to Byzantine attacks lies in the aggregation rule: even one worker can arbitrarily distort the average. Therefore, many papers on Byzantine robustness focus on the application of robust aggregation rules, such as the geometric median \citep{pillutla2022robust}, coordinate-wise median, trimmed median \citep{yin2018byzantine}, Krum \citep{blanchard2017machine}, and Multi-Krum \citep{damaskinos2019aggregathor}. However, simply robustifying the aggregation rule is insufficient to achieve provable Byzantine robustness, as illustrated in \citep{baruch2019little} and \citep{xie2020fall}, who design special Byzantine attacks that can bypass standard defenses. This implies that more significant algorithmic changes are required to achieve Byzantine robustness, a point also formally proven in \citep{karimireddy2021learning}, who demonstrate that permutation-invariant algorithms -- i.e., algorithms independent of the order of stochastic gradients at each step -- cannot provably converge to any predefined accuracy in the presence of Byzantines.

In \citep{wu2020federated}, the authors are the first to exploit variance reduction to tolerate Byzantine attacks. They propose and analyze the method called \algname{Byrd-SAGA}, which uses \algname{SAGA}-type \citep{defazio2014saga} gradient estimators on the good workers and geometric median for the aggregation. In \citep{gorbunov2023variance}, the authors develop another variance-reduced method called \gls{Byz-VR-MARINA}, which is based on (conditionally biased) \algname{GeomSARAH}/\algname{PAGE}-type \citep{DIANA2, li2021page} gradient estimator and any robust aggregation in the sense of the definition from \citep{karimireddy2021learning, karimireddy2020byzantine}, and derive the improved convergence guarantees that are the current SOTA in the literature. There are also many other approaches and we discuss some of them in Appendix~\ref{appendix:extra_related_work}.

\textbf{Partial Participation.} In the context of Byzantine robust learning, there exist several works that develop and analyze methods with partial participation \citep{data2021byzantine, el2021collaborative, boubouh2022democratizing, allouah2024tackling}. However, these works rely on the restrictive assumption that the number of participating clients at each round is larger than the number of Byzantine workers. In this case, Byzantines cannot form a majority, and standard methods can be applied without any changes. In contrast, our method converges in more challenging scenarios, e.g., \gls{Byz-VR-MARINA-PP} provably converges even when the server samples one client, which can be Byzantine. If the number of participating clients is such that Byzantine clients can form a majority, these methods have a certain probability of divergence, and this probability grows with each communication round. We provide a more detailed discussion in Appendix~\ref{appendix:extra_related_work}.

\section{Preliminaries}\label{section:prelim}

In this section, we formally introduce the problem, main definition, and assumptions used in the analysis. That is, we consider a finite-sum distributed optimization problem\footnote{For simplicity, we assume that all regular workers have the same size of local datasets. Our analysis can be easily generalized to the case of different sizes of local datasets: this will affect only the value of $\cL_{\pm}$ from Assumption~\ref{assm:local} for some sampling strategies.}
\begin{align}
\label{eq:main_problem}
    \min_{x\in \R^d}\left\{f(x) \eqdef \frac{1}{G}\sum_{m\in \mathcal{G}}f_m(x)\right\},\quad    f_m(x) \eqdef \frac{1}{n}\sum_{i=1}^n f_{m,i}(x)  \quad \forall m\in \mathcal{G}, 
\end{align}
where $\mathcal{G}$ is a set of regular clients of size $G \eqdef |\mathcal{G}|$. In the context of distributed learning, $f_m:\R^d \to \R$ corresponds to the loss function on the data of client $m$, and $f_{m,i}:\R^d \to \R$ is the loss computed on the $i$-th sample from the dataset of client $m$. Next, we assume that the set of all clients taking part in the training is $[M] = \{1,2,\ldots, M\}$ and $\mathcal{G} \subseteq [M]$. The remaining clients $\cB \eqdef [M]\setminus \mathcal{G}$ are Byzantine ones. We assume that $B \eqdef |\cB| \eqdef \deltar M \leq \delta M$, where $\deltar$ is an exact ratio of Byzantine workers and $\delta$ is a known upper bound for $\deltar$. We also assume that $0 \leq \deltar \leq \delta <\nicefrac{1}{2}$ since otherwise, Byzantine workers form a majority and problem \eqref{eq:main_problem} becomes impossible to solve in general.

\textbf{Notation.} We use a standard notation in the literature on distributed stochastic optimization. Everywhere in the text $\|x\|$ denotes a standard $\ell_2$-norm of $x\in\R^d$, $\langle a, b \rangle$ refers to the standard inner product of vectors $a,b \in \R^d$. The clipping operator is defined as follows: $\clip_\lambda(x) \eqdef \min\{1, \nicefrac{\lambda}{\|x\|}\}x$ for $x\neq 0$ and $\clip_\lambda(0) \eqdef 0$. Finally, $\PP\{A\}$ denotes the probability of event $A$, $\mathbb{E}[\xi]$ is the full expectation of random variable $\xi$, $\mathbb{E}[\xi \mid A]$ is the expectation of $\xi$ conditioned on the event $A$. We also sometimes use $\mathbb{E}_{t}[\xi]$ to denote the expectation of $\xi$ w.r.t.\ the randomness coming from step $t$.

\textbf{Robust aggregator.} We follow the definition from \citep{gorbunov2023variance} of $(\delta,c)$-robust aggregation, which is a generalization of the definitions proposed in \citep{karimireddy2021learning, karimireddy2020byzantine}.

\begin{definition}[$(\delta, c)$-Robust Aggregator] 
\label{def:aragg}
Assume that $\left\{x_1, x_2, \ldots, x_M\right\}$ is such that there exists a subset $\mathcal{G} \subseteq[M]$ of size $|\mathcal{G}|=G \geq(1-\delta) n$ for $\delta\leq \delta_{\max} < 0.5$ and there exists $\sigma \geq 0$ such that $\frac{1}{G(G-1)} \sum_{m, l \in \mathcal{G}} \mathbb{E}\left[\left\|x_m-x_l\right\|^2\right] \leq \sigma^2$ where the expectation is taken w.r.t.\ the randomness of $\left\{x_m\right\}_{m \in \mathcal{G}}$. We say that the quantity $\widehat{x}$ is $(\delta, c)$-\texttt{\gls{RAgg}}) and write $\widehat{x} = \texttt{RAgg}\left(x_1, \ldots, x_M\right)$ for some $c>0$, if the following inequality holds:
\begin{equation}
    \mathbb{E}\left[\|\widehat{x}-\bar{x}\|^2\right] \leq c \delta \sigma^2, \label{eq:robust_aggr_definition}
\end{equation}
where $\bar{x}\eqdef\frac{1}{|\mathcal{G}|} \sum_{m \in \mathcal{G}} x_m$. If additionally $\widehat{x}$ is computed without the knowledge of $\sigma^2$, we say that $\widehat{x}$ is $(\delta, c)$-Agnostic Robust Aggregator $\left(\delta, c)\right.$-\gls{ARAgg} and write $\widehat{x} = \gls{ARAgg}\left(x_1, \ldots, x_M\right)$.
\end{definition}

One can interpret the definition as follows. Ideally, we would like to filter out all Byzantine workers and compute just an average $\bar x$ over the set of good clients. However, this is impossible in general since we do not know a priori who are Byzantine workers. Instead of this, it is natural to expect that the aggregation rule approximates the ideal average in a certain sense, e.g., in terms of the expected squared distance to $\bar x$. In  \citep{karimireddy2021learning} it is formally shown that, in terms of such criterion ($\mathbb{E}[\|\widehat x - \bar x\|^2]$), the definition of $\left(\delta, c)\right.$-\gls{ARAgg} cannot be improved (up to the numerical constant). Moreover, standard aggregators such as Krum \citep{blanchard2017machine}, geometric median, and coordinate-wise median do not satisfy Definition~\ref{def:aragg} \citep{karimireddy2021learning}, though another popular standard aggregation rule called coordinate-wise trimmed mean \citep{yin2018byzantine} satisfies Definition~\ref{def:aragg} as shown in \citep{allouah2023fixing} through the more general definition of robust aggregation. To address this issue, in \citep{karimireddy2021learning} the authors develop the aggregator called \algname{CenteredClip} and prove that it fits the definition of $\left(\delta, c)\right.$-\texttt{\gls{RAgg}}. In \citep{karimireddy2020byzantine} the authors propose a procedure called \algname{Bucketing} that fixes Krum, geometric median, and coordinate-wise median, i.e., with \algname{Bucketing} Krum, geometric median, and coordinate-wise median become $\left(\delta, c)\right.$-\gls{ARAgg}, which is important for our algorithm since the variance of the vectors received from regular workers changes over time in our method. We notice here that $\delta$ is a part of the input that should satisfy $\deltar \leq \delta \leq \delta_{\max}$.

\textbf{Compression operators.} In our work, we use standard unbiased compression operators with relatively bounded variance \citep{khirirat2018distributed, DIANA2}.

\begin{definition}[Unbiased compression]
\label{def:Q}
Stochastic mapping $\mathcal{Q}: \mathbb{R}^d \rightarrow \mathbb{R}^d$ is called unbiased compressor/compression operator if there exists $\omega \geq 0$ such that for any $x \in \mathbb{R}^d$
$$\mathbb{E}[\mathcal{Q}(x)]=x, \quad \mathbb{E}\left[\|\mathcal{Q}(x)-x\|^2\right] \leq \omega\|x\|^2 .$$
For the given unbiased compressor $\mathcal{Q}(x)$, one can define the expected density\footnote{This quantity is well-suited for sparsification-type compression operators like random sparsification \citep{stich2018sparsified} and $1$-level $\ell_2$-quantization \citep{alistarh2017qsgd}. For other compressors, such as quantization with more than one level \citep{goodall1951television, roberts1962picture},  $\zeta_{\cQ}$ is not the main characteristic describing their properties.} as $\zeta_{\mathcal{Q}}\eqdef$ $\sup _{x \in \mathbb{R}^d} \mathbb{E}\left[\|\mathcal{Q}(x)\|_0\right]$, where $\|y\|_0$ is the number of non-zero components of $y \in \mathbb{R}^d$.
\end{definition}

In this definition, parameter $\omega$ reflects how lossy the compression operator is: the larger $\omega$ the more lossy the compression. For example, this class of compression operators includes random sparsification (RandK) \citep{stich2018sparsified} and quantization \citep{goodall1951television, roberts1962picture, alistarh2017qsgd}. For RandK compression $\omega = \frac{d}{K} - 1, \zeta_{\cQ} = K$ and for $\ell_2$-quantization $\omega = \sqrt{d}-1, \zeta_{\cQ} = \sqrt{d}$, see the proofs in \citep{beznosikov20_biased_compr_distr_learn}.

\textbf{Assumptions.} Up to a couple of assumptions that are specific to our work, we use the same assumptions as in \citep{gorbunov2023variance}. We start with two new assumptions.

\begin{assumption}[Bounded \texttt{ARAgg}]
\label{assm:bounded-aggr}
    We assume that the server applies aggregation rule $\cA$ such that $\cA$ is $(\delta,c)$-\gls{ARAgg} and there exists a constant $F_{\cA} > 0$ such that for any inputs $x_1,\ldots, x_M \in \R^d$ the norm of the aggregator is not greater than the maximal norm of the inputs:
        $$\left\| \cA\left(x_1, \ldots, x_M\right)  \right\| \leq F_{\cA} \max_{m\in [M]} \|x_m\|.$$
\end{assumption}

The above assumption is satisfied for popular $(\delta,c)$-robust aggregation rules presented in the literature \citep{karimireddy2021learning, karimireddy2020byzantine}. Therefore, this assumption is more a formality than a real limitation: it is needed to exclude some pathological examples of $(\delta,c)$-robust aggregation rules, e.g., for any $\cA$ that is $(\delta,c)$-\texttt{\gls{RAgg}} one can construct an unbounded $(\delta,2c)$-\texttt{\gls{RAgg}} as $\overline\cA = \cA + X$, where $X$ is a random sample from the Gaussian distribution $\cN(0, c\delta\sigma^2)$.

Next, for part of our results, we also make the following assumption.

\begin{assumption}[Bounded compressor (optional)]
\label{assm:bounded-compressor}
    We assume that workers use compression operator $\cQ$ satisfying Definition~\ref{def:Q} and bounded as follows:
        $$\left\| \cQ(x)\right\| \leq D_{Q} \|x\| \quad \forall x \in \R^d.$$
\end{assumption}

For example, RandK and $\ell_2$-quantization meet this assumption with $D_\cQ = \frac{d}{K}$ and $D_{\cQ} = \sqrt{d}$ respectively. In general, constant $D_{\cQ}$ can be large (proportional to $d$). However, in practice, one can use RandK with $K = \frac{d}{100}$ and, thus, have a moderate $D_{\cQ} = 100$. We also have the results without Assumption~\ref{assm:bounded-compressor}, but with worse dependence on some other parameters, see  Section~\ref{section:convergence_results}.

Next, we assume that good workers have $\zeta^2$-heterogeneous local loss functions.

\begin{assumption}[$\zeta^2$-heterogeneity] 
\label{assm:het_simplified}
We assume that good clients have $\zeta^2$ heterogeneous local loss functions for some $\zeta \geq 0$, i.e.,
$$
\frac{1}{G} \sum_{m \in \mathcal{G}}\left\|\nabla f_m(x)-\nabla f(x)\right\|^2 \leq \zeta^2 \quad \forall x \in \mathbb{R}^d.
$$
\end{assumption}

The above assumption is quite standard in the literature on Byzantine robustness \citep{wu2020federated, karimireddy2020byzantine, gorbunov2023variance, allouah2023fixing}. Moreover, some kind of a bound on the heterogeneity of good clients is necessary since otherwise Byzantine robustness cannot be achieved in general. In the appendix, all proofs are given under a more general version of Assumption~\ref{assm:het_simplified} (see Assumption~\ref{assm:het}). Finally, the case of homogeneous data ($\zeta = 0$) is also quite popular for collaborative learning \citep{diskin2021distributed, kijsipongse2018hybrid}.

The following assumption is classical in the literature on non-convex optimization.
\begin{assumption}[Smoothness (simplified)]
\label{assm:smoothness_simplified}
We assume that for all $m\in \mathcal{G}$ and $i\in [n]$ there exists $\cL \geq 0$ such that $f_{m,i}$ is $\cL$-smooth, i.e., for all $x, y \in \mathbb{R}^d$
\begin{equation}
    \|\nabla f_{m,i}(x) - \nabla f_{m,i}(y)\| \leq \cL \|x - y\|. \label{eq:f_m_j_smooth_simple}
\end{equation}
Moreover, we assume that $f$ is uniformly lower bounded by $f^{\star} \in \mathbb{R}$, i.e., $f^{\star}\eqdef\inf _{x \in \mathbb{R}^d} f(x)$.
\end{assumption}
For the sake of simplicity, we do not differentiate between various notions of smoothness in the main text. However, our analysis takes into account the differences between smoothness constants, similarity of local functions, and sampling strategy (see Appendix~\ref{appendix:refined_assumptions}).

\begin{algorithm*}[t]
   \caption{\gls{Byz-VR-MARINA-PP}: Byzantine-tolerant \algname{VR-MARINA} with Partial Participation}\label{alg:byz_vr_marina}
\begin{algorithmic}[1]
   \STATE {\bfseries Input:} vectors $x^0, g^0 \in \R^d$, stepsize $\gamma$, mini-batch size $b$, probability $p\in(0,1]$, number of iterations $T$, $(\delta,c)$-\texttt{ARAgg}, clients' sample size $1 \leq C \leq \widehat{C} \leq M$, clipping coefficients $\{\alpha_{t}\}_{t\geq 1}$ 
   \FOR{$k=0,1,\ldots,T-1$}
   \STATE Get a sample from Bernoulli distribution with parameter $p$: $c_t \sim \text{Be}(p)$
   \STATE Sample the set of clients $\set \subseteq [M]$, $|\set| = C$ if $c_t = 0$; otherwise $|\set| = \widehat{C}$
   \STATE Broadcast $g^t$, $c_t$ to all workers
   \FOR{$m \in \mathcal{G} \cap \set$ in parallel} 
   \STATE $x^{t+1} = x^t - \gamma g^t$ and $\lambda_{t+1} = \alpha_{t+1}\|x^{t+1} - x^t\|$
   \STATE\label{line:g_i^k+1_main}  Set $g_m^{t+1} = \begin{cases} \nabla f_m(x^{t+1}),& \text{if } c_t = 1,\\ g^t + \clip_{\lambda_{t+1}}\left(\cQ\left(\widehat{\Delta}_m(x^{t+1}, x^t)\right)\right),& \text{otherwise,}\end{cases}$\\ where $\widehat{\Delta}_m(x^{t+1}, x^t)$ is a mini-batched estimator of $\nabla f_m(x^{t+1}) - \nabla f_m(x^t)$, $\cQ(\cdot)$ for $m\in\mathcal{G} \cap \set$ are computed independently
   \ENDFOR
   \STATE\label{line:g^t+1} $g^{t+1} = \begin{cases} \texttt{ARAgg}\left(\{g_m^{t+1}\}_{m\in \set}\right),& \text{if } c_t = 1, \\
       g^t + \texttt{ARAgg}\left(\left\{\clip_{\lambda_{t+1}}\left(\cQ\left(\widehat{\Delta}_m(x^{t+1}, x^t)\right)\right)\right\}_{m\in \set}\right),&\text{otherwise}\end{cases}$
   \ENDFOR
\end{algorithmic}
\end{algorithm*}

Finally, we also consider functions satisfying the Polyak-Łojasiewicz (\gls{PL}) condition \citep{polyak1963gradient, lojasiewicz1963topological}. This assumption belongs to the class of assumptions on the structured non-convexity that allows for achieving linear convergence \citep{necoara2019linear}.

\begin{assumption}[PŁ condition (optional)]
\label{assm:PL}
We assume that function $f$ satisfies Polyak-Łojasiewicz \gls{PL} condition with parameter $\mu > 0$, i.e., for all $x \in \mathbb{R}^d$ there exists  $f^{\star}\eqdef\inf _{x \in \mathbb{R}^d} f(x)$ such that
  $$ \|\nabla f(x)\|^2 \geq 2 \mu\left(f(x)-f^{\star}\right) .$$

\end{assumption}
\section{New Method: Byz-VR-MARINA-PP}\label{section:method}

We propose a new method called Byzantine-tolerant Variance-Reduced \algname{MARINA} with Partial Participation (\gls{Byz-VR-MARINA-PP}, Algorithm~\ref{alg:byz_vr_marina}). Our method extends \gls{Byz-VR-MARINA} \citep{gorbunov2023variance} to the partial participation case via the proper usage of the clipping operator. To illustrate how \gls{Byz-VR-MARINA-PP} works, we first consider a special case of full participation.

\textbf{Special case: \gls{Byz-VR-MARINA}.} If all clients participate at each round ($\set \equiv [M]$) and clipping is turned off ($\lambda_{k} \equiv +\infty$), then \gls{Byz-VR-MARINA-PP} reduces to \gls{Byz-VR-MARINA} that works as follows. Consider the case when no compression is applied ($\cQ(x) = x$) and $\widehat{\Delta}_m(x^{t+1}, x^t) = \nabla f_{m,i^t}(x^{t+1}) - \nabla f_{m,i^t}(x^{t})$, where $i^t$ is sampled uniformly at random from $[n]$, $m\in \cG$. Then, regular workers compute \algname{GeomSARAH}/\algname{PAGE} gradient estimator at each step: for $m\in \cG$
\begin{equation*}
    g_m^{t+1} = \begin{cases}
        \nabla f_m(x^{t+1}), \quad \text{with probability } p,\\ g^t + \nabla f_{m,i^t}(x^{t+1}) - \nabla f_{m,i^t}(x^{t}),\quad\text{otherwise} 
    \end{cases}
\end{equation*}
With small probability $p$, good workers compute full gradients, and with larger probability $1-p$ they update their estimator via adding stochastic gradient difference. To balance the oracle cost of these two cases, one can choose $p \sim \nicefrac{1}{n}$ (for $b$-size mini-batched estimator -- $p\sim \nicefrac{b}{n}$). Such estimators are known to be optimal for finding stationary points in the stochastic first-order optimization \citep{fang2018spider, arjevani2023lower}. Next, good workers send $g_m^{t+1}$ or $\nabla f_{m,i^t}(x^{t+1}) - \nabla f_{m,i^t}(x^{t})$ to the server who robustly aggregates the received vectors. Since estimators are conditionally biased, i.e., $\mathbb{E}[g_m^{t+1}\mid x^{t+1}, x^t] \neq \nabla f_m(x^{t+1})$, the additional bias coming from the aggregation does not cause significant issues in the analysis or practice. Moreover, the variance of $\{g_m^{t+1}\}_{m\in \cG}$ w.r.t.\ the sampling of the stochastic gradients is proportional to $\|x^{t+1} - x^t\|^2 \to 0$ with probability $1-p$ (due to Assumption~\ref{assm:local}) that progressively limits the effect of Byzantine attacks. For a more detailed explanation of why recursive variance reduction works better than \algname{SAGA}/\algname{\gls{SVRG}}-type variance reduction, we refer to \citep{gorbunov2023variance}. Arbitrary sampling allows the improvement of the dependence on the smoothness constants. Unbiased communication compression also naturally fits the framework since it is applied to the stochastic gradient difference, meaning that the variance of $\{g_m^{t+1}\}_{m\in \cG}$ w.r.t.\ the sampling of the stochastic gradients and compression remains proportional to $\|x^{t+1} - x^t\|^2$ with probability $1-p$.

\textbf{New ingredients: Partial Participation and clipping.} The algorithmic novelty of \gls{Byz-VR-MARINA-PP} in comparison to \gls{Byz-VR-MARINA} is twofold: with (typically large) probability $1-p$ only $C$ clients sampled uniformly at random from the set of all clients participate at each round, and clipping is applied to the compressed stochastic gradient differences. With a small probability $p$, a larger number\footnote{As one can see from our analysis, it is sufficient to take $\widehat C \geq \max\{1, \nicefrac{\deltar M}{\delta}\}$ similarly to \citep{data2021byzantine}. However, in contrast to the approach in \citep{data2021byzantine}, \gls{Byz-VR-MARINA-PP} requires such communications only with small probability $p$.} of clients $\widehat C \leq M$ take part in the communication. The main role of clipping is to ensure that the method can withstand the attacks of Byzantines when they form a majority or, more precisely, when there are more than $\delta C$ Byzantine workers among the sampled ones. \emph{Indeed, without clipping (or some other algorithmic changes), such situations are critical for convergence: Byzantine workers can shift the method arbitrarily far from the solution, e.g., they can collectively send some vector with an arbitrarily large norm}. In contrast, \gls{Byz-VR-MARINA-PP} tolerates any attacks even when all sampled clients are Byzantine workers since the update remains bounded due to the clipping. Via choosing $\lambda_{t+1}\sim \|x^{t+1} - x^t\|$ we ensure that the norm of transmitted vectors decreases with the same rate as it does in \gls{Byz-VR-MARINA} with full client participation. Finally, with probability $1-p$ regular workers can transmit just compressed vectors and leave the clipping operation to the server since Byzantines can ignore the clipping operation.

\section{Convergence Results}\label{section:convergence_results}

We define $\cG_C^t = \cG\cap \set$ and $G_{C}^t = |\cG_C^t|$ and $\binom{n}{k} = \frac{n !}{k !(n-k) !}$ represents the binomial coefficient. We also use the following probabilities:
\begin{align*}
p_G &\eqdef  \PP\left\lbrace G_C^t \geq\left(1-\delta\right) C\right\rbrace = \sum_{\lceil(1-\delta)C\rceil\leq m^\prime \leq C} \frac{\binom{G}{m^\prime}\binom{M-G}{C-m^\prime}}{\binom{M}{C}},\\
\mathcal{P}_{\mathcal{G}^t_C} &\eqdef  \PP\left\lbrace m \in \mathcal{G}_C^t \mid G_C^t \geq\left(1-\delta\right) C\right\rbrace = \frac{C}{Mp_G} \cdot \sum_{\lceil(1-\delta)C\rceil\leq m^\prime \leq C} \frac{\binom{G-1}{m^\prime-1}\binom{M-G}{C-m^\prime}}{\binom{M-1}{C-1}}.
\end{align*}
These probabilities naturally appear in the analysis and statements of the theorems. When $c_t = 0$, then server samples $C$ clients, and two situations can appear: either $G_C^t$ is at least $\left(1-\delta\right) C$ meaning that the aggregator can ensure robustness according to Definition~\ref{def:aragg} or $G_C^t < \left(1-\delta\right) C$. Probability $p_G$ is the probability of the first event, and the second event implies that the aggregation can be spoiled by Byzantine workers (but clipping bounds the ``harm''). Finally, we use $\mathcal{P}_{\mathcal{G}^t_C}$ in the computation of some conditional expectations when the first event occurs. The mentioned probabilities can be easily computed for some special cases. For example, if $C = 1$, then $p_G = \nicefrac{G}{M}$ and $\mathcal{P}_{\mathcal{G}^t_C} = \nicefrac{1}{G}$; if $C = 2$, then $p_G = \nicefrac{G(G-1)}{M(M-1)}$ and $\mathcal{P}_{\mathcal{G}^t_C} = \nicefrac{2}{G}$; finally, if $C = M$, then  $p_G = 1$ and $\mathcal{P}_{\mathcal{G}^t_C} = 1$.

The next theorem is our main convergence result for general unbiased compression operators.
\begin{theorem}\label{thm:main_result_1}
 Let Assumptions \ref{assm:bounded-aggr}, \ref{assm:het_simplified}, \ref{assm:smoothness_simplified} hold, $\lambda_{t+1} = 2\cL \left\|x^{t+1} - x^t\right\|$, and $\widehat C \geq \max\{1, \nicefrac{\deltar M}{\delta}\}$. Assume that $0<\gamma \leq \nicefrac{1}{\cL(1+\sqrt{A})},$ 
where constant $A$ is defined as 
\begin{eqnarray}
    A &=& \frac{32p_G G\mathcal{P}_{\mathcal{G}^t_C}}{p^2(1-\delta)C} \left( 30\omega + 11\right) (1 + 2c\delta) + \frac{16(1-p_G)(1+4F_{\cA}^2)}{p^2}. \label{eq:A_unbounded_compr}
\end{eqnarray}
Then for all $T \geq 0$ the iterates produced by \gls{Byz-VR-MARINA-PP} (Algorithm \ref{alg:byz_vr_marina}) satisfy
\begin{equation}
    \mathbb{E}\left[\left\|\nabla f\left(\widehat{x}^T\right)\right\|^2\right] \leq \frac{2 \Phi^{(0)}}{\gamma(T+1)}+\frac{4 \widehat{D} \zeta^2}{p}, \label{eq:main_result_non_convex}
\end{equation}
where $\widehat{D} = \frac{ 2\delta\mathcal{P}_{\mathcal{G}^t_{\widehat{C}}} }{1-\delta} \left(\frac{6cG}{\widehat C} + p \right) + \widetilde{D}$, where $\widetilde{D} = 0$ when $\widehat{C} = M$, $\widetilde{D} = \frac{\cP_{\cG_{\widehat{C}}^t}G}{(1-\delta)\widehat{C}}$ when $\widehat{C} < M$, and $\widehat{x}^T$ is chosen uniformly at random from $x^0, x^1, \ldots, x^t$, and $$\Phi^{(0)}=f\left(x^0\right)-f^{\star}+\frac{2\gamma}{p}\left\|g^0-\nabla f\left(x^0\right)\right\|^2.$$ If, in addition, Assumption~\ref{assm:PL} holds and $0<\gamma \leq \nicefrac{1}{\cL(1+\sqrt{2 A})},$ then for all $T \geq 0$ the iterates produced by \gls{Byz-VR-MARINA-PP} (Algorithm \ref{alg:byz_vr_marina}) with $\rho = \min\left\{\gamma\mu, \frac{p}{8}\right\}$ satisfy
\begin{equation}
    \mathbb{E}\left[f\left(x^t\right)-f\left(x^{\star}\right)\right] \leq\left(1-\rho\right)^T \Phi^{(0)}+\frac{4\widehat{D}\zeta^2\gamma}{p\rho}, \label{eq:main_result_PL}
\end{equation}
where $$\Phi^{(0)}=f\left(x^0\right)-f^{\star}+\frac{4\gamma}{p}\left\|g^0-\nabla f\left(x^0\right)\right\|^2.$$
\end{theorem}

The above theorem establishes similar guarantees to the current SOTA ones obtained for \gls{Byz-VR-MARINA}. That is, in the general non-convex case, we prove $\cO(\nicefrac{1}{T})$ rate, which is optimal \citep{arjevani2023lower}, and for P\L-functions we derive linear convergence result to the neighborhood depending on the heterogeneity. The size of this neighborhood matches the one derived for \gls{Byz-VR-MARINA} in \citep{gorbunov2023variance}. However, since our result is obtained considering the challenging scenario of partial participation of clients, the maximal theoretically allowed stepsize in our analysis of \gls{Byz-VR-MARINA-PP} is smaller than the one from \citep{gorbunov2023variance}.


In particular, the second term in the constant $A$ appears due to the partial participation, and the whole expression for $A$ is proportional to $\nicefrac{1}{p^2}$. In contrast, a similar constant $A$ from the result for \gls{Byz-VR-MARINA} is proportional to $\nicefrac{1}{p}$, which can be noticeably smaller than $\nicefrac{1}{p^2}$. Indeed, to make the expected number of clients participating in the communication round equal to $\cO(C)$, to make the expected number of stochastic oracle calls equal to $\cO(b)$, and to make the expected number of transmitted components for each worker taking part in the communication round equal $\cO(\zeta_{\cQ})$, parameter $p$ should be chosen as $p = \min\{\nicefrac{C}{M}, \nicefrac{b}{n}, \nicefrac{\zeta_{\cQ}}{d}\}$, where the latter term in the minimum often equals to $\Theta(\nicefrac{1}{(\omega+1)})$ \citep{gorbunov2021marina}. Therefore, in some scenarios, $p$ can be small.

Next, in the special case of full participation, we have $C=\widehat C=M$, $p_G = \cP_{\cG_{C}^t} = 1$, meaning that $A = \Theta(\nicefrac{(1+\omega)(1+c\delta)}{p^2})$ for \gls{Byz-VR-MARINA-PP}. In contrast, the corresponding constant for \gls{Byz-VR-MARINA} is of the order $\Theta(\nicefrac{(1+\omega)}{pM} + \nicefrac{(1+\omega)c\delta}{p^2})$, which is strictly better than our bound. In this special case, we do not recover the result for \gls{Byz-VR-MARINA}.

Such a complexity deterioration can be explained as follows: the presence of clipping introduces additional technical difficulties in the analysis, resulting in a reduced step size compared to \gls{Byz-VR-MARINA}, even when $C = \widehat{C} = M$. To achieve a more favorable convergence rate, particularly in scenarios of complete participation, we also establish the results under Assumption~\ref{assm:bounded-compressor}.
\begin{theorem}\label{thm:main_result_2}
 Let Assumptions \ref{assm:bounded-aggr}, \ref{assm:bounded-compressor}, \ref{assm:het_simplified}, \ref{assm:smoothness_simplified} hold, $\lambda_{t+1} = D_Q\cL \left\|x^{t+1} - x^t\right\|$, and $\widehat C \geq \max\{1, \nicefrac{\deltar M}{\delta}\}$. Assume that $0<\gamma \leq \nicefrac{1}{\cL(1+\sqrt{A})},$ 
where constant $A$ equals
\begin{eqnarray}
    A &=& \frac{4p_G  G\mathcal{P}_{\mathcal{G}^t_C}}{p(1-\delta)C}\left( \frac{3\omega + 2}{(1-\delta)C} + \frac{8(5\omega + 4)c\delta}{p}\right) + \frac{8(1-p_G) (2 + F_{\cA}^2 D_Q^2)}{p^2}.\label{eq:A_bounded_compr}
\end{eqnarray}
Then for all $T \geq 0$ the iterates produced by \gls{Byz-VR-MARINA-PP} (Algorithm \ref{alg:byz_vr_marina}) satisfy
\begin{equation}
    \mathbb{E}\left[\left\|\nabla f\left(\widehat{x}^T\right)\right\|^2\right] \leq \frac{2 \Phi^{(0)}}{\gamma(T+1)}+\frac{2 \widehat{D} \zeta^2}{p}, \label{eq:main_result_non_convex-q}
\end{equation}
where $\widehat{D} = \frac{ 2\delta\mathcal{P}_{\mathcal{G}^t_{\widehat{C}}} }{1-\delta} \left(\frac{6cG}{\widehat C} + p \right) + \widetilde{D}$, where $\widetilde{D} = 0$ when $\widehat{C} = M$, $\widetilde{D} = \frac{\cP_{\cG_{\widehat{C}}^t}G}{(1-\delta)\widehat{C}}$ when $\widehat{C} < M$, and $\widehat{x}^T$ is chosen uniformly at random from $x^0, x^1, \ldots, x^t$, and $$\Phi^{(0)}=f\left(x^0\right)-f^{\star}+\frac{\gamma}{p}\left\|g^0-\nabla f\left(x^0\right)\right\|^2.$$ If, in addition, Assumption~\ref{assm:PL} holds and $0<\gamma \leq \nicefrac{1}{\cL(1+\sqrt{2 A})},$ then for all $T \geq 0$ the iterates produced by \gls{Byz-VR-MARINA-PP} (Algorithm \ref{alg:byz_vr_marina}) satisfy with $\rho = \min\left\{\gamma\mu, \frac{p}{4}\right\}$
\begin{equation}
    \mathbb{E}\left[f\left(x^t\right)-f\left(x^{\star}\right)\right] \leq\left(1-\rho\right)^T \Phi^{(0)}+\frac{2\widehat{D}\zeta^2\gamma}{p\rho}, \label{eq:main_result_PL-q}
\end{equation}
where 
$$\Phi^{(0)}=f\left(x^0\right)-f^{\star}+\frac{2\gamma}{p}\left\|g^0-\nabla f\left(x^0\right)\right\|^2.$$
\end{theorem}

With Assumption~\ref{assm:bounded-compressor}, vectors $\{\cQ(\widehat{\Delta}_m(x^{t+1}, x^t))\}_{m\in \cG_C^t}$ can be upper bounded by $D_Q\cL \left\|x^{t+1} - x^t\right\|$. Using this fact, one can take the clipping level sufficiently large such that it is turned off for the regular workers. This allows us to simplify the proof and remove $\nicefrac{1}{p}$ factor in front of the terms not proportional to $\delta$ or to $1-p_G$ in the expression for $A$ that can make the stepsize larger. However, the second term in \eqref{eq:A_bounded_compr} can be larger than \eqref{eq:A_unbounded_compr}, since it depends on potentially large constant $D_Q$. Therefore, the rates of convergence from Theorems \ref{thm:main_result_1} and \ref{thm:main_result_2} cannot be compared directly. We also highlight that the clipping level from Theorem~\ref{thm:main_result_2} is in general larger than the clipping level from Theorem~\ref{thm:main_result_1} and, thus, it is expected that with full participation Theorem~\ref{thm:main_result_2} gives better results than Theorem~\ref{thm:main_result_1}: the bias introduced due to the clipping becomes smaller with the increase of the clipping level. However, in the partial participation regime, the price for this is a decrease of the stepsize to compensate for the increased harm from Byzantine clients in situations when they form a majority. Further discussion of the technical challenges we overcame is deferred to Appendix~\ref{appendix:technical_challenges}.

Nevertheless, in the case of full participation, we have $C=\widehat C=M$, $p_G = \cP_{\cG_{C}^t} = \cP_{\cG_{\widehat C}^t} = 1$, meaning that $A = \Theta(\nicefrac{(1+\omega)}{pM} + \nicefrac{(1+\omega)c\delta}{p^2})$ in Theorem~\ref{thm:main_result_2}. That is, in this case, we recover the result of \gls{Byz-VR-MARINA}. More generally, if $p_G = 1$, which is equivalent to $C \geq \max\{1, \nicefrac{\deltar M}{\delta}\}$, then $\cP_{\cG_C^t} = \PP\{m \in \cG_C^t\} = \min\{1, \nicefrac{C}{G}\}$, $\cP_{\cG_{\widehat{C}}^t} = \PP\{m \in \cG_{\widehat C}^t\} = \min\{1, \nicefrac{\widehat C}{G}\}$ and we have $A = \Theta(\nicefrac{(1+\omega)}{p C} + \nicefrac{(1+\omega)c\delta}{p^2})$. Here, the first term in $A$ is $\nicefrac{M}{C}$ worse than the corresponding term for \gls{Byz-VR-MARINA}. However, the second term in $A$ matches the corresponding term for \gls{Byz-VR-MARINA}. Moreover, this term is the main one if $c\delta \geq \nicefrac{p}{C}$, which is typically the case since parameter $p$ is often small ($p = \min\{\nicefrac{C}{\widehat C}, \nicefrac{b}{n}, \nicefrac{\zeta_{\cQ}}{d}\}$). In such cases, \gls{Byz-VR-MARINA-PP} has the same rate of convergence as \gls{Byz-VR-MARINA} while utilizing, on average, just $\cO(C)$ workers at each step in contrast to \gls{Byz-VR-MARINA} that uses $n$ workers at each step. \emph{That is, in some cases, partial participation is provably beneficial for \gls{Byz-VR-MARINA-PP}.} 

\paragraph{\algname{Byz-VR-MARINA+}: simplified version of \gls{Byz-VR-MARINA}.} In Appendix~\ref{appendix:simplified+}, we propose a simplified version of \gls{Byz-VR-MARINA} called \algname{Byz-VR-MARINA+} (Algorithm~\ref{alg:byz_vr_marina+}). The only difference is related to Line~\ref{line:g^t+1} of the method: when $c_t = 0$, \algname{Byz-VR-MARINA+} computes just the average of clipped compressed vectors $$\left\{\clip_{\lambda_{t+1}}\left(\cQ\left(\widehat{\Delta}_m(x^{t+1}, x^t)\right)\right)\right\}_{m\in \set}$$ instead of robust aggregation, while keeping \gls{ARAgg} when $c_t = 1$. Of course, when $c_t = 0$ and at least one Byzantine worker is sampled, then the step can be useless, but the ``harm'' of this step is bounded due to the clipping. However, in certain regimes (e.g., when $C$ is small enough and the number of Byzantine workers is much smaller than the number of regular workers), the probability of sampling only regular workers is larger than sampling at least one Byzantine worker when $c_t = 0$, meaning that with high enough probability the resulting estimator has no additional bias coming from the robust aggregation. We formally analyze \algname{Byz-VR-MARINA+} and show that such a modification of the method leads to better theoretical results (especially when $C$ is small). In particular, in the settings of Theorem~\ref{thm:main_result_2}, we prove that \algname{Byz-VR-MARINA+} exhibits the same $\cO(\nicefrac{1}{T})$ rate but converges to $\cO(\nicefrac{1}{p})$ smaller neighborhood when $\widehat{C} = M$, i.e., the neighborhood term for \algname{Byz-VR-MARINA+} is optimal \citep{karimireddy2020byzantine, allouah2024robust}. Moreover, our results for \algname{Byz-VR-MARINA+} allow larger stepsizes when $C$ is small enough. For further details and complete proofs, we refer to Appendix~\ref{appendix:simplified+}.

\paragraph{Extensions without full-batch gradient computations.} The proposed new methods -- \gls{Byz-VR-MARINA} and \algname{Byz-VR-MARINA+} -- have a common limitation related to the full-batch gradient computation with probability $p$. Although this probability is typically small, even one full-gradient computation can be very expensive for certain problems. To address this issue, we propose the modifications of \gls{Byz-VR-MARINA} and \algname{Byz-VR-MARINA+} without full-batch gradient computations at all (see Algorithms~\ref{alg:byz_vr_marina_no_full} and \ref{alg:byz_vr_marina+_no_full} in Appendix~\ref{appendix:no_full_grads}). That is, these modifications differ from \gls{Byz-VR-MARINA} and \algname{Byz-VR-MARINA+} in Line~\ref{line:g_i^k+1_main} only: when $c_t = 1$, every good worker $m$ from $\set$ computes and sends to the server $b'$-size mini-batched stochastic gradient estimator $\widetilde \nabla f_m(x^{t+1})$ of $\nabla f_m(x^{t+1})$. Under the additional assumption that the variance of $\widetilde \nabla f_m(x^{t+1})$ is uniformly bounded by $\nicefrac{\sigma^2}{b'}$, which is a standard assumption for variance-reduced methods without full-batch gradient computations \citep{fang2018spider, cutkosky2019momentum, li2021page, gorbunov2021marina}, we prove that both methods converge similarly as in the case of the (periodical) full-batch gradient computations but to the neighborhood having an additional term proportional to $\left(c\delta + \nicefrac{\cP_{\cG_{\widehat{C}}^t}G}{\widehat{C}^2} \right)\nicefrac{\sigma^2}{b'}$. For further details and complete proofs, we refer to Appendix~\ref{appendix:no_full_grads}.

\paragraph{Heuristic extension of \gls{Byz-VR-MARINA-PP}.} In this short remark, we illustrate how the proposed clipping technique can be applied to a general class of Byzantine-robust methods to adapt them to the case of partial participation. Consider the methods having the following update rule: $x^{t+1} = x^t - \gamma \cdot \texttt{Agg}(\{g_m^t\}_{m\in[M]})$, where $\{g_m^t\}_{m\in[M]}$ are the vectors received from workers at iteration $t$ and $\texttt{Agg}$ is some aggregation rule. The vast majority of existing Byzantine-robust methods fit this scheme. In the case of partial participation of clients, we propose to modify the scheme as follows:
\begin{equation}
    x^{t+1} = x^t - \gamma g^t,\quad \text{where } g^t \eqdef g^{k-1} + \texttt{Agg}\left(\left\{\clip_{\lambda_k}(g_m^t - g^{k-1})\right\}_{m\in \set}\right), \label{eq:heuristic}
\end{equation}
where $\set \subseteq [M]$ is a subset of clients participating in round $t$ and $\{\lambda_k\}_{k\geq 0}$ is a sequence of clipping parameters specified by the server. In particular, \gls{Byz-VR-MARINA-PP} can be seen as an application of scheme \eqref{eq:heuristic} to \gls{Byz-VR-MARINA} (up to a minor modification when $c_t = 1$ in \gls{Byz-VR-MARINA}) with $\lambda_{t+1} = \lambda\|x^{t+1} - x^{t}\|$. We suggest using $\lambda_{t+1} = \lambda\|x^{t+1} - x^{t}\|$ with tunable parameter $\lambda > 0$ for other methods as well.

\begin{figure*}[t]
\centering
\includegraphics[width=0.32\textwidth]{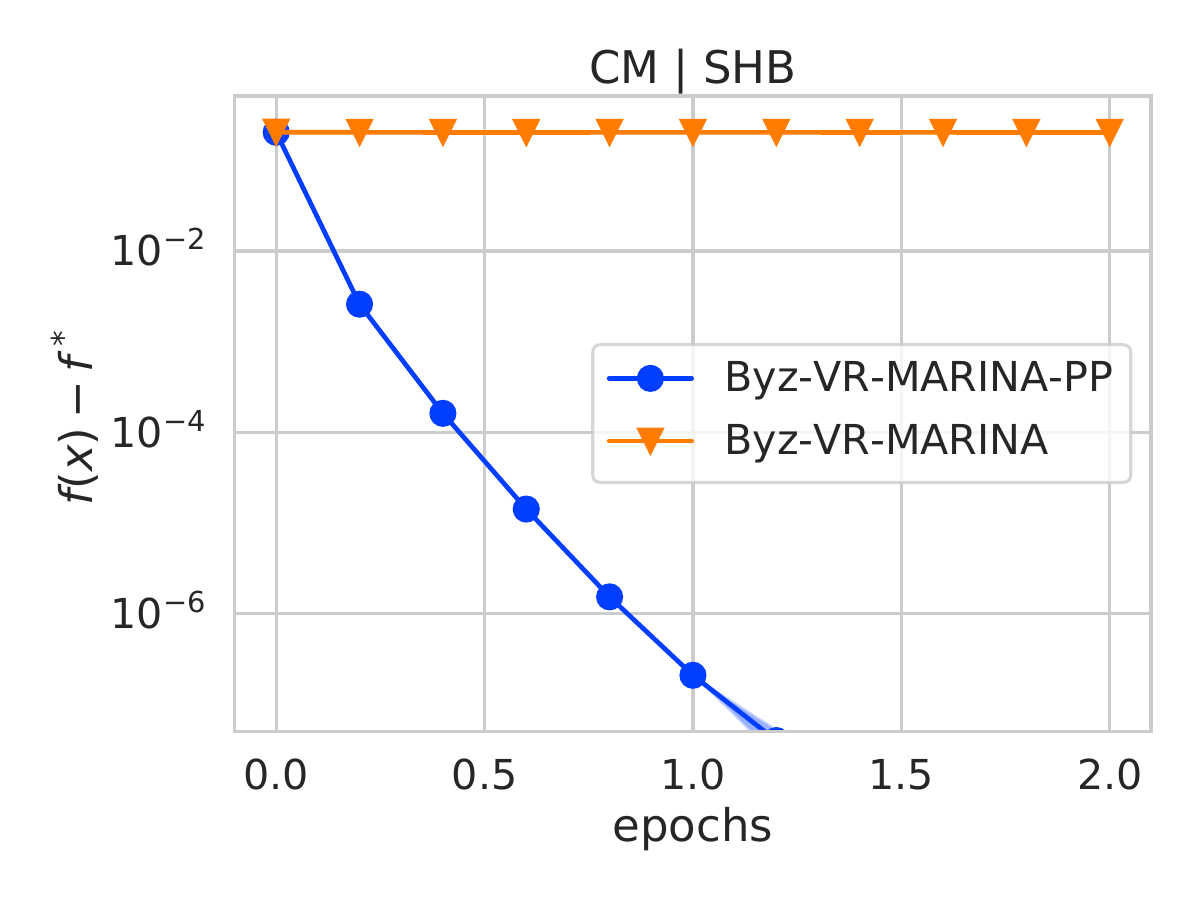}
\includegraphics[width=0.32\textwidth]{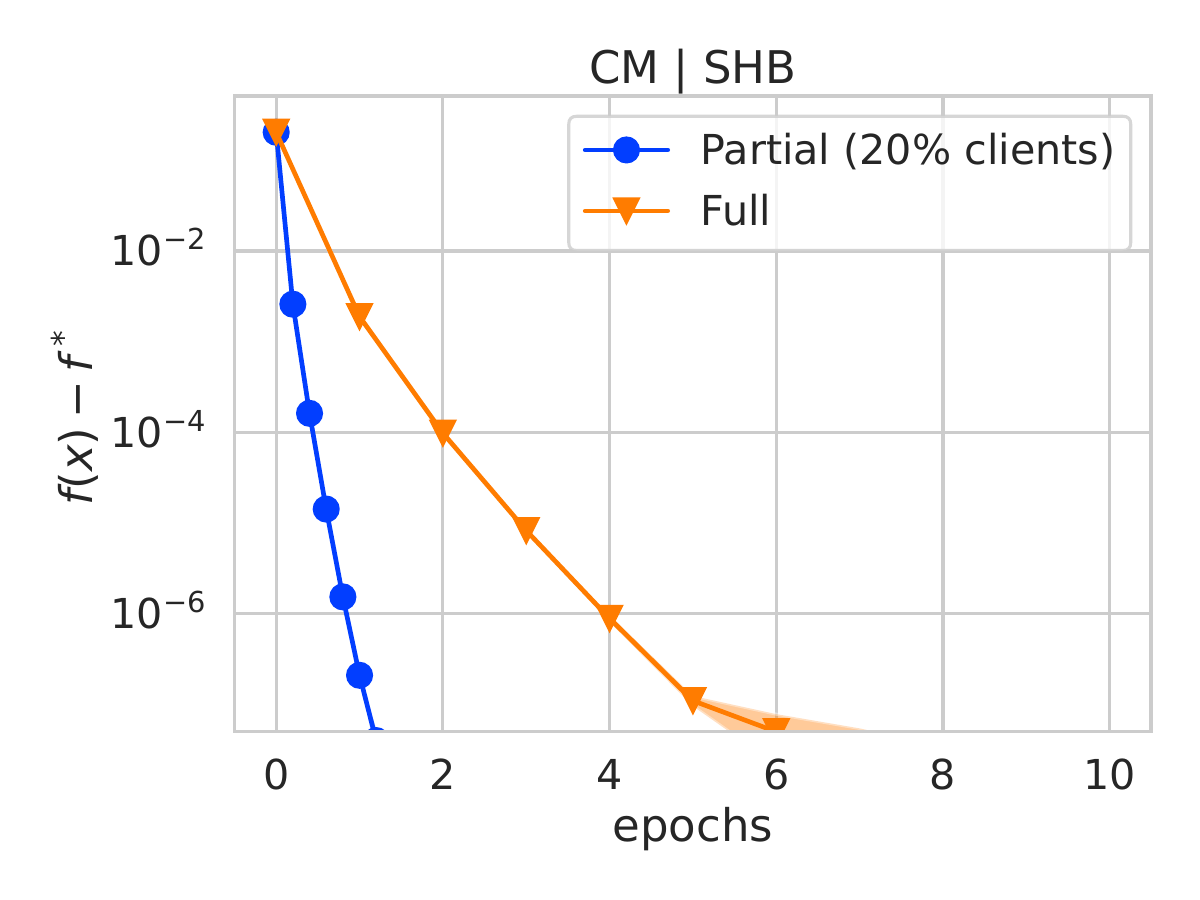}
\includegraphics[width=0.32\textwidth]{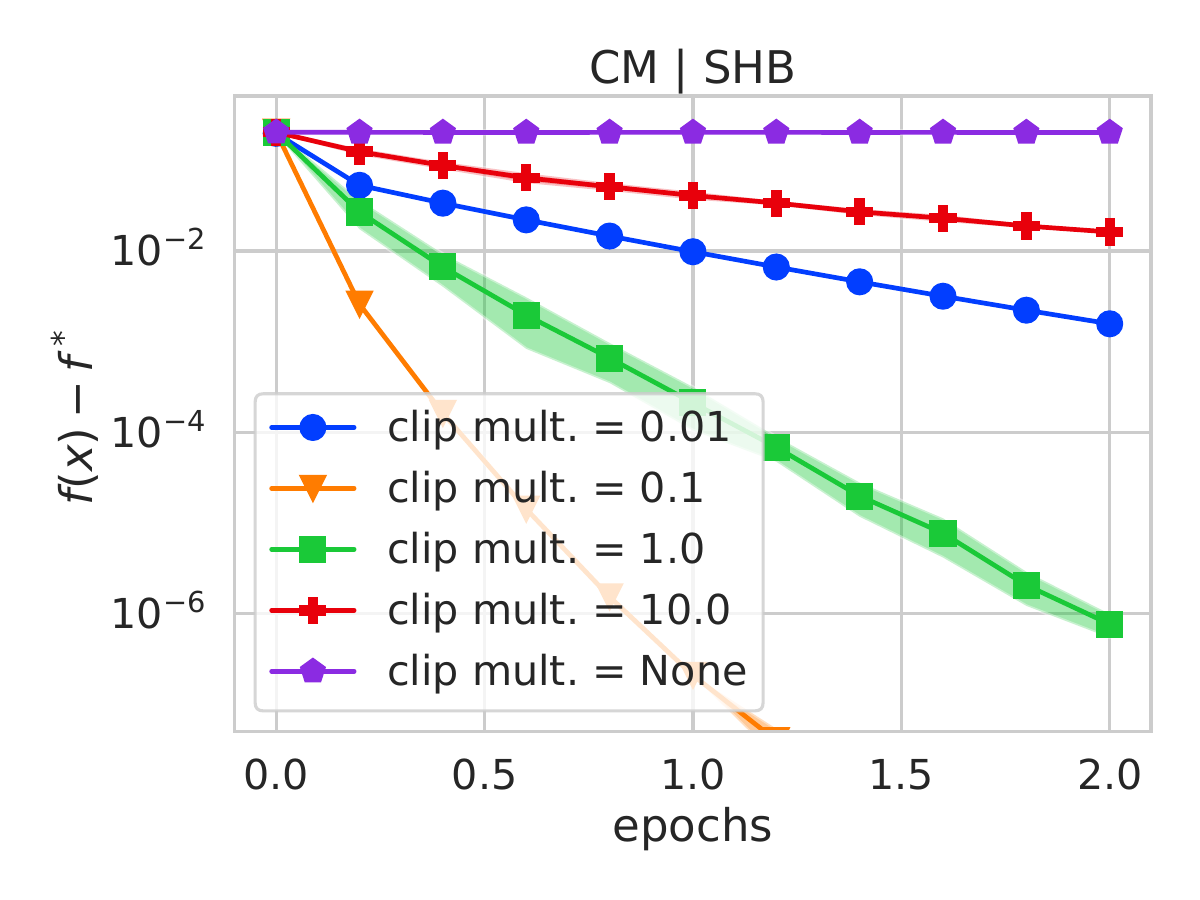}
\caption{
The optimality gap $f(x^t) - f(x^{\star})$ for 3 different scenarios. We use coordinate-wise mean with bucketing equal to 2 as an aggregation and shift-back as an attack.  We use the a9a dataset, where each worker accesses the full dataset with 15 good and 5 Byzantine workers. We do not use any compression. In each step, we sample 20\% of clients uniformly at random to participate in the given round unless we specifically mention that we use full participation. Left: Linear convergence of \gls{Byz-VR-MARINA-PP} with clipping versus non-convergence without clipping. Middle: Full versus partial participation, showing faster convergence with clipping. Right: Clipping multiplier $\lambda$ sensitivity, demonstrating consistent linear convergence across varying $\lambda$ values.} 
\label{fig:a9a_full}
\end{figure*}

\section{Experiments}\label{section:experiments}
\vspace{-0.5em}

Firstly, we showcase the benefits of employing clipping to mitigate the presence of Byzantine workers and partial participation. For this task, we consider the standard logistic regression model with $\ell_2$-regularization, i.e., $f_{m,i} (x) =  - y_{m,i}\log(h(x, a_{m,i})) - (1 - y_{m,i}) \log(1 - h(x, a_{m,i})) + \eta \|x\|^2,$ 
where $y_{m,i} \in \{0, 1\}$ is the label, $a_{m,i} \in \R^d$ represents the feature vector, $\eta$ is the regularization parameter, and $h(x, a) = \nicefrac{1}{(1 + e^{-a^\top x})}$. This objective is smooth, and for $\eta > 0$, it is also strongly convex, satisfying the P\L-condition. We consider the \textit{a9a} LIBSVM dataset~\citep{chang2011libsvm} and set $\eta = 0.01$. In the experiments, we focus on an important feature of \gls{Byz-VR-MARINA-PP}: it has linear convergence for homogeneous datasets across clients even in the presence of Byzantine workers and partial participation, as shown in Theorems~\ref{thm:main_result_1} and \ref{thm:main_result_2}. 

To demonstrate this experimentally, we consider the setup with 15 good workers and 5 Byzantines, where \textit{each worker can access the entire dataset}, and where the server uses coordinate-wise median with bucketing as the aggregator (see also Appendix~\ref{appendix:justification_of_Assumption_1}). For the attack, we propose a new attack that we refer to as the \textit{shift-back} attack, which acts in the following way. If Byzantine workers are in the majority in the current round $t$, then each Byzantine worker sends $x^0 - x^t$. Otherwise, they follow protocol and act as benign workers. Further experimental details are deferred to Appendix~\ref{app:experiments}.


We compare our \gls{Byz-VR-MARINA-PP} with its version without clipping. We note that the setup that we consider is the most favorable in terms of minimized variance in terms of data and gradient heterogeneity. We show that even in this simplest setup, the method without clipping does not converge since there is no method that can withstand a Byzantine majority. Therefore, any more complex scenario would also fall short using our simple attack. On the other hand, we show that once clipping is applied, \gls{Byz-VR-MARINA-PP} is able to converge linearly to the exact solution, complementing our theoretical results.

Figure~\ref{fig:a9a_full} showcases these observations. On the left, we can see \gls{Byz-VR-MARINA-PP} converges linearly to the optimal solution, while the version without clipping remains stuck at the starting point since Byzantines are always able to push the solution back to the origin because they can create a majority in some rounds. In the middle plot, we compare the full participation scenario in which all the clients participate in each round, which does not require clipping since, in each step, we are guaranteed that Byzantines are not in the majority, to partial participation with clipping. When comparing the total number of computations (measured in epochs), we can see that \gls{Byz-VR-MARINA-PP} leads to faster convergence even though we need to employ clipping. Finally, in the right plot, we measure the sensitivity of clipping multiplier $\lambda$. We can see that \gls{Byz-VR-MARINA-PP} is not very sensitive to $\lambda$ in terms of convergence, i.e., for all the values of $\lambda$, we still converge linearly. However, the suboptimal choice of $\lambda$ leads to slower convergence.

\begin{figure*}[t]
\centering
\includegraphics[width=0.45\textwidth]{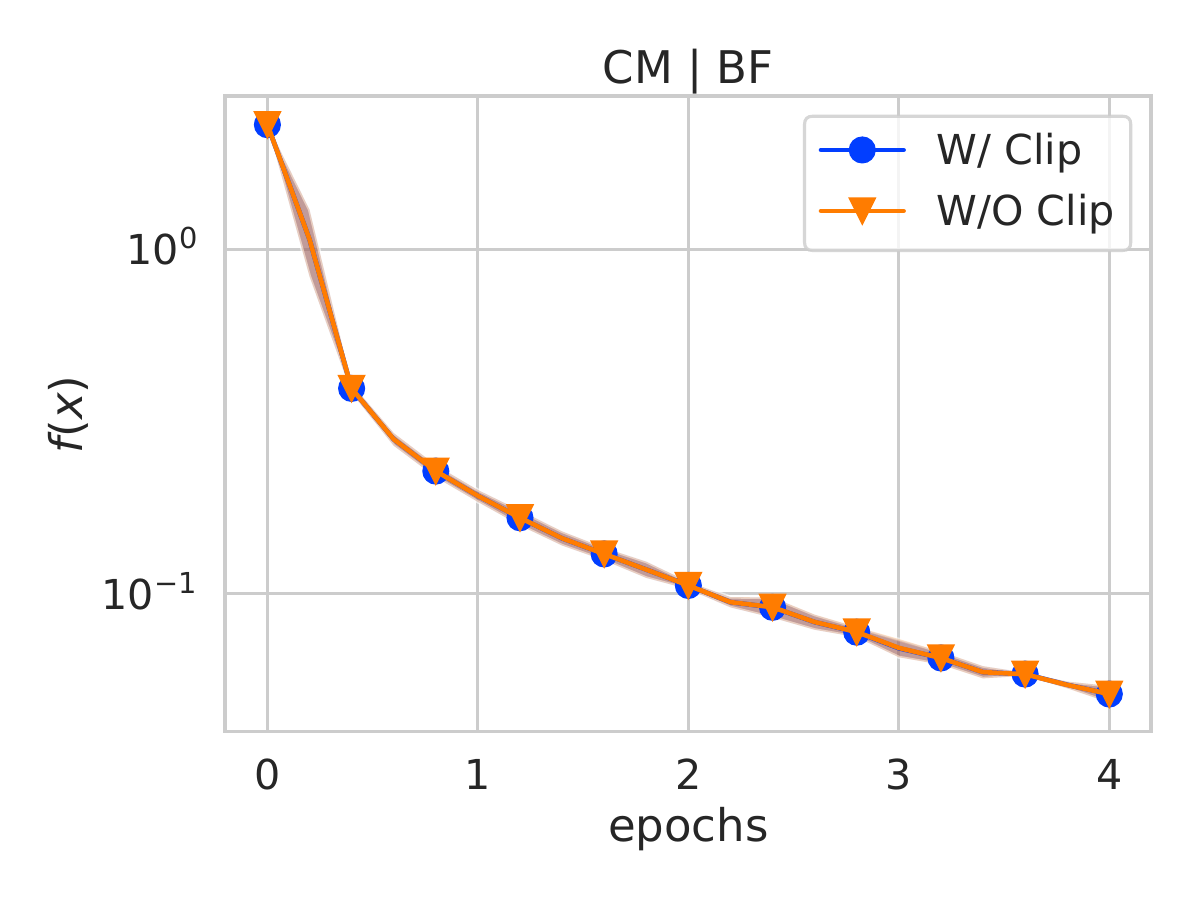}
\includegraphics[width=0.45\textwidth]{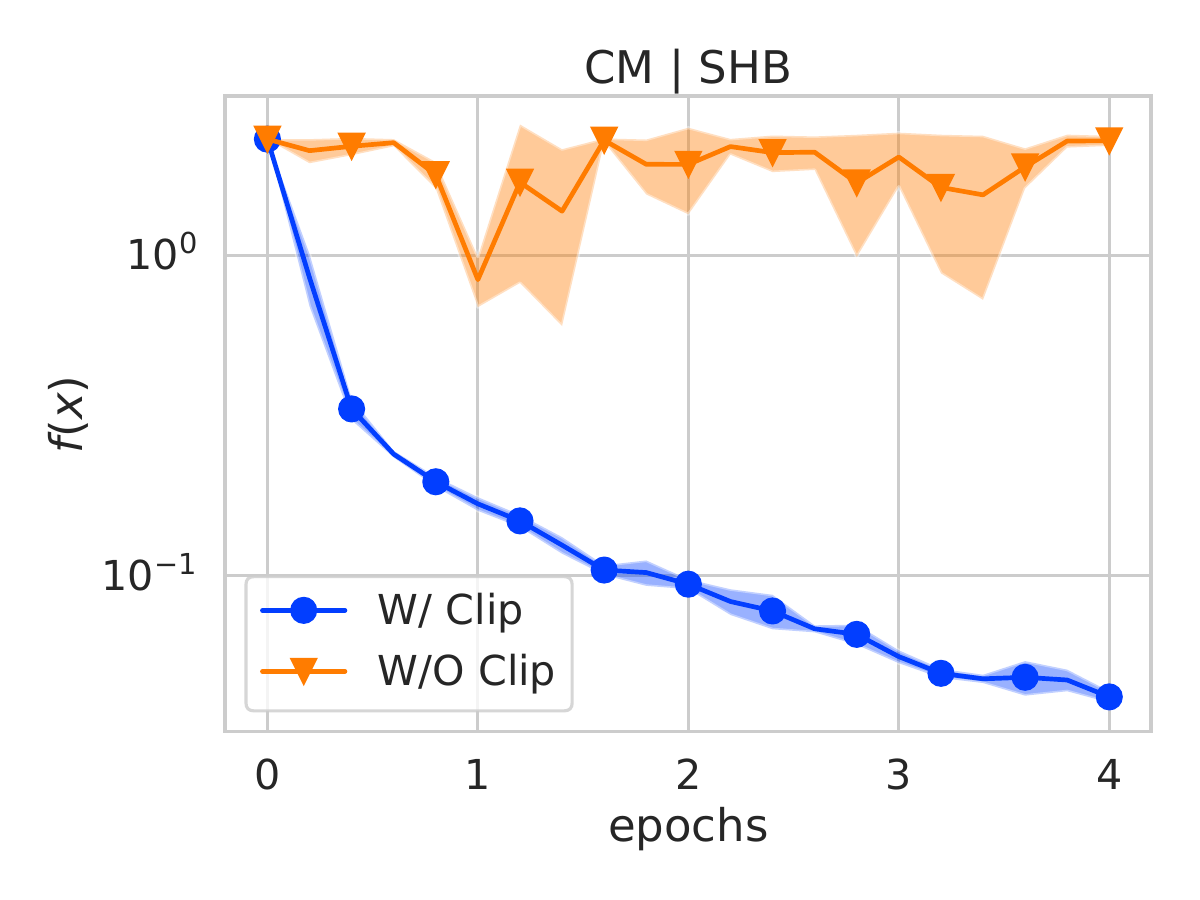}\\
\includegraphics[width=0.45\textwidth]{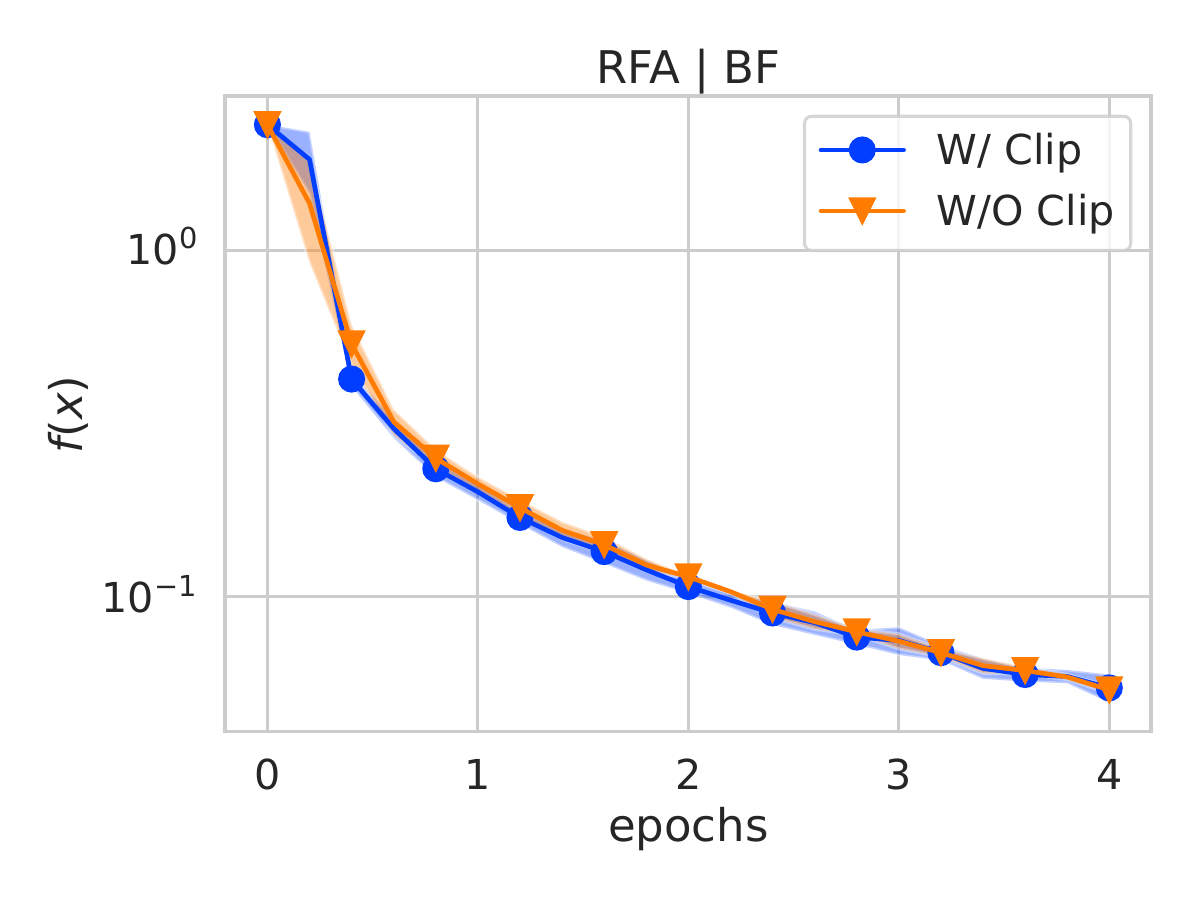}
\includegraphics[width=0.45\textwidth]{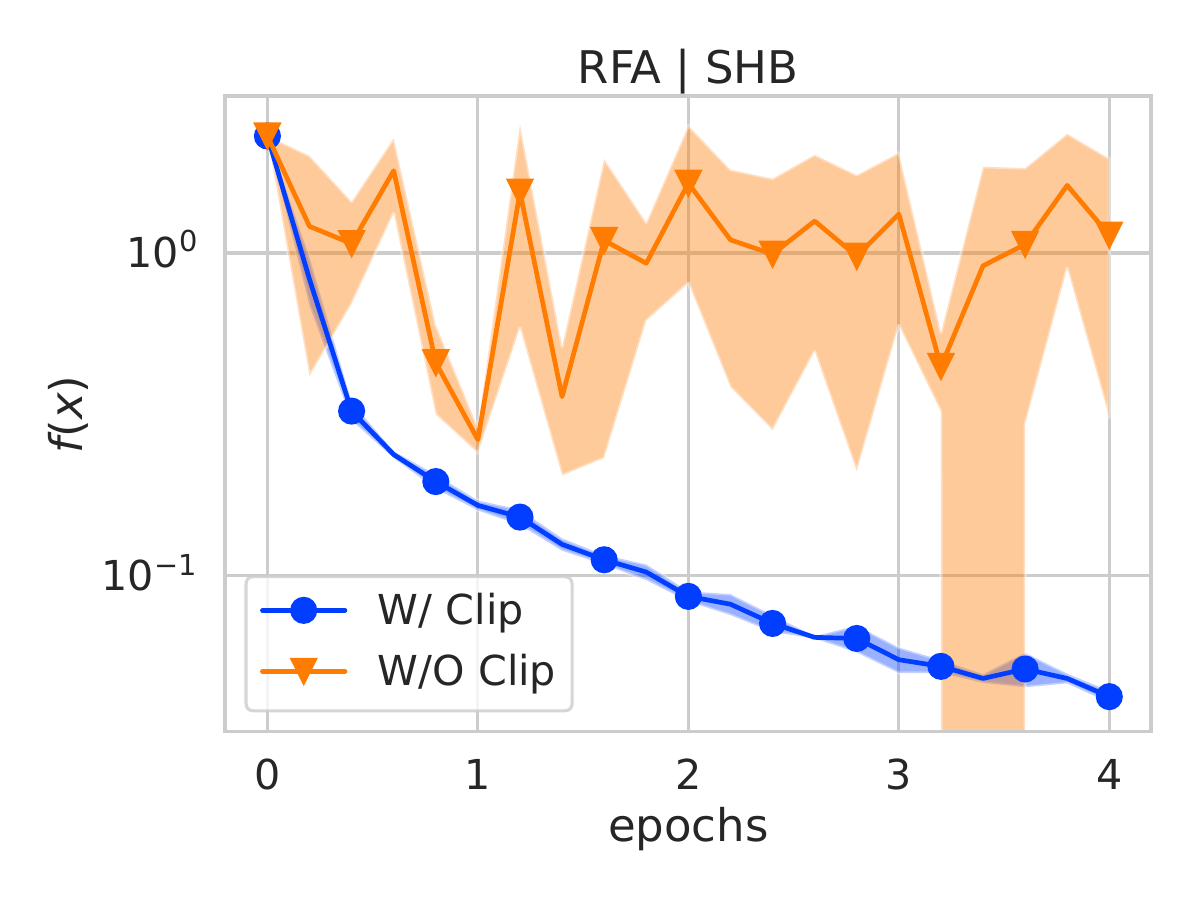}
\caption{
Training loss of 2 aggregation rules (CM, RFA) under 2 attacks (BF, SHB) on the MNIST dataset under heterogeneous data split with 20 clients, 
5 of which are malicious. Additional experiments on CIFAR10 are provided in Appendix~\ref{app:experiments}.} 
\label{fig:nn}
\vspace{-1.5em}
\end{figure*}

Furthermore, we also realize that other attacks and more complicated experiments could potentially damage clipping more than methods not using clipping. Therefore, we provide additional experiments with neural networks and different attacks in heterogeneous settings. For our experimental setup, we follow \citep{karimireddy2021learning}. However, when working with neural networks, the choice of standard variance reduction is not effective~\citep{defazio2019ineffectiveness}. Therefore, we use Byzantine Robust Momentum \gls{SGD}~\citep{karimireddy2021learning} as an underlying optimization method; see \eqref{eq:heuristic}. 

We consider the MNIST dataset~\citep{mnist} with heterogeneous splits of the dataset with 20 clients, 5 of which are malicious. For the attacks of malicious clients, we consider a specific strategy: \gls{ALIE}~\citep{baruch2019little}, \gls{BF}, and \gls{SHB}. For the aggregations, we consider coordinate median (CM)~\citep{chen2017distributed} and \gls{RFA}~\citep{pillutla2022robust} with bucketing. 

From Figure~\ref{fig:nn}, we can see that clipping does not lead to performance degradation. On the contrary, clipping performs on par or better than its variant without clipping. Furthermore, we can see that no robust aggregator is able to withstand the shift-back attack without clipping.


\chapter{RAC-LoRA: Provable Convergence for Low Rank Adaptation via Randomized Asymmetric Chains}
\label{chapter8}
\thispagestyle{empty}

\section{Introduction}
\label{introduction_}

Many real-world Deep Learning (\gls{DL}) applications require adapting a large pre-trained model to specific tasks in order to improve its performance \citep{church2021emerging}. This process, known as fine-tuning, involves adjusting the model from its pre-trained state to better handle the nuances of particular tasks or domains. Fine-tuning is a specialized form of transfer learning, where knowledge gained during pre-training is adapted for new, specific applications \citep{vrbanvcivc2020transfer}.

\textbf{Parameter-Efficient Fine-Tuning.}
While fine-tuning all model parameters has been effective, modern models with billions of parameters pose significant challenges due to their scale. Full-parameter fine-tuning is often computationally impractical with standard resources. To address this challenge, \gls{PEFT} \citep{he2021towards} proposes updating a subset of parameters only \citep{PCDM}, or adding task-specific modules \citep{xu2023parameter}. \gls{PEFT} reduces computational costs by modifying fewer parameters or adding external modules, enabling more efficient resource use and lowering storage requirements. This approach significantly reduces both training time and computational demands, making it a practical solution for adapting large models to new tasks \citep{han2024parameter}.

\subsection{Low-Rank adaptation (LoRA)}
One of the most popular \gls{PEFT} methods is \gls{LoRA} \citep{hu2021lora}. The core idea behind \algname{\gls{LoRA}} is that fine-tuning large pre-trained models can be effectively achieved by utilizing lower-dimensional parameter spaces \citep{li2018measuring, aghajanyan2020intrinsic}. Instead of updating all parameters of a large and potentially dense matrix associated with the weights of a linear layer, \algname{\gls{LoRA}} works with the product of two trainable low-rank matrices, which significantly reduces the number of parameters updated during fine-tuning. These matrices are trained such that their product is added to the pre-trained model weights.


In \algname{\gls{LoRA}}  \citep{hu2021lora}, the weight adaptation is represented as the product of two low-rank matrices (and a scalar multiplier), resulting in the final model 
\begin{align*}
        W = W^0 + \frac{\alpha}{r}BA,
\end{align*}
where $ W^0 \in \mathbb{R}^{m \times n} $, $ B \in \mathbb{R}^{m \times r} $, and $ A \in \mathbb{R}^{r \times n} $. Here, $r$ and $\alpha$  respectively denote the \algname{\gls{LoRA}} rank and its scaling factor. Typically, since the dimensions of (particularly deep learning) models are enormous, we have rank $ r \ll \min \{m, n\} $.
 This approach saves computational resources and 
minimizes the risk of overfitting or catastrophic forgetting \citep{biderman2024lora}. Hence, \algname{\gls{LoRA}} has become a lightweight and efficient technique for adapting large models to various tasks, particularly in resource-constrained environments \citep{sun2022recent}.
It is important to note that $ W^0 $ remains fixed and does not receive updates, while $ A $ and $ B $ are optimized during the training process. The scaling factor $ \alpha $ serves as a ``step size'' for the adaptation, and it is normalized by rank $r$. The matrix \( A \) is typically initialized with random Gaussian values, while the matrix \( B \) is set to zero, ensuring \( \Delta W = 0 \) at the start of training. Alternative initialization strategies were explored in \citep{zhu2024asymmetry}.

\subsection{Chain of LoRA (COLA)}
While \algname{\gls{LoRA}} offers significant computational advantages in practice, it remains less effective than \gls{FPFT} if efficiency is not a major concern \citep{biderman2024lora}.
To balance efficiency and performance, an iterative method called Chain of LoRA (\algname{COLA}) is proposed. Essentially, \algname{COLA} simply means the successive application of several \algname{\gls{LoRA}} updates \citep{xia2024chain}.  

Chain of LoRA (\algname{COLA}) constructs a sequence of \algname{\gls{LoRA}} modules through an iterative process of parameter fine-tuning, merging, and extending. The chain length is defined by the number of optimized \algname{\gls{LoRA}} modules. \algname{COLA}'s central concept involves applying \algname{\gls{LoRA}} adaptations iteratively $T$ times. \algname{COLA} can be summarized as training a \algname{\gls{LoRA}} module, merging the updates with the fixed parameters, reinitializing the \algname{\gls{LoRA}} matrices, and repeating the process \citep{xia2024chain}. The resulting model can be represented by:
$$
    W = W^0 + \frac{\alpha}{r} \sum_{t=0}^{T-1} B^t A^t,
$$
where $A^t$ and $B^t$ indicate the low-rank matrices in the $t$-th block in the chain, which are typically initialized in the same manner as in standard \algname{\gls{LoRA}}. The motivation behind \algname{COLA} is that standard \algname{\gls{LoRA}} may clearly fail to find the optimal adaptation since such an adaptation may not in general be of a low rank. To address this, \algname{COLA} proposes using a sequence of low-rank matrix decompositions to approximate a middle-to-high-rank update. The hypothesis is that this sequence of updates can provide a better approximation than a single \algname{\gls{LoRA}} adaptation and may be easier to optimize compared to learning the optimal adaptation from scratch. 

\section{Problem Formulation and Summary of Contributions}

\subsection{Problem formulation}
The primary approach for training supervised Machine Learning models is to formulate the task as an optimization problem where the goal is to minimize a loss function, which measures the discrepancy between the model's predictions and the actual outcomes. In this work, we explore this optimization problem in the specific context of fine-tuning, where a pre-trained model is adapted to a new task or dataset, requiring efficient adjustments to its parameters to achieve better performance on the target task. In particular, we consider the model-agnostic problem formulation
\begin{align}
\label{eq:main_Lora}
 \min_{\Delta W \in \mathbb{R}^{m\times n}}   f(W^0 + \Delta W) , 
\end{align}
where \( W^0 \in \mathbb{R}^{m \times n} \) represents the parameters of a pre-trained model (or of a single linear layer, with the others being fixed), and \( \Delta W \in \mathbb{R}^{m \times n} \) denotes the adaptation term. 
The function \( f: \mathbb{R}^{m \times n} \to \R \) corresponds to the empirical loss over the adaptation dataset, or any other loss function of interest. As the total dimensionality \( m \times n \) is typically very large for deep learning models, the adaptation term $\Delta W$ needs to have a specific structure to be feasible in real-world applications. 


\subsection{No reasonable theory for Low-Rank adaptation}

We claim that a satisfying theoretical understanding of prevalent fine-tuning methods based on low-rank updates, such as \algname{\gls{LoRA}} and \algname{COLA}, is lacking. 

\setlist{nolistsep}
\begin{itemize}[noitemsep]
\item  First, as already noted in \citep{sun2024improving}, the \algname{\gls{LoRA}} re-parameterization of the domain effectively transforms a  {\em smooth} Lipschitz  loss into a {\em non-smooth} Lipschitz  loss, which poses {\em additional} theoretical challenges to those related to proper handling of the low-rank structure of the updates. While this hints at a possible source of issues with providing a good theory for methods based on low-rank adaptation, this observation does not on its own mean that a good theory is impossible to obtain. 

\item More importantly, the  existing theoretical analysis of \algname{COLA}  \citep{xia2024chain} replaces low-rank optimization over matrices \(A\) and \(B\) with full-rank matrix optimization (\(\Delta W\)). This makes the theoretical analysis irrelevant at worst and unsatisfactory at best as it completely fails to model and to explain the key component of \algname{\gls{LoRA}}: low-rank updates.

\item Third, it is known that \algname{\gls{LoRA}} can be highly sensitive to the choice of the hyper-parameters \citep{khodak2021federated, kuang2024federatedscope}. A good theory should be able to explain or remove this issue. No such theory exists, to the best of our knowledge.

\item Finally, and this is the true starting point of our exploration in this work, we observe that \algname{COLA} may simply {\em fail to converge} to the optimal solution. We give a simple example (with \(3 \times 3\) matrices) of this divergence behavior 
in Section~\ref{sec:lora_convergence_issue}. Hence, \algname{COLA} is merely a {\em heuristic}. Providing a fix is an open problem -- the problem we address in this work.

\end{itemize}

While \algname{\gls{LoRA}} and \algname{COLA}  are useful in practice, these methods remain mere {\em heuristics} since they do not come with solid theoretical backing. This is problematic and raises valid concerns about the robustness and reliability of \algname{\gls{LoRA}}-type methods in scenarios {\em beyond} current datasets, models and practice.

\subsection{Contributions}

To address the aforementioned fundamental issues of \algname{\gls{LoRA}}-type heuristics, and to firmly ground the fine-tuning-via-low-rank adaptation line of work in a theoretically sound algorithmic framework,  we propose a new generic low-rank adaptation framework for which we coin the name \algname{\gls{RAC-LoRA}}; see Algorithm~\ref{alg:RAC-LoRA}.

\begin{table}[t]
\centering
\scriptsize
\caption{Summary of our theoretical convergence results for \algname{\gls{RAC-LoRA}} for solving the problem in Equation (\ref{eq:main_Lora}) when using a specific optimizer for approximately solving the subproblem in Step 4. The results for the \algname{\gls{RAC-LoRA}} + \algname{\gls{GD}} combination are described in Section~\ref{sec:theory_Lora}, while the proofs can be found in Appendix~\ref{sec:GD-proofs}. The results and proofs for all other combinations can be found in the indicated appendices.}
\begin{tabular}{llccc}
\toprule
{\bf Problem} & {\bf Fine-tuner} & \textbf{Subproblem Optimizer} & \textbf{Non-convex} & \textbf{\gls{PL}} \\\midrule
(\ref{eq:main_Lora}) & \algname{\glsentryshort{RAC-LoRA}} & \begin{tabular}{c} {\small Gradient Descent } \\ (\algname{GD}) \end{tabular} & \begin{tabular}{c} ${\cal O}\left(1/T\right)$ \\ 
Sec.~\ref{sec:GD-pr} \end{tabular} & \begin{tabular}{c} ${\cal O}\left(\exp(-T)\right)$ \\Sec.~\ref{sec:GD-PL}\end{tabular}    \\
\hline
(\ref{eq:main_Lora})+(\ref{eq:finite}) & \algname{\glsentryshort{RAC-LoRA}} & \begin{tabular}{c} {\small Random Reshuffling} \\ (\algname{\gls{RR}}) \end{tabular}  &  \begin{tabular}{c}  ${\cal O}\left(1/T^{\frac{2}{3}}\right)$  \\ Sec.~\ref{sec:RR-gen} \end{tabular} &  \begin{tabular}{c}  ${\cal O}(1/T^2)$  \\ Sec.~\ref{sec:RR-PL} \end{tabular}   \\
\hline
(\ref{eq:main_Lora}) & \algname{\glsentryshort{RAC-LoRA}} & \begin{tabular}{c} {\small Stoch. Gradient Descent }\\ (\algname{\gls{SGD}}) \end{tabular} & \begin{tabular}{c} ${\cal O}\left(1/T^{\frac{1}{2}}\right)$ \\ Sec.~\ref{sec:SGD-gen} \end{tabular} &  \begin{tabular}{c}  ${\cal O}\left(1/T\right)$ \\ Sec.~\ref{sec:SGD-PL}\end{tabular}  \\ 
\hline
(\ref{eq:main_Lora})+(\ref{eq:fed}) & \glsentryshort{Fed-RAC-LoRA} & \begin{tabular}{c} {\small Random Reshuffling}  \\(\algname{\gls{RR}}) \end{tabular}& \begin{tabular}{c}  ${\cal O}\left(1/T^{\frac{1}{2}}\right)$ \\ Sec.~\ref{sec:Fed-gen} \end{tabular} &  \begin{tabular}{c} ${\cal O}\left(1/T\right)$ \\ Sec.~\ref{sec:Fed-PL} \end{tabular} \\ 
\bottomrule
\end{tabular}
\label{tab:theory}
\end{table}

\begin{itemize}
\item Similarly to \algname{COLA} \citep{xia2024chain}, our method is iterative: we perform a chain of low-rank updates (see Step 2 in Algorithm~\ref{alg:RAC-LoRA}). In each step of the chain, 
one matrix (e.g., $A$) is chosen randomly from a pre-defined distribution, and the other (e.g., $B$) is  trainable (see Step 3 in Algorithm~\ref{alg:RAC-LoRA}). Which of these two update matrices is chosen randomly and which one is trainable is decided a priori, and hence our method is asymmetric in nature, similarly to \algname{AsymmLoRA} \citep{zhu2024asymmetry}. We propose two options, depending on which matrix is trainable and which one is chosen randomly: in Option 1, $A$ is trainable, and in Option 2, $B$ is trainable. 
\item 
In order to make our framework flexible, we offer a variety of strategies for updating the trainable matrix in each step of the chain. This is possible since in each such step we formulate an auxiliary optimization subproblem in the trainable matrix, and one can thus choose essentially {\em any optimizer} for approximately solving it (see Step 4 in Algorithm~\ref{alg:RAC-LoRA}). We theoretically analyze several such optimizers within our \algname{\gls{RAC-LoRA}} framework, including Gradient Descent (\algname{\gls{GD}}) in Appendix~\ref{sec:GD-proofs} (however, we include and describe the theorems in Section~\ref{sec:GD}),  Random Reshuffling (\algname{\gls{RR}}) in Appendix~\ref{sec:RR-LoRA}, and Stochastic Gradient Descent (\algname{\gls{SGD}}) in Appendix~\ref{sec:SGD}. In the case of \algname{\gls{GD}} and \algname{\gls{SGD}}, just a single step of the optimizer is sufficient, and this is what our analysis accounts for. In the case of \algname{\gls{RR}}, we apply a single pass over the data in a randomly reshuffled order. See Table~\ref{tab:theory} for a quick overview. Our analysis applies to the smooth non-convex regime, in which we prove fast sublinear (i.e., ${\cal O}(1/\sqrt{T})$, ${\cal O}(1/T)$ or ${\cal O}(1/T^2)$) convergence rates to a stationary point, and fast linear (i.e., ${\cal O}(\exp(-T))$) rates to the globally optimal solution under the Polyak-Łojasiewicz (\gls{PL}) condition. 

\item
The update is applied (see Step 5 in Algorithm~\ref{alg:RAC-LoRA}), and the method moves on to the next step of the chain.
\end{itemize}

\textbf{Experiments.} We apply our method to several Machine Learning tasks. We start from convex problems with traditional models, such as logistic and linear regression, to provide clear illustrations of our theoretical findings. 
    In addition, we present empirical analyses for \gls{MLP} on MNIST and RoBERTa on the GLUE benchmark tasks \citep{wang2018glue}. See Appendix~\ref{sec:exp}.

\textbf{Federated Learning.} Furthermore, we extend our findings from the simple unstructured problem (\ref{eq:main_Lora}) to the more challenging distributed/federated problem where $f$ has the special form described in (\ref{eq:fed}); there we consider solving a distributed optimization problem via our new \gls{Fed-RAC-LoRA} method (Algorithm~\ref{alg:Fed-RAC-LoRA}). These additional results can be found in Section~\ref{sec:FL}. For illustrative purposes, we provide an analysis for \algname{\gls{RR}} as the optimizer for the subproblem; see also Table~\ref{tab:theory}. Previous research \citep{sun2024improving} has shown that using a single learnable matrix in this context provides several key advantages, particularly in terms of preserving privacy, ensuring the correctness of model aggregation, and maintaining stability when adjusting the scaling factor. These benefits are crucial in Federated Learning \citep{FedLearn2016}, where data is distributed across multiple clients, and privacy constraints must be upheld while performing model updates. Building on this asymmetric approach, we integrate the concept of chained updates to develop \gls{Fed-RAC-LoRA}, a more robust and scalable distributed method. Our approach maintains the computational efficiency of the original \algname{\gls{RAC-LoRA}} while ensuring rigorous convergence properties in the distributed setting, offering a theoretically sound method for large-scale Federated Learning scenarios.

\section{Shining Some Light on LoRA's Convergence Issues}
\label{sec:lora_convergence_issue}

In contemporary Machine Learning, loss function minimization is primarily accomplished using gradient-based (first-order) optimization techniques \citep{ruder2016overview}. Most advanced methods build on the vanilla Gradient Descent (\algname{\gls{GD}}) in various ways, e.g., by adding support for stochastic approximation, momentum, adaptive stepsizes and more \citep{shapiro1996convergence,gower2019sgd}.

It is therefore meaningful to start our exploration of \algname{\gls{LoRA}}-style methods in connection with  \algname{\gls{GD}} steps. In particular, we analyze the update process of \algname{\gls{LoRA}} matrices through a \algname{\gls{GD}} step, focusing on the application of the chain rule of differentiation. The gradient with respect to the low-rank matrices \(B\) and \(A\) consists of two components,
\begin{align*}
    \nabla_{B,A} f(W+\frac{\alpha}{r}BA) = \begin{pmatrix}
    \nabla_A f(W+\frac{\alpha}{r}BA)\\
    \nabla_B f(W+\frac{\alpha}{r}BA)
\end{pmatrix} = \frac{\alpha}{r} \begin{pmatrix}
    B^\top \nabla  f(W+\frac{\alpha}{r}BA)\\
    \nabla f(W+\frac{\alpha}{r}BA) A^\top
\end{pmatrix}.
\end{align*}
and hence the update rules for the matrices \( A \) and \( B \) are given by
\[
A^+ = A - \eta \frac{\alpha}{r} B^\top \nabla f(W + \frac{\alpha}{r}BA), \qquad B^+ = B - \eta \frac{\alpha}{r} \nabla f(W + \frac{\alpha}{r}BA) A^\top,
\]

where \(\eta>0\) is a step size, and \(A^+\) and \(B^+\) are the updated matrices. Since both \(A\) and \(B\) are trainable, the gradients are multiplied by \( B^\top \) and \( A^\top \), which adds complexity to the optimization process and complicates the interpretation of its evolution. This interaction between low-rank matrices and gradients creates a non-trivial structure that challenges rigorous analysis and may disrupt Lipschitz continuity, raising concerns about convergence guarantees. While \algname{\gls{LoRA}} is effective for deep learning adaptation, a deeper understanding of this process is needed to ensure that the optimization scheme is theoretically sound.


\paragraph{Loss of Lipschitz smoothness.}  Lipschitz continuity of the gradient is a commonly invoked  assumption in the theoretical analysis of gradient-based optimization methods \citep{zhou2018fenchel, khaled2023better, demidovich2023guide}. This property ensures that the gradient does not change too rapidly, which in turn guarantees a controlled behavior of the optimization process, and plays a key role in establishing convergence rates and in providing stability guarantees for various optimization algorithms \citep{NesterovBook,sun2020optimization}. A formal definition follows.
\begin{assumption}[Lipschitz Gradient]
\label{asm:L-smooth}
Function $f$ is  differentiable, and there exists $L>0$ such that
\begin{align*}
\|\nabla f(W)-\nabla f(V)\| \leq L\|W-V\|, \qquad \forall  W, V \in \mathbb{R}^{m\times n},
\end{align*}
where $\|\cdot\|$ denotes the Frobenius matrix norm, and the gradient is computed w.r.t.\ the trace inner product.
\end{assumption}
However, the property of Lipschitz smoothness does not necessarily hold when applying \algname{\gls{LoRA}} adaptation. Specifically, even if the original function $ f(W) $ is Lipschitz smooth, meaning that the gradient of $ f(W) $ satisfies the Lipschitz continuity condition (as stated in Assumption \ref{asm:L-smooth}), this smoothness property is generally lost when the function is expressed in the adapted form $ f(W^0 + BA) $. In particular, the function $ f(W^0 + BA) $ is not Lipschitz smooth with respect to the set of variables $\{B, A\}$ for any constant. This breakdown of smoothness is a significant limitation, as it complicates the theoretical analysis of optimization algorithms when using \algname{\gls{LoRA}}. The formal proof of this result is provided in Theorem 2 of the work in \citep{sun2024improving}, highlighting the challenges in extending standard gradient-based methods to such adaptations.

\begin{figure}
\includegraphics[width=0.32\textwidth]{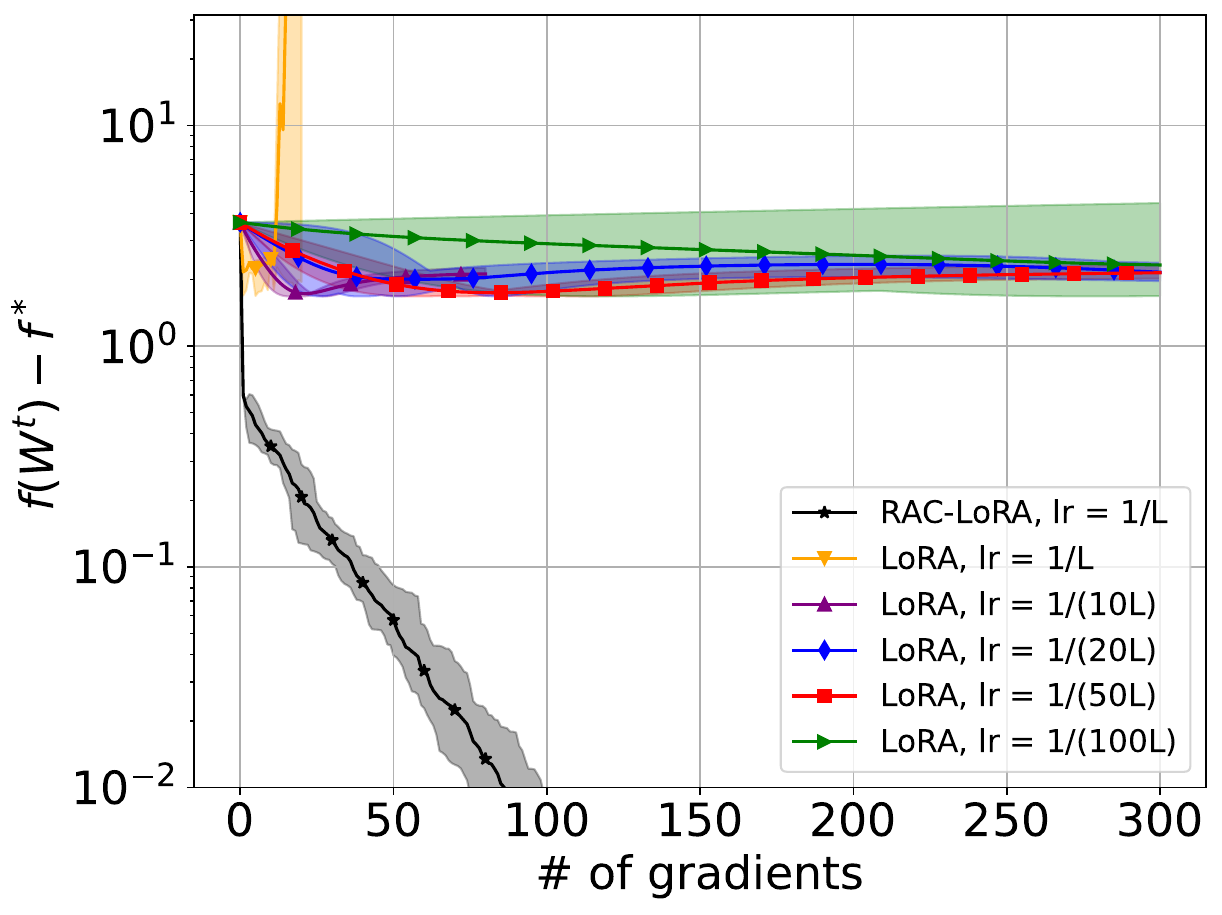}
\includegraphics[width=0.32\textwidth]{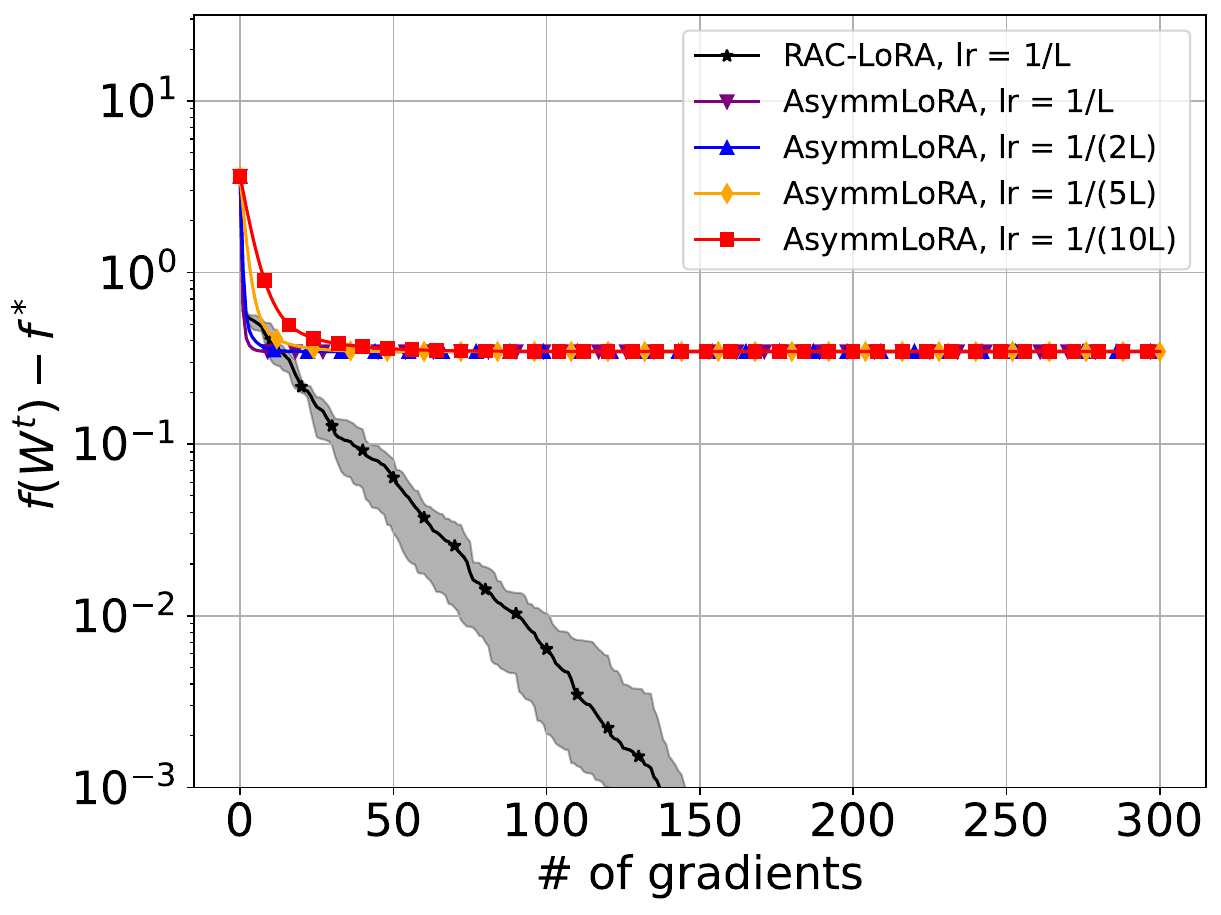}
\includegraphics[width=0.32\textwidth]{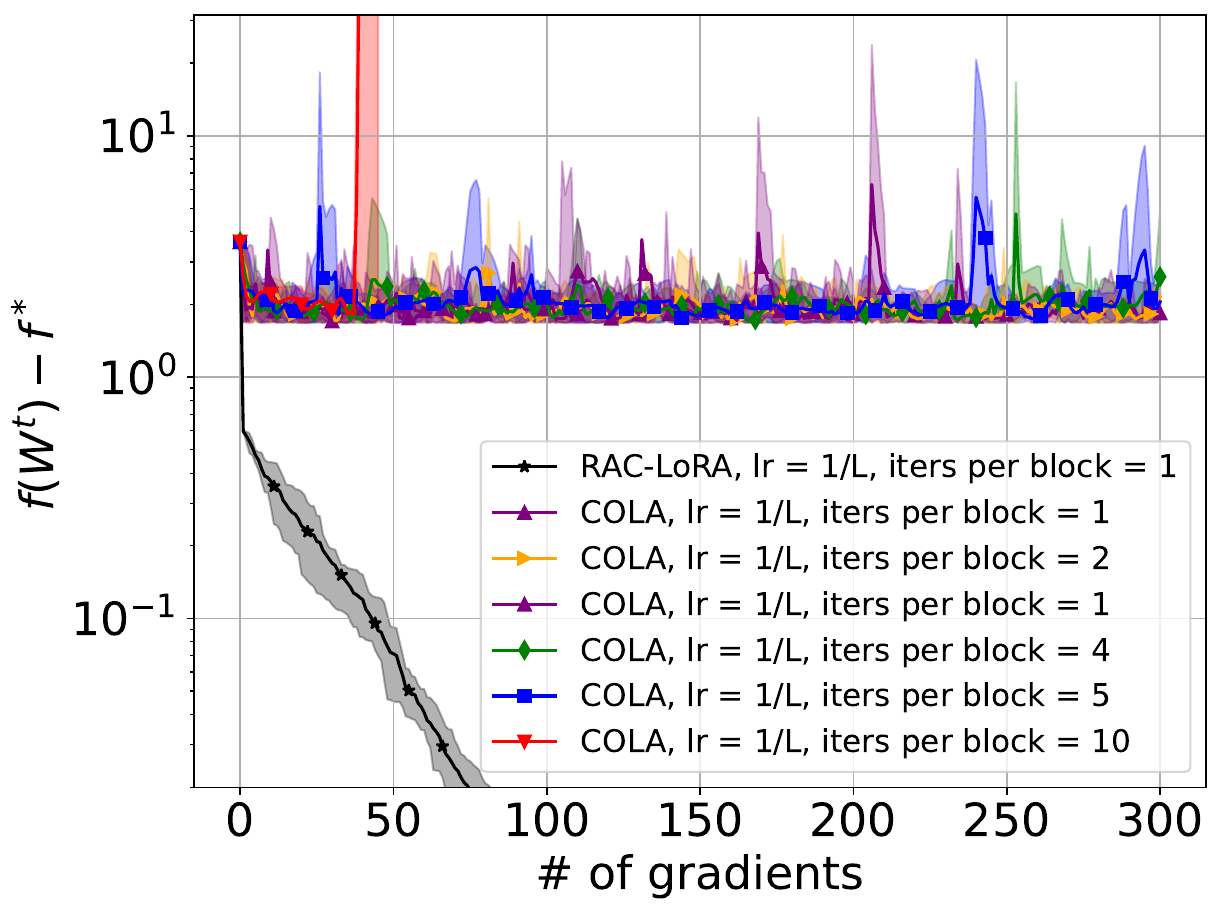}
\caption{Convergence of \algname{LoRA}, Asymmetric LoRA (\algname{AsymmLoRA}), Chain of LoRA (\algname{COLA}), and our proposed Randomized Asymmetric Chain of LoRA (\algname{RAC-LoRA}) on the problem in Equation (\ref{eq:counter}).}\label{fig:counterexample}
\end{figure}

\paragraph{Numerical counterexample.} We present a clear and illustrative example demonstrating that the \algname{\gls{LoRA}} and \algname{COLA} methods may not converge to the solution of the optimization problem. To illustrate this, let us consider a quadratic function of the following form:
\begin{align}
    f(x) = x^\top M x + b^\top x,
    \label{eq:counter}
\end{align}
where \( x \in \mathbb{R}^d \) is a vector of parameters, \( M \in \mathbb{R}^{d \times d} \) is a positive definite matrix, and \( b \in \mathbb{R}^d \) is a vector corresponding to the linear term. In our numerical example, we consider \(d = 9\), \(M = \operatorname{Diag}(10, 1, 1, 1, 1, 1, 1, 1, 1)\), and \(b = (1, 1, 1, 1, 1, 1, 1, 1, 1)^\top\). This function has a Lipschitz gradient (Assumption \ref{asm:L-smooth}) with \(L = 10\). We represent the vector \(x \in \mathbb{R}^9\) as a matrix \(W \in \mathbb{R}^{3 \times 3}\). In the \algname{\gls{LoRA}} adaptation, we use a rank \(r = 1\) and set \(\alpha = r\).

Figure~\ref{fig:counterexample} shows experiments on \algname{\gls{LoRA}}, \algname{AsymmLoRA}, \algname{COLA} and our new method \algname{\gls{RAC-LoRA}}. In the case of \algname{COLA}, we varied the step sizes and the number of gradients per block. Our results indicate that, when using the theoretical step size $\frac{1}{L}$, both \algname{\gls{LoRA}} and \algname{COLA} may diverge, while \algname{AsymmLoRA} converges to a different stationary point. When smaller step sizes are applied to \algname{\gls{LoRA}} and \algname{COLA}, these methods do converge, but to a stationary point that is significantly distant from the optimal solution. In contrast, our \algname{\gls{RAC-LoRA}} converges linearly to the optimal solution without such issues. These results provide clear evidence that the choice of \algname{\gls{LoRA}}-type updates has a significant impact on both the convergence and the quality of the final solution. The divergence, convergence to suboptimal points, and sensitivity to step sizes in traditional methods underscore the need for careful selection and design of update mechanisms. Our findings suggest that \algname{\gls{RAC-LoRA}} offers a more reliable approach for achieving optimal solutions in the context of \algname{\gls{LoRA}}-based adaptations.

\section{Randomized Asymmetric Chain of LoRA}

	\begin{algorithm}[t]
	\caption{Randomized Asymmetric Chain of LoRA (\algname{\gls{RAC-LoRA}})}\label{alg:RAC-LoRA}
	\begin{algorithmic}[1]
		\STATE	\textbf{Parameters:}  pre-trained model $W^0 \in \R^{m\times n}$, rank $r \ll \min\{m,n\}$, learning rate $\gamma > 0$, scaling factor $\alpha>0$, chain length $T$,  sketch distribution $\mathcal{D}^B_S$ (Option 1) or $\mathcal{D}^A_S$ (Option 2).
		\FOR{$t = 0, 1, \ldots , T-1$}\do\\
          \STATE Sample a sketch matrix $$ \text{(Option 1)} \quad B^t_S\sim \mathcal{D}^B_S \qquad  \text{(Option 2)} \quad  A^t_S \sim \mathcal{D}^A_S $$
		\STATE  Using some iterative solver, approximately solve the subproblem $$ \text{\small (Option 1) } \text{\small$\hat{A}^t \approx \min \limits_{A} f(W^t+ \frac{\alpha}{r} B^t_S A$} \text{ \quad \small (Option 2) } \text{\small$\hat{B}^t \approx \min \limits_{B} f(W^t+ \frac{\alpha}{r} B A^t_S)$}$$
		\STATE  Apply the update 
  $$ \text{(Option 1) } W^{t+1} = W^t + \frac{\alpha}{r} B^t_S \hat{A}^t \text{\quad (Option 2) } W^{t+1} = W^t + \frac{\alpha}{r} \hat{B}^t A^t_S$$
		\ENDFOR
	\end{algorithmic}
\end{algorithm}

To address the convergence issues in \algname{\gls{LoRA}} updates, we propose \algname{Randomized Asymmetric Chain of LoRA} (\algname{\gls{RAC-LoRA}}). This method introduces an asymmetric \algname{\gls{LoRA}} mechanism with a chain-based structure to enhance convergence while preserving model flexibility and efficiency. The method is summarized in Algorithm~\ref{alg:RAC-LoRA}.

\textbf{Description of the algorithm.} At the start of each iteration (or block), one matrix is randomly initialized and fixed throughout training, while the other remains fully trainable. This strategy prevents optimization within a restricted subspace, reducing the risk of convergence to suboptimal points. There are two configurations: freeze matrix \( B \) and train \( A \), or freeze \( A \) and train \( B \). We now formally define the sampling/sketch schemes.

\begin{definition}[Left Sketch]
\label{def:left}
By a ``left sketch'' (of rank $r$) we refer to the update rule 
$$ \Delta W = \frac{\alpha}{r} B_{S} \hat{A}, $$
where \( B_{S} \sim \mathcal{D}_B \) is sampled from some fixed distribution over matrices of dimensions \( m \times r \), and only the matrix \( \hat{A} \) is adjustable. 
\end{definition}

\begin{definition}[Right Sketch]
\label{def:right}
By a ``right sketch'' (of rank $r$) we refer to the update rule 
$$ \Delta W = \frac{\alpha}{r} \hat{B} A_{S}, $$
where \( A_{S} \sim \mathcal{D}_A \) is sampled from some fixed distribution over matrices of dimensions \( r \times n \), and only the matrix \( \hat{B} \) is adjustable. 
\end{definition}

In both sampling schemes, we update the trainable matrix over several epochs. This step effectively corresponds to training a \algname{\gls{LoRA}} block within the chain, following the standard \algname{\gls{LoRA}} approach. While this procedure mirrors the conventional \algname{\gls{LoRA}} method, we can formally characterize it as an approximate optimization problem, allowing for a structured analysis of the training process. These procedures for both matrices can be formally expressed via
\begin{align*}
     \text{(Option 1) }  \hat{A}^t \approx \min \limits_{A} f(W^t+ \frac{\alpha}{r} B^t_S A)\quad \text{(Option 2) }  \hat{B}^t \approx \min \limits_{B} f(W^t+ \frac{\alpha}{r} B A^t_S).
\end{align*}
Similarly to \algname{COLA}, $t$ identifies the block in the chain.
Next, we incorporate the product of the trained matrix and the sampled matrix into the current model. The merging process involves adding the product of the two matrices—one sampled and the other trained. This addition is scaled by a factor of \(\frac{\alpha}{r}\), ensuring the appropriate weighting of the update within the model:
\begin{align*}
     \text{(Option 1)} \quad W^{t+1} = W^t + \frac{\alpha}{r} B^t_S \hat{A}^t \qquad \text{(Option 2)} \quad W^{t+1} = W^t + \frac{\alpha}{r} \hat{B}^t A^t_S.
\end{align*}

\section{Theory}\label{sec:theory_Lora}



\subsection{Derivation of the update step}
\label{sec:der}
Without loss of generality, let us focus on the Left Sketch scheme (Definition~\ref{def:left}). Specifically, for each model in the chain, the update rule is given as follows:
\begin{align*}
    W^{t+1} = W^t + \frac{\alpha}{r} B^t_S \hat{A}^t. 
\end{align*}
Next, we apply the Lipschitz gradient condition (Assumption \ref{asm:L-smooth}) to the loss function $f$:
\begin{align*}
    f(U) \leq f(V)+\langle\nabla f(V), U-V \rangle+\frac{L}{2}\|U-V\|_F^2, \quad \forall U,V \in \mathbb{R}^{m\times n}
\end{align*}
Applying this with $U = W^{t+1}$, $V = W^t $ and $\gamma\leq \frac{1}{L}$ leads to
\begin{align*}
    f(W^{t+1}) &\leq f(W^t)+\left\langle\nabla f(W^t), \frac{\alpha}{r} B^t_S \hat{A}^t \right\rangle+\frac{L}{2}\left\| \frac{\alpha}{r}B^t_S \hat{A}^t\right\|_F^2\\
    &\leq f(W^t)+\frac{\alpha}{r}\left\langle (B^t_S)^\top \nabla f(W^t), \hat{A}^t \right\rangle+\frac{\alpha^2}{2\gamma r^2}\langle (B^t_S)^\top B^t_S \hat{A}^t , \hat{A}^t\rangle. 
\end{align*}
Note that $\left\|B_S^t \hat{A}^t\right\|_F^2=\left\langle B_S^t \hat{A}^t, B_S^t \hat{A}^t\right\rangle=\left\langle\left(B_S^t\right)^{\top} B_S^t \hat{A}^t, \hat{A}^t\right\rangle$. Let us minimize the right-hand side term in $\hat{A}^t$, when the gradient vanishes: $$
    \frac{\alpha}{r}(B^t_S)^\top \nabla f(W^t) + \frac{\alpha^2}{\gamma r^2} (B^t_S)^\top (B^t_S) \hat{A}^t = 0. 
$$
One such solution is given by \footnote{The dagger notation refers to the Moore-Penrose pseudoinverse.}
\begin{align*}
\hat{A}^t = -\gamma \frac{r}{\alpha}  \left((B^t_S)^\top (B^t_S) \right)^\dagger (B^t_S)^\top  \nabla f(W^t),
\end{align*}
and this leads to the following gradient update:
\begin{align}
\label{eq:left_GD}
 \notag   W^{t+1} &= W^t + \frac{\alpha}{r} B^t_S \hat{A}^t = W^t -\frac{\alpha}{r} \frac{r}{\alpha} \gamma  B^t_S \left((B^t_S)^\top (B^t_S) \right)^\dagger (B^t_S)^\top  \nabla f(W^t)\\
    & = W^t - \gamma H^t_B \nabla f(W^t),
\end{align}
where $H^t_B =  B^t_S \left((B^t_S)^\top (B^t_S) \right)^\dagger (B^t_S)^\top$ is a projection matrix. Similarly, we can obtain the update for the Right Sketch scheme (Definition~\ref{def:right}):
\begin{align}
\label{eq:right_GD}
    W^{t+1} &= W^t - \gamma \nabla f(W^t) (A^t_S)^\top \left(A_S^t(A^t_S)^\top\right)^\dagger A^t_S  = W^t - \gamma \nabla f(W^t) H^t_A, 
\end{align}
where $H^t_A = (A^t_S)^\top \left(A_S^t(A^t_S)^\top\right)^\dagger A^t_S$ is also a projection matrix. Notably, the scaling factor \(\frac{\alpha}{r}\) does not affect the gradient step. This simplifies the learning process by unifying the scaling and learning rate. Using this type of update, we provide convergence results for both standard and stochastic gradient descent methods.

\subsection{Convergence results} 
\label{sec:GD}
To derive the convergence results, a key factor in our analysis is the smallest eigenvalue of the expected value of the projection matrix. The projection matrix, which is generated by the sketch, is formally introduced in Section~\ref{sec:der}. This eigenvalue plays a critical role in shaping the optimization process. As we will show, a well-conditioned projection matrix—with a sufficiently large smallest eigenvalue—ensures more efficient and reliable convergence. Therefore, we make an important assumption that this smallest eigenvalue must remain strictly positive.

\begin{assumption}
\label{asm:lambda}
Consider a projection matrix $H$ generated by Left Sketch distribution (Def.~\ref{def:left}) or Right Sketch distribution (Def.~\ref{def:right}). Assume that the sampling distributions \(\mathcal{D}^B_S\) and \(\mathcal{D}^A_S\) are such that the smallest eigenvalue of the expected projection matrix $H$ generated by sampled matrix is positive:
\begin{align*}
    \lambda_{\min}^H = \lambda_{\min}\left[ \mathbb{E}\left[H\right] \right] >0.
\end{align*}
    
\end{assumption}

In particular, it is important to observe that the eigenvalues of the projection matrix are either zero or one, with the smallest eigenvalue being zero. However, the smallest eigenvalue of the expected value of the projection matrix can be strictly greater than zero. Additionally, it is essential to establish a lower bound for the loss function.
\begin{remark}
    Assumption \ref{asm:lambda} is easily satisfied.
    Let $H$ be the projection matrix as
    defined below~\Cref{eq:right_GD} and
    assume that the $A$ matrices are
    drawn from an isotropic distribution
    (the rows of $A$ are isotropic).
    Then $H$ is the projection onto the rank
    of $A$, which is a subspace of dimension
    $r$ distributed isotropically in $\mathbb{R}^n$.
    The matrix $\mathbb{E}[H]$ is then invariant
    under rotations, so it must be a scalar multiple of the identity. By taking
    traces, one finds that $\mathbb{E}[H] = \frac{r}{n}I$ so $\lambda_{\min}^H = \frac{r}{n}$.
\end{remark}
\begin{assumption}
    Function $f$ is bounded from below by an infimum $f^\star \in \mathbb{R}$. 
    \end{assumption}
We now present the convergence result for \algname{\gls{RAC-LoRA}} with Gradient Descent (\algname{GD}) updates.

\begin{theorem}
\label{thm:GD}
Let Assumptions \ref{asm:L-smooth} and \ref{asm:lambda} hold, and let the step size satisfy $0<\gamma\leq \frac{1}{L}$.  Then, the iterates of \algname{\gls{RAC-LoRA}}  (Algorithm \ref{alg:RAC-LoRA}) with \algname{\gls{GD}} updates (Equation (\ref{eq:left_GD}) or (\ref{eq:right_GD})) satisfy
                   \begin{align*}
 \mathbb{E}\left[\left\| \nabla f(\widetilde{W}^T)\right\|^2 \right] \leq \frac{2 (f(W^0) - f^\star)}{\lambda^H_{\min} \gamma T},
   \end{align*}
where the output $\widetilde{W}^T$ is chosen  uniformly at random from $W^0, W^1,\ldots,W^{T-1}$.   
\end{theorem}
We obtain a sub-linear convergence rate, as is expected in general non-convex settings. To achieve a stronger convergence result, we employ an additional assumption: the Polyak-Łojasiewicz (\gls{PL}) condition. This assumption generalizes strong convexity but applies to certain non-convex functions.

\begin{assumption}[PL-condition] 
    \label{asm:PL}
Function $f$ satisfies the Polyak-Łojasiewicz (\gls{PL}) condition with parameter $\mu > 0$  if
\begin{align*}
    \frac{1}{2}\|\nabla f(W)\|^2 \geq \mu\left(f(W)-f^{\star}\right)
\end{align*}
for all $W \in \mathbb{R}^{m\times n}$,
where $f^\star = \inf f$, assumed to be finite.
\end{assumption}
Next, we establish a convergence rate for \algname{\gls{RAC-LoRA}} in the Polyak-Łojasiewicz (\gls{PL}) setting.

\begin{theorem}
\label{thm:PL-GD}
Let Assumptions \ref{asm:L-smooth}, \ref{asm:lambda} and  \ref{asm:PL} hold, and let the step size satisfy  $0<\gamma\leq \frac{1}{L}$. Then, for each $T\geq 0$, the iterates of \algname{\gls{RAC-LoRA}} (Algorithm \ref{alg:RAC-LoRA}) with \algname{\gls{GD}} updates (Equation (\ref{eq:left_GD}) or (\ref{eq:right_GD}))  satisfy
     \begin{align*}
         \mathbb{E}\left[ f(W^{T}) \right] -f^\star  \leq  \left(1 - \gamma \mu \lambda^H_{\min}\right)^T\left( f(W^0) - f^\star \right).
     \end{align*}
\end{theorem}

We achieved a linear convergence rate, which is significantly better than previous results; however, this improvement applies to a more limited class of functions. Importantly, we can recover the classical results of \algname{\gls{GD}} by setting \(\lambda_{\min}^H = 1\), which corresponds to the full-rank scenario. 

The comprehensive analysis of different optimizers and their performance across various settings is provided in the appendix, as summarized in Table \ref{tab:theory}.

\section{Experiments} \label{sec:exp}

In this section, we explore the performance of 
\algname{\gls{RAC-LoRA}} as an optimization algorithm in Machine Learning applications.
In~\Cref{sec:exp-convex} we validate the theoretical results in convex problems,
while in~\Cref{sec:exp-nonconvex} we evaluate the method applied to neural networks.


\subsection{Convex optimization problems}\label{sec:exp-convex}
\begin{figure}
\centering
\includegraphics[width=0.4\textwidth]{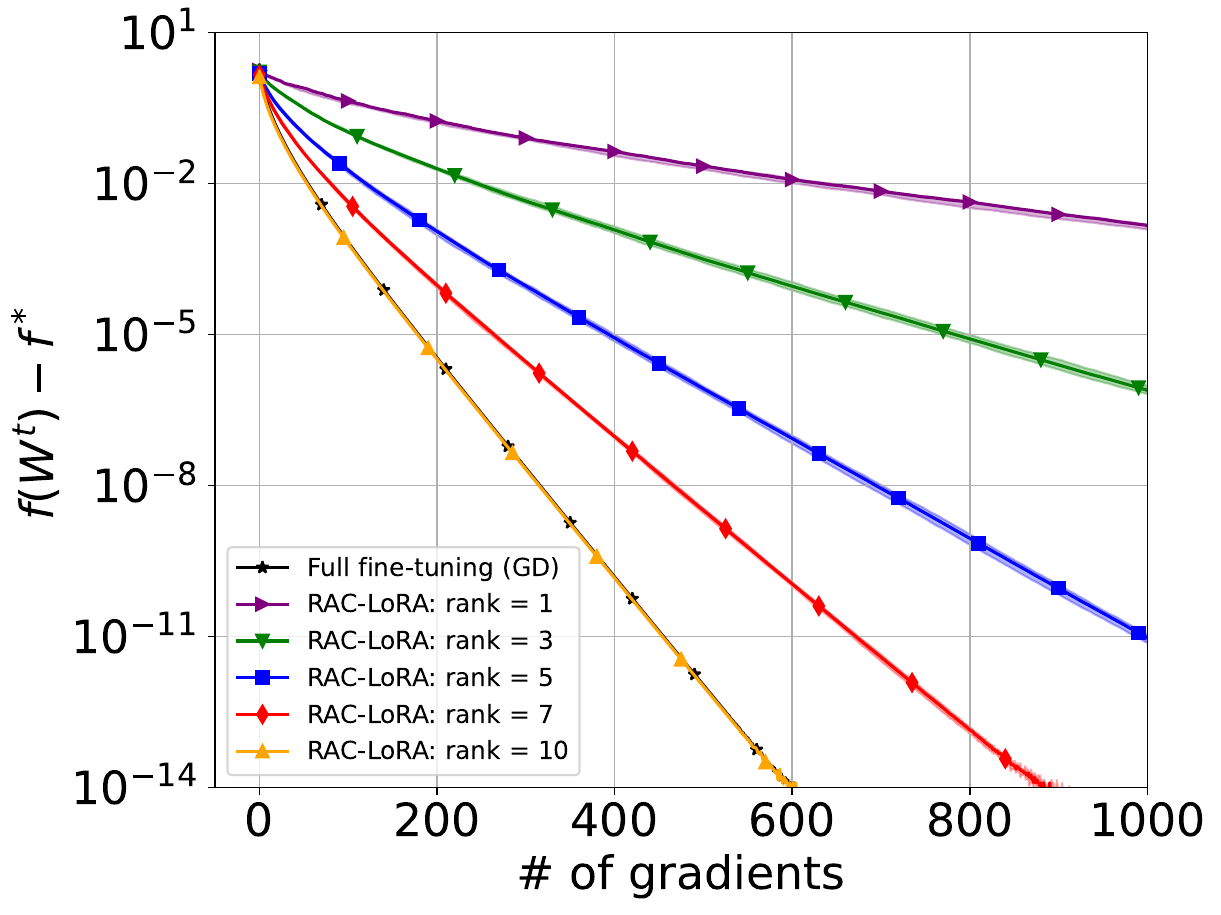}\qquad\qquad
\includegraphics[width=0.4\textwidth]{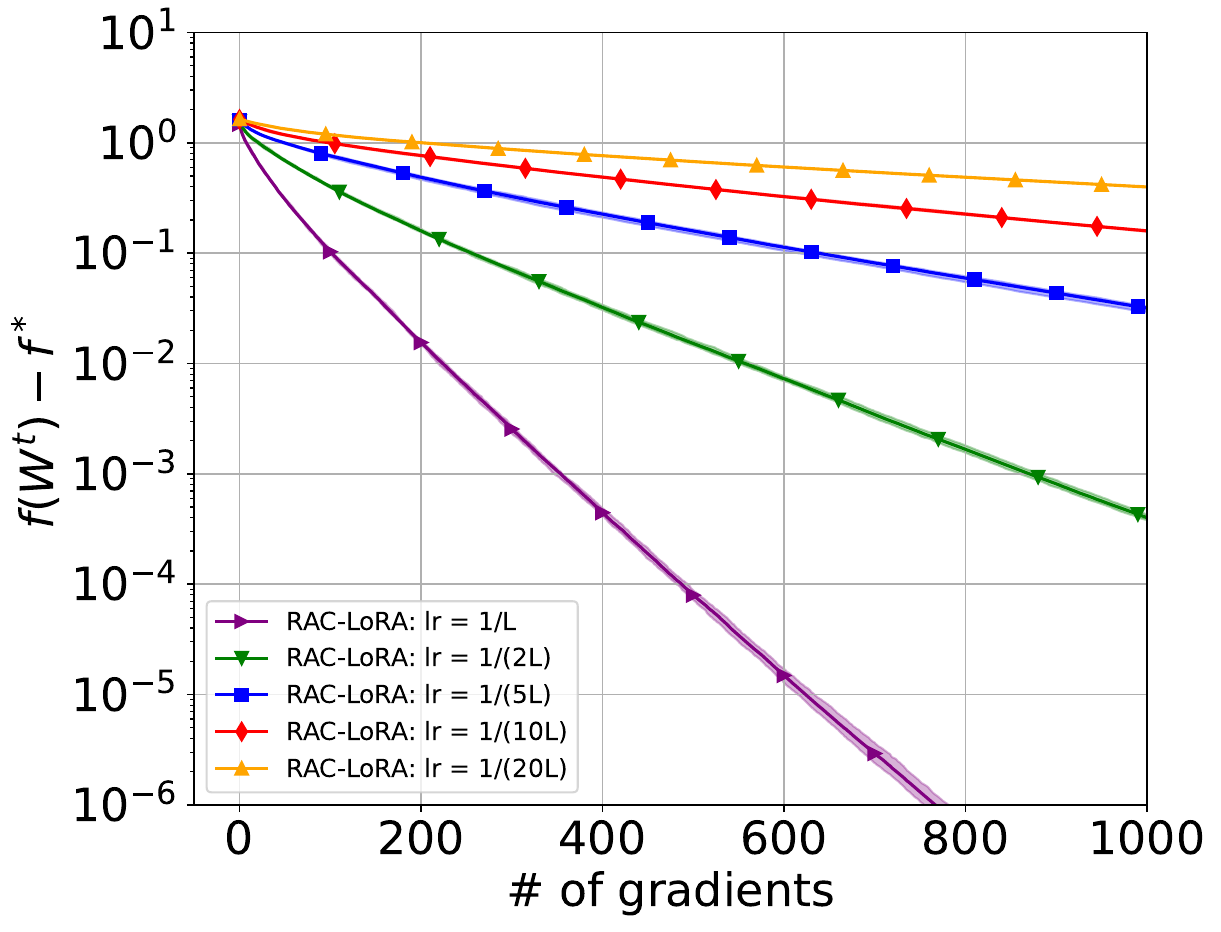}
\vspace{-0.25cm}
\caption{\algname{RAC-LoRA} convergence with varying ranks and step sizes on a linear regression problem.}\label{fig:linereg}
\end{figure}



\textbf{Linear Regression.} We conducted our analysis in a controlled setting using linear regression with quadratic regularization and synthetically generated data. Specifically, we utilized 3,000 samples for pre-training the model and 1,000 samples for fine-tuning. In this setup, we have \(d = 100\) with weight matrices of size \(10 \times 10\), and the regularization term is set to \(0.0001\). As illustrated in Figure~\ref{fig:linereg}, the method converges for various ranks and the convergence speed is proportional to $\frac{n}{r}$, and when the rank is set to the full rank, we observe convergence identical to that of \algname{\gls{FPFT}}. We remark that \algname{COLA} would suffer from the same divergence behavior as in \Cref{fig:counterexample} on this quadratic problem.

\textbf{Logistic Regression.} Analogous results for logistic regression are shown in Appendix~\ref{sec:app:exp-convex}.



\subsection{Non-Convex optimization problems}\label{sec:exp-nonconvex}

Further experimental results are provided in Appendix~\ref{sec:app:nonconvex}.

\subsubsection{Results of RoBERTa on NLP tasks}

As in prior work  \citep{zhu2024asymmetry,xia2024chain}, we evaluate 
low-rank adaptation methods for LLMs using the GLUE dataset \citep{wang2018glue}.

\textbf{Setup.}
We fine-tuned the \texttt{roberta-base} model \citep{liu2019roberta} on four of the smallest GLUE tasks to study the behavior of low-rank methods in practical scenarios.
For the chained methods, we use a range of values for the number of chains and
epochs per chain hyperparameters.
In each experiment
we used rank 2 for the adaptations and trained using the \algname{AdamW} optimizer~\citep{loshchilov2017decoupled} with $\beta$ parameters 0.9 and 0.999,
$\epsilon=1e-8$, a learning rate of 4e-4 with a linear schedule and a training batch size of 8.

\textbf{Discussion.}
The results are presented in~\Cref{tab:main:results_roberta_base_rank=2}.
We find that \algname{\gls{RAC-LoRA}} performs competitively with other low-rank adaptation methods, but
does not outperform \algname{Asymmetric LoRA} despite having greater capacity.
We expect \algname{\gls{RAC-LoRA}} to outperform \algname{Asymmetric LoRA} in settings where there
is a benefit to the additional capacity, i.e., those where a full parameter
fine tune (\algname{\gls{FPFT}}) is much better than \algname{Asymmetric LoRA}. 
The performance of the \algname{\gls{FPFT}}
in~\Cref{tab:main:results_roberta_base_rank=2} shows that
the selected GLUE tasks do not provide such a setting. Here,
a single low-rank adaptation is already enough to obtain
performance close to that of \algname{\gls{FPFT}}.
However, this intuition motivates the experiments in~\Cref{sec:exp-mnist}
where we intentionally restrict the capacity of the adaptations
to isolate the effect of the chaining procedure.

\begin{table*}[htb!]
\caption{Results with RoBERTa-base for rank 2 on tasks from the GLUE benchmark. *: results taken from the work of~\citep{hu2021lora}. We report Matthews correlation coefficient for \algname{COLA}, Pearson correlation coefficient for STS-B, and accuracy for the remaining tasks. Results are averaged over 3 seeds and standard deviations are given in the subscript.}
\resizebox{\textwidth}{!}{%
\centering
{

\begin{tabular}{lccccccc}
 \toprule
{{\bf Method}} & {{\bf \# Chains}} & {{\bf \# Epochs}} & \textcolor{black}{{\bf MRPC}} & \textcolor{black}{{\bf CoLA}} & \textcolor{black}{{\bf RTE}} & \textcolor{black}{{\bf STS-B}} &  {\bf Avg}\\
 \midrule

\algname{FPFT}* &  \multirow{2}{*}{1} & \multirow{2}{*}{30, 80, 80, 40} & 
90.2\textsubscript{$\pm$0.0} &
63.6\textsubscript{$\pm$0.0} &
78.7\textsubscript{$\pm$0.0} & 
91.2\textsubscript{$\pm$0.0} & 
80.9\\

\algname{LoRA}* & & & 
89.7\textsubscript{$\pm$0.7} &
63.4\textsubscript{$\pm$1.2} &
86.6\textsubscript{$\pm$0.7} & 
91.5\textsubscript{$\pm$0.2} & 
82.8\\

\midrule

\algname{LoRA} & \multirow{2}{*}{1} & \multirow{2}{*}{100}  
&
87.7\textsubscript{$\pm$0.2} &
{60.8\textsubscript{$\pm$0.2}} &
{75.2\textsubscript{$\pm$1.5}} & 
90.2\textsubscript{$\pm$0.1} & 
78.5\\
 
\algname{AsymmLoRA} & & 
&
86.9\textsubscript{$\pm$0.3} &
58.7\textsubscript{$\pm$1.0} &
71.0\textsubscript{$\pm$3.3} & 
90.4\textsubscript{$\pm$0.0} & 
76.8\\

\algname{COLA} & 10 & 10
& 
{88.0\textsubscript{$\pm$0.8}} &
59.5\textsubscript{$\pm$1.0} &
72.1\textsubscript{$\pm$0.9}&
{90.7\textsubscript{$\pm$0.2}} & 
77.6\\

\algname{\gls{RAC-LoRA}} & 10 & 10
& 
87.0\textsubscript{$\pm$0.7} &
58.5\textsubscript{$\pm$0.1} &
72.3\textsubscript{$\pm$1.5}&
90.3\textsubscript{$\pm$0.0} & 
77.0\\

\bottomrule
\end{tabular}
}
}

\label{tab:main:results_roberta_base_rank=2}
\vspace{-5.0pt}
\end{table*}

\begin{table}{Rt}{8.5cm}
\setlength{\tabcolsep}{3pt}
\centering
\caption{\gls{MLP} results on MNIST with rank $r$ and $\alpha$ set to 1. In the case of \algname{AsymmLoRA} and \algname{\gls{RAC-LoRA}}, only the zero-initialized matrix is trained.}
\begin{tabular}{lcccccc}
\toprule
\textbf{Method} & $\mathcal{D}_A$ & $\mathcal{D}_B$ & \textbf{Acc} & \textbf{Train Params}\\\midrule
\algname{FPFT} & - & - & 98.0 & 54,700  \\\midrule
\algname{LoRA} & Gaussian & Zero & 83.8 & 1K  \\
\algname{COLA} & Gaussian & Zero & 92.6 & 1K \\ \midrule

\algname{LoRA} & Zero & Gaussian & 87.0 & 1K \\ 
\algname{COLA}  & Zero & Gaussian & 96.2 & 1K \\
\midrule

\algname{AsymmLoRA}  & Gaussian & Zero & 62.3 & 133 \\ 
\algname{\gls{RAC-LoRA}} & Gaussian & Zero & 92.0 & 133\\ 
\midrule
\algname{AsymmLoRA} & Zero & Gaussian & 81.6 & 912 \\
\algname{\gls{RAC-LoRA}} & Zero & Gaussian & 96.1 & 912\\
\bottomrule
\end{tabular}
\label{tab:mnist_rank1}
\end{table}

\subsubsection{Results of \gls{MLP}s on MNIST}\label{sec:exp-mnist}

In this section, we seek to isolate the effect of the chaining procedure on generalization performance by restricting the capacity of the low-rank adaptations. 
This ensures that a single adaptation is not sufficient to reach performance
comparable with \algname{\gls{FPFT}}, allowing us to explore how chaining adaptations can bridge this gap.

\textbf{Setup.}
We first pre-train a 3-layer \gls{MLP} on the first five classes (digits 0-4) and then adapt the network using \algname{\gls{LoRA}}-based methods for recognizing the remaining five unseen classes (digits 5-9). The model is evaluated solely on these unseen classes\footnote{The setup is inspired by \url{https://github.com/sunildkumar/lora_from_scratch/}.}. We used rank 1 for the adaptations and trained using the \algname{AdamW} optimizer~\citep{loshchilov2017decoupled} with $\beta$ parameters 0.9 and 0.999,
$\epsilon=1e-8$, a constant learning rate of 2e-4 and a training batch size of 128.

\textbf{Discussion.}
Table~\ref{tab:mnist_rank1} shows results for MNIST with different ranks and initialization. \algname{\gls{LoRA}} reaches around 90\% of the accuracy of \algname{\gls{FPFT}}, leaving some margin for improvement when using the chains. \algname{COLA} constructs a sequence of \algname{\gls{LoRA}} modules, delivering significant accuracy
improvements over \algname{\gls{LoRA}} due to the chaining procedure.
The chaining allows \algname{COLA} to capture richer features (at the cost of training more parameters). 
However, both \algname{LoRA} and \algname{COLA} lack rigorous convergence guarantees. \algname{AsymmLoRA} has been shown empirically to approximate the performance of \algname{\gls{LoRA}} \citep{sun2024improving} --- but again no convergence result is provided. Our proposed method (\algname{\gls{RAC-LoRA}}) enjoys significant accuracy improvements over \algname{AsymmLoRA}, again due to the chaining procedure. \algname{\gls{RAC-LoRA}} leverages a diverse learning process across different \algname{\gls{LoRA}} blocks, which intuitively allows the model to capture a broader range of features. Crucially, \algname{\gls{RAC-LoRA}} comes with convergence guarantees (Theorems~\ref{thm:GD} and  \ref{thm:PL-GD}).
Finally, we note that each iteration of \algname{\gls{RAC-LoRA}} requires training only one matrix per \algname{\gls{LoRA}} block, while \algname{COLA} needs to train two matrices. This reduction in trainable parameters may offer advantages in resource-constrained settings, such as Federated Learning, where minimizing communication costs is critical (see also Appendix~\ref{sec:FL}).


\chapter{Concluding Remarks}
\thispagestyle{empty}
In this concluding chapter, we summarize the obtained results and outline potential future directions and open problems.
\section{Summary}

In this thesis, we have addressed several fundamental issues arising when applying optimization to large-scale machine learning problems. Our particular interest was in stochastic first order methods that scale efficiently in distributed networks and the federated learning community, too, where communication constraints, data heterogeneity, and adversarial presence pose significant hurdles.

In Chapter 2, we resolved a major open question regarding local training by introducing the \gls{ProxSkip} algorithm. We formally established that local gradient steps provably lead to communication acceleration, reducing the expected communication complexity to $\mathcal{O}(\sqrt{\kappa} \log 1/\varepsilon)$ without explicitly applying Nesterov-style acceleration. This provided rigorous theoretical justification for local training without requiring heterogeneity bounding assumptions.

To improve the algorithms proposed in Chapter 2, in Chapter 3 we developed Variance Reduced ProxSkip (\gls{ProxSkip-VR}). By incorporating variance reduction mechanisms, we successfully neutralized stochastic noise, achieving linear convergence even in the stochastic regime and overcoming the lack of a linear speedup. Furthermore, we expanded the standard federated topology by introducing a hierarchical structure with regional hubs.

In Chapter 4, we designed the \gls{5GCS} algorithm, the first fifth generation local training method that explicitly achieves accelerated communication complexity while supporting partial client participation. We demonstrated that taking only a relatively small number of local gradient steps is entirely sufficient to preserve the accelerated communication complexity of exact proximity evaluation, theoretically matching established lower bounds.

In Chapter 5, we addressed the theoretical gap concerning the combined use of server-side stepsizes and sampling without replacement. We proved that scaling aggregated updates via a server-side stepsize yields improved convergence complexities, effectively reducing the nonconvex convergence complexity from $\mathcal{O}(1/\varepsilon^3)$ to $\mathcal{O}(1/\varepsilon^2)$, and establishing a clearer theoretical understanding of these common optimization heuristics.

In Chapter 6, we proposed techniques to successfully integrate gradient compression with Random Reshuffling and local computation. After demonstrating that naive compression injects noise that overwhelms the benefits of sampling without replacement, we developed the \gls{DIANA-RR} and \gls{DIANA-NASTYA} algorithms. By compressing gradient differences and utilizing multiple shift vectors, we successfully eliminated the added compression variance, providing a provably efficient method for federated learning.

In Chapter 7, we developed the first distributed method that simultaneously accommodates partial participation and guarantees provable tolerance to Byzantine workers. By applying adaptive clipping directly to stochastic gradient differences within a recursive variance reduction framework, we strictly bounded the harm of malicious participants, matching state-of-the-art theoretical results under general assumptions.

Finally, in Chapter 8, we introduced a novel theoretical framework for Low-Rank Adaptation using randomized asymmetric chains. We constructed a precise counterexample to illustrate the fundamental instability of existing approaches, and then proved that our combined asymmetric chain perfectly stabilizes the optimization trajectory. This provided the first provable convergence guarantees establishing that parameter-efficient methods can reliably reach the exact same optimal solution as full-parameter fine-tuning.

\section{Future Directions}

As we discovered answers to some of the important challenges or their aspects, we also faced new challenges and can see new gaps between theory and practice. Below, we briefly provide a few directions which we personally consider to be important and challenging.

\subsection{Optimization for foundation models and adaptive methods}
While stochastic gradient descent is extensively studied, the theoretical foundations for the adaptive methods required to train large language models and vision language models remain limited. Methods such as \algname{Adam}, \algname{AdamW}, and newer orthogonalization-based optimizers like Muon demonstrate exceptional empirical performance, yet they often lack rigorous convergence or generalization guarantees in these complex settings. We believe that scientific progress must go beyond empirical observation. Advancing the theory of adaptive algorithms, alongside memory-efficient optimization strategies, is essential for explaining practical phenomena and guiding the principled design of reliable machine learning systems. By establishing exact convergence rates and understanding the structural properties that allow these optimizers to successfully navigate complicated loss landscapes, we can develop new algorithms that drastically reduce the computational resources required for model training. Furthermore, extending these theoretical insights to accommodate sparse memory states will directly address the hardware limitations currently restricting the scale of artificial intelligence research.

\subsection{Federated fine-tuning and parameter efficiency}

Distributed pretraining of massive models is frequently constrained by prohibitive computational costs and severe memory limits. Conversely, federated fine-tuning presents a highly realistic paradigm where our theoretical advances in federated optimization can have an immediate, practical impact. A critical open direction is the seamless integration of parameter-efficient techniques, specifically low-rank adaptation, directly into diverse federated environments. Developing explicit theoretical insights for these combinations will guide the efficient and scalable adaptation of foundation models across decentralized networks. Specifically, we must investigate how the structural properties of randomized asymmetric chains behave under extreme data heterogeneity and partial participation. Resolving the optimization trajectories for these parameter-efficient methods across varying local data distributions will unlock the ability to collaboratively train highly specialized models without exchanging sensitive raw data or requiring massive server infrastructure.

\subsection{Realistic Byzantine robustness}
In this thesis, we made significant progress in securing distributed systems under partial client participation. However, standard theoretical settings for Byzantine robustness frequently assume omniscient adversaries possessing full knowledge of all client data and system updates. This results in overly pessimistic guarantees that fail to reflect the reality of deployed systems and often force algorithms to adopt extremely conservative learning rates. A major future challenge is to define and explore more realistic adversarial models, such as attackers with delayed information or restricted collusion capabilities. Developing optimization theory under these practical assumptions will yield robust algorithms that offer meaningful protection without imposing severe theoretical or computational penalties. Applying these refined robust techniques to multi-agent systems will be highly valuable for ensuring artificial intelligence safety, providing a mathematical guarantee that collaborative agents remain aligned even when a subset acts unpredictably or submits corrupted information.

\subsection{Differential privacy in machine learning systems}

While differential privacy provides a mathematically rigorous framework for data protection, integrating it into complex multi-agent systems presents unique challenges. For example, standard private stochastic gradient descent often suffers from bias introduced by necessary clipping mechanisms. While error compensation techniques can mitigate this bias, they substantially increase memory usage, which is a critical constraint in large-scale applications. Future research must deeply investigate the complex relationship between utility, privacy, and memory limits to design algorithms that are simultaneously secure and resource-efficient. Furthermore, extending these private and communication-efficient techniques to support machine unlearning mechanisms will be crucial for ensuring safety and alignment in collaborative artificial intelligence frameworks. By formalizing the unlearning problem within a distributed optimization context, we can leverage existing tools from meta-learning and federated learning to allow multi-agent systems to efficiently forget restricted data without requiring a complete retraining of the global model.


\begin{onehalfspacing}
	\renewcommand*\bibname{\centerline{REFERENCES}} 
    \phantomsection
	\addcontentsline{toc}{chapter}{References}
	\newcommand{\BIBdecl}{\setlength{\itemsep}{0pt}}
		\bibliographystyle{plainnat}
		\bibliography{Cleaned_References_new}
\end{onehalfspacing}



\appendix
		\newpage
		\begingroup
			\let\clearpage\relax
			\begin{center}
			\vspace*{2\baselineskip}
			{ \textbf{{\large APPENDICES}}} 
            \phantomsection
			\addcontentsline{toc}{chapter}{Appendices} 
			\end{center}
            

\refstepcounter{chapter}%
\chapter*{\thechapter \quad Appendix B for Chapter 2}
\label{appendixB}

\section{Basic Facts}
\label{appendix:facts}

The Bregman divergence of a differentiable function $f\colon\mathbb{R}^d \to \mathbb{R}$ is defined by
\[D_f(x,y) \eqdef f(x) - f(y) - \< \nabla f(y), x-y >. \]
It is easy to see that
\begin{equation} \label{eq:sym_Bregman} \< \nabla f(x) - \nabla f(y), x-y > = D_f(x,y) + D_f(y,x), \quad \forall x,y\in \mathbb{R}^d\end{equation}

For an $L$-smooth and  $\mu$-strongly convex function $f\colon\mathbb{R}^d\to \mathbb{R}$, we have
\begin{equation}\label{eq:bi87fgddf-1}\frac{\mu}{2} \|x-y\|^2 \leq D_f(x,y) \leq \frac{L}{2} \|x-y\|^2, \quad \forall x,y\in \mathbb{R}^d\end{equation}
and
\begin{equation}\label{eq:bi87fgddf-2}\frac{1}{2L} \|\nabla f(x)-\nabla f(y)\|^2 \leq D_f(x,y) \leq \frac{1}{2\mu} \|\nabla f(x)-\nabla f(y)\|^2, \quad \forall x,y\in \mathbb{R}^d.\end{equation}
Given $\psi\colon \mathbb{R}^d\to \mathbb{R}$, we define $\psi^*(y) \eqdef \sup_{x\in\mathbb{R}^d}\{\<x, y> - \psi(x)\}$ to be its Fenchel conjugate. The proximity operator of $\psi^*$ satisfies for any $\tau>0$
\begin{equation}
	\mathrm{if}\quad u=\prox_{\tau \psi^*}(y), \quad\mathrm{then}\quad u \in y - \tau\partial \psi^*(u). \label{eq:prox_implicit}
\end{equation}

\section{Analysis of ProxSkip~(\texorpdfstring{\Cref{alg:ProxSkip}}{Algorithm ProxSkip})}

\subsection{Proof of Lemma~\ref{lem:A}}

\begin{proof}
In order to simplify notation, let $P(\cdot)\eqdef\prox_{\frac{\gamma}{p}\psi}(\cdot)$, and  \begin{equation}\label{eq:x_and_y}x\eqdef\hat x^{t+1} - \frac{\gamma}{p}h^{t}, \qquad y\eqdef x^{\star} - \frac{\gamma}{p}h^{\star}.\end{equation}

{\bf STEP 1 (Optimality conditions).} Using the first-order optimality conditions for $f+\psi$ and using $h^{\star}\eqdef \nabla f(x^{\star})$, we obtain the following fixed-point identity for $x^{\star}$:
	\begin{equation}\label{eq:n08fhd90fd}
		x^{\star}
		= \prox_{\frac{\gamma}{p} \psi}\left(x^{\star} - \frac{\gamma}{p}h^{\star} \right) \overset{\eqref{eq:x_and_y}}{=} P(y).
	\end{equation}
	
{\bf STEP 2 (Recalling the steps of the method).}
Recall that the vectors $x^t$ and $h^{t}$ are in \Cref{alg:ProxSkip} updated as follows:
\begin{equation} \label{eq:step2a} x^{t+1} = \begin{cases} 
P\bigl(x \bigr) & \text{with probability} \quad p \\
 \hat x^{t+1} & \text{with probability} \quad 1- p 
\end{cases},\end{equation}
and
\begin{equation}\label{eq:step2b}  h^{t+1} = h^{t} + \frac{p}{\gamma}(x^{t+1} - \hat x^{t+1}) =\begin{cases}  h^{t} + \frac{p}{\gamma}(P(x)- \hat x^{t+1}) & \text{with prob.} \quad p\\ h^{t} &  \text{with prob.} \quad 1- p 
\end{cases}.
\end{equation}

{\bf STEP 3 (One-step expectation of the Lyapunov function).}

The expected value of the Lyapunov function 
\begin{equation}\label{eq:Lyapunov-proof} \Psi^{t} \eqdef \|x^{t} - x^{\star}\|^2 + \frac{\gamma^2}{p^2}\|h^{t} - h^{\star}\|^2\end{equation}
at time $t+1$, with respect to the coin toss at iteration $t$, is
{\footnotesize
\begin{eqnarray*}\E{\Psi^{t+1}} &\overset{\eqref{eq:step2a}+\eqref{eq:step2b}+\eqref{eq:Lyapunov-proof}}{=} & p \left(\|P(x) - x^{\star}\|^2 + \frac{\gamma^2}{p^2}\left\|h^{t} + \frac{p}{\gamma}(P(x)- \hat x^{t+1})- h^{\star} \right\|^2\right)\\
&+& (1-p) \left(\| \hat x^{t+1}  - x^{\star}\|^2 + \frac{\gamma^2}{p^2}\|h^{t} - h^{\star}\|^2\right)\\
&\overset{\eqref{eq:n08fhd90fd}}{=}& p \left(\|P(x) - P(y)\|^2 + \left\| {\color{red}\frac{\gamma}{p}h^{t} + P(x)- \hat x^{t+1}} - {\color{blue} \frac{\gamma}{p}h^{\star} }\right\|^2\right)\\
&+& (1-p) \left(\| \hat x^{t+1}  - x^{\star}\|^2 + \frac{\gamma^2}{p^2}\|h^{t} - h^{\star}\|^2\right)\\
&\overset{\eqref{eq:x_and_y}+\eqref{eq:n08fhd90fd}}{=} & p \left( \left\| P(x) - P(y) \right\|^2 + \underbrace{\left\| {\color{red}P(x) - x}   + {\color{blue} y-P(y)} \right\|^2}_{\|Q(x)-Q(y)\|^2} \right)\\  &+& (1-p) \left(\| \hat x^{t+1}  - x^{\star}\|^2 + \frac{\gamma^2}{p^2}\|h^{t} - h^{\star}\|^2\right).
\end{eqnarray*}
}

{\bf STEP 4 (Applying firm nonexpansiveness).}
Applying firm nonexpansiveness of $P$ (Lemma~\ref{lem:prox-contraction}), this leads to
 the inequality
\begin{eqnarray*}\E{\Psi^{t+1}} &\overset{\eqref{eq:prox_firm_non_exp}}{\le}  & p  \left\|x-y \right\|^2  + (1-p) \left(\| \hat x^{t+1}  - x^{\star}\|^2 + \frac{\gamma^2}{p^2}\|h^{t} - h^{\star}\|^2\right)\\
&\overset{\eqref{eq:x_and_y}}{=} & p \left\|\hat x^{t+1} - \frac{\gamma}{p}h^{t} - \left(x^{\star} - \frac{\gamma}{p}h^{\star} \right) \right\|^2\\
&+& (1-p) \left(\| \hat x^{t+1}  - x^{\star}\|^2 + \frac{\gamma^2}{p^2}\|h^{t} - h^{\star}\|^2\right).
\end{eqnarray*}

	
{\bf STEP 5 (Simple algebra).}	
Next, we expand the squared norm and collect the terms, obtaining
	\begin{eqnarray}
	\notag	\mathbb{E}\left[\Psi^{t+1} \right] &\leq & p\|\hat x^{t+1} - x^{\star}\|^2 + p\frac{\gamma^2}{p^2}\|h^{t}-h^{\star}\|^2 - 2\gamma\<\hat x^{t+1}-x^{\star}, h^{t}-h^{\star}>\\
        &+& (1-p)\Bigl( \|\hat x^{t+1} - x^{\star}\|^2 + \frac{\gamma^2}{p^2}\|h^{t} - h^{\star}\|^2\Bigr)  \notag \\
		&=& \|\hat x^{t+1} - x^{\star}\|^2 - 2\gamma\<\hat x^{t+1}-x^{\star}, h^{t}-h^{\star}> + \frac{\gamma^2}{p^2}\|h^{t}-h^{\star}\|^2. \label{eq:FINAL-A}
	\end{eqnarray}
Finally, note that by our definition of $w^t$, we have the identity $\hat x^{t+1} = w^t + \gamma h^{t}$. Therefore, the first two terms above can be rewritten as
	\begin{eqnarray}
		\|\hat x^{t+1} - x^{\star}\|^2 &-& 2\gamma\<\hat x^{t+1}-x^{\star}, h^{t}-h^{\star}>\\&=& \|w^t - w^{\star} + \gamma(h^{t}-h^{\star})\|^2 - 2\gamma\<w^t-w^{\star} + \gamma (h^{t} - h^{\star}), h^{t}-h^{\star}> \notag \\
		& =&  \|w^t - w^{\star}\|^2 + 2\gamma\<w^t-w^{\star}, h^{t}-h^{\star}> + \gamma^2 \|h^{t}-h^{\star}\|^2 \notag \\
		&&\qquad  - 2\gamma\<w^t-w^{\star}, h^{t}-h^{\star}> - 2\gamma^2 \|h^{t} - h^{\star}\|^2  \notag  \\
		& =& \|w^t - w^{\star}\|^2 - \gamma^2 \|h^{t}-h^{\star}\|^2. \label{eq:FINAL-B}
	\end{eqnarray}
	It remains to plug \eqref{eq:FINAL-B} into \eqref{eq:FINAL-A}.
\end{proof}

\subsection{Proof of Lemma~\ref{lem:B}}

\begin{proof}
	Recall the definition of $w^t$ and $w^{\star}$ in~\eqref{eq:98g9gbjfd8d}.
	Plugging these expressions into $\|w^t-w^{\star}\|^2$, expanding the square, and applying properties of $f$ as a $\mu$-strongly convex and $L$-smooth function, we get 
	\begin{eqnarray*}
		\|w^t - w^{\star}\|^2 &\overset{\eqref{eq:98g9gbjfd8d}}{=} &
		\|x^t - x^{\star} - \gamma (\nabla f(x^t)   - \nabla f(x^{\star}))\|^2 \\
		&=	&\|x^t - x^{\star}\|^2 + \gamma^2\|\nabla f(x^t) - \nabla f(x^{\star})\|^2\\
        &-&2\gamma\<\nabla f(x^t) - \nabla f(x^{\star}), x^t - x^{\star}>\\
		&\overset{\eqref{eq:bi87fgddf-1}}{\le} & (1-\gamma\mu)\|x^t - x^{\star}\|^2\\
        &-& 2\gamma  D_f(x^t,x^{\star}) +\gamma^2\|\nabla f(x^t) - \nabla f(x^{\star})\|^2 \\
		& = & (1-\gamma\mu)\|x^t - x^{\star}\|^2\\
        &-& 2\gamma \left(D_f(x^t,x^{\star}) -  \frac{\gamma}{2} \|\nabla f(x^t) - \nabla f(x^{\star})\|^2 \right) \\
		& \overset{\eqref{eq:bi87fgddf-2}}{\le} &(1-\gamma\mu)\|x^t - x^{\star}\|^2,
	\end{eqnarray*}
	\normalsize
	where the last inequality holds if $0\leq \gamma \leq \frac{1}{L}$.
\end{proof}

\clearpage
\section{Analysis of SProxSkip\texorpdfstring{~(\Cref{alg:stoch_rand_prox})}{ (Algorithm)}}

\subsection{The algorithm}

We consider a variant of \gls{ProxSkip} which uses a {\color{blue}{\em stochastic} gradient $g^t(x^t)$} instead of $\nabla f(x^t)$; see Algorithm~\ref{alg:stoch_rand_prox}.


\subsection{Two lemmas}

Lemma~\ref{lem:A-stochastic} is an extension of Lemma~\ref{lem:A} to the stochastic case. In this result, we work with  \begin{equation}\label{eq:tilde{w}}\tilde{w}^t= x^t - \gamma {\color{blue} g^t(x^t)} \end{equation} instead of $w^t = x^t - \gamma \nabla f(x^t)$.

\begin{lemma}\label{lem:A-stochastic}
If Assumptions \ref{as:f} and \ref{as:proper_psi} hold, $\gamma >0$ and $0<p\leq 1$, then
\begin{equation} \label{eq:b8f9d89fd8df_09_} \E{ \Psi^{t+1} } \leq  \|\tilde{w}^t- w^{\star}\|^2 +  (1-p^2)\frac{\gamma^2}{p^2}\|h^{t} - h^{\star}\|^2 \,, \end{equation}
where the expectation is taken over the $\theta_t$ in \Cref{alg:stoch_rand_prox}.
\end{lemma}
\begin{proof} The proof is identical to the proof of Lemma~\ref{lem:A}. 
\end{proof}

Likewise, Lemma~\ref{lem:B-stochastic} is an extension of  Lemma~\ref{lem:B} to the stochastic case.

\begin{lemma} \label{lem:B-stochastic} Let \Cref{as:f} hold with any $\mu\geq 0$. If $0< \gamma \leq \frac{1}{A}$, then \begin{equation}\label{eq:nbo98fd8f_09uf00}\E{\|\tilde{w}^t- w^{\star}\|^2} \le (1-\gamma\mu)\|x^t - x^{\star}\|^2 + \gamma^2C,\end{equation}
where the expectation is w.r.t.\ the randomness in the stochastic gradient ${\color{blue} g^t(\cdot)}$.
\end{lemma}

\begin{proof}
	Recall the definition of $\tilde{w}^t$ in \eqref{eq:tilde{w}} and $w^{\star}$ in~\eqref{eq:98g9gbjfd8d}.
	Plugging these expressions into $\|\tilde{w}^t-w^{\star}\|^2$ and expanding the square, we get 
	\begin{eqnarray}
		\|\tilde{w}^t- w^{\star}\|^2 &\overset{\eqref{eq:98g9gbjfd8d}+\eqref{eq:tilde{w}}}{=} &
		\|x^t - x^{\star} - \gamma (g^t(x^t)   - \nabla f(x^{\star}))\|^2 \notag \\
		&=	&\|x^t - x^{\star}\|^2 + \gamma^2\|{\color{blue} g^t(x^t)} - \nabla f(x^{\star})\|^2\\
        &-&2\gamma\< {\color{blue} g^t(x^t)} - \nabla f(x^{\star}), x^t - x^{\star}>. \label{eq: bjhbidys_09}
	\end{eqnarray}
Taking expectation w.r.t.\ the randomness of the stochastic gradient $g^t(x^t)$, and using unbiasedness (\Cref{as:exp_smooth})  	and expected smoothness (\Cref{as:unbias}), we get	
	\begin{eqnarray*}
	\E{ \|\tilde{w}^t- w^{\star}\|^2} & \overset{\eqref{eq: bjhbidys_09}}{=} & \|x^t - x^{\star}\|^2 + \gamma^2 \E{\|{\color{blue} g^t(x^t)} - \nabla f(x^{\star})\|^2}\\
    &-&2\gamma\< \E{{\color{blue} g^t(x^t)}} - \nabla f(x^{\star}), x^t - x^{\star}> \\		
	&=&  \|x^t - x^{\star}\|^2 -2\gamma\< \nabla f(x^t)- \nabla f(x^{\star}), x^t - x^{\star}>\\
    &+& \gamma^2 \E{\|{\color{blue} g^t(x^t)} - \nabla f(x^{\star})\|^2}  .	
	\end{eqnarray*}	

The second term can be decomposed using the identity:  
$$\left\langle \nabla f(x^t) - \nabla f(x^{\star}) , x^t - x^{\star}\right\rangle = D_f(x^t,x^{\star})+D_f(x^{\star},x^t)$$ (see \eqref{eq:sym_Bregman}), and the third term can be bounded via $\E{\|{\color{blue} g^t(x^t)} - \nabla f(x^{\star})\|^2} \leq 2AD_f(x^t,x^{\star})+C$ (see expected smoothness; \Cref{as:exp_smooth}), which leads to
\begin{eqnarray}
    \label{eq:noihfd_9u0fd9}
	\E{\|\tilde{w}^t- w^{\star}\|^2} &\leq&  \|x^t - x^{\star}\|^2 -  2\gamma ( D_f(x^t,x^{\star})+D_f(x^{\star},x^t))\\ 
    &+& \gamma^2\left( 2AD_f(x^t,x^{\star})+C \right).
\end{eqnarray}

By plugging the inequality $\mu\|x^t - x^{\star}\|^2\leq 2D_f(x^{\star},x^t)$ (see \eqref{eq:bi87fgddf-1}) into \eqref{eq:noihfd_9u0fd9}, we get
\begin{eqnarray*}
	\E{ \|\tilde{w}^t- w^{\star}\|^2} &\leq & (1-\gamma\mu)\|x^t - x^{\star}\|^2  - 2\gamma  D_f(x^t,x^{\star})\\
    &+& \gamma^2\left( 2AD_f(x^t,x^{\star})+C \right)\\
	&\leq & (1-\gamma\mu)\|x^t - x^{\star}\|^2  - 2\gamma (1-\gamma A)  D_f(x^t,x^{\star})+ \gamma^2C.
\end{eqnarray*}
Finally, the stepsize restriction $\gamma\leq \frac{1}{A}$ allows us to produce the estimate
\begin{equation}\label{eq:090fd9hf}
	\E{\|\tilde{w}^t- w^{\star}\|^2}  \leq   (1-\gamma\mu) \|x^t - x^{\star}\|^2 + \gamma^2C,
\end{equation}
which is what we wanted to show.
\end{proof}

		

\subsection{Proof of Theorem~\ref{thm:main-stoch}}

\begin{proof}

Combining Lemma~\ref{lem:A-stochastic} and Lemma~\ref{lem:B-stochastic}, we get
\begin{eqnarray*}  \E{ \Psi^{t+1} } & \overset{\eqref{eq:b8f9d89fd8df_09}+\eqref{eq:090fd9hf}}{\leq} &  (1-\gamma\mu) \|x^t - x^{\star}\|^2  +  (1-p^2)\frac{\gamma^2}{p^2}\|h^{t} - h^{\star}\|^2 + \gamma^2C \\
&\leq & \max \{ 1-\gamma\mu, 1-p^2\} \Psi^{t} +\gamma^2 C\\
&=& (1-\zeta)\Psi^{t} +\gamma^2 C\, ,
 \end{eqnarray*}
 where $\zeta\eqdef \min\{\gamma\mu, p^2\}$. Taking full expectation, we get
$ \E{ \Psi^{t+1} } \leq  (1-\zeta) \E{\Psi^{t}} +\gamma^2 C $,
 and unrolling the recurrence, we finally obtain
 \begin{equation}\label{eq:yu8tgufd090f}\E{ \Psi^{T} } \leq (1 - \zeta)^T \Psi^0 + \frac{\gamma^2 C}{\zeta} \,.\end{equation}

\end{proof}

\subsection{Proof of Corollary~\ref{cor:0099887766}}
Recall that Theorem~\ref{thm:main-stoch} requires the stepsize $\gamma$ to satisfy \begin{equation} \label{eq:xx_one}0<\gamma \le \frac{1}{A}.\end{equation}
Pick $0<\varepsilon<1$. We will now choose $\gamma$ and $T$ such that $\E{ \Psi^{T} } \leq \varepsilon $. We shall do so by bounding both terms on the right-hand side of \eqref{eq:yu8tgufd090f}  by $\frac{\varepsilon}{2}$. 
\begin{itemize}
\item In order to minimize the number of prox evaluations, whatever the choice of $\gamma$ will be, we choose the smallest probability $p$ which does not lead to any degradation of the rate $\zeta\eqdef \min\{\gamma\mu, p^2\}$. That is, we choose \begin{equation}\label{eq:xx_p}p=\sqrt{\gamma \mu},\end{equation} in which case $\zeta = \gamma \mu$.
\item The first term on the right-hand side of \eqref{eq:yu8tgufd090f} can be bounded as follows:
\begin{equation}\label{eq:xx^three}T \geq \frac{1}{\gamma \mu} \log \left(\frac{2 \Psi^0}{\varepsilon}\right)  \quad \Longrightarrow \quad  (1 - \zeta)^T \Psi^0 \leq \frac{\varepsilon}{2}  .\end{equation}
\item The second term on the right-hand side of \eqref{eq:yu8tgufd090f} can be bounded as follows:
\begin{equation}\label{eq:xx^two}\gamma \leq \frac{ \varepsilon \mu  }{2 C} \quad \Longrightarrow \quad \frac{\gamma^2 C }{\zeta} = \frac{\gamma C }{\mu} \leq \frac{\varepsilon}{2} .\end{equation}

\end{itemize}

Since the number of iterations \eqref{eq:xx^three} depends inversely on the stepsize $\gamma$, we choose the largest stepsize consistent with the bounds \eqref{eq:xx_one} and \eqref{eq:xx^two}: \begin{equation}\label{eq:xx_four}\gamma = \min\left\{  \frac{1}{A}, \frac{ \varepsilon \mu}{2 C} \right\}.\end{equation}

By plugging this into \eqref{eq:xx^three}, we get the iteration complexity bound 
\[T \geq \max \left\{ \frac{A}{\mu}, \frac{2C}{\varepsilon \mu^2}\right\}\log \left(\frac{2 \Psi^0}{\varepsilon}\right) \quad \Longrightarrow \quad \E{ \Psi^{T} } \leq \varepsilon\,.\]

Since in each iteration we evaluate the prox with probability 
$p$ given by \eqref{eq:xx_p}, the expected number of prox evaluations is given by
\begin{eqnarray*}
    p T &\overset{\eqref{eq:xx^three}}{\geq}& p \frac{1}{\gamma \mu} \log \left(\frac{2 \Psi^0}{\varepsilon}\right)  \overset{\eqref{eq:xx_p}}{=} \sqrt{\frac{1}{\gamma \mu}} \log \left(\frac{2 \Psi^0}{\varepsilon}\right) \\
    &\overset{\eqref{eq:xx_four}}{=}&\max\left\{ \sqrt{\frac{A}{\mu}}, \sqrt{\frac{2C}{\varepsilon \mu^2}}\right\} \log \left(\frac{2 \Psi^0}{\varepsilon}\right). 
    \end{eqnarray*}

\clearpage

\section{Decentralized Analysis}

Let us now analyze the convergence of \Cref{alg:dist_gd} by introducing an algorithm for the problem $\min_x f(x)+\psi(\mL x)$ and studying its properties.

\begin{algorithm*}[t]
	\caption{\gls{SplitSkip}}
	\label{alg:split_skip}
	\begin{algorithmic}[1]
		\STATE stepsizes $\gamma > 0$ and $\tau>0$, matrix $\mL\in\mathbb{R}^{m\times d}$,  probability $p>0$, initial iterate $x^0\in \mathbb{R}^d$, initial control variate ${\red y_0}=0 \in \mathbb{R}^m$, number of iterations $T\geq 1$
		\FOR{$t=0,1,\dotsc,T-1$}
		\STATE $\hat x^{t+1} = x^t - \gamma (\nabla f (x^t) +\mL^\top {\red y^{t}})$ \hfill $\diamond$ Take a gradient-type step adjusted via the control variate ${\red y^{t}}$
		\STATE Flip a coin $\theta_t \in \{0,1\}$ where $\mathop{\rm Prob}(\theta_t =1) = p$ \hfill $\diamond$ Flip a coin that decides whether to skip the prox or not
		\IF{$\theta_t=1$}
		\STATE  ${\red y^{t+1}} = \prox_{\tau\psi^*}\bigl( {\red y^{t}} +\tau \mL\hat{x}^{t+1} \bigr)$ \hfill $\diamond$ Apply prox, but only very rarely! (with small probability $p$)
		\STATE $x^{t+1} = \hat{x}^{t+1} - \frac{\gamma}{p}\mL^\top ({\red y^{t+1}} - {\red y^{t}})$
		\ELSE
		\STATE $x^{t+1} = \hat x^{t+1}$
		\STATE ${\red y^{t+1}} = {\red y^{t}}$ \hfill $\diamond$ Skip the prox!
		\ENDIF
		\ENDFOR
	\end{algorithmic}
\end{algorithm*}

 Notice that \Cref{alg:dist_gd} is a special case of \Cref{alg:split_skip} with $\mL=(\mI-\mW)^{1/2}$ and $\psi$ being the indicator function of 0. Indeed, the conjugate of $\psi$ is simply 0 everywhere, so $\prox_{\tau\psi^*}(y)=y$ for any $y$. Thus, if we define $h^{t} \eqdef -\mL^\top y^{t}$, we obtain \Cref{alg:dist_gd}, as shown in more detail in Section~\ref{sec:formal_decentral}. For this reason, we will do the analysis for the more general \Cref{alg:split_skip}. 

 Let us also add a few connections of this method to the existing literature on primal--dual algorithms. When $p=1$, \Cref{alg:split_skip} reverts to a primal--dual algorithm first proposed by Loris and Verhoven for least squares problems~\citep{lor11}, and rediscovered later under the names \algname{PDFP2O}~\citep{chen2013primal} and \algname{PAPC}~\citep{DRORI2015209}. The convergence of this algorithm has been analyzed in~\citep{combettes2014forward,condat2019proximal,condat2022distributed} and generalized to the case of stochastic gradients in~\citep{PDDY2020}, who also studied its linear convergence under similar assumptions, albeit without skipping the prox step.
 
 Our analysis is based on the following Lyapunov function:
\begin{align} 
 \Phi^t \eqdef \|x^{t} - x^{\star}\|^2 + \frac{\gamma}{p\tau}\|y^{t} - y_{\star}\|^2 , \label{eq:Lyapunov_decent}
\end{align}
where $y^{t}$ is the dual variable from \Cref{alg:split_skip}. 
\begin{theorem}\label{th:split_skip}
	Let \Cref{as:f} and \Cref{as:proper_psi} hold, and assume that for any $y$, we have $\partial\psi^*(y)\subseteq \range(\mL)$. If we choose $p\in(0, 1]$, $\gamma\le \frac{1}{L}$, $\gamma\tau \le \frac{p}{\|\mL\mL^\top\|}$, then
		\[
		\mathbb{E}\left[\|x^T-x^\star\|^2\right]
		\le (1 - \zeta)^T\Phi^0,
	\]
	where $\zeta=\min\{\gamma\mu, p\gamma\tau\lambda_{\min}^+(\mL\mL^\top)\}$.
\end{theorem}
\begin{proof}
	Let us define $\hat y^{t+1} \eqdef \prox_{\tau \psi^*}(y^{t} + \tau \mL\hat x^{t+1})$. As stated in equation~\eqref{eq:prox_implicit}, this definition implies the following implicit representation of $\hat y^{t+1}$:
	\[
		\hat y^{t+1}=y^{t} + \tau \mL\hat x^{t+1} - \tau (\psi^*) '(y^{t+1}),
	\]
	where $(\psi^*) '(y^{t+1})\in \partial \psi^*(y^{t+1})$ is a subgradient of $\psi^*$ at $y^{t+1}$.
	With the help of $\hat y^{t+1}$, we can expand the expected distance to the solution,
	\begin{align*}
		&\mathbb{E}\left[\|x^{t+1} - x^\star\|^2\right]\\
		&= p \Bigl\|\hat x^{t+1} - x^\star - \frac{\gamma}{p}\mL^\top (\hat y^{t+1} - y^{t})\Bigr\|^2 + (1-p)\|\hat x^{t+1} - x^\star\|^2 \\
		&= p \left[\|\hat x^{t+1} - x^\star\|^2 - 2\frac{\gamma}{p}\<\hat x^{t+1} - x^\star, \mL^\top (\hat y^{t+1} - y^{t})> + \frac{\gamma^2}{p^2}\|\mL^\top (\hat y^{t+1} - y^{t})\|^2\right]\\
        &+ (1-p)\|\hat x^{t+1} - x^\star\|^2 \\
		&= \|\hat x^{t+1} - x^\star\|^2 - 2\gamma \<\hat x^{t+1} - x^\star, \mL^\top (\hat y^{t+1} - y^{t})> + \frac{\gamma^2}{p}\|\mL^\top (\hat y^{t+1} - y^{t})\|^2.
	\end{align*}
	Next, let us rewrite the first term as $\|w^t- w^\star\|$ by using the expansion $\|a+b\|^2 = \|a\|^2 + 2\<a+b, b> - \|b\|^2$,
	\begin{align*}
		\|\hat x^{t+1} - x^\star\|^2
		&= \|w^t - w^\star - \gamma \mL^\top (y^{t} - y^\star)\|^2 \\
		&= \|w^t - w^\star\|^2 - 2\gamma \<\hat x^{t+1} - x^\star, \mL^\top (y^{t} - y^\star)> - \gamma^2\|\mL^\top (y^{t} - y^\star)\|^2.
	\end{align*}
	Thus,
	\begin{eqnarray*}
		\mathbb{E}\left[\|x^{t+1} - x^\star\|^2\right]
		&=& \|w^t - w^\star\|^2 - 2\gamma \<\hat x^{t+1} - x^\star, \mL^\top (y^{t} - y^\star)>\\
        &-& 2\gamma \<\hat x^{t+1} - x^\star, \mL^\top (\hat y^{t+1} - y^{t})> \\
		&  -& \gamma^2\|\mL^\top (y^{t} - y^\star)\|^2 + \frac{\gamma^2}{p}\|\mL^\top (\hat y^{t+1} - y^{t})\|^2 \\
		&=&\|w^t - w^\star\|^2 - 2\gamma \<\hat x^{t+1} - x^\star, \mL^\top (\hat y^{t+1} - y^\star)>\\
        &-& \gamma^2\|\mL^\top (y^{t} - y^\star)\|^2 + \frac{\gamma^2}{p}\|\mL^\top (\hat y^{t+1} - y^{t})\|^2.
	\end{eqnarray*}
	Now, we can turn our attention to the convergence of the dual variable $y^{t}$. Since $y^{t+1}$ is updated with probability $p$, it holds
	\begin{align*}
		\mathbb{E}\left[\|y^{t+1} - y_{\star}\|^2\right]
		&= p \|y^{t} - y^\star + (\hat y^{t+1} - y^{t})\|^2 + (1-p)\|y^{t}-y^\star\|^2 \\
		&= p \|y^{t} - y^\star\|^2 + 2p\<\hat y^{t+1}-y^\star,\hat y^{t+1} - y^{t}>\\
        &- p\|\hat y^{t+1} - y^{t}\|^2 + (1-p)\|y^{t}-y^\star\|^2 \\
		&= \|y^{t} - y^\star\|^2 + 2p\<\hat y^{t+1}-y^\star,\hat y^{t+1} - y^{t}>  - p\|\hat y^{t+1} - y^{t}\|^2 \\
		&= \|y^{t} - y^\star\|^2\\
        &+ 2p\tau\<\hat y^{t+1}-y^\star,\mL \hat x^{t+1} - (\psi^*) '(\hat y^{t+1})>  - p\|\hat y^{t+1} - y^{t}\|^2.
	\end{align*}
	By the first-order optimality conditions, we get $\mL x^\star = (\psi^*)'(x^\star)$, where $(\psi^*)'(x^\star)$ is some subgradient of $\psi^*$ at $x^\star$. Therefore, convexity of $\psi^*$ gives
	\begin{align*}
		&\<\hat y^{t+1}-y^\star, \mL \hat x^{t+1} - (\psi^*) '(\hat y^{t+1})>\\
		&= \<\hat y^{t+1}-y^\star,\mL (\hat x^{t+1} - x^\star) - (\psi^*) '(\hat y^{t+1}) + (\psi^*)'(y^\star)> \\
		&\le \<\hat y^{t+1}-y^\star,\mL (\hat x^{t+1} - x^\star)>.
	\end{align*}
	Combining the recursions for the iterates $x^{t+1}$ and $y^{t+1}$, we obtain
	\begin{align*}
		\mathbb{E}\left[\Phi^{t+1}\right]
		&= \mathbb{E}\left[\|x^{t+1}-x^\star\|^2 + \frac{\gamma}{p\tau}\|y^{t+1} - y^\star\|^2\right]\\
		&\le \|w^t - w^\star\|^2 - 2\gamma \<\hat x^{t+1} - x^\star, \mL^\top (\hat y^{t+1} - y^\star)> - \gamma^2\|\mL^\top (y^{t} - y^\star)\|^2\\
        &+ \frac{\gamma^2}{p}\|\mL^\top (\hat y^{t+1} - y^{t})\|^2\\
		&\quad + \frac{\gamma}{p\tau}\|y^{t} - y^\star\|^2 + 2\gamma\<\hat y^{t+1}-y^\star,\mL (\hat x^{t+1} - x^\star)>  - \frac{\gamma}{\tau}\|\hat y^{t+1} - y^{t}\|^2 \\
		&= \|w^t - w^\star\|^2 +  \frac{\gamma}{p\tau}\|y^{t} - y^\star\|^2  - \gamma^2\|\mL^\top (y^{t} - y^\star)\|^2\\
        &+ \frac{\gamma^2}{p}\|\mL^\top (\hat y^{t+1} - y^{t})\|^2  - \frac{\gamma}{\tau}\|\hat y^{t+1} - y^{t}\|^2.
	\end{align*}
	Using the assumption that $\tau \le \frac{p}{\gamma\|\mL \mL^\top\|}$, we get
	\[
		\frac{\gamma^2}{p}\|\mL^\top (\hat y^{t+1} - y^{t})\|^2
		\le \frac{\gamma^2}{p}\|\mL\mL^\top\| \|\hat y^{t+1} - y^{t}\|^2
		\le \frac{\gamma}{\tau}\|\hat y^{t+1} - y^{t}\|^2.
	\]
	Plugging this back, we derive
	\begin{align*}
		\mathbb{E}\left[\Phi^{t+1}\right]
		&\le \|w^t - w^\star\|^2 +  \frac{\gamma}{p\tau}\|y^{t} - y^\star\|^2 - \gamma^2\|\mL^\top (y^{t} - y^\star)\|^2\\
        &
		\overset{\eqref{eq:nbo98fd8f_09uf}}{\le} (1-\gamma\mu)\|x^t - x^{\star}\|^2 +  \frac{\gamma}{p\tau}\|y^{t} - y^\star\|^2 - \gamma^2\|\mL^\top (y^{t} - y^\star)\|^2.	    
	\end{align*}
	Since we assume that $\partial \psi^*(y)\subseteq \range(\mL)$ and $y_0=0\in\mathbb{R}^d$, we have that $y^{t}-y^\star\in \range(\mL)$. Therefore, $\|\mL^\top (y^{t}-y^\star)\|^2\ge \lambda_{\min}^+(\mL\mL^\top)\|y^{t}- y^\star\|^2$, where $\lambda_{\min}^+$ is the smallest positive eigenvalue. Combining these results, we get
	\begin{align*}
		\mathbb{E}\left[\|x^t-x^\star\|^2\right]
		&\le \mathbb{E}\left[\Phi^t\right]
		\le (1-\gamma\mu)\|x^{t-1} - x^\star\|^2\\
        &+  \frac{\gamma}{p\tau}\left(1 - p\gamma\tau\lambda_{\min}^+(\mL\mL^\top)\right)\|y^{t-1} - y^\star\|^2 \\
		&\le (1 - \min(\gamma\mu, p\gamma\tau\lambda_{\min}^+(\mL\mL^\top))^t\Phi^0.  
	\end{align*}
\end{proof}
\subsection{Proof of \texorpdfstring{\Cref{th:decentralized}}{Theorem}}\label{sec:formal_decentral}
\begin{proof}
	To obtain the communication step as a special case of the proximity operator, we set $\psi$ to be the indicator function of the singleton $\{0\}\subseteq \mathbb{R}^d$,
	\[
		\psi(x)= \begin{cases} 0 & x = 0 \\ +\infty & x\neq 0\end{cases}.
	\]
	Its Fenchel conjugate equals, by definition, $\psi^*(y)=\sup_{x\in\mathbb{R}^d}\{\<x, y> - \psi(x)\} = \<0, y>=0$. Therefore, for any $y$, $\prox_{\tau \psi^*}(y)=y$ and $\partial \psi^*(y)=\{0\}\subseteq \range(\mL)$, and the conditions of \Cref{th:split_skip} hold. Next, let us establish that \Cref{alg:dist_gd} is a special case of \Cref{alg:split_skip}. If we consider the iterates of \Cref{alg:split_skip} and define $h^{t} \eqdef -\mL^\top y^{t}$, then its first step can be rewritten as
	\[
		\hat x^{t+1} 
		= x^t - \gamma (\nabla f(x^t) + \mL^\top y^{t})
		= x^t - \gamma (\nabla f(x^t) - h^{t}),
	\]
	which is exactly the first step of \Cref{alg:dist_gd}. The second step of \Cref{alg:split_skip} is either to do nothing or to update $y^{t+1}$. Using the fact that $\prox_{\tau \psi^*}(y^{t} + \tau \mL \hat x^{t+1})= y^{t} + \tau \mL \hat x^{t+1}$, it is easy to see that
	\[
		h^{t+1}
		\eqdef -\mL^\top y^{t+1}
		= -\mL^\top(y^{t} + \tau \mL \hat x^{t+1})
		= h^{t} - \tau \mL^\top \mL \hat x^{t+1}.
	\]
	By setting $\mL=(\mI-\mW)^{1/2}$, we get $\mL^\top \mL = \mI - \mW$, and we recover the second step of \Cref{alg:dist_gd} in an equivalent form:
	\begin{align*}
	&\begin{cases}
		h_{m}^{t+1} &= h_{m}^{t} + \tau\bigl( \hat h_{m}^{t+1} - \sum_{m^\prime=1}^M W_{m,m^\prime}\hat x_{m^\prime}^{t+1}\bigr), \\
		h_{m}^{t+1} &= \hat h_{m}^{t+1} + \frac{\gamma}{p}(h_{m}^{t+1} - h_{m}^{t}).
	\end{cases}
\\
	&\begin{cases}
		x_{m}^{t+1} &= \left(1 - \frac{\gamma\tau}{p}\right)\hat x_{m}^{t+1} + \frac{\gamma\tau}{p}\sum_{m^\prime=1}^M W_{m,m^\prime}\hat x_{m^\prime}^{t+1}, \\
		h_{m}^{t+1} &= h_{m}^{t} + \frac{p}{\gamma}(x_{m}^{t+1} -  \hat h_{m}^{t+1}).
	\end{cases}	    
	\end{align*}
	Finally, note $ \lambda_{\min}^+(\mL^\top \mL) = \lambda_{\min}^+(\mI - \mW) = 1-\lambda_2(\mW) = \delta$, $\|\mL^\top \mL\| = \|\mI - \mW\| < 1$, and $y^0_{m}=0$, so applying \Cref{th:split_skip} yields
	\begin{align*}
		\mathbb{E}\left[\frac{1}{M}\sum_{m=1}^M \|x_{m}^{T}-x^\star\|^2 \right]
		&\le (1 - \zeta)^T\Phi^0\\
		&= (1 - \zeta)^T\left(\|x^0 - x^\star\|^2 + \frac{\gamma}{p\tau}\frac{1}{M}\sum_{m=1}^M\|y_{m}^{\star}\|^2 \right).
	\end{align*}
	By Jensen's inequality, the average iterate $\overline x^T \eqdef \frac{1}{M}\sum_{m=1}^M x_{m}^{T}$ satisfies
	\begin{align*}
		\mathbb{E}\left[\|\overline x^T - x^\star\|^2 \right]
		&\le \mathbb{E}\left[\frac{1}{M}\sum_{m=1}^M \|x_{m}^{T}-x^\star\|^2 \right]\\
		&\le (1 - \zeta)^T \left(\|x^0 - x^\star\|^2 + \frac{\gamma}{p\tau}\frac{1}{M}\sum_{m=1}^M\|y_{m}^{\star}\|^2 \right). 
	\end{align*}
	Finally, notice that $$\|y_{m}^{\star}\|^2 = \|\mL^{\dagger}\mL y_{m}^{\star}\|^2 = \|\mL^{\dagger}\nabla f_m(x^{\star})\|^2\le \frac{1}{\lambda_{\min}^+(\mL^\top \mL)}\|\nabla f_m(x^{\star})\|^2= \frac{1}{\delta}\|\nabla f_m(x^{\star})\|^2.$$
\end{proof}

\newpage


\refstepcounter{chapter}%
\chapter*{\thechapter \quad Appendix C for Chapter 3}
\label{appendixC}

\section{Additional Experiments}

\subsection{Additional experiments with ProxSkip-VR}

We conduct further experiments to validate the efficiency of our proposed method \gls{ProxSkip-VR}. We instantiate our variance reduction design with \algname{LSVRG} and compare our method with several baselines across various  datasets (\textbf{w8a}/\textbf{a9a}), different numbers of workers (10/20), and different batch sizes (16/32/64). All results in Figures~\ref{fig:053},~\ref{fig:054},~\ref{fig:051},~\ref{fig:052} show that \gls{ProxSkip-VR}  achieves linear convergence and outperforms the baselines. 

\begin{figure}[!htbp]
	\centering
	\begin{subfigure}[b]{0.32\textwidth}
		\centering
		\includegraphics[trim=20 10 40 40, clip, width=\textwidth]{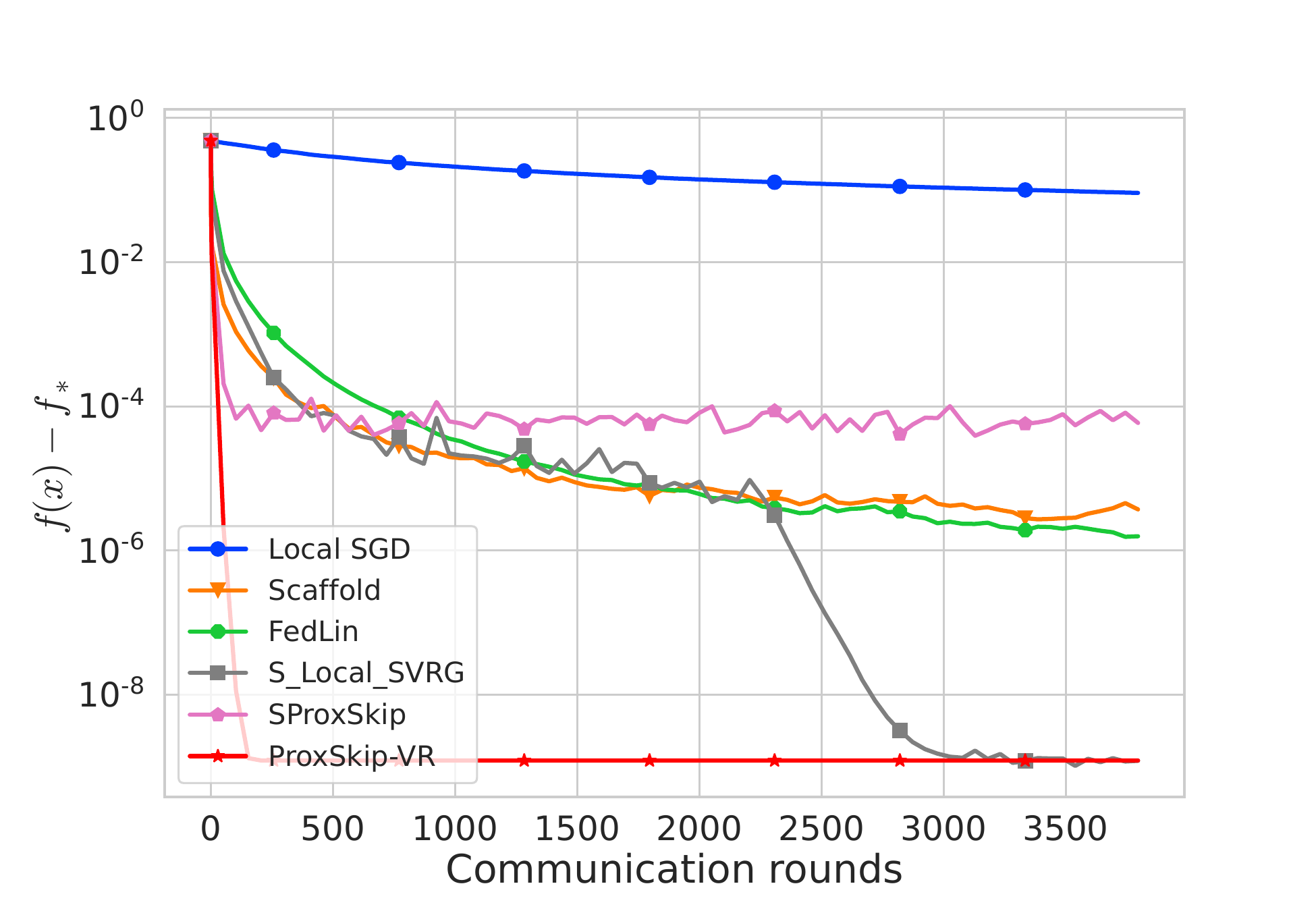}
		\caption{$\tau=16$.}
	\end{subfigure}
	\hfill
	\begin{subfigure}[b]{0.32\textwidth}
		\centering
		\includegraphics[trim=20 10 40 40, clip, width=\textwidth]{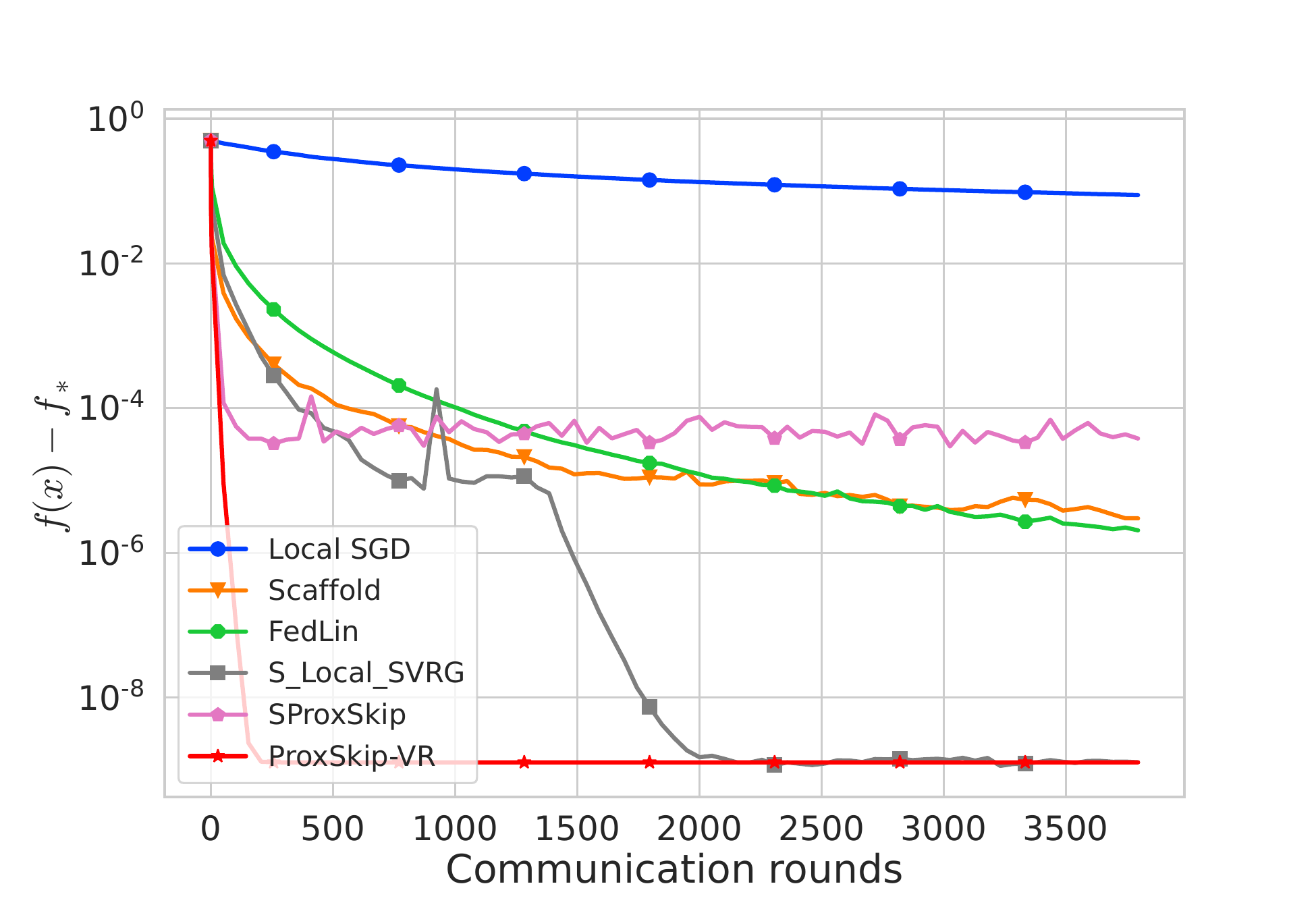}
		\caption{$\tau=32$.}
	\end{subfigure}
	\hfill
	\begin{subfigure}[b]{0.32\textwidth}
		\centering
		\includegraphics[trim=20 10 40 40, clip, width=\textwidth]{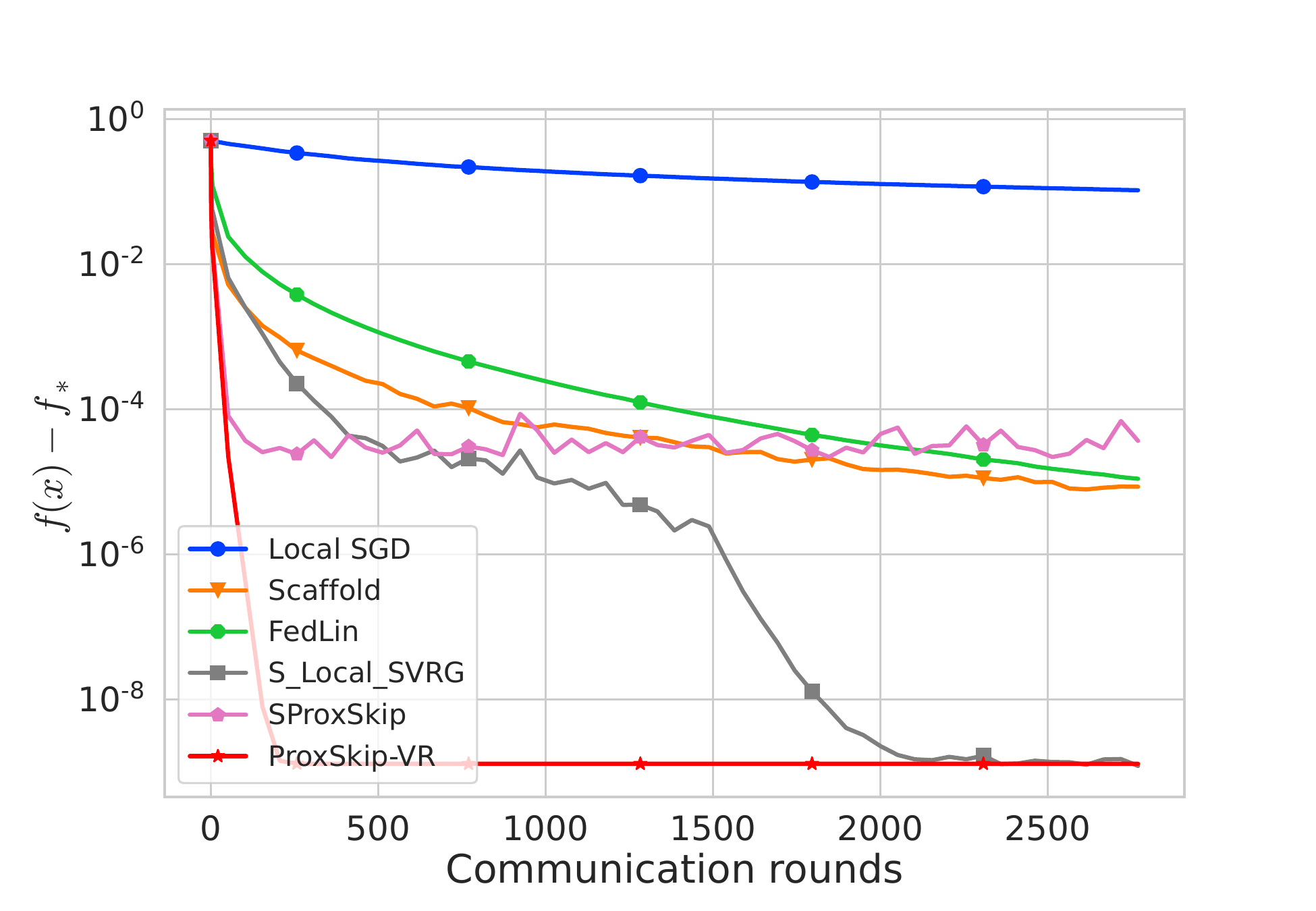}
		\caption{$\tau=64$.}
	\end{subfigure}
	\caption{Convergence results with 20 distributed workers on \textbf{w8a} dataset, $\kappa=1e3$.}
	\label{fig:053}
\end{figure} 

\begin{figure}[!htbp]
	\centering
	\begin{subfigure}[b]{0.32\textwidth}
		\centering
		\includegraphics[trim=20 10 40 40, clip, width=\textwidth]{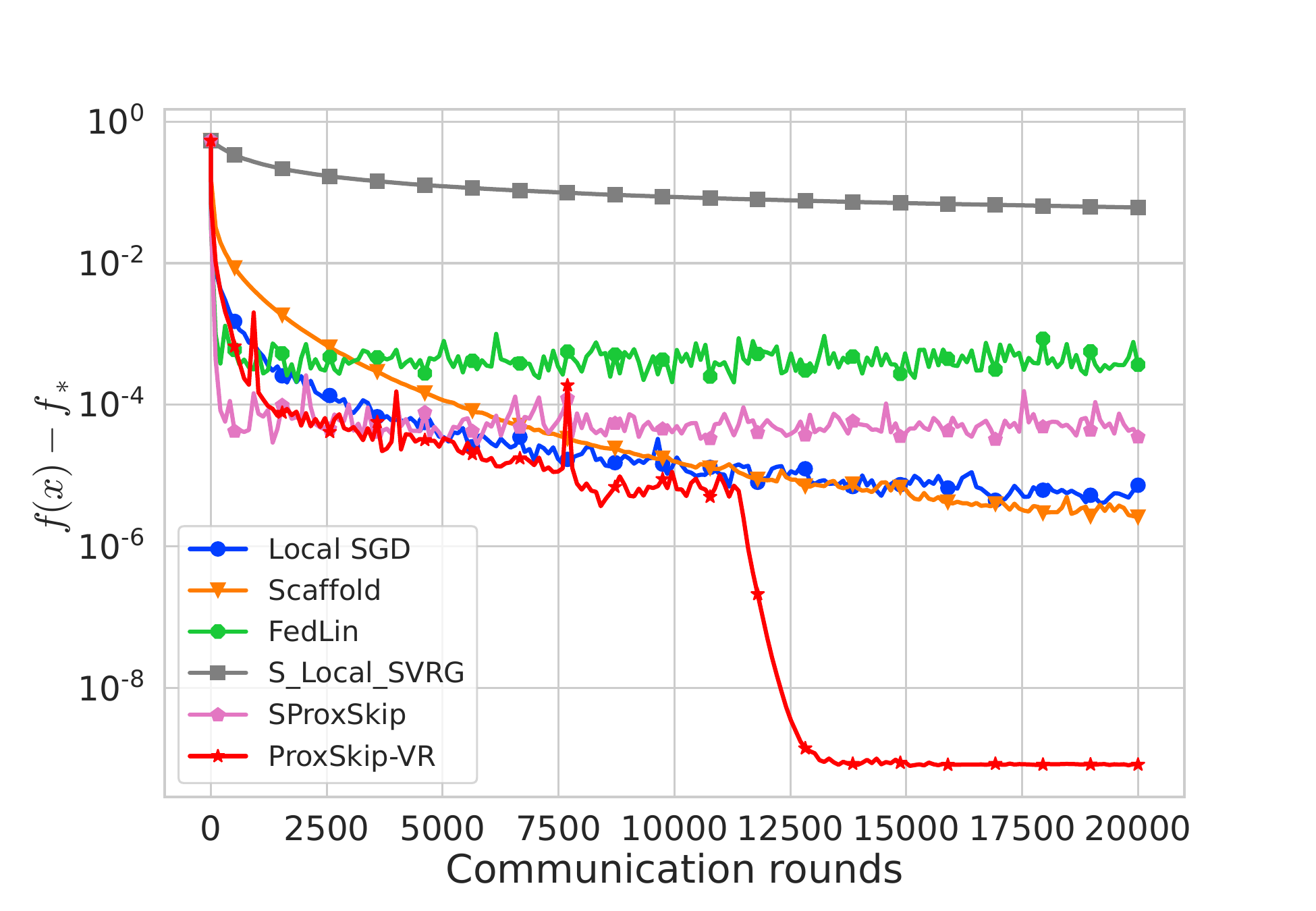}
		\caption{$\tau=16$.}
	\end{subfigure}
	\hfill
	\begin{subfigure}[b]{0.32\textwidth}
		\centering
		\includegraphics[trim=20 10 40 40, clip, width=\textwidth]{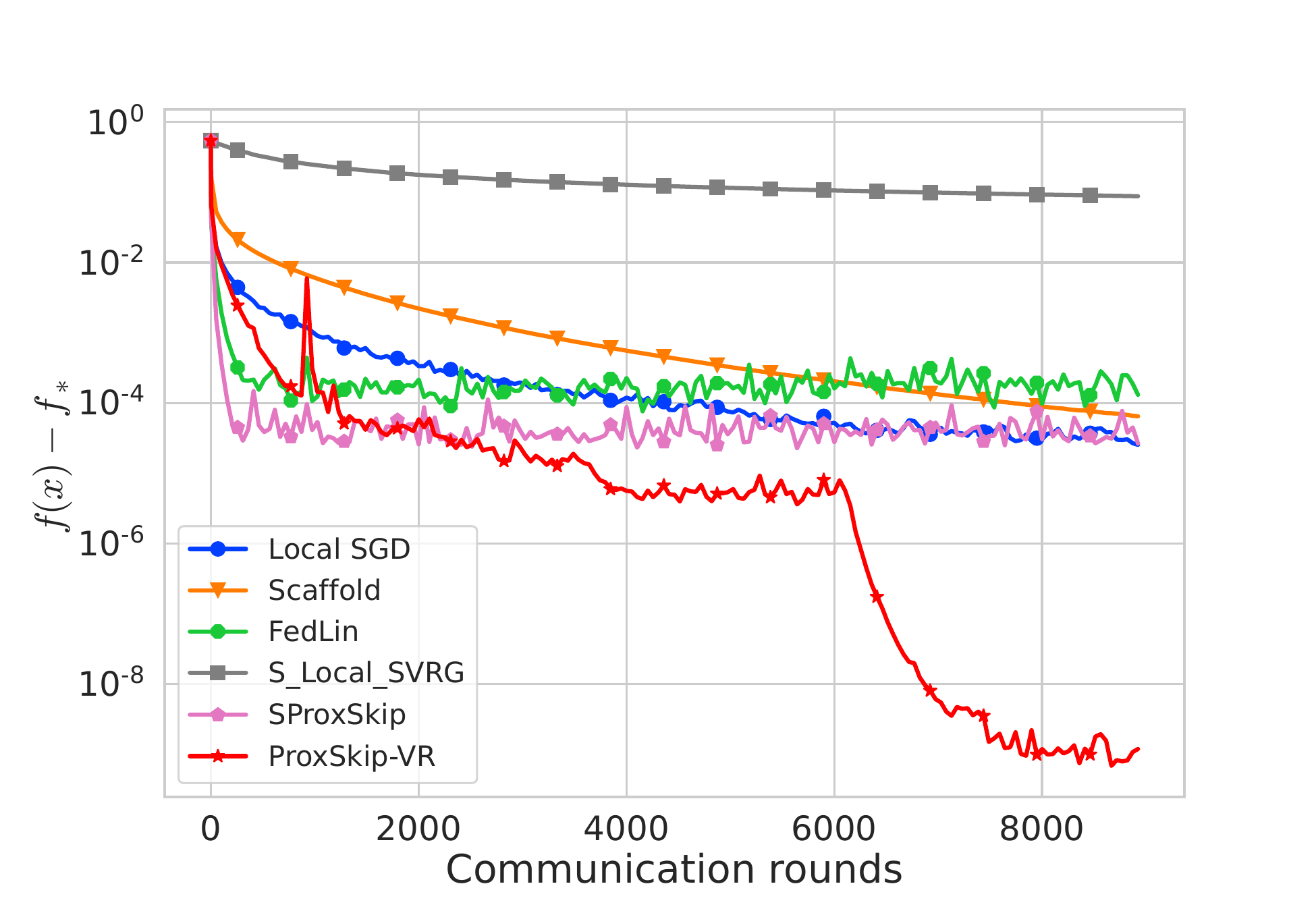}
		\caption{$\tau=32$.}
	\end{subfigure}
	\hfill
	\begin{subfigure}[b]{0.32\textwidth}
		\centering
		\includegraphics[trim=20 10 40 40, clip, width=\textwidth]{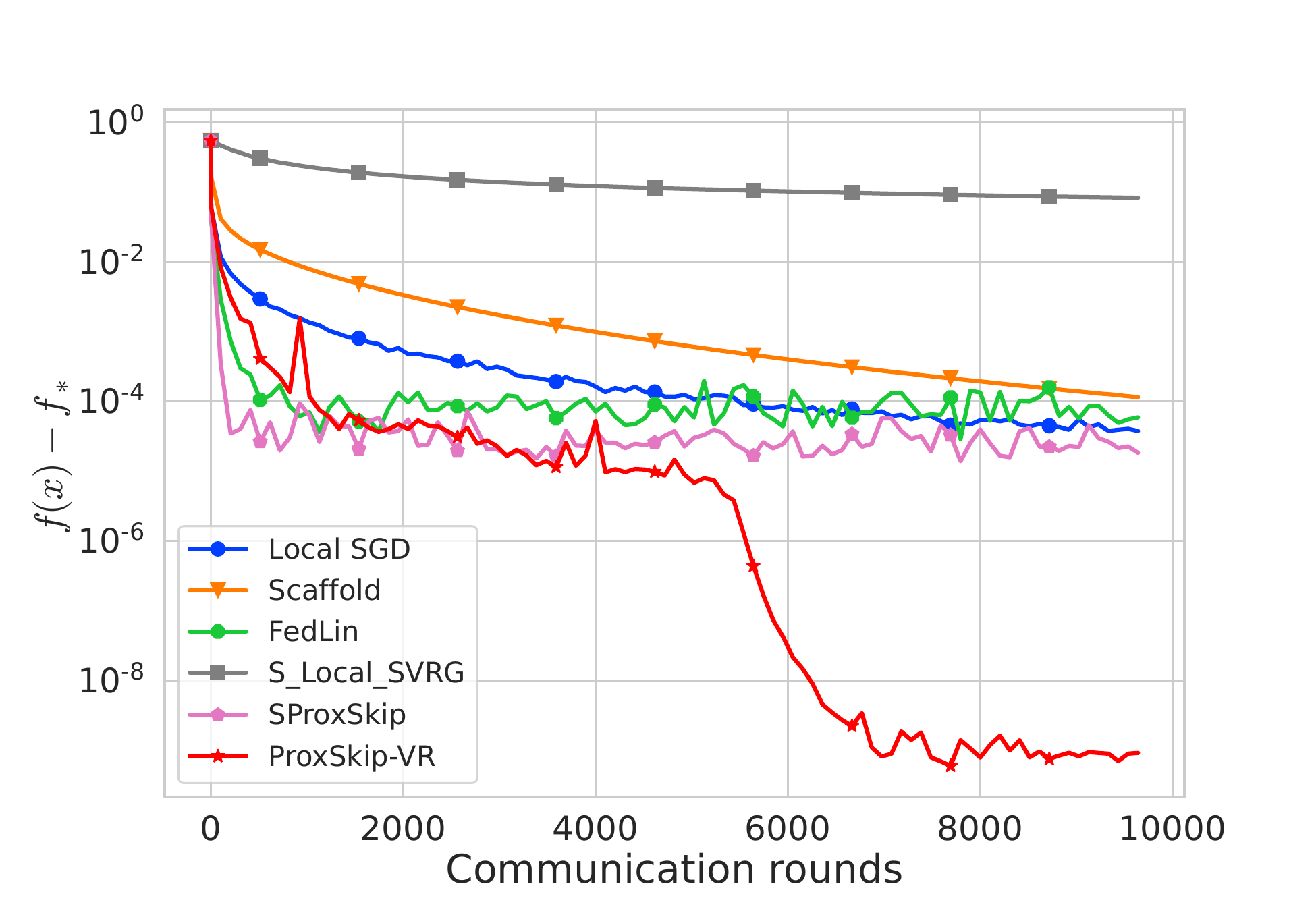}
		\caption{$\tau=64$.}
	\end{subfigure}
	\caption{Convergence results with 20 distributed workers on \textbf{w8a} dataset, $\kappa=1e4$.}
	\label{fig:054}
\end{figure}

\begin{figure}[!htbp]
	\centering
	\begin{subfigure}[b]{0.32\textwidth}
		\centering
		\includegraphics[trim=20 10 40 40, clip, width=\textwidth]{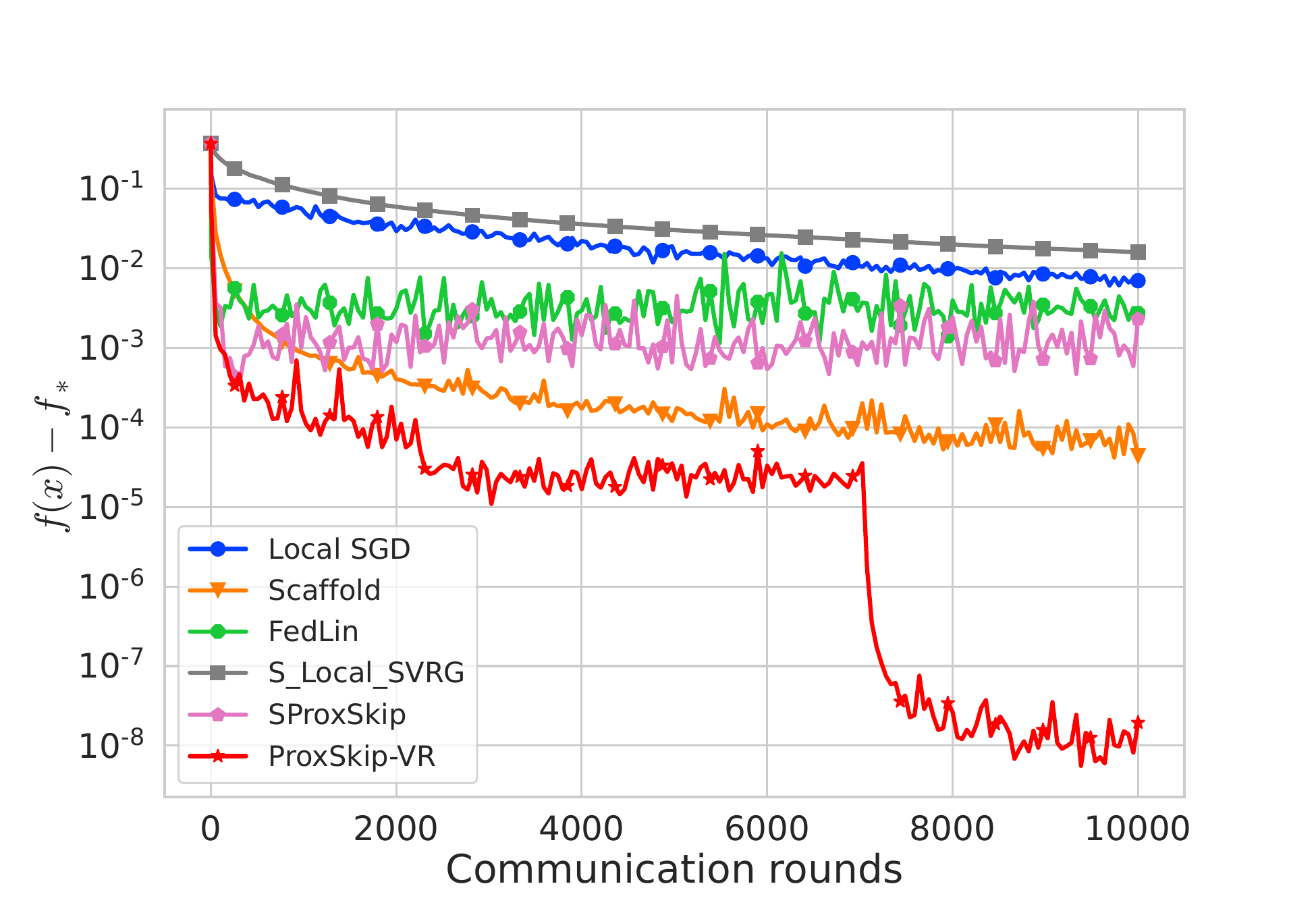}
		\caption{$\tau=16$.}
	\end{subfigure}
	\hfill
	\begin{subfigure}[b]{0.32\textwidth}
		\centering
		\includegraphics[trim=20 10 40 40, clip, width=\textwidth]{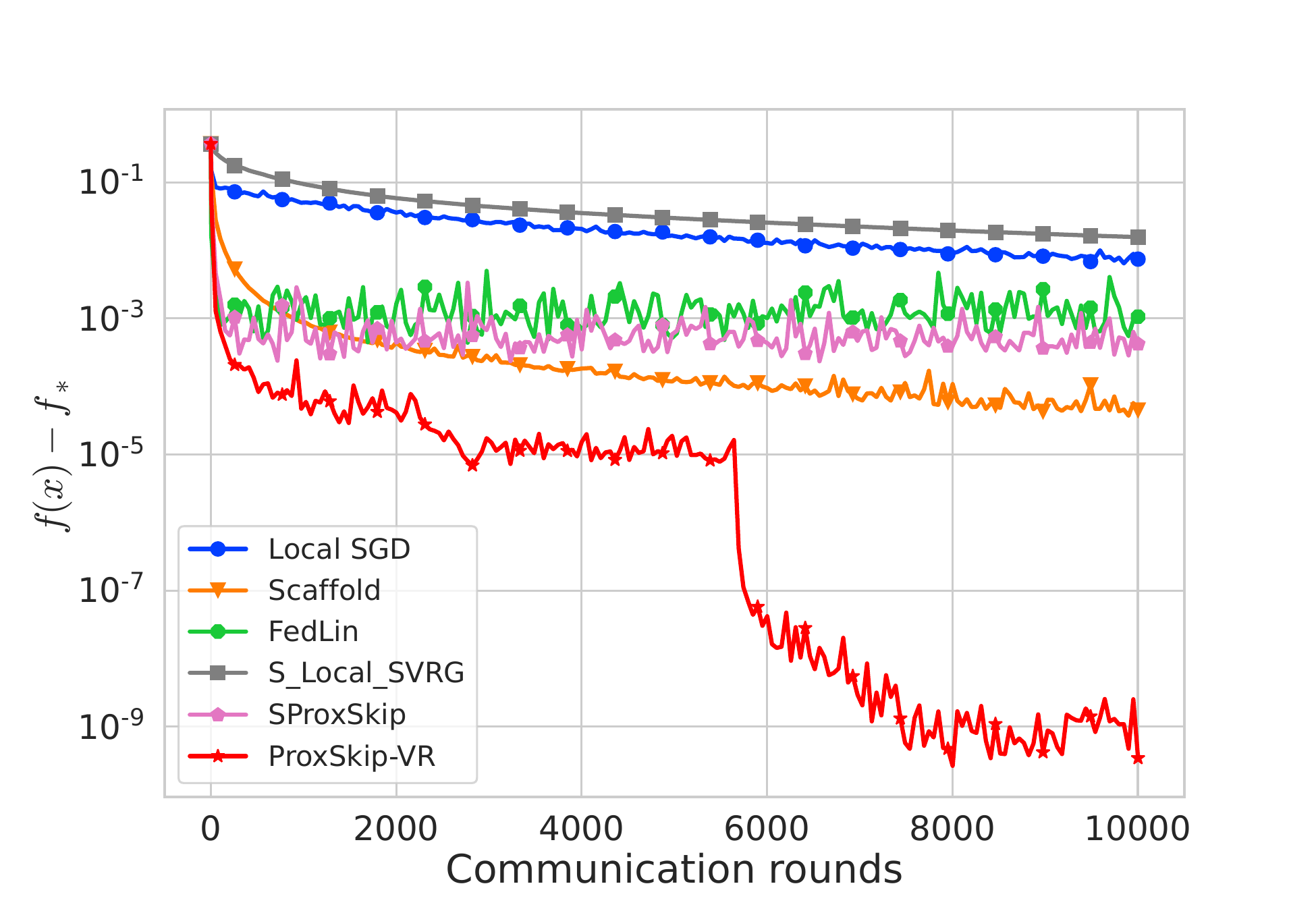}
		\caption{$\tau=32$.}
	\end{subfigure}
	\hfill
	\begin{subfigure}[b]{0.32\textwidth}
		\centering
		\includegraphics[trim=20 10 40 40, clip, width=\textwidth]{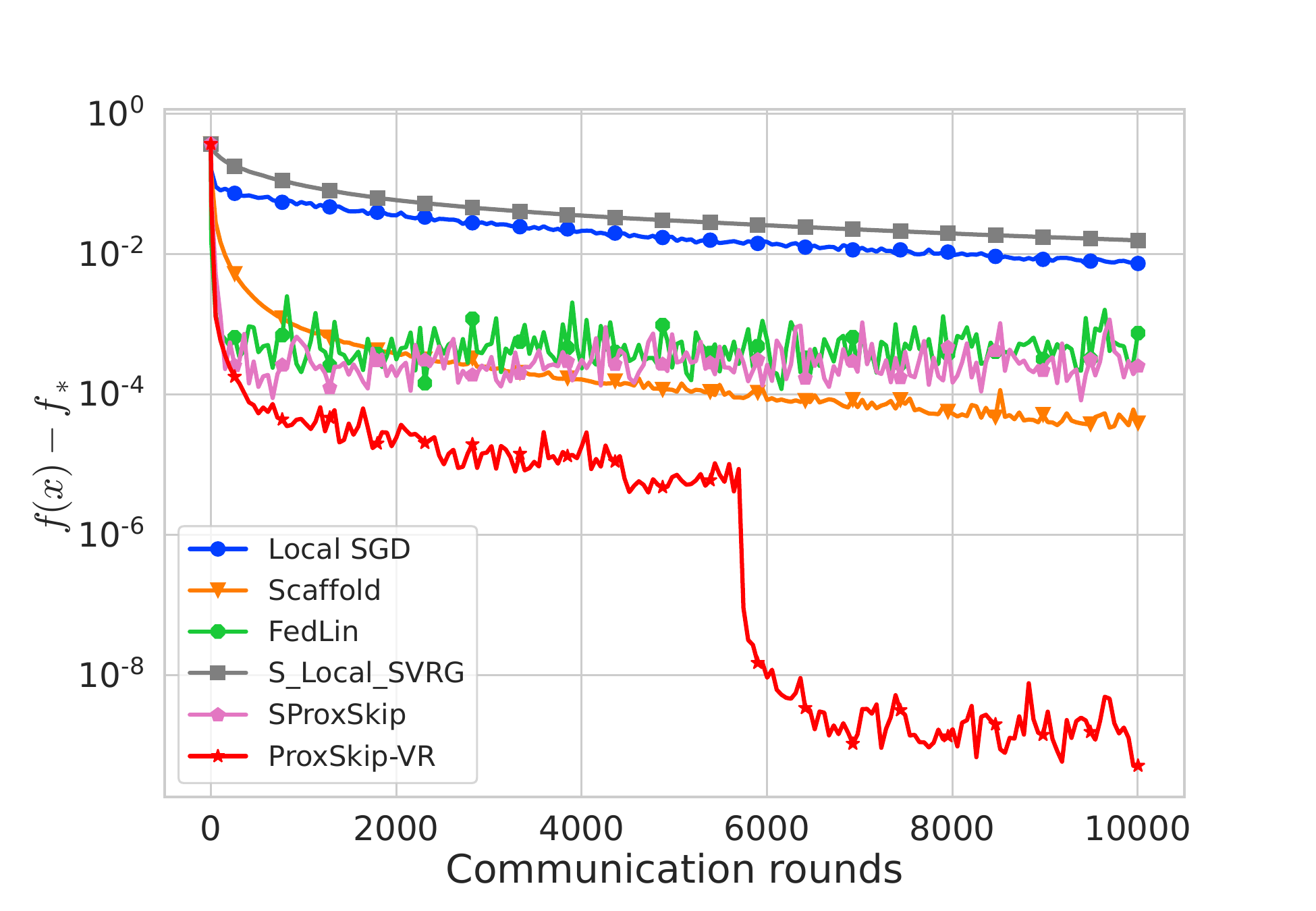}
		\caption{$\tau=64$.}
	\end{subfigure}
	\caption{Convergence results with 10 distributed workers on \textbf{a9a} dataset, $\kappa=1e4$.}
	\label{fig:051}
\end{figure} 

\begin{figure}[!htbp]
	\centering
	\begin{subfigure}[b]{0.32\textwidth}
		\centering
		\includegraphics[trim=20 10 40 40, clip, width=\textwidth]{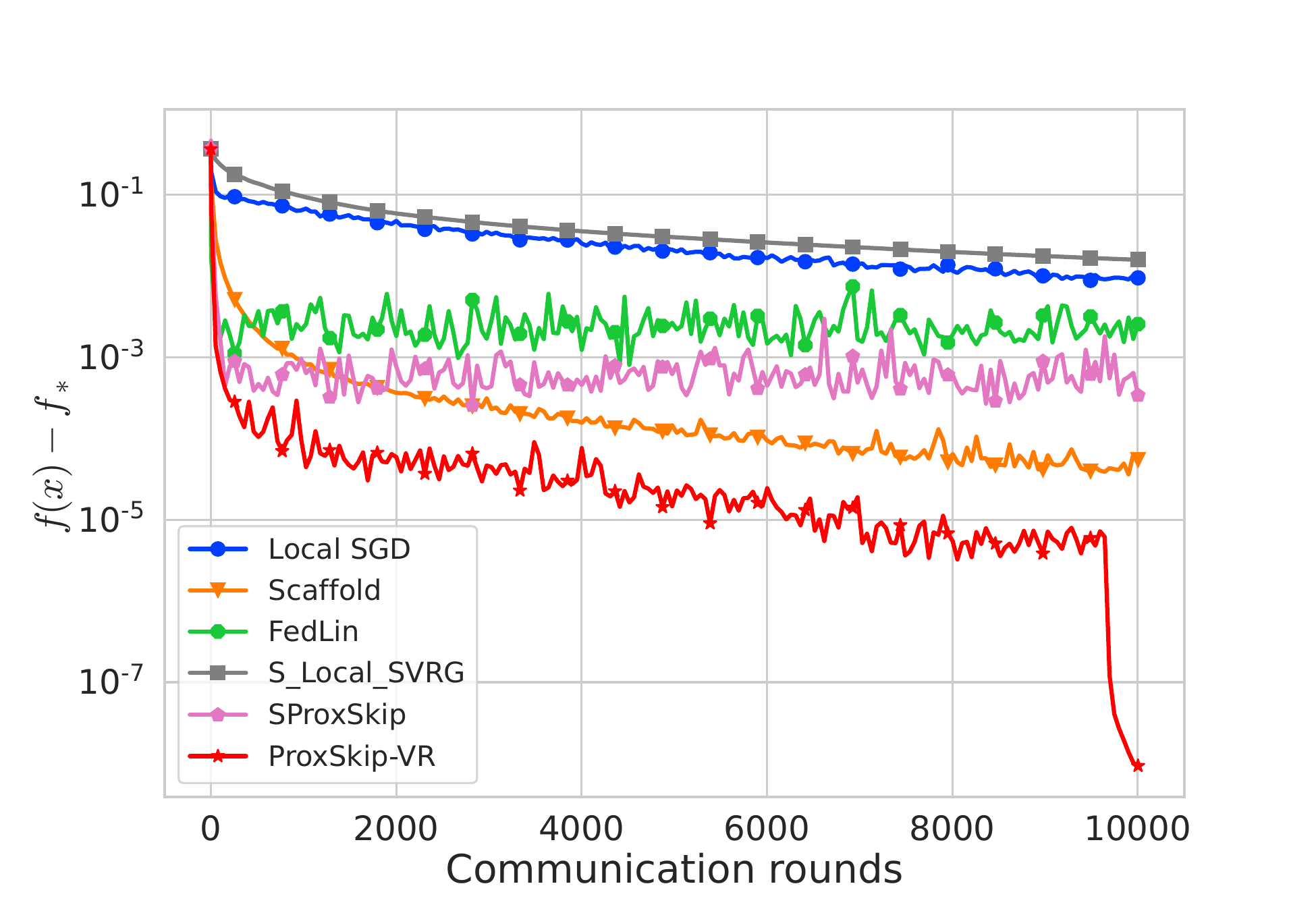}
		\caption{$\tau=16$.}
	\end{subfigure}
	\hfill
	\begin{subfigure}[b]{0.32\textwidth}
		\centering
		\includegraphics[trim=20 10 40 40, clip, width=\textwidth]{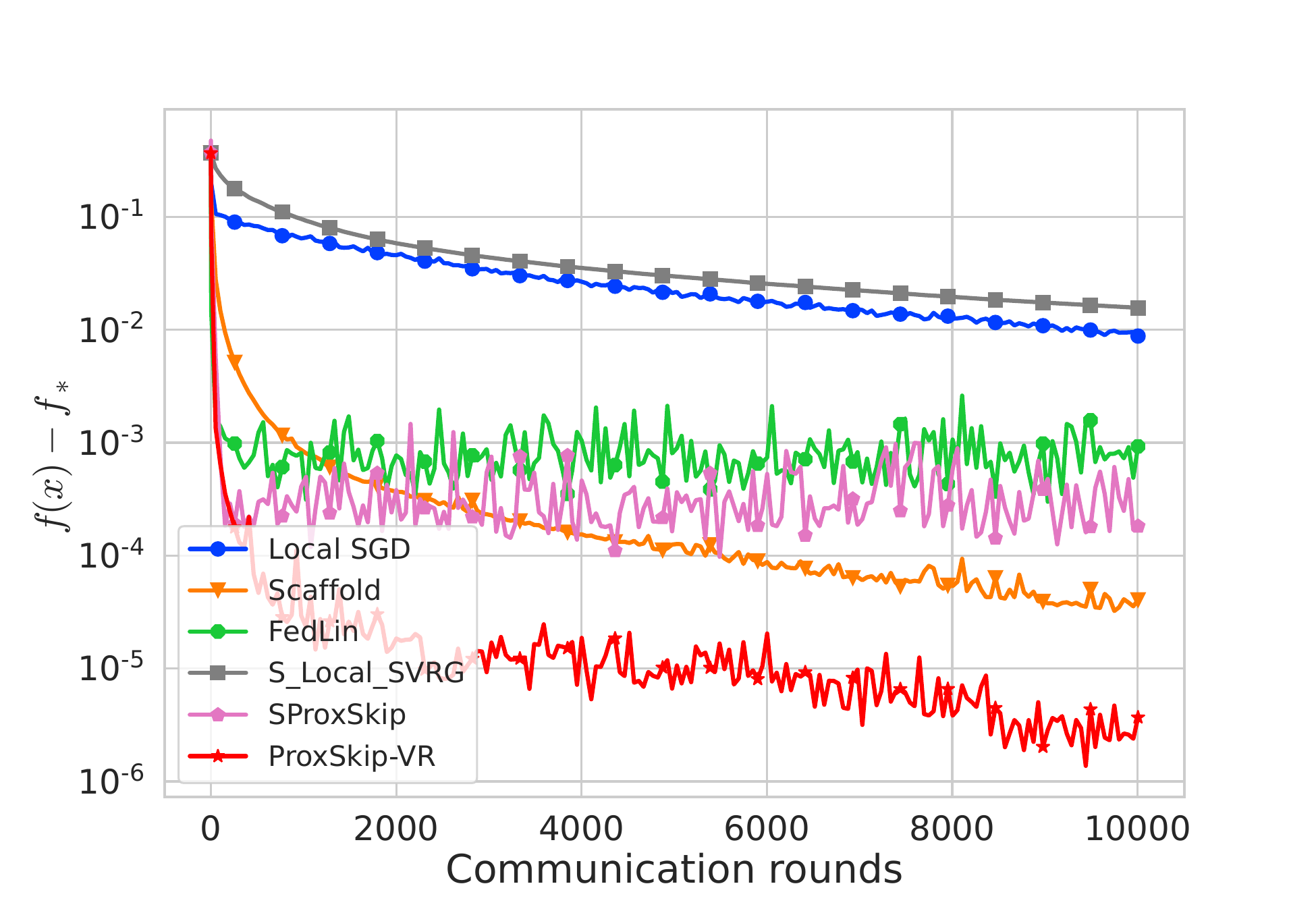}
		\caption{$\tau=32$.}
	\end{subfigure}
	\hfill
	\begin{subfigure}[b]{0.32\textwidth}
		\centering
		\includegraphics[trim=20 10 40 40, clip, width=\textwidth]{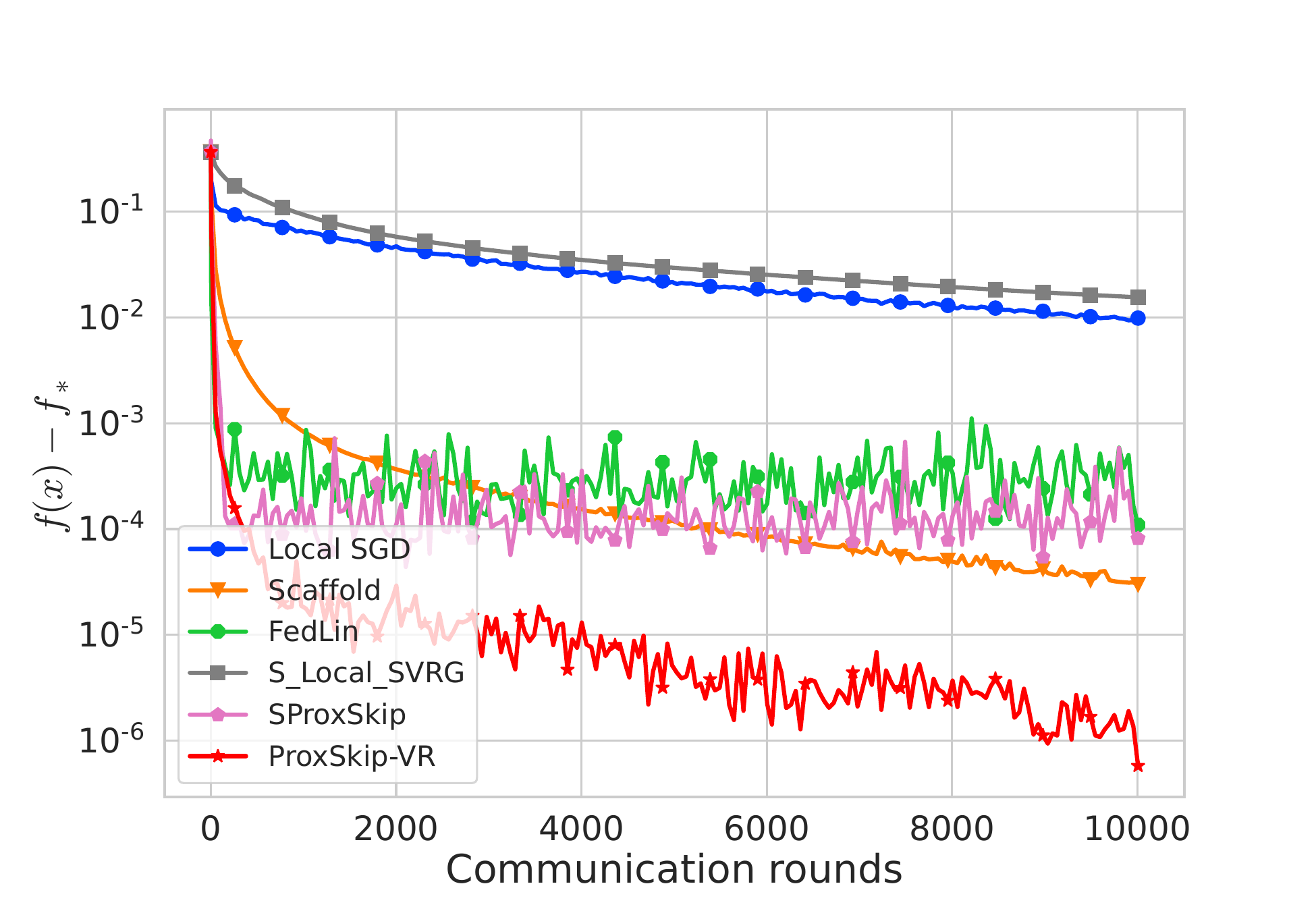}
		\caption{$\tau=64$.}
	\end{subfigure}
	\caption{Convergence results with 20 distributed workers on \textbf{a9a} dataset, $\kappa=1e4$.}
	\label{fig:052}
\end{figure} 

\subsection{Cost ratio: ProxSkip vs. ProxSkip-VR
}

Here we report on more experiments related to the total cost ratio of \gls{ProxSkip} over \gls{ProxSkip-VR} in Figures~\ref{fig:055},~\ref{fig:056},~\ref{fig:057},~\ref{fig:058}. When \algname{\gls{GD}} is used as the subroutine, i.e., when \gls{ProxSkip-VR} reduces to \gls{ProxSkip}, the ratio is equal to one, and  is depicted by the red horizontal dashed line. Any cost ratio above one means that \gls{ProxSkip-VR} benefits over \gls{ProxSkip} (e.g., a cost ratio of $10$ means $10\times$ speedup in favor of our method). As seen in the plots,  acceleration of our method over \gls{ProxSkip} is clearly visible, and improves as the local computation cost $\delta$ per sample increases. For large values of $\delta$ (i.e., $\delta\approx 10^{-1}$), the acceleration can reach $20\times$ to $85\times$. Note also that acceleration is more significant for small minibatch sizes. That is, it is better for $\tau=16$ than for $\tau=32$, which is better than in the $\tau=64$ case.  This means that in terms of total cost, it is beneficial for the workers to use smaller minibatch sizes, i.e., it is beneficial to be as far from the full batch regime employed by \gls{ProxSkip} as possible. 

\begin{figure}[!htbp]
	\centering
	\begin{subfigure}[b]{0.32\textwidth}
		\centering
		\includegraphics[trim=20 10 40 40, clip, width=\textwidth]{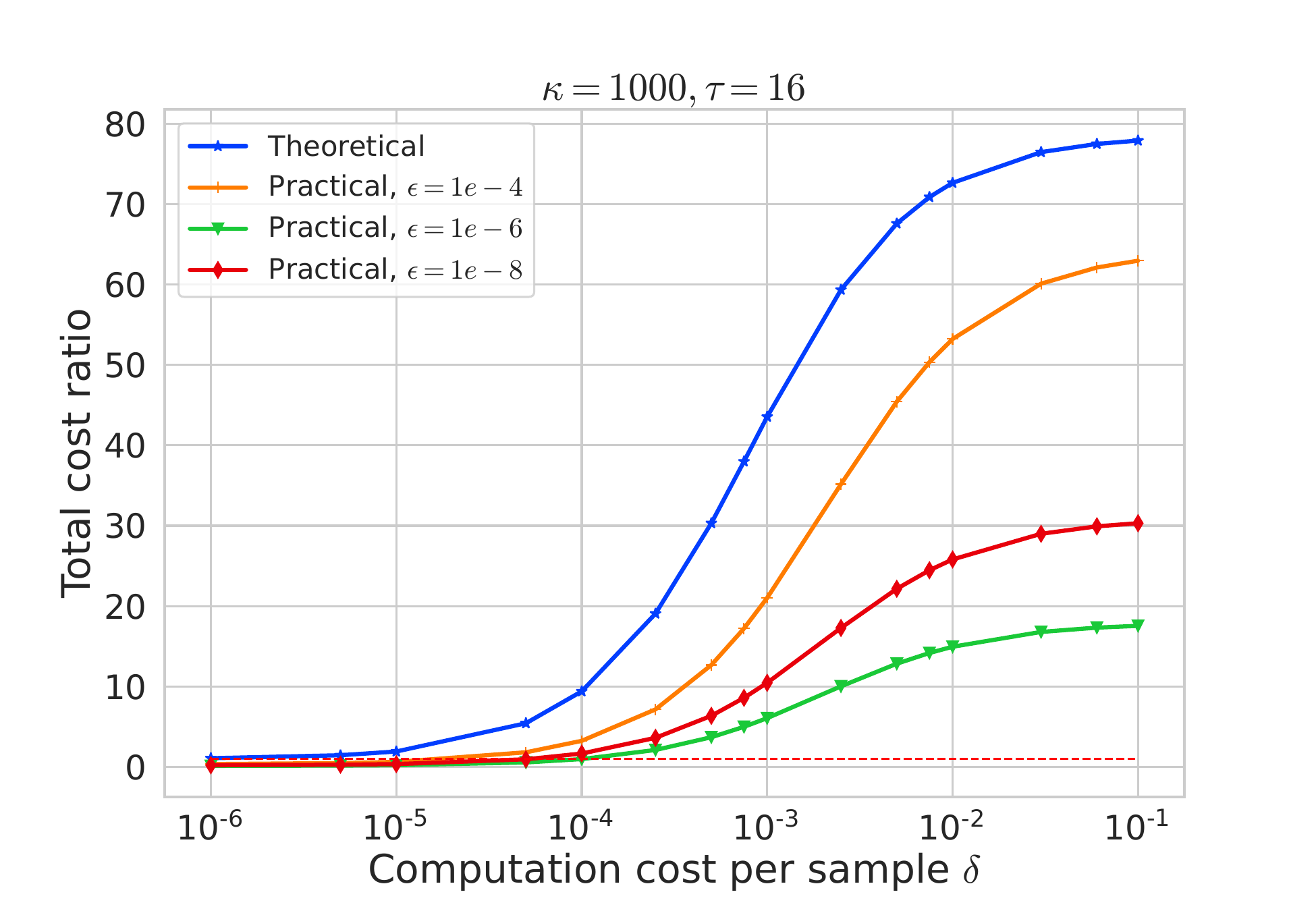}
		\caption{$\tau=16$.}
	\end{subfigure}
	\hfill
	\begin{subfigure}[b]{0.32\textwidth}
		\centering
		\includegraphics[trim=20 10 40 40, clip, width=\textwidth]{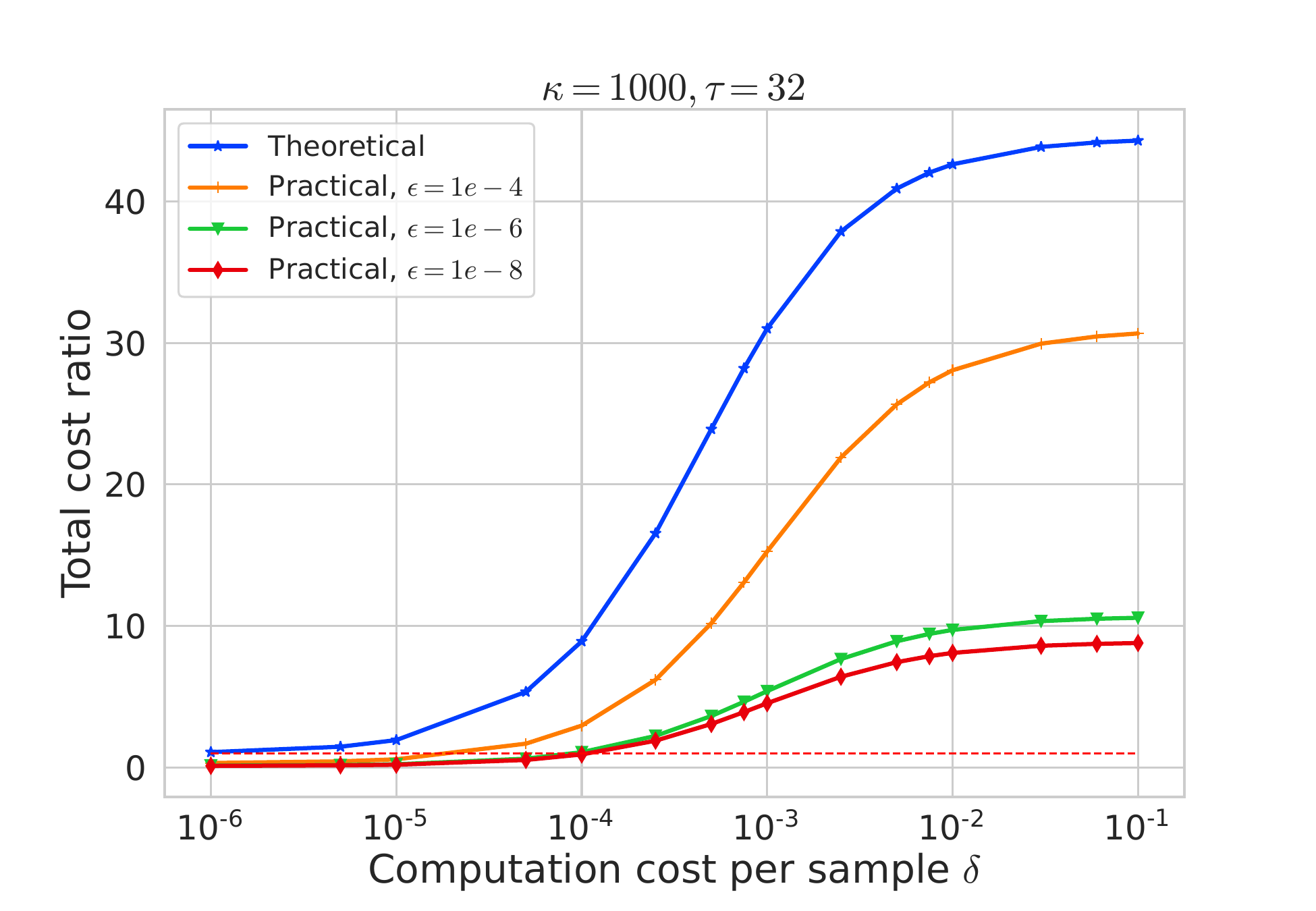}
		\caption{$\tau=32$.}
	\end{subfigure}
	\hfill
	\begin{subfigure}[b]{0.32\textwidth}
		\centering
		\includegraphics[trim=20 10 40 40, clip, width=\textwidth]{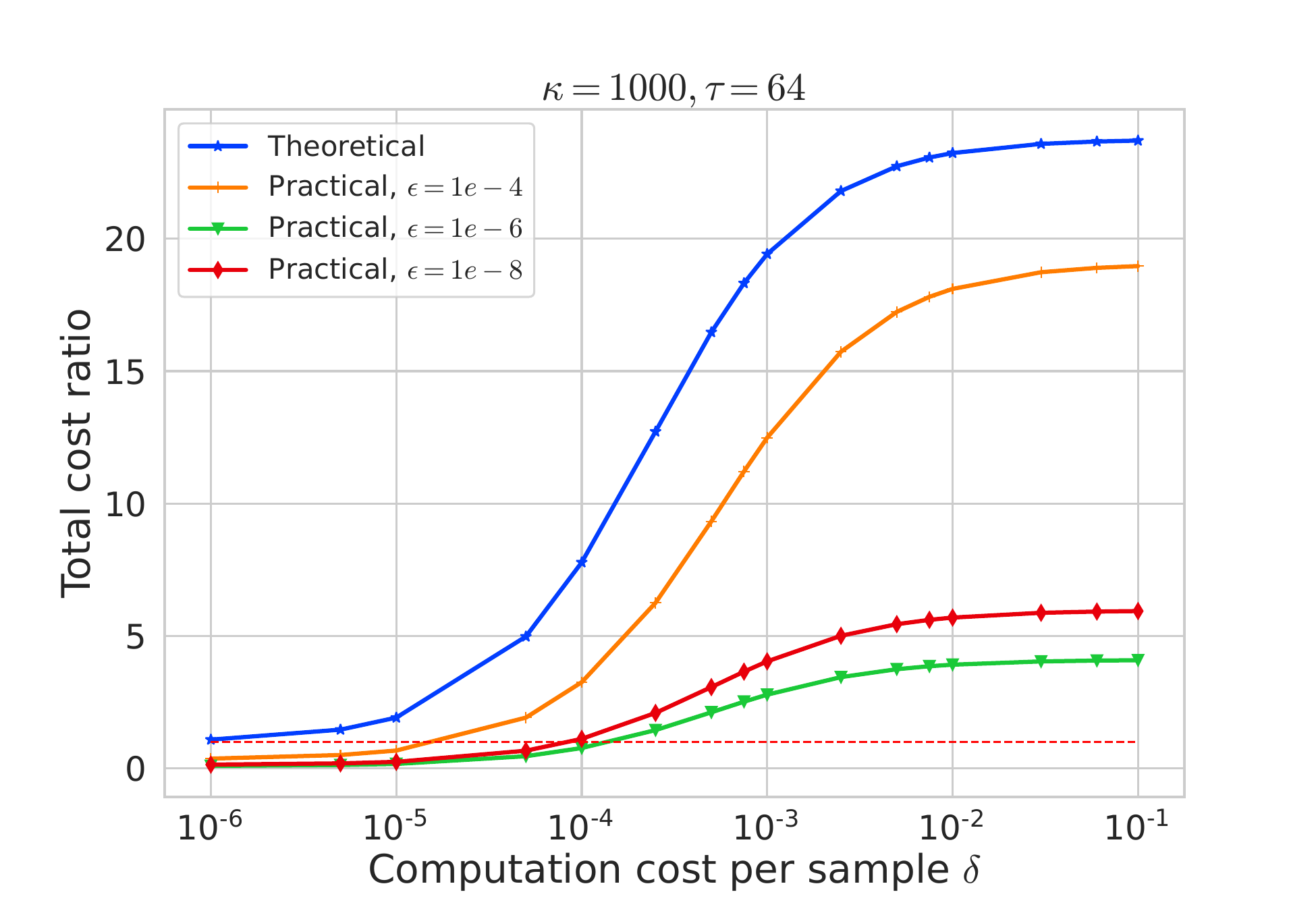}
		\caption{$\tau=64$.}
	\end{subfigure}
	\caption{Acceleration with 10 distributed workers on \textbf{a9a} dataset, $\kappa=1e3$.}
	\label{fig:055}
\end{figure} 

\begin{figure}[!htbp]
	\centering
	\begin{subfigure}[b]{0.32\textwidth}
		\centering
		\includegraphics[trim=20 10 40 40, clip, width=\textwidth]{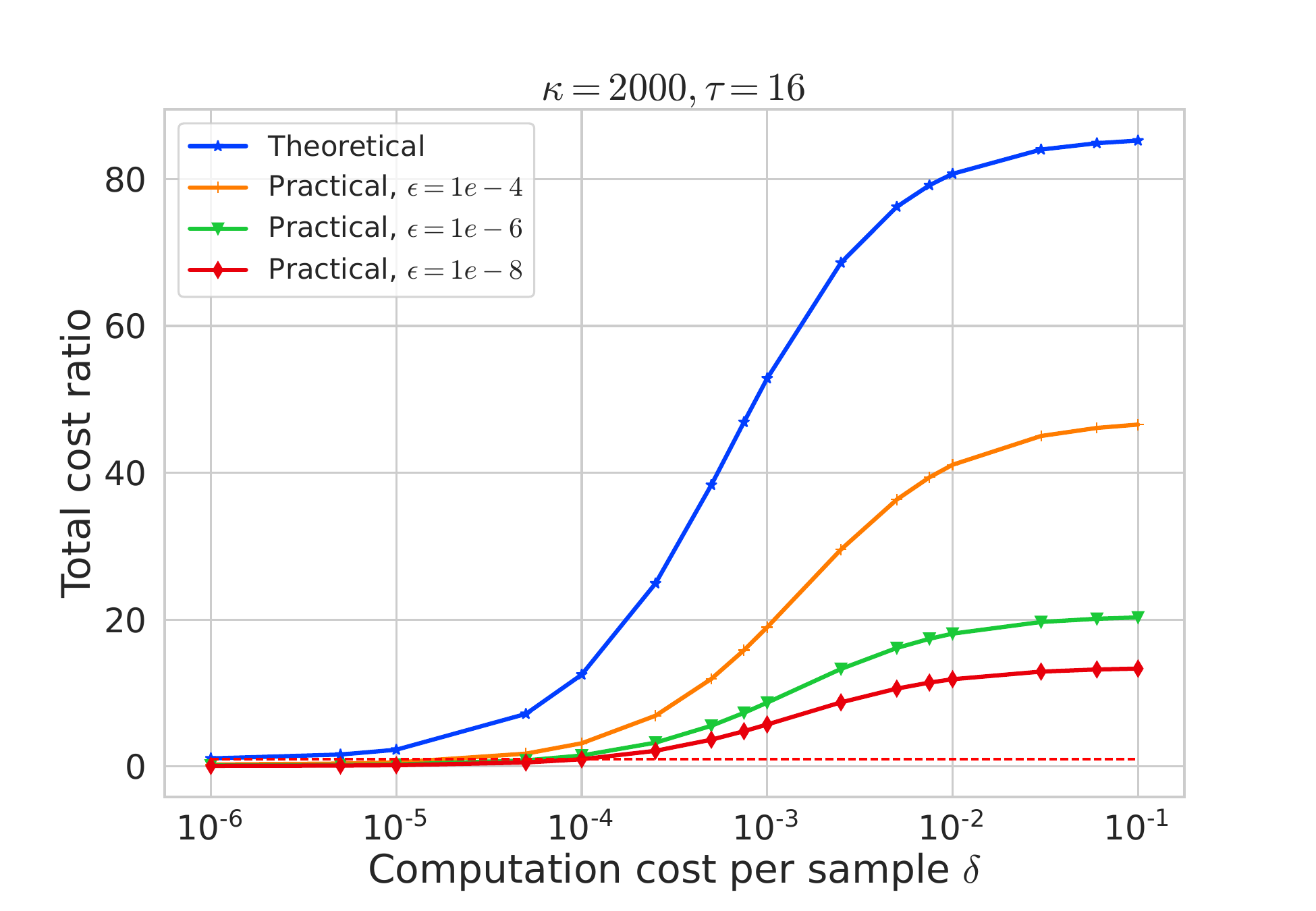}
		\caption{$\tau=16$.}
	\end{subfigure}
	\hfill
	\begin{subfigure}[b]{0.32\textwidth}
		\centering
		\includegraphics[trim=20 10 40 40, clip, width=\textwidth]{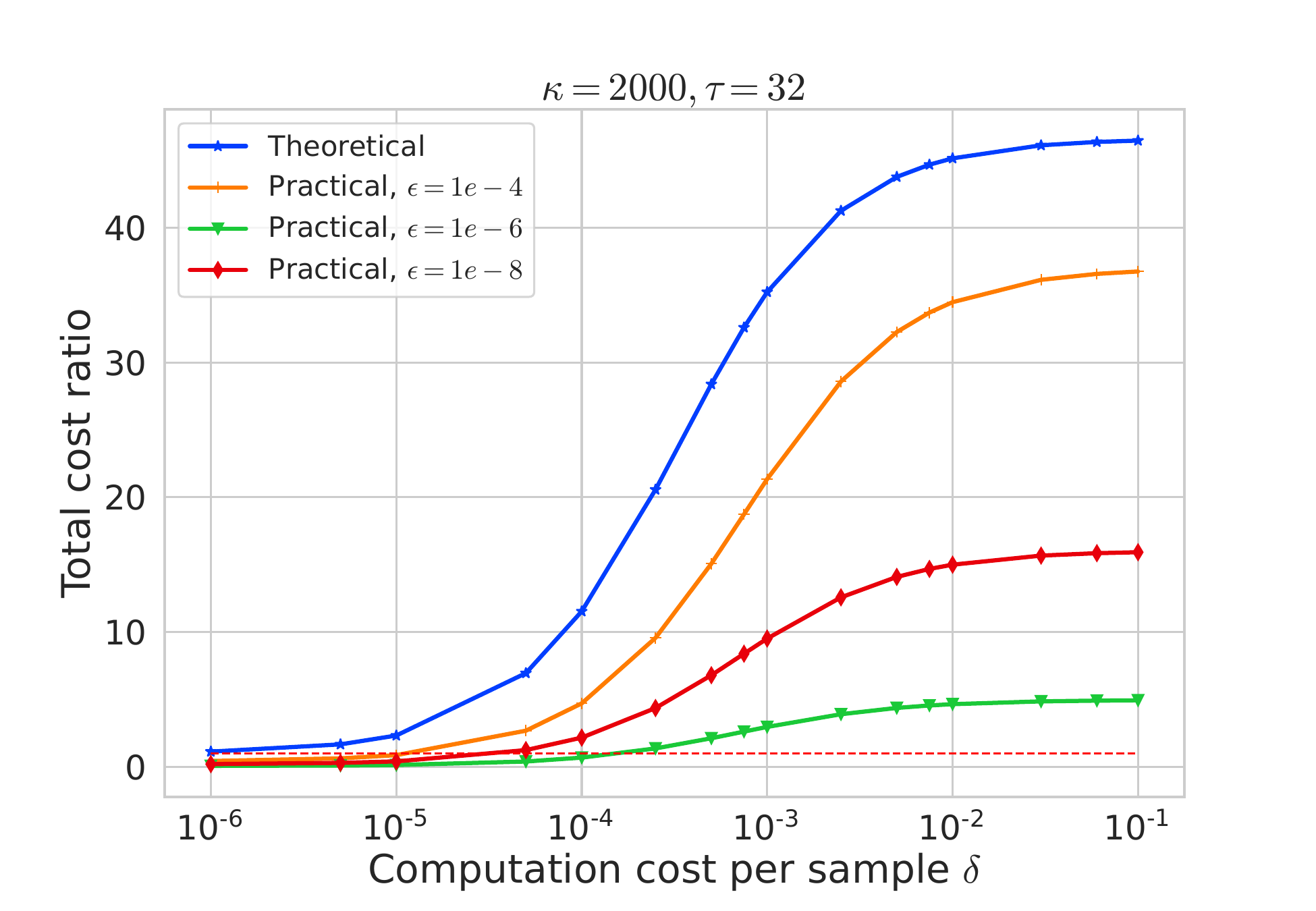}
		\caption{$\tau=32$.}
	\end{subfigure}
	\hfill
	\begin{subfigure}[b]{0.32\textwidth}
		\centering
		\includegraphics[trim=20 10 40 40, clip, width=\textwidth]{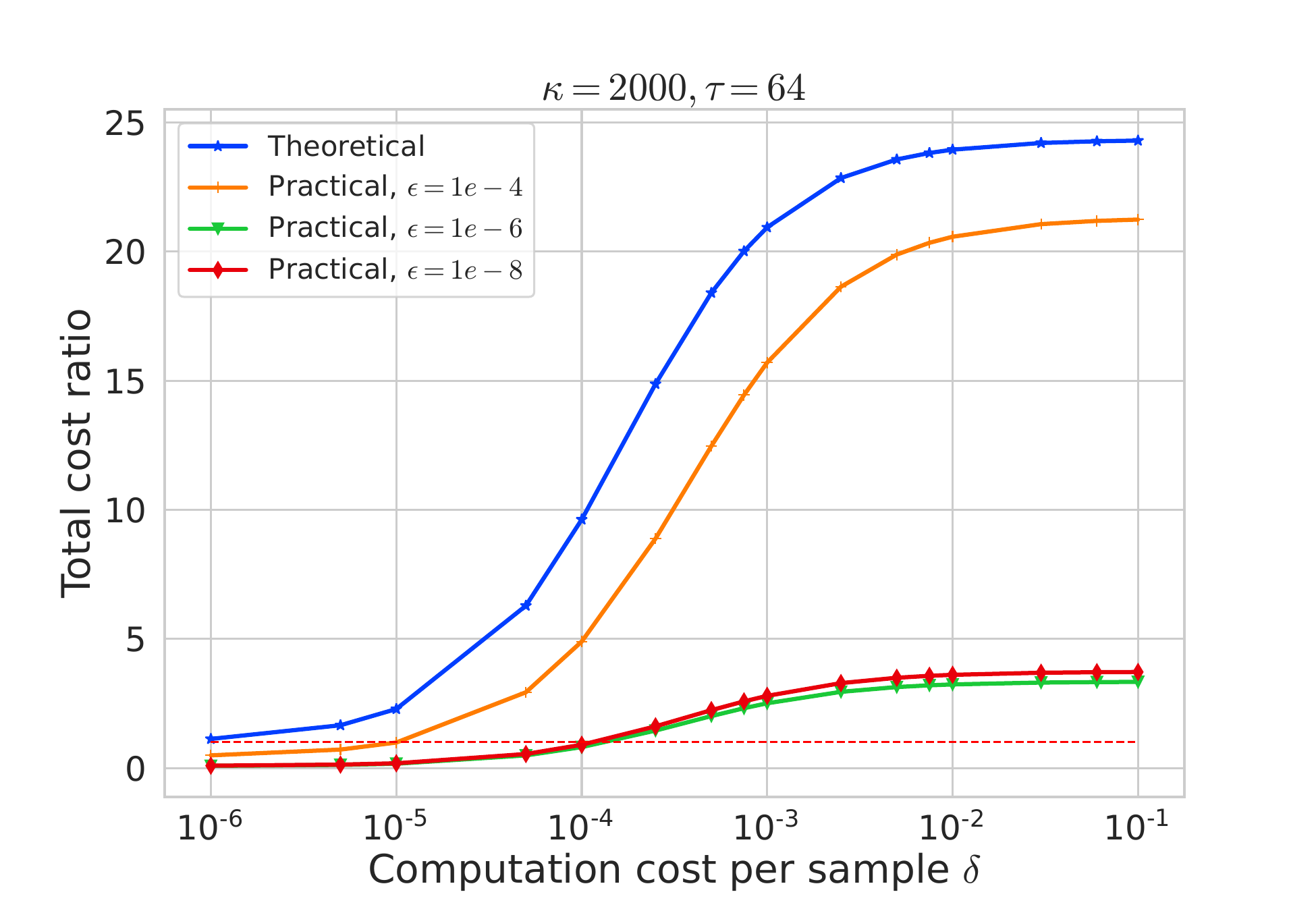}
		\caption{$\tau=64$.}
	\end{subfigure}
	\caption{Acceleration with 10 distributed workers on \textbf{a9a} dataset, $\kappa=2e3$.}
	\label{fig:056}
\end{figure} 

\begin{figure}[!htbp]
	\centering
	\begin{subfigure}[b]{0.32\textwidth}
		\centering
		\includegraphics[trim=20 10 40 40, clip, width=\textwidth]{proxskip-VR/img/0019_a9a_n10_bs16_cosize10_cc1_kappa_kappa1000.0_error_0.0001.txt.pdf}
		\caption{$\tau=16$.}
	\end{subfigure}
	\hfill
	\begin{subfigure}[b]{0.32\textwidth}
		\centering
		\includegraphics[trim=20 10 40 40, clip, width=\textwidth]{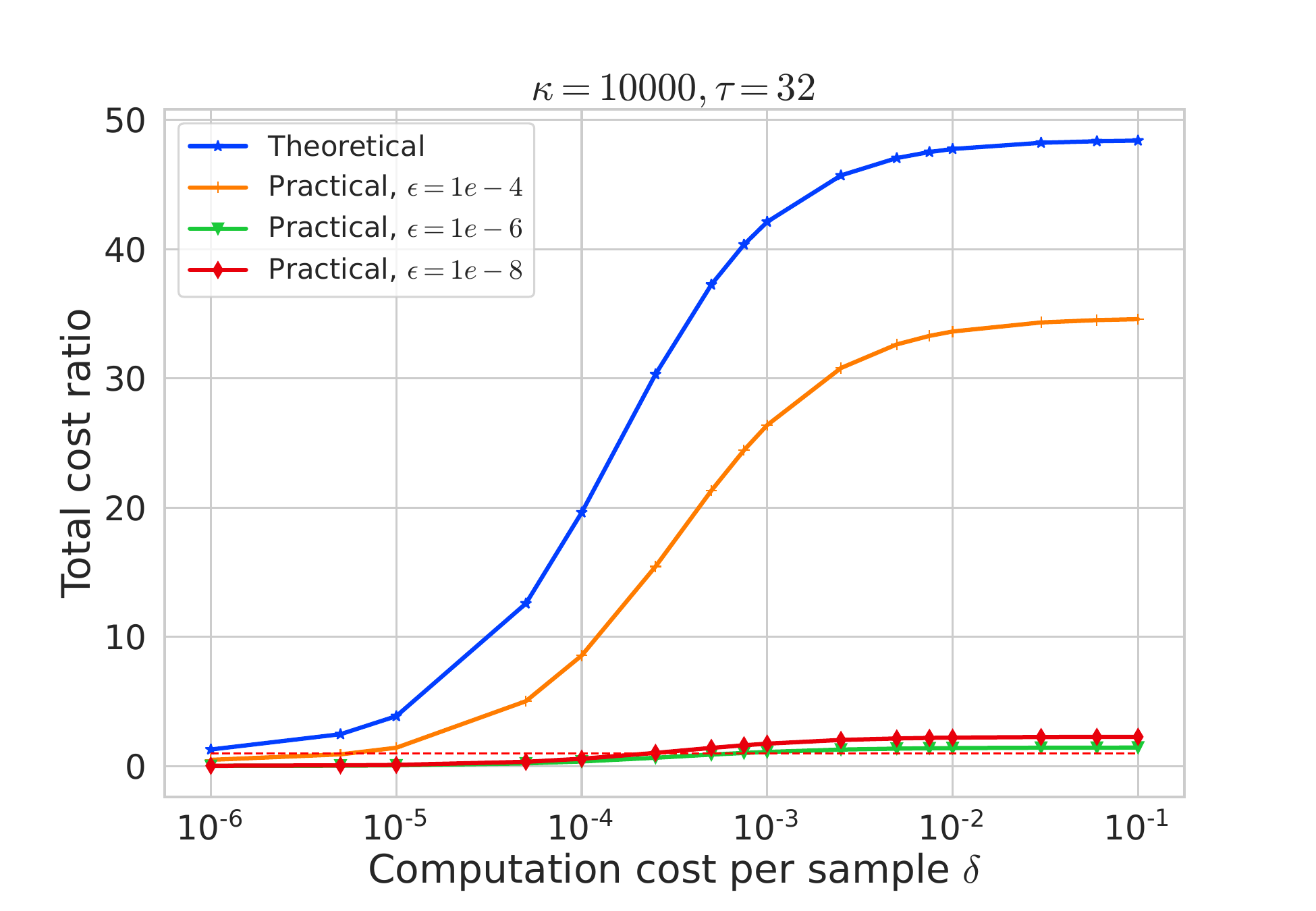}
		\caption{$\tau=32$.}
	\end{subfigure}
	\hfill
	\begin{subfigure}[b]{0.32\textwidth}
		\centering
		\includegraphics[trim=20 10 40 40, clip, width=\textwidth]{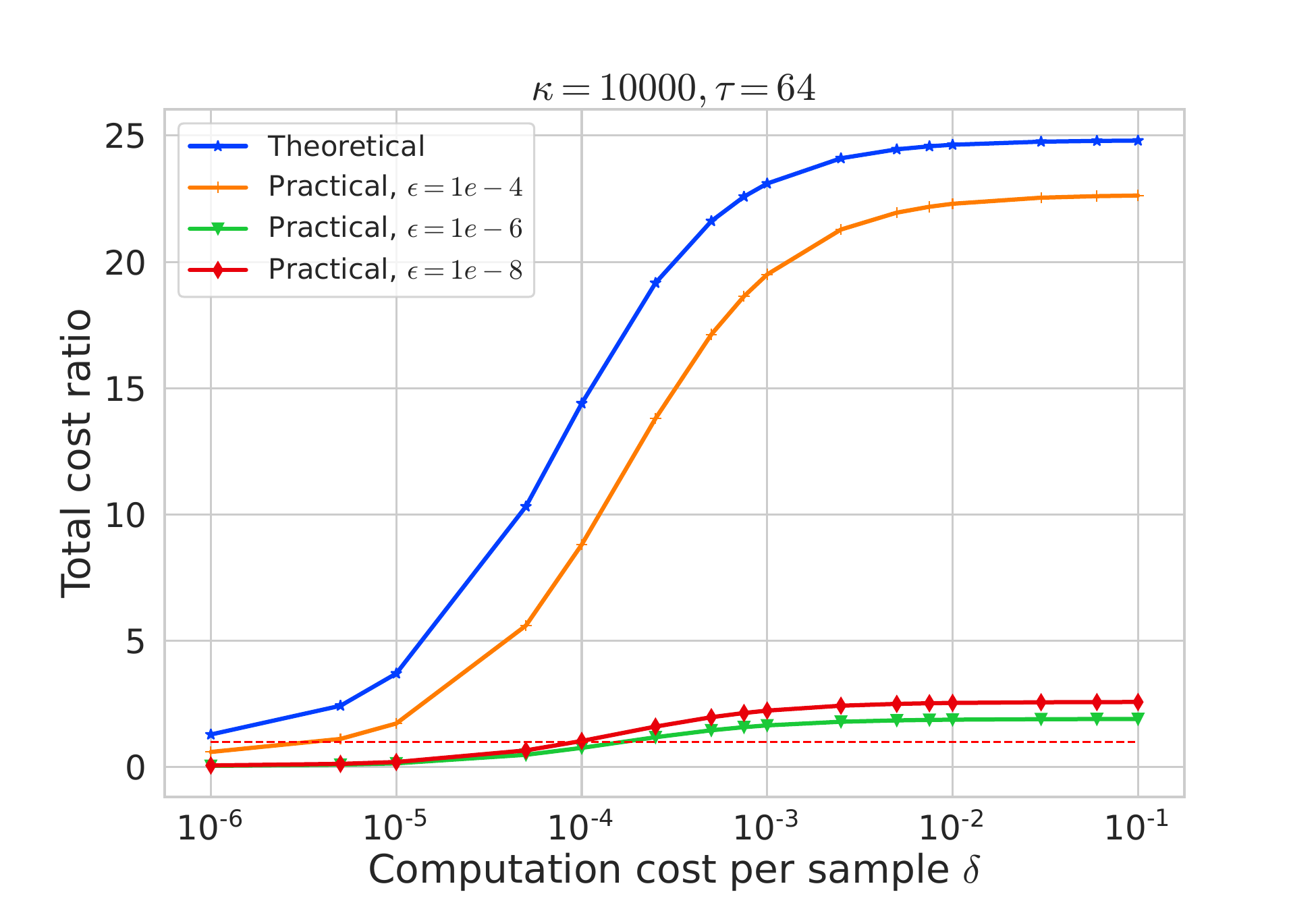}
		\caption{$\tau=64$.}
	\end{subfigure}
	\caption{Acceleration with 10 distributed workers on \textbf{a9a} dataset, $\kappa=1e4$.}
	\label{fig:057}
\end{figure} 

\begin{figure}[!htbp]
	\centering
	\begin{subfigure}[b]{0.32\textwidth}
		\centering
		\includegraphics[trim=20 10 40 40, clip, width=\textwidth]{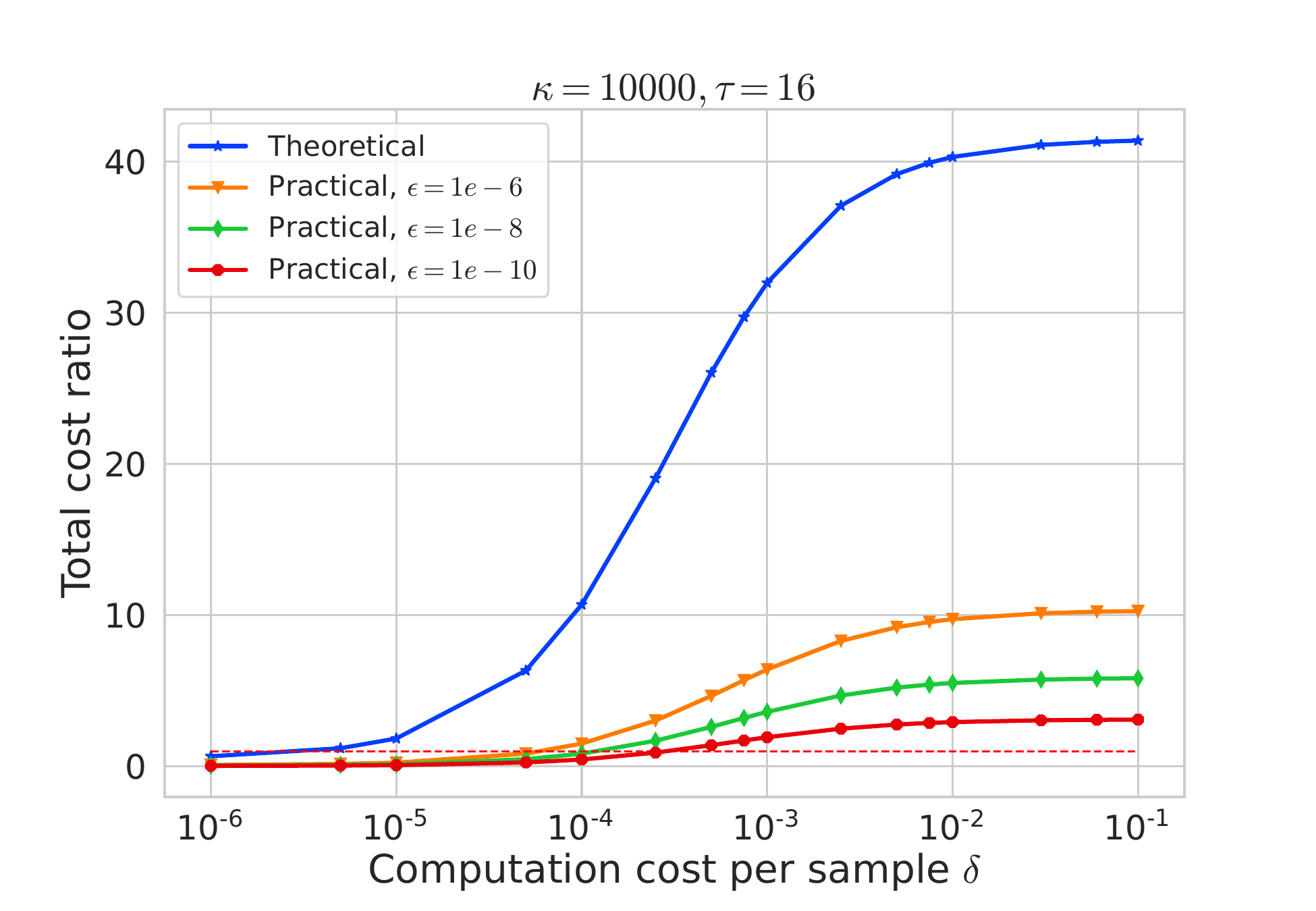}
		\caption{$\tau=16$.}
	\end{subfigure}
	\hfill
	\begin{subfigure}[b]{0.32\textwidth}
		\centering
		\includegraphics[trim=20 10 40 40, clip, width=\textwidth]{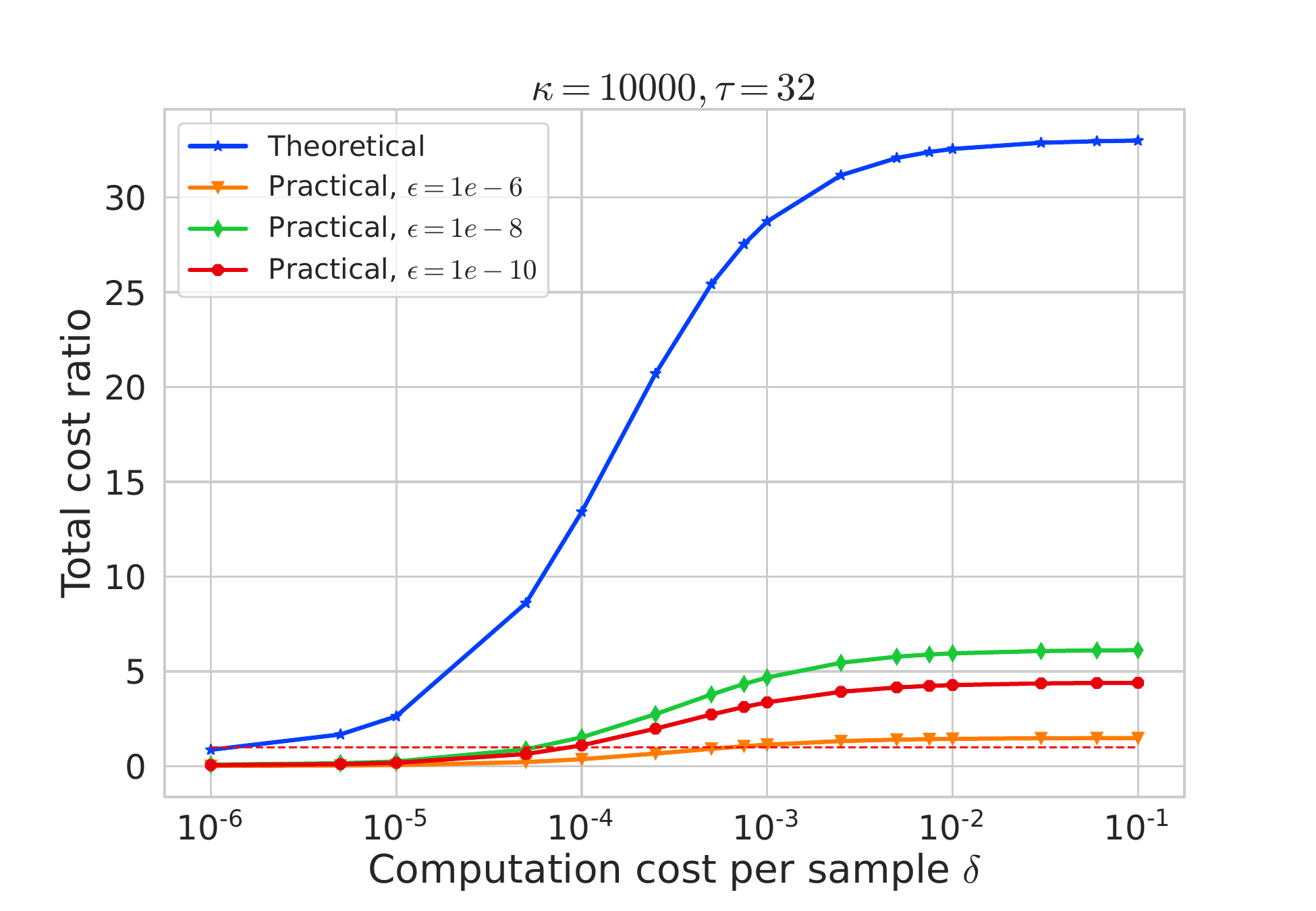}
		\caption{$\tau=32$.}
	\end{subfigure}
	\hfill
	\begin{subfigure}[b]{0.32\textwidth}
		\centering
		\includegraphics[trim=20 10 40 40, clip, width=\textwidth]{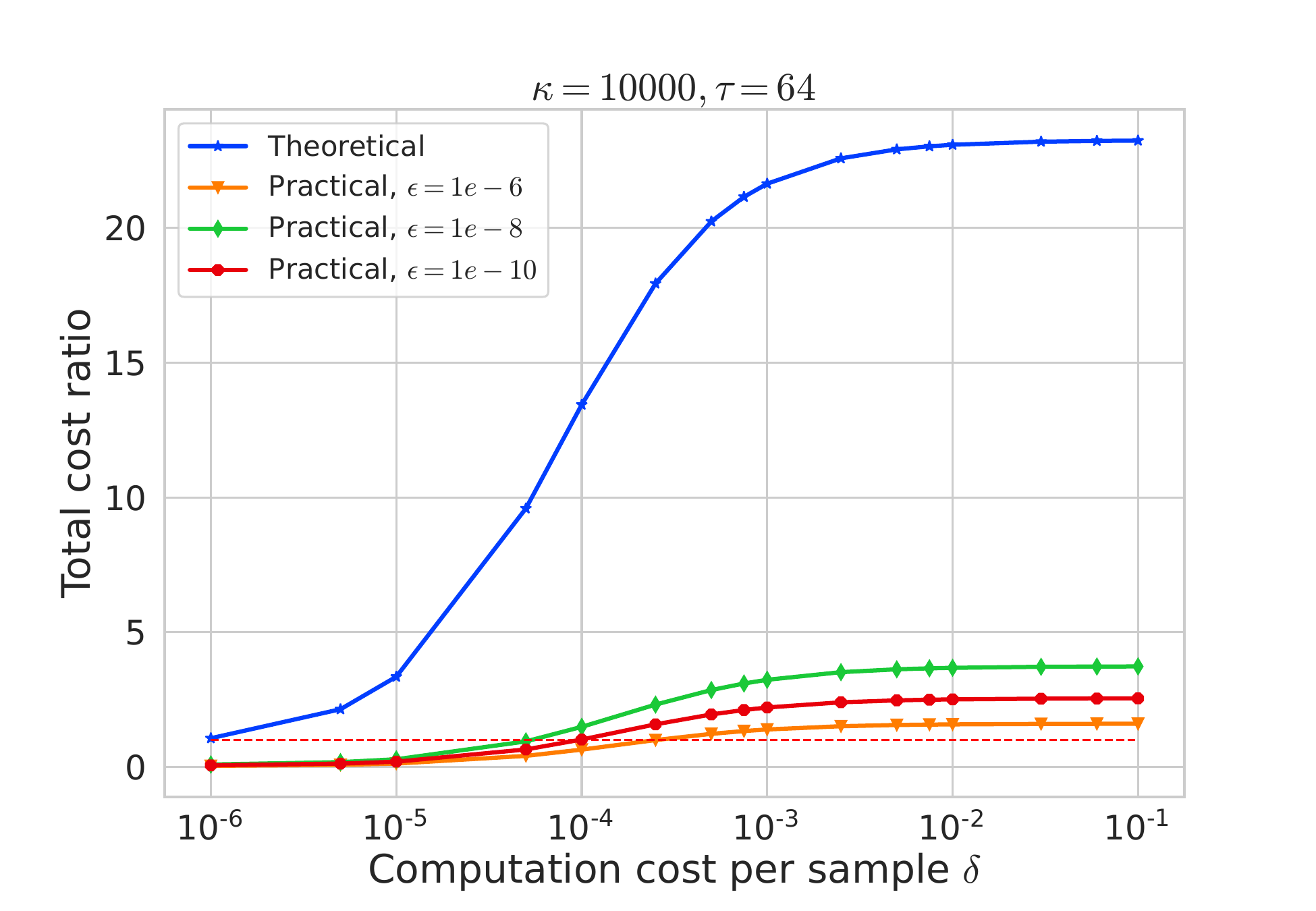}
		\caption{$\tau=64$.}
	\end{subfigure}
	\caption{Acceleration with 10 distributed workers on \textbf{w8a} dataset, $\kappa=1e4$.}
	\label{fig:058}
\end{figure} 

\subsection{Experiments with ProxSkip-HUB}
In this work we introduced a new \gls{FL} architecture: regional hubs connecting the clients to the server; see Section~\ref{sec:tree}.  For conceptual simplicity, and in order to facilitate fair comparison with \gls{ProxSkip-LSVRG}, we assume that the number of hubs equals to  the number of clients, and that each client owns a single datapoint only. We compare \gls{ProxSkip-HUB} with \gls{ProxSkip-LSVRG} to check whether communication compression leads to any benefits in terms of  total costs. Theoretically, and similarly to our analysis in Section~\ref{sec:experiments}, the total cost for \gls{ProxSkip-LSVRG} is 
\begin{align*}
    \text{Cost}(\text{\gls{ProxSkip-LSVRG}}) &\eqdef T_{\text{comm.}}(\text{\gls{ProxSkip-LSVRG}})\\
&+ \delta   \left(q m + (1-q)\tau+\tau\right)T_{\text{iter}}({\text{\gls{ProxSkip-LSVRG}}}).
\end{align*}


Recall that we assume the communication cost from every worker/hub to the master is equal to 1, and the computation cost per sample is equal to $\delta$. Here we generalize to the multi-level structure. We assume that the communication cost from every client to hub is equal to $\delta'$. We choose the Rand-$k$ sparsification for \gls{ProxSkip-HUB}; this compressor selects $k$-entries of the gradient vector, uniformly at random from the full $d$-dimensional gradient. The total cost of \gls{ProxSkip-HUB} is
\begin{equation}
	\begin{aligned}
		\text{Cost}(\text{\gls{ProxSkip-HUB}}) &:= T_{\text{comm.}} (\text{\gls{ProxSkip-HUB}})\\
		&\quad + \delta' \left( q m+  \frac{k}{d} \left( (1-q) \tau+\tau\right)  \right) T_{\text{iter}} (\text{\gls{ProxSkip-HUB}}).\\
	\end{aligned}
\end{equation}

Note that our experimental results are summarized in Figure~\ref{fig:060}; we use the values $\delta = \delta^\prime = 10^{-2}$. Clearly, and thanks to communication compression,   \gls{ProxSkip-HUB} has  benefit in terms of the total cost compared to \gls{ProxSkip-LSVRG}, and can reach up to three orders of magnitude! 

\begin{figure}[!htbp]
	\centering
	\begin{subfigure}[b]{0.32\textwidth}
		\centering
		\includegraphics[trim=20 10 40 40, clip, width=\textwidth]{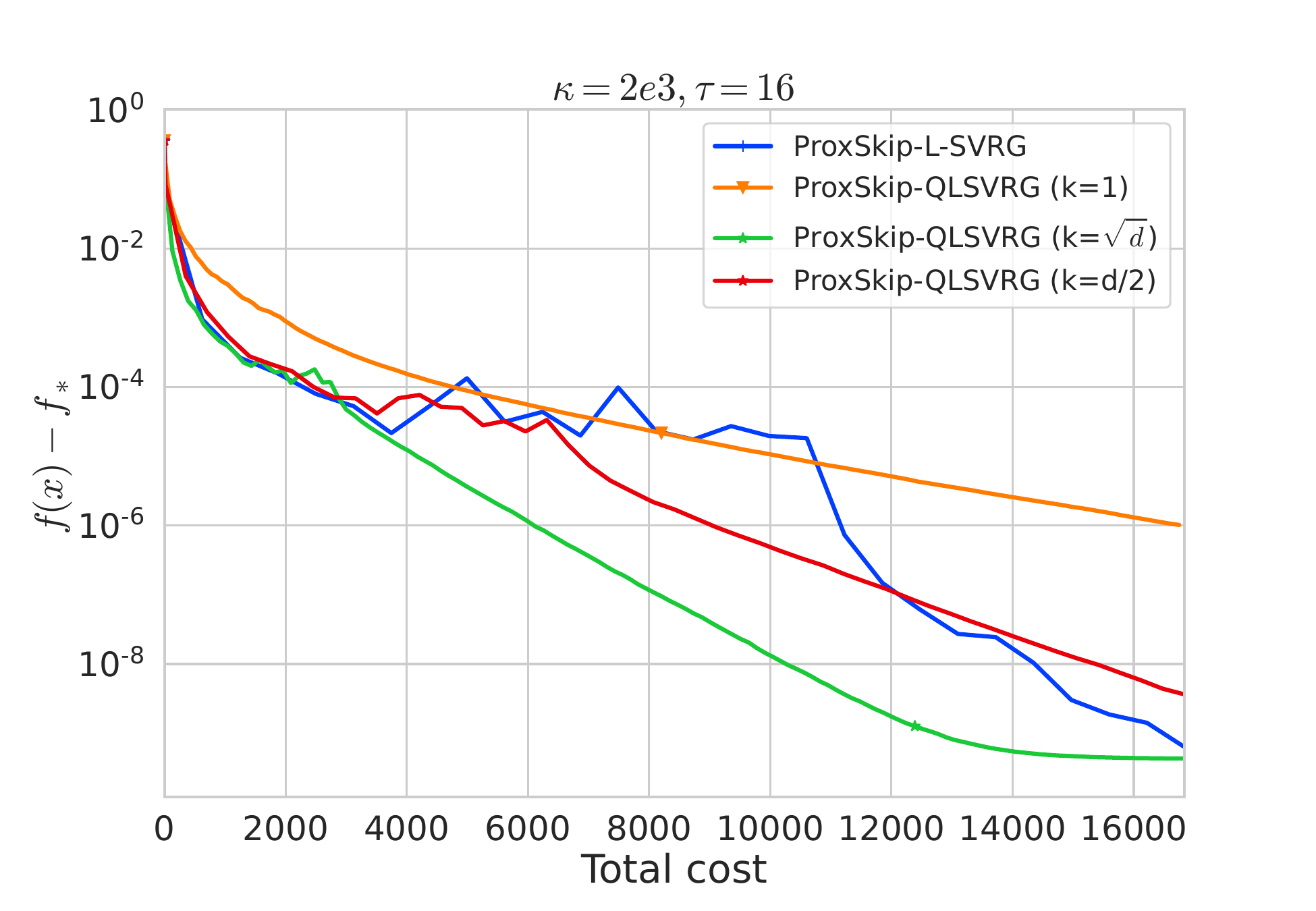}
		\caption{$\tau=16$.}
	\end{subfigure}
	\hfill
	\begin{subfigure}[b]{0.32\textwidth}
		\centering
		\includegraphics[trim=20 10 40 40, clip, width=\textwidth]{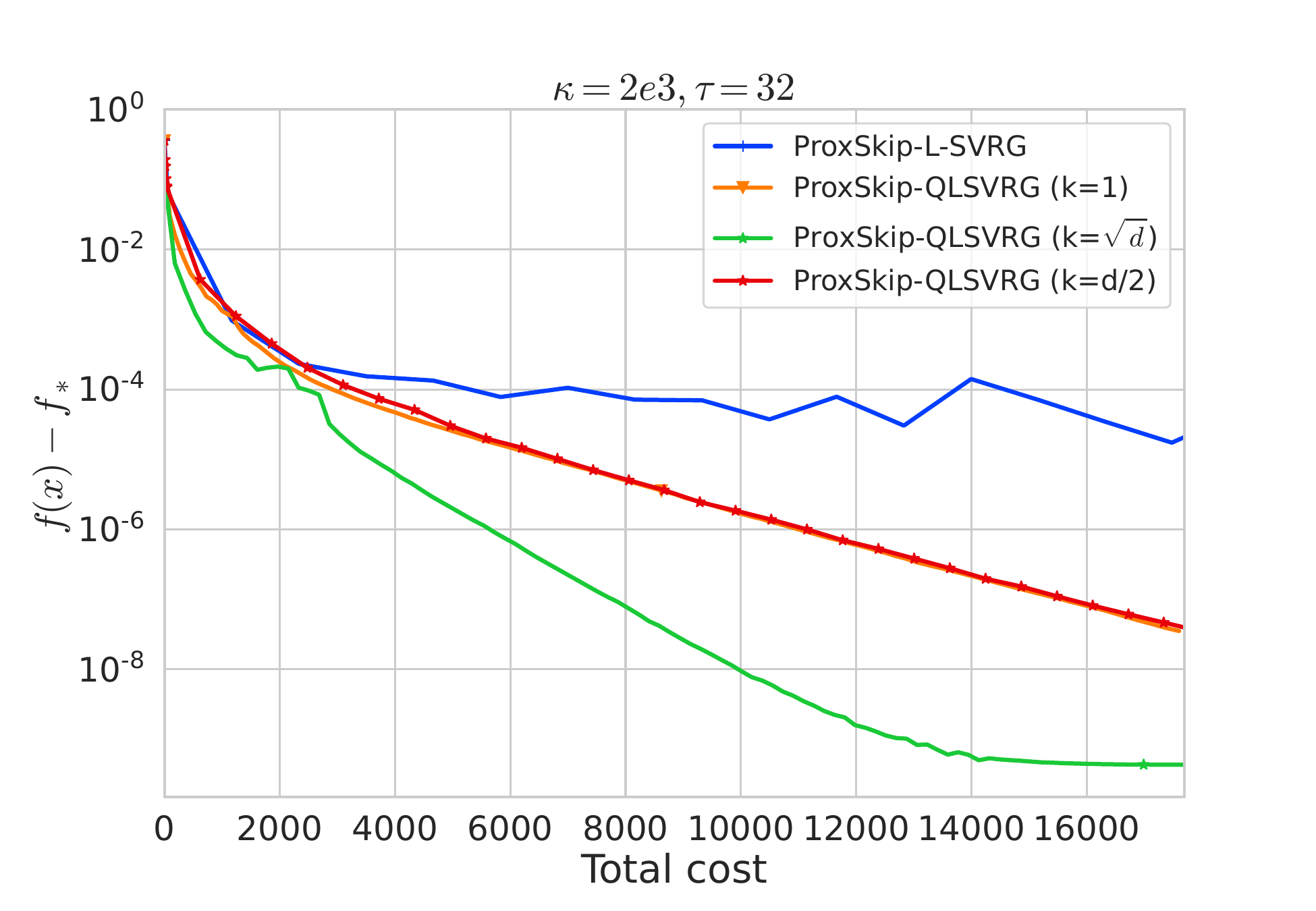}
		\caption{$\tau=32$.}
	\end{subfigure}
	\hfill
	\begin{subfigure}[b]{0.32\textwidth}
		\centering
		\includegraphics[trim=20 10 40 40, clip, width=\textwidth]{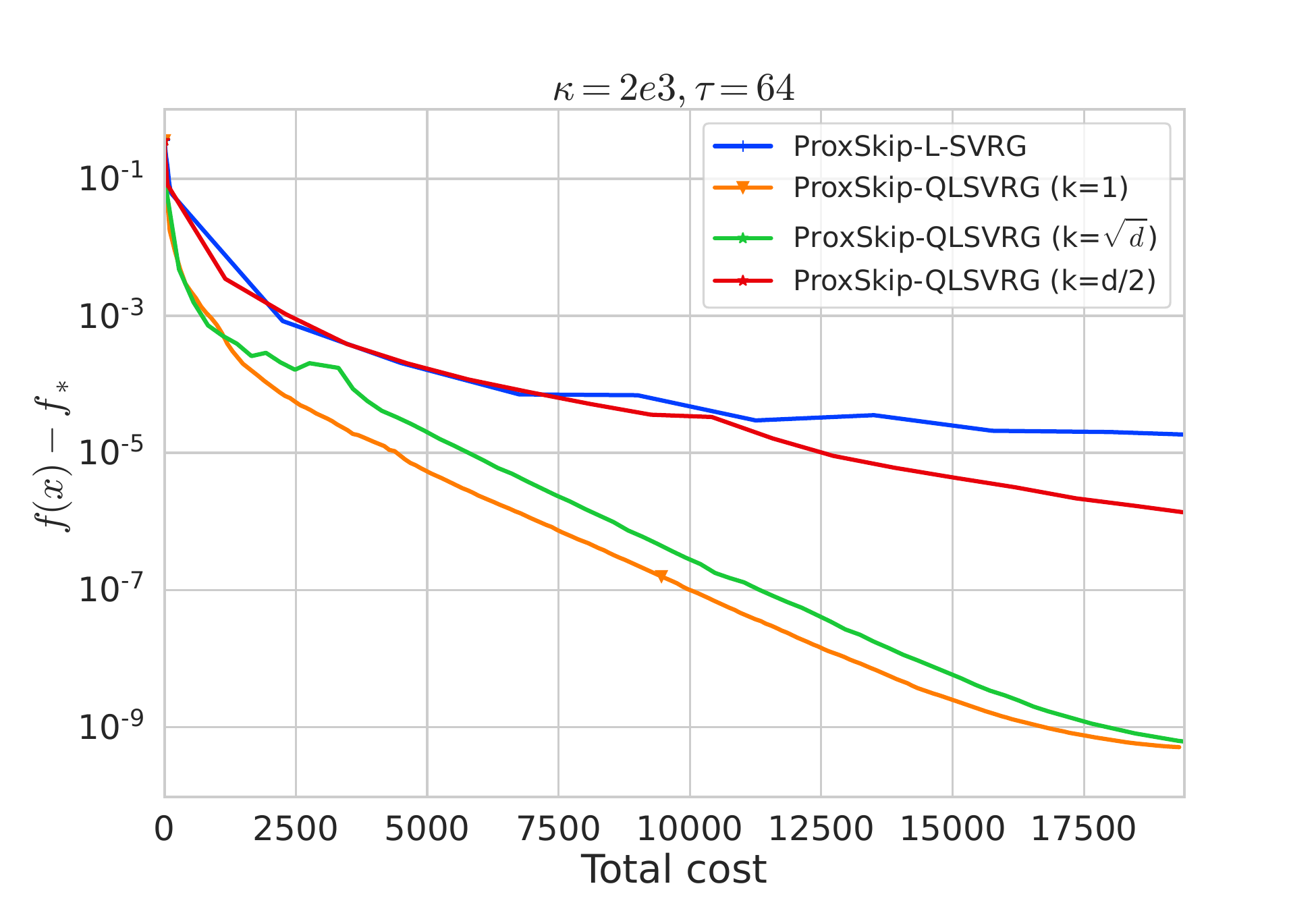}
		\caption{$\tau=64$.}
	\end{subfigure}
	\caption{Convergence results with different batch sizes and sparsification parameter  $k$ on \textbf{a9a}, $\kappa=2e3$.}
	\label{fig:060}
\end{figure} 

\section{Basic Facts}

\subsection{Bregman divergence, Lipschitz smoothness and \texorpdfstring{$\mu$}{mu}-strong convexity}
The Bregman divergence of a differentiable function $f: \mathbb{R}^d \rightarrow \mathbb{R}$ is defined by
\begin{align}
	\label{eq:bregman_divergence}
	D_{f}(x, y):=f(x)-f(y)-\langle\nabla f(y), x-y\rangle.
\end{align}
It is easy to see that 
\begin{align}
	\label{eq:sum_bregman}
\langle\nabla f(x)-\nabla f(y), x-y\rangle=D_{f}(x, y)+D_{f}(y, x), \quad \forall x, y \in \mathbb{R}^{d}.
\end{align}
For an $L$-smooth and $\mu$-strongly convex function $f: \mathbb{R}^d \rightarrow \mathbb{R}$, we have
\begin{align}
	\label{eq:smooth-norms}
\frac{\mu}{2}\|x-y\|^{2} \leq D_{f}(x, y) \leq \frac{L}{2}\|x-y\|^{2}, \quad \forall x, y \in \mathbb{R}^{d}
\end{align}
and
\begin{align}
	\label{eq:smooth-grad}
	\frac{1}{2 L}\|\nabla f(x)-\nabla f(y)\|^{2} \leq D_{f}(x, y) \leq \frac{1}{2 \mu}\|\nabla f(x)-\nabla f(y)\|^{2}, \quad \forall x, y \in \mathbb{R}^{d}.
\end{align}

\subsection{Firm-Nonexpansiveness of the proximity operator}
Given $\psi: \mathbb{R}^{d} \rightarrow \mathbb{R}$, we define $\psi^{*}(y):=\sup _{x \in \mathbb{R}^{d}}\{\langle x, y\rangle-\psi(x)\}$ to be its Fenchel conjugate. The proximity operator of $\psi^{*}$ satisfies for any $\tau>0$
\begin{align}
	\label{eq:fenhel}
	\text{ if } u=\operatorname{prox}_{\tau \psi^{*}}(y), \quad \text{ then } \quad u \in y-\tau \partial \psi^{*}(u).
\end{align}
If Assumption~\ref{ass:Reg} is satisfied, then firm nonexpansiveness of the proximity operator implies~\citep{ProxSkip}  that
\begin{align}
	\label{eq:firm-prox}
\notag	\left\|\operatorname{prox}_{\frac{\gamma}{p} \psi}(x)-\operatorname{prox}_{\frac{\gamma}{p} \psi}(y) \right\|^{2}&+ \left\| \left(x- \operatorname{prox}_{\frac{\gamma}{p} \psi}(x) \right) - \left(y-\operatorname{prox}_{\frac{\gamma}{p} \psi}(y)\right) \right\|^{2}\\
    &\leq\|x-y\|^{2},
\end{align}
for all $x, y \in \mathbb{R}^{d}$ and any $\gamma, p>0$.

\subsection{Young's inequality} For any two vectors $ a, b \in \mathbb{R}^{d}$, we have 
\begin{equation}
	\label{youngs}
	\|a+b\|^{2} \leq 2\|a\|^{2}+2\|b\|^{2}.
\end{equation}

\subsection{Jensen’s inequality} For a convex function $h : \mathbb{R}^d \leftarrow \mathbb{R}$ and any vectors $x_1, \ldots , x_n \in \mathbb{R}^d$, we have 
\begin{align}
	h\left(\frac{1}{n} \sum_{i=1}^{n} x_{i}\right) \leq \frac{1}{n} \sum_{i=1}^{n} h\left(x_{i}\right).
\end{align}
Applying this to the squared norm, $h(x) = \|x\|^2,$ we get
\begin{align}
	\label{jensen}
	\left\|\frac{1}{n} \sum_{i=1}^{n} y_{i}\right\|^{2} \leq \frac{1}{n} \sum_{i=1}^{n}\left\|y_{i}\right\|^{2}.
\end{align}

\clearpage
\section{Analysis of ProxSkip-VR}
In this section we provide the proof of  Theorem~\ref{thm:main_vr_proxskip}.
\subsection{Main Lemma of ProxSkip}
 We start from  Lemma~\ref{lem:main_lemma} initially introduced in~\citep{ProxSkip}; for completeness we provide the whole proof. Let us define two additional sequences:
 \begin{align}
 	\label{eq:seqs}
 \hat{w}^{t}=x^t-\gamma \hat{g}^{t}\left(x^t\right), \qquad  \hat{w}^{\star}=x^\star-\gamma \hat{g}^{t}\left(x^\star\right)  .
 \end{align}
\begin{lemma}
	\label{lem:main_lemma}
	If Assumption~\ref{ass:Reg} holds, $\gamma > 0$ and $0 < p \leq 1$, then the iterates of \gls{ProxSkip-VR} satisfy
	\begin{align}
		\label{eq:lemma1}
		\Exp{\|x^{t+1} - x^\star\|^2+\frac{\gamma^2}{p^2}\|h^{t+1} - h^\star\|^2} \leq\left\|\hat{w}^{t}-w^{\star}\right\|^{2}+\left(1-p^{2}\right) \frac{\gamma^{2}}{p^{2}}\left\|h^t-h^\star\right\|^{2}.
	\end{align}
\end{lemma}
\begin{proof}
	 In order to simplify, let us define two points:
	 \begin{align}
	 	\label{eq:two_points}
	 	x\eqdef\hat{x}^{t+1}-\frac{\gamma}{p} h^t, \quad y\eqdef x^\star-\frac{\gamma}{p} h^\star.
	 \end{align}
 \paragraph{STEP 1 (Optimality conditions).} Using the first-order optimality conditions for $f + \psi$ and using $h^\star \eqdef \nabla f (x^\star)$, we obtain the following fixed-point identity for $x^\star$:
 \begin{align}
 	\label{eq:opt}
 	x^\star=\operatorname{prox}_{\frac{\gamma}{p} \psi}\left(x^\star-\frac{\gamma}{p} h^\star\right)=	\operatorname{prox}_{\frac{\gamma}{p} \psi}(y) .
 \end{align}

\paragraph{STEP 2 (Recalling the steps of the method).}
Recall that the vectors $x^t$ and $h^t$ are in Algorithm~\ref{alg:ProxSkip-VR} updated as follows:
\begin{align}
 	\label{eq:x_upd}
	x^{t+1}=\left\{\begin{array}{lll}
		\operatorname{prox}_{\frac{\gamma}{p} \psi}(x) & \text { with probability } & p \\
		\hat{x}^{t+1} & \text { with probability } & 1-p
	\end{array}\right.
\end{align}
and 
\begin{align}
	\notag	 	\label{eq:h_upd}
	h^{t+1}&=h^t+\frac{p}{\gamma}\left(x^{t+1}-\hat{x}^{t+1}\right)\\
    &= \left\{\begin{array}{lll}
		h^t+\frac{p}{\gamma}\left(	\operatorname{prox}_{\frac{\gamma}{p} \psi}(x) -\hat{x}^{t+1}\right) & \text { with probability } p \\
		h^t & \text { with probability } 1-p
	\end{array}\right..
\end{align}
\paragraph{STEP 3 (One-step expectation of the Lyapunov function).}
Let us consider the expected value $V^{(t+1)}\eqdef \Exp{\left\|x^{t+1}-x^\star\right\|^{2}+\frac{\gamma^{2}}{p^{2}}\left\|h^{t+1}-h^\star\right\|^{2}}$:
\begin{align}
\notag	&V^{(t+1)}\\
    & \stackrel{(\ref{eq:x_upd})+(\ref{eq:h_upd})}{=} p\left(\left\|	\operatorname{prox}_{\frac{\gamma}{p} \psi}(x)-x^\star\right\|^{2}+\frac{\gamma^{2}}{p^{2}}\left\|h^t+\frac{p}{\gamma}\left(	\operatorname{prox}_{\frac{\gamma}{p} \psi}(x)-\hat{x}^{t+1}\right)-h^\star\right\|^{2}\right) \notag \\
	& +(1-p)\left(\left\|\hat{x}^{t+1}-x^\star\right\|^{2}+\frac{\gamma^{2}}{p^{2}}\left\|h^t-h^\star\right\|^{2}\right)\notag \\
	&\stackrel{(\ref{eq:opt})}{=} p\left( \left\|	\operatorname{prox}_{\frac{\gamma}{p} \psi}(x)-	\operatorname{prox}_{\frac{\gamma}{p} \psi}(y) \right\|^{2}+\left\|\frac{\gamma}{p} h^t+	\operatorname{prox}_{\frac{\gamma}{p} \psi}(x)-\hat{x}^{t+1}-\frac{\gamma}{p} h^\star\right\|^{2}\right)\notag \\
	& +(1-p)\left(\left\|\hat{x}^{t+1}-x^\star\right\|^{2}+\frac{\gamma^{2}}{p^{2}}\left\|h^t-h^\star\right\|^{2}\right)\notag \\
	&\stackrel{(\ref{eq:two_points})+(\ref{eq:opt})}{=} p\left\|	\operatorname{prox}_{\frac{\gamma}{p} \psi}(x)-	\operatorname{prox}_{\frac{\gamma}{p} \psi}(y)\right\|^{2}\notag\\
    &+p\left\|	\operatorname{prox}_{\frac{\gamma}{p} \psi}(x)-x+y-	\operatorname{prox}_{\frac{\gamma}{p} \psi}(y)\right\|^{2}\notag \\
	& +(1-p)\left(\left\|\hat{x}^{t+1}-x^\star\right\|^{2}+\frac{\gamma^{2}}{p^{2}}\left\|h^t-h^\star\right\|^{2}\right).\label{eq:9u0fd9h0fd9h}
\end{align}
\paragraph{STEP 4 (Applying firm nonexpansiveness).}Applying firm nonexpansiveness of the proximal operator~\eqref{eq:firm-prox}, this leads to the inequality
\begin{eqnarray*}
V^{(t+1)}& \stackrel{\eqref{eq:9u0fd9h0fd9h}+(\ref{eq:firm-prox})}{\leq} & p\|x-y\|^{2} +(1-p)\left(\left\|\hat{x}^{t+1}-x^\star\right\|^{2}+\frac{\gamma^{2}}{p^{2}}\left\|h^t-h^\star\right\|^{2}\right) \\ \notag
	& \stackrel{(\ref{eq:two_points})}{=}&  p\left\|\hat{x}^{t+1}-\frac{\gamma}{p} h^t-\left(x^\star-\frac{\gamma}{p} h^\star\right)\right\|^{2}\\
    &+&(1-p)\left(\left\|\hat{x}^{t+1}-x^\star\right\|^{2}+\frac{\gamma^{2}}{p^{2}}\left\|h^t-h^\star\right\|^{2}\right) .
\end{eqnarray*}
\paragraph{STEP 5 (Simple Algebra).} Next, we expand the squared norm and collect the terms, obtaining
\begin{eqnarray}
V^{(t+1)}& \leq & p\left\|\hat{x}^{t+1}-x^{\star}\right\|^{2} +p \frac{\gamma^{2}}{p^{2}}\left\|h^t-h^\star\right\|^{2}-2 \gamma\left\langle\hat{x}^{t+1}-x^{\star}, h^t-h^\star\right\rangle \notag\\
&&\quad +(1-p)\left(\left\|\hat{x}^{t+1}-x^{\star}\right\|^{2}  +\frac{\gamma^{2}}{p^{2}}\left\|h^t-h^\star\right\|^{2}\right) \notag \\
	&=& \left\|\hat{x}^{t+1}-x^{\star}\right\|^{2}-2 \gamma\left\langle\hat{x}^{t+1}-x^{\star}, h^{t}-h^\star\right\rangle +\frac{\gamma^{2}}{p^{2}}\left\|h^{t}-h^\star\right\|^{2}.\label{eq:last1}	
\end{eqnarray}
Finally, note that by our definition of $\hat{w}^t$, we have the identity $\hat{x}^{t+1}=\hat{w}^{t}+\gamma h^{t}$. Therefore, the first two terms above can be rewritten as
\begin{align}
		\label{eq:last2}
\notag	\left\|\hat{x}^{t+1}-x^{\star}\right\|^{2}&-2 \gamma\left\langle\hat{x}^{t+1}-x^{\star}, h^{t}-h^\star\right\rangle\\
\notag&=\left\|\hat{w}^{t}-w^{\star}+\gamma\left(h^{t}-h^\star\right)\right\|^{2}\\
\notag	&\quad -2 \gamma\left\langle \hat{w}^{t}-w^{\star}+\gamma\left(h^{t}-h^\star\right), h^{t}-h^\star\right\rangle \\
\notag	&=\left\|\hat{w}^{t}-w^{\star}\right\|^{2}+2 \gamma\left\langle \hat{w}^{t}-w^{\star}, h^{t}-h^\star\right\rangle\\
\notag	&\quad +\gamma^{2}\left\|h^{t}-h^\star\right\|^{2} 
	-2 \gamma\left\langle \hat{w}^{t}-w^{\star}, h^{t}-h^\star\right\rangle\\
\notag	&\quad -2 \gamma^{2}\left\|h^{t}-h^\star\right\|^{2} \\
	&=\left\|\hat{w}^{t}-w^{\star}\right\|^{2}-\gamma^{2}\left\|h^{t}-h^\star\right\|^{2}.
\end{align}
Finally, plugging \eqref{eq:last2} into \eqref{eq:last1}, we get:
	\begin{align*}
	V^{(t+1)} \leq\left\|\hat{w}^{t}-w^{\star}\right\|^{2}+\left(1-p^{2}\right) \frac{\gamma^{2}}{p^{2}}\left\|h^{t}-h^\star\right\|^{2}.
\end{align*}
\end{proof}
\subsection{Main Lemma}
This lemma allows us to obtain a useful recursion for variance-reduced stochastic estimators used in our \gls{ProxSkip-VR} algorithm. 
\begin{lemma}
	Let Assumptions~\ref{ass:mu-strongly-convex} and \ref{sigma_t} hold. Then the iterates of \gls{ProxSkip-VR} satisfy 
	\begin{align*}
\notag	\Exp{\|\hat{w}^t - w^\star\|^2} & \leq  (1-\gamma\mu)\|x^t - x^\star\|^2\\
&- 2\gamma D_f(x^t,x^\star)\left(1-\gamma A\right) + \gamma^2B\sigma^{(t)} + \gamma^2C.
	\end{align*}
\end{lemma}
\begin{proof}
	We start from the definitions of the auxiliary sequence $\hat{w}^t$ (see \eqref{eq:seqs}):
	\begin{eqnarray}
		\label{32}
\notag		\|\hat{w}^t - w^\star\|^2 &\overset{\eqref{eq:seqs}}{=}& \| x^t - \gamma\hat{g}^t - (x^\star - \gamma\nabla f(x^\star) ) \|^2\\
	\notag	&=&\|  (x^t - x^\star ) - \gamma(\hat{g}^t - \nabla f(x^\star))  \|^2\\
\notag			& =& \|x^t - x^\star\|^2 - 2\gamma\left\langle x^t - x^\star, \hat{g}^t - \nabla f(x^\star) \right\rangle\\
            &+& \gamma^2 \| \hat{g}^t - \nabla f(x^\star) \|^2.
	\end{eqnarray}
Taking expectation in~\eqref{32} and using unbiasedness of $\hat{g}^t$ (see \eqref{eq:unbiased} in Assumption~\ref{sigma_t}), we get 
\begin{align}
	\label{33}
\notag	\Exp{\|\hat{w}^t - w^\star\|^2}  &\overset{\eqref{32}+\eqref{eq:unbiased}}{=} \|x^t - x^\star\|^2 - 2\gamma\left\langle x^t - x^\star, \nabla f(x^t) - \nabla f(x^\star) \right\rangle\\
    &+ \gamma^2 \Exp{\| \hat{g}^t - \nabla f(x^\star) \|^2}.
\end{align}
Let us now consider the inner product in~\eqref{33}. Using~\eqref{eq:smooth-norms} and~\eqref{eq:sum_bregman}, we obtain
\begin{align}
		\label{34}
	\Exp{\|\hat{w}^t - w^\star\|^2}  &\leq (1-\gamma\mu)\|x^t - x^\star\|^2 - 2\gamma D_f(x^t,x^\star)\notag\\
    &+ \gamma^2 \Exp{\| \hat{g}^t - \nabla f(x^\star) \|^2}.
\end{align}
To bound the last term in \eqref{34}, we can apply Assumption~\ref{sigma_t}:
\begin{align}
	\label{35}
	\Exp{\| \hat{g}^t - \nabla f(x^\star) \|^2} \leq 2AD_f(x^t,x^\star)+B\sigma^{(t)} + C.
\end{align}
Plugging~\eqref{35} into~\eqref{34} gives us
\begin{align}
	\label{eq:le2}
\notag	\Exp{\|\hat{w}^t - w^\star\|^2}  &\leq (1-\gamma\mu)\|x^t - x^\star\|^2\\
&- 2\gamma D_f(x^t,x^\star) + \gamma^2 \left(2AD_f(x^t,x^\star)+B\sigma^{(t)} + C\right)\notag\\
	& \leq  (1-\gamma\mu)\|x^t - x^\star\|^2\notag\\
    &- 2\gamma D_f(x^t,x^\star)\left(1-\gamma A\right) + \gamma^2B\sigma^{(t)} + \gamma^2C,
\end{align}
which is what we set out to prove.
\end{proof}

\subsection{Proof of Theorem~\ref{thm:main_vr_proxskip}}


\begin{proof}
	Using definition of the Lyapunov function $\Psi^{(t)}$, and the tower property of conditional expectation, we obtain
	\begin{align}
\notag		\Exp{\Psi^{(t+1)}} &= 	\Exp{\|x^{t+1} - x^{\star}\|^2+\frac{\gamma^2}{p^2}\|h^{t+1} - h^\star\|^2+\MM\gamma^2\sigma^{(t+1)}} \notag \\
\notag		 &\stackrel{\eqref{eq:lemma1}}{\leq}  \Exp{\|\hat{w}^t - w^\star \|^2} + (1-p^2)\frac{\gamma^2}{p^2}\|h^t - h^\star\|^2 \notag  \\
	&  +\MM \gamma^2 \left( 2\tilde{A}D_f(x^t,x^\star) + \tilde{B}\sigma^{(t)} + \tilde{C} \right) \notag \\
		&\stackrel{\eqref{eq:le2}}{\leq}		(1-\gamma\mu)\|x^t - x^\star\|^2 - 2\gamma D_f(x^t,x^\star)\left(1-\gamma A\right) + \gamma^2B\sigma^{(t)} + \gamma^2C	\notag  \\
			&  + (1-p^2)\frac{\gamma^2}{p^2}\|h^t - h^\star\|^2 +\MM \gamma^2 \left( 2\tilde{A}D_f(x^t,x^\star) + \tilde{B}\sigma^{(t)} + \tilde{C} \right) \notag \\
		&\leq 	(1-\gamma\mu)\|x^t - x^\star\|^2 - 2\gamma D_f(x^t,x^\star)\left(1-\gamma (A+\MM\tilde{A})\right) \notag \\
	\notag		&  + \gamma^2\MM \sigma^{(t)}\left(\frac{B+\MM \tilde{B}}{\MM}\right) + \gamma^2(C+\MM\tilde{C})\\
            &+(1-p^2)\frac{\gamma^2}{p^2}\|h^t - h^\star\|^2 .
	\end{align}

Using the stepsize bound $\gamma \leq \frac{1}{A+\MM\tilde{A}}$, this leads to 
	\begin{align}
	\notag		\Exp{\Psi^{(t+1)}} 	&\leq 	(1-\gamma\mu)\|x^t - x^\star\|^2  + \gamma^2\MM \sigma^{(t)}\left(\frac{B+\MM \tilde{B}}{\MM}\right)\\
    &+ \gamma^2(C+\MM\tilde{C})+(1-p^2)\frac{\gamma^2}{p^2}\|h^t - h^\star\|^2.
\end{align}

Let us denote $\beta \eqdef \frac{B+\MM \tilde{B}}{\MM}$. In order to obtain a contraction, we need to have $\beta< 1$, which is satisfied when  $ \MM > \frac{B}{1-\tilde{B}}$, and we get
	\begin{align}
		\notag\Exp{\Psi^{(t+1)}} 	&\leq  	(1-\gamma\mu)\|x^t - x^\star\|^2\\
        &+ \gamma^2\MM \sigma^{(t)}\beta 			+ \gamma^2(C+\MM\tilde{C})+(1-p^2)\frac{\gamma^2}{p^2}\|h^t - h^\star\|^2 \notag\\
	&\leq  \max\left( 1-p^2,\beta,1-\gamma\mu \right)\Psi^{(t)} +  \gamma^2(C+\MM\tilde{C}) \label{eq:final} .
\end{align}

Finally, using the tower property of expectation and unrolling our recursion~\eqref{eq:final}, we get 
	\begin{equation*}
	\Exp{ \Psi^{(T)} } \leq \max \left\{(1-\gamma \mu)^{T},\beta^{T},(1-p^2)^T\right\} \Psi^{(0)}+\frac{\left(C+\MM \tilde{C}\right) \gamma^{2}}{\min \left\{\gamma \mu,p^2, 1 - \beta \right\}}.
\end{equation*}

\end{proof}

\clearpage
\section{Examples of Methods Without Variance Reduction}
\subsection{Proof of Theorem~\ref{thm:proxskip} (GD estimator)}
\label{sec:GD_est}


\begin{proof}
	Let us show that \algname{\gls{GD}} estimator $(\hat{g}^t = \nabla f(x^t))$ satisfies Assumption~\ref{sigma_t}
	\begin{align}
	\notag	\Exp{\|\hat{g}^t - \nabla f(x^\star)\|^2} = \|\nabla f(x^t)- \nabla f(x^\star)\|^2 \stackrel{\eqref{eq:smooth-norms}}{\leq} 2 L D_f(x^t,x^\star).
	\end{align}
This means that Assumption~\ref{sigma_t} is satisfied with the following constant:
\begin{align*}
	A = L, \quad B = 0, \quad C = 0, \quad \tilde{A} = 0, \quad \tilde{B} = 0,  \quad \tilde{C} = 0, \quad \sigma_t \equiv 0.
\end{align*}
Applying Theorem~\ref{thm:main_vr_proxskip} leads to final recursion:
 \begin{equation}
 	\label{eq:req_gd}
 	\Exp{\Psi^{(T)}} \leq \max \left\{(1-\gamma \mu)^{T},(1-p^2)^T\right\} \Psi^{(0)},
 \end{equation}
By inspecting \eqref{eq:req_gd} it is easy to see that
 \begin{equation}
T \geq \max \left\{\frac{1}{\gamma \mu}, \frac{1}{p^{2}}\right\} \log \frac{1}{\varepsilon} \qquad \Longrightarrow \qquad\Exp{\Psi^{(T)}} \leq \varepsilon \Psi^{(0)}.
 \end{equation}
Then the communication complexity is equal to
\begin{equation}
	p T \geq \max \left\{\frac{p}{\gamma \mu}, \frac{1}{p}\right\} \log \frac{1}{\varepsilon} .
\end{equation}
Setting $\gamma = \frac{1}{L}$ and solving $\frac{p L}{\mu}=\frac{1}{p}$ gives the optimal probability
\begin{equation}
	p=\sqrt{\frac{\mu}{L}}=\frac{1}{\sqrt{\kappa}}
\end{equation}
Finally, the iteration complexity and communication complexity have the following form:
\begin{align}
	T &\geq \max \left\{\frac{1}{\gamma \mu}, \frac{1}{p^{2}}\right\} \log \frac{1}{\varepsilon} = \kappa \log \frac{1}{\varepsilon},\\
	p T &\geq \max \left\{\frac{p}{\gamma \mu}, \frac{1}{p}\right\} \log \frac{1}{\varepsilon} = \sqrt{\kappa} \log \frac{1}{\varepsilon}.
\end{align}

\end{proof}
This recovers the result obtained in~\citep{ProxSkip}.

\subsection{Proof of Theorem~\ref{thm:sproxskip} (SGD estimator)}

 \begin{proof}
 		Let us show that the \algname{SGD} estimator $\hat{g}^t = g(x^t,\xi_t)$ satisfying Assumption~\ref{Expected_smoothness} also satisfies Assumption~\ref{sigma_t}. Using Young's inequality we get 
 		\begin{eqnarray}
 \notag			\Exp{\|\hat{g}^t - \nabla f(x^\star)\|^2} &=& \Exp{\|g(x^t,\xi_t)- \nabla f(x^\star)\|^2} \\
 	 \notag			& =&  \Exp{\|g(x^t,\xi_t)- g(x^\star,\xi_t) + g(x^\star,\xi_t) -\nabla f(x^\star)\|^2}\\
 		 \notag		&\stackrel{\eqref{youngs}}{\leq}& 2\Exp{\|g(x^t,\xi_t)- g(x^\star,\xi_t)\|^2}\\
       \notag  &+& 2\Exp{g(x^\star,\xi_t) -\nabla f(x^\star)\|^2}\\
 			& \stackrel{\eqref{Expected_smoothness}}{\leq}& 4 A^{\prime\prime} D_f(x^t,x^\star) + 2 {\rm Var}(g(x^\star,\xi)) .
 		\end{eqnarray}
 	This means that Assumption~\ref{sigma_t} is satisfied with the following constants:
 	 \begin{align*}
 		A = 2A^{\prime\prime}, \quad B = 0, \quad C = 2{\rm Var}(g(x^\star,\xi)), \quad \tilde{A} = 0, \quad \tilde{B} = 0,  \quad \tilde{C} = 0, \quad \sigma_t \equiv 0.
 	\end{align*}
 Applying Theorem~\ref{thm:main_vr_proxskip} leads to the final bound:
  \begin{equation}
  	\label{sgd_final}
 	\Exp{\Psi^{(T)}} \leq \max \left\{(1-\gamma \mu)^{T},(1-p^2)^T\right\} \Psi^{(0)} + \gamma^2 \frac{2{\rm Var}(g(x^\star,\xi))}{\min\left\lbrace \gamma\mu,p^2 \right\rbrace}.
 \end{equation}
In order to minimize the number of prox evaluations, whatever the choice of $\gamma$ will be, we choose the smallest probability $p$ which does not lead to any degradation of the rate $\min\{\gamma\mu, p^2\}.$ That is, we choose $p = \sqrt{\gamma \mu}$. The first term on the right-hand side of~\eqref{sgd_final} can be bounded as follows:
\begin{equation*}
T \geq \frac{1}{\gamma \mu} \log \left(\frac{2 \Psi^{(0)}}{\varepsilon}\right) \quad \Longrightarrow \quad(1-\gamma\mu)^{T} \Psi^{(0)} \leq \frac{\varepsilon}{2}.
\end{equation*}
The second term on the right-hand side of~\eqref{sgd_final} can be bounded as follows:
\begin{equation*}
	\gamma \leq \frac{\varepsilon \mu}{2 C} \quad \Longrightarrow \quad \frac{\gamma C}{\mu} \leq \frac{\varepsilon}{2}.
\end{equation*}
We choose the largest stepsize consistent with bounds $	\gamma \leq \frac{\varepsilon \mu}{2 C}$ and $\gamma\leq \frac{1}{A}$:
\begin{align*}
	\gamma=\min \left\{\frac{1}{A}, \frac{\varepsilon \mu}{2 C}\right\}.
\end{align*}
Using this stepsize, we get the following iteration and (expected) communication complexities: 
\begin{align*}
	T \geq \max \left\{\frac{A}{\mu}, \frac{2 C}{\varepsilon \mu^{2}}\right\} \log \left(\frac{2 \Psi^{(0)}}{\varepsilon}\right),\qquad  	pT \geq \max \left\{\sqrt{\frac{A}{\mu}}, \sqrt{\frac{2 C}{\varepsilon \mu^{2}}}\right\} \log \left(\frac{2 \Psi^{(0)}}{\varepsilon}\right).
\end{align*}
This recovers the result obtained in~\citep{ProxSkip}.
 \end{proof}

 \clearpage
\section{Analysis of ProxSkip-HUB}

In this section we provide analysis of the new algorithm \gls{ProxSkip-HUB}, which works for the new \gls{FL} formulation described in Section~\ref{sec:tree}. The pseudocode is presented in Algorithm~\ref{alg:ProxSkip-HUB}.

\subsection{Lemma for minibatch sampling}
Fix a batch size $\tau \in \{1,2,\ldots,n\}$ and let $\set$ be a random subset of
$\{1,2,\ldots,n\}$ of size $\tau$, chosen uniformly at random. Define the gradient estimator via 
\begin{equation}
	\label{nice_est}
	g(x) \eqdef \frac{1}{\tau} \sum_{j \in \set} \nabla \widetilde{\phi}_{j}(x)
\end{equation}
\begin{lemma}
	\label{tau-nice}
	The gradient estimator $g(x)$ defined in~\eqref{nice_est} is unbiased. If we further assume that $n \geq 2$, $\widetilde{\phi}_j$ is convex and $L_j$-smooth for all $j$, and $f$ is $L$-smooth, then
	\begin{align*}
		\Exp{\|g(x^t)-g(x^\star)\|^{2}} \leq 2 L(\tau)D_{f}(x^t, x^\star),
	\end{align*}
	where 
	\begin{align*}
		L(\tau)\eqdef \frac{n-\tau}{\tau(n-1)} \max _{j} L_{j}+\frac{n(\tau-1)}{\tau(n-1)} L.
	\end{align*}
\end{lemma}
\begin{proof}
	Let $\chi_j$ be the random variable defined by
	\begin{align*}
		\chi_{j}= \begin{cases}1 & j \in S \\ 0 & j \notin S\end{cases}.
	\end{align*}
	It is easy to show that 
	\begin{align}
		\label{prob}
		\Exp{\chi_{j}}=\operatorname{Prob}(j \in S)=\frac{\tau}{n}.
	\end{align}
	Unbiasedness of $g(x)$ now follows via direct computation:
	\begin{eqnarray*}
		\notag	\Exp{g(x)} & \stackrel{\eqref{nice_est}}{=} & \Exp{\frac{1}{\tau} \sum_{j \in S} \nabla \widetilde{\phi}_{j}(x)}\\
     \notag   &=& \Exp{\frac{1}{\tau} \sum_{j=1}^{n} \chi_{i} \nabla \widetilde{\phi}_{j}(x)}\\
    \notag    &=&\frac{1}{\tau} \sum_{j=1}^{n} \Exp{\chi_{i}}\nabla \widetilde{\phi}_{j}(x) \\
		\notag	&=& \frac{1}{\tau} \sum_{j=1}^{n} \operatorname{Prob}(j \in S) \nabla \widetilde{\phi}_{j}(x) \\
        &\stackrel{\eqref{prob}}{=}& \frac{1}{\tau} \sum_{j=1}^{n} \frac{\tau}{n} \nabla \widetilde{\phi}_{j}(x) = \nabla f(x).
	\end{eqnarray*}
	Let us define 
	\begin{equation}
		\label{a_j}
		a_{j} \eqdef \nabla \widetilde{\phi}_{j}(x)-\nabla \widetilde{\phi}_{j}(x^\star).
	\end{equation}
	Let $\chi_{j,k}$ be the random variable defined by
	\begin{align*}
		\chi_{j,k}= \begin{cases}1 & j \in \set \text { and } k \in \set \\ 0 & \text { otherwise }\end{cases}.
	\end{align*}
	Note that
	\begin{equation}
		\label{indet}
		\chi_{j,k}=\chi_{j} \chi_{k}.
	\end{equation}
	Further, it is easy to show that
	\begin{equation}
		\label{proba-2}
		\Exp{\chi_{j,k}}=\operatorname{Prob}(j \in \set, k\in \set)=\frac{\tau(\tau-1)}{n(n-1)}.
	\end{equation}
	Let us consider 
	\begin{eqnarray}
		\notag		\Exp{\|g(x^t)-g(x^\star)\|^{2}}
       \notag &=& 	\Exp{	\left\|\frac{1}{\tau} \sum_{j \in \set} \nabla \widetilde{\phi}_{j}(x)-\frac{1}{\tau} \sum_{j \in \set} \nabla \widetilde{\phi}_{j}(x^\star)\right\|^{2}}\\
		\notag	&=&	\Exp{	\left\|\frac{1}{\tau} \sum_{j \in \set} \left(\nabla \widetilde{\phi}_{j}(x)- \nabla \widetilde{\phi}_{j}(x^\star)\right)\right\|^{2}}\\
		\notag &\stackrel{\eqref{a_j}}{=}& \Exp{	\left\|\frac{1}{\tau} \sum_{j \in \set} a_j\right\|^{2}}\\
		\notag & =& \frac{1}{\tau^{2}} \Exp{\left\|\sum_{j=1}^{n} \chi_{j} a_{j}\right\|^{2}}\\
		\notag&=&	\frac{1}{\tau^{2}} \Exp{\sum_{j=1}^{n}\left\|\chi_{j} a_{j}\right\|^{2}+\sum_{k \neq j}\left\langle\chi_{j} a_{j}, \chi_{k} a_{k}\right\rangle}\\
		&\stackrel{\eqref{indet}}{=}&\frac{1}{\tau^{2}} \Exp{\sum_{j=1}^{n}\left\|\chi_{j} a_{j}\right\|^{2}+\sum_{k \neq j} \chi_{j,k}\left\langle a_{j}, a_{k}\right\rangle}.
	\end{eqnarray}
	Using the formulas \eqref{prob} and \eqref{proba-2} we can continue:
	\begin{align}
		\notag	\Exp{\|g(x^t)-g(x^\star)\|^{2}} &= \frac{1}{\tau^{2}}\left(\frac{\tau}{n} \sum_{j=1}^{n}\left\|a_{j}\right\|^{2}+\frac{\tau(\tau-1)}{n(n-1)} \sum_{j \neq k}\left\langle a_{j}, a_{k}\right\rangle\right)\\
		\notag	&=\frac{1}{\tau n} \sum_{j=1}^{n}\left\|a_{j}\right\|^{2}+\frac{\tau-1}{\tau n(n-1)} \sum_{j \neq k}\left\langle a_{j}, a_{k}\right\rangle\\
		\notag	&=\frac{1}{\tau n} \sum_{j=1}^{n}\left\|a_{j}\right\|^{2}+\frac{\tau-1}{\tau n(n-1)}\left(\left\|\sum_{j=1}^{n} a_{j}\right\|^{2}-\sum_{j=1}^{n}\left\|a_{j}\right\|^{2}\right)\\
		&=\frac{n-\tau}{\tau(n-1)} \frac{1}{n} \sum_{j=1}^{n}\left\|a_{j}\right\|^{2}+\frac{n(\tau-1)}{\tau(n-1)}\left\|\frac{1}{n} \sum_{j=1}^{n} a_{j}\right\|^{2}.
	\end{align}
	Since $\widetilde{\phi}_j$ is convex and $L_j$-smooth, we know that
	\begin{equation}
		\left\|a_{j}\right\|^{2} =\left\|\nabla f_{j}(x^t)-\nabla f_{j}(x^\star)\right\|^{2} \leq 2 L_{j} D_{\widetilde{\phi}_{j}}(x^t, x^\star).
	\end{equation}
	Since $f$ is convex and $L$-smooth, we know that
	\begin{align}
		\left\|\frac{1}{n} \sum_{i=1}^{n} a_{i}\right\|^{2} =\|\nabla f(x^t)-\nabla f(x^\star)\|^{2} \leq 2 L D_{f}(x^t, x^\star).
	\end{align}
	Let us apply the bound $L_j \leq \max_j L_j$ and use the following identity $D_f(x^t, x^\star) = \frac{1}{n}\sum_{j=1}^{n} D_{\widetilde{\phi}_j}(x^t, x^\star):$
	\begin{align}
		\notag	&\Exp{	\left\|\frac{1}{\tau} \sum_{j \in \set} \nabla \widetilde{\phi}_{j}(x)-\frac{1}{\tau} \sum_{j \in \set} \nabla \widetilde{\phi}_{j}(x^\star)\right\|^{2}}\\
        \notag&\leq \frac{n-\tau}{\tau(n-1)} \frac{1}{n} \sum_{j=1}^{n} 2 L_{j} D_{\widetilde{\phi}_{j}}(x^t, x^\star)\\
		\notag	&\quad +\frac{n(\tau-1)}{\tau(n-1)} 2 L D_{f}(x^t, x^\star)\\
		\notag	&\leq 2 \frac{n-\tau}{\tau(n-1)} \max _{j} L_{j} \frac{1}{n} \sum_{j=1}^{n} D_{f_{j}}(x^t, x^\star)\\
		\notag	&\quad +2 \frac{n(\tau-1)}{\tau(n-1)} L D_{f}(x, y)\\
		\notag	&=2 \frac{n-\tau}{\tau(n-1)} \max _{j} L_{j} D_{f}(x, y)\\
		\notag&\quad +2 \frac{n(\tau-1)}{\tau(n-1)} L D_{f}(x^t, x^\star)\\
		\notag	&=2\left(\frac{n-\tau}{\tau(n-1)} \max _{j} L_{j}+\frac{n(\tau-1)}{\tau(n-1)} L\right) D_{f}(x^t, x^\star).
	\end{align}	
\end{proof}

\subsection{Proof of Theorem~\ref{thm:QLSVRG}}
As in previous analysis we need to show that Assumption~\ref{sigma_t} is satisfied for the \gls{ProxSkip-HUB} method. 
%
\begin{proof}
	Let us consider the first inequality in Assumption~\ref{sigma_t} and show that it holds for the new gradient estimator $\hat{g}^t = \frac{1}{\tau}\sum_{j\in \set} Q(\nabla \widetilde{\phi}_j(x^t) - \nabla \widetilde{\phi}_j(y^t)) + \nabla f(y) $:
	\begin{align*}
		&\Exp{\|\hat{g}^t - \nabla f(x^\star) \|^2} = \Exp{\left\| \frac{1}{\tau}\sum_{j\in \set} Q(\nabla \widetilde{\phi}_j(x^t) - \nabla \widetilde{\phi}_j(y^t)) + \nabla f(y^t) - \nabla f(x^\star) \right\|^2}.
	\end{align*}
Let $\Delta^{t} =  \frac{1}{\tau}\sum_{j\in \set} (\nabla \widetilde{\phi}_j(x^t) - \nabla \widetilde{\phi}_j(y^t))$ and $ \hat{\Delta}_{t} = \frac{1}{\tau}\sum_{j\in \set} Q(\nabla \widetilde{\phi}_j(x^t) - \nabla \widetilde{\phi}_j(y^t))$. Using smart zero $0=\Delta^{t} -\Delta^{t} $ we have
\begin{eqnarray}
	&&\notag		\Exp{\|\hat{g}^t - \nabla f(x^\star) \|^2}\\
 \notag   & = & \Exp{\left\| \hat{\Delta}_{t}  -  \Delta^{t} + \Delta^{t} + \nabla f(y) - \nabla f(x^\star) \right\|^2}\\
\notag	&\stackrel{\eqref{youngs}}{\leq}& 2\Exp{\|\hat{\Delta}^t - \Delta^{t}\|^2} + 2\Exp{\|\Delta^{t}+\nabla f(y^t) - \nabla f(x^\star)\|^2}\\
\notag	&	\leq	&	2\Exp{\left\| \frac{1}{\tau}\sum_{j\in \set} Q(\nabla \widetilde{\phi}_j(x^t) - \nabla \widetilde{\phi}_j(y^t)) -  \frac{1}{\tau}\sum_{j\in \set} (\nabla \widetilde{\phi}_j(x^t) - \nabla \widetilde{\phi}_j(y^t))\right\|^2}\\
\label{eq:12121}
	&  +&2\Exp{\left\| \frac{1}{\tau}\sum_{j\in \set} (\nabla \widetilde{\phi}_j(x^t) - \nabla \widetilde{\phi}_j(y^t))+ \nabla f(y) - \nabla f(x^\star) \right\|^2}.
\end{eqnarray}
Let us consider the first term~\eqref{eq:12121}, let us define $\theta^t_i = Q(\nabla \widetilde{\phi}_j(x^t) - \nabla \widetilde{\phi}_j(y^t)) - (\nabla \widetilde{\phi}_j(x^t) - \nabla \widetilde{\phi}_j(y^t))$:
\begin{align}
\notag\Exp{\|\hat{\Delta}^t - \Delta^{t}\|^2}  &= 	\Exp{\left\| \frac{1}{\tau}\sum_{j\in \set} Q(\nabla \widetilde{\phi}_j(x^t) - \nabla \widetilde{\phi}_j(y^t)) - (\nabla \widetilde{\phi}_j(x^t) - \nabla \widetilde{\phi}_j(y^t)) \right\|^2}\\
\notag& = 	\Exp{\left\| \frac{1}{\tau}\sum_{j\in \set}\theta_i^t \right\|^2}\\
\notag& = 	\Exp{ \frac{1}{\tau^2}\left( \sum_{j\in \set}\left\|\theta_i^t \right\|^2 + \sum_{i\neq j} \left\langle \theta_i^t,\theta_j^t  \right\rangle \right)}\\
\notag& = 	\frac{1}{\tau^2}\left( \sum_{j\in \set}\Exp{ \left\|\theta_i^t \right\|^2} + \sum_{i\neq j}\Exp{  \left\langle \theta_i^t,\theta_j^t  \right\rangle} \right).
\end{align}
Using independence and unbiasedness of compressors we have 
\begin{eqnarray}
\notag&&	\Exp{\|\hat{\Delta}^t - \Delta^{t}\|^2}\\
\notag &=& \frac{1}{\tau^2}\left( \sum_{j\in \set}\Exp{ \left\|\theta_i^t \right\|^2} + \sum_{i\neq j}\Exp{  \left\langle \theta_i^t,\theta_j^t  \right\rangle} \right)\\
\notag		&=&	 \frac{1}{\tau^2}\left( \sum_{j\in \set}\Exp{ \left\|\theta_i^t \right\|^2} + \sum_{i\neq j}  \left\langle \Exp{\theta_i^t},\Exp{\theta_t^j}  \right\rangle \right)\\
\notag		&=&	 \frac{1}{\tau^2} \sum_{j\in \set}\Exp{ \left\|\theta_i^t \right\|^2} \\
\notag		&=& \frac{1}{\tau^2} \sum_{j\in \set} \Exp{\|Q(\nabla \widetilde{\phi}_j(x^t) - \nabla \widetilde{\phi}_j(y^t)) - (\nabla \widetilde{\phi}_j(x^t) - \nabla \widetilde{\phi}_j(y^t))\|^2}\\
	&\stackrel{\eqref{compress}}{\leq}& \frac{\omega}{\tau} \Exp{\frac{1}{\tau}\sum_{j\in \set} \|\nabla \widetilde{\phi}_j(x^t) - \nabla \widetilde{\phi}_j(y^t)\|^2}.
\end{eqnarray}
Using Young's inequality and expectation of Partial Participation we get 
\begin{align}
		\label{part1}
\notag &	\Exp{\|\hat{\Delta}^t - \Delta^{t}\|^2}  \\		
\notag&\stackrel{\eqref{youngs}}{\leq} \frac{2\omega}{\tau} \Exp{\frac{1}{\tau}\sum_{j\in \set} \|\nabla \widetilde{\phi}_j(x^t) - \nabla \widetilde{\phi}_j(x^\star)\|^2}\\
\notag	&\quad +\frac{2\omega}{\tau} \Exp{\frac{1}{\tau}\sum_{j\in \set} \|\nabla \widetilde{\phi}_j(y^t) - \nabla \widetilde{\phi}_j(x^\star)\|^2}\\
	\notag	&\stackrel{\eqref{youngs}}{\leq} \frac{2\omega}{\tau} \frac{1}{n}\sum_{j=1}^n  \|\nabla \widetilde{\phi}_j(x^t) - \nabla \widetilde{\phi}_j(x^\star)\|^2\\
\notag	&\quad +\frac{2\omega}{\tau} \frac{1}{n}\sum_{j=1}^n \|\nabla \widetilde{\phi}_j(y^t) - \nabla \widetilde{\phi}_j(x^\star)\|^2\\
	&\stackrel{\eqref{eq:smooth-grad}}{\leq} \frac{4\omega}{\tau}L_{\max}D_f(x^t,x^\star)+\frac{2\omega}{\tau} \frac{1}{n}\sum_{j=1}^n \|\nabla \widetilde{\phi}_j(y^t) - \nabla \widetilde{\phi}_j(x^\star)\|^2 .
\end{align}
Let us consider the second term in~\eqref{eq:12121}:
\begin{align}
	\label{eq:aqaqaq}
\notag	&\Exp{\left\| \frac{1}{\tau}\sum_{j\in \set} (\nabla \widetilde{\phi}_j(x^t) - \nabla \widetilde{\phi}_j(y^t))+ \nabla f(y) - \nabla f(x^\star) \right\|^2}\\
\notag	& = 	\Exp{\left\| \frac{1}{\tau}\sum_{j\in \set} (\nabla \widetilde{\phi}_j(x^t) - \nabla \widetilde{\phi}_j(y^t)+\nabla \widetilde{\phi}_j(x^\star)-\nabla \widetilde{\phi}_j(x^\star))+ \nabla f(y) - \nabla f(x^\star) \right\|^2} \\
\notag& \stackrel{\eqref{youngs}}{\leq} 2\Exp{\left\|\frac{1}{\tau}\sum_{j\in \set} \nabla \widetilde{\phi}_j(x^t) - \frac{1}{\tau}\sum_{j\in \set} \nabla \widetilde{\phi}_j(x^\star)  \right\|^2}\\
\notag & \quad + 2\Exp{\left\|\frac{1}{\tau}\sum_{j\in \set} \nabla \widetilde{\phi}_j(x^\star) - \frac{1}{\tau}\sum_{j\in \set} \nabla \widetilde{\phi}_j(y^t) - \left(\nabla f(x^\star)-\frac{1}{\tau}\sum_{j\in \set} \nabla \widetilde{\phi}_j(y^t)  \right) \right\|^2} \\
\notag& \leq 2\Exp{\left\|\frac{1}{\tau}\sum_{j\in \set} \nabla \widetilde{\phi}_j(x^t) - \frac{1}{\tau}\sum_{j\in \set} \nabla \widetilde{\phi}_j(x^\star)  \right\|^2}\\
&+ 2\Exp{\left\|\frac{1}{\tau}\sum_{j\in \set} \nabla \widetilde{\phi}_j(y^t) - \frac{1}{\tau}\sum_{j\in \set} \nabla \widetilde{\phi}_j(x^\star)  \right\|^2}.
\end{align}
Using Lemma~\ref{tau-nice} and Jensen's inequality~\eqref{jensen} we have
\begin{align}
	\label{part2}
	\notag	&\Exp{\left\| \frac{1}{\tau}\sum_{j\in \set} (\nabla \widetilde{\phi}_j(x^t) - \nabla \widetilde{\phi}_j(y^t))+ \nabla f(y) - \nabla f(x^\star) \right\|^2}\\
	 \notag& \leq 2\Exp{\left\|\frac{1}{\tau}\sum_{j\in \set} \nabla \widetilde{\phi}_j(x^t) - \frac{1}{\tau}\sum_{j\in \set} \nabla \widetilde{\phi}_j(x^\star)  \right\|^2}\\
     \notag &+ 2\Exp{\left\|\frac{1}{\tau}\sum_{j\in \set} \nabla \widetilde{\phi}_j(y^t) - \frac{1}{\tau}\sum_{j\in \set} \nabla \widetilde{\phi}_j(x^\star)  \right\|^2}\\
	&\leq 4L(\tau)D_f(x^t,x^\star) + \frac{2}{n}\sum_{j=1}^n \left\|\nabla \widetilde{\phi}_j(y^t) -  \nabla \widetilde{\phi}_j(x^\star)  \right\|^2 .
\end{align}
Combining two parts \eqref{part1}, \eqref{part2} and plugging into~\eqref{eq:12121} we get
\begin{align}
		\label{eq:first_in}
\notag	&\Exp{\|\hat{g}^t - \nabla f(x^\star) \|^2}\\
\notag &	\leq		2\Exp{\left\| \frac{1}{\tau}\sum_{j\in \set} Q(\nabla \widetilde{\phi}_j(x^t) - \nabla \widetilde{\phi}_j(y^t)) -  \frac{1}{\tau}\sum_{j\in \set} (\nabla \widetilde{\phi}_j(x^t) - \nabla \widetilde{\phi}_j(y^t))\right\|^2}\\
\notag&\quad +2\Exp{\left\| \frac{1}{\tau}\sum_{j\in \set} (\nabla \widetilde{\phi}_j(x^t) - \nabla \widetilde{\phi}_j(y^t))+ \nabla f(y) - \nabla f(x^\star) \right\|^2}\\
\notag&\leq 8L(\tau)D_f(x^t,x^\star) + \frac{4}{n}\sum_{j=1}^n \left\|\nabla \widetilde{\phi}_j(y^t) -  \nabla \widetilde{\phi}_j(x^\star)  \right\|^2\\
\notag	& \quad + \frac{8\omega}{\tau}L_{\max}D_f(x^t,x^\star)+\frac{4\omega}{\tau} \frac{1}{n}\sum_{j=1}^n \|\nabla \widetilde{\phi}_j(y^t) - \nabla \widetilde{\phi}_j(x^\star)\|^2\\
&\leq 2\cdot 4\left( L(\tau) + \frac{\omega}{\tau}L_{\max} \right)D_f(x^t,x^\star) + 4\left(1+\frac{\omega}{\tau}\right)\sigma^{(t)},
\end{align}
where $\sigma^{(t)} =  \frac{1}{n}\sum_{j=1}^n \|\nabla \widetilde{\phi}_j(y^t) - \nabla \widetilde{\phi}_j(x^\star)\|^2$.
Let us consider update of control variable $y^t$:
\begin{align}
	y_{t+1}=\left\{\begin{array}{lll}
		x^t & \text { with probability } & q \\
		y_{t} & \text { with probability } & 1-q
	\end{array}\right. .
\end{align}
Let us show that second inequality in Assumption~\ref{sigma_t}:
\begin{align}
	\label{eq:second_in}
\notag	\Exp{ \sigma^{(t+1)} } &= \Exp{\frac{1}{n} \sum_{j=1}^{n}  \|\nabla \widetilde{\phi}_j(y_{t+1}) - \nabla \widetilde{\phi}_j(x^\star)\|^2 }\\
\notag	&=\left(1-q\right)\frac{1}{n} \sum_{j=1}^{n}  \|\nabla \widetilde{\phi}_j(y_{t}) - \nabla \widetilde{\phi}_j(x^\star)\|^2 + q \frac{1}{n} \sum_{j=1}^{n}  \|\nabla \widetilde{\phi}_j(x^t) - \nabla \widetilde{\phi}_j(x^\star)\|^2\\
	& = \left(1-q\right)\sigma^{(t)} + 2 qL_{\max} D_f(x^t,x^\star).
\end{align}
Using \eqref{eq:first_in} and \eqref{eq:second_in} bounds we can confirm that Assumption~\ref{sigma_t} is satisfied with the following constants:
\begin{align*}	
	&A = 4\left(L(\tau)+\frac{\omega}{\tau}L_{\max}\right) , 
	\quad B = 4\left(1+\frac{\omega}{\tau}\right), \quad C = 0, \\&\ \tilde{A} = q L_{\max},\quad \tilde{B} = 1-q, \quad \tilde{C} = 0.
\end{align*}
Applying Theorem~\ref{thm:main_vr_proxskip} leads to the final result
	\begin{align}
		\squeeze 	\Exp{\Psi^{(T)}} \leq \max \left\{(1-\gamma \mu)^{T},(1-p^2)^T,\left(1-\nicefrac{q}{2}\right)^{T}\right\} \Psi^{(0)},
	\end{align}
	where the Lyapunov function is defined by $$	\Psi^{(t)} \eqdef \|x^t - x^\star\|^2 + \frac{\gamma^2}{p^2}\|h^t - h^\star\|^2 + \gamma^{2} \frac{8}{q}\left(1+\frac{\omega}{\tau}\right) \sigma^{(t)} .$$
	Let us set $\gamma = \frac{1}{A+\MM \tilde{A}}$, $p = \sqrt{\gamma\mu}$ and $q = 2\gamma\mu$. Using the same proof as for \gls{ProxSkip} in Section~\ref{sec:GD_est} and $L(\tau) \leq L_{\max}$ we get communication and iteration complexities:  
\begin{align*}
	T_{\text{comm.}}=\mathcal{O}\left(\sqrt{\frac{L_{\max}}{\mu}\left(1+\frac{\omega}{\tau}\right)}\log\frac{1}{\varepsilon}\right), \qquad 
		T_{\text{iter.}} = \mathcal{O}\left(\frac{L_{\max}}{\mu}\left(1+\frac{\omega}{\tau}\right)\log\frac{1}{\varepsilon}\right).
\end{align*}

\end{proof}
If we use full participation $\tau = n$ and $q = \frac{1}{\omega+1}$ and $r(x) \equiv 0$ then we get the same rate as for \gls{DIANA}~\citep{mishchenko2019distributed,DIANA2} and \algname{RAND-DIANA}~\citep{Shifted}.

\section{Analysis of ProxSkip-LSVRG}
The analysis of \gls{ProxSkip-LSVRG} is almost the same as the analysis of \gls{ProxSkip-HUB}, with one exception. We use a different sigma component: 
\begin{align}
	\sigma^{(t)} = \Exp{\left\| \frac{1}{\tau} \sum_{j\in \set}(\nabla \widetilde{\phi}_j(y^t) - \nabla \widetilde{\phi}_j(x^\star)) \right\|^2}.
\end{align}
Let us consider $\Exp{\| \hat{g}^t - \nabla f(x^\star) \|^2}$:
\begin{align}
	\Exp{\| \hat{g}^t - \nabla f(x^\star) \|^2} = \Exp{\left\|\frac{1}{\tau}\sum_{j \in S_t} \left( \nabla \widetilde{\phi}_j(x^t) - \nabla \widetilde{\phi}_j(y^t) \right) + \nabla f(y)   - \nabla f(x^\star) \right\|^2} 
\end{align}
Using~\eqref{eq:aqaqaq} we have 
\begin{align}
\notag		\Exp{\| \hat{g}^t - \nabla f(x^\star) \|^2} &\leq 2\Exp{\left\|\frac{1}{\tau}\sum_{j\in \set} \nabla \widetilde{\phi}_j(x^t) - \frac{1}{\tau}\sum_{j\in \set} \nabla \widetilde{\phi}_j(x^\star)  \right\|^2}\\
\notag			& + 2\Exp{\left\|\frac{1}{\tau}\sum_{j\in \set} \nabla \widetilde{\phi}_j(y^t) - \frac{1}{\tau}\sum_{j\in \set} \nabla \widetilde{\phi}_j(x^\star)  \right\|^2}\\
		&\leq 4L(\tau) D_f(x^t,x^\star) +2\sigma^{(t)} .
\end{align}
Let us show that the second inequality in Assumption~\ref{sigma_t} holds; using Lemma~\ref{tau-nice}, we get
\begin{align}
	\label{eq:second_in_}
	\notag	\Exp{ \sigma^{(t+1)} } &=  \Exp{\left\| \frac{1}{\tau} \sum_{j\in \set}(\nabla \widetilde{\phi}_j(y_{t+1}) - \nabla\widetilde{\phi}_j(x^\star)) \right\|^2}\\
	\notag	&=\left(1-q\right) \Exp{\left\| \frac{1}{\tau} \sum_{j\in \set}(\nabla\widetilde{\phi}_j(y^t) - \nabla\widetilde{\phi}_j(x^\star)) \right\|^2}\\
   \notag &+ q  \Exp{\left\| \frac{1}{\tau} \sum_{j\in \set}(\nabla \widetilde{\phi}_j(x^t) - \nabla \widetilde{\phi}_j(x^\star)) \right\|^2}\\
	& = \left(1-q\right)\sigma^{(t)} + 2 qL(\tau) D_f(x^t,x^\star).
\end{align}
We showed that Assumption~\ref{sigma_t} is satisfied with the following constants:
\begin{align}
		A = 2 L(\tau) , 
	\quad B = 2, \quad C = 0, \quad \tilde{A} = q L(\tau),\quad \tilde{B} = 1-q, \quad \tilde{C} = 0.
\end{align}
Applying Theorem~\ref{thm:main_vr_proxskip} with $\gamma = \frac{1}{6L(\tau)}$  we get the final bound:
\begin{align*}
	\squeeze 	\Exp{\Psi^{(T)}} \leq \max \left\{(1-\gamma \mu)^{T},(1-p^2)^T,\left(1-\frac{q}{2}\right)^{T}\right\} \Psi^{(0)},
\end{align*}
where the Lyapunov function is defined as $$	\Psi^{(t)} \eqdef \|x^t - x^\star\|^2 + \frac{\gamma^2}{p^2}\|h^t - h^\star\|^2 + \gamma^{2} \frac{4}{q}\sigma^{(t)}.$$

Using the same argument as for \gls{ProxSkip} and setting $\frac{q}{2} = \gamma\mu$, we get 
\begin{align}
T_{\text{comms}} =	\mathcal{O}\left(\sqrt{\frac{L(\tau)}{\mu}}\log\frac{1}{\varepsilon}\right), \qquad T_{\text{iters}} =\mathcal{O}\left(\frac{L(\tau)}{\mu}\log\frac{1}{\varepsilon}\right).
\end{align}
If $r(x) \equiv 0$, this recovers results in \citep{kovalev2020don}.

\newpage
            \refstepcounter{chapter}%
\chapter*{\thechapter \quad Appendix D Title}
\label{appendixD}

\section{Basic Inequalities}

\subsection{Young's inequalities} For all $x,y\in \mathbb{R}^d$ and all $a>0$, we have
	\begin{eqnarray}
		&&\langle x, y\rangle\leq \frac{a\sqnorm{x}}{2}+\frac{\sqnorm{y}}{2a},\label{yi1}\\
		&&\sqnorm{x+y}\leq2\sqnorm{x}+2\sqnorm{y}\label{yi2},\\
		&&\frac{1}{2}\sqnorm{x}-\sqnorm{y}\leq \sqnorm{x+y}.\label{yi3}
	\end{eqnarray}

\subsection{Variance decomposition} For a random vector $\mathrm{X}\in\mathbb{R}^d$ (with finite second moment) and any $c\in\mathbb{R}^d$, the variance of $X$ can be decomposed as
	\begin{eqnarray}
		\Exp{\sqnorm{\mathrm{X}-\Exp{\mathrm{X}}}}=\Exp{\sqnorm{\mathrm{X}-c}}-\sqnorm{\Exp{\mathrm{X}}-c}.\label{vardec}
	\end{eqnarray}

\subsection{Compressor variance}	An unbiased randomized mapping  $\cC: \mathbb{R}^d\to \mathbb{R}^d$ has conic variance if there exists $\omega\geq 0$ such that
	\begin{equation}
		\Exp{\sqnorm{\mathcal{C}(x)-x}}\leq \omega \sqnorm{x}\label{cvar}
	\end{equation}
	for all $x\in \mathbb{R}^d.$

\subsection{Convexity and \texorpdfstring{$L$}{l}-smoothness}

	Suppose $\phi\colon\mathbb{R}^d\to\mathbb{R}$ is $L$-smooth and convex. Then
	\begin{equation}
		\frac{1}{L}\sqnorm{\nabla \phi(x)-\nabla \phi(y)}\leq \langle \nabla \phi(x)-\nabla \phi(y),x-y \rangle\label{strmono}
	\end{equation}
	 for all $x,y\in\mathbb{R}^d$.

\subsection{Partial participation operator}

\begin{definition}[Partial Participation Operator]
	The Partial Participation operator is the randomized mapping $\Rop:\mathbb{R}^{\Mx d}\to \mathbb{R}^{\Mx d}$ defined as follows. We choose a random subset $S\subseteq \left\{1,\dots,\Mx\right\}$ of size $\Cx \in \{1,\dots,\Mx\}$ uniformly at random, and for  $v=\left(v_1,\dots,v_\Mx\right) \in \mathbb{R}^{\Mx  d}$, where $v_\iclient\in\mathbb{R}^d$ for all $\iclient$, we define $$\Rop(v)\eqdef \left(\Rop_1(v_1),\dots,\Rop_\Mx(v_\Mx)\right),$$ where
	\begin{eqnarray*}
		\Rop_\iclient(v_\iclient)\eqdef \begin{cases}
			\frac{\Mx}{\Cx}v_\iclient \in \mathbb{R}^d & \quad\text{for}\quad \iclient\in S,\\
			 0  \in \mathbb{R}^d &\quad\text{otherwise.}
		\end{cases}
	\end{eqnarray*}
\end{definition}
The Partial Participation operator admits the following identity:
	\begin{equation}
		\Exp{\sqnorm{\Koper^\top\left(\Rop(v)-v\right)}}=\frac{\Mx}{\Cx}\frac{\Mx-\Cx}{\Mx-1}\sum_{\iclient=1}^{\Mx}\sqnorm{v_\iclient}-\frac{\Mx-\Cx}{\Cx\left(\Mx-1\right)}\sqnorm{\sum_{\iclient=1}^{\Mx}v_\iclient},\label{PP}
	\end{equation} 
	where $\Koper$ was defined in Section~\ref{sec:H}, and $v=\left(v_1,\dots,v_\Mx\right)\in \mathbb{R}^{\Mx  d}$ and $v_{\iclient}\in\mathbb{R}^d$.
\begin{proof}
	Let $\mathbb{E}_S$ denote expectation with respect to the random set $S$. We can write 
	\begin{eqnarray*}
		&&\Exp{\sqnorm{\Koper^\top\left(\Rop(v)-v\right)}}\\&=&\Exp{\sqnorm{\sum_{\iclient=1}^{\Mx}\left(\Rop_\iclient(v_\iclient)-v_\iclient\right)}}\\
        &=&\Exps{\sqnorm{\sum_{\iclient\in S}\frac{\Mx}{\Cx}v_\iclient-\sum_{\iclient=1}^{\Mx}v_\iclient}}\\
		&=&\frac{\Mx^2}{\Cx^2}\Exps{\sqnorm{\sum_{\iclient\in S}v_\iclient}}+\sqnorm{\sum_{\iclient=1}^{\Mx}v_\iclient}-\frac{2\Mx}{\Cx}\Exps{\left\langle \sum_{\iclient\in S}v_\iclient, \sum_{\iclient=1}^{\Mx}v_\iclient\right\rangle}\\
		&=&\frac{\Mx^2}{\Cx^2}\Exps{\sum_{\iclient\in S}\sqnorm{v_\iclient}}+\frac{\Mx^2}{\Cx^2}\Exps{\sum_{\iclient\in S}\sum_{\iclient'\in S,\neq\iclient} \left\langle v_\iclient,v_{\iclient'}\right\rangle}-\sqnorm{\sum_{\iclient=1}^{\Mx}v_\iclient}.
	\end{eqnarray*}
	By computing the expectation on the right hand side, we get
	\begin{eqnarray*}
&&\Exp{\sqnorm{\sum_{\iclient=1}^{\Mx}\left(\Rop_\iclient(v_\iclient)-v_\iclient\right)}}\\
		&=&\frac{\Mx}{\Cx}\sqnorm{\sum_{\iclient=1}^{\Mx}v_\iclient}+\frac{\Mx}{\Cx}\frac{\Cx-1}{\Mx-1}\sum_{\iclient=1}^{\Mx}\sum_{\iclient'=1,\neq\iclient}^{\Mx} \left\langle v_\iclient,v_{\iclient'}\right\rangle-\sqnorm{\sum_{\iclient=1}^{\Mx}v_\iclient}\\
		&=&\frac{\Mx}{\Cx}\left(1-\frac{\Cx-1}{\Mx-1}\right)\sqnorm{\sum_{\iclient=1}^{\Mx}v_\iclient}+\left(\frac{\Mx\left(\Cx-1\right)}{\Cx\left(\Mx-1\right)}-1\right)\sqnorm{\sum_{\iclient=1}^{\Mx}v_\iclient}\\
		&=&\frac{\Mx}{\Cx}\left(\frac{\Mx-\Cx}{\Mx-1}\right)\sqnorm{\sum_{\iclient=1}^{\Mx}v_\iclient}-\frac{\Mx-\Cx}{\Cx\left(\Mx-1\right)}\sqnorm{\sum_{\iclient=1}^{\Mx}v_\iclient}.
	\end{eqnarray*}
	
\end{proof}

\subsection{Dual problem and Saddle-Point reformulation}

Then the saddle function reformulation of \myref{eq:main-new} is:
\begin{equation}
  (x^\star,(u_\iclient^\star)_{\iclient=1}^\Mx) \in \arg\min_{x\in\mathbb{R}^d}\max_{u\in\mathbb{R}^{\Mx d}} \, \left( \frac{\mu}{2}\sqnorm{x}+\sum_{\iclient=1}^\Mx \left\langle  x,u_\iclient \right\rangle -\sum_{\iclient=1}^\Mx F_\iclient^*(u_\iclient)\right).
\label{saddlenew}
\end{equation}
To ensure well-posedness of these problems, we need to assume that there exists $x^\star\in\mathbb{R}^d$ s.t.:
\begin{align}
	0=\mu x^\star+\sum_{\iclient=1}^{\Mx} \nabla F_\iclient(x^\star).
\end{align}
This is equivalent to \myref{eq:main} having a solution, which it does (unique in fact) as each $f_\iclient$ is $\mu$-strongly convex.
By first order optimality condition $x^\star$ and $u^\star$ that are solutions to \myref{saddlenew}, satisfy:
\begin{equation}
	\left\{ \begin{array}{l}
		0= \mu x^\star + \sum_{\iclient=1}^\Mx u_\iclient^{\star}\\
		\Koper x^\star\in\partial F^* (u^{\star})
	\end{array}\right..\label{fooc}
\end{equation}
where the latter in \myref{fooc} is equivalent to:
\begin{equation}
	\nabla F(\Koper x^\star)=u^\star.
\end{equation}
Throughout this section, we will denote by $\mathcal{F}^t$ for all $ t\geq 0$ the $\sigma$-algebra generated by the collection of $\left(\mathbb{R}^d\times\mathbb{R}^{d \Mx}\right)$-valued random variables $\left(x^0,u^0\right),\dots,\left(x^t,u^t\right).$

\clearpage
\section{Analysis of 5GCS\texorpdfstring{${}_\infty$}{\_infty}}

\begin{algorithm*}
	\caption{\gls{5GCS} with $\infty$ local \algname{GD} steps (a.k.a.\ \algname{Minibatch Point-SAGA})}
	\begin{algorithmic}[1]\label{alg:5GCS-infty}
		\STATE  \textbf{input:} initial points $x^0\in\mathbb{R}^d$, $u_\iclient^0\in\mathbb{R}^d$ for all $\iclient=\{1,\dots,\Mx\}$; 
		\STATE stepsize $\gammaM>0$, $\tauM>0$; $\Cx\in \{1,\dots,\Mx\}$
		\STATE $v^0\eqdef \sum_{\iclient=1}^\Mx u_\iclient^0$
		\FOR{$\kstep=0, 1, \ldots$}
		\STATE $\hat{x}^{\kstep} \eqdef \frac{1}{1+\gammaM\mu} \left(x^\kstep - \gammaM v^\kstep\right)  $
		\STATE Pick $\set \subset \{1,\ldots,\Mx\}$ of size $\Cx$ uniformly at random
		\FOR{$\iclient\in\set$}
		\STATE $u_{\iclient}^{\kstep+1}\eqdef  u_{\iclient}^\kstep + \tauM  \hat{x}^{\kstep} - \tauM \mathrm{prox}_{\frac{1}{\tauM} F_\iclient} \left(\hat{x}^{\kstep}+ \frac{1}{\tauM} u_{\iclient}^\kstep  \right)$
		\ENDFOR
		\FOR{$\iclient\in\{1, \ldots,\Mx\}\backslash \set$}
		\STATE $u_{\iclient}^{\kstep+1}\eqdef u_{\iclient}^\kstep $
		\ENDFOR
		\STATE $v^{\kstep+1}\eqdef \sum_{\iclient=1}^\Mx u_\iclient^{\kstep+1}$
		\STATE  $x^{\kstep+1} \eqdef \hat{x}^{\kstep}- \gammaM \frac{\Mx}{\Cx} (v^{\kstep+1}-v^\kstep)$
		\ENDFOR
	\end{algorithmic}
\end{algorithm*}
\begin{theorem} Consider Algorithm~\ref{alg:5GCS} (\gls{5GCS}) with the LT solver being \algname{\gls{GD}} run for $K=+\infty$ iterations (this is equivalent to Algorithm~\ref{alg:5GCS-infty}; we shall also call the method \algname{5GCS${}_\infty$}).
	Let $\gammaM>0$, $\tauM>0$ and $\gammaM \tauM \leq \frac{1}{\Mx}$. Then for the Lyapunov function
	\begin{equation*}
		\Psi^{\kstep}\eqdef  \frac{1}{\gammaM}\sqnorm{x^{\kstep}-x^\star}+\frac{\Mx}{\Cx}\left(\frac{1}{\tauM}+2\frac{1}{L_F}\right)\sqnorm{u^{\kstep}-u^\star},
	\end{equation*}
	the iterates of the method satisfy
	$$
	\Exp{\Psi^{\Tx}}\leq (1-\rho)^\Tx \Psi^0,
	$$
	where 
	$ 	\rho \eqdef  \min\left(\frac{\gammaM\mu}{1+\gammaM\mu}, \frac{\Cx}{\Mx}\frac{2\tauM }{L_F+2\tauM }\right)<1.
	$

\end{theorem}
\begin{proof}
	Noting that updates for $u^{t+1}$ and $x^{t+1}$ can be written as
	\begin{eqnarray}
		&u^{t+1}\eqdef u^t + \tfrac{1}{1+\omega} \Rop^t\left(\hat{u}^{t+1}-u^t\right),&\\
		&x^{t+1}=\hat{x}^t-\gammaM\frac{\Mx}{\Cx}\Koper^\top\left(u^{t+1}-u^t \right)&\label{xupdate}
	\end{eqnarray}
	where $\Rop^t$ is the sampling operator, $\omega=\frac{\Mx}{\Cx}-1$ and $\hat{u}^{t+1} = \mathrm{prox}_{\tauM F^*}\left(u^{t}+\tauM \Koper \hat{x}^{t}\right)$. 
	Using variance decomposition and Proposition 1 from \citep{condat2021murana} to write
	\begin{eqnarray}
	\notag	&&\Exp{\sqnorm{x^{t+1}-x^\star}\;|\;\mathcal{F}^t}\\
        &\overset{(\ref{vardec})}{=}&\sqnorm{\Exp{x^{t+1}\;|\;\mathcal{F}^t}-x^\star}+\Exp{\sqnorm{x^{t+1}-\Exp{x^{t+1}\;|\;\mathcal{F}^t}}\;|\;\mathcal{F}^t}\notag\\
	\notag	&\overset{(\ref{xupdate})}{=}& \sqnorm{\Exp{\hat{x}^{t}-\gammaM\frac{\Mx}{\Cx} (v^{t+1}-v^t)\;|\;\mathcal{F}^t}-x^\star}\\
       \notag &+&\Exp{\sqnorm{x^{t+1}-\Exp{x^{t+1}\;|\;\mathcal{F}^t}}\;|\;\mathcal{F}^t}\notag\\
		\notag &=& \sqnorm{\hat{x}^{t}-x^\star-\gammaM\frac{\Mx}{\Cx} \Exp{\Koper^\top\left(u^{t+1}-u^t \right)\;|\;\mathcal{F}^t}}\\
        &+&\Exp{\sqnorm{x^{t+1}-\Exp{x^{t+1}\;|\;\mathcal{F}^t}}\;|\;\mathcal{F}^t}\notag\\
		&=& \sqnorm{\hat{x}^{t}-x^\star-\gammaM\Koper^\top\left(\hat{u}^{t+1}-u^t \right)}+\Exp{\sqnorm{x^{t+1}-\Exp{x^{t+1}\;|\;\mathcal{F}^t}}\;|\;\mathcal{F}^t}\notag\\
		&\overset{(\ref{PP})}{=}& \underbrace{ \sqnorm{\hat{x}^{t}-x^\star-\gammaM \Koper^\top\left(\hat{u}^{t+1}-u^t \right)}}_X+\gammaM^2\oma\sqnorm{\hat{u}^{t+1}-u^t}\notag\\
		&\quad &- \gammaM^2\zeta\sqnorm{\Koper^\top\left(\hat{u}^{t+1}-u^t \right)}.\label{xbound31}
	\end{eqnarray}
	where
	\begin{equation*}
		\oma=\frac{\Mx(\Mx-\Cx)}{\Cx(\Mx-1)},\quad \zeta=\frac{\Mx-\Cx}{\Cx(\Mx-1)}.
	\end{equation*}
	Moreover, using \myref{fooc} and the definition of $\hat{x}^t$, we have
	\begin{align}
		&(1+\gammaM\mu)\hat{x}^t=x^t-\gammaM \Koper^\top u^{t},\label{opt13}\\
		&(1+\gammaM\mu)x^\star= x^\star -\gammaM\Koper^\top u^\star.\label{opt23}
	\end{align}
	Using \myref{opt13} and \myref{opt23} we obtain
	\begin{eqnarray}
	\notag	X&=&\sqnorm{\hat{x}^{t}-x^\star} +\gammaM^2\sqnorm{\Koper^\top\left(\hat{u}^{t+1}-u^t \right)}\\
        &&-2\gammaM \left\langle 
		\hat{x}^{t}-x^\star,\Koper^\top\left(\hat{u}^{t+1}-u^t \right) \right\rangle  \notag\\
		&\leq &(1+\gammaM\mu) \sqnorm{\hat{x}^{t}-x^\star}
        +\gammaM^2\sqnorm{\Koper^\top\left(\hat{u}^{t+1}-u^t \right)}\notag\\
	\notag	&&-2\gammaM  \left\langle 
		\hat{x}^{t}-x^\star,\Koper^\top\left(\hat{u}^{t+1}-u^\star \right) \right\rangle\\
        &&+2\gammaM  \left\langle 
		\hat{x}^{t}-x^\star,\Koper^\top\left(u^{t}-u^\star \right) \right\rangle  \notag\\
		&\overset{(\ref{opt13})+(\ref{opt23})}{=} &  \left\langle  x^t-x^\star-\gammaM \Koper^\top\left(u^{t}-u^\star \right),\hat{x}^{t}-x^\star \right\rangle  \notag\\
        &&+\gammaM^2\sqnorm{\Koper^\top\left(\hat{u}^{t+1}-u^t \right)}\notag\\
	\notag	&\quad&-2\gammaM  \left\langle 
		\hat{x}^{t}-x^\star,\Koper^\top\left(\hat{u}^{t+1}-u^\star \right) \right\rangle\\
        \notag&&+  \left\langle 
		\hat{x}^{t}-x^\star,2\gammaM\Koper^\top\left(u^{t}-u^\star \right)\right\rangle  \notag\\
		&= &\left\langle  x^t-x^\star+\gammaM \Koper^\top\left(u^{t}-u^\star \right),\hat{x}^{t}-x^\star \right\rangle  +\gammaM^2\sqnorm{\Koper^\top\left(\hat{u}^{t+1}-u^t \right)}\notag\\
		&\quad&-2\gammaM  \left\langle 
		\hat{x}^{t}-x^\star,\Koper^\top\left(\hat{u}^{t+1}-u^\star \right) \right\rangle  \notag\\
		&\overset{(\ref{opt13})+(\ref{opt23})}{=}&\frac{1}{1+\gammaM\mu} \left\langle  x^t-x^\star+\gammaM \Koper^\top\left(u^{t}-u^\star \right),x^t-x^\star-\gammaM \Koper^\top\left(u^{t}-u^\star \right) \right\rangle  \notag\\
		&\quad&+\gammaM^2\sqnorm{\Koper^\top\left(\hat{u}^{t+1}-u^t \right)}-2\gammaM  \left\langle 
		\hat{x}^{t}-x^\star,\Koper^\top\left(\hat{u}^{t+1}-u^\star \right) \right\rangle  \notag\\
		&= &\frac{1}{1+\gammaM\mu}\sqnorm{x^t-x^\star}-\frac{\gammaM^2}{1+\gammaM\mu}\sqnorm{\Koper^\top\left(u^{t}-u^\star \right)}\notag\\
    \notag&\quad&+\gammaM^2\sqnorm{\Koper^\top\left(\hat{u}^{t+1}-u^t \right)}\\
        &&-2\gammaM  \left\langle 
		\hat{x}^{t}-x^\star,\Koper^\top\left(\hat{u}^{t+1}-u^\star \right) \right\rangle  . \label{long31} 
	\end{eqnarray}
	Combining \myref{xbound31} and \myref{long31}
	\begin{eqnarray}
		\Exp{\sqnorm{x^{t+1}-x^\star}\;|\;\mathcal{F}^t}&\leq&  \frac{1}{1+\gammaM\mu}\sqnorm{x^t-x^\star}-\frac{\gammaM^2}{1+\gammaM\mu}\sqnorm{\Koper^\top\left(u^{t}-u^\star \right)}\notag\\
		&\quad&+\gammaM^2(1-\zeta)\sqnorm{\Koper^\top\left(\hat{u}^{t+1}-u^t \right)}\notag\\
        &&-2\gammaM  \left\langle 
		\hat{x}^{t}-x^\star,\Koper^\top\left(\hat{u}^{t+1}-u^\star \right) \right\rangle \notag\\
		&\quad&+\gammaM^2\oma\sqnorm{\hat{u}^{t+1}-u^t}.\label{xbnd31}
	\end{eqnarray}
	On the other hand using the variance decomposition and conic variance of $\Rop^t$ 
	\begin{eqnarray}
		&&\Exp{\sqnorm{u^{t+1}-u^\star}\;|\;\mathcal{F}^t}\notag \\
        &\overset{(\ref{vardec})+(\ref{cvar})}{\leq}& \sqnorm{u^{t}-u^\star+\frac{1}{1+\omega}\left(\hat{u}^{t+1} -u^t\right)}
		+\frac{\omega}{(1+\omega)^2}\sqnorm{\hat{u}^{t+1} -u^t }\notag\\
		&= &\sqnorm{\frac{\omega}{1+\omega}(u^{t}-u^\star)+\frac{1}{1+\omega}\left(\hat{u}^{t+1} -u^\star\right)}\notag\\
		&&+\frac{\omega}{(1+\omega)^2}\sqnorm{\hat{u}^{t+1} - u^\star - (u^t - u^\star) }\notag\\
		&=&\frac{\omega^2}{(1+\omega)^2}\sqnorm{u^{t}-u^\star}+\frac{1}{(1+\omega)^2}\sqnorm{\hat{u}^{t+1}-u^\star}\notag\\
		&\quad&+\frac{2\omega}{(1+\omega)^2} \left\langle  u^{t}-u^\star,
		\hat{u}^{t+1}-u^\star \right\rangle +\frac{\omega}{(1+\omega)^2}\sqnorm{\hat{u}^{t+1} -u^\star }\notag\\
		&\quad&+\frac{\omega}{(1+\omega)^2}\sqnorm{u^{t} -u^\star }-\frac{2\omega}{(1+\omega)^2} \left\langle  u^{t}-u^\star,
		\hat{u}^{t+1}-u^\star \right\rangle \notag\\
		&=&\frac{1}{1+\omega}\sqnorm{\hat{u}^{t+1} -u^\star }+\frac{\omega}{1+\omega}\sqnorm{u^{t} -u^\star }. \label{ubound}
	\end{eqnarray}
	Let $(s_\iclient^{t+1})_{\iclient=1}^\Mx\in \partial F^*(\hat{u}^{t+1})$ be such that $\hat{u}_\iclient^{t+1}=u_\iclient^t + \tauM \hat{x}^{t}-\tauM s_\iclient^{t+1}$;  $s^{t+1}$ exists and is unique. We also define $s_\iclient^\star\eqdef  x^\star$; we have $s^\star\in \partial F^*(u^\star)$. 
	Therefore,
	\begin{eqnarray}
		\sqnorm{\hat{u}^{t+1}-u^\star}&=&\sqnorm{(u^t-u^\star)+(\hat{u}^{t+1}-u^t)}\notag\\
		&=&\sqnorm{u^t-u^\star}+\sqnorm{\hat{u}^{t+1}-u^t}+2 \left\langle  u^t-u^\star,\hat{u}^{t+1}-u^t \right\rangle \notag \\
		&=&\sqnorm{u^t-u^\star}+2  \left\langle  \hat{u}^{t+1}-u^\star,\hat{u}^{t+1}-u^t \right\rangle  - \sqnorm{\hat{u}^{t+1}-u^t} \notag\\
		\notag&=&\sqnorm{u^t-u^\star} - \sqnorm{\hat{u}^{t+1}-u^t}\\
        &&+2\tauM  \left\langle  \Koper^\top\left(\hat{u}^{t+1}-u^\star \right), \hat{x}^{t}-x^\star \right\rangle \notag \\
		&\quad& -2\tauM  \left\langle  \hat{u}^{t+1}-u^\star,s^{t+1}-s^\star \right\rangle .\label{ubound31}
	\end{eqnarray}
	Combining \myref{ubound}, \myref{ubound31} and \myref{xbnd31} gives
	\begin{eqnarray*}
		&&\frac{1}{\gammaM}\Exp{\sqnorm{x^{t+1}-x^\star}\;|\;\mathcal{F}^t}+\frac{1+\omega}{\tauM}\Exp{\sqnorm{u^{t+1}-u^\star}\;|\;\mathcal{F}^t}\\
		&\leq&  \frac{1}{\gammaM(1+\gammaM\mu)}\sqnorm{x^t-x^\star}-\frac{\gammaM}{1+\gammaM\mu}\sqnorm{\Koper^\top\left(u^{t}-u^\star \right)}\\
		&\quad&+\gammaM(1-\zeta)\sqnorm{\Koper^\top\left(\hat{u}^{t+1}-u^t \right)}-2  \left\langle 
		\hat{x}^{t}-x^\star,\Koper^\top\left(\hat{u}^{t+1}-u^\star \right) \right\rangle \\
		&\quad&+\gammaM\oma\sqnorm{\hat{u}^{t+1}-u^t}+ \frac{1}{\tauM}\sqnorm{u^t-u^\star} - \frac{1}{\tauM}\sqnorm{\hat{u}^{t+1}-u^t}\\
		&\quad&+2  \left\langle  \Koper^\top\left(\hat{u}^{t+1}-u^\star \right), \hat{x}^{t}-x^\star \right\rangle  -2  \left\langle  \hat{u}^{t+1}-u^\star,s^{t+1}-s^\star \right\rangle  \\
		&\quad&+\frac{\omega}{\tauM}\sqnorm{u^{t} -u^\star }\\
		&\leq& \frac{1}{\gammaM(1+\gammaM\mu)}\sqnorm{x^t-x^\star}-\frac{\gammaM}{1+\gammaM\mu}\sqnorm{\Koper^\top\left(u^{t}-u^\star \right)}\\
		&\quad&+\frac{1+\omega}{\tauM}\sqnorm{u^{t} -u^\star }+ \left(\gammaM \left((1-\zeta)\Mx+\oma\right) - \frac{1}{\tauM}\right) \sqnorm{\hat{u}^{t+1}-u^t} \\
		&\quad&-2  \left\langle  \hat{u}^{t+1}-u^\star,s^{t+1}-s^\star \right\rangle \\
		&\leq& \frac{1}{\gammaM(1+\gammaM\mu)}\sqnorm{x^t-x^\star}-\frac{\gammaM}{1+\gammaM\mu}\sqnorm{\Koper^\top\left(u^{t}-u^\star \right)}\\
		&\quad&+\frac{1+\omega}{\tauM}\sqnorm{u^{t} -u^\star }-2  \left\langle  \hat{u}^{t+1}-u^\star,s^{t+1}-s^\star \right\rangle .
	\end{eqnarray*}
	By $\frac{1}{L_{F}}$-strong monotonicity of $\partial F^*$, $ \left\langle  \hat{u}^{t+1}-u^\star,s^{t+1}-s^\star \right\rangle  \geq \frac{1}{L_{F}} \sqnorm{\hat{u}^{t+1}-u^\star}$, and using \myref{ubound},
	\begin{equation*}
		\left\langle  \hat{u}^{t+1}-u^\star,s^{t+1}-s^\star \right\rangle  \geq \frac{1}{L_{F}} \left((1+\omega)\Exp{\sqnorm{u^{t+1}-u^\star}\;|\;\mathcal{F}^t}-\omega \sqnorm{u^{t} -u^\star }\right).
	\end{equation*}
	Hence, 
	\begin{eqnarray}
		&&\frac{1}{\gammaM}\Exp{\sqnorm{x^{t+1}-x^\star}\;|\;\mathcal{F}^t}+(1+\omega)\left(\frac{1}{\tauM}+2\frac{1}{L_{F}}\right)\Exp{\sqnorm{u^{t+1}-u^\star}\;|\;\mathcal{F}^t}\notag\\
		& \leq& \frac{1}{\gammaM(1+\gammaM\mu)}\sqnorm{x^t-x^\star}+\left(\frac{1+\omega}{\tauM}+2\omega \frac{1}{L_{F}} \right)
		\sqnorm{u^{t} -u^\star }\notag\\
		&\quad &-\frac{\gammaM}{1+\gammaM\mu}\sqnorm{\Koper^\top\left(u^{t}-u^\star \right)}.\label{kkkk}
	\end{eqnarray}
	Ignoring the last term in \myref{kkkk}, we obtain
	\begin{equation}
		\Exp{\Psi^{t+1}}\leq \max\left(\frac{1}{1+\gammaM\mu},1-\frac{2\tauM}{(1+\omega)(L_{F}+2\tauM)}\right)\Exp{\Psi^{t}}.\label{finresult}
	\end{equation}
\end{proof}

\subsection{Proof of Corollary~\ref{cor:5GCS-infty}}
\begin{corollary}
	Choose any $0<\varepsilon<1$. If
	we choose $\gammaM=\sqrt{\frac{2\Cx}{L_F\mu \Mx^2}}$ and $\tauM=\sqrt{\frac{L_F\mu}{2\Cx}}$, then  in order to guarantee $\Exp{\Psi^{\Tx}}\leq \varepsilon \Psi^0$, it suffices to take
	\[
		\Tx \geq	\left(\frac{\Mx}{\Cx}+\sqrt{\frac{\Mx}{ \Cx} \frac{L-\mu}{2\mu}}\right)\log \frac{1}{\varepsilon} =\tilde{\cO}\left(\frac{\Mx}{\Cx}+\sqrt{\frac{\Mx}{\Cx}\frac{L}{\mu}}\right)
	\]
	communication rounds.
\end{corollary}
\begin{proof}
	Firstly, note that choosing $\gammaM=\sqrt{\frac{2\Cx}{L_{F}\mu \Mx^2}}$ and $\tauM=\sqrt{\frac{L_{F}\mu}{2\Cx}}$ we satisfy $\gammaM\tauM=\frac{1}{\Mx}$, then we get the contraction constant from the proof to be equal to:
	\begin{eqnarray} &&\max\left\{1-\frac{\sqrt{\frac{2\Cx\mu}{L_{F}\Mx^2}}}{1+\sqrt{\frac{2\Cx\mu}{L_{F}\Mx^2}}},1-\frac{\sqrt{\frac{2L_{h}\mu}{\Cx}}}{\frac{\Mx}{\Cx}\left(L_{F}+\sqrt{\frac{2L_{F}\mu}{\Cx}}\right)}\right\}\notag\\
    & = &
		\max\left\{1-\frac{\sqrt{2\Cx\mu}}{\Mx\sqrt{L_{F}}+\sqrt{2\Cx\mu}},1-\frac{\sqrt{2\Cx\mu}}{\Mx\sqrt{L_{F}}+\sqrt{\frac{2\mu \Mx^2}{\Cx}}}\right\}\notag\\
		& =&1-\frac{\sqrt{2\Cx\mu}}{\Mx\sqrt{L_{F}}+\sqrt{\frac{2\mu \Mx^2}{\Cx}}}\notag.
	\end{eqnarray}
	This  gives a rate of 
	\begin{align*}
		T = \mathcal{O}\left(\frac{\Mx\sqrt{L_{F}}+\sqrt{\frac{2\mu \Mx^2}{\Cx}}}{\sqrt{2\Cx\mu}}\log\frac{1}{\varepsilon}\right)=	\mathcal{O}\left(\left(\frac{\Mx}{\Cx}+\sqrt{\frac{(L-\mu)\Mx}{2\mu \Cx}}\right)\log \frac{1}{\varepsilon} \right) .
	\end{align*}
\end{proof}

\clearpage

\section{Analysis of 5GCS}\label{proof35}
\begin{theorem} Consider Algorithm~\ref{alg:5GCS} (\gls{5GCS}) with the LT solver being \algname{\gls{GD}} run for $$\Kx \geq \left(\frac{3}{4}\sqrt{\frac{\Cx}{\Mx}\frac{L}{\mu}}+2\right)\log\left(4\frac{L}{\mu}\right)$$ iterations.
	Let $0<\gammaM\leq \frac{3}{16}\sqrt{\frac{\Cx}{L\mu \Mx}}$ and $\tauM=\frac{1}{2\gammaM \Mx}$. Then for the Lyapunov function
	\begin{equation*}
		\Psi^{\kstep}\eqdef \frac{1}{\gammaM}\sqnorm{x^{\kstep}-x^\star}+\frac{\Mx}{\Cx}\left(\frac{1}{\tauM}+\frac{1}{L_F}\right)\sqnorm{u^{\kstep}-u^\star},
	\end{equation*}
	the iterates of  the method satisfy
	$$		\Exp{\Psi^{\Tx}}\leq (1-\rho)^\Tx \Psi^0,
	$$	where 
	$
	\rho\eqdef  \max\left\{\frac{\gammaM\mu}{1+\gammaM\mu},\frac{\Cx}{\Mx}\frac{\tauM}{(L_F+\tauM)}\right\}<1.
	$

\end{theorem}
\begin{proof}
	Noting that updates for $u^{t+1}$ and $x^{t+1}$ can be written as
	\begin{eqnarray}
		&u^{t+1}\eqdef u^t + \frac{1}{1+\omega} \Rop^t\left(\bar{u}^{t+1}-u^t\right),&\\
		&x^{t+1}=\hat{x}^t-\gammaM\left(\omega+1\right)\Koper^\top\left(u^{t+1}-u^t\right)&\label{xupdate35}
	\end{eqnarray}
	where $\Rop^t$ is the Partial Participation operator, $\omega=\frac{\Mx}{\Cx}-1$ and  $\bar{u}^{t+1} = \nabla F(\lastlocitterk)$. 
	Using variance decomposition and Proposition 1 from \citep{condat2021murana}, we obtain
	\begin{eqnarray}
		\label{eq:starting_eq3.5}
	&&	\Exp{\sqnorm{x^{t+1}-x^\star}\;|\;\mathcal{F}^t}\\
    \notag&\overset{(\ref{vardec})}{=}&\sqnorm{\Exp{x^{t+1}\;|\;\mathcal{F}^t}-x^\star}+\Exp{\sqnorm{x^{t+1}-\Exp{x^{t+1}\;|\;\mathcal{F}^t}}\;|\;\mathcal{F}^t}\notag\\
		&\overset{(\ref{xupdate35})+(\ref{PP})}{=}& \underbrace{\sqnorm{\hat{x}^{t}-x^\star-\gammaM \Koper^\top(\bar{u}^{t+1}-u^t)}}_{X}+\gammaM^2\oma\sqnorm{\bar{u}^{t+1}-u^t}\notag\\
		&\quad &- \gammaM^2\zeta\sqnorm{\Koper^\top(\bar{u}^{t+1}-u^t)},
	\end{eqnarray}
	where
	\begin{equation*}
		\oma=\frac{\Mx(\Mx-\Cx)}{\Cx(\Mx-1)},\quad \zeta=\frac{\Mx-\Cx}{\Cx(\Mx-1)}.
	\end{equation*}
	Moreover, using \myref{fooc} and the definition of $\hat{x}^t$, we have
	\begin{align}
		&(1+\gammaM\mu)\hat{x}^t=x^t-\gammaM \Koper^\top u^{t},\label{opt1}\\
		&(1+\gammaM\mu)x^\star= x^\star -\gammaM \Koper^\top u^\star.\label{opt2}
	\end{align}
	Using \myref{opt1} and \myref{opt2} we obtain
	\begin{eqnarray}
		\label{eq:long3.5}
		X &=&	\sqnorm{\hat{x}^{t}-x^\star-\gammaM \Koper^\top(\bar{u}^{t+1}-u^t)}\notag\\
		&=&\sqnorm{\hat{x}^{t}-x^\star} +\gammaM^2\sqnorm{\Koper^\top (\bar{u}^{t+1}-u^t)}\notag\\
        &&-2\gammaM \left\langle 
		\hat{x}^{t}-x^\star,\Koper^\top(\bar{u}^{t+1}-u^t) \right\rangle  \notag\\
		&=& (1+\gammaM\mu) \sqnorm{\hat{x}^{t}-x^\star} +\gammaM^2\sqnorm{\Koper^\top(\bar{u}^{t+1}-u^t)}\notag\\
		&\quad&-2\gammaM  \left\langle 
		\hat{x}^{t}-x^\star,\Koper^\top (\bar{u}^{t+1}-u^\star) \right\rangle\notag\\
        &&+2\gammaM  \left\langle 
		\hat{x}^{t}-x^\star,\Koper^\top (u^{t}-u^\star) \right\rangle  -\gammaM\mu\sqnorm{\hat{x}^{t}-x^\star} \notag\\
		&\overset{(\ref{opt1})+(\ref{opt2})}{=}&   \left\langle  x^t-x^\star-\gammaM \Koper^\top (u^t-u^\star),\hat{x}^{t}-x^\star \right\rangle +\gammaM^2\sqnorm{\Koper^\top (\bar{u}^{t+1}-u^t)}\notag\\
		&\quad&-2\gammaM  \left\langle 
		\hat{x}^{t}-x^\star,\Koper^\top (\bar{u}^{t+1}-u^\star) \right\rangle\notag\\
        &&+  \left\langle 
		\hat{x}^{t}-x^\star,2\gammaM \Koper^\top (u^{t}-u^\star) \right\rangle  -\gammaM\mu\sqnorm{\hat{x}^{t}-x^\star} \notag.
			\end{eqnarray}
This leads to	
			\begin{eqnarray}
	X	&= &\left\langle  x^t-x^\star+\gammaM \Koper^\top (u^t-u^\star),\hat{x}^{t}-x^\star \right\rangle  \notag\\
		&\quad&+\gammaM^2\sqnorm{\Koper^\top (\bar{u}^{t+1}-u^t)}\notag\\
        &&-2\gammaM  \left\langle 
		\hat{x}^{t}-x^\star, \Koper^\top(\bar{u}^{t+1}-u^\star) \right\rangle -\gammaM\mu\sqnorm{\hat{x}^{t}-x^\star}\notag \\
		&\overset{(\ref{opt1})+(\ref{opt2})}{=}&\frac{1}{1+\gammaM\mu} \left\langle  x^t-x^\star+\gammaM \Koper^\top (u^t-u^\star),x^t-x^\star-\gammaM \Koper^\top (u^t-u^\star) \right\rangle  \notag\\
		&\quad&+\gammaM^2\sqnorm{\Koper^\top (\bar{u}^{t+1}-u^t)}\notag\\
        &&-2\gammaM  \left\langle 
		\hat{x}^{t}-x^\star, \Koper^\top (\bar{u}^{t+1}-u^\star) \right\rangle -\gammaM\mu\sqnorm{\hat{x}^{t}-x^\star} \notag\\
		&=& \frac{1}{1+\gammaM\mu}\sqnorm{x^t-x^\star}-\frac{\gammaM^2}{1+\gammaM\mu}\sqnorm{\Koper^\top (u^t-u^\star)}\notag\\
		&\quad&+\gammaM^2\sqnorm{\Koper^\top (\bar{u}^{t+1}-u^t)}\notag\\
        &&-2\gammaM  \left\langle 
		\hat{x}^{t}-x^\star,\Koper^\top (\bar{u}^{t+1}-u^\star) \right\rangle -\gammaM\mu\sqnorm{\hat{x}^{t}-x^\star}   . 
	\end{eqnarray}
	Combining \myref{eq:starting_eq3.5} and \myref{eq:long3.5}, we get
	\begin{eqnarray*}
		&&\Exp{\sqnorm{x^{t+1}-x^\star}\;|\;\mathcal{F}^t}\notag\\
        &\leq&  \frac{1}{1+\gammaM\mu}\sqnorm{x^t-x^\star}-\frac{\gammaM^2}{1+\gammaM\mu}\sqnorm{\Koper^\top (u^t-u^\star)}\\
		&\quad&+\gammaM^2(1-\zeta)\sqnorm{\Koper^\top (\bar{u}^{t+1}-u^t)}-2\gammaM  \left\langle 
		\hat{x}^{t}-x^\star,\Koper^\top (\bar{u}^{t+1}-u^\star) \right\rangle \\
		&\quad&+\gammaM^2\oma\sqnorm{\bar{u}^{t+1}-u^t}-\frac{\gammaM\mu}{\Mx}\sqnorm{\Koper\hat{x}^t-\Koper x^\star}.
	\end{eqnarray*}
	Note that we can have the update rule for $u$ as: 
	\begin{equation*}
		u^{t+1}\eqdef u^t + \frac{1}{1+\omega} \Rop^t\left(\bar{u}^{t+1}-u^t\right),
	\end{equation*}
	where $\Rop^t$ is Partial Participation operator with parameter  $\omega=\frac{\Mx}{\Cx}-1$. Using conic variance formula \myref{cvar} of $\Rop^t$, we obtain
	\begin{eqnarray}
		&&\Exp{\sqnorm{u^{t+1}-u^\star}\;|\;\mathcal{F}^t}\notag\\
        &\overset{(\ref{vardec})+(\ref{cvar})}{\leq}& \sqnorm{u^{t}-u^\star+\frac{1}{1+\omega}\left(\bar{u}^{t+1} -u^t\right)}
		+\frac{\omega}{(1+\omega)^2}\sqnorm{\bar{u}^{t+1} -u^t }\notag\\
		&=&\frac{\omega^2}{(1+\omega)^2}\sqnorm{u^{t}-u^\star}+\frac{1}{(1+\omega)^2}\sqnorm{\bar{u}^{t+1}-u^\star}\notag\\
		&\quad&+\frac{2\omega}{(1+\omega)^2} \left\langle  u^{t}-u^\star,
		\bar{u}^{t+1}-u^\star \right\rangle +\frac{\omega}{(1+\omega)^2}\sqnorm{\bar{u}^{t+1} -u^\star }\notag\\
		&\quad&+\frac{\omega}{(1+\omega)^2}\sqnorm{u^{t} -u^\star }-\frac{2\omega}{(1+\omega)^2} \left\langle  u^{t}-u^\star,
		\bar{u}^{t+1}-u^\star \right\rangle \notag\\
		&=&\frac{1}{1+\omega}\sqnorm{\bar{u}^{t+1} -u^\star }+\frac{\omega}{1+\omega}\sqnorm{u^{t} -u^\star }.\label{expu}
	\end{eqnarray}
	Let us consider the first term in \myref{expu}:
	\begin{eqnarray*}
		\sqnorm{\bar{u}^{t+1}-u^\star}&=&\sqnorm{(u^t-u^\star)+(\bar{u}^{t+1}-u^t)}\\
		&=&\sqnorm{u^t-u^\star}+\sqnorm{\bar{u}^{t+1}-u^t}+2 \left\langle  u^t-u^\star,\bar{u}^{t+1}-u^t \right\rangle \\
		&=&\sqnorm{u^t-u^\star}+2  \left\langle  \bar{u}^{t+1}-u^\star,\bar{u}^{t+1}-u^t \right\rangle  - \sqnorm{\bar{u}^{t+1}-u^t}.
	\end{eqnarray*}
	Combining the terms together, we get
	\begin{eqnarray*}
		\Exp{\sqnorm{u^{t+1}-u^\star}\;|\;\mathcal{F}^t}&\leq &\sqnorm{u^{t} -u^\star }\\
        &&+\frac{1}{1+\omega}\left(2  \left\langle  \bar{u}^{t+1}-u^\star,\bar{u}^{t+1}-u^t \right\rangle  - \sqnorm{\bar{u}^{t+1}-u^t}\right).
	\end{eqnarray*}
	Finally, we obtain
	\begin{eqnarray*}
		\frac{1}{\gammaM}\Exp{\sqnorm{x^{t+1}-x^\star}\;|\;\mathcal{F}^t}&+&\frac{1+\omega}{\tauM}\Exp{\sqnorm{u^{t+1}-u^\star}\;|\;\mathcal{F}^t}\\
		&\leq&  \frac{1}{\gammaM(1+\gammaM\mu)}\sqnorm{x^t-x^\star}-\frac{\gammaM}{1+\gammaM\mu}\sqnorm{\Koper^\top (u^t-u^\star)}\\
		&\quad&+\gammaM(1-\zeta)\sqnorm{\Koper^\top (\bar{u}^{t+1}-u^t)}\\
		&\quad&+\gammaM\oma\sqnorm{\bar{u}^{t+1}-u^t}-\frac{\mu}{\Mx}\sqnorm{\Koper \hat{x}^t- \Koper x^\star}\\
		&\quad&+\frac{1+\omega}{\tauM}\sqnorm{u^{t} -u^\star }-2  \left\langle 
		\hat{x}^{t}-x^\star,\Koper^\top (\bar{u}^{t+1}-u^\star) \right\rangle \\ &\quad&+\frac{1}{\tauM}\left(2  \left\langle  \bar{u}^{t+1}-u^\star,\bar{u}^{t+1}-u^t \right\rangle  - \sqnorm{\bar{u}^{t+1}-u^t}\right).
	\end{eqnarray*}
	Ignoring $-\frac{\gammaM}{1+\gammaM\mu}\sqnorm{\Koper^\top (u^t-u^\star)}$ and noting that
	\begin{eqnarray*}
		&&-\left\langle \hat{x}^{t}-x^\star,\Koper^\top (\bar{u}^{t+1}-u^\star) \right\rangle\notag\\ &+&\frac{1}{\tauM}\left\langle  \bar{u}^{t+1}-u^\star,\bar{u}^{t+1}-u^t \right\rangle  \\
		&=&-\left\langle \lastlocitterk-\Koper x^\star,\bar{u}^{t+1}-u^\star \right\rangle +\frac{1}{\tauM}\left\langle  \nabla\localfun (\lastlocitterk),\bar{u}^{t+1}-u^\star \right\rangle\\
		&\overset{(\ref{yi1})+(\ref{strmono})}{\leq}& -\frac{1}{L_F}\sqnorm{\bar{u}^{t+1}-u^\star}+\frac{a}{2\tauM}\sqnorm{\nabla\localfun (\lastlocitterk)}+\frac{1}{2a\tauM}\sqnorm{\bar{u}^{t+1}-u^\star }\\
		&=& -\left(\frac{1}{L_F}-\frac{1}{2a\tauM}\right)\sqnorm{\bar{u}^{t+1}-u^\star}+\frac{a}{2\tauM}\sqnorm{\nabla\localfun (\lastlocitterk)}\\
		&\overset{(\ref{expu})}{\leq}& -\left(\frac{1}{L_F}-\frac{1}{2a\tauM}\right)\left((1+\omega)\Exp{\sqnorm{u^{t+1}-u^\star}\;|\;\mathcal{F}^t}-\omega\sqnorm{u^t-u^\star}\right)\\
		&\quad&+\frac{a}{2\tauM}\sqnorm{\nabla\localfun (\lastlocitterk)},
	\end{eqnarray*}
	we get
	\begin{eqnarray*}
		\frac{1}{\gammaM}\Exp{\sqnorm{x^{t+1}-x^\star}\;|\;\mathcal{F}^t}&+&\left(1+\omega\right)\left(\frac{1}{\tauM}+\frac{1}{L_F}\right)\Exp{\sqnorm{u^{t+1}-u^\star}\;|\;\mathcal{F}^t}\\
		&\leq&  \frac{1}{\gammaM(1+\gammaM\mu)}\sqnorm{x^t-x^\star}\\
		&\quad&+\left(1+\omega\right)\left(\frac{1}{\tauM}+\frac{\omega}{1+\omega}\frac{1}{L_F}\right)\sqnorm{u^{t} -u^\star } \\
		&\quad&+\left(\gammaM\left(1-\zeta\right)\Mx+\gammaM\oma-\frac{1}{\tauM}\right)\sqnorm{\bar{u}^{t+1}-u^t}\\
		&\quad& +\frac{L_F}{\tauM^2}\sqnorm{\nabla\localfun (\lastlocitterk)}-\frac{\mu}{\Mx}\sqnorm{\Koper\hat{x}^t-\Koper x^\star}.
	\end{eqnarray*}
	Where we made the choice $a=\frac{L_F}{\tau}$.
	Using Young's inequality we have
	\begin{equation*}
		-\frac{\mu}{3\Mx}\sqnorm{\Koper \hat{x}^t-\localsolk+\localsolk-\Koper x^\star}\overset{(\ref{yi3})}{\leq}\frac{\mu}{3\Mx}\sqnorm{\localsolk-\Koper x^\star}-\frac{\mu}{6\Mx}\sqnorm{\Koper\hat{x}^t-\localsolk}.	\end{equation*}
	Noting the fact that $\localsolk=\Koper\hat{x}^t-\frac{1}{\tauM}(\hat{u}^{t+1}-u^t)$, we have
	$$\frac{\mu}{3\Mx}\sqnorm{\localsolk-\Koper x^\star}\overset{(\ref{yi2})}{\leq} 2\frac{\mu}{3\Mx}\sqnorm{\Koper\hat{x}^t-\Koper x^\star}+\frac{2}{\tauM^2}\frac{\mu}{3\Mx}\sqnorm{\hat{u}^{t+1}-u^t}.$$
	Combining those inequalities, we get
	\begin{eqnarray*}
		\frac{1}{\gammaM}\Exp{\sqnorm{x^{t+1}-x^\star}\;|\;\mathcal{F}^t}&+&\left(1+\omega\right)\left(\frac{1}{\tauM}+\frac{1}{L_F}\right)\Exp{\sqnorm{u^{t+1}-u^\star}\;|\;\mathcal{F}^t}\\
		&\leq&  \frac{1}{\gammaM(1+\gammaM\mu)}\sqnorm{x^t-x^\star}\\
		&\quad&+\left(1+\omega\right)\left(\frac{1}{\tauM}+\frac{\omega}{1+\omega}\frac{1}{L_F}\right)\sqnorm{u^{t} -u^\star } \\
		&\quad&+\frac{2}{\tauM^2}\frac{\mu}{3\Mx}\sqnorm{\hat{u}^{t+1}-u^t}\\
		&\quad&-\left(\frac{1}{\tauM}-\left(\gammaM\left(1-\zeta\right)\Mx+\gammaM\oma\right)\right)\sqnorm{\bar{u}^{t+1}-u^t}\\
		&\quad&+\frac{L_F}{\tauM^2}\sqnorm{\nabla\localfun (\lastlocitterk)}-\frac{\mu}{6\Mx}\sqnorm{\Koper\hat{x}^t-\localsolk}.
	\end{eqnarray*}
	Assuming $\gammaM$ and $\tauM$ can be chosen so that $\frac{1}{\tauM}-(\gammaM(1-\zeta)\Mx+\gammaM\oma))\geq \frac{4}{\tauM^2}\frac{\mu}{3\Mx}$ we obtain
	\begin{eqnarray*}
		\frac{1}{\gammaM}\Exp{\sqnorm{x^{t+1}-x^\star}\;|\;\mathcal{F}^t}&+&\left(1+\omega\right)\left(\frac{1}{\tauM}+\frac{1}{L_F}\right)\Exp{\sqnorm{u^{t+1}-u^\star}\;|\;\mathcal{F}^t}\\
		&\leq&  \frac{1}{\gammaM(1+\gammaM\mu)}\sqnorm{x^t-x^\star}\\
		&\quad&+\left(1+\omega\right)\left(\frac{1}{\tauM}+\frac{\omega}{1+\omega}\frac{1}{L_F}\right)\sqnorm{u^{t} -u^\star } \\
		&\quad&+\frac{4}{\tauM^2}\frac{\mu L_F^2}{3\Mx}\sqnorm{\lastlocitterk-\localsolk}+\frac{L_F}{\tauM^2}\sqnorm{\nabla\localfun (\lastlocitterk)}\\
		&\quad&-\frac{\mu}{6\Mx}\sqnorm{\Koper\hat{x}^t-\localsolk}.
	\end{eqnarray*}
	Where the point $\lastlocitterk$ is assumed to satisfy 
	\begin{eqnarray*}
		\frac{4}{\tauM^2}\frac{\mu L_F^2}{3\Mx}\sqnorm{\lastlocitterk-\localsolk}+\frac{L_F}{\tauM^2}\sqnorm{\nabla\localfun (\lastlocitterk)}\leq\frac{\mu}{6\Mx}\sqnorm{\Koper\hat{x}^t-\localsolk}.
	\end{eqnarray*}
	Thus
	\begin{eqnarray*}
		\frac{1}{\gammaM}\Exp{\sqnorm{x^{t+1}-x^\star}\;|\;\mathcal{F}^t}&+&\left(1+\omega\right)\left(\frac{1}{\tauM}+\frac{1}{L_F}\right)\Exp{\sqnorm{u^{t+1}-u^\star}\;|\;\mathcal{F}^t}\\
		&\leq&  \frac{1}{\gammaM(1+\gammaM\mu)}\sqnorm{x^t-x^\star}\\
		&\quad&+\left(1+\omega\right)\left(\frac{1}{\tauM}+\frac{\omega}{1+\omega}\frac{1}{L_F}\right)\sqnorm{u^{t} -u^\star }.
	\end{eqnarray*}
	
	By taking the expectation on both sides we get
	\begin{equation*}
		\Exp{\Psi^{t+1}}\leq  \max\left\{\frac{1}{1+\gammaM\mu} ,\frac{ L_F+\frac{\Mx-\Cx}{\Cx}\tauM}{L_F+\tauM} \right\}\Exp{\Psi^{t}},
	\end{equation*}
	which finishes the proof. 
	Note that our standard choice of constants is $$\omega=\frac{\Mx}{\Cx}-1,\quad \oma=\frac{\Mx(\Mx-\Cx)}{\Cx(\Mx-1)},\quad \zeta=\frac{\Mx-\Cx}{\Cx(\Mx-1)}.$$ Using these parameters the requirement for stepsizes becomes:$$\frac{1}{\tauM}-\gammaM \Mx\geq \frac{4\mu}{3\Mx\tauM^2}.$$
	This inequality is satisfied, when $0<\gammaM\leq \frac{3}{16}\sqrt{\frac{\Cx}{L\mu \Mx}}$ and $\tauM=\frac{1}{2\Mx\gammaM}.$
\end{proof}
\subsection{Proof of Corollary \ref{cor:5GCS}}

\begin{corollary}
Choose any $0<\varepsilon<1$ and $\gammaM = \frac{3}{16}\sqrt{\frac{\Cx}{L\mu \Mx}}$. In order to guarantee $\Exp{\Psi^{\Tx}}\leq \varepsilon \Psi^0$, it suffices to take 
\begin{eqnarray}
	T &\geq	&
	\max\left\{1+\frac{16}{3}\sqrt{\frac{\Mx}{\Cx}\frac{L}{\mu}},\frac{\Mx}{\Cx}+\frac{3}{8}\sqrt{\frac{\Mx}{\Cx}\frac{L}{\mu} }\right\}\log\frac{1}{\varepsilon} =
	\tilde{\cO}\left(\frac{\Mx}{\Cx}+\sqrt{\frac{\Mx}{\Cx}\frac{L}{\mu }}\right) \notag
\end{eqnarray}
communication rounds.
\end{corollary}
\begin{proof}
	Choosing the maximal $\gammaM=\frac{3}{16}\sqrt{\frac{\Cx}{L\mu \Mx}}$ and $a=\frac{L_F}{\tauM}$ we have 
	\begin{eqnarray*}
		\max\left\{\frac{1}{1+\gammaM\mu},\frac{\frac{1}{\tauM}+\frac{\Mx-\Cx}{\Mx}\frac{1}{L_F}}{\frac{1}{\tauM}+\frac{1}{L_F}}\right\}&=&\max\left\{\frac{1}{1+\frac{3}{16}\sqrt{\frac{\mu \Cx}{L \Mx}}},\frac{\frac{1}{\tauM}+\frac{\Mx-\Cx}{\Mx}\frac{1}{L_F}}{\frac{1}{\tauM}+\frac{1}{L_F}}\right\}\\
		&=&\max\left\{\frac{1}{1+\frac{3}{16}\sqrt{\frac{\mu \Cx}{L \Mx}}},1-\frac{\frac{8\Cx}{3\Mx L_F}\sqrt{\frac{L\mu}{\Mx \Cx}}}{1+\frac{8\Mx}{3\Mx L_F}\sqrt{\frac{L\mu}{\Mx \Cx}}}\right\}\\
		&\leq& \max\left\{\frac{1}{1+\frac{3}{16}\sqrt{\frac{\mu \Cx}{L \Mx}}},1-\frac{\frac{8}{3}\sqrt{\frac{\Cx\mu}{\Mx L}}}{1+\frac{8}{3}\sqrt{\frac{\Mx\mu}{L\Cx}}}\right\}.
	\end{eqnarray*}
	Thus Algorithm~\ref{alg:5GCS} finds $\varepsilon$-solution in:
	\begin{equation*}
		T\geq\mathcal{O}\left(\max\left\{1+\frac{16}{3}\sqrt{\frac{L\Mx}{\mu \Cx}},\frac{\Mx}{\Cx}+\frac{3}{8}\sqrt{\frac{L\Mx}{\mu \Cx}}\right\}\log\frac{1}{\varepsilon}\right)
	\end{equation*}
	communications.	
\end{proof}

\clearpage
\section{Analysis of 5GCS\texorpdfstring{${}_0$}{0}}
\subsection{Proof of Theorem~\ref{thm:5GCS-0}}
\begin{algorithm*}
	\caption{\gls{5GCS} with 0 local \algname{GD} steps }
	\begin{algorithmic}[1]\label{alg:5GCS-0}
		\STATE  \textbf{input:} initial points $x^0\in\mathbb{R}^d$, $u_\iclient^0\in\mathbb{R}^d$ for all $\iclient=\{1,\dots,\Mx\}$; 
		\STATE stepsize $\gammaM>0$, $\tauM>0$
		\STATE $v^0\eqdef \sum_{\iclient=1}^\Mx u_\iclient^0$
		\FOR{$\kstep=0, 1, \ldots$}
		\STATE $\hat{x}^{\kstep} \eqdef \frac{1}{1+\gammaM\mu} \left(x^\kstep - \gammaM v^\kstep\right)$
		\STATE Pick $\set\subset \{1,\ldots,\Mx\}$ of size $\Cx$ uniformly at random
		\FOR{$\iclient\in \set$}
		\STATE $u_\iclient^{\kstep+1}\eqdef \nabla F_\iclient(\hat{x}^\kstep)=\frac{1}{\Mx}\left(\nabla f_\iclient(\hat{x}^\kstep)-\mu\hat{x}^\kstep\right)$
		\ENDFOR
		\FOR{$\iclient\in\{1, \ldots,\Mx\}\backslash \set$}
		\STATE $u_{\iclient}^{\kstep+1}\eqdef u_{\iclient}^\kstep $
		\ENDFOR
		\STATE $v^{\kstep+1}\eqdef \sum_{\iclient=1}^\Mx u_\iclient^{\kstep+1}$
		\STATE  $x^{\kstep+1} \eqdef \hat{x}^{\kstep}- \gammaM \frac{\Mx}{\Cx}(v^{\kstep+1}-v^\kstep)$
		\ENDFOR
	\end{algorithmic}
\end{algorithm*}

\begin{theorem}
	Consider Algorithm~\ref{alg:5GCS} (\gls{5GCS}) with the LT solver being \algname{\gls{GD}} run for $\Kx=0$ iterations (this is equivalent to Algorithm~\ref{alg:5GCS-0}; we shall also call the method \algname{5GCS${}_0$}).	Let $0<\gammaM\leq \frac{\Cx}{4L\Mx}$. Then for the Lyapunov function
	\begin{equation*}
		\Psi^{\kstep}\eqdef  \frac{\Cx}{\Mx^2\gammaM^2}\left(1-\sqrt{\frac{\gammaM \Mx L_F}{2}}\right)\sqnorm{x^{\kstep}-x^\star}+\sqnorm{u^{\kstep}-u^\star},
	\end{equation*}
	the iterates of the method satisfy
	$$		\Exp{\Psi^{\Tx}}\leq \left(1-\rho\right)^\Tx \Psi^0,
	$$	
	where 
	$
	\rho\eqdef  \min\left(\frac{\gammaM\mu}{1+\gammaM\mu},\frac{\Cx}{\Mx+2\gammaM  L_F\Mx^2}\right)<1.
	$

\end{theorem}
\begin{proof}
	Noting that updates for $u^{t+1}$ and $x^{t+1}$ can be written as
	\begin{eqnarray}
		&u^{t+1}\eqdef u^t + \frac{1}{1+\omega} \Rop^t\left(\bar{u}^{t+1}-u^t\right),&\\
		&x^{t+1}=\hat{x}^t-\gammaM\left(\omega+1\right)\Koper^\top\left(u^{t+1}-u^t\right)&\label{xupdate33}
	\end{eqnarray}
	where $\Rop^t$ is a Partial Participation operator, $\omega=\frac{\Mx}{\Cx}-1$ and  $\bar{u}^{t+1} = \nabla F(\Koper\hat{x}^t)$. 
	Using variance decomposition and Proposition 1 from \citep{condat2021murana}, we obtain
	\begin{eqnarray}
		\label{eq:starting_eq3.3}
		&&\Exp{\sqnorm{x^{t+1}-x^\star}\;|\;\mathcal{F}^t}\notag\\
        &\overset{(\ref{vardec})}{=}&\sqnorm{\Exp{x^{t+1}\;|\;\mathcal{F}^t}-x^\star}+\Exp{\sqnorm{x^{t+1}-\Exp{x^{t+1}\;|\;\mathcal{F}^t}}\;|\;\mathcal{F}^t}\notag\\
		&\overset{(\ref{xupdate33})+(\ref{PP})}{=}& \underbrace{\sqnorm{\hat{x}^{t}-x^\star-\gammaM \Koper^\top(\bar{u}^{t+1}-u^t)}}_{X}+\gammaM^2\oma\sqnorm{\bar{u}^{t+1}-u^t}\notag\\
		&\quad &- \gammaM^2\zeta\sqnorm{\Koper^\top(\bar{u}^{t+1}-u^t)},
	\end{eqnarray}
	where
	\begin{equation*}
		\oma=\frac{\Mx(\Mx-\Cx)}{\Cx(\Mx-1)},\quad \zeta=\frac{\Mx-\Cx}{\Cx(\Mx-1)}.
	\end{equation*}
	Moreover, using \myref{fooc} and the definition of $\hat{x}^t$, we have
	\begin{align}
		&(1+\gammaM\mu)\hat{x}^t=x^t-\gammaM \Koper^\top u^{t},\label{opt12}\\
		&(1+\gammaM\mu)x^\star= x^\star -\gammaM \Koper^\top u^\star.\label{opt22}
	\end{align}
	Using \myref{opt12} and \myref{opt22} we obtain
	\begin{eqnarray}\label{eq:long3.3}
		X &=&	\sqnorm{\hat{x}^{t}-x^\star-\gammaM \Koper^\top(\bar{u}^{t+1}-u^t)}\notag\\
		&=&\sqnorm{\hat{x}^{t}-x^\star} +\gammaM^2\sqnorm{\Koper^\top (\bar{u}^{t+1}-u^t)}\notag\\
		&\quad&-2\gammaM \left\langle 
		\hat{x}^{t}-x^\star,\Koper^\top(\bar{u}^{t+1}-u^t) \right\rangle  \notag\\
		&\leq& (1+\gammaM\mu) \sqnorm{\hat{x}^{t}-x^\star} +\gammaM^2\sqnorm{\Koper^\top(\bar{u}^{t+1}-u^t)}\notag\\
		&\quad&-2\gammaM  \left\langle 
		\hat{x}^{t}-x^\star,\Koper^\top (\bar{u}^{t+1}-u^\star) \right\rangle\notag\\
        &&+2\gammaM  \left\langle 
		\hat{x}^{t}-x^\star,\Koper^\top (u^{t}-u^\star) \right\rangle 
        \end{eqnarray}
Next, we get
        \begin{eqnarray}
        \label{eq:long3.3_sec}
	X	&\overset{(\ref{opt12})+(\ref{opt22})}{=}&   \left\langle  x^t-x^\star-\gammaM \Koper^\top (u^t-u^\star),\hat{x}^{t}-x^\star \right\rangle +\gammaM^2\sqnorm{\Koper^\top (\bar{u}^{t+1}-u^t)}\notag\\
		&\quad&-2\gammaM  \left\langle 
		\hat{x}^{t}-x^\star,\Koper^\top (\bar{u}^{t+1}-u^\star) \right\rangle  +  \left\langle 
		\hat{x}^{t}-x^\star,2\gammaM \Koper^\top (u^{t}-u^\star) \right\rangle  \notag\\
		&= &\left\langle  x^t-x^\star+\gammaM \Koper^\top (u^t-u^\star),\hat{x}^{t}-x^\star \right\rangle  \notag\\
		&\quad&+\gammaM^2\sqnorm{\Koper^\top (\bar{u}^{t+1}-u^t)}-2\gammaM  \left\langle 
		\hat{x}^{t}-x^\star, \Koper^\top(\bar{u}^{t+1}-u^\star) \right\rangle  \notag\\
		&\overset{(\ref{opt12})+(\ref{opt22})}{=}&\frac{1}{1+\gammaM\mu} \left\langle  x^t-x^\star+\gammaM \Koper^\top (u^t-u^\star),x^t-x^\star-\gammaM \Koper^\top (u^t-u^\star) \right\rangle  \notag\\
		&\quad&+\gammaM^2\sqnorm{\Koper^\top (\bar{u}^{t+1}-u^t)}-2\gammaM  \left\langle 
		\hat{x}^{t}-x^\star, \Koper^\top (\bar{u}^{t+1}-u^\star) \right\rangle  \notag\\
		&=& \frac{1}{1+\gammaM\mu}\sqnorm{x^t-x^\star}-\frac{\gammaM^2}{1+\gammaM\mu}\sqnorm{\Koper^\top (u^t-u^\star)}\notag\\
		&\quad&+\gammaM^2\sqnorm{\Koper^\top (\bar{u}^{t+1}-u^t)}-2\gammaM  \left\langle 
		\hat{x}^{t}-x^\star,\Koper^\top (\bar{u}^{t+1}-u^\star) \right\rangle.
	\end{eqnarray}
	Combining \myref{eq:starting_eq3.3} and \myref{eq:long3.3} we have
	\begin{eqnarray}
		\Exp{\sqnorm{x^{t+1}-x^\star}\;|\;\mathcal{F}^t}&\leq&  \frac{1}{1+\gammaM\mu}\sqnorm{x^t-x^\star}-\frac{\gammaM^2}{1+\gammaM\mu}\sqnorm{\Koper^\top (u^t-u^\star)}\notag\\
		&\quad&+\gammaM^2(1-\zeta)\sqnorm{\Koper^\top (\bar{u}^{t+1}-u^t)}\notag\\
        &&-2\gammaM  \left\langle 
		\hat{x}^{t}-x^\star,\Koper^\top (\bar{u}^{t+1}-u^\star) \right\rangle \notag\\
		&\quad&+\gammaM^2\oma\sqnorm{\bar{u}^{t+1}-u^t}.\label{xbnd35}
	\end{eqnarray}
	On the other hand using the variance decomposition and conic variance of $\Rop^t$ 
	\begin{eqnarray}
		&&\Exp{\sqnorm{u^{t+1}-u^\star}\;|\;\mathcal{F}^t}\notag\\&\overset{(\ref{vardec})+(\ref{cvar})}{\leq}& \sqnorm{u^{t}-u^\star+\frac{1}{1+\omega}\left(\bar{u}^{t+1} -u^t\right)}\notag\\
        &&+\frac{\omega}{(1+\omega)^2}\sqnorm{\bar{u}^{t+1} -u^t }\notag\\ 
		&=& \sqnorm{\frac{\omega}{1+\omega}(u^{t}-u^\star)+\frac{1}{1+\omega}\left(\bar{u}^{t+1} -u^\star\right)}\notag\\
		&&+\frac{\omega}{(1+\omega)^2}\sqnorm{\bar{u}^{t+1} - u^\star - (u^t - u^\star) }.\label{ubound33_prev}
			\end{eqnarray}
            Next, we obtain
            \begin{eqnarray}
		&&\Exp{\sqnorm{u^{t+1}-u^\star}\;|\;\mathcal{F}^t}\notag\\&=&\frac{\omega^2}{(1+\omega)^2}\sqnorm{u^{t}-u^\star}+\frac{1}{(1+\omega)^2}\sqnorm{\bar{u}^{t+1}-u^\star}\notag\\
		&\quad&+\frac{2\omega}{(1+\omega)^2} \left\langle  u^{t}-u^\star,
		\bar{u}^{t+1}-u^\star \right\rangle +\frac{\omega}{(1+\omega)^2}\sqnorm{\bar{u}^{t+1} -u^\star }\notag\\
		&\quad&+\frac{\omega}{(1+\omega)^2}\sqnorm{u^{t} -u^\star }-\frac{2\omega}{(1+\omega)^2} \left\langle  u^{t}-u^\star,
		\bar{u}^{t+1}-u^\star \right\rangle\notag \\
		&=&\frac{1}{1+\omega}\sqnorm{\bar{u}^{t+1} -u^\star }+\frac{\omega}{1+\omega}\sqnorm{u^{t} -u^\star },\label{ubound33}
	\end{eqnarray}
	where
	\begin{eqnarray}
		\sqnorm{\bar{u}^{t+1}-u^\star}&=&\sqnorm{(u^t-u^\star)+(\bar{u}^{t+1}-u^t)}\notag\\
		&=&\sqnorm{u^t-u^\star}+\sqnorm{\bar{u}^{t+1}-u^t}\notag\\
        &&+2 \left\langle  u^t-u^\star,\bar{u}^{t+1}-u^t \right\rangle \notag\\
		&=&\sqnorm{u^t-u^\star}\notag\\
        &&+2  \left\langle  \bar{u}^{t+1}-u^\star,\bar{u}^{t+1}-u^t \right\rangle  - \sqnorm{\bar{u}^{t+1}-u^t}.\label{ueq}
	\end{eqnarray}
	Combining \myref{ubound33} and \myref{ueq}, we get
	\begin{eqnarray}
		\Exp{\sqnorm{u^{t+1}-u^\star}\;|\;\mathcal{F}^t}&\leq &\sqnorm{u^{t} -u^\star }\notag\\
        &&+\frac{2}{1+\omega}  \left\langle  \bar{u}^{t+1}-u^\star,\bar{u}^{t+1}-u^t \right\rangle  \notag\\
                &&-\frac{1}{1+\omega}   \sqnorm{\bar{u}^{t+1}-u^t}.\label{ubound35}	    
	\end{eqnarray}
	Now let $c>0$ and combine \myref{xbnd35} with \myref{ubound35} to get
	\begin{eqnarray*}
		&&c\Exp{\sqnorm{x^{t+1}-x^\star}\;|\;\mathcal{F}^t}+\Exp{\sqnorm{u^{t+1}-u^\star}\;|\;\mathcal{F}^t}\\
		&\leq&  \frac{c}{1+\gammaM\mu}\sqnorm{x^t-x^\star}+c\gammaM^2(1-\zeta)\sqnorm{\Koper^\top(\bar{u}^{t+1}-u^t)}\\
		&\quad& -2c\gammaM  \left\langle \Koper(\hat{x}^{t}-x^\star),\bar{u}^{t+1}-u^\star \right\rangle+c\gammaM^2\oma\sqnorm{\bar{u}^{t+1}-u^t}\\
		&\quad&+ \sqnorm{u^{t} -u^\star }+\frac{1}{1+\omega}\left(2  \left\langle  \bar{u}^{t+1}-u^\star,\bar{u}^{t+1}-u^t \right\rangle  - \sqnorm{\bar{u}^{t+1}-u^t}\right)\\
		&\overset{(\ref{yi1})}{\leq}& \frac{c}{1+\gammaM\mu}\sqnorm{x^t-x^\star}+c\gammaM^2(1-\zeta)\sqnorm{\Koper^\top(\bar{u}^{t+1}-u^t)}\\
		&\quad& -\frac{2c\gammaM}{L_F}  \sqnorm{\bar{u}^{t+1}-u^\star }+c\gammaM^2\oma\sqnorm{\bar{u}^{t+1}-u^t}+ \sqnorm{u^{t} -u^\star }\\
		&\quad&+\frac{1}{1+\omega}\left(a\sqnorm{ \bar{u}^{t+1}-u^\star}+\frac{1}{a}\sqnorm{\bar{u}^{t+1}-u^t} - \sqnorm{\bar{u}^{t+1}-u^t}\right)\\
		&\leq &\frac{c}{1+\gammaM\mu}\sqnorm{x^t-x^\star}+ \sqnorm{u^{t} -u^\star }\\
		&\quad& +\left(c\gammaM^2\left(1-\zeta\right)\Mx+c\gammaM^2\oma+\frac{1}{1+\omega}\left(\frac{1}{a}-1\right)\right)\sqnorm{\bar{u}^{t+1}-u^t}\\
		&\quad&-\left(\frac{2c\gammaM}{L_F}-\frac{1}{1+\omega}a\right)  \sqnorm{\bar{u}^{t+1}-u^\star }.
	\end{eqnarray*}
	Using \myref{ubound33} and assuming $a$,$c$ and $\gammaM$ can be chosen so that $\frac{2c\gammaM}{L_F}-\frac{1}{1+\omega}a\geq 0 $
	\begin{eqnarray*}
		&&c\Exp{\sqnorm{x^{t+1}-x^\star}\;|\;\mathcal{F}^t}+\left(1+\left(1+\omega\right)\frac{2c\gammaM}{L_F}-a\right)\Exp{\sqnorm{u^{t+1}-u^\star}\;|\;\mathcal{F}^t}\\
		&\leq &\frac{c}{1+\gammaM\mu}\sqnorm{x^t-x^\star}+\left(1+\omega\left(\frac{2c\gammaM}{L_F}-\frac{1}{1+\omega}a\right)\right) \sqnorm{u^{t} -u^\star }\\
		&\quad &+\left(c\gammaM^2\left(1-\zeta\right)\Mx+c\gammaM^2\oma+\frac{1}{1+\omega}\left(\frac{1}{a}-1\right)\right)\sqnorm{\bar{u}^{t+1}-u^t}.
	\end{eqnarray*}
	In our case we have
	
	\begin{equation}
		\omega=\frac{\Mx}{\Cx}-1,\quad \oma=\frac{\Mx(\Mx-\Cx)}{\Cx(\Mx-1)},\quad \zeta=\frac{\Mx-\Cx}{\Cx(\Mx-1)},
	\end{equation}

	the term next to $\sqnorm{\hat{u}^{t+1}-u^t}$ becomes
	\begin{equation*}
		c\gammaM^2\Mx+\frac{\Cx}{\Mx}\left(\frac{1}{a}-1\right),
	\end{equation*}
	to get rid of it, we set
	\begin{equation*}
		c=\frac{\frac{\Cx}{\Mx}\left(1-\frac{1}{a}\right)}{\gammaM^2\Mx}\quad ,\quad a\geq1.
	\end{equation*}
	An $a$ that maximizes the contraction on $\Exp{\sqnorm{u^{t+1}-u^\star}\;|\;\mathcal{F}^t}$ is given by $a=\sqrt{\frac{2}{\gammaM \Mx L_F}}$, thus we need $\gammaM\leq\frac{2}{\Mx L_F}$ and
	\begin{equation*}
		\frac{1}{\gammaM L_F \Mx}-\sqrt{\frac{2}{L_F\gammaM \Mx}}>0.
	\end{equation*} 
	Thus we need $\gammaM<\frac{1}{2\Mx L_F}$
	and we can write a contraction constant of Lyapunov function as
	\begin{eqnarray*}
	C &=	&\max\left\{\frac{1}{1+\gammaM\mu},\frac{1+\omega(\frac{2c\gammaM}{L_F}-\frac{1}{1+\omega}a)}{1+(1+\omega)(\frac{2c\gammaM}{L_F}-\frac{1}{1+\omega}a)}\right\}\\
    &=&\max\left\{\frac{1}{1+\gammaM\mu},\frac{1+\frac{\Mx-\Cx}{\Cx}\left(\frac{2\Cx}{\gammaM L_F \Mx^2}-\frac{2\Cx}{\Mx}\sqrt{\frac{2}{L_F\gammaM \Mx}}\right)}{1+\frac{\Mx}{\Cx}\left(\frac{2\Cx}{\gammaM L_F \Mx^2}-\frac{2\Cx}{\Mx}\sqrt{\frac{2}{L_F\gammaM \Mx}}\right)}\right\}.
	\end{eqnarray*}
\end{proof}
\subsection{Proof of Corollary~\ref{cor:5GCS-0}}

\begin{corollary}
	Choose any $0<\varepsilon<1$ and
	$\gammaM=\frac{\Cx}{4 L \Mx}$. In order to guarantee $\Exp{\Psi^{\Tx}}\leq \varepsilon \Psi^0$, it suffices to take
	\[
		\Tx \geq 	\max\left\{1+\frac{4\Mx}{\Cx}\frac{L}{\mu},\frac{\Mx}{\Cx}+\frac{L_F\Mx}{L}\right\}\log \frac{1}{\varepsilon}= \tilde{O}\left(\frac{\Mx}{\Cx}\frac{L}{\mu}\right)
	\]
	communication rounds.
\end{corollary}
\begin{proof}
	If we let $\gammaM=\frac{\Cx}{4 L_F \Mx^2}\theta$, where $\theta=\frac{\Mx L_F}{L}$ then
	\begin{eqnarray*}
	C&=	&\max\left\{\frac{1}{1+\gammaM\mu},\frac{1+\frac{\Mx-\Cx}{\Cx}\left(\frac{2\Cx}{\gammaM L_F \Mx^2}-\frac{2\Cx}{\Mx}\sqrt{\frac{2}{L_F\gammaM \Mx}}\right)}{1+\frac{\Mx}{\Cx}\left(\frac{2\Cx}{\gammaM L_F \Mx^2}-\frac{2\Cx}{\Mx}\sqrt{\frac{2}{L_F\gammaM \Mx}}\right)}\right\}\\
		&=&\max\left\{\frac{1}{1+\frac{\Cx}{4 L_F \Mx^2}\mu},\frac{1+\frac{\Mx-\Cx}{\Cx}\left(8\frac{1}{\theta}-8\sqrt{\frac{\Cx}{2\Mx\theta}}\right)}{1+\frac{\Mx}{\Cx}\left(8\frac{1}{\theta}-8\sqrt{\frac{\Cx}{2\Mx\theta}}\right)}\right\}\\
		&\leq&\max\left\{\frac{1}{1+\frac{\Cx}{4 L_F \Mx^2}\mu},1-\frac{8-8\sqrt{\frac{1}{2}}}{\theta+\frac{\Mx}{\Cx}\left(8-8\sqrt{\frac{1}{2}}\right)}\right\}\\
		&\leq &\max\left\{\frac{1}{1+\frac{\mu \Cx}{4 L\Mx}},1-\frac{2\Cx}{2\Mx+2\Cx\frac{\Mx L_F}{L}}\right\} \\
		&\leq&\frac{1}{1+\mathcal{O}\left(\frac{\mu \Cx}{L\Mx}\right)}.
	\end{eqnarray*}
	Thus Algorithm~\ref{alg:5GCS-0} finds $\varepsilon$-solution in 
	$$T = \mathcal{O}\left(\frac{\Mx L}{\Cx\mu}\log\frac{1}{\varepsilon}\right)$$
	iterations.
\end{proof}

\clearpage
\section{Analysis of 5GCS for Arbitary Solvers \texorpdfstring{$\cA_m$}{Am}}

In real-life applications we might be in a situation where we would want local solvers to be personalized to each client, one such reason might be the amount of data or the type of software on a local machine. Thanks to the structure of the lifted space, the inner problem is separable which allows us to use arbitrary solvers to minimize the local function.
The general local problem 
\begin{equation*}
	\argmin \limits_{y\in\mathbb{R}^{dn}}\left\{\localfun(y) \eqdef F(y)+\frac{\tauM}{2} \sqnorm{y-\left(\Koper\hat{x}^\kstep+\frac{1}{\tauM}u^t\right)}\right\},
\end{equation*}
can be separated into
\begin{eqnarray*}
	\argmin \limits_{y\in\mathbb{R}^d}\left\{\localfuni(y) \eqdef  F_\iclient(y)+\frac{\tauM}{2} \sqnorm{y-\left(\hat{x}^\kstep+\frac{1}{\tauM}u_\iclient^t\right)}\right\},
\end{eqnarray*}
for $\iclient\in\left\{1,\dots,\Mx\right\}$ as the vector components are independent. This means that the Algorithm $\localsolver$ can be interpreted as concatenation of solutions that Algorithms $\localsolver_m$ find to respective local problems $\localfuni$. Noting that Assumption~\ref{ass:TPS} implies Assumption~\ref{ass:GTPS}, we can note that since local problems are independent there is no constraint on what local solver each client uses nor on a shared number of local steps that each method uses. 

\subsection{Proof of Theorem~\ref{thm:INEXACTPPanyM}}
\begin{theorem}
	Consider Algorithm~\ref{alg:5GCS} (\gls{5GCS}) with the LT solvers $\{\cA_1,\dots,\cA_\Mx\}$ satisfying Assumption~\ref{ass:GTPS}. Let $0<\gammaM\leq \frac{3}{16}\sqrt{\frac{\Cx}{L\mu \Mx}}$ and $\tauM=\frac{1}{2\gammaM \Mx}$. 
	Then for the Lyapunov function
	\begin{equation*}
			\Psi^{\kstep}\eqdef \frac{1}{\gammaM}\sqnorm{x^{\kstep}-x^\star}+\frac{\Mx}{\Cx}\left(\frac{1}{\tauM}+\frac{1}{L_F}\right)\sqnorm{u^{\kstep}-u^\star}\label{PPINEXACTpsianyM_2},
	\end{equation*}
	the iterates of the method satisfy
	$$
	\Exp{\Psi^{\Tx}}\leq (1-\rho)^\Tx \Psi^0,
	$$
	where 
	$	\rho\eqdef  \max\left\{\frac{\gammaM\mu}{1+\gammaM\mu},\frac{\Cx}{\Mx}\frac{\tauM}{(L_F+\tauM)}\right\}<1.
	$

\end{theorem}

\begin{proof}
	Noting that updates for $u^{t+1}$ and $x^{t+1}$ can be written as
	\begin{eqnarray}
		&u^{t+1}\eqdef u^t + \frac{1}{1+\omega} \Rop^t\left(\bar{u}^{t+1}-u^t\right),&\\
		&x^{t+1}=\hat{x}^t-\gammaM\left(\omega+1\right)\Koper^\top\left(u^{t+1}-u^t\right),&\label{xupdate35_1}
	\end{eqnarray}
	where $\Rop^t$ is the Partial Participation operator, $\omega=\frac{\Mx}{\Cx}-1$ and  $\bar{u}^{t+1} = \nabla F(\lastlocitterk)$. 
	Then using variance decomposition and Proposition~1 from work \citep{condat2021murana}, we obtain
	\begin{eqnarray}
		\label{eq:starting_eq3.5_1}
		&&\Exp{\sqnorm{x^{t+1}-x^\star}\;|\;\mathcal{F}^t}\notag\\
        &\overset{(\ref{vardec})}{=}&\sqnorm{\Exp{x^{t+1}\;|\;\mathcal{F}^t}-x^\star}+\Exp{\sqnorm{x^{t+1}-\Exp{x^{t+1}\;|\;\mathcal{F}^t}}\;|\;\mathcal{F}^t}\notag\\
		&\overset{(\ref{xupdate35_1})+(\ref{PP})}{=}& \underbrace{\sqnorm{\hat{x}^{t}-x^\star-\gammaM \Koper^\top(\bar{u}^{t+1}-u^t)}}_{X}+\gammaM^2\oma\sqnorm{\bar{u}^{t+1}-u^t}\notag\\
		&\quad &- \gammaM^2\zeta\sqnorm{\Koper^\top(\bar{u}^{t+1}-u^t)},
	\end{eqnarray}
	where
	\begin{equation*}
		\oma=\frac{\Mx(\Mx-\Cx)}{\Cx(\Mx-1)},\quad \zeta=\frac{\Mx-\Cx}{\Cx(\Mx-1)}.
	\end{equation*}
	Moreover, using \myref{fooc} and the definition of $\hat{x}^t$, we have
	\begin{align}
		&(1+\gammaM\mu)\hat{x}^t=x^t-\gammaM \Koper^\top u^{t},\label{opt1_1}\\
		&(1+\gammaM\mu)x^\star= x^\star -\gammaM \Koper^\top u^\star.\label{opt2_2}
	\end{align}
	Using \myref{opt1_1} and \myref{opt2_2} we obtain
	\begin{eqnarray}
		\label{eq:long3.5_1}
		X &=&	\sqnorm{\hat{x}^{t}-x^\star-\gammaM \Koper^\top(\bar{u}^{t+1}-u^t)}\notag\\
		&=&\sqnorm{\hat{x}^{t}-x^\star} +\gammaM^2\sqnorm{\Koper^\top (\bar{u}^{t+1}-u^t)}\notag\\
		&\quad&-2\gammaM \left\langle 
		\hat{x}^{t}-x^\star,\Koper^\top(\bar{u}^{t+1}-u^t) \right\rangle  \notag\\
		&=& (1+\gammaM\mu) \sqnorm{\hat{x}^{t}-x^\star} +\gammaM^2\sqnorm{\Koper^\top(\bar{u}^{t+1}-u^t)}\notag\\
		&\quad&-2\gammaM  \left\langle 
		\hat{x}^{t}-x^\star,\Koper^\top (\bar{u}^{t+1}-u^\star) \right\rangle    \notag\\
		&\quad&+2\gammaM  \left\langle 
		\hat{x}^{t}-x^\star,\Koper^\top (u^{t}-u^\star) \right\rangle-\gammaM\mu\sqnorm{\hat{x}^{t}-x^\star} \notag\\
		&\overset{(\ref{opt1_1})+(\ref{opt2_2})}{=}&   \left\langle  x^t-x^\star-\gammaM \Koper^\top (u^t-u^\star),\hat{x}^{t}-x^\star \right\rangle +\gammaM^2\sqnorm{\Koper^\top (\bar{u}^{t+1}-u^t)}\notag\\
		&\quad&-2\gammaM  \left\langle 
		\hat{x}^{t}-x^\star,\Koper^\top (\bar{u}^{t+1}-u^\star) \right\rangle  +  \left\langle 
		\hat{x}^{t}-x^\star,2\gammaM \Koper^\top (u^{t}-u^\star) \right\rangle  \notag\\
		&\quad&-\gammaM\mu\sqnorm{\hat{x}^{t}-x^\star} \notag.
	\end{eqnarray}
	It leads to	
	\begin{eqnarray}
		X	&= &\left\langle  x^t-x^\star+\gammaM \Koper^\top (u^t-u^\star),\hat{x}^{t}-x^\star \right\rangle  \notag\\
		&\quad&+\gammaM^2\sqnorm{\Koper^\top (\bar{u}^{t+1}-u^t)}-2\gammaM  \left\langle 
		\hat{x}^{t}-x^\star, \Koper^\top(\bar{u}^{t+1}-u^\star) \right\rangle  \notag\\
		&\quad&-\gammaM\mu\sqnorm{\hat{x}^{t}-x^\star}\notag \\
		&\overset{(\ref{opt1_1})+(\ref{opt2_2})}{=}&\frac{1}{1+\gammaM\mu} \left\langle  x^t-x^\star+\gammaM \Koper^\top (u^t-u^\star),x^t-x^\star-\gammaM \Koper^\top (u^t-u^\star) \right\rangle  \notag\\
		&\quad&+\gammaM^2\sqnorm{\Koper^\top (\bar{u}^{t+1}-u^t)}-2\gammaM  \left\langle 
		\hat{x}^{t}-x^\star, \Koper^\top (\bar{u}^{t+1}-u^\star) \right\rangle  \notag\\
		&\quad&-\gammaM\mu\sqnorm{\hat{x}^{t}-x^\star} \notag\\
		&=& \frac{1}{1+\gammaM\mu}\sqnorm{x^t-x^\star}-\frac{\gammaM^2}{1+\gammaM\mu}\sqnorm{\Koper^\top (u^t-u^\star)}\notag\\
		&\quad&+\gammaM^2\sqnorm{\Koper^\top (\bar{u}^{t+1}-u^t)}-2\gammaM  \left\langle 
		\hat{x}^{t}-x^\star,\Koper^\top (\bar{u}^{t+1}-u^\star) \right\rangle\notag\\
		&\quad&-\gammaM\mu\sqnorm{\hat{x}^{t}-x^\star}   . 
	\end{eqnarray}
	Combining \myref{eq:starting_eq3.5_1} and \myref{eq:long3.5_1} we have
	\begin{eqnarray*}
		&&\Exp{\sqnorm{x^{t+1}-x^\star}\;|\;\mathcal{F}^t}\notag\\
        &\leq&  \frac{1}{1+\gammaM\mu}\sqnorm{x^t-x^\star}-\frac{\gammaM^2}{1+\gammaM\mu}\sqnorm{\Koper^\top (u^t-u^\star)}\\
		&\quad&+\gammaM^2(1-\zeta)\sqnorm{\Koper^\top (\bar{u}^{t+1}-u^t)}-2\gammaM  \left\langle 
		\hat{x}^{t}-x^\star,\Koper^\top (\bar{u}^{t+1}-u^\star) \right\rangle \\
		&\quad&+\gammaM^2\oma\sqnorm{\bar{u}^{t+1}-u^t}-\frac{\gammaM\mu}{\Mx}\sqnorm{\Koper\hat{x}^t-\Koper x^\star}.
	\end{eqnarray*}
	Note that we can have the update rule for $u$ as: 
	\begin{equation*}
		u^{t+1}\eqdef u^t + \tfrac{1}{1+\omega} \Rop^t\left(\bar{u}^{t+1}-u^t\right),
	\end{equation*}
	where $\Rop^t$ is the Partial Participation operator with parameter  $\omega=\frac{\Mx}{\Cx}-1$. Using conic variance formula \myref{cvar} of $\Rop^t$ we obtain
	\begin{eqnarray}
		&&\Exp{\sqnorm{u^{t+1}-u^\star}\;|\;\mathcal{F}^t}\notag\\&\overset{(\ref{vardec})+(\ref{cvar})}{\leq}& \sqnorm{u^{t}-u^\star+\frac{1}{1+\omega}\left(\bar{u}^{t+1} -u^t\right)}
		+\frac{\omega}{(1+\omega)^2}\sqnorm{\bar{u}^{t+1} -u^t }\notag\\
		&=&\frac{\omega^2}{(1+\omega)^2}\sqnorm{u^{t}-u^\star}+\frac{1}{(1+\omega)^2}\sqnorm{\bar{u}^{t+1}-u^\star}\notag\\
		&\quad&+\frac{2\omega}{(1+\omega)^2} \left\langle  u^{t}-u^\star,
		\bar{u}^{t+1}-u^\star \right\rangle +\frac{\omega}{(1+\omega)^2}\sqnorm{\bar{u}^{t+1} -u^\star }\notag\\
		&\quad&+\frac{\omega}{(1+\omega)^2}\sqnorm{u^{t} -u^\star }-\frac{2\omega}{(1+\omega)^2} \left\langle  u^{t}-u^\star,
		\bar{u}^{t+1}-u^\star \right\rangle \notag\\
		&=&\frac{1}{1+\omega}\sqnorm{\bar{u}^{t+1} -u^\star }+\frac{\omega}{1+\omega}\sqnorm{u^{t} -u^\star }.\label{expu_1}
	\end{eqnarray}
	Let us consider the first term in \myref{expu_1}:
	\begin{eqnarray*}
		\sqnorm{\bar{u}^{t+1}-u^\star}&=&\sqnorm{(u^t-u^\star)+(\bar{u}^{t+1}-u^t)}\\
		&=&\sqnorm{u^t-u^\star}+\sqnorm{\bar{u}^{t+1}-u^t}+2 \left\langle  u^t-u^\star,\bar{u}^{t+1}-u^t \right\rangle \\
		&=&\sqnorm{u^t-u^\star}+2  \left\langle  \bar{u}^{t+1}-u^\star,\bar{u}^{t+1}-u^t \right\rangle  - \sqnorm{\bar{u}^{t+1}-u^t}.
	\end{eqnarray*}
	Combining the terms together, we get
	\begin{eqnarray*}
		\Exp{\sqnorm{u^{t+1}-u^\star}\;|\;\mathcal{F}^t}&\leq &\sqnorm{u^{t} -u^\star }\notag\\
        &&+\frac{1}{1+\omega}\left(2  \left\langle  \bar{u}^{t+1}-u^\star,\bar{u}^{t+1}-u^t \right\rangle  - \sqnorm{\bar{u}^{t+1}-u^t}\right).
	\end{eqnarray*}
	Finally, we obtain
	\begin{eqnarray*}
		\frac{1}{\gammaM}\Exp{\sqnorm{x^{t+1}-x^\star}\;|\;\mathcal{F}^t}&+&\frac{1+\omega}{\tauM}\Exp{\sqnorm{u^{t+1}-u^\star}\;|\;\mathcal{F}^t}\\
		&\leq&  \frac{1}{\gammaM(1+\gammaM\mu)}\sqnorm{x^t-x^\star}-\frac{\gammaM}{1+\gammaM\mu}\sqnorm{\Koper^\top (u^t-u^\star)}\\
		&\quad&+\gammaM(1-\zeta)\sqnorm{\Koper^\top (\bar{u}^{t+1}-u^t)}\\
		&\quad&+\gammaM\oma\sqnorm{\bar{u}^{t+1}-u^t}-\frac{\mu}{\Mx}\sqnorm{\Koper \hat{x}^t- \Koper x^\star}\\
		&\quad&+\frac{1+\omega}{\tauM}\sqnorm{u^{t} -u^\star }-2  \left\langle 
		\hat{x}^{t}-x^\star,\Koper^\top (\bar{u}^{t+1}-u^\star) \right\rangle \\ &\quad&+\frac{1}{\tauM}\left(2  \left\langle  \bar{u}^{t+1}-u^\star,\bar{u}^{t+1}-u^t \right\rangle  - \sqnorm{\bar{u}^{t+1}-u^t}\right).
	\end{eqnarray*}
	Ignoring $-\frac{\gammaM}{1+\gammaM\mu}\sqnorm{\Koper^\top (u^t-u^\star)}$ and noting
	\begin{eqnarray*}
		&&-\left\langle \hat{x}^{t}-x^\star,\Koper^\top (\bar{u}^{t+1}-u^\star) \right\rangle +\frac{1}{\tauM}\left\langle  \bar{u}^{t+1}-u^\star,\bar{u}^{t+1}-u^t \right\rangle  \\
		&=&-\left\langle \lastlocitterk-\Koper x^\star,\bar{u}^{t+1}-u^\star \right\rangle +\frac{1}{\tauM}\left\langle  \nabla\localfun (\lastlocitterk),\bar{u}^{t+1}-u^\star \right\rangle\\
		&\overset{(\ref{yi1})+(\ref{strmono})}{\leq}& -\frac{1}{L_F}\sqnorm{\bar{u}^{t+1}-u^\star}+\frac{a}{2\tauM}\sqnorm{\nabla\localfun (\lastlocitterk)}+\frac{1}{2a\tauM}\sqnorm{\bar{u}^{t+1}-u^\star }\\
		&=& -\left(\frac{1}{L_F}-\frac{1}{2a\tauM}\right)\sqnorm{\bar{u}^{t+1}-u^\star}+\frac{a}{2\tauM}\sqnorm{\nabla\localfun (\lastlocitterk)}\\
		&\overset{(\ref{expu_1})}{\leq}& -\left(\frac{1}{L_F}-\frac{1}{2a\tauM}\right)\left((1+\omega)\Exp{\sqnorm{u^{t+1}-u^\star}\;|\;\mathcal{F}^t}-\omega\sqnorm{u^t-u^\star}\right)\\
		&\quad&+\frac{a}{2\tauM}\sqnorm{\nabla\localfun (\lastlocitterk)},
	\end{eqnarray*}
	we get
	\begin{eqnarray*}
		\frac{1}{\gammaM}\Exp{\sqnorm{x^{t+1}-x^\star}\;|\;\mathcal{F}^t}&+&\left(1+\omega\right)\left(\frac{1}{\tauM}+\frac{1}{L_F}\right)\Exp{\sqnorm{u^{t+1}-u^\star}\;|\;\mathcal{F}^t}\\
		&\leq&  \frac{1}{\gammaM(1+\gammaM\mu)}\sqnorm{x^t-x^\star}\\
		&\quad&+\left(1+\omega\right)\left(\frac{1}{\tauM}+\frac{\omega}{1+\omega}\frac{1}{L_F}\right)\sqnorm{u^{t} -u^\star } \\
		&\quad&+\left(\gammaM\left(1-\zeta\right)\Mx+\gammaM\oma-\frac{1}{\tauM}\right)\sqnorm{\bar{u}^{t+1}-u^t}\\
		&\quad& +\frac{L_F}{\tauM^2}\sqnorm{\nabla\localfun (\lastlocitterk)}-\frac{\mu}{\Mx}\sqnorm{\Koper\hat{x}^t-\Koper x^\star}.
	\end{eqnarray*}
	Where we made the choice $a=\frac{L_F}{\tau}$.
	Using Young's inequality we have
	\begin{equation*}
		-\frac{\mu}{3\Mx}\sqnorm{\Koper \hat{x}^t-\localsolk+\localsolk-\Koper x^\star}\overset{(\ref{yi3})}{\leq}\frac{\mu}{3\Mx}\sqnorm{\localsolk-\Koper x^\star}-\frac{\mu}{6\Mx}\sqnorm{\Koper\hat{x}^t-\localsolk}.	\end{equation*}
	Noting the fact that $\localsolk=\Koper\hat{x}^t-\frac{1}{\tauM}(\hat{u}^{t+1}-u^t)$, we have
	$$\frac{\mu}{3\Mx}\sqnorm{\localsolk-\Koper x^\star}\overset{(\ref{yi2})}{\leq} 2\frac{\mu}{3\Mx}\sqnorm{\Koper\hat{x}^t-\Koper x^\star}+\frac{2}{\tauM^2}\frac{\mu}{3\Mx}\sqnorm{\hat{u}^{t+1}-u^t}.$$
	Combining those inequalities we get
	\begin{eqnarray*}
		\frac{1}{\gammaM}\Exp{\sqnorm{x^{t+1}-x^\star}\;|\;\mathcal{F}^t}&+&\left(1+\omega\right)\left(\frac{1}{\tauM}+\frac{1}{L_F}\right)\Exp{\sqnorm{u^{t+1}-u^\star}\;|\;\mathcal{F}^t}\\
		&\leq&  \frac{1}{\gammaM(1+\gammaM\mu)}\sqnorm{x^t-x^\star}\\
		&\quad&+\left(1+\omega\right)\left(\frac{1}{\tauM}+\frac{\omega}{1+\omega}\frac{1}{L_F}\right)\sqnorm{u^{t} -u^\star } \\
		&\quad&+\frac{2}{\tauM^2}\frac{\mu}{3\Mx}\sqnorm{\hat{u}^{t+1}-u^t}\\
		&\quad&-\left(\frac{1}{\tauM}-\left(\gammaM\left(1-\zeta\right)\Mx+\gammaM\oma\right)\right)\sqnorm{\bar{u}^{t+1}-u^t}\\
		&\quad&+\frac{L_F}{\tauM^2}\sqnorm{\nabla\localfun (\lastlocitterk)}-\frac{\mu}{6\Mx}\sqnorm{\Koper\hat{x}^t-\localsolk}.
	\end{eqnarray*}
	Assuming $\gammaM$ and $\tauM$ can be chosen so that $\frac{1}{\tauM}-(\gammaM(1-\zeta)\Mx+\gammaM\oma))\geq \frac{4}{\tauM^2}\frac{\mu}{3\Mx}$ we obtain
	\begin{eqnarray*}
		\frac{1}{\gammaM}\Exp{\sqnorm{x^{t+1}-x^\star}\;|\;\mathcal{F}^t}&+&\left(1+\omega\right)\left(\frac{1}{\tauM}+\frac{1}{L_F}\right)\Exp{\sqnorm{u^{t+1}-u^\star}\;|\;\mathcal{F}^t}\\
		&\leq&  \frac{1}{\gammaM(1+\gammaM\mu)}\sqnorm{x^t-x^\star}\\
		&\quad&+\left(1+\omega\right)\left(\frac{1}{\tauM}+\frac{\omega}{1+\omega}\frac{1}{L_F}\right)\sqnorm{u^{t} -u^\star } \\
		&\quad&+\frac{4}{\tauM^2}\frac{\mu L_F^2}{3\Mx}\sqnorm{\lastlocitterk-\localsolk}+\frac{L_F}{\tauM^2}\sqnorm{\nabla\localfun (\lastlocitterk)}\\
		&\quad&-\frac{\mu}{6\Mx}\sqnorm{\Koper\hat{x}^t-\localsolk}.
	\end{eqnarray*}
Using Assumption~\ref{ass:GTPS}, we have 
	\begin{align*}
	\sum\limits_{\iclient=1}^{\Mx}\frac{4}{\tauM^2}\frac{\mu L_F^2}{3\Mx}\sqnorm{\lastlocittermk-\localsolmk} +\sum\limits_{\iclient=1}^{\Mx}\frac{L_F}{\tauM^2}\sqnorm{\nabla\localfuni(\lastlocittermk)} \leq\sum\limits_{\iclient=1}^{\Mx}\frac{\mu}{6\Mx}\sqnorm{\hat{x}^\kstep-\localsolmk},
\end{align*}
This is enough to have similar bound in lifted space for the point $\lastlocitterk$:
	\begin{eqnarray*}
		\frac{4}{\tauM^2}\frac{\mu L_F^2}{3\Mx}\sqnorm{\lastlocitterk-\localsolk}+\frac{L_F}{\tauM^2}\sqnorm{\nabla\localfun (\lastlocitterk)}\leq\frac{\mu}{6\Mx}\sqnorm{\Koper\hat{x}^t-\localsolk}.
	\end{eqnarray*}
	Thus
	\begin{eqnarray*}
		\frac{1}{\gammaM}\Exp{\sqnorm{x^{t+1}-x^\star}\;|\;\mathcal{F}^t}&+&\left(1+\omega\right)\left(\frac{1}{\tauM}+\frac{1}{L_F}\right)\Exp{\sqnorm{u^{t+1}-u^\star}\;|\;\mathcal{F}^t}\\
		&\leq&  \frac{1}{\gammaM(1+\gammaM\mu)}\sqnorm{x^t-x^\star}\\
		&\quad&+\left(1+\omega\right)\left(\frac{1}{\tauM}+\frac{\omega}{1+\omega}\frac{1}{L_F}\right)\sqnorm{u^{t} -u^\star }.
	\end{eqnarray*}
	
	By taking the expectation on both sides we get
	\begin{equation*}
		\Exp{\Psi^{t+1}}\leq  \max\left\{\frac{1}{1+\gammaM\mu} ,\frac{ L_F+\frac{\Mx-\Cx}{\Cx}\tauM}{L_F+\tauM} \right\}\Exp{\Psi^{t}},
	\end{equation*}
	which finishes the proof. 
	Note that our standard choice of constants is $$\omega=\frac{\Mx}{\Cx}-1,\quad \oma=\frac{\Mx(\Mx-\Cx)}{\Cx(\Mx-1)},\quad \zeta=\frac{\Mx-\Cx}{\Cx(\Mx-1)}.$$ Using these parameters the requirement for stepsizes becomes:$$\frac{1}{\tauM}-\gammaM \Mx\geq \frac{4\mu}{3\Mx\tauM^2}.$$
	This inequality is satisfied, when $0<\gammaM\leq \frac{3}{16}\sqrt{\frac{\Cx}{L\mu \Mx}}$ and $\tauM=\frac{1}{2\Mx\gammaM}.$
\end{proof}

\subsection{Reallocation of resources}

\begin{assumption}\label{ass:TPS}
	Let $\mathcal{A}_\iclient$ be an Algorithm that can find a point $\lastlocittermk$ after $\Kx$ local steps applied to the local function $\localfuni$ from \myref{localprob} and starting point $y_{\iclient}^{0,\kstep}=\hat{x}^\kstep$, which satisfies
	\begin{eqnarray*}
		\frac{4}{\tauM^2}\frac{\mu L_F^2}{3\Mx}\sqnorm{\lastlocittermk-\localsolmk}	+ \frac{L_F}{\tauM^2}\sqnorm{\nabla\localfuni(\lastlocittermk)}
		\leq	\squeeze \frac{\mu}{6\Mx}\sqnorm{\hat{x}^\kstep-\localsolmk},
	\end{eqnarray*}
	where $\localsolmk$ is the unique minimizer of $\localfuni$, and $\tauM\geq \frac{8}{3}\sqrt{\frac{L\mu}{\Mx \Cx}}.$
\end{assumption}

The general local problem is
\begin{equation}
	\argmin \limits_{y\in\mathbb{R}^{dn}}\left\{\localfun(y) \eqdef  F(y)+\frac{\tauM}{2} \sqnorm{y-\left(\Koper\hat{x}^\kstep+\frac{1}{\tauM}u^t\right)}\right\}\label{locprobG},
\end{equation}
and the condition necessary for Theorem~\ref{thm:INEXACTPPanyM} is
\begin{align*}
	\frac{4}{\tauM^2}\frac{\mu L_F^2}{3\Mx}\sqnorm{\lastlocitterk-\localsolk}+\squeeze\frac{L_F}{\tauM^2}\sqnorm{\nabla\localfun (\lastlocitterk)}\leq\frac{\mu}{6\Mx}\sqnorm{\Koper\hat{x}^t-\localsolk}.
\end{align*}
This is actually a restriction in $\mathbb{R}^{dn}$ (a dual/lifted space), which can be equivalently written as
\begin{align*}
	\sum_{\iclient=1}^{\Mx}\frac{4}{\tauM^2}\frac{\mu L_F^2}{3\Mx}\sqnorm{\lastlocittermk-\localsolmk}+\sum_{\iclient=1}^{\Mx}\frac{L_F}{\tauM^2}\sqnorm{\nabla\localfuni(\lastlocittermk)}\leq\sum_{\iclient=1}^{\Mx}\frac{\mu}{6\Mx}\sqnorm{\hat{x}^\kstep-\localsolmk}.
\end{align*}

Assumption~\ref{ass:GTPS}, which is necessary to hold for Theorem~\ref{thm:INEXACTPPanyM} arises due to the definition of the lifted space. The strength of this condition is that it allows for provable convergence even in situations where some clients can not find the required by Assumption~\ref{ass:TPS} accuracy as long other clients compensate for it by doing more iterations.

\subsection{Number of local steps in LT subroutine of 5GCS}
\label{mylabel}
In this section, we would like to present different guarantees that various Algorithms $\localsolver$ can give us.
Algorithm~$\mathcal{A}$ is simply taking current iterates $\hat{x}^t$ and $u^t$ and applying Algorithms $\localsolver_m$ to the local problem \myref{localprob}  (at each client), and finally concatenating the result in $\localsolk$. 
To guarantee convergence of Algorithm~\ref{alg:5GCS}, we need to do locally $\Kx$ iterations of Algorithm~$\localsolver$ which would guarantee:
\begin{eqnarray*}
	&&\frac{4}{\tauM^2}\frac{\mu L_F^2}{3\Mx}\sqnorm{\lastlocitterk-\localsolk}+\frac{a}{\tauM}\sqnorm{\nabla\localfun (\lastlocitterk)}\\	&\leq&\left(\frac{4\mu L_F^2}{3\Mx\tauM^2}+\frac{a(L_F+\tauM)^2}{\tauM}\right)\sqnorm{\lastlocitterk-\localsolk}\\
	&\leq&\frac{\mu}{6\Mx}\sqnorm{\Koper\hat{x}^t-\localsolk}.
\end{eqnarray*}
Thus, we need:
\begin{equation}
	\sqnorm{\lastlocitterk-\localsolk}
	\leq\betaM\sqnorm{\Koper\hat{x}^t-\localsolk}.\label{betasol}  
\end{equation}
Where $$\betaM=\frac{\frac{\mu}{6\Mx}}{\left(\frac{4\mu L_F^2}{3\Mx\tauM^2}+\frac{a(L_F+\tauM)^2}{\tauM}\right)}.$$
For $a=\frac{L_F}{\tauM}$, the term that will appear in most of those analysis is 
\begin{equation*}
	\frac{1}{\betaM}=\frac{\left(\frac{4\mu L_F^2}{3\Mx\tauM^2}+\frac{a(L_F+\tauM)^2}{\tauM}\right)}{\frac{\mu}{6\Mx}}\leq\frac{8L_F^2}{\tauM^2}+\frac{12 L_F^3\Mx}{\tauM^2\mu}+\frac{12 L_F}{\mu}.
\end{equation*}  
Note that $\tauM$ is smallest for the optimal choice of $\gammaM$, thus
\begin{equation*}
	\frac{1}{\betaM}\leq\frac{9 L_F^2\Cx\Mx}{8L\mu}+\frac{108 L_F^3\Mx^2\Cx}{64L\mu^2}+\frac{12 L_F \Mx}{\mu}\leq\left(4\frac{L}{\mu}\right)^2,
\end{equation*}  
where in the last inequality we used  bounds such as $\Mx\geq \Cx$, $L\geq \Mx L_F$ and $\frac{L}{\mu}\geq 1$.

\subsubsection{Gradient descent for local problem}
\algname{\gls{GD}} with stepsize $\frac{1}{L_F+\tauM}$ would need:
\begin{equation*}
	\Kx\geq \left(\frac{L_F+\tauM}{\tauM}\right)\log\left(\frac{1}{\betaM}\right).
\end{equation*}
Again noting that $\tauM$ is smallest when we choose stepsizes optimally:
$$\frac{L_F+\tauM}{\tauM}\leq \frac{3}{8}\sqrt{\frac{L\Cx}{\mu \Mx}}+1.$$
Thus, if $\mathcal{A}$ is \algname{\gls{GD}}, then:
\begin{equation*}
	\Kx\geq \left(\frac{3}{4}\sqrt{\frac{L\Cx}{\mu \Mx}}+2\right)\log\left(4\frac{L}{\mu}\right).
\end{equation*}

\subsection{Local speedup induced by Client-Specific condition numbers
}
Dependence of the local condition number on $\tauM$ and how can we use this dependence to control the speed of local convergence is described in Section~\ref{mylabel}. Here we would like to focus on the case where each function has a different smoothness parameter.
Suppose each $f_\iclient$ is $L_m$-smooth and $\mu$-convex. If we let $L=\max_\iclient L_\iclient$ then we can note that each $f_\iclient$ is $L$-smooth, thus we have that $L_F=\frac{1}{\nclients}\left(L-\mu\right)$ and we recover the whole communication result for our Algorithm. However, locally we can note that each client needs to find $\betaM$-solution to the local problem \myref{localprob}, which is $\left(\frac{1}{\nclients}\left(L_\iclient-\mu\right)+\tauM\right)$-smooth and $\tauM$-convex.
Remembering $\tauM\geq\frac{8}{3}\sqrt{\frac{\mu L}{\nclients\kcohort}} $, \algname{\gls{GD}} needs
\begin{eqnarray*}
	2\left(\frac{1}{\nclients}\left(L_\iclient-\mu\right)\frac{1}{\tauM}+1\right)\log\left(4\frac{L}{\mu}\right)&\leq&2\left(\frac{3}{8}\sqrt{\frac{L_\iclient\kcohort}{\mu \nclients }}+1\right)\log\left(4\frac{L}{\mu}\right),
\end{eqnarray*}
iterations. This is better than if we were using the upper bound $\max_\iclient L_\iclient$ on each $L_\iclient$. To illustrate this we can formulate the following Corollary~\ref{cor:personal} to the general Theorem~\ref{thm:5GCS}
\begin{corollary}\label{cor:personal}
	Consider Algorithm~\ref{alg:5GCS} with LT solver being \gls{GD}. In the new personalized setting with $L=\max_{\iclient}L_{\iclient}$, we can run the LT for $$\localsteps\geq 2\left(\frac{3}{8}\sqrt{\frac{L_\iclient\kcohort}{\mu \nclients }}+1\right)\log\left(4\frac{L}{\mu}\right)$$ and still accomplish guarantees of Theorem \ref{thm:5GCS}.
\end{corollary}

\clearpage
\subsection{Local solvers \texorpdfstring{$\localsolver_m$}{local solvers m} may be stochastic}
Until now we assumed that Algorithms $\localsolver_m$ were deterministic (in a sense that they do not introduce any randomness to the system). However, with a small change in the analysis from Section~\ref{proof35}, we can allow for local solvers to be stochastic, we can present a more general condition which includes stochastic local solvers. To analyze the stochastic local solvers we need to modify Assumption~\ref{ass:GTPS} with respect to stochasticity. We introduce a new assumption, where the inequality appearing in Assumption~\ref{ass:GTPS} should be satisfied in expectation. 
\begin{assumption}\label{ass:STPS}
	Let $\mathcal{A}$ be stochastic Algorithm that can find a point $\lastlocitterk$ in $\Kx$ local steps applied to the local function $\localfun$ from \myref{localprob} and starting point $y_{\iclient}^{0,\kstep}=\hat{x}^\kstep$, which satisfies
	\begin{align*}
		&\Exp{\sum_{\iclient=1}^{\Mx}\frac{4}{\tauM^2}\frac{\mu L_F^2}{3\Mx}\sqnorm{\lastlocittermk-\localsolmk}+\sum_{\iclient=1}^{\Mx}\frac{L_F}{\tauM^2}\sqnorm{\nabla\localfuni(\lastlocittermk)}\;|\;\mathcal{F}^t}\\
&\qquad\qquad\qquad\qquad\qquad\qquad\leq\sum_{\iclient=1}^{\Mx}\frac{\mu}{6\Mx}\sqnorm{\hat{x}^\kstep-\localsolmk},
	\end{align*}
	where $\localsolmk$ is the unique minimizer of $\localfuni$, and $\tauM\geq \frac{8}{3}\sqrt{\frac{L\mu}{\Mx \Cx}}.$
\end{assumption}
 The conditioning on $\mathcal{F}^t$ simply means that $\hat{x}^t$ is not treated as a random vector and the only randomness comes from the local solvers. Let us consider $\Exp{X\;|\;A}$, which represents the expectation of a random variable $X$ conditioned on the randomness accumulated due to local solvers being stochastic. Then conditioning on both $A^t$ and $\mathcal{F}^t$, we can get
	\begin{eqnarray*}
		&&\frac{1}{\gammaM}\Exp{\sqnorm{x^{t+1}-x^\star}\;|\;\mathcal{F}^t\cup A^t}\\
        &&+\left(1+\omega\right)\left(\frac{1}{\tauM}+\frac{1}{L_F}\right)\Exp{\sqnorm{u^{t+1}-u^\star}\;|\;\mathcal{F}^t\cup A^t}\\
		&\leq&  \frac{1}{\gammaM(1+\gammaM\mu)}\sqnorm{x^t-x^\star}\\
		&\quad&+\left(1+\omega\right)\left(\frac{1}{\tauM}+\frac{\omega}{1+\omega}\frac{1}{L_F}\right)\sqnorm{u^{t} -u^\star } \\
		&\quad&+\frac{4}{\tauM^2}\frac{\mu L_F^2}{3\Mx}\sqnorm{\lastlocitterk-\localsolk}+\frac{L_F}{\tauM^2}\sqnorm{\nabla\localfun (\lastlocitterk)}\\
		&\quad&-\frac{\mu}{6\Mx}\sqnorm{\Koper\hat{x}^t-\localsolk}.
	\end{eqnarray*}
	Taking expectation conditioned on $\mathcal{F}^t$ on both sides we get
	\begin{eqnarray*}
		&&\frac{1}{\gammaM}\Exp{\sqnorm{x^{t+1}-x^\star}\;|\;\mathcal{F}^t}\\
        &&+\left(1+\omega\right)\left(\frac{1}{\tauM}+\frac{1}{L_F}\right)\Exp{\sqnorm{u^{t+1}-u^\star}\;|\;\mathcal{F}^t}\\
		&\leq&  \frac{1}{\gammaM(1+\gammaM\mu)}\sqnorm{x^t-x^\star}\\
		&\quad&+\left(1+\omega\right)\left(\frac{1}{\tauM}+\frac{\omega}{1+\omega}\frac{1}{L_F}\right)\sqnorm{u^{t} -u^\star } \\
		&\quad&+\Exp{\frac{4}{\tauM^2}\frac{\mu L_F^2}{3\Mx}\sqnorm{\lastlocitterk-\localsolk}+\frac{L_F}{\tauM^2}\sqnorm{\nabla\localfun (\lastlocitterk)}\;|\;\mathcal{F}^t}\\
		&\quad&-\frac{\mu}{6\Mx}\sqnorm{\Koper\hat{x}^t-\localsolk}.
	\end{eqnarray*}
Crucial practical benefit comes from the expected improvement in gradient calculation, when each local function has a finite sum structure, which is common in practice.

\subsubsection{L-SVRG for local problem}
\begin{algorithm}[H]
	\caption{\algname{L-SVRG}}
	\begin{algorithmic}[1]\label{alg:LSVRG}
		\STATE  \textbf{input:} initial points $x^0\in\mathbb{R}^d$, $y^0=x^0$, gradient estimator $g$; 
		\STATE stepsize $\gammaM>0$
		\FOR{$k=0, 1, \ldots$}
		\STATE$g^k=g(x^k)-g(y^k)+\nabla f(y^k)$
		\STATE$x^{k+1}=x^k-\gammaM g^k$
		\STATE$y^{k+1}=\begin{cases}
			x^k \quad \text{with probability $p$}\\
			y^k \quad \text{with probability $1-p$}
		\end{cases}$
		\ENDFOR
	\end{algorithmic}
\end{algorithm}
In this section we consider \algname{\gls{L-SVRG}}\citep{kovalev2020don} method as local stochastic solver with variance reduction mechanism. Our analysis is based on general expected smoothness assumption \citep{gower2019sgd}.
\begin{assumption}
The gradient estimator $g$ is unbiased, and satisfies the expected smoothness bound\label{ass:expsmooth}
	\begin{align*}
		\Exp{g(x)} &= \nabla f(x),\\
		\Exp{\sqnorm{g(x)-g(x^\star)}}&\leq 2A''\mathcal{D}_f(x,x^\star).
	\end{align*}
\end{assumption}
We apply convergence guarantees of \algname{\gls{L-SVRG}} for the subproblem in Algorithm~\ref{alg:5GCS} for a gradient estimator $g$ satisfying Assumption~\ref{ass:expsmooth} and stepsize $\gammaM_2=\frac{1}{6A''}$. We obtain the following bound:
\begin{equation*}
	\Exp{\sqnorm{\lastlocitterk-\localsolk}}\leq\left(1-\min\left\{\gammaM_2\tauM,\frac{p}{2}\right\}\right)^T\left(1+2\gammaM_2^2\frac{L_F+\tauM}{p}\right)\sqnorm{\Koper\hat{x}^t-\localsolk} .
\end{equation*}
This means that Algorithm~\ref{alg:LSVRG} with $p=2\tauM\gammaM_2$ finds $\betaM$-solution to the local problem of Algorithm~\ref{alg:5GCS} in
\begin{equation*}
K=	\frac{6A''}{\tauM}\log\left(\frac{\tauM+\left(L_F + \tauM\right)\gammaM_2}{\tauM}\frac{1}{\betaM}\right)
\end{equation*}
local steps. Particularly interesting and practical example of $g$ in the Algorithm~\ref{alg:LSVRG} is mini-batch gradient estimator.
Thus, we assume the finite sum structure:
\begin{eqnarray*}
	f_\iclient(x)=\frac{1}{n_\iclient}\sum_{i=1}^{n_\iclient}f_{\iclient,i}(x),
\end{eqnarray*}
where each $f_{\iclient,i}$ is convex and $L_{i}$ smooth. Then $\localfuni$ can be put in the finite sum structure, by writing
\begin{eqnarray*}
	\localfuni(y)=\frac{1}{n_\iclient}\sum_{i=1}^{n_\iclient}g_i(y),
\end{eqnarray*}
where
\begin{eqnarray*}
	g_i(y)=\frac{1}{\Mx}\left(f_{\iclient,i}(y)-\frac{\mu}{2}\sqnorm{y}\right)+\frac{\tauM}{2}\sqnorm{y-(\hat{x}^t+\frac{1}{\tauM}u_{\iclient}^t)}.
\end{eqnarray*}
Since $\tau\geq\frac{4\mu}{3\Mx}$, $g_{i}$ is $\left(\frac{1}{\Mx}\left(L_{i}-\mu\right)+\tauM\right)$-smooth and $\left(\tauM-\frac{\mu}{\Mx}\right)$-convex. 
Fix a mini-batch size $b_\iclient\in\{1,2,\dots,\Mx_\iclient\}$ and let $S_\iclient$ be a random subset of $ \{1,\ldots,\Mx_\iclient\}$ of size $\Cx$, chosen uniformly at random, then the mini-batch gradient estimator is 
\begin{eqnarray*}
	g(y)=\frac{1}{b_\iclient}\sum_{i\in S_\iclient} \nabla g_i(y).
\end{eqnarray*}
For this gradient estimator 
\begin{eqnarray*}
	A''=\frac{n_\iclient-b_\iclient}{b_\iclient\left(n_\iclient-1\right)}\underset{i}{\max}L_{g_i}+\frac{n_\iclient\left(b_\iclient-1\right)}{b_\iclient\left(n_\iclient-1\right)}\left(L_F+\tauM\right),
\end{eqnarray*}
where $L_{g_i}=\frac{1}{\Mx}\left(L_{i}-\mu\right)+\tau$.

\clearpage

\section{Relation Between the \# of Communication Rounds \texorpdfstring{$\Tx$}{Tx} and the \# of Local Steps \texorpdfstring{$\Kx$}{Kx}}
\subsection{Proof of Theorem~\ref{thm:relation2}}
\begin{theorem}
	Consider Algorithm~\ref{alg:5GCS} (\gls{5GCS}) with the LT solver being \algname{\gls{GD}}. Let $\gammaM=\frac{3}{16L}$  and $\tauM=\frac{8L}{3\Mx}.$
	With such chosen stepsizes, it is enough to run \algname{\gls{GD}} for
	\begin{eqnarray*}
		\squeeze		\Kx\geq \left(2+\frac{3\Mx L_F}{4L}\right)\log\left(4\frac{L}{\mu}\right)=\mathcal{O}\left(\log\frac{L}{\mu}\right).
	\end{eqnarray*}
	Whereas, the number of communication rounds to reach $\varepsilon$-solution is
	\begin{equation*}
		\squeeze	T\geq\max\left\{1+\frac{16}{3}\frac{L}{\mu}, \frac{\Mx}{\Cx}+\frac{3\Mx}{8\Cx}\frac{\Mx L_F}{L}\right\}\log\frac{1}{\epsilon}=\tilde{\mathcal{O}}\left(\frac{\Mx}{\Cx}+\frac{L}{\mu}\right).
	\end{equation*}
\end{theorem}
\begin{proof}
	Note that by choosing $\tauM=\frac{8L}{3\Mx}$ and $\gammaM=\frac{3}{16L}$ stepsizes satisfy the condition from Theorem~\ref{thm:INEXACTPPanyM} and the number of local iterations of \algname{\gls{GD}} to guarantee convergence is:
	\begin{eqnarray*}
		\Kx\geq 2\frac{L_F+\frac{8L}{3\Mx}}{\frac{8L}{3\Mx}}\log\left(4\frac{L}{\mu}\right) = \left(2+\frac{3\Mx L_F}{4L}\right)\log\left(4\frac{L}{\mu}\right)=\mathcal{O}\left(\log\frac{L}{\mu}\right).
	\end{eqnarray*}
	Whereas, the number of communication rounds to reach $\varepsilon$-solution is:
	\begin{equation*}
		\max\left\{1+\frac{16}{3}\frac{L}{\mu}, \frac{\Mx}{\Cx}+\frac{3\Mx}{8\Cx}\frac{\Mx L_F}{L}\right\}\log\frac{1}{\epsilon}=\mathcal{O}\left(\left(\frac{\Mx}{\Cx}+\frac{L}{\mu}\right)\log\frac{1}{\epsilon}\right).
	\end{equation*}
\end{proof}
\subsection{Proof of Theorem~\ref{thm:relation}}
\begin{theorem}
	Consider Algorithm~\ref{alg:5GCS} (\gls{5GCS}) with the LT solver being \algname{\gls{GD}} run for $\Kx \geq \Kx(\deltaM)\eqdef 2\deltaM \log\left(\nicefrac{4L}{\mu}\right)$ iterations, where  $1<\deltaM< 1+\frac{3}{8}\sqrt{\frac{L\Cx}{\mu \Mx}}$ . Let $\gammaM=\frac{1}{2\Mx\tauM}$  and $\tauM=\max\left\{\frac{L}{\Mx (\deltaM-1)},\frac{8}{3}\sqrt{\frac{L\mu}{\Mx \Cx}}\right\}.$
	Then for the Lyapunov function
	\begin{equation*}
			\Psi^{\kstep}\eqdef \frac{1}{\gammaM}\sqnorm{x^{\kstep}-x^\star}+\frac{\Mx}{\Cx}\left(\frac{1}{\tauM}+\frac{1}{L_F}\right)\sqnorm{u^{\kstep}-u^\star}\label{PPINEXACTpsianyM_3},
	\end{equation*}
	the iterates of the method satisfy
	$$
	\Exp{\Psi^{\Tx}}\leq (1-\rho)^\Tx \Psi^0,
	$$
	where 
	$	\rho\eqdef  \max\left\{\frac{\gammaM\mu}{1+\gammaM\mu},\frac{\Cx}{\Mx}\frac{\tauM}{(L_F+\tauM)}\right\}<1.
	$

\end{theorem}
\begin{proof}
	Firstly, we can note that at each step we need to find $\betaM$-solution to the local problem \myref{localprob}. Here, noting that we can restrict ourself to $\tauM\geq \frac{8}{3}\sqrt{\frac{L\mu}{\Mx \Cx}}$ since for this choice we get optimal number of communication rounds, thus we can note:
	\begin{eqnarray*}
		&6\frac{L}{\mu} \leq \frac{1}{\betaM}\leq \left(4\frac{L}{\mu}\right)^2&\\
		& \log\left(6\frac{L}{\mu}\right)\leq \log\frac{1}{\betaM}\leq 2\log\left(4\frac{L}{\mu}\right)&.
	\end{eqnarray*}
	Thus, the speed of local convergence depends fully on the condition number of the local problem $\left(\text{i.e., on }\frac{L_F+\tauM}{\tauM}\right)$. 
	For general result we can ask for the guarantee such that
	\begin{equation*}
		\Kx\geq \Kx(\deltaM)\eqdef \deltaM\left(2\log\left(4\frac{L}{\mu}\right)\right),\quad\deltaM>1.
	\end{equation*}
 For that we would need:
	\begin{equation*}
		\frac{L_F+\tauM}{\tauM}\leq\frac{\frac{L}{\Mx}+\tauM}{\tauM} \leq\deltaM \implies \tauM\geq \frac{\frac{L}{\Mx}}{\deltaM-1}.
	\end{equation*}
We use the choice 
		\begin{equation*}
			\gammaM\leq \frac{1}{\Mx\tauM}\left(1-\frac{4\mu}{3\Mx\tauM}\right).
		\end{equation*}
		Thus if $\tauM\geq \frac{8\mu}{3\Mx}$, then we can choose $\gammaM=\frac{1}{2\Mx\tauM}$.	Thus, let us take dual stepsize as $\tauM=\max\left\{\frac{\frac{L}{\Mx}}{\deltaM-1},\frac{8}{3}\sqrt{\frac{L\mu}{\Mx \Cx}}\right\}$, so that we can choose $\gammaM=\frac{1}{2\Mx\tauM}$. With $\Kx$ \algname{\gls{GD}} local iterations and this stepsize choice the contraction of the Lyapunov function follows from Theorem \ref{thm:INEXACTPPanyM}.
	\end{proof}
\subsection{Proof of Corollary~\ref{cor:rela}}
\begin{corollary}
	Choose any $0<\varepsilon<1$. In order to guarantee $\Exp{\Psi^{\Tx}}\leq \varepsilon \Psi^0$, it suffices to take 
	\begin{equation*}
		\squeeze	T\geq	\max\left\{1+\frac{2L}{\left(\deltaM-1\right)\mu},\frac{\Mx}{\Cx}\deltaM\right\}\log\frac{1}{\epsilon}.
	\end{equation*}
	We can note that when $\deltaM\leq \frac{\Mx+\Cx}{2\Mx}+\sqrt{\frac{2L\Cx}{\mu \Mx}+\left(\frac{\Mx-\Cx}{2\Mx}\right)^2}$, then
	\begin{equation*}
		\squeeze 	\Tx \geq	T(\alpha)\eqdef \left(1+\frac{2}{\deltaM-1} \frac{L}{\mu}\right)\log \frac{1}{\varepsilon}.
	\end{equation*}
\end{corollary}
\begin{proof}
		To satisfy Assumption \ref{ass:GTPS}, assume that the Local Solver is \algname{\gls{GD}} run for 
		\begin{equation*}
			\Kx\geq \Kx(\deltaM)\eqdef \deltaM\left(2\log\left(4\frac{L}{\mu}\right)\right),\quad\deltaM>1.
		\end{equation*}
	To ensure this, choose $\tauM=\max\left\{\frac{\frac{L}{\Mx}}{\deltaM-1},\frac{8}{3}\sqrt{\frac{L\mu}{\Mx \Cx}}\right\}$ and $\gammaM=\frac{1}{2\Mx\tauM}$. Then the communication complexity is:
	\begin{eqnarray*}
		&&\max\left\{1+\frac{1}{\gammaM\mu},\frac{\Mx}{\Cx}+\frac{\Mx}{\Cx}\frac{L_F}{\tauM}\right\}\\
		&\leq&\max\left\{\max\left\{1+\frac{2L}{\left(\deltaM-1\right)\mu},1+\frac{16}{3}\sqrt{\frac{L\Mx}{\mu \Cx}}\right\},\frac{\Mx}{\Cx}\min\left\{\deltaM,1+\frac{3}{8}\sqrt{\frac{L\Cx}{\mu \Mx}}\right\}\right\}.
	\end{eqnarray*}
	For $\deltaM\leq 1+\frac{3}{8}\sqrt{\frac{L\Cx}{\mu \Mx}}$, this simplifies to:
	\begin{equation*}
	T\geq	\max\left\{1+\frac{2L}{\left(\deltaM-1\right)\mu},\frac{\Mx}{\Cx}\deltaM\right\}\log\frac{1}{\epsilon}.
	\end{equation*}
	We can note that when $\deltaM\leq \frac{\Mx+\Cx}{2\Mx}+\sqrt{\frac{2L\Cx}{\mu \Mx}+\left(\frac{\Mx-\Cx}{2\Mx}\right)^2}$, then:
	\begin{equation*}
	T\geq	\max\left\{1+\frac{2L}{\left(\deltaM-1\right)\mu},\frac{\Mx}{\Cx}\deltaM\right\}\log\frac{1}{\epsilon}=\left(1+\frac{2L}{\left(\deltaM-1\right)\mu}\right)\log\frac{1}{\epsilon}.
	\end{equation*}
	Thus, we get a relation between the number of local steps and communication rounds.
\end{proof}

\clearpage
\section{Implementation-Friendly Version of Algorithm~\ref{alg:5GCS}}

\begin{algorithm}[H]
	\caption{\algname{Partial participation with a new rule for $u$ and memory-efficient update for $v$}}
	\begin{algorithmic}[1]\label{alg:membet}
		\STATE  \textbf{input:} initial points $x^0\in\mathbb{R}^d$, $u_\iclient^0\in\mathbb{R}^d$ for all $\iclient=\{1,\dots,\Mx\}$; 
		\STATE stepsize $\gammaM>0$, $\tauM>0$;  $\Cx\in \{1,\dots,\Mx\}$
		\STATE $v^0\eqdef \sum_{\iclient=1}^\Mx u_\iclient^0$
		\FOR{$t=0, 1, \ldots$}
		\STATE $\hat{x}^{t} \eqdef \frac{1}{1+\gammaM\mu} \left(x^t - \gammaM v^t\right)$
		\STATE Pick $\set\subset \{1,\ldots,\Mx\}$ of size $\Cx$ uniformly at random
		\FOR{$\iclient\in \set$}
		\STATE Find $\lastlocittermk$ as a final point of $\Kx$ iteration of some Algorithm~$\mathcal{A}_\iclient$ starting with $y_\iclient^0=\hat{x}^t$ for following problem:
		\begin{eqnarray*}
			\lastlocittermk \approx \argmin_{y\in\mathbb{R}^d}\left\{\localfuni(y) =  F_\iclient(y)+\frac{\tauM}{2} \sqnorm{y-\left(\hat{x}^\kstep+\frac{1}{\tauM}u_\iclient^t\right)}\right\}
		\end{eqnarray*}
		\STATE $u_\iclient^{t+1}=\nabla F_\iclient(\lastlocittermk)$
		\STATE $\Delta u_\iclient^{t+1}=u_\iclient^{t+1}-u_\iclient^t$
		\ENDFOR
		\FOR{$\iclient\in\{1, \ldots,\Mx\}\backslash \set$}
		\STATE $u_{\iclient}^{t+1}\eqdef u_{\iclient}^t $
		\ENDFOR
		\STATE $\Delta v^{t+1}\eqdef \sum_{\iclient\in\set} \Delta u_\iclient^{t+1}$
		\STATE  $x^{t+1} \eqdef \hat{x}^{t}- \gammaM \frac{\Mx}{\Cx} \Delta v^{t+1}$
		\STATE $v^{t+1}=v^t+\Delta v^{t+1} $
		\ENDFOR
	\end{algorithmic}
\end{algorithm}

We now present Algorithm~\ref{alg:membet}, which is Algorithm~\ref{alg:5GCS} written in a memory-efficient manner. We use the fact that we do not need any information on specific $u_\iclient^t$ and that not all $u_\iclient^t$ are updated in each communication round.

\newpage
            \refstepcounter{chapter}%
\chapter*{\thechapter \quad Appendix E Title}
\label{appendixE}

\section{Basic Facts and Notation}
\subsection{Basic facts}
For any two vectors $a, b \in \R^d$ and any $\zeta > 0$,
\begin{equation}
	\label{eq:young}
	2 \ev{a, b} \leq \frac{\sqn{a}}{\zeta} + \zeta \sqn{b}.
\end{equation}
A consequence of \eqref{eq:young} is that for any $a, b \in \R^d$, we have
\begin{equation}
	\label{eq:sqnorm-triangle}
	\sqn{a + b} \leq \br{1 + \zeta} \sqn{a} + \br{1 + \zeta^{-1}} \sqn{b}.
\end{equation}
Using $\zeta = 1$ specifically yields,
\begin{equation}
	\label{eq:sqnorm-triangle-2}
	\sqn{a + b} \leq 2 \sqn{a} + 2 \sqn{b}.
\end{equation}
A function $h:\R^d\to \R$ is called $\mu$-convex if for some $\mu \geq 0$ and for all $x, y \in \R^d$, we have
\begin{equation}
	\label{eq:mu-convexity}
	h(x) + \ev{\nabla h(x), y - x} + \frac{\mu}{2} \sqn{y - x} \leq h(y).
\end{equation}
A function $h:\R^d\to \R$ is called $L$-smooth if for some $L\geq 0$ and for all $x, y \in \R^d$, we have
\begin{equation}
	\label{eq:nabla-Lip}
	\norm{\nabla h(x) - \nabla h(y)} \leq L \norm{x - y}.
\end{equation}
A useful consequence of $L$-smoothness is the inequality
\begin{equation}
	\label{eq:L-smoothness}
	h(x) \leq h(y) + \ev{\nabla h(y), x - y} + \frac{L}{2} \sqn{x - y},
\end{equation}
holding for all $x,y\in \R^d$. If $h$ is $L$-smooth and lower bounded by $h_\ast$, then
\begin{equation}
	\label{eq:grad-bound}
	\sqn{\nabla h(x)} \leq 2 L \br{h(x) - h_\ast}.
\end{equation}
For any convex and $L$-smooth function $h$ it holds that that
\begin{align}
	\norm{\nabla h(x) - \nabla h(y)}^2 \le 2L D_h(x, y). \label{eq:grad_dif_to_bregman}
\end{align}

For a convex function $h\colon \R^d \to \R$ and any vectors $y_1,\dots,y_n \in \R^d$,  Jensen's inequality states that
\begin{equation}
	\label{eq:jensen}
	h\br{\frac{1}{n} \sum\limits_{i=1}^{n} y_i} \leq \frac{1}{n} \sum\limits_{i=1}^{n} h(y_i).
\end{equation}
Applying this to the squared norm, $h(y) = \sqn{y}$, we get
\begin{equation}
	\label{eq:sqnorm-jensen}
	\sqn{ \frac{1}{n} \sum\limits_{i=1}^{n} y_i } \leq \frac{1}{n} \sum\limits_{i=1}^{n} \sqn{y_i}.
\end{equation}
Simple multiplication on both sides of \eqref{eq:sqnorm-jensen} also yields,
\begin{equation}
	\label{eq:sqnorm-sum-bound}
	\sqn{\sum\limits_{i=1}^{n} y_i} \leq n \sum\limits_{i=1}^{n} \sqn{y_i}.
\end{equation}
We use the following decomposition that holds for any random variable $X$ with $\ec{\norm{X}^2}<+\infty$,
\begin{align}
	\ec{\norm{X}^2}=\norm{\ec{X}}^2 + \ec{\norm{X-\ec{X}}^2}. \label{eq:rv_moments}
\end{align}
We will make use of the particularization of \eqref{eq:rv_moments} to the discrete case: Let $y_{1}, \ldots, y_{n} \in \R^d$ be given vectors and let $\bar{y} = \frac{1}{n} \sum\limits_{i=1}^{n} y_i$ be their average. Then,
\begin{equation}
	\label{eq:variance-decomp}
	\frac{1}{n} \sum\limits_{i=1}^{n} \sqn{y_{i}} = \sqn{ \bar{y} } + \frac{1}{n} \sum\limits_{i=1}^{n} \sqn{y_i - \bar{y}}.
\end{equation}

\subsection{Notation}
We define the variance of the local gradients from their average at a point $x^{t}$ as
\[ \sigma_{t}^2 \eqdef \frac{1}{n} \sum\limits_{j=1}^{n} \sqn{\nabla f_{j} (x^{t}) - \nabla f(x^{t})}. \]

A summary of the notation used is given in Table~\ref{tab:notation}.

\begin{table}[h]
	\centering
	\caption{Summary of notation used.}
    \small
	\label{tab:notation}
	\begin{tabular}{@{}cl@{}}
		\toprule
		Symbol               & Description                                                                                                                                    \\ \midrule
		$x^{t}$                & The iterate used at the start of epoch $t$.                                                                                                    \\ \midrule
		$\pi_m$ &
		\begin{tabular}[c]{@{}l@{}}A permutation $\pi_m = \br{ \pi^{0}_m, \pi^{1}_m, \ldots, \pi_m^{n-1} }$ of $\{ 1, 2, \ldots, n \}$,\\ which is resampled every epoch for Random Reshuffling.\end{tabular} \\ \midrule
		$\cstep$             & The stepsize used when taking descent steps in an epoch.                                                                                       \\ \midrule
		$x^{t}_{m,i}$              & The current iterate after $i$ steps in epoch $t$, for $0 \leq i \leq n$.                                                                       \\ \midrule
		$g^t$                & The sum of gradients used over epoch $t$ such that $x^{t+1}=x^{t}-\sstep g^t$.                                                                                                      \\ \midrule
		$\beta$              & The epoch jumping parameter.                                                                                                                   \\ \midrule
		$\sstep$               & The effective epoch stepsize, defined as $\sstep \eqdef \cstep \br{1 + \beta}n$.                                                                  \\ \midrule
		$\sigma_{t}^2$       & The variance of the individual loss gradients from the average loss at $x^{t}$.                                                                                                                                                     \\ \midrule
		$L$                  & The smoothness constant of $f$ and each $f_{m,i}$.                                                                              \\ \midrule
		$\delta^{t}$         & \begin{tabular}[c]{@{}c@{}}Functional suboptimality, $\delta^{t} = f(x^{t}) - f^\star$, where $f^\star= \inf_{x} f(x)$.\end{tabular}     \\ \bottomrule
	\end{tabular}
\end{table}
\subsection{Sampling without replacement}
We provide the full proof of Lemma~\ref{lem:sampling_wo_replacement}.
\begin{lemma_empt}
	Let $X_1,\dotsc, X_n\in \R^d$ be fixed vectors, $\overline X\eqdef \frac{1}{n}\sum\limits_{i=1}^n X_i$ be their average and $\sigma^2 \eqdef \frac{1}{n}\sum\limits_{i=1}^n \norm{X_i-\overline X}^2$ be the population variance. Fix any $k\in\{1,\dotsc, n\}$, let $X_{\pi_1}, \dotsc X_{\pi_k}$ be sampled uniformly without replacement from $\{X_1,\dotsc, X_n\}$ and $\overline X_\pi$ be their average. Then, it holds
	\begin{align}
		\ec{\overline X_\pi}=\overline X, && \ec{\norm{\overline X_{\pi} - \overline X}^2}= \frac{n-k}{k(n-1)}\sigma^2.
	\end{align}
\end{lemma_empt}
\begin{proof}
	The first claim follows by linearity of the expectation and uniformity of the sampling,
	\begin{align*}
		\ec{\overline X_\pi} 
		= \frac{1}{k}\sum\limits_{i=1}^k \ec{X_{\pi_i}}
		= \frac{1}{k}\sum\limits_{i=1}^k \overline X
		= \overline X.
	\end{align*}
	To show the second claim, let us first establish that for any $i\neq j$ it holds $\mathrm{cov}(X_{\pi_i}, X_{\pi_j})=-\frac{\sigma^2}{n-1}$. Indeed,
	\begin{align*}
		\mathrm{cov}(X_{\pi_i}, X_{\pi_j})
		&= \ec{ \ev{X_{\pi_i} - \overline X, X_{\pi_j} - \overline X}}\\
		&= \frac{1}{n(n-1)}\sum\limits_{l=1}^n\sum\limits_{m\neq l}\ev{X_l - \overline X, X_m - \overline X} \\
		&= \frac{1}{n(n-1)}\sum\limits_{l=1}^n\sum\limits_{m=1}^n\ev{X_l - \overline X, X_m - \overline X} - \frac{1}{n(n-1)}\sum\limits_{l=1}^n \norm{X_l - \overline X}^2 \\
		&= \frac{1}{n(n-1)}\sum\limits_{l=1}^n \ev{X_l - \overline X, \sum\limits_{m=1}^n(X_m - \overline X)} - \frac{\sigma^2}{n-1} \\
		&=-\frac{\sigma^2}{n-1}.
	\end{align*}
	Therefore,
	\begin{align*}
		\ecn{\overline X_{\pi} - \overline X}
		&= \frac{1}{k^2} \sum\limits_{i=1}^k\sum\limits_{j=1}^k \mathrm{cov}(X_{\pi_i}, X_{\pi_j}) \\
		&= \frac{1}{k^2}\ec{\sum\limits_{i=1}^k \sqn{X_{\pi_i} - \overline X}} + \sum\limits_{i=1}^k\sum\limits_{j=1,j\neq i}^{n} \mathrm{cov}(X_{\pi_i}, X_{\pi_j})  \\
		&=\frac{1}{k^2}\left(k\sigma^2 - k(k-1)\frac{\sigma^2}{n-1}\right)
		= \frac{n-k}{k(n-1)}\sigma^2.
	\end{align*}
\end{proof}

\section{Large Server Stepsize}
\subsection{Strongly convex and general convex case}\label{section:C.1}
\begin{lemma}
	\label{lemma:inner-product-convex}
	Let Assumption~\ref{assump: L-smooth_1} hold and further assume $f$ is $\mu$-strongly convex and each $f_m^i$ is convex. Then
	\begin{align*}
		- \frac{1}{Mn}\sum\limits_{m=1}^{M}\sum\limits_{i=0}^{n-1}\left\langle f_{m, \pi^i_m} \left(x_{m,i}^t\right),x^{t} - x^\star  \right\rangle &\leq -\frac{\mu}{4}\|x^{t} - x^\star\|^2 - \frac{1}{2}\left(f\left(x^{t}\right) - f\left(x^\star\right)\right)\\
        &+\frac{L}{2Mn}\sum\limits_{m=1}^{M}\sum\limits_{i=0}^{n-1} \left\|x^{t} - x_{m,i}^t\right\|^2.  
	\end{align*}
\end{lemma}

\begin{proof}
	We start with the inner product and decompose it using the three-point identity:
	\begin{align}
	\notag	\left\langle \nabla f_{m, \pi^i_m} \left(x_{m,i}^t\right),x^{t} - x^\star \right\rangle &= f_{m, \pi^i_m}\left(x^{t}\right) - f_{m, \pi^i_m}\left(x^\star\right) + f_{m, \pi^i_m}\left(x^\star\right) - f_{m, \pi^i_m}\left(x_{m,i}^t\right)\\\notag
		&+   	\left\langle \nabla f_{m, \pi^i_m} \left(x_{m,i}^t\right),x_{m,i}^t - x^\star \right\rangle - f_{m, \pi^i_m}\left(x^{t}\right)\\\notag
		&+f_{m, \pi^i_m}\left(x_{m,i}^t\right)+	\left\langle \nabla f_{m, \pi^i_m} \left(x_{m,i}^t\right),x^{t} - x_{m,i}^t \right\rangle \\
		&\notag = f_{m, \pi^i_m}\left(x^{t}\right) - f_{m, \pi^i_m}\left(x^\star\right) + D_{f_{m, \pi^i_m}}\left(x^\star,x_{m,i}^t\right)\\
        &- D_{f_{m, \pi^i_m}}\left(x^{t},x_{m,i}^t\right).
		\label{eq:3point}
	\end{align}
	Using the representation~\eqref{eq:3point}, $L$-smoothness and $\mu$-strong convexity we have a bound:
	\begin{align*}
		&- \frac{1}{Mn}\sum\limits_{m=1}^{M}\sum\limits_{i=0}^{n-1}\left\langle f_{m, \pi^i_m} \left(x_{m,i}^t\right),x^{t} - x^\star  \right\rangle\\ &\leq -  \frac{1}{Mn}\sum\limits_{m=1}^{M}\sum\limits_{i=0}^{n-1} \left(f_{m, \pi^i_m}\left(x^{t}\right) - f_{m, \pi^i_m}\left(x^\star\right) + D_{f_{m, \pi^i_m}}\left(x^\star,x_{m,i}^t\right)\right.\\
        &\left.- D_{f_{m, \pi^i_m}}\left(x^{t},x_{m,i}^t\right)\right)\\
		&\overset{\eqref{eq:L-smoothness}}{\leq} -\left(f\left(x^{t}\right) - f\left(x^\star\right)\right) - \frac{1}{Mn}\sum\limits_{m=1}^{M}\sum\limits_{i=0}^{n-1} D_{f_{m, \pi^i_m}}\left(x^\star,x_{m,i}^t\right)\\
        &+ \frac{L}{2Mn}\sum\limits_{m=1}^{M}\sum\limits_{i=0}^{n-1} \left\|x^{t} - x_{m,i}^t\right\|^2\\
		&\overset{\eqref{eq:mu-convexity}}{\leq}-\frac{\mu}{4}\|x^{t} - x^\star\|^2 - \frac{1}{2}\left(f\left(x^{t}\right) - f\left(x^\star\right)\right)+\frac{L}{2Mn}\sum\limits_{m=1}^{M}\sum\limits_{i=0}^{n-1} \left\|x^{t} - x_{m,i}^t\right\|^2.
	\end{align*}
\end{proof}

\begin{lemma}
	\label{thm:sqrt}
	Assume that Assumption~\ref{assump: L-smooth_1} holds, then 
	\begin{align*}
		\left\| \frac{1}{Cn} \sum_{m\in \set} \sum\limits_{i=0}^{n-1} \nabla f_{m, \pi^i_m} \left(x_{m,i}^t\right)\right\|^2 &\leq2\frac{L^2}{Cn} \sum_{m\in \set} \sum\limits_{i=0}^{n-1}\| x_{m,i}^t - x^{t} \|^2\\
        &+ 4\left\|\frac{1}{C}\sum_{m\in \set} \nabla f_m(x^\star)\right\|^2 + 8L(f_m(x^{t}) - f_m(x^\star)).
	\end{align*}	
\end{lemma}
\begin{proof}
	We start with Young's inequality. Note that $f_m(x^{t}) = \frac{1}{n} \sum_{i=0}^{n-1}\nabla f_{m, \pi_m^i}(x^{t})$:
	\begin{align*}
		&\left\| \frac{1}{Cn}  \sum_{m\in \set} \sum\limits_{i=0}^{n-1} \nabla f_{m, \pi^i_m} \left(x_{m,i}^t\right)\right\|^2 \\
        &\overset{\eqref{eq:sqnorm-triangle-2}}{\leq} 2 \left\| \frac{1}{Cn} \sum_{m\in \set}\sum\limits_{i=0}^{n-1}\left( \nabla f_{m, \pi^i_m} \left(x_{m,i}^t\right)- \nabla f_{m, \pi^i_m}(x^{t})\right)\right\|^2\\
        &+ 2\left\|\frac{1}{C} \sum_{m\in \set}\nabla f_m(x^{t}) \right\|^2\\
		&\overset{\eqref{eq:jensen},\eqref{eq:nabla-Lip}}{\leq}  2L^2\frac{1}{Cn}\sum_{m\in \set} \sum\limits_{i=0}^{n-1}\| x_{m,i}^t - x^{t} \|^2\\
        &+ 2\left\|\frac{1}{C} \sum_{m\in \set}\nabla f_m(x^{t}) \right\|^2.
	\end{align*}
	We use Young's inequality and $L$-smoothness again:
	\begin{align*}
		\left\|  \frac{1}{Cn} \sum_{m\in \set}\sum\limits_{i=0}^{n-1} \nabla f_{m, \pi^i_m} \left(x_{m,i}^t\right)\right\|^2& \overset{\eqref{eq:jensen},\eqref{eq:sqnorm-triangle-2}}{\leq} 2L^2\frac{1}{Cn} \sum_{m\in \set}\sum\limits_{i=0}^{n-1}\| x_{m,i}^t - x^{t} \|^2 \\
        &+ 4\left\|\frac{1}{C}\sum_{m\in \set}\nabla f_m(x^\star)\right\|^2\\
		& + 4\frac{1}{C}\sum_{m\in \set}\| \nabla f_m(x^{t}) - \nabla f_m(x^\star) \|^2\\
		& \overset{\eqref{eq:grad_dif_to_bregman}}{\leq} 2L^2\frac{1}{Cn} \sum_{m\in \set}\sum\limits_{i=0}^{n-1}\| x^i_{m,t} - x^{t} \|^2\\
        &+ 4\left\|\frac{1}{C}\sum_{m\in \set}\nabla f_m(x^\star)\right\|^2\\
		& + 8L\frac{1}{C}\sum_{m\in \set}(f_m(x^{t}) - f_m(x^\star)).
	\end{align*}
\end{proof}
\begin{lemma}
	\label{lemma:V_t}
	Suppose that Algorithm~\ref{alg:pp-jumping} is used and Assumption~\ref{assump: L-smooth_1} holds. If $\cstep \leq \frac{1}{2Ln}$, then
	\begin{align*}
		\frac{1}{Mn}\sum\limits_{m=1}^{M}\sum\limits_{i=0}^{n-1} \mathbb{E}\left[\left\|x^{t} - x_{m,i}^t\right\|^2|x^{t}\right]\ &\leq 8\cstepsquared n^2L \left(f(x^{t}) - f(x^\star)\right)\\
        &+2\cstepsquared n^2\frac{1}{M}\sum\limits_{m=1}^{M}\left\| \nabla f_m(x^\star) \right\|^2\\
        &+ 2 \cstepsquared  n \frac{1}{M}\sum\limits_{m=1}^{M}\sigma^2_{*,m}.
	\end{align*}
\end{lemma}
\begin{proof}
	We start from the definition of $x_{m,i}^t$:
	\begin{align*}
		\mathbb{E}\left[ \left\|x_{m,i}^t - x^{t}\right\|^2|x^{t}\right] &= \mathbb{E}\left[\left\| \cstep \sum\limits_{j=0}^{i-1}\nabla f_{m}^{\pi^j_m}\left(x^t_{m,j}\right) \right\|^2\vert x^{t} \right]\\
		&\overset{\eqref{eq:sqnorm-triangle-2}}{\leq} 2\cstepsquared  \mathbb{E}\left[\left\|\sum\limits_{j=0}^{i-1}\left(\nabla f_{m}^{\pi^j_m}\left(x^t_{m,j}\right) - \nabla f_{m}^{\pi^j_m}\left(x^{t}\right)\right)\right\|^2|x^{t}\right]\\
        &+2\cstepsquared \mathbb{E}\left[\left\|\sum\limits_{j=0}^{i-1}\nabla f_{m}^{\pi^j_m}\left(x^{t}\right)\right\|^2|x^{t}\right]\\
		&\overset{\eqref{eq:jensen}}{\leq}  2\cstepsquared  i \sum\limits_{j=0}^{i-1}\mathbb{E}\left[\left\| \nabla f_{m}^{\pi^j_m} \left(x^t_{m,j}\right) - \nabla f_{m}^{\pi^j_m}(x^{t}) \right\|^2|x^{t}\right]\\
        &+2\cstepsquared \mathbb{E}\left[\left\|\sum\limits_{j=0}^{i-1}\nabla f_{m}^{\pi^j_m}\left(x^{t}\right)\right\|^2|x^{t}\right]\\
		&\overset{\eqref{eq:nabla-Lip}}{\leq} 2\cstepsquared L^2 i \sum\limits_{j=0}^{i-1}\mathbb{E}\left[\left\| x^t_{m,j} - x^{t} \right\|^2|x^{t}\right]\\
        &+2\cstepsquared \mathbb{E}\left[\left\|\sum\limits_{j=0}^{i-1}\nabla f_{m}^{\pi^j_m}\left(x^{t}\right)\right\|^2|x^{t}\right].
	\end{align*}
	Now let us look at the last term. We can apply Lemma~\ref{lem:sampling_wo_replacement} and get
	\begin{align*}
		\mathbb{E}\left[\left\| \sum\limits_{j=0}^{i-1}\nabla f_{m}^{\pi^j_m}(x^{t}) \right\|^2|x^{t}\right] &= i^2\left\| \nabla f_m(x^{t}) \right\|^2\\
        &+ i^2\mathbb{E}\left[\left\| \frac{1}{i}\sum\limits_{j=0}^{i-1}\left( \nabla f_{m}^{\pi^j_m}(x^{t}) - \nabla f_m(x^{t}) \right) \right\|^2|x^{t}\right]\\
		&= i^2\left\| \nabla f_m(x^{t}) \right\|^2 + \frac{i(n-i)}{n-1}\sigma_{t,m}^2,
	\end{align*}
where $\sigma_{t, m}^{2} \stackrel{\text { def }}{=} \frac{1}{n} \sum_{i=1}^{n}\left\|\nabla f_{m,i}\left(x^{t}\right) - \nabla f_{m}(x^{t})\right\|^{2}$.

	Let us go back:
	\begin{align*}
		\mathbb{E}\left[ \left\|x_{m,i}^t - x^{t}\right\|^2 |x^{t}\right]&\leq 2\cstepsquared L^2 i \sum\limits_{j=0}^{i-1}\mathbb{E}\left[\left\| x^t_{m,j} - x^{t} \right\|^2|x^{t}\right]\\
        &+ 2\cstepsquared \left( i^2\left\| \nabla f_m(x^{t}) \right\|^2 + \frac{i(n-i)}{n-1}\sigma_{t,m}^2 \right).
	\end{align*}
	Summing the terms leads to
	\begin{align*}
		\frac{1}{Mn}\sum\limits_{m=1}^{M}\sum\limits_{i=0}^{n-1}&\mathbb{E}\left[\left\|x_{m,i}^t - x^{t}\right\|^2|x^{t}\right]\\
        &\leq 2\cstepsquared L^2\frac{1}{Mn}\sum\limits_{m=1}^{M}\sum\limits_{i=0}^{n-1} i \sum\limits_{j=0}^{i-1}\mathbb{E}\left[\left\| x^t_{m,j} - x^{t} \right\|^2|x^{t}\right]\\
		& + \frac{2\cstepsquared }{Mn}\sum\limits_{m=1}^{M}\sum\limits_{i=0}^{n-1}i^2\left\| \nabla f_m(x^{t}) \right\|^2\\
        &+ \frac{2\cstepsquared }{Mn}\sum\limits_{m=1}^{M}\sum\limits_{i=0}^{n-1}\frac{i(n-i)\sigma_{t,m}^2}{n-1}\\
		&\leq 2\cstepsquared  L^2 \frac{1}{Mn}\sum\limits_{m=1}^{M}\sum\limits_{i=0}^{n-1}\mathbb{E}\left[\left\|x_{m,i}^t - x^{t}\right\|^2|x^{t}\right] \cdot \frac{n(n-1)}{2}\\
		&+\frac{2\cstepsquared }{Mn}\sum\limits_{m=1}^{M}\left\|\nabla f_m(x^{t})\right\|^2 \cdot \frac{n(n-1)(2n-1)}{6} \\
        &+\frac{\cstepsquared n(n+1)}{3}\frac{1}{Mn}\sum\limits_{m=1}^{M}\sigma_{t,m}^2.
	\end{align*}
	Choosing $\cstep\leq \frac{1}{2Ln}$, we verify
	\begin{align}
		\label{eq:32}
	\notag	\frac{1}{Mn}\sum\limits_{m=1}^{M}\sum\limits_{i=0}^{n-1}&\mathbb{E}\left[\left\|x_{m,i}^t - x^{t}\right\|^2|x^{t}\right]\\
  \notag  &\leq \frac{4}{3}\left( 1 - \cstepsquared  L^2 n(n-1) \right)	\frac{1}{Mn}\sum\limits_{m=1}^{M}\sum\limits_{i=0}^{n-1}\mathbb{E}\left[\left\|x_{m,i}^t - x^{t}\right\|^2|x^{t}\right]\\
	\notag	&\leq \frac{4\cstepsquared }{9}\frac{1}{M}\sum\limits_{m=1}^{M}\left\|\nabla f_m(x^{t})\right\|^2 \cdot (n-1)(2n-1) \\
  \notag  &+\frac{4\cstepsquared (n+1)}{9}\frac{1}{M}\sum\limits_{m=1}^{M}\sigma_{t,m}^2\\
		&\leq \cstepsquared n^2\frac{1}{M}\sum\limits_{m=1}^{M}\left\|\nabla f_m(x^{t})\right\|^2 + \cstepsquared  n \frac{1}{M}\sum\limits_{m=1}^{M}\sigma_{t,m}^2.
	\end{align}
	Using Young's inequality, we get
	\begin{align*}
	\notag	\frac{1}{Mn}\sum\limits_{m=1}^{M}\sum\limits_{i=0}^{n-1}&\mathbb{E}\left[\left\|x_{m,i}^t - x^{t}\right\|^2|x^{t}\right]\\
    &\overset{\eqref{eq:sqnorm-triangle-2},\eqref{eq:variance-decomp}}{\leq} 2\cstepsquared n^2 \frac{1}{M}\sum\limits_{m=1}^{M}\left\|\nabla f_m(x^{t}) - \nabla f_m(x^\star)\right\|^2\\
		\notag	& + 2\cstepsquared n^2\frac{1}{M}\sum\limits_{m=1}^{M}\left\| \nabla f_m(x^\star) \right\|^2-  \frac{\cstepsquared  n}{M}\sum\limits_{m=1}^{M}\left\| \nabla f_{m}(x^{t})  \right\|^2\\
		\notag	&+2\cstepsquared  n \frac{1}{M}\sum\limits_{m=1}^{M}\frac{1}{n}\sum\limits_{i=0}^{n-1}\mathbb{E}\left[\left\| \nabla f_{m, \pi^i_m}(x^{t}) - \nabla  f_{m, \pi^i_m}(x^\star) \right\|^2\right]\\
		&  + 2\cstepsquared  n \frac{1}{M}\sum\limits_{m=1}^{M}\frac{1}{n}\sum\limits_{i=0}^{n-1}\mathbb{E}\left[\left\| \nabla  f_{m, \pi^i_m}(x^\star) \right\|^2\right].
	\end{align*}
	Using $L$-smoothness, we obtain
	\begin{align*}
		\frac{1}{Mn}\sum\limits_{m=1}^{M}\sum\limits_{i=0}^{n-1}&\mathbb{E}\left[\left\|x_{m,i}^t - x^{t}\right\|^2\right]\\&\leq 4\cstepsquared n^2  L \frac{1}{M}\sum\limits_{m=1}^{M} D_{f_m}(x^{t},x^\star)  + 2 \cstepsquared  n \frac{1}{M}\sum\limits_{m=1}^{M}\sigma^2_{*,m} \\
		&+ 4 \cstepsquared n L\frac{1}{M}\sum\limits_{m=1}^{M}\frac{1}{n}\sum\limits_{i=0}^{n-1}D_{f_{m, \pi^i_m}}(x^{t},x^\star)\\
        &+2\cstepsquared n^2\frac{1}{M}\sum\limits_{m=1}^{M}\left\| \nabla f_m(x^\star) \right\|^2 \\
		&\overset{\eqref{eq:L-smoothness}}{\leq} 8\cstepsquared n^2L \left(f(x^{t}) - f(x^\star)\right)+2\cstepsquared n^2\frac{1}{M}\sum\limits_{m=1}^{M}\left\| \nabla f_m(x^\star) \right\|^2\\
        &+ 2 \cstepsquared  n \frac{1}{M}\sum\limits_{m=1}^{M}\sigma^2_{*,m}.
	\end{align*}
\end{proof}
\subsubsection{Proof of Theorem~\ref{thm:PP-SC}}
\begin{theorem_empt}
	Assume that Assumption~\ref{assump: L-smooth_1} holds and $f$ is $\mu$-strongly convex function. Let $\cstep n\leq \sstep\leq \frac{1}{16L}$. Then for iterates $x^{t}$ generated by Algorithm~\ref{alg:pp-jumping} we have 
	\begin{align*}
		\mathbb{E}\left[\| x^{T} - x^\star \|^2\right] &\leq \left(1 - \frac{\sstep\mu}{2}\right)^T  \left\| x^{0} - x^\star \right\|^2\\
        &+ \frac{5\cstepsquared nL}{\mu}\frac{1}{M}\sum\limits_{m=1}^{M}\left(\sigma^2_{*,m} + n \|\nabla f_m(x^\star)\|^2 \right)\\
        &+\frac{8\sstep}{\mu}\sum\limits_{m=1}^{M}\|\nabla f_m(x^\star)\|^2.
	\end{align*}
\end{theorem_empt}
\begin{proof}
	We start from definition of $x^{t+1}$,
	\begin{align*}
		\|x^{t+1} - x^\star\|^2 &= \left\| x^{t} - \sstep\frac{1}{Cn}\sum_{m\in \set}\sum\limits_{i=0}^{n-1} \nabla f_{m, \pi^i_m}\left(x_{m,i}^t \right) - x^\star \right\|^2\\
		&= \|x^{t} - x^\star\|^2 - 2\sstep \left\langle \frac{1}{Cn}\sum_{m\in \set} \sum\limits_{i=0}^{n-1} \nabla f_{m, \pi^i_m} \left(x_{m,i}^t\right), x^{t} - x^\star \right\rangle\\
        &+\sstepsquared\left\| \frac{1}{Cn} \sum_{m\in \set} \sum\limits_{i=0}^{n-1} \nabla f_{m, \pi^i_m} \left(x_{m,i}^t\right)\right\|^2.
	\end{align*}
Using Lemma~\ref{thm:sqrt}, we get 
	\begin{align*}
		\|x^{t+1} - x^\star\|^2	&\leq \|x^{t} - x^\star\|^2 - 2\sstep \left\langle \frac{1}{Cn}\sum_{m\in \set} \sum\limits_{i=0}^{n-1} \nabla f_{m, \pi^i_m} \left(x_{m,i}^t\right), x^{t} - x^\star \right\rangle\\
		&+\sstepsquared \left( 2L^2\frac{1}{Cn} \sum_{m\in \set}\sum\limits_{i=0}^{n-1}\| x^i_{m,t} - x^{t} \|^2 + 4\left\|\frac{1}{C}\sum_{m\in \set}\nabla f_m(x^\star)\right\|^2 \right)\\
        &+ 8L \sstepsquared\frac{1}{C}\sum_{m\in \set}(f_m(x^{t}) - f_m(x^\star)).
	\end{align*}
Taking conditional expectation over sampling $\set$, we get
	\begin{align*}
	\mathbb{E}_{\set}\left[\|x^{t+1} - x^\star\|^2\right]	&\leq \|x^{t} - x^\star\|^2\\
    &- 2\sstep\mathbb{E}_{\set}\left[ \left\langle \frac{1}{Cn}\sum_{m\in \set} \sum\limits_{i=0}^{n-1} \nabla f_{m, \pi^i_m} \left(x_{m,i}^t\right), x^{t} - x^\star \right\rangle\right]\\
	&+  2\sstepsquared L^2\mathbb{E}_{\set}\left[\frac{1}{Cn} \sum_{m\in \set}\sum\limits_{i=0}^{n-1}\| x_{m,i}^t - x^{t} \|^2\right] \\
    &+ 4\sstepsquared\left\|\frac{1}{C}\sum_{m\in \set}\nabla f_m(x^\star)\right\|^2\\
    &+ 8\sstepsquared L\frac{1}{C}\sum_{m\in \set}(f_m(x^{t}) - f_m(x^\star))\\
	&\leq \|x^{t} - x^\star\|^2\\
    &- 2\sstep \left\langle \frac{1}{Mn}\sum^{M}_{m=1} \sum\limits_{i=0}^{n-1} \nabla f_{m, \pi^i_m} \left(x_{m,i}^t\right), x^{t} - x^\star \right\rangle\\
	&+\sstepsquared 2L^2\frac{1}{Mn} \sum_{m=1}^{M}\sum\limits_{i=0}^{n-1}\| x_{m,i}^t - x^{t} \|^2\\
    &+ \sstepsquared\left( 4\mathbb{E}_{\set}\left[\left\|\frac{1}{C}\sum_{m\in \set}\nabla f_m(x^\star)\right\|^2\right] + 8L(f(x^{t}) - f(x^\star))\right)\\
	&\overset{\eqref{lem:sampling_wo_replacement}}{\leq} \|x^{t} - x^\star\|^2 - 2\sstep \left\langle \frac{1}{Mn}\sum^{M}_{m=1} \sum\limits_{i=0}^{n-1} \nabla f_{m, \pi^i_m} \left(x_{m,i}^t\right), x^{t} - x^\star \right\rangle\\
	&+ 2\sstepsquared L^2\frac{1}{Mn} \sum_{m=1}^{M}\sum\limits_{i=0}^{n-1}\| x_{m,i}^t - x^{t} \|^2\\
    &+ 4\sstepsquared\frac{M-C}{C\max\left\lbrace M-1,1 \right\rbrace}\sigma_{\star}^2\\
    &+ 8\sstepsquared L(f(x^{t}) - f(x^\star)).
\end{align*}
Using Lemma~\ref{lemma:inner-product-convex}, we obtain 
\begin{align*}
	\mathbb{E}_{\set}\left[\|x^{t+1} - x^\star\|^2\right]&\leq \|x^{t} - x^\star\|^2\\
    &- 2\sstep \left(-\frac{\mu}{4}\left\|x^{t}-x^{\star}\right\|^{2}-\frac{1}{2}\left(f\left(x^{t}\right)-f\left(x^{\star}\right)\right)\right)\\
    &-\sstep \frac{L}{ M n} \sum_{m=1}^{M} \sum_{i=0}^{n-1}\left\|x^{t}-x_{t, m}^{i}\right\|^{2}\\
&+2\sstepsquared L^2\frac{1}{Mn} \sum_{m=1}^{M}\sum\limits_{i=0}^{n-1}\| x_{m,i}^t - x^{t} \|^2\\
&\sstepsquared\left( 4\frac{M-C}{C\max\left\lbrace M-1,1 \right\rbrace}\sigma_{\star}^2+ 8L(f(x^{t}) - f(x^\star))\right).
\end{align*}

	Rearranging the terms, we obtain:
	\begin{align*}
	\mathbb{E}_{\set}\left[\|x^{t+1} - x^\star\|^2\right]
		&	 \leq \left\| x^{t} - x^\star \right\|^2\left(1 - \frac{\sstep\mu}{2}\right)  - \sstep\left(1 - 8\sstep L \right)\left(f(x^{t}) - f(x^\star)\right)\\
		&+\sstep L\left(1+2\sstep L\right)\frac{1}{Mn}\sum\limits_{m=1}^{M}\sum\limits_{i=0}^{n-1}\|x^i_{m,t}-x^{t}\|^2\\
        &+4\sstepsquared\frac{M-C}{C\max\left\lbrace M-1,1 \right\rbrace}\sigma_{\star}^2.
\end{align*}
Using the tower property of conditional expectation and Lemma~\ref{lemma:V_t}, we get
\begin{align}
	\label{eq:need for C}
	\notag	\mathbb{E}\left[\|x^{t+1} - x^\star\|^2|x^{t}\right]	&\leq \left\| x^{t} - x^\star \right\|^2\left(1 - \frac{\sstep\mu}{2}\right)+4\sstepsquared\frac{M-C}{C\max\left\lbrace M-1,1 \right\rbrace}\sigma_{\star}^2\\
	&\notag - \sstep \left( 1 - 8\sstep L - \left(1+2\sstep L\right)8\cstepsquared n^2L^2 \right) \left( f(x^{t}) - f(x^\star) \right)\\
			&+2\sstep \left(1+2\sstep L\right)\cstepsquared nL\frac{1}{M}\sum\limits_{m=1}^{M}\left(\sigma^2_{*,m} + n \|\nabla f_m(x^\star)\|^2 \right).
	\end{align}
	Taking $\cstep \leq \frac{1}{16nL}$ and $\sstep \leq \frac{1}{16L}$, we derive
	\begin{align*}
		\sstep \left( 1 - 8\sstep L - \left(1+2\sstep L\right)8\cstepsquared n^2L^2 \right) \left( f(x^{t}) - f(x^\star) \right) \ge 0.
	\end{align*} 
	Taking full expectation yields 
	\begin{align*}
		\mathbb{E}\left[\| x^{t+1} - x^\star \|^2\right] &\leq \mathbb{E}\left[ \left\| x^{t} - x^\star \right\|^2\left(1 - \frac{\sstep\mu}{2}\right)\right]\\
        &+ \frac{5}{2}\sstep\cstepsquared  n L \frac{1}{M}\sum\limits_{m=1}^{M}\left(\sigma^2_{*,m} + n \|\nabla f_m(x^\star)\|^2 \right)\\
        &+4\frac{M-C}{C\max\left\lbrace M-1,1 \right\rbrace}\sigma_{\star}^2.
	\end{align*}
	Unrolling this recursion, we have 
	\begin{align*}
		\mathbb{E}\left[\| x^{T} - x^\star \|^2\right] &\leq \left(1 - \frac{\sstep\mu}{2}\right)^T  \left\| x^{0} - x^\star \right\|^2\\
        &+ \frac{5\cstepsquared nL}{\mu}\frac{1}{M}\sum\limits_{m=1}^{M}\left(\sigma^2_{*,m} + n \|\nabla f_m(x^\star)\|^2 \right)\\
        &+\frac{8\sstep}{\mu}\sum\limits_{m=1}^{M}\|\nabla f_m(x^\star)\|^2.
	\end{align*}
\end{proof}
\subsection{General convex case}
\subsubsection{Proof of Theorem~\ref{thm:PP-C}}
\begin{theorem_empt}
	Let Assumption~\ref{assump: L-smooth_1} hold, each $f_{m,i}$ be convex function. Let $\cstep n \leq \sstep\leq \frac{1}{16L}$. Let $\hat{x}_{T} \eqdef \frac{1}{T} \sum_{t=1}^{T} x^{t}$. Then for iterates $x^{t}$ of Algorithm~\ref{alg:pp-jumping}, we have 
	\begin{align*}
		\mathbb{E}& [ f(\hat{x}^T) - f(x^\star)] \leq  \frac{ 5\left\| x^{0} - x^\star \right\|^2}{2\sstep  T}  + 7\cstepsquared nL \left(\frac{1}{M}\sum\limits_{m=1}^{M}\sigma^2_{*,m} + n \sigma_{\star}^2 \right)\\
        &+10\sstep \frac{M-C}{C\max\{M-1,1\}} \sigma_{\star}^2.
	\end{align*}

\end{theorem_empt}
\begin{proof}
	We start from equation~\eqref{eq:need for C} with $\mu = 0$:
	\begin{align*}
			\notag	\mathbb{E}\left[\|x^{t+1} - x^\star\|^2|x^{t}\right]	&\leq \left\| x^{t} - x^\star \right\|^2+4\sstepsquared\frac{M-C}{C\max\left\lbrace M-1,1 \right\rbrace}\sigma_{\star}^2\\
		& - \sstep \left( 1 - 8\sstep L - \left(1+2\sstep L\right)8\cstepsquared n^2L^2 \right) \left( f(x^{t}) - f(x^\star) \right)\\
		\notag	&+2\sstep \left(1+2\sstep L\right)\cstepsquared nL\frac{1}{M}\sum\limits_{m=1}^{M}\left(\sigma^2_{*,m} + n \|\nabla f_m(x^\star)\|^2 \right).
	\end{align*}
Using $\cstep n \leq \sstep\leq \frac{1}{16L}$, we obtain $ - \left( 1 - 8\sstep L - \left(1+2\sstep L\right)8\cstepsquared n^2L^2 \right) \leq -\frac{4}{10}$
	\begin{align*}
	\notag	\mathbb{E}\left[\|x^{t+1} - x^\star\|^2|x^{t}\right]	&\leq \left\| x^{t} - x^\star \right\|^2+4\sstepsquared\frac{M-C}{C\max\left\lbrace M-1,1 \right\rbrace}\sigma_{\star}^2\\
    &-  \frac{4\sstep}{10} \left( f(x^{t}) - f(x^\star) \right)\\
	\notag	&+\frac{5}{2}\sstep\cstepsquared nL\frac{1}{M}\sum\limits_{m=1}^{M}\left(\sigma^2_{*,m} + n \|\nabla f_m(x^\star)\|^2 \right).
\end{align*}
Taking full expectation, we get
	\begin{align*}
	\notag	\mathbb{E}\left[\|x^{t+1} - x^\star\|^2\right]	&\leq \mathbb{E}\left[\left\| x^{t} - x^\star \right\|^2\right]+4\sstepsquared\frac{M-C}{C\max\left\lbrace M-1,1 \right\rbrace}\sigma_{\star}^2\\
    &m-  \frac{4\sstep}{10} \mathbb{E}\left[\left( f(x^{t}) - f(x^\star) \right)\right]\\
	\notag	&+\frac{5}{2}\sstep\cstepsquared nL\frac{1}{M}\sum\limits_{m=1}^{M}\left(\sigma^2_{*,m} + n \|\nabla f_m(x^\star)\|^2 \right).
\end{align*}
Rearranging the terms leads us to
	\begin{align*}
	\notag	  \frac{4\sstep}{10} \mathbb{E}\left[\left( f(x^{t}) - f(x^\star) \right)\right]&\leq \mathbb{E}\left[\left\| x^{t} - x^\star \right\|^2\right] - \mathbb{E}\left[\|x^{t+1} - x^\star\|^2\right]\\
    &+4\sstepsquared\frac{M-C}{C\max\left\lbrace M-1,1 \right\rbrace}\sigma_{\star}^2\\ 
	\notag	&+\frac{5}{2}\sstep\cstepsquared nL\frac{1}{M}\sum\limits_{m=1}^{M}\left(\sigma^2_{*,m} + n \|\nabla f_m(x^\star)\|^2 \right).
\end{align*}
Averaging from $0$ to $T-1$, we get 
	\begin{align*}
 \frac{4\sstep}{10} \frac{1}{T}\sum_{t=0}^{T-1} \left[\left( f(x^{t}) - f(x^\star) \right)\right]	&\leq \frac{1}{T}\sum_{t=0}^{T-1} \left(\mathbb{E}\left[\left\| x^{t} - x^\star \right\|^2\right] - \mathbb{E}\left[\|x^{t+1} - x^\star\|^2\right]\right)\\
 &+4\sstepsquared\frac{M-C}{C\max\left\lbrace M-1,1 \right\rbrace}\sigma_{\star}^2\\
	\notag	&+\frac{5}{2}\sstep\cstepsquared nL\frac{1}{M}\sum\limits_{m=1}^{M}\left(\sigma^2_{*,m} + n \|\nabla f_m(x^\star)\|^2 \right)\\
	&\leq \frac{1}{T} \left(\mathbb{E}\left[\left\| x^{0} - x^\star \right\|^2\right] - \mathbb{E}\left[\|x^{T} - x^\star\|^2\right]\right)\\
    &+4\sstepsquared\frac{M-C}{C\max\left\lbrace M-1,1 \right\rbrace}\sigma_{\star}^2\\
	\notag	&+\frac{5}{2}\sstep\cstepsquared nL\frac{1}{M}\sum\limits_{m=1}^{M}\left(\sigma^2_{*,m} + n \|\nabla f_m(x^\star)\|^2 \right).
\end{align*}
Using Jensen's inequality~\eqref{eq:jensen}, we have

	\begin{align*}
	\mathbb{E} [ f(\hat{x}^T) - f(x^\star)] &\leq  \frac{ 5\left\| x^{0} - x^\star \right\|^2}{2\sstep  T}\\
    &+ 7\cstepsquared nL \left(\frac{1}{M}\sum\limits_{m=1}^{M}\sigma^2_{*,m} + n \sigma_{\star}^2 \right)  + 10\sstep \frac{M-C}{C\max\{M-1,1\}} \sigma_{\star}^2.
\end{align*}

\end{proof}

\subsection{General non-convex case}
Finally, we provide  guarantees in the non-convex case.  

\begin{lemma}
	\label{lemma:\set}
	Assume that Assumption~\ref{assump: L-smooth_1} holds. For uniform sampling of cohort $\set$ we have 
	\begin{align*}
							\frac{L}{2}\sstepsquared&	\mathbb{E}_{\set}\left[\left\| \frac{1}{Cn} \sum_{m\in \set} \sum_{i=0}^{n-1}  \nabla f_{m, \pi_m^i} (x^{t}_{m,i}) \right\|^2 \right]\\
                            &\leq  L^3\sstepsquared	\mathbb{E}_{\set} \left[\frac{1}{Mn} \sum^M_{m=1} \sum_{i=0}^{n-1}\left\|   x^{t}_{m,i} - x^{t} \right\|^2\right]+L\sstepsquared \|\nabla f(x^{t})\|^2\\
&	+L\sstepsquared \frac{M-C}{C\max\left\lbrace M-1,1  \right\rbrace}\left( 2L(f(x^{t}) - f(x^\star))+2L\Delta^\star\right).
\end{align*}
\end{lemma}
\begin{proof}
	We start from Young's inequality and then we use Jensen's inequality:
\begin{align*}
	\frac{L}{2}\sstepsquared&\mathbb{E}_{\set}\left[\left\| \frac{1}{Cn} \sum_{m\in \set} \sum_{i=0}^{n-1}  \nabla f_{m, \pi_m^i} (x^{t}_{m,i}) \right\|^2\right]\\
    &\overset{\eqref{eq:sqnorm-triangle-2}}{\leq} L\sstepsquared	\mathbb{E}_{\set}\left[\left\| \frac{1}{Cn} \sum_{m\in \set} \sum_{i=0}^{n-1}  \left(\nabla f_{m, \pi_m^i} (x^{t}_{m,i}) - \nabla f_{m, \pi_m^i} (x^{t}) \right)\right\|^2\right]\\
	&+L\sstepsquared	\mathbb{E}_{\set}\left[\left\| \frac{1}{Cn} \sum_{m\in \set} \sum_{i=0}^{n-1}  \nabla f_{m, \pi_m^i} (x^{t}) \right\|^2\right]\\
	&\overset{\eqref{eq:jensen}}{\leq}L\sstepsquared	\mathbb{E}_{\set}\left[ \frac{1}{Cn} \sum_{m\in \set} \sum_{i=0}^{n-1}\left\|  \nabla f_{m, \pi_m^i} (x^{t}_{m,i}) - \nabla f_{m, \pi_m^i} (x^{t}) \right\|^2\right]\\
		&+L\sstepsquared	\mathbb{E}_{\set}\left[\left\| \frac{1}{Cn} \sum_{m\in \set} \sum_{i=0}^{n-1}  \nabla f_{m, \pi_m^i} (x^{t}) \right\|^2\right]\\
			&\overset{\eqref{eq:nabla-Lip}}{\leq}L^3\sstepsquared	\mathbb{E}_{\set}\left[ \frac{1}{Cn} \sum_{m\in \set} \sum_{i=0}^{n-1}\left\|   x^{t}_{m,i} - x^{t} \right\|^2\right]\\
		&+L\sstepsquared	\mathbb{E}_{\set}\left[\left\| \frac{1}{Cn} \sum_{m\in \set} \sum_{i=0}^{n-1}  \nabla f_{m, \pi_m^i} (x^{t}) \right\|^2\right].
\end{align*}
Taking expectations and using Lemma~\ref{lem:sampling_wo_replacement} we get
\begin{align*}
		\frac{L}{2}\sstepsquared&\mathbb{E}_{\set}\left[\left\| \frac{1}{Cn} \sum_{m\in \set} \sum_{i=0}^{n-1}  \nabla f_{m, \pi_m^i} (x^{t}_{m,i}) \right\|^2 \right]\\
        &\overset{\eqref{eq:sampling_wo_replacement}}{\leq}L^3\sstepsquared	\frac{1}{Mn} \sum^M_{m=1} \sum_{i=0}^{n-1}\left\|   x^{t}_{m,i} - x^{t} \right\|^2\\
		&+L\sstepsquared\left( \left\| \nabla f(x^{t})\right\|^2 + \frac{M-C}{C\max\left\lbrace M-1,1  \right\rbrace} \sigma_{t}^2\right)
\end{align*}
Next, we follow steps of Proposition 2 from the work~\citep{MKR2020rr}. Using the definition $\sigma_{t}^2 = \frac{1}{M} \sum_{m=1}^{M}\left\|\nabla f_{m}\left(x^{t}\right)-\nabla f\left(x^{t}\right)\right\|^{2} $ we obtain
\begin{align*}
			\frac{L}{2}\sstepsquared&\mathbb{E}_{\set}\left[\left\| \frac{1}{Cn} \sum_{m\in \set} \sum_{i=0}^{n-1}  \nabla f_{m, \pi_m^i} (x^{t}_{m,i}) \right\|^2 \right]	\leq L^3\sstepsquared	 \frac{1}{Mn} \sum^M_{m=1} \sum_{i=0}^{n-1}\left\|   x^{t}_{m,i} - x^{t} \right\|^2\\
			&+L\sstepsquared\left( \|\nabla f(x^{t})\|^2 + \frac{M-C}{C\max\left\lbrace M-1,1  \right\rbrace} \frac{1}{M} \sum_{m=1}^{M}\left\|\nabla f_{m}\left(x^{t}\right)-\nabla f\left(x^{t}\right)\right\|^{2}\right)\\
			&\overset{\eqref{eq:variance-decomp}}{=} L^3\sstepsquared\frac{1}{Mn} \sum^M_{m=1} \sum_{i=0}^{n-1}\left\|   x^{t}_{m,i} - x^{t} \right\|^2\\
            &+L\sstepsquared\| \nabla f(x^{t})\|^2\\
			&+L\sstepsquared \frac{M-C}{C\max\left\lbrace M-1,1  \right\rbrace}\left( \frac{1}{M} \sum_{m=1}^{M}\left\|\nabla f_{m}\left(x^{t}\right)\right\|^{2}-\|\nabla f\left(x^{t}\right)\|^2\right)\\
			&\leq  L^3\sstepsquared	 \frac{1}{Mn} \sum^M_{m=1} \sum_{i=0}^{n-1}\left\|   x^{t}_{m,i} - x^{t} \right\|^2\\
			&+L\sstepsquared\left( \|\nabla f(x^{t})\|^2 + \frac{M-C}{C\max\left\lbrace M-1,1  \right\rbrace} \frac{1}{M} \sum_{m=1}^{M}\left\|\nabla f_{m}\left(x^{t}\right)\right\|^{2}\right)\\
			&\leq  L^3\sstepsquared	 \frac{1}{Mn} \sum^M_{m=1} \sum_{i=0}^{n-1}\left\|   x^{t}_{m,i} - x^{t} \right\|^2+L\sstepsquared \|\nabla f(x^{t})\|^2\\
			&+L\sstepsquared \frac{M-C}{C\max\left\lbrace M-1,1  \right\rbrace}\left( 2L(f(x^{t}) - f^\star)+2L\left(f_{*}-\frac{1}{M} \sum_{m=1}^{M} f_{*,m}\right)\right).
			\end{align*}
Finally, we get 
\begin{align*}
								\frac{L}{2}\sstepsquared&\mathbb{E}_{\set}\left[\left\| \frac{1}{Cn} \sum_{m\in \set} \sum_{i=0}^{n-1}  \nabla f_{m, \pi_m^i} (x^{t}_{m,i}) \right\|^2 \right]\\	&\leq  L^3\sstepsquared	\mathbb{E}_{\set}\left[ \frac{1}{Mn} \sum^M_{m=1} \sum_{i=0}^{n-1}\left\|   x^{t}_{m,i} - x^{t} \right\|^2\right]+L\sstepsquared \|\nabla f(x^{t})\|^2\\
								&	+L\sstepsquared \frac{M-C}{C\max\left\lbrace M-1,1  \right\rbrace}\left( 2L(f(x^{t}) - f(x^\star))+2L\Delta^\star\right).
\end{align*}
\end{proof}

\begin{lemma}
	\label{lemma:V_t_non}
	Suppose that Algorithm~\ref{alg:pp-jumping} is used and Assumption~\ref{assump: L-smooth_1} holds. If $\cstep \leq \frac{1}{2Ln}$, then
	\begin{align*}
		\frac{1}{Mn}\sum\limits_{m=1}^{M}\sum\limits_{i=0}^{n-1} \mathbb{E}\left[\left\|x^{t} - x_{m,i}^t\right\|^2|x^{t}\right]\ &\leq 4\cstepsquared n^2L \left(f(x^{t}) - f^\star\right) \\
        &+2\cstepsquared n^2 L\Delta^{\star} + 2 \cstepsquared  n L \frac{1}{M}\sum\limits_{m=1}^{M}\Delta^{\star}_{m}.
	\end{align*}
\end{lemma}
\begin{proof}
	We start from equation~\eqref{eq:32}. It is proved in section~\ref{section:C.1} but it does not require convexity:
	\begin{align*}
		\frac{1}{Mn}\sum\limits_{m=1}^{M}\sum\limits_{i=0}^{n-1}\mathbb{E}\left[\left\|x_{m,i}^t - x^{t}\right\|^2|x^{t}\right] 	&\leq \cstepsquared n^2\frac{1}{M}\sum\limits_{m=1}^{M}\left\|\nabla f_m(x^{t})\right\|^2 + \cstepsquared  n \frac{1}{M}\sum\limits_{m=1}^{M}\sigma_{t,m}^2.
	\end{align*}
	Using $L$-smoothness, we get
	\begin{align*}
		\frac{1}{Mn}\sum\limits_{m=1}^{M}\sum\limits_{i=0}^{n-1}\mathbb{E}\left[\left\|x_{m,i}^t - x^{t}\right\|^2|x^{t}\right] &\leq 2\cstepsquared n^2 L \frac{1}{M}\sum_{m=1}^{M} (f_m(x^{t}) - f_{*,m} )\\
        &+ 2\cstepsquared n L \frac{1}{M}\sum_{m=1}^{M} \frac{1}{n}\sum_{i=0}^{n-1} (f^{i}_m(x^{t}) - f^\star_{m,i} )\\
		&\leq  2\cstepsquared n^2 L \frac{1}{M}\sum_{m=1}^{M} (f_m(x^{t}) - f_{*} )\\
        &+  2\cstepsquared n^2 L \frac{1}{M}\sum_{m=1}^{M} (f^\star - f_{*,m} )\\
		&+ 2\cstepsquared n L \frac{1}{M}\sum_{m=1}^{M} \frac{1}{n}\sum_{i=0}^{n-1} (f^{i}_m(x^{t}) - f^\star )\\
        &+ 2\cstepsquared n L \frac{1}{M}\sum_{m=1}^{M} \frac{1}{n}\sum_{i=0}^{n-1} (f^\star - f^\star_{m,i} )\\
		&\leq  4 L\cstepsquared n^2 (f(x^{t}) - f^\star)\\
        &+ 2\cstepsquared n^2 L \Delta^{\star} + 2\cstepsquared nL \frac{1}{M}\sum_{m=1}^{M}\Delta^{\star}_{m}.
	\end{align*}

\end{proof}
\begin{lemma}
	\label{lemma:noncvx-recursion-solution}
	Suppose that there exists constants $a, b, c \geq 0$ and nonnegative sequences $(s^{t})_{t=0}^{T}, (q^{t})_{t=0}^{T}$ such that for any $t \in \{ 0, 1, \ldots, T \}$
	\begin{equation}
		\label{eq:non-convex-recursion-init}
		s^{t+1} \leq \br{1 + a} s^{t} - b q^{t} + c.
	\end{equation}
	Then if $a > 0$ we have,
	\begin{equation}
		\label{eq:non-convex-recursion-soln}
		\min_{t=0, \ldots, T-1} q^{t} \leq \frac{\br{1+a}^T}{b T} s^{0} + \frac{c}{b}.
	\end{equation}
	And if $a = 0$ we have,
	\begin{equation}
		\label{eq:weakly-convex-recursion-soln}
		\frac{1}{T} \sum\limits_{t=0}^{T-1} q^{t} \leq \frac{s^0}{b T} + \frac{c}{b}.
	\end{equation}
\end{lemma}
\begin{proof}
	The first part of the proof (for $a > 0$) is a distillation of the recursion solution in Lemma~2 \citep{khaled2023better} and we closely follow their proof.  Let $w_{-1} = w^{0} > 0$ be arbitrary. Define 
	\[ w^{t} \eqdef \frac{w^0}{\br{1+a}^{t}}. \]
	Note that $w^{t} \br{1+a} = w^{t-1}$. Multiplying both sides of \eqref{eq:non-convex-recursion-init} by $w^{t}$,
	\begin{align*}
		w^{t} s^{t+1} &\leq \br{1 + a} w^t \set - b w^t q^t + c w^t \\
		&= w^{t-1} s^{t} - b w^{t} q^t + c w^t.
	\end{align*}
	Rearranging,
	\begin{align*}
		b w^{t} q^{t} &\leq w^{t-1} s^{t} - w^{t} s^{t+1} + c w^{t}.
	\end{align*}
	Summing up as $t$ varies from $0$ to $T-1$ and noting that the sum telescopes,
	\begin{align*}
		\sum\limits_{t=0}^{T-1} b w^{t} q^{t} &\leq \sum\limits_{t=0}^{T-1} \br{w^{t-1} s^{t} - w^{t} s^{t+1}} + c \sum\limits_{t=0}^{T-1} w^{t}\\
        &= w^{0} s^{0} - w^{T-1} s_{T} + c \sum\limits_{t=0}^{T-1} w^{t}\\
        &\leq w^{0} s^{0} + c \sum\limits_{t=0}^{T-1} w^{t}.
	\end{align*}
	Let $W^{T} = \sum\limits_{t=0}^{T-1} w^{t}$. Dividing both sides by $W^{T}$ we have,
	\begin{align}
		\frac{1}{W^{T}} \sum\limits_{t=0}^{T-1} b w^{t} q^{t} \leq \frac{w^0 s^0}{W^{T}} + c.
		\label{eq:nc-rec-1}
	\end{align}
	We now separate the proof into two cases:
	\begin{itemize}[leftmargin=0.2in,itemsep=0.01in]
		\item \textbf{If $a > 0$}: Note that the left-hand side of \eqref{eq:nc-rec-1} satisfies
		\begin{equation}
			b \min_{t=0, \ldots, T-1} q^t \leq \frac{1}{W^{T}} \sum\limits_{t=0}^{T-1} b w^t q^t.
			\label{eq:nc-rec-2}
		\end{equation}
		And for the right hand-side of \eqref{eq:nc-rec-1} we have,
		\begin{equation}
			W^{T} = \sum\limits_{t=0}^{T-1} w^{t} \geq T \min_{t=0, \ldots, T-1} w^{t} = T w^{T-1} \geq T w^{T} = \frac{T w^0}{\br{1+a}^{T}}.
			\label{eq:nc-rec-3}
		\end{equation}
		Substituting with \eqref{eq:nc-rec-3} in \eqref{eq:nc-rec-2} and dividing both sides by $b$ we get,
		\begin{align*}
			\min_{t=0, \ldots, T-1} q^{t} \leq \frac{\br{1 + a}^T}{bT} s^0 + \frac{c}{b}.
		\end{align*}
		\item \textbf{If $a = 0$}: then $w^{t} = w^0$ for all $t$ and hence $w^{T} = T$, then \eqref{eq:nc-rec-2} is equivalent to
		\[ \frac{1}{T} \sum\limits_{t=0}^{T-1} b q^{t} \leq \frac{s^0}{T} + c.  \]
		Dividing both sides by $b$ yields the lemma's claim.
		\qedhere
	\end{itemize}
\end{proof}

\subsubsection{Proof of Theorem~\ref{thm:PP-NC}}

\begin{theorem_empt}
	Let the Assumption of smoothness hold. Let $\delta^0 = f(x^{0}) - f^\star$ and $\Delta^{\star}_{m} = \frac{1}{n}\sum\limits_{i=1}^{n}(f^\star - f^\star_{m,i})$. Let $\cstep \leq \frac{1}{2nL}$ and $\sstep \leq \frac{1}{4L}$. Then for iterates $x^{t}$ of Algorithm~\ref{alg:pp-jumping}, we have
	\begin{align*}
		\min _{t = 0, \ldots, T-1}  \mathbb{E}\left[\left\|\nabla f\left(x^{t}\right)\right\|^{2} \right]&\leq  \squeeze 8L^2\sstep  \frac{M-C}{C\max\{M-1,1\}} \Delta^\star\\
		&  +6\cstepsquared  n L^3 \left(\frac{1}{M}\sum\limits_{m=1}^{M}\Delta^{\star}_{m}+n\Delta^{\star}\right)\\
        &+\frac{4\left(1+\frac{2L^2\sstepsquared(M-C)}{C\max\left\lbrace M-1,1 \right\rbrace }+\frac{3}{2}\sstep\cstepsquared n^2L^3\right)^T}{\sstep  T} \delta^0.
	\end{align*}

\end{theorem_empt}
\begin{proof}
	We start from $L$-smoothness~\eqref{eq:L-smoothness}:
		\begin{align*}
		f(x^{t+1}) &\overset{\eqref{eq:L-smoothness}}{\leq} f(x^{t})+\left\langle \nabla f(x^{t}), x^{t+1} - x^{t} \right\rangle + \frac{L}{2}\| x^{t+1} - x^{t} \|^2\\
		&= f(x^{t}) - \left\langle \nabla f(x^{t}), \sstep \frac{1}{Cn} \sum_{m\in \set} \sum\limits_{i=0}^{n-1} \nabla f_{m, \pi^i_m}\left(x_{m,i}^t\right) \right\rangle\\
        &+ \frac{L}{2}\left\| \sstep \frac{1}{Cn}\sum_{m\in \set}\sum\limits_{n=0}^{n-1}\nabla f_{m, \pi^i_m}\left(x_{m,i}^t\right) \right\|^2\\
		&= f(x^{t}) -  \sstep \left\langle \nabla f(x^{t}), \frac{1}{Cn} \sum_{m\in \set} \sum\limits_{i=0}^{n-1} \nabla f_{m, \pi^i_m}\left(x_{m,i}^t\right) \right\rangle\\
        &+ \frac{L}{2} \sstepsquared\left\| \frac{1}{Cn}\sum_{m\in \set}\sum\limits_{n=0}^{n-1}\nabla f_{m, \pi^i_m}\left(x_{m,i}^t\right) \right\|^2.
	\end{align*}
Taking conditional expectation over cohort $\set$, we get 
\begin{align*}
			\mathbb{E}_{\set}\left[f(x^{t+1})\right] &\leq f(x^{t}) -  \sstep\mathbb{E}_{\set}\left[ \left\langle \nabla f(x^{t}), \frac{1}{Cn} \sum_{m\in \set} \sum\limits_{i=0}^{n-1} \nabla f_{m, \pi^i_m}\left(x_{m,i}^t\right) \right\rangle \right]\\
            &+ \frac{L}{2} \sstepsquared\mathbb{E}_{\set}\left[\left\| \frac{1}{Cn}\sum_{m\in \set}\sum\limits_{n=0}^{n-1}\nabla f_{m, \pi^i_m}\left(x_{m,i}^t\right) \right\|^2\right]\\
			&= f(x^{t}) -  \sstep \left\langle \nabla f(x^{t}), \frac{1}{Mn} \sum^{M}_{m=1} \sum\limits_{i=0}^{n-1} \nabla f_{m, \pi^i_m}\left(x_{m,i}^t\right) \right\rangle\\
            &+ \frac{L}{2} \sstepsquared\mathbb{E}_{\set}\left[\left\| \frac{1}{Cn}\sum_{m\in \set}\sum\limits_{n=0}^{n-1}\nabla f_{m, \pi^i_m}\left(x_{m,i}^t\right) \right\|^2\right].
\end{align*}
	Using $2\left\langle a,b \right\rangle =  \|a+b\|^2 - \|a\|^2 - \|b\|^2$, we have 
\begin{align*}
\mathbb{E}_{\set}\left[	f(x^{t+1})\right]	& = f(x^{t}) +  \frac{L}{2} \sstepsquared\mathbb{E}_{\set}\left[\left\| \frac{1}{Cn}\sum_{m\in \set}\sum\limits_{n=0}^{n-1}\nabla f_{m, \pi^i_m}\left(x_{m,i}^t\right) \right\|^2 \right]\\
	& -\left( \frac{\sstep}{2}\|\nabla f(x^{t})\|^2 + \frac{\sstep}{2}\left\| \frac{1}{Mn} \sum\limits_{m=1}^{M}\sum\limits_{i=0}^{n-1} \nabla f_{m, \pi^i_m}\left( x_{m,i}^t \right)  \right\|^2\right)\\
    &+ \frac{\sstep}{2} \left\|\nabla f(x^{t}) -  \frac{1}{Mn} \sum\limits_{m=1}^{M}\sum\limits_{i=0}^{n-1} \nabla f_{m, \pi^i_m}\left( x_{m,i}^t \right) \right\|^2 \\
	&\leq f(x^{t}) +  \frac{L}{2} \sstepsquared \mathbb{E}_{\set}\left[\left\| \frac{1}{Cn}\sum_{m\in \set}\sum\limits_{n=0}^{n-1}\nabla f_{m, \pi^i_m}\left(x_{m,i}^t\right) \right\|^2\right] \\
	& -\left( \frac{\sstep}{2}\|\nabla f(x^{t})\|^2 + \frac{\sstep}{2}\left\| \frac{1}{Mn} \sum\limits_{m=1}^{M}\sum\limits_{i=0}^{n-1} \nabla f_{m, \pi^i_m}\left( x_{m,i}^t \right)  \right\|^2\right)\\
    &+ \frac{\sstep}{2} \left\|\frac{1}{Mn} \sum\limits_{m=1}^{M}\sum\limits_{i=0}^{n-1} \left( \nabla f_{m, \pi^i_m}\left( x_{m,i}^t \right) - \nabla f_{m, \pi^i_m}\left( x^{t} \right) \right)   \right\|^2.
\end{align*}
Using $L$-smoothness, we get 
\begin{align*}
	\mathbb{E}_{\set}\left[ f(x^{t+1})\right] &\leq f(x^{t}) + \frac{L}{2} \sstepsquared\mathbb{E}_{\set}\left[\left\| \frac{1}{Cn}\sum_{m\in \set}\sum\limits_{n=0}^{n-1}\nabla f_{m, \pi^i_m}\left(x_{m,i}^t\right) \right\|^2\right]\\
	&- \frac{\sstep}{2}\|\nabla f(x^{t})\|^2 +\frac{\sstep}{2} L^2 \frac{1}{Mn}\sum\limits_{m=1}^{M}\sum\limits_{i=0}^{n-1} \left\| x_{m,i}^t - x^{t} \right\|^2.
\end{align*}
Utilizing Lemma~\ref{lemma:\set} and taking conditional expectation, we get
\begin{align*}
		\mathbb{E}\left[ f(x^{t+1}) | x^{t}\right]  &\leq f(x^{t}) +  L^3\sstepsquared	\frac{1}{Mn}\sum\limits_{m=1}^{M}\sum\limits_{i=0}^{n-1} 		\mathbb{E}\left[ \left\| x_{m,i}^t - x^{t} \right\|^2 |x^{t}\right]+L\sstepsquared \|\nabla f(x^{t})\|^2\\
		&	+L\sstepsquared \frac{M-C}{C\max\left\lbrace M-1,1  \right\rbrace}\left( 2L(f(x^{t}) - f(x^\star))+2L\Delta^\star\right)\\
		& - \frac{\sstep}{2}\|\nabla f(x^{t})\|^2 +\frac{\sstep}{2} L^2 \frac{1}{Mn}\sum\limits_{m=1}^{M}\sum\limits_{i=0}^{n-1} 		\mathbb{E}\left[ \left\| x_{m,i}^t - x^{t} \right\|^2 |x^{t}\right]\\
		&\leq f(x^{t}) + \frac{3}{4}\sstep L^2 \frac{1}{Mn}\sum\limits_{m=1}^{M}\sum\limits_{i=0}^{n-1} 		\mathbb{E}\left[ \left\| x_{m,i}^t - x^{t} \right\|^2 |x^{t}\right] - \frac{\sstep}{4}\|\nabla f(x^{t})\|^2\\
				&	+L\sstepsquared \frac{M-C}{C\max\left\lbrace M-1,1  \right\rbrace}\left( 2L(f(x^{t}) - f(x^\star))+2L\Delta^\star\right).\\
\end{align*}
Applying Lemma~\ref{lemma:V_t_non} and using $\sstep\leq \frac{1}{4L}$ we get 
\begin{align*}
	\mathbb{E}\left[ f(x^{t+1}) | x^{t}\right]  &\leq f(x^{t}) +L\sstepsquared \frac{M-C}{C\max\left\lbrace M-1,1  \right\rbrace}\left( 2L(f(x^{t}) - f(x^\star))+2L\Delta^\star\right)\\
	& - \frac{\sstep}{4}\|\nabla f(x^{t})\|^2\\
    &+\frac{3\sstep}{4} L^2 \left(4 L\cstepsquared n^2 (f(x^{t}) - f^\star) + 2\cstepsquared n^2 L \Delta^{\star} + 2\cstepsquared nL \frac{1}{M}\sum_{m=1}^{M}\Delta^{\star}_{m}\right).
\end{align*}
Subtracting $f^\star$ from both sides leads to
\begin{align*}
	\mathbb{E}\left[ f(x^{t+1}) | x^{t}\right]& - f^\star \leq f(x^{t}) - f^\star\\
    &+L\sstepsquared \frac{M-C}{C\max\left\lbrace M-1,1  \right\rbrace}\left( 2L(f(x^{t}) - f(x^\star))+2L\Delta^\star\right)\\
	& - \frac{\sstep}{4}\|\nabla f(x^{t})\|^2\\
    &+\frac{3\sstep}{4} L^2 \left(4 L\cstepsquared n^2 (f(x^{t}) - f^\star) + 2\cstepsquared n^2 L \Delta^{\star} + 2\cstepsquared nL \frac{1}{M}\sum_{m=1}^{M}\Delta^{\star}_{m}\right).
\end{align*}

 Taking full expectation, we have 
\begin{align*}
	\mathbb{E}\left[ \delta_{t+1}\right]  &\leq \left(1+\frac{2L^2\sstepsquared}{C}+\frac{3}{2}\sstep\cstepsquared n^2L^3\right)\mathbb{E}\left[\delta^{t}\right]- \frac{\sstep}{4}\mathbb{E}\left[\|\nabla f(x^{t})\|^2\right]\\
	 &+ 2L^2\sstepsquared  \frac{M-C}{C\max\left\lbrace M-1,1  \right\rbrace} \Delta^\star  + \frac{3}{2}\sstep\cstepsquared n^2 L^3 \Delta^{\star} + \frac{3}{2}\sstep\cstepsquared nL^3 \frac{1}{M}\sum_{m=1}^{M}\Delta^{\star}_{m} .
\end{align*}
Applying Lemma~\ref{lemma:noncvx-recursion-solution} from the paper \citep{MKR2020rr}, we get 
	\begin{align*}
	\min _{t = 0, \ldots, T-1} &\mathbb{E}\left[\left\|\nabla f\left(x^{t}\right)\right\|^{2} \right]\leq \frac{4\left(1+\frac{2L^2\sstepsquared}{C}+\frac{3}{2}\sstep\cstepsquared n^2L^3\right)^T}{\sstep  T} \delta^0\\
    &+ 6\cstepsquared  n L^3 \left(\frac{1}{M}\sum\limits_{m=1}^{M}\Delta^{\star}_{m}+n\Delta^{\star}\right)\\
	& + 8L^2\sstep  \frac{M-C}{C\max\{M-1,1\}} \Delta^\star.
\end{align*}

\end{proof}

\section{Small Server Stepsize}
In this section, we present a result when it is useful to pull back the last iterates of local passes. In particular, we show that one can reduce the variance of FedAvg with uniform partial participation.

\begin{theorem}
	Assume that all losses $f_{m,i}$ are $L$-smooth and $\mu$-strongly convex. Define $\alpha  = \frac{\sstep}{\cstep n}$. Let $\cstep\leq \frac{1}{L}$ and $0\leq \alpha <1$. Then, for iterates $x^{t}$ generated by Algorithm~\ref{alg:pp-jumping}, we have 
	\begin{align*}
		\mathbb{E}\left[\left\|x^{T}-x^{\star}\right\|^{2}\right]& \leq\left(1-\alpha+\alpha(1-\cstep \mu)^{n}\right)^{T}\left\|x_{0}-x^{\star}\right\|^{2}\\
		& +\frac{\alpha}{(1-\alpha)\left(1-(1-\cstep \mu)^{n}\right)} \cstepsquared \frac{M-C}{C\max\left\lbrace M-1,1 \right\rbrace} \sigma_{*}^{2}\\
        &+2 \cstepcubed \sigma_{\operatorname{rad}}^{2} \frac{1}{1-(1-\cstep \mu)^{n}} \sum\limits_{i=0}^{n-1}(1-\cstep \mu)^{i}.
	\end{align*}
\end{theorem}
\begin{proof}
	Let us denote $f_{\set} = \frac{1}{C}\sum\limits_{m\in \set} f_m$. We start by rewriting the distance to the optimum in the following way:
	\begin{align*}
		x^{t+1} - x^\star
		&= (1-\alpha) x^{t} + \alpha x^{t}_n - x^\star \\
		&= (1-\alpha) x^{t} + \alpha x^{t}_n - (1-\alpha) \left(x^\star + \frac{\alpha}{1-\alpha}\cstep n\nabla f_{\set}(x^\star)\right)\\
        &- \alpha (x^\star - \cstep n \nabla f_{\set}(x^\star)).
	\end{align*}
	Therefore, by convexity of the squared norm,
	\begin{align*}
		\|x^{t+1} - x^\star\|^2
		&\le (1-\alpha) \left \|x^{t} - \left(x^\star + \frac{\alpha}{1-\alpha}\cstep n\nabla f_{\set}(x^\star)\right) \right \|^2\\
        &+ \alpha \|x^{t}_n - (x^\star - \cstep n\nabla f_{\set}(x^\star))\|^2.
	\end{align*}
	We bound the two terms in the right-hand side separately. For the first term, it suffices to take expectation over the sampling of client cohort $\set$,
	\begin{align*}
		\mathbb{E}_{\set}\left\|x^{t} - \left(x^\star + \frac{\alpha}{1-\alpha}\cstep n\nabla f_{\set}(x^\star)\right) \right\|^2
		&\overset{\eqref{eq:rv_moments}}{=} \|x^{t} - x^\star\|^2\\
        &+ \frac{\alpha^2}{(1-\alpha)^2}\cstepsquared  n^2\mathbb{E}_{\set}\| \nabla f_{\set}(x^\star)	\|^2  
        \\
        		&= \|x^{t} - x^\star\|^2\\
                &+ \frac{\alpha^2}{(1-\alpha)^2}\cstepsquared  n^2\frac{M-C}{C\max\left\lbrace M-1,1\right\rbrace}\sigma_{\star}^2.
	\end{align*}
	For the second term, we use the results of prior work on convergence of \gls{RR} that gives
	\begin{align*}
		\|x^{t}_n - (x^\star - \cstep \nabla f_{\set}(x^\star))\|^2
		\le (1-\cstep \mu)^n \|x^{t}-x^\star\|^2 + 2\cstepcubed  \sigma_{\mathrm{rad}}^2\sum\limits_{i=0}^{n-1}(1-\cstep\mu)^i,
	\end{align*}
	where, as shown by \citep{mishchenko2022proximal}, $\sigma_{\mathrm{rad}}\ge 0$ is some constant satisfying
	\begin{align*}
		\sigma_{\mathrm{rad}}^2
		\le L\sum\limits_{m=1}^M(n^2\|\nabla f_m(x^\star)\|^2 + \frac{n}{4}\sigma_{*,m}^2).
	\end{align*}
Notice that the upper bound depends on $\alpha$ in a nonlinear way, so the optimal value of $\alpha$ would often lie somewhere in the interval $(0, 1)$. Recurrence $a_{t+1}\le (1-\rho)a_t + c$ implies by induction $a_t\le (1-\rho)^ta_0 + \frac{c}{\rho}$, so by propagating the bound above to $x^{0}$, we obtain
\begin{align*}
	\E\|x^{t}-x^\star\|^2
	&\le (1-\alpha +\alpha(1-\cstep \mu)^n)^t\|x^{0}-x^\star\|^2\\
    &+ \frac{\alpha}{(1-\alpha)(1-(1-\cstep\mu)^n)}\cstepsquared \frac{M-C}{C\max\left\lbrace M-1,1 \right\rbrace}\sigma_{\star}^2\\
	& + 2\cstepcubed  \sigma_{\mathrm{rad}}^2\frac{1}{1-(1-\cstep\mu)^n}\sum\limits_{i=0}^{n-1}(1-\cstep\mu)^i.
\end{align*}
Notice that the last term does not change with $\alpha$, so its optimal value is completely determined by the first two terms.
\end{proof}

\newpage
            \refstepcounter{chapter}%
\chapter*{\thechapter \quad Appendix E Title}
\label{appendixF}

\section{Extra Related Works}

Federated optimization has been the subject of intense study, with many open questions even in the setting when all clients have identical data~\citep{woodworth2020local, woodworth2020minibatch, woodworth21_min_max_compl_distr_stoch}. The \algname{FedAvg} algorithm (also known as \algname{Local SGD}) has also been a subject of intense study, with tight bounds obtained only very recently \citep{glasgow2022sharp}. It is now understood that using many local steps adds bias to distributed \algname{SGD}, and hence several methods have been developed to mitigate it, e.g.~\citep{karimireddy2020scaffold, murata21_bias_varian_reduc_local_sgd}, see the work \citep{gorbunov2021local} for a unifying lens on many variants of \algname{Local SGD}. Note that despite the bias, even vanilla \algname{FedAvg}/\algname{Local SGD} still reduces the overall communication overhead in practice~\citep{ortiz21_trade_offs_local_sgd_at_scale}.

The success of \algname{\gls{RR}} in the single-machine setting has inspired several recent methods that use it as a local update method as part of distributed training: in \citep{mishchenko2022proximal} authors developed a distributed variant of random reshuffling, \algname{FedRR}. \algname{FedRR} uses \gls{RR} as a local client update method in lieu of \algname{SGD}. They show that \algname{FedRR} can improve upon the convergence of \algname{Local SGD} when the number of local steps is fixed as the local dataset size, i.e.\ when $H = n$. In work \citep{yun2021minibatch} authors study the same method under the name \algname{Local RR} under a more restrictive assumption of bounded inter-machine gradient deviation and show that by varying $H$ to be smaller than $n$ better rates can be obtained in this setting than the rates~\citep{mishchenko2022proximal}. Other work has explored more such combinations between \algname{\gls{RR}} and distributed training algorithms~\citep{huang21_distr_random_reshuf_over_networ,malinovsky2023server, horvath2022fedshuffle}.

There are several methods that combine compression or quantization and local steps: both works \citep{basu2019qsparse} and \citep{reisizadeh19_fedpaq} combined \algname{Local SGD} with quantization and sparsification, and work \citep{haddadpour2021federated} later improved their results using a gradient tracking method, achieving linear convergence under strong convexity. In parallel, in work \citep{mitra2021linear} authors also developed a variance-reduced method, \algname{FedLin}, that achieves linear convergence under strong convexity despite using local steps and compression. The paper most related to our work is \citep{malinovsky2022federated} in which the authors combine \emph{iterate} compression, random reshuffling, and local steps. We study \emph{gradient} compression instead, which is a more common form of compression in both theory and practice~\citep{kairouz2019advances}. We compare our results against \citep{malinovsky2022federated} and show we obtain better rates compared to their work.

\section{Experimental Details}\label{subsec:extra_exp}
In this section, we provide missing details on the experimental setting from Section \ref{sec:experiments}.  The codes are provided in the following anonymous repository: \url{https://anonymous.4open.science/r/diana_rr-B0A5}.

\subsection{Logistic regression}

To confirm our theoretical results we conducted several numerical experiments on binary classification problem with L2 regularized logistic regression of the form
\begin{eqnarray}\label{eq:log-reg}
\min _{x \in \mathbb{R}^{d}}\left[f(x) \eqdef \frac{1}{M} \sum_{m=1}^{M} \frac{1}{n_m} \sum_{i=1}^{n_m} f_{m,i} \right] \text {, }
\end{eqnarray}
where $f_{m,i}\eqdef \log \left(1+\exp({-y_{mi} a_{mi}^{\top} x})\right)+\lambda \|x\|^2_2$ $(a_{mi},  y_{mi}) \in \mathbb{R}^{d} \times \{-1,1\}, i =1,\dots,n_m$ are the training data samples stored on machines $m =1,\dots,M$, and $\lambda>0$ is a regularization parameter. In all experiments, for each method, we used the largest stepsize allowed by its theory multiplied by some individually tuned constant multiplier. For better parallelism, each worker $m$ uses mini-batches of size $\approx 0.1 n_m$. In all algorithms, as a compression operator $\cQ$, we use Rand-$k$ \citep{beznosikov20_biased_compr_distr_learn}  with fixed compression ratio $\nicefrac{k}{d} \approx 0.02$, where $d$ is the number of features in the dataset. 

\paragraph{Hardware and Software.}
All algorithms were written in Python 3.8. We used three different CPU cluster node types:
\begin{enumerate}
	\item AMD EPYC 7702 64-Core; 
	\item Intel(R) Xeon(R) Gold 6148 CPU @ 2.40GHz;
	\item Intel(R) Xeon(R) Gold 6248 CPU @ 2.50GHz.
\end{enumerate}
\paragraph{Datasets.}
The datasets were taken from open LibSVM library \citep{chang2011libsvm}, sorted in ascending order of labels, and equally split among 20 machines $\backslash$clients$\backslash$workers. The remaining part of size $N - 20\cdot \left \lfloor{\nicefrac{N}{20}}\right \rfloor $ was assigned to the last worker,  where $N = \sum_{m=1}^{M}n_m$ is the total size of the dataset. 
A summary of the splitting and the data samples distribution between clients can be found in Tables \ref{tbl:datasets_summary}, \ref{tbl:mushrooms_partition}, \ref{tbl:w8a_partition}, \ref{tbl:a9a_partition}. 
\begin{table}[t]
	\caption{Summary of the datasets and splitting of the data samples among clients.}
	\label{tbl:datasets_summary}
	\centering
	\begin{tabular}{l l l l l }
		\toprule
		Dataset  & $M$ & $N$&  $d$   & $n_m$  \\
		\midrule			
		\texttt{mushrooms} & $20$ & $8120$ & $112$  & $406$\\ 
		\texttt{w8a} & $20$  &$49749$ & $300$ &  $2487$\\
		\texttt{a9a} & $20$ &$32560$ & $123$  &$1628$ \\ 
		\bottomrule
	\end{tabular}
\end{table}
\begin{table}[h!]
	\caption{Partition of the \texttt{mushrooms} dataset among clients.}
	\label{tbl:mushrooms_partition}
	\centering
	\begin{tabular}{l l l }
		\toprule
		Client's \textnumero &\# of datasamples of class "-1" & \# of datasamples of class "+1"  \\
		\midrule			
		$1$ -- $9$& $406$& $0$\\
		$10$& $262$& $144$\\
		$11$ -- $19$& $0$& $406$\\
		$20$& $0$& $410$\\
		\bottomrule
	\end{tabular}
\end{table}
\begin{table}[h!]
	\caption{Partition of the \texttt{w8a} dataset among clients.}
	\label{tbl:w8a_partition}
	\centering
	\begin{tabular}{l l l }
		\toprule
		Client's \textnumero &\# of datasamples of class "-1" & \# of datasamples of class "+1"  \\
		\midrule			
		$1$ -- $19$& $2487$& $0$\\
		$20$& $1017$& $1479$\\
		\bottomrule
	\end{tabular}
\end{table}
\begin{table}[h!]
	\caption{Partition of the \texttt{a9a} dataset among clients.}
	\label{tbl:a9a_partition}
	\centering
	\begin{tabular}{l l l }
		\toprule
		Client's \textnumero &\# of datasamples of class "-1" & \# of datasamples of class "+1"  \\
		\midrule			
		$1$ -- $14$& $1628$& $0$\\
		$15$& $1328$& $300$\\
		$16$ -- $19$& $0$& $1628$\\
		$20$& $0$& $1629$\\
		\bottomrule
	\end{tabular}
\end{table}
\paragraph{Hyperparameters.}
Regularization parameter $\lambda$ was chosen individually for each dataset to guarantee the condition number $\nicefrac{L}{\mu}$ to be approximately $10^4$,  where $L$ and $\mu$ are the smoothness and strong-convexity constants of function $f$. For the chosen logistic regression problem of the form \eqref{eq:log-reg}, smoothness and strong convexity constants $L$, $L_m$, $L_{i,m}$, $\mu$, $ \widetilde{\mu}$ of functions  $f$, $f_m$ and $f_m^i$ were computed explicitly as 

\begin{eqnarray*}
L &=& \lambda_{\max}\rb{\frac{1}{M}\sum_{m=1}^{M}\frac{1}{4n_m}\mA_m^{\top}\mA_m + 2\lambda \mI} \\
L_m &=& \lambda_{\max}\rb{\frac{1}{4n_m}\mA_m^{\top}\mA_m + 2\lambda \mI}  \\L_{i,m} &=& \lambda_{\max}\rb{ \frac{1}{4} a_{mi}a_{mi}^{\top} + 2\lambda \mI} \\
 \mu &=& 2\lambda \\
 \widetilde{\mu}  &=& 2\lambda, 
\end{eqnarray*}  
where $\mA_m $ is the dataset associated with client $m$, and $a_{mi}$ is the $i$-th row of data matrix $\mA_m$. In general, the fact that $f$ is $L$-smooth with $$L \le \frac{1}{M}\sum_{m=1}^{M}\frac{1}{n_m}\sum_{i=1}^{n_m} L_{i,m}$$ follows from the $L_{i,m}$-smoothness of $f_m^i$ (see Assumption \ref{asm:lip_max_f_m}). 

In all algorithms, as a compression operator $\cQ$, we use Rand-$k$  as a canonical example of unbiased compressor with relatively bounded variance, and fix the compression parameter $k = \lfloor 0.02 d\rfloor$, where $d$ is the number of features in the dataset. 

In addition, in all algorithms, for all clients $m=1,\dots,M$, we set the batch size for the \algname{SGD} estimator to be $b_m = \lfloor 0.1 n_m\rfloor$, where $n_m$ is the size of the local dataset. 

The summary of the values $L$, $L_m$, $L_{i,m}$ $L_{\max}$, $\mu$, $b_m$ and $k$ for each dataset can be found in Table \ref{tbl:hyperparameters_per_dataset_summary}.

\begin{table}[h!]
	\caption{Summary of the hyperparameters.}
	\label{tbl:hyperparameters_per_dataset_summary}
	\centering
	\begin{tabular}{l l l l l l l}
		\toprule
		Dataset  & $L$ & $L_{\max}$& $\mu$ & $\lambda$ & $k$&$b_m$ (batchsize)\\
		\midrule			
		\texttt{mushrooms} & $2.59$ & $5.25$ & $2.58\cdot10^{-4}$ & $1.29\cdot10^{-4}$ & $2$&$40$\\ 
		\texttt{w8a} & $0.66$ &$28.5$  & $6.6\cdot10^{-5}$ & $3.3\cdot10^{-5}$ &$6$ & $248$\\
		\texttt{a9a} & $1.57$  & $3.5$ &  $1.57\cdot10^{-4}$ & $7.85\cdot10^{-5}$& $2$& $162$\\ 
		\bottomrule
	\end{tabular}
\end{table}

In all experiments, we follow a constant stepsize strategy within the whole iteration procedure. For each method, we set the largest possible stepsize predicted by its theory multiplied by some individually tuned constant multiplier. 
For a more detailed explanation of the tuning routine, see Sections \ref{subsec:non_local_methods} and \ref{subsec:local_methods}.

\paragraph{SGD implementation.} 
We considered two approaches to minibatching: random reshuffling and with-replacement sampling. In the first, all clients $m=1,\dots, M$ independently permute their local datasets and pass through them within the next subsequent  $\lfloor \frac{n_m}{b_m} \rfloor$ steps. In our implementations of \gls{Q-RR}, \gls{Q-NASTYA} and \gls{DIANA-NASTYA}, all clients permuted their datasets in the beginning of every new epoch, whereas for the \gls{DIANA-RR} method they do so only once in the beginning of the iteration procedure. The second approach to minibatching is called with-replacement sampling, and it requires every client to draw $b_m$ 
data samples from the local dataset uniformly at random. We used this strategy in the baseline algorithms (\algname{QSGD}, \gls{DIANA}, \algname{FedCOM} and \algname{FedPAQ}) we compared our proposed methods to.

\paragraph{Experimental setup.}
To compare the performance of methods within the whole optimization process, we track the functional suboptimality metric $f(x^t) - f(x^{\star})$ that was recomputed after each epoch. For each dataset, the value $f(x^{\star})$ was computed once at the preprocessing stage with $10^{-16}$ tolerance via conjugate gradient method.
We terminate our algorithms after performing $5000$ epochs.

\subsubsection{Experiment 1: Comparison of the Non-Local methods with existing baselines }\label{subsec:non_local_methods} 

\begin{figure*}[t]
	\centering
		\includegraphics[width=0.825\linewidth]{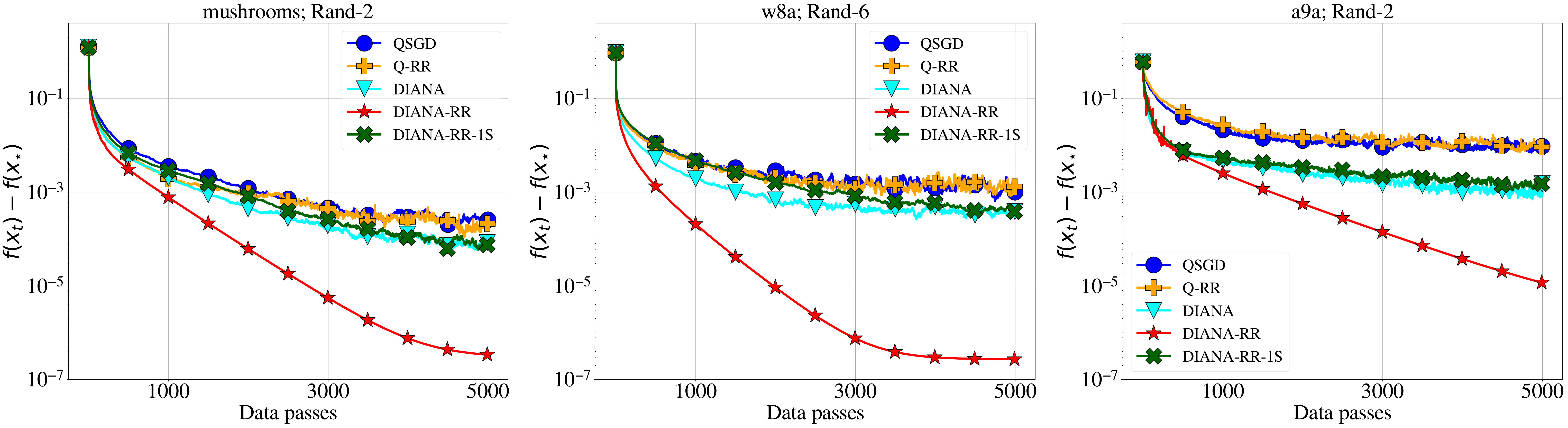} 
		\caption{Non-local methods}\label{fig:non_local_1}
		\includegraphics[width=0.825\linewidth]{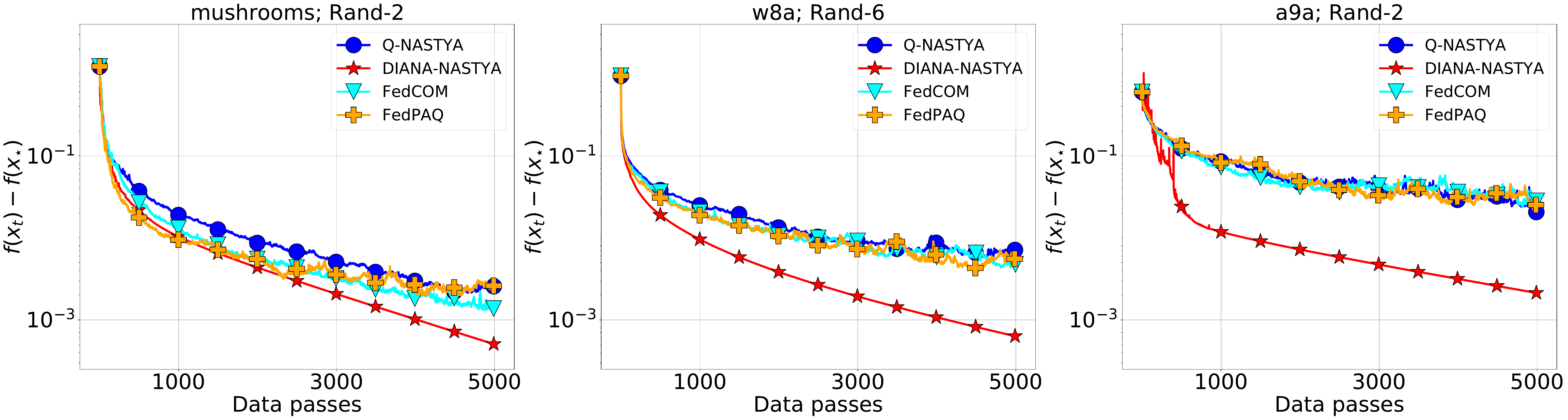} 
		\caption{Local methods}\label{fig:local_1}
  \caption{The comparison of the proposed methods (\gls{Q-NASTYA}, \gls{DIANA-NASTYA}, \gls{Q-RR}, \gls{DIANA-RR}), \algname{DIANA-RR-1S} (a version of \gls{DIANA-RR}), and existing baselines (\algname{QSGD}, \gls{DIANA}, \algname{FedCOM}, \algname{FedPAQ}). All methods use tuned stepsizes and the Rand-$k$ compressor.}
 \label{fig:tuned experimets_1}
\end{figure*}
In our first experiment (see Figure~\ref{fig:non_local}), we compare the new methods \gls{Q-RR}, \gls{DIANA-RR}, and \algname{DIANA-RR-1S} with classical baselines (\algname{QSGD} \citep{alistarh2017qsgd}, \gls{DIANA} \citep{mishchenko2019distributed}) that use a with-replacement mini-batch \algname{SGD} estimator. \algname{DIANA-RR-1S} is a memory-friendly version of \gls{DIANA-RR} that stores and uses a single shift $h^{t}_{m}$ on the worker side rather than $n$ individual shifts $h^{t}_{m,\pi_m^i}$. Figure~\ref{fig:non_local} illustrates that \gls{Q-RR} exhibits similar behavior to \algname{QSGD}, with both methods being slower than \gls{DIANA} methods across all considered datasets.
\algname{DIANA-RR-1S} and \gls{DIANA} show comparable convergence rates, indicating that random reshuffling alone, without introducing additional shifts, does not make a significant difference.
Finally, \gls{DIANA-RR} achieves the best rate among all considered non-local methods, efficiently reducing the variance and reaching the lowest functional sub-optimality tolerance. These experimental results align perfectly with our theoretical analysis.
For each of the considered non-local methods, we take the stepsize as the largest one predicted by the theory premultiplied by the individually tuned constant factor from the set \{0.000975, 0.00195, 0.0039, 0.0078, 0.0156, 0.0312, 0.0625, 0.125, 0.25, 0.5, 1, 2, 4, 8, 16, 32, 64, 128, 256, 512, 1024, 2048, 4096\}.

Therefore, for each local method on every dataset, we performed $20$ launches to find the stepsize multiplier showing the best convergence behavior (the fastest reaching the lowest possible level of functional suboptimality $f(x^t) - f(x^{\star})$).

Theoretical stepsizes for methods \gls{Q-RR} and \gls{DIANA-RR} are provided by the Theorems \ref{th_conv_new_rr_q} and \ref{th_conv_rr_diana}, whereas stepsizes for \algname{QSGD} and \gls{DIANA} were taken from the paper \citep{gorbunov2020unified}.

\subsubsection{Experiment 2: Comparison of the proposed local methods with existing baselines }\label{subsec:local_methods}

The second experiment shows that the \gls{DIANA}-based method can significantly outperform in practice when one applies it to local methods as well. In particular, whereas \gls{Q-NASTYA} shows comparable behavior to existing methods \algname{FedCOM} \citep{haddadpour2021federated}, \algname{FedPAQ} \citep{reisizadeh19_fedpaq} in all considered datasets, \gls{DIANA-NASTYA} noticeably outperforms other methods.

In this set of experiments, we tuned stepsizes similarly to the non-local methods. However, for algorithms \gls{Q-NASTYA}, \gls{DIANA-NASTYA}, and \algname{FedCOM} we needed to independently adjust the client and server stepsizes, leading to a more extensive tuning routine. 

As before, for each local method on every dataset, tuned client and server stepsizes are defined by the theoretical one and adjusted constant multiplier.
Theoretical stepsizes for methods \gls{Q-NASTYA} and \gls{DIANA-NASTYA} are given by the Theorems \ref{thm:convergence_Q_NASTYA} and \ref{thm:diana-nastya-conv}, whereas \algname{FedCOM} and \algname{FedPAQ} stepsizes  were taken from the papers \citep{haddadpour2021federated} and \citep{reisizadeh19_fedpaq} respectively.
We now list all the considered multipliers of client and server stepsizes for every method (i.e. $\gamma$ and $\eta$ respectively):
\begin{itemize}
	\item \gls{Q-NASTYA}:
	\begin{itemize}
		\item
		Multipliers for $\gamma:$
		$\{$0.000975, 0.00195, 0.0039,	0.0078,	0.0156,	0.0312,	0.0625,	0.125,	0.25, 0.5, 1,	2,	4,	8,	16,	32,	64,128$\}$;
		\item
		Multipliers for $\eta:$
		$\{$0.0039,	0.0078,	0.0156,	0.0312,	0.0625,	0.125,	0.25, 0.5, 1,	2,	4,	8,	16,	32,	64, 128$\}$. 
	\end{itemize}
	\item \gls{DIANA-NASTYA}:
	\begin{itemize}
		\item Multipliers for $\gamma$ and $\eta:$
		$\{$0.000975, 0.00195, 0.0039,	0.0078,	0.0156,	0.0312,	0.0625,	0.125,	0.25, 0.5, 1,	2,	4,	8,	16,	32,	64,128$\}$;
	\end{itemize}
	\item \algname{FedCOM}:
	\begin{itemize}
		\item
		Multipliers for $\gamma:$
		$\{$0.0312,	0.0625,	0.125,	0.25, 0.5, 1,	2,	4,	8,	16,	32,	64, 128, 256, 512,1024, 2048, 4096, 8192, 16384, 32768 $\}$;
		\item
		Multipliers for $\eta:$
		$\{$0.000975, 0.00195, 0.0039,	0.0078,	0.0156,	0.0312,	0.0625,	0.125,	0.25, 0.5, 1,	2,	4,	8,	16,	32,	64, 128$\}$.
	\end{itemize}
	\item \algname{FedPAQ}:
	\begin{itemize}
		\item 
		Multipliers for $\gamma:$
		$\{$0.00195, 0.0039, 0.0078, 0.0156, 0.0312, 0.0625, 0.125, 0.25, 0.5, 1, 2, 4, 8, 16, 32, 64, 128, 256, 512, 1024, 2048, 4096, 8192, 16384, 32768, 65536, 131072, 262144, 524288, 1048576 $\}$.
	\end{itemize}
\end{itemize}

For example, to find the best pair $(\gamma, \eta)$ for \algname{FedCOM} method on each dataset, we performed $378$ launches. A similar subroutine was executed for all algorithms on all datasets independently.

\subsubsection{Experiment 3: Comparison of DIANA-RR with EF21 and DIANA}\label{subsec:non-local-methods-vs-ef21}

In our third experiment (see Figure~\ref{fig:experimets_3}), we compared \gls{DIANA-RR} with \gls{DIANA} and \algname{EF21-SGD} \citep{fatkhullin2021ef21}. The \algname{EF21-SGD} is the state-of-the-art algorithm for contractive compressors in distributed non-convex settings. All compared algorithms used a with-replacement mini-batch \algname{SGD} estimator, consistent with the setup in Section~\ref{subsec:non_local_methods}. However, in this experiment in contrast with Section ~\ref{subsec:non_local_methods}, we employed reliable theoretical step sizes that ensure guaranteed convergence.

The \algname{EF21} algorithm family is designed for usage with contraction compressors in non-convex optimization for $L$-smooth objective functions in the form of Equation \ref{eq:erm-opt-orig}. For scenarios involving unbiased compressors such as Rand-$k$, the \algname{EF21-SGD} can be adapted through scaling \citep{fatkhullin2021ef21}. More specifically, an unbiased compressor $\mathcal{C}(x): \mathbb{R}^d \to \mathbb{R}^d$ which satisfied Assumption~\ref{asm:quantization_operators} via applying the transformation $C'(x) \eqdef (\omega + 1)^{-1} \cdot \mathcal{C}(x)$ yields a contraction compressor $\mathbb{E}\left[\|C'(x)-x\|^2\right] \le (1-\alpha) \|x\|^2, \forall x \in \mathbb{R}^d$ with $\alpha = \nicefrac{1}{\omega + 1}$. In particular, this procedure makes Rand-$k$ compatible with the \algname{EF21} algorithm family. In this experiment, we implemented \algname{EF21-SGD} following the refined analysis from \citep{richtarik2024error}, which offers a strictly better convergence guarantee through improved bounds on the theoretical step size compared to \citep{fatkhullin2021ef21}.

As shown in Figure~\ref{fig:experimets_3} \algname{EF21-SGD} does not perform fast enough in scenarios involving using unbiased compressors for strongly-convex optimization problems compared to \gls{DIANA-RR} and \gls{DIANA}.

\begin{figure*}[t]
  \centering
  \includegraphics[width=0.48\textwidth]{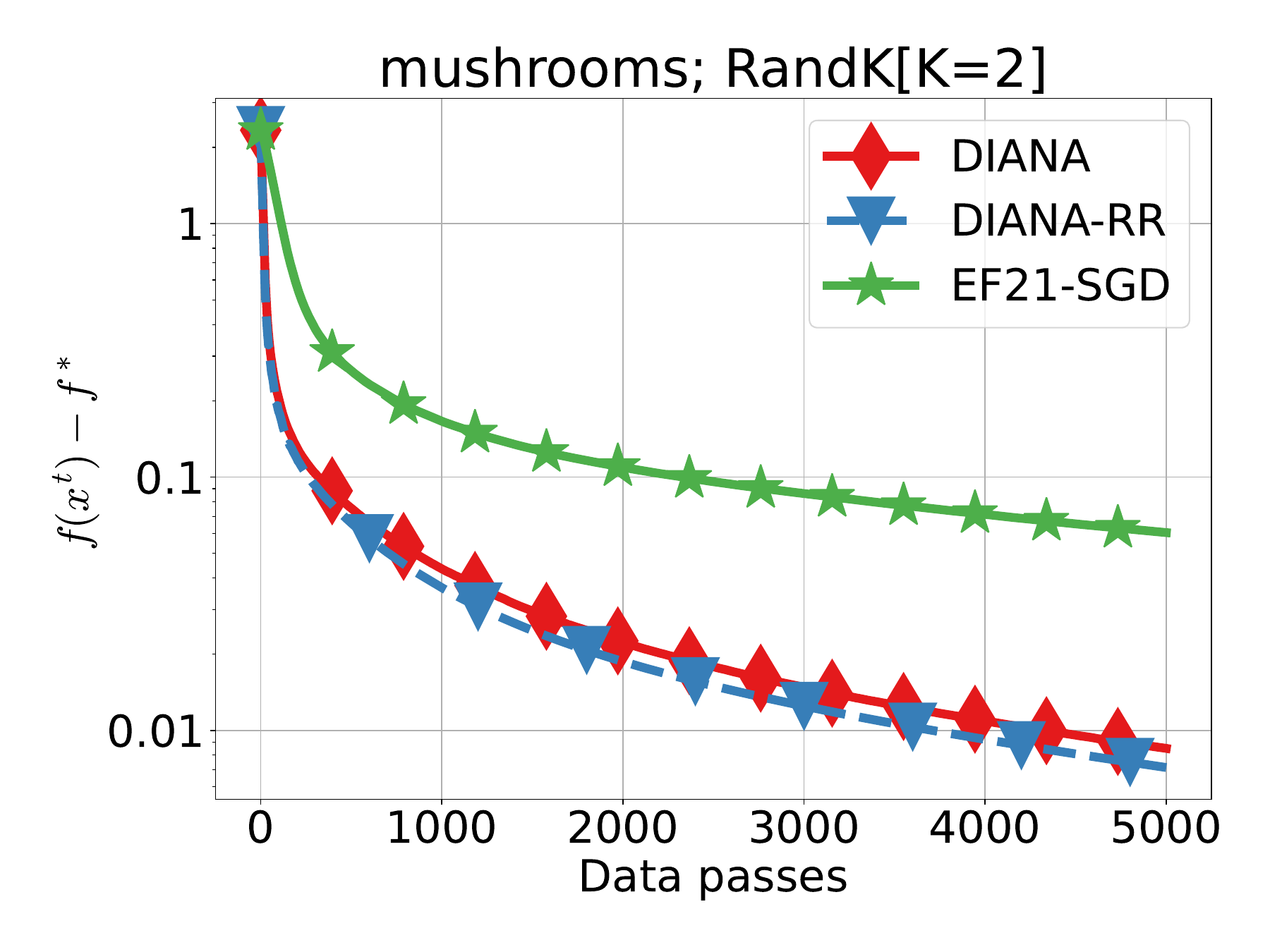}
  \includegraphics[width=0.48\textwidth]{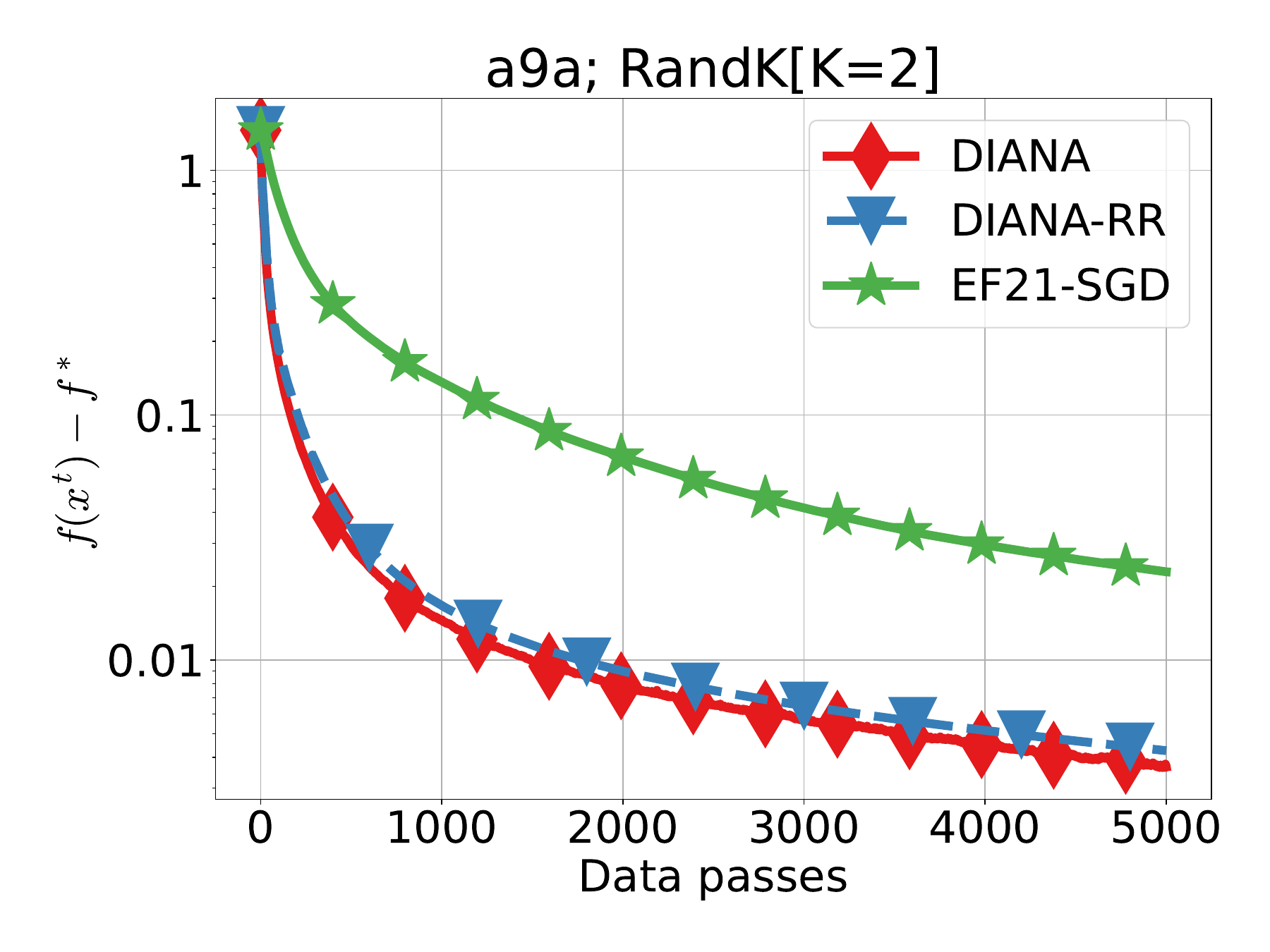} 

  \caption{The comparison of the proposed variance-reduced \gls{DIANA-RR} and baselines \gls{DIANA}, \algname{EF21-SGD}. All algorithms use theoretical step-sizes, Rand-$k$ compressor, number of workers is $20$.}
 \label{fig:experimets_3}
\end{figure*}


\subsection{Training deep neural network model: ResNet-18 on CIFAR-10}

\begin{figure*}[t]
	\centering
	\captionsetup[sub]{font=small,labelfont={}}	
		\includegraphics[width=0.48\textwidth]{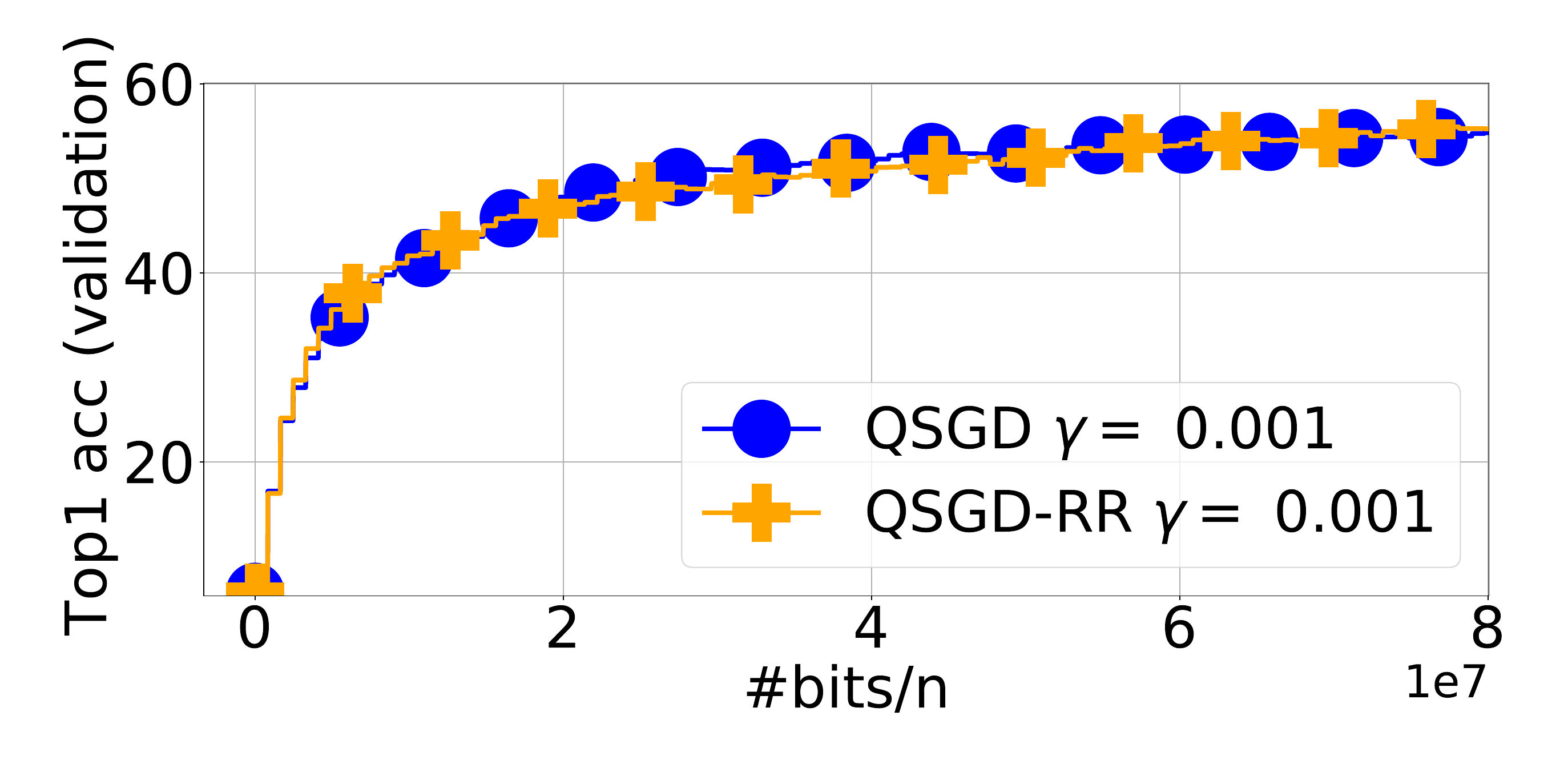} 
		\includegraphics[width=0.48\textwidth]{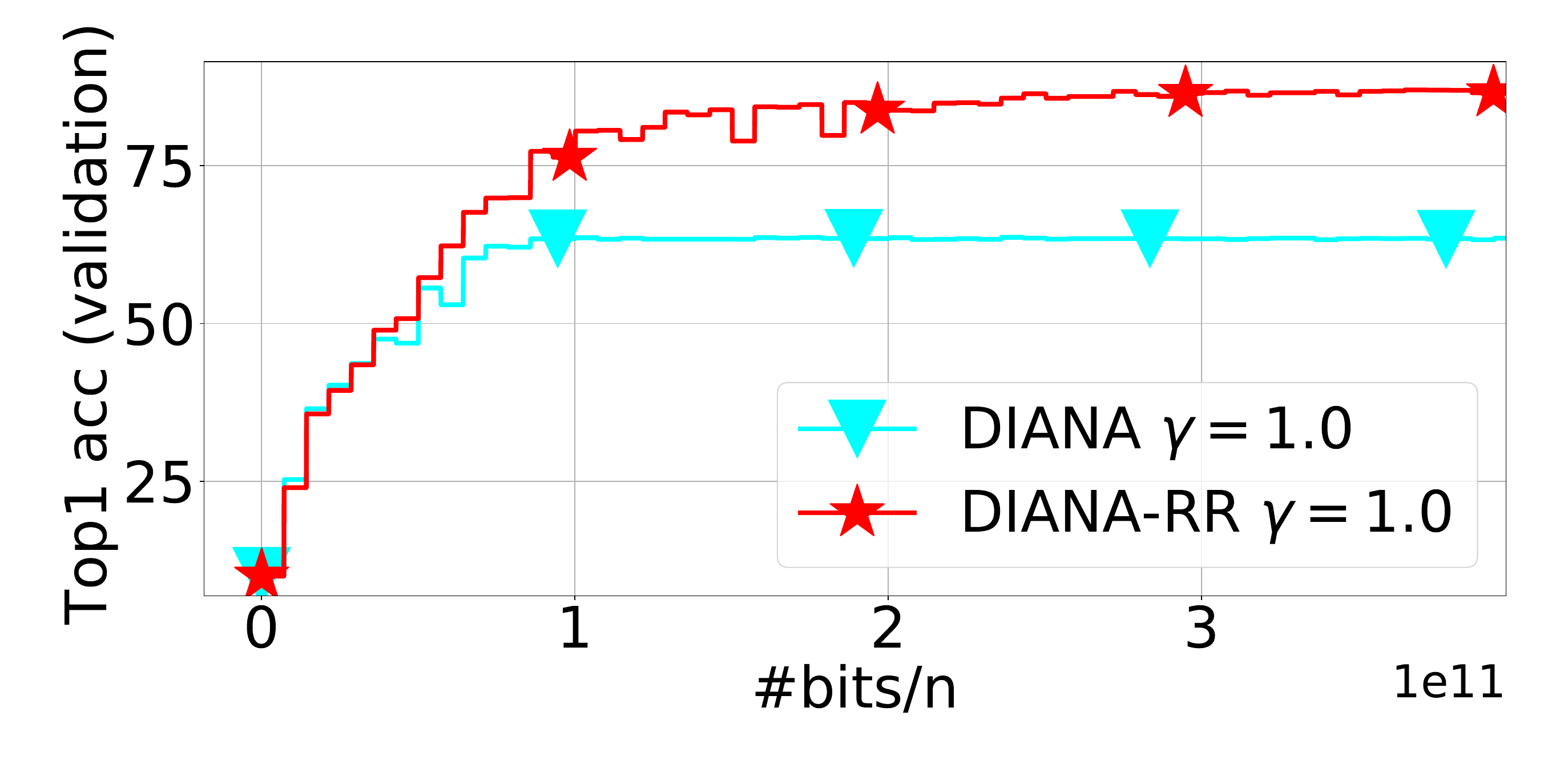} 
	\caption{The comparison of \gls{Q-RR}, \algname{QSGD}, \gls{DIANA}, and \gls{DIANA-RR} on the task of training \texttt{ResNet-18} on \texttt{CIFAR-10} with $n=10$ workers. Top-1 accuracy on test set is reported. Stepsizes were tuned and workers used Rand-$k$ compressor with $\nicefrac{k}{d} \approx 0.05$. 
 }
	\label{fig:NN_plots_main}
\end{figure*}
Since random reshuffling is a very popular technique in training neural networks, it is natural to test the proposed methods on such problems. Therefore, in the second set of experiments, we consider training \texttt{ResNet-18}~\citep{resnet18} model on the \texttt{CIFAR10} dataset \citep{cifar}. To conduct these experiments we use \texttt{FL\_PyTorch} simulator \citep{burlachenko2021fl_pytorch}. 

The main goal of this experiment is to verify the phenomenon observed in Experiment 1 on the training of a deep neural network. That is, we tested \gls{Q-RR}, \algname{QSGD}, \gls{DIANA}, and \gls{DIANA-RR} in the distributed training of \texttt{ResNet-18} on \texttt{CIFAR10}, see Figure~\ref{fig:NN_plots_main}. As in the logistic regression experiments, we observe that (i) \gls{Q-RR} and \algname{QSGD} behave similarly and (ii) \gls{DIANA-RR} outperforms \gls{DIANA}.

To illustrate the behavior of the proposed methods in training Deep Neural Networks (DNN), we consider the \texttt{ResNet-18}~\citep{resnet18} model. This model is used for image classification, feature extraction for image segmentation, object detection, image embedding, and image captioning. We train all layers of the \texttt{ResNet-18} model, meaning that the dimension of the optimization problem equals $d=11,173,962$. During the training, the \texttt{ResNet-18} model normalizes layer inputs via exploiting $20$ Batch Normalization ~\citep{ioffe2015batch} layers that are applied directly before nonlinearity in the computation graph of this model. Batch normalization (BN) layers add $9600$ trainable parameters to the model. Besides trainable parameters, a BN layer has its internal state that is used for computing the running mean and variance of inputs due to its own specific regime of working. 
We use \textit{He} initialization~\citep{he2015delving}.

\subsubsection{Computing environment} 
We performed numerical experiments on a server-grade machine running Ubuntu 18.04 and Linux Kernel v5.4.0, equipped with 16-cores (2 sockets by 16 cores per socket) 3.3 GHz Intel Xeon, and four NVIDIA A100 GPU with 40GB of GPU memory. The distributed environment is simulated in Python 3.9 via using the software suite \texttt{FL\_PyTorch} ~\citep{burlachenko2021fl_pytorch} that serves for carrying complex Federated Learning experiments. \texttt{FL\_PyTorch} allowed us to simulate the distributed environment in the local machine. Besides storing trainable parameters per client, this simulator stores all not trainable parameters including BN statistics per client.

\subsubsection{Loss function} 
Training of \texttt{ResNet-18} can be formalized as problem \eqref{eq:erm-opt-orig} with the following choice of $f_m^i$ 
\begin{equation}
	f_m(x)=\frac{1}{|n_{m}|}\sum_{j=1}^{|n_{m}|} CE(b^{(j)}, g(a^{(j)},x)),
\end{equation}
where $CE(p, q) \eqdef -\sum_{k=1}^{\mathrm{\#classes}} p_i \cdot \log(q_i)$ with agreement $0\cdot\log(0) = 0$ is a standard cross-entropy loss, function $g: \mathbb{R}^{28 \times 28} \times \mathbb{R}^d \to [0,1]^{\mathrm{\#classes}}$ is a neural network taking image $a^{(j)}$ and vector of parameters $x$ as an input and returning a vector in probability simplex, and $n_m$ is the size of the dataset on worker $m$. 

\subsubsection{Dataset and metric} 
In our experiments, we used \texttt{CIFAR10} dataset \citep{cifar}. The dataset consists of input variables $a_i \in \mathbb{R}^{28 \times 28 \times 3}$, and response variables $b_i \in \{0,1\}^{10}$ and is used for training 10-way classification. The sizes of training and validation set are $5\times 10^4$ and $10^4$ respectively. The training set is partitioned heterogeneously across $10$ clients. 
To measure the performance, we evaluate the loss function value $f(x)$, norm of the gradient $\|\nabla f(x)\|_2$ and the Top-1 accuracy of the obtained model as a function of passed epochs and the normalized number of bits sent from clients to the server.

\subsubsection{Tuning process}  
In this set of experiments, we tested \algname{QSGD}~\citep{alistarh2017qsgd}, \gls{Q-RR} (Algorithm \ref{alg_new_Q_RR}), \gls{DIANA}~\citep{mishchenko2019distributed} and \gls{DIANA-RR} (Algorithm \ref{alg_new_RR_DIANA}) algorithms. For all algorithms, we tuned the strategy $\in \{A, B, C\}$ of decaying stepsize model via selecting the best in terms of the norm of the full gradient on the train set in the final iterate produced after $20 000$ rounds. The stepsize policies are described below.
\begin{enumerate}[label=\Alph*.]
	\item Stepsizes decaying as inverse square root of the number epochs
	\[
	\gamma_e = \begin{cases}
		\gamma_{init}  \cdot \dfrac{1}{\sqrt{e - s + 1}}, &\text{if } e \ge s,\\
		\gamma_{init}, &\text{if } e < s,
	\end{cases}
	\]
	where $\gamma_e$ denotes the stepsize used during epoch $e+1$, $s$ is a fixed shift.
	\item Stepsizes decaying as inverse of number epochs
	\[
	\gamma=\begin{cases}
		\gamma_{init}  \cdot \dfrac{1}{{e - s + 1}}, &\text{if } e \ge s,\\
		\gamma_{init}, &\text{if } e < s.
	\end{cases}
	\]
	\item Fixed stepsize
	\[
	\gamma=\gamma_{init}.	
	\]
	
\end{enumerate}

We say that the algorithm passed $e$ epochs if the total number of computed gradient oracles lies between $e \sum_{m=1}^M n_m$ and $(e+1) \sum_{m=1}^M n_m$. For each algorithm the used stepsize $\gamma_{init}$ and shift parameter $s$ were tuned via selecting from the following sets:
\begin{eqnarray*}
	\gamma_{init} \in \gamma_{set} \eqdef \{4.0, 3.75, 3.00, 2.5, 2.00, 1.25, 1.0, 0.75, 0.5, 0.25,\\
	0.2, 0.1, 0.06, 0.03, 0.01, 0.003, 0.001, 0.0006\}.
\end{eqnarray*}
\begin{eqnarray*}
	s \in s_{set} \eqdef \{50, 100, 200, 500, 1000\}.
\end{eqnarray*}


In all tested methods, clients independently apply Rand-$k$  compression with cardinality $k= \lfloor 0.05 d \rfloor$. Computation for all gradient oracles is carried out in single precision float (FP64) arithmetic.

\begin{figure}[t]
	\centering
	\captionsetup[sub]{font=small,labelfont={}}	
		\includegraphics[width=0.48\textwidth]{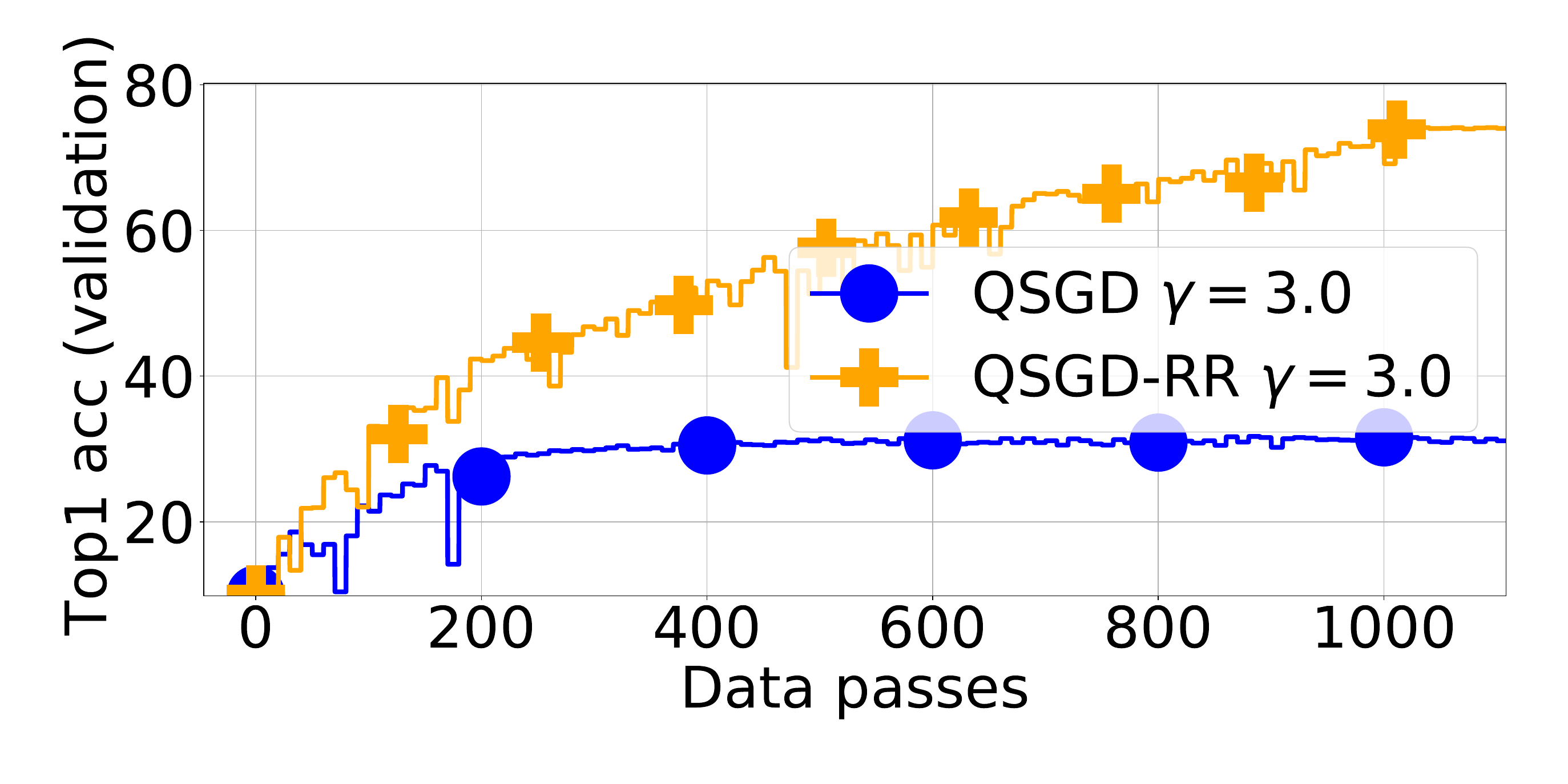} \label{fig:train_resnet18_qsgd_a_1}
		\includegraphics[width=0.48\textwidth]{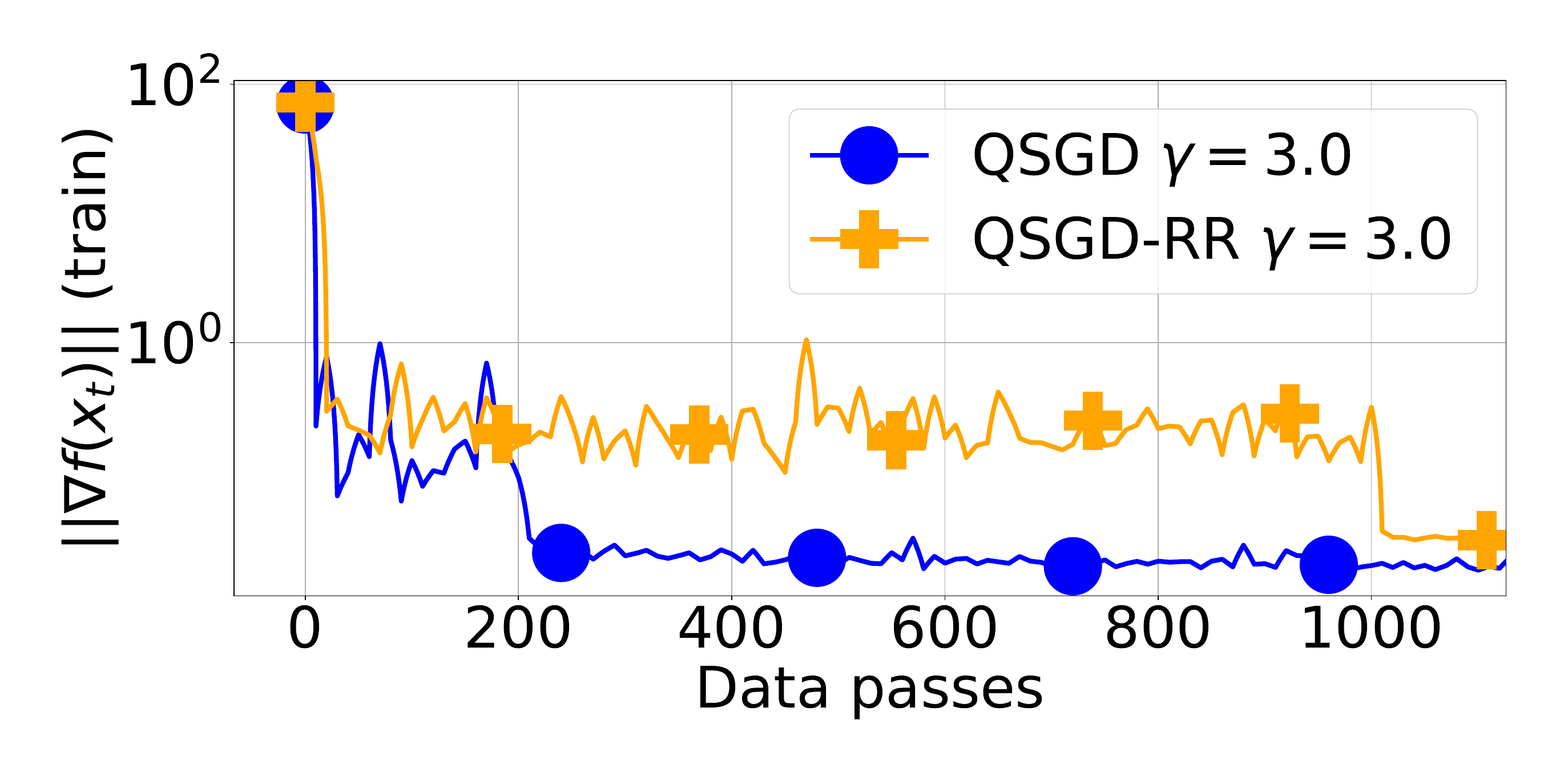} \label{fig:train_resnet18_qsgd_b_1}
		\includegraphics[width=0.48\textwidth]{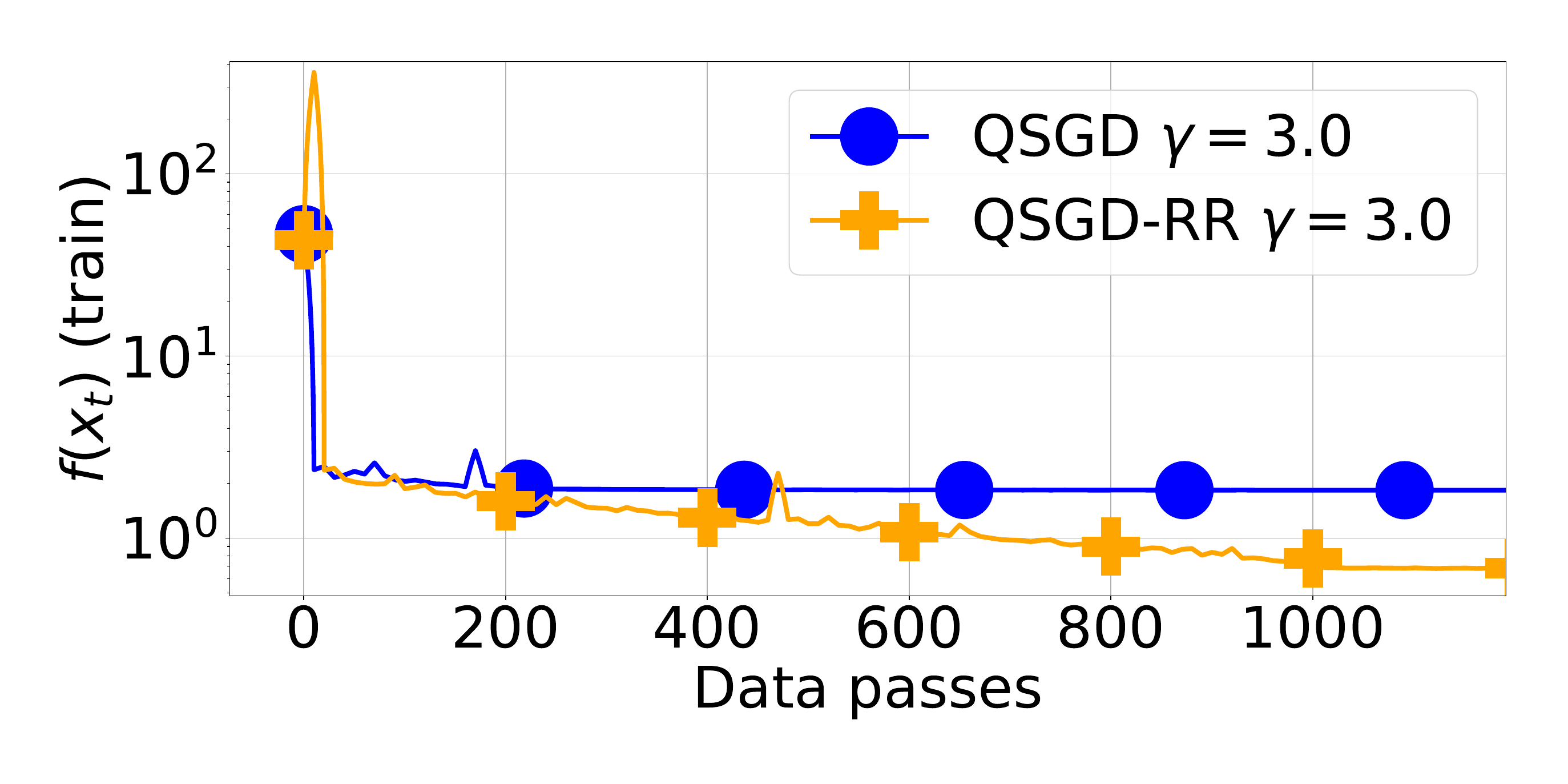}
  \label{fig:train_resnet18_qsgd_c_1}
		\includegraphics[width=0.48\textwidth]{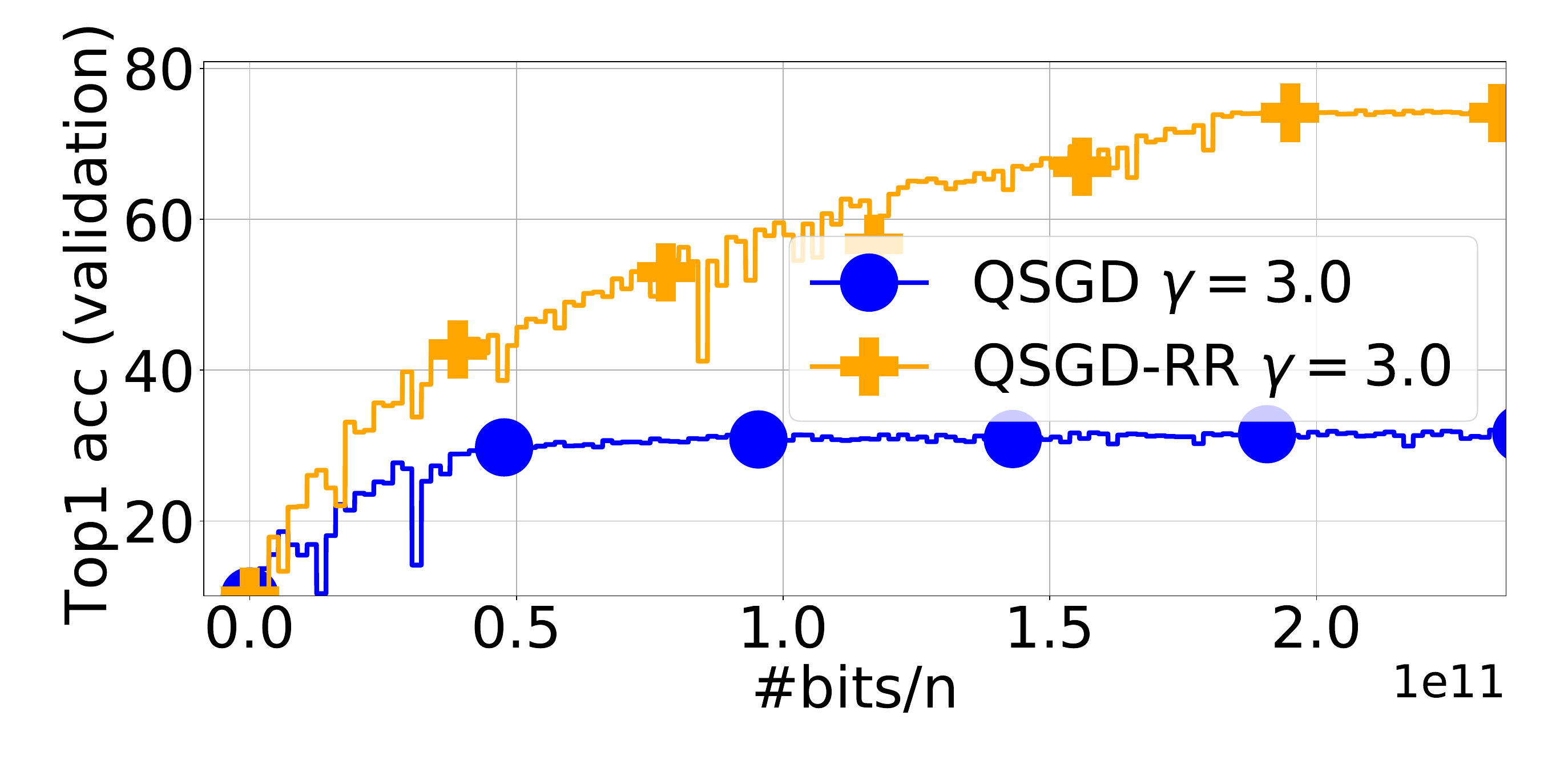} \label{fig:train_resnet18_qsgd_d_1}
		\includegraphics[width=0.48\textwidth]{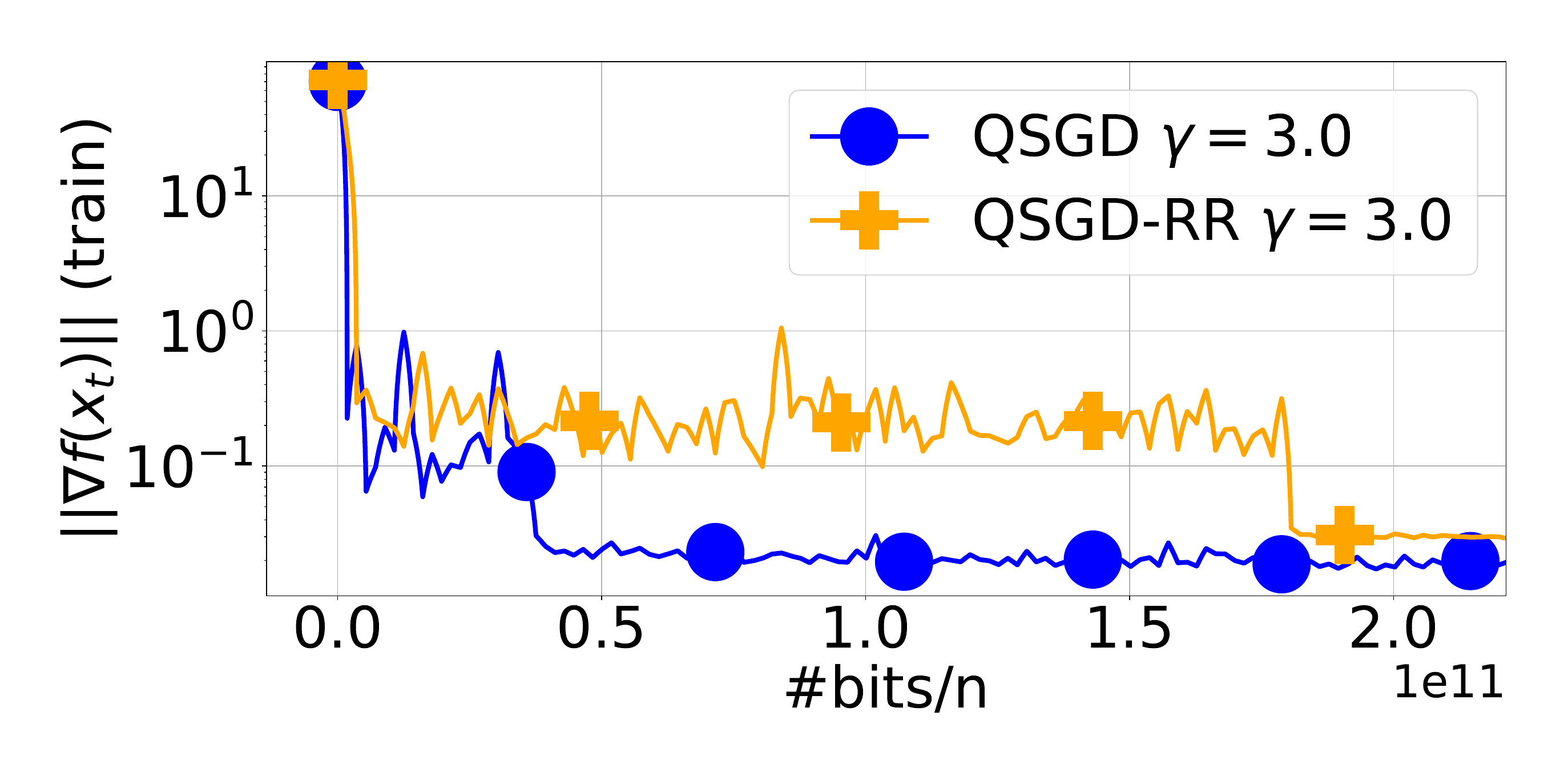}\label{fig:train_resnet18_qsgd_e_1}
		\includegraphics[width=0.48\textwidth]{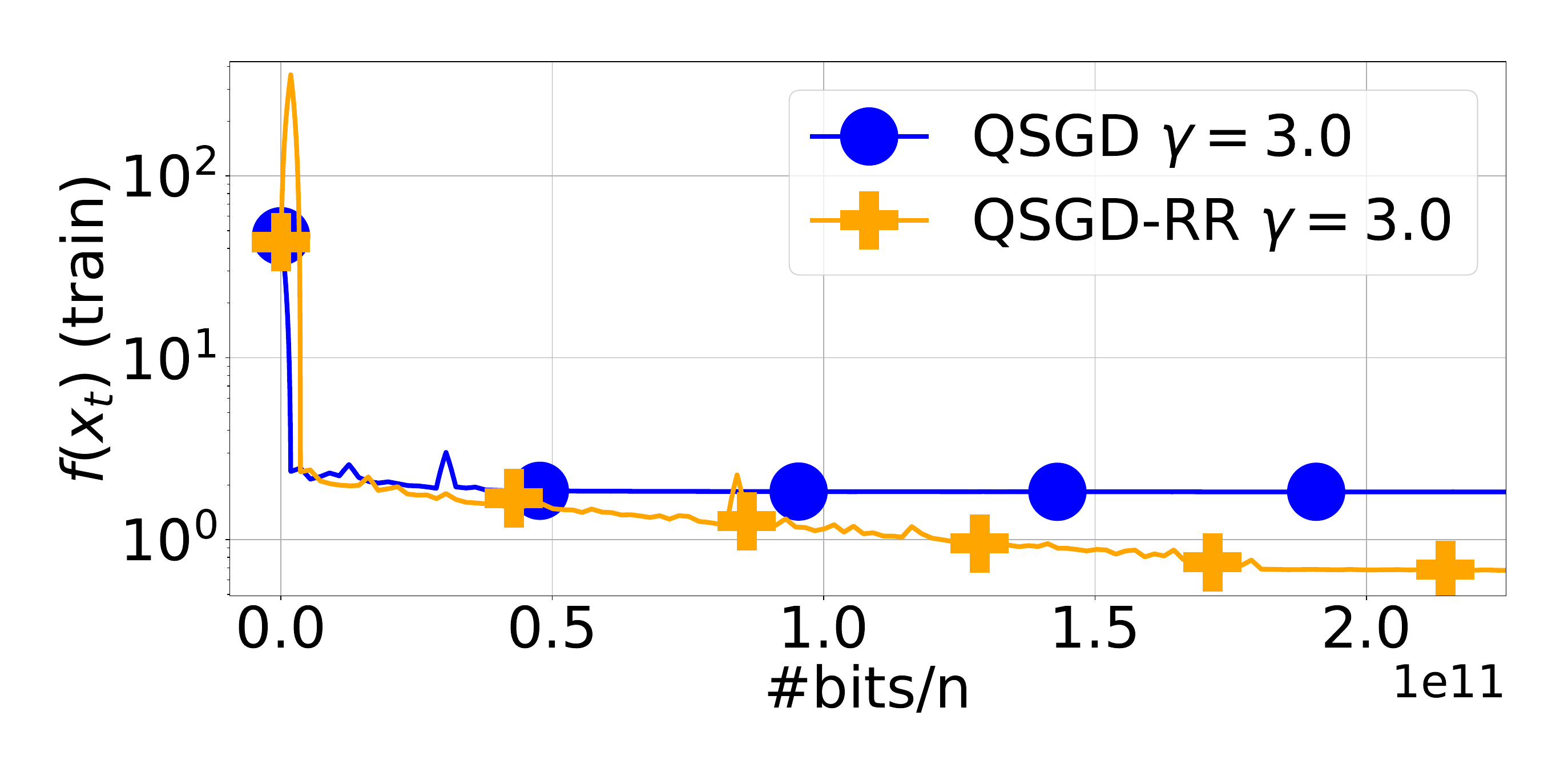}\caption{}\label{fig:train_resnet18_qsgd_f_1}
	\caption{{Comparison of \algname{QSGD} and \gls{Q-RR} in the training of \texttt{ResNet-18} on \texttt{CIFAR-10}, with $n=10$ workers. Here (a) and (d) show Top-1 accuracy on test set, (b) and (e) -- norm of full gradient on the train set, (c) and (f) -- loss function value on the train set. Stepsizes and decay shifts have been tuned from $s_{set}$ and $\gamma_{set}$ based on minimum achievable value of loss function on the train set. 
	}}
	\label{fig:train_resnet18_qsgd}
\end{figure}

\begin{figure}[t]
	\centering
	\captionsetup[sub]{font=small,labelfont={}}	
		\includegraphics[width=0.48\textwidth]{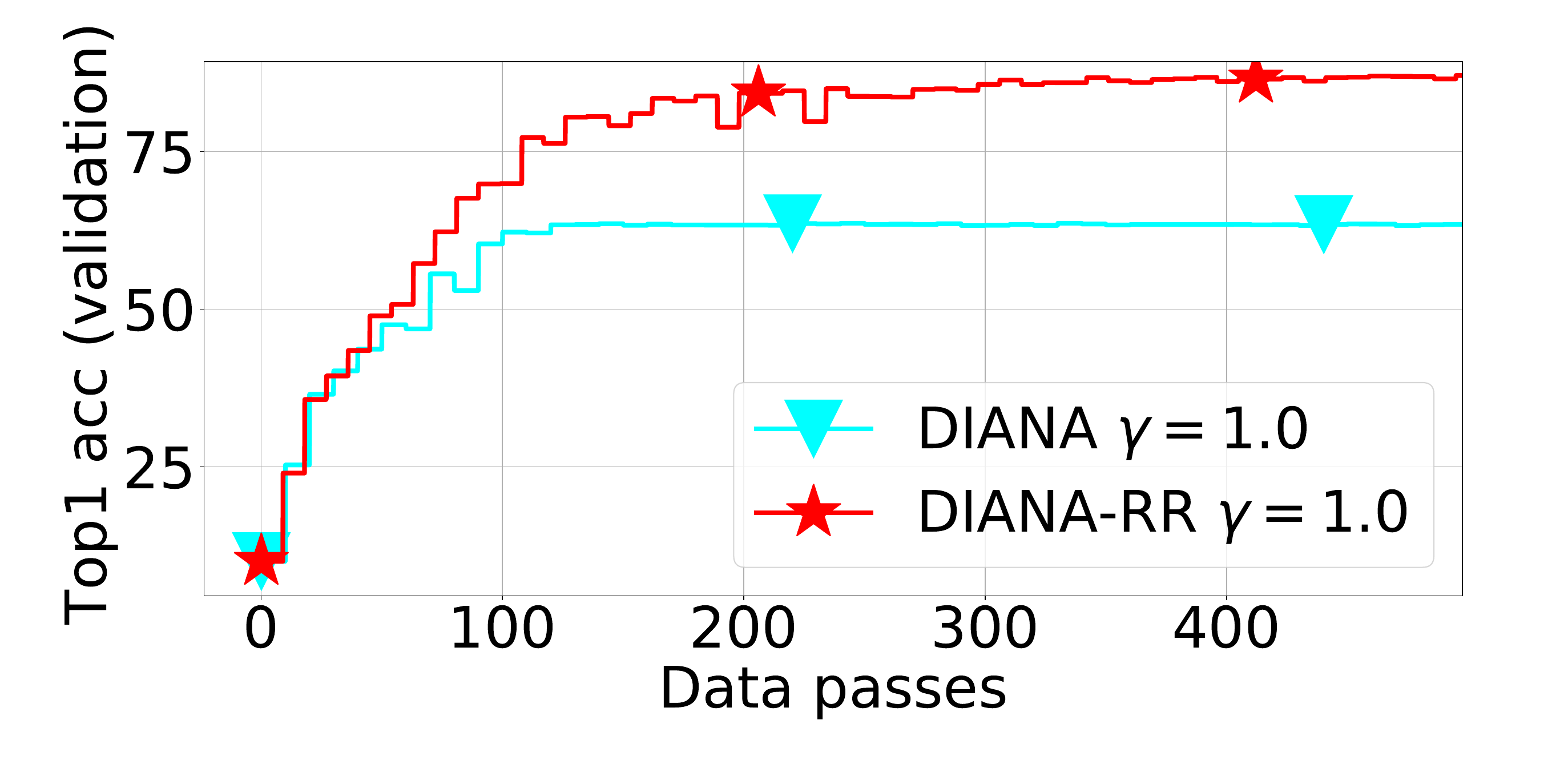} \label{fig:train_resnet18_diana_a}
		\includegraphics[width=0.48\textwidth]{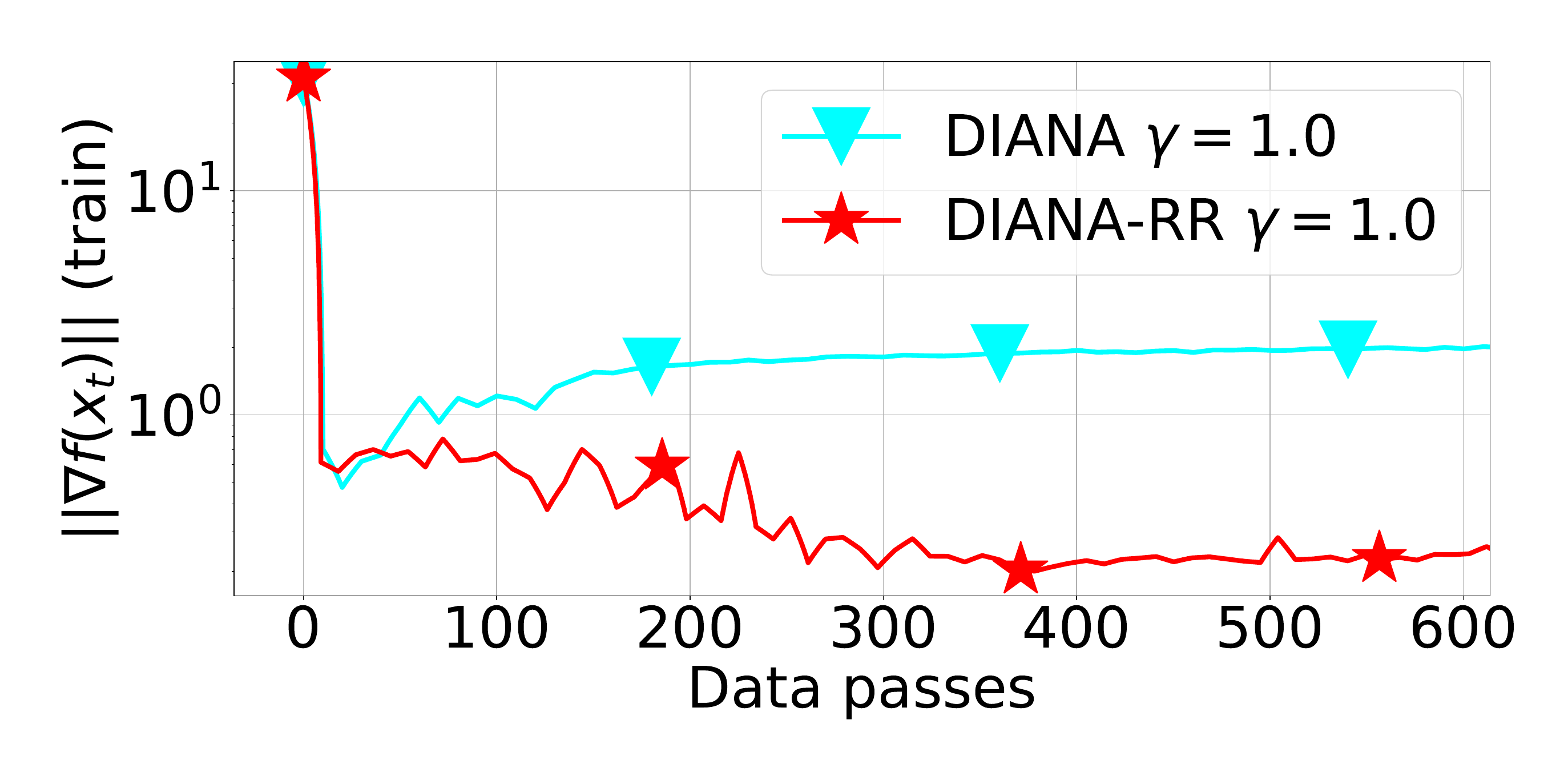} \label{fig:train_resnet18_diana_b}
		\includegraphics[width=0.48\textwidth]{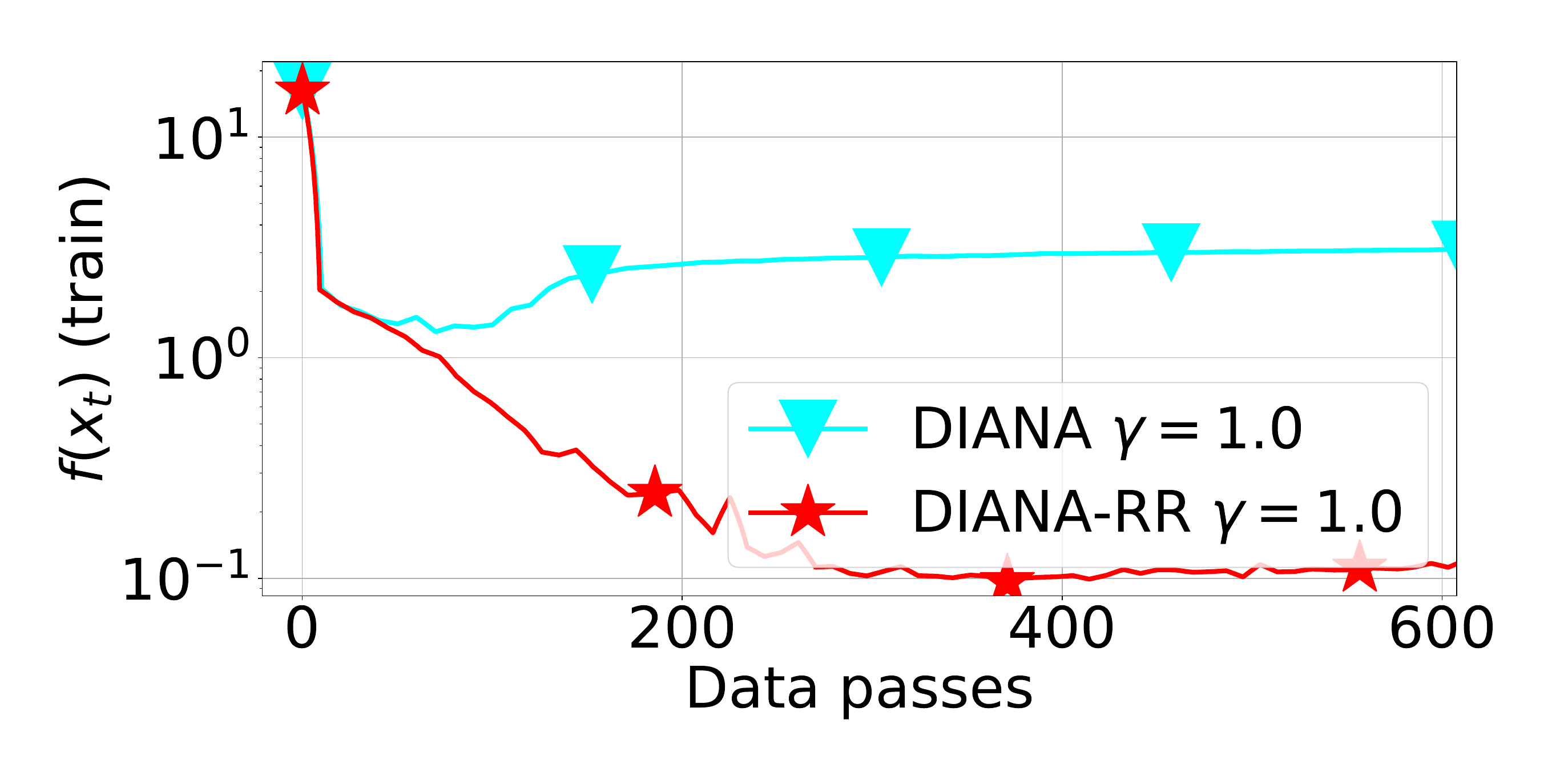} \label{fig:train_resnet18_diana_c}
		\includegraphics[width=0.48\textwidth]{RR-DIANA/plots_nn/diana_fig_2.pdf} \label{fig:train_resnet18_diana_d}
		\includegraphics[width=0.48\textwidth]{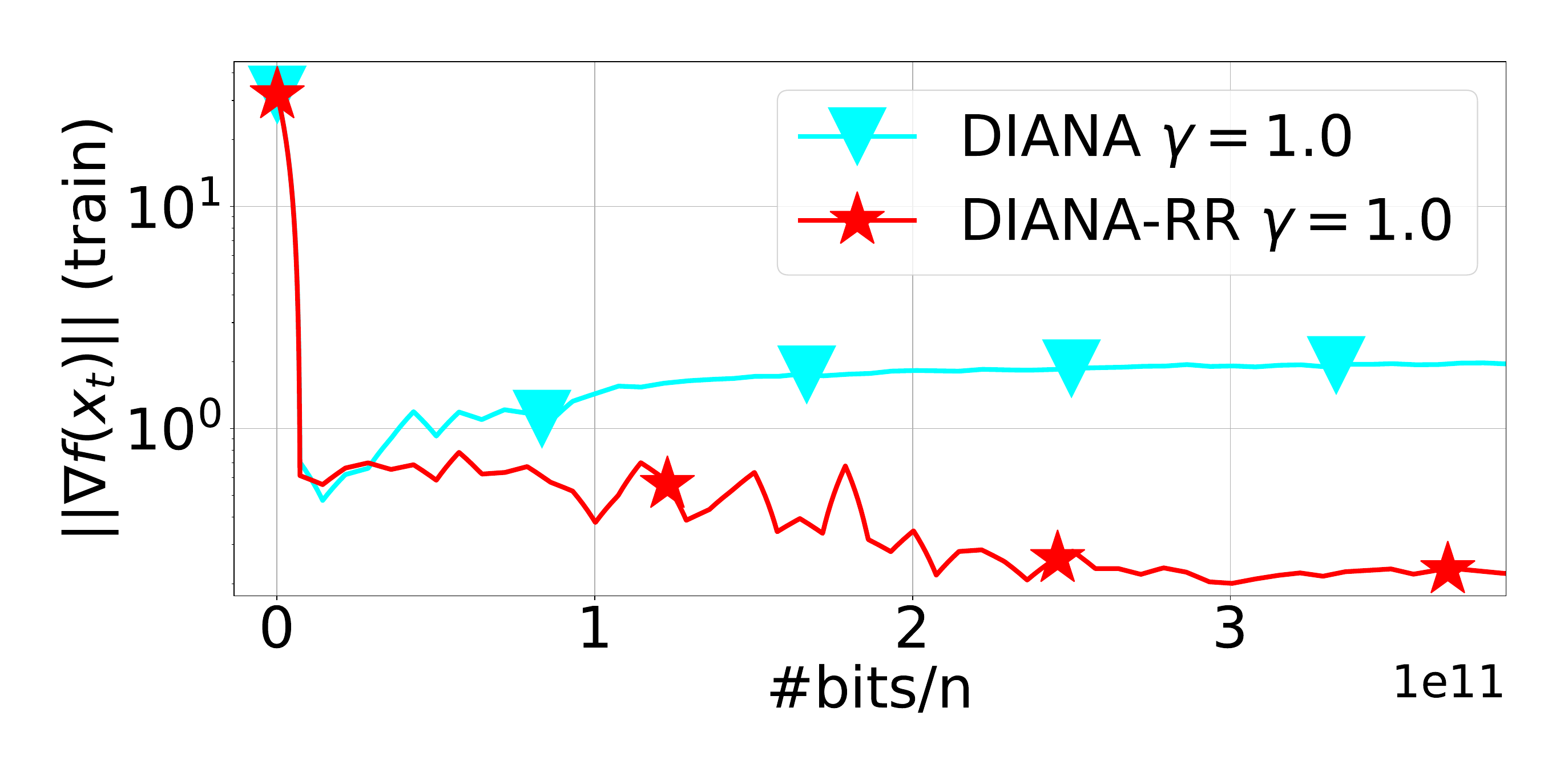}\label{fig:train_resnet18_diana_e}
		\includegraphics[width=0.48\textwidth]{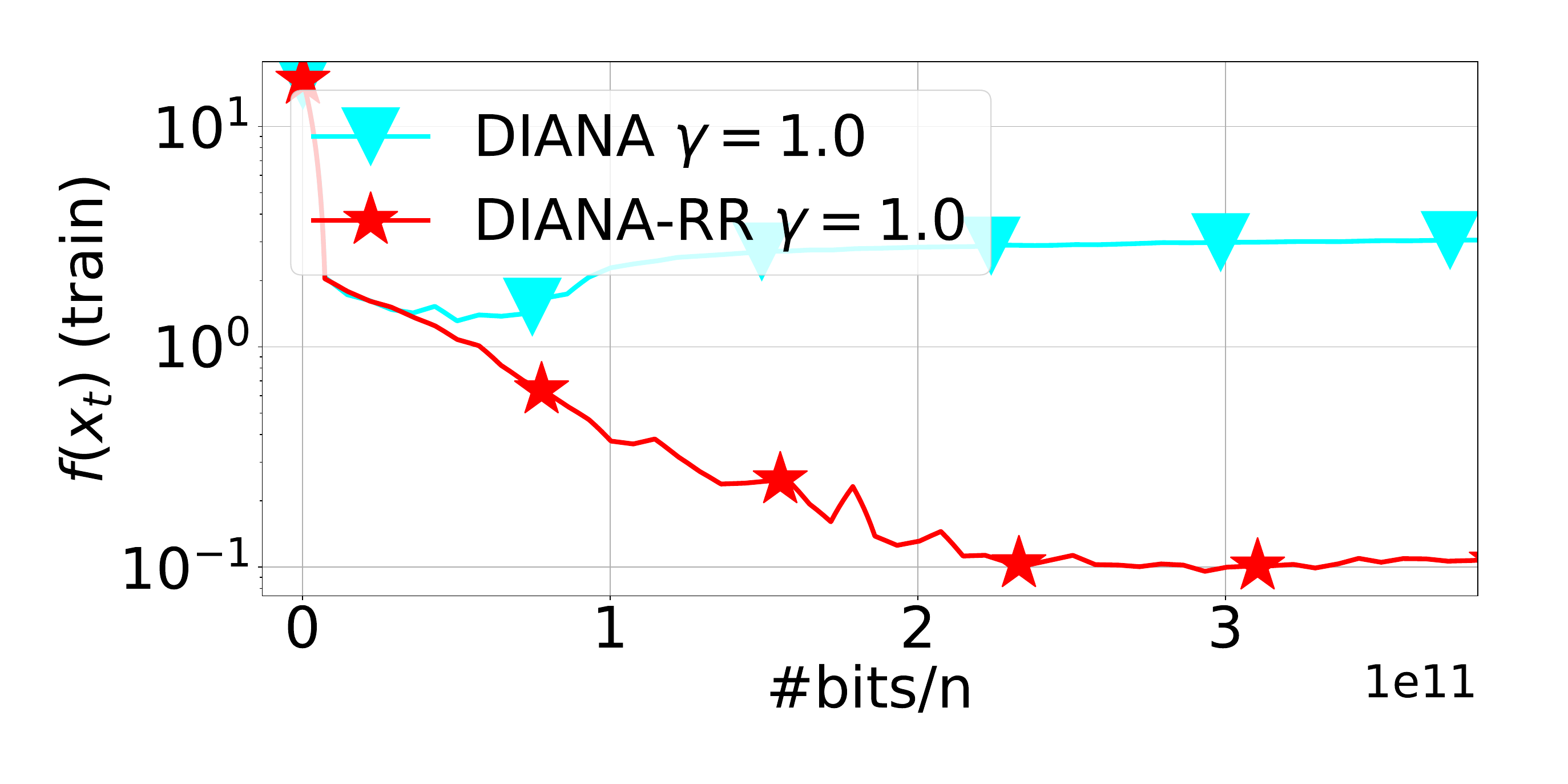}\label{fig:train_resnet18_diana_f}
	\caption{{Comparison of \gls{DIANA} and \gls{DIANA-RR} in the training of \texttt{ResNet-18} on \texttt{CIFAR-10}, with $n=10$ workers. Here (a) and (d) show Top-1 accuracy on test set, (b) and (e) -- norm of full gradient on the train set, (c) and (f) -- loss function value on the train set. Stepsizes and decay shifts have been tuned from $s_{set}$ and $\gamma_{set}$ based on minimum achievable value of loss function on the train set. For both algorithms stepsize is fixed. For both algorithms stepsize is decaying according to strategy $B$. 
	}}
	\label{fig:train_resnet18_diana}
\end{figure}
\subsubsection{Optimization-Based Fine-Tuning for pretrained \texttt{ResNet-18}.}
In this setting, we trained \texttt{ResNet-18} image classification in a distributed way across $n=10$ clients. In this experiment, we have trained only the last linear layer. 

Next, we have turned off batch normalization. Turning off batch normalization implies that the computation graph of NN $g(a, x)$ with weights of NN denoted as $x$ is a deterministic function and does not include any internal state.

The loss function is a standard cross-entropy loss augmented with extra $\ell_2$-regularization $\alpha \nicefrac{\|x\|^2}{2}$ with $\alpha=0.0001$. Initially used weights of NN are pretrained parameters after training the model on ImageNet. 

The dataset distribution across clients has been set in a heterogeneous manner via presorting dataset $D$ by label class and after this, it was split across $10$ clients. 

The comparison of stepsizes policies used in \algname{QSGD} and \gls{Q-RR} is presented in Figure~\ref{fig:train_resnet18_qsgd_all_bad}. The behavior of the algorithms with best tuned step sizes is presented in Figure~\ref{fig:train_resnet18_qsgd_best_to_best_bad}. These results demonstrate that in this setting there is no real benefit of using \gls{Q-RR} in comparison to \algname{QSGD}.

\subsubsection{Experiments}
\label{subs:experiments}

The gradient sampling strategy in client-side during local gradient computation step for \texttt{QSGD} and \texttt{DIANA} uses \texttt{SGD-MULTI} with $\tau=10\%$ samples. In that sampling strategy, the \gls{LSGD} is estimated by sampling uniformly at random $10\%$ of local samples possibly with repetition. 


The comparison of \algname{QSGD} and \gls{Q-RR} is presented in Figure \ref{fig:train_resnet18_qsgd}. In particular, Figure \ref{fig:NN_plots_main} shows that in terms of the convergence to stationary points both algorithms exhibit similar behavior. However, \gls{Q-RR} has better generalization and in fact, converges to the better loss function value. This experiment demonstrates that \gls{Q-RR} with manually tuned stepsize can be better compared to \algname{QSGD} in terms of the final quality of obtained Deep Learning model. For \algname{QSGD} the tuned meta parameters are: $\gamma_{init}=3.0$,$s=200$, $\text{strategy}=B$. For  \texttt{QSGD-RR} tuned meta parameters are: $\gamma_{init}=3.0$, $s=1000$, $\text{strategy}=B$.

The results of comparison of \gls{DIANA} and \gls{DIANA-RR} are presented in Figure~\ref{fig:train_resnet18_diana}. For \gls{DIANA} the tuned meta parameters are: $\gamma_{init}=1.0$,$s=0$, $\text{strategy}=C$ and for  \gls{DIANA-RR} tuned meta parameters are: $\gamma_{init}=1.0$, $s=0$, $\text{strategy}=C$. These results show that \gls{DIANA-RR} outperforms \gls{DIANA} in terms of all reported metrics.

\subsection{Discussion}
\label{subs:discussion}

\paragraph{More about used arithmetics.}

We used \texttt{FP64} (IEEE 754) due to its superior numerical stability compared to \texttt{FP32}, \texttt{FP16}, and \texttt{BFloat16}. While \texttt{FP32} and \texttt{FP16} are commonly used for inference tasks, the choice of precision for training depends on the specific requirements of the task. In certain cases, \texttt{FP32} may be sufficient, but for others, \texttt{FP64} is necessary to ensure stability.

The performance gain from switching from \texttt{FP64} to \texttt{FP32} can indeed vary based on the GPU model. For instance, the NVIDIA A100 40GB GPU used in our experiments offers approximately a two-fold increase in computational throughput with \texttt{FP32} compared to \texttt{FP64}. The specific architecture of the GPU influences the choice of precision, and these characteristics can differ across various GPU models and updates.

\paragraph{The computational burden.}

The primary focus of the paper is to highlight the fundamental complexities and limits of algorithmic behavior. The experiments presented in our paper are intended for illustrative purposes.

The computational demands of our work are significant. Performing experiments beyond \texttt{ResNet-18/CIFAR-10/FP64} with 10 clients is near the limit of what is feasible with our computational resources. In our simulation involving 10 clients sharing a common dataset, we ran 2000 rounds/epochs for fine-tuning. Based on an estimate of 2 minutes per epoch, the total computation time would be approximately 66 hours per run (2 minutes/epoch × 2000 epochs = 66 hours). Taking into account the grid search with 18 preset learning rates, 5 sets of decay parameters, and 4 algorithms, the total estimated computation time would be around 23760 hours (66 hours × 18 × 5 × 4). This represents a substantial amount of computation time. Therefore, conducting a comprehensive comparison involving four algorithms with an extensive grid of hyperparameters is already challenging for models larger than ResNet-18 on CIFAR10. To cover 23760 hours of training would indeed require approximately 40 GPUs running continuously for about 25 days. Nonetheless, we have conducted numerous experiments to ensure a thorough and fair comparison.

\paragraph{Training in overparameterized regime.}

\begin{figure}[t]
	\centering
	\captionsetup[sub]{font=small,labelfont={}}	
		\includegraphics[width=0.48\textwidth]{RR-DIANA/plots_nn_not_good/best2best_fig1.pdf} 
		\includegraphics[width=0.48\textwidth]{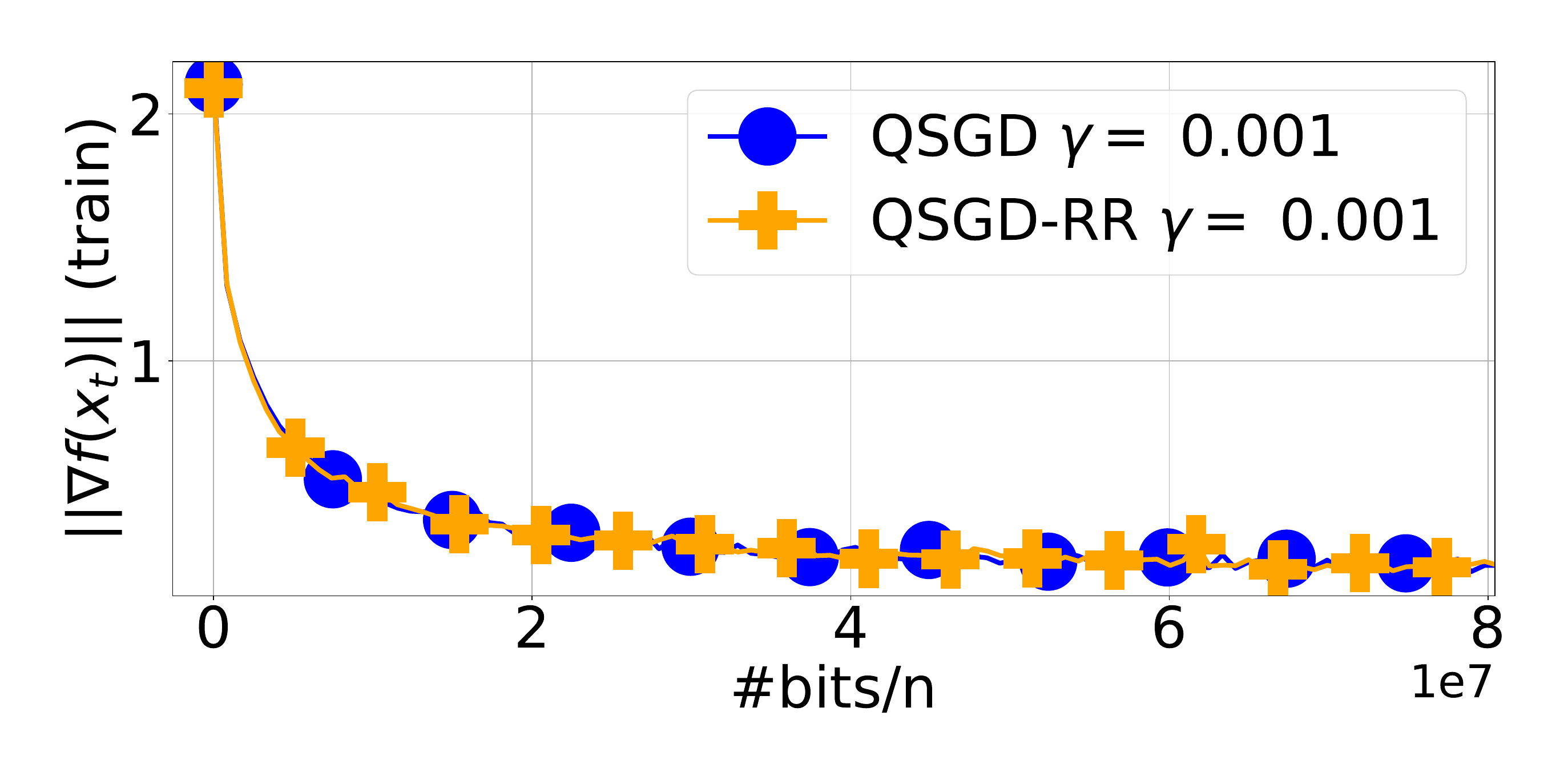} 
		\includegraphics[width=0.48\textwidth]{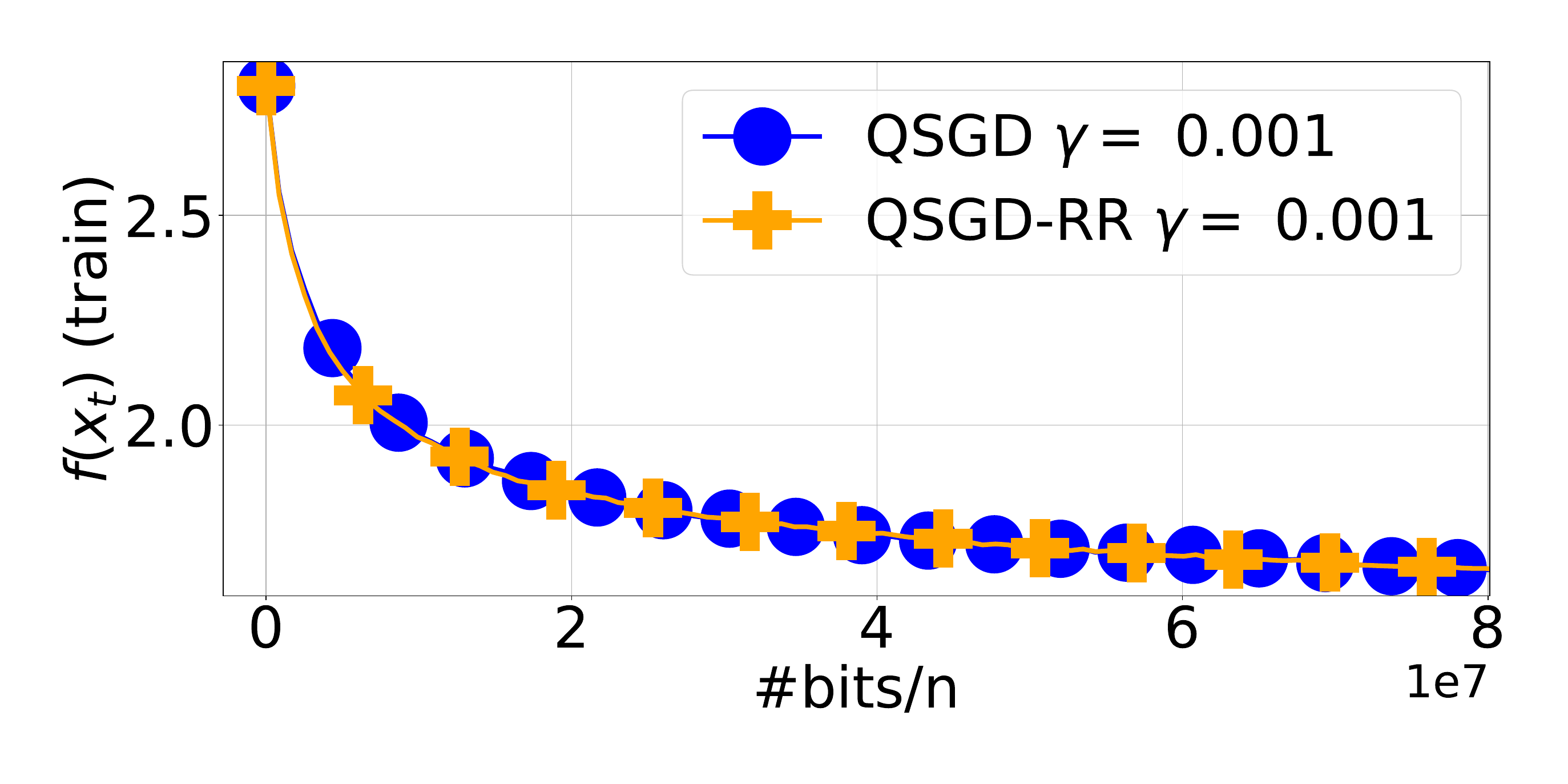} 
	\caption{{Comparison of \algname{QSGD} and \gls{Q-RR} in the training of the last linear layer of \texttt{ResNet-18} on \texttt{CIFAR-10}, with $n=10$ workers. Here (a) shows Top-1 accuracy on test set, (b) -- norm of full gradient on the train set, (c) -- loss function value on the train set. Stepsizes and decay shifts have been tuned from $s_{set}$ and $\gamma_{set}$ based on minimum achievable value of loss function on the train set. Both algorithms used fixed stepsize during training. 
	}}
	\label{fig:train_resnet18_qsgd_best_to_best_bad}
\end{figure}

During training image classification Convolutional Neural Networks, we got two results for \algname{QSGD-RR} as an improvement of \algname{QSGD}. During training only the last layer (see Fig. \ref{fig:train_resnet18_qsgd_best_to_best_bad}) there are no benefits of \algname{QSGD-RR}, but \algname{QSGD} does not behave worse.

When training the whole network (Fig. \ref{fig:train_resnet18_qsgd}), the results suggest that \gls{Q-RR} is much better than \algname{Q-SGD}. Although we do not have formal proof explaining this phenomenon, we conjecture that this can be related to the significant overparameterization occurring during the training of a large model on a relatively small dataset. That is, the model can almost perfectly fit the training data on all clients, leading to a decrease in the heterogeneity parameter. In this case, there is no need for shifts since the variance coming from compression naturally goes to zero, and the complexities of \algname{QSGD} and \gls{DIANA} match (see Table \ref{tab:comparison_of_rates}). In this situation, \gls{Q-RR} performs better than \algname{QSGD} since the compression does not spoil the convergence of \texttt{RR}. Therefore, \gls{DIANA}-type shifts are not always necessary to get improvements. Nevertheless, we conjecture that they are necessary when the datasets are larger and more complex.


\begin{figure}[t]
	\centering
	\captionsetup[sub]{font=small,labelfont={}}	
		\includegraphics[width=0.49\textwidth]{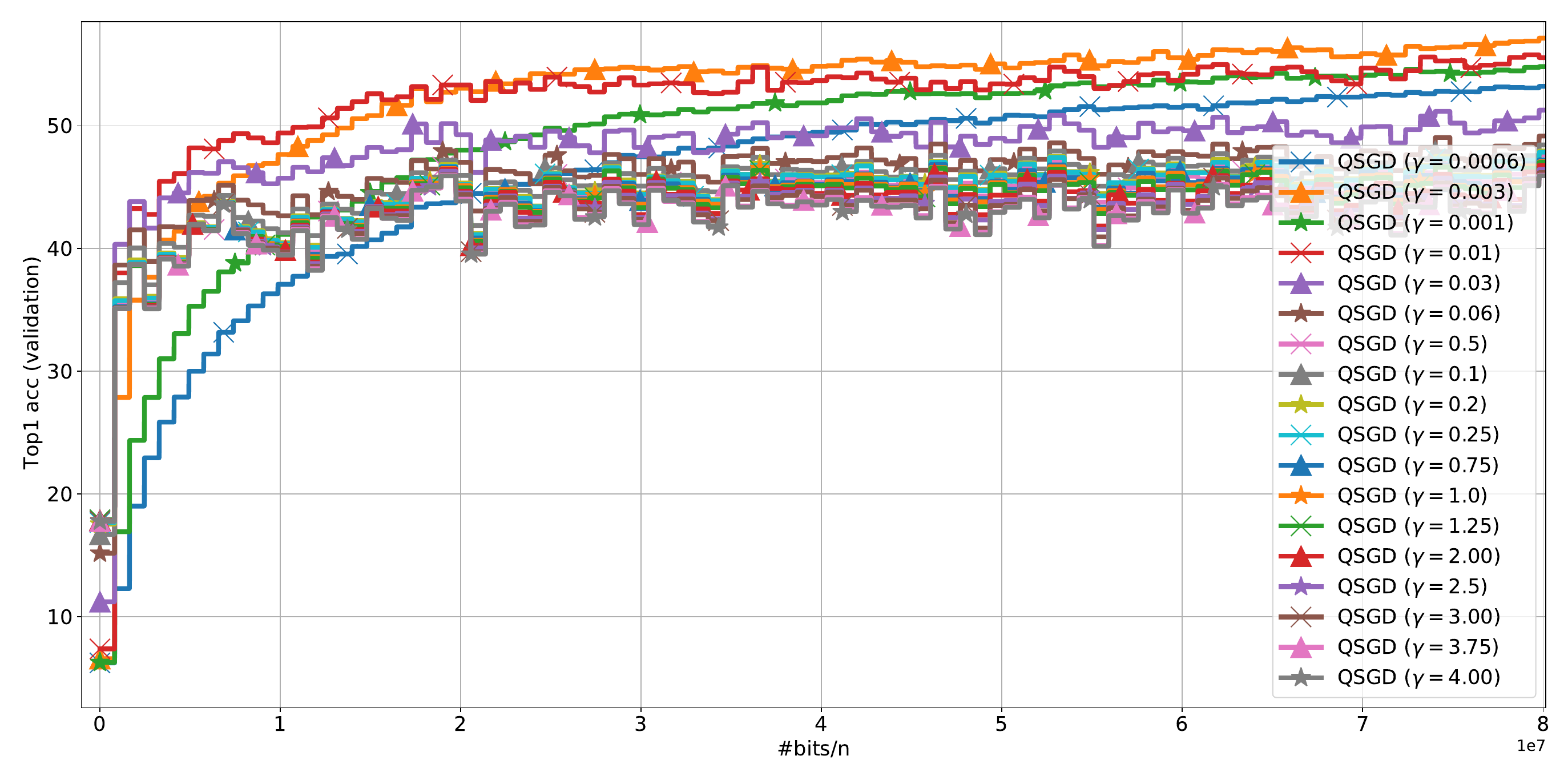} 
		\includegraphics[width=0.49\textwidth]{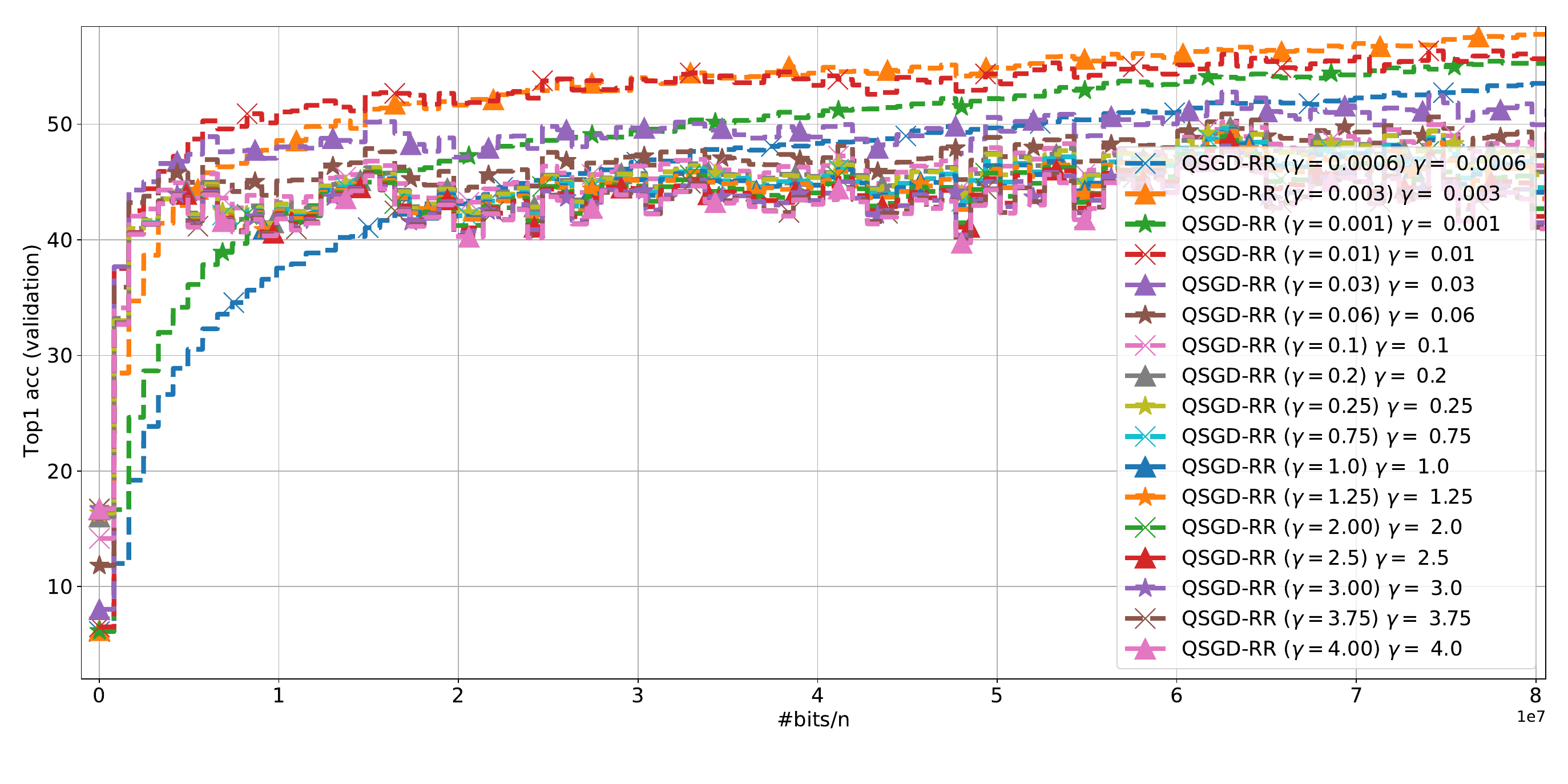} 
		\includegraphics[width=0.49\textwidth]{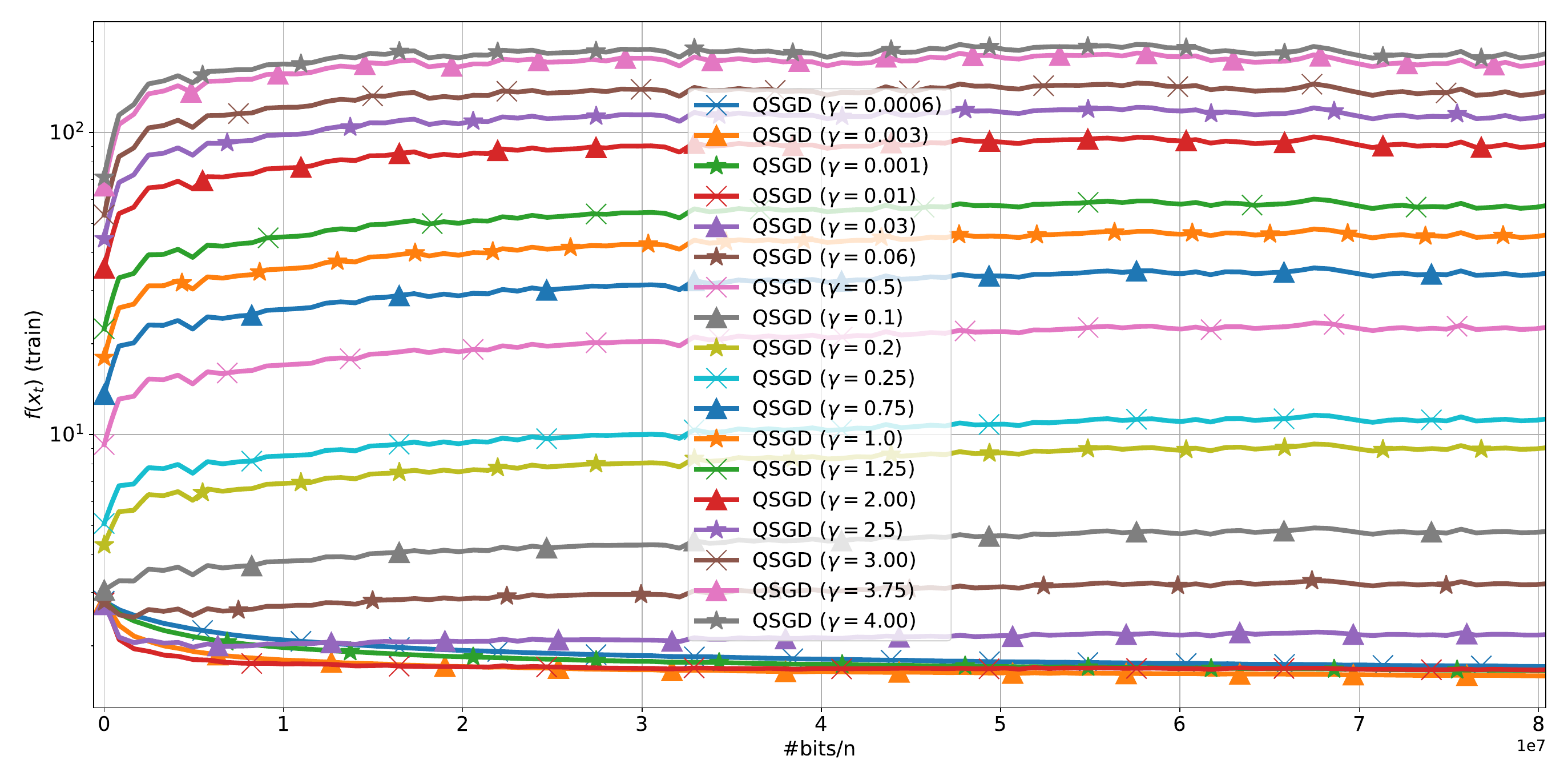} 
		\includegraphics[width=0.49\textwidth]{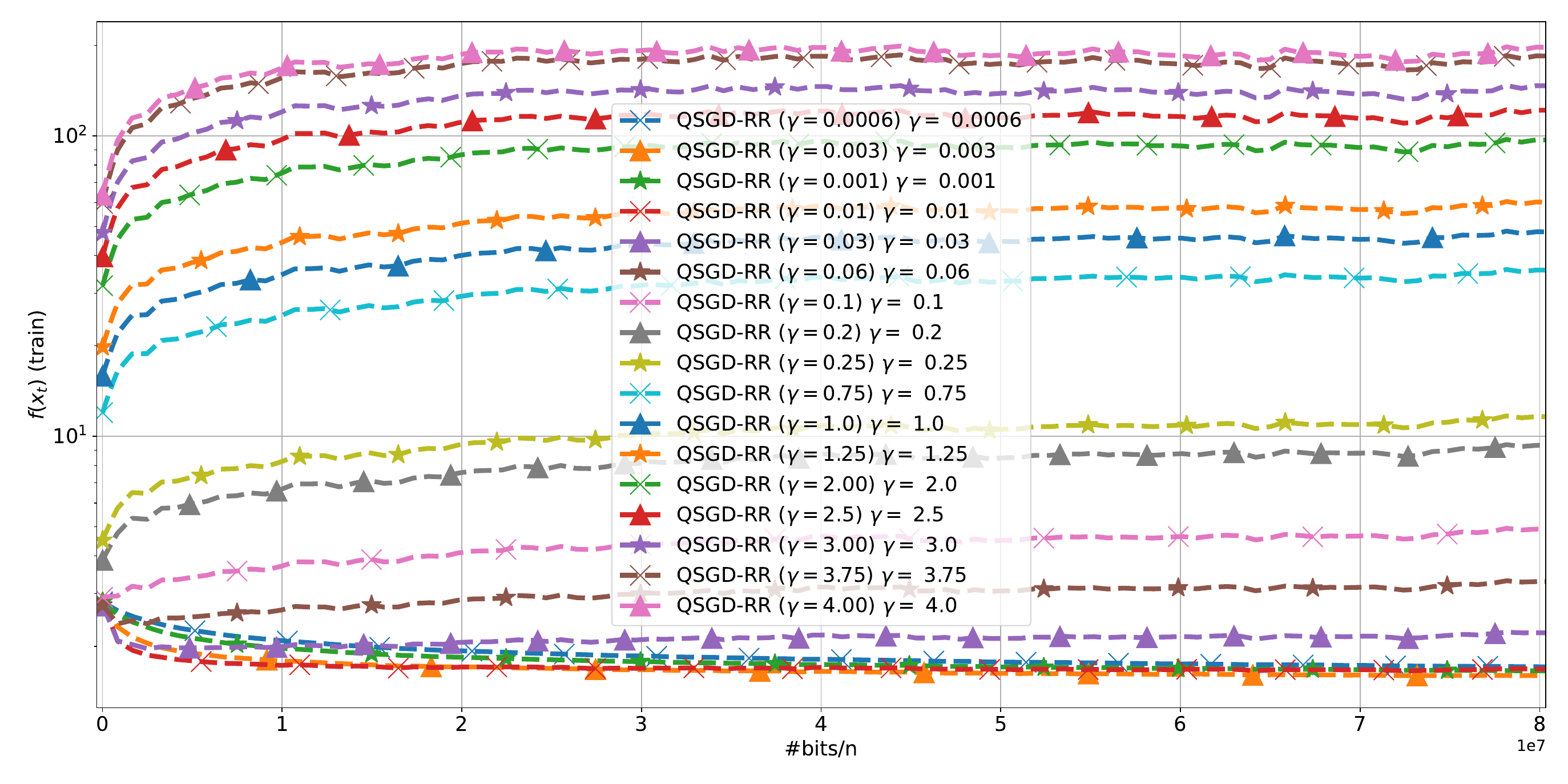} 
		\includegraphics[width=0.49\textwidth]{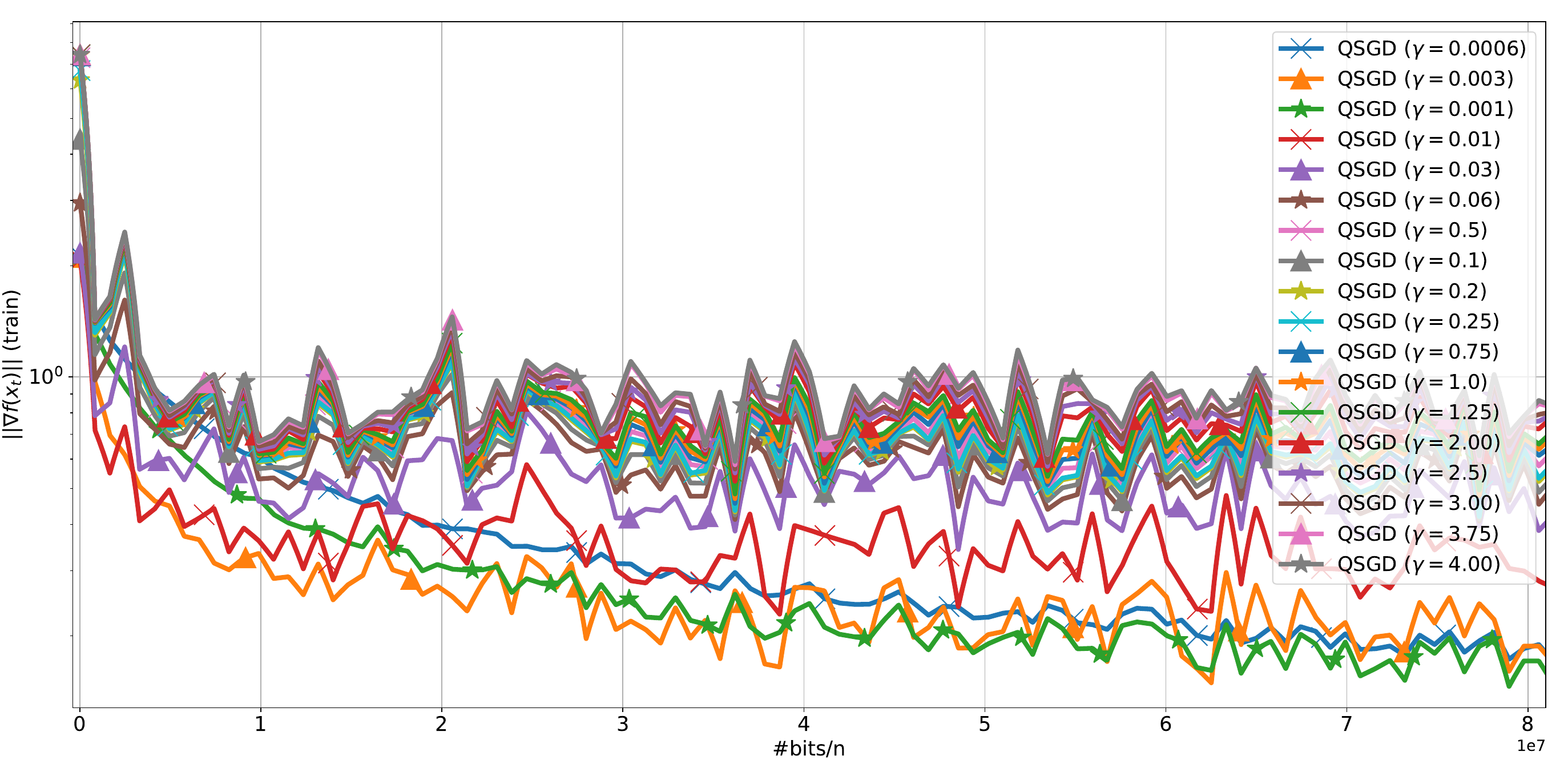} 
		\includegraphics[width=0.49\textwidth]{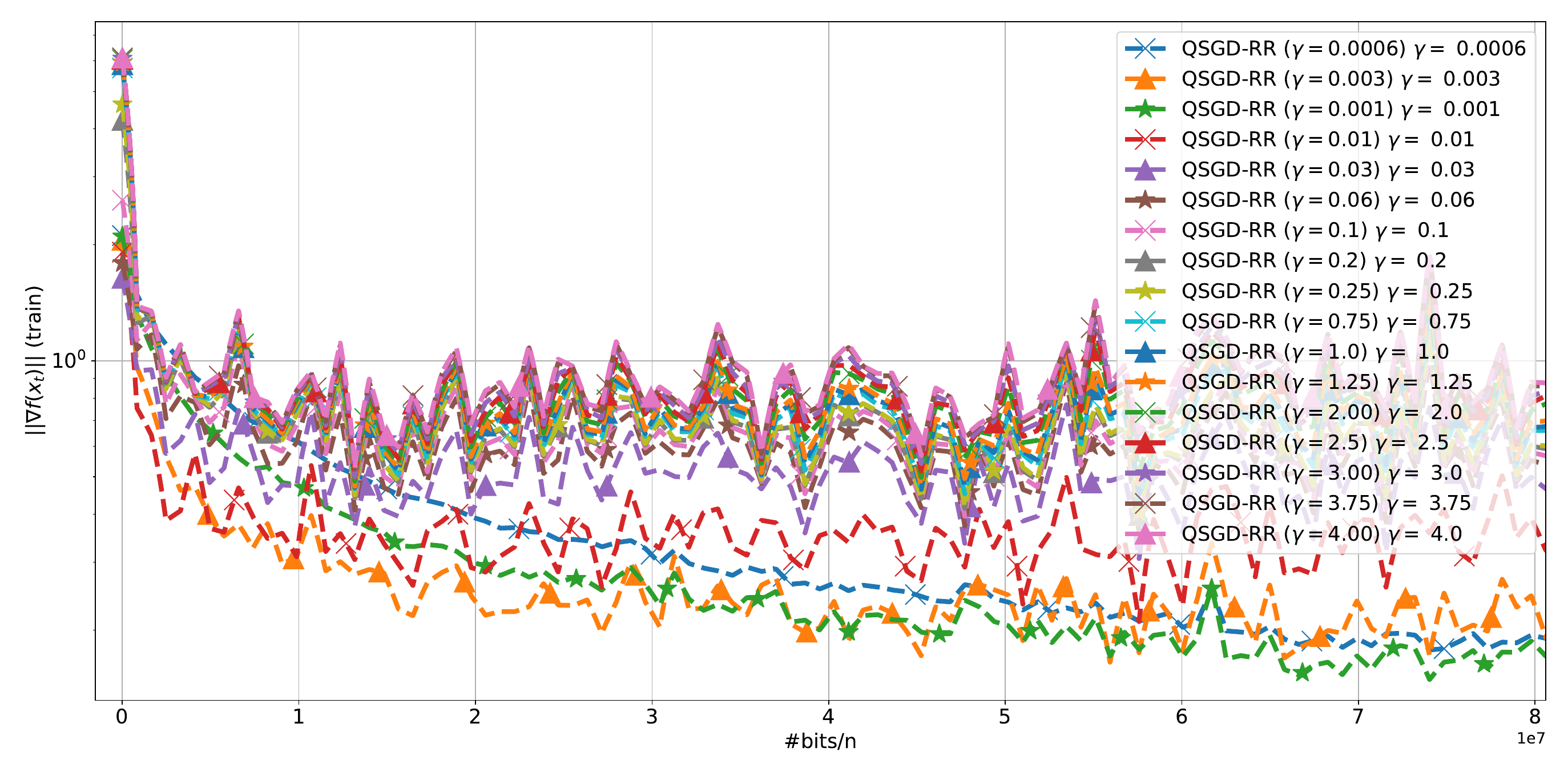} 
	\caption{{Comparison of \algname{QSGD} and \gls{Q-RR} in the training of the last linear layer of \texttt{ResNet-18} on \texttt{CIFAR-10}, with $n=10$ workers. Here (a) and (b) show Top-1 accuracy on test set, (c) and (d) -- loss function value on the train set, (e) and (f) -- norm of full gradient on the train set. Stepsizes and decay shifts have been tuned from $s_{set}$ and $\gamma_{set}$ based on minimum achievable value of loss function on the train set. During training stepsize was fixed. Batch Normalization was turned off. 
	}}
	\label{fig:train_resnet18_qsgd_all_bad}
\end{figure}

\section{Missing Proofs for Q-RR}

In the main part of the paper, we introduce Assumptions~\ref{asm:lip_max_f_m} and \ref{asm:sc_each_f_m} for the analysis of \gls{Q-RR} and \gls{DIANA-RR}. These assumptions can be refined as follows.

\begin{assumption}
	\label{asm:lip_avr_f}
	Function $f_{\pi^i} = \frac{1}{M}\sum^M_{i=1}f^{\pi_m^i}_m: \R^d \rightarrow \R$ is $\widetilde{L}$-smooth for all sets of permutations $\pi = (\pi_1,\ldots,\pi_m)$ from $[n]$ and all $i \in [n]$, i.e.,
	\begin{equation}
		\max_{i \in [n], \pi}\|\nabla f_{\pi^i}(x) - \nabla f_{\pi^i}(y)\|\leq \widetilde{L}\|x - y\| \quad \forall x, y \in \R^d.\notag
	\end{equation}
\end{assumption}

\begin{assumption}
\label{asm:sc_each_f_m_refined}
Function $f_{\pi^i} = \frac{1}{M}\sum^M_{i=1}f^{\pi_m^i}_m: \R^d \rightarrow \R$ is $\widetilde{\mu}$-strongly convex for all sets of permutations $\pi = (\pi_1,\ldots,\pi_m)$ from $[n]$ and all $i \in [n]$, i.e.,
\begin{equation}
	\min_{i \in [n], \pi}\left\{f_{\pi^i}(x) - f_{\pi^i}(y) - \la\nabla f_{\pi^i}(y), x - y\ra\right\} \geq \frac{\widetilde{\mu}}{2}\|x - y\|^2 \quad \forall x, y \in \R^d. \notag
\end{equation}
Moreover, functions $f_{1,i}, f_{2,i}, \dots, f_{M,i}: \R^d \rightarrow \R$  are convex for all $i = 1,\dots, n$.
\end{assumption}

We notice that Assumptions~\ref{asm:lip_max_f_m} and \ref{asm:sc_each_f_m} imply Assumptions~\ref{asm:lip_avr_f} and \ref{asm:sc_each_f_m_refined}. In the proofs of the results for \gls{Q-RR} and \gls{DIANA-RR}, we use Assumptions~\ref{asm:lip_avr_f} in addition to Assumptions~\ref{asm:lip_max_f_m} and we use Assumption~\ref{asm:sc_each_f_m_refined} instead of Assumption~\ref{asm:sc_each_f_m}.

\subsection{Shuffle radius clarification}\label{appendix:shuffling_radius}

Our results depend on the so-called shuffling radius proposed in \citep{mishchenko2022proximal}:
$$\sigma^2_{\text{rad}} \eqdef \max_{i}\left\{\frac{1}{\gamma^2 M}\sum^M_{m=1}\mathbb{E} D_{f_{m,\pi^i}}(x^{\star}_i, x^{\star})\right\},$$
where \(x_{i+1}^{\star} = x^{\star}_{i} -\frac{\gamma}{M}\sum^M_{m=1}\nabla f_{m,\pi^i_m}(x^{\star})\).


One can think of the shuffling radius as a counterpart to the variance term in \algname{SGD}. Both concepts measure how much the algorithm’s performance can fluctuate near the optimal solution, but the cause of these fluctuations is different: in \algname{SGD}, it is due to random sampling, and in \algname{\gls{RR}}, it is due to reshuffling. Additionally, Lemma~\ref{lem:shuffling_radius} provides bounds for the shuffling radius --- showing the maximum and minimum possible values --- based on the variance at the optimum, reinforcing the shuffling radius as a useful way to understand how \algname{\gls{RR}} behaves. This relationship helps clarify how the reshuffling process influences the algorithm's path and its efficiency in reaching an optimal point.

\subsection{Proof of Theorem \ref{th_conv_new_rr_q}}
For convenience, we restate the theorem below.
\begin{theorem}[Theorem~\ref{th_conv_new_rr_q}]
	\label{thm:advaced_conv_Q_RR}
	Let Assumptions \ref{asm:quantization_operators}, \ref{asm:lip_max_f_m}, \ref{asm:lip_avr_f}, \ref{asm:sc_each_f_m_refined} hold and $0 < \gamma \leq \frac{1}{\widetilde{L}+2\frac{\omega}{M}L_{\max}}$. Then, for all $T \geq 0$ the iterates produced by \gls{Q-RR} satisfy
	\begin{equation}
		\mathbb{E}\|x_{T}-x^{\star}\|^2\leq \left(1-\gamma\widetilde{\mu}\right)^{nT}\|x^0-x^{\star}\|^2 + \frac{2\gamma^2\sigma^2_{\text{rad}}}{\widetilde{\mu}}+\frac{2\gamma\omega}{\widetilde{\mu} M}\left(\zeta^2_{\star}+\sigma^2_{\star}\right),\notag
	\end{equation}
	where $\zeta^2_{\star} = \frac{1}{M}\sum\limits_{m=1}^M\|\nabla f_m (x^{\star})\|^2,$ and $\sigma_{\star}^2 = \frac{1}{Mn}\sum\limits_{m=1}^M\sum\limits_{i=1}^n \|\nabla f_m^i(x^\star) - \nabla f_m(x^\star)\|^2.$
\end{theorem}
\begin{proof}
	Using \(x_{i+1}^{\star} = x^{\star}_{i} -\frac{\gamma}{M}\sum^M_{m=1}\nabla f_{m,\pi^i_m}(x^{\star})\) and line 7 of Algorithm \ref{alg_new_Q_RR}, we get
	\begin{eqnarray*}
		\|x_{i+1}^t-x_{i+1}^{\star}\|^2 &=& \left\|x^t_i-x^{\star}_{i}- \gamma\frac{1}{M}\sum^M_{m=1} \left(\cQ\left(\nabla f_{m,\pi^{i}_m}(x^{t}_{i})\right) - \nabla f_{m,\pi^{i}_m}(x^{\star})\right)\right\|^2\\
		&=& \left\|x^t_i-x^{\star}_{i}\right\|^2\\
        &&- 2\gamma\left\la\frac{1}{M}\sum^M_{m=1} \left(\cQ\left(\nabla f_{m,\pi^{i}_m}(x^{t}_{i})\right) - \nabla f_{m,\pi^{i}_m}(x^{\star})\right), x^t_i-x^{\star}_{i} \right\ra\\
		&&+\gamma^2\left\|\frac{1}{M}\sum^M_{m=1} \left(\cQ\left(\nabla f_{m,\pi^{i}_m}(x^{t}_{i})\right) - \nabla f_{m,\pi^{i}_m}(x^{\star})\right)\right\|^2.
	\end{eqnarray*}
	Taking the expectation w.r.t.\ $\cQ$, we obtain 
	\begin{eqnarray*}
		\mathbb{E}_{\cQ}\left[\|x_{i+1}^t-x_{i+1}^{\star}\|^2 \right]
		&=& \left\|x^t_i-x^{\star}_{i}\right\|^2\\
        &&- 2\gamma\left\la\frac{1}{M}\sum^M_{m=1} \left(\nabla f_{m,\pi^{i}_m}(x^{t}_{i}) - \nabla f_{m,\pi^{i}_m}(x^{\star})\right), x^t_i-x^{\star}_{i} \right\ra\\
		&&+\gamma^2\mathbb{E}_{\cQ}\left[\left\|\frac{1}{M}\sum^M_{m=1} \left(\cQ\left(\nabla f_{m,\pi^{i}_m}(x^{t}_{i})\right) - \nabla f_{m,\pi^{i}_m}(x^{\star})\right)\right\|^2\right].
	\end{eqnarray*}
	In view of Assumption \ref{asm:quantization_operators} and $\mathbb{E}_{\xi}\|\xi - c\|^2 = \mathbb{E}_{\xi}\|\xi - \mathbb{E}_{\xi}\xi\|^2+ \|\mathbb{E}_{\xi}\xi - c\|^2 $, we have 
	\begin{eqnarray*}
		&&\mathbb{E}_{\cQ}\left[\|x_{i+1}^t-x_{i+1}^{\star}\|^2 \right]\\
		&=& \left\|x^t_i-x^{\star}_{i}\right\|^2\\
        &&- \frac{2\gamma}{M}\sum^M_{m=1}\left\la \nabla f_{m,\pi^{i}_m}(x^{t}_{i}) - \nabla f_{m,\pi^{i}_m}(x^{\star}), x^t_i-x^{\star}_{i} \right\ra\\
		&&+\gamma^2\mathbb{E}_{\cQ}\left[\left\|\frac{1}{M}\sum^M_{m=1} \left(\cQ\left(\nabla f_{m,\pi^{i}_m}(x^{t}_{i})\right) - \nabla f_{m,\pi^{i}_m}(x^{t}_{i})\right)\right\|^2\right]\\
		&&+ \gamma^2\left\|\frac{1}{M}\sum^M_{m=1} \left(\nabla f_{m,\pi^{i}_m}(x^{t}_{i}) - \nabla f_{m,\pi^{i}_m}(x^{\star})\right)\right\|^2\\
		&\leq& \left\|x^t_i-x^{\star}_{i}\right\|^2 - \frac{2\gamma}{M}\sum^M_{m=1}\left\la \nabla f_{m,\pi^{i}_m}(x^{t}_{i}) - \nabla f_{m,\pi^{i}_m}(x^{\star}), x^t_i-x^{\star}_{i} \right\ra\\
		&&+ \gamma^2\left\|\frac{1}{M}\sum^M_{m=1} \left(\nabla f_{m,\pi^{i}_m}(x^{t}_{i}) - \nabla f_{m,\pi^{i}_m}(x^{\star})\right)\right\|^2\\
  &&+\frac{\gamma^2\omega}{M^2}\sum^M_{m=1}\left\| \nabla f_{m,\pi^{i}_m}(x^{t}_{i})\right\|^2,
	\end{eqnarray*}
	where in the last step we apply independence of $\cQ\left(\nabla f_{m,\pi^{i}_m}(x^{t}_{i})\right)$ for $m\in [M]$. Next, we use three-point identity\footnote{For any differentiable function $f : \R^d \to \R^d$ we have: $\langle \nabla f(x) - \nabla f(y), x- z \rangle = D_{f}(z, x) + D_f(x, y) - D_f(z, y)$.} and obtain 
	\begin{eqnarray*}
		&&\mathbb{E}_{\cQ}\left[\|x_{i+1}^t-x_{i+1}^{\star}\|^2 \right]\\
		&\leq& \left\|x^t_i-x^{\star}_{i}\right\|^2\\
  &&- \frac{2\gamma}{M}\sum^M_{m=1}\left( D_{f_{m,\pi^{i}_m}}(x^{\star}_{i},x^{t}_{i}) +D_{f_{m,\pi^{i}_m}}(x^{t}_{i},x^{\star})- D_{f_{m,\pi^{i}_m}}(x^{\star}_{i},x^{\star})\right) \\
		&&+ \gamma^2\left\|\frac{1}{M}\sum^M_{m=1} \left(\nabla f_{m,\pi^{i}_m}(x^{t}_{i}) - \nabla f_{m,\pi^{i}_m}(x^{\star})\right)\right\|^2\\
  &&+\frac{\gamma^2\omega}{M^2}\sum^M_{m=1}\left\| \nabla f_{m,\pi^{i}_m}(x^{t}_{i})\right\|^2.
	\end{eqnarray*}
	Applying $\widetilde{L}$-smoothness and convexity of $\frac{1}{M}\sum_{m= 1}^m f_{m,\pi_m^i}$, $\widetilde\mu$-strong convexity of $\frac{1}{M}\sum_{m= 1}^m f_{m,\pi_m^i}$, and $L_{\max}$-smoothness and convexity of $f_m^i$, we get
	\begin{eqnarray*}
		\mathbb{E}_{\cQ}\left[\|x_{i+1}^t-x_{i+1}^{\star}\|^2 \right]
		&\leq& \left(1-\gamma\widetilde\mu\right)\left\|x^t_i-x^{\star}_{i}\right\|^2\\
        &&- 2\gamma \left(1 -\widetilde{L}\gamma\right)\frac{1}{M}\sum^M_{m=1}D_{f_{m,\pi^{i}_m}}(x^{t}_{i},x^{\star}) \\
		&&+ 2\gamma\frac{1}{M}\sum^M_{m=1}D_{f_{m,\pi^{i}_m}}(x^{\star}_{i},x^{\star}) +\frac{\gamma^2\omega}{M^2}\sum^M_{m=1}\left\| \nabla f_{m,\pi^{i}_m}(x^{t}_{i})\right\|^2\\
		&\leq& \left(1-\gamma\widetilde\mu\right)\left\|x^t_i-x^{\star}_{i}\right\|^2\\
        &&- 2\gamma \left(1 -
		\widetilde{L}\gamma\right)\frac{1}{M}\sum^M_{m=1}D_{f_{m,\pi^{i}_m}}(x^{t}_{i},x^{\star}) \\
		&&+ 2\gamma\frac{1}{M}\sum^M_{m=1}D_{f_{m,\pi^{i}_m}}(x^{\star}_{i},x^{\star}) + \frac{2\gamma^2\omega}{M^2}\sum^M_{m=1}\left\| \nabla f_{m,\pi^{i}_m}(x^{\star})\right\|^2\\
		&&+ \frac{2\gamma^2\omega}{M^2}\sum^M_{m=1}\left\| \nabla f_{m,\pi^{i}_m}(x^{t}_{i}) - \nabla f_{m,\pi^{i}_m}(x^{\star})\right\|^2. 
  \end{eqnarray*}
  So, we get
  \begin{eqnarray*}
			\mathbb{E}_{\cQ}\left[\|x_{i+1}^t-x_{i+1}^{\star}\|^2 \right]	&\leq& \left(1-\gamma\widetilde\mu\right)\left\|x^t_i-x^{\star}_{i}\right\|^2  + \frac{2\gamma^2\omega}{M^2}\sum^M_{m=1}\left\| \nabla f_{m,\pi^{i}_m}(x^{\star})\right\|^2\\
  &&+ \frac{2\gamma}{M}\sum^M_{m=1}D_{f_{m,\pi^{i}_m}}(x^{\star}_{i},x^{\star})\\
		&&- 2\gamma \left(1 -\gamma\left(\widetilde{L} +\frac{2\omega L_{\max}}{M}\right)\right)\frac{1}{M}\sum^M_{m=1}D_{f_{m,\pi^{i}_m}}(x^{t}_{i},x^{\star}). 
	\end{eqnarray*}
	Taking the  full  expectation and using a definition of shuffle radius, $0<\gamma \leq \frac{1}{\left(\widetilde{L}+2\frac{\omega}{M}L_{\max}\right)}$, and $D_{f_{m,\pi^{i}_m}}(x^{t}_{i},x^{\star}) \geq 0$, we obtain
	\begin{eqnarray*}
		\mathbb{E}\left[\|x_{i+1}^t-x_{i+1}^{\star}\|^2 \right]
		&\leq& \left(1-\gamma\widetilde\mu\right)\mathbb{E}\left[\left\|x^t_i-x^{\star}_{i}\right\|^2\right] + 2\gamma^3\sigma^2_{\text{rad}}\\
        &&+ \frac{2\gamma^2\omega}{M^2}\sum^M_{m=1}\mathbb{E}\left[\left\| \nabla f_{m,\pi^{i}_m}(x^{\star})\right\|^2\right]\\
		&=& \left(1-\gamma\widetilde\mu\right)\mathbb{E}\left[\left\|x^t_i-x^{\star}_{i}\right\|^2\right] + 2\gamma^3 \sigma^2_{\text{rad}}\\
        &&+ \frac{2\gamma^2\omega}{M^2n}\sum^M_{m=1}\sum^n_{j=1}\left\| \nabla f^{j}_m(x^{\star})\right\|^2\\
		&\leq& \left(1-\gamma\widetilde\mu\right)\mathbb{E}\left[\left\|x^t_i-x^{\star}_{i}\right\|^2\right] + 2\gamma^3 \sigma^2_{\text{rad}}\\
        &&+ \frac{2\gamma^2\omega}{M}\left(\zeta^2_{\star}+\sigma^2_{\star}\right).
	\end{eqnarray*}
	Unrolling the recurrence in $i$, we derive 
	\begin{eqnarray*}
		\mathbb{E}\left[\|x^{t+1}-x^{\star}\|^2 \right]
		&\leq& \left(1-\gamma\widetilde\mu\right)^n\mathbb{E}\left[\left\|x^t-x^{\star}\right\|^2\right] + 2\gamma^3 \sigma^2_{\text{rad}}\sum^{n-1}_{j=0}(1-\gamma\widetilde\mu)^j\\
		&&+ \frac{2\gamma^2\omega}{M}\left(\zeta^2_{\star}+\sigma^2_{\star}\right)\sum^{n-1}_{j=0}(1-\gamma\widetilde\mu)^j.
	\end{eqnarray*}
	Unrolling the recurrence in $t$, we derive 
	\begin{eqnarray*}
		\mathbb{E}\left[\|x_{T}-x^{\star}\|^2 \right]
		&\leq& \left(1-\gamma\widetilde\mu\right)^{nT}\left\|x^0-x^{\star}\right\|^2 + 2\gamma^3\sigma^2_{\text{rad}}\sum^{T-1}_{t=0}(1-\gamma\widetilde\mu)^{nt}\sum^{n-1}_{j=0}(1-\gamma\widetilde\mu)^j \\
		&&+ \frac{2\gamma^2\omega}{M}\left(\zeta^2_{\star}+\sigma^2_{\star}\right)\sum^{nT-1}_{j=0}(1-\gamma\widetilde\mu)^{nt}\sum^{n-1}_{j=0}(1-\gamma\widetilde\mu)^j.
	\end{eqnarray*}
	Since $\sum^{nT-1}_{j=0}(1-\gamma\widetilde\mu)^{j} \leq \frac{1}{\gamma\widetilde\mu}$, we get the result.
\end{proof}

\begin{corollary}
\label{cor_convergence_new_Q_RR}
Let the assumptions of Theorem~\ref{thm:advaced_conv_Q_RR} hold and
\begin{equation}
	\label{new_gamma_new_Q_RR}
	\gamma = \min\left\{\frac{1}{\widetilde{L}+2\frac{\omega}{M}L_{\max}}, \sqrt{\frac{\varepsilon\widetilde{\mu}}{6\sigma^2_{\text{rad}}}}, \frac{\varepsilon\widetilde{\mu}M}{6\omega\left(\zeta^2_{\star}+\sigma^2_{\star}\right)}\right\}.
\end{equation}
Then, \gls{Q-RR} finds a solution with accuracy $\varepsilon>0$ after the following number of communication rounds: 
\begin{equation}
	\widetilde{\cO}\left(\frac{\widetilde{L}}{\widetilde{\mu}}+\frac{\omega}{M}\frac{L_{\max}}{\widetilde{\mu}} + \frac{\omega}{M}\frac{\zeta^2_{\star}+\sigma^2_{\star}}{\varepsilon\widetilde{\mu}^2} + \frac{\sigma_{\text{rad}}}{\sqrt{\varepsilon\widetilde{\mu}^3}}\right).\notag
\end{equation}
\end{corollary}
\begin{proof}
Theorem~\ref{thm:advaced_conv_Q_RR} implies
\begin{equation}
	\mathbb{E}\|x_{T}-x^{\star}\|^2\leq \left(1-\gamma\widetilde{\mu}\right)^{nT}\|x^0-x^{\star}\|^2 + \frac{2\gamma^2\sigma^2_{\text{rad}}}{\widetilde{\mu}}+\frac{2\gamma\omega}{\widetilde{\mu} M}\left(\zeta^2_{\star}+\sigma^2_{\star}\right).
\end{equation}
To estimate the number of communication rounds required to find a solution with accuracy $\varepsilon >0$, we need to upper-bound each term from the right-hand side by $\nicefrac{\varepsilon}{3}$. Thus, we get additional conditions on $\gamma$:
\begin{equation*}
	\frac{2\gamma^2\sigma^2_{\text{rad}}}{\widetilde{\mu}} < \frac{\varepsilon}{3},\quad\frac{2\gamma\omega}{\widetilde{\mu} M}\left(\zeta^2_{\star}+\sigma^2_{\star}\right)<\frac{\varepsilon}{3}
\end{equation*}
and also the upper bound on the number of communication rounds $nT$
\begin{equation*}
	nT = \widetilde{\cO}\left(\frac{1}{\gamma\widetilde{\mu}}\right).
\end{equation*}
Substituting \eqref{new_gamma_new_Q_RR}, we get a final result.
\end{proof}

\subsection{Non-Strongly convex summands}
In this section, we provide the analysis of \gls{Q-RR} without using Assumptions~\ref{asm:sc_each_f_m}, \ref{asm:sc_each_f_m_refined}. Before we move on to the proofs, we would like to emphasize that 
\begin{equation*}
	x_{i+1}^t = x^{i}_t - \gamma \frac{1}{M}\sum^{M}_{m=1}\cQ\left(\nabla f_{m,\pi^i_m}(x^t_i)\right).
\end{equation*}
Then we have 
\begin{equation*}
	x^{t+1} = x^t - \gamma \sum^{n-1}_{i=0}\frac{1}{M}\sum^{M}_{m=1}\cQ\left(\nabla f_{m,\pi^i_m}(x^t_i)\right) = x^t - \tau \frac{1}{Mn}\sum^{n-1}_{i=0} \sum^{M}_{m=1}\cQ\left(\nabla f_{m,\pi^i_m}(x^t_i)\right),
\end{equation*}
where $\tau = \gamma n$. For convenience, we denote
\begin{equation*}
	g^t = \frac{1}{Mn}\sum^{n-1}_{i=0} \sum^{M}_{m=1}\cQ\left(\nabla f_{m,\pi^i_m}(x^t_i)\right) 
\end{equation*}
allowing to write the update rule as $x^{t+1} = x^t - \tau g^t$.

\begin{lemma}[Lemma 1 from ~\citep{malinovsky2023server}]
	\label{Sampling_without_replacement}
	For any $k \in [n]$, let $\xi_{\pi_1}, \dots,\xi_{\pi_k}$ be sampled uniformly without replacement from a set of vectors $\{\xi_1, \dots, \xi_n\}$ and $\bar{\xi}_{\pi}$ be their average. Then, it holds
	\begin{equation}
		\mathbb{E} \bar{\xi}_{\pi} = \bar{\xi},\quad \mathbb{E}\left[\|\bar{\xi}_{\pi} - \bar{\xi}\|^2\right] = \frac{n-k}{k(n-1)}\sigma^2, 
	\end{equation}
	where $\bar{\xi} = \frac{1}{n}\sum^n_{i=1}\xi_i$, $\bar{\xi}_{\pi} = \frac{1}{k}\sum^k_{i=1}\xi_{\pi_i}$, $\sigma^2 = \frac{1}{n}\sum^n_{i=1}\|\xi_i - \bar{\xi}\|^2$
\end{lemma}

\begin{lemma}
	\label{lem_inner_product_cvx_case} 
	Under Assumptions \ref{asm:quantization_operators}, \ref{asm:sc_general_f},  \ref{asm:lip_max_f_m}, \ref{asm:lip_avr_f}, the following inequality holds
	\begin{equation*}
		\mathbb{E}_{\cQ}\left[-2\tau\la g^t, x^t-x^{\star}\ra\right] \leq -\frac{\tau\mu}{2}\|x^t-x^{\star}\|^2-\tau(f(x^t)- f(x^{\star}))+\frac{\tau\widetilde{L}}{n}\sum^{n-1}_{i=0}\|x^t_i-x^t\|^2.
	\end{equation*}
\end{lemma}
\begin{proof}Using that $\mathbb{E}_{\cQ}\left[g^t\right] = \frac{1}{Mn}\sum^{n-1}_{i=0} \sum^{M}_{m=1}\nabla f_{m,\pi^i_m}(x^t_i) $ and definition of $h^{\star}$, we get
		\begin{eqnarray*}
			-2\tau\mathbb{E}_{\cQ}\left[\left\la g^t, x^t - x^{\star}\right\ra\right] 
			&=& -\frac{1}{Mn}\sum^{n-1}_{i=0} \sum^{M}_{m=1}\left\la \nabla f_{m,\pi^i_m}(x^t_i), x^t - x^{\star}\right\ra\\
			&=& -\frac{1}{Mn}\sum^M_{m=1}\sum^{n-1}_{i=0}\left\la \nabla f_{m,\pi^i_m}(x^{t}_{i}) - \nabla f_{m,\pi^i_m}(x^{\star}), x^t - x^{\star}\right\ra.
		\end{eqnarray*}
		
		Using three-point identity, we obtain 
		\begin{eqnarray*}
			&&-2\tau\mathbb{E}_{\cQ}\left[\left\la g^t , x^t - x^{\star}\right\ra\right] \\
			&=& -\frac{2\tau}{Mn}\sum^M_{m=1}\sum^{n-1}_{i=0}\left( D_{f_{m,\pi^i_m}}(x^t, x^{\star})+ D_{f_{m,\pi^i_m}}( x^{\star}, x^{t}_{i}) - D_{f_{m,\pi^i_m}}(x^t, x^{t}_{i})\right)\\
			&=& - 2\tau D_{f}(x^t, x^{\star}) - \frac{2\tau}{n}\sum^{n-1}_{i=0} D_{f_{\pi^i}}(x^{\star}, x^{t}_{i})+ \frac{2\tau}{n}\sum^{n-1}_{i=0} D_{f_{\pi^i}}(x^{t}, x^{t}_{i})\\
			&\leq& - 2\tau D_{f}(x^t, x^{\star}) + \frac{\tau\widetilde{L}}{n}\sum^{n-1}_{i=0} \| x^{t}_{i} - x^{t} \|^2,
		\end{eqnarray*}
		where in the last inequality we apply $\widetilde{L}$-smoothness  and convexity of each function $f_{\pi^i}$. Finally, using $\mu$-strong convexity of $f$, we finish the proof of the lemma.

\end{proof}

\begin{lemma}
	\label{lem_norm_grad_cvx_case}
	Under Assumptions \ref{asm:quantization_operators}, \ref{asm:sc_general_f},  \ref{asm:lip_max_f_m}, \ref{asm:lip_avr_f}, the following inequality holds
	\begin{eqnarray*}
		\mathbb{E}_{\cQ}\left[\|g^t\|^2\right] 
		&\leq& 2\widetilde{L}\left(\widetilde{L} + \frac{\omega}{Mn}L_{\max}\right)\frac{1}{n}\sum^{n-1}_{i=0}\mathbb{E}\left[\|x^t_i-x^t\|^2\right]+ \frac{4\omega}{Mn}\left(\zeta^2_{\star}+\sigma^2_{\star}\right)\\
		&&+8\left(\widetilde{L}+\frac{\omega}{Mn}L_{\max}\right)(f(x^t)-f(x^{\star})).
	\end{eqnarray*}
\end{lemma}
\begin{proof}
	Taking the expectation w.r.t.\ $\cQ$ and using variance decomposition identity $\mathbb{E}\left[\|\xi\|^2\right] = \mathbb{E}\left[\|\xi-\mathbb{E}\left[\xi\right]\|^2\right] +\|\mathbb{E}\xi\|^2$, we get 
	\begin{eqnarray*}
		\mathbb{E}_{\cQ}\left[\|g^t\|^2\right]
		&=& \mathbb{E}_{\cQ}\left[\left\|\frac{1}{Mn}\sum^{n-1}_{i=0} \sum^{M}_{m=1}\cQ\left(\nabla f_{m,\pi^i_m}(x^t_i)\right)\right\|^2\right]\\
		&=& \mathbb{E}_{\cQ}\left[\left\|\frac{1}{Mn}\sum^{n-1}_{i=0} \sum^{M}_{m=1}\left(\cQ\left(\nabla f_{m,\pi^i_m}(x^t_i)\right) - \nabla f_{m,\pi^i_m}(x^t_i)\right)\right\|^2\right]\\
		&&+ \left\|\frac{1}{Mn}\sum^{n-1}_{i=0} \sum^{M}_{m=1}\nabla f_{m,\pi^i_m}(x^t_i)\right\|^2.
	\end{eqnarray*}
	Next, Assumption \ref{asm:quantization_operators} and conditional independence of $\cQ\left(\nabla f_{m,\pi^i_m}(x^t_i)\right)$ for $m = 1,\ldots, M, i = 0,\ldots, n-1$ imply
	\begin{eqnarray*}
		\mathbb{E}_{\cQ}\left[\|g^t\|^2\right] &=& \frac{1}{M^2n^2}\sum^{n-1}_{i=0} \sum^{M}_{m=1}\mathbb{E}_{\cQ}\left[\left\|\cQ\left(\nabla f_{m,\pi^i_m}(x^t_i)\right) - \nabla f_{m,\pi^i_m}(x^t_i)\right\|^2\right]\\
		&&+ \left\|\frac{1}{Mn}\sum^{n-1}_{i=0} \sum^{M}_{m=1}\nabla f_{m,\pi^i_m}(x^t_i)\right\|^2\\
		&\leq& \frac{\omega}{M^2n^2} \sum^{n-1}_{i=0} \sum^{M}_{m=1}\left\|\nabla f_{m,\pi^i_m}(x^t_i)\right\|^2 + \left\|\frac{1}{Mn}\sum^{n-1}_{i=0} \sum^{M}_{m=1}\nabla f_{m,\pi^i_m}(x^t_i)\right\|^2\\
		&\leq& \frac{2\omega}{M^2n^2} \sum^{n-1}_{i=0} \sum^{M}_{m=1}\left\|\nabla f_{m,\pi^i_m}(x^t_i) - \nabla f_{m,\pi^i_m}(x^t)\right\|^2\\
        &&+ \frac{2\omega}{M^2n^2} \sum^{n-1}_{i=0} \sum^{M}_{m=1}\left\|\nabla f_{m,\pi^i_m}(x^t)\right\|^2\\
		&&+ 2\left\|\frac{1}{Mn}\sum^{n-1}_{i=0} \sum^{M}_{m=1}\left(\nabla f_{m,\pi^i_m}(x^t_i) - \nabla f_{m,\pi^i_m}(x^t)\right)\right\|^2\\
  &&+ 2\left\|\frac{1}{Mn}\sum^{n-1}_{i=0} \sum^{M}_{m=1}\nabla f_{m,\pi^i_m}(x^t)\right\|^2.
	\end{eqnarray*}
	Using $L_{\max}$-smoothness and convexity of $f_{m, i}$ and $\widetilde{L}$-smoothness and convexity of $f_{\pi^i} = \frac{1}{M}\sum_{m=1}^M f_{m,\pi_m^i}$, we derive 
	\begin{eqnarray*}
		\mathbb{E}_{\cQ}\left[\|g^t\|^2\right]&\leq& \frac{4\omega}{M^2n^2}L_{\max} \sum^{n-1}_{i=0} \sum^{M}_{m=1}D_{f_{m,\pi^i_m}}(x^t_i, x^t)\\
        &&+ \frac{2\omega}{M^2n^2} \sum^{n-1}_{i=0} \sum^{M}_{m=1}\left\|\nabla f_{m,\pi^i_m}(x^t)\right\|^2\\
		&&+ 4\widetilde{L}\frac{1}{n}\sum^{n-1}_{i=0}D_{f_{\pi^i}}(x^t_i,x^t) + 2\left\|\nabla f(x^t)\right\|^2\\
		&\leq&4\left(\widetilde{L} + \frac{\omega}{Mn}L_{\max}\right) \frac{1}{n}\sum^{n-1}_{i=0} D_{f_{\pi^i}}(x^t_i, x^t)\\
        &&+ \frac{4\omega}{M^2n^2} \sum^{n-1}_{i=0} \sum^{M}_{m=1}\left\|\nabla f_{m,\pi^i_m}(x^{\star})\right\|^2  \\
		&& + \frac{4\omega}{M^2n^2} \sum^{n-1}_{i=0} \sum^{M}_{m=1}\left\|\nabla f_{m,\pi^i_m}(x^t) - \nabla f_{m,\pi^i_m}(x^{\star})\right\|^2\\
        &&+ 2\left\|\nabla f(x^t) - \nabla f(x^{\star})\right\|^2\\
		&\leq& 2\widetilde{L}\left(\widetilde{L} + \frac{\omega}{Mn}L_{\max}\right)\frac{1}{n}\sum^{n-1}_{i=0}\left\| x^t_i -  x^t\right\|^2\\
        &&+ \frac{4\omega}{M^2n^2} \sum^{n-1}_{i=0} \sum^{M}_{m=1}\left\|\nabla f_{m,\pi^i_m}(x^{\star})\right\|^2  \\
		&& + \frac{8\omega}{M^2n^2}L_{\max} \sum^{n-1}_{i=0} \sum^{M}_{m=1}D_{f_{m,\pi^i_m}}(x^t, x^{\star}) + 4\widetilde{L}\left( f(x^t) -  f(x^{\star})\right).
	\end{eqnarray*}
	Taking the full expectation, we obtain
	\begin{eqnarray*}
		\mathbb{E}\left[\|g^t\|^2\right]
		&\leq& 2\widetilde{L}\left(\widetilde{L} + \frac{\omega}{Mn}L_{\max}\right)\frac{1}{n}\sum^{n-1}_{i=0}\mathbb{E}\left[\left\| x^t_i -  x^t\right\|^2\right]\\
        &&+ \frac{4\omega}{M^2n^2} \sum^{n-1}_{i=0} \sum^{M}_{m=1}\mathbb{E}\left[\left\|\nabla f_{m,\pi^i_m}(x^{\star})\right\|^2\right]  \\
		&& +\left(4\widetilde{L}+ \frac{8\omega}{Mn}L_{\max}\right)\mathbb{E}\left[f(x^t) -  f(x^{\star})\right]\\
		&=& 2\widetilde{L}\left(\widetilde{L} + \frac{\omega}{Mn}L_{\max}\right)\frac{1}{n}\sum^{n-1}_{i=0}\mathbb{E}\left[\left\| x^t_i -  x^t\right\|^2\right]+ \frac{4\omega}{Mn} \left(\zeta_{\star}^2 + \sigma_{\star}^2\right) \\
		&& +\left(4\widetilde{L}+ \frac{8\omega}{Mn}L_{\max}\right)\mathbb{E}\left[f(x^t) -  f(x^{\star})\right].
	\end{eqnarray*}
\end{proof}

\begin{lemma}
	\label{lem_dinst_cvx_case} 
	Let Assumptions \ref{asm:quantization_operators}, \ref{asm:sc_general_f},  \ref{asm:lip_max_f_m}, \ref{asm:lip_avr_f} hold and $\tau \leq \frac{1}{2\sqrt{\widetilde{L}\left(\widetilde{L}+\frac{\omega}{Mn}L_{\max}\right)}}$. Then, the following inequality holds
	\begin{eqnarray*}
		\frac{1}{n}\sum^{n-1}_{i=0}\mathbb{E}\left[\|x^t_i-x^t\|^2\right] 
		&\leq& 24\tau^2\left(\widetilde{L}+\frac{\omega}{Mn}L_{\max}\right)\mathbb{E}\left[f(x^t)-f(x^{\star})\right] \\
		&&+ 8\tau^2\frac{\omega}{Mn}\left(\zeta^2_{\star}+\sigma^2_{\star}\right) +8\tau^2\frac{\sigma^2_{\star,n}}{n},
	\end{eqnarray*}
	where $\sigma^2_{\star,n} = \frac{1}{n}\sum^n_{i=1}\|\nabla f_i(x^{\star})\|^2$, $f_i(x) = \frac{1}{M}\sum_{m=1}^M f_m^i(x)$, $i \in [n]$.
\end{lemma}
\begin{proof}
	Since $x^t_i= x^t -\frac{\tau}{Mn}\sum^M_{m=1}\sum^{i-1}_{j=0}\cQ\left(\nabla f_{m,\pi^{j}_m}(x^{t}_{j})\right)$, we have 
	\begin{eqnarray*}
		\mathbb{E}_{\cQ}\left[\|x^t_i -x^t\|^2\right]
		&=& \tau^2\mathbb{E}_{\cQ}\left[\left\|\frac{1}{Mn}\sum^M_{m=1}\sum^{i-1}_{j=0}\cQ\left(\nabla f_{m,\pi^{j}_m}(x^{t}_{j})\right)\right\|^2\right]\\
		&=&\tau^2\mathbb{E}_{\cQ}\left[\left\|\frac{1}{Mn}\sum^M_{m=1}\sum^{i-1}_{j=0}\left(\cQ\left(\nabla f_{m,\pi^{j}_m}(x^{t}_{j})\right) - \nabla f_{m,\pi^{j}_m}(x^{t}_{j})\right)\right\|^2\right]\\
		&&+ \tau^2\left\|\frac{1}{Mn}\sum^M_{m=1}\sum^{i-1}_{j=0}\nabla f_{m,\pi^{j}_m}(x^{t}_{j})\right\|^2\\
		&\leq& \frac{\tau^2}{M^2n^2}\sum^M_{m=1}\sum^{i-1}_{j=0}\mathbb{E}_{\cQ}\left[\left\|\cQ\left(\nabla f_{m,\pi^{j}_m}(x^{t}_{j})\right) - \nabla f_{m,\pi^{j}_m}(x^{t}_{j})\right\|^2\right]\\
		&&+ \tau^2\left\|\frac{1}{Mn}\sum^M_{m=1}\sum^{i-1}_{j=0}\nabla f_{m,\pi^{j}_m}(x^{t}_{j})\right\|^2.
	\end{eqnarray*}
	Using Assumption \ref{asm:quantization_operators}, $\widetilde{L}$-smoothness  and convexity of $f_{\pi^i} = \frac{1}{M}\sum_{m=1}^M f_{m,\pi_m^i}$ and $L_{\max}$-smoothness  and convexity of $f^{i}_m$, we obtain 
	\begin{eqnarray}
		\mathbb{E}_{\cQ}\left[\|x^t_i -x^t\|^2\right]
		&\leq& \frac{\tau^2\omega}{M^2n^2}\sum^M_{m=1}\sum^{i-1}_{j=0}\left\|\nabla f_{m,\pi^{j}_m}(x^{t}_{j})\right\|^2\notag\\
        &&+ \tau^2\left\|\frac{1}{Mn}\sum^M_{m=1}\sum^{i-1}_{j=0}\nabla f_{m,\pi^{j}_m}(x^{t}_{j})\right\|^2\notag\\
		&\leq& \frac{2\tau^2\omega}{M^2n^2}\sum^M_{m=1}\sum^{i-1}_{j=0}\left\|\nabla f_{m,\pi^{j}_m}(x^{t}_{j}) - \nabla f_{m,\pi^{j}_m}(x^t)\right\|^2\notag\\
        &&+ 2\tau^2\left\|\frac{1}{n}\sum^{i-1}_{j=0}\nabla f_{\pi^j}(x^t)\right\|^2\notag\\
		&& + 2\tau^2\left\|\frac{1}{n}\sum^{i-1}_{j=0}\left(\nabla f_{\pi^j}(x^{t}_{j})-\nabla f_{\pi^j}(x^t)\right)\right\|^2\notag\\
  &&+ \frac{2\tau^2\omega}{M^2n^2}\sum^M_{m=1}\sum^{i-1}_{j=0}\left\|\nabla f_{m,\pi^{j}_m}(x^t)\right\|^2\notag \\
		&\leq& \frac{4\tau^2\omega}{M^2n^2}\sum^M_{m=1}\sum^{n-1}_{j=0}L_{\max}D_{f^{\pi_m^j}_m}(x^{t}_{j},x^t)\notag\\
        &&+ 2\tau^2\left\|\frac{1}{n}\sum^{i-1}_{j=0}\nabla f_{\pi^j}(x^t)\right\|^2\notag \\
		&&+ 4\widetilde{L}\tau^2\frac{1}{n}\sum^{n-1}_{j=0}D_{f_{\pi^j}}(x^{t}_{j}, x^t)\notag\\
        &&+ \frac{2\tau^2\omega}{M^2n^2}\sum^M_{m=1}\sum^{n-1}_{j=0}\left\|\nabla f_{m,\pi^{j}_m}(x^t)\right\|^2\notag\\
		&=& 4\tau^2\left(\widetilde{L} + \frac{\omega}{Mn}L_{\max}\right)\frac{1}{n}\sum^{n-1}_{j=0}D_{f_{\pi^j}}(x^{t}_{j}, x^t)\notag\\
		&&+ 2\tau^2\left\|\frac{1}{n}\sum^{i-1}_{j=0}\nabla f_{\pi^j}(x^t)\right\|^2\notag\\
        &&+ \frac{2\tau^2\omega}{M^2n^2}\sum^M_{m=1}\sum^{n-1}_{j=0}\left\|\nabla f_{m,\pi^{j}_m}(x^t)\right\|^2. \label{eq:cjshjdjsdbhs}
	\end{eqnarray}
	
	Next, we need to estimate the second term from the previous inequality. Taking the full expectation and using Lemma \ref{Sampling_without_replacement}  and using new notation $\sigma_t^2 = \frac{1}{n}\sum_{j=1}^n \mathbb{E}[\|\nabla f_j (x^t) - \nabla f(x^t)\|^2]$, we get 
	\begin{eqnarray}
		\mathbb{E}\left[\left\|\frac{1}{n}\sum^{i-1}_{j=0}\nabla f_{\pi^j}(x^t)\right\|^2 \right]&=& \frac{i^2}{n^2}\mathbb{E}\left[\|\nabla f(x^t)\|^2\right]\notag\\
        &&+ \frac{i^2}{n^2}\mathbb{E}\left[\left\|\frac{1}{i}\sum^{i-1}_{j=0}\left(\nabla f_{\pi^j}(x^t) - \nabla f(x^t)\right)\right\|^2\right]\notag\\
		&\leq& \frac{i^2}{n^2}\mathbb{E}\left[\|\nabla f(x^t)\|^2\right]\notag\\
        &&+\frac{i^2}{n^3}\frac{n-i}{i(n-1)}\sum^n_{j=1}\mathbb{E}\left[\|\nabla f_j(x^t) - \nabla f(x^t)\|^2\right]\notag \\
		&\leq& \mathbb{E}\left[\|\nabla f(x^t)\|^2\right] +\frac{1}{n}\sigma^2_{t}. \label{eq:kdjkcnsdbciscsnd}
	\end{eqnarray}
	Taking the full expectation from \eqref{eq:cjshjdjsdbhs} and using \eqref{eq:kdjkcnsdbciscsnd}, we obtain 
	\begin{eqnarray*}
		\mathbb{E}\left[\|x^t_i -x^t\|^2\right]
		&\leq& 4\tau^2\left(\widetilde{L} + \frac{\omega}{Mn}L_{\max}\right)\sum^{n-1}_{j=0}\mathbb{E}\left[D_{f_{\pi^j}}(x^{t}_{j}, x^t)\right]\\
		&&+ 2\tau^2\mathbb{E}\left[\|\nabla f(x^t)\|^2\right] +\frac{2\tau^2}{n}\sigma^2_{t}\\
        &&+ \frac{2\tau^2\omega}{M^2n^2}\sum^M_{m=1}\sum^{n-1}_{j=0}\mathbb{E}\left[\left\|\nabla f_{m,\pi^{j}_m}(x^t)\right\|^2\right].
	\end{eqnarray*}
	Using $\widetilde{L}$-smoothness of $f_{\pi^j}$, we get
	\begin{eqnarray*}
		\mathbb{E}\left[\|x^t_i -x^t\|^2\right]
		&\leq& 2\widetilde{L}\tau^2\left(\widetilde{L} + \frac{\omega}{Mn}L_{\max}\right)\sum^{n-1}_{j=0}\mathbb{E}\left[\|x^{t}_{j} - x^t\|^2\right]\\
		&&+ 2\tau^2\mathbb{E}\left[\|\nabla f(x^t)\|^2\right] +\frac{2\tau^2}{n}\sigma^2_{t}\\
        &&+ \frac{2\tau^2\omega}{M^2n^2}\sum^M_{m=1}\sum^{n-1}_{j=0}\mathbb{E}\left[\left\|\nabla f_{m,\pi^{j}_m}(x^t)\right\|^2\right].
	\end{eqnarray*}
	Since $\tau \leq\frac{1}{2\sqrt{\widetilde{L}\left(\widetilde{L}+\frac{\omega}{Mn}L_{\max}\right)}}$, we have 
	\begin{eqnarray*}
		\mathbb{E}\left[\|x^t_i -x^t\|^2\right]
		&\leq& 2\left(1-2\widetilde{L}\tau^2\left(\widetilde{L} + \frac{\omega}{Mn}L_{\max}\right)\right)\sum^{n-1}_{j=0}\mathbb{E}\left[\|x^{t}_{j} - x^t\|^2\right]\\
		&\leq& 4\tau^2\mathbb{E}\left[\|\nabla f(x^t)\|^2\right] +\frac{4\tau^2}{n}\sigma^2_{t}\\
        &&+ \frac{4\tau^2\omega}{M^2n^2}\sum^M_{m=1}\sum^{n-1}_{j=0}\mathbb{E}\left[\left\|\nabla f_{m,\pi^{j}_m}(x^t)\right\|^2\right]\\
		&\leq& \frac{8\tau^2\omega}{M^2n^2}\sum^M_{m=1}\sum^{n-1}_{j=0}\mathbb{E}\left[\left\|\nabla f_{m,\pi^{j}_m}(x^t) - \nabla f_{m,\pi^{j}_m}(x^{\star})\right\|^2\right] \\
		&& + \frac{8\tau^2\omega}{M^2n^2}\sum^M_{m=1}\sum^{n-1}_{j=0}\mathbb{E}\left[\left\|\nabla f_{m,\pi^{j}_m}(x^{\star})\right\|^2\right]\\
        &&+ 4\tau^2\mathbb{E}\left[\|\nabla f(x^t) - \nabla f(x^{\star})\|^2\right] \\
		&& + \frac{4\tau^2}{n}\left(\frac{1}{n}\sum^{n}_{j=1}\mathbb{E}\left[\|\nabla f_j(x^t)\|\right] -\mathbb{E}\left[\|\nabla f(x^t)\|^2\right]\right)\\
		&\leq& \frac{8\tau^2\omega}{M^2n^2}\sum^M_{m=1}\sum^{n-1}_{j=0}\mathbb{E}\left[\left\|\nabla f_{m,\pi^{j}_m}(x^t) - \nabla f_{m,\pi^{j}_m}(x^{\star})\right\|^2\right] \\
		&& + \frac{8\tau^2\omega}{M^2n^2}\sum^M_{m=1}\sum^{n-1}_{j=0}\mathbb{E}\left[\left\|\nabla f_{m,\pi^{j}_m}(x^{\star})\right\|^2\right]\\
        &&+ 8\tau^2\mathbb{E}\left[\|\nabla f(x^t) - \nabla f(x^{\star})\|^2\right]\\
		&& + \frac{8\tau^2}{n^2}\sum^{n}_{j=1}\mathbb{E}\left[\|\nabla f_j(x^t) - \nabla f_j(x^{\star})\|^2\right]\\
        &&+ \frac{8\tau^2}{n^2}\sum^{n}_{j=1}\mathbb{E}\left[\|\nabla f_j(x^{\star})\|^2\right].
	\end{eqnarray*}
	Summing from $i=0$ to $n-1$ and using $\widetilde{L}$-smoothness of $f_i$ and $L_{\max}$-smoothness of $f_{m, i}$, we obtain 
	\begin{eqnarray*}
		\frac{1}{n}\sum^{n-1}_{i=0}\mathbb{E}\left[\|x^t_i -x^t\|^2\right]
		&\leq& \frac{16\tau^2\omega}{Mn}L_{\max}\mathbb{E}\left[f(x^t) - f(x^{\star})\right]\\
        &&+ \frac{16\tau^2}{n}\widetilde{L}\mathbb{E}\left[ f(x^t) -  f(x^{\star})\right] \\
		&& + \frac{8\tau^2\omega}{Mn}\left(\zeta_{\star}^2+\sigma^2_{\star}\right) + \frac{8\tau^2}{n}\sigma^2_{\star, n}\\
        &&+ 8\tau^2\widetilde{L}\mathbb{E}\left[f(x^t) - f(x^{\star})\right].
	\end{eqnarray*}
\end{proof}

\begin{theorem}
	\label{th_convergence_Q_RR_cvx_case}
	Let Assumptions \ref{asm:quantization_operators}, \ref{asm:sc_general_f},  \ref{asm:lip_max_f_m}, \ref{asm:lip_avr_f} hold and stepsize $\gamma$ satisfy
	\begin{equation}
		\label{stepsize_q_rr_cvx_case}
		0 < \gamma \leq \frac{1}{16n\left(\widetilde{L}+\frac{\omega}{Mn}L_{\max}\right)}.
	\end{equation}
	Then, for all $T \geq 0$ the iterates produced by \gls{Q-RR} satisfy
	\begin{eqnarray*}
		\mathbb{E}\left[\|x^T-x^{\star}\|^2\right]
		&\leq& \left(1-\frac{n\gamma\mu}{2}\right)^T\|x^0 - x^{\star}\|^2 +18\frac{\gamma^2n\widetilde{L}}{\mu}\left(\frac{\omega}{M}(\zeta^2_{\star}+\sigma_{\star}^2) +\sigma^2_{\star,n}\right)\\
		&& +  8\frac{\gamma\omega}{\mu M}(\zeta^2_{\star}+\sigma_{\star}^2),
	\end{eqnarray*}
	where 
	\begin{equation}
		\label{def_sigma_n}
		\sigma^2_{\star,n} = \frac{1}{n}\sum^n_{i=1}\|\nabla f_i(x^{\star})\|^2.
	\end{equation}
\end{theorem}
\begin{proof}
	Taking the expectation w.r.t.\ $\cQ$ and using Lemma \ref{lem_norm_grad_cvx_case}, we get 
	\begin{eqnarray*}
		\mathbb{E}_{\cQ}\left[\|x^{t+1} - x^{\star}\|^2\right] 
		&=& \|x^t - x^{\star}\|^2 -2\tau\mathbb{E}_{\cQ}\left[\la g^t, x^t- x^{\star}\ra\right] + \tau^2\mathbb{E}_{\cQ}\left[\|g^t\|^2\right]\\
		&\leq& \|x^t - x^{\star}\|^2 -2\tau\mathbb{E}_{\cQ}\left[\left\la g^t, x^t- x^{\star}\right\ra\right]\\
		&& + 2\tau^2\widetilde{L}\left(\widetilde{L} + \frac{\omega}{Mn}L_{\max}\right)\frac{1}{n}\sum^{n-1}_{i=0}\mathbb{E}\left[\|x^t_i-x^t\|^2\right]\\
		&&+ 8\tau^2\left(\widetilde{L}+\frac{\omega}{Mn}L_{\max}\right)(f(x^t)-f(x^{\star}))\\
        &&+ \frac{4\tau^2\omega}{Mn}(\zeta^2_{\star} + \sigma_{\star}^2).
	\end{eqnarray*}
	Using Lemma \ref{lem_inner_product_cvx_case}, we obtain 
	\begin{eqnarray*}
		\mathbb{E}_{\cQ}\left[\|x^{t+1} - x^{\star}\|^2\right] 
		&\leq& \|x^t - x^{\star}\|^2\\
  &&-\frac{\tau\mu}{2}\|x^t-x^{\star}\|^2-\tau(f(x^t)- f(x^{\star}))\\
  &&+\frac{\tau\widetilde{L}}{n}\sum^{n-1}_{i=0}\|x^t_i-x^t\|^2\\
		&& + 2\tau^2\widetilde{L}\left(\widetilde{L} + \frac{\omega}{Mn}L_{\max}\right)\frac{1}{n}\sum^{n-1}_{i=0}\mathbb{E}\left[\|x^t_i-x^t\|^2\right]\\
		&&+ 8\tau^2\left(\widetilde{L}+\frac{\omega}{Mn}L_{\max}\right)(f(x^t)-f(x^{\star}))\\
        &&+ \frac{4\tau^2\omega}{Mn}(\zeta^2_{\star} + \sigma_{\star}^2)\\
		&\leq& \left(1 -\frac{\tau\mu}{2}\right)\|x^t-x^{\star}\|^2\\
  &&-\tau\left(1- 8\tau\left(\widetilde{L}+\frac{\omega}{Mn}L_{\max}\right)\right)(f(x^t)- f(x^{\star}))\\
		&& + \tau\widetilde{L}\left(1+ 2\tau\left(\widetilde{L} + \frac{\omega}{Mn}L_{\max}\right)\right)\frac{1}{n}\sum^{n-1}_{i=0}\mathbb{E}\left[\|x^t_i-x^t\|^2\right]\\
  &&+ \frac{4\tau^2\omega}{Mn}(\zeta^2_{\star} + \sigma_{\star}^2).
	\end{eqnarray*}
	Next, we take the full expectation and apply Lemma \ref{lem_dinst_cvx_case}: 
	\begin{eqnarray*}
		&&\mathbb{E}\left[\|x^{t+1} - x^{\star}\|^2\right] 
		\leq \left(1 -\frac{\tau\mu}{2}\right)\mathbb{E}\left[\|x^t-x^{\star}\|^2\right]\\
  &&-\tau\left(1- 8\tau\left(\widetilde{L}+\frac{\omega}{Mn}L_{\max}\right)\right)\mathbb{E}\left[f(x^t)- f(x^{\star})\right]\\
		&&+ 24\tau^3\widetilde{L}\left(1+ 2\tau\left(\widetilde{L} + \frac{\omega}{Mn}L_{\max}\right)\right)\left(\widetilde{L}+\frac{\omega}{Mn}L_{\max}\right)(f(x^t)-f(x^{\star})) \\
		&&+ 8\tau^3\widetilde{L}\left(1+ 2\tau\left(\widetilde{L} + \frac{\omega}{Mn}L_{\max}\right)\right)\left(\frac{\omega}{Mn}(\zeta^2_{\star}+\sigma_{\star}^2) +\frac{\sigma^2_{\star,n}}{n}\right)+ \frac{4\tau^2\omega}{Mn}(\zeta^2_{\star} + \sigma_{\star}^2).
	\end{eqnarray*}
	Using \eqref{stepsize_q_rr_cvx_case}, we derive
	\begin{eqnarray*}
		\mathbb{E}\left[\|x^{t+1} - x^{\star}\|^2\right] 
		&\leq& \left(1 -\frac{\tau\mu}{2}\right)\mathbb{E}\left[\|x^t-x^{\star}\|^2\right]\\
  &&+ 9\tau^3\widetilde{L}\left(\frac{\omega}{Mn}(\zeta^2_{\star} + \sigma_{\star}^2) +\frac{\sigma^2_{\star,n}}{n}\right)+ \frac{4\tau^2\omega}{Mn}(\zeta^2_{\star} + \sigma_{\star}^2).
	\end{eqnarray*}
	Recursively unrolling the inequality, substituting $\tau=n\gamma$ and using inequality $\sum\limits^{+\infty}_{t = 0}\left(1-\frac{\tau\mu}{2}\right)^t \leq \frac{2}{\mu\tau}$, we get the result.
\end{proof}

\begin{corollary}
	\label{cor_convergence_new_Q_RR_cvx_case}
	Let the assumptions of Theorem~\ref{th_convergence_Q_RR_cvx_case} hold and
	\begin{equation}
		\label{new_gamma_new_Q_RR_cvx_case}
		\gamma = \min\left\{\frac{1}{16n\left(\widetilde{L}+\frac{\omega}{Mn}L_{\max}\right)}, \sqrt{\frac{\varepsilon\mu}{8^2n\widetilde{L}}}\left(\frac{\omega}{M}\Delta^2_{\star} +\sigma^2_{\star,n}\right)^{-\frac{1}{2}}, \frac{\varepsilon\mu M}{24\omega\Delta^2_{\star}}\right\},
	\end{equation}
	where $\Delta^2_{\star} = \zeta^2_{\star}+\sigma^2_{\star}$. Then, \gls{Q-RR} finds a solution with accuracy $\varepsilon>0$ after the following number of communication rounds: 
	\begin{equation}
		\widetilde{\cO}\left(\frac{n\widetilde{L}}{\mu}+\frac{\omega}{M}\frac{L_{\max}}{\mu}+ \frac{\omega}{M}\frac{\zeta^2_{\star}+\sigma^2_{\star}}{\varepsilon\mu^2} + \sqrt{\frac{n\widetilde{L}}{\varepsilon\mu^3}}\sqrt{\frac{\omega}{M}\left(\zeta^2_{\star}+\sigma^2_{\star}\right) +\sigma^2_{\star,n}}\right).\notag
	\end{equation}
\end{corollary}
\begin{proof}
	Theorem~\ref{th_convergence_Q_RR_cvx_case} implies
	\begin{eqnarray*}
		\mathbb{E}\left[\|x^T-x^{\star}\|^2\right]
		&\leq& \left(1-\frac{n\gamma\mu}{2}\right)^T\|x^0 - x^{\star}\|^2 +18\frac{\gamma^2n\widetilde{L}}{\mu}\left(\frac{\omega}{M}\left(\zeta^2_{\star}+\sigma^2_{\star}\right) +\sigma^2_{\star,n}\right)\\
		&&+  8\frac{\gamma\omega}{\mu M}\left(\zeta^2_{\star}+\sigma^2_{\star}\right).
	\end{eqnarray*}
	To estimate the number of communication rounds required to find a solution with accuracy $\varepsilon >0$, we need to upper bound each term from the right-hand side by $\nicefrac{\varepsilon}{3}$. Thus, we get additional conditions on $\gamma$:
	\begin{equation*}
		18\frac{\gamma^2n\widetilde{L}}{\mu}\left(\frac{\omega}{M}\left(\zeta^2_{\star}+\sigma^2_{\star}\right) +\sigma^2_{\star,n}\right) < \frac{\varepsilon}{3},\quad 8\frac{\gamma\omega}{\mu M}\left(\zeta^2_{\star}+\sigma^2_{\star}\right)<\frac{\varepsilon}{3},
	\end{equation*}
	and also the upper bound on the number of communication rounds $nT$
	\begin{equation*}
		nT = \widetilde{\cO}\left(\frac{1}{\gamma\mu}\right).
	\end{equation*}
Substituting \eqref{new_gamma_new_Q_RR_cvx_case} in the previous equation, we get the result. 
\end{proof}

\section{Missing Proofs for DIANA-RR}

\subsection{Proof of Theorem \ref{th_conv_rr_diana}}

\begin{lemma}
\label{lem_rr_diana_conv_h}
Let Assumptions \ref{asm:quantization_operators}, \ref{asm:lip_max_f_m}, \ref{asm:lip_avr_f}, \ref{asm:sc_each_f_m_refined} hold  and $\alpha\leq\frac{1}{1+\omega}$. Then, the iterates of \gls{DIANA-RR} satisfy
\begin{eqnarray*}
	\frac{1}{M}\sum^M_{m=1}\mathbb{E}_{\cQ}\left[\|h^{t+1}_{m, \pi^i_m} - \nabla f_{m,\pi^i_m}(x^{\star})\|^2\right] &\leq& \frac{1-\alpha}{M}\sum^M_{m=1}\|h^{t}_{m,\pi^i_m} -\nabla f_{m,\pi^i_m}(x^{\star})\|^2\\
	&& + \frac{2\alpha L_{\max}}{M}\sum^M_{m=1} D_{f_{m,\pi^{i}_m}}(x^{t}_{i},x^{\star}).
\end{eqnarray*}
\end{lemma}
\begin{proof}
Taking the expectation w.r.t.\ $\cQ$, we obtain
\begin{eqnarray*}
	&&\mathbb{E}_{\cQ}\left[\|h^{t+1}_{m, \pi^i_m} - \nabla f_{m,\pi^i_m}(x^{\star})\|^2\right]\\
    &=& \mathbb{E}_{\cQ}\left[\|h^{t}_{m,\pi^i_m} +\alpha\cQ(\nabla f_{m,\pi^i_m}(x^{t}_{i})-h^{t}_{m,\pi^i_m})- \nabla f_{m,\pi^i_m}(x^{\star})\|^2\right]\\
	&=& \|h^{t}_{m,\pi^i_m}- \nabla f_{m,\pi^i_m}(x^{\star})\|^2\\
	&& +2\alpha\mathbb{E}_{\cQ}\left[\left\la\cQ(\nabla f_{m,\pi^i_m}(x^{t}_{i})-h^{t}_{m,\pi^i_m}),h^{t}_{m,\pi^i_m}- \nabla f_{m,\pi^i_m}(x^{\star})\right\ra\right] \\
	&&+ \alpha^2 \mathbb{E}_{\cQ}\left[\| \cQ(\nabla f_{m,\pi^i_m}(x^{t}_{i})-h^{t}_{m,\pi^i_m})\|^2\right]\\
	&=& \|h^{t}_{m,\pi^i_m}- \nabla f_{m,\pi^i_m}(x^{\star})\|^2 \\&&+2\alpha\left\la \nabla f_{m,\pi^i_m}(x^{t}_{i})-h^{t}_{m,\pi^i_m},h^{t}_{m,\pi^i_m}- \nabla f_{m,\pi^i_m}(x^{\star})\right\ra \\
	&&+ \alpha^2 \mathbb{E}_{\cQ}\left[\| \cQ(\nabla f_{m,\pi^i_m}(x^{t}_{i})-h^{t}_{m,\pi^i_m})\|^2\right].
\end{eqnarray*}
Assumption \ref{asm:quantization_operators}, $L_{\max}$-smoothness and convexity of $f_{m, i}$ and $\alpha \leq \nicefrac{1}{(1+\omega)}$ imply
\begin{eqnarray}
	&&\mathbb{E}_{\cQ}\left[\|h^{t+1}_{m, \pi^i_m} - \nabla f_{m,\pi^i_m}(x^{\star})\|^2\right]\notag\\
	&\leq& \|h^{t}_{m,\pi^i_m}- \nabla f_{m,\pi^i_m}(x^{\star})\|^2  \notag\\
	&&+2\alpha\left\la \nabla f_{m,\pi^i_m}(x^{t}_{i})-h^{t}_{m,\pi^i_m},h^{t}_{m,\pi^i_m}- \nabla f_{m,\pi^i_m}(x^{\star})\right\ra \notag\\
	&&+ \alpha^2(1+\omega)\| \nabla f_{m,\pi^i_m}(x^{t}_{i}) -h^{t}_{m,\pi^i_m}\|^2 \notag\\
	&\leq& \|h^{t}_{m,\pi^i_m}- \nabla f_{m,\pi^i_m}(x^{\star})\|^2 \notag\\
	&& + \alpha\left\la\nabla f_{m,\pi^i_m}(x^{t}_{i}) -h^{t}_{m,\pi^i_m}, h^{t}_{m,\pi^i_m} +\nabla f_{m,\pi^i_m}(x^{t}_{i}) - 2\nabla f_{m,\pi^i_m}(x^{\star})\right\ra \notag \\
	&\leq& \|h^{t}_{m,\pi^i_m}- \nabla f_{m,\pi^i_m}(x^{\star})\|^2 \notag \\
	&& + \alpha\|\nabla f_{m,\pi^i_m}(x^{t}_{i}) - \nabla f_{m,\pi^i_m}(x^{\star})\|^2\notag\\
    &&- \alpha\|h^{t}_{m,\pi^i_m}-\nabla f_{m,\pi^i_m}(x^{\star})\|^2\notag\\
	&\leq& (1-\alpha)\|h^{t}_{m,\pi^i_m}- \nabla f_{m,\pi^i_m}(x^{\star})\|^2 \notag\\
	&&+ \alpha\|\nabla f_{m,\pi^i_m}(x^{t}_{i}) - \nabla f_{m,\pi^i_m}(x^{\star})\|^2 \notag\\
	&\leq& (1-\alpha)\|h^{t}_{m,\pi^i_m}- \nabla f_{m,\pi^i_m}(x^{\star})\|^2\notag\\
    &&+ 2\alpha L_{\max} D_{f_{m,\pi^{i}_m}}(x^{t}_{i},x^{\star}).
    \label{eq:ndskjjdcidscuidbid}
\end{eqnarray}
Summing up the above inequality for $m=1, \ldots, M$, we get the result.
\end{proof}

\begin{theorem}
	\label{thm:advanced_conv_diana_rr_sc}
	Let Assumptions \ref{asm:quantization_operators}, \ref{asm:lip_max_f_m}, \ref{asm:lip_avr_f}, \ref{asm:sc_each_f_m_refined}  hold and select stepsize as $0<\gamma \leq \min\left\{\frac{\alpha}{2n\widetilde\mu}, \frac{1}{\widetilde{L}+\frac{6\omega}{M}L_{\max}}\right\}$, $\alpha \leq \frac{1}{1+\omega}$.
	Then, for all $T \geq 0$ the iterates produced by \gls{DIANA-RR} satisfy
	\begin{equation*}
		\mathbb{E}\left[\Psi^{(t)}\right] \leq \left(1-\gamma\widetilde{\mu}\right)^{nT}\Psi^0 +\frac{2\gamma^2\sigma^2_{\text{rad}}}{\widetilde\mu},
	\end{equation*}
	where $\Psi^{(t)}$ is defined in \eqref{lyapunov_func_rr_diana}.
\end{theorem}
\begin{proof}
Using  \(x_{i+1}^{\star} = x^{\star}_{i} -\frac{\gamma}{M}\sum^M_{m=1}\nabla f_{m,\pi^i_m}(x^{\star})\)  and line 9 of Algorithm \ref{alg_new_RR_DIANA}, we derive
\begin{eqnarray*}
\|x_{i+1}^t-x_{i+1}^{\star}\|^2 
&=& \left\|x^t_i-x^{\star}_{i}- \gamma\frac{1}{M}\sum^M_{m=1} \left(\hat{g}^{\pi^{i}_m}_{t,m} - \nabla f_{m,\pi^{i}_m}(x^{\star})\right)\right\|^2\\
&=& \left\|x^t_i-x^{\star}_{i}\right\|^2 - \frac{2\gamma}{M}\sum^M_{m=1}\left\la \left(\hat{g}^{\pi^{i}_m}_{t,m} - \nabla f_{m,\pi^{i}_m}(x^{\star})\right), x^t_i-x^{\star}_{i} \right\ra \\
&&+\gamma^2\left\|\frac{1}{M}\sum^M_{m=1} \left(\hat{g}^{\pi^{i}_m}_{t,m} - \nabla f_{m,\pi^{i}_m}(x^{\star})\right)\right\|^2.
\end{eqnarray*}
Taking  expectation w.r.t.\ $\cQ$ and using $\mathbb{E}\|\xi - c\|^2 = \mathbb{E}\|\xi - \mathbb{E}\xi\|^2 + \|\mathbb{E}\xi - c\|^2$, we obtain 
\begin{eqnarray*}
&&\mathbb{E}_{\cQ}\left[\|x_{i+1}^t-x_{i+1}^{\star}\|^2 \right]
= \left\|x^t_i-x^{\star}_{i}\right\|^2\\
&&- \frac{2\gamma}{M}\sum^M_{m=1} \left\la\nabla f_{m,\pi^{i}_m}(x^{t}_{i}) - \nabla f_{m,\pi^{i}_m}(x^{\star}), x^t_i-x^{\star}_{i} \right\ra\\
&&+\gamma^2\mathbb{E}_{\cQ}\left[\left\|\frac{1}{M}\sum^M_{m=1} \left(\cQ\left(\nabla f_{m,\pi^{i}_m}(x^{t}_{i}) - h^{\pi^{i}_m}_{t,m}\right) + h^{t}_{m, \pi^{i}_m} - \nabla f_{m,\pi^{i}_m}(x^{\star})\right)\right\|^2\right]\\
&\leq& \left\|x^t_i-x^{\star}_{i}\right\|^2 - \frac{2\gamma}{M}\sum^M_{m=1} \left\la\nabla f_{m,\pi^{i}_m}(x^{t}_{i}) - \nabla f_{m,\pi^{i}_m}(x^{\star}), x^t_i-x^{\star}_{i} \right\ra\\
&&+\gamma^2\mathbb{E}_{\cQ}\left[\left\|\frac{1}{M}\sum^M_{m=1} \left(\cQ\left(\nabla f_{m,\pi^{i}_m}(x^{t}_{i}) - h^{\pi^{i}_m}_{t,m}\right) - \nabla f_{m,\pi^{i}_m}(x^{t}_{i}) + h^{t}_{m, \pi^{i}_m} \right)\right\|^2\right]\\
&&+\gamma^2\left\|\frac{1}{M}\sum^M_{m=1} \left( \nabla f_{m,\pi^{i}_m}(x^{\star})  - \nabla f_{m,\pi^{i}_m}(x^{t}_{i})\right)\right\|^2.
\end{eqnarray*}
Independence of $\cQ\left(\nabla f_{m,\pi^{i}_m}(x^{t}_{i}) - h^{\pi^{i}_m}_{t,m}\right)$, $m \in [M]$, assumption \ref{asm:quantization_operators}, and three-point identity imply
\begin{align*}
&\mathbb{E}_{\cQ}\left[\|x_{i+1}^t-x_{i+1}^{\star}\|^2 \right]\\
&\leq \left\|x^t_i-x^{\star}_{i}\right\|^2\\
&- \frac{2\gamma}{M}\sum^M_{m=1} \left(D_{f_{m,\pi^{i}_m}}(x^{\star}_{i},x^{t}_{i}) +D_{f_{m,\pi^{i}_m}}( x^t_i, x^{\star}) -D_{f_{m,\pi^{i}_m}}(x^{\star}_{i},x^{\star})  \right)\\
&+\frac{\gamma^2\omega}{M^2}\sum^M_{m=1}\left\|\nabla f_{m,\pi^{i}_m}(x^{t}_{i}) - h^{\pi^{i}_m}_{t,m}\right\|^2\\
& +\gamma^2\left\|\frac{1}{M}\sum^M_{m=1} \left( \nabla f_{m,\pi^{i}_m}(x^{\star})  - \nabla f_{m,\pi^{i}_m}(x^{t}_{i})\right)\right\|^2\\
&\leq \left\|x^t_i-x^{\star}_{i}\right\|^2 \\
&- \frac{2\gamma}{M}\sum^M_{m=1} \left(D_{f_{m,\pi^{i}_m}}(x^{\star}_{i},x^{t}_{i}) +D_{f_{m,\pi^{i}_m}}( x^t_i, x^{\star}) -D_{f_{m,\pi^{i}_m}}(x^{\star}_{i},x^{\star})  \right)\\
&+\frac{2\gamma^2\omega}{M}\frac{1}{M}\sum^M_{m=1}\left\|\nabla f_{m,\pi^{i}_m}(x^{t}_{i}) - \nabla f_{m,\pi^{i}_m}(x^{\star})\right\|^2\\
& + \gamma^2\left\|\frac{1}{M}\sum^M_{m=1} \left( \nabla f_{m,\pi^{i}_m}(x^{\star})  - \nabla f_{m,\pi^{i}_m}(x^{t}_{i})\right)\right\|^2\\
&+\frac{2\gamma^2\omega}{M^2}\sum^M_{m=1}\left\|h^{t}_{m, \pi^{i}_m}- \nabla f_{m,\pi^{i}_m}(x^{\star})\right\|^2.
\end{align*}
Using $L_{\max}$-smoothness and $\mu$-strong convexity of functions $f_{m, i}$ and $\widetilde{L}$ smoothness and $\widetilde{\mu}$-strong convexity of $f_{\pi^i} = \frac{1}{M}\sum^M_{i=1}f_{m,\pi_m^i}$, we obtain 
\begin{eqnarray*}
&&\mathbb{E}_{\cQ}\left[\|x_{i+1}^t-x_{i+1}^{\star}\|^2 \right]\\
&\leq& (1-\gamma\widetilde\mu)\left\|x^t_i-x^{\star}_{i}\right\|^2  \\
&&-2\gamma\left(1 - \gamma\left(\widetilde{L}+\frac{2\omega}{M}L_{\max}\right)\right) \frac{1}{M}\sum^M_{m=1} D_{f_{m,\pi^i_m}}(x^t_i,x^{\star}) \\
&& + \frac{2\gamma}{M}\sum^M_{m=1} D_{f_{m,\pi^{i}_m}}(x^{\star}_{i},x^{\star}) +\frac{2\gamma^2\omega}{M^2}\sum^M_{m=1}\left\|h^{t}_{m, \pi^{i}_m}- \nabla f_{m,\pi^{i}_m}(x^{\star})\right\|^2.
\end{eqnarray*}
Taking the full expectation and using Definition 2, we derive
\begin{eqnarray*}
&&\mathbb{E}\left[\|x_{i+1}^t-x_{i+1}^{\star}\|^2 \right]\\
&\leq& (1-\gamma\widetilde\mu)\mathbb{E}\left[\left\|x^t_i-x^{\star}_{i}\right\|^2\right]\\
&&-2\gamma\left(1 - \gamma \left(\widetilde{L}+\frac{2\omega}{M}L_{\max}\right)\right) \frac{1}{M}\sum^M_{m=1}\mathbb{E}\left[D_{f_{m,\pi^{i}_m}}(x^{t}_{i},x^{\star})\right]  \\
&& + 2\gamma^3\sigma^2_{\text{rad}} +\frac{2\gamma^2\omega}{M^2}\sum^M_{m=1}\mathbb{E}\left[\left\|h^{t}_{m, \pi^{i}_m}- \nabla f_{m,\pi^{i}_m}(x^{\star})\right\|^2 \right].
\end{eqnarray*}
Recursively unrolling the inequality, we get
\begin{eqnarray*}
&&\mathbb{E}\left[\|x^{t+1}-x^{\star}\|^2 \right]\\
&\leq& (1-\gamma\widetilde\mu)^n\mathbb{E}\left[\left\|x^t-x^{\star}\right\|^2\right] \\
&&+\frac{2\gamma^2\omega}{M^2}\sum^M_{m=1}\sum^{n-1}_{j=0}(1-\gamma\widetilde\mu)^j\mathbb{E}\left[\left\|h^{t}_{m, \pi^{i}_m}- \nabla f_{m,\pi^{i}_m}(x^{\star})\right\|^2 \right]  \\
&& -2\gamma\left(1 - \gamma \left(\widetilde{L}+\frac{2\omega}{M}L_{\max}\right)\right)\times\\
&&\times\frac{1}{M}\sum^M_{m=1}\sum^{n-1}_{j=0}(1-\gamma\widetilde\mu)^j\mathbb{E}\left[D_{f_{m,\pi^{i}_m}}(x^{t}_{i},x^{\star})\right]\\
&& + 2\gamma^3\sigma^2_{\text{rad}}\sum^{n-1}_{j=0}(1-\gamma\widetilde\mu)^j.
\end{eqnarray*}
Next, we apply \eqref{lyapunov_func_rr_diana} and Lemma \ref{lem_rr_diana_conv_h}: 
\begin{eqnarray*}
&&\mathbb{E}\left[\Psi^{(t+1)}\right]
\leq (1-\gamma\widetilde\mu)^n\mathbb{E}\left[\left\|x^t-x^{\star}\right\|^2\right] + 2\gamma^3\sigma^2_{\text{rad}}\sum^{n-1}_{j=0}(1-\gamma\widetilde\mu)^j\\ &&+\left(c(1-\alpha)+\frac{2\omega}{M}\right)\times\\
&&\times\frac{\gamma^2}{M}\sum^M_{m=1}\sum^{n-1}_{j=0}(1-\gamma\widetilde\mu)^j\mathbb{E}\left[\left\|h^{t}_{m, \pi^{i}_m}- \nabla f_{m,\pi^{i}_m}(x^{\star})\right\|^2 \right]  \\
&&  -2\gamma\left(1 -c\gamma\alpha L_{\max}-  \gamma 
\left(\widetilde{L}+\frac{2\omega}{M}L_{\max}\right) \right)\times\\
&&\times\frac{1}{M}\sum^M_{m=1}\sum^{n-1}_{j=0}(1-\gamma\widetilde\mu)^j\mathbb{E}\left[D_{f_{m,\pi^{i}_m}}(x^{\star}_{i},x^{\star})\right],
\end{eqnarray*}
where $c = \frac{4\omega}{\alpha M^2}$. Using $\alpha \leq \frac{1}{1+\omega}$ and $\gamma \leq \min\left\{\frac{\alpha}{2n\mu}, \frac{1}{\left(\widetilde{L}+\nicefrac{6\omega}{M}L_{\max}\right)}\right\}$, we obtain
\begin{eqnarray*}
&&\mathbb{E}\left[\Psi^{(t+1)} \right]\\
&\leq& (1-\gamma\widetilde\mu)^n\mathbb{E}\left[\left\|x^t-x^{\star}\right\|^2\right] \\
&&+\left(1-\frac{\alpha}{2}\right)\frac{4\omega\gamma^2}{\alpha M^2}\sum^M_{m=1}\sum^{n-1}_{j=0}(1-\gamma\widetilde\mu)^j\mathbb{E}\left[\left\|h^{t}_{m, \pi^{i}_m}- \nabla f_{m,\pi^{i}_m}(x^{\star})\right\|^2 \right] \\
&& + 2\gamma^2\sigma^3_{\text{rad}}\sum^{n-1}_{j=0}(1-\gamma\widetilde\mu)^j\\
&\leq&\max\left\{(1-\gamma\widetilde\mu)^n, \left(1-\frac{\alpha}{2}\right)\right\}\mathbb{E}\left[\Psi^{(t)} \right]\\
&&+ 2\gamma^2\sigma^3_{\text{rad}}\sum^{n-1}_{j=0}(1-\gamma\widetilde\mu)^j\\
&\leq&(1-\gamma\widetilde\mu)^n\mathbb{E}\left[\Psi^{(t)} \right] + 2\gamma^3\sigma^2_{\text{rad}}\sum^{n-1}_{j=0}(1-\gamma\widetilde\mu)^j.
\end{eqnarray*}

Recursively rewriting the inequality, we obtain 
\begin{eqnarray*}
\mathbb{E}\left[\Psi^{(T)} \right]
&\leq&(1-\gamma\widetilde\mu)^{nT}\Psi^{0}\\
&&+ 2\gamma^3\sigma^2_{\text{rad}}\sum^{T-1}_{t=0}(1-\gamma\widetilde\mu)^{tn}\sum^{n-1}_{j=0}(1-\gamma\widetilde\mu)^j\\
&\leq&(1-\gamma\widetilde\mu)^{nT}\Psi^{0}\\
&&+ 2\gamma^3\sigma^2_{\text{rad}}\sum^{nT-1}_{k=0}(1-\gamma\widetilde\mu)^{k}
\end{eqnarray*}
Using that $\sum\limits^{+\infty}_{k = 0}\left(1-\frac{\gamma\widetilde\mu}{2}\right)^k \leq \frac{2}{\widetilde\mu\gamma}$, we finish the proof.
\end{proof}

\begin{corollary}
	\label{cor_convergence_DIANA_RR}
	Let the assumptions of Theorem~\ref{thm:advanced_conv_diana_rr_sc} hold, $\alpha = \frac{1}{1+\omega}$ and
	\begin{equation}
		\label{new_gamma_DIANA_RR}
		\gamma = \min\left\{\frac{\alpha}{2n\widetilde{\mu}},\frac{1}{\widetilde{L}+\frac{6\omega}{M}L_{\max}}, \frac{\sqrt{\varepsilon\widetilde{\mu}}}{2\sigma_{\text{rad}}}\right\}.
	\end{equation}
	Then \gls{DIANA-RR} finds a solution with accuracy $\varepsilon>0$ after the following number of communication rounds: 
	\begin{equation}
		\widetilde{\cO}\left(n(1+\omega)+\frac{\widetilde{L}}{\widetilde{\mu}}+\frac{\omega}{M}\frac{L_{\max}}{\widetilde{\mu}} + \frac{\sigma_{\text{rad}}}{\sqrt{\varepsilon\widetilde{\mu}^3}}\right).\notag
	\end{equation}
\end{corollary}
\begin{proof}
	Theorem~\ref{thm:advanced_conv_diana_rr_sc} implies
	\begin{equation*}
		\mathbb{E}\left[\Psi^{(t)}\right] \leq \left(1-\gamma\widetilde{\mu}\right)^{nT}\Psi^0 +\frac{2\gamma^2\sigma^2_{\text{rad}}}{\widetilde\mu}.
	\end{equation*}
	To estimate the number of communication rounds required to find a solution with accuracy $\varepsilon >0$, we need to upper bound each term from the right-hand side by $\frac{\varepsilon}{2}$. Thus, we get an additional condition on $\gamma$:
	\begin{equation*}
		\frac{2\gamma^2\sigma^2_{\text{rad}}}{\widetilde\mu}<\frac{\varepsilon}{2},
	\end{equation*}
	and also the upper bound on the number of communication rounds $nT$
	\begin{equation*}
		nT = \widetilde{\cO}\left(\frac{1}{\gamma\mu}\right).
	\end{equation*}
Substituting \eqref{new_gamma_DIANA_RR} in the previous equation, we get the result.
\end{proof}

\subsection{Non-Strongly convex summands}
In this section, we provide the analysis of \gls{DIANA-RR} without using Assumptions~\ref{asm:sc_each_f_m}, \ref{asm:sc_each_f_m_refined}. We emphasize that $x_{i+1}^t = x^{i}_t - \gamma \frac{1}{M}\sum^{M}_{m=1}\hat{g}_{m,\pi^i_m}^{t}$. Then we have 
\begin{equation*}
	x^{t+1} = x^t - \gamma \sum^{n-1}_{i=0}\frac{1}{M}\sum^{M}_{m=1}\hat{g}_{m,\pi^i_m}^{t} = x^t -\tau \frac{1}{Mn}\sum^{n-1}_{i=0} \sum^{M}_{m=1}\hat{g}_{m,\pi^i_m}^{t}.
\end{equation*}
We denote $\hat{g}^t = \frac{1}{Mn}\sum^{n-1}_{i=0} \sum^{M}_{m=1}\hat{g}_{m,\pi^i_m}^{t}$.

\begin{lemma}
	\label{lem_inner_product_diana_rr_cvx_case}
	 Let Assumptions \ref{asm:quantization_operators}, \ref{asm:sc_general_f},  \ref{asm:lip_max_f_m}, \ref{asm:lip_avr_f} hold. Then, the following inequality holds
	\begin{align*}
		-2\tau\mathbb{E}_{\cQ}\left[\la\hat{g}_t- h^{\star}, x^t-x^{\star}\ra \right] &\leq -\frac{\tau\mu}{2}\|x^t-x^{\star}\|^2 - \tau\left(f(x^t)-f(x^{\star})\right)\\
        &+ \tau\widetilde{L}\frac{1}{n}\sum^{n-1}_{i=1}\|x^t-x^t_i\|^2,
	\end{align*}
	where $h^{\star} = \nabla f(x^{\star}) = 0$.
\end{lemma}
\begin{proof}
	Since $h^{\star} = \nabla f(x^{\star}) = 0$, the proof of Lemma \ref{lem_inner_product_diana_rr_cvx_case} is identical to the proof of Lemma \ref{lem_inner_product_cvx_case}.
\end{proof}

\begin{lemma}
	\label{lem_norm_of_grad_diana_rr_cvx_case}
	 Let Assumptions \ref{asm:quantization_operators}, \ref{asm:sc_general_f},  \ref{asm:lip_max_f_m}, \ref{asm:lip_avr_f} hold. Then, the following inequality holds
	\begin{eqnarray*}
		\mathbb{E}_{\cQ}\left[\|\hat{g}^t - h^{\star}\|^2\right] &\leq& 2\widetilde{L}\left(\widetilde{L} + \frac{\omega}{Mn}L_{\max}\right)\frac{1}{n}\sum^{n-1}_{i=0}\|x^t_i-x^t\|^2\\
        &&+ 8\left(\widetilde{L} + \frac{\omega}{Mn}L_{\max}\right)\left(f(x^t)-f(x^{\star})\right)\\
		&&+ \frac{4\omega}{M^2n^2}\sum^{n-1}_{i=0}\sum^M_{m=1}\|h^{t}_{m,\pi^i_m}-\nabla f_{m, \pi^i_m}(x^{\star})\|^2
	\end{eqnarray*}
\end{lemma}
\begin{proof}
	Taking the expectation w.r.t. $\cQ$, we get 
	\begin{eqnarray*}
		&&\mathbb{E}_{\cQ}\left[\left\|\hat{g}^t - h^{\star}\right\|^2\right] \\
		&=& \mathbb{E}_{\cQ}\left[\left\|\frac{1}{Mn}\sum^{n-1}_{i=0} \sum^{M}_{m=1}\hat{g}_{m,\pi^i_m}^{t} - h^{\star}\right\|^2\right]\\
		&=& \mathbb{E}_{\cQ}\left[\left\|\frac{1}{Mn}\sum^{n-1}_{i=0} \sum^{M}_{m=1}\left(h^{t}_{m,\pi^i_m} + \cQ\left(\nabla f_{m,\pi^i_m}(x^{t}_{i}) - h^{t}_{m,\pi^i_m}\right)\right) - h^{\star}\right\|^2\right]\\
		&=& \mathbb{E}_{\cQ}\left[\left\|\frac{1}{Mn}\sum^{n-1}_{i=0} \sum^{M}_{m=1}\left(h^{t}_{m,\pi^i_m} - \nabla f_{m,\pi^i_m}(x^{t}_{i}) + \cQ\left(\nabla f_{m,\pi^i_m}(x^{t}_{i}) - h^{t}_{m,\pi^i_m}\right)\right) \right\|^2\right]\\
		&&+ \left\|\frac{1}{Mn}\sum^{n-1}_{i=0} \sum^{M}_{m=1}\nabla f_{m,\pi^i_m}(x^{t}_{i})  - h^{\star}\right\|^2.
	\end{eqnarray*}
	Independence of $\cQ\left(\nabla f_{m,\pi^i_m}(x^{t}_{i}) - h^{t}_{m,\pi^i_m}\right)$, $m \in [M]$ and Assumption \ref{asm:quantization_operators} imply 
	\begin{eqnarray*}
		&&\mathbb{E}_{\cQ}\left[\|\hat{g}^t - h^{\star}\|^2\right] \\
		&=& \frac{1}{M^2n^2}\sum^{n-1}_{i=0} \sum^{M}_{m=1}\mathbb{E}_{\cQ}\left[\left\|h^{t}_{m,\pi^i_m} - \nabla f_{m,\pi^i_m}(x^{t}_{i}) + \cQ\left(\nabla f_{m,\pi^i_m}(x^{t}_{i}) - h^{t}_{m,\pi^i_m}\right)\right\|^2\right]\\
		&&+ \left\|\frac{1}{Mn}\sum^{n-1}_{i=0} \sum^{M}_{m=1}\nabla f_{m,\pi^i_m}(x^{t}_{i})  - h^{\star}\right\|^2\\
		&\leq& \frac{\omega}{M^2n^2}\sum^{n-1}_{i=0} \sum^{M}_{m=1}\left\|\nabla f_{m,\pi^i_m}(x^{t}_{i}) -  h^{t}_{m,\pi^i_m} \right\|^2 + \left\|\frac{1}{n}\sum^{n-1}_{i=0}\nabla f_{\pi^i}(x^{t}_{i})  - h^{\star}\right\|^2\\
		&\leq& \frac{2\omega}{M^2n^2}\sum^{n-1}_{i=0} \sum^{M}_{m=1}\left\|\nabla f_{m,\pi^i_m}(x^{t}_{i}) - \nabla f_{m,\pi^i_m}(x^{t})\right\|^2\\
        &&+ \frac{2}{n}\sum^{n-1}_{i=0}\left\|\nabla f_{\pi^i}(x^{t}_{i})  - \nabla f_{\pi^i}(x^{t})\right\|^2\\
		&&+ \frac{2\omega}{M^2n^2}\sum^{n-1}_{i=0} \sum^{M}_{m=1}\left\|h^{t}_{m,\pi^i_m} - \nabla f_{m,\pi^i_m}(x^{t}) \right\|^2 + 2\left\|\frac{1}{n}\sum^{n-1}_{i=0}\nabla f_{\pi^i}(x^{t})  - h^{\star}\right\|^2.
	\end{eqnarray*}
	Using $L_{\max}$-smoothness and convexity of $f_{m, i}$ and $\widetilde{L}$-smoothness and convexity of $f_{\pi^i}$, we obtain 
	\begin{eqnarray*}
		\mathbb{E}_{\cQ}\left[\|\hat{g}^t - h^{\star}\|^2\right] 
		&\leq& \frac{4\omega L_{\max}}{M^2n^2}\sum^{n-1}_{i=0} \sum^{M}_{m=1}D_{f_{m,\pi^i_m}}(x^{t}_{i}, x^{t}) + \frac{4\widetilde{L}}{n}\sum^{n-1}_{i=0}D_{f_{\pi^i}}(x^{t}_{i},x^{t})\\
		&&+ \frac{4\omega}{M^2n^2}\sum^{n-1}_{i=0} \sum^{M}_{m=1}\left\|h^{t}_{m,\pi^i_m} - \nabla f_{m,\pi^i_m}(x^{\star}) \right\|^2\\
        &&+ 4\widetilde{L}\left(f(x^{t})  - f(x^{\star})\right)\\
		&&+ \frac{4\omega}{M^2n^2}\sum^{n-1}_{i=0} \sum^{M}_{m=1}\left\|\nabla f_{m,\pi^i_m}(x^{t}) - \nabla f_{m,\pi^i_m}(x^{\star}) \right\|^2 \\
		&\leq& 2\widetilde{L}\left(\widetilde{L} +\frac{\omega}{Mn}L_{\max}\right)\frac{1}{n}\sum^{n-1}_{i=0} \|x^{t}_{i} - x^{t}\|^2\\
        &&+ 4\widetilde{L}\left(f(x^{t})  - f(x^{\star})\right)\\
		&&+ \frac{8\omega}{Mn}L_{\max}\frac{1}{Mn}\sum^{n-1}_{i=0} \sum^{M}_{m=1}D_{f_{m,\pi^i_m}}(x^{t},x^{\star})\\
		&&+ \frac{4\omega}{M^2n^2}\sum^{n-1}_{i=0} \sum^{M}_{m=1}\left\|h^{t}_{m,\pi^i_m} - \nabla f_{m,\pi^i_m}(x^{\star}) \right\|^2. 
	\end{eqnarray*}
\end{proof}

\begin{lemma}
	\label{lem_conv_h_diana_rr_cvx_case}
	Let $\alpha\leq\frac{1}{1+\omega}$ and  Assumptions \ref{asm:quantization_operators}, \ref{asm:sc_general_f},  \ref{asm:lip_max_f_m}, \ref{asm:lip_avr_f} hold. Then, the iterates produced by \gls{DIANA-RR} satisfy
	\begin{eqnarray*}
		\frac{1}{Mn}\sum^{n-1}_{i=0}\sum^M_{m=1}\mathbb{E}_{\cQ}\left[\|h^{t+1}_{m, \pi^i_m} - \nabla f_{m,\pi^i_m}(x^{\star})\|^2\right] &\leq& \frac{1-\alpha}{Mn}\sum^{n-1}_{i=0}\sum^M_{m=1}\|c -\nabla f_{m,\pi^i_m}(x^{\star})\|^2\\
		&&+ \frac{2\alpha\widetilde{L} L_{\max}}{n}\sum^{n-1}_{i=0} \|x^t_i-x^t\|^2 \\
		&&+ 4 \alpha L_{\max}\left(f(x^t)-f(x^{\star})\right).
	\end{eqnarray*}
\end{lemma}
\begin{proof}
	First of all, we introduce new notation: $$\mathcal{H}_{t} = \frac{1}{Mn}\sum^{n-1}_{i=0}\sum^M_{m=1}\mathbb{E}_{\cQ}\left[\|h^{t}_{m,\pi^i_m} - \nabla f_{m,\pi^i_m}(x^{\star})\|^2\right].$$ Using \eqref{eq:ndskjjdcidscuidbid} and summing it up for $i=0,\ldots, n-1$, we obtain 
	\begin{eqnarray*}
		\mathcal{H}_{t+1} &\leq& \frac{1-\alpha}{Mn}\sum^{n-1}_{i=0}\sum^M_{m=1}\|h^{t}_{m,\pi^i_m}- \nabla f_{m,\pi^i_m}(x^{\star})\|^2\\
        &&+ \frac{\alpha}{Mn}\sum^{n-1}_{i=0}\sum^M_{m=1}\|\nabla f_{m,\pi^i_m}(x^{t}_{i}) - \nabla f_{m,\pi^i_m}(x^{\star})\|^2\\
		&\leq& \frac{1-\alpha}{Mn}\sum^{n-1}_{i=0}\sum^M_{m=1}\|h^{t}_{m,\pi^i_m}- \nabla f_{m,\pi^i_m}(x^{\star})\|^2\\
        &&+ \frac{2\alpha}{Mn}\sum^{n-1}_{i=0}\sum^M_{m=1}\|\nabla f_{m,\pi^i_m}(x^{t}_{i}) - \nabla f_{m,\pi^i_m}(x^{t}) \|^2\\
		&&+ \frac{2\alpha}{Mn}\sum^{n-1}_{i=0}\sum^M_{m=1}\|\nabla f_{m,\pi^i_m}(x^{t}) - \nabla f_{m,\pi^i_m}(x^{\star})\|^2.
	\end{eqnarray*}
	Next, we apply $L_{\max}$-smoothness and convexity of $f_{m,i}$ and $\widetilde{L}$-smoothness and convexity of $f_{\pi^i}$:
	\begin{eqnarray*}
		\mathcal{H}_{t+1} 
		&\leq& \frac{1-\alpha}{Mn}\sum^{n-1}_{i=0}\sum^M_{m=1}\|h^{t}_{m,\pi^i_m}- \nabla f_{m,\pi^i_m}(x^{\star})\|^2\\
        &&+ \frac{4\alpha}{Mn}L_{\max}\sum^{n-1}_{i=0}\sum^M_{m=1}D_{f_{m,\pi^i_m}}(x^{t}_{i}, x^{t})\\
		&&+ \frac{4\alpha}{Mn}L_{\max}\sum^{n-1}_{i=0}\sum^M_{m=1} D_{f_{m,\pi^i_m}}(x^{t}, x^{\star})\\
		&\leq& \frac{1-\alpha}{Mn}\sum^{n-1}_{i=0}\sum^M_{m=1}\|h^{t}_{m,\pi^i_m}- \nabla f_{m,\pi^i_m}(x^{\star})\|^2 + \frac{2\alpha}{n}\widetilde{L}L_{\max}\sum^{n-1}_{i=0}\|x^{t}_{i} - x^{t}\|^2\\
		&&+ \frac{4\alpha}{Mn}L_{\max}\sum^{n-1}_{i=0}\sum^M_{m=1} D_{f_{m,\pi^i_m}}(x^{t}, x^{\star}).
	\end{eqnarray*}
\end{proof}

\begin{lemma}
	\label{lem_dinst_diana_rr_cvx_case} 
	Let Assumptions \ref{asm:quantization_operators}, \ref{asm:sc_general_f},  \ref{asm:lip_max_f_m}, \ref{asm:lip_avr_f} and $\tau \leq \frac{1}{2\sqrt{\widetilde{L}\left(\widetilde{L}+\frac{\omega}{Mn}L_{\max}\right)}}$. Then, the following inequality holds
	\begin{eqnarray*}
		\frac{1}{n}\sum^{n-1}_{i=0}\mathbb{E}\left[\|x^t_i-x^t\|^2\right] &\leq& 24\tau^2\left(\widetilde{L}+\frac{\omega}{Mn}L_{\max}\right)\mathbb{E}\left[f(x^t)-f(x^{\star})\right]  +8\tau^2\frac{\sigma^2_{\star,n}}{n}\\
		&&+ 8\frac{\tau^2\omega}{M^2n^2}\sum^{n-1}_{i=0}\sum^M_{m=1}\mathbb{E}\left[\|h^{t}_{m,\pi^i_m}-\nabla f_{m,\pi^i_m} (x^{\star})\|^2\right],
	\end{eqnarray*}
	where $\sigma^2_{\star,n} = \frac{1}{n}\sum^n_{i=1}\|\nabla f_i(x^{\star})\|^2$.
\end{lemma}
\begin{proof}
	Since $x^t_i= x^t -\frac{\tau}{Mn}\sum^M_{m=1}\sum^{i-1}_{j=0}\left(h_{m,\pi^j_m}^{t} + \cQ\left(\nabla f_{m,\pi^j_m}(x_j^{t}) - h^{t}_{m,\pi^i_m}\right)\right)$, we have 
	\begin{align*}
		&\mathbb{E}_{\cQ}\left[\|x^t_i -x^t\|^2\right]\\
		&= \tau^2\mathbb{E}_{\cQ}\left[\left\|\frac{1}{Mn}\sum^M_{m=1}\sum^{i-1}_{j=0}\left(h_{m,\pi^j_m}^{t} + \cQ\left(\nabla f_{m,\pi^j_m}(x_j^{t}) - h_{m,\pi^j_m}^{t}\right)\right)\right\|^2\right]\\
		&=\tau^2\mathbb{E}_{\cQ}\left[\left\|\frac{1}{Mn}\sum^M_{m=1}\sum^{i-1}_{j=0}\left(h_{m,\pi^j_m}^{t}- \nabla f_{m,\pi^{j}_m}(x^{t}_{j}) + \cQ\left(\nabla f_{m,\pi^i_m}(x_j^{t}) - h_{m,\pi^j_m}^{t}\right)\right)\right\|^2\right]\\
		&+ \tau^2\left\|\frac{1}{Mn}\sum^M_{m=1}\sum^{i-1}_{j=0}\nabla f_{m,\pi^{j}_m}(x^{t}_{j})\right\|^2.
	\end{align*}
	Independence of $\cQ\left(\nabla f_{m,\pi^i_m}(x_j^{t}) - h_{m,\pi^j_m}^{t}\right)$, $m \in [M]$ and Assumption \ref{asm:quantization_operators} imply 
	\begin{eqnarray*}
		&&\mathbb{E}_{\cQ}\left[\|x^t_i -x^t\|^2\right]\\
		&\leq&\frac{\tau^2}{M^2n^2}\sum^M_{m=1}\sum^{i-1}_{j=0}\mathbb{E}_{\cQ}\left[\left\|h_{m,\pi^j_m}^{t}- \nabla f_{m,\pi^{j}_m}(x^{t}_{j}) + \cQ\left(\nabla f_{m,\pi^i_m}(x_j^{t}) - h_{m,\pi^j_m}^{t}\right)\right\|^2\right]\\
		&&+ \tau^2\left\|\frac{1}{Mn}\sum^M_{m=1}\sum^{i-1}_{j=0}\nabla f_{m,\pi^{j}_m}(x^{t}_{j})\right\|^2\\
		&\leq& \frac{\tau^2\omega}{M^2n^2}\sum^M_{m=1}\sum^{i-1}_{j=0}\left\|\nabla f_{\pi^{j}_m}(x^{t}_{j}) - h_{m,\pi^j_m}^{t} \right\|^2 + \tau^2\left\|\frac{1}{n}\sum^{i-1}_{j=0}\nabla f_{\pi^j}(x^{t}_{j})\right\|^2\\
		&\leq& \frac{2\tau^2\omega}{M^2n^2}\sum^M_{m=1}\sum^{n-1}_{j=0}\left\|\nabla f_{m,\pi^{j}_m}(x^{t}_{j}) - \nabla f_{m,\pi^{j}_m}(x^t)\right\|^2 + 2\tau^2\left\|\frac{1}{n}\sum^{i-1}_{j=0}\nabla f_{\pi^j}(x^t)\right\|^2\\
		&& + \frac{2\tau^2\omega}{M^2n^2}\sum^M_{m=1}\sum^{n-1}_{j=0}\left\|h_{m,\pi^j_m}^{t} - \nabla f_{m,\pi^{j}_m}(x^t) \right\|^2\\
        &&+ \frac{2\tau^2}{n}\sum^{n-1}_{j=0}\left\|\nabla f_{\pi^j}(x^{t}_{j}) - \nabla f_{\pi^j}(x^t)\right\|^2.
	\end{eqnarray*}
	Using $L_{\max}$-smoothness and convexity of $f_{m, i}$ and $\widetilde{L}$-smoothness and convexity of $f_{\pi^j}$, we obtain
	\begin{eqnarray*}
		\mathbb{E}_{\cQ}\left[\|x^t_i -x^t\|^2\right]
		&\leq& \frac{4\tau^2\omega}{M^2n^2}L_{\max}\sum^M_{m=1}\sum^{n-1}_{j=0}D_{f_{m,\pi^{j}_m}}(x^{t}_{j},x^t) + 2\tau^2\left\|\frac{1}{n}\sum^{i-1}_{j=0}\nabla f_{\pi^j}(x^t)\right\|^2\\
		&& + \frac{2\tau^2\omega}{M^2n^2}\sum^M_{m=1}\sum^{n-1}_{j=0}\left\|h_{m,\pi^j_m}^{t} - \nabla f_{m,\pi^{j}_m}(x^t) \right\|^2\\
        &&+ \frac{2\tau^2\widetilde{L}^2}{n}\sum^{n-1}_{j=0}\left\|x^{t}_{j} - x^t\right\|^2\\
		&\leq& 2\tau^2\widetilde{L}\left(\widetilde{L}+\frac{\omega}{Mn}L_{\max}\right)\frac{1}{n}\sum^{n-1}_{j=0}\|x^{t}_{j} - x^t\|^2\\
        &&+ 2\tau^2\left\|\frac{1}{n}\sum^{i-1}_{j=0}\nabla f_{\pi^j}(x^t)\right\|^2\\
		&&+ \frac{2\tau^2\omega}{M^2n^2}\sum^M_{m=1}\sum^{n-1}_{j=0}\left\|h_{m,\pi^j_m}^{t} - \nabla f_{m,\pi^{j}_m}(x^t) \right\|^2. 
	\end{eqnarray*}
	Taking the full expectation and using \eqref{eq:kdjkcnsdbciscsnd}, we derive
	\begin{eqnarray*}
		\mathbb{E}\left[\|x^t_i -x^t\|^2\right]
		&\leq& 2\tau^2\widetilde{L}\left(\widetilde{L}+\frac{\omega}{Mn}L_{\max}\right)\frac{1}{n}\sum^{n-1}_{j=0}\mathbb{E}\left[\|x^{t}_{j} - x^t\|^2\right]\\
        &&+ 2\tau^2 \mathbb{E}\left[\|\nabla f(x^t)\|^2\right]\\
		&&+ \frac{4\tau^2\omega}{M^2n^2}\sum^M_{m=1}\sum^{n-1}_{j=0}\mathbb{E}\left[\left\|h_{m,\pi^j_m}^{t} - \nabla f_{m,\pi^{j}_m}(x^{\star}) \right\|^2\right]  + \frac{2\tau^2}{n}\mathbb{E}\left[\sigma^2_t\right] \\
		&&+ \frac{8\tau^2\omega}{M^2n^2}L_{\max}\sum^M_{m=1}\sum^{n-1}_{j=0}\mathbb{E}\left[D_{f_{m,\pi^{j}_m}}(x^t,x^{\star})\right].
	\end{eqnarray*}
	Using $L_{\max}$-smoothness and convexity of $f_{m, i}$ and $\widetilde{L}$-smoothness and convexity of $f_{\pi^j}$, we obtain
	\begin{eqnarray*}
		\mathbb{E}\left[\|x^t_i -x^t\|^2\right]
		&\leq& 2\tau^2\widetilde{L}\left(\widetilde{L}+\frac{\omega}{Mn}L_{\max}\right)\frac{1}{n}\sum^{n-1}_{j=0}\mathbb{E}\left[\|x^{t}_{j} - x^t\|^2\right] \\
		&&+ \frac{4\tau^2\omega}{M^2n^2}\sum^M_{m=1}\sum^{n-1}_{j=0}\mathbb{E}\left[\left\|h_{m,\pi^j_m}^{t} - \nabla f_{m,\pi^{j}_m}(x^{\star}) \right\|^2\right]  + \frac{2\tau^2}{n}\mathbb{E}\left[\sigma^2_t\right] \\
		&&+ 4\tau^2\left(\widetilde{L} + \frac{2\omega}{M^2n^2}L_{\max}\right)\mathbb{E}\left[f(x^t)-f(x^{\star})\right].
	\end{eqnarray*}
	Now we need to estimate $\frac{2\tau^2}{n}\mathbb{E}\left[\sigma^2_t\right]$. Due to $\mathbb{E}\left[\sigma^2_t\right] \leq \frac{1}{n}\sum^{n}_{i=1}\mathbb{E}\left[\|\nabla f_i(x^t)\|^2\right]$, we get
	\begin{eqnarray*}
		\frac{2\tau^2}{n}\mathbb{E}\left[\sigma^2_t\right]
		&\leq& \frac{2\tau^2}{n^2}\sum^{n}_{j=1}\mathbb{E}\left[\|\nabla f_j(x^t)\|^2\right] \\
		&\leq& \frac{4\tau^2}{n^2}\sum^{n}_{j=1}\mathbb{E}\left[\|\nabla f_j(x^t) - \nabla f_j(x^{\star})\|^2\right] + \frac{4\tau^2}{n^2}\sum^{n}_{j=1}\mathbb{E}\left[\|\nabla f_j(x^{\star})\|^2\right]\\
		&\leq&  \frac{8\tau^2}{n^2}\widetilde{L}\sum^{n}_{j=1}\mathbb{E}\left[D_{f_j}(x^t,x^{\star})\right] + \frac{4\tau^2}{n^2}\sum^{n}_{j=1}\sigma^2_{n,\star}.
	\end{eqnarray*}
	Combining two previous inequalities, we get 
	\begin{eqnarray*}
		\mathbb{E}\left[\|x^t_i -x^t\|^2\right]
		&\leq& 2\tau^2\widetilde{L}\left(\widetilde{L}+\frac{\omega}{Mn}L_{\max}\right)\frac{1}{n}\sum^{n-1}_{j=0}\mathbb{E}\left[\|x^{t}_{j} - x^t\|^2\right] \\
		&&+ \frac{4\tau^2\omega}{M^2n^2}\sum^M_{m=1}\sum^{n-1}_{j=0}\mathbb{E}\left[\left\|h_{m,\pi^j_m}^{t} - \nabla f_{m,\pi^{j}_m}(x^{\star}) \right\|^2\right]   \\
		&&+ 4\tau^2\left(\widetilde{L} + \frac{2\omega}{M^2n^2}L_{\max}\right)\mathbb{E}\left[f(x^t)-f(x^{\star})\right]\\
		&&+ \frac{8\tau^2}{n}\widetilde{L}\mathbb{E}\left[f(x^t)- f(x^{\star})\right] + \frac{4\tau^2}{n^2}\sum^{n}_{j=1}\sigma^2_{n,\star}.
	\end{eqnarray*}
	Summing from $i=0$ to $n-1$ and using $\tau \leq \frac{1}{2\sqrt{\widetilde{L}\left(\widetilde{L}+\frac{\omega}{Mn}L_{\max}\right)}}$, we obtain
	\begin{eqnarray*}
		\frac{1}{n}\sum^{n-1}_{i=0}\mathbb{E}\left[\|x^t_i -x^t\|^2\right]
		&\leq& 2\left(1-2\tau^2\widetilde{L}\left(\widetilde{L}+\frac{\omega}{Mn}L_{\max}\right)\right)\frac{1}{n}\sum^{n-1}_{i=0}\mathbb{E}\left[\|x^t_i - x^t\|^2\right] \\
		&\leq& \frac{8\tau^2\omega}{M^2n^2}\sum^M_{m=1}\sum^{n-1}_{j=0}\mathbb{E}\left[\left\|h_{m,\pi^j_m}^{t} - \nabla f_{m,\pi^{j}_m}(x^{\star}) \right\|^2\right]  \\
		&&+ 8\tau^2\left(\widetilde{L} + \frac{2\omega}{M^2n^2}L_{\max}\right)\mathbb{E}\left[f(x^t)-f(x^{\star})\right]\\
		&&+ \frac{16\tau^2}{n}\widetilde{L}\mathbb{E}\left[f(x^t)- f(x^{\star})\right] + \frac{8\tau^2}{n^2}\sum^{n}_{j=1}\sigma^2_{n,\star}.
	\end{eqnarray*}
\end{proof}

We consider the following Lyapunov function:
\begin{equation}
	\label{lyapunov_func_diana_rr_cvx_case}
	\Psi^{(t+1)} = \|x^{t+1}-x^{\star}\|^2 +\frac{c\tau^2}{Mn}\sum^M_{m=1}\sum^{n-1}_{j=0}\left\|h^{t+1}_{m, \pi^i_m} - \nabla f_{m,\pi^i_m}(x^{\star})\right\|^2.
\end{equation}

\begin{theorem}
	\label{thm:advanced_conv_for_DIANA_RR}
	 Let Assumptions \ref{asm:quantization_operators}, \ref{asm:sc_general_f},  \ref{asm:lip_max_f_m}, \ref{asm:lip_avr_f} hold  and
	\begin{equation*}
		\gamma \leq \min\left\{\frac{\alpha}{n\mu}, \frac{1}{12n\left(\widetilde{L}+\frac{11\omega}{Mn}L_{\max}\right)}\right\},\quad \alpha \leq \frac{1}{1+\omega},\quad c = \frac{10\omega}{\alpha Mn}.
	\end{equation*}
	Then, for all $T \geq 0$ the iterates produced by \gls{DIANA-RR} satisfy
	\begin{equation*}
		\mathbb{E}\left[\Psi^{(t)}\right] \leq \left(1-\frac{n\gamma\mu}{2}\right)^{T}\Psi^0 +20\frac{\gamma^2n\widetilde{L}}{\mu}\sigma^2_{\star,n}.
	\end{equation*}
\end{theorem}
\begin{proof}
	Taking the expectation w.r.t.\ $\cQ$ and using Lemma \ref{lem_inner_product_diana_rr_cvx_case}, we get 
	\begin{eqnarray*}
		\mathbb{E}_{\cQ}\left[\|x^{t+1} - x^{\star}\|^2\right]
		&=& \|x^t -\tau\hat{g}^t - x^{\star}+\tau h^{\star}\|^2\\
		&=& \|x^t - x^{\star}\|^2 -2\tau\mathbb{E}_{\cQ}\left[\la \hat{g}^t - h^{\star}, x^t- x^{\star}\ra\right]\\
        &&+ \tau^2\mathbb{E}_{\cQ}\left[\|\hat{g}^t - h^{\star}\|^2\right]\\
		&\leq& \|x^t - x^{\star}\|^2 -\frac{\tau\mu}{2}\|x^t-x^{\star}\|^2\\
        &&+ \tau^2\mathbb{E}_{\cQ}\left[\|\hat{g}^t - h^{\star}\|^2\right]\\
		&& - \tau\left(f(x^t)-f(x^{\star})\right) + \tau\widetilde{L}\frac{1}{n}\sum^{n-1}_{i=1}\|x^t-x^t_i\|^2.
	\end{eqnarray*}
	Next, due to Lemma \ref{lem_norm_of_grad_diana_rr_cvx_case} we have 
	\begin{eqnarray*}
		\mathbb{E}_{\cQ}\left[\|x^{t+1} - x^{\star}\|^2\right]
		&\leq& \left(1 -\frac{\tau\mu}{2}\right)\|x^t-x^{\star}\|^2 - \tau\left(f(x^t)-f(x^{\star})\right)\\
        &&+ \tau\widetilde{L}\frac{1}{n}\sum^{n-1}_{i=1}\|x^t-x^t_i\|^2\\
		&&+ 2\tau^2\widetilde{L}\left(\widetilde{L} + \frac{\omega}{Mn}L_{\max}\right)\frac{1}{n}\sum^{n-1}_{i=0}\|x^t_i-x^t\|^2 \\
		&&+ 8\tau^2\left(\widetilde{L} + \frac{\omega}{Mn}L_{\max}\right)\left(f(x^t)-f(x^{\star})\right)\\
		&&+ \frac{4\omega\tau^2}{M^2n^2}\sum^{n-1}_{i=0}\sum^M_{m=1}\|h^{t}_{m,\pi^i_m}-\nabla f_{m, \pi^i_m}(x^{\star})\|^2\\
		&\leq& \left(1 -\frac{\tau\mu}{2}\right)\|x^t-x^{\star}\|^2\\
        &&+ \frac{4\omega\tau^2}{M^2n^2}\sum^{n-1}_{i=0}\sum^M_{m=1}\|h^{t}_{m,\pi^i_m}-\nabla f_{m, \pi^i_m}(x^{\star})\|^2\\
		&&- \tau\left(1-8\tau\left(\widetilde{L} + \frac{\omega}{Mn}L_{\max}\right)\right)\left(f(x^t)-f(x^{\star})\right)\\
		&&+ \tau\widetilde{L}\left(1+2\tau\left(\widetilde{L} + \frac{\omega}{Mn}L_{\max}\right)\right)\frac{1}{n}\sum^{n-1}_{i=0}\|x^t_i-x^t\|^2.
	\end{eqnarray*}
	Using \eqref{lyapunov_func_diana_rr_cvx_case}, we obtain
	\begin{eqnarray*}
		\mathbb{E}_{\cQ}\left[\Psi^{(t+1)}\right]
		&\leq& \left(1 -\frac{\tau\mu}{2}\right)\|x^t-x^{\star}\|^2  + \frac{4\omega\tau^2}{M^2n^2}\sum^{n-1}_{i=0}\sum^M_{m=1}\|h^{t}_{m,\pi^i_m}-\nabla f_{m, \pi^i_m}(x^{\star})\|^2\\
		&&- \tau\left(1-8\tau\left(\widetilde{L} + \frac{\omega}{Mn}L_{\max}\right)\right)\left(f(x^t)-f(x^{\star})\right)\\
		&&+ \tau\widetilde{L}\left(1+2\tau\left(\widetilde{L} + \frac{\omega}{Mn}L_{\max}\right)\right)\frac{1}{n}\sum^{n-1}_{i=0}\|x^t_i-x^t\|^2 \\
		&&+ \frac{c\tau^2}{Mn}\sum^M_{m=1}\sum^{n-1}_{j=0}\mathbb{E}\left[\left\|h^{t+1}_{m, \pi^i_m} - \nabla f_{m,\pi^i_m}(x^{\star})\right\|^2\right].
	\end{eqnarray*}
	To estimate the last term in the above inequality, we apply Lemma \ref{lem_conv_h_diana_rr_cvx_case}:
	\begin{eqnarray*}
		\mathbb{E}_{\cQ}\left[\Psi^{(t+1)}\right]
		&\leq& \left(1 -\frac{\tau\mu}{2}\right)\|x^t-x^{\star}\|^2  + \frac{4\omega\tau^2}{M^2n^2}\sum^{n-1}_{i=0}\sum^M_{m=1}\|h^{t}_{m,\pi^i_m}-\nabla f_{m, \pi^i_m}(x^{\star})\|^2\\
		&&- \tau\left(1-8\tau\left(\widetilde{L} + \frac{\omega}{Mn}L_{\max}\right)\right)\left(f(x^t)-f(x^{\star})\right)\\
		&&+ \tau\widetilde{L}\left(1+2\tau\left(\widetilde{L} + \frac{\omega}{Mn}L_{\max}\right)\right)\frac{1}{n}\sum^{n-1}_{i=0}\|x^t_i-x^t\|^2 \\
		&&+ c\tau^2\frac{1-\alpha}{Mn}\sum^{n-1}_{i=0}\sum^M_{m=1}\|h^{t}_{m,\pi^i_m} -\nabla f_{m,\pi^i_m}(x^{\star})\|^2\\
		&&+ c\tau^2\frac{2\alpha\widetilde{L} L_{\max}}{n}\sum^{n-1}_{i=0} \|x^t_i-x^t\|^2 + 4 c\tau^2\alpha L_{\max}\left(f(x^t)-f(x^{\star})\right)\\
		&\leq& \left(1 -\frac{\tau\mu}{2}\right)\|x^t-x^{\star}\|^2\\
        &&+ \left(1-\alpha+ \frac{4\omega}{cMn}\right)\frac{c\tau^2}{Mn}\sum^{n-1}_{i=0}\sum^M_{m=1}\|h^{t}_{m,\pi^i_m}-\nabla f_{m, \pi^i_m}(x^{\star})\|^2\\
		&&- \tau\left(1-4c\tau\alpha L_{\max}-8\tau\left(\widetilde{L} + \frac{\omega}{Mn}L_{\max}\right)\right)\left(f(x^t)-f(x^{\star})\right)\\
		&& + \tau\widetilde{L}\left(1+ 2c\tau\alpha L_{\max}+2\tau\left(\widetilde{L} + \frac{\omega}{Mn}L_{\max}\right)\right)\frac{1}{n}\sum^{n-1}_{i=0}\|x^t_i-x^t\|^2.
	\end{eqnarray*}	
	Let $\mathcal{H}^t = \frac{c\tau^2}{Mn}\sum^{n-1}_{i=0}\sum^M_{m=1}\mathbb{E}\left[\|h^{t}_{m,\pi^i_m}-\nabla f_{m, \pi^i_m}(x^{\star})\|^2\right]$. Taking the full expectation and using Lemma \ref{lem_dinst_diana_rr_cvx_case}, we get 
	\begin{align*}
		&\mathbb{E}\left[\Psi^{(t+1)}\right]\\
		&\leq \left(1 -\frac{\tau\mu}{2}\right)\mathbb{E}\left[\|x^t-x^{\star}\|^2\right] + \left(1-\alpha+ \frac{4\omega}{cMn}\right)\mathcal{H}^t\\
		&- \tau\left(1-4c\tau\alpha L_{\max}-8\tau\left(\widetilde{L} + \frac{\omega}{Mn}L_{\max}\right)\right)\mathbb{E}\left[f(x^t)-f(x^{\star})\right]\\
		& + \tau\widetilde{L}\left(1+ 2c\tau\alpha L_{\max}+2\tau\left(\widetilde{L} + \frac{\omega}{Mn}L_{\max}\right)\right)\frac{1}{n}\sum^{n-1}_{i=0}\mathbb{E}\left[\|x^t_i-x^t\|^2\right] \\
		&\leq \left(1 -\frac{\tau\mu}{2}\right)\mathbb{E}\left[\|x^t-x^{\star}\|^2\right] + \left(1-\alpha+ \frac{4\omega}{cMn}\right)\mathcal{H}^t\\
		&- \tau\left(1-4c\tau\alpha L_{\max}-8\tau\left(\widetilde{L} + \frac{\omega}{Mn}L_{\max}\right)\right)\mathbb{E}\left[f(x^t)-f(x^{\star})\right]\\
		&+24\tau^3 \widetilde{L}
\left(1 + 2c\tau\alpha L_{\max} + 2\tau\widetilde{L} + \frac{2\tau\omega}{Mn}L_{\max}\right)
\left(\widetilde{L}+\frac{\omega}{Mn}L_{\max}\right)
\mathbb{E}[f(x^t)-f(x^{\star})]\\  &+8\tau^3\widetilde{L}\left(1+ 2c\tau\alpha L_{\max}+2\tau\left(\widetilde{L} + \frac{\omega}{Mn}L_{\max}\right)\right)\frac{\sigma^2_{\star,n}}{n}\\
		&+ \frac{8\tau\widetilde{L}\omega}{cMn}\left(1+ 2c\tau\alpha L_{\max}+2\tau\left(\widetilde{L} + \frac{\omega}{Mn}L_{\max}\right)\right)\mathcal{H}^t.
	\end{align*}
	Selecting $c = \frac{A\omega}{\alpha Mn}$, where $A$ is a positive number to be specified later, we have 
	\begin{equation*}
		1+ 2c\tau\alpha L_{\max}+2\tau\left(\widetilde{L} + \frac{\omega}{Mn}L_{\max}\right) = 1 + 2\tau\left(\widetilde{L} + \frac{(A+1)\omega}{Mn}L_{\max}\right),
	\end{equation*}
	\begin{equation*}
		1-4c\tau\alpha L_{\max}-8\tau\left(\widetilde{L} + \frac{\omega}{Mn}L_{\max}\right) \geq 1 - 8\tau\left(\widetilde{L} + \frac{(A+1)\omega}{Mn}L_{\max}\right).
	\end{equation*}
	Then, we have 
	\begin{align*}
		&\mathbb{E}\left[\Psi^{(t+1)}\right]\\
		&\leq \left(1 -\frac{\tau\mu}{2}\right)\mathbb{E}\left[\|x^t-x^{\star}\|^2\right] + \left(1-\alpha+ \frac{4\alpha}{A} \right)\mathcal{H}^t\\
		&- \tau\left(1 - 8\tau\left(\widetilde{L} + \frac{(A+1)\omega}{Mn}L_{\max}\right)\right)\mathbb{E}\left[f(x^t)-f(x^{\star})\right]\\	&+24\tau^3 \widetilde{L}
\left(\widetilde{L}+\frac{\omega L_{\max}}{Mn}\right)
\left(1 + 2\tau\widetilde{L} + \frac{2\tau(A+1)\omega L_{\max}}{Mn}\right)
\mathbb{E}[f(x^t)-f(x^{\star})]\\  &+8\tau^3\widetilde{L}\left(1 + 2\tau\left(\widetilde{L} + \frac{(A+1)\omega}{Mn}L_{\max}\right)\right)\frac{\sigma^2_{\star,n}}{n}\\
		&+ \frac{8\alpha}{A}\tau\widetilde{L}\left(1 + 2\tau\left(\widetilde{L} + \frac{(A+1)\omega}{Mn}L_{\max}\right)\right)\mathcal{H}^t.
	\end{align*}
	Taking $\tau = \frac{1}{B\left(\widetilde{L} + \frac{(A+1)\omega}{Mn}L_{\max}\right)}$, where $B$ is some positive constant, we obtain
	\begin{eqnarray*}
		\mathbb{E}\left[\Psi^{(t+1)}\right]
		&\leq& \left(1 -\frac{\tau\mu}{2}\right)\mathbb{E}\left[\|x^t-x^{\star}\|^2\right] + \left(1-\alpha+ \frac{4\alpha}{A} + \frac{8\alpha}{A}\tau\widetilde{L}\left(1 + \frac{2}{B}\right) \right)\mathcal{H}^t\\
		&&- \tau\left(1 - \frac{8}{B}- \frac{24}{B^2}\left(1 + \frac{2}{B}\right)\right)\mathbb{E}\left[f(x^t)-f(x^{\star})\right]\\
		&&+8\tau^3\widetilde{L}\left(1 + \frac{2}{B}\right)\frac{\sigma^2_{\star,n}}{n}.
	\end{eqnarray*}
	Choosing $A=10$, $B = 12$, $\tau \leq \frac{\alpha}{\mu}$, we have 
	\begin{eqnarray*}
		\mathbb{E}\left[\Psi^{(t+1)}\right]
		&\leq& \left(1 -\min\left\{\frac{\tau\mu}{2}, \frac{\alpha}{2}\right\}\right)\mathbb{E}\left[\Psi^{(t)}\right] +10\tau^3\widetilde{L}\frac{\sigma^2_{\star,n}}{n}\\
		&\leq& \left(1 -\frac{\tau\mu}{2}\right)\mathbb{E}\left[\Psi^{(t)}\right] +10\tau^3\widetilde{L}\frac{\sigma^2_{\star,n}}{n}\\
	\end{eqnarray*}
	Recursively unrolling the inequality, substituting $\tau=n\gamma$ and using inequality $\sum\limits^{+\infty}_{t = 0}\left(1-\frac{\tau\mu}{2}\right)^t \leq \frac{2}{\mu\tau}$, we finish the proof.
\end{proof}

\begin{corollary}
	\label{cor_convergence_DIANA_RR_cvx_case}
	Let the assumptions of Theorem~\ref{thm:advanced_conv_for_DIANA_RR} hold, $\alpha = \frac{1}{1+\omega}$, and
	\begin{equation}
		\label{new_gamma_DIANA_RR_cvx_case}
		\gamma = \min\left\{\frac{\alpha}{2n\mu},\frac{1}{12n\left(\widetilde{L}+\frac{11\omega}{Mn}L_{\max}\right)}, \sqrt{\frac{\varepsilon\mu}{40n\widetilde{L}\sigma^2_{\star,n}}}\right\}.
	\end{equation} 
	Then, \gls{DIANA-RR} finds a solution with accuracy $\varepsilon>0$ after the following number of communication rounds: 
	\begin{equation}
		\widetilde{\cO}\left(n(1+\omega)+\frac{n\widetilde{L}}{\mu}+\frac{\omega}{M}\frac{L_{\max}}{\mu} + \sqrt{\frac{n\widetilde{L}}{\varepsilon\mu^3}}\sigma_{\star,n}\right).\notag
	\end{equation}
\end{corollary}
\begin{proof}
	Theorem~\ref{thm:advanced_conv_for_DIANA_RR} implies
	\begin{equation*}
		\mathbb{E}\left[\Psi^{(t)}\right] \leq \left(1-\gamma\mu\right)^{nT}\Psi^0 +20\frac{\gamma^2n\widetilde{L}}{\mu}\sigma^2_{\star,n}.
	\end{equation*}
	To estimate the number of communication rounds required to find a solution with accuracy $\varepsilon >0$, we need to upper bound each term from the right-hand side by $\frac{\varepsilon}{2}$. Thus, we get an additional condition on $\gamma$:
	\begin{equation*}
		20\frac{\gamma^2n\widetilde{L}}{\mu}\sigma^2_{\star,n} <\frac{\varepsilon}{2},
	\end{equation*}
	and also the upper bound on the number of communication rounds $nT$
	\begin{equation*}
		nT = \widetilde{\cO}\left(\frac{1}{\gamma\mu}\right).
	\end{equation*}
Substituting \eqref{new_gamma_DIANA_RR_cvx_case} in the previous equation, we obtain the result.
\end{proof}

\section{Results for Q-NASTYA and DIANA-NASTYA}

\begin{theorem}
\label{thm:convergence_Q_NASTYA_}
Let Assumptions~\ref{asm:quantization_operators}, \ref{asm:sc_general_f}, \ref{asm:lip_max_f_m} hold. Let the stepsizes $\gamma$, $\eta$ satisfy $0 < \eta \leq \frac{1}{16L_{\max}\left(1+\frac{\omega}{M}\right)},$ $0 < \gamma \leq \frac{1}{5nL_{\max}}.$
Then, for all $T \geq 0$ the iterates produced by \gls{Q-NASTYA} (Algorithm~\ref{alg:Q_NASTYA}) satisfy
\begin{align*}
	\mathbb{E}\left[\|x^T-x^{\star}\|^2\right] &\leq \left(1-\frac{\eta\mu}{2}\right)^T\|x^0 - x^{\star}\|^2+  8\frac{\eta\omega}{\mu M}\zeta^2_{\star}\\
    &+\frac{9}{2}\frac{\gamma^2nL_{\max}}{\mu}\left((n+1)\zeta^2_{\star} + \sigma_{\star}^2\right).
\end{align*}
\end{theorem}

\begin{corollary}
Under the same conditions as Theorem~\ref{thm:convergence_Q_NASTYA} and for Algorithm~\ref{alg:Q_NASTYA}, there exist stepsizes \(\gamma = \nicefrac{\eta}{n}\) and \(\eta > 0\) such that the number of communication rounds $T$ to find a solution with accuracy $\varepsilon > 0$ is $$			\widetilde{\cO}\left(\frac{L_{\max}}{\mu}\left(1+\frac{\omega}{M}\right)+ \frac{\omega}{M}\frac{\zeta^2_{\star}}{\varepsilon\mu^3}+\sqrt{\frac{ L_{\max} }{\varepsilon\mu^3}} \sqrt{\zeta^2_{\star}+\frac{\sigma_{\star}^2}{n}}\right).$$
If $\gamma \rightarrow 0$, one can choose $\eta > 0$ such that the above complexity bound improves to $			\widetilde{\cO}\left( \frac{L_{\max}}{\mu}\left(1+\frac{\omega}{M}\right)+\frac{\omega}{M}\frac{\zeta^2_{\star}}{\varepsilon\mu^3}\right).$
\end{corollary}
We emphasize several differences with the known theoretical  results. First, the \algname{FedCOM} method \citep{haddadpour2021federated} was analyzed in the homogeneous setting only, i.e., $f_m(x) = f(x)$ for all $m \in [M]$, which is an unrealistic assumption for \gls{FL} applications. In contrast, our result holds in the fully heterogeneous case. Next, the analysis of \algname{FedPAQ} ~\citep{reisizadeh19_fedpaq} uses a bounded variance assumption, which is also known to be restrictive. Nevertheless, let us compare to their result. In \citep{reisizadeh19_fedpaq} authors derive the following complexity for their method: $
	\widetilde{\cO}\left( \frac{L_{\max}}{\mu}\left(1+\frac{\omega}{M}\right)+\frac{\omega}{M}\frac{\sigma^2}{\mu^2\varepsilon} + \frac{\sigma^2}{M\mu^2\varepsilon}\right).
$  This result is inferior to the one we show for \gls{Q-NASTYA}: when $\omega$ is small, the main term in the complexity bound of \algname{FedPAQ} is $\widetilde{\cO}\left(\nicefrac{1}{\varepsilon}\right)$, while for \gls{Q-NASTYA} the dominating term is of the order  $\widetilde{\cO}\left(\nicefrac{1}{\sqrt{\varepsilon}}\right)$ (when $\omega$ and $\varepsilon$ are sufficiently small). 
We also highlight that \algname{FedCRR}~\citep{malinovsky2022federated} does not converge if $\omega > \nicefrac{M^2 \gamma \mu \varepsilon}{(2\left\|x_{*, m}^{n}\right\|^{2})}$, while \gls{Q-NASTYA} does for any $\omega \geq 0$. Finally, when $\omega = 0$ (no compression) we recover \gls{NASTYA} as a special case, and using $\gamma = \nicefrac{\eta}{n}$, we recover the rate of \algname{FedRR}~\citep{mishchenko2022proximal}.

\begin{theorem}
    \label{thm:diana-nastya-conv_}
    Let Assumptions~\ref{asm:quantization_operators}, \ref{asm:sc_general_f}, \ref{asm:lip_max_f_m} hold. Suppose the stepsizes $\gamma$, $\eta, \alpha$ satisfy $
        0 < \gamma \leq  \frac{1}{16L_{\max}n}$, $0 < \eta \leq \min\left\{\frac{\alpha}{2\mu}, \frac{1}{16L_{\max}\left(1+\frac{9\omega}{M}\right)}\right\},$ and $\alpha \leq \frac{1}{1+\omega}.$
    Define the following Lyapunov function:
    \begin{equation}
    \label{eq:diana-nastya-lyapunov-function_}
    \Psi^{(t+1)} \eqdef \|x^{t+1}-x^{\star}\|^2 +\frac{8\omega\eta^2}{\alpha M^2}\sum^M_{m=1}\|h^{t+1}_{m}-h^{\star}_m\|^2.
    \end{equation}
    Then, for all $T \geq 0$ the iterates produced by \gls{DIANA-NASTYA} (Algorithm~\ref{alg:diana-nastya}) satisfy
    \begin{equation}
    \mathbb{E}\left[\Psi^{(t)}\right] \leq \left(1-\frac{\eta\mu}{2}\right)^T\Psi^0 + \frac{9}{2}\frac{\gamma^2 nL}{\mu}\left((n+1)\zeta^2_{\star} + \sigma_{\star}^2\right).
    \end{equation}
\end{theorem}

\begin{corollary}
Under the same conditions as Theorem~\ref{thm:diana-nastya-conv} and for Algorithm~\ref{alg:diana-nastya}, there exist stepsizes \(\gamma = \nicefrac{\eta}{n}\), \(\eta > 0\), \(\alpha > 0\) such that the number of communication rounds $T$ to find a solution with accuracy $\varepsilon > 0$ is $$	\widetilde{\cO}\left(\omega + \frac{L_{\max}}{\mu}\left(1+\frac{\omega}{M}\right)+\sqrt{\frac{ L_{\max} }{\varepsilon\mu^3}} \sqrt{\zeta^2_{\star}+\frac{\sigma_{\star}^2}{n}}\right).$$
If $\gamma \rightarrow 0$, one can choose $\eta > 0$ such that the above complexity bound improves to $		\widetilde{\cO}\left(\omega + \frac{L_{\max}}{\mu}\left(1+\frac{\omega}{M}\right)\right).$
\end{corollary}

In contrast to \gls{Q-NASTYA}, \gls{DIANA-NASTYA} does not suffer from the $\widetilde{\cO}(\nicefrac{1}{\varepsilon})$ term in the complexity bound. This shows the superiority of \gls{DIANA-NASTYA} to \gls{Q-NASTYA}. Next, \algname{FedCRR-VR}~\citep{malinovsky2022federated} has the rate $
	\widetilde{\mathcal{O}}\left(\frac{(\omega+1)\left(1-\frac{1}{\kappa}\right)^{n}}{\left(1-\left(1-\frac{1}{\kappa}\right)^{n}\right)^{2}}+\frac{\sqrt{\kappa}\left(\zeta_{\star}+ \sigma_{\star}\right)}{\mu \sqrt{\varepsilon}}\right),$ which depends on $\widetilde{\cO}\left(\nicefrac{1}{\sqrt{\varepsilon}}\right)$. However, the first term is close to $\widetilde{\cO}\left((\omega+1)\kappa^2\right)$ for a large condition number. \algname{FedCRR-VR-2} utilizes variance reduction technique from work~\citep{malinovsky2023random} and it allows to get rid of permutation variance. This method has $\widetilde{\cO}\left(\frac{(\omega+1)\left(1-\frac{1}{\kappa \sqrt{\kappa n}}\right)^{\frac{n}{2}}}{\left(1-\left(1-\frac{1}{\kappa \sqrt{\kappa n}}\right)^{\frac{n}{2}}\right)^{2}}+\frac{\sqrt{\kappa}\zeta_{\star}}{\mu \sqrt{\varepsilon}}\right) $ complexity, but it requires additional assumption on number of functions $n$ and thus not directly comparable with our result. Note that if we have no compression $(\omega = 0)$, \gls{DIANA-NASTYA} recovers rate of \gls{NASTYA}.

\newpage 
\section{Missing Proofs for Q-NASTYA}

We start with deriving a technical lemma along with stating several useful results from \citep{malinovsky2023server}. For convenience, we also introduce the following notation:
\begin{equation*}
	g^{t}_{m} = \frac{1}{n}\sum^{n-1}_{i = 0}\nabla f_{m,\pi^i_m}(x^i_{t,m}).
\end{equation*}

\begin{lemma}
	\label{lem_norm_g^t}
	Let Assumptions \ref{asm:quantization_operators}, \ref{asm:sc_general_f}, \ref{asm:lip_max_f_m} hold. Then, for all $t \geq 0$ the iterates produced by \gls{Q-NASTYA} satisfy
	\begin{align*}
		\mathbb{E}_{\cQ}\left[\|g^t\|^2\right] &\leq \frac{2L_{\max}^2\left(1+\frac{\omega}{M}\right)}{Mn}\sum^{M}_{m = 1}\sum^{n-1}_{i = 0}\left\|x^i_{t,m} - x^t\right\|^2\\
        &+ 8L_{\max}\left(1+\frac{\omega}{M}\right)\left(f(x^t) -  f(x^{\star})\right) + \frac{4\omega}{M}\zeta^2_{\star},
	\end{align*}
	where $\mathbb{E}_{\cQ}$ is the expectation w.r.t.\ $\cQ,$ and $\zeta^2_{\star} = \frac{1}{M}\sum^M_{m = 1}\left\|\nabla f_m(x^{\star})\right\|^2$.
	
\end{lemma}

\begin{proof} Using the variance decomposition $\mathbb{E}\left[\|\xi\|^2\right] = \mathbb{E}\left[\|\xi-\mathbb{E}\left[\xi\right]\|^2\right] +\|\mathbb{E}\xi\|^2$, we obtain 
	\begin{eqnarray*}
		&&\mathbb{E}_{\cQ}\left[\|g^t\|^2\right]\\
		&=& \frac{1}{M^2}\sum^M_{m=1}\mathbb{E}_{\cQ}\left[\left\|\cQ\left(\frac{1}{n}\sum^{n-1}_{i = 0}\nabla f_{m,\pi^i_m}(x^i_{t,m})\right) - \frac{1}{n}\sum^{n-1}_{i = 0}\nabla f_{m,\pi^i_m}(x^i_{t,m})\right\|^2\right]\\
		&&+ \left\| \frac{1}{Mn}\sum^M_{m=1}\sum^{n-1}_{i = 0}\nabla f_{m,\pi^i_m}(x^i_{t,m})\right\|^2\\
		&\overset{\text{Asm.} \ref{asm:quantization_operators}}{\leq}&\frac{\omega}{M^2}\sum^M_{m=1}\left\|\frac{1}{n}\sum^{n-1}_{i = 0}\nabla f_{m,\pi^i_m}(x^i_{t,m})\right\|^2   + \left\| \frac{1}{Mn}\sum^M_{m =1}\sum^{n-1}_{i = 0}\nabla f_{m,\pi^i_m}(x^i_{t,m})\right\|^2.
	\end{eqnarray*}
	Next, we use $\nabla f_m(x^t) = \frac{1}{n}\sum^{n-1}_{i=0}\nabla f_{m,\pi^i_m}(x^t)$ and $\|a+b\|^2 \leq 2\|a\|^2 + 2\|b\|^2$:  
	\begin{eqnarray*}
		\mathbb{E}_{\cQ}\left[\|g^t\|^2\right]
		&\leq& \frac{2\omega}{M^2}\sum^M_{m=1}\left\|\frac{1}{n}\sum^{n-1}_{i = 0}\left(\nabla f_{m,\pi^i_m}(x^i_{t,m}) - \nabla f_{m,\pi^i_m}(x^t)\right)\right\|^2\\
        &&+ \frac{2\omega}{M^2}\sum^M_{m=1}\left\|\nabla f_m(x^t)\right\|^2 \\
		&& + 2\left\| \frac{1}{Mn}\sum^M_{m=1}\sum^{n-1}_{i = 0}\left(\nabla f_{m,\pi^i_m}(x^i_{t,m}) - \nabla f_{m,\pi^i_m}(x^{t}) \right)\right\|^2\\
        &&+ 2\left\| \frac{1}{M}\sum^M_{m=1}\nabla f_m(x^{t})\right\|^2\\
		&\leq& \frac{2\left(1+\frac{\omega}{M}\right)}{M}\sum_{m = 1}^M\left\|\frac{1}{n}\sum^{n-1}_{i = 0}\left(\nabla f_{m,\pi^i_m}(x^i_{t,m}) - \nabla f_{m,\pi^i_m}(x^t)\right)\right\|^2\\
		&&+ \frac{2\omega}{M^2}\sum_{m = 1}^M\left\|\nabla f_m(x^t)\right\|^2 + 2\left\| \nabla f(x^{t})\right\|^2.
	\end{eqnarray*}
	Using $L_{i,m}$-smoothness of $f_{m, i}$ and $f$ and also convexity of $f_m$, we obtain
	\begin{eqnarray*}
		\mathbb{E}_{\cQ}\left[\|g^t\|^2\right]
		&\leq& \frac{2\left(1+\frac{\omega}{M}\right)}{Mn}\sum_{m = 1}^M\sum^{n-1}_{i = 0}\left\|\nabla f_{m,\pi^i_m}(x^i_{t,m}) - \nabla f_{m,\pi^i_m}(x^t)\right\|^2\\
        &&+ \frac{4\omega}{M^2}\sum_{m = 1}^M\left\|\nabla f_m(x^t) -\nabla f_m(x^{\star})\right\|^2 \\
		&&+ \frac{4\omega}{M^2}\sum_{m = 1}^M\left\|\nabla f_m(x^{\star})\right\|^2 + 2\left\| \nabla f(x^{t}) - \nabla f(x^{\star})\right\|^2\\
		&\leq& \frac{2L_{\max}^2\left(1+\frac{\omega}{M}\right)}{Mn}\sum_{m = 1}^M\sum^{n-1}_{i = 0}\left\|x^i_{t,m} - x^t\right\|^2\\
        &&+ \frac{8L_{\max}\left(1+\frac{\omega}{M}\right)}{M}\sum_{m = 1}^M D_{f_m}(x^t, x^{\star}) + \frac{4\omega}{M}\zeta_{\star}^2.
	\end{eqnarray*}
\end{proof}

\begin{lemma}[see \citep{malinovsky2023server}]
	\label{lem_inner_product}
	Under Assumptions \ref{asm:quantization_operators}, \ref{asm:sc_general_f}, \ref{asm:lip_max_f_m}, it holds
	\begin{align*}
		&-\frac{1}{Mn}\sum^{M}_{m = 1}\sum^{n-1}_{i = 0}\left\la f_{m,\pi^i_m}(x^i_{t,m}),x^t-x^{\star}\right\ra\\
        &\leq -\frac{\mu}{4}\|x^t - x^{\star}\|^2- \frac{1}{2}\left(f(x^t)-f(x^{\star})\right) + \frac{L_{\max}}{2Mn}\sum^{M}_{m = 1}\sum^{n-1}_{i = 0}\left\|x^i_{t,m} - x^t\right\|^2.
	\end{align*}
\end{lemma}

\begin{lemma}[see \citep{malinovsky2023server}]
	\label{lem_dist}
	Under Assumptions \ref{asm:quantization_operators}, \ref{asm:sc_general_f}, \ref{asm:lip_max_f_m} and $\gamma\leq \frac{1}{2L_{\max}n}$, it holds
	\begin{equation}
		\frac{1}{Mn}\sum^{M}_{m = 1}\sum^{n-1}_{i = 0}\left\|x^i_{t,m} - x^t\right\|^2 \leq 8\gamma^2n^2L_{\max}\left(f(x^t)-f(x^{\star})\right) + 2\gamma^2n\left(\sigma^2_{\star} + (n+1)\zeta^2_{\star}\right).\notag
	\end{equation}
\end{lemma}

\begin{theorem}
	\label{th_convergence_Q_RR}
	Let Assumptions \ref{asm:quantization_operators}, \ref{asm:sc_general_f}, \ref{asm:lip_max_f_m} hold and stepsizes $\gamma$, $\eta$ satisfy
	\begin{equation}
		\label{step_sizes}
		0 < \eta \leq \frac{1}{16L_{\max}\left(1+\frac{\omega}{M}\right)}, \quad 0 < \gamma \leq \frac{1}{5nL_{\max}}.
	\end{equation}
	Then, for all $T \geq 0$ the iterates produced by \gls{Q-NASTYA} satisfy
	\begin{eqnarray*}
		\mathbb{E}\left[\|x^T-x^{\star}\|^2\right]
		&\leq& \left(1-\frac{\eta\mu}{2}\right)^T\|x^0 - x^{\star}\|^2\\
        &&+\frac{9}{2}\frac{\gamma^2nL_{\max}}{\mu}\left(\sigma^2_{\star} +(n+1)\zeta^2_{\star}\right) +  8\frac{\eta\omega}{\mu M}\zeta^2_{\star}.
	\end{eqnarray*}
\end{theorem}
\begin{proof} Taking the expectation w.r.t.\ $\cQ$ and using Lemma \ref{lem_norm_g^t}, we get 
\begin{eqnarray*}
	&&\mathbb{E}_{\cQ}\left[\|x^{t+1} - x^{\star}\|^2\right]\\
	&=& \|x^t - x^{\star}\|^2 -2\eta\mathbb{E}_{\cQ}\left[\la g^t, x^t- x^{\star}\ra\right] + \eta^2\mathbb{E}_{\cQ}\left[\|g^t\|^2\right]\\
	&\leq& \|x^t - x^{\star}\|^2 -2\eta\mathbb{E}_{\cQ}\left[\left\la \frac{1}{M}\sum_{m=1}^M\cQ\left(\frac{1}{n}\sum^{n-1}_{i = 0}\nabla f_{m,\pi^i_m}(x^i_{t,m})\right), x^t- x^{\star}\right\ra\right]\\
	&& + \frac{2\eta^2L^2_{\max}\left(1+\frac{\omega}{M}\right)}{Mn}\sum^{M}_{m = 1}\sum^{n-1}_{i = 0}\left\|x^i_{t,m} - x^t\right\|^2\\
	&& + 8\eta^2L_{\max}\left(1+\frac{\omega}{M}\right)\left(f(x^t) -  f(x^{\star})\right) + 4\eta^2\frac{\omega}{M}\zeta^2_{\star}\\
	&\leq& \|x^t - x^{\star}\|^2 -2\eta\frac{1}{Mn}\sum^M_{m = 1}\sum^{n-1}_{i = 0}\left\la \nabla f_{m,\pi^i_m}(x^i_{t,m}), x^t- x^{\star}\right\ra\\
	&& + \frac{2\eta^2L^2_{\max}\left(1+\frac{\omega}{M}\right)}{Mn}\sum^{M}_{m = 1}\sum^{n-1}_{i = 0}\left\|x^i_{t,m} - x^t\right\|^2\\
	&& + 8\eta^2L_{\max}\left(1+\frac{\omega}{M}\right)\left(f(x^t) -  f(x^{\star})\right) + 4\eta^2\frac{\omega}{M}\zeta^2_{\star}.
\end{eqnarray*}
Next, Lemma \ref{lem_inner_product} implies 
\begin{eqnarray*}
	&&\mathbb{E}_{\cQ}\left[\|x^{t+1} - x^{\star}\|^2\right]\\ 
	&\leq&  \|x^t - x^{\star}\|^2  - \frac{\eta\mu}{2}\|x^t - x^{\star}\|^2 - \eta\left(f(x^t)-f(x^{\star})\right)\\
	&&+ 8\eta^2L_{\max}\left(1+\frac{\omega}{M}\right)\left(f(x^t) -  f(x^{\star})\right) + \frac{\eta L_{\max}}{Mn}\sum^{M}_{m = 1}\sum^{n-1}_{i = 0}\left\|x^i_{t,m} - x^t\right\|^2\\
	&& + \frac{2\eta^2L^2_{\max}\left(1+\frac{\omega}{M}\right)}{Mn}\sum^{M}_{m = 1}\sum^{n-1}_{i = 0}\left\|x^i_{t,m} - x^t\right\|^2 + 4\eta^2\frac{\omega}{M}\zeta^2_{\star}\\
	&\leq& \left(1 -\frac{\eta\mu}{2}\right)\|x^t - x^{\star}\|^2 - \eta\left(1-8\eta L_{\max}\left(1+\frac{\omega}{M}\right)\right)\left(f(x^t) -  f(x^{\star})\right) \\
	&& + \frac{\eta L_{\max}\left(1+ 2\eta L_{\max}\left(1+\frac{\omega}{M}\right)\right)}{Mn}\sum^{M}_{m = 1}\sum^{n-1}_{i = 0}\left\|x^i_{t,m} - x^t\right\|^2 + 4\eta^2\frac{\omega}{M}\zeta^2_{\star}.
\end{eqnarray*}
Using Lemma \ref{lem_dist}, we get 
\begin{eqnarray*}
	&&\mathbb{E}_{\cQ}\left[\|x^{t+1} - x^{\star}\|^2\right] \\
	&\leq& \left(1 -\frac{\eta\mu}{2}\right)\|x^t - x^{\star}\|^2 - \eta\left(1-8\eta L\left(1+\frac{\omega}{M}\right)\right)\left(f(x^t) -  f(x^{\star})\right)   \\
	&& + \eta L_{\max}\left(1+ 2\eta L_{\max}\left(1+\frac{\omega}{M}\right)\right) \cdot 8\gamma^2n^2L_{\max}\left(f(x^t)-f(x^{\star})\right)\\
	&& + \eta L_{\max}\left(1+ 2\eta L_{\max}\left(1+\frac{\omega}{M}\right)\right) \cdot 2\gamma^2n\left(\sigma^2_{\star} + (n+1)\zeta^2_{\star}\right) \\
	&&  + 4\eta^2\frac{\omega}{M}\zeta^2_{\star}.
\end{eqnarray*}
In view of \eqref{step_sizes}, we have
\begin{eqnarray*}
	&&\mathbb{E}_{\cQ}\left[\|x^{t+1} - x^{\star}\|^2\right] \\
	&\leq& \left(1 -\frac{\eta\mu}{2}\right)\|x^t - x^{\star}\|^2 + 4\eta^2\frac{\omega}{M}\zeta^2_{\star}  \\
	&& - \eta\left(1-8\eta L_{\max}\left(1+\frac{\omega}{M}\right) - 8\gamma^2n^2L^2_{\max}\left(1+2L_{\max}\eta \left(1 +\frac{\omega}{M}\right)\right)\right)\times\\
    &&\times\left(f(x^t) -  f(x^{\star})\right)  \\
	&& + 2\gamma^2n\eta L_{\max}\left(1+ 2\eta L\left(1+\frac{\omega}{M}\right)\right) \left(\sigma^2_{\star} + n\zeta^2_{\star}\right)\\
	&\leq& \left(1 -\frac{\eta\mu}{2}\right)\|x^t - x^{\star}\|^2 + 4\eta^2\frac{\omega}{M}\zeta^2_{\star} + \frac{9}{4}\eta L_{\max}\gamma^2n \left(\sigma^2_{\star} + (n+1)\sigma^2_{\star}\right).
\end{eqnarray*}
Recursively unrolling the inequality and using $\sum\limits^{+\infty}_{t = 0}\left(1-\frac{\eta\mu}{2}\right)^t \leq \frac{2}{\mu\eta}$, we get the result.
\end{proof}

\begin{corollary}
	\label{cor_convergence_Q_NASTYA}
	Let the assumptions of Theorem~\ref{thm:convergence_Q_NASTYA} hold, \(\gamma = \nicefrac{\eta}{n}\), and 
	\begin{equation}
		\label{new_gamma_Q_NASTAY}
		\eta = \min\left\{ \frac{1}{16L_{\max}\left(1+\frac{\omega}{M}\right)}, \sqrt{\frac{\varepsilon\mu n}{9L_{\max}}}\left( (n+1)\zeta^2_{\star} + \sigma^2_{\star}\right)^{-\nicefrac{1}{2}}, \frac{\varepsilon\mu M}{24\omega\zeta^2_{\star}}\right\}.
	\end{equation}
	Then, \gls{Q-NASTYA} finds a solution with accuracy $\varepsilon > 0$ after the following number of communication rounds:
	\begin{equation*}
		\widetilde{\cO}\left(\frac{L_{\max}}{\mu}\left(1+\frac{\omega}{M}\right)+ \frac{\omega}{M}\frac{\zeta^2_{\star}}{\varepsilon\mu^3}+\sqrt{\frac{ L_{\max} }{\varepsilon\mu^3}} \sqrt{\zeta^2_{\star}+\nicefrac{\sigma_{\star}^2}{n}}\right).
	\end{equation*}
	If $\gamma \rightarrow 0$, one can choose $\eta = \min\left\{\frac{1}{16L_{\max}\left(1+\frac{\omega}{M}\right)},\frac{\varepsilon\mu M}{24\omega\zeta^2_{\star}}\right\}$ such that the above complexity bound improves to
	\begin{equation*}
		\widetilde{\cO}\left( \frac{L_{\max}}{\mu}\left(1+\frac{\omega}{M}\right)+\frac{\omega}{M}\frac{\zeta^2_{\star}}{\varepsilon\mu^3}\right).
	\end{equation*}
\end{corollary}
\begin{proof}
	Theorem~\ref{thm:convergence_Q_NASTYA} implies
	\begin{equation}
		\mathbb{E}\left[\|x^T-x^{\star}\|^2\right] \leq \left(1-\frac{\eta\mu}{2}\right)^T\|x^0 - x^{\star}\|^2 +\frac{9}{2}\frac{\gamma^2nL_{\max}}{\mu}\left((n+1)\zeta^2_{\star} + \sigma_{\star}^2\right) +  8\frac{\eta\omega}{\mu M}\zeta^2_{\star}.\notag
	\end{equation}
	To estimate the number of communication rounds required to find a solution with accuracy $\varepsilon >0$, we need to upper bound each term from the right-hand side by $\nicefrac{\varepsilon}{3}$. Thus, we get additional conditions on $\eta$:
	\begin{equation*}
		\frac{9}{2}\frac{\eta^2 L_{\max}}{n\mu}\left( (n+1)\zeta^2_{\star} + \sigma^2_{\star}\right) <\frac{\varepsilon}{3},\quad 8\frac{\eta\omega}{\mu M}\zeta^2_{\star} < \frac{\varepsilon}{3}
	\end{equation*}
	and also the upper bound on the number of communication rounds $T$
	\begin{equation*}
		T = \widetilde{\cO}\left(\frac{1}{\eta\mu}\right).
	\end{equation*}
	Substituting \eqref{new_gamma_DIANA_NASTAY} in the previous equation, we get the first part of the result. 
	When $\gamma \rightarrow 0$, the proof follows similar steps. 
\end{proof}

\section{Missing Proofs for DIANA-NASTYA}

\begin{lemma}
	\label{rr_diana_inner_product}
	Under Assumptions \ref{asm:quantization_operators}, \ref{asm:sc_general_f}, \ref{asm:lip_max_f_m}, the iterates produced by \gls{DIANA-NASTYA} satisfy
	\begin{eqnarray*}
		-\mathbb{E}_{\cQ}\left[\frac{1}{M}\sum^M_{m=1}\left\la \hat{g}_{t,m} - h^{\star}, x^t - x^{\star}\right\ra\right] 
		&\leq& -\frac{\mu}{4}\|x^t - x^{\star}\|^2 -\frac{1}{2}\left(f(x^t) - f(x^{\star})\right)\\
		&& - \frac{1}{Mn}\sum^M_{m=1}\sum^{n-1}_{i=0} D_{f_{m,\pi^i_m}}(x^{\star}, x^i_{t,m})\\
		&& + \frac{L_{\max}}{2Mn}\sum^M_{m=1}\sum^{n-1}_{i=0}\|x^t - x^i_{t,m}\|^2,
	\end{eqnarray*}
	where $h^{\star} = \nabla f(x^{\star})$.
\end{lemma}
\begin{proof} Using that $\mathbb{E}_{\cQ}\left[\hat{g}_{t,m}\right] = g^{t}_{m}$ and definition of $h^{\star}$, we get
	\begin{eqnarray*}
		&-&\mathbb{E}_{\cQ}\left[\frac{1}{M}\sum^M_{m=1}\left\la \hat{g}_{t,m} - h^{\star}, x^t - x^{\star}\right\ra\right] \\
		&=& -\frac{1}{M}\sum^M_{m=1}\left\la g^{t}_{m} - h^{\star}, x^t - x^{\star}\right\ra\\
		&=& -\frac{1}{Mn}\sum^M_{m=1}\sum^{n-1}_{i=0}\left\la \nabla f_{m,\pi^i_m}(x^i_{t,m}) - \nabla f_{m,\pi^i_m}(x^{\star}), x^t - x^{\star}\right\ra.
	\end{eqnarray*}
	Next, three-point identity and $L_{\max}$-smoothness of each function $f^{i}_m$ imply 
	\begin{eqnarray*}
		&-&\mathbb{E}_{\cQ}\left[\frac{1}{M}\sum^M_{m=1}\left\la \hat{g}_{t,m} - h^{\star}, x^t - x^{\star}\right\ra\right] \\
		&=& -\frac{1}{Mn}\sum^M_{m=1}\sum^{n-1}_{i=0}\left( D_{f_{m,\pi^i_m}}(x^t, x^{\star})+ D_{f_{m,\pi^i_m}}( x^{\star}, x^{t}_{m,i}) - D_{f_{m,\pi^i_m}}(x^t, x^{t}_{m,i})\right)\\
		&\leq& - D_{f}(x^t, x^{\star}) - \frac{1}{Mn}\sum^M_{m=1}\sum^{n-1}_{i=0} D_{f_{m,\pi^i_m}}(x^{\star}, x^i_{t,m})\\
		&& + \frac{L_{\max}}{2Mn}\sum^M_{m=1}\sum^{n-1}_{i=0} \|x^t - x^{t}_{m,i}\|^2
	\end{eqnarray*}
	Finally, using $\mu$-strong convexity of $f$, we finish the proof of the lemma.
\end{proof}

\begin{lemma}
	\label{rr_diana_dinst}
	Under Assumptions \ref{asm:quantization_operators}, \ref{asm:sc_general_f}, \ref{asm:lip_max_f_m}, the iterates produced by \gls{DIANA-NASTYA} satisfy
	\begin{eqnarray*}
		\mathbb{E}_{\cQ}\left[\|\hat{g}^t - h^{\star}\|^2\right] 
		&\leq& \frac{2L^2_{\max}\left(1 + \frac{\omega}{M}\right)}{Mn}\sum^M_{m=1}\sum^{n-1}_{i=0}\|x^{t}_{m,i} - x^t\|^2\\
        &&+ 8L_{\max}\left(1+\frac{\omega}{M}\right)\left(f(x^t) - f(x^{\star})\right)\\ 
		&&+ \frac{4\omega}{M^2}\sum^M_{m=1}\|h^{t}_{m}-h^{\star}_m\|^2.
	\end{eqnarray*}
\end{lemma}
\begin{proof}
	Since $g^t = \frac{1}{M}\sum^M_{m=1}g^{t}_{m}$ and $\mathbb{E}\|\xi - c\|^2 = \mathbb{E}\|\xi - \mathbb{E}\xi\|^2 + \mathbb{E}\|\mathbb{E}\xi - c\|^2$, we have
	\begin{eqnarray*}
		\mathbb{E}_{\cQ}\left[\|\hat{g}^t - h^{\star}\|^2\right] 
		&=& \mathbb{E}_{\cQ}\left[\left\|\frac{1}{M}\sum^M_{m=1}\left( h^{t}_{m} + \cQ\left(g^{t}_{m} - h^{t}_{m}\right) - h^{\star}_m\right)\right\|^2\right] \\
		&=& \mathbb{E}_{\cQ}\left[\left\|\frac{1}{M}\sum^M_{m=1}\left( h^{t}_{m} + \cQ\left(g^{t}_{m} - h^{t}_{m}\right)\right) - g^t\right\|^2\right]+ \left\|g^t-h^{\star}\right\|^2.
	\end{eqnarray*}
	Next, independence of $\cQ\left(g^{t}_{m} - h^{t}_{m}\right)$, $m \in M$, Assumption \ref{asm:quantization_operators}, and $L_{\max}$ smoothness  and convexity of each function $f^{i}_{m}$ imply
	\begin{eqnarray*}
		&&\mathbb{E}_{\cQ}\left[\|\hat{g}^t - h^{\star}\|^2\right] \\
		&\leq& \frac{\omega}{M^2}\sum^M_{m=1}\left\| g^{t}_{m} - h^{t}_{m}\right\|^2 + \left\|g^t-h^{\star}\right\|^2\\
		&\leq&\frac{2\omega}{M^2}\sum^M_{m=1}\left\| \frac{1}{n}\sum^{n-1}_{i=0}\nabla f_{m,\pi^i_m}(x^i_{t,m}) - \nabla f_m(x^t)\right\|^2\\
        &&+ \frac{2\omega}{M^2}\sum^M_{m=1}\left\|\nabla f_m(x^t) - h^{t}_{m}\right\|^2+ 2\left\|g^t-\nabla f(x^t)\right\|^2 + 2\left\|\nabla f(x^t)-h^{\star}\right\|^2\\
		&\leq& \frac{2\omega}{M^2}\sum^M_{m=1}\left\| \frac{1}{n}\sum^{n-1}_{i=0}\nabla f_{m,\pi^i_m}(x^i_{t,m}) - \nabla f_m(x^t)\right\|^2+ \frac{2\omega}{M^2}\sum^M_{m=1}\left\|\nabla f_m(x^t) - h^{t}_{m}\right\|^2\\
		&&+ 2\left\|\frac{1}{M}\sum^M_{m=1}\left(\frac{1}{n}\sum^{n-1}_{i=0}\nabla f_{m,\pi^i_m}(x^i_{t,m}) - \nabla f_m(x^t)\right)\right\|^2+ 2\left\|\nabla f(x^t)-h^{\star}\right\|^2\\
		&\leq& \frac{2L^2_{\max}\left(1+\frac{\omega}{M}\right)}{Mn}\sum^M_{m=1}\sum^{n-1}_{i=0}\|x^i_{t,m} - x^t\|^2\\
        &&+ \frac{2\omega}{M^2}\sum^M_{m=1}\|\nabla f_m(x^t) - h^{t}_{m}\|^2  + 2\left\|\nabla f(x^t)-h^{\star}\right\|^2.
	\end{eqnarray*}
	Using $L_{\max}$-smoothness and convexity of $f_m$, we get
	\begin{eqnarray*}
		\mathbb{E}_{\cQ}\left[\|\hat{g}^t - h^{\star}\|^2\right] 
		&\leq& \frac{2L^2_{\max}\left(1+\frac{\omega}{M}\right)}{Mn}\sum^M_{m=1}\sum^{n-1}_{i=0}\|x^i_{t,m} - x^t\|^2\\
        &&+ \frac{2\omega}{M^2}\sum^M_{m=1}\|\nabla f_m(x^t) - h^{t}_{m}\|^2\\
		&&+ 4L_{\max}\left(f(x^t)-f(x^{\star})\right)\\
		&\leq& \frac{2L^2_{\max}\left(1+\frac{\omega}{M}\right)}{Mn}\sum^M_{m=1}\sum^{n-1}_{i=0}\|x^i_{t,m} - x^t\|^2\\
        &&+ \frac{4\omega}{M^2}\sum^M_{m=1}\|\nabla f_m(x^t) - h^{\star}_{m}\|^2\\
		&&+\frac{4\omega}{M^2}\sum^M_{m=1}\|h^{t}_{m}- h^{\star}_{m}\|^2+ 4L_{\max}\left(f(x^t)-f(x^{\star})\right)\\
		&\leq& \frac{2L^2_{\max}\left(1+\frac{\omega}{M}\right)}{Mn}\sum^M_{m=1}\sum^{n-1}_{i=0}\|x^i_{t,m} - x^t\|^2\\
        &&+ \frac{8L_{\max}\omega}{M^2}\sum^M_{m=1}D_{f_m}(x^t, x^{\star})\\
		&&+\frac{4\omega}{M^2}\sum^M_{m=1}\|h^{t}_{m}- h^{\star}_{m}\|^2+ 4L_{\max}\left(f(x^t)-f(x^{\star})\right).
	\end{eqnarray*}
\end{proof}

\begin{lemma}
	\label{rr_diana_descent_lemma_for_h}
	Under Assumptions \ref{asm:quantization_operators}, \ref{asm:sc_general_f}, \ref{asm:lip_max_f_m}, and $\alpha\leq\frac{1}{1+\omega}$, the iterates produced by \gls{DIANA-NASTYA} satisfy 
	\begin{eqnarray*}
		\frac{1}{M}\sum^M_{m=1}\mathbb{E}_{\cQ}\left[\|h^{t+1}_{m} - h^{\star}_m\|^2\right] &\leq& \frac{1-\alpha}{M}\sum^M_{m=1}\|h^{t}_{m} - h^{\star}_m\|^2\\
        &&+ \frac{2\alpha L^2_{\max}}{Mn}\sum^M_{m=1}\sum^{n-1}_{i=0}\|x^i_{t,m} - x^t\|^2 \\
		&&+ 4\alpha L_{\max}\left(f(x^t) - f(x^{\star})\right).
	\end{eqnarray*}
\end{lemma}
\begin{proof}
	Taking the expectation w.r.t.\ $\cQ$ and using Assumption \ref{asm:quantization_operators}, we obtain
	\begin{eqnarray*}
		&&\frac{1}{M}\sum^M_{m=1}\mathbb{E}_{\cQ}\left[\|h^{t+1}_{m} - h^{\star}_m\|^2\right]\\ &=& \frac{1}{M}\sum^M_{m=1}\mathbb{E}_{\cQ}\left[\|h^{t}_{m} +\alpha\cQ(g^{t}_{m}-h^{t}_{m})- h^{\star}_m\|^2\right]\\
		&\leq& \frac{1}{M}\sum^M_{m=1}\left(\|h^{t}_{m}- h^{\star}_m\|^2 +2\alpha\mathbb{E}_{\cQ}\left[\left\la\cQ(g^{t}_{m}-h^{t}_{m}),h^{t}_{m}- h^{\star}_m\right\ra\right]\right) \\
		&&+ \alpha^2 \frac{1}{M}\sum^M_{m=1}\mathbb{E}_{\cQ}\left[\| \cQ(g^{t}_{m}-h^{t}_{m})\|^2\right]\\
		&\leq& \frac{1}{M}\sum^M_{m=1}\left(\|h^{t}_{m}- h^{\star}_m\|^2 +2\alpha\left\la g^{t}_{m}-h^{t}_{m},h^{t}_{m}- h^{\star}_m\right\ra\right) \\
		&&+ \alpha^2(1+\omega) \frac{1}{M}\sum^M_{m=1}\| g^{t}_{m}-h^{t}_{m}\|^2
	\end{eqnarray*}
	Using $\alpha \leq \frac{1}{1+\omega}$, we get
	\begin{eqnarray*}
		&&\frac{1}{M}\sum^M_{m=1}\mathbb{E}_{\cQ}\left[\|h^{t+1}_{m} - h^{\star}_m\|^2\right]\\ 
		&\leq& \frac{1}{M}\sum^M_{m=1}\left(\|h^{t}_{m}- h^{\star}_m\|^2 + \alpha\left\la g^{t}_{m}-h^{t}_{m},h^{t}_{m} +g^{t}_{m} - 2h^{\star}_m\right\ra\right) \\
		&\leq& \frac{1}{M}\sum^M_{m=1}\left(\|h^{t}_{m}- h^{\star}_m\|^2 + \alpha\|g^{t}_{m} - h^{\star}_m\|^2 - \alpha\|h^{t}_{m}-h^{\star}_m\|^2\right)\\
		&\leq& \frac{1-\alpha}{M}\sum^M_{m=1}\|h^{t}_{m}- h^{\star}_m\|^2 + \frac{\alpha}{M}\sum^M_{m=1}\|g^{t}_{m} - h^{\star}_m\|^2.
	\end{eqnarray*}
	Finally, $L_{\max}$-smoothness and convexity of $f_m$ imply
	\begin{eqnarray*}
		&&\frac{1}{M}\sum^M_{m=1}\mathbb{E}_{\cQ}\left[\|h^{t+1}_{m} - h^{\star}_m\|^2\right]\\
		&\leq& \frac{1-\alpha}{M}\sum^M_{m=1}\|h^{t}_{m}- h^{\star}_m\|^2\\
		&&+\frac{2\alpha}{M}\sum^M_{m=1}\left(\|g^{t}_{m} - \nabla f_{m}(x^t)\|^2 + \|\nabla f_{m}(x^t) - h^{\star}_m\|^2\right)\\
		&\leq& \frac{1-\alpha}{M}\sum^M_{m=1}\|h^{t}_{m}- h^{\star}_m\|^2 + \frac{4L_{\max}\alpha}{M}\sum^M_{m=1}D_{f_{m}}(x^t, x^{\star})\\
		&& + \frac{2\alpha}{M}\sum^M_{m=1}\left\|\frac{1}{n}\sum^{n-1}_{i=0}(\nabla f_{m,\pi^i_m}(x^{t}_{m,i}) - \nabla f_{m, i}(x^t))\right\|^2\\
		&\leq& \frac{1-\alpha}{M}\sum^M_{m=1}\|h^{t}_{m}- h^{\star}_m\|^2 + 4L_{\max}\alpha\left(f(x^t)-f(x^{\star})\right) \\
		&&+ \frac{2L^2_{\max}\alpha}{Mn}\sum^M_{m=1}\sum^{n-1}_{i=0}\left\|x^{t}_{m,i}-x^t\right\|^2.
	\end{eqnarray*}
\end{proof}

\begin{theorem}
	\label{rr_diana_conv_th}
	Let Assumptions \ref{asm:quantization_operators}, \ref{asm:sc_general_f}, \ref{asm:lip_max_f_m} hold and stepsizes $\gamma$, $\eta, \alpha$ satisfy
	\begin{equation}
		\label{step_sizes_diana_nastya}
		0 < \gamma \leq  \frac{1}{16L_{\max}n},\quad 0 < \eta \leq \min\left\{\frac{\alpha}{2\mu}, \frac{1}{16L_{\max}\left(1+\frac{9\omega}{M}\right)}\right\},\quad \alpha \leq \frac{1}{1+\omega}.
	\end{equation}
	Then, for all $T \geq 0$ the iterates produced by \gls{DIANA-NASTYA} satisfy
	\begin{equation}
		\mathbb{E}\left[\Psi^{(t)}\right] \leq \left(1-\frac{\eta\mu}{2}\right)^T\Psi^0 + \frac{9}{2}\frac{\gamma^2 nL}{\mu}\left( \sigma^2_{\star} + (n+1)\zeta^2_{\star}\right).
	\end{equation}
\end{theorem}
\begin{proof}
We have
\begin{eqnarray*}
	\|x^{t+1}-x^{\star}\|^2 
	&=& \|x^t -\eta \hat{g}^t - x^{\star} + \eta h^{\star}\|^2\\
	&=& \|x^t - x^{\star}\|^2 - 2\eta\la\hat{g}^t - h^{\star},x^t-x^{\star}\ra + \eta^2 \|\hat{g}^t - h^{\star}\|^2.
\end{eqnarray*}
Taking the expectation w.r.t.\ $\cQ$ and using Lemma \ref{rr_diana_inner_product}, we obtain
\begin{eqnarray*}
	\mathbb{E}_{\cQ}\left[\|x^{t+1}-x^{\star}\|^2\right]
	&=& \|x^t - x^{\star}\|^2 - 2\eta\mathbb{E}_{\cQ}\left[\la\hat{g}^t - h^{\star},x^t-x^{\star}\ra\right]\\
    &&+ \eta^2 \mathbb{E}_{\cQ}\left[\|\hat{g}^t - h^{\star}\|^2\right]\\
	&\leq& \left(1-\frac{\eta\mu}{2}\right)\|x^t - x^{\star}\|^2 - \eta(f(x^t)-f(x^{\star}))\\
    &&- \frac{2\eta}{Mn}\sum^M_{m=1}\sum^{n-1}_{i=0}D_{f_{m,\pi^i_m}}(x^{\star}, x^i_{t,m}) \\
	&&+\frac{L_{\max}\eta}{Mn}\sum^M_{m=1}\sum^{n-1}_{i=0}\|x^i_{t,m} - x^t\|^2 + \eta^2 \mathbb{E}_{\cQ}\left[\|\hat{g}^t - h^{\star}\|^2\right].
\end{eqnarray*}
Next, Lemma \ref{rr_diana_dinst} implies
\begin{eqnarray*}
	&&\mathbb{E}_{\cQ}\left[\|x^{t+1}-x^{\star}\|^2\right]\\
	&\leq& \left(1-\frac{\eta\mu}{2}\right)\|x^t - x^{\star}\|^2 - \eta(f(x^t)-f(x^{\star}))\\
    &&- \frac{2\eta}{Mn}\sum^M_{m=1}\sum^{n-1}_{i=0}D_{f_{m,\pi^i_m}}(x^{\star}, x^i_{t,m}) \\
	&&+\frac{L_{\max}\eta}{Mn}\sum^M_{m=1}\sum^{n-1}_{i=0}\|x^i_{t,m} - x^t\|^2\\
    &&+ \frac{2\eta^2L^2_{\max}\left(1 + \frac{\omega}{M}\right)}{Mn}\sum^M_{m=1}\sum^{n-1}_{i=0}\|x^{t}_{m,i} - x^t\|^2\\
	&& + \eta^2\left( 8L_{\max}\left(1+\frac{\omega}{M}\right)\left(f(x^t) - f(x^{\star})\right)+ \frac{4\omega}{M^2}\sum^M_{m=1}\|h^{t}_{m}-h^{\star}_m\|^2\right)\\
	&\leq& \left(1-\frac{\eta\mu}{2}\right)\|x^t - x^{\star}\|^2 - \eta\left(1- 8\eta L_{\max}\left(1+\frac{\omega}{M}\right)\right)\left(f(x^t) - f(x^{\star})\right)\\
	&&+ L_{\max}\eta\left(1 +2\eta L_{\max}\left(1+\frac{\omega}{M}\right)\right) \frac{1}{Mn}\sum^M_{m=1}\sum^{n-1}_{i=0}\|x^{t}_{m,i} - x^t\|^2\\
	&&- \frac{2\eta}{Mn}\sum^M_{m=1}\sum^{n-1}_{i=0}D_{f_{m,\pi^i_m}}(x^{\star}, x^i_{t,m}) + \frac{4\eta^2\omega}{M^2}\sum^M_{m=1}\|h^{t}_{m}-h^{\star}_m\|^2.
\end{eqnarray*}
Using \eqref{lyapunov_func_rr_diana} and Lemma \ref{rr_diana_descent_lemma_for_h}, we get
\begin{eqnarray*}
	&&\mathbb{E}_{\cQ}\left[\Psi^{(t+1)}\right]\\
	&\leq& \left(1-\frac{\eta\mu}{2}\right)\|x^t - x^{\star}\|^2 - \eta\left(1- 8\eta L_{\max}\left(1+\frac{\omega}{M}\right)\right)\left(f(x^t) - f(x^{\star})\right)\\
	&&+ L_{\max}\eta\left(1 +2\eta L_{\max}\left(1+\frac{\omega}{M}\right)\right) \frac{1}{Mn}\sum^M_{m=1}\sum^{n-1}_{i=0}\|x^{t}_{m,i} - x^t\|^2\\
	&&- \frac{2\eta}{Mn}\sum^M_{m=1}\sum^{n-1}_{i=0}D_{f_{m,\pi^i_m}}(x^{\star}, x^i_{t,m}) + \frac{4\eta^2\omega}{M^2}\sum^M_{m=1}\|h^{t}_{m}-h^{\star}_m\|^2\\
	&& + c\eta^2\left(\frac{1-\alpha}{M}\sum^M_{m=1}\|h^{t}_{m} - h^{\star}_m\|^2 + \frac{2\alpha L^2_{\max}}{Mn}\sum^M_{m=1}\sum^{n-1}_{i=0}\|x^i_{t,m} - x^t\|^2 \right)\\
    &&+4c\eta^2 \alpha L_{\max}\left(f(x^t) - f(x^{\star})\right)\\
	&\leq& \left(1-\frac{\eta\mu}{2}\right)\|x^t - x^{\star}\|^2 + \eta^2\left(c(1-\alpha)+\frac{4\omega}{M}\right)\frac{1}{M}\sum^M_{m=1}\|h^{t}_{m}-h^{\star}_m\|^2 \\
	&&- \eta\left(1- 8\eta L_{\max}\left(1+\frac{\omega}{M}\right) - 4\alpha\eta cL_{\max}\right)\left(f(x^t) - f(x^{\star})\right)\\
	&&+ L\eta\left(1 +2\eta L_{\max}\left(1+\frac{\omega}{M}\right)+ 2\alpha \eta cL_{\max}\right) \frac{1}{Mn}\sum^M_{m=1}\sum^{n-1}_{i=0}\|x^{t}_{m,i} - x^t\|^2.
\end{eqnarray*}
Taking the full expectation, we derive 
\begin{eqnarray*}
	&&\mathbb{E}\left[\Psi^{(t+1)}\right]\\
	&\leq& \left(1-\frac{\eta\mu}{2}\right)\mathbb{E}\left[\|x^t - x^{\star}\|^2\right]\\
    &&+ \eta^2\left(c(1-\alpha)+\frac{4\omega}{M}\right)\frac{1}{M}\sum^M_{m=1}\mathbb{E}\left[\|h^{t}_{m}-h^{\star}_m\|^2\right] \\
	&&- \eta\left(1- 8\eta L_{\max}\left(1+\frac{\omega}{M}\right) - 4\alpha\eta cL\right)\mathbb{E}\left[f(x^t) - f(x^{\star})\right]\\
	&&+ L_{\max}\eta\left(1 +2\eta L_{\max}\left(1+\frac{\omega}{M}\right)+ 2\alpha \eta cL_{\max}\right)\times\\
    &&\times\frac{1}{Mn}\sum^M_{m=1}\sum^{n-1}_{i=0}\mathbb{E}\left[\|x^{t}_{m,i} - x^t\|^2\right].
\end{eqnarray*}
Using Lemma \ref{lem_dist}, we get
\begin{eqnarray*}
	&&\mathbb{E}\left[\Psi^{(t+1)}\right]\\
	&\leq& \left(1-\frac{\eta\mu}{2}\right)\mathbb{E}\left[\|x^t - x^{\star}\|^2\right]\\
    &&+ \eta^2\left(c(1-\alpha)+\frac{4\omega}{M}\right)\frac{1}{M}\sum^M_{m=1}\mathbb{E}\left[\|h^{t}_{m}-h^{\star}_m\|^2\right] \\
	&&- \eta\left(1- 8\eta L_{\max}\left(1+\frac{\omega}{M}\right) - 4\alpha\eta cL_{\max}\right)\mathbb{E}\left[f(x^t) - f(x^{\star})\right]\\
	&&+ 8\gamma^2n^2L^2_{\max}\eta\left(1 +2\eta L_{\max}\left(1+\frac{\omega}{M}\right)+ 2\alpha \eta cL_{\max}\right)\mathbb{E}\left[f(x^t) - f(x^{\star})\right]\\
	&&+ 2\gamma^2nL_{\max}\eta\left(1 +2\eta L_{\max}\left(1+\frac{\omega}{M}\right)+ 2\alpha \eta cL_{\max}\right) \left(\sigma^2_{\star} + (n+1)\zeta^2_{\star}\right).
\end{eqnarray*}
In view of \eqref{step_sizes_diana_nastya}, we have
\begin{eqnarray*}
	\mathbb{E}\left[\Psi^{(t+1)}\right]
	&\leq& \left(1-\frac{\eta\mu}{2}\right)\mathbb{E}\left[\|x^t - x^{\star}\|^2\right] + \left(1-\frac{\alpha}{2}\right)\frac{c\eta^2}{M}\sum^M_{m=1}\mathbb{E}\left[\|h^{t}_{m}-h^{\star}_m\|^2\right] \\
	&&+ \frac{9}{4}\gamma^2nL_{\max}\eta \left(\sigma^2_{\star} + (n+1)\zeta^2_{\star}\right)
\end{eqnarray*}
Using definition of Lyapunov function and using $\sum\limits^{+\infty}_{t = 0}\left(1-\frac{\eta\mu}{2}\right)^t \leq \frac{2}{\mu\eta}$, we get the result.
\end{proof}

\begin{corollary}
	\label{cor_convergence_DIANA_NASTYA}
	Let the assumptions of Theorem~\ref{thm:diana-nastya-conv} hold, $\gamma = \nicefrac{\eta}{n}$, $\alpha = \frac{1}{1+\omega}$, and
	\begin{equation}
		\label{new_gamma_DIANA_NASTAY}
		\eta = \min\left\{\frac{\alpha}{2\mu}, \frac{1}{16L_{\max}\left(1+\frac{9\omega}{M}\right)}, \sqrt{\frac{\varepsilon\mu n}{9L_{\max}}}\left( (n+1)\zeta^2_{\star} + \sigma^2_{\star}\right)^{-\nicefrac{1}{2}}\right\}.
	\end{equation}
	Then, \gls{DIANA-NASTYA} finds a solution with accuracy $\varepsilon>0$ after the following number of communication rounds: 
	\begin{equation}
		\widetilde{\cO}\left(\omega+\frac{L_{\max}}{\mu}\left(1+\frac{\omega}{M}\right) + \sqrt{\frac{L_{\max}}{\varepsilon\mu^3}}\sqrt{ \zeta^2_{\star} + \nicefrac{\sigma^2_{\star}}{n}}\right).\notag
	\end{equation}
	If $\gamma \rightarrow 0$, one can choose $\eta = \min\left\{\frac{\alpha}{2\mu}, \frac{1}{16L_{\max}\left(1+\frac{9\omega}{M}\right)}\right\}$
	such that the number of communication rounds $T$ to find a solution with accuracy $\varepsilon>0$ is 
	\begin{equation}
		\widetilde{\cO}\left(\omega +\frac{L_{\max}}{\mu}\left(1+\frac{\omega}{M}\right)\right).\notag
	\end{equation}
\end{corollary}
\begin{proof}
	Theorem~\ref{thm:diana-nastya-conv} implies
	\begin{equation}
		\mathbb{E}\left[\Psi^{(t)}\right] \leq \left(1-\frac{\eta\mu}{2}\right)^T\Psi^0 + \frac{9}{2}\frac{\gamma^2 nL_{\max}}{\mu}\left( (n+1)\zeta^2_{\star} + \sigma^2_{\star}\right).\notag
	\end{equation}
	To estimate the number of communication rounds required to find a solution with accuracy $\varepsilon >0$, we need to upper bound each term from the right-hand side by $\frac{\varepsilon}{2}$. Thus, we get an additional restriction on $\eta$:
	\begin{equation*}
		\frac{9}{2}\frac{\eta^2 L_{\max}}{n\mu}\left( (n+1)\zeta^2_{\star} + \sigma^2_{\star}\right) <\frac{\varepsilon}{2},
	\end{equation*}
	and also the upper bound on the number of communication rounds $T$
	\begin{equation*}
		T = \widetilde{\cO}\left(\frac{1}{\eta\mu}\right).
	\end{equation*}
	Substituting \eqref{new_gamma_DIANA_NASTAY} in the previous equation, we get the first part of the result. 
	When $\gamma \rightarrow 0$, the proof follows similar steps. 
\end{proof}

\section{Alternative Analysis of Q-NASTYA}
In this analysis, we will use an additional sequence:
\begin{equation}
	x^i_{\star,m} = x^\star - \gamma\sum_{j=0}^{i-1}\nabla f_m(x^\star).
\end{equation}
\begin{theorem}\label{thm:Q_NASTYA_alternative_proof}
	Let Assumptions \ref{asm:quantization_operators}, \ref{asm:lip_max_f_m}, \ref{asm:sc_each_f_m} hold. Moreover, we assume that $(1-\gamma\mu)^n\leq \frac{\nicefrac{9}{10} - \nicefrac{1}{C}}{1+\nicefrac{1}{C}} = \widehat{C}<1$ for some numerical constant $C > 1$. Also let $\beta = \frac{\eta}{\gamma n} \leq \frac{1}{3C\frac{\omega}{M}+1}$ and $\gamma \leq \frac{1}{L_{\max}}$. Then, for all $T \geq 0$ the iterates produced by \gls{Q-NASTYA} satisfy
	\begin{align}
			&\leq \max \left(1-\frac{\beta}{10}, 1-\frac{\alpha}{2}\right) \Psi^{(t)} + \frac{2}{\mu}\beta\gamma^2\hat{\sigma}_{\text{rad}}^2
	\end{align}
\end{theorem}
\begin{align*}
	\mathbb{E}\left[\|x^T - x^\star\|^2\right]  &\leq	\left(1-\frac{\beta}{10}\right)\|x^t - x^{\star}\|^2+\frac{4}{\mu}\beta\gamma^2 \hat\sigma_{\mathrm{rad}}^{2} + 3\beta^2 \frac{\omega}{M} \frac{1}{M}\hat{\Delta}_\star,
\end{align*}
where $\hat{\Delta}_\star=\frac{1}{M}\sum_{m=1}^{M}\|x^n_{\star,m} - x^\star\|^2$ and $\hat{\sigma}^2_{\text{rad}} \leq L_{\max}\left( \zeta_{\star}^2 + \nicefrac{n\sigma_{\star}^2}{4} \right)$.
\begin{proof}
	The update rule for one epoch can be rewritten as
	\begin{align*}
		x^{t+1} = x^t - \eta \frac{1}{M}\sum_{M}^{m=1}Q\left( \frac{x^t - x^t_{m,n}}{\gamma n} \right).
	\end{align*}
	Using this, we derive
	\begin{align*}
		\| x^{t+1} - x^{\star} \|^2 &= \left\| x^t - \eta \frac{1}{M} \sum_{m=1}^{M}Q\left( \frac{x^t - x^t_{m,n}}{\gamma n}\right) - x^{\star} \right\|^2\\
		&=\|x^t - x^{\star}\|^2 - 2\eta\left\langle x^t - x^{\star},\frac{1}{M} \sum_{m=1}^{M}Q\left( \frac{x^t - x^t_{m,n}}{\gamma n} \right) \right\rangle\\
		&+ \eta^2\left\| \frac{1}{M} \sum_{m=1}^{M}Q\left( \frac{x^t - x^t_{m,n}}{\gamma n} \right)\right\|^2.
	\end{align*}
	Taking conditional expectation w.r.t.\ the randomness coming from compression, we get 
	\begin{align*}
		\mathbb{E}_Q\|x^{t+1} - x^{\star}\|^2 &= \|x^t - x^{\star}\|^2  - 2\eta\left\langle x^t - x^{\star},\frac{1}{M} \sum_{m=1}^{M}\left( \frac{x^t - x^t_{m,n}}{\gamma n} \right) \right\rangle\\
		&+ \eta^2	\mathbb{E}_Q\left\| \frac{1}{M} \sum_{m=1}^{M}Q\left( \frac{x^t - x^t_{m,n}}{\gamma n} \right)\right\|^2.
	\end{align*}
	Next, we use the definition of quantization operator and independence of compressed differences $Q\left( \frac{x^t - x^t_{m,n}}{\gamma n} \right)$, $m \in [M]$: 
	\begin{align*}
		\mathbb{E}_Q \|x^{t+1} - x^{\star}\|^2 & \leq \|x^t - x^{\star}\|^2  - 2\eta\left\langle x^t - x^{\star},\frac{1}{M} \sum_{m=1}^{M}\left( \frac{x^t - x^t_{m,n}}{\gamma n} \right) \right\rangle\\
		& + \eta^2\left( \frac{\omega}{M} \frac{1}{M}\sum_{m=1}^M\left\|\frac{x^t - x^t_{m,n}}{\gamma n}\right\|^2 + \left\| \frac{1}{M}\sum^M_{m=1}\frac{x^t - x^t_{m,n}}{\gamma n}  \right\|^2\right).
	\end{align*}
	Since $\beta = \frac{\eta}{\gamma n}$, we obtain
	\begin{align*}
		\mathbb{E}_Q \|x^{t+1} - x^{\star}\|^2 & \leq \|x^t - x^{\star}\|^2  - 2\beta\left\langle x^t - x^{\star},\frac{1}{M} \sum_{m=1}^{M}\left( x^t - x^t_{m,n} \right) \right\rangle\\
		& + \beta^2 \frac{\omega}{M} \frac{1}{M}\sum_{m=1}^M\left\|x^t - x^t_{m,n}\right\|^2\\
        &+\beta^2 \left\| \frac{1}{M}\sum^M_{m=1}\left(x^t - x^t_{m,n}\right)  \right\|^2\\
		&=\|x^t - x^{\star}\|^2  + 2\beta\left\langle x^t - x^{\star},\frac{1}{M} \sum_{m=1}^{M}\left( x^t_{m,n} - x^t   \right) \right\rangle\\
		& + \beta^2 \frac{\omega}{M} \frac{1}{M}\sum_{m=1}^M\left\|x^t - x^t_{m,n}\right\|^2 +\beta^2 \left\| \frac{1}{M}\sum^M_{m=1}\left(x^t_{m,n} - x^t\right)  \right\|^2 \\
		&=\left\| x^t - x^{\star} + \beta\left( \frac{1}{M}\sum_{m=1}^M\left( x^t_{m,n} - x^t  \right) \right)\right\|^2\\
        &+\beta^2 \frac{\omega}{M} \frac{1}{M}\sum_{m=1}^M\left\|x^t - x^t_{m,n}\right\|^2  \\
		&=\left\| (1-\beta)(x^t - x^{\star}) + \beta\left( \frac{1}{M}\sum_{m=1}^M\left( x^t_{m,n}   \right) - x^{\star} \right)\right\|^2\\
		&+\beta^2 \frac{\omega}{M} \frac{1}{M}\sum_{m=1}^M\left\|x^t - x^t_{m,n}\right\|^2 .
	\end{align*}
	Using the condition that $x^{\star} = \frac{1}{M}\sum_{m=1}^M x^\star_{m,n}$ we have:
	\begin{align*}
		\mathbb{E}_Q \|x^{t+1} - x^{\star}\|^2 & \leq\left\| (1-\beta)(x^t - x^{\star}) + \beta\left( \frac{1}{M}\sum_{m=1}^M\left( x^t_{m,n} - x^\star_{m,n}  \right)  \right)\right\|^2\\
        &+\beta^2 \frac{\omega}{M} \frac{1}{M}\sum_{m=1}^M\left\|x^t - x^t_{m,n}\right\|^2 .
	\end{align*}
	Convexity of squared norm and Jensen's inequality imply
	\begin{align*}
		\mathbb{E}_Q \|x^{t+1} - x^{\star}\|^2 &\leq (1-\beta)\|x^t - x^{\star}\|^2 + \beta \left\|  \frac{1}{M}\sum_{m=1}^M\left( x^t_{m,n} - x^\star_{m,n}  \right) \right\|^2\\
        &+\beta^2 \frac{\omega}{M} \frac{1}{M}\sum_{m=1}^M\left\|x^t - x^t_{m,n}\right\|^2 .
	\end{align*}
	Next, from Young's inequality we get
	\begin{align*}
		\mathbb{E}_Q \|x^{t+1} - x^{\star}\|^2 &\leq (1-\beta)\|x^t - x^{\star}\|^2 + \beta \left\|  \frac{1}{M}\sum_{m=1}^M\left( x^t_{m,n} - x^\star_{m,n}  \right) \right\|^2\\
        &+ 3\beta^2 \frac{\omega}{M}\|x^t - x^{\star}\|^2\\
		& + 3\beta^2 \frac{\omega}{M} \frac{1}{M}\sum_{m=1}^{M}\|x^t_{m,n} - x^\star_{m,n}\|^2\\
        &+ 3\beta^2 \frac{\omega}{M} \frac{1}{M}\sum_{m=1}^{M}\| x^\star_{m,n} - x^{\star} \|^2.
	\end{align*}
	Theorem 4 from \citep{mishchenko2022proximal} gives 
	\begin{align*}
	\mathbb{E}\left[	\frac{1}{M}\sum_{m=1}^{M}\|x^t_{m,n} - x^\star_{m,n}\|^2\right]& \leq(1-\gamma \mu)^{n} \left[\left\|x^{t}-x^{\star}\right\|^{2}\right]\\
    &+2 \gamma^{3} \hat\sigma_{\mathrm{rad}}^{2}\left(\sum_{j=0}^{n-1}(1-\gamma \mu)^{j}\right)\\
		& = (1-\gamma \mu)^{n} \left[\left\|x^{t}-x^{\star}\right\|^{2}\right]+2 \gamma^{2} \hat\sigma_{\mathrm{rad}}^{2}\frac{1}{\gamma\mu}.
	\end{align*}
	It leads to
	\begin{align*}
		&\mathbb{E} \|x^{t+1} - x^{\star}\|^2\\
        &\leq (1-\beta)\|x^t - x^{\star}\|^2 + \beta \left((1-\gamma \mu)^{n} \left[\left\|x^{t}-x^{\star}\right\|^{2}\right]+2 \gamma^{3} \hat\sigma_{\mathrm{rad}}^{2}\frac{1}{\gamma\mu}\right)\\
		&+ 3\beta^2 \frac{\omega}{M}\|x^t - x^{\star}\|^2 + 3\beta^2 \frac{\omega}{M} \left( (1-\gamma \mu)^{n} \left[\left\|x^{t}-x^{\star}\right\|^{2}\right]+2 \gamma^{3} \hat\sigma_{\mathrm{rad}}^{2}\frac{1}{\gamma\mu}\right)\\
		&+ 3\beta^2 \frac{\omega}{M} \frac{1}{M}\sum_{m=1}^{M}\| x^\star_{m,n} - x^{\star} \|^2\\
		&\leq \left(1-\beta+\beta(1-\gamma\mu)^n+3\beta^2\frac{\omega}{M}+3\beta^2\frac{\omega}{M}(1-\gamma\mu)^n\right)\|x^t - x^{\star}\|^2\\
		&+2\beta\gamma^3 \hat\sigma_{\mathrm{rad}}^{2}\frac{1}{\gamma\mu}\left( 1+3\beta\frac{\omega}{M} \right) + 3\beta^2 \frac{\omega}{M} \frac{1}{M}\sum_{m=1}^{M}\| x^\star_{m,n} - x^{\star} \|^2.
	\end{align*}
	Using $(1-\gamma\mu)^n\leq \frac{\nicefrac{9}{10} - \nicefrac{1}{C}}{1+\nicefrac{1}{C}} $, we have 
	\begin{align*}
		&(1-\gamma\mu)^n\leq \frac{\nicefrac{9}{10} - \nicefrac{1}{C}}{1+\nicefrac{1}{C}} \\
		&(1-\gamma\mu)^n\left(1+\frac{1}{C}\right) \leq \frac{9}{10} - \frac{1}{C}\\
		&-\frac{9}{10}\beta+\beta(1-\gamma\mu)^n+\frac{\beta}{C}+\frac{\beta}{C}(1-\gamma\mu)^n\leq 0\\
		&1-\beta+\beta(1-\gamma\mu)^n + \frac{\beta}{C}+\frac{\beta}{C}(1-\gamma\mu)^n \leq 1 - \frac{\beta}{10}.
	\end{align*}
	Next, applying $\beta \leq \frac{1}{1+3C\frac{\omega}{M}}$, we derive 
	\begin{align*}
		1-\beta+\beta(1-\gamma\mu)^n+3\beta^2\frac{\omega}{M}+3\beta^2\frac{\omega}{M}(1-\gamma\mu)^n \leq 1- \frac{\beta}{10}.
	\end{align*}
	Finally, we have 
	\begin{align*}
		\mathbb{E}\|x^{t+1} - x^\star\|^2&\leq	\left(1-\frac{\beta}{10}\right)\|x^t - x^{\star}\|^2+2\beta\gamma^2 \hat\sigma_{\mathrm{rad}}^{2}\frac{1}{\mu}\left( 1+\frac{1}{C} \right)\\
		& + 3\beta^2 \frac{\omega}{M} \frac{1}{M}\sum_{m=1}^{M}\| x^\star_{m,n} - x^{\star} \|^2\\
		&\leq	\left(1-\frac{\beta}{10}\right)\|x^t - x^{\star}\|^2+\frac{4}{\mu}\beta\gamma^2 \hat\sigma_{\mathrm{rad}}^{2}\\
		& + 3\beta^2 \frac{\omega}{M} \frac{1}{M}\sum_{m=1}^{M}\| x^\star_{m,n} - x^{\star} \|^2.
	\end{align*}
\end{proof}

\section{Alternative Analysis of DIANA-NASTYA}
\begin{theorem}\label{thm:Q_NASTYA_alternative_proof_1}
	Let Assumptions \ref{asm:quantization_operators}, \ref{asm:lip_max_f_m}, \ref{asm:sc_each_f_m} hold. Moreover, we assume that $(1-\gamma\mu)^n\leq \frac{\nicefrac{9}{10} - \nicefrac{1}{B}}{1+\nicefrac{1}{B}} = \widehat{B}<1$ for some numerical constant $B > 1$. Also let $\beta = \frac{\eta}{\gamma n} \leq \frac{1}{12B\frac{\omega}{M}+1}$ and $\gamma \leq \frac{1}{L_{\max}}$ and also $\alpha\leq \frac{1}{\omega+1}$. Then, for all $T \geq 0$ the iterates produced by \gls{DIANA-NASTYA} satisfy
	\begin{align}
		\mathbb{E}\Psi^{(T)}	&\leq \max \left(1-\frac{\beta}{10}, 1-\frac{\alpha}{2}\right)^T \Psi^{(0)} + \frac{2}{\mu\min(\frac{\beta}{10},\frac{\alpha}{2})}\beta\gamma^2\hat{\sigma}_{\text{rad}}^2.
	\end{align}
\end{theorem}
\begin{proof}
We start with expanding the square:
\begin{align*}
	\|x^{t+1} - x^{\star}\|^2 &= \| x^t - \eta \hat{g}^t - x^{\star} \|^2\\
	&= \left\| x^t - \eta \frac{1}{M}\sum_{m=1}^M\left( h^{t}_{m} + Q(g^{t}_{m} - h^{t}_{m}) \right) - x^{\star} \right\|^2\\
	& = \|x^t - x^{\star}\|^2 - 2\eta \left\langle \frac{1}{M}\sum_{m=1}^M\left(h^{t}_{m}+Q(g^{t}_{m} - h^{t}_{m})\right),x^t - x^{\star} \right\rangle\\
	& + \eta^2\left\| \frac{1}{M}\sum_{m=1}^M \left(h^{t}_{m}+Q(g^{t}_{m} - h^{t}_{m})\right) \right\|^2.
\end{align*}
Taking the expectation w.r.t.\ $\cQ$, we get
\begin{align*}
	\mathbb{E}_Q	\|x^{t+1} - x^{\star}\|^2 & = \|x^t - x^{\star}\|^2 - 2\eta \left\langle \frac{1}{M}\sum_{m=1}^M g^{t}_{m} ,x^t - x^{\star} \right\rangle\\
	& + \eta^2\mathbb{E}_Q\left\| \frac{1}{M}\sum_{m=1}^M \left(h^{t}_{m}+Q(g^{t}_{m} - h^{t}_{m})\right) \right\|^2\\
	&= \|x^t - x^{\star}\|^2 - 2\eta \left\langle \frac{1}{M}\sum_{m=1}^M g^{t}_{m} ,x^t - x^{\star} \right\rangle\\
	& + \eta^2\mathbb{E}_Q\left\| \frac{1}{M}\sum_{m=1}^M \left(h^{t}_{m}+Q(g^{t}_{m} - h^{t}_{m}) - g^{t}_{m}\right) \right\|^2\\
    &+\eta^2\left\| \frac{1}{M}\sum_{m=1}^{M}g^{t}_{m} \right\|^2\\
	&\leq\|x^t - x^{\star}\|^2 - 2\eta \left\langle \frac{1}{M}\sum_{m=1}^M g^{t}_{m} ,x^t - x^{\star} \right\rangle\\
	& + \eta^2\frac{\omega}{M^2}\sum_{m=1}^{M}\|g^{t}_{m} - h^{t}_{m}\|^2+\eta^2\left\| \frac{1}{M}\sum_{m=1}^{M}g^{t}_{m} \right\|^2\\
	&\leq\|x^t - x^{\star}\|^2 - 2\eta \left\langle \frac{1}{M}\sum_{m=1}^M g^{t}_{m} ,x^t - x^{\star} \right\rangle\\
	& + \eta^2\frac{2\omega}{M^2}\sum_{m=1}^{M}\|g^{t}_{m} - h_{*,m}\|^2+\eta^2\frac{2\omega}{M^2}\sum_{m=1}^{M}\|h^{t}_{m} - h_{*,m}\|^2\\
    &+\eta^2\left\| \frac{1}{M}\sum_{m=1}^{M}g^{t}_{m} \right\|^2.
\end{align*}
Next, using definition of $g^{t}_{m}$, we obtain 
\begin{align*}
	&\mathbb{E}	\|x^{t+1} - x^{\star}\|^2\\
    &\leq	\|x^t - x^{\star}\|^2 - 2\eta \left\langle \frac{1}{M}\sum_{m=1}^M \frac{x^t - x^t_{m,n}}{\gamma n},x^t - x^{\star} \right\rangle+\eta^2\left\| \frac{1}{M}\sum_{m=1}^{M} \frac{x^t - x^t_{m,n}}{\gamma n} \right\|^2\\
	& + \eta^2\frac{2\omega}{M^2}\sum_{m=1}^{M}\|g^{t}_{m} - h_{*,m}\|^2+\eta^2\frac{2\omega}{M^2}\sum_{m=1}^{M}\|h^{t}_{m} - h_{*,m}\|^2\\
	&= \|x^t - x^{\star}\|^2 + 2\alpha  \left\langle \frac{1}{M}\sum_{m=1}^M \left(x^t_{m,n}-x^t\right),x^t - x^{\star} \right\rangle + \alpha^2\left\|  \frac{1}{M}\sum_{m=1}^{M} \left(  x^t_{m,n}-x^t \right) \right\|^2\\
	& + \eta^2\frac{2\omega}{M^2}\sum_{m=1}^{M}\|g^{t}_{m} - h_{*,m}\|^2+\eta^2\frac{2\omega}{M^2}\sum_{m=1}^{M}\|h^{t}_{m} - h_{*,m}\|^2\\
	&=\left\| x^t - x^{\star} + \alpha  \frac{1}{M}\sum_{m=1}^{M} \left(  x^t_{m,n}-x^t \right)  \right\|^2\\
	& + \eta^2\frac{2\omega}{M^2}\sum_{m=1}^{M}\|g^{t}_{m} - h_{*,m}\|^2+\eta^2\frac{2\omega}{M^2}\sum_{m=1}^{M}\|h^{t}_{m} - h_{*,m}\|^2\\
	&=\left\| (1-\beta)(x^t - x^{\star}) + \beta  \left(\frac{1}{M}\sum_{m=1}^{M} \left(  x^t_{m,n}-x^\star_{m,n} \right)\right)  \right\|^2\\
	&\leq (1-\beta)\|x^t - x^{\star}\|^2 + \beta \frac{1}{M}\sum_{m=1}^{M}\| x^t_{m,n}-x^\star_{m,n} \|^2\\
	&+\eta^2\frac{2\omega}{M^2}\sum_{m=1}^{M}\|g^{t}_{m} - h_{*,m}\|^2\\
    &+\eta^2\frac{2\omega}{M^2}\sum_{m=1}^{M}\|h^{t}_{m} - h_{*,m}\|^2.
\end{align*}
Let us consider recursion for control variable:
\begin{align*}
	\| h^{t+1}_{m} - h_{*,m} \|^2 &= \|h^{t}_{m} + \alpha Q(g^{t}_{m} - h^{t}_{m}) - h_{*,m}\|^2 \\
	&=\|h^{t}_{m} - h_{*,m}\|^2 + \alpha \left\langle Q(g^{t}_{m} - h^{t}_{m}),h^{t}_{m} - h_{*,m} \right\rangle\\
    &+\alpha^2\|Q(g^{t}_{m} - h^{t}_{m})\|^2.
\end{align*}
Taking the expectation w.r.t.\ $\cQ$, we have 
\begin{align*}
	\mathbb{E}_{\cQ}	\| h^{t+1}_{m} - h_{*,m} \|^2 &\leq \|h^{t}_{m} - h_{*,m}\|^2 + 2\alpha \left\langle g^{t}_{m} - h^{t}_{m},h^{t}_{m} - h_{*,m} \right\rangle\\
    &+ \alpha^2\left( \omega + 1 \right)\left\|g^{t}_{m} - h^{t}_{m}\right\|^2.
\end{align*}
Using $\alpha \leq \frac{1}{\omega+1}$ we have
\begin{align*}
	\mathbb{E}	\| h^{t+1}_{m} - h_{*,m} \|^2 &\leq \|h^{t}_{m} - h_{*,m}\|^2\\
	& + 2\alpha \left\langle g^{t}_{m} - h^{t}_{m},h^{t}_{m} - h_{*,m} \right\rangle + \alpha\left\|g^{t}_{m} - h^{t}_{m}\right\|^2\\
	& = \|h^{t}_{m} - h_{*,m}\|^2\\
	& + 2\alpha \left\langle g^{t}_{m} - h^{t}_{m},h^{t}_{m} - h_{*,m} \right\rangle + \alpha\left\langle g^{t}_{m} - h^{t}_{m},g^{t}_{m} - h^{t}_{m}\right\rangle\\
	& = \| h^{t}_{m} - h_{*,m} \|^2\\
	& + \alpha\left\langle g^{t}_{m} - h^{t}_{m},g^{t}_{m} - h^{t}_{m}+2h^{t}_{m}-2h_{*,m}\right\rangle\\	
	& = \| h^{t}_{m} - h_{*,m} \|^2\\
	& + \alpha\left\langle g^{t}_{m} - h^{t}_{m},g^{t}_{m} +h^{t}_{m}-2h_{*,m}\right\rangle\\
	& = \| h^{t}_{m} - h_{*,m} \|^2\\
	& + \alpha\left\langle g^{t}_{m} - h^{t}_{m}-h_{*,m}+h_{*,m},g^{t}_{m} +h^{t}_{m}-2h_{*,m}\right\rangle\\
	& = \| h^{t}_{m} - h_{*,m} \|^2\\
	& + \alpha\left\langle g^{t}_{m} -h_{*,m}- (h^{t}_{m}-h_{*,m}),(g^{t}_{m}-h_{*,m}) +(h^{t}_{m}-h_{*,m})\right\rangle\\
	&= \| h^{t}_{m} - h_{*,m} \|^2+\alpha\| g^{t}_{m} - h_{*,m} \|^2 - \alpha \| h^{t}_{m} - h_{*,m} \|^2\\
	& = (1-\alpha) \| h^{t}_{m} - h_{*,m} \|^2 +\alpha \| g^{t}_{m} - h_{*,m} \|^2.
\end{align*}
Using this bound we get that 
\begin{align*}
	\frac{1}{M}\sum_{m=1}^{M}		\mathbb{E}_{\cQ}	\| h^{t+1}_{m} - h_{*,m} \|^2 &\leq (1-\alpha) \frac{1}{M}\sum_{m=1}^{M}\| h^{t}_{m} - h_{*,m} \|^2\\
    &+\alpha \frac{1}{M}\sum_{m=1}^{M} \| g^{t}_{m} - h_{*,m} \|^2.
\end{align*}
Let us consider Lyapunov function:
\begin{align*}
	\Psi^{(t)} = \|x^t - x^{\star}\|^2+\frac{4\omega\eta^2}{\alpha M} \frac{1}{M}\sum_{m=1}^{M}\| h^{t}_{m} - h_{*,m} \|^2.
\end{align*}
Using previous bounds and Theorem 4 from \citep{mishchenko2022proximal} we have 
\begin{align*}
	\mathbb{E}\Psi^{(t+1)}&\leq (1-\beta)\|x^t - x^{\star}\|^2 + \beta\left( (1-\gamma\mu)^n \mathbb{E}\|x^t - x^{\star}\|^2 + \gamma^3\frac{1}{\gamma\mu}\hat{\sigma}^2_{rad} \right)\\
	&+\eta^2\frac{2\omega}{M}\frac{1}{M}\sum_{m=1}^{M}\mathbb{E}\| g^{t}_{m} - h_{*,m} \|^2
	 + \eta^2\frac{2\omega}{M}\frac{1}{M}\sum_{m=1}^{M}\mathbb{E}\| h^{t}_{m} - h_{*,m} \|^2\\&+(1-\alpha)\frac{4\omega\eta^2}{\alpha M}\frac{1}{M}\sum_{m=1}^{M} \mathbb{E}\|h^{t}_{m} - h_{*,m}\|^2 + \alpha \frac{4\omega\eta^2}{\alpha M}\frac{1}{M}\sum_{m=1}^{M} \mathbb{E}\|g^{t}_{m} - h_{*,m}\|^2\\
	&\leq \left(1-\frac{\alpha}{2}\right)\frac{4\omega\eta^2}{\alpha M} \frac{1}{M}\sum_{m=1}^{M}\mathbb{E}\|h^{t}_{m} - h_{*,m}\|^2 + \eta^2\frac{6\omega}{M}\frac{1}{M}\sum_{m=1}^{M}\mathbb{E}\| g^{t}_{m} - h_{*,m} \|^2\\
	&+ (1-\beta)\mathbb{E}\|x^t - x^{\star}\|^2 + \beta\left( (1-\gamma\mu)^n\mathbb{E}\|x^t - x^{\star}\|^2 + \gamma^3\frac{1}{\gamma\mu}\hat{\sigma}^2_{rad}. \right)
\end{align*}
Let us consider 
\begin{align*}
	&\eta^2\frac{1}{M}\sum_{m=1}^{M}\mathbb{E}\| g^{t}_{m} - h_{*,m} \|^2\\
    &= \eta^2\frac{1}{M}\sum_{m=1}^{M}\mathbb{E}\left\| \frac{x^t - x^t_{m,n}}{\gamma n} - \frac{x^{\star} - x^\star_{m,n}}{\gamma n}  \right\|^2 \\
	&\leq 2\eta^2 \frac{1}{M}\sum_{m=1}^{M}\mathbb{E}\left\| \frac{x^t - x^{\star}}{\gamma n} \right\|^2 + 2\eta^2 \frac{1}{M}\sum_{m=1}^{M} \mathbb{E}\left\|\frac{x^t_{m,n} - x^\star_{m,n}}{\gamma n} \right\|^2\\
	&\leq  2\beta^2 \frac{1}{M}\sum_{m=1}^{M}\mathbb{E}\left\| x^t - x^{\star} \right\|^2 + 2\beta^2 \frac{1}{M}\sum_{m=1}^{M} \mathbb{E}\left\|x^t_{m,n} - x^\star_{m,n} \right\|^2\\
	&\leq  2\beta^2 \mathbb{E}\left\| x^t - x^{\star} \right\|^2 + 2\beta^2 \frac{1}{M}\sum_{m=1}^{M} \mathbb{E}\left\|x^t_{m,n} - x^\star_{m,n} \right\|^2.\\
\end{align*}
Putting all the terms together and using $(1-\gamma\mu)^n\leq \frac{\nicefrac{9}{10} - \nicefrac{1}{B}}{1+\nicefrac{1}{B}} = \widehat{B}<1$, $\beta \leq \frac{1}{12B\frac{\omega}{M}+1}$  we have 
\begin{align*}
	\mathbb{E}\Psi^{(t+1)} &\leq \left(1-\beta +12\frac{\omega}{M}\beta^2 + 12\frac{\omega}{M}\beta^2(1-\gamma\mu)^n + \beta(1-\gamma \mu)^n  \right)\mathbb{E}\|x^t - x^{\star}\|^2\\
    &+ \beta \gamma^3 \frac{1}{\gamma\mu}\hat{\sigma}^2_{rad}\\ 
	&+ 2\beta^2\frac{6\omega}{M}\gamma^3 \frac{1}{\gamma\mu} \hat{\sigma}^2_{rad}
	+\left(1-\frac{\alpha}{2}\right)\frac{4\omega\eta^2}{\alpha M}\frac{1}{M}\sum_{m=1}^M\mathbb{E}\|h^{t}_{m} - h_{*,m}\|^2\\
	&\leq \left(1-\frac{\beta}{10}\right)\mathbb{E}\|x^t - x^\star\|^2 + \frac{2}{\mu}\beta\gamma^2\hat{\sigma}_{\text{rad}}^2\\
    &+ \left(1-\frac{\alpha}{2}\right)\frac{4\omega\eta^2}{\alpha M} \frac{1}{M}\sum_{m=1}^M\mathbb{E}\|h^{t}_{m} - h_{*,m}\|^2\\
	&\leq \max \left(1-\frac{\beta}{10}, 1-\frac{\alpha}{2}\right) \Psi^{(t)} + \frac{2}{\mu}\beta\gamma^2\hat{\sigma}_{\text{rad}}^2.
\end{align*}
Unrolling this recursion we get the final result.
\end{proof}

\section{Partial Participation for Method with Local Steps}\label{appendix:PP}
\subsection{Analysis of Q-NASTYA with partial participation}

\begin{algorithm}[t]
\caption{\algname{Q-NASTYA-PP}}
\label{alg:Q_NASTYA-PP}
	\begin{algorithmic}[1]
		\REQUIRE $x^0$ -- starting point, $\gamma > 0$ -- local stepsize, $\eta > 0$ -- global stepsize
	    \FOR{$t =0,1,\dots, T-1$}
            \STATE Sample a cohort $\set$ with cardinality $C$ uniformly
    		\FOR{$m \in \set$ in parallel}
    		    \STATE Receive $x^t$ from the server and set $x^0_{t,m} = x^t$
    		    \STATE Sample random permutation of $[n]$: $\pi_m = (\pi^0_m, \dots, \pi^{n-1}_m)$
    		    \FOR{$i = 0, 1,\dots, n-1$}
    		        \STATE Set $x^{t}_{m,i+1} = x^{t}_{m,i} - \gamma \nabla f_{m,\pi^{i}_m}(x^{t}_{m,i})$
    		    \ENDFOR
    		    \STATE Compute $g^{t}_{m} = \frac{1}{\gamma n}\left(x^t - x^t_{m,n}\right)$ and send $\cQ_t(g^{t}_{m})$ to the server
    		\ENDFOR
    		\STATE Compute $g^t = \frac{1}{C}\sum_{m\in \set}\mathcal{Q}_t(g^{t}_{m})$
    		\STATE Compute $x^{t+1} = x^{t} - \eta g^t$ and send $x^{t+1}$ to the workers
    	\ENDFOR
    	\ENSURE $x^T$
	\end{algorithmic}
\end{algorithm}

\begin{lemma}
\label{lemma:F1} Let Assumptions \ref{asm:quantization_operators}, \ref{asm:sc_general_f}, \ref{asm:lip_max_f_m} hold. Then, for all $t \geq 0$ the iterates produced by Q-NASTYA-PP (Algorithm \ref{alg:Q_NASTYA-PP}) satisfy

\begin{align*}
   \mathbb{E}_{\mathcal{Q}, \set}\left[\left\|g^t\right\|^2\right] &\leq \frac{2 L_{\max }^2\left(1+\frac{\omega}{C}\right)}{M n} \sum_{m=1}^M \sum_{i=0}^{n-1}\left\|x^t_{m, i}-x^t\right\|^2\\
   &+8 L_{\max }\left(1+\frac{\omega}{C}\right)\left(f\left(x^t\right)-f\left(x^{\star}\right)\right)\\
   &+4\left(\frac{\omega}{C}+\frac{M-C}{C \max M-1,1}\right) \sigma_{\star}^2, 
\end{align*}

where $\mathbb{E}_{\mathcal{Q}, \set}$ is the expectation w.r.t. $\mathcal{Q}, \set$ and $\sigma_{\star}^2=\frac{1}{M} \sum_{m=1}^M\left\|\nabla f_m\left(x^{\star}\right)\right\|^2$.

\end{lemma}
\begin{proof}
{\footnotesize
$\mathbb{E}\left[\|\xi\|^2\right]=\mathbb{E}\left[\|\xi-\mathbb{E}[\xi]\|^2\right]+\|\mathbb{E} \xi\|^2 \text {, we obtain }$
    \begin{align*}
&\mathbb{E}_{\mathcal{Q}}\left[\left\|g^t\right\|^2\right]\\
= & \frac{1}{C^2} \mathbb{E}_{\mathcal{Q}} \| \sum_{m \in \set}(\underbrace{\left(\frac{1}{n} \sum_{i=0}^{n-1} \nabla f_{m,\pi_m^i}\left(x^t_{m, i}\right)\right)-\frac{1}{n} \sum_{i=0}^{n-1} \nabla f_{m,\pi_m^i}\left(x^t_{m, i}\right)}_{=\xi_m} \|^2] \\
& +\left\|\frac{1}{C n} \sum_{m \in \set} \sum_{i=0}^{n-1} \nabla f_{m,\pi_m^i}\left(x^t_{m, i}\right)\right\|^2 \\
= & \frac{1}{C^2} \mathbb{E}_{\mathcal{Q}}\left[\sum_{m \in \set}\left\|\xi_m\right\|^2+\sum_{m, l \in S_{t: m \neq l}} 2\left\langle\xi_m, \xi_l\right\rangle\right]+\left\|\frac{1}{C n} \sum_{m \in \set} \sum_{i=0}^{n-1} \nabla f_{m,\pi_m^i}\left(x^t_{m, i}\right)\right\|^2.
\end{align*}
}
Using independence between $\xi_m$ and $\xi_l$ for different  $m, l$ we get 
\begin{align*}
\mathbb{E}_{\mathcal{Q}}\left[\left\|g^t\right\|^2\right]= & \frac{1}{C^2} \sum_{m \in \set} \mathbb{E}_{\mathcal{Q}}\left[\left\|\mathcal{Q}\left(\frac{1}{n} \sum_{i=0}^{n-1} \nabla f_{m,\pi_m^i}\left(x^t_{m, i}\right)\right)-\frac{1}{n} \sum_{i=0}^{n-1} \nabla f_{m,\pi_m^i}\left(x^t_{m, i}\right)\right\|^2\right] \\
& +\left\|\frac{1}{C n} \sum_{m \in \set} \sum_{i=0}^{n-1} \nabla f_{m,\pi_m^i}\left(x^t_{m, i}\right)\right\|^2 \\
\leq & \frac{\omega}{C^2} \sum_{m \in \set}\left\|\frac{1}{n} \sum_{i=0}^{n-1} \nabla f_{m,\pi_m^i}\left(x^t_{m, i}\right)\right\|^2+\left\|\frac{1}{C n} \sum_{m \in \set} \sum_{i=0}^{n-1} \nabla f_{m,\pi_m^i}\left(x^t_{m, i}\right)\right\|^2.
\end{align*}
Rewriting previous inequality and using $ \nabla f_m(x)=\frac{1}{n} \sum_{i=0}^{n-1} \nabla f_{m,\pi_m^i}\left(x^t\right)$, we have 
\begin{align*}
\mathbb{E}_{\mathcal{Q}}\left[\left\|g^t\right\|^2\right] \leq & \frac{2 \omega}{C^2} \sum_{m \in \set}\left\|\frac{1}{n} \sum_{i=0}^{n-1}\left(\nabla f_{m,\pi_m^i}\left(x^t_{m, i}\right)-\nabla f_{m,\pi_m^i}\left(x^t\right)\right)\right\|^2\\
&+\frac{2 \omega}{C^2} \sum_{m \in \set}\left\|\nabla f_m\left(x^t\right)\right\|^2 \\
& +2\left\|\frac{1}{C n} \sum_{m \in \set} \sum_{i=0}^{n-1}\left(\nabla f_{m,\pi_m^i}\left(x^t_{m, i}\right)-\nabla f_{m,\pi_m^i}\left(x^t\right)\right)\right\|^2\\
&+2\left\|\frac{1}{C} \sum_{m \in \set} \nabla f_m\left(x^t\right)\right\|^2 \\
\leq & \frac{2\left(1+\frac{\omega}{C}\right)}{C} \sum_{m \in \set}\left\|\frac{1}{n} \sum_{i=0}^{n-1}\left(\nabla f_{m,\pi_m^i}\left(x^t_{m, i}\right)-\nabla f_{m,\pi_m^i}\left(x^t\right)\right)\right\|^2 \\
& +\frac{2 \omega}{C^2} \sum_{m \in \set}\left\|\nabla f_m\left(x^t\right)\right\|^2+2\left\|\frac{1}{C} \sum_{m \in \set} \nabla f_m\left(x^t\right)\right\|^2
\end{align*}

$$
\begin{aligned}
&\text { Using } L \text {-smoothness of } f_m^i \text { and } f \text { and also convexity of } f_m \text {, we obtain }\\
&\begin{aligned}
\mathbb{E}_{\mathcal{Q}}\left[\left\|g^t\right\|^2\right] \leq & \frac{2\left(1+\frac{\omega}{C}\right)}{C n} \sum_{m \in \set} \sum_{i=0}^{n-1}\left\|\nabla f_{m,\pi_m^i}\left(x^t_{m, i}\right)-\nabla f_{m,\pi_m^i}\left(x^t\right)\right\|^2\\
&+\frac{4 \omega}{C^2} \sum_{m \in \set}\left\|\nabla f_m\left(x^t\right)-\nabla f_m\left(x^{\star}\right)\right\|^2 \\
& +\frac{4 \omega}{C^2} \sum_{m \in \set}\left\|\nabla f_m\left(x^{\star}\right)\right\|^2\\
&+4\left\|\frac{1}{C} \sum_{m \in \set}\left(\nabla f_m\left(x^t\right)-\nabla f_m\left(x^{\star}\right)\right)\right\|^2\\
&+4\left\|\frac{1}{C} \sum_{m \in \set} \nabla f_m\left(x^{\star}\right)\right\|^2 \\
\leq & \frac{2 L_{\max }^2\left(1+\frac{\omega}{C}\right)}{C n} \sum_{m \in \set} \sum_{i=0}^{n-1}\left\|x^t_{m, i}-x^t\right\|^2\\
&+\frac{8 L_{\max }\left(1+\frac{\omega}{C}\right)}{C} \sum_{m \in \set} D_{f_m}\left(x^t, x^{\star}\right) \\
& +\frac{4 \omega}{C^2} \sum_{m \in \set}\left\|\nabla f_m\left(x^{\star}\right)\right\|^2+4\left\|\frac{1}{C} \sum_{m \in \set} \nabla f_m\left(x^{\star}\right)\right\|^2 .
\end{aligned}
\end{aligned}
$$

Taking the expectation w.r.t. $\set$ and using uniform sampling, we receive

$$
\begin{aligned}
\mathbb{E}_{\mathcal{Q}, \set}\left[\left\|g^t\right\|^2\right] \leq & \frac{2 L_{\max }^2\left(1+\frac{\omega}{C}\right)}{n} \mathbb{E}_{\set}\left[\frac{1}{C} \sum_{m \in \set} \sum_{i=0}^{n-1} \| x^t_{m, i}-\left.x^t\right|^2\right]\\
&+8 L_{\max }\left(1+\frac{\omega}{C}\right) \mathbb{E}_{\set}\left[\frac{1}{C} \sum_{m \in \set} D_{f_m}\left(x^t, x^{\star}\right)\right] \\
& +\frac{4 \omega}{C} \mathbb{E}_{\set}\left[\frac{1}{C} \sum_{m \in \set}\left\|\nabla f_m\left(x^{\star}\right)\right\|^2\right]\\
&+4 \mathbb{E}_{\set}\left[\left\|\frac{1}{C} \sum_{m \in \set} \nabla f_m\left(x^{\star}\right)\right\|^2\right] \\
\leq & \frac{2 L_{\max }^2\left(1+\frac{\omega}{C}\right)}{M n} \sum_{m=1}^M \sum_{i=0}^{n-1}\left\|x^t_{m, i}-x^t\right\|^2\\
&+\frac{8 L_{\max }\left(1+\frac{\omega}{C}\right)}{M} \sum_{m=1}^M D_{f_m}\left(x^t, x^{\star}\right) \\
& +\frac{4 \omega}{C} \frac{1}{M} \sum_{m=1}^M\left\|\nabla f_m\left(x^{\star}\right)\right\|^2\\
&+4 \frac{M-C}{M C \max M-1,1} \sum_{m=1}^M\left\|\nabla f_m\left(x^{\star}\right)\right\|^2 .
\end{aligned}
$$
\end{proof}

\begin{theorem}
\label{thm:pp-q-nastya}
    Let step sizes $\eta, \gamma$ satisfy the following equations

$$
\eta=\frac{1}{16 L_{\max }\left(1+\frac{\omega}{C}\right)}, \quad \gamma=\frac{1}{5 n L_{\max }}
$$

Under Assumptions \ref{asm:quantization_operators}, \ref{asm:sc_general_f}, \ref{asm:lip_max_f_m} iterates of \algname{Q-NASTYA-PP} (Algorithm \ref{alg:Q_NASTYA-PP}) satisfy 

\begin{align*}
    \mathbb{E}\left[\left\|x^T-x^{\star}\right\|^2\right] &\leq\left(1-\frac{\eta \mu}{2}\right)^T\left\|x^0-x^{\star}\right\|^2+\frac{9}{2} \frac{\gamma^2 n L_{\max }}{\mu}\left(\frac{1}{M} \sum_{m=1}^M \sigma_{\star, m}^2+n \sigma_{\star}^2\right)\\
    &+8 \frac{\eta}{\mu}\left(\frac{\omega}{C} \sigma_{\star}^2+\frac{M-C}{C \max (M-1,1)} \sigma_{\star}^2\right),
\end{align*}


\end{theorem}

As we can see, there is an additional error term proportional to $\frac{M-C}{C \max (M-1,1)}$ that arises due to Partial participation setting. Note that when $C=M$ (all clients are participating), this error term vanishes, allowing us to recover the previous result for the full participation case. This shows the consistency of our theoretical framework across different participation scenarios.

\begin{proof}
    \begin{align*}
        \begin{aligned}
&\text { Taking the expectation w.r.t. } \mathcal{Q}, \set \text { and using Lemma ~\ref{lemma:F1} }  \text {updated, we get }\\
&\begin{aligned}
&\mathbb{E}_{\mathcal{Q}, \set}\left[\| x^{t+1}-\left.x^{\star}\right|^2\right]\\
= & \| x^t-\left.x^{\star}\right|^2-2 \eta \mathbb{E}_{\mathcal{Q}, \set}\left[\left\langle g^t, x^t-x^{\star}\right\rangle\right]+\eta^2 \mathbb{E}_{\mathcal{Q}, \set}\left[\left\|g^t\right\|^2\right] \\
\leq & \left\|x^t-x^{\star}\right\|^2-2 \eta \mathbb{E}_{\mathcal{Q}, \set}\left[\left\langle\frac{1}{C} \sum_{m \in \set} \mathcal{Q}\left(\frac{1}{n} \sum_{i=0}^{n-1} \nabla f_{m,\pi_m^i}\left(x^t_{m, i}\right)\right), x^t-x^{\star}\right\rangle\right] \\
& +\frac{2 \eta^2 L_{\max }^2\left(1+\frac{\omega}{C}\right)}{M n} \sum_{m=1}^M \sum_{i=0}^{n-1}\left\|x^t_{m, i}-x^t\right\|^2\\
&+8 \eta^2 L_{\max }\left(1+\frac{\omega}{C}\right)\left(f\left(x^t\right)-f\left(x^{\star}\right)\right) \\
& +4 \eta^2\left(\frac{\omega}{C}+\frac{M-C}{C \max M-1,1}\right) \sigma_{\star}^2 \\
\leq & \left\|x^t-x^{\star}\right\|^2-2 \eta \frac{1}{M n} \sum_{m=1}^M \sum_{i=0}^{n-1}\left\langle\nabla f_m^{\pi i}\left(x^t_{m, i}\right), x^t-x^{\star}\right\rangle \\
& +\frac{2 \eta^2 L_{\max }^2\left(1+\frac{\omega}{C}\right)}{M n} \sum_{m=1}^M \sum_{i=0}^{n-1}\left\|x^t_{m, i}-x^t\right\|^2\\
&+8 \eta^2 L_{\max }\left(1+\frac{\omega}{C}\right)\left(f\left(x^t\right)-f\left(x^{\star}\right)\right) \\
& +4 \eta^2\left(\frac{\omega}{C}+\frac{M-C}{C \max M-1,1}\right) \sigma_{\star}^2 .
\end{aligned}
\end{aligned}
    \end{align*}
\begin{align*}
&\text { Using Lemma \ref{lem_inner_product}, we obtain }\\
&\begin{aligned}
&\mathbb{E}_{\mathcal{Q}, \set}\left[\left\|x^{t+1}-x^{\star}\right\|^2\right]\\
\leq & \left\|x^t-x^{\star}\right\|^2-\frac{\eta \mu}{2}\left\|x^t-x^{\star}\right\|^2-\eta\left(f\left(x^t\right)-f\left(x^{\star}\right)\right) \\
& +8 \eta^2 L_{\max }\left(1+\frac{\omega}{C}\right)\left(f\left(x^t\right)-f\left(x^{\star}\right)\right)\\
&+\frac{\eta L_{\max }}{M n} \sum_{m=1}^M \sum_{i=0}^{n-1}\left\|x^t_{m, i}-x^t\right\|^2 \\
& +\frac{2 \eta^2 L_{\max }^2\left(1+\frac{\omega}{C}\right)}{M n} \sum_{m=1}^M \sum_{i=0}^{n-1}\left\|x^t_{m, i}-x^t\right\|^2\\
&+4 \eta^2\left(\frac{\omega}{C}+\frac{M-C}{C \max M-1,1}\right) \sigma_{\star}^2 \\
\leq & \left(1-\frac{\eta \mu}{2}\right)\left\|x^t-x^{\star}\right\|^2-\eta\left(1-8 \eta L_{\max }\left(1+\frac{\omega}{C}\right)\right)\left(f\left(x^t\right)-f\left(x^{\star}\right)\right) \\
& +\frac{\eta L_{\max }\left(1+2 \eta L_{\max }\left(1+\frac{\omega}{C}\right)\right)}{M n} \sum_{m=1}^M \sum_{i=0}^{n-1}\left\|x^t_{m, i}-x^t\right\|^2+ \\
& 4 \eta^2\left(\frac{\omega}{C}+\frac{M-C}{C \max M-1,1}\right) \sigma_{\star}^2 .
\end{aligned}
\end{align*}

Using Lemma \ref{lem_dist}, we have 
\begin{align*}
&\mathbb{E}_{\mathcal{Q}, \set}\left[\left\|x^{t+1}-x^{\star}\right\|^2\right]\\
\leq & \left(1-\frac{\eta \mu}{2}\right)\left\|x^t-x^{\star}\right\|^2-\eta\left(1-8 \eta L\left(1+\frac{\omega}{C}\right)\right)\left(f\left(x^t\right)-f\left(x^{\star}\right)\right) \\
& +\eta L_{\max }\left(1+2 \eta L_{\max }\left(1+\frac{\omega}{C}\right)\right) \cdot 8 \gamma^2 n^2 L_{\max }\left(f\left(x^t\right)-f\left(x^{\star}\right)\right) \\
& +\eta L_{\max }\left(1+2 \eta L_{\max }\left(1+\frac{\omega}{C}\right)\right) \cdot 2 \gamma^2 n\left(\frac{1}{M} \sum_{m=1}^M \sigma_{\star, m}^2+n \sigma_{\star}^2\right) \\
& +4 \eta^2\left(\frac{\omega}{C}+\frac{M-C}{C \max M-1,1}\right) \sigma_{\star}^2 .
\end{align*}

Finally, we receive

$$
\begin{aligned}
&\mathbb{E}_{\mathcal{Q}, \set}\left[\left\|x^{t+1}-x^{\star}\right\|^2\right]\\
\leq & \left(1-\frac{\eta \mu}{2}\right)\left\|x^t-x^{\star}\right\|^2+4 \eta^2\left(\frac{\omega}{C}+\frac{M-C}{C \max M-1,1}\right) \sigma_{\star}^2 \\
& -\eta\left(1-8 \eta L_{\max }\left(1+\frac{\omega}{C}\right)\right)\left(f\left(x^t\right)-f\left(x^{\star}\right)\right) \\
&+8 \gamma^2 n^2 L_{\max }^2\left(1+2 L_{\max } \eta\left(1+\frac{\omega}{C}\right)\right)\left(f\left(x^t\right)-f\left(x^{\star}\right)\right)\\
& +2 \gamma^2 n \eta L_{\max }\left(1+2 \eta L\left(1+\frac{\omega}{C}\right)\right)\left(\frac{1}{M} \sum_{m=1}^M \sigma_{\star, m}^2+n \sigma_{\star}^2\right) \\
\leq & \left(1-\frac{\eta \mu}{2}\right)\left\|x^t-x^{\star}\right\|^2\\
&+4 \eta^2\left(\frac{\omega}{C}+\frac{M-C}{C \max M-1,1}\right) \sigma_{\star}^2 \\
& +\frac{9}{4} \eta L_{\max } \gamma^2 n\left(\frac{1}{M} \sum_{m=1}^M \sigma_{\star, m}^2+n \sigma_{\star}^2\right)
\end{aligned}
$$

Recursively rewriting the inequality and using $\sum_{t=0}^{+\infty}\left(1-\frac{\eta \mu}{2}\right)^t \leq \frac{2}{\mu \eta}$, we finish the proof.

\end{proof}
\clearpage

\subsection{Analysis of DIANA-NASTYA with partial participation}

\begin{algorithm}[t]
\caption{\algname{DIANA-NASTYA-PP}}
\label{alg:diana-nastya-pp}
	\begin{algorithmic}[1]
		\REQUIRE $x^0$ -- starting point, $\{h_{0,m}\}_{m=1}^M$ -- initial shift-vectors, $\gamma > 0$ -- local stepsize, $\eta > 0$ -- global stepsize, $\alpha > 0$ -- stepsize for learning the shifts
	    \FOR{$t =0,1,\dots, T-1$}
       \STATE Sample a cohort $\set$ with cardinality $C$ uniformly
    	   \FOR{$m \in \set$ in parallel}
    		    \STATE Receive $x^t$ from the server and set $x^0_{t,m} = x^t$
    		    \STATE Sample random permutation of $[n]$: $\pi_m = (\pi^0_m, \dots, \pi^{n-1}_m)$
    		    \FOR{$i = 0, 1,\dots, n-1$}
    		        \STATE Set $x^{t}_{m,i+1} = x^{t}_{m,i} - \gamma \nabla f_{m,\pi^{i}_m}(x^{t}_{m,i})$
    		    \ENDFOR
    		    \STATE Compute $g^{t}_{m} = \frac{1}{\gamma n}\left(x^t - x^t_{m,n}\right)$ and send $\cQ_t\left(g^{t}_{m} - h^{t}_{m}\right)$ to the server
    		    \STATE Set $h^{t+1}_{m} = h^{t}_{m} + \alpha\cQ_t\left(g^{t}_{m} - h^{t}_{m}\right)$
    		    \STATE Set $\hat{g}_{t,m} = h^{t}_{m} + \cQ_t\left(g^{t}_{m} - h^{t}_{m}\right)$
    		\ENDFOR
			\STATE $h_{t+1} = \frac{1}{C}\sum_{m\in \set}h^{t+1}_{m} = h_t + \frac{\alpha}{C}\sum_{m\in \set} \cQ_t\left(g^{t}_{m} - h^{t}_{m}\right)$
    		\STATE $\hat{g}^t = \frac{1}{C}\sum_{m\in \set}\hat{g}_{t,m} = h_t + \frac{1}{C}\sum_{m\in \set}\cQ_t\left(g^{t}_{m} - h^{t}_{m}\right)$
    		\STATE $x^{t+1} = x^{t} - \eta \hat{g}_t$
    	\ENDFOR
    	\ENSURE $x^T$
	\end{algorithmic}
\end{algorithm}

\begin{theorem}
    Let step sizes $\eta, \gamma$ satisfy the following equations

$$
\eta=\min \left(\frac{1}{80 L_{\max }\left(1+\frac{\omega}{C}\right)}, \frac{C}{\mu(1+\omega) M}\right), \quad \gamma=\frac{1}{5 n L_{\max }}
$$
Define the Lyapunov function:
$$\Psi^{(t)}=\left\|x^t-x^{\star}\right\|^2+\frac{A}{M} \sum_{m=1}^M\left\|h^{t}_{m}-h_m^{\star}\right\|^2,$$
where $A=\lambda \eta^2.$ Selecting parameters $\alpha=\frac{1}{1+\omega}$ ; $ \lambda=\frac{8 \omega}{\alpha M}$,$\gamma=\frac{1}{5 n L_{\max }}$, also using stepsize $ \eta \leq \min \left[ \frac{C}{\mu(1+\omega) M}, \frac{1}{80 L_{\max }\left(1+\frac{\omega}{C}\right)}\right]  $  Under Assumptions \ref{asm:quantization_operators}, \ref{asm:sc_general_f}, \ref{asm:lip_max_f_m} iterates of \algname{DIANA-NASTYA-PP} (Algorithm \ref{alg:diana-nastya-pp}) satisfy 

\begin{align*}
\mathbb{E}\left[\Psi^{(t)}\right]& \leq\left(1-\frac{\eta \mu}{2}\right)^T \mathbb{E}\left[\Psi_0\right]+\frac{3 \gamma^2 n^2 L_{\max }^2}{\mu}\left(\frac{1}{M} \sum_{m=1}^M \sigma_{\star, m}^2+n \sigma_{\star}^2\right)\\
&+\frac{2 \eta(M-C)}{\mu C \max (1, M-1)} \sigma_{\star}^2 .
\end{align*}

\end{theorem}

Note that we eliminate the variance term proportional to $\omega: 8 \frac{\eta}{\mu} \frac{\omega}{C} \sigma_{\star}^2$. In the Partial Participation regime, we have a variance term proportional to $\frac{(M-C)}{C \max (1, M-1)}$, which equals zero if $C=M$. This term decreases as $\mathcal{O}\left(\frac{1}{C}\right)$, so we achieve the expected linear speedup.

\begin{proof}
 STEP 1: we need to estimate the inner product. By $\hat{g}_t=\frac{1}{C} \sum_{m \in \set} \hat{g}_{t, m}$, we have 
    \begin{align*}
        \begin{aligned}
-\mathbb{E}_t\left[\left\langle\frac{1}{C} \sum_{m \in \set} \hat{g}_{t, m}, x^t-x^{\star}\right\rangle\right]= & -\left\langle\frac{1}{C} \mathbb{E}_t\left[\sum_{m \in \set} \hat{g}_{t, m}\right], x^t-x^{\star}\right\rangle \\
= & -\left\langle\frac{1}{M} \sum_{m=1}^M \mathbb{E}_t\left[\hat{g}_{t, m}\right], x^t-x^{\star}\right\rangle \\
= & -\frac{1}{M} \sum_{m=1}^M\left\langle g^{t}_{m}, x^t-x^{\star}\right\rangle \\
= & -\frac{1}{M} \sum_{m=1}^M\left\langle g^{t}_{m}-h_m^{\star}, x^t-x^{\star}\right\rangle \\
\leq & -\frac{\mu}{4}\left\|x^t-x^{\star}\right\|^2-\frac{1}{2}\left(f\left(x^t\right)-f\left(x^{\star}\right)\right)\\
&-\frac{1}{M n} \sum_{m=1}^M \sum_{i=0}^{n-1} D_{f_m^{\pi_m^{i m}}}\left(x^{\star}, x^t_{m, i}\right) \\
& +\frac{L_{\max }}{2 M n} \sum_{m=1}^M \sum_{i=0}^{n-1}\left\|x^t-x^t_{m, i}\right\|^2 .
\end{aligned}
    \end{align*}

\begin{align*}
    \begin{aligned}
&\text {STEP 2: We need to bound } \mathbb{E}\left\|\hat{g}_t\right\|^2 \text {. By } \hat{g}_t=\frac{1}{C} \sum_{m \in \set} \hat{g}_{t, m} \text {, we have }\\
&\begin{aligned}
\mathbb{E}_{\mathcal{Q}}\left[\left\|\hat{g}_t\right\|^2\right]= & \mathbb{E}_{\mathcal{Q}}\left[\left\|\frac{1}{C} \sum_{m \in \set}\left(h^{t}_{m}+\mathcal{Q}\left(g^{t}_{m}-h^{t}_{m}\right)-g^{t}_{m}+g^{t}_{m}\right)\right\|^2\right] \\
= & \mathbb{E}_{\mathcal{Q}}\left[\left\|\frac{1}{C} \sum_{m \in \set}\left(h^{t}_{m}+\mathcal{Q}\left(g^{t}_{m}-h^{t}_{m}\right)-g^{t}_{m}\right)\right\|^2\right]+\left\|\frac{1}{C} \sum_{m \in \set} g^{t}_{m}\right\|^2 \\
= & \frac{1}{C^2} \sum_{m \in \set} \mathbb{E}_{\mathcal{Q}}\left[\left\|h^{t}_{m}+\mathcal{Q}\left(g^{t}_{m}-h^{t}_{m}\right)\right\|^2\right]+\left\|\frac{1}{C} \sum_{m \in \set} g^{t}_{m}\right\|^2 \\
\leq & \frac{\omega}{C^2} \sum_{m \in \set}\left\|g^{t}_{m}-h^{t}_{m}\right\|^2+\left\|\frac{1}{C} \sum_{m \in \set} g^{t}_{m}\right\|^2 \\
\leq & \frac{2 \omega}{C^2} \sum_{m \in \set}\left\|g^{t}_{m}-\nabla f_m\left(x^t\right)\right\|^2+\frac{2 \omega}{C^2} \sum_{m \in \set}\left\|\nabla f_m\left(x^t\right)-h^{t}_{m}\right\|^2 \\
& +2\left\|\frac{1}{C} \sum_{m \in \set} g^{t}_{m}-h_m^{\star}\right\|^2+2\left\|\frac{1}{C} \sum_{m \in \set} h_m^{\star}\right\|^2
\end{aligned}
\end{aligned}
\end{align*}

Taking the expectation by subsampling, we have

\begin{align*}
&\mathbb{E}_{\mathcal{Q}, \set}\left[\left\|\hat{g}_t\right\|^2\right]\\ \leq & \frac{2 \omega}{C} \frac{1}{M} \sum_{m=1}^M\left\|g^{t}_{m}-\nabla f_m\left(x^t\right)\right\|^2+\frac{2 \omega}{C} \frac{1}{M} \sum_{m=1}^M\left\|\nabla f_m\left(x^t\right)-h^{t}_{m}\right\|^2 \\
& +\frac{2}{M} \sum_{m=1}^M\left\|g^{t}_{m}-h_m^{\star}\right\|^2+\frac{2(M-C)}{C(M-1) M} \sum_{m=1}^M\left\|h_m^{\star}\right\|^2 \\
\leq & \frac{2 \omega}{C} \frac{1}{M} \sum_{m=1}^M\left\|g^{t}_{m}-\nabla f_m\left(x^t\right)\right\|^2+\frac{2 \omega}{C} \frac{1}{M} \sum_{m=1}^M\left\|\nabla f_m\left(x^t\right)-h^{t}_{m}\right\|^2 \\
& +\frac{4}{M} \sum_{m=1}^M\left\|g^{t}_{m}-\nabla f_m\left(x^t\right)\right\|^2+\frac{4}{M} \sum_{m=1}^M\left\|\nabla f_m\left(x^t\right)-h_m^{\star}\right\|^2 \\
& +\frac{2(M-C)}{C(M-1) M} \sum_{m=1}^M\left\|h_m^{\star}\right\|^2 \\
\leq & 4\left(1+\frac{\omega}{C}\right) \frac{L_{\max }^2}{M n} \sum_{m=1}^M \sum_{i=0}^{n-1}\left\|x^t_{m, i}-x^t\right\|^2+\frac{2 \omega}{C} \frac{1}{M} \sum_{m=1}^M\left\|\nabla f_m\left(x^t\right)-h^{t}_{m}\right\|^2 \\
& +\frac{8 L_{\max }}{M} \sum_{m=1}^M D_{f_m}\left(x^t, x^{\star}\right)+\frac{2(M-C)}{C(M-1) M} \sum_{m=1}^M\left\|h_m^{\star}\right\|^2 \\
= & \left(1+\frac{\omega}{C}\right) \frac{L_{\max }^2}{M n} \sum_{m=1}^M \sum_{i=0}^{n-1}\left\|x^t_{m, i}-x^t\right\|^2+\frac{2 \omega}{C} \frac{1}{M} \sum_{m=1}^M\left\|\nabla f_m\left(x^t\right)-h^{t}_{m}\right\|^2 \\
& +8 L_{\max }\left(f\left(x^t\right)-f\left(x^{\star}\right)\right)+\frac{2(M-C)}{C(M-1) M} \sum_{m=1}^M\left\|h_m^{\star}\right\|^2
\end{align*}

Thus, we have
\begin{align*}
&\mathbb{E}_{\mathcal{Q}, \set}\left[\left\|x^{t+1}-x^{\star}\right\|^2\right]\\ \leq & \left(1-\frac{\eta \mu}{2}\right)\left\|x^t-x^{\star}\right\|^2-\eta\left(1-4 L_{\max } \eta\right)\left(f\left(x^t\right)-f\left(x^{\star}\right)\right) \\
& +\eta L_{\max }\left(1+4\left(1+\frac{\omega}{C}\right) L_{\max } \eta\right) \frac{1}{M n} \sum_{m=1}^M \sum_{i=0}^{n-1}\left\|x^t_{m, i}-x^t\right\|^2 \\
& +\frac{2 \eta^2 \omega}{C} \frac{1}{M} \sum_{m=1}^M\left\|\nabla f_m\left(x^t\right)-h^{t}_{m}\right\|^2+\frac{2 \eta^2(M-C)}{C(M-1) M} \sum_{m=1}^M\left\|h_m^{\star}\right\|^2.
\end{align*}

STEP 3: Note that
\begin{align*}
\frac{1}{M} \sum_{m=1}^M\left\|h_{t+1, m}-h_m^{\star}\right\|^2&=\frac{C}{M} \frac{1}{C} \sum_{m \in \set}\left\|h_{t+1, m}-h_m^{\star}\right\|^2\\
&+\frac{M-C}{M} \frac{1}{M-C} \sum_{m \notin \set}\left\|h_{t+1, m}-h_m^{\star}\right\|^2 .
\end{align*}

Taking the expectation by compression, we have

\begin{align*}
& \mathbb{E}_{\mathcal{Q}}\left[\frac{1}{C} \sum_{m \in \set}\left\|h_{t+1, m}-h_m^{\star}\right\|^2\right]=\mathbb{E}_{\mathcal{Q}}\left[\frac{1}{C} \sum_{m \in \set}\left\|h^{t}_{m}+\alpha \mathcal{Q}\left(g^{t}_{m}-h^{t}_{m}\right)-h_m^{\star}\right\|^2\right] \\
& =\frac{1}{C} \sum_{m \in \set}\left(\left\|h^{t}_{m}-h_m^{\star}\right\|^2+2 \alpha\left\langle g^{t}_{m}-h^{t}_{m}, h^{t}_{m}-h_m^{\star}\right\rangle\right) \\
&+\frac{1}{C} \sum_{m \in \set}\alpha^2(1+\omega)\left\|g^{t}_{m}-h^{t}_{m}\right\|^2\\
& \stackrel{\alpha \leq 1 / 1+\omega}{\leq} \frac{1}{C} \sum_{m \in \set}\left(\left\|h^{t}_{m}-h_m^{\star}\right\|^2+2 \alpha\left\langle g^{t}_{m}-h^{t}_{m}, h^{t}_{m}-h_m^{\star}\right\rangle+\alpha\left\|g^{t}_{m}-h^{t}_{m}\right\|^2\right) \\
& =\frac{1-\alpha}{C} \sum_{m \in \set}\left\|h^{t}_{m}-h_m^{\star}\right\|^2+\frac{\alpha}{C} \sum_{m \in \set}\left\|g^{t}_{m}-h^{t}_{m}\right\|^2 .
\end{align*}

Taking the expectation by subsampling, we have
\begin{align*}
&\mathbb{E}_{\mathcal{Q}, \set}\left[\frac{1}{C} \sum_{m \in \set}\left\|h_{t+1, m}-h_m^{\star}\right\|^2\right]\\
& \leq \mathbb{E}_{\set}\left[\frac{1-\alpha}{C} \sum_{m \in \set}\left\|h^{t}_{m}-h_m^{\star}\right\|^2+\frac{\alpha}{C} \sum_{m \in \set}\left\|g^{t}_{m}-h_m^{\star}\right\|^2\right] \\
& =\frac{1-\alpha}{M} \sum_{m=1}^M\left\|h^{t}_{m}-h_m^{\star}\right\|^2+\frac{\alpha}{M} \sum_{m=1}^M\left\|g^{t}_{m}-h_m^{\star}\right\|^2 .
\end{align*}

Thus, we have
\begin{align*}
&\mathbb{E}_{\set, \mathcal{Q}^t}\left[\frac{1}{M} \sum_{m=1}^M\left\|h_{t+1, m}-h_m^{\star}\right\|^2\right]\\
= & \frac{(1-\alpha) C}{M^2} \sum_{m=1}^M\left\|h^{t}_{m}-h_m^{\star}\right\|^2+\frac{\alpha C}{M^2} \sum_{m=1}^M\left\|g^{t}_{m}-h_m^{\star}\right\|^2 \\
& +\frac{M-C}{M} \mathbb{E}_{\set, \mathcal{Q}^t}\left[\frac{1}{M-C} \sum_{m \notin \set}\left\|h^{t}_{m}-h_m^{\star}\right\|^2\right] \\
= & \frac{(1-\alpha) C}{M^2} \sum_{m=1}^M\left\|h^{t}_{m}-h_m^{\star}\right\|^2+\frac{\alpha C}{M^2} \sum_{m=1}^M\left\|g^{t}_{m}-h_m^{\star}\right\|^2 \\
& +\frac{M-C}{M} \frac{1}{M} \sum_{m=1}^M\left\|h^{t}_{m}-h_m^{\star}\right\|^2 \\
\leq & \left(1-\frac{\alpha C}{M}\right) \frac{1}{M} \sum_{m=1}^M\left\|h^{t}_{m}-h_m^{\star}\right\|^2 \\
& +\frac{2 \alpha L_{\max }^2 C}{M^2 n} \sum_{m=1}^M \sum_{i=0}^{n-1}\left\|x^t_{m, i}-x^t\right\|^2\\
&+\frac{4 L_{\mathrm{max}} \alpha C}{M^2} \sum_{m=1}^M D_{f_m}\left(x^t, x^{\star}\right).
\end{align*}

STEP 4: Defining Lyapunov function as follows
    \begin{align*}
&\Psi^{(t)}=\left\|x^t-x^{\star}\right\|^2+\frac{A}{M} \sum_{m=1}^M\left\|h^{t}_{m}-h_m^{\star}\right\|^2,\\
\end{align*}
 we have 
\begin{align*}
&\mathbb{E}_{\mathcal{Q}, \set}\left[\Psi^{(t+1)}\right]\\
\leq & \left(1-\frac{\eta \mu}{2}\right)\left\|x^t-x^{\star}\right\|^2-\eta\left(1-4 L_{\max } \eta\right)\left(f\left(x^t\right)-f\left(x^{\star}\right)\right) \\
& +\eta L_{\max }\left(1+4\left(1+\frac{\omega}{C}\right) L_{\max } \eta\right) \frac{1}{M n} \sum_{m=1}^M \sum_{i=0}^{n-1}\left\|x^t_{m, i}-x^t\right\|^2 \\
& +\frac{2 \eta^2 \omega}{C} \frac{1}{M} \sum_{m=1}^M\left\|\nabla f_m\left(x^t\right)-h^{t}_{m}\right\|^2+\frac{2 \eta^2(M-C)}{C(M-1) M} \sum_{m=1}^M\left\|h_m^{\star}\right\|^2 \\
& +\left(1-\frac{\alpha C}{M}\right) \frac{A}{M} \sum_{m=1}^M\left\|h^{t}_{m}-h_m^{\star}\right\|^2 \\
& +\frac{2 \alpha L_{\max }^2 A C}{M^2 n} \sum_{m=1}^M \sum_{i=0}^{n-1}\left\|x^t_{m, i}-x^t\right\|^2+\frac{4 L_{\max } \alpha A C}{M}\left(f\left(x^t\right)-f\left(x^{\star}\right)\right).
\end{align*}

$\text { Setting } A=\lambda \eta^2 \text { and using Lemma \ref{lem_dist}, we have }$

\begin{align*}
  &  \mathbb{E}\left[\Psi^{(t+1)}\right]\\
  \leq & \left(1-\frac{\eta \mu}{2}\right) \mathbb{E}\left[\left\|x^t-x^{\star}\right\|^2\right]+\left(1-\frac{\alpha C}{M}+\frac{4 \omega}{\lambda C}\right) \frac{\lambda \eta^2}{M} \sum_{m=1}^M \mathbb{E}\left[\left\|h^{t}_{m}-h_m^{\star}\right\|^2\right] \\
& -\eta\left(1-8 \eta L_{\max }\left(1+\frac{\omega}{C}\right)-4 \eta L_{\max } \alpha \lambda \frac{C}{M}\right) \mathbb{E}\left[f\left(x^t\right)-f\left(x^{\star}\right)\right] \\
& +8 \gamma^2 n^2 L_{\max }^2 \eta\left(1+4 \eta L_{\max }\left(1+\frac{\omega}{C}\right)+2 \eta L_{\max } \alpha \lambda \frac{C}{M}\right) \mathbb{E}\left[f\left(x^t\right)-f\left(x^{\star}\right)\right] \\
& +2 \gamma^2 n^2 L_{\max }^2 \eta\left(1+4 \eta L_{\max }\left(1+\frac{\omega}{C}\right)+2 \eta L_{\max } \alpha \lambda \frac{C}{M}\right)\left(\frac{1}{M} \sum_{m=1}^M \sigma_{\star, m}^2+n \sigma_{\star}^2\right) \\
& +\frac{2 \eta^2(M-C)}{C(M-1)} \sigma_{\star}^2.
\end{align*}

 Set  $\alpha=\frac{1}{1+\omega}$, $\lambda=\frac{8 \omega}{\alpha M}$, $\eta \leq \frac{C}{\mu(1+\omega) M}$, also set  $\eta=\frac{1}{80 L_{\max }\left(1+\frac{\omega}{C}\right)}$, $\gamma=\frac{1}{5 n L_{\max }}$ and applying previous steps we obtain
\begin{align*}
\mathbb{E}\left[\Psi^{(t+1)}\right] \leq & \left(1-\frac{\eta \mu}{2}\right) \mathbb{E}\left[\Psi^{(t)}\right]\\
&+3 \gamma^2 n^2 L_{\max }^2 \eta\left(\frac{1}{M} \sum_{m=1}^M \sigma_{\star, m}^2+n \sigma_{\star}^2\right)+\frac{2 \eta^2(M-C)}{C(M-1)} \sigma_{\star}^2 \\
& -\eta\left(\frac{1}{2}-10 \gamma^2 n^2 L_{\max }^2\right) \mathbb{E}\left[f\left(x^t\right)-f\left(x^{\star}\right)\right] \\
\leq & \left(1-\frac{\eta \mu}{2}\right) \mathbb{E}\left[\Psi^{(t)}\right]\\
&+3 \gamma^2 n^2 L_{\max }^2 \eta\left(\frac{1}{M} \sum_{m=1}^M \sigma_{\star, m}^2+n \sigma_{\star}^2\right)+\frac{2 \eta^2(M-C)}{C(M-1)} \sigma_{\star}^2,
\end{align*}

\end{proof}

\newpage

            \refstepcounter{chapter}%
\chapter*{\thechapter \quad Appendix G Title}
\label{appendixG}

\section{Extra Related Work}\label{appendix:extra_related_work}

\paragraph{Further Comparison with work \citep{data2021byzantine}.} As we mention in the main text, in \citep{data2021byzantine} it is assumed that $3B$ is smaller than $C$. More precisely, in \citep{data2021byzantine} the authors assume that $B \leq \epsilon C$, where $\epsilon \leq \frac{1}{3} - \epsilon'$ for some parameter $\epsilon' > 0$ that will be explained later. That is, the results in \citep{data2021byzantine} do not hold when $C$ is smaller than $3B$, and, in particular, their algorithm cannot tolerate the situation when the server samples only Byzantine workers at some particular communication round. We also notice that when $C \geq 4B$, then existing methods such as \gls{Byz-VR-MARINA} \citep{gorbunov2023variance} or Client Momentum \citep{karimireddy2021learning, karimireddy2020byzantine} can be applied without any changes to get a provable convergence.

Next, in \citep{data2021byzantine} the authors derive the upper bounds for the expected squared distance to the solution (in the strongly convex case) and the averaged expected squared norm of the gradient (in the non-convex case), where the expectation is taken w.r.t.\ the sampling of stochastic gradients only and the bounds themselves hold with probability at least $1 - \frac{K}{H}\exp\left( - \frac{\epsilon'^2(1 - \epsilon)C}{16}\right)$, where $H$ is the number of local steps. For simplicity consider the best-case scenario: $H = 1$ (local steps deteriorate the results in \citep{data2021byzantine}). Then, the lower bound for this probability becomes negative when either $C$ is not large enough or when $K$ is large or when $\epsilon$ is close to $\frac{1}{3}$, e.g., for $K = 10^6, \epsilon = \epsilon' = \frac{1}{6}, C = 5000$ this lower bound is smaller than $-720$, meaning that in this case, the result does not guarantee convergence. In contrast, our results have classical convergence criteria, where the expectations are taken w.r.t.\ all randomness.

Finally, the bounds in \citep{data2021byzantine} have irreducible terms even for the homogeneous data case: these terms are proportional to $\frac{\sigma^2}{b}$, where $\sigma^2$ is the upper bound for the variance of the stochastic estimator on regular clients and $b$ is the batchsize. In contrast, our results have only decreasing terms in the upper bounds when the data is homogeneous.

\paragraph{Byzantine robustness.} There exist various approaches to achieving Byzantine robustness \citep{lyu2020privacy}. Results in \citep{alistarh2018byzantine, allen2020byzantine} rely on the concentration inequalities for the stochastic gradients with bounded noise to iteratively remove them from the training. In work \citep{karimireddy2021learning} the authors formalize the definition of robust aggregation and propose the first provably robust aggregation rule called \algname{CenteredClip} and the first provably Byzantine robust method under bounded variance assumption for homogeneous problems, i.e., when all good workers share one dataset. In particular, the method from \citep{karimireddy2021learning} uses client momentum on the clients that helps to memorize previous steps for good workers and withstand time-coupled attacks. This approach is extended in \citep{he2022byzantine} to the setup of decentralized learning. In work \citep{allouah2023fixing} the authors develop an alternative definition for robust aggregation and propose a new aggregation rule satisfying their definition. Work \citep{karimireddy2020byzantine} generalizes these results to the heterogeneous data case and derives lower bounds for the optimization error that one can achieve in the heterogeneous case. Based on the formalism in \citep{karimireddy2021learning}, in work \citep{gorbunov2021secure} the authors propose a server-free approach that uses random checks of computations and bans of peers. This trick allows the elimination of all Byzantine workers after a finite number of steps on average. There are also many other approaches, e.g., one can use redundant computations of the stochastic gradients \citep{chen2018draco, rajput2019detox} or introduce reputation metrics \citep{rodriguez2020dynamic, regatti2020bygars, xu2020towards} to achieve some robustness, see also a recent survey in \citep{lyu2020privacy}.

\paragraph{Variance reduction.} The literature on variance-reduced methods is very rich \citep{gower2020variance}. The first variance-reduced methods are designed to fix the convergence of standard Stochastic Gradient Descent (\algname{SGD}) and make it convergent to any predefined accuracy even with constant stepsizes. Such methods as \algname{SAG} \citep{schmidt2017minimizing}, \algname{SVRG} \citep{johnson2013accelerating}, \algname{SAGA} \citep{defazio2014saga} are developed mainly for (strongly) convex smooth optimization problems, while methods like \algname{SARAH} \citep{nguyen2017sarah}, \algname{STORM} \citep{cutkosky2019momentum}, \algname{GeomSARAH} \citep{DIANA2}, \algname{PAGE} \citep{li2021page} are designed for general smooth non-convex problems. In this paper, we use \algname{GeomSARAH}/\algname{PAGE}-type variance reduction as the main building block of the method that makes the method robust to Byzantine attacks.

\paragraph{Partial Participation.} In the context of Byzantine robust learning, there exists one work that develops and analyzes the method with partial participation \citep{data2021byzantine}. However, this work relies on the restrictive assumption that the number of participating clients at each round is at least three times larger than the number of Byzantine workers. In this case, Byzantines cannot form a majority, and standard methods can be applied without any changes. In contrast, our method converges in more challenging scenarios, e.g., \gls{Byz-VR-MARINA-PP} provably converges even when the server samples one client, which can be Byzantine. The results in \citep{data2021byzantine} have some other noticeable limitations that we discuss in Appendix~\ref{appendix:extra_related_work}.

\paragraph{Communication compression.} The literature on communication compression can be roughly divided into two huge groups. The first group studies the methods with unbiased communication compression. Different compression operators in the application to Distributed \algname{SGD}/\algname{GD} are studied in \citep{alistarh2017qsgd, wen2017terngrad, khirirat2018distributed}. To improve the convergence rate by fixing the error coming from the compression it is proposed to apply compression to the special gradient differences \citep{mishchenko2019distributed}. Multiple extensions and generalizations of mentioned techniques are proposed and analyzed in the literature, e.g., see \citep{DIANA2, gorbunov2021marina, li2020acceleration, qian2021error, basu2019qsparse, haddadpour2021federated, sadiev2022federated, islamov2021distributed, safaryan2021fednl}. 

Another large part of the literature on compressed communication is devoted to biased compression operators \citep{ajalloeian2020convergence, demidovich2023guide}. Typically, such compression operators require more algorithmic changes than unbiased compressors since na\"ive combinations of biased compression with standard methods (e.g., Distributed \algname{\gls{GD}}) can diverge \citep{beznosikov20_biased_compr_distr_learn}. Error feedback is one of the most popular ways of utilizing biased compression operators in practice \citep{seide20141, stich2018sparsified, vogels2019powersgd}, see also \citep{richtarik2021ef21, fatkhullin2021ef21} for the modern version of error feedback with better theoretical guarantees for non-convex problems.

In the context of Byzantine robustness, methods with communication compression are also studied. The existing approaches are based on aggregation rules based on the norms of the updates \citep{ghosh2020distributed, ghosh2021communication}, \algname{SignSGD} and majority vote \citep{bernstein2018signsgd}, \algname{SAGA}-type variance reduction coupled with unbiased compression \citep{zhu2021broadcast}, and \algname{GeomSARAH}/\algname{PAGE}-type variance reduction combined with unbiased compression \citep{gorbunov2023variance}.

\paragraph{Gradient clipping.} Gradient clipping has multiple useful properties and applications. Originally it was used in work \citep{pascanu2013difficulty} to reduce the effect of exploding gradients during the training of RNNs. Gradient clipping is also a popular tool for achieving provable differential privacy \citep{abadi2016deep, chen2020understanding}, convergence under generalized notions of smoothness \citep{zhang2019gradient, mai2021stability} and better (high-probability) convergence under heavy-tailed noise assumption \citep{zhang2020adaptive, nazin2019algorithms, gorbunov2020stochastic, sadiev2023high, nguyen2023improved}. In the context of Byzantine-robust learning, gradient clipping is also utilized to design provably robust aggregation \citep{karimireddy2021learning}. Our work proposes a novel useful application of clipping, i.e., we utilize clipping to achieve Byzantine robustness with partial participation of clients.

\paragraph{Byzantine-robust asynchronous methods.} Byzantine-robust asynchronous methods are also very relevant to the problem of partial participation in Byzantine-robust learning. Indeed, the asynchronous methods like Asynchronous \algname{SGD} \citep{agarwal2011distributed, nedic2001distributed} naturally have partial participation since whenever some worker finishes the computation (of the stochastic gradients), this worker immediately sends the update to the server and the server applies this update without waiting for all other clients. However, without extra assumptions asynchronous methods cannot tolerate Byzantine attacks: Byzantine clients could immediately send any vector to the server to guarantee that their update is received earlier than the updates from regular clients. Clearly, such a behavior of Byzantine workers leads to the divergence of the method unless the server has additional information that can be used for acceptance/rejection of the update or some other alteration of the communication protocol preventing the situations when some client updates the model too many times in a row is applied.

Therefore, the existing approaches addressing this important problem rely on extra assumptions. In work \citep{damaskinos2018asynchronous} the authors propose to use a Lipschitz filter and frequency filters in order to filter out Byzantine workers. Next, work \citep{xie2020zeno++, fang2022aflguard} uses additional validation data on the server to decide whether to accept the update from workers. This assumption is restrictive for many \gls{FL} applications when the data on clients is private and is not available on the server. In work \citep{yang2023buffered} authors propose so-called \algname{BASGD} (and its momentum version) where the key idea is to split workers into buffers and wait until each buffer gets at least one gradient update. In the case when the number of buffers is sufficiently large (at least $2B$, where $B$ is the number of Byzantine workers), the authors show that \algname{BASGD} converges. However, this means that to make the step \algname{BASGD} requires collecting a sufficiently large number of gradients such that the good buffers form a majority, which is closer to full participation than to the partial participation in the worst case.

We emphasize that in our work we consider a different setup of synchronous communications with partial participation. The approaches discussed in the above paragraph cannot be directly applied to the problem considered in this paper without extra assumptions.

\clearpage

\section{Useful Facts}

For all $a, b \in \mathbb{R}^d$ and $\alpha>0, p \in(0,1]$ the following relations hold:

\begin{align}
\label{eq:quadratic} 2\langle a, b\rangle & =\|a\|^2+\|b\|^2-\|a-b\|^2 \\
\label{eq:yung-1}\|a+b\|^2 & \leq(1+\alpha)\|a\|^2+\left(1+\alpha^{-1}\right)\|b\|^2 \\
\label{eq:yung-2}-\|a-b\|^2 & \leq-\frac{1}{1+\alpha}\|a\|^2+\frac{1}{\alpha}\|b\|^2, \\
\label{eq:contract}(1-p)\left(1+\frac{p}{2}\right) & \leq 1-\frac{p}{2}, \quad p\geq 0 \\
\label{eq:contract-2}(1-p)\left(1+\frac{p}{2}\right)\left(1+\frac{p}{4}\right) & \leq 1-\frac{p}{4}\quad p\geq 0 .
\end{align}

\begin{lemma} (Lemma 5 from \citep{richtarik2021ef21}).
\label{lemma:peter}
Let $a, b>0$. If $0 \leq \gamma \leq \frac{1}{\sqrt{a}+b}$, then $a \gamma^2+b \gamma \leq 1$. The bound is tight up to the factor of 2 since $\frac{1}{\sqrt{a}+b} \leq \min \left\{\frac{1}{\sqrt{a}}, \frac{1}{b}\right\} \leq \frac{2}{\sqrt{a}+b}$.
\end{lemma}

\clearpage

\section{Justification of Assumption~\ref{assm:bounded-aggr}}\label{appendix:justification_of_Assumption_1}

\begin{algorithm}[h]
   \caption{Bucketing Algorithm \citep{karimireddy2020byzantine}}\label{alg:bucketing}
\begin{algorithmic}[1]
   \STATE {\bfseries Input:} $\{x_1,\ldots,x_n\}$, $s \in \mathbb{N}$ -- bucket size, \texttt{Aggr} -- aggregation rule 
   \STATE Sample random permutation $\pi = (\pi(1),\ldots, \pi(n))$ of $[n]$
   \STATE Compute $y_i = \frac{1}{s}\sum_{k = s(i-1)+1}^{\min\{si, n\}} x_{\pi(k)}$ for $i = 1, \ldots, \lceil \nicefrac{n}{s} \rceil$
   \STATE {\bfseries Return:} $\widehat x = \texttt{Aggr}(y_1, \ldots, y_{\lceil \nicefrac{n}{s} \rceil})$
\end{algorithmic}
\end{algorithm}

\paragraph{Krum and Krum $\circ$ Bucketing.} Krum aggregation rule is defined as
\begin{equation}
    \text{Krum}(x_1,\ldots, x_n) = \argmin\limits_{x_m \in \{x_1,\ldots, x_n\}} \sum\limits_{j \in S_i}\|x_j - x_i\|^2, \notag
\end{equation}
where $S_i \subset \{x_1,\ldots, x_n\}$ is the subset of $n-B-2$ closest vectors to $x_i$. By definition, $\text{Krum}(x_1,\ldots, x_n) \in \{x_1,\ldots, x_n\}$ and, thus we have the inequality $\|\text{Krum}(x_1,\ldots, x_n)\| \leq \max_{i\in [n]} \|x_i\|$, i.e., Assumption~\ref{assm:bounded-aggr} holds with $F_{\cA} = 1$. Since Krum $\circ$ Bucketing applies Krum aggregation to averages $y_i$ over the buckets and $\|y_i\| \leq \frac{1}{s}\sum_{k = s(i-1)+1}^{\min\{si, n\}} \|x_{\pi(k)}\| \leq \max_{i\in [n]} \|x_i\|$, we have that $\|\text{Krum}\circ\text{Bucketing}(x_1,\ldots, x_n)\| \leq \max_{i\in [n]} \|x_i\|$.

\paragraph{Geometric median (GM) and GM $\circ$ Bucketing.} Geometric median is defined as follows:
\begin{equation}
    \text{GM}(x_1,\ldots, x_n) = \argmin\limits_{x \in \R^d} \sum\limits_{i=1}^n \|x - x_i\|. \label{eq:GM}
\end{equation}
One can show that $\text{GM}(x_1,\ldots, x_n) \in \text{Conv}(x_1, \ldots, x_n) \eqdef \{x \in \R^d \mid x = \sum_{i=1}^n \alpha_i x_i \text{ for some } \alpha_1, \ldots, \alpha_n \geq 1 \text{ such that } \sum_{i=1}^n\alpha_i = 1\}$, i.e., geometric median belongs to the convex hull of the inputs. Indeed, let $\text{GM}(x_1,\ldots, x_n) = x = \hat x + \tilde x$, where $\hat x$ is the projection of $x$ on $\text{Conv}(x_1, \ldots, x_n)$ and $\tilde x = x - 
\hat x$. Then, the optimality condition implies that $\langle \hat x - x, y - \hat x \rangle \geq 0$ for all $y \in \text{Conv}(x_1, \ldots, x_n)$. In particular, for all $m \in [n]$ we have $\langle \hat x - x, x_i - \hat x \rangle \geq 0$. Since
\begin{eqnarray*}
\langle \hat x - x, x_i - \hat x \rangle &=& \langle \tilde x, \hat x - x_i \rangle = \frac{1}{2}\|\tilde x + \hat x - x_i\|^2 - \frac{1}{2}\|\tilde x\|^2 - \frac{1}{2}\|\hat x - x_i\|^2\\
&=& \frac{1}{2}\|x - x_i\|^2 - \frac{1}{2}\|\tilde x\|^2 - \frac{1}{2}\|\hat x - x_i\|^2\\
&\leq& \frac{1}{2}\|x - x_i\|^2 - \frac{1}{2}\|\hat x - x_i\|^2,
\end{eqnarray*}
we get that $\|x - x_i\| \geq \|\hat x - x_i\|$ for all $i\in [n]$ and the equality holds if and only if $\tilde x = 0$. Therefore, $\argmin$ from \eqref{eq:GM} is achieved for $x$ such that $x = \hat x$, meaning that $\text{GM}(x_1,\ldots, x_n) \in \text{Conv}(x_1, \ldots, x_n)$. Therefore, there exist some coefficients $\alpha_1, \ldots, \alpha_n \geq 0$ such that $\sum_{i=1}^n \alpha_i = 1$ and $\text{GM}(x_1,\ldots, x_n) = \sum_{i=1}^n \alpha_i x_i$, implying that
\begin{equation}
\|\text{GM}(x_1,\ldots, x_n)\| \leq \sum\limits_{i=1}^n \alpha_i \|x_i\| \leq \max\limits_{i\in [n]} \|x_i\|. \notag
\end{equation}
That is, GM satisfies Assumption~\ref{assm:bounded-aggr} with $F_{\cA} = 1$. Similarly to the case of Krum $\circ$ Bucketing, we also have $\|\text{GM}\circ\text{Bucketing}(x_1,\ldots, x_n)\| \leq \max_{i\in [n]} \|x_i\|$.

\paragraph{Coordinate-wise median (CM) and CM $\circ$ Bucketing.} \gls{CM} is formally defined as
\begin{equation}
    \text{CM}(x_1,\ldots, x_n) = \argmin\limits_{x \in \R^d} \sum\limits_{i=1}^n \|x - x_i\|_1, \label{eq:CM}
\end{equation}
where $\|\cdot\|_1$ denotes $\ell_1$-norm. This is equivalent to geometric median/median applied to vectors $x_1,\ldots, x_n$ component-wise. Therefore, from the above derivations for GM we have
\begin{eqnarray*}
    \|\text{CM}(x_1,\ldots, x_n)\|_{\infty} &\leq& \max_{i\in [n]}\|x_i\|_{\infty},\\
    \|\text{CM}\circ\text{Bucketing}(x_1,\ldots, x_n)\|_{\infty} &\leq& \max_{i\in [n]}\|x_i\|_{\infty},
\end{eqnarray*}
where $\|\cdot\|_{\infty}$ denotes $\ell_{\infty}$-norm. Therefore, due to the standard relations between $\ell_2$- and $\ell_\infty$-norms, i.e., $\|a\|_\infty \leq \|a\| \leq \sqrt{d}\|a\|_{\infty}$ for any $a \in \R^d$, we have
\begin{eqnarray*}
    \|\text{CM}(x_1,\ldots, x_n)\| &\leq& \sqrt{d}\max_{i\in [n]}\|x_i\|,\\
    \|\text{CM}\circ\text{Bucketing}(x_1,\ldots, x_n)\| &\leq& \sqrt{d}\max_{i\in [n]}\|x_i\|,
\end{eqnarray*}
i.e., Assumption~\ref{assm:bounded-aggr} is satisfied with $F_{\cA} = \sqrt{d}$.

\clearpage

\section{General Analysis}

\subsection{Refined assumptions}\label{appendix:refined_assumptions}

For simplicity, in the main part of our paper, we present simplified versions of our main results. However, our analysis works under more general assumptions presented in this section.

\paragraph{Assumption on $\widehat{C}$.} In all the results of this paper, we assume that $n \geq \widehat C \geq \max\{1, \nicefrac{\deltar M}{\delta}\}$. This condition ensures that the robust aggregation makes sense when $c_t = 1$, i.e., at least $1-\delta$ proportion of sampled workers are not Byzantine ones when $c_t = 1$.

\paragraph{Refined smoothness.} The following assumption is classical for the literature on non-convex optimization.
\begin{assumption}[$L$-smoothness]
\label{assm:L-smoothness}
 We assume that function $f: \mathbb{R}^d \rightarrow \mathbb{R}$ is L-smooth, i.e., for all $x, y \in \mathbb{R}^d$ we have
 \begin{equation}
     \|\nabla f(x)-\nabla f(y)\| \leq L\|x-y\|. \label{eq:f_smooth}
 \end{equation}
 Moreover, we assume that $f$ is uniformly lower bounded by $f^{\star} \in \mathbb{R}$, i.e., $f^{\star}\eqdef\inf _{x \in \mathbb{R}^d} f(x)$. In addition, we assume that $f_m$ is $L_m$-smooth for all $m\in\cG$, i.e., for all $x, y \in \mathbb{R}^d$
\begin{equation}
    \|\nabla f_m(x) - \nabla f_m(y)\| \leq L_m \|x - y\|. \label{eq:f_m_smooth}
\end{equation}
\end{assumption}
We notice here that \eqref{eq:f_m_smooth} implies $L$-smoothness of $f$ with $L \leq \frac{1}{G}\sum_{m \in \cG} L_m$, i.e., smoothness constant of $f$ can be better than the averaged smoothness constant of the local loss functions on the regular clients.

Following ideas in \citep{gorbunov2023variance}, we consider refined assumptions on the smoothness.
\begin{assumption}[Global Hessian variance assumption \citep{szlendak2021permutation}] 
\label{assm:global} We assume that there exists $L_{ \pm} \geq 0$ such that for all $x, y \in \mathbb{R}^d$
\begin{align}
   \frac{1}{G} \sum_{m \in \mathcal{G}}\left\|\nabla f_m(x)-\nabla f_m(y)\right\|^2-\|\nabla f(x)-\nabla f(y)\|^2\leq L_{ \pm}^2\|x-y\|^2. \label{eq:global_hessian_variance}
\end{align}
\end{assumption}

We notice that \eqref{eq:f_m_smooth} implies \eqref{eq:global_hessian_variance} with $L_\pm \leq \max_{m\in \cG} L_m$. It is proved that $L_\pm$ satisfies the following relation: $L_{\mathrm{avg}}^2 - L^2 \leq L_\pm^2 \leq L_{\mathrm{avg}}^2$, where $L_{\mathrm{avg}}^2 \eqdef \frac{1}{G}\sum_{m\in \cG}L_m^2$ \citep{szlendak2021permutation}. In particular, it is possible that $L_\pm = 0$ even if the data on the good workers is heterogeneous.

\begin{assumption}[Local Hessian variance assumption \citep{gorbunov2023variance}]
\label{assm:local}
We assume that there exists $\mathcal{L}_{ \pm} \geq 0$ such that for all $x, y \in \mathbb{R}^d$
\begin{equation}
    \frac{1}{G} \sum_{m \in \mathcal{G}} \mathbb{E}\left\|\widehat{\Delta}_m(x, y)-\Delta_m(x, y)\right\|^2 \leq \frac{\mathcal{L}_{ \pm}^2}{b}\|x-y\|^2, \label{eq:local_hessian_var}
\end{equation}
where $\Delta_m(x, y)\eqdef\nabla f_m(x)-\nabla f_m(y)$ and $\widehat{\Delta}_m(x, y)$ is an unbiased mini-batched estimator of $\Delta_m(x, y)$ with batch size $b$.
\end{assumption}

This assumption incorporates considerations for the smoothness characteristics inherent in all functions $\{f_{m,i}\}_{m\in \cG, i\in [n]}$, the sampling policy, and the similarity among the functions $\{f_{m,i}\}_{m\in \cG, i\in [n]}$. The authors have demonstrated that, assuming smoothness of $\{f_{m,i}\}_{m\in \cG, i\in [n]}$, Assumption~\ref{assm:local} holds for various standard sampling strategies, including uniform and importance samplings \citep{gorbunov2023variance}.

For part of our results, we also need to assume smoothness of all $\{f_{m,i}\}_{m\in \cG, i\in [n]}$ explicitly.

\begin{assumption}[Smoothness of $f_{m,i}$ (optional)]
\label{assm:local_smooth_all}
We assume that for all $m\in \cG$ and $i\in [n]$ there exists $L_{m,i} \geq 0$ such that $f_{m,i}$ is $L_{m,i}$-smooth, i.e., for all $x, y \in \mathbb{R}^d$
\begin{equation}
    \|\nabla f_{m,i}(x) - \nabla f_{m,i}(y)\| \leq L_{m,i} \|x - y\|. \label{eq:f_m_j_smooth}
\end{equation}
\end{assumption}

\paragraph{Refined heterogeneity.} Instead of Assumption~\ref{assm:het_simplified}, we consider a more generalized one. 

\begin{assumption}[$(B, \zeta^2)$-heterogeneity] 
\label{assm:het}
We assume that our good clients have $\left(B, \zeta^2\right)$ heterogeneous local loss functions for some $B \geq 0, \zeta \geq 0$, i.e.,
\begin{align*}
\frac{1}{G} \sum_{m \in \mathcal{G}}\left\|\nabla f_m(x)-\nabla f(x)\right\|^2 \leq B\|\nabla f(x)\|^2+\zeta^2 \quad \forall x \in \mathbb{R}^d
\end{align*}
\end{assumption}

When $B = 0$, the above assumption recovers Assumption~\ref{assm:het_simplified}. However, it also covers some situations when the model is over-parameterized \citep{vaswani2019fast} and can hold with smaller values of $\zeta^2$. This assumption is also used in \citep{karimireddy2020byzantine, gorbunov2023variance}.

\subsection{Technical lemmas}\label{appendix:technical_lemmas_general}

\begin{lemma} \label{lemma:clipping} Let  $X \text { be a random vector in } \mathbb{R}^d \text { and } \widetilde{X}=\clip_{\lambda}(X).$ Assume that $\mathbb{E}[X]=x \in \mathbb{R}^d$ and $\|x\| \leq \lambda / 2,$ then 

\begin{equation*}
    \mathbb{E}\left[\|\widetilde{X}-x\|^2\right] \leq 10\mathbb{E}\left\| X - x\right\|^2.
\end{equation*}
\end{lemma}
\begin{proof}
The proof follows a similar procedure to that presented in Lemma F.5 from \citep{gorbunov2020stochastic}. To commence the proof, we introduce two indicator random variables:
\begin{align*}
\chi=\mathbb{I}_{\{X:\|X\|>\lambda\}}=\left\{\begin{array}{ll}
1, & \text { if }\|X\|>\lambda, \\
0, & \text { otherwise }
\end{array}\right.
\end{align*}
and 
\begin{align*}
\eta=\mathbb{I}_{\left\{X:\|X-x\|>\frac{\lambda}{2}\right\}}= \begin{cases}1, & \text { if }\|X-x\|>\frac{\lambda}{2} \\
0, & \text { otherwise }\end{cases}.
\end{align*}
Moreover, since $\|X\| \leq\|x\|+\|X-x\| \stackrel{\|x\| \leq \lambda / 2}{\leq} \frac{\lambda}{2}+\|X-x\|$, we have $\chi \leq \eta$. Using that we get
$$
\widetilde{X}=\min \left\{1, \frac{\lambda}{\|X\|}\right\} X=\chi \frac{\lambda}{\|X\|} X+(1-\chi) X.
$$
By Markov’s inequality,
\begin{align}
\label{eq:exp_eta}
\mathbb{E}[\eta]&=\mathbb{P}\left\{\|X-x\|>\frac{\lambda}{2}\right\}=\mathbb{P}\left\{\|X-x\|^2>\frac{\lambda^2}{4}\right\} \leq \frac{4}{\lambda^2} \mathbb{E}\left[\|X-x\|^2\right].
\end{align}
$\text { Using }\|\widetilde{X}-x\| \leq\|\widetilde{X}\|+\|x\| \leq \lambda+\frac{\lambda}{2}=\frac{3 \lambda}{2} \text {, we obtain }$
\begin{align*}
\mathbb{E}\left[\|\widetilde{X}-x\|^2\right] &= \mathbb{E}\left[\|\widetilde{X}-x\|^2 \chi+\|\widetilde{X}-x\|^2(1-\chi)\right] \\
& =\mathbb{E}\left[\chi\left\|\frac{\lambda}{\|X\|} X-x\right\|^2+\|X-x\|^2(1-\chi)\right]\\
&\leq\mathbb{E}\left[\chi\left(\left\|\frac{\lambda}{\|X\|} X\right\|+\|x\|\right)^2+\|X-x\|^2(1-\chi)\right]\\
&\stackrel{\|x\| \leq \frac{\lambda}{2}}{\leq}\left(\mathbb{E}\left[\chi\left(\frac{3 \lambda}{2}\right)^2+\|X-x\|^2\right]\right),
\end{align*}
where in the last inequality we applied $1 - \chi \leq 1.$ Using (\ref{eq:exp_eta}) and $\chi \leq \eta$ we get 
\begin{align*}
\mathbb{E}\left[\|\widetilde{X}-x\|^2\right] & \leq \frac{9 \lambda^2}{4}\left(\frac{2 }{\lambda}\right)^2\mathbb{E}\left[\|X-x\|^2\right]+\mathbb{E}\left[\|X-x\|^2\right] \\
& \leq 10\mathbb{E}\left[\|X-x\|^2\right].
\end{align*}
\end{proof}

\begin{lemma}[Lemma 2 \citep{li2021page}]
\label{lemma:page}
Assume that function $f$ is L-smooth (Assumption~\ref{assm:L-smoothness}) and $x^{t+1}=$ $x^t-\gamma g^t$. Then
\begin{align*}
f\left(x^{t+1}\right) &\leq f\left(x^t\right)-\frac{\gamma}{2}\left\|\nabla f\left(x^t\right)\right\|^2-\left(\frac{1}{2 \gamma}-\frac{L}{2}\right)\left\|x^{t+1}-x^t\right\|^2\\
&+\frac{\gamma}{2}\left\|g^t-\nabla f\left(x^t\right)\right\|^2 .
\end{align*}
\end{lemma}

\begin{lemma}
\label{lemma:premainA1}
Let Assumptions~\ref{assm:L-smoothness}, \ref{assm:global}, \ref{assm:local} hold and the Compression Operator satisfy Definition~\ref{def:Q}. Let us define "ideal" estimator:
\begin{equation*}
    \overline{g}^{t+1}= \begin{cases}
        \frac{1}{G^t_C}\sum \limits_{m \in \mathcal{G}^t_C} \nabla f_m(x^{t+1}),& c_n=1,\quad\quad\quad\quad\quad\quad~\hspace{0.45cm} [1]\\
        g^t+\nabla f\left(x^{t+1}\right)-\nabla f\left(x^t\right),& c_n=0, \text{ } G^t_C < (1-\delta)C, \text{ } [2]\\
        g^t+\frac{1}{G^t_C} \sum \limits_{m \in \mathcal{G}^t_C}\clip_{\lambda}\left(\mathcal{Q}\left(\widehat{\Delta}_m\left(x^{t+1}, x^t\right)\right)\right),& c_n=0, \text{ } G^t_C \geq (1-\delta)C.\text{ }[3]
    \end{cases}
\end{equation*}
Then for all $t\geq 0$ the iterates produced by \gls{Byz-VR-MARINA-PP} (Algorithm~\ref{alg:byz_vr_marina}) satisfy
\begin{align*}
 A_1 &=    \mathbb{E}\left[\left\|\overline{g}^{t+1}-\nabla f\left(x^{t+1}\right)\right\|^2\right]\\& \leq (1-p) \left(1+ \frac{p}{4}\right)\mathbb{E}\left[\left\|g^{t}-\nabla f(x^{t})\right\|^2\right]+p\frac{ \delta\cdot\mathcal{P }_{\mathcal{G}^t_{\widehat{C}}} }{(1-\delta)}  \mathbb{E}\left[ B\|\nabla f(x)\|^2+\zeta^2 \right]\\
    &+ \text{\scriptsize $(1-p)p_G\left(1+ \frac{4}{p}\right) \frac{2\cdot\mathcal{P}_{\mathcal{G}^t_C}M}{C} \left( 10 \omega L^2+(10\omega+1) L_{ \pm}^2+\frac{10(\omega+1) \mathcal{L}_{ \pm}^2}{b} \right)\mathbb{E}\left[\|x^{t+1} - x^t\|^2\right]$},
\end{align*}
where $p_G = \operatorname{Prob}\left\lbrace G^t_C \geq (1-\delta)C \right\rbrace $ and $\mathcal{P}_{\mathcal{G}^t_C} =  \operatorname{Prob}\left\lbrace m \in \mathcal{G}^t_C \mid G^t_C \geq\left(1-\delta\right) C\right\rbrace$.
\end{lemma}

\begin{proof}

Let us examine the expected value of the squared difference between ideal estimator and full gradient:
\begin{align*}
    A_1 &= \mathbb{E}\left[\left\|\overline{g}^{t+1}-\nabla f\left(x^{t+1}\right)\right\|^2\right]\\
&=\mathbb{E}\left[\mathbb{E}_t\left[\left\|\overline{g}^{t+1}-\nabla f\left(x^{t+1}\right)\right\|^2\right]\right]\\
    & = \left(1-p\right)p_{G}\mathbb{E}\left[\mathbb{E}_t\left[\left\|g^t+\frac{1}{G^t_C} \sum \limits_{m \in \mathcal{G}^t_C}\clip_{\lambda}\left(\mathcal{Q}\left(\widehat{\Delta}_m\left(x^{t+1}, x^t\right)\right)\right) -\right.\right.\right.\\
    &\left.\left.\left.-\nabla f\left(x^{t+1}\right)\right\|^2\right]\mid [3]\right]\\
    & + (1-p)(1-p_{G})\mathbb{E}\left[\mathbb{E}_t\left[\left\|g^t -\nabla f(x^{t})\right\|^2\right]\mid [2]\right]\\
    &+ p\mathbb{E}\left[ \left\|\frac{1}{G^t_{\widehat{C}}}\sum \limits_{m \in \cG_{\widehat{C}}^t} \nabla f_m(x^{t+1}) - \nabla f(x^{t+1})\right\|^2 \right].
\end{align*}
Using (\ref{eq:yung-1}) and $\nabla f\left(x^{t}\right) - \nabla f\left(x^{t}\right) = 0$ we obtain
\begin{align*}
B_1 &= \mathbb{E}\left[\mathbb{E}_t\left[\left\|g^t+\frac{1}{G^t_C} \sum \limits_{m \in \mathcal{G}^t_C}\clip_{\lambda}\left(\mathcal{Q}\left(\widehat{\Delta}_m\left(x^{t+1}, x^t\right)\right)\right) -\nabla f\left(x^{t+1}\right)\right\|^2\right]\mid [3]\right]\\
&=\mathbb{E}\left[\mathbb{E}_t\left[\left\|g^t+\frac{1}{G^t_C} \sum \limits_{m \in \mathcal{G}^t_C}\clip_{\lambda}\left(\mathcal{Q}\left(\widehat{\Delta}_m\left(x^{t+1}, x^t\right)\right)\right) -\right.\right.\right.\\
&\left.\left.\left.-\nabla f\left(x^{t+1}\right)+\nabla f\left(x^{t}\right) - \nabla f\left(x^{t}\right)\right\|^2\right]\mid [3]\right]\\
     &\stackrel{(\ref{eq:yung-1})}{\leq} \left(1+ \frac{p}{4}\right)\mathbb{E}\left[\left\|g^{t} -\nabla f\left(x^{t}\right)\right\|^2\right]\\
    &+  \left(1+ \frac{4}{p}\right) \mathbb{E}\left[\mathbb{E}_t\left[\left\|\frac{1}{G^t_C} \sum \limits_{m \in \mathcal{G}^t_C}\clip_{\lambda}\left(\mathcal{Q}\left(\widehat{\Delta}_m\left(x^{t+1}, x^t\right)\right)\right)  -\right.\right.\right.\\
&\left.\left.\left.- \left( \nabla f(x^{t+1}) - \nabla f(x^{t}) \right)\right\|^2\right]\mid [3]\right]\\
    &= \left(1+ \frac{p}{4}\right)\mathbb{E}\left[\left\|g^{t}-\nabla f(x^{t})\right\|^2\right]\\
    &+ \left(1+ \frac{4}{p}\right)\mathbb{E}\left[ \mathbb{E}_t\left[\left\|\frac{1}{G^t_C} \sum \limits_{m \in \mathcal{G}^t_C}\clip_{\lambda}\left(\mathcal{Q}\left(\widehat{\Delta}_m\left(x^{t+1}, x^t\right)\right)\right) -\right.\right.\right.\\
&\left.\left.\left. - \Delta\left(x^{t+1}, x^t\right)\right\|^2\right]\mid [3]\right].
\end{align*}
Let us consider the last part of the inequality:
\begin{align}
\notag  B^\prime_1 &= \mathbb{E}\left[ \mathbb{E}_t\left[\left\|\frac{1}{G^t_C} \sum \limits_{m \in \mathcal{G}^t_C}\clip_{\lambda}\left(\mathcal{Q}\left(\widehat{\Delta}_m\left(x^{t+1}, x^t\right)\right)\right) - \right.\right.\right.\\
\notag&\left.\left.\left. - \Delta\left(x^{t+1}, x^t\right)\right\|^2\right]\mid [3] \right]\\
 \notag &= \mathbb{E}\left[\mathbb{E}_{\set}\left[\mathbb{E}_t\left[\left\|\frac{1}{G^t_C} \sum \limits_{m \in \mathcal{G}^t_C}\clip_{\lambda}\left(\mathcal{Q}\left(\widehat{\Delta}_m\left(x^{t+1}, x^t\right)\right)\right) -\right.\right.\right.\right.\\
&\left.\left.\left.\left. - \Delta\left(x^{t+1}, x^t\right)\right\|^2\right]\mid [3]\right]\right].
 \end{align}
 Note that $G^t_C \geq (1-\delta)C$ in this case:
 \begin{align}
\notag B^\prime_1  &\leq  \frac{1}{C(1-\delta)}\mathbb{E}\left[\mathbb{E}_{\set}\left[ \sum \limits_{m \in \mathcal{G}^t_C}\mathbb{E}_t\left[\left\|\clip_{\lambda}\left(\mathcal{Q}\left(\widehat{\Delta}_m\left(x^{t+1}, x^t\right)\right)\right)  -\right.\right.\right.\right.\\
\notag&\left.\left.\left.\left. - \Delta\left(x^{t+1}, x^t\right)\right\|^2\right]\mid [3]\right]\right]\\
 \notag   &\leq  \frac{1}{C(1-\delta)}\mathbb{E}\left[\sum \limits_{m \in \mathcal{G}}\mathbb{E}_{\set}\left[\mathcal{I}_{\mathcal{G}^t_C}\right]\mathbb{E}_t\left[\left\|\clip_{\lambda}\left(\mathcal{Q}\left(\widehat{\Delta}_m\left(x^{t+1}, x^t\right)\right)\right)  -\right.\right.\right.\\\notag
\notag&\left.\left.\left. - \Delta\left(x^{t+1}, x^t\right)\right\|^2\right]\mid [3] \right]\\
       \notag &= \frac{1}{C(1-\delta)}\mathbb{E}\left[\sum \limits_{m \in \mathcal{G}}\mathcal{P}_{\mathcal{G}^t_C}\cdot\mathbb{E}_t\left[\left\|\clip_{\lambda}\left(\mathcal{Q}\left(\widehat{\Delta}_m\left(x^{t+1}, x^t\right)\right)\right)  - \right.\right.\right.\\
&\left.\left.\left. -\Delta\left(x^{t+1}, x^t\right)\right\|^2\right]\mid [3]\right],\label{eq:bvdjbjdfbvjdf}
\end{align}
where $\mathcal{I}_{\mathcal{G}^t_C}$ is an indicator function for the event $\left\lbrace m \in \mathcal{G}^t_C \mid G^t_C \geq\left(1-\delta\right) C\right\rbrace$ and $\mathcal{P}_{\mathcal{G}^t_C} =  \operatorname{Prob}\left\lbrace m \in \mathcal{G}^t_C \mid G^t_C \geq\left(1-\delta\right) C\right\rbrace$ is the probability of such an event. Note that 
$\mathbb{E}_{\set}\left[\mathcal{I}_{\mathcal{G}^t_C}\right] = \mathcal{P}_{\mathcal{G}^t_C}$. In case of uniform sampling of clients we have 
\begin{align*}
    \forall m \in \mathcal{G} \quad \mathcal{P}_{\mathcal{G}^t_C} &=\operatorname{Prob}\left\lbrace m \in \mathcal{G}^t_C \mid G^t_C \geq\left(1-\delta\right) C\right\rbrace\\
    & = \frac{C}{M p_G} \cdot \sum_{(1-\delta)C\leq m^\prime \leq C} \left(\left(\begin{array}{l}
G \\
m^\prime
\end{array}\right)\left(\begin{array}{l}
M-G \\
C-m^\prime
\end{array}\right) \left(\left(\begin{array}{l}
M \\
C
\end{array}\right)\right)^{-1} \right),\\
p_G &= \sum_{(1-\delta)C\leq m^\prime \leq C} \left(\left(\begin{array}{l}
G-1 \\
t-1
\end{array}\right)\left(\begin{array}{l}
M-G \\
C-m^\prime
\end{array}\right) \left(\left(\begin{array}{l}
M-1 \\
C-1
\end{array}\right)\right)^{-1} \right)
\end{align*}
Now we can continue with inequalities:
\begin{align*}
B_1^\prime &\leq      \frac{\mathcal{P}_{\mathcal{G}^t_C}}{C(1-\delta)} \mathbb{E}\left[ \sum \limits_{m \in \mathcal{G}}\mathbb{E}_t\left[\left\|\clip_{\lambda}\left(\mathcal{Q}\left(\widehat{\Delta}_m\left(x^{t+1}, x^t\right)\right)\right) -\right.\right.\right.\\
&\left.\left.\left. - \Delta\left(x^{t+1}, x^t\right)\right\|^2\right]\mid [3]\right]\\
 &\leq         \frac{\mathcal{P}_{\mathcal{G}^t_C}}{C(1-\delta)}\mathbb{E}\left[\sum \limits_{m \in \mathcal{G}}\mathbb{E}_t\left[\mathbb{E}_{Q}\left[\left\|\clip_{\lambda}\left(\mathcal{Q}\left(\widehat{\Delta}_m\left(x^{t+1}, x^t\right)\right)\right)  -\right.\right.\right.\right.\\
&\left.\left.\left.\left.- \Delta\left(x^{t+1}, x^t\right)\right\|^2\right] \right]\mid [3]  \right]\\
 &\stackrel{(\ref{eq:yung-1})}{\leq}      \frac{\mathcal{P}_{\mathcal{G}^t_C}}{C(1-\delta)}\mathbb{E}\left[\sum \limits_{m \in \mathcal{G}}2\mathbb{E}_t\left[\mathbb{E}_{Q}\left[\left\|\clip_{\lambda}\left(\mathcal{Q}\left(\widehat{\Delta}_m\left(x^{t+1}, x^t\right)\right)\right)  - \right.\right.\right.\right.\\
&\left.\left.\left.\left.-\Delta_m\left(x^{t+1}, x^t\right)\right\|^2\right]\right]\mid [3]\right]\\ 
& +   \frac{\mathcal{P}_{\mathcal{G}^t_C}}{C(1-\delta)}\mathbb{E}\left[\sum \limits_{m \in \mathcal{G}}2\mathbb{E}_t\left[\left\|  \Delta_m\left(x^{t+1}, x^t\right) - \Delta\left(x^{t+1}, x^t\right)\right\|^2\right]\mid [3]\right].
\end{align*}
Using Lemma \ref{lemma:clipping} we have 
\begin{align*}
B_1^\prime&\stackrel{\text{Lemma \ref{lemma:clipping}}}{\leq}    \frac{\mathcal{P}_{\mathcal{G}^t_C}}{C(1-\delta)}\mathbb{E}\left[\sum \limits_{m \in \mathcal{G}}20\mathbb{E}_t\left[\mathbb{E}_{Q}\left[\left\|\mathcal{Q}\left(\widehat{\Delta}_m\left(x^{t+1}, x^t\right)\right)   -\right.\right.\right.\right.\\
&\left.\left.\left.\left.- \Delta_m\left(x^{t+1}, x^t\right)\right\|^2\right]\right]\mid [3] \right]\\
&+\frac{\mathcal{P}_{\mathcal{G}^t_C}}{C(1-\delta)}\mathbb{E}\left[\sum \limits_{m \in \mathcal{G}}2\mathbb{E}_t\left[\left\|  \Delta_m\left(x^{t+1}, x^t\right) -  \Delta\left(x^{t+1}, x^t\right)\right\|^2\right]\mid [3]\right]\\
&\leq  \frac{20\cdot\mathcal{P}_{\mathcal{G}^t_C}}{C(1-\delta)} \mathbb{E}\left[ \sum \limits_{m \in \mathcal{G}}\mathbb{E}_t\left[\mathbb{E}_{Q}\left[\left\|\mathcal{Q}\left(\widehat{\Delta}_m\left(x^{t+1}, x^t\right)\right) -\right.\right.\right.\right.\\
&\left.\left.\left.\left.  - \Delta_m\left(x^{t+1}, x^t\right)\right\|^2\right]\right]\mid [3]\right]\\
&+\frac{2\cdot\mathcal{P}_{\mathcal{G}^t_C}}{C(1-\delta)} \mathbb{E}\left[ \sum \limits_{m \in \mathcal{G}}\mathbb{E}_t\left[\left\|  \Delta_m\left(x^{t+1}, x^t\right) -  \Delta\left(x^{t+1}, x^t\right)\right\|^2\right]\mid [3]\right]\\
&\leq \frac{20\cdot\mathcal{P}_{\mathcal{G}^t_C}}{C(1-\delta)} \mathbb{E}\left[ \sum \limits_{m \in \mathcal{G}}\mathbb{E}_t\left[\mathbb{E}_{Q}\left[\left\|\mathcal{Q}\left(\widehat{\Delta}_m\left(x^{t+1}, x^t\right)\right)\right\|^2\right]\right]- \right.\\
&\left.- \sum \limits_{m \in \mathcal{G}}\left\|\Delta_m\left(x^{t+1}, x^t\right)\right\|^2\mid [3]\right]\\
&+\frac{2\cdot\mathcal{P}_{\mathcal{G}^t_C}}{C(1-\delta)}\mathbb{E}\left[\sum \limits_{m \in \mathcal{G}}\mathbb{E}_t\left[\left\|  \Delta_m\left(x^{t+1}, x^t\right) -\Delta\left(x^{t+1}, x^t\right)\right\|^2\right]\mid [3]\right].
\end{align*}
Applying Definition~\ref{def:Q} of Unbiased Compressor we have 
\begin{align*}
B_1^\prime&\leq  \frac{20\cdot\mathcal{P}_{\mathcal{G}^t_C}}{C(1-\delta)} \mathbb{E}\left[ \sum \limits_{m \in \mathcal{G}}(1+\omega)\mathbb{E}_t\left\|\widehat{\Delta}_m\left(x^{t+1}, x^t\right)\right\|^2 - \sum \limits_{m \in \mathcal{G}}\left\|\Delta_m\left(x^{t+1}, x^t\right)\right\|^2\mid [3]\right]\\
&+\frac{2\cdot\mathcal{P}_{\mathcal{G}^t_C}}{C(1-\delta)}\mathbb{E}\left[\sum \limits_{m \in \mathcal{G}}\left\|  \Delta_m\left(x^{t+1}, x^t\right) -  \Delta\left(x^{t+1}, x^t\right)\right\|^2\mid [3]\right]\\
&\leq  \frac{20\cdot\mathcal{P}_{\mathcal{G}^t_C}}{C(1-\delta)} \mathbb{E} \left[ \sum \limits_{m \in \mathcal{G}}(1+\omega)\mathbb{E}_t\left\|\widehat{\Delta}_m\left(x^{t+1}, x^t\right) - \Delta_m\left(x^{t+1}, x^t\right) \right\|^2 \right]\\
&+ \frac{20\cdot\mathcal{P}_{\mathcal{G}^t_C}}{C(1-\delta)} \mathbb{E} \left[ \sum \limits_{m \in \mathcal{G}}(1+\omega)\mathbb{E}_t\left\|\Delta_m\left(x^{t+1}, x^t\right)\right\|^2- \sum \limits_{m \in \mathcal{G}}\mathbb{E}_t\left\|\Delta_m\left(x^{t+1}, x^t\right)\right\|^2\mid [3]\right]\\
&+\frac{2\cdot\mathcal{P}_{\mathcal{G}^t_C}}{C(1-\delta)}\mathbb{E}\left[\sum \limits_{m \in \mathcal{G}}\left\|  \Delta_m\left(x^{t+1}, x^t\right) -  \Delta\left(x^{t+1}, x^t\right)\right\|^2\mid [3]\right].
\end{align*}
Now we combine terms and have 
\begin{align*}
    B_1^\prime &\leq  \frac{20\cdot\mathcal{P}_{\mathcal{G}^t_C}}{C(1-\delta)} (1+\omega) \mathbb{E} \left[ \sum \limits_{m \in \mathcal{G}}\mathbb{E}_t\left[\left\|\widehat{\Delta}_m\left(x^{t+1}, x^t\right) - \Delta_m\left(x^{t+1}, x^t\right) \right\|^2\right] \mid [3] \right]\\
&+ \frac{20\cdot\mathcal{P}_{\mathcal{G}^t_C}}{C(1-\delta)}\omega \mathbb{E} \left[ \sum \limits_{m \in \mathcal{G}}\left\|\Delta_m\left(x^{t+1}, x^t\right)\right\|^2\mid [3]\right]\\
&+\frac{2\cdot\mathcal{P}_{\mathcal{G}^t_C}}{C(1-\delta)}\mathbb{E} \left[ \sum \limits_{m \in \mathcal{G}}\left\|  \Delta_m\left(x^{t+1}, x^t\right) -  \Delta\left(x^{t+1}, x^t\right)\right\|^2\mid [3]\right]\\
 &= \frac{20\cdot\mathcal{P}_{\mathcal{G}^t_C}}{C(1-\delta)} (1+\omega) \mathbb{E} \left[ \sum \limits_{m \in \mathcal{G}}\mathbb{E}_t\left[\left\|\widehat{\Delta}_m\left(x^{t+1}, x^t\right) - \Delta_m\left(x^{t+1}, x^t\right) \right\|^2\right] \mid [3] \right] \\
&+ \frac{20\cdot\mathcal{P}_{\mathcal{G}^t_C}}{C(1-\delta)}\omega \mathbb{E} \left[  \sum \limits_{m \in \mathcal{G}}\left\|\Delta_m\left(x^{t+1}, x^t\right) - \Delta\left(x^{t+1}, x^t\right)\right\|^2 + \|\Delta\left(x^{t+1}, x^t\right)\|^2\mid [3]\right]\\
&+\frac{2\cdot\mathcal{P}_{\mathcal{G}^t_C}}{C(1-\delta)} \mathbb{E} \left[ \sum \limits_{m \in \mathcal{G}}\left\|  \Delta_m\left(x^{t+1}, x^t\right) -  \Delta\left(x^{t+1}, x^t\right)\right\|^2 \mid [3] \right].
\end{align*}
Rearranging terms leads to 
\begin{align*}
   B_1^\prime  &\leq \frac{20\cdot\mathcal{P}_{\mathcal{G}^t_C}}{C(1-\delta)} (1+\omega) \mathbb{E} \left[  \sum \limits_{m \in \mathcal{G}}\mathbb{E}_t\left[\left\|\widehat{\Delta}_m\left(x^{t+1}, x^t\right) - \Delta_m\left(x^{t+1}, x^t\right) \right\|^2\right] \mid [3] \right] \\
&+ \frac{2\cdot\mathcal{P}_{\mathcal{G}^t_C}}{C(1-\delta)}(10\omega+1) \mathbb{E}\left[ \sum \limits_{m \in \mathcal{G}}\left\|\Delta_m\left(x^{t+1}, x^t\right) - \Delta\left(x^{t+1}, x^t\right)\right\|^2\mid[3]\right] \\
&+\frac{20\cdot\mathcal{P}_{\mathcal{G}^t_C}}{C(1-\delta)} \omega \mathbb{E}\left[\sum \limits_{m \in \mathcal{G}}\left\|  \Delta\left(x^{t+1}, x^t\right)\right\|^2 \mid [3]\right].
\end{align*}
Now we apply Assumptions \ref{assm:L-smoothness}, \ref{assm:global}, \ref{assm:local}:
\begin{align*}
   B_1^\prime  &\leq \frac{20\cdot\mathcal{P}_{\mathcal{G}^t_C}}{C(1-\delta)} (1+\omega) \mathbb{E} \left[ G \frac{\mathcal{L}_{ \pm}^2}{b} \|x^{t+1} - x^t\|^2\right]\\
&+ \frac{2\cdot\mathcal{P}_{\mathcal{G}^t_C}}{C(1-\delta)}(10\omega+1) \mathbb{E} \left[ G L_{ \pm}^2 \|x^{t+1} - x^t\|^2\right]\\
&+\frac{20\cdot\mathcal{P}_{\mathcal{G}^t_C}}{C(1-\delta)} \omega \mathbb{E} \left[ G L^2\left\|  x^{t+1} - x^t\right\|^2\right].
\end{align*}
Finally, we have
\begin{align*}
B_1^\prime&\leq \frac{2\cdot\mathcal{P}_{\mathcal{G}^t_C}\cdot G}{C(1-\delta)} \left( 10 \omega L^2+(10\omega+1) L_{ \pm}^2+\frac{10(\omega+1) \mathcal{L}_{ \pm}^2}{b} \right)\mathbb{E}\left[ \|x^{t+1} - x^t\|^2\right].
\end{align*}
Let us plug obtained results:
\begin{align*}
  B_1      &\leq \left(1+ \frac{p}{4}\right) \mathbb{E}\left[\left\|g^{t}-\nabla f(x^{t})\right\|^2\right]\\
    &+ \left(1+ \frac{4}{p}\right) \frac{2\cdot\mathcal{P}_{\mathcal{G}^t_C}\cdot G}{C(1-\delta)} \left( 10 \omega L^2+(10\omega+1) L_{ \pm}^2+\right.\\
    &\left.+\frac{10(\omega+1) \mathcal{L}_{ \pm}^2}{b} \right)\mathbb{E}\left[\|x^{t+1} - x^t\|^2\right].
\end{align*}

Let us consider the term $\mathbb{E}\left[ \left\|\frac{1}{G^t_{\widehat{C}}}\sum \limits_{m \in \cG_{\widehat{C}}^t} \nabla f_m(x^{t+1}) - \nabla f(x^{t+1})\right\|^2 \right]$:

\begin{align*}
    &\mathbb{E}\left[ \left\|\frac{1}{G^t_{\widehat{C}}}\sum \limits_{m \in \cG_{\widehat{C}}^t} \nabla f_m(x^{t+1}) - \nabla f(x^{t+1})\right\|^2 \right]\\
    &\leq     \mathbb{E}\left[ \frac{1}{G^t_{\widehat{C}}}\sum \limits_{m \in \cG_{\widehat{C}}^t} \left\|\nabla f_m(x^{t+1}) - \nabla f(x^{t+1})\right\|^2 \right]\\
    &\leq \frac{1}{(1-\delta)\widehat{C}}  \mathbb{E}\left[ \sum \limits_{m \in \cG_{\widehat{C}}^t} \left\|\nabla f_m(x^{t+1}) - \nabla f(x^{t+1})\right\|^2 \right]\\
    & = \frac{1}{(1-\delta)\widehat{C}}  \mathbb{E}\left[ \sum \limits_{m \in \mathcal{G}} \mathcal{I}_{\mathcal{G}^t_{\widehat{C}}} \left\|\nabla f_m(x^{t+1}) - \nabla f(x^{t+1})\right\|^2 \right]
\end{align*}
Using definition of $\mathcal{P}_{\mathcal{G}^t_C}$ we get
\begin{align*}
    &\mathbb{E}\left[ \left\|\frac{1}{G^t_{\widehat{C}}}\sum \limits_{m \in \cG_{\widehat{C}}^t} \nabla f_m(x^{t+1}) - \nabla f(x^{t+1})\right\|^2 \right]\\
    &\leq  \frac{\mathcal{P}_{\mathcal{G}^t_{\widehat{C}}} }{(1-\delta)\widehat{C}}  \mathbb{E}\left[ \sum \limits_{m \in \mathcal{G}} \left\|\nabla f_m(x^{t+1}) - \nabla f(x^{t+1})\right\|^2 \right]\\
    &\leq \frac{ G\cdot\mathcal{P }_{\mathcal{G}^t_{\widehat{C}}} }{(1-\delta)\widehat{C}G}  \mathbb{E}\left[ \sum \limits_{m \in \mathcal{G}} \left\|\nabla f_m(x^{t+1}) - \nabla f(x^{t+1})\right\|^2 \right]\\ 
\end{align*}
Using Assumption \ref{assm:het} we get
\begin{align}
    &\mathbb{E}\left[ \left\|\frac{1}{G^t_{\widehat{C}}}\sum \limits_{m \in \cG_{\widehat{C}}^t} \nabla f_m(x^{t+1}) - \nabla f(x^{t+1})\right\|^2 \right]\notag\\
    &\leq  \frac{ G\cdot\mathcal{P }_{\mathcal{G}^t_{\widehat{C}}} }{(1-\delta)\widehat{C}}  \mathbb{E}\left[ B\|\nabla f(x)\|^2+\zeta^2 \right] \notag \\
    & \leq  \frac{ \deltar n \cdot\mathcal{P }_{\mathcal{G}^t_{\widehat{C}}} }{(1-\delta)\frac{\deltar n }{\delta}}  \mathbb{E}\left[ B\|\nabla f(x)\|^2+\zeta^2 \right] \notag\\
    &= \frac{ \delta\cdot\mathcal{P }_{\mathcal{G}^t_{\widehat{C}}} }{(1-\delta)}  \mathbb{E}\left[ B\|\nabla f(x)\|^2+\zeta^2 \right] \label{eq:nsjknbvsbicusd}
    \end{align}
Also, we have 
\begin{align*}
    A_1 &= \mathbb{E}\left[\left\|\overline{g}^{t+1}-\nabla f(x^{t+1})\right\|^2\right]\\
    &\leq (1-p)p_G B_1 + (1-p)(1-p_{G})\mathbb{E}\left[\left\|g^t -\nabla f(x^{t})\right\|^2\right]\\
    &+p\frac{ \delta\cdot\mathcal{P }_{\mathcal{G}^t_{\widehat{C}}} }{(1-\delta)}  \mathbb{E}\left[ B\|\nabla f(x)\|^2+\zeta^2 \right]\\
    &\leq (1-p)p_G \left(1+ \frac{p}{4}\right)\mathbb{E}\left[\left\|g^{t}-\nabla f(x^{t})\right\|^2\right]\\
    &+ (1-p)p_G\left(1+ \frac{4}{p}\right) \frac{2\cdot\mathcal{P}_{\mathcal{G}^t_C}\cdot G}{C(1-\delta)} \left( 10 \omega L^2+(10\omega+1) L_{ \pm}^2+\right.\\
    &\left.+\frac{10(\omega+1) \mathcal{L}_{ \pm}^2}{b} \right)\mathbb{E}\left[\|x^{t+1} - x^t\|^2\right]\\
    &+(1-p)(1-p_{G})\mathbb{E}\left[\left\|g^t -\nabla f(x^{t})\right\|^2\right]\\
    &+p\frac{ \delta\cdot\mathcal{P }_{\mathcal{G}^t_{\widehat{C}}} }{(1-\delta)}  \mathbb{E}\left[ B\|\nabla f(x)\|^2+\zeta^2 \right].
\end{align*}
To simplify the bound we use $\left(1+\frac{p}{4}>1\right)$ and obtain 
\begin{align*} 
    A_1 
    &\leq (1-p)p_G \left(1+ \frac{p}{4}\right)\mathbb{E}\left[\left\|g^{t}-\nabla f(x^{t})\right\|^2\right]\\
    &+p\frac{ \delta\cdot\mathcal{P }_{\mathcal{G}^t_{\widehat{C}}} }{(1-\delta)}  \mathbb{E}\left[ B\|\nabla f(x)\|^2+\zeta^2 \right]\\
    &+  (1-p)p_G\left(1+ \frac{4}{p}\right) \frac{2\cdot\mathcal{P}_{\mathcal{G}^t_C}\cdot G}{C(1-\delta)} \left( 10 \omega L^2+(10\omega+1) L_{ \pm}^2+\right.\\
    &+\left.\frac{10(\omega+1) \mathcal{L}_{ \pm}^2}{b} \right)\mathbb{E}\left[\|x^{t+1} - x^t\|^2\right]\\
    &+(1-p)(1-p_{G})\mathbb{E}\left[\left\|g^t -\nabla f(x^{t})\right\|^2\right]\\
    &\leq (1-p)p_G \left(1+ \frac{p}{4}\right)\mathbb{E}\left[\left\|g^{t}-\nabla f(x^{t})\right\|^2\right]\\
    &+ (1-p)p_G\left(1+ \frac{4}{p}\right) \frac{2\cdot\mathcal{P}_{\mathcal{G}^t_C}\cdot G}{C(1-\delta)} \left( 10 \omega L^2+(10\omega+1) L_{ \pm}^2+\right.\\
    &+\left.\frac{10(\omega+1) \mathcal{L}_{ \pm}^2}{b} \right)\mathbb{E}\left[\|x^{t+1} - x^t\|^2\right]\\
    &+(1-p)(1-p_{G})\left(1+\frac{p}{4}\right)\mathbb{E}\left[\left\|g^t -\nabla f(x^{t})\right\|^2\right]\\
    &+p\frac{ \delta\cdot\mathcal{P }_{\mathcal{G}^t_{\widehat{C}}} }{(1-\delta)}  \mathbb{E}\left[ B\|\nabla f(x)\|^2+\zeta^2 \right]\\
      &\leq (1-p) \left(1+ \frac{p}{4}\right)\mathbb{E}\left[\left\|g^{t}-\nabla f(x^{t})\right\|^2\right]\\
      &+p\frac{ \delta\cdot\mathcal{P }_{\mathcal{G}^t_{\widehat{C}}} }{(1-\delta)}  \mathbb{E}\left[ B\|\nabla f(x)\|^2+\zeta^2 \right]\\
    &+ (1-p)p_G\left(1+ \frac{4}{p}\right) \frac{2\cdot\mathcal{P}_{\mathcal{G}^t_C}M}{C} \left( 10 \omega L^2+(10\omega+1) L_{ \pm}^2+\right.\\
    &+\left.\frac{10(\omega+1) \mathcal{L}_{ \pm}^2}{b} \right)\mathbb{E}\left[\|x^{t+1} - x^t\|^2\right].
    \end{align*}
\end{proof}

\begin{lemma}\label{lemma:premainA2}
 Let us define "ideal" estimator:
 \begin{equation*}
    \overline{g}^{t+1}= \begin{cases}
        \frac{1}{G^t_C}\sum \limits_{m \in \mathcal{G}^t_C} \nabla f_m(x^{t+1}),& c_n=1,\quad\quad\quad\quad\quad\quad\quad~\hspace{0.06cm} [1]\\
        g^t+\nabla f\left(x^{t+1}\right)-\nabla f\left(x^t\right),& c_n=0, \text{ } G^t_C < (1-\delta)C, \text{ }[2]\\
        g^t+\frac{1}{G^t_C} \sum \limits_{m \in \mathcal{G}^t_C}\clip_{\lambda}\left(\mathcal{Q}\left(\widehat{\Delta}_m\left(x^{t+1}, x^t\right)\right)\right),& c_n=0, \text{ } G^t_C \geq (1-\delta)C. \text{ }[3]
    \end{cases}
\end{equation*}
Also let us introduce the notation
\begin{align*}
\texttt{ARAgg}_Q^{t+1} = \texttt{ARAgg}\left(\clip_{\lambda_{t+1}}\left(\cQ\left(\widehat{\Delta}_1(x^{t+1}, x^t)\right)\right),\right.\\
\left.\ldots, \clip_{\lambda_{t+1}}\left(\cQ\left(\widehat{\Delta}_C(x^{t+1}, x^t)\right)\right)\right).
\end{align*}
Then for all $t\geq 0$ the iterates produced by \gls{Byz-VR-MARINA-PP} (Algorithm~\ref{alg:byz_vr_marina}) satisfy

    \begin{align*}
        A_2&= \mathbb{E}\left[\left\|g^{t+1}-\overline{g}^{t+1}\right\|^2\right]\\
        &\leq p\mathbb{E}\left[\mathbb{E}_t\left[\left\|\texttt{ARAgg}\left(\{g_m^{t+1}\}_{m\in \set}\right) - \nabla f(x^{t+1})\right\|^2\right]\mid [1]\right]\\
   & + (1-p)p_G \mathbb{E}\left[ \mathbb{E}_t\left[\left\| \frac{1}{G^t_C} \sum \limits_{m \in \mathcal{G}^t_C}\clip_{\lambda}\left(\mathcal{Q}\left(\widehat{\Delta}_m\left(x^{t+1}, x^t\right)\right)\right)  -\right.\right.\right.\\
   &-\left.\left.\left.\texttt{ARAgg}_Q^{t+1}\right\|^2\mid [3]\right]\right]\\
   &+ (1-p)(1-p_G)\mathbb{E}\left[\mathbb{E}_t\left[\left\| \nabla f(x^{t+1}) - \nabla f(x^{t}) - \texttt{ARAgg}_Q^{t+1}\right\|^2\mid [2]\right]\right],
    \end{align*}
    where $p_G = \operatorname{Prob}\left\lbrace G^t_C \geq (1-\delta)C \right\rbrace $. 
\end{lemma}

\begin{proof}
    Using conditional expectations we have 
\begin{align*}
   A_2 &=  \mathbb{E}\left[\mathbb{E}_t\left[\left\|g^{t+1}-\overline{g}^{t+1}\right\|^2\right]\right]\\
   & = p\mathbb{E}\left[\mathbb{E}_t\left[\left\|\texttt{ARAgg}\left(\{g_m^{t+1}\}_{m\in \set}\right) - \nabla f(x^{t+1})\right\|^2\right]\mid [1]\right]\\
   & + (1-p)p_G \mathbb{E}\left[ \mathbb{E}_t\left[\left\| g^t+\frac{1}{G^t_C} \sum \limits_{m \in \mathcal{G}^t_C}\clip_{\lambda}\left(\mathcal{Q}\left(\widehat{\Delta}_m\left(x^{t+1}, x^t\right)\right)\right) -\right.\right.\right.\\
   &\left.\left.\left.-\left(g^t + \texttt{ARAgg}_Q^{t+1}\right)\right\|^2\right]\mid [3]\right]\\
   &+ (1-p)(1-p_G)\mathbb{E}\left[\mathbb{E}_t\left[\left\| g^t+\nabla f(x^{t+1}) - \nabla f(x^{t})\right.\right.\right.\\
   &-\left.\left.\left. \left(g^t + \texttt{ARAgg}_Q^{t+1}\right) \right\|^2\right]\mid [2]\right].
   \end{align*}
After simplification, we get the following bound:
   \begin{align*}
A_2   &\leq  p\mathbb{E}\left[\mathbb{E}_t\left[\left\|\texttt{ARAgg}\left(\{g_m^{t+1}\}_{m\in \set}\right) - \nabla f(x^{t+1})\right\|^2\right]\mid [1]\right]\\
   & + (1-p)p_G \mathbb{E}\left[ \mathbb{E}_t\left[\left\| \frac{1}{G^t_C} \sum \limits_{m \in \mathcal{G}^t_C}\clip_{\lambda}\left(\mathcal{Q}\left(\widehat{\Delta}_m\left(x^{t+1}, x^t\right)\right)\right) - \right.\right.\right.\\
   &-\left.\left.\left. \texttt{ARAgg}_Q^{t+1}\right\|^2\mid [3]\right]\right]\\
   &+ (1-p)(1-p_G)\mathbb{E}\left[\mathbb{E}_t\left[\left\| \nabla f(x^{t+1}) - \nabla f(x^{t}) - \texttt{ARAgg}_Q^{t+1}\right\|^2\mid [2]\right]\right].
\end{align*} 
\end{proof}

\begin{lemma} 
\label{lemma:full_aggr}
Let Assumptions~\ref{assm:L-smoothness} and \ref{assm:het} hold and Aggregation Operator ($\texttt{ARAgg}$) satisfy Definition~\ref{def:aragg}. Then for all $t\geq 0$ the iterates produced by \gls{Byz-VR-MARINA-PP} (Algorithm~\ref{alg:byz_vr_marina}) satisfy
\begin{align*}
T_1 &=  \mathbb{E}\left[\mathbb{E}_t\left[\left\|\texttt{ARAgg}\left(\{g_m^{t+1}\}_{m\in \set}\right) - \nabla f(x^{t+1})\right\|^2\right]\mid [1] \right]\\ 
&\leq \left(\frac{8 G \mathcal{P}_{\mathcal{G}^t_{\widehat{C}}} c\delta B}{(1-\delta) \widehat{C}} + 2\widetilde{B}\right)  \mathbb{E}\left[\left\|\nabla f\left(x^t\right)\right\|^2  + L^2\left\|x^{t+1}-x^t\right\|^2\right]\\
&+ \frac{4 G \mathcal{P}_{\mathcal{G}^t_{\widehat{C}}} c\delta \zeta^2}{(1-\delta) \widehat{C}} + \widetilde{\zeta}^2,
\end{align*}
where $\widetilde{B} \eqdef 0$ and $\widetilde{\zeta}^2 \eqdef 0$ when $\widehat{C} = M$, and $\widetilde{B} \eqdef \frac{\cP_{\cG_{\widehat{C}}^t}GB}{(1-\delta)\widehat{C}}$ and $\widetilde{\zeta}^2 \eqdef \frac{\cP_{\cG_{\widehat{C}}^t}G\zeta^2}{(1-\delta)\widehat{C}}$ when $\widehat{C} < M$.
\end{lemma}

\begin{proof}
Using the definition of aggregation operator, we have
    \begin{align*}
T_1 &= \mathbb{E}\left[\mathbb{E}_t\left[\left\|\texttt{ARAgg}\left(\{g_m^{t+1}\}_{m\in \set}\right) - \nabla f(x^{t+1})\right\|^2\right]\mid [1]\right]\\
&\overset{\eqref{eq:yung-1}}{\leq} \mathbb{E}\left[\mathbb{E}_t\left[\left\|\texttt{ARAgg}\left(\{g_m^{t+1}\}_{m\in \set}\right) - \frac{1}{G_{\widehat{C}}^t}\sum\limits_{m \in \cG_{\widehat{C}}^t}\nabla f_m(x^{t+1})\right\|^2\right]\mid [1]\right]\\
&\quad + \mathbb{E}\left[\mathbb{E}_t\left[\left\|\frac{1}{G_{\widehat{C}}^t}\sum\limits_{m \in \cG_{\widehat{C}}^t}\nabla f_m(x^{t+1}) - \nabla f(x^{t+1})\right\|^2\right]\mid [1]\right].
\end{align*}
Since $\frac{1}{G_{\widehat{C}}^t}\sum\limits_{m \in \cG_{\widehat{C}}^t}\nabla f_m(x^{t+1}) = \nabla f(x^{t+1})$ with probability $1$ when $\widehat{C} = M$, we can estimate the last term as
\begin{align*}
    \mathbb{E}&\left[\mathbb{E}_t\left[\left\|\frac{1}{G_{\widehat{C}}^t}\sum\limits_{m \in \cG_{\widehat{C}}^t}\nabla f_m(x^{t+1}) - \nabla f(x^{t+1})\right\|^2\right] \mid\, [1]\right] \\
    &\leq \begin{cases}
        0,& \text{if } \widehat{C} = M\\ \mathbb{E}\left[\frac{1}{G_{\widehat{C}}^t}\sum\limits_{m \in \cG_{\widehat{C}}^t}\mathbb{E}_t\left[\left\|\nabla f_m(x^{t+1}) - \nabla f(x^{t+1})\right\|^2\right]\mid [1]\right], & \text{if } \widehat{C} < M
    \end{cases}\\
    &\leq \begin{cases}
        0,& \text{if } \widehat{C} = M\\ \frac{\cP_{\cG_{\widehat{C}}^t}}{(1-\delta)\widehat{C}}\sum\limits_{m \in \cG}\mathbb{E}\left[\left\|\nabla f_m(x^{t+1}) - \nabla f(x^{t+1})\right\|^2\right], & \text{if } \widehat{C} < M
    \end{cases}\\
    &\stackrel{(\text{As.~\ref{assm:het}})}{\leq} \begin{cases}
        0,& \text{if } \widehat{C} = M\\ \frac{\cP_{\cG_{\widehat{C}}^t}G}{(1-\delta)\widehat{C}}\left(B\mathbb{E}\left[\|\nabla f(x^{t+1})\|^2\right] + \zeta^2\right), & \text{if } \widehat{C} < M
    \end{cases}\\
    &= \widetilde{B}\mathbb{E}\left[\|\nabla f(x^{t+1})\|^2\right] + \widetilde{\zeta}^2,
\end{align*}
where
\begin{equation*}
    \widetilde{B} \eqdef \begin{cases}
        0,& \text{if } \widehat{C} = M,\\ \frac{\cP_{\cG_{\widehat{C}}^t}GB}{(1-\delta)\widehat{C}},& \text{if } \widehat{C} < M,
    \end{cases} \quad \text{and}\quad  \widetilde{\zeta}^2 \eqdef \begin{cases}
        0,& \text{if } \widehat{C} = M,\\ \frac{\cP_{\cG_{\widehat{C}}^t}G\zeta^2}{(1-\delta)\widehat{C}},& \text{if } \widehat{C} < M.
    \end{cases}
\end{equation*}
Using the above bound, we continue the estimation of $T_1$ as follows:
\begin{align*}
T_1 & \stackrel{(\text{Def.~\ref{def:aragg}})}{\leq}\mathbb{E}\left[\frac{c\delta}{G_{\widehat{C}}^t(G_{\widehat{C}}^t-1)} \sum_{\substack{i, l \in \mathcal{G}^t_{\widehat{C}} \\
m \neq l}} \mathbb{E}_t\left[\left\|\nabla f_m\left(x^{t+1}\right)-\nabla f_l\left(x^{t+1}\right)\right\|^2\mid [1]\right]\right]\\
&\quad + \widetilde{B}\mathbb{E}\left[\|\nabla f(x^{t+1})\|^2\right] + \widetilde{\zeta}^2\\
& \stackrel{(\ref{eq:yung-1})}{\leq} \mathbb{E}\left[\frac{c\delta}{G_{\widehat{C}}^t(G_{\widehat{C}}^t-1)}\sum_{\substack{i, l \in \mathcal{G}^t_{\widehat{C}} \\
m \neq l}} \mathbb{E}\left[2\left\|\nabla f_m\left(x^{t+1}\right)-\nabla f\left(x^{t+1}\right)\right\|^2\mid [1]\right]\right]\\
&\quad + \mathbb{E}\left[\frac{c\delta}{G_{\widehat{C}}^t(G_{\widehat{C}}^t-1)}\sum_{\substack{i, l \in \mathcal{G}^t_{\widehat{C}} \\
m \neq l}} \mathbb{E}\left[2\left\|\nabla f_l\left(x^{t+1}\right)-\nabla f\left(x^{t+1}\right)\right\|^2\mid [1]\right]\right]\\
&\quad + \widetilde{B}\mathbb{E}\left[\|\nabla f(x^{t+1})\|^2\right] + \widetilde{\zeta}^2\\
&=\mathbb{E}\left[\frac{c\delta}{G^t_{\widehat{C}}} \sum_{m \in \mathcal{G}^t_{\widehat{C}}} 4\mathbb{E}_t\left[\left\|\nabla f_m\left(x^{t+1}\right)-\nabla f\left(x^{t+1}\right)\right\|^2\mid [1]\right]\right]\\
&+ \widetilde{B}\mathbb{E}\left[\|\nabla f(x^{t+1})\|^2\right] + \widetilde{\zeta}^2\\
&\leq \frac{\mathcal{P}_{\mathcal{G}^t_{\widehat{C}}} c\delta}{(1-\delta) \widehat{C}} \sum_{m \in \mathcal{G}} 4\mathbb{E}_t\left[\left\|\nabla f_m\left(x^{t+1}\right)-\nabla f\left(x^{t+1}\right)\right\|^2 \right]\\
&+ \widetilde{B}\mathbb{E}\left[\|\nabla f(x^{t+1})\|^2\right] + \widetilde{\zeta}^2\\
&\stackrel{(\text{As.~\ref{assm:het}})}{\leq} \left(\frac{4 G \mathcal{P}_{\mathcal{G}^t_{\widehat{C}}} c\delta B}{(1-\delta) \widehat{C}} + \widetilde{B}\right)  \mathbb{E}\left[\left\|\nabla f\left(x^{t+1}\right)\right\|^2\right] + \frac{4 G \mathcal{P}_{\mathcal{G}^t_{\widehat{C}}} c\delta \zeta^2}{(1-\delta) \widehat{C}} + \widetilde{\zeta}^2\\
&\stackrel{(\ref{eq:yung-1})}{\leq} \left(\frac{8 G \mathcal{P}_{\mathcal{G}^t_{\widehat{C}}} c\delta B}{(1-\delta) \widehat{C}} + 2\widetilde{B}\right)  \mathbb{E}\left[\left\|\nabla f\left(x^t\right)\right\|^2 + \left\|\nabla f\left(x^{t+1}\right)-\nabla f\left(x^t\right)\right\|^2\right]\\
& + \frac{4 G \mathcal{P}_{\mathcal{G}^t_{\widehat{C}}} c\delta \zeta^2}{(1-\delta) \widehat{C}} + \widetilde{\zeta}^2\\
&\leq \quad  \left(\frac{8 G \mathcal{P}_{\mathcal{G}^t_{\widehat{C}}} c\delta B}{(1-\delta) \widehat{C}} + 2\widetilde{B}\right)  \mathbb{E}\left[\left\|\nabla f\left(x^t\right)\right\|^2  + L^2\left\|x^{t+1}-x^t\right\|^2\right]\\
&+ \frac{4 G \mathcal{P}_{\mathcal{G}^t_{\widehat{C}}} c\delta \zeta^2}{(1-\delta) \widehat{C}} + \widetilde{\zeta}^2,
\end{align*}
which concludes the proof.
\end{proof}

\begin{lemma}
\label{lemma:good_aggr}
Let Assumptions~\ref{assm:L-smoothness}, \ref{assm:global}, \ref{assm:local} hold and the Compression Operator satisfy Definition~\ref{def:Q}. Also let us introduce the notation
\begin{align*}\texttt{ARAgg}_Q^{t+1} = \texttt{ARAgg}\left(\clip_{\lambda_{t+1}}\left(\cQ\left(\widehat{\Delta}_1(x^{t+1}, x^t)\right)\right),\right.\\
\left.\ldots, \clip_{\lambda_{t+1}}\left(\cQ\left(\widehat{\Delta}_C(x^{t+1}, x^t)\right)\right)\right).\end{align*} 
Then for all $t\geq 0$ the iterates produced by \gls{Byz-VR-MARINA-PP} (Algorithm~\ref{alg:byz_vr_marina}) satisfy

    \begin{align*}
       T_2 &=   \mathbb{E}\left[ \mathbb{E}_t\left[\left\| \frac{1}{G^t_C} \sum \limits_{m \in \mathcal{G}^t_C}\clip_{\lambda}\left(\mathcal{Q}\left(\widehat{\Delta}_m\left(x^{t+1}, x^t\right)\right)\right)  -   \texttt{ARAgg}_Q^{t+1}\right\|^2\mid [3]\right]\right]\\
       &\leq \frac{8G\mathcal{P}_{\mathcal{G}^t_C}}{(1-\delta)C} \left( 10(1+\omega)\frac{\mathcal{L}_{ \pm}^2}{b} + (10\omega+1) L_{ \pm}^2 + 10\omega  L^2 \right)c\delta\mathbb{E}\left[\|x^{t+1} - x^t\|^2\right],
\end{align*}
where $\mathcal{P}_{\mathcal{G}^t_C} =  \operatorname{Prob}\left\lbrace m \in \mathcal{G}^t_C \mid G^t_C \geq\left(1-\delta\right) C\right\rbrace$.
\end{lemma}
\begin{proof}
By the definition of robust aggregation, we have 
\begin{align*}
   T_2 &=   \mathbb{E}\left[ \mathbb{E}_t\left[\left\| \frac{1}{G^t_C} \sum \limits_{m \in \mathcal{G}^t_C}\clip_{\lambda}\left(\mathcal{Q}\left(\widehat{\Delta}_m\left(x^{t+1}, x^t\right)\right)\right)  -   \texttt{ARAgg}_Q^{t+1}\right\|^2\mid [3]\right]\right]\\
   &\text{ \scriptsize $\leq  \mathbb{E}\left[  \frac{c \delta}{D_2} \sum_{\substack{m, l \in \mathcal{G}_C^t \\
m \neq l}}    
\mathbb{E}_t\left[\left\| \clip_{\lambda}\left(\mathcal{Q}\left(\widehat{\Delta}_m\left(x^{t+1}, x^t\right)\right)\right) - \clip_{\lambda}\left(\mathcal{Q}\left(\widehat{\Delta}_l\left(x^{t+1}, x^t\right)\right)\right) \right\|^2\mid [3]\right]\right]$},
\end{align*}
where $D_2 = G^t_C(G^t_C-1)$. Next, we consider pair-wise differences:
\begin{align*}
T_{2}^\prime(i,l) &= \mathbb{E}_t\left[\left\| \clip_{\lambda}\left(\mathcal{Q}\left(\widehat{\Delta}_m\left(x^{t+1}, x^t\right)\right)\right) - \clip_{\lambda}\left(\mathcal{Q}\left(\widehat{\Delta}_l\left(x^{t+1}, x^t\right)\right)\right) \right\|^2\mid [3]\right]\\
& \stackrel{(\ref{eq:yung-1})}{\leq}   2\mathbb{E}_t\left[\left\| \clip_{\lambda}\left(\mathcal{Q}\left(\widehat{\Delta}_m\left(x^{t+1}, x^t\right)\right)\right)  - \Delta_m\left(x^{t+1}, x^t\right) + \Delta_l\left(x^{t+1}, x^t\right) -\right.\right.\\
&\left.\left.-\clip_{\lambda}\left(\mathcal{Q}\left(\widehat{\Delta}_l\left(x^{t+1}, x^t\right)\right)\right) \right\|^2\mid [3]\right]\\
&+ 2   \mathbb{E}_t\left[\left\|  \Delta_m\left(x^{t+1}, x^t\right) - \Delta_l\left(x^{t+1}, x^t\right) \right\|^2\mid [3]\right]\\
& \stackrel{(\ref{eq:yung-1})}{\leq}  4 \mathbb{E}_t\left[\left\| \clip_{\lambda}\left(\mathcal{Q}\left(\widehat{\Delta}_m\left(x^{t+1}, x^t\right)\right)\right) -\Delta_m\left(x^{t+1}, x^t\right) \right\|^2\mid [3]\right]\\
&+ 4  \mathbb{E}_t\left[\left\| \Delta_l\left(x^{t+1}, x^t\right) - \clip_{\lambda}\left(\mathcal{Q}\left(\widehat{\Delta}_l\left(x^{t+1}, x^t\right)\right)\right) \right\|^2\mid [3]\right]\\
&+ 2  \mathbb{E}_t\left[\left\|  \Delta_l\left(x^{t+1}, x^t\right) - \Delta_m\left(x^{t+1}, x^t\right) \right\|^2\mid [3]\right]]\\
& \stackrel{(\ref{eq:yung-1})}{\leq}  4  \mathbb{E}_t\left[\left\| \clip_{\lambda}\left(\mathcal{Q}\left(\widehat{\Delta}_m\left(x^{t+1}, x^t\right)\right)\right) -\Delta_m\left(x^{t+1}, x^t\right)  \right\|^2\mid [3]\right]\\
&+ 4 \mathbb{E}_t\left[\left\| \Delta_l\left(x^{t+1}, x^t\right) - \clip_{\lambda}\left(\mathcal{Q}\left(\widehat{\Delta}_l\left(x^{t+1}, x^t\right)\right)\right) \right\|^2\mid [3]\right]\\
&+ 4  \mathbb{E}_t\left[\left\|  \Delta_l\left(x^{t+1}, x^t\right) - \Delta\left(x^{t+1}, x^t\right) \right\|^2\mid [3]\right]\\
&+ 4 \mathbb{E}_t\left[\left\|  \Delta_m\left(x^{t+1}, x^t\right) - \Delta\left(x^{t+1}, x^t\right)\right\|^2\mid [3]\right].
\end{align*}
Now we can combine all parts together:
\begin{align*}
    \widehat{T}_2 & = \mathbb{E}\left[\frac{1}{G^t_C(G^t_C - 1)}  \sum_{\substack{m, l \in \mathcal{G}_C^t \\
m \neq l}} T_{2}^\prime(i,l) \right]  \\
& \leq   \mathbb{E}\left[ \frac{1}{D_2} \sum_{\substack{m, l \in \mathcal{G}_C^t \\
m \neq l}} 4  \mathbb{E}_t\left[\left\| \clip_{\lambda}\left(\mathcal{Q}\left(\widehat{\Delta}_m\left(x^{t+1}, x^t\right)\right)\right) -\Delta_m\left(x^{t+1}, x^t\right)  \right\|^2\mid [3]\right]\right]\\
&+  \mathbb{E}\left[  \frac{1}{D_2}  \sum_{\substack{m, l \in \mathcal{G}_C^t \\
m \neq l}} 4\mathbb{E}_t\left[\left\| \Delta_l\left(x^{t+1}, x^t\right) - \clip_{\lambda}\left(\mathcal{Q}\left(\widehat{\Delta}_l\left(x^{t+1}, x^t\right)\right)\right) \right\|^2\mid [3]\right]\right]\\
&+    \mathbb{E}\left[  \frac{1}{D_2}  \sum_{\substack{m, l \in \mathcal{G}_C^t \\
m \neq l}} 4  \mathbb{E}_t\left[\left\|  \Delta_l\left(x^{t+1}, x^t\right) - \Delta\left(x^{t+1}, x^t\right) \right\|^2\mid [3]\right]\right]\\
&+   \mathbb{E}\left[  \frac{1}{D_2}  \sum_{\substack{m, l \in \mathcal{G}_C^t \\
m \neq l}} 4 \mathbb{E}_t\left[\left\|  \Delta_m\left(x^{t+1}, x^t\right) - \Delta\left(x^{t+1}, x^t\right)\right\|^2\mid [3]\right]\right].
\end{align*}
Rearranging the terms, we obtain
\begin{align*}
    \widehat{T}_2& \leq   \mathbb{E}\left[ \frac{1}{D_2} \sum_{\substack{m, l \in \mathcal{G}_C^t \\
m \neq l}} 8  \mathbb{E}_t\left[\left\| \clip_{\lambda}\left(\mathcal{Q}\left(\widehat{\Delta}_m\left(x^{t+1}, x^t\right)\right)\right) -\Delta_m\left(x^{t+1}, x^t\right)  \right\|^2\mid [3]\right]\right]\\
&+    \mathbb{E}\left[  \frac{1}{D_2}  \sum_{\substack{m, l \in \mathcal{G}_C^t \\
m \neq l}} 8 \mathbb{E}_t\left[\left\|  \Delta_m\left(x^{t+1}, x^t\right) - \Delta\left(x^{t+1}, x^t\right) \right\|^2\mid [3]\right]\right].
\end{align*}
It leads to 
\begin{align*}
    \widehat{T}_2&\leq \mathbb{E}\left[ \frac{1}{G^t_C} \sum_{m \in \mathcal{G}^t_C} 8  \mathbb{E}_t\left[\left\| \clip_{\lambda}\left(\mathcal{Q}\left(\widehat{\Delta}_m\left(x^{t+1}, x^t\right)\right)\right) -\Delta_m\left(x^{t+1}, x^t\right)  \right\|^2\mid [3]\right] \right]\\
    &+\mathbb{E}\left[ \frac{1}{G^t_C} \sum_{m \in \mathcal{G}^t_C} 8 \mathbb{E}_t\left[\left\|  \Delta_m\left(x^{t+1}, x^t\right) - \Delta\left(x^{t+1}, x^t\right) \right\|^2\mid [3]\right] \right]\\
    &\overset{\text{Lemma~\ref{lemma:clipping}}}{\leq} \mathbb{E}\left[ \frac{1}{G^t_C} \sum_{m \in \mathcal{G}^t_C} 80  \mathbb{E}_t\left[\left\| \mathcal{Q}\left(\widehat{\Delta}_m\left(x^{t+1}, x^t\right)\right) -\Delta_m\left(x^{t+1}, x^t\right)  \right\|^2\mid [3]\right] \right]\\
    &+\mathbb{E}\left[ \frac{1}{G^t_C} \sum_{m \in \mathcal{G}^t_C} 8 \mathbb{E}_t\left[\left\|  \Delta_m\left(x^{t+1}, x^t\right) - \Delta\left(x^{t+1}, x^t\right) \right\|^2\mid [3]\right] \right].
\end{align*}
Using variance decomposition we get
\begin{align*}
        \widehat{T}_2&\leq \mathbb{E}\left[ \frac{1}{G^t_C} \sum_{m \in \mathcal{G}^t_C} 80  \mathbb{E}_t\left[\left\| \mathcal{Q}\left(\widehat{\Delta}_m\left(x^{t+1}, x^t\right)\right) \right\|^2\mid [3] \right]\right]\\
        &-\mathbb{E}\left[ \frac{1}{G^t_C} \sum_{m \in \mathcal{G}^t_C} 80  \mathbb{E}_t\left[\left\| \Delta_m\left(x^{t+1}, x^t\right) \right\|^2\mid [3]\right] \right]\\
&+\mathbb{E}\left[ \frac{1}{G^t_C} \sum_{m \in \mathcal{G}^t_C} 8 \mathbb{E}_t\left[\left\|  \Delta_m\left(x^{t+1}, x^t\right) - \Delta\left(x^{t+1}, x^t\right) \right\|^2\mid [3]\right] \right].
\end{align*}    
Using properties of unbiased compressors (Definition~\ref{def:Q}) we have 
\begin{align*}
        \widehat{T}_2&\leq \mathbb{E}\left[ \frac{1}{G^t_C} \sum_{m \in \mathcal{G}^t_C} 80(1+\omega)  \mathbb{E}_t\left[\left\| \widehat{\Delta}_m\left(x^{t+1}, x^t\right) \right\|^2\mid [3] \right]\right]\\
        &-\mathbb{E}\left[ \frac{1}{G^t_C} \sum_{m \in \mathcal{G}^t_C} 80  \mathbb{E}_t\left[\left\| \Delta_m\left(x^{t+1}, x^t\right) \right\|^2\mid [3]\right] \right]\\
&+\mathbb{E}\left[ \frac{1}{G^t_C} \sum_{m \in \mathcal{G}^t_C} 8 \mathbb{E}_t\left[\left\|  \Delta_m\left(x^{t+1}, x^t\right) - \Delta\left(x^{t+1}, x^t\right) \right\|^2\mid [3]\right] \right].
\end{align*}
Also we have
\begin{align*}
\widehat{T}_2&\leq \mathbb{E}\left[ \frac{1}{G^t_C} \sum_{m \in \mathcal{G}^t_C} 80(1+\omega)  \mathbb{E}_t\left[\left\| \widehat{\Delta}_m\left(x^{t+1}, x^t\right) - \Delta_m\left(x^{t+1}, x^t\right)\right\|^2\mid [3] \right]\right]\\
        &+\mathbb{E}\left[ \frac{1}{G^t_C} \sum_{m \in \mathcal{G}^t_C} 80(1+\omega)  \mathbb{E}_t\left[\left\| \Delta_m\left(x^{t+1}, x^t\right) \right\|^2\mid [3]\right] \right]\\
        &-\mathbb{E}\left[ \frac{1}{G^t_C} \sum_{m \in \mathcal{G}^t_C} 80  \mathbb{E}_t\left[\left\| \Delta_m\left(x^{t+1}, x^t\right) \right\|^2\mid [3]\right] \right]\\
&+\mathbb{E}\left[ \frac{1}{G^t_C} \sum_{m \in \mathcal{G}^t_C} 8 \mathbb{E}_t\left[\left\|  \Delta_m\left(x^{t+1}, x^t\right) - \Delta\left(x^{t+1}, x^t\right) \right\|^2\mid [3]\right] \right].
\end{align*} 
Let us simplify the inequality:
\begin{align*}
    \widehat{T}_2 &\leq \mathbb{E}\left[ \frac{1}{G^t_C} \sum_{m \in \mathcal{G}^t_C} 80(1+\omega)  \mathbb{E}_t\left[\left\| \widehat{\Delta}_m\left(x^{t+1}, x^t\right) - \Delta_m\left(x^{t+1}, x^t\right)\right\|^2\mid [3] \right]\right]\\
        &+\mathbb{E}\left[ \frac{1}{G^t_C} \sum_{m \in \mathcal{G}^t_C} 80\omega  \mathbb{E}_t\left[\left\| \Delta_m\left(x^{t+1}, x^t\right) \right\|^2\mid [3]\right] \right]\\
&+\mathbb{E}\left[ \frac{1}{G^t_C} \sum_{m \in \mathcal{G}^t_C} 8 \mathbb{E}_t\left[\left\|  \Delta_m\left(x^{t+1}, x^t\right) - \Delta\left(x^{t+1}, x^t\right) \right\|^2\mid [3]\right] \right].
\end{align*}
Using a variance decomposition once again, we get
\begin{align*}
    \widehat{T}_2 &\leq \mathbb{E}\left[ \frac{1}{G^t_C} \sum_{m \in \mathcal{G}^t_C} 80(1+\omega)  \mathbb{E}_t\left[\left\| \widehat{\Delta}_m\left(x^{t+1}, x^t\right) - \Delta_m\left(x^{t+1}, x^t\right)\right\|^2\mid [3] \right]\right]\\
        &+\mathbb{E}\left[ \frac{1}{G^t_C} \sum_{m \in \mathcal{G}^t_C} 80\omega  \mathbb{E}_t\left[\left\| \Delta_m\left(x^{t+1}, x^t\right) - \Delta\left(x^{t+1}, x^t\right)  \right\|^2\mid [3]\right] \right]\\
&+\mathbb{E}\left[ \frac{1}{G^t_C} \sum_{m \in \mathcal{G}^t_C} 8 \mathbb{E}_t\left[\left\|  \Delta_m\left(x^{t+1}, x^t\right) - \Delta\left(x^{t+1}, x^t\right) \right\|^2\mid [3]\right] \right]\\
&+ \mathbb{E}\left[ \frac{1}{G^t_C} \sum_{m \in \mathcal{G}^t_C} 80\omega  \mathbb{E}_t\left[\left\| \Delta\left(x^{t+1}, x^t\right)  \right\|^2\mid [3]\right] \right].
\end{align*}
Using a similar argument to the one used in the previous lemma, we obtain
\begin{align*}
    &\widehat{T}_2\\
    &\leq \mathbb{E}\left[ \frac{\mathcal{P}_{\mathcal{G}^t_C}}{(1-\delta)C} \sum_{m \in \mathcal{G}} 80(1+\omega)  \mathbb{E}_t\left[\left\| \widehat{\Delta}_m\left(x^{t+1}, x^t\right) - \Delta_m\left(x^{t+1}, x^t\right)\right\|^2\mid [3] \right]\right]\\
        &+\mathbb{E}\left[ \frac{\mathcal{P}_{\mathcal{G}^t_C}}{(1-\delta)C} \sum_{m \in \mathcal{G}} 80\omega  \mathbb{E}_t\left[\left\| \Delta_m\left(x^{t+1}, x^t\right) - \Delta\left(x^{t+1}, x^t\right)  \right\|^2\mid [3]\right] \right]\\
&+\mathbb{E}\left[ \frac{\mathcal{P}_{\mathcal{G}^t_C}}{(1-\delta)C} \sum_{m \in \mathcal{G}} 8 \mathbb{E}_t\left[\left\|  \Delta_m\left(x^{t+1}, x^t\right) - \Delta\left(x^{t+1}, x^t\right) \right\|^2\mid [3]\right] \right]\\
&+ \mathbb{E}\left[ \frac{\mathcal{P}_{\mathcal{G}^t_C}}{(1-\delta)C} \sum_{m \in \mathcal{G}} 80\omega  \mathbb{E}_t\left[\left\| \Delta\left(x^{t+1}, x^t\right)  \right\|^2\mid [3]\right] \right].
\end{align*}
Using Assumptions \ref{assm:L-smoothness}, \ref{assm:global}, \ref{assm:local}: 
    \begin{align*}
    \widehat{T}_2 &\leq \mathbb{E}\left[  \frac{80(1+\omega)  G\mathcal{P}_{\mathcal{G}^t_C}\mathcal{L}_{ \pm}^2}{(1-\delta)Cb}\|x^{t+1}-x^t\|^2 \right]\\
    &+\mathbb{E}\left[  \frac{8(10\omega+1)G\mathcal{P}_{\mathcal{G}^t_C}   L_{ \pm}^2}{(1-\delta)C} \|x^{t+1}-x^t\|^2\right]\\
        &+\mathbb{E}\left[  \frac{80G\mathcal{P}_{\mathcal{G}^t_C} \omega  L^2}{(1-\delta)C} \|x^{t+1}-x^t\|^2\right].
\end{align*}
Finally, we obtain
\begin{align*}
       T_2 &=   \mathbb{E}\left[ \mathbb{E}_t\left[\left\| \frac{1}{G^t_C} \sum \limits_{m \in \mathcal{G}^t_C}\clip_{\lambda}\left(\mathcal{Q}\left(\widehat{\Delta}_m\left(x^{t+1}, x^t\right)\right)\right)  -   \texttt{ARAgg}_Q^{t+1}\right\|^2\mid [3]\right]\right]\\
       &\leq \frac{8G\mathcal{P}_{\mathcal{G}^t_C}}{(1-\delta)C} \left( 10(1+\omega)\frac{\mathcal{L}_{ \pm}^2}{b} + (10\omega+1) L_{ \pm}^2 + 10\omega  L^2 \right)c\delta \mathbb{E}\left[\|x^{t+1} - x^t\|^2\right].
\end{align*}

\end{proof}

\begin{lemma}\label{lemma:bad_aggr}
Let Assumptions~\ref{assm:bounded-aggr} and \ref{assm:L-smoothness} hold. Also let us introduce the notation
\begin{align*}
\texttt{ARAgg}_Q^{t+1} = \texttt{ARAgg}\left(\clip_{\lambda_{t+1}}\left(\cQ\left(\widehat{\Delta}_1(x^{t+1}, x^t)\right)\right),\right.\\
\left.\ldots, \clip_{\lambda_{t+1}}\left(\cQ\left(\widehat{\Delta}_C(x^{t+1}, x^t)\right)\right)\right).
\end{align*}
Assume that $\lambda_{t+1} = \alpha_{\lambda_{t+1}} \|x^{t+1} - x^t\| $.
Then for all $t\geq 0$ the iterates produced by \gls{Byz-VR-MARINA-PP} (Algorithm~\ref{alg:byz_vr_marina}) satisfy

        \begin{align*}
        T_3 &= \mathbb{E}\left[\mathbb{E}_t\left[\left\| \nabla f(x^{t+1}) - \nabla f(x^{t}) - \texttt{ARAgg}_Q^{t+1}\right\|^2\mid [2]\right]\right]\\
       &\leq 2(L^2+ F_{\cA}^2\alpha^2_{\lambda_{t+1}})\mathbb{E}\left[\left\|  x^{t+1} - x^t\right\|^2\right]
        \end{align*}
\end{lemma}
\begin{proof}
      \begin{align*}
    T_3 &=   \mathbb{E}\left[\mathbb{E}_t\left[\left\| \nabla f(x^{t+1}) - \nabla f(x^{t}) - \texttt{ARAgg}_Q^{t+1}\right\|^2\mid [2]\right]\right]\\
    &\stackrel{(\ref{eq:yung-1})}{\leq} \mathbb{E}\left[\mathbb{E}_t\left[2\left\| \nabla f(x^{t+1}) - \nabla f(x^{t}) \right\|^2+2\left\| \texttt{ARAgg}_Q^{t+1}\right\|^2\mid [2]\right]\right]\\
            \end{align*}
            Using $L$-smoothness and Assumption~\ref{assm:bounded-aggr} we have 

          \begin{align*}
    T_3 
    &\stackrel{(\ref{eq:yung-1})}{\leq} \mathbb{E}\left[\mathbb{E}_t\left[2L^2\left\|  x^{t+1} - x^t\right\|^2+2F_{\cA}^2\lambda^2_{t+1}\mid [2]\right]\right]\\
     &\leq\mathbb{E}\left[\mathbb{E}_t\left[2L^2\left\|  x^{t+1} - x^t\right\|^2+2F_{\cA}^2\alpha^2_{\lambda_{t+1}}\|x^{t+1} - x^t\|^2\mid [2]\right]\right]\\
      &\leq 2(L^2+F_{\cA}^2\alpha^2_{\lambda_{t+1}})\mathbb{E}\left[\left\|  x^{t+1} - x^t\right\|^2\right].
            \end{align*}        
\end{proof}

\begin{lemma}
\label{lemma:final_lemma}
    Let Assumptions ~\ref{assm:bounded-aggr}, \ref{assm:L-smoothness}, \ref{assm:global}, \ref{assm:local}, \ref{assm:het} hold and Compression Operator satisfy Definition~\ref{def:Q}.  Also let us introduce the notation
\begin{align*}
    \texttt{ARAgg}_Q^{t+1} = \texttt{ARAgg}\left(\clip_{\lambda_{t+1}}\left(\cQ\left(\widehat{\Delta}_1(x^{t+1}, x^t)\right)\right),\right.\\
    \left.\ldots, \clip_{\lambda_{t+1}}\left(\cQ\left(\widehat{\Delta}_C(x^{t+1}, x^t)\right)\right)\right).
    \end{align*}
Then for all $t\geq 0$ the iterates produced by \gls{Byz-VR-MARINA-PP} (Algorithm~\ref{alg:byz_vr_marina}) satisfy

    \begin{align*}
    \mathbb{E}\left[\left\|g^{t+1}-\nabla f\left(x^{t+1}\right)\right\|^2\right] &\leq  \left(1-\frac{p}{4}\right) \mathbb{E}\left[\left\|g^{t}-\nabla f\left(x^{t}\right)\right\|^2\right]\\
    & +  \widehat{B}\mathbb{E}\left[\left\|\nabla f\left(x^t\right)\right\|^2\right]  + \widehat{D}\zeta^2+\frac{pA}{4}\|x^{t+1} - x^t\|^2,
\end{align*}
where \begin{align*}
    A& = \frac{4}{p}\left( \frac{80}{p} \frac{p_G \mathcal{P}_{\mathcal{G}^t_C} n}{C} \omega + 24 \frac{ G\mathcal{P}_{\mathcal{G}^t_{\widehat{C}}} c\delta}{(1-\delta)\widehat C} B + 6\widetilde{B} + \frac{4}{p}(1-p_G)+\right.\\
    &+\left.\frac{160}{p}p_G \frac{G\mathcal{P}_{\mathcal{G}^t_C} }{(1-\delta)C}c\delta\omega  \right) L^2\\
    &+\frac{4}{p}\left( \frac{8}{p} \frac{p_G \mathcal{P}_{\mathcal{G}^t_C} n}{C} \left( 10\omega + 1 \right) + \frac{16}{p}p_G \frac{G\mathcal{P}_{\mathcal{G}^t_C} }{(1-\delta)C}c\delta(10\omega+1) \right)L_{ \pm}^2\\
    &+\frac{4}{p}\left( \frac{160}{p} p_G \frac{G \mathcal{P}_{\mathcal{G}^t_C}}{(1-\delta) C} (1+\omega)c\delta +\frac{80}{p} p_G \mathcal{P}_{\mathcal{G}^t_C} (1+\omega) \frac{n}{C} \right)\frac{\mathcal{L}_{ \pm}^2}{b}\\
    &+\frac{4}{p}\left( \frac{4}{p}(1-p_G)F_{\cA}^2\alpha^2_{\lambda_{t+1}} \right),
\end{align*}
\begin{align*}
    \widehat{B} &= 2 \frac{ \delta\mathcal{P}_{\mathcal{G}^t_{\widehat{C}}} }{1-\delta} B\left(\frac{12cG}{\widehat C} + p \right) + 6\widetilde{B},\\ \widehat{D} &= 2 \frac{ \delta\mathcal{P}_{\mathcal{G}^t_{\widehat{C}}} }{1-\delta}\left(\frac{6cG}{\widehat C} + p \right) + \widetilde{D},
\end{align*}
and where $\widetilde{B} \eqdef 0$ and $\widetilde{D} \eqdef 0$ when $\widehat{C} = M$, $\widetilde{B} \eqdef \frac{\cP_{\cG_{\widehat{C}}^t}GB}{(1-\delta)\widehat{C}}$ and $\widetilde{D} \eqdef \frac{\cP_{\cG_{\widehat{C}}^t}G}{(1-\delta)\widehat{C}}$ when $\widehat{C} < M$, $p_G = \operatorname{Prob}\left\lbrace G^t_C \geq (1-\delta)C \right\rbrace $, and 
$$\mathcal{P}_{\mathcal{G}^t_C} =  \operatorname{Prob}\left\lbrace m \in \mathcal{G}^t_C \mid G^t_C \geq\left(1-\delta\right) C\right\rbrace$$.
\end{lemma}

\begin{proof}

Let us combine bounds for $A_1$ and $A_2$ together:
\begin{align*}
     A_0 &=    \mathbb{E}\left[\left\|g^{t+1}-\nabla f\left(x^{t+1}\right)\right\|^2\right]\\
     & \leq \left(1+\frac{p}{2}\right)\mathbb{E}\left[\left\|\overline{g}^{t+1}-\nabla f\left(x^{t+1}\right)\right\|^2\right]\\
     &+\left(1+\frac{2}{p}\right)\mathbb{E}\left[\left\|g^{t+1}-\overline{g}^{t+1}\right\|^2\right]\\
     &\leq \left(1+\frac{p}{2}\right)A_1 + \left(1+\frac{2}{p}\right)A_2\\
       & \leq \left(1+\frac{p}{2}\right)(1-p) \left(1+ \frac{p}{4}\right)\mathbb{E}\left[\left\|g^{t}-\nabla f\left(x^{t}\right)\right\|^2\right]\\
    &+ \left(1+\frac{p}{2}\right)(1-p)p_G\left(1+ \frac{4}{p}\right) \frac{2\cdot\mathcal{P}_{\mathcal{G}^t_C}M}{C} \left( 10 \omega L^2+(10\omega+1) L_{ \pm}^2\right.+\\
    &\left.+\frac{10(\omega+1) \mathcal{L}_{ \pm}^2}{b} \right) \mathbb{E} \left[ \|x^{t+1} - x^t\|^2\right]\\
    &+ \left(1+\frac{p}{2}\right)p\left(\frac{ \delta\cdot\mathcal{P }_{\mathcal{G}^t_{\widehat{C}}} }{(1-\delta)}  \mathbb{E}\left[ B\|\nabla f(x)\|^2+\zeta^2 \right]\right)\\
    &+ \left(1+\frac{2}{p}\right)p\mathbb{E}\left[\mathbb{E}_t\left[\left\|\texttt{ARAgg}\left(\nabla f_1(x^{t+1}), \ldots, \nabla f_{\widehat{C}}(x^{t+1})\right) \right.\right.\right.\\
    &-\left.\left.\left.  \nabla f(x^{t+1})\right\|^2\right]\mid [1]\right]\\
   & + \left(1+\frac{2}{p}\right)(1-p)p_G \mathbb{E}\left[ \mathbb{E}_t\left[\left\| \frac{1}{G^t_C} \sum \limits_{m \in \mathcal{G}^t_C}\clip_{\lambda}\left(\mathcal{Q}\left(\widehat{\Delta}_m\left(x^{t+1}, x^t\right)\right)\right)  - \right.\right.\right.\\
   &-\left.\left.\left.\texttt{ARAgg}_Q^{t+1}\right\|^2\mid [3]\right]\right]\\
   &+ \left(1+\frac{2}{p}\right)(1-p)(1-p_G)\mathbb{E}\left[\mathbb{E}_t\left[\left\| \nabla f(x^{t+1}) - \nabla f(x^{t}) - \right.\right.\right.\\ 
   -&\left.\left.\left.\texttt{ARAgg}_Q^{t+1}\right\|^2\mid [2]\right]\right].
   \end{align*}
   Finally, we obtain the following bound:
   \begin{align*}
A_0  & \stackrel{(\ref{eq:yung-1})}{\leq}  \left(1-\frac{p}{4}\right) \mathbb{E}\left[\left\|g^{t}-\nabla f\left(x^{t}\right)\right\|^2\right]\\
  &+  \frac{8}{p} \frac{\mathcal{P}_{\mathcal{G}^t_C}M}{C}p_G  \left( 10 \omega L^2+(10\omega+1) L_{ \pm}^2+\frac{10(\omega+1) \mathcal{L}_{ \pm}^2}{b} \right)\mathbb{E}\left[\|x^{t+1} - x^t\|^2\right]\\
  &+2p\left(\frac{ \delta\cdot\mathcal{P }_{\mathcal{G}^t_{\widehat{C}}} }{1-\delta}  \mathbb{E}\left[ B\|\nabla f(x)\|^2+\zeta^2 \right]\right)\\
      &+ \left(p+2\right)\mathbb{E}\left[\mathbb{E}_t\left[\left\|\texttt{ARAgg}\left(\nabla f_1(x^{t+1}), \ldots, \nabla f_n(x^{t+1})\right) - \nabla f(x^{t+1})\right\|^2\right]\mid [1] \right]\\
   & + \frac{2}{p}p_G \mathbb{E}\left[ \mathbb{E}_t\left[\left\| \frac{1}{G^t_C} \sum \limits_{m \in \mathcal{G}^t_C}\clip_{\lambda}\left(\mathcal{Q}\left(\widehat{\Delta}_m\left(x^{t+1}, x^t\right)\right)\right)  -   \texttt{ARAgg}_Q^{t+1}\right\|^2\mid [3]\right]\right]\\
   &+ \frac{2}{p}(1-p_G)\mathbb{E}\left[\mathbb{E}_t\left[\left\| \nabla f(x^{t+1}) - \nabla f(x^{t}) - \texttt{ARAgg}_Q^{t+1}\right\|^2\mid [2]\right]\right]
\end{align*}
Now, we can apply Lemmas \ref{lemma:full_aggr}, \ref{lemma:good_aggr}, \ref{lemma:bad_aggr}:
   \begin{align*}
   A_0 &=  \mathbb{E}\left[\left\|g^{t+1}-\nabla f\left(x^{t+1}\right)\right\|^2\right]\\ 
   & \leq \left(1-\frac{p}{4}\right) \mathbb{E}\left[\left\|g^{t}-\nabla f\left(x^{t}\right)\right\|^2\right]\\
  &+  \frac{8}{p} \frac{\mathcal{P}_{\mathcal{G}^t_C}M}{C}p_G  \left( 10 \omega L^2+(10\omega+1) L_{ \pm}^2+\frac{10(\omega+1) \mathcal{L}_{ \pm}^2}{b} \right)\mathbb{E}\left[\|x^{t+1} - x^t\|^2\right]\\
      &+ \left(p+2\right)\left(\frac{8 G \mathcal{P}_{\mathcal{G}^t_{\widehat{C}}} c\delta B}{(1-\delta) \widehat{C}} + 2\widetilde{B}\right)\mathbb{E}\left[\left\|\nabla f\left(x^t\right)\right\|^2 + L^2\left\|x^{t+1}-x^t\right\|^2\right]\\
      &+4\left(p+2\right) \frac{G \mathcal{P}_{\mathcal{G}^t_{\widehat{C}}} c\delta}{(1-\delta)\widehat{C}}  \zeta^2 + (p+2)\widetilde{\zeta}^2\\
   & + \frac{2}{p}p_G\mathbb{E}\left[  80(1+\omega) \frac{G\mathcal{P}_{\mathcal{G}^t_C} }{(1-\delta)C} \frac{\mathcal{L}_{ \pm}^2}{b}c\delta \|x^{t+1}-x^t\|^2 \right] \\
    &    + \frac{2}{p}p_G \mathbb{E}\left[  8(10\omega+1)\frac{G\mathcal{P}_{\mathcal{G}^t_C} }{(1-\delta)C} L_{ \pm}^2c\delta  \|x^{t+1}-x^t\|^2\right]\\
     &   + \frac{2}{p}p_G \mathbb{E}\left[  80 \frac{G\mathcal{P}_{\mathcal{G}^t_C} }{(1-\delta)C} \omega  L^2 c\delta  \|x^{t+1}-x^t\|^2\right]\\
   &+ \frac{2}{p}(1-p_G)2(L^2+F_{\cA}^2\alpha^2_{\lambda_{t+1}})\mathbb{E}\left[\left\|  x^{t+1} - x^t\right\|^2\right]\\
 &  +2p\frac{ \delta\mathcal{P }_{\mathcal{G}^t_{\widehat{C}}} }{1-\delta}  \mathbb{E}\left[ B\|\nabla f(x)\|^2+\zeta^2 \right].
\end{align*}
Finally, we have 
\begin{align*}
    \mathbb{E}\left[\left\|g^{t+1}-\nabla f\left(x^{t+1}\right)\right\|^2\right]   &\leq  \left(1-\frac{p}{4}\right) \mathbb{E}\left[\left\|g^{t}-\nabla f\left(x^{t}\right)\right\|^2\right]\\
    &+\widehat{B}\mathbb{E}\left[\left\|\nabla f\left(x^t\right)\right\|^2\right] + \widehat{D} \zeta^2+\frac{p A}{4} \mathbb{E}\left[\|x^{t+1} - x^t\|^2\right],
\end{align*}
where 
\begin{align*}
    A &= \text{\small $\frac{32p_G }{p^2} \frac{\mathcal{P}_{\mathcal{G}^t_C}M}{C} \left( 10 \omega L^2+(10\omega+1) L_{ \pm}^2+\frac{10(\omega+1) \mathcal{L}_{ \pm}^2}{b} \right)$}\\
    &+\text{\small $\frac{8}{p^2}\frac{G\mathcal{P}_{\mathcal{G}^t_C} }{(1-\delta)C}p_G c\delta\left( 80(1+\omega)  \frac{\mathcal{L}_{ \pm}^2}{b} + 8(10\omega+1)L_{ \pm}^2  
       + 80\omega  L^2\right)$}\\
       &+\text{\small $\frac{4}{p} \left(\frac{24 G \mathcal{P}_{\mathcal{G}^t_{\widehat{C}}} c\delta B}{(1-\delta) \widehat{C}} + 6\widetilde{B}\right)  L^2 +\frac{16(1-p_G)}{p^2}(L^2+F_{\cA}^2\alpha^2_{\lambda_{t+1}})$},
\end{align*}
and
\begin{align*}
    \widehat{B} = 2 \frac{ \delta\mathcal{P}_{\mathcal{G}^t_{\widehat{C}}} }{1-\delta} B\left(\frac{12cG}{\widehat C} + p \right) + 6\widetilde{B}, \quad \widehat{D} = 2 \frac{ \delta\mathcal{P}_{\mathcal{G}^t_{\widehat{C}}} }{1-\delta}\left(\frac{6cG}{\widehat C} + p \right) + \widetilde{D},
\end{align*}
where $\widetilde{D} \eqdef 0$ when $\widehat{C} = M$, and $\widetilde{D} \eqdef \frac{\cP_{\cG_{\widehat{C}}^t}G}{(1-\delta)\widehat{C}}$ when $\widehat{C} < M$.
Once we simplify the equation, we obtain
\begin{align*}
    A& = \frac{4}{p}\left( \frac{80}{p} \frac{p_G \mathcal{P}_{\mathcal{G}^t_C} n}{C} \omega + 24 \frac{ G\mathcal{P}_{\mathcal{G}^t_{\widehat{C}}} c\delta}{(1-\delta)\widehat C} B + 6\widetilde{B} + \frac{4}{p}(1-p_G)+\right.\\
    &\left.+\frac{160}{p}p_G \frac{G\mathcal{P}_{\mathcal{G}^t_C} }{(1-\delta)C}c\delta\omega  \right) L^2\\
    &+\frac{4}{p}\left( \frac{8}{p} \frac{p_G \mathcal{P}_{\mathcal{G}^t_C} n}{C} \left( 10\omega + 1 \right) + \frac{16}{p}p_G \frac{G\mathcal{P}_{\mathcal{G}^t_C} }{(1-\delta)C}c\delta(10\omega+1) \right)L_{ \pm}^2\\
    &+\frac{4}{p}\left( \frac{160}{p} p_G \frac{G \mathcal{P}_{\mathcal{G}^t_C}}{(1-\delta) C} (1+\omega)c\delta +\frac{80}{p} p_G \mathcal{P}_{\mathcal{G}^t_C} (1+\omega) \frac{n}{C} \right)\frac{\mathcal{L}_{ \pm}^2}{b}\\
    &+\frac{4}{p}\left( \frac{4}{p}(1-p_G)F_{\cA}^2\alpha^2_{\lambda_{t+1}} \right).
\end{align*}
\end{proof}

\subsection{Main results}

\begin{theorem}
\label{them:1}
 Let Assumptions \ref{assm:bounded-aggr}, \ref{assm:L-smoothness}, \ref{assm:global}, \ref{assm:local}, \ref{assm:het} hold. Setting clipping parameter $\lambda_{t+1} = 2\max_{i\in \mathcal{G}} L_m \left\|x^{t+1} - x^t\right\|$. Assume that
$$
0<\gamma \leq \frac{1}{L+\sqrt{A}}, \quad 4\widehat{B}<p,
$$
where 
\begin{align*}
    A& = \frac{4}{p}\left( \frac{80}{p} \frac{p_G \mathcal{P}_{\mathcal{G}^t_C} n}{C} \omega + 24 \frac{ G\mathcal{P}_{\mathcal{G}^t_{\widehat{C}}} c\delta}{(1-\delta)\widehat C} B + 6\widetilde{B} + \frac{4}{p}(1-p_G)\right.\\
    &\left.+\frac{160}{p}p_G \frac{G\mathcal{P}_{\mathcal{G}^t_C} }{(1-\delta)C}c\delta\omega  \right) L^2\\
    &+\frac{4}{p}\left( \frac{8}{p} \frac{p_G \mathcal{P}_{\mathcal{G}^t_C} n}{C} \left( 10\omega + 1 \right) + \frac{16}{p}p_G \frac{G\mathcal{P}_{\mathcal{G}^t_C} }{(1-\delta)C}c\delta(10\omega+1) \right)L_{ \pm}^2\\
    &+\frac{4}{p}\left( \frac{160}{p} p_G \frac{G \mathcal{P}_{\mathcal{G}^t_C}}{(1-\delta) C} (1+\omega)c\delta +\frac{80}{p} p_G \mathcal{P}_{\mathcal{G}^t_C} (1+\omega) \frac{n}{C} \right)\frac{\mathcal{L}_{ \pm}^2}{b}\\
    &+\frac{4}{p}\left( \frac{4}{p}(1-p_G)F_{\cA}^2\alpha^2_{\lambda_{t+1}} \right),
\end{align*}
\begin{align*}
    \widehat{B} = 2 \frac{ \delta\mathcal{P}_{\mathcal{G}^t_{\widehat{C}}} }{1-\delta} B\left(\frac{12cG}{\widehat C} + p \right) + 6\widetilde{B}, \quad \widehat{D} = 2 \frac{ \delta\mathcal{P}_{\mathcal{G}^t_{\widehat{C}}} }{1-\delta}\left(\frac{6cG}{\widehat C} + p \right) + \widetilde{D},
\end{align*}
and where $\widetilde{B} \eqdef 0$ and $\widetilde{D} \eqdef 0$ when $\widehat{C} = M$, and $\widetilde{B} \eqdef \frac{\cP_{\cG_{\widehat{C}}^t}GB}{(1-\delta)\widehat{C}}$ and $\widetilde{D} \eqdef \frac{\cP_{\cG_{\widehat{C}}^t}G}{(1-\delta)\widehat{C}}$ when $\widehat{C} < M$, and 
\begin{align*}
 \mathcal{P}_{\mathcal{G}^t_C} &=   \frac{C}{Mp_G} \cdot \sum_{(1-\delta)C\leq m^\prime \leq C} \left(\left(\begin{array}{l}
G-1 \\
t-1
\end{array}\right)\left(\begin{array}{l}
M-G \\
C-m^\prime
\end{array}\right) \left(\left(\begin{array}{l}
n \\
C
\end{array}\right)\right)^{-1} \right),\\
p_G &=  \PP\left\lbrace G_C^t \geq\left(1-\delta\right) C\right\rbrace\\
&= \sum_{\lceil(1-\delta)C\rceil\leq t \leq C} \left(\left(\begin{array}{l}
G \\
t
\end{array}\right)\left(\begin{array}{l}
M-G \\
C-m^\prime
\end{array}\right) \left(\begin{array}{l}
M-1 \\
C-1
\end{array}\right)^{-1} \right).
\end{align*}
Then for all $T \geq 0$ the iterates produced by \gls{Byz-VR-MARINA} (Algorithm \ref{alg:byz_vr_marina}) satisfy
$$
\mathbb{E}\left[\left\|\nabla f\left(\widehat{x}^T\right)\right\|^2\right] \leq \frac{2 \Phi^{(0)}}{\gamma\left(1-\frac{4 \widehat{B}}{p}\right)(T+1)}+\frac{4 \widehat{D}  \zeta^2}{p-4 \widehat{B} },
$$
where $\widehat{x}^T$ is chosen uniformly at random from $x^0, x^1, \ldots, x^T$, and $\Phi^{(0)}=$ $f\left(x^0\right)-f^{\star}+\frac{2\gamma}{p}\left\|g^0-\nabla f\left(x^0\right)\right\|^2$ . 
\end{theorem}
\begin{proof}[Proof of Theorem~\ref{them:1}]
     For all $T \geq 0$ we also introduce our Lyapunov function $\Phi^{(t)}=f\left(x^t\right)-f^{\star}+\frac{2\gamma}{p}\left\|g^t-\nabla f\left(x^t\right)\right\|^2$. Using the results of Lemmas \ref{lemma:final_lemma} and \ref{lemma:page}, we derive
     \begin{align*}
         \mathbb{E}\left[\Phi^{(t+1)}\right] &\stackrel{(\ref{lemma:page})}{\leq} \mathbb{E}\left[f\left(x^t\right)-f^{\star}-\left(\frac{1}{2 \gamma}-\frac{L}{2}\right)\left\|x^{t+1}-x^t\right\|^2+\frac{\gamma}{2}\left\|g^t-\nabla f\left(x^t\right)\right\|^2\right]\\
         &-\frac{\gamma}{2} \mathbb{E}\left[\left\|\nabla f\left(x^t\right)\right\|^2\right] + \frac{2\gamma}{p}   \mathbb{E}\left[\left\|g^{t+1}-\nabla f\left(x^{t+1}\right)\right\|^2\right]\\
         &\stackrel{(\ref{lemma:final_lemma})}{\leq}  \mathbb{E}\left[f\left(x^t\right)-f^{\star}-\left(\frac{1}{2 \gamma}-\frac{L}{2}\right)\left\|x^{t+1}-x^t\right\|^2+\frac{\gamma}{2}\left\|g^t-\nabla f\left(x^t\right)\right\|^2\right]\\
         &-\frac{\gamma}{2} \mathbb{E}\left[\left\|\nabla f\left(x^t\right)\right\|^2\right] + \frac{2\gamma}{p}     \left(1-\frac{p}{4}\right) \mathbb{E}\left[\left\|g^{t}-\nabla f\left(x^{t}\right)\right\|^2\right]\\
         &+ \frac{2\gamma}{p}\left( \widehat{B}\mathbb{E}\left[\left\|\nabla f\left(x^t\right)\right\|^2\right] + \widehat{D} \zeta^2+\frac{p A}{4}\|x^{t+1} - x^t\|^2\right)\\
         &=\mathbb{E}\left[f\left(x^t\right)-f^{\star}\right] + \frac{2\gamma}{p}\left(\left(1-\frac{p}{4}\right) + \frac{p}{4}\right) \mathbb{E}\left[\left\|g^{t}-\nabla f\left(x^{t}\right)\right\|^2\right]+\frac{2\widehat{D}\zeta^2\gamma}{p}\\
         &+\frac{1}{2\gamma}\left(1-L\gamma-A\gamma^2\right)\mathbb{E}\left[ \|x^{t+1} - x^t\|^2 \right] - \frac{\gamma}{2}\left(1-\frac{4\widehat{B}}{p}\right)\mathbb{E}\left[\left\|\nabla f\left(x^t\right)\right\|^2\right]\\
         &=\mathbb{E}\left[\Phi^{(t)}\right] +\frac{2 \widehat{D}\zeta^2\gamma}{p}+\frac{1}{2\gamma}\left(1-L\gamma-A\gamma^2\right)\mathbb{E}\left[ \|x^{t+1} - x^t\|^2 \right]\\
         &- \frac{\gamma}{2}\left(1-\frac{4\widehat{B}}{p}\right)\mathbb{E}\left[\left\|\nabla f\left(x^t\right)\right\|^2\right].
     \end{align*}
     Using choice of stepsize and second condition: $
0<\gamma \leq \frac{1}{L+\sqrt{A}}, 4\widehat{B}<p
$ and lemma \ref{lemma:peter} we have
\begin{align*}
       \mathbb{E}\left[\Phi^{(t+1)}\right]     &\leq\mathbb{E}\left[\Phi^{(t)}\right] +\frac{2 \widehat{D}\zeta^2\gamma}{p}- \frac{\gamma}{2}\left(1-\frac{4\widehat{B}}{p}\right)\mathbb{E}\left[\left\|\nabla f\left(x^t\right)\right\|^2\right]
     \end{align*}
      Next, we have $\frac{\gamma}{2}\left(1-\frac{4 \widehat{B} }{p}\right)>0$ and $\Phi^{(t+1)}\geq0$. Therefore, summing up the above inequality for $k=0,1, \ldots, K$ and rearranging the terms, we get
\begin{align*}
\frac{1}{T+1} \sum_{t=0}^T \mathbb{E}\left[\left\|\nabla f\left(x^t\right)\right\|^2\right] \leq & \frac{2}{\gamma\left(1-\frac{4 \widehat{B}}{p}\right)(T+1)} \sum_{t=0}^T\left(\mathbb{E}\left[\Phi^{(t)}\right]-\mathbb{E}\left[\Phi^{(t+1)}\right]\right) \\
& +\frac{4 \widehat{D}  \zeta^2}{p-4\widehat{B} } \\
= & \frac{2\left(\mathbb{E}\left[\Phi^{(0)}\right]-\mathbb{E}\left[\Phi^{(t+1)}\right]\right)}{\gamma\left(1-\frac{4 \widehat{B} }{p}\right)(T+1)}+\frac{4\widehat{D}  \zeta^2}{p-4 \widehat{B} } \\
\leq& \frac{2 \mathbb{E}\left[\Phi^{(0)}\right]}{\gamma\left(1-\frac{4\widehat{B} }{p}\right)(T+1)}+\frac{4 \widehat{D}  \zeta^2}{p-4\widehat{B} }.
\end{align*}

\end{proof}

\begin{theorem}
\label{them:2}
Let Assumptions  \ref{assm:bounded-aggr}, \ref{assm:L-smoothness}, \ref{assm:global}, \ref{assm:local}, \ref{assm:het}, \ref{assm:PL} hold. Set clipping parameter  $\lambda_{t+1} = \max_{i\in \mathcal{G}} L_m \left\|x^{t+1} - x^t\right\|$. Assume that
$$
0<\gamma \leq \frac{1}{L+\sqrt{2 A}}, \quad 8\widehat{B}<p
$$
where 
\begin{align*}
    A& = \frac{4}{p}\left( \frac{80}{p} \frac{p_G \mathcal{P}_{\mathcal{G}^t_C} n}{C} \omega + 24 \frac{ G\mathcal{P}_{\mathcal{G}^t_{\widehat{C}}} c\delta}{(1-\delta)\widehat C} B + 6\widetilde{B} + \frac{4}{p}(1-p_G)\right.\\
    &\left.+\frac{160}{p}p_G \frac{G\mathcal{P}_{\mathcal{G}^t_C} }{(1-\delta)C}c\delta\omega  \right) L^2\\
    &+\frac{4}{p}\left( \frac{8}{p} \frac{p_G \mathcal{P}_{\mathcal{G}^t_C} n}{C} \left( 10\omega + 1 \right) + \frac{16}{p}p_G \frac{G\mathcal{P}_{\mathcal{G}^t_C} }{(1-\delta)C}c\delta(10\omega+1) \right)L_{ \pm}^2\\
    &+\frac{4}{p}\left( \frac{160}{p} p_G \frac{G \mathcal{P}_{\mathcal{G}^t_C}}{(1-\delta) C} (1+\omega)c\delta +\frac{80}{p} p_G \mathcal{P}_{\mathcal{G}^t_C} (1+\omega) \frac{n}{C} \right)\frac{\mathcal{L}_{ \pm}^2}{b}\\
    &+\frac{4}{p}\left( \frac{4}{p}(1-p_G)F_{\cA}^2\alpha^2_{\lambda_{t+1}} \right),
\end{align*}
\begin{align*}
    \widehat{B} = 2 \frac{ \delta\mathcal{P}_{\mathcal{G}^t_{\widehat{C}}} }{1-\delta} B\left(\frac{12cG}{\widehat C} + p \right) + 6\widetilde{B}, \quad \widehat{D} = 2 \frac{ \delta\mathcal{P}_{\mathcal{G}^t_{\widehat{C}}} }{1-\delta}\left(\frac{6cG}{\widehat C} + p \right) + \widetilde{D},
\end{align*}
and where $\widetilde{B} \eqdef 0$ and $\widetilde{D} \eqdef 0$ when $\widehat{C} = M$, and $\widetilde{B} \eqdef \frac{\cP_{\cG_{\widehat{C}}^t}GB}{(1-\delta)\widehat{C}}$ and $\widetilde{D} \eqdef \frac{\cP_{\cG_{\widehat{C}}^t}G}{(1-\delta)\widehat{C}}$ when $\widehat{C} < M$, and 
\begin{align*}
 \mathcal{P}_{\mathcal{G}^t_C} &=   \frac{C}{Mp_G} \cdot \sum_{(1-\delta)C\leq m^\prime \leq C} \left(\left(\begin{array}{l}
G-1 \\
t-1
\end{array}\right)\left(\begin{array}{l}
M-G \\
C-m^\prime
\end{array}\right) \left(\left(\begin{array}{l}
n \\
C
\end{array}\right)\right)^{-1} \right),\\
p_G &=  \PP\left\lbrace G_C^t \geq\left(1-\delta\right) C\right\rbrace\\
&= \sum_{\lceil(1-\delta)C\rceil\leq t \leq C} \left(\left(\begin{array}{l}
G \\
t
\end{array}\right)\left(\begin{array}{l}
M-G \\
C-m^\prime
\end{array}\right) \left(\begin{array}{l}
M-1 \\
C-1
\end{array}\right)^{-1} \right).
\end{align*}
Then for all $T \geq 0$ the iterates produced by \gls{Byz-VR-MARINA} (Algorithm \ref{alg:byz_vr_marina}) satisfy
$$
\mathbb{E}\left[f\left(x^T\right)-f\left(x^{\star}\right)\right] \leq\left(1-\rho\right)^T \Phi^{(0)}+\frac{4 \widehat{D} \gamma \zeta^2}{p\rho},
$$
where $\rho = \min\left[\gamma\mu\left(1-\frac{8\widehat{B}}{p}\right), \frac{p}{8}\right]$ and $\Phi^{(0)}=$ $f\left(x^0\right)-f^{\star}+\frac{4\gamma}{p}\left\|g^0-\nabla f\left(x^0\right)\right\|^2$.
\end{theorem}
\begin{proof}
     For all $T \geq 0$ we introduce $\Phi^{(t)}=f\left(x^t\right)-f^{\star}+\frac{4\gamma}{p}\left\|g^t-\nabla f\left(x^t\right)\right\|^2$. Using the results of Lemmas \ref{lemma:final_lemma} and \ref{lemma:page}, we derive
     \begin{align*}
         \mathbb{E}\left[\Phi^{(t+1)}\right] &\stackrel{(\ref{lemma:page})}{\leq} \mathbb{E}\left[f\left(x^t\right)-f^{\star}-\left(\frac{1}{2 \gamma}-\frac{L}{2}\right)\left\|x^{t+1}-x^t\right\|^2+\frac{\gamma}{2}\left\|g^t-\nabla f\left(x^t\right)\right\|^2\right]\\
         &-\frac{\gamma}{2} \mathbb{E}\left[\left\|\nabla f\left(x^t\right)\right\|^2\right] + \frac{4\gamma}{p}   \mathbb{E}\left[\left\|g^{t+1}-\nabla f\left(x^{t+1}\right)\right\|^2\right]\\
         &\stackrel{(\ref{lemma:final_lemma})}{\leq}  \mathbb{E}\left[f\left(x^t\right)-f^{\star}-\left(\frac{1}{2 \gamma}-\frac{L}{2}\right)\left\|x^{t+1}-x^t\right\|^2+\frac{\gamma}{2}\left\|g^t-\nabla f\left(x^t\right)\right\|^2\right]\\
         &-\frac{\gamma}{2} \mathbb{E}\left[\left\|\nabla f\left(x^t\right)\right\|^2\right] + \frac{4\gamma}{p}     \left(1-\frac{p}{4}\right) \mathbb{E}\left[\left\|g^{t}-\nabla f\left(x^{t}\right)\right\|^2\right]\\
         &+ \frac{4\gamma}{p}\left( \widehat{B}\mathbb{E}\left[\left\|\nabla f\left(x^t\right)\right\|^2\right] + \widehat{D} \zeta^2+\frac{p A}{4}\|x^{t+1} - x^t\|^2\right)\\
         &=\mathbb{E}\left[f\left(x^t\right)-f^{\star}\right] + \frac{4\gamma}{p}\left(\left(1-\frac{p}{4}\right) + \frac{p}{8}\right) \mathbb{E}\left[\left\|g^{t}-\nabla f\left(x^{t}\right)\right\|^2\right]\\
         &+\frac{4 \widehat{D}\zeta^2\gamma}{p}\\
         &+\frac{1}{2\gamma}\left(1-L\gamma-2A\gamma^2\right)\mathbb{E}\left[ \|x^{t+1} - x^t\|^2 \right]\\
         &- \frac{\gamma}{2}\left(1-\frac{8\widehat{B}}{p}\right)\mathbb{E}\left[\left\|\nabla f\left(x^t\right)\right\|^2\right].
\end{align*}
Using Assumption~\ref{assm:PL} we obtain
\begin{align*}
\mathbb{E}\left[\Phi^{(t+1)}\right] &\leq\mathbb{E}\left[f\left(x^t\right)-f^{\star}\right]+\left(1-\frac{p}{8}\right) \frac{4\gamma}{p} \mathbb{E}\left[\left\|g^{t}-\nabla f\left(x^{t}\right)\right\|^2\right]+\frac{4 \widehat{D}\zeta^2\gamma}{p}\\
         &+\frac{1}{2\gamma}\left(1-L\gamma-2A\gamma^2\right)\mathbb{E}\left[ \|x^{t+1} - x^t\|^2 \right]\\ &- \gamma\mu\left(1-\frac{8\widehat{B}}{p}\right)\mathbb{E}\left[f\left(x^t\right)-f^{\star}\right].
\end{align*}
Finally, we have 
\begin{align*}
\mathbb{E}\left[\Phi^{(t+1)}\right] &\leq\left(1 - \min\left[\gamma\mu\left(1-\frac{\widehat{B}}{p}\right), \frac{p}{8}\right]\right)\mathbb{E}\left[\Phi^{(t)}\right]+\frac{4 \widehat{D}\zeta^2\gamma}{p}.
\end{align*}
Unrolling the recurrence with $\rho = \min\left[\gamma\mu\left(1-\frac{8\widehat{B}}{p}\right), \frac{p}{8}\right] $, we obtain
$$
\begin{aligned}
\mathbb{E}\left[\Phi^{(t)}\right] & \leq\left(1-\rho\right)^T \mathbb{E}\left[\Phi^{(0)}\right]+\frac{4 \widehat{D} \zeta^2 \gamma}{p} \sum_{k=0}^{K-1}\left(1-\rho\right)^t \\
& \leq\left(1-\rho\right)^T \mathbb{E}\left[\Phi^{(0)}\right]+\frac{4 \widehat{D} \zeta^2 \gamma}{p} \sum_{k=0}^{\infty}\left(1-\rho\right)^t \\
& =\left(1-\rho\right)^T \mathbb{E}\left[\Phi^{(0)}\right]+\frac{4 \widehat{D} \gamma\zeta^2}{p\rho}
\end{aligned}
$$
Taking into account $\Phi^{(t)} \geq f\left(x^t\right)-f\left(x^{\star}\right)$, we get the result.
\end{proof}

\clearpage
\section{Analysis for Bounded Compressors}

\subsection{Technical lemmas}

\begin{lemma}
\label{lemma:premainA1_Q}
Let Assumptions~\ref{assm:L-smoothness}, \ref{assm:global}, \ref{assm:local} and \ref{assm:bounded-compressor} hold and the Compression Operator satisfy Definition~\ref{def:Q}. We set $\lambda_{t+1} = D_Q \max_{m,i} L_{m,i}$.  Let us define "ideal" estimator:
\begin{equation*}
    \overline{g}^{t+1}= \begin{cases}
        \frac{1}{G^t_{\widehat{C}}}\sum \limits_{m \in \mathcal{G}^t_{\widehat{C}}} \nabla f_m(x^{t+1}),& c_n=1,\quad\quad\quad\quad\quad\quad\quad~\hspace{0.05cm} [1]\\
        g^t+\nabla f\left(x^{t+1}\right)-\nabla f\left(x^t\right),& c_n=0, \text{ } G^t_C < (1-\delta)C, \text{ }[2]\\
        g^t+\frac{1}{G^t_C} \sum \limits_{m \in \mathcal{G}^t_C}\clip_{\lambda}\left(\mathcal{Q}\left(\widehat{\Delta}_m\left(x^{t+1}, x^t\right)\right)\right),& c_n=0, \text{ } G^t_C \geq (1-\delta)C.\text{ }   [3]
    \end{cases}
\end{equation*}
Then for all $t\geq 0$ the iterates produced by \gls{Byz-VR-MARINA-PP} (Algorithm~\ref{alg:byz_vr_marina}) satisfy
\begin{align*}
   A_1  &= \mathbb{E}\left[\left\|\overline{g}^{t+1}-\nabla f\left(x^{t+1}\right)\right\|^2\right]\\
   &\leq (1-p) \mathbb{E}\left[\left\|g^{t}-\nabla f(x^{t})\right\|^2\right] + p\frac{ \delta\mathcal{P }_{\mathcal{G}^t_{\widehat{C}}} }{(1-\delta)}  \mathbb{E}\left[ B\|\nabla f(x)\|^2+\zeta^2 \right]\\
    &+ \text{\normalsize $(1-p)p_G\frac{\mathcal{P}_{\mathcal{G}^t_C} G}{C^2(1-\delta)^2} \left(  \omega L^2+(\omega+1) L_{ \pm}^2+\frac{(\omega+1) \mathcal{L}_{ \pm}^2}{b} \right)\mathbb{E}\left[\|x^{t+1} - x^t\|^2\right]$}.
    \end{align*}
where $p_G = \operatorname{Prob}\left\lbrace G^t_C \geq (1-\delta)C \right\rbrace $ and $\mathcal{P}_{\mathcal{G}^t_C} =  \operatorname{Prob}\left\lbrace m \in \mathcal{G}^t_C \mid G^t_C \geq\left(1-\delta\right) C\right\rbrace$.
\end{lemma}

\begin{proof}
Similarly to general analysis, we start from conditional expectations:
\begin{align}
    A_1 &= \mathbb{E}\left[\left\|\overline{g}^{t+1}-\nabla f\left(x^{t+1}\right)\right\|^2\right] \notag\\
&=\mathbb{E}\left[\mathbb{E}_t\left[\left\|\overline{g}^{t+1}-\nabla f\left(x^{t+1}\right)\right\|^2\right]\right] \notag\\
    & = \left(1-p\right)p_{G}\mathbb{E}\left[\mathbb{E}_t\left[\left\|g^t+\frac{1}{G^t_C} \sum \limits_{m \in \mathcal{G}^t_C}\clip_{\lambda}\left(\mathcal{Q}\left(\widehat{\Delta}_m\left(x^{t+1}, x^t\right)\right)\right) -\right.\right.\right. \notag \\
    &-\left.\left.\left. \nabla f\left(x^{t+1}\right)\right\|^2\right]\mid [3]\right] \notag\\
    & + (1-p)(1-p_{G})\mathbb{E}\left[\mathbb{E}_t\left[\left\|g^t -\nabla f(x^{t})\right\|^2\right]\mid [2]\right] \notag\\
    &+ p\mathbb{E}\left[ \left\|\frac{1}{G^t_{\widehat{C}}}\sum \limits_{m \in \cG_{\widehat{C}}^t} \nabla f_m(x^{t+1})
    - \nabla f(x^{t+1})\right\|^2 \right]. \label{eq:ncsjindisbciusdbhc}
\end{align}
Using (\ref{eq:yung-1}) and $\nabla f\left(x^{t}\right) - \nabla f\left(x^{t}\right) = 0$ we obtain
\begin{align*}
B_1 &= \mathbb{E}\left[\mathbb{E}_t\left[\left\|g^t+\frac{1}{G^t_C} \sum \limits_{m \in \mathcal{G}^t_C}\clip_{\lambda}\left(\mathcal{Q}\left(\widehat{\Delta}_m\left(x^{t+1}, x^t\right)\right)\right) -\nabla f\left(x^{t+1}\right)\right\|^2\right]\mid [3]\right]\\
&\text{\scriptsize$ =\mathbb{E}\left[\mathbb{E}_t\left[\left\|g^t+\frac{1}{G^t_C} \sum \limits_{m \in \mathcal{G}^t_C}\clip_{\lambda}\left(\mathcal{Q}\left(\widehat{\Delta}_m\left(x^{t+1}, x^t\right)\right)\right) -\nabla f\left(x^{t+1}\right)+\nabla f\left(x^{t}\right) - \nabla f\left(x^{t}\right)\right\|^2\right]\mid [3]\right]$}
\end{align*}
Using $\lambda_{t+1} = D_Q\max_{m,i} L_{m,i} \|x^{t+1} - x^t\|$ we can guarantee that clipping operator becomes identical since we have 
\begin{align*}
   \left \|\mathcal{Q}\left(\widehat{\Delta}_m\left(x^{t+1}, x^t\right)\right)\right\| &\leq D_Q\left\|\widehat{\Delta}_m\left(x^{t+1}, x^t\right)\right\|\\
   &\leq D_Q \left\|\frac{1}{b} \sum_{j\in m} \nabla f_{m,i}(x^{t+1})-\nabla f_{m,i}(x^t) \right\|\\
    &\leq D_Q \frac{1}{b}\sum_{j\in m}\left\|  \nabla f_{m,i}(x^{t+1})-\nabla f_{m,i}(x^t) \right\|\\
    &\leq D_Q \max_j L_{m,i}\left\|  x^{t+1}-x^t \right\|\\
    &\leq D_Q \max_{m,i} L_{m,i}\left\|  x^{t+1}-x^t \right\|.
\end{align*}
Therefore, we can continue as follows
\begin{align*}
B_1 &= \mathbb{E}\left[\mathbb{E}_t\left[\left\|g^t+\frac{1}{G^t_C} \sum \limits_{m \in \mathcal{G}^t_C}\mathcal{Q}\left(\widehat{\Delta}_m\left(x^{t+1}, x^t\right)\right) -\nabla f\left(x^{t+1}\right)\right\|^2\right]\mid [3]\right]\\
&\text{\scriptsize$ =\mathbb{E}\left[\mathbb{E}_t\left[\left\|g^t+\frac{1}{G^t_C} \sum \limits_{m \in \mathcal{G}^t_C}\mathcal{Q}\left(\widehat{\Delta}_m\left(x^{t+1}, x^t\right)\right)-\nabla f\left(x^{t+1}\right)+\nabla f\left(x^{t}\right) - \nabla f\left(x^{t}\right)\right\|^2\right]\mid [3]\right]$}.
\end{align*}
Moreover, we can avoid application of Young's inequality and use variance decomposition instead:
 \begin{align}  
     B_1 &\leq \mathbb{E}\left[\left\|g^{t} -\nabla f\left(x^{t}\right)\right\|^2\right] \notag \\
    &+ \text{\scriptsize$ \mathbb{E}\left[\mathbb{E}_t\left[\left\|\frac{1}{G^t_C} \sum \limits_{m \in \mathcal{G}^t_C}\mathcal{Q}\left(\widehat{\Delta}_m\left(x^{t+1}, x^t\right)\right)  - \left( \nabla f(x^{t+1}) - \nabla f(x^{t}) \right)\right\|^2\right]\mid [3]\right]$} \notag\\
    &\leq \mathbb{E}\left[\left\|g^{t}-\nabla f(x^{t})\right\|^2\right] \notag \\
    &+ \mathbb{E}\left[ \mathbb{E}_t\left[\left\|\frac{1}{G^t_C} \sum \limits_{m \in \mathcal{G}^t_C}\mathcal{Q}\left(\widehat{\Delta}_m\left(x^{t+1}, x^t\right)\right) - \Delta\left(x^{t+1}, x^t\right)\right\|^2\right]\mid [3]\right]. \label{eq:djbfjnskcnksbfjvbsd}
\end{align}
Let us consider the last part of the inequality. Note that $G^t_C \geq (1-\delta)C$ in this case and
\begin{align}\label{eq:indicator_1}
\notag  B^\prime_1 &= \mathbb{E}\left[ \mathbb{E}_t\left[\left\|\frac{1}{G^t_C} \sum \limits_{m \in \mathcal{G}^t_C}\mathcal{Q}\left(\widehat{\Delta}_m\left(x^{t+1}, x^t\right)\right)  - \Delta\left(x^{t+1}, x^t\right)\right\|^2\right]\mid [3] \right]\\
 \notag &= \mathbb{E}\left[\mathbb{E}_{\set}\left[\mathbb{E}_t\left[\left\|\frac{1}{G^t_C} \sum \limits_{m \in \mathcal{G}^t_C}\mathcal{Q}\left(\widehat{\Delta}_m\left(x^{t+1}, x^t\right)\right) - \Delta\left(x^{t+1}, x^t\right)\right\|^2\right]\mid [3]\right]\right]\\
\notag   &\leq \text{\small $\frac{1}{C^2(1-\delta)^2}\mathbb{E}\left[\mathbb{E}_{\set}\left[ \sum \limits_{m \in \mathcal{G}^t_C}\mathbb{E}_t\left[\left\|\mathcal{Q}\left(\widehat{\Delta}_m\left(x^{t+1}, x^t\right)\right) - \Delta\left(x^{t+1}, x^t\right)\right\|^2\right]\mid [3]\right]\right]$}\\
 \notag   &\leq \text{\small $\frac{1}{C^2(1-\delta)^2}\mathbb{E}\left[\sum \limits_{m \in \mathcal{G}}\mathbb{E}_{\set}\left[\mathcal{I}_{\mathcal{G}^t_C}\right]\mathbb{E}_t\left[\left\|\mathcal{Q}\left(\widehat{\Delta}_m\left(x^{t+1}, x^t\right)\right)  - \Delta\left(x^{t+1}, x^t\right)\right\|^2\right]\mid [3] \right]$}\\
        &\text{\small$= \frac{1}{C^2(1-\delta)^2}\mathbb{E}\left[\sum \limits_{m \in \mathcal{G}}\mathcal{P}_{\mathcal{G}^t_C}\cdot\mathbb{E}_t\left[\left\|\mathcal{Q}\left(\widehat{\Delta}_m\left(x^{t+1}, x^t\right)\right) - \Delta\left(x^{t+1}, x^t\right)\right\|^2\right]\mid [3]\right]$},
\end{align}
where $\mathcal{I}_{\mathcal{G}^t_C}$ is an indicator function for the event $\left\lbrace m \in \mathcal{G}^t_C \mid G_C^t \geq\left(1-\delta\right) C\right\rbrace$ and $\mathcal{P}_{\mathcal{G}^t_C} =  \operatorname{Prob}\left\lbrace m \in \mathcal{G}^t_C \mid G_C^t \geq\left(1-\delta\right) C\right\rbrace$ is the probability of such an event. Note that 
$\mathbb{E}_{\set}\left[\mathcal{I}_{\mathcal{G}^t_C}\right] = \mathcal{P}_{\mathcal{G}^t_C}$. In the case of uniform sampling of clients, we have 
\begin{align*}
    \forall m \in \mathcal{G} \quad &\mathcal{P}_{\mathcal{G}^t_C}=\operatorname{Prob}\left\lbrace m \in \mathcal{G}^t_C \mid G^t_C\geq\left(1-\delta\right) C\right\rbrace\\
    & = \frac{C}{n} \frac{1}{p_G}\cdot \sum_{(1-\delta)C\leq m^\prime \leq C} \left(\left(\begin{array}{l}
G-1 \\
t-1
\end{array}\right)\left(\begin{array}{l}
M-G \\
C-m^\prime
\end{array}\right) \left(\left(\begin{array}{l}
M-1 \\
C-1
\end{array}\right)\right)^{-1} \right). 
\end{align*}
Now, we can continue with inequalities:
\begin{align*}
B_1^\prime &\leq  \text{\small $       \frac{\mathcal{P}_{\mathcal{G}^t_C}}{C^2(1-\delta)^2} \mathbb{E}\left[ \sum \limits_{m \in \mathcal{G}}\mathbb{E}_t\left[\left\|\mathcal{Q}\left(\widehat{\Delta}_m\left(x^{t+1}, x^t\right)\right)  - \Delta\left(x^{t+1}, x^t\right)\right\|^2\right]\mid [3]\right]$}\\
 &\leq     \text{\small $      \frac{\mathcal{P}_{\mathcal{G}^t_C}}{C^2(1-\delta)^2}\mathbb{E}\left[\sum \limits_{m \in \mathcal{G}}\mathbb{E}_t\left[\mathbb{E}_{Q}\left[\left\|\mathcal{Q}\left(\widehat{\Delta}_m\left(x^{t+1}, x^t\right)\right)  - \Delta\left(x^{t+1}, x^t\right)\right\|^2\right] \right]\mid [3]  \right]$}\\
 &\leq     \text{\small $      \frac{\mathcal{P}_{\mathcal{G}^t_C}}{C^2(1-\delta)^2}\mathbb{E}\left[\sum \limits_{m \in \mathcal{G}}\mathbb{E}_t\left[\mathbb{E}_{Q}\left[\left\|\mathcal{Q}\left(\widehat{\Delta}_m\left(x^{t+1}, x^t\right)\right)  - \Delta_m\left(x^{t+1}, x^t\right)\right\|^2\right]\right]\mid [3]\right]$}\\ 
& +   \frac{\mathcal{P}_{\mathcal{G}^t_C}}{C^2(1-\delta)^2}\mathbb{E}\left[\sum \limits_{m \in \mathcal{G}}\mathbb{E}_t\left[\left\|  \Delta_m\left(x^{t+1}, x^t\right) - \Delta\left(x^{t+1}, x^t\right)\right\|^2\right]\mid [3]\right].
\end{align*}
Using variance decomposition, we have 
\begin{align*}
B_1^\prime &\leq \frac{\mathcal{P}_{\mathcal{G}^t_C}}{C^2(1-\delta)^2} \mathbb{E}\left[ \sum \limits_{m \in \mathcal{G}}\mathbb{E}_t\left[\mathbb{E}_{Q}\left[\left\|\mathcal{Q}\left(\widehat{\Delta}_m\left(x^{t+1}, x^t\right)\right)\right\|^2\right]\right] -\right.\\
&\left.-\sum \limits_{m \in \mathcal{G}}\left\|\Delta_m\left(x^{t+1}, x^t\right)\right\|^2\mid [3]\right]\\
&+\frac{\mathcal{P}_{\mathcal{G}^t_C}}{C^2(1-\delta)^2}\mathbb{E}\left[\sum \limits_{m \in \mathcal{G}}\mathbb{E}_t\left[\left\|  \Delta_m\left(x^{t+1}, x^t\right) -  \Delta\left(x^{t+1}, x^t\right)\right\|^2\right]\mid [3]\right].
\end{align*}

Applying the definition of unbiased compressor, we get
\begin{align*}
B_1^\prime&\leq  \frac{\mathcal{P}_{\mathcal{G}^t_C}}{C^2(1-\delta)^2} \mathbb{E}\left[ \sum \limits_{m \in \mathcal{G}}(1+\omega)\mathbb{E}_t\left\|\widehat{\Delta}_m\left(x^{t+1}, x^t\right)\right\|^2-\right.\\
&\left.- \sum \limits_{m \in \mathcal{G}}\left\|\Delta_m\left(x^{t+1}, x^t\right)\right\|^2\mid [3]\right]\\
&+\frac{\mathcal{P}_{\mathcal{G}^t_C}}{C^2(1-\delta)^2}\mathbb{E}\left[\sum \limits_{m \in \mathcal{G}}\left\|  \Delta_m\left(x^{t+1}, x^t\right) -  \Delta\left(x^{t+1}, x^t\right)\right\|^2\mid [3]\right]\\
&\leq  \frac{\mathcal{P}_{\mathcal{G}^t_C}}{C^2(1-\delta)^2} \mathbb{E} \left[ \sum \limits_{m \in \mathcal{G}}(1+\omega)\mathbb{E}_t\left\|\widehat{\Delta}_m\left(x^{t+1}, x^t\right) - \Delta_m\left(x^{t+1}, x^t\right) \right\|^2 \right]\\
&+ \frac{\mathcal{P}_{\mathcal{G}^t_C}}{C^2(1-\delta)^2} \mathbb{E} \left[ \sum \limits_{m \in \mathcal{G}}(1+\omega)\mathbb{E}_t\left\|\Delta_m\left(x^{t+1}, x^t\right)\right\|^2 - \right.\\
&-\left.\sum \limits_{m \in \mathcal{G}}\mathbb{E}_t\left\|\Delta_m\left(x^{t+1}, x^t\right)\right\|^2\mid [3]\right]\\
&+\frac{\mathcal{P}_{\mathcal{G}^t_C}}{C^2(1-\delta)^2}\mathbb{E}\left[\sum \limits_{m \in \mathcal{G}}\left\|  \Delta_m\left(x^{t+1}, x^t\right) -  \Delta\left(x^{t+1}, x^t\right)\right\|^2\mid [3]\right].
\end{align*}
Next, we rearrange terms and derive 
\begin{align*}
    B_1^\prime &\leq  \frac{\mathcal{P}_{\mathcal{G}^t_C}}{C^2(1-\delta)^2} (1+\omega) \mathbb{E} \left[ \sum \limits_{m \in \mathcal{G}}\mathbb{E}_t\left[\left\|\widehat{\Delta}_m\left(x^{t+1}, x^t\right) - \Delta_m\left(x^{t+1}, x^t\right) \right\|^2\right] \mid [3] \right]\\
&+ \frac{\mathcal{P}_{\mathcal{G}^t_C}}{C^2(1-\delta)^2}\omega \mathbb{E} \left[ \sum \limits_{m \in \mathcal{G}}\left\|\Delta_m\left(x^{t+1}, x^t\right)\right\|^2\mid [3]\right]\\
&+\frac{\mathcal{P}_{\mathcal{G}^t_C}}{C^2(1-\delta)^2}\mathbb{E} \left[ \sum \limits_{m \in \mathcal{G}}\left\|  \Delta_m\left(x^{t+1}, x^t\right) -  \Delta\left(x^{t+1}, x^t\right)\right\|^2\mid [3]\right]\\
 &= \frac{\mathcal{P}_{\mathcal{G}^t_C}}{C^2(1-\delta)^2} (1+\omega) \mathbb{E} \left[ \sum \limits_{m \in \mathcal{G}}\mathbb{E}_t\left[\left\|\widehat{\Delta}_m\left(x^{t+1}, x^t\right) - \Delta_m\left(x^{t+1}, x^t\right) \right\|^2\right] \mid [3] \right] \\
&+ \frac{\mathcal{P}_{\mathcal{G}^t_C}}{C^2(1-\delta)^2}\omega \mathbb{E} \left[  \sum \limits_{m \in \mathcal{G}}\left\|\Delta_m\left(x^{t+1}, x^t\right) - \Delta\left(x^{t+1}, x^t\right)\right\|^2 +\right.\\ &\left.+\|\Delta\left(x^{t+1}, x^t\right)\|^2\mid [3]\right]\\
&+\frac{\mathcal{P}_{\mathcal{G}^t_C}}{C^2(1-\delta)^2} \mathbb{E} \left[ \sum \limits_{m \in \mathcal{G}}\left\|  \Delta_m\left(x^{t+1}, x^t\right) -  \Delta\left(x^{t+1}, x^t\right)\right\|^2 \mid [3] \right].
\end{align*}
Rearranging terms leads to 
\begin{align*}
   B_1^\prime  &\leq \frac{\mathcal{P}_{\mathcal{G}^t_C}}{C^2(1-\delta)^2} (1+\omega) \mathbb{E} \left[  \sum \limits_{m \in \mathcal{G}}\mathbb{E}_t\left[\left\|\widehat{\Delta}_m\left(x^{t+1}, x^t\right) - \Delta_m\left(x^{t+1}, x^t\right) \right\|^2\right] \mid [3] \right] \\
&+ \frac{\mathcal{P}_{\mathcal{G}^t_C}}{C^2(1-\delta)^2}(\omega+1) \mathbb{E}\left[ \sum \limits_{m \in \mathcal{G}}\left\|\Delta_m\left(x^{t+1}, x^t\right) - \Delta\left(x^{t+1}, x^t\right)\right\|^2\mid[3]\right] \\
&+\frac{\mathcal{P}_{\mathcal{G}^t_C}}{C^2(1-\delta)^2} \omega \mathbb{E}\left[\sum \limits_{m \in \mathcal{G}}\left\|  \Delta\left(x^{t+1}, x^t\right)\right\|^2 \mid [3]\right].
\end{align*}
Now we apply Assumptions \ref{assm:L-smoothness}, \ref{assm:global}, \ref{assm:local}:
\begin{align*}
   B_1^\prime  &\leq \frac{\mathcal{P}_{\mathcal{G}^t_C}}{C^2(1-\delta)^2} (1+\omega) \mathbb{E} \left[ G \frac{\mathcal{L}_{ \pm}^2}{b} \|x^{t+1} - x^t\|^2\right]\\
&+ \frac{\mathcal{P}_{\mathcal{G}^t_C}}{C^2(1-\delta)^2}(\omega+1) \mathbb{E} \left[ G L_{ \pm}^2 \|x^{t+1} - x^t\|^2\right]\\
&+\frac{\mathcal{P}_{\mathcal{G}^t_C}}{C^2(1-\delta)^2} \omega \mathbb{E} \left[ G L^2\left\|  x^{t+1} - x^t\right\|^2\right].
\end{align*}
Finally, we have
\begin{align*}
B_1^\prime&\leq \frac{\mathcal{P}_{\mathcal{G}^t_C}\cdot G}{C^2(1-\delta)^2} \left(  \omega L^2+(\omega+1) L_{ \pm}^2+\frac{(\omega+1) \mathcal{L}_{ \pm}^2}{b} \right)\mathbb{E}\left[ \|x^{t+1} - x^t\|^2\right].
\end{align*}
Let us plug the obtained results in \eqref{eq:djbfjnskcnksbfjvbsd}:
\begin{align*}
  B_1      &\leq \mathbb{E}\left[\left\|g^{t}-\nabla f(x^{t})\right\|^2\right]\\
    &+ \frac{\mathcal{P}_{\mathcal{G}^t_C}\cdot G}{C^2(1-\delta)^2} \left(  \omega L^2+(\omega+1) L_{ \pm}^2+\frac{(\omega+1) \mathcal{L}_{ \pm}^2}{b} \right)\mathbb{E}\left[\|x^{t+1} - x^t\|^2\right].
\end{align*}
Also, we have 
\begin{align*}
    A_1 &= \mathbb{E}\left[\left\|\overline{g}^{t+1}-\nabla f(x^{t+1})\right\|^2\right]\\
    &\overset{\eqref{eq:ncsjindisbciusdbhc}, \eqref{eq:nsjknbvsbicusd}}{\leq} (1-p)p_G B_1 + (1-p)(1-p_{G})\mathbb{E}\left[\left\|g^t -\nabla f(x^{t})\right\|^2\right] \\
    &+ p\frac{ \delta\cdot\mathcal{P }_{\mathcal{G}^t_{\widehat{C}}} }{(1-\delta)}  \mathbb{E}\left[ B\|\nabla f(x)\|^2+\zeta^2 \right]\\
    &\leq (1-p)p_G \mathbb{E}\left[\left\|g^{t}-\nabla f(x^{t})\right\|^2\right] + p\frac{ \delta\cdot\mathcal{P }_{\mathcal{G}^t_{\widehat{C}}} }{(1-\delta)}  \mathbb{E}\left[ B\|\nabla f(x)\|^2+\zeta^2 \right]\\
    &+ \text{\small $(1-p)p_G\frac{\mathcal{P}_{\mathcal{G}^t_C}\cdot G}{C^2(1-\delta)^2} \left(  \omega L^2+(\omega+1) L_{ \pm}^2+\frac{(\omega+1) \mathcal{L}_{ \pm}^2}{b} \right)\mathbb{E}\left[\|x^{t+1} - x^t\|^2\right]$}\\
    &+(1-p)(1-p_{G})\mathbb{E}\left[\left\|g^t -\nabla f(x^{t})\right\|^2\right].
\end{align*}
Rearranging the terms, we get
\begin{align*}
   A_1  &\leq (1-p) \mathbb{E}\left[\left\|g^{t}-\nabla f(x^{t})\right\|^2\right] + \frac{ \delta\mathcal{P }_{\mathcal{G}^t_{\widehat{C}}} }{(1-\delta)}  \mathbb{E}\left[ B\|\nabla f(x)\|^2+\zeta^2 \right]\\
    &+ \text{\normalsize $(1-p)p_G\frac{\mathcal{P}_{\mathcal{G}^t_C}G}{C^2(1-\delta)^2} \left(  \omega L^2+(\omega+1) L_{ \pm}^2+\frac{(\omega+1) \mathcal{L}_{ \pm}^2}{b} \right)\mathbb{E}\left[\|x^{t+1} - x^t\|^2\right]$}.
    \end{align*}

\end{proof}

\begin{lemma}
\label{lemma:good_aggr_new}
Let Assumptions~\ref{assm:L-smoothness}, \ref{assm:global}, \ref{assm:local}, \ref{assm:local_smooth_all}, \ref{assm:bounded-compressor} hold and the compression operator satisfy Definition~\ref{def:Q}. Also, let us introduce the notation
\begin{align*}
\texttt{ARAgg}_Q^{t+1} = \texttt{ARAgg}\left(\clip_{\lambda_{t+1}}\left(\cQ\left(\widehat{\Delta}_1(x^{t+1}, x^t)\right)\right),\right.\\
\left.\ldots, \clip_{\lambda_{t+1}}\left(\cQ\left(\widehat{\Delta}_C(x^{t+1}, x^t)\right)\right)\right).
\end{align*}
Then for all $t\geq 0$ the iterates produced by \gls{Byz-VR-MARINA-PP} (Algorithm~\ref{alg:byz_vr_marina}) satisfy

\begin{align*}
       T_2 &=   \mathbb{E}\left[ \mathbb{E}_t\left[\left\| \frac{1}{G^t_C} \sum \limits_{m \in \mathcal{G}^t_C}\clip_{\lambda}\left(\mathcal{Q}\left(\widehat{\Delta}_m\left(x^{t+1}, x^t\right)\right)\right)  -   \texttt{ARAgg}_Q^{t+1}\right\|^2\mid [3]\right]\right]\\
       &\leq 4\frac{G\mathcal{P}_{\mathcal{G}^t_C}}{C(1-\delta)} c\delta \left( (1+\omega)\frac{\mathcal{L}_{ \pm}^2}{b} + (\omega+1) L_{ \pm}^2 + \omega  L^2 \right) \mathbb{E}\left[\|x^{t+1} - x^t\|^2\right],
\end{align*}
where $\mathcal{P}_{\mathcal{G}^t_C} =  \operatorname{Prob}\left\lbrace m \in \mathcal{G}^t_C \mid G^t_C \geq\left(1-\delta\right) C\right\rbrace$.
\end{lemma}
\begin{proof}
By definition of the robust aggregation, we have 
\begin{align*}
   T_2 &=   \mathbb{E}\left[ \mathbb{E}_t\left[\left\| \frac{1}{G^t_C} \sum \limits_{m \in \mathcal{G}^t_C}\clip_{\lambda}\left(\mathcal{Q}\left(\widehat{\Delta}_m\left(x^{t+1}, x^t\right)\right)\right)  -   \texttt{ARAgg}_Q^{t+1}\right\|^2\mid [3]\right]\right]\\
   &\text{ \scriptsize $\leq  \mathbb{E}\left[  \frac{c \delta}{D_2} \sum_{\substack{m, l \in \mathcal{G}_C^t \\
m \neq l}}    
\mathbb{E}_t\left[\left\| \clip_{\lambda}\left(\mathcal{Q}\left(\widehat{\Delta}_m\left(x^{t+1}, x^t\right)\right)\right) - \clip_{\lambda}\left(\mathcal{Q}\left(\widehat{\Delta}_l\left(x^{t+1}, x^t\right)\right)\right) \right\|^2\mid [3]\right]\right]$},
\end{align*}
where $D_2 = G^t_C(G^t_C-1)$.

Using $\lambda_{t+1} = D_Q\max_{m,i} L_{m,i} \|x^{t+1} - x^t\|$ we can guarantee that clipping operator becomes identical since we have $\forall m\in \cG$ 
\begin{align}
   \left \|\mathcal{Q}\left(\widehat{\Delta}_m\left(x^{t+1}, x^t\right)\right)\right\| &\leq D_Q\left\|\widehat{\Delta}_m\left(x^{t+1}, x^t\right)\right\| \notag\\
   &\leq D_Q \left\|\frac{1}{b} \sum_{j\in m} \nabla f_{m,i}(x^{t+1})-\nabla f_{m,i}(x^t) \right\| \notag\\
    &\leq D_Q \frac{1}{b}\sum_{j\in m}\left\|  \nabla f_{m,i}(x^{t+1})-\nabla f_{m,i}(x^t) \right\| \notag\\
    &\leq D_Q \max_j L_{m,i}\left\|  x^{t+1}-x^t \right\|. \label{eq:clipping_turned_off}
\end{align}

Let us consider pair-wise differences: $\forall i,l \in \cG$
\begin{align*}
T_{2}^\prime(i,l) &= \mathbb{E}_t\left[\left\| \clip_{\lambda}\left(\mathcal{Q}\left(\widehat{\Delta}_m\left(x^{t+1}, x^t\right)\right)\right) - \clip_{\lambda}\left(\mathcal{Q}\left(\widehat{\Delta}_l\left(x^{t+1}, x^t\right)\right)\right) \right\|^2\mid [3]\right]\\
&= \mathbb{E}_t\left[\left\| \mathcal{Q}\left(\widehat{\Delta}_m\left(x^{t+1}, x^t\right)\right) - \mathcal{Q}\left(\widehat{\Delta}_l\left(x^{t+1}, x^t\right)\right) \right\|^2\mid [3]\right]\\
& =   \mathbb{E}_t\left[\left\| \mathcal{Q}\left(\widehat{\Delta}_m\left(x^{t+1}, x^t\right)\right)  - \Delta_m\left(x^{t+1}, x^t\right) + \Delta_l\left(x^{t+1}, x^t\right)-\right.\right.\\
&\left.\left.-\mathcal{Q}\left(\widehat{\Delta}_l\left(x^{t+1}, x^t\right)\right) \right\|^2\mid [3]\right]\\
&+   \mathbb{E}_t\left[\left\|  \Delta_m\left(x^{t+1}, x^t\right) - \Delta_l\left(x^{t+1}, x^t\right) \right\|^2\mid [3]\right]\\
& \stackrel{(\ref{eq:yung-1})}{\leq}  2 \mathbb{E}_t\left[\left\| \mathcal{Q}\left(\widehat{\Delta}_m\left(x^{t+1}, x^t\right)\right) -\Delta_m\left(x^{t+1}, x^t\right) \right\|^2\mid [3]\right]\\
&+ 2 \mathbb{E}_t\left[\left\| \Delta_l\left(x^{t+1}, x^t\right) - \mathcal{Q}\left(\widehat{\Delta}_l\left(x^{t+1}, x^t\right)\right) \right\|^2\mid [3]\right]\\
&+   \mathbb{E}_t\left[\left\|  \Delta_l\left(x^{t+1}, x^t\right) - \Delta_m\left(x^{t+1}, x^t\right) \right\|^2\mid [3]\right]]\\
& \stackrel{(\ref{eq:yung-1})}{\leq}  2 \mathbb{E}_t\left[\left\| \mathcal{Q}\left(\widehat{\Delta}_m\left(x^{t+1}, x^t\right)\right) -\Delta_m\left(x^{t+1}, x^t\right)  \right\|^2\mid [3]\right]\\
&+ 2  \mathbb{E}_t\left[\left\| \Delta_l\left(x^{t+1}, x^t\right) - \mathcal{Q}\left(\widehat{\Delta}_l\left(x^{t+1}, x^t\right)\right) \right\|^2\mid [3]\right]\\
&+ 2  \mathbb{E}_t\left[\left\|  \Delta_l\left(x^{t+1}, x^t\right) - \Delta\left(x^{t+1}, x^t\right) \right\|^2 +\right.\\
&\left.+\left\|  \Delta_m\left(x^{t+1}, x^t\right) - \Delta\left(x^{t+1}, x^t\right)\right\|^2\mid [3]\right].
\end{align*}
Now we can combine all the parts together:
\begin{align*}
    \widehat{T}_2 & = \mathbb{E}\left[\frac{1}{G^t_C(G^t_C - 1)}  \sum_{\substack{m, l \in \mathcal{G}_C^t \\
m \neq l}} T_{2}^\prime(i,l) \right]  \\
& \leq   \mathbb{E}\left[ \frac{1}{D_2} \sum_{\substack{m, l \in \mathcal{G}_C^t \\
m \neq l}} 2  \mathbb{E}_t\left[\left\| \mathcal{Q}\left(\widehat{\Delta}_m\left(x^{t+1}, x^t\right)\right) -\Delta_m\left(x^{t+1}, x^t\right)  \right\|^2\mid [3]\right]\right]\\
&+  \mathbb{E}\left[  \frac{1}{D_2}  \sum_{\substack{m, l \in \mathcal{G}_C^t \\
m \neq l}} 2 \mathbb{E}_t\left[\left\| \Delta_l\left(x^{t+1}, x^t\right) - \mathcal{Q}\left(\widehat{\Delta}_l\left(x^{t+1}, x^t\right)\right) \right\|^2\mid [3]\right]\right]\\
&+    \mathbb{E}\left[  \frac{1}{D_2}  \sum_{\substack{m, l \in \mathcal{G}_C^t \\
m \neq l}} 2  \mathbb{E}_t\left[\left\|  \Delta_l\left(x^{t+1}, x^t\right) - \Delta\left(x^{t+1}, x^t\right) \right\|^2\mid [3]\right]\right]\\
&+   \mathbb{E}\left[  \frac{1}{D_2}  \sum_{\substack{m, l \in \mathcal{G}_C^t \\
m \neq l}} 2 \mathbb{E}_t\left[\left\|  \Delta_m\left(x^{t+1}, x^t\right) - \Delta\left(x^{t+1}, x^t\right)\right\|^2\mid [3]\right]\right].
\end{align*}
Rearranging the terms, we get
\begin{align*}
    \widehat{T}_2& \leq   \mathbb{E}\left[ \frac{4}{G_C^t} \sum_{m \in \mathcal{G}_C^t}   \mathbb{E}_t\left[\left\| \mathcal{Q}\left(\widehat{\Delta}_m\left(x^{t+1}, x^t\right)\right) -\Delta_m\left(x^{t+1}, x^t\right)  \right\|^2\mid [3]\right]\right]\\
&+    \mathbb{E}\left[  \frac{4}{G_C^t} \sum_{m \in \mathcal{G}_C^t} \mathbb{E}_t\left[\left\|  \Delta_m\left(x^{t+1}, x^t\right) - \Delta\left(x^{t+1}, x^t\right) \right\|^2\mid [3]\right]\right].
\end{align*}
Using variance decomposition, we get
\begin{align*}
        \widehat{T}_2&\leq \mathbb{E}\left[ \frac{1}{G^t_C} \sum_{m \in \mathcal{G}^t_C} 4  \mathbb{E}_t\left[\left\| \mathcal{Q}\left(\widehat{\Delta}_m\left(x^{t+1}, x^t\right)\right) \right\|^2\mid [3] \right]\right]\\
        &-\mathbb{E}\left[ \frac{1}{G^t_C} \sum_{m \in \mathcal{G}^t_C} 4  \mathbb{E}_t\left[\left\| \Delta_m\left(x^{t+1}, x^t\right) \right\|^2\mid [3]\right] \right]\\
&+\mathbb{E}\left[ \frac{1}{G^t_C} \sum_{m \in \mathcal{G}^t_C} 4 \mathbb{E}_t\left[\left\|  \Delta_m\left(x^{t+1}, x^t\right) - \Delta\left(x^{t+1}, x^t\right) \right\|^2\mid [3]\right] \right].
\end{align*}    
Using the properties of unbiased compressors, we obtain 
\begin{align*}
        \widehat{T}_2&\leq \mathbb{E}\left[ \frac{1}{G^t_C} \sum_{m \in \mathcal{G}^t_C} 4(1+\omega)  \mathbb{E}_t\left[\left\| \widehat{\Delta}_m\left(x^{t+1}, x^t\right) \right\|^2\mid [3] \right]\right]\\
        &-\mathbb{E}\left[ \frac{1}{G^t_C} \sum_{m \in \mathcal{G}^t_C} 4  \mathbb{E}_t\left[\left\| \Delta_m\left(x^{t+1}, x^t\right) \right\|^2\mid [3]\right] \right]\\
&+\mathbb{E}\left[ \frac{1}{G^t_C} \sum_{m \in \mathcal{G}^t_C} 4 \mathbb{E}_t\left[\left\|  \Delta_m\left(x^{t+1}, x^t\right) - \Delta\left(x^{t+1}, x^t\right) \right\|^2\mid [3]\right] \right]\\
&\leq \mathbb{E}\left[ \frac{1}{G^t_C} \sum_{m \in \mathcal{G}^t_C} 4(1+\omega)  \mathbb{E}_t\left[\left\| \widehat{\Delta}_m\left(x^{t+1}, x^t\right) - \Delta_m\left(x^{t+1}, x^t\right)\right\|^2\mid [3] \right]\right]\\
        &+\mathbb{E}\left[ \frac{1}{G^t_C} \sum_{m \in \mathcal{G}^t_C} 4(1+\omega)  \mathbb{E}_t\left[\left\| \Delta_m\left(x^{t+1}, x^t\right) \right\|^2\mid [3]\right] \right]\\
        &-\mathbb{E}\left[ \frac{1}{G^t_C} \sum_{m \in \mathcal{G}^t_C} 4  \mathbb{E}_t\left[\left\| \Delta_m\left(x^{t+1}, x^t\right) \right\|^2\mid [3]\right] \right]\\
&+\mathbb{E}\left[ \frac{1}{G^t_C} \sum_{m \in \mathcal{G}^t_C} 4 \mathbb{E}_t\left[\left\|  \Delta_m\left(x^{t+1}, x^t\right) - \Delta\left(x^{t+1}, x^t\right) \right\|^2\mid [3]\right] \right].
\end{align*} 
Let us simplify the inequality:
\begin{align*}
    \widehat{T}_2 &\leq \mathbb{E}\left[ \frac{1}{G^t_C} \sum_{m \in \mathcal{G}^t_C} 4(1+\omega)  \mathbb{E}_t\left[\left\| \widehat{\Delta}_m\left(x^{t+1}, x^t\right) - \Delta_m\left(x^{t+1}, x^t\right)\right\|^2\mid [3] \right]\right]\\
        &+\mathbb{E}\left[ \frac{1}{G^t_C} \sum_{m \in \mathcal{G}^t_C} 4\omega  \mathbb{E}_t\left[\left\| \Delta_m\left(x^{t+1}, x^t\right) \right\|^2\mid [3]\right] \right]\\
&+\mathbb{E}\left[ \frac{1}{G^t_C} \sum_{m \in \mathcal{G}^t_C} 4 \mathbb{E}_t\left[\left\|  \Delta_m\left(x^{t+1}, x^t\right) - \Delta\left(x^{t+1}, x^t\right) \right\|^2\mid [3]\right] \right].
\end{align*}
Using variance decomposition once again, we get
\begin{align*}
    \widehat{T}_2 &\leq \mathbb{E}\left[ \frac{1}{G^t_C} \sum_{m \in \mathcal{G}^t_C} 4(1+\omega)  \mathbb{E}_t\left[\left\| \widehat{\Delta}_m\left(x^{t+1}, x^t\right) - \Delta_m\left(x^{t+1}, x^t\right)\right\|^2\mid [3] \right]\right]\\
        &+\mathbb{E}\left[ \frac{1}{G^t_C} \sum_{m \in \mathcal{G}^t_C} 4\omega  \mathbb{E}_t\left[\left\| \Delta_m\left(x^{t+1}, x^t\right) - \Delta\left(x^{t+1}, x^t\right)  \right\|^2\mid [3]\right] \right]\\
&+\mathbb{E}\left[ \frac{1}{G^t_C} \sum_{m \in \mathcal{G}^t_C} 4 \mathbb{E}_t\left[\left\|  \Delta_m\left(x^{t+1}, x^t\right) - \Delta\left(x^{t+1}, x^t\right) \right\|^2\mid [3]\right] \right]\\
&+ \mathbb{E}\left[ \frac{1}{G^t_C} \sum_{m \in \mathcal{G}^t_C} 4\omega  \mathbb{E}_t\left[\left\| \Delta\left(x^{t+1}, x^t\right)  \right\|^2\mid [3]\right] \right].
\end{align*}
Then, we apply similar arguments to the ones used in deriving \eqref{eq:indicator_1}:
\begin{align*}
    \widehat{T}_2 &\leq \mathbb{E}\left[ \frac{\mathcal{P}_{\mathcal{G}^t_C}}{C(1-\delta)} \sum_{m \in \mathcal{G}} 4(1+\omega)  \mathbb{E}_t\left[\left\| \widehat{\Delta}_m\left(x^{t+1}, x^t\right) - \Delta_m\left(x^{t+1}, x^t\right)\right\|^2\mid [3] \right]\right]\\
        &+\mathbb{E}\left[ \frac{\mathcal{P}_{\mathcal{G}^t_C}}{C(1-\delta)} \sum_{m \in \mathcal{G}} 4\omega  \mathbb{E}_t\left[\left\| \Delta_m\left(x^{t+1}, x^t\right) - \Delta\left(x^{t+1}, x^t\right)  \right\|^2\mid [3]\right] \right]\\
&+\mathbb{E}\left[ \frac{\mathcal{P}_{\mathcal{G}^t_C}}{C(1-\delta)} \sum_{m \in \mathcal{G}} 4 \mathbb{E}_t\left[\left\|  \Delta_m\left(x^{t+1}, x^t\right) - \Delta\left(x^{t+1}, x^t\right) \right\|^2\mid [3]\right] \right]\\
&+ \mathbb{E}\left[ \frac{\mathcal{P}_{\mathcal{G}^t_C}}{C(1-\delta)} \sum_{m \in \mathcal{G}} 4\omega  \mathbb{E}_t\left[\left\| \Delta\left(x^{t+1}, x^t\right)  \right\|^2\mid [3]\right] \right].
\end{align*}
Using Assumptions \ref{assm:L-smoothness}, \ref{assm:global}, \ref{assm:local}: 
    \begin{align*}
    \widehat{T}_2 &\leq \mathbb{E}\left[  4(1+\omega)  \frac{G\mathcal{P}_{\mathcal{G}^t_C}}{C(1-\delta)} \frac{\mathcal{L}_{ \pm}^2}{b}\|x^{t+1}-x^t\|^2 \right] \\
    &+\mathbb{E}\left[  4(\omega+1)\frac{G\mathcal{P}_{\mathcal{G}^t_C}}{C(1-\delta)} \omega  L_{ \pm}^2 \|x^{t+1}-x^t\|^2\right]\\
        &+\mathbb{E}\left[  4\frac{G\mathcal{P}_{\mathcal{G}^t_C}}{C(1-\delta)} \omega  L^2 \|x^{t+1}-x^t\|^2\right].
\end{align*}
Finally, we obtain
\begin{align*}
       T_2 &=   \mathbb{E}\left[ \mathbb{E}_t\left[\left\| \frac{1}{G^t_C} \sum \limits_{m \in \mathcal{G}^t_C}\clip_{\lambda}\left(\mathcal{Q}\left(\widehat{\Delta}_m\left(x^{t+1}, x^t\right)\right)\right)  -   \texttt{ARAgg}_Q^{t+1}\right\|^2\mid [3]\right]\right]\\
       &\leq 4\frac{G\mathcal{P}_{\mathcal{G}^t_C}}{C(1-\delta)} c\delta \left( (1+\omega)\frac{\mathcal{L}_{ \pm}^2}{b} + (\omega+1) L_{ \pm}^2 + \omega  L^2 \right) \mathbb{E}\left[\|x^{t+1} - x^t\|^2\right].
\end{align*}
\end{proof}

\begin{lemma}
\label{lemma:final_lemma_Q}
    Let Assumptions ~\ref{assm:bounded-aggr}, \ref{assm:L-smoothness}, \ref{assm:global}, \ref{assm:local}, \ref{assm:local_smooth_all}, \ref{assm:het}, \ref{assm:bounded-compressor} hold and the compression operator satisfy Definition~\ref{def:Q}. We set $\lambda_{t+1} = D_Q \max_{m,i} L_{m,i}\|x^{t+1} - x^t\|$. Also, let us introduce the notation
\begin{align*}\texttt{ARAgg}_Q^{t+1} = \texttt{ARAgg}\left(\clip_{\lambda_{t+1}}\left(\cQ\left(\widehat{\Delta}_1(x^{t+1}, x^t)\right)\right),\right.\\
\left.\ldots, \clip_{\lambda_{t+1}}\left(\cQ\left(\widehat{\Delta}_C(x^{t+1}, x^t)\right)\right)\right).\end{align*}
Then for all $t\geq 0$ the iterates produced by \gls{Byz-VR-MARINA-PP} (Algorithm~\ref{alg:byz_vr_marina}) satisfy

    \begin{align*}
    \mathbb{E}\left[\left\|g^{t+1}-\nabla f\left(x^{t+1}\right)\right\|^2\right] &\leq  \left(1-\frac{p}{2}\right) \mathbb{E}\left[\left\|g^{t}-\nabla f\left(x^{t}\right)\right\|^2\right]\\
    &+\widehat{B}\mathbb{E}\left[\left\|\nabla f\left(x^t\right)\right\|^2\right] + \widehat{D} \zeta^2+\frac{pA}{4}\|x^{t+1} - x^t\|^2,
\end{align*}
with
\begin{align*}
    A& = \frac{4}{p}\left(  \frac{p_G \mathcal{P}_{\mathcal{G}^t_C} G}{C^2(1-\delta)^2} \omega + \frac{8G\cP_{\cG_{\widehat{C}}^t}c\delta}{(1-\delta)\widehat{C}} B + 6\widetilde{B} + \frac{4}{p}(1-p_G) + \frac{8}{p}p_G \frac{G\mathcal{P}_{\mathcal{G}^t_C}}{C(1-\delta)} c \delta \omega  \right) L^2\\
    &+\frac{4}{p}\left( \frac{p_G \mathcal{P}_{\mathcal{G}^t_C} G}{C^2(1-\delta)^2} \left( \omega + 1 \right) + \frac{8}{p}p_G \frac{G\mathcal{P}_{\mathcal{G}^t_C}}{C(1-\delta)} c \delta (\omega+1) \right)\left(L_{ \pm}^2 + \frac{\mathcal{L}_{ \pm}^2}{b}\right)\\
    &+\frac{16}{p^2}(1-p_G)F_{\cA}^2 \left(D_Q \max_{m,i} L_{m,i} \right)^2
\end{align*}
\textbf{\begin{align*}
    \widehat{B} = 2 \frac{ \delta\mathcal{P}_{\mathcal{G}^t_{\widehat{C}}} }{1-\delta} B\left(\frac{12cG}{\widehat C} + p \right) + 6\widetilde{B}, \quad \widehat{D} = 2 \frac{ \delta\mathcal{P}_{\mathcal{G}^t_{\widehat{C}}} }{1-\delta} \left(\frac{6cG}{\widehat C} + p \right) + \widetilde{D},
\end{align*}}
where $\widetilde{B} \eqdef 0$ and $\widetilde{D} \eqdef 0$ when $\widehat{C} = M$, and $\widetilde{B} \eqdef \frac{\cP_{\cG_{\widehat{C}}^t}GB}{(1-\delta)\widehat{C}}$ and $\widetilde{D} \eqdef \frac{\cP_{\cG_{\widehat{C}}^t}G}{(1-\delta)\widehat{C}}$ when $$\widehat{C} < M, p_G = \operatorname{Prob}\left\lbrace G^t_C \geq (1-\delta)C \right\rbrace $$ and $$\mathcal{P}_{\mathcal{G}^t_C} =  \operatorname{Prob}\left\lbrace m \in \mathcal{G}^t_C \mid G^t_C \geq\left(1-\delta\right) C\right\rbrace.$$
\end{lemma}
\begin{proof}
    Let us combine bounds for $A_1$ and $A_2$ together:
\begin{align*}
     A_0 &=    \mathbb{E}\left[\left\|g^{t+1}-\nabla f\left(x^{t+1}\right)\right\|^2\right]\\
     & \leq \left(1+\frac{p}{2}\right)\mathbb{E}\left[\left\|\overline{g}^{t+1}-\nabla f\left(x^{t+1}\right)\right\|^2\right]+\left(1+\frac{2}{p}\right)\mathbb{E}\left[\left\|g^{t+1}-\overline{g}^{t+1}\right\|^2\right]\\
     &\leq \left(1+\frac{p}{2}\right)A_1 + \left(1+\frac{2}{p}\right)A_2\\
       & \leq \left(1+\frac{p}{2}\right)(1-p) \mathbb{E}\left[\left\|g^{t}-\nabla f(x^{t})\right\|^2\right]\\
       &+ \left(1+\frac{p}{2}\right)p\frac{ \delta\mathcal{P }_{\mathcal{G}^t_{\widehat{C}}} }{(1-\delta)}  \mathbb{E}\left[ B\|\nabla f(x)\|^2+\zeta^2 \right]\\
        &+ \left(1+\frac{p}{2}\right)(1-p)p_G\frac{\mathcal{P}_{\mathcal{G}^t_C} G}{C^2(1-\delta)^2} \times\\
        &\times\left(  \omega L^2+(\omega+1) L_{ \pm}^2+\frac{(\omega+1) \mathcal{L}_{ \pm}^2}{b} \right)\mathbb{E}\left[\|x^{t+1} - x^t\|^2\right]\\
    &+ \left(1+\frac{2}{p}\right)p\mathbb{E}\left[\mathbb{E}_t\left[\left\|\texttt{ARAgg}\left(\nabla f_1(x^{t+1}), \ldots, \nabla f_n(x^{t+1})\right) - \nabla f(x^{t+1})\right\|^2\right]\mid [1]\right]\\
   & + \left(1+\frac{2}{p}\right)(1-p)p_G \times\\ &\times\mathbb{E}\left[ \mathbb{E}_t\left[\left\| \frac{1}{G^t_C} \sum \limits_{m \in \mathcal{G}^t_C}\clip_{\lambda}\left(\mathcal{Q}\left(\widehat{\Delta}_m\left(x^{t+1}, x^t\right)\right)\right)  -   \texttt{ARAgg}_Q^{t+1}\right\|^2\mid [3]\right]\right]\\
   &+ \left(1+\frac{2}{p}\right)(1-p)(1-p_G)\mathbb{E}\left[\mathbb{E}_t\left[\left\| \nabla f(x^{t+1}) - \nabla f(x^{t}) - \texttt{ARAgg}_Q^{t+1}\right\|^2\mid [2]\right]\right].
   \end{align*}
   Using Lemma \ref{lemma:good_aggr_new} and lemmas from General Analysis (Lemmas~\ref{lemma:full_aggr} and \ref{lemma:bad_aggr}) we have 
     \begin{align*}
   A_0 &=  \mathbb{E}\left[\left\|g^{t+1}-\nabla f\left(x^{t+1}\right)\right\|^2\right]\\ 
   & \leq \left(1-\frac{p}{2}\right) \mathbb{E}\left[\left\|g^{t}-\nabla f\left(x^{t}\right)\right\|^2\right] + 2p\frac{ \delta\mathcal{P }_{\mathcal{G}^t_{\widehat{C}}} }{(1-\delta)}  \mathbb{E}\left[ B\|\nabla f(x)\|^2+\zeta^2 \right]\\
            &+ \text{\normalsize $\left(1-\frac{p}{2}\right)p_G\frac{\mathcal{P}_{\mathcal{G}^t_C} G}{C^2(1-\delta)^2} \left(  \omega L^2+(\omega+1) L_{ \pm}^2+\frac{(\omega+1) \mathcal{L}_{ \pm}^2}{b} \right)\mathbb{E}\left[\|x^{t+1} - x^t\|^2\right]$}\\
      &+ \left(p+2\right)\left(\frac{8 G \mathcal{P}_{\mathcal{G}^t_{\widehat{C}}} c\delta B}{(1-\delta) \widehat{C}} + 2\widetilde{B}\right)\mathbb{E}\left[\left\|\nabla f\left(x^t\right)\right\|^2 + L^2\left\|x^{t+1}-x^t\right\|^2\right]\\
      &+4\left(p+2\right) \frac{G \mathcal{P}_{\mathcal{G}^t_{\widehat{C}}} c\delta}{(1-\delta)\widehat{C}}  \zeta^2 + (p+2)\widetilde{\zeta}^2\\
   & + \frac{2}{p}p_G\mathbb{E}\left[  4(1+\omega)  \frac{G\mathcal{P}_{\mathcal{G}^t_C}}{C(1-\delta)} c\delta \frac{\mathcal{L}_{ \pm}^2}{b} \|x^{t+1}-x^t\|^2 \right] \\
    &    + \frac{2}{p}p_G \mathbb{E}\left[  4(\omega+1)\frac{G\mathcal{P}_{\mathcal{G}^t_C}}{C(1-\delta)} c\delta L_{ \pm}^2  \|x^{t+1}-x^t\|^2\right]\\
     &   + \frac{2}{p}p_G \mathbb{E}\left[  4\omega\frac{G\mathcal{P}_{\mathcal{G}^t_C}}{C(1-\delta)} c\delta   L^2   \|x^{t+1}-x^t\|^2\right]\\
     &+ \frac{2}{p}(1-p_G)2(L^2+F_{\cA}^2\alpha^2_{\lambda_{t+1}})\mathbb{E}\left[\left\|  x^{t+1} - x^t\right\|^2\right].
\end{align*}
Finally, we have 
    \begin{align*}
    \mathbb{E}\left[\left\|g^{t+1}-\nabla f\left(x^{t+1}\right)\right\|^2\right] &\leq  \left(1-\frac{p}{2}\right) \mathbb{E}\left[\left\|g^{t}-\nabla f\left(x^{t}\right)\right\|^2\right]\\
    &+\widehat{B}\mathbb{E}\left[\left\|\nabla f\left(x^t\right)\right\|^2\right] + \widehat{D}\zeta^2+\frac{pA}{4}\|x^{t+1} - x^t\|^2,
\end{align*}
where 
\begin{align*}
    A& = \frac{4}{p}\left(  \frac{p_G \mathcal{P}_{\mathcal{G}^t_C} G}{C^2(1-\delta)^2} \omega + \frac{8G\cP_{\cG_{\widehat{C}}^t}c\delta}{(1-\delta)\widehat{C}} B + 6\widetilde{B} + \frac{4}{p}(1-p_G) + \frac{8}{p}p_G \frac{G\mathcal{P}_{\mathcal{G}^t_C}}{C(1-\delta)} c \delta \omega  \right) L^2\\
    &+\frac{4}{p}\left( \frac{p_G \mathcal{P}_{\mathcal{G}^t_C} G}{C^2(1-\delta)^2} \left( \omega + 1 \right) + \frac{8}{p}p_G \frac{G\mathcal{P}_{\mathcal{G}^t_C}}{C(1-\delta)} c \delta (\omega+1) \right)\left(L_{ \pm}^2 + \frac{\mathcal{L}_{ \pm}^2}{b}\right)\\
    &+\frac{16}{p^2}(1-p_G)F_{\cA}^2 \left(D_Q \max_{m,i} L_{m,i} \right)^2
\end{align*}
and
\begin{align*}
    \widehat{B} = 2 \frac{ \delta\mathcal{P}_{\mathcal{G}^t_{\widehat{C}}} }{1-\delta} B\left(\frac{12cG}{\widehat C} + p \right) + 6\widetilde{B}, \quad \widehat{D} = 2 \frac{ \delta\mathcal{P}_{\mathcal{G}^t_{\widehat{C}}} }{1-\delta} \left(\frac{6cG}{\widehat C} + p \right) + \widetilde{D}.
\end{align*}
\end{proof}

\subsection{Main results}

\begin{theorem}\label{them:1_Q}
 Let Assumptions \ref{assm:bounded-aggr}, \ref{assm:L-smoothness}, \ref{assm:global}, \ref{assm:local}, \ref{assm:local_smooth_all}, \ref{assm:het}, \ref{assm:bounded-compressor} hold. Setting $\lambda_{t+1} = \max_{m,i} L_{m,i} \left\|x^{t+1} - x^t\right\|$. Assume that
$$
0<\gamma \leq \frac{1}{L+\sqrt{A}}, \quad 4\widehat{B}<p,
$$ 
where 
\begin{align*}
    A& = \frac{4}{p}\left(  \frac{p_G \mathcal{P}_{\mathcal{G}^t_C} G}{C^2(1-\delta)^2} \omega + \frac{8G\cP_{\cG_{\widehat{C}}^t}c\delta}{(1-\delta)\widehat{C}} B + 6\widetilde{B} + \frac{4}{p}(1-p_G) + \frac{8}{p}p_G \frac{G\mathcal{P}_{\mathcal{G}^t_C}}{C(1-\delta)} c \delta \omega  \right) L^2\\
    &+\frac{4}{p}\left( \frac{p_G \mathcal{P}_{\mathcal{G}^t_C} G}{C^2(1-\delta)^2} \left( \omega + 1 \right) + \frac{8}{p}p_G \frac{G\mathcal{P}_{\mathcal{G}^t_C}}{C(1-\delta)} c \delta (\omega+1) \right)\left(L_{ \pm}^2 + \frac{\mathcal{L}_{ \pm}^2}{b}\right)\\
    &+\frac{16}{p^2}(1-p_G)F_{\cA}^2 \left(D_Q \max_{m,i} L_{m,i} \right)^2
\end{align*}
\begin{align*}
    \widehat{B} = 2 \frac{ \delta\mathcal{P}_{\mathcal{G}^t_{\widehat{C}}} }{1-\delta} B\left(\frac{12cG}{\widehat C} + p \right) + 6\widetilde{B}, \quad \widehat{D} = 2 \frac{ \delta\mathcal{P}_{\mathcal{G}^t_{\widehat{C}}} }{1-\delta} \left(\frac{6cG}{\widehat C} + p \right) + \widetilde{D},
\end{align*}
where $\widetilde{B} \eqdef 0$ and $\widetilde{D} \eqdef 0$ when $\widehat{C} = M$, and $\widetilde{B} \eqdef \frac{\cP_{\cG_{\widehat{C}}^t}GB}{(1-\delta)\widehat{C}}$ and $\widetilde{D} \eqdef \frac{\cP_{\cG_{\widehat{C}}^t}G}{(1-\delta)\widehat{C}}$ when $\widehat{C} < M$, and 
\begin{align*}
 \mathcal{P}_{\mathcal{G}^t_C} &=   \frac{C}{Mp_G} \cdot \sum_{(1-\delta)C\leq m^\prime \leq C} \left(\left(\begin{array}{l}
G-1 \\
t-1
\end{array}\right)\left(\begin{array}{l}
M-G \\
C-m^\prime
\end{array}\right) \left(\left(\begin{array}{l}
n \\
C
\end{array}\right)\right)^{-1} \right),\\
p_G &=  \PP\left\lbrace G_C^t \geq\left(1-\delta\right) C\right\rbrace\\
&= \sum_{\lceil(1-\delta)C\rceil\leq t \leq C} \left(\left(\begin{array}{l}
G \\
t
\end{array}\right)\left(\begin{array}{l}
M-G \\
C-m^\prime
\end{array}\right) \left(\begin{array}{l}
n \\
C
\end{array}\right)^{-1} \right).
\end{align*}
Then for all $T \geq 0$ the iterates produced by \gls{Byz-VR-MARINA} (Algorithm \ref{alg:byz_vr_marina}) satisfy
$$
\mathbb{E}\left[\left\|\nabla f\left(\widehat{x}^T\right)\right\|^2\right] \leq \frac{2 \Phi^{(0)}}{\gamma\left(1-\frac{4\widehat{B}}{p}\right)(T+1)}+\frac{2 \widehat{D} \zeta^2}{p-4 \widehat{B}},
$$
where $\widehat{x}^T$ is chosen uniformly at random from $x^0, x^1, \ldots, x^T$, and $\Phi^{(0)}=$ $f\left(x^0\right)-f^{\star}+\frac{\gamma}{p}\left\|g^0-\nabla f\left(x^0\right)\right\|^2$ . 
\end{theorem}
\begin{proof}
    The proof is analogous to the proof of Theorem~\ref{them:1}.
\end{proof}

\begin{theorem}\label{them:2_Q}
Let Assumptions  \ref{assm:bounded-aggr}, \ref{assm:bounded-compressor}, \ref{assm:L-smoothness}, \ref{assm:global}, \ref{assm:local}, \ref{assm:local_smooth_all}, \ref{assm:het}, \ref{assm:PL} hold. Setting $\lambda_{t+1} = \max_{m,i} L_{m,i} \left\|x^{t+1} - x^t\right\|$. Assume that
$$
0<\gamma \leq \frac{1}{L+\sqrt{2 A}}, \quad 8\widehat{B}<p,
$$
where
\begin{align*}
    A& = \frac{4}{p}\left(  \frac{p_G \mathcal{P}_{\mathcal{G}^t_C} G}{C^2(1-\delta)^2} \omega + \frac{8G\cP_{\cG_{\widehat{C}}^t}c\delta}{(1-\delta)\widehat{C}} B + 6\widetilde{B} + \frac{4}{p}(1-p_G) + \frac{8}{p}p_G \frac{G\mathcal{P}_{\mathcal{G}^t_C}}{C(1-\delta)} c \delta \omega  \right) L^2\\
    &+\frac{4}{p}\left( \frac{p_G \mathcal{P}_{\mathcal{G}^t_C} G}{C^2(1-\delta)^2} \left( \omega + 1 \right) + \frac{8}{p}p_G \frac{G\mathcal{P}_{\mathcal{G}^t_C}}{C(1-\delta)} c \delta (\omega+1) \right)\left(L_{ \pm}^2 + \frac{\mathcal{L}_{ \pm}^2}{b}\right)\\
    &+\frac{16}{p^2}(1-p_G)F_{\cA}^2 \left(D_Q \max_{m,i} L_{m,i} \right)^2
\end{align*}
\begin{align*}
    \widehat{B} = 2 \frac{ \delta\mathcal{P}_{\mathcal{G}^t_{\widehat{C}}} }{1-\delta} B\left(\frac{12cG}{\widehat C} + p \right) + 6\widetilde{B}, \quad \widehat{D} = 2 \frac{ \delta\mathcal{P}_{\mathcal{G}^t_{\widehat{C}}} }{1-\delta} \left(\frac{6cG}{\widehat C} + p \right) + \widetilde{D},
\end{align*}
where $\widetilde{B} \eqdef 0$, $\widetilde{D} \eqdef 0$ when $\widehat{C} = M$, $\widetilde{B} \eqdef \frac{\cP_{\cG_{\widehat{C}}^t}GB}{(1-\delta)\widehat{C}}$, $\widetilde{D} \eqdef \frac{\cP_{\cG_{\widehat{C}}^t}G}{(1-\delta)\widehat{C}}$ when $\widehat{C} < M$, and where $p_G = \operatorname{Prob}\left\lbrace G^t_C \geq (1-\delta)C \right\rbrace $ and $\mathcal{P}_{\mathcal{G}^t_C} =  \operatorname{Prob}\left\lbrace m \in \mathcal{G}^t_C \mid G^t_C \geq\left(1-\delta\right) C\right\rbrace$. Then for all $T \geq 0$ the iterates produced by \gls{Byz-VR-MARINA} (Algorithm \ref{alg:byz_vr_marina}) satisfy
$$
\mathbb{E}\left[f\left(x^T\right)-f\left(x^{\star}\right)\right] \leq\left(1-\rho\right)^T \Phi^{(0)}+\frac{2\widehat{D}\zeta^2}{p\rho},
$$
where $\rho = \min\left[\gamma\mu\left(1-\frac{8\widehat{B}}{p}\right), \frac{p}{4}\right]$ and $\Phi^{(0)}=$ $f\left(x^0\right)-f^{\star}+\frac{2\gamma}{p}\left\|g^0-\nabla f\left(x^0\right)\right\|^2$.
\end{theorem}
\begin{proof}
    The proof is analogous to the proof of Theorem~\ref{them:2}.
\end{proof}

\subsection{On the technical Non-Triviality of the analysis}\label{appendix:technical_challenges}

As we explain in the main part of the paper, the main reason why we propose to use clipping is to handle the situations when Byzantine workers form a majority during some communication rounds since the existing approaches are vulnerable to such scenarios. However, the introduction of the clipping does not come for free: if the clipping level is too small, clipping can create a noticeable bias to the updates. Because of this issue, existing works such as \citep{zhang2020adaptive, gorbunov2020stochastic} use non-trivial policies for the choice of the clipping level, and the analysis in these works differs significantly from the existing analysis for the methods without clipping. The analysis of \gls{Byz-VR-MARINA} is based on the unbiasedness of vectors $\mathcal{Q}(\hat \Delta_m(x^{t+1}, x^t))$, i.e., on the following identity: $\mathbb{E}[\mathcal{Q}(\hat \Delta_m(x^{t+1}, x^t)) \mid x^{t+1}, x^t] = \Delta_m(x^{t+1}, x^t) = \nabla f_m(x^{t+1}) - \nabla f_m(x^t)$. Since $\mathbb{E}[\clip_{\lambda_{t+1}}(\mathcal{Q}(\hat \Delta_m(x^{t+1}, x^t))) \mid x^{t+1}, x^t] \neq \nabla f_m(x^{t+1}) - \nabla f_m(x^t)$ in general, to analyze \gls{Byz-VR-MARINA-PP} we also use a special choice of the clipping level: $\lambda_{t+1} = \alpha_{t+1} \|x^{t+1} - x^t\|$. To illustrate the main reasons for that, let us consider the case of uncompressed communication ($\mathcal{Q}(x) \equiv x$). In this setup, for large enough $\alpha_{t+1}$ we have $\clip_{\lambda_{t+1}}\hat \Delta_m(x^{t+1}, x^t) = \hat \Delta_m(x^{t+1}, x^t)$ for all $i\in \mathcal{G}$ (due to Assumption~\ref{assm:smoothness_simplified}), which allows us using a similar proof to the one for Byz-VR-MARINA when good workers form a majority in a round. Moreover, when Byzantine workers form a majority, our choice of the clipping level allows us to bound the second moment of the shift from the Byzantine workers as $\sim \| x^{t+1} - x^t \|^2$ (see Lemmas~\ref{lemma:premainA2} and \ref{lemma:bad_aggr}), i.e., the second moment of the shift is of the same scale as the variance of $\lbrace g_i \rbrace_{i\in \mathcal{G}}$, which goes to zero. Next, to properly analyze these two situations, we overcame another technical challenge related to the estimation of the conditional expectations and probabilities of corresponding events (see Lemmas \ref{lemma:premainA2} and \ref{lemma:full_aggr} and formulas for $p_G$ and $\mathcal{P}_{\mathcal{G}_C^t}$ at the beginning of Section~\ref{section:convergence_results}). In particular, the derivation of formula \eqref{eq:bvdjbjdfbvjdf} is quite non-standard for stochastic optimization literature: there are two sources of stochasticity – one comes from the sampling of clients, and the other one comes from the sampling of stochastic gradients and compression. This leads to the estimation of variance of the average of the random number of random vectors, which is novel on its own. In addition, when the compression operator is used, the analysis becomes even more involved since one cannot directly apply the main property of unbiased compression (Definition~\ref{def:Q}), and we use Lemma~\ref{lemma:clipping} in the proof to address this issue. It is also worth mentioning that in contrast to \gls{Byz-VR-MARINA}, our method does not require full participation even with a small probability $p$. Instead, it is sufficient for \gls{Byz-VR-MARINA-PP} to sample a large enough cohort of $\widehat{C}$ clients with probability $p$ to ensure that Byzantine workers form a minority in such rounds.

\clearpage

\section{Byz-VR-MARINA-PP+: Simplified Version of Byz-VR-MARINA-PP}\label{appendix:simplified+}

In this section, we present a simplified version of \gls{Byz-VR-MARINA-PP} called \algname{Byz-VR-MARINA-PP+} (see Algorithm~\ref{alg:byz_vr_marina+}). The only difference between the two methods is in Line~\ref{line:aggregation+}: \algname{Byz-VR-MARINA+} does not apply robust aggregation when $c_t = 0$ and just averages the clipped vectors received from the set of clients $\set$. Nevertheless, when $c_t = 1$, i.e., a large cohort of clients is sampled, the method still uses robust aggregation.

\begin{algorithm*}[h]
   \caption{\algname{Byz-VR-MARINA-PP+}: Simplified \gls{Byz-VR-MARINA-PP}}\label{alg:byz_vr_marina+}
\begin{algorithmic}[1]
   \STATE {\bfseries Input:} vectors $x^0, g^0 \in \R^d$, stepsize $\gamma$, mini-batch size $b$, probability $p\in(0,1]$, number of iterations $K$, $(\delta,c)$-\texttt{ARAgg}, clients' sample size $1 \leq C \leq \widehat{C} \leq n$, clipping coefficients $\{\alpha_{t}\}_{t\geq 1}$ 
   \FOR{$k=0,1,\ldots,K-1$}
   \STATE Get a sample from Bernoulli distribution with parameter $p$: $c_t \sim \text{Be}(p)$
   \STATE Sample the set of clients $\set \subseteq [n]$, $|\set| = C$ if $c_t = 0$; otherwise $|\set| = \widehat{C}$
   \STATE Broadcast $g^t$, $c_t$ to all workers
   \FOR{$m \in \cG \cap \set$ in parallel} 
   \STATE $x^{t+1} = x^t - \gamma g^t$ and $\lambda_{t+1} = \alpha_{t+1}\|x^{t+1} - x^t\|$
   \STATE  Set $g_m^{t+1} = \begin{cases} \nabla f_m(x^{t+1}),& \text{if } c_t = 1,\\ g^t + \clip_{\lambda_{t+1}}\left(\cQ\left(\widehat{\Delta}_m(x^{t+1}, x^t)\right)\right),& \text{otherwise,}\end{cases}$\\ where $\widehat{\Delta}_m(x^{t+1}, x^t)$ is a mini-batched estimator of $\nabla f_m(x^{t+1}) - \nabla f_m(x^t)$, $\cQ(\cdot)$ for $m\in\cG \cap \set$ are computed independently
   \ENDFOR
   \STATE\label{line:aggregation+} $g^{t+1} = \begin{cases} \texttt{ARAgg}\left(\{g_m^{t+1}\}_{m\in \set}\right),& \text{if } c_t = 1, \\
       g^t + \frac{1}{C}\sum\limits_{m\in \set}\clip_{\lambda_{t+1}}\left(\cQ\left(\widehat{\Delta}_m(x^{t+1}, x^t)\right)\right),&\text{otherwise}\end{cases}$
   \ENDFOR
\end{algorithmic}
\end{algorithm*}

The key idea behind this modification can be explained as follows. For simplicity, let us assume that $C$ is small and $\deltar$ is also small. Then, for the communication rounds with $c_t = 0$, with a large probability, only good clients will be sampled. In this case, the method can use just an average of the received vectors and benefit from the lack of bias appearing due to the robust aggregation. Moreover, when $c_t = 0$ and at least one of the sampled clients is Byzantine, the method will tolerate due to the clipping. That is, when $C$ is small, the method can potentially benefit from the lack of robust aggregation when $c_t = 0$. However, for the rounds with $c_t = 1$, in the worst case, $\widehat{C} = M$, meaning that all Byzantines workers are guaranteed to be sampled. To tolerate such situations, we keep the robust aggregation in the method when $c_t = 1$.

\subsection{Analysis for bounded compressors}

For simplicity, we analyze \algname{Byz-VR-MARINA-PP+} for bounded compressors only. The analysis is very similar to the one we provide for \gls{Byz-VR-MARINA-PP}, but several steps are significantly simpler. In particular, the central part in the analysis of \gls{Byz-VR-MARINA-PP} is in deriving a good recursive inequality for $\mathbb{E}[\|g^t - \nabla f(x^t)\|^2]$, which requires several quite technical steps. For \algname{Byz-VR-MARINA-PP+}, one can obtain a similar inequality much easier as shown in the next lemma.

\begin{lemma}
\label{lemma:final_lemma_simplified}
    Let Assumptions ~\ref{assm:L-smoothness}, \ref{assm:global}, \ref{assm:local}, \ref{assm:local_smooth_all}, \ref{assm:het}, \ref{assm:bounded-compressor} hold and the compression operator satisfy Definition~\ref{def:Q}. Assume that $C \leq G$. We set clipping parameter $\lambda_{t+1} = D_Q \max_{m,i} L_{m,i}\|x^{t+1}-x^t\|$. Then for all $t\geq 0$ the iterates produced by \algname{Byz-VR-MARINA-PP+} (Algorithm~\ref{alg:byz_vr_marina+}) satisfy
    \begin{align}
    \mathbb{E}\left[\left\|g^{t+1}-\nabla f\left(x^{t+1}\right)\right\|^2\right] &\leq  \left(1-\frac{p}{2}\right) \mathbb{E}\left[\left\|g^{t}-\nabla f\left(x^{t}\right)\right\|^2\right] \label{eq:simplified_main_lemma}\\
    &+\widehat{B}\mathbb{E}\left[\left\|\nabla f\left(x^t\right)\right\|^2\right] + \widehat{D} \zeta^2+\frac{pA}{4}\|x^{t+1} - x^t\|^2, \notag
\end{align}
with
\begin{align*}
    A& = \frac{4}{p}\left(  \frac{8p B G \mathcal{P}_{\mathcal{G}^t_{\widehat{C}}} c\delta}{(1-\delta)\widehat{C}} + 6p\widetilde{B}  +  \frac{(1-p)p_{\cG}^t\omega}{C} + \frac{6(1-p)(1- p_{\cG}^t)}{p} \right) L^2\\
    &+\frac{4(1-p)p_{\cG}^t}{p}\left( 1+\frac{\omega}{C} \right)L_{ \pm}^2 + \frac{4(1-p)p_{\cG}^t(1+\omega)}{pC} \frac{\mathcal{L}_{ \pm}^2}{b}\\
    &+\frac{24(1-p)(1-p_{\cG}^t)}{p^2} \left(D_Q \max_{m,i} L_{m,i} \right)^2,
\end{align*}
\textbf{\begin{align*}
    \widehat{B} = \frac{8 G \mathcal{P}_{\mathcal{G}^t_{\widehat{C}}} c\delta Bp}{(1-\delta)\widehat{C}} + 6p\widetilde{B} , \quad \widehat{D} = \frac{4 G \mathcal{P}_{\mathcal{G}^t_{\widehat{C}}} c\delta p}{(1-\delta)\widehat{C}} + p\widetilde{D},
\end{align*}}
where $\widetilde{B}$, $\widetilde{D}$, $p_{\cG}^t$, $\mathcal{P}_{\mathcal{G}^t_{\widehat C}}$ are defined in Lemma~\ref{lemma:final_lemma}.
\end{lemma}
\begin{proof}
    From the update rule of $g^{t+1}$, we have
    \begin{align}
        &\mathbb{E}\left[\left\|g^{t+1} - \nabla f(x^{t+1})\right\|^2\right]\notag\\
        &= p\underbrace{\mathbb{E}\left[\left\|\texttt{ARAgg}\left(\{g_m^{t+1}\}_{m\in \set}\right) - \nabla f(x^{t+1})\right\|^2\mid c_t = 1\right]}_{T_1} \label{eq:first_bound_simplified}\notag\\
        & + (1-p)\times\\
        &\times\underbrace{\mathbb{E}\left[\left\|g^t + \frac{1}{C}\sum\limits_{m\in \set}\clip_{\lambda_{t+1}}\left(\cQ\left(\widehat{\Delta}_m(x^{t+1}, x^t)\right)\right) - \nabla f(x^{t+1})\right\|^2\mid c_t = 0\right]}_{T_2}. \notag
    \end{align}
    Next, we bound $T_1$ and $T_2$ separately. From Lemma~\ref{lemma:full_aggr}, we have
    \begin{equation*}
        T_1 \leq \left(\frac{8 G \mathcal{P}_{\mathcal{G}^t_{\widehat{C}}} c\delta B}{(1-\delta) \widehat{C}} + 2\widetilde{B}\right)  \mathbb{E}\left[\left\|\nabla f\left(x^t\right)\right\|^2  + L^2\left\|x^{t+1}-x^t\right\|^2\right] + \frac{4 G \mathcal{P}_{\mathcal{G}^t_{\widehat{C}}} c\delta \zeta^2}{(1-\delta) \widehat{C}} + \widetilde{\zeta}^2.
    \end{equation*}
    As for $T_2$, we consider two possible situations: either $\set \cap \cB = \varnothing$ (no Byzantine workers are among sampled ones) or $\set \cap \cB \neq \varnothing$ (at least one Byzantine worker is sampled). Then, $T_2$ equals
    \begin{align*}
        &T_2 = \text{\footnotesize$p_{\cG}^t\underbrace{\mathbb{E}\left[\left\|g^t+\frac{1}{C}\sum\limits_{m\in \set}\clip_{\lambda_{t+1}}\left(\cQ\left(\widehat{\Delta}_m(x^{t+1}, x^t)\right)\right) - \nabla f(x^{t+1})\right\|^2\mid c_t = 0, \set \cap \cB = \varnothing\right]}_{\widehat T_2}$}\\
        &+ \text{\scriptsize$ (1-p_{\cG}^t) \underbrace{\mathbb{E}\left[\left\|g^t+\frac{1}{C}\sum\limits_{m\in \set}\clip_{\lambda_{t+1}}\left(\cQ\left(\widehat{\Delta}_m(x^{t+1}, x^t)\right)\right) - \nabla f(x^{t+1})\right\|^2\mid c_t = 0, \set \cap \cB \neq \varnothing\right]}_{\widetilde T_2}$},
    \end{align*}
    where
    \begin{align*}
        p_{\cG}^t &\eqdef \PP\{\set \cap \cB = \varnothing \mid c_t = 0\}\\
        &= \frac{\binom{G}{C}}{\binom{n}{C}} = \frac{(G-C+1)(G-C+2)\cdot\ldots\cdot (n-C)}{(G+1)(G+2)\cdot \ldots \cdot n}.
    \end{align*}
    The choice of the clipping level $\lambda_{t+1} = D_Q\max_{m,i} L_{m,i} \|x^{t+1} - x^t\|$ and inequality \eqref{eq:clipping_turned_off} imply that $\clip_{\lambda_{t+1}}\left(\cQ\left(\widehat{\Delta}_m(x^{t+1}, x^t)\right)\right)$ for all $m \in \cG$. Therefore, for $\widehat T_2$, we have
    \begin{align*}
        &\widehat T_2 = \mathbb{E}\left[\left\|g^t+\frac{1}{C}\sum\limits_{m\in \set}\cQ\left(\widehat{\Delta}_m(x^{t+1}, x^t)\right) - \nabla f(x^{t+1})\right\|^2\mid c_t = 0, \set \cap \cB = \varnothing\right]\\
        &= \mathbb{E}\left[\|g^t - \nabla f(x^t)\|^2\right]\\
        &+ \text{\small$ \mathbb{E}\left[\left\|\frac{1}{C}\sum\limits_{m\in \set}\cQ\left(\widehat{\Delta}_m(x^{t+1}, x^t)\right) - (\nabla f(x^{t+1}) - \nabla f(x^t))\right\|^2\mid c_t = 0, \set \cap \cB = \varnothing\right]$},
    \end{align*}
    where we use that $$\mathbb{E}\left[\frac{1}{C}\sum_{m\in \set}\cQ\left(\widehat{\Delta}_m(x^{t+1}, x^t)\right) \mid c_t = 0, \set \cap \cB = \varnothing\right] = \nabla f(x^{t+1}) - \nabla f(x^t)$$. Moreover, since \begin{align*}
&\mathbb{E}\left[\frac{1}{C}\sum_{m\in \set}\cQ\left(\widehat{\Delta}_m(x^{t+1}, x^t)\right) \mid c_t = 0, \set \cap \cB = \varnothing, \set\right]\\& = \frac{1}{C}\sum_{m\in \set} (\nabla f_m(x^{t+1})- \nabla f_m(x^t))\\
&=: \frac{1}{C}\sum_{m\in \set} \Delta_m(x^{t+1},x^t)        
    \end{align*}, we can decompose the last term in the upper-bound for $\widehat{T}_2$ as follows:
    \begin{align*}
       & \widehat{T}_2 = \mathbb{E}\left[\|g^t - \nabla f(x^t)\|^2\right]\\
        &+ \text{\small $\mathbb{E}\left[\mathbb{E}\left[\left\|\frac{1}{C}\sum\limits_{m\in \set}\left(\cQ\left(\widehat{\Delta}_m(x^{t+1}, x^t)\right) - \Delta_m(x^{t+1}, x^t) \right)\right\|^2\mid  \set\right]\mid c_t = 0, \set \cap \cB = \varnothing\right]$}\\
        &+ \mathbb{E}\left[\left\|\frac{1}{C}\sum\limits_{m\in \set}\Delta_m(x^{t+1}, x^t)  - (\nabla f(x^{t+1}) - \nabla f(x^t))\right\|^2\mid c_t = 0, \set \cap \cB = \varnothing\right].
    \end{align*}
    Since the compression operator computations are independent on each client, we have
    \begin{align*}
        \widehat{T}_2 &= \mathbb{E}\left[\|g^t - \nabla f(x^t)\|^2\right]\\
        &+ \text{\footnotesize$\frac{1}{C^2}\mathbb{E}\left[\mathbb{E}\left[\sum\limits_{m\in \set}\left\|\cQ\left(\widehat{\Delta}_m(x^{t+1}, x^t)\right) - \Delta_m(x^{t+1}, x^t) \right\|^2\mid  \set\right]\mid c_t = 0, \set \cap \cB = \varnothing\right]$}\\
        &+ \mathbb{E}\left[\left\|\frac{1}{C}\sum\limits_{m\in \set}\Delta_m(x^{t+1}, x^t)\right\|^2\mid c_t = 0, \set \cap \cB = \varnothing\right]\\
        &- \mathbb{E}\left[\left\|\nabla f(x^{t+1}) - \nabla f(x^t)\right\|^2\right]\\
        &\leq \mathbb{E}\left[\|g^t - \nabla f(x^t)\|^2\right]\\
        &+ \text{\footnotesize$ \frac{1}{C^2}\mathbb{E}\left[\mathbb{E}\left[\sum\limits_{m\in \set}\left\|\cQ\left(\widehat{\Delta}_m(x^{t+1}, x^t)\right) - \widehat{\Delta}_m(x^{t+1}, x^t) \right\|^2\mid  \set\right]\mid c_t = 0, \set \cap \cB = \varnothing\right]$}\\
        &+ \text{\footnotesize$ \frac{1}{C^2}\mathbb{E}\left[\mathbb{E}\left[\sum\limits_{m\in \set}\left\| \widehat{\Delta}_m(x^{t+1}, x^t) - \Delta_m(x^{t+1}, x^t) \right\|^2\mid  \set\right]\mid c_t = 0, \set \cap \cB = \varnothing\right]$}\\
        &+ \mathbb{E}\left[\left\|\frac{1}{C}\sum\limits_{m\in \set}\Delta_m(x^{t+1}, x^t)\right\|^2\mid c_t = 0, \set \cap \cB = \varnothing\right]\\
        &- \mathbb{E}\left[\left\|\nabla f(x^{t+1}) - \nabla f(x^t)\right\|^2\right]\\
        &\overset{(\text{Def.~\ref{def:Q}})}{\leq}  \mathbb{E}\left[\|g^t - \nabla f(x^t)\|^2\right]\\
        &+ \frac{\omega}{C^2}\mathbb{E}\left[\sum\limits_{m\in \set}\left\|\widehat{\Delta}_m(x^{t+1}, x^t) \right\|^2\mid c_t = 0, \set \cap \cB = \varnothing\right]\\
        &+ \frac{1}{C^2}\mathbb{E}\left[\sum\limits_{m\in \set}\left\| \widehat{\Delta}_m(x^{t+1}, x^t) - \Delta_m(x^{t+1}, x^t) \right\|^2\mid c_t = 0, \set \cap \cB = \varnothing\right]\\
        &+ \frac{1}{C}\mathbb{E}\left[\sum\limits_{m\in \set}\left\|\Delta_m(x^{t+1}, x^t)\right\|^2\mid c_t = 0, \set \cap \cB = \varnothing\right]\\
        &- \mathbb{E}\left[\left\|\nabla f(x^{t+1}) - \nabla f(x^t)\right\|^2\right]\\
        &= \mathbb{E}\left[\|g^t - \nabla f(x^t)\|^2\right]+ \frac{\omega}{CG}\sum\limits_{m\in \cG}\mathbb{E}\left[\left\|\widehat{\Delta}_m(x^{t+1}, x^t) \right\|^2\right]\\
        &+ \frac{1}{CG}\sum\limits_{m\in \cG}\mathbb{E}\left[\left\| \widehat{\Delta}_m(x^{t+1}, x^t) - \Delta_m(x^{t+1}, x^t) \right\|^2\right]\\
        &+ \frac{1}{G}\sum\limits_{m\in \cG}\mathbb{E}\left[\left\| \Delta_m(x^{t+1}, x^t) \right\|^2\right] - \mathbb{E}\left[\left\|\nabla f(x^{t+1}) - \nabla f(x^t)\right\|^2\right].
    \end{align*}
    Using
    \begin{align*}\mathbb{E}\left[\left\| \widehat{\Delta}_m(x^{t+1}, x^t) \right\|^2\right]& = \mathbb{E}\left[\left\| \widehat{\Delta}_m(x^{t+1}, x^t) - \Delta_m(x^{t+1}, x^t) \right\|^2\right]\\
    &+ \mathbb{E}\left[\left\| \Delta_m(x^{t+1}, x^t) \right\|^2\right],\end{align*} we continue the derivation as follows:
    \begin{align*}
        \widehat{T}_2 &\leq \mathbb{E}\left[\|g^t - \nabla f(x^t)\|^2\right] + \frac{1+\omega}{CG}\sum\limits_{m\in \cG}\mathbb{E}\left[\left\| \widehat{\Delta}_m(x^{t+1}, x^t) - \Delta_m(x^{t+1}, x^t) \right\|^2\right]\\
        &+ \left(1 + 
        \frac{\omega}{C}\right)\frac{1}{G}\sum\limits_{m\in \cG}\mathbb{E}\left[\left\| \Delta_m(x^{t+1}, x^t) \right\|^2\right] - \mathbb{E}\left[\left\|\nabla f(x^{t+1}) - \nabla f(x^t)\right\|^2\right]\\
        &\overset{\eqref{eq:local_hessian_var}}{\leq} \mathbb{E}\left[\|g^t - \nabla f(x^t)\|^2\right] + \frac{(1+\omega)\cL_{\pm}^2}{bC}\mathbb{E}\left[\|x^{t+1} - x^t\|^2\right]\\
        &+ \left(1 + 
        \frac{\omega}{C}\right)\mathbb{E}\left[\frac{1}{G}\sum\limits_{m\in \cG}\left\| \Delta_m(x^{t+1}, x^t) \right\|^2 - \left\|\nabla f(x^{t+1}) - \nabla f(x^t)\right\|^2\right]\\
        &+ \frac{\omega}{C}\mathbb{E}\left[\left\|\nabla f(x^{t+1}) - \nabla f(x^t)\right\|^2\right]\\
        &\overset{\eqref{eq:global_hessian_variance}, \eqref{eq:f_smooth}}{\leq} \mathbb{E}\left[\|g^t - \nabla f(x^t)\|^2\right]\\
        &+ \left(\frac{\omega}{C}L^2 + \left(1+\frac{\omega}{C}\right)L_{\pm}^2+ \frac{(1+\omega)\cL_{\pm}^2}{bC}\right)\mathbb{E}\left[\|x^{t+1} - x^t\|^2\right].
    \end{align*}
    Next, we estimate $\widetilde T_2$ using Young's inequality and the choice of the clipping level:
    \begin{align*}
        &\widetilde T_2 \leq (1+\beta)\mathbb{E}\left[\left\|g^t - \nabla f(x^{t})\right\|^2\right] + 2(1+\beta^{-1})\mathbb{E}\left[\left\|\nabla f(x^{t+1}) - \nabla f(x^{t})\right\|^2\right]\\
        &+ 2(1+\beta^{-1})\mathbb{E}\left[\left\|\frac{1}{C}\sum\limits_{m\in \set}\clip_{\lambda_{t+1}}\left(\cQ\left(\widehat{\Delta}_m(x^{t+1}, x^t)\right)\right)\right\|^2\mid c_t = 0, \set \cap \cB \neq \varnothing\right]\\
        &\overset{\eqref{eq:f_smooth}}{\leq} (1+\beta)\mathbb{E}\left[\left\|g^t - \nabla f(x^{t})\right\|^2\right] + 2(1+\beta^{-1})\left(L^2\mathbb{E}\left[\|x^{t+1} - x^t\|^2\right] + \mathbb{E}[\lambda_{t+1}^2]\right)\\
        &= (1+\beta)\mathbb{E}\left[\left\|g^t - \nabla f(x^{t})\right\|^2\right]\\
        &+ 2(1+\beta^{-1})\left(L^2 + D_Q^2 \max_{m,i}L_{m,i}^2\right)\mathbb{E}\left[\|x^{t+1} - x^t\|^2\right],
    \end{align*}
    where $\beta > 0$ will be specified later in the proof. Combining the derived upper bounds for $\widehat{T}_2$ and $\widetilde{T}_2$, we get
    \begin{align*}
        T_2 &\leq \left(p_{\cG}^t + (1-p_{\cG}^t)(1+\beta)\right)\mathbb{E}\left[\|g^t - \nabla f(x^t)\|^2\right] \\
        &+  p_{\cG}^t\left(\frac{\omega}{C}L^2 + \left(1+\frac{\omega}{C}\right)L_{\pm}^2 + \frac{(1+\omega)\cL_{\pm}^2}{bC}\right)\mathbb{E}\left[\|x^{t+1} - x^t\|^2\right]\\
        &+ 2(1-p_{\cG}^t)(1+\beta^{-1})\left(L^2 + D_Q^2 \max_{m,i}L_{m,i}^2\right)\mathbb{E}\left[\|x^{t+1} - x^t\|^2\right].
    \end{align*}
    Plugging the obtained bounds for $T_1$ and $T_2$ into \eqref{eq:first_bound_simplified}, we obtain
    \begin{align*}
        \mathbb{E}&\left[\|g^{t+1} - \nabla f(x^{t+1})\|^2\right]\\
        &\leq (1-p)\left(p_{\cG}^t + (1-p_{\cG}^t)(1+\beta)\right)\mathbb{E}\left[\|g^t - \nabla f(x^t)\|^2\right]\\
        &+ p\left(\left(\frac{8 G \mathcal{P}_{\mathcal{G}^t_{\widehat{C}}} c\delta B}{(1-\delta) \widehat{C}} + 2\widetilde{B}\right)  \mathbb{E}\left[\left\|\nabla f\left(x^t\right)\right\|^2  + L^2\left\|x^{t+1}-x^t\right\|^2\right] +\right.\\
        &\left.+\frac{4 G \mathcal{P}_{\mathcal{G}^t_{\widehat{C}}} c\delta \zeta^2}{(1-\delta) \widehat{C}} + \widetilde{\zeta}^2\right)\\
        & + (1-p)p_{\cG}^t\left(\frac{\omega}{C}L^2 + \left(1+\frac{\omega}{C}\right)L_{\pm}^2 + \frac{(1+\omega)\cL_{\pm}^2}{bC}\right)\mathbb{E}\left[\|x^{t+1} - x^t\|^2\right]\\
        &+ 2(1-p)(1-p_{\cG}^t)(1+\beta^{-1})\left(L^2 + D_Q^2 \max_{m,i}L_{m,i}^2\right)\mathbb{E}\left[\|x^{t+1} - x^t\|^2\right].
    \end{align*}
    Taking
    \begin{equation*}
        \beta \eqdef \begin{cases}
            \frac{p}{2(1-p_{\cG}^t)},& \text{if } p_{\cG}^t < 1,\\
            1,& \text{if } p_{\cG}^t = 1,
        \end{cases}
    \end{equation*}
    we ensure that $p_{\cG}^t + (1-p_{\cG}^t)(1+\beta) \leq 1 + \frac{p}{2}$ and $(1-p_{\cG}^t)(1+\beta^{-1}) \leq \frac{(1-p_{\cG}^t)(p + 2(1-p_{\cG}^t))}{p} \leq \frac{3(1-p_{\cG}^t)}{p}$. Using these inequalities and $(1-p)\left(1-\frac{p}{2}\right) \leq 1 - \frac{p}{2}$, we simplify the upper bound for $\mathbb{E}\left[\|g^{t+1} - \nabla f(x^{t+1})\|^2\right]$ as follows:
    \begin{align*}
        \mathbb{E}&\left[\|g^{t+1} - \nabla f(x^{t+1})\|^2\right] \leq \left(1 - \frac{p}{2}\right)\mathbb{E}\left[\|g^t - \nabla f(x^t)\|^2\right]\\
        &+ p\left(\left(\frac{8 G \mathcal{P}_{\mathcal{G}^t_{\widehat{C}}} c\delta B}{(1-\delta) \widehat{C}} + 2\widetilde{B}\right)  \mathbb{E}\left[\left\|\nabla f\left(x^t\right)\right\|^2  + L^2\left\|x^{t+1}-x^t\right\|^2\right] +\right.\\
        &+\left.\frac{4 G \mathcal{P}_{\mathcal{G}^t_{\widehat{C}}} c\delta \zeta^2}{(1-\delta) \widehat{C}} + \widetilde{\zeta}^2\right)\\
        & + (1-p)p_{\cG}^t\left(\frac{\omega}{C}L^2 + \left(1+\frac{\omega}{C}\right)L_{\pm}^2 + \frac{(1+\omega)\cL_{\pm}^2}{bC}\right)\mathbb{E}\left[\|x^{t+1} - x^t\|^2\right]\\
        &+ \frac{6(1-p)(1-p_{\cG}^t)}{p}\left(L^2 + D_Q^2 \max_{m,i}L_{m,i}^2\right)\mathbb{E}\left[\|x^{t+1} - x^t\|^2\right].
    \end{align*}
    Rearranging the terms, we get \eqref{eq:simplified_main_lemma}.
\end{proof}

Then, similarly to the analysis of \gls{Byz-VR-MARINA}, we get the following result.

\begin{theorem}\label{thm:non_convex+}
 Let Assumptions \ref{assm:L-smoothness}, \ref{assm:global}, \ref{assm:local}, \ref{assm:local_smooth_all}, \ref{assm:het}, \ref{assm:bounded-compressor} hold. Set clipping parameter $\lambda_{t+1} = \max_{m,i} L_{m,i} \left\|x^{t+1} - x^t\right\|$. Assume that
$$
0<\gamma \leq \frac{1}{L+\sqrt{A}}, \quad 4\widehat{B}<p,
$$ 
where
\begin{align*}
    A& = \frac{4}{p}\left(  \frac{8p B G \mathcal{P}_{\mathcal{G}^t_{\widehat{C}}} c\delta}{(1-\delta)\widehat{C}} + 6p\widetilde{B} +  \frac{(1-p)p_{\cG}^t\omega}{C} + \frac{6(1-p)(1- p_{\cG}^t)}{p} \right) L^2\\
    &+\frac{4(1-p)p_{\cG}^t}{p}\left( 1+\frac{\omega}{C} \right)L_{ \pm}^2 + \frac{4(1-p)p_{\cG}^t(1+\omega)}{pC} \frac{\mathcal{L}_{ \pm}^2}{b}\\
    &+\frac{24(1-p)(1-p_{\cG}^t)}{p^2} \left(D_Q \max_{m,i} L_{m,i} \right)^2,
\end{align*}
\textbf{\begin{align*}
    \widehat{B} = \frac{8 G \mathcal{P}_{\mathcal{G}^t_{\widehat{C}}} c\delta Bp}{(1-\delta)\widehat{C}} + 6p\widetilde{B}, \quad \widehat{D} = \frac{4 G \mathcal{P}_{\mathcal{G}^t_{\widehat{C}}} c\delta p}{(1-\delta)\widehat{C}} + p\widetilde{D},
\end{align*}}
and 
\begin{align*}
 \mathcal{P}_{\mathcal{G}^t_C} &=   \frac{C}{Mp_G} \cdot \sum_{(1-\delta)C\leq m^\prime \leq C} \left(\left(\begin{array}{l}
G-1 \\
t-1
\end{array}\right)\left(\begin{array}{l}
M-G \\
C-m^\prime
\end{array}\right) \left(\left(\begin{array}{l}
n \\
C
\end{array}\right)\right)^{-1} \right),\\
p_{\cG}^t &=  \PP\{\set \cap \cB = \varnothing \mid c_t = 0\}  = \frac{(G-C+1)(G-C+2)\cdot\ldots\cdot (n-C)}{(G+1)(G+2)\cdot \ldots \cdot n}.
\end{align*}
Then for all $T \geq 0$ the iterates produced by \algname{Byz-VR-MARINA+} (Algorithm \ref{alg:byz_vr_marina+}) satisfy
$$
\mathbb{E}\left[\left\|\nabla f\left(\widehat{x}^T\right)\right\|^2\right] \leq \frac{2 \Phi^{(0)}}{\gamma\left(1-\frac{4\widehat{B}}{p}\right)(T+1)}+\frac{2 \widehat{D} \zeta^2}{p-4 \widehat{B}},
$$
where $\widehat{x}^T$ is chosen uniformly at random from $x^0, x^1, \ldots, x^T$, and $\Phi^{(0)}=$ $f\left(x^0\right)-f^{\star}+\frac{\gamma}{p}\left\|g^0-\nabla f\left(x^0\right)\right\|^2$ . 
\end{theorem}
\begin{proof}
    The proof is analogous to the proof of Theorem~\ref{them:1}.
\end{proof}

\begin{theorem}\label{thm:PL+}
Let Assumptions  \ref{assm:bounded-compressor}, \ref{assm:L-smoothness}, \ref{assm:global}, \ref{assm:local}, \ref{assm:local_smooth_all}, \ref{assm:het}, \ref{assm:PL} hold. Set $\lambda_{t+1} = \max_{m,i} L_{m,i} \left\|x^{t+1} - x^t\right\|$. Assume that
$$
0<\gamma \leq \min \left\{\frac{1}{L+\sqrt{2 A}} \right\}, \quad 8\widehat{B}<p,
$$
where
\begin{align*}
    A& = \frac{4}{p}\left(  \frac{8p B G \mathcal{P}_{\mathcal{G}^t_{\widehat{C}}} c\delta}{(1-\delta)\widehat{C}} + 6p\widetilde{B} +  \frac{(1-p)p_{\cG}^t\omega}{C} + \frac{6(1-p)(1- p_{\cG}^t)}{p} \right) L^2\\
    &+\frac{4(1-p)p_{\cG}^t}{p}\left( 1+\frac{\omega}{C} \right)L_{ \pm}^2 + \frac{4(1-p)p_{\cG}^t(1+\omega)}{pC} \frac{\mathcal{L}_{ \pm}^2}{b}\\
    &+\frac{24(1-p)(1-p_{\cG}^t)}{p^2} \left(D_Q \max_{m,i} L_{m,i} \right)^2,
\end{align*}
\textbf{\begin{align*}
    \widehat{B} = \frac{8 G \mathcal{P}_{\mathcal{G}^t_{\widehat{C}}} c\delta Bp}{(1-\delta)\widehat{C}} + 6p\widetilde{B}, \quad \widehat{D} = \frac{4 G \mathcal{P}_{\mathcal{G}^t_{\widehat{C}}} c\delta p}{(1-\delta)\widehat{C}} + p\widetilde{D},
\end{align*}}
and 
\begin{align*}
 \mathcal{P}_{\mathcal{G}^t_C} &=   \frac{C}{Mp_G} \cdot \sum_{(1-\delta)C\leq m^\prime \leq C} \left(\left(\begin{array}{l}
G-1 \\
t-1
\end{array}\right)\left(\begin{array}{l}
M-G \\
C-m^\prime
\end{array}\right) \left(\left(\begin{array}{l}
n \\
C
\end{array}\right)\right)^{-1} \right),\\
p_{\cG}^t &=  \PP\{\set \cap \cB = \varnothing \mid c_t = 0\}  = \frac{(G-C+1)(G-C+2)\cdot\ldots\cdot (n-C)}{(G+1)(G+2)\cdot \ldots \cdot n}.
\end{align*} 
Then for all $T \geq 0$ the iterates produced by \algname{Byz-VR-MARINA+} (Algorithm \ref{alg:byz_vr_marina+}) satisfy
$$
\mathbb{E}\left[f\left(x^T\right)-f\left(x^{\star}\right)\right] \leq\left(1-\rho\right)^T \Phi^{(0)}+\frac{2\widehat{D}\zeta^2}{p\rho},
$$
where $\rho = \min\left[\gamma\mu\left(1-\frac{8\widehat{B}}{p}\right), \frac{p}{4}\right]$ and $\Phi^{(0)}=$ $f\left(x^0\right)-f^{\star}+\frac{2\gamma}{p}\left\|g^0-\nabla f\left(x^0\right)\right\|^2$.
\end{theorem}
\begin{proof}
    The proof is analogous to the proof of Theorem~\ref{them:2}.
\end{proof}

\subsection{Discussion of the results}

\paragraph{Improved neighborhood term and bound on $\delta$.} The key property of \algname{Byz-VR-MARINA+} is its better neighborhood terms, and maximal allowed fraction of Byzantine workers $\delta$ in comparison to \gls{Byz-VR-MARINA}. To illustrate it, consider the non-P\L setting (the discussion for the P\L~case is similar). For both algorithms, the neighborhood term in the convergence bounds equals $\cO\left(\frac{\widehat{D}\zeta^2}{p - 4\widehat{B}}\right)$, but corresponding constants $\widehat{B}$ and $\widehat{D}$ are different:
\begin{align*}
    &\widehat{B} = 2 \frac{ \delta\mathcal{P}_{\mathcal{G}^t_{\widehat{C}}} }{1-\delta} B\left(\frac{12cG}{\widehat C} + p \right) + 6\widetilde{B},\\ 
    &\widehat{D} = 2 \frac{ \delta\mathcal{P}_{\mathcal{G}^t_{\widehat{C}}} }{1-\delta} \left(\frac{6cG}{\widehat C} + p \right) + \widetilde{D} \quad \text{for\quad\gls{Byz-VR-MARINA}},\\
    &\widehat{B} = \frac{8 G \mathcal{P}_{\mathcal{G}^t_{\widehat{C}}} c\delta Bp}{(1-\delta)\widehat{C}} + 6p\widetilde{B}\\
    & \widehat{D} = \frac{4 G \mathcal{P}_{\mathcal{G}^t_{\widehat{C}}} c\delta p}{(1-\delta)\widehat{C}} + p\widetilde{D} \quad \text{for\quad\algname{Byz-VR-MARINA+}.}
\end{align*}
For the simplicity of the comparison, consider the case of $\widehat{C} = M$. Then, $\mathcal{P}_{\mathcal{G}^t_{\widehat{C}}} = 1$ and
\begin{gather*}
    \widehat{B} = \Theta\left(c\delta B\right) \quad \text{and} \quad \widehat{D} = \Theta(c\delta) \quad \text{for\quad\gls{Byz-VR-MARINA}},\\
    \widehat{B} = \Theta\left(c\delta Bp\right) \quad \text{and} \quad \widehat{D} = \Theta(c\delta p) \quad \text{for\quad\algname{Byz-VR-MARINA+},}
\end{gather*}
implying that the neighborhood term for \algname{Byz-VR-MARINA+} is $\nicefrac{1}{p}$ times smaller than the neighborhood term for \gls{Byz-VR-MARINA}. Moreover, the restriction $4\widehat{B} < p$ used in the analysis of both methods implies
\begin{gather*}
    c\delta = \cO\left(\frac{1}{Bp}\right) \quad \text{for\quad\gls{Byz-VR-MARINA}},\\
    c\delta = \cO\left(\frac{1}{B}\right) \quad \text{for\quad\algname{Byz-VR-MARINA+},}
\end{gather*}
i.e., the result for \algname{Byz-VR-MARINA+} allows $(\nicefrac{1}{p})$-times more Byzantine workers when $B > 0$. We emphasize that the neighborhood term and the bound on $\delta$ in the results for \algname{Byz-VR-MARINA+} cannot be improved up to the numerical factors \citep{allouah2024robust}.

\paragraph{Comparison of stepsizes when $\widehat{C} = M$ and $C = 1$.} For simplicity, to compare the stepsize restrictions for \gls{Byz-VR-MARINA} and \algname{Byz-VR-MARINA+}, we consider the case when $\widehat{C} = M$ and $C = 1$. Moreover, let us assume that $b=1$ and let us ignore the differences between smoothness constants and replace them with their upper bound $\cL$ from Assumption~\ref{assm:smoothness_simplified}. Then, for both methods, the results in the non-P\L setting (the discussion for the P\L~case is similar) with $B = 0$ hold for $0<\gamma \leq \nicefrac{1}{\cL(1+\sqrt{A})}$, where
\begin{gather*}
    A = \Theta\left(\frac{1}{p}\left( 1 + \omega + \frac{(1+\omega)c\delta}{p}\right) + \frac{\deltar (1 + F_{\cA}^2 D_Q^2)}{p^2}\right)\quad \text{for\quad\gls{Byz-VR-MARINA}},\\
    A = \Theta\left(\frac{1+\omega}{p} + \frac{\deltar D_Q^2}{p^2}\right)\quad \text{for\quad\algname{Byz-VR-MARINA+}},
\end{gather*}
where we use $p_{G} = \nicefrac{G}{n} = 1 - \deltar$, $\cP_{\cG_C^t} = \nicefrac{1}{G}$, $p_{\cG}^t = \nicefrac{G}{n} = 1 - \deltar$. That is, the result for \algname{Byz-VR-MARINA+} allows using larger stepsizes than in \gls{Byz-VR-MARINA} (though the methods are equivalent when $C = 1$). A similar comparison holds for small enough $C$ as well. Therefore, we recommend using \algname{Byz-VR-MARINA+} instead of \gls{Byz-VR-MARINA} when $C$ is small. We also highlight that the result for \algname{Byz-VR-MARINA+} does not require Assumption~\ref{assm:bounded-aggr}.

\clearpage

\section{Analysis Without Full-Batch Gradient Computations}\label{appendix:no_full_grads}

In this section, we consider versions of \gls{Byz-VR-MARINA-PP} and \algname{Byz-VR-MARINA-PP+} that do not use full-batch gradient computations at all -- see Algorithms~\ref{alg:byz_vr_marina_no_full} and \ref{alg:byz_vr_marina+_no_full}. These variants of \gls{Byz-VR-MARINA-PP} and \algname{Byz-VR-MARINA-PP+} use $b'$-size mini-batched estimator $\widetilde \nabla f_m(x^{t+1})$ when $c_t = 1$ for every $m \in \cG \cap \set$ in line~\ref{line:g_i^t+1} and are identical to their original versions in all other steps/computations. This modification reduces the computation cost of iterations when $c_t = 1$, making the methods more practical.

\begin{algorithm*}[h]
   \caption{\gls{Byz-VR-MARINA-PP} without full-batch gradient computations}\label{alg:byz_vr_marina_no_full}
\begin{algorithmic}[1]
   \STATE {\bfseries Input:} vectors $x^0, g^0 \in \R^d$, stepsize $\gamma$, mini-batch size $b$, mini-batch size $b'$, probability $p\in(0,1]$, number of iterations $K$, $(\delta,c)$-\texttt{ARAgg}, clients' sample size $1 \leq C \leq \widehat{C} \leq n$, clipping coefficients $\{\alpha_{t}\}_{t\geq 1}$ 
   \FOR{$k=0,1,\ldots,K-1$}
   \STATE Get a sample from Bernoulli distribution with parameter $p$: $c_t \sim \text{Be}(p)$
   \STATE Sample the set of clients $\set \subseteq [n]$, $|\set| = C$ if $c_t = 0$; otherwise $|\set| = \widehat{C}$
   \STATE Broadcast $g^t$, $c_t$ to all workers
   \FOR{$m \in \cG \cap \set$ in parallel} 
   \STATE $x^{t+1} = x^t - \gamma g^t$ and $\lambda_{t+1} = \alpha_{t+1}\|x^{t+1} - x^t\|$
   \STATE  Set $g_m^{t+1} = \begin{cases} \widetilde\nabla f_m(x^{t+1}),& \text{if } c_t = 1,\\ g^t + \clip_{\lambda_{t+1}}\left(\cQ\left(\widehat{\Delta}_m(x^{t+1}, x^t)\right)\right),& \text{otherwise,}\end{cases}$\\ where $\widetilde\nabla f_m(x^{t+1})$ is a $b'$-size mini-batched estimator of $\nabla f_m(x^{t+1})$, $\widehat{\Delta}_m(x^{t+1}, x^t)$ is a $b$-size mini-batched estimator of $\nabla f_m(x^{t+1}) - \nabla f_m(x^t)$, $\cQ(\cdot)$ for $m\in\cG \cap \set$ are computed independently
   \ENDFOR
   \STATE\label{line:g^t+1-full} $g^{t+1} = \begin{cases} \texttt{ARAgg}\left(\{g_m^{t+1}\}_{m\in \set}\right),& \text{if } c_t = 1, \\
       g^t + \texttt{ARAgg}\left(\left\{\clip_{\lambda_{t+1}}\left(\cQ\left(\widehat{\Delta}_m(x^{t+1}, x^t)\right)\right)\right\}_{m\in \set}\right),&\text{otherwise}\end{cases}$
   \ENDFOR
\end{algorithmic}
\end{algorithm*}

\begin{algorithm*}[h]
   \caption{\algname{Byz-VR-MARINA-PP+}: without full-batch gradient computations}\label{alg:byz_vr_marina+_no_full}
\begin{algorithmic}[1]
   \STATE {\bfseries Input:} vectors $x^0, g^0 \in \R^d$, stepsize $\gamma$, mini-batch size $b$, mini-batch size $b'$, probability $p\in(0,1]$, number of iterations $K$, $(\delta,c)$-\texttt{ARAgg}, clients' sample size $1 \leq C \leq \widehat{C} \leq n$, clipping coefficients $\{\alpha_{t}\}_{t\geq 1}$ 
   \FOR{$k=0,1,\ldots,K-1$}
   \STATE Get a sample from Bernoulli distribution with parameter $p$: $c_t \sim \text{Be}(p)$
   \STATE Sample the set of clients $\set \subseteq [n]$, $|\set| = C$ if $c_t = 0$; otherwise $|\set| = \widehat{C}$
   \STATE Broadcast $g^t$, $c_t$ to all workers
   \FOR{$m \in \cG \cap \set$ in parallel} 
   \STATE $x^{t+1} = x^t - \gamma g^t$ and $\lambda_{t+1} = \alpha_{t+1}\|x^{t+1} - x^t\|$
   \STATE\label{line:g_i^t+1}  Set $g_m^{t+1} = \begin{cases} \widetilde\nabla f_m(x^{t+1}),& \text{if } c_t = 1,\\ g^t + \clip_{\lambda_{t+1}}\left(\cQ\left(\widehat{\Delta}_m(x^{t+1}, x^t)\right)\right),& \text{otherwise,}\end{cases}$\\ where $\widetilde\nabla f_m(x^{t+1})$ is a $b'$-size mini-batched estimator of $\nabla f_m(x^{t+1})$, $\widehat{\Delta}_m(x^{t+1}, x^t)$ is a $b$-size mini-batched estimator of $\nabla f_m(x^{t+1}) - \nabla f_m(x^t)$, $\cQ(\cdot)$ for $m\in\cG \cap \set$ are computed independently
   \ENDFOR
   \STATE\label{line:aggregation+_f} $g^{t+1} = \begin{cases} \texttt{ARAgg}\left(\{g_m^{t+1}\}_{m\in \set}\right),& \text{if } c_t = 1, \\
       g^t + \frac{1}{C}\sum\limits_{m\in \set}\clip_{\lambda_{t+1}}\left(\cQ\left(\widehat{\Delta}_m(x^{t+1}, x^t)\right)\right),&\text{otherwise}\end{cases}$
   \ENDFOR
\end{algorithmic}
\end{algorithm*}

However, our analysis of \gls{Byz-VR-MARINA-PP}/\algname{Byz-VR-MARINA-PP+} without full-batch gradient computations requires the following additional assumption.

\begin{assumption}\label{assm:bounded_variance}
    We assume that there exist $\sigma \geq 0$ such that for all $x\in \R^d$ and $i\in [n]$
    \begin{equation}
        \mathbb{E}\left[\|\widetilde\nabla f_m(x) - \nabla f_m(x)\|^2\right] \leq \frac{\sigma^2}{b'}, \label{eq:bounded_variance}
    \end{equation}
    where $\widetilde\nabla f_m(x)$ is an unbiased $b'$-size mini-batched estimator of $\nabla f_m(x)$.
\end{assumption}

In particular, when $\frac{1}{n}\sum_{i=1}^n \|\nabla f_{m,i}(x) - \nabla f_m(x)\|^2 \leq \sigma$, which is a standard assumption for variance-reduced methods without full-batch gradient computations \citep{cutkosky2019momentum, li2021page, gorbunov2021marina}, estimator $\widetilde \nabla f_m(x) = \frac{1}{b'}\sum_{i = 1}^{b'} \nabla f_{m,\xi_m^i}(x)$ with $\{\xi_i^j\}_{m\in [M], i\in [n]}$ being i.i.d.\ samples from the uniform distribution over $[n]$ satisfies \eqref{eq:bounded_variance}. Assumption~\ref{assm:bounded_variance} is also standard for general stochastic optimization \citep{nemirovski2009robust, ghadimi2013stochastic}.

\subsection{New Lemma}

The main change in the analysis is related to Lemma~\ref{lemma:full_aggr} since it is the only lemma that relies on the full-batch gradient computation. Nevertheless, it can be easily generalized to the case of Algorithms~\ref{alg:byz_vr_marina_no_full} and \ref{alg:byz_vr_marina+_no_full}, as shown in the next result.

\begin{lemma} 
\label{lemma:full_aggr_no_full_grad}
Let Assumptions~\ref{assm:L-smoothness}, \ref{assm:het}, \ref{assm:bounded_variance} hold and Aggregation Operator ($\texttt{ARAgg}$) satisfy Definition~\ref{def:aragg}. Then for all $t\geq 0$ the iterates produced by \gls{Byz-VR-MARINA-PP}/\algname{Byz-VR-MARINA-PP+} (Algorithms~\ref{alg:byz_vr_marina_no_full} and \ref{alg:byz_vr_marina+_no_full}) satisfy
\begin{align*}
T_1 &=  \mathbb{E}\left[\mathbb{E}_t\left[\left\|\texttt{ARAgg}\left(\{g_m^{t+1}\}_{m\in \set}\right) - \nabla f(x^{t+1})\right\|^2\right]\mid [1] \right]\\ 
&\leq \left(\frac{8 G \mathcal{P}_{\mathcal{G}^t_{\widehat{C}}} c\delta B}{(1-\delta) \widehat{C}} + 2\widetilde{B}\right)  \mathbb{E}\left[\left\|\nabla f\left(x^t\right)\right\|^2  + L^2\left\|x^{t+1}-x^t\right\|^2\right] \\
&\quad + \frac{4 G \mathcal{P}_{\mathcal{G}^t_{\widehat{C}}} c\delta B}{(1-\delta) \widehat{C}}\zeta^2 + \widetilde{\zeta}^2  + \left(\frac{\cP_{\cG_{\widehat{C}}^t}G}{(1-\delta)^2\widehat{C}^2} + 4c\delta\right)\frac{\sigma^2}{b'},
\end{align*}
where $\widetilde{B} \eqdef 0$ and $\widetilde{\zeta}^2 \eqdef 0$ when $\widehat{C} = M$, and $\widetilde{B} \eqdef \frac{\cP_{\cG_{\widehat{C}}^t}GB}{(1-\delta)\widehat{C}}$ and $\widetilde{\zeta}^2 \eqdef \frac{\cP_{\cG_{\widehat{C}}^t}G\zeta^2}{(1-\delta)\widehat{C}}$ when $\widehat{C} < M$.
\end{lemma}
\begin{proof}
Using the definition of aggregation operator, we have
\begin{align}
T_1 &= \mathbb{E}\left[\mathbb{E}_t\left[\left\|\texttt{ARAgg}\left(\{g_m^{t+1}\}_{m\in \set}\right) - \nabla f(x^{t+1})\right\|^2\right]\mid [1]\right] \notag\\
&\overset{\eqref{eq:yung-1}}{\leq} \mathbb{E}\left[\mathbb{E}_t\left[\left\|\texttt{ARAgg}\left(\{g_m^{t+1}\}_{m\in \set}\right) - \frac{1}{G_{\widehat{C}}^t}\sum\limits_{m \in \cG_{\widehat{C}}^t}\widetilde\nabla f_m(x^{t+1})\right\|^2\right]\mid [1]\right] \notag\\
&\quad + \mathbb{E}\left[\mathbb{E}_t\left[\left\|\frac{1}{G_{\widehat{C}}^t}\sum\limits_{m \in \cG_{\widehat{C}}^t}\widetilde\nabla f_m(x^{t+1}) - \nabla f(x^{t+1})\right\|^2\right]\mid [1]\right]. \label{eq:djkcdvdcdnjdnfkjdn}
\end{align}
To proceed, we estimate the second term in the right-hand side of the above inequality first. From variance decomposition, we have
\begin{align}
    \mathbb{E}&\left[\mathbb{E}_t\left[\left\|\frac{1}{G_{\widehat{C}}^t}\sum\limits_{m \in \cG_{\widehat{C}}^t}\widetilde\nabla f_m(x^{t+1}) - \nabla f(x^{t+1})\right\|^2\right]\mid [1]\right] \notag\\
    &= \mathbb{E}\left[\mathbb{E}_t\left[\left\|\frac{1}{G_{\widehat{C}}^t}\sum\limits_{m \in \cG_{\widehat{C}}^t}(\widetilde\nabla f_m(x^{t+1}) - \nabla f_m(x^{t+1}))\right\|^2\right]\mid [1]\right] \notag\\
    &\quad + \mathbb{E}\left[\mathbb{E}_t\left[\left\|\frac{1}{G_{\widehat{C}}^t}\sum\limits_{m \in \cG_{\widehat{C}}^t}\nabla f_m(x^{t+1}) - \nabla f(x^{t+1})\right\|^2\right]\mid [1]\right]. \label{eq:djvbcdkcndkfn}
\end{align}
The choice of $\widehat{C}$ implies that $G_{\widehat{C}}^t \geq (1-\delta)\widehat{C}$. Moreover, due to the independence of stochastic gradient computations on different workers, we have
\begin{align*}
    \mathbb{E}&\left[\mathbb{E}_t\left[\left\|\frac{1}{G_{\widehat{C}}^t}\sum\limits_{m \in \cG_{\widehat{C}}^t}(\widetilde\nabla f_m(x^{t+1}) - \nabla f_m(x^{t+1}))\right\|^2\right]\mid [1]\right]\\
    &\leq \frac{1}{(1-\delta)^2\widehat{C}^2}\mathbb{E}\left[\mathbb{E}_t\left[\left\|\sum\limits_{m \in \cG_{\widehat{C}}^t}(\widetilde\nabla f_m(x^{t+1}) - \nabla f_m(x^{t+1}))\right\|^2\right]\mid [1]\right]\\
    &= \frac{1}{(1-\delta)^2\widehat{C}^2}\mathbb{E}\left[\sum\limits_{m \in \cG_{\widehat{C}}^t}\mathbb{E}_t\left[\left\|\widetilde\nabla f_m(x^{t+1}) - \nabla f_m(x^{t+1})\right\|^2\right]\mid [1]\right]\\
    &= \frac{\cP_{\cG_{\widehat{C}}^t}}{(1-\delta)^2\widehat{C}^2}\sum\limits_{m \in \cG}\mathbb{E}\left[\left\|\widetilde\nabla f_m(x^{t+1}) - \nabla f_m(x^{t+1})\right\|^2\right] \overset{\eqref{eq:bounded_variance}}{\leq} \frac{\cP_{\cG_{\widehat{C}}^t}G\sigma^2}{(1-\delta)^2\widehat{C}^2b'}.
\end{align*}
Next, since $\frac{1}{G_{\widehat{C}}^t}\sum\limits_{m \in \cG_{\widehat{C}}^t}\nabla f_m(x^{t+1}) = \nabla f(x^{t+1})$ with probability $1$ when $\widehat{C} = M$, we can estimate the last term in \eqref{eq:djvbcdkcndkfn} as
\begin{align*}
    \mathbb{E}&\left[\mathbb{E}_t\left[\left\|\frac{1}{G_{\widehat{C}}^t}\sum\limits_{m \in \cG_{\widehat{C}}^t}\nabla f_m(x^{t+1}) - \nabla f(x^{t+1})\right\|^2\right]\mid [1]\right] \\
    &\leq \begin{cases}
        0,& \text{if } \widehat{C} = M\\ \mathbb{E}\left[\frac{1}{G_{\widehat{C}}^t}\sum\limits_{m \in \cG_{\widehat{C}}^t}\mathbb{E}_t\left[\left\|\nabla f_m(x^{t+1}) - \nabla f(x^{t+1})\right\|^2\right]\mid [1]\right], & \text{if } \widehat{C} < M
    \end{cases}\\
    &\leq \begin{cases}
        0,& \text{if } \widehat{C} = M\\ \frac{\cP_{\cG_{\widehat{C}}^t}}{(1-\delta)\widehat{C}}\sum\limits_{m \in \cG}\mathbb{E}\left[\left\|\nabla f_m(x^{t+1}) - \nabla f(x^{t+1})\right\|^2\right], & \text{if } \widehat{C} < M
    \end{cases}\\
    &\stackrel{(\text{As.~\ref{assm:het}})}{\leq} \begin{cases}
        0,& \text{if } \widehat{C} = M\\ \frac{\cP_{\cG_{\widehat{C}}^t}G}{(1-\delta)\widehat{C}}\left(B\mathbb{E}\left[\|\nabla f(x^{t+1})\|^2\right] + \zeta^2\right), & \text{if } \widehat{C} < M
    \end{cases}\\
    &= \widetilde{B}\mathbb{E}\left[\|\nabla f(x^{t+1})\|^2\right] + \widetilde{\zeta}^2,
\end{align*}
where
\begin{equation*}
    \widetilde{B} \eqdef \begin{cases}
        0,& \text{if } \widehat{C} = M,\\ \frac{\cP_{\cG_{\widehat{C}}^t}GB}{(1-\delta)\widehat{C}},& \text{if } \widehat{C} < M,
    \end{cases} \quad \text{and}\quad  \widetilde{\zeta}^2 \eqdef \begin{cases}
        0,& \text{if } \widehat{C} = M,\\ \frac{\cP_{\cG_{\widehat{C}}^t}G\zeta^2}{(1-\delta)\widehat{C}},& \text{if } \widehat{C} < M.
    \end{cases}
\end{equation*}
Plugging the derived bounds in \eqref{eq:djvbcdkcndkfn}, we get
\begin{align*}
    \mathbb{E}\left[\mathbb{E}_t\left[\left\|\frac{1}{G_{\widehat{C}}^t}\sum\limits_{m \in \cG_{\widehat{C}}^t}\widetilde\nabla f_m(x^{t+1}) - \nabla f(x^{t+1})\right\|^2\right]\mid [1]\right] &\leq \frac{\cP_{\cG_{\widehat{C}}^t}G\sigma^2}{(1-\delta)^2\widehat{C}^2b'} \\
    &\quad + \widetilde{B}\mathbb{E}\left[\|\nabla f(x^{t+1})\|^2\right] + \widetilde{\zeta}^2.
\end{align*}
Using the above bound in \eqref{eq:djkcdvdcdnjdnfkjdn}, we continue the estimation of $T_1$ as follows:
    \begin{align*}
&T_1  \stackrel{(\text{Def.~\ref{def:aragg}})}{\leq}\mathbb{E}\left[\frac{c\delta}{G_{\widehat{C}}^t(G_{\widehat{C}}^t-1)} \sum_{\substack{i, l \in \mathcal{G}^t_{\widehat{C}} \\
m \neq l}} \mathbb{E}_t\left[\left\|\widetilde\nabla f_m\left(x^{t+1}\right)-\widetilde\nabla f_l\left(x^{t+1}\right)\right\|^2\mid [1]\right]\right] \\
    &\quad + \frac{\cP_{\cG_{\widehat{C}}^t}G\sigma^2}{(1-\delta)^2\widehat{C}^2b'} + \widetilde{B}\mathbb{E}\left[\|\nabla f(x^{t+1})\|^2\right] + \widetilde{\zeta}^2\\
& = \mathbb{E}\left[\frac{c\delta}{G_{\widehat{C}}^t(G_{\widehat{C}}^t-1)} \sum_{\substack{i, l \in \mathcal{G}^t_{\widehat{C}} \\
m \neq l}} \mathbb{E}_t\left[\left\|\nabla f_m\left(x^{t+1}\right)-\nabla f_l\left(x^{t+1}\right)\right\|^2\mid [1]\right]\right]\\
&\quad  + \mathbb{E}\left[\frac{c\delta}{G_{\widehat{C}}^t(G_{\widehat{C}}^t-1)} \sum_{\substack{i, l \in \mathcal{G}^t_{\widehat{C}} \\
m \neq l}} \mathbb{E}_t\left[\left\|\widetilde\nabla f_m\left(x^{t+1}\right) - \nabla f_m(x^{t+1})\right.\right.\right.-\\
&\left.\left.\left.-\widetilde\nabla f_l\left(x^{t+1}\right) + \nabla f_l\left(x^{t+1}\right)\right\|^2\mid [1]\right]\right]\\
&\quad + \frac{\cP_{\cG_{\widehat{C}}^t}G\sigma^2}{(1-\delta)^2\widehat{C}^2b'} + \widetilde{B}\mathbb{E}\left[\|\nabla f(x^{t+1})\|^2\right] + \widetilde{\zeta}^2,
\end{align*}
where in the last inequality, we use the conditional independence of gradients $\{\widetilde \nabla f_m(x^{t+1})\}_{i \ \in \cG_{\widehat C}^t}$ for fixed $x^{t+1}$. Next, using Young's inequality, we derive
\begin{align*}
T_1 & \stackrel{(\ref{eq:yung-1})}{\leq} \mathbb{E}\left[\frac{c\delta}{G_{\widehat{C}}^t(G_{\widehat{C}}^t-1)}\sum_{\substack{i, l \in \mathcal{G}^t_{\widehat{C}} \\
m \neq l}} \mathbb{E}\left[2\left\|\nabla f_m\left(x^{t+1}\right)-\nabla f\left(x^{t+1}\right)\right\|^2\mid [1]\right]\right]\\
&\quad + \mathbb{E}\left[\frac{c\delta}{G_{\widehat{C}}^t(G_{\widehat{C}}^t-1)}\sum_{\substack{i, l \in \mathcal{G}^t_{\widehat{C}} \\
m \neq l}} \mathbb{E}\left[2\left\|\nabla f_l\left(x^{t+1}\right)-\nabla f\left(x^{t+1}\right)\right\|^2\mid [1]\right]\right]\\
&\quad + \mathbb{E}\left[\frac{c\delta}{G_{\widehat{C}}^t(G_{\widehat{C}}^t-1)}\sum_{\substack{i, l \in \mathcal{G}^t_{\widehat{C}} \\
m \neq l}} \mathbb{E}\left[2\left\|\widetilde\nabla f_m\left(x^{t+1}\right)-\nabla f_m\left(x^{t+1}\right)\right\|^2\mid [1]\right]\right]\\
&\quad + \mathbb{E}\left[\frac{c\delta}{G_{\widehat{C}}^t(G_{\widehat{C}}^t-1)}\sum_{\substack{i, l \in \mathcal{G}^t_{\widehat{C}} \\
m \neq l}} \mathbb{E}\left[2\left\|\widetilde\nabla f_l\left(x^{t+1}\right)-\nabla f_l\left(x^{t+1}\right)\right\|^2\mid [1]\right]\right]\\
&\quad + \frac{\cP_{\cG_{\widehat{C}}^t}G\sigma^2}{(1-\delta)^2\widehat{C}^2b'} + \widetilde{B}\mathbb{E}\left[\|\nabla f(x^{t+1})\|^2\right] + \widetilde{\zeta}^2\\
&\overset{\eqref{eq:bounded_variance}}{\leq} \mathbb{E}\left[\frac{c\delta}{G^t_{\widehat{C}}} \sum_{m \in \mathcal{G}^t_{\widehat{C}}} 4\mathbb{E}_t\left[\left\|\nabla f_m\left(x^{t+1}\right)-\nabla f\left(x^{t+1}\right)\right\|^2\mid [1]\right]\right] + \frac{4c\delta\sigma^2}{b'}\\
&\quad + \frac{\cP_{\cG_{\widehat{C}}^t}G\sigma^2}{(1-\delta)^2\widehat{C}^2b'} + \widetilde{B}\mathbb{E}\left[\|\nabla f(x^{t+1})\|^2\right] + \widetilde{\zeta}^2\\
&\leq \frac{\mathcal{P}_{\mathcal{G}^t_{\widehat{C}}} c\delta}{(1-\delta) \widehat{C}} \sum_{m \in \mathcal{G}} 4\mathbb{E}_t\left[\left\|\nabla f_m\left(x^{t+1}\right)-\nabla f\left(x^{t+1}\right)\right\|^2 \right] + \frac{4c\delta\sigma^2}{b'}\\
&\quad + \frac{\cP_{\cG_{\widehat{C}}^t}G\sigma^2}{(1-\delta)^2\widehat{C}^2b'} + \widetilde{B}\mathbb{E}\left[\|\nabla f(x^{t+1})\|^2\right] + \widetilde{\zeta}^2\\
&\stackrel{(\text{As.~\ref{assm:het}})}{\leq} \left(\frac{4 G \mathcal{P}_{\mathcal{G}^t_{\widehat{C}}} c\delta B}{(1-\delta) \widehat{C}} + \widetilde{B}\right)  \mathbb{E}\left[\left\|\nabla f\left(x^{t+1}\right)\right\|^2\right] + \frac{4 G \mathcal{P}_{\mathcal{G}^t_{\widehat{C}}} c\delta B}{(1-\delta) \widehat{C}}\zeta^2 + \widetilde{\zeta}^2\\
&\quad + \left(\frac{\cP_{\cG_{\widehat{C}}^t}G}{(1-\delta)^2\widehat{C}^2} + 4c\delta\right)\frac{\sigma^2}{b'}
\end{align*}
Thus, we have
\begin{align*}
    T_1&\stackrel{(\ref{eq:yung-1})}{\leq} \left(\frac{8 G \mathcal{P}_{\mathcal{G}^t_{\widehat{C}}} c\delta B}{(1-\delta) \widehat{C}} + 2 \widetilde{B}\right) \mathbb{E}\left[\left\|\nabla f\left(x^t\right)\right\|^2 + \left\|\nabla f\left(x^{t+1}\right)-\nabla f\left(x^t\right)\right\|^2\right]\\
&\quad + \frac{4 G \mathcal{P}_{\mathcal{G}^t_{\widehat{C}}} c\delta B}{(1-\delta) \widehat{C}}\zeta^2 + \widetilde{\zeta}^2  + \left(\frac{\cP_{\cG_{\widehat{C}}^t}G}{(1-\delta)^2\widehat{C}^2} + 4c\delta\right)\frac{\sigma^2}{b'}\\
&\leq  \left(\frac{8 G \mathcal{P}_{\mathcal{G}^t_{\widehat{C}}} c\delta B}{(1-\delta) \widehat{C}} + 2 \widetilde{B}\right) \mathbb{E}\left[\left\|\nabla f\left(x^t\right)\right\|^2 + L^2\left\|x^{t+1}- x^t\right\|^2\right]\\
&\quad + \frac{4 G \mathcal{P}_{\mathcal{G}^t_{\widehat{C}}} c\delta B}{(1-\delta) \widehat{C}}\zeta^2 + \widetilde{\zeta}^2  + \left(\frac{\cP_{\cG_{\widehat{C}}^t}G}{(1-\delta)^2\widehat{C}^2} + 4c\delta\right)\frac{\sigma^2}{b'},
\end{align*}
which concludes the proof.
\end{proof}

\subsection{Main results for Byz-VR-MARINA without Full-Batch gradient computations}

\subsubsection{General results}

All the lemmas derived in Appendix~\ref{appendix:technical_lemmas_general} hold for Algorithm~\ref{alg:byz_vr_marina_no_full} as well except Lemma~\ref{lemma:full_aggr}, which can be replaced with Lemma~\ref{lemma:full_aggr_no_full_grad}, and Lemma~\ref{lemma:final_lemma} that has the following analog.
\begin{lemma}
\label{lemma:final_lemma_no_full_grad}
    Let Assumptions ~\ref{assm:bounded-aggr}, \ref{assm:L-smoothness}, \ref{assm:global}, \ref{assm:local}, \ref{assm:het}, \ref{assm:bounded_variance} hold and Compression Operator satisfy Definition~\ref{def:Q}.  Also, let us introduce the notation
\begin{align*}\texttt{ARAgg}_Q^{t+1} = \texttt{ARAgg}\left(\clip_{\lambda_{t+1}}\left(\cQ\left(\widehat{\Delta}_1(x^{t+1}, x^t)\right)\right),\right.\\
\left.\ldots, \clip_{\lambda_{t+1}}\left(\cQ\left(\widehat{\Delta}_C(x^{t+1}, x^t)\right)\right)\right).\end{align*}
Then for all $t\geq 0$ the iterates produced by \gls{Byz-VR-MARINA-PP} without full-batch gradient computations (Algorithm~\ref{alg:byz_vr_marina_no_full}) satisfy

    \begin{align*}
    \mathbb{E}\left[\left\|g^{t+1}-\nabla f\left(x^{t+1}\right)\right\|^2\right] &\leq  \left(1-\frac{p}{4}\right) \mathbb{E}\left[\left\|g^{t}-\nabla f\left(x^{t}\right)\right\|^2\right]\\
    &+ \left(\frac{3\cP_{\cG_{\widehat{C}}^t}G}{(1-\delta)^2\widehat{C}^2} + 12c\delta\right)\frac{\sigma^2}{b'}\\
    & +  \widehat{B}\mathbb{E}\left[\left\|\nabla f\left(x^t\right)\right\|^2\right]  + \widehat{D}\zeta^2+\frac{pA}{4}\|x^{t+1} - x^t\|^2,
\end{align*}
where $A, \widehat{B}, \widehat{D}, p_G, \mathcal{P}_{\mathcal{G}^t_C}$ are defined in Lemma~\ref{lemma:final_lemma}.
\end{lemma}
\begin{proof}
    Up to the replacement of the bound from Lemma~\ref{lemma:full_aggr} with the bound from Lemma~\ref{lemma:full_aggr_no_full_grad}, the proof of the result is identical to the proof of Lemma~\ref{lemma:final_lemma}.
\end{proof}

\begin{theorem}
\label{them:1_no_full_grad}
 Let Assumptions \ref{assm:bounded-aggr}, \ref{assm:L-smoothness}, \ref{assm:global}, \ref{assm:local}, \ref{assm:het}, \ref{assm:bounded_variance} hold. Set clipping parameter $\lambda_{t+1} = 2\max_{i\in \mathcal{G}} L_m \left\|x^{t+1} - x^t\right\|$. Assume that
$$
0<\gamma \leq \frac{1}{L+\sqrt{A}}, \quad 4\widehat{B}<p,
$$
where $A$ and $\widehat{B}$ are defined in Theorem~\ref{them:1}. Then for all $T \geq 0$ the iterates produced by \gls{Byz-VR-MARINA} without full-batch gradient computations (Algorithm \ref{alg:byz_vr_marina_no_full}) satisfy
\begin{align*}
\mathbb{E}\left[\left\|\nabla f\left(\widehat{x}^T\right)\right\|^2\right] &\leq \frac{2 \Phi^{(0)}}{\gamma\left(1-\frac{4 \widehat{B}}{p}\right)(T+1)}+\frac{4 \widehat{D}  \zeta^2}{p-4 \widehat{B} }\\
&+ \left(\frac{12\cP_{\cG_{\widehat{C}}^t}G}{(1-\delta)^2\widehat{C}^2} + 48c\delta\right)\frac{\sigma^2}{b'(p - 4\widehat{B})},
\end{align*}
where $\widehat{x}^T$ is chosen uniformly at random from $x^0, x^1, \ldots, x^T$, and $\Phi^{(0)}=$ $f\left(x^0\right)-f^{\star}+\frac{2\gamma}{p}\left\|g^0-\nabla f\left(x^0\right)\right\|^2$ . 
\end{theorem}
\begin{proof}
    The proof is identical to the proof of Theorem~\ref{them:1} up to the replacement of Lemma~\ref{lemma:final_lemma} with Lemma~\ref{lemma:final_lemma_no_full_grad}.
\end{proof}

\begin{theorem}
\label{them:2_no_full_grad}
Let Assumptions  \ref{assm:bounded-aggr}, \ref{assm:L-smoothness}, \ref{assm:global}, \ref{assm:local}, \ref{assm:het}, \ref{assm:PL} hold. Set clipping parameter $\lambda_{t+1} = \max_{i\in \mathcal{G}} L_m \left\|x^{t+1} - x^t\right\|$. Assume that
$$
0<\gamma \leq \min \left\{\frac{1}{L+\sqrt{2 A}} \right\}, \quad 8\widehat{B}<p
$$
where $A$ and $\widehat{B}$ are defined in Theorem~\ref{them:2}. Then for all $T \geq 0$ the iterates produced by \gls{Byz-VR-MARINA} without full-batch gradient computations (Algorithm \ref{alg:byz_vr_marina_no_full}) satisfy
$$
\mathbb{E}\left[f\left(x^T\right)-f\left(x^{\star}\right)\right] \leq\left(1-\rho\right)^T \Phi^{(0)} +\frac{4 \widehat{D} \gamma \zeta^2}{p\rho} + \left(\frac{12\cP_{\cG_{\widehat{C}}^t}G}{(1-\delta)^2\widehat{C}^2} + 48c\delta\right)\frac{\gamma\sigma^2}{b'p\rho},
$$
where $\rho = \min\left[\gamma\mu\left(1-\frac{8\widehat{B}}{p}\right), \frac{p}{8}\right]$ and $\Phi^{(0)}=$ $f\left(x^0\right)-f^{\star}+\frac{4\gamma}{p}\left\|g^0-\nabla f\left(x^0\right)\right\|^2$.
\end{theorem}
\begin{proof}
    The proof is identical to the proof of Theorem~\ref{them:2} up to the replacement of Lemma~\ref{lemma:final_lemma} with Lemma~\ref{lemma:final_lemma_no_full_grad}.
\end{proof}

In contrast to their counterparts for \gls{Byz-VR-MARINA-PP} with (periodical) full-batch gradient computations (Theorems~\ref{them:1} and \ref{them:2}), the above results have additional terms proportional to $\frac{\sigma^2}{b'}$ in the upper bounds. These terms cannot be reduced with the decrease of the stepsize but can be made smaller via the increase of $b'$. A similar phenomenon appears in the analysis of the methods with recursive variance reduction even in the Byzantine-free case \citep{fang2018spider, li2021page, gorbunov2021marina}, and to address it, $b'$ is typically chosen to be large.

\subsubsection{Results for bounded compressors}

Similarly to the previous section, we start with an adaptation of Lemma~\ref{lemma:final_lemma_Q} to the case without full-batch gradient computations.
\begin{lemma}
\label{lemma:final_lemma_Q_no_full_grad}
    Let Assumptions ~\ref{assm:bounded-aggr}, \ref{assm:L-smoothness}, \ref{assm:global}, \ref{assm:local}, \ref{assm:local_smooth_all}, \ref{assm:het}, \ref{assm:bounded_variance}, \ref{assm:bounded-compressor} hold and the compression operator satisfy Definition~\ref{def:Q}. We set $\lambda_{t+1} = D_Q \max_{m,i} L_{m,i}\|x^{t+1} - x^t\|$. Also, let us introduce the notation
\begin{align*}\texttt{ARAgg}_Q^{t+1} = \texttt{ARAgg}\left(\clip_{\lambda_{t+1}}\left(\cQ\left(\widehat{\Delta}_1(x^{t+1}, x^t)\right)\right),\right.\\
\left.\ldots, \clip_{\lambda_{t+1}}\left(\cQ\left(\widehat{\Delta}_C(x^{t+1}, x^t)\right)\right)\right).\end{align*}
Then for all $t\geq 0$ the iterates produced by \gls{Byz-VR-MARINA-PP} without full-batch gradient computations (Algorithm~\ref{alg:byz_vr_marina_no_full}) satisfy

    \begin{align*}
    \mathbb{E}\left[\left\|g^{t+1}-\nabla f\left(x^{t+1}\right)\right\|^2\right] &\leq  \left(1-\frac{p}{2}\right) \mathbb{E}\left[\left\|g^{t}-\nabla f\left(x^{t}\right)\right\|^2\right]\\
    &+ \left(\frac{3\cP_{\cG_{\widehat{C}}^t}G}{(1-\delta)^2\widehat{C}^2} + 12c\delta\right)\frac{\sigma^2}{b'}\\
    &+\widehat{B}\mathbb{E}\left[\left\|\nabla f\left(x^t\right)\right\|^2\right] + \widehat{D} \zeta^2+\frac{pA}{4}\|x^{t+1} - x^t\|^2,
\end{align*}
where $A, \widehat{B}, \widehat{D}, p_G, \mathcal{P}_{\mathcal{G}^t_C}$ are defined in Lemma~\ref{lemma:final_lemma}.
\end{lemma}
\begin{proof}
    Up to the replacement of the bound from Lemma~\ref{lemma:full_aggr} with the bound from Lemma~\ref{lemma:full_aggr_no_full_grad}, the proof of the result is identical to the proof of Lemma~\ref{lemma:final_lemma_Q}.
\end{proof}

\begin{theorem}\label{them:1_Q_no_full_grad}
 Let Assumptions \ref{assm:bounded-aggr}, \ref{assm:L-smoothness}, \ref{assm:global}, \ref{assm:local}, \ref{assm:local_smooth_all}, \ref{assm:het}, \ref{assm:bounded_variance}, \ref{assm:bounded-compressor} hold. Setting $\lambda_{t+1} = \max_{m,i} L_{m,i} \left\|x^{t+1} - x^t\right\|$. Assume that
$$
0<\gamma \leq \frac{1}{L+\sqrt{A}}, \quad 4\widehat{B}<p,
$$ 
where $A$ and $\widehat{B}$ are defined in Theorem~\ref{them:1_Q}. Then for all $T \geq 0$ the iterates produced by \gls{Byz-VR-MARINA} without full-batch computations (Algorithm \ref{alg:byz_vr_marina_no_full}) satisfy
\begin{align*}
\mathbb{E}\left[\left\|\nabla f\left(\widehat{x}^T\right)\right\|^2\right] &\leq \frac{2 \Phi^{(0)}}{\gamma\left(1-\frac{4\widehat{B}}{p}\right)(T+1)}+\frac{2 \widehat{D} \zeta^2}{p-4 \widehat{B}}\\
&+ \left(\frac{6\cP_{\cG_{\widehat{C}}^t}G}{(1-\delta)^2\widehat{C}^2} + 24c\delta\right)\frac{\sigma^2}{b'(p - 4\widehat{B})},
\end{align*}
where $\widehat{x}^T$ is chosen uniformly at random from $x^0, x^1, \ldots, x^T$, and $\Phi^{(0)}=$ $f\left(x^0\right)-f^{\star}+\frac{\gamma}{p}\left\|g^0-\nabla f\left(x^0\right)\right\|^2$ . 
\end{theorem}
\begin{proof}
    The proof is identical to the proof of Theorem~\ref{them:1_Q} up to the replacement of Lemma~\ref{lemma:final_lemma_Q} with Lemma~\ref{lemma:final_lemma_Q_no_full_grad}.
\end{proof}

\begin{theorem}\label{them:1_Q_no_full_grad_}
 Let Assumptions \ref{assm:bounded-aggr}, \ref{assm:L-smoothness}, \ref{assm:global}, \ref{assm:local}, \ref{assm:local_smooth_all}, \ref{assm:het}, \ref{assm:bounded_variance}, \ref{assm:bounded-compressor}, \ref{assm:PL} hold. Setting $\lambda_{t+1} = \max_{m,i} L_{m,i} \left\|x^{t+1} - x^t\right\|$. Assume that
$$
0<\gamma \leq \frac{1}{L+\sqrt{2A}}, \quad 8\widehat{B}<p,
$$ 
where $A$ and $\widehat{B}$ are defined in Theorem~\ref{them:2_Q}. Then for all $T \geq 0$ the iterates produced by \gls{Byz-VR-MARINA} without full-batch computations (Algorithm \ref{alg:byz_vr_marina_no_full}) satisfy
$$
\mathbb{E}\left[f\left(x^T\right)-f\left(x^{\star}\right)\right] \leq\left(1-\rho\right)^T \Phi^{(0)}+\frac{2\widehat{D}\zeta^2}{p\rho} + \left(\frac{6\cP_{\cG_{\widehat{C}}^t}G}{(1-\delta)^2\widehat{C}^2} + 24c\delta\right)\frac{\gamma\sigma^2}{b'p\rho},
$$
where $\rho = \min\left[\gamma\mu\left(1-\frac{8\widehat{B}}{p}\right), \frac{p}{4}\right]$ and $\Phi^{(0)}=$ $f\left(x^0\right)-f^{\star}+\frac{2\gamma}{p}\left\|g^0-\nabla f\left(x^0\right)\right\|^2$. 
\end{theorem}
\begin{proof}
    The proof is identical to the proof of Theorem~\ref{them:2_Q} up to the replacement of Lemma~\ref{lemma:final_lemma_Q} with Lemma~\ref{lemma:final_lemma_Q_no_full_grad}.
\end{proof}

\subsection{Main results for Byz-VR-MARINA+ without full batch gradient computations}

\subsubsection{Results for bounded compressors}

Similarly to the analysis of \gls{Byz-VR-MARINA} without full-batch gradient computations, we start with the adaptation of Lemma~\ref{lemma:final_lemma_simplified} to the no-full-batch gradient computations case.
\begin{lemma}
\label{lemma:final_lemma_simplified_no_full}
    Let Assumptions ~\ref{assm:L-smoothness}, \ref{assm:global}, \ref{assm:local}, \ref{assm:local_smooth_all}, \ref{assm:het}, \ref{assm:bounded_variance}, \ref{assm:bounded-compressor} hold and the compression operator satisfy Definition~\ref{def:Q}. Assume that $C \leq G$. We set $\lambda_{t+1} = D_Q \max_{m,i} L_{m,i}\|x^{t+1}-x^t\|$. Then for all $t\geq 0$ the iterates produced by \algname{Byz-VR-MARINA-PP+} without full-batch gradient computations (Algorithm~\ref{alg:byz_vr_marina+_no_full}) satisfy
    \begin{align}
   \notag \mathbb{E}\left[\left\|g^{t+1}-\nabla f\left(x^{t+1}\right)\right\|^2\right] &\leq  \left(1-\frac{p}{2}\right) \mathbb{E}\left[\left\|g^{t}-\nabla f\left(x^{t}\right)\right\|^2\right]\\
    &+ \left(\frac{\cP_{\cG_{\widehat{C}}^t}G}{(1-\delta)^2\widehat{C}^2} + 4c\delta\right)\frac{p\sigma^2}{b'}\notag\\
    &\notag+\widehat{B}\mathbb{E}\left[\left\|\nabla f\left(x^t\right)\right\|^2\right] + \widehat{D} \zeta^2\\
    &+\frac{pA}{4}\|x^{t+1} - x^t\|^2, \label{eq:simplified_main_lemma_}
\end{align}
where $A, \widehat{B}, \widehat{D}, p_G, \mathcal{P}_{\mathcal{G}^t_C}$ are defined in Lemma~\ref{lemma:final_lemma_simplified}.
\end{lemma}
\begin{proof}
    Up to the replacement of the bound from Lemma~\ref{lemma:full_aggr} with the bound from Lemma~\ref{lemma:full_aggr_no_full_grad}, the proof of the result is identical to the proof of Lemma~\ref{lemma:final_lemma_simplified}.
\end{proof}

\begin{theorem}\label{thm:non_convex+_no_full}
 Let Assumptions \ref{assm:L-smoothness}, \ref{assm:global}, \ref{assm:local}, \ref{assm:local_smooth_all}, \ref{assm:het}, \ref{assm:bounded_variance}, \ref{assm:bounded-compressor} hold. Set $\lambda_{t+1} = \max_{m,i} L_{m,i} \left\|x^{t+1} - x^t\right\|$. Assume that
$$
0<\gamma \leq \frac{1}{L+\sqrt{A}}, \quad 4\widehat{B}<p,
$$ 
where $A$ and $\widehat{B}$ are defined in Theorem~\ref{thm:non_convex+}. Then for all $T \geq 0$ the iterates produced by \algname{Byz-VR-MARINA+} without full-batch gradient computations (Algorithm \ref{alg:byz_vr_marina+_no_full}) satisfy
\begin{align*}
\mathbb{E}\left[\left\|\nabla f\left(\widehat{x}^T\right)\right\|^2\right] &\leq \frac{2 \Phi^{(0)}}{\gamma\left(1-\frac{4\widehat{B}}{p}\right)(T+1)}+\frac{2 \widehat{D} \zeta^2}{p-4 \widehat{B}}\\
&+ \left(\frac{2\cP_{\cG_{\widehat{C}}^t}G}{(1-\delta)^2\widehat{C}^2} + 8c\delta\right)\frac{p\sigma^2}{b'(p - 4\widehat{B})},
\end{align*}
where $\widehat{x}^T$ is chosen uniformly at random from $x^0, x^1, \ldots, x^T$, and $\Phi^{(0)}=$ $f\left(x^0\right)-f^{\star}+\frac{\gamma}{p}\left\|g^0-\nabla f\left(x^0\right)\right\|^2$ . 
\end{theorem}
\begin{proof}
    The proof is analogous to the proof of Theorem~\ref{them:1}.
\end{proof}

\begin{theorem}\label{thm:PL+_no_full}
Let Assumptions  \ref{assm:bounded-compressor}, \ref{assm:L-smoothness}, \ref{assm:global}, \ref{assm:local}, \ref{assm:local_smooth_all}, \ref{assm:het}, \ref{assm:bounded_variance} \ref{assm:PL} hold. Set $\lambda_{t+1} = \max_{m,i} L_{m,i} \left\|x^{t+1} - x^t\right\|$. Assume that
$$
0<\gamma \leq \min \left\{\frac{1}{L+\sqrt{2 A}} \right\}, \quad 8\widehat{B}<p,
$$
where $A$ and $\widehat{B}$ are defined in Theorem~\ref{thm:PL+}. Then for all $T \geq 0$ the iterates produced by \algname{Byz-VR-MARINA+} without full-batch gradient computations (Algorithm \ref{alg:byz_vr_marina+_no_full}) satisfy
$$
\mathbb{E}\left[f\left(x^T\right)-f\left(x^{\star}\right)\right] \leq\left(1-\rho\right)^T \Phi^{(0)}+\frac{2\widehat{D}\zeta^2}{p\rho} + \left(\frac{2\cP_{\cG_{\widehat{C}}^t}G}{(1-\delta)^2\widehat{C}^2} + 8c\delta\right)\frac{\gamma \sigma^2}{b'\rho},
$$
where $\rho = \min\left[\gamma\mu\left(1-\frac{8\widehat{B}}{p}\right), \frac{p}{4}\right]$ and $\Phi^{(0)}=$ $f\left(x^0\right)-f^{\star}+\frac{2\gamma}{p}\left\|g^0-\nabla f\left(x^0\right)\right\|^2$.
\end{theorem}
\begin{proof}
    The proof is analogous to the proof of Theorem~\ref{them:2}.
\end{proof}
As in the case of \gls{Byz-VR-MARINA}, the above upper bounds for \algname{Byz-VR-MARINA+} without full-batch gradient computations have additional terms proportional to $\frac{\sigma^2}{b'}$. In contrast to the results for \algname{Byz-VR-MARINA+} without full-batch gradient computations, these terms for \algname{Byz-VR-MARINA+} are $\nicefrac{1}{p}$ times smaller.

\clearpage

\section{Experimental Details and Extra Experiments}
\label{app:experiments}

\subsection{Experimental details}

For each experiment, we tune the step size using the following set of candidates $\{0.1, 0.01, 0.001\}$. The step size is fixed. We do not use learning rate warmup or decay. We use batches of size $32$ for all methods. For partial participation, in each round, we sample $20 \%$ of clients uniformly at random.  For $\lambda_t = \lambda \|x^t - x^{t-1}\|$ used for clipping, we select $\lambda$ from $\{0.1, 1., 10.\}$. Each experiment is run with three varying random seeds, and we report the mean optimality gap with one standard error. The optimal value is obtained by running gradient descent (\algname{\gls{GD}}) on the complete dataset for 1000 epochs. Our implementation of attacks and robust aggregation schemes is based on the public implementation from \citep{gorbunov2023variance}. %
\vspace{-1.em}
\subsection{Extra experiments}
\vspace{-.5em}
Below we provide the missing neural network experiments from the main paper. We consider the MNIST dataset \citep{mnist} and CIFAR10 \citep{cifar} (as in \citep{karimireddy2021learning}) with 20 clients, 5 of which are malicious, and 4 clients are sampled in each step. For the attacks, we consider\gls{ALIE}~\citep{baruch2019little} and the aforementioned Shift-Back (\gls{SHB}). For the aggregations, we consider \gls{CM} \citep{chen2017distributed} and \gls{RFA} \citep{pillutla2022robust} with bucketing. For the MNIST dataset, we use a simple neural network with two convolution layers followed by two fully connected. For CIFAR 10, we use ResNet18~\citep{resnet18} architecture with layer norm. One can note that the results are consistent with the ones provided in the main paper, i.e., clipping performs on par or better than its variant without clipping, and no robust aggregator is able to withstand the shift-back attack without clipping.

\begin{figure}[H]
\centering
\includegraphics[width=0.24\textwidth]{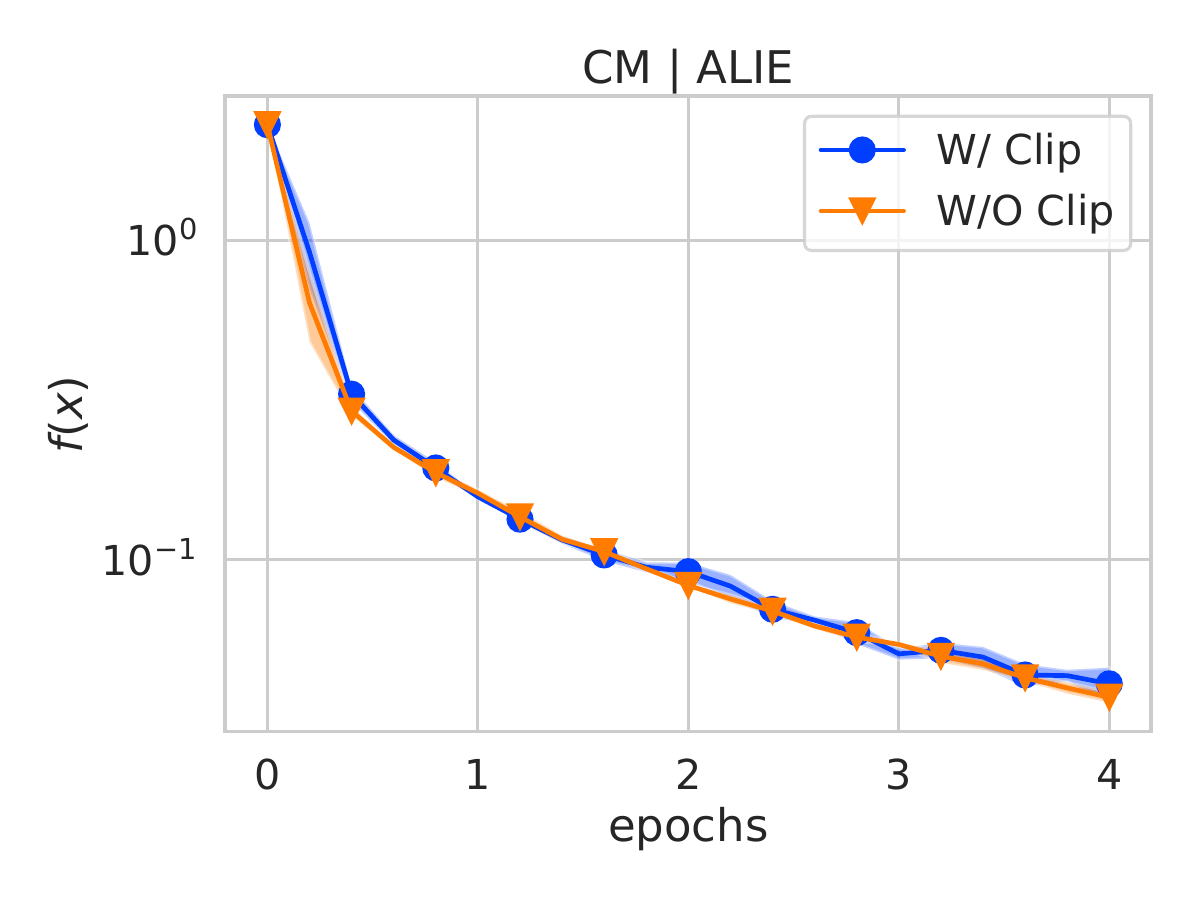}
\includegraphics[width=0.24\textwidth]{Byz-Marina-PP/figures/MNIST_non_iid_comp=none_agg=cm_attack_SHB_clip_sensitivity.pdf}
\includegraphics[width=0.24\textwidth]{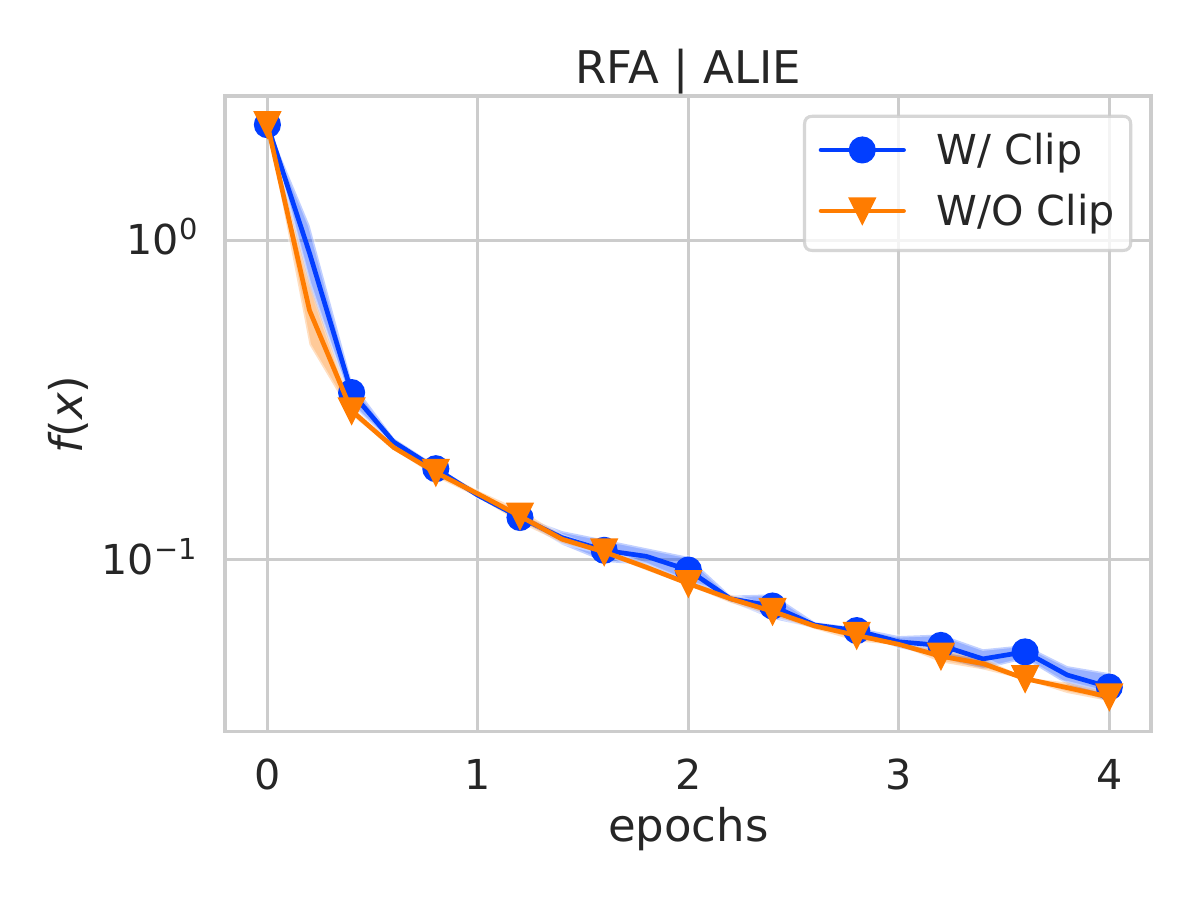}
\includegraphics[width=0.24\textwidth]{Byz-Marina-PP/figures/MNIST_non_iid_comp=none_agg=rfa_attack_SHB_clip_sensitivity.pdf}
\hfill
\includegraphics[width=0.24\textwidth]{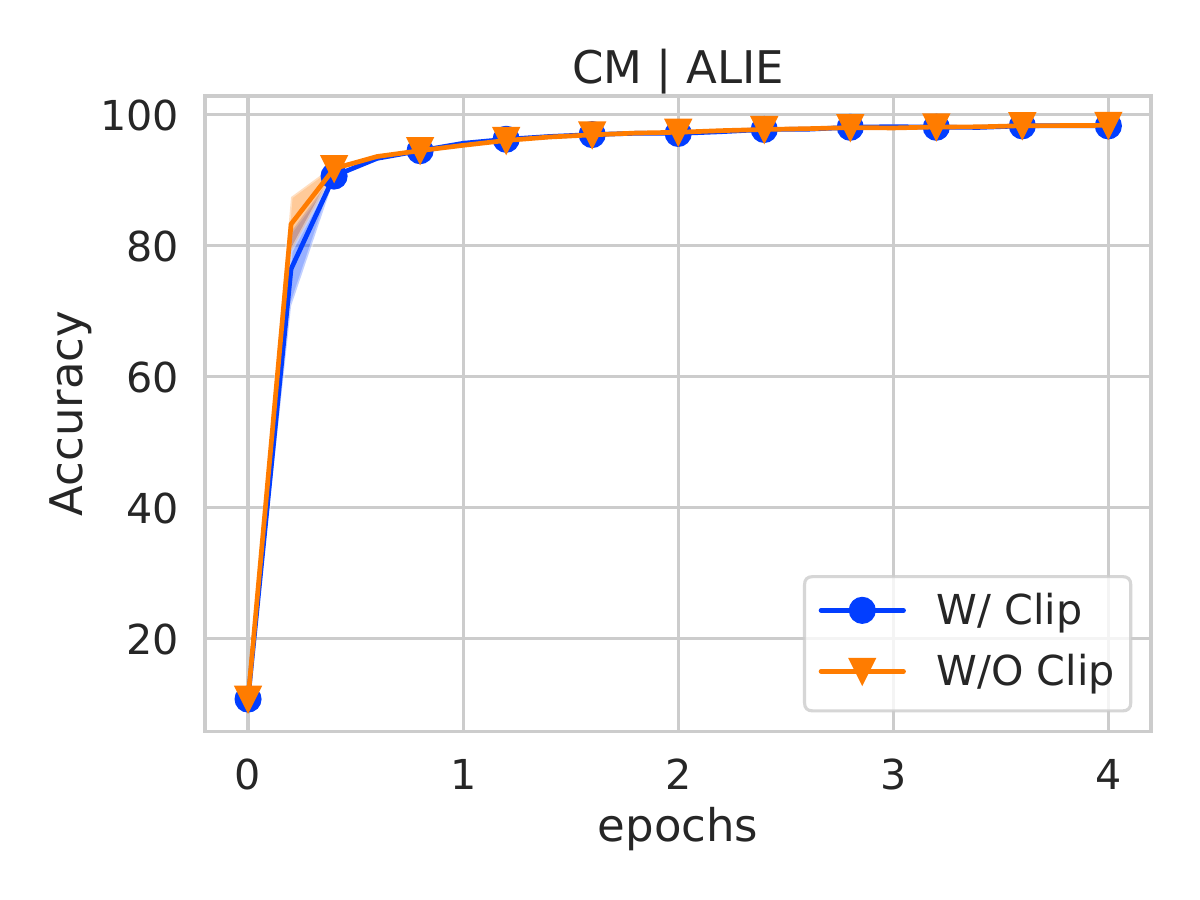}
\includegraphics[width=0.24\textwidth]{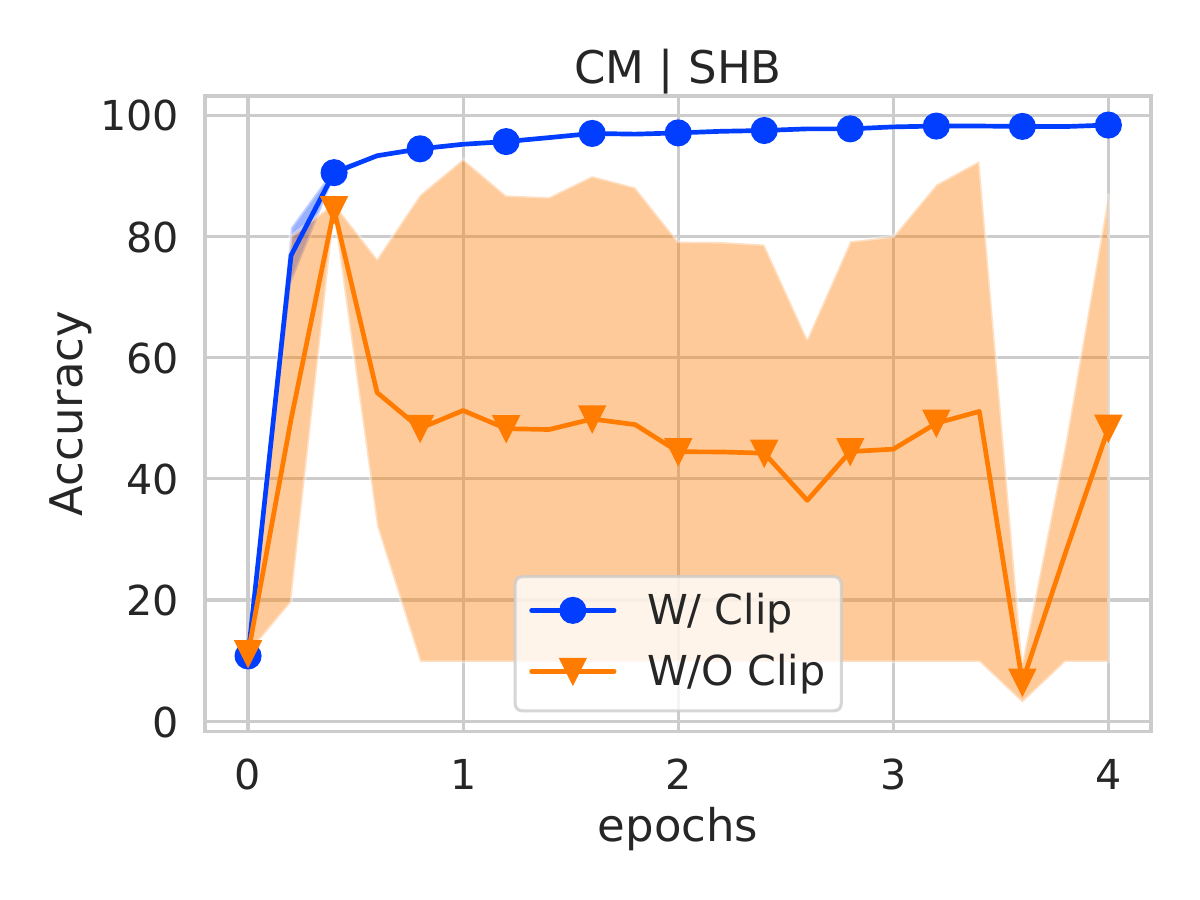}
\includegraphics[width=0.24\textwidth]{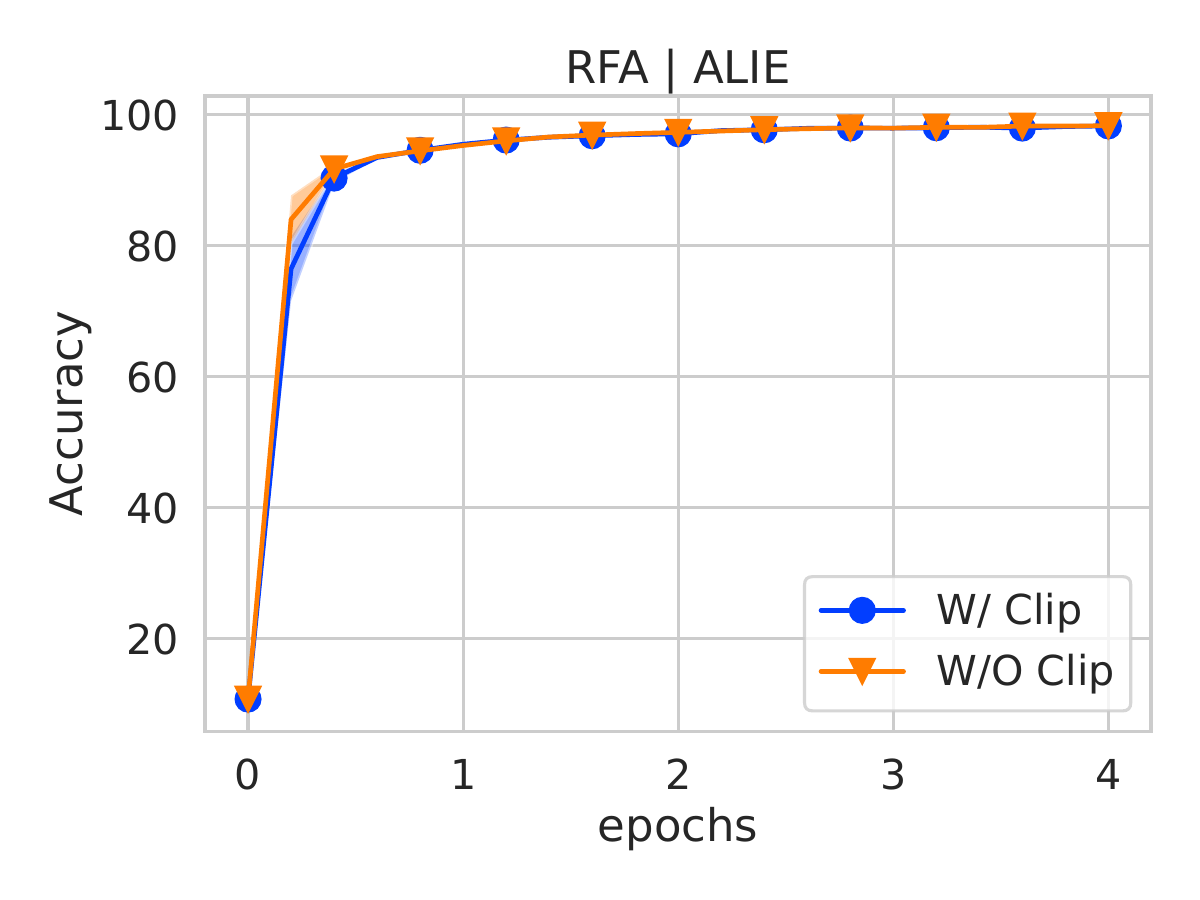}
\includegraphics[width=0.24\textwidth]{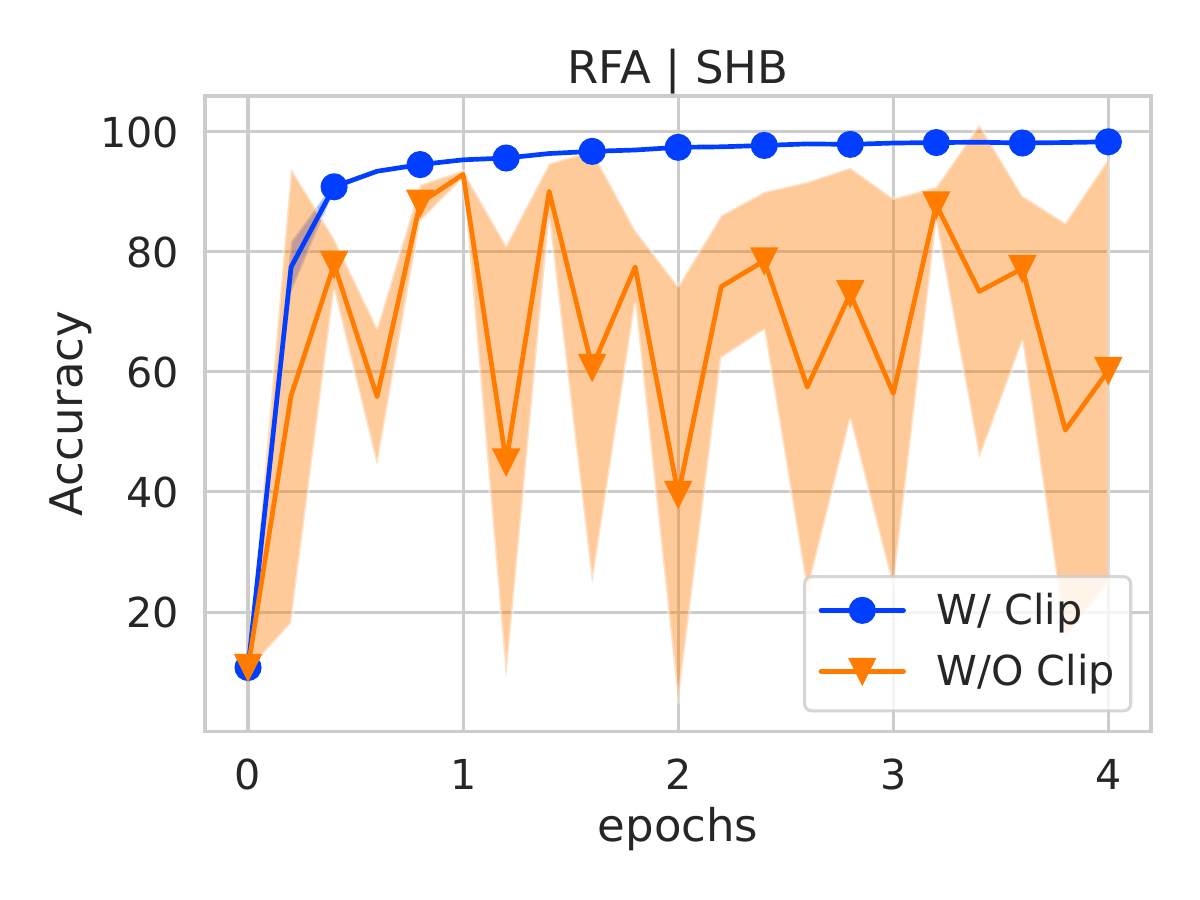}
\caption{
Training loss (top) and test accuracy (bottom) of 2 aggregation rules (\gls{CM}, \gls{RFA}) under 4 attacks (\gls{BF}, LF, \gls{ALIE}, \gls{SHB}) on the MNIST dataset under heterogeneous data split with 20 clients,  5 of which are malicious, 4 clients sampled per round. } 
\label{fig:nn_app}
\vspace{-1.5em}
\end{figure}

\begin{figure}[H]
\centering
\includegraphics[width=0.24\textwidth]{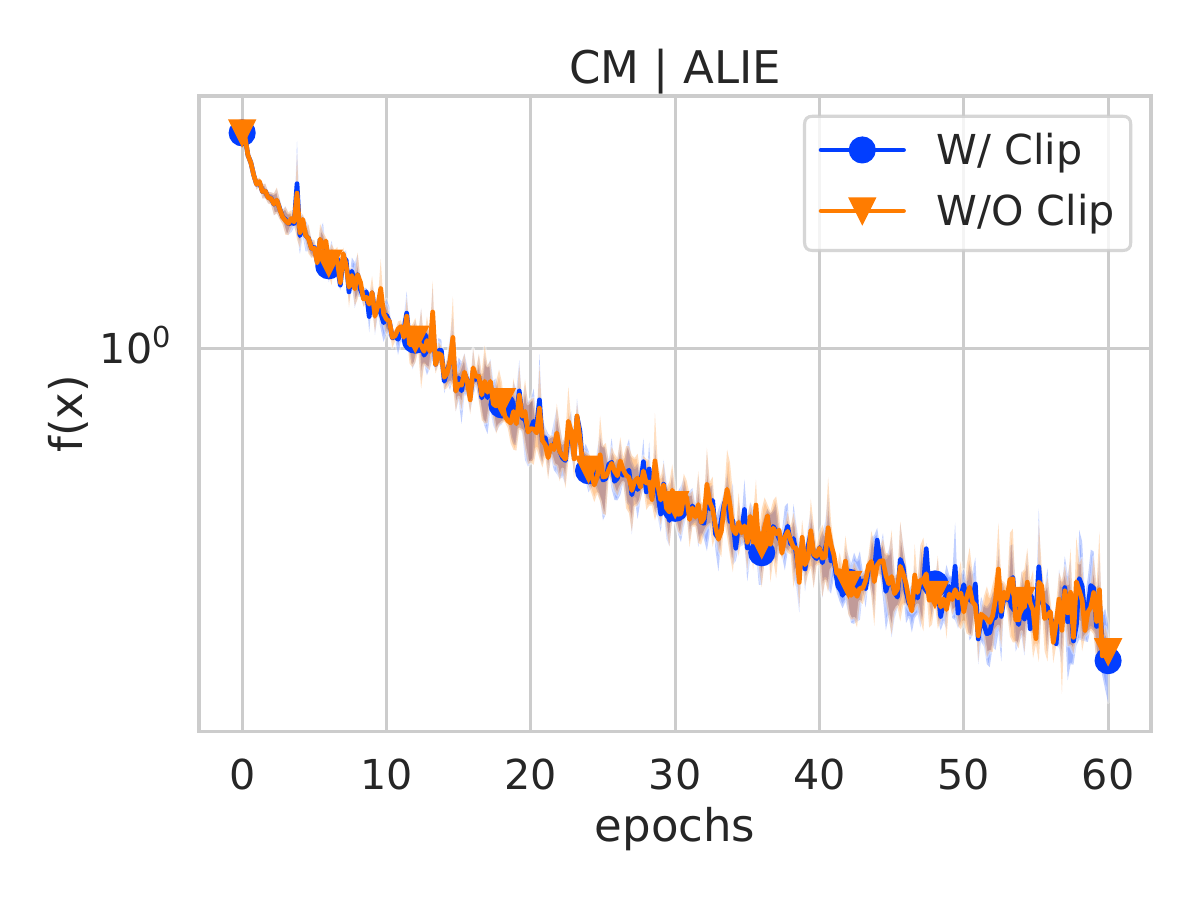}
\includegraphics[width=0.24\textwidth]{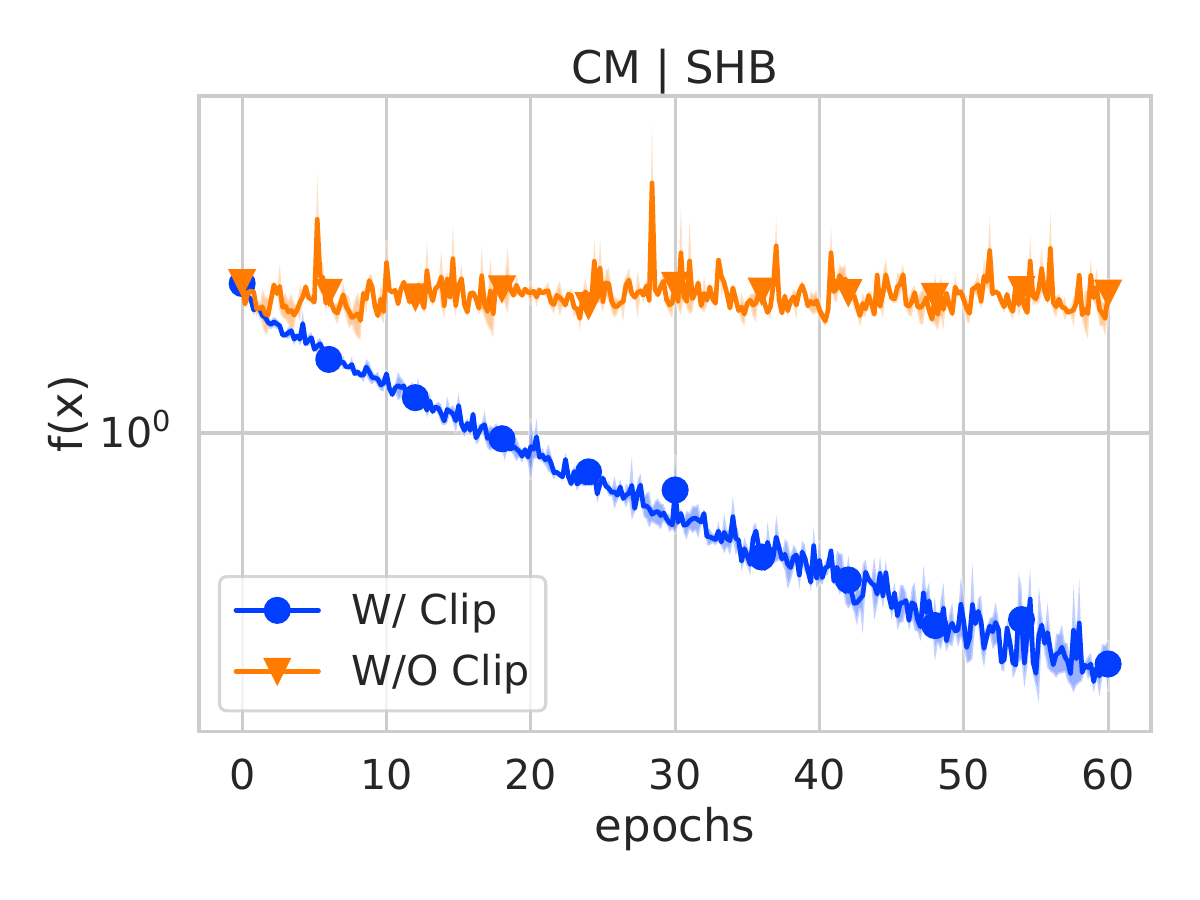}
\includegraphics[width=0.24\textwidth]{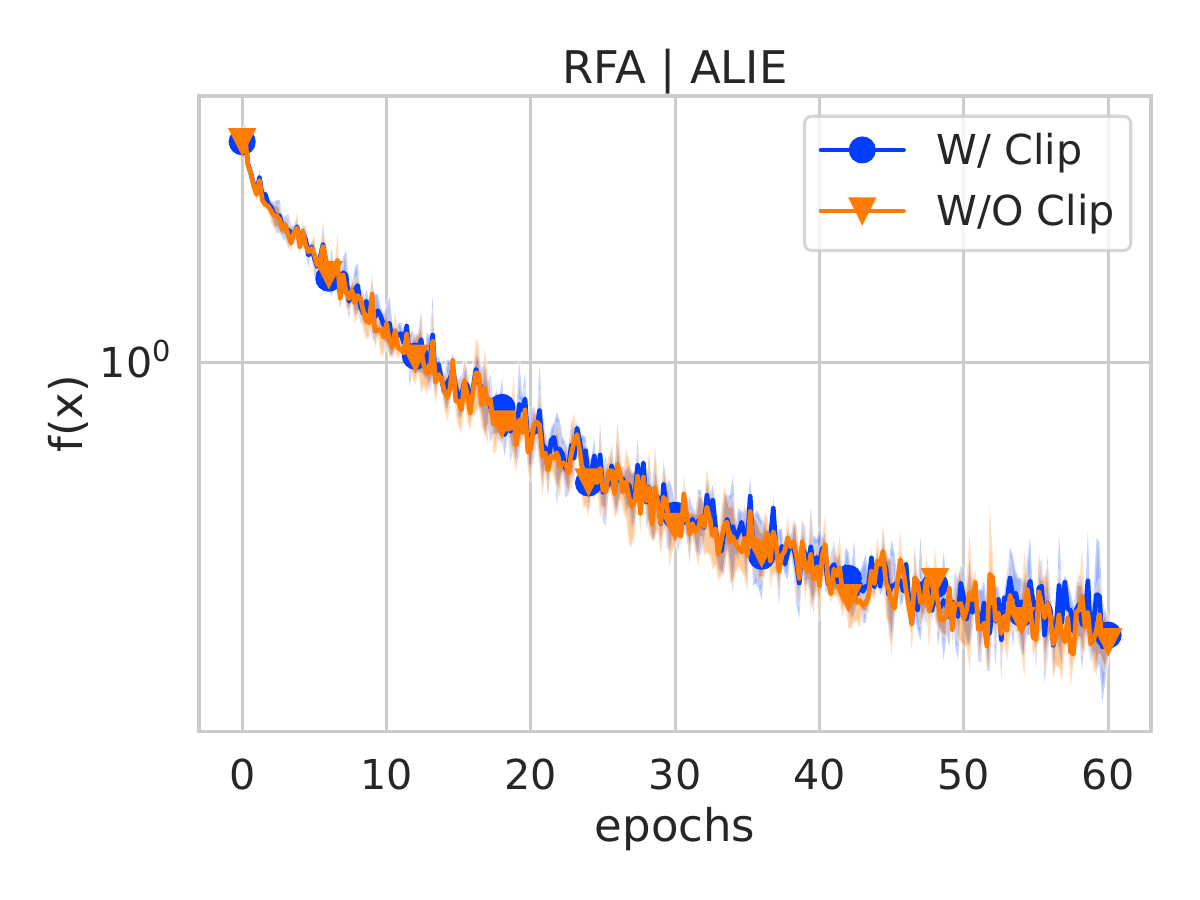}
\includegraphics[width=0.24\textwidth]{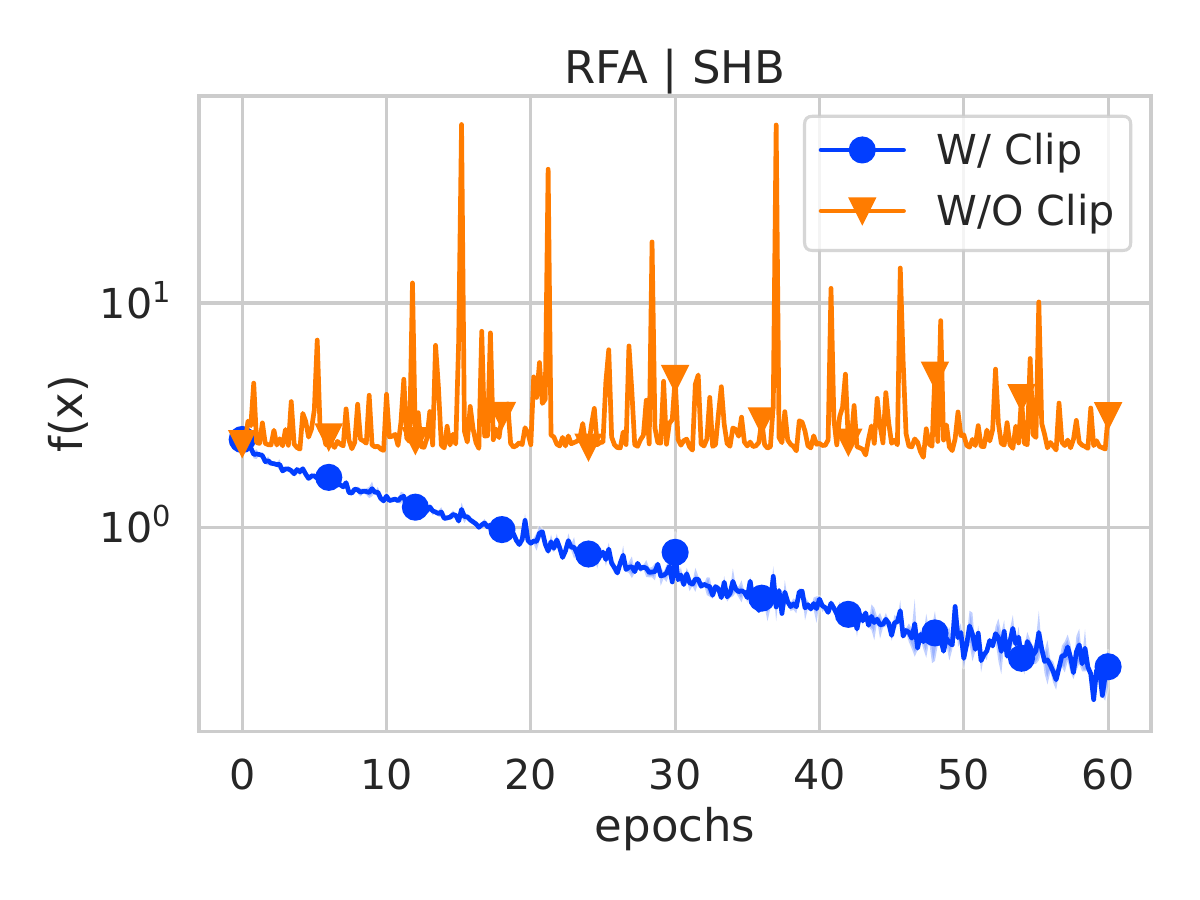}
\hfill
\includegraphics[width=0.24\textwidth]{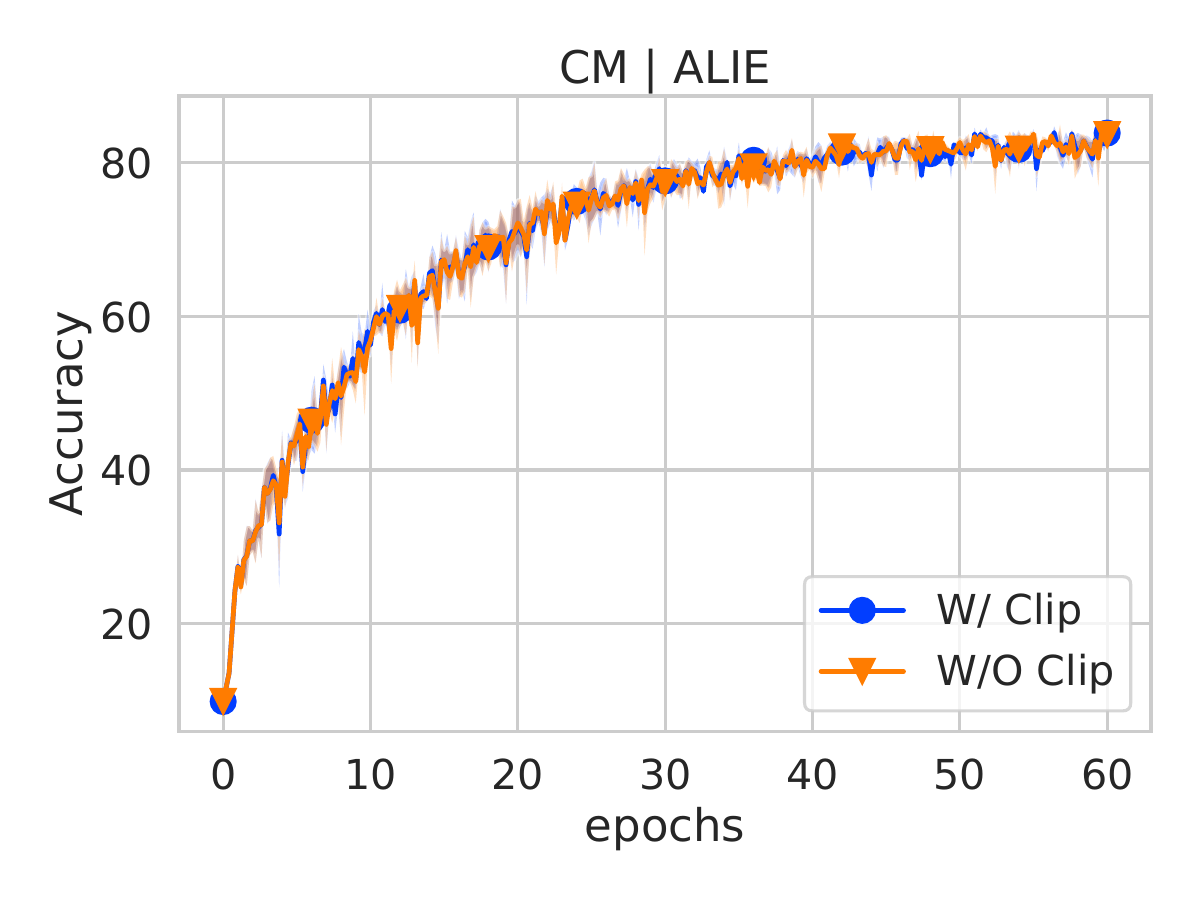}
\includegraphics[width=0.24\textwidth]{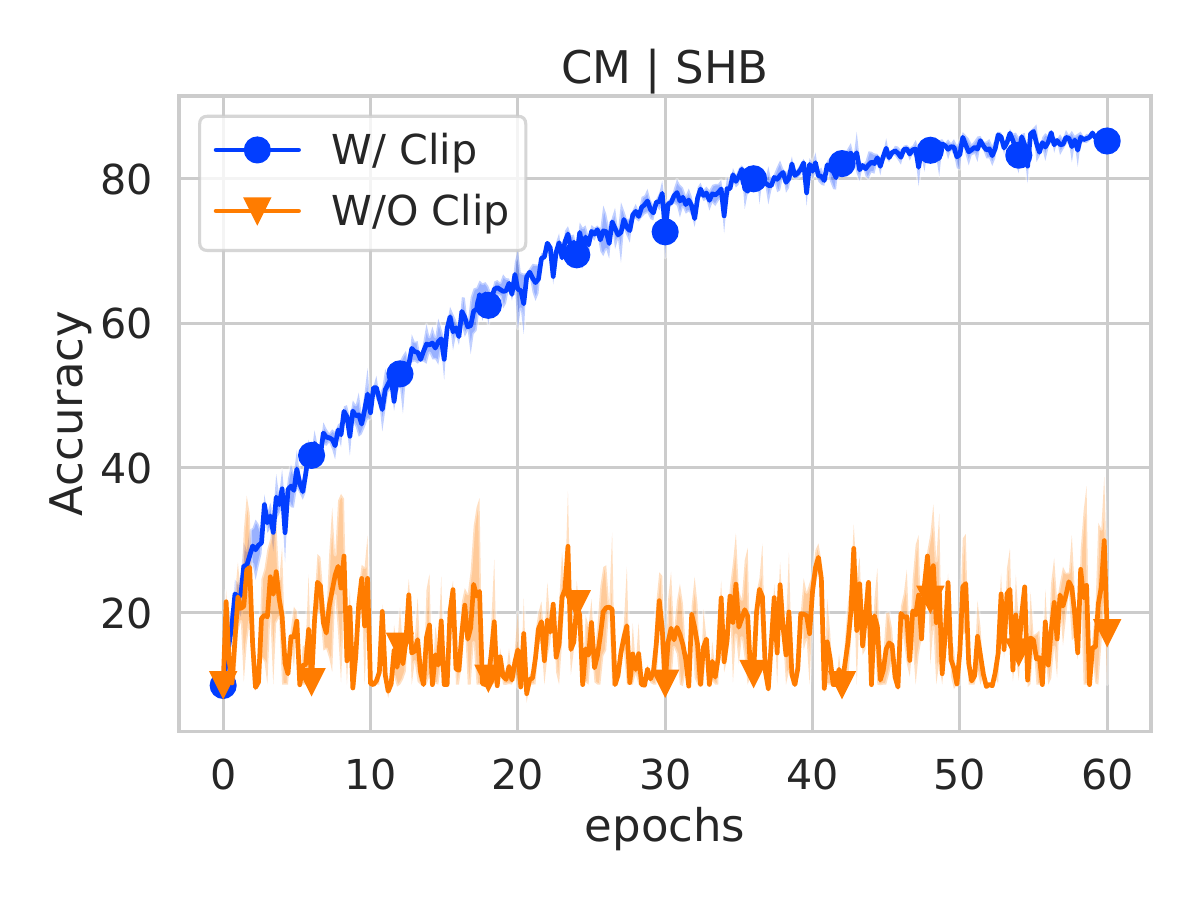}
\includegraphics[width=0.24\textwidth]{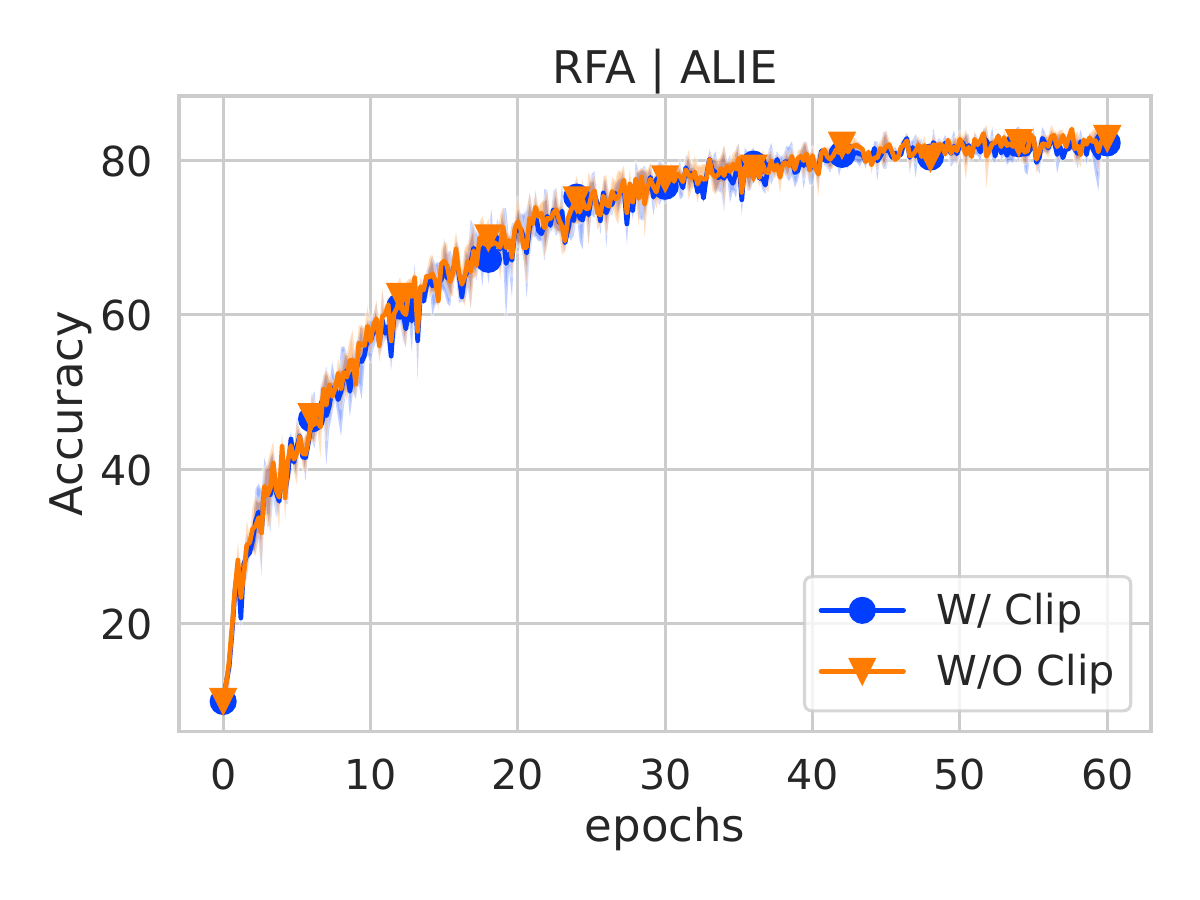}
\includegraphics[width=0.24\textwidth]{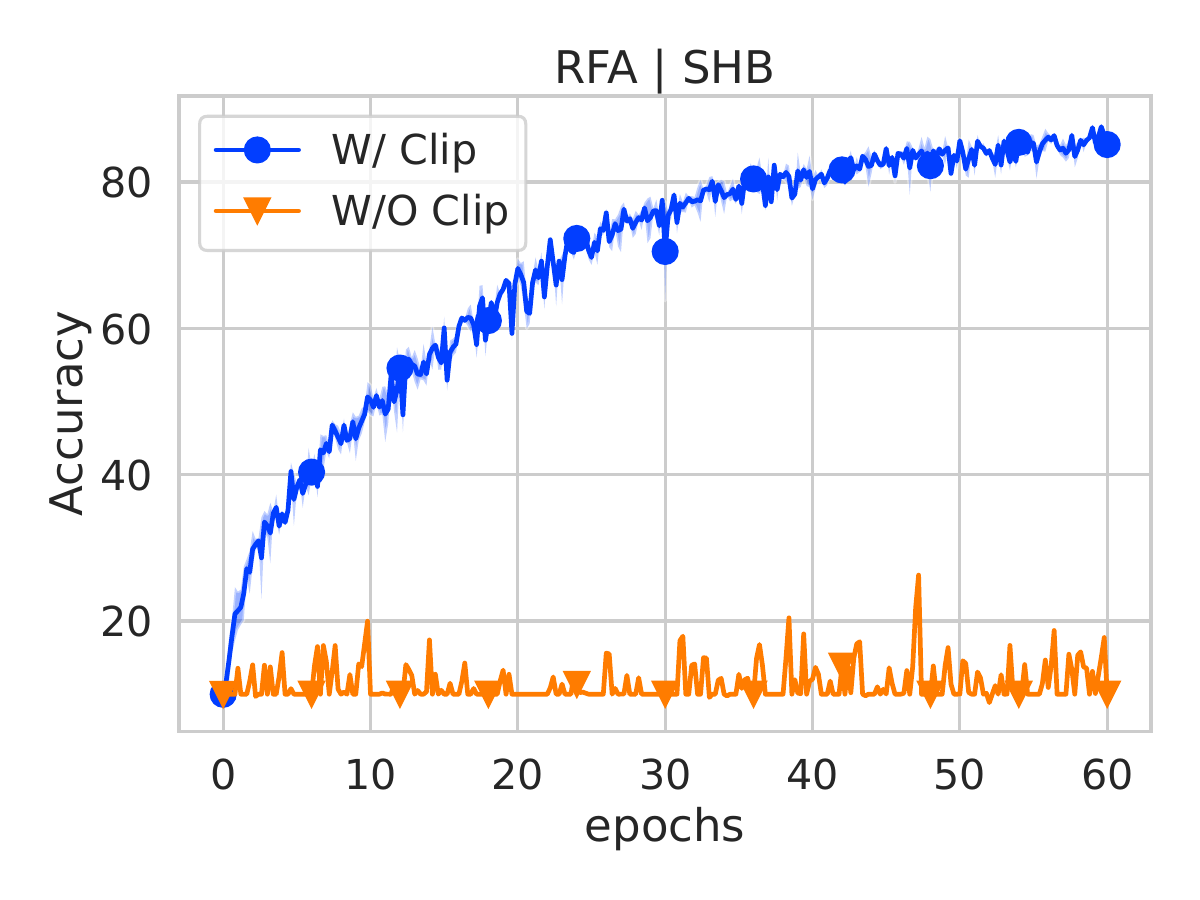}
\caption{
Training loss (top) and test accuracy (bottom) of 2 aggregation rules (\gls{CM}, \gls{RFA}) under 4 attacks (\gls{BF}, \gls{LF}, \gls{ALIE}, \gls{SHB}) on the CIFAR10 dataset under heterogeneous data split with 20 clients,  5 of which are malicious, 4 clients sampled per round. } 
\label{fig:nn_app_}
\vspace{-1.5em}
\end{figure}

\newpage
            \refstepcounter{chapter}%
\chapter*{\thechapter \quad Appendix H Title}
\label{appendixH}

\section{Results on Convex Optimization Problems}
\label{sec:app:exp-convex}
\subsection{Logistic regression}
We performed our analysis in a controlled environment using logistic regression with quadratic regularization on synthetic data. In this configuration, we set \(d = 100\), employed weight matrices of size \(10 \times 10\), and used 2,000 samples, with the regularization term fixed at \(0.1\). As shown in Figure~\ref{fig:logreg}, the method demonstrates convergence across different ranks, and when the rank is set to full rank, we observe convergence that mirrors that of \algname{\gls{FPFT}}.

\begin{figure}[h]
\centering
\includegraphics[width=0.7\textwidth]{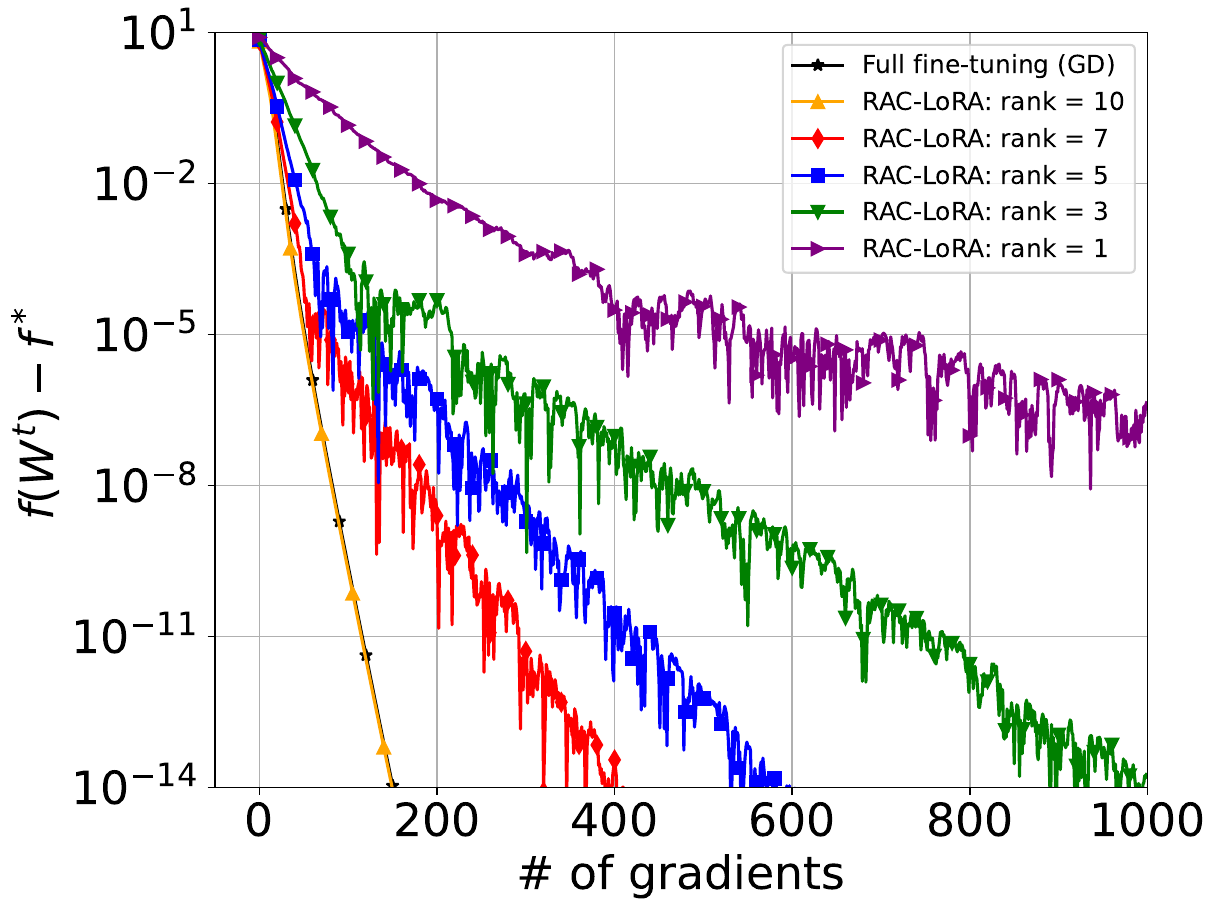}

\caption{\algname{RAC-LoRA} convergence with varying ranks and step sizes on a logistic regression problem.}\label{fig:logreg}
\end{figure}

\section{Results on Non-Convex Optimization Problems}
\label{sec:app:nonconvex}

\subsection{Additional results of RoBERTa on NLP tasks}

Table~\ref{results_roberta_base_rank=2} reports additional configurations of the number of epochs per chain and the number of chains on the GLUE benchmark. These results further corroborate the discussion in Section~\ref{sec:exp-nonconvex}.

\begin{table*}[htb!]
\resizebox{\textwidth}{!}{%
\centering
{
\begin{tabular}{lccccccc}
 \toprule
{{Method}} & {{\# Chains}} & {{\# Epochs}} & \textcolor{black}{{MRPC}} & \textcolor{black}{{CoLA}} & \textcolor{black}{{RTE}} & \textcolor{black}{{STS-B}} &  {Avg}\\
 \midrule
\algname{FPFT}* &  \multirow{2}{*}{1} & \multirow{2}{*}{30, 80, 80, 40} & 
90.2\textsubscript{$\pm$0.0} &
63.6\textsubscript{$\pm$0.0} &
78.7\textsubscript{$\pm$0.0} & 
91.2\textsubscript{$\pm$0.0} & 
80.9\\
\algname{LoRA}* & & & 
89.7\textsubscript{$\pm$0.7} &
63.4\textsubscript{$\pm$1.2} &
86.6\textsubscript{$\pm$0.7} & 
91.5\textsubscript{$\pm$0.2} & 
82.8\\
\midrule
\algname{LoRA} & \multirow{2}{*}{1} & \multirow{2}{*}{20} & 
86.8\textsubscript{$\pm$0.8} &
58.0\textsubscript{$\pm$0.4} &
{71.4\textsubscript{$\pm$0.7}} & 
{90.3\textsubscript{$\pm$0.1}} & 
76.6\\
\algname{AsymmLoRA} & & & 
85.5\textsubscript{$\pm$0.5} &
56.5\textsubscript{$\pm$1.5} &
69.2\textsubscript{$\pm$0.2} & 
89.6\textsubscript{$\pm$0.1} & 
75.2\\
\cdashline{1-8}
\multirow{2}{*}{\algname{COLA}} & 2 & 10 & 
{87.1\textsubscript{$\pm$0.2}} &
{58.4\textsubscript{$\pm$1.5}} &
69.9\textsubscript{$\pm$0.9} &
90.3\textsubscript{$\pm$0.2} & 
76.4
\\
& 10 & 2 & 
84.2\textsubscript{$\pm$1.1} &
54.2\textsubscript{$\pm$0.4}&
64.6\textsubscript{$\pm$1.3}& 
89.1\textsubscript{$\pm$0.1} & 
73.0\\
\cdashline{1-8}
\multirow{2}{*}{\algname{RAC-LoRA}} & 2 & 10
&
85.6\textsubscript{$\pm$1.7} &
55.3\textsubscript{$\pm$1.2} &
68.6\textsubscript{$\pm$1.0} & 
89.4\textsubscript{$\pm$0.2} & 
74.7\\
& 10 & 2
&
85.4\textsubscript{$\pm$0.4} &
55.1\textsubscript{$\pm$1.2} &
65.5\textsubscript{$\pm$0.9} & 
89.3\textsubscript{$\pm$0.1} & 
73.8\\
\\[-2.5ex]
\midrule
\algname{LoRA} & \multirow{2}{*}{1} & \multirow{2}{*}{50}
&
{88.2\textsubscript{$\pm$0.3}} &
{60.1\textsubscript{$\pm$0.4}} &
{74.4\textsubscript{$\pm$0.9}} & 
{90.6\textsubscript{$\pm$0.1}} & 
78.3\\
\algname{AsymmLoRA} &  & 
&
86.4\textsubscript{$\pm$1.0} &
57.4\textsubscript{$\pm$0.3} &
69.9\textsubscript{$\pm$1.8} & 
90.3\textsubscript{$\pm$0.1} & 
76.0\\
\cdashline{1-8}
\multirow{2}{*}{\algname{COLA}} & 5 & 10
& 
87.8\textsubscript{$\pm$1.1} &
59.3\textsubscript{$\pm$2.1} &
71.2\textsubscript{$\pm$1.2} &
90.6\textsubscript{$\pm$0.2} & 
77.2\\
& 10 & 5 
&
87.7\textsubscript{$\pm$0.5} &
58.1\textsubscript{$\pm$1.2} &
70.9\textsubscript{$\pm$0.5} &
90.2\textsubscript{$\pm$0.2} & 
76.7\\
\cdashline{1-8}
\multirow{2}{*}{\algname{RAC-LoRA}} & 5 & 10
& 
87.2\textsubscript{$\pm$0.6} &
57.6\textsubscript{$\pm$0.5} &
70.6\textsubscript{$\pm$0.7} &
90.2\textsubscript{$\pm$0.1} & 
76.4\\
& 10 & 5
&
87.5\textsubscript{$\pm$0.4} &
57.8\textsubscript{$\pm$1.0} &
70.3\textsubscript{$\pm$1.2} &
90.2\textsubscript{$\pm$0.2} & 
76.5\\
\midrule
\algname{LoRA} & \multirow{2}{*}{1} & \multirow{2}{*}{100}  
&
87.7\textsubscript{$\pm$0.2} &
{60.8\textsubscript{$\pm$0.2}} &
{75.2\textsubscript{$\pm$1.5}} & 
90.2\textsubscript{$\pm$0.1} & 
78.5\\
\algname{AsymmLoRA} & & 
&
86.9\textsubscript{$\pm$0.3} &
58.7\textsubscript{$\pm$1.0} &
71.0\textsubscript{$\pm$3.3} & 
90.4\textsubscript{$\pm$0.0} & 
76.8\\
\algname{COLA} & 10 & 10
& 
{88.0\textsubscript{$\pm$0.8}} &
59.5\textsubscript{$\pm$1.0} &
72.1\textsubscript{$\pm$0.9}&
{90.7\textsubscript{$\pm$0.2}} & 
77.6\\
\algname{RAC-LoRA} & 10 & 10
& 
87.0\textsubscript{$\pm$0.7} &
58.5\textsubscript{$\pm$0.1} &
72.3\textsubscript{$\pm$1.5}&
90.3\textsubscript{$\pm$0.0} & 
77.0\\
\bottomrule
\end{tabular}
}
}
\caption{{Performance of the methods using RoBERTa-base for rank 2}. The experiments are based on 4 tasks from the GLUE benchmark. * denotes the results reported in \cite{hu2021lora}. We report Matthews correlation coefficient for the CoLA dataset, Pearson correlation coefficient for STS-B, and accuracy for the remaining tasks, with the standard deviations given in the subscript. The results are obtained using 3 random seeds.}
\label{results_roberta_base_rank=2}
\vspace{-5.0pt}
\end{table*}

\subsection{Ablation on number of epochs per block in the chains}

The convergence proof for \algname{\gls{RAC-LoRA}} (Corollary \ref{cor-RR-gen} and Corollary \ref{cor-RR-PL}) states that each \algname{\gls{LoRA}} module shall be optimized for one epoch only. However, good approximations can also be obtained using more epochs per block and hence fewer blocks (i.e., fewer parameters), as we show in Table~\ref{tab:varying-epochs} for the case of \gls{MLP} on MNIST. 

Similarly, we plot the training loss curves for RoBERTa-base on the RTE dataset in Figure~\ref{fig:train_loss_rte}. We observe that all setups reach the same value at convergence with similar speed.



\begin{table}[ht]
\centering
\begin{tabular}{ccccccc}
\toprule
& \multicolumn{6}{c}{\textbf{Number of epochs per block}} \\\cmidrule(lr){2-7}
& \textbf{1} & \textbf{2} & \textbf{3} & \textbf{4} & \textbf{5} & \textbf{10} \\
\midrule

\algname{COLA} & 96.2 & 95.8 & 95.9 & 95.1 & 95.4 & 94.5 \\

\algname{\gls{RAC-LoRA}} & 96.1 & 95.6 & 95.6 & 94.9 & 94.7 & 93.9 \\

\bottomrule
\end{tabular}
\caption{Accuracy at varying epochs for each block in the chained methods (\algname{COLA} and \algname{\gls{RAC-LoRA}}). The setup is the same as in Table~\ref{tab:mnist_rank1}, with a zero-initialized $A$ matrix and a Gaussian-initialized $B$ matrix. To ensure a fair comparison, the product of the number of epochs per block and the number of blocks is kept constant at 50. The number of trainable parameters for \algname{COLA} and \algname{\gls{RAC-LoRA}} are 1K and 912, respectively.}
\label{tab:varying-epochs}
\end{table}

\begin{figure}
    \centering
    \includegraphics[width=0.8\linewidth]{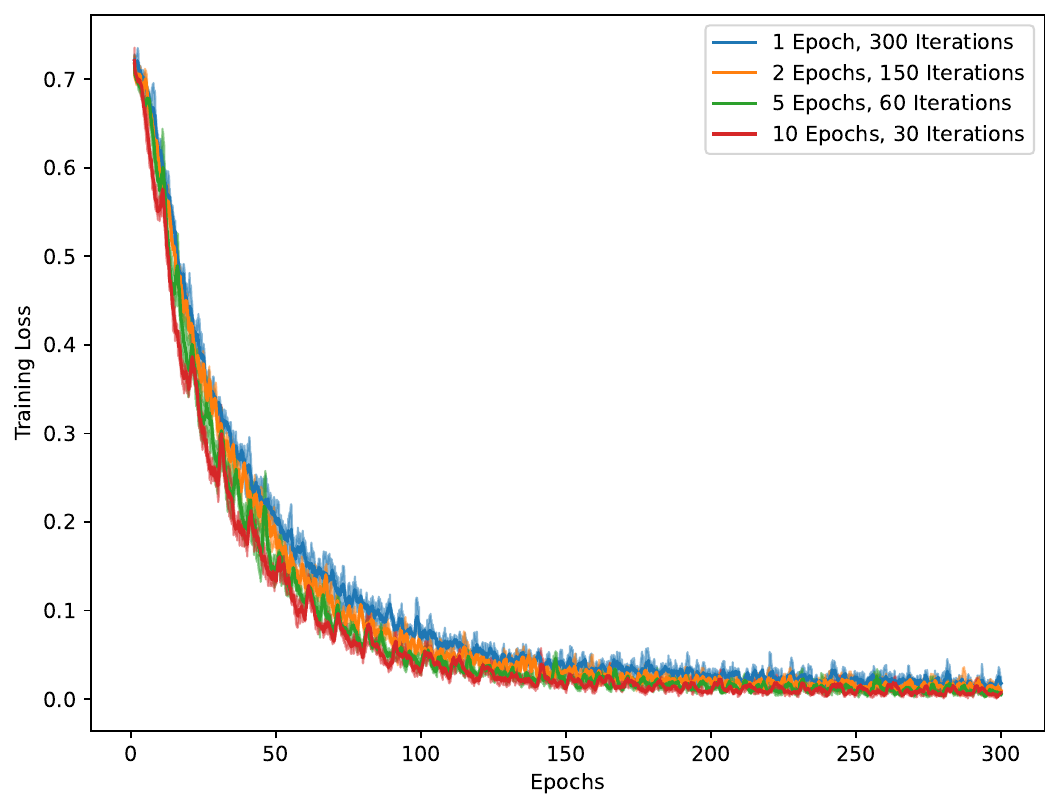}
    \caption{\algname{\gls{RAC-LoRA}} training loss curves at a fixed computational budget for varying epochs for each block in the chain. RoBERTa-base with rank 2.}
    \label{fig:train_loss_rte}
\end{figure}



\section{Analysis of RAC-LoRA with Gradient Descent} \label{sec:GD-proofs}
\subsection{Proof of Theorem \ref{thm:GD}}
\label{sec:GD-pr}


The proof is provided for Left Sketch (Definition \ref{def:left}). The result for Right Sketch (Definition \ref{def:right}) can be derived by following the same steps.
\begin{proof}
We begin by examining the implications of Assumption~\ref{asm:L-smooth}. The relationships between various conditions associated with Assumption~\ref{asm:L-smooth} are discussed in detail in \citep{NesterovBook}.
\begin{align*}
    f(W^{t+1}) \leq& f(W^t) + \left\langle \nabla f(W^t), W^{t+1} - W^t \right \rangle + \frac{L}{2}\left\| W^{t+1} - W^t \right\|^2
\end{align*}
Using the update rule $W^{t+1} = W^t - \gamma H^t_B \nabla f(W^t)$ we get
\begin{align*}
   f(W^{t+1})      \leq& f(W^t) + \left\langle \nabla f(W^t), - \gamma H^t_B \nabla f(W^t) \right \rangle + \frac{L}{2}\left\|  - \gamma H^t_B \nabla f(W^t) \right\|^2\\
   \leq& f(W^t) -\gamma \left\langle \nabla f(W^t),  H^t_B \nabla f(W^t) \right \rangle + \frac{L}{2} \gamma^2 \left\|    H^t_B \nabla f(W^t) \right\|^2\\
      \leq& f(W^t) -\gamma \left\langle \nabla f(W^t),  H^t_B \nabla f(W^t) \right \rangle\\
      &+ \frac{L}{2} \gamma^2 \left\langle H^t_B \nabla f(W^t),  H^t_B \nabla f(W^t) \right \rangle\\
            \leq& f(W^t) -\gamma \left\langle \nabla f(W^t),  H^t_B \nabla f(W^t) \right \rangle\\
            &+ \frac{L}{2} \gamma^2 \left\langle  \nabla f(W^t),  (H^t_B)^\top H^t_B \nabla f(W^t) \right \rangle.
\end{align*}
Since matrix $H^t_B$ is projection matrix, we have $(H^t_B)^\top H^t_B = (H^t_B)^2 = H^t_B$:
\begin{align*}
   f(W^{t+1})  \leq& f(W^t) -\gamma \left\langle \nabla f(W^t),  H^t_B \nabla f(W^t) \right \rangle + \frac{L}{2} \gamma^2 \left\langle  \nabla f(W^t),  H^t_B \nabla f(W^t) \right \rangle.
\end{align*}
Using the fact that $\gamma\leq \frac{1}{L}$ we have 
\begin{align*}
   f(W^{t+1})  \leq& f(W^t) -\frac{\gamma}{2} \left\langle \nabla f(W^t),  H^t_B \nabla f(W^t) \right \rangle.
\end{align*}
Taking expectation we get
\begin{align*}
   \mathbb{E} \left[ f(W^{t+1}) \mid W^t \right]  \leq&    \mathbb{E} \left[f(W^t) -\frac{\gamma}{2} \left\langle \nabla f(W^t),  H^t_B \nabla f(W^t) \right \rangle  \mid W^t \right]\\
   \leq & f(W^t) -\frac{\gamma}{2} \left\langle \nabla f(W^t),  \mathbb{E} \left[ H^t_B \right] \nabla f(W^t) \right \rangle  
   \end{align*}
   Using Assumption \ref{asm:lambda} we have 
   \begin{align*}
   \mathbb{E} \left[ f(W^{t+1}) \mid W^t \right]  \leq&    \mathbb{E} \left[f(W^t) -\frac{\gamma}{2} \left\langle \nabla f(W^t),  H^t_B \nabla f(W^t) \right \rangle  \mid W^t \right]\\
   \leq & f(W^t) -\frac{\gamma}{2}\lambda_{\min}^{H_B}\left\| \nabla f(W^t)\right\|^2.
   \end{align*}
    Subtracting $f^\star$ from both sides we get
       \begin{align}
       \label{eq:GD for PL}
   \mathbb{E} \left[ f(W^{t+1}) \mid W^t \right]  - f^\star
   \leq & f(W^t) - f^\star -\frac{\gamma}{2}\lambda^{H_B}_{\min}\left\| \nabla f(W^t)\right\|^2.
   \end{align}
  Now we can rewrite as 
   \begin{align*}
       \frac{\gamma}{2}\lambda^{H_B}_{\min}\left\| \nabla f(W^t)\right\|^2 \leq    \left(  f(W^t) - f^\star  \right) - \left( \mathbb{E} \left[ f(W^{t+1}) \mid W^t \right]  - f^\star\right)
   \end{align*}
   Taking expectation and using tower property we obtain
      \begin{align*}
       \frac{\gamma}{2}\lambda^{H_B}_{\min}\mathbb{E}\left[\left\| \nabla f(W^t)\right\|^2 \right] \leq    e^t - e^{t+1},
   \end{align*}
   where $e^t =\mathbb{E} \left[ f(W^{t}) \right]  - f^\star $. Now we can sum these inequalities together and get
         \begin{align*}
     \sum^{T-1}_{t=0}  \frac{\gamma}{2}\lambda^{H_B}_{\min}\mathbb{E}\left[\left\| \nabla f(W^t)\right\|^2 \right] \leq  \sum^{T-1}_{t=0}  \left(  e^t - e^{t+1} \right),
   \end{align*}
   Using telescoping property of $e^t - e^{t+1}$ we get 
         \begin{align*}
     \sum^{T-1}_{t=0}  \frac{\gamma}{2}\lambda^{H_B}_{\min}\mathbb{E}\left[\left\| \nabla f(W^t)\right\|^2 \right] \leq e^0 - e^T.
   \end{align*}
   Once we divide by $T$ we obtain
            \begin{align*}
   \frac{1}{T}  \sum^{T-1}_{t=0}  \frac{\gamma}{2}\lambda^{H_B}_{\min}\mathbb{E}\left[\left\| \nabla f(W^t)\right\|^2 \right] &\leq \frac{e^0 - e^T}{T} \leq \frac{e^0 }{T}.
   \end{align*}
   Finally, we get
               \begin{align*}
   \frac{1}{T}  \sum^{T-1}_{t=0}  \mathbb{E}\left[\left\| \nabla f(W^t)\right\|^2 \right] \leq \frac{2 (f(W^0) - f^\star)}{\lambda^{H_B}_{\min} \gamma T}.
   \end{align*}
   Applying the argument from \citep{danilova2022recent}, we obtain the result for a uniformly chosen point.

\end{proof}

\subsection{Proof of Theorem \ref{thm:PL-GD}}
\label{sec:GD-PL}


The proof is provided for Left Sketch (Definition \ref{def:left}). The result for Right Sketch (Definition \ref{def:right}) can be derived by following the same steps.


\begin{proof}

We start from the inequality (\ref{eq:GD for PL}):
       \begin{align*}
   \mathbb{E} \left[ f(W^{t+1}) \mid W^t \right]  - f^\star
   \leq & f(W^t) - f^\star -\frac{\gamma}{2}\lambda^{H_B}_{\min}\left\| \nabla f(W^t)\right\|^2.
   \end{align*}
       Using \gls{PL} condition $\left\| \nabla f(W^t) \right\|^2 \geq 2\mu\left( f(W^t) - f^\star \right)$ we have
     \begin{align*}
            \mathbb{E} \left[ f(W^{t+1}) \mid W^t \right]  - f^\star
   \leq & f(W^t) - f^\star -\gamma \mu\lambda^{H_B}_{\min}\left( f(W^t) - f^\star \right)\\
    \leq & \left(1 - \gamma \mu \lambda^{H_B}_{\min}\right)\left( f(W^t) - f^\star \right).
     \end{align*}
     Once we unroll the recursion we get
     \begin{align*}
         \mathbb{E}\left[ f(W^{T}) \right] -f^\star  \leq  \left(1 - \gamma \mu \lambda^{H_B}_{\min}\right)^T\left( f(W^0) - f^\star \right).
     \end{align*}
     In order to obtain $\varepsilon$ solution we need to take
     \begin{align*}
    T \geq \mathcal{O} \left( \frac{L}{\mu} \frac{1}{\lambda^{H_B}_{\min}} \log \frac{1}{\varepsilon}\right).
\end{align*}

\end{proof}

\clearpage
\section{Analysis of RAC-LoRA with Random Reshuffling} \label{sec:RR-LoRA}

The previous results were obtained using full gradients. However, this approach is impractical in deep learning settings, where calculating full gradients is often infeasible. To analyze stochastic methods, we consider problem (\ref{eq:main}), where $f$ has the following special structure: 
\begin{align}
\label{eq:finite}
f(W^0 + \Delta W) := \frac{1}{N}\sum_{i=1}^{N} f_i(W^0+\Delta W) , 
\end{align}
where each function \(f_i\) represents the individual loss function for one sample and $N$ is total number of datapoints. Next, we analyze a practical variant of stochastic gradient descent (\algname{SGD}) known as Random Reshuffling (\algname{\gls{RR}}), which involves sampling without replacement. In this method, the dataset is shuffled according to a permutation, ensuring that each training sample is used exactly once during each epoch. 

\algname{\gls{RR}} is a variant of \algname{SGD} in which each data point is used exactly once per epoch, also known as \algname{SGD} with sampling without replacement. Many efforts have been made to explain why gradient methods with reshuffling perform so well in practice, across different types of problems. The convergence rates for incremental gradient methods with random reshuffling in convex optimization were first explored in \citep{nedic2001convergence} and later in \citep{bertsekas2011incremental}. In recent years, a lot of focus has shifted toward strongly convex problems, with studies showing that \algname{\gls{RR}} can outperform \algname{SGD}. For example, work \citep{recht2012toward} was among the first to analyze this for quadratic least squares problems. 

Researchers have also managed to improve results and remove some of the earlier assumptions, such as second-order smoothness, as seen in works \citep{jain2019sgd}, \citep{safran2021random} and \citep{MKR2020rr}. These studies introduced a new way to account for the random permutation's variance, making it easier to analyze both convex and strongly convex cases. There have even been extensions into non-convex settings, with results under the \gls{PL} condition \citep{ahn2020sgd, nguyen2021unified} and general non-convex smooth cases \citep{lu2022general, MKR2020rr, malinovsky2023random}. More recently, tighter lower bounds for strongly convex and \gls{PL} functions have been developed \citep{cha2023tighter}.

In recent years, there’s also been growing interest in applying these reshuffling techniques to distributed and Federated Learning, which is crucial for training large-scale, decentralized models \citep{yun2021minibatch, malinovsky2023server, sadiev2022federated,mishchenko2022proximal, cho2023convergence, malinovsky2022federated, malinovsky2023federated, horvath2022fedshuffle}.


To analyze stochastic methods, we need to make assumptions about the variance. The standard assumption is that the variance is bounded:
\begin{assumption}
\label{asm:var}
There exist nonnegative constants $\sigma\geq 0$ such that for any $W^t \in \mathbb{R}^{m\times n}$ we have,
\begin{align*}
 \frac{1}{N} \sum_{i=1}^N\left\|\nabla f_i\left(W^t\right)-\nabla f\left(W^t\right)\right\|^2 \leq \sigma^2.   
\end{align*}

\end{assumption}


The proof is provided for Left Sketch (Definition \ref{def:left}). The result for Right Sketch (Definition \ref{def:right}) can be derived by following the same steps.

We consider a method belonging to the class of data permutation methods which is the \algname{\gls{RR}} algorithm. In each epoch $t$ of \algname{\gls{RR}}, we sample indices $\pi_0, \pi_1, \ldots, \pi_{N-1}$ without replacement from $\{1,2, \ldots, N\}$, i.e., $\left\{\pi_0, \pi_1, \ldots, \pi_{N-1}\right\}$ is a random permutation of the set $\{1,2 \ldots N\}$ and proceed with $N$ iterates of the form:
\begin{align*}
    W^t_{i+1} &= W^t_i - \gamma H^t_B \nabla f_{\pi_i}(W^t_i).
\end{align*}
We then set $W^{t+1} = W^t_N$ , and repeat the process for a total of $T$ \algname{LoRA} blocks. We can derive the effective step:
 \begin{align}   
 \label{eq:RR}
     W^{t+1} &= W^t - \gamma H^t_B\sum^{N-1}_{i=0} \nabla f(W^t_i)= W^t - \gamma H^t_B N \hat{g}^t,
\end{align}
where $\hat{g}^t = \frac{1}{N} \sum^{N-1}_{i=0} \nabla f(W^t_i). $ 

\subsection{Analysis of general non-convex setting}
\label{sec:RR-gen}
\begin{theorem}
\label{thm:RR}
Suppose that Assumption \ref{asm:L-smooth} and Assumption \ref{asm:lambda} hold. Suppose that a stepsize $\gamma > 0$ is chosen such that $\gamma\leq \frac{1}{2LN}$. We choose the output of the method $\widetilde{W}^T$ uniformly at random from $W^0, W^1,\ldots,W^{T-1}$ Then, the iterate $\widetilde{W}^T$ of the \algname{\gls{RAC-LoRA}} method (Algorithm \ref{alg:RAC-LoRA}) with \algname{\gls{RR}} updates (Equation (\ref{eq:RR})) satisfies
\begin{align*}
 \mathbb{E}\left[\left\| \nabla f(\widetilde{W}^T)\right\|^2 \right]\leq & \frac{2}{\gamma N T} \frac{f(W^0) - f^\star }{\left( 1- \lambda_{\max} \left[ \mathbb{E}\left[I-H^t\right] \right] - \frac{1 }{4}\lambda_{\max}^{H} \right)   } \\
 &+ \frac{L^2\gamma^2 \lambda_{\max}^{H} N \sigma^2}{\left( 1- \lambda_{\max} \left[ \mathbb{E}\left[I-H^t\right] \right] - \frac{1}{4}\lambda_{\max}^{H} \right)   }   .
\end{align*}
\end{theorem}

Remark: Notice that if we choose $\gamma = {\cal O}(1/T)$, the above result yields the rate ${\cal O}(1/T^2)$.

\begin{proof}
In this context, and in subsequent discussions, the notation \(\|\cdot\|\) refers to the Frobenius norm, while \(\langle \cdot \rangle\) denotes the inner product associated with the Frobenius norm.

Now we can apply the $L$-smoothness:
\begin{align*}
    f(W^{t+1}) &\leq f(W^t) + \left\langle \nabla f(W^t), W^{t+1} - W^t \right\rangle + \frac{L}{2}\left\| W^{t+1} - W^t \right\|^2\\
    & = f(W^t) + \left\langle \nabla f(W^t), -\gamma H^t_B N \hat{g}^t\right\rangle + \frac{L}{2}\left\| \gamma H^t_B N \hat{g}^t\right\|^2\\
    &=f(W^t)  -\gamma N\left\langle \nabla f(W^t), H^t_B  \hat{g}^t\right\rangle + \frac{L}{2} \gamma^2 N^2\left\| H^t_B \hat{g}^t\right\|^2\\
        &=f(W^t)  -\frac{\gamma N}{2}\left( \left\| \nabla f(W^t) \right\|^2 + \left\| H_B^t \hat{g}^t \right\|^2 - \left\| \nabla f(W^t) - H_B^t \hat{g}^t \right\|^2 \right)\\
        &+ \frac{L}{2} \gamma^2 N^2\left\| H_B^t \hat{g}^t\right\|^2\\
                &=f(W^t)  -\frac{\gamma N}{2}\left( \left\| \nabla f(W^t) \right\|^2 + \left\| H_B^t \hat{g}^t \right\|^2 - \left\| \nabla f(W^t) - H_B^t \hat{g}^t \right\|^2 \right)\\
                &+ \frac{L}{2} \gamma^2 N^2\left\| H_B^t \hat{g}^t\right\|^2\\
                & = f(W^t) - \frac{\gamma N}{2}\left\| \nabla f(W^t) \right\|^2 - \frac{\gamma N}{2}\left\| H_B^t \hat{g}^t \right\|^2\left( 1-\gamma L N \right)\\
                &+ \frac{\gamma N}{2} \left\| \nabla f(W^t) - H_B^t \hat{g}^t \right\|^2.
\end{align*}
Using $\gamma \leq \frac{1}{L N}$ we get 
\begin{align*}
    f(W^{t+1}) 
                & \leq f(W^t) - \frac{\gamma N}{2}\left\| \nabla f(W^t) \right\|^2 + \frac{\gamma N}{2} \left\| \nabla f(W^t) - H^t_B \hat{g}^t \right\|^2.
\end{align*}
Let us take expectation and subtract $f^\star$:
\begin{align*}
\mathbb{E}\left[    f(W^{t+1}) \mid W^t \right] - f^\star
                & \leq f(W^t) - f^\star - \frac{\gamma N}{2}\left\| \nabla f(W^t) \right\|^2\\
                &+ \frac{\gamma N}{2} \mathbb{E} \left[ \left\| \nabla f(W^t) - H_B^t \hat{g}^t \right\|^2 \mid W^t \right].
\end{align*}
Let us consider the last term:
\begin{align*}
 \mathbb{E}&\left[    \left\| \nabla f(W^t) - H_B^t \hat{g}^t \right\|^2 \mid W^t \right]\\
 & =  \mathbb{E}\left[   \left\| \frac{1}{N} \sum^{N-1}_{i=0} \nabla f_{\pi_i}(W^t) - H^t_B\frac{1}{N}\sum^{N-1}_{i=0}\nabla f_{\pi_i}(W^t_i) \right\|^2 \mid W^t \right]\\
     & =  \mathbb{E}\left[\left\| \frac{1}{N} \sum^{N-1}_{i=0} \nabla f_{\pi_i}(W^t) + H^t_B\frac{1}{N} \sum^{N-1}_{i=0} \nabla f_{\pi_i}(W^t) -\right.\right.\\
     &\left.\left. - H^t_B\frac{1}{N} \sum^{N-1}_{i=0} \nabla f_{\pi_i}(W^t) - H_B^t \frac{1}{N}\sum^{N-1}_{i=0}\nabla f_{\pi_i}(W^t_i) \right\|^2 \mid W^t \right]
\end{align*}
Since $I-H^t_B$ and $H^t_B$ are projection matrices generating  perpendicular subspaces we have 
\begin{align*}
 \mathbb{E}&\left[    \left\| \nabla f(W^t) - H_B^t \hat{g}^t \right\|^2 \mid W^t \right]\\
     &=  \mathbb{E}\left[ \left\| \left(I - H_B^t\right) \nabla f(W^t) \right\|^2 + \left\| H_B^t \frac{1}{n} \sum^{ n-1}_{i=0}\left( \nabla f_{\pi_i}(W^t) - f_{\pi_i}(W^t_i) \right) \right\|^2 \mid W^t\right]\\
     & =  \mathbb{E}\left[ \left\langle \left(I - H_B^t\right) \nabla f(W^t) , \left(I - H_B^t\right) \nabla f(W^t) \right\rangle + \right.\\
     &+\left.\left\| H^t_B \frac{1}{N} \sum^{ N-1}_{i=0}\left( \nabla f_{\pi_i}(W^t) - f_{\pi_i}(W^t_i) \right) \right\|^2 \mid W^t\right].
\end{align*}
Using the property that $H^t_B$ and $I-H^t_B$ are projection matrices we obtain
\begin{align*}
 \mathbb{E}&\left[    \left\| \nabla f(W^t) - H^t \hat{g}^t \right\|^2 \mid W^t \right]\\
 & \leq \lambda_{\max} \left[ \mathbb{E}\left[I-H^t\right] \right] \left\| \nabla f(W^t) \right\|^2 + \mathbb{E}\left[ \lambda_{\max}[H^t] L^2 \frac{1}{N} \sum^{N-1}_{i=0} \left\| W^t - W^t_i  \right\|^2 \right].
 \end{align*}
 Since $\lambda_{\max}[H^t] = 1$ for projections matrix we get
 \begin{align*}
 \mathbb{E}&\left[    \left\| \nabla f(W^t) - H^t_B \hat{g}^t \right\|^2 \mid W^t \right] \leq \lambda_{\max} \left[ \mathbb{E}\left[I-H^t_B\right] \right] \left\| \nabla f(W^t) \right\|^2\\
 &+  L^2 \frac{1}{N} \sum^{N-1}_{i=0} \mathbb{E}\left[ \left\| W^t - W^t_i  \right\|^2 \mid W^t \right].
 \end{align*}
 Now let us consider the last term:
\begin{align*}
  &  \mathbb{E}\left[ \left\| W^t - W^t_k \right\|^2 \right]\\
  &= \gamma^2 \mathbb{E}\left[ \left\| \sum^{k-1}_{i=0} H^t_B \nabla f_{\pi_i}(W^t_i) \right\|^2 \mid W^t \right]\\
    & = \gamma^2  \mathbb{E}\left[ \left\| \sum^{k-1}_{i=0} H^t_B \left( \nabla f_{\pi_i}(W^t_i)  - \nabla f_{\pi_i}(W^t) \right) + \sum^{k-1}_{i=0} H^t_B \nabla f_{\pi_i}(W^t)  \right\|^2 \mid W^t\right]\\
    & \leq 2 \gamma^2 k \mathbb{E}\left[\sum^{k-1}_{i=0} \left(  \left\| H^t_B \left( \nabla f_{\pi_i}(W^t_i)  - \nabla f_{\pi_i}(W^t) \right)  \right\|^2 +\right.\right.\\
    &\left.\left.+ 2\gamma^2 k^2 \left\| H^t_B \nabla f_{\pi_i}(W^t) \right\|^2\right)\mid W^t \right]\\
        & \leq 2 \gamma^2 k \mathbb{E}\left[ \sum^{k-1}_{i=0} \left( \lambda_{\max}\left[ H^t \right] \left\| W^t_i  - W^t   \right\|^2  + 2\gamma^2 k^2 \left\| H^t_B  \nabla f_{\pi_i}(W^t) \right\|^2\right) \mid W^t \right]\\
                & \leq 2 \gamma^2 k \mathbb{E}\left[ \sum^{k-1}_{i=0}  \left( \left\| W^t_i  - W^t   \right\|^2  + 2\gamma^2 k^2 \lambda_{\max}\left[ \mathbb{E}\left[H^t_B\right] \right]\left\|  \nabla f_{\pi_i}(W^t) \right\|^2\right)\mid W^t \right]. 
\end{align*}
Now, we are ready to sum the inequalities. By using the definition of the expected smallest eigenvalue \( \lambda_{\max}\left[ \mathbb{E}\left[H^t\right] \right] = \lambda_{\max}^{H_B} \) and applying Lemma 1 from \citep{MKR2020rr} with Assumption \ref{asm:var}, we obtain:
 \begin{align*}
    & \sum^{N-1}_{i=0}    \mathbb{E}\left[ \left\| W^t - W^t_i \right\|^2 \right]\\
     &\leq     \mathbb{E}\left[  \sum^{N-1}_{i=0} \left( 2 \gamma^2 k \sum^{k-1}_{i=0}   \left\| W^t_i  - W^t   \right\|^2  + 2\gamma^2 k^2 \lambda_{\max}^{H_B}\left\|  \nabla f_{\pi_i}(W^t) \right\|^2 \right)\mid W^t \right]\\
      \leq & \gamma^2 L^2 N(N-1)      \sum^{N-1}_{i=0}    \mathbb{E}\left[ \left\| W^t - W^t_k \right\|^2 \right]\\
     &+ \frac{1}{3}\gamma^2(N-1)N(2N-1)\lambda_{\max}^{H_B}\left\|  \nabla f(W^t) \right\|^2 + \frac{1}{3} \lambda_{\max}^{H_B} \gamma^2 N(N+1) \sigma^2.
 \end{align*}
 Using $\gamma\leq \frac{1}{2LN}$ we get 
 \begin{align*}
         & \sum^{N-1}_{i=0}    \mathbb{E}\left[ \left\| W^t - W^t_i \right\|^2 \right]\\
         &\leq \frac{4}{3} \left( 1-\gamma^2 L^2 N(N-1) \right)           \sum^{N-1}_{i=0}    \mathbb{E}\left[ \left\| W^t - W^t_i \right\|^2 \right] \\
          &\leq \frac{4}{3} \left( \frac{1}{3}\gamma^2(N-1)N(2N-1)\lambda_{\max}^{H_B}\left\|  \nabla f(W^t) \right\|^2  + \frac{1}{3} \lambda_{\max}^{H_B} \gamma^2 N(N+1) \sigma^2\right)\\
          &\leq \gamma^2 N^3 \lambda_{\max}^{H_B} \left\|  \nabla f(W^t) \right\|^2 +  \gamma^2 \lambda_{\max}^{H_B} N^2 \sigma^2
 \end{align*}
 Plugging to the previous bound we obtain:
 \begin{align*}
      \mathbb{E}\left[    \left\| \nabla f(W^t) - H^t \hat{g}^t \right\|^2 \mid W^t \right]&\leq \lambda_{\max} \left[ \mathbb{E}\left[I-H^t_B\right] \right] \left\| \nabla f(W^t) \right\|^2\\
      &+  L^2 \gamma^2 N^2 \lambda_{\max}^{H_B} \left\|  \nabla f(W^t) \right\|^2\\
      &+L^2\gamma^2 \lambda_{\max}^{H_B} N \sigma^2.
\end{align*}
Now we have the following 
\begin{align*}
&\mathbb{E}\left[    f(W^{t+1}) \mid W^t \right] - f^\star \\
               & \leq  f(W^t) - f^\star - \frac{\gamma N}{2}\left\| \nabla f(W^t) \right\|^2\\
                &+ \frac{\gamma N}{2} \left( \lambda_{\max} \left[ \mathbb{E}\left[I-H^t_B\right] \right] \left\| \nabla f(W^t) \right\|^2 +  L^2 \gamma^2 N^2 \lambda_{\max}^{H_B} \left\|  \nabla f(W^t) \right\|^2\right)\\
                &+ \frac{\gamma N}{2} L^2\gamma^2 \lambda_{\max}^{H_B} N \sigma^2.
\end{align*}
Using $\gamma\leq \frac{1}{2LN}$ we get 
\begin{align}
\label{eq:rr-pl-1}
&\notag \mathbb{E}\left[    f(W^{t+1}) \mid W^t \right] - f^\star \\
                &\notag \leq  f(W^t) - f^\star - \frac{\gamma N}{2}\left\| \nabla f(W^t) \right\|^2 \left( 1- \lambda_{\max} \left[ \mathbb{E}\left[I-H^t_B\right] \right] - \frac{1}{4}\lambda_{\max}^{H_B} \right)\\
                                &+ \frac{\gamma N}{2} L^2\gamma^2 \lambda_{\max}^{H_B} N \sigma^2.
\end{align}
After rearranging the terms, we have 
\begin{align*}
& \frac{\gamma N}{2}\left\| \nabla f(W^t) \right\|^2 \left( 1- \lambda_{\max} \left[ \mathbb{E}\left[I-H^t_B\right] \right] - \frac{1}{4}\lambda_{\max}^{H_B} \right)\\
           &\leq  \left(f(W^t) - f^\star \right) - \left(\mathbb{E}\left[    f(W^{t+1}) \mid W^t \right] - f^\star \right) \\
                                &+ \frac{\gamma N}{2} L^2\gamma^2 \lambda_{\max}^{H_B} N \sigma^2.
\end{align*}
Next, we have 
\begin{align*}
\left\| \nabla f(W^t) \right\|^2 
                \leq & \frac{2}{\gamma N} \frac{1}{\left( 1- \lambda_{\max} \left[ \mathbb{E}\left[I-H^t_B\right] \right] - \frac{1}{4}\lambda_{\max}^{H_B} \right)   } \times\\
                &\times\left( \left(f(W^t) - f^\star \right) - \left(\mathbb{E}\left[    f(W^{t+1}) \mid W^t \right] - f^\star \right) \right) \\
                                &+ \frac{2}{\gamma N} \frac{1}{\left( 1- \lambda_{\max} \left[ \mathbb{E}\left[I-H^t_B\right] \right] - \frac{1}{4}\lambda_{\max}^{H_B} \right)   }   \frac{\gamma N}{2} L^2\gamma^2 \lambda_{\max}^{H_B} N \sigma^2.
\end{align*}
Using telescoping property and taking expectation we get
\begin{align*}
    \frac{1}{T} \sum^{T-1}_{t=0} \left\| \nabla f(W^t) \right\|^2 &\leq \frac{2}{\gamma N T} \frac{f(W^0) - f^\star }{\left( 1- \lambda_{\max} \left[ \mathbb{E}\left[I-H^t_B\right] \right] - \frac{1 }{4}\lambda_{\max}^{H_B} \right)   } \\
    & + \frac{L^2\gamma^2 \lambda_{\max}^{H_B} N \sigma^2}{\left( 1- \lambda_{\max} \left[ \mathbb{E}\left[I-H^t_B\right] \right] - \frac{1}{4}\lambda_{\max}^{H_B} \right)   }   .
\end{align*}
   Applying argument from \citep{danilova2022recent} we obtain the result for uniformly chosen point.
   \end{proof}

   \begin{corollary}
       \label{cor-RR-gen}
       Suppose that Assumption \ref{asm:L-smooth} and Assumption \ref{asm:lambda} hold. Suppose that a stepsize $\gamma > 0$ is chosen such that $\gamma\leq \frac{1}{2LN}$. Let the updates have a form of several gradient steps (variance $\sigma^2 = 0$). We choose the output of the method $\widetilde{W}^T$ uniformly at random from $W^0, W^1,\ldots,W^{T-1}$ Then, the iterate $\widetilde{W}^T$ of the \algname{\gls{RAC-LoRA}} method (Algorithm \ref{alg:RAC-LoRA}) with several \algname{\gls{GD}} updates (Equation (\ref{eq:left_GD})) satisfies
\begin{align*}
 \mathbb{E}\left[\left\| \nabla f(\widetilde{W}^T)\right\|^2 \right]\leq & \frac{2}{\gamma N T} \frac{f(W^0) - f^\star }{\left( 1- \lambda_{\max} \left[ \mathbb{E}\left[I-H^t\right] \right] - \frac{1 }{4}\lambda_{\max}^{H} \right)   }.
\end{align*}
   \end{corollary}
   
Given that the step size is divided by the number of gradient steps allocated for each \algname{LoRA} block, employing multiple gradient steps for a single \algname{LoRA} block does not provide any significant benefits. This observation suggests that a single gradient step is adequate for each \algname{LoRA} block. Therefore, in practical applications, it is more advantageous to utilize only one epoch per \algname{LoRA} block within the training chain. This approach not only streamlines the training process but also optimizes computational efficiency, allowing for more effective resource allocation without compromising the performance of the model.

\subsection{Analysis of Polyak-Łojasiewicz setting}
\label{sec:RR-PL}
Next, we establish the convergence rate for the Polyak-Łojasiewicz setting (Assumption \ref{asm:PL}).

\begin{theorem}
\label{thm:RR-PL}
Suppose that Assumption \ref{asm:L-smooth}, Assumption \ref{asm:PL} and Assumption \ref{asm:lambda} hold. Suppose that a stepsize $\gamma \geq 0$ is chosen such that $\gamma\leq \frac{1}{2NL}$. Then, the iterates of \algname{\gls{RAC-LoRA}} method (Algorithm~\ref{alg:RAC-LoRA}) with \algname{\gls{RR}} updates (Equation (\ref{eq:RR}))  satisfy
\begin{align*}
  &  \mathbb{E}\left[    f(W^{T})  - f^\star\right]\\
                &\leq  \left(1 - \gamma N \mu\left( 1- \lambda_{\max} \left[ \mathbb{E}\left[I-H^t_B\right] \right] - \frac{1}{4}\lambda_{\max}^{H_B} \right) \right)^T\mathbb{E}\left[f(W^0) - f^\star \right]\\
                                &+ \frac{L^2\gamma^2 \lambda_{\max}^{H_B} N \sigma^2}{2\left( 1- \lambda_{\max} \left[ \mathbb{E}\left[I-H^t_B\right] \right] - \frac{1}{4}\lambda_{\max}^{H_B} \right)} .
\end{align*}
\end{theorem}
\begin{proof}

We start from Equation (\ref{eq:rr-pl-1}):
\begin{align*}
\mathbb{E}\left[    f(W^{t+1}) \mid W^t \right] - f^\star
                \leq & f(W^t) - f^\star\\
                &- \frac{\gamma N}{2}\left\| \nabla f(W^t) \right\|^2 \left( 1- \lambda_{\max} \left[ \mathbb{E}\left[I-H^t_B\right] \right] - \frac{1}{4}\lambda_{\max}^{H_B} \right)\\
                                &+ \frac{\gamma N}{2} L^2\gamma^2 \lambda_{\max}^{H_B} N \sigma^2.
\end{align*}
       Using \gls{PL} condition $\left\| \nabla f(W^t) \right\|^2 \geq 2\mu\left( f(W^t) - f^\star \right)$ we have
\begin{align*}
&\mathbb{E}\left[    f(W^{t+1}) \mid W^t \right] - f^\star\\
                &\leq  f(W^t) - f^\star\\
                &- \gamma N\mu\left( f(W^t) - f^\star \right) \left( 1- \lambda_{\max} \left[ \mathbb{E}\left[I-H^t_B\right] \right] - \frac{1}{4}\lambda_{\max}^{H_B} \right)\\
                                &+ \frac{\gamma N}{2} L^2\gamma^2 \lambda_{\max}^{H_B} N \sigma^2.
\end{align*}
Taking full expectation we obtain:
\begin{align*}
  &  \mathbb{E}\left[    f(W^{t+1})  - f^\star\right]\\
               & \leq  \left(1 - \gamma N \mu\left( 1- \lambda_{\max} \left[ \mathbb{E}\left[I-H^t_B\right] \right] - \frac{1}{4}\lambda_{\max}^{H_B} \right) \right)\mathbb{E}\left[f(W^t) - f^\star \right]\\
                                &+ \frac{\gamma N}{2} L^2\gamma^2 \lambda_{\max}^{H_B} N \sigma^2.
\end{align*}
After unrolling the recursion we obtain
\begin{align*}
    &\mathbb{E}\left[    f(W^{T})  - f^\star\right]\\
                &\leq  \left(1 - \gamma N \mu\left( 1- \lambda_{\max} \left[ \mathbb{E}\left[I-H^t_B\right] \right] - \frac{1}{4}\lambda_{\max}^{H_B} \right) \right)^T\mathbb{E}\left[f(W^0) - f^\star \right]\\
                                &+ \frac{L^2\gamma^2 \lambda_{\max}^{H_B} N \sigma^2}{2\left( 1- \lambda_{\max} \left[ \mathbb{E}\left[I-H^t_B\right] \right] - \frac{1}{4}\lambda_{\max}^{H_B} \right)} .
\end{align*}
This finishes the proof.
\end{proof}

\begin{corollary}
\label{cor-RR-PL}
    Suppose that Assumption \ref{asm:L-smooth}, Assumption \ref{asm:PL} and Assumption \ref{asm:lambda} hold. Let the updates have a form of several gradient steps (variance $\sigma^2 = 0$). Suppose that a stepsize $\gamma \geq 0$ is chosen such that $\gamma\leq \frac{1}{2NL}$. Then, the iterates of \algname{\gls{RAC-LoRA}} method (Algorithm \ref{alg:RAC-LoRA}) with several \algname{\gls{GD}} updates (Equation (\ref{eq:left_GD}))  satisfy

    \begin{align*}
    \mathbb{E}&\left[    f(W^{T})  - f^\star\right]\\
               & \leq  \left(1 - \gamma N \mu\left( 1- \lambda_{\max} \left[ \mathbb{E}\left[I-H^t_B\right] \right] - \frac{1}{4}\lambda_{\max}^{H_B} \right) \right)^T\mathbb{E}\left[f(W^0) - f^\star \right].
\end{align*}
\end{corollary}
Since the step size is divided by the number of gradient steps for each \algname{LoRA} block, using multiple gradient steps does not offer significant advantages. Thus, a single gradient step per \algname{LoRA} block is sufficient. Practically, it is more efficient to use only one epoch per \algname{LoRA} block in the training chain.

\clearpage
\section{Analysis of RAC-LoRA with SGD Under the Arbitrary Data Sampling Paradigm}\label{sec:SGD}

In the previous section, we introduced the Random Reshuffling (\algname{\gls{RR}}) method, where each data point is used exactly once during each epoch, also known as sampling without replacement. This method has demonstrated strong empirical performance across various optimization tasks. However, in this section, we shift our focus to the \algname{\gls{RAC-LoRA}} framework, where Stochastic Gradient Descent (\algname{SGD}) is applied with a more general, arbitrary data sampling procedure, allowing for broader flexibility in how data is selected and used during training.

The analysis of general sampling schemes in \algname{SGD} has garnered significant attention in the literature, particularly in understanding its impact on convergence rates and optimization performance across different problem classes. For strongly convex functions, general sampling methods have been rigorously studied in works such as \citep{gower2019sgd}, which provide detailed convergence guarantees and bounds. In the case of general convex optimization problems, work \citep{khaled2023unified} offers a thorough analysis of the performance of \algname{SGD} under various sampling strategies. Furthermore, for non-convex settings, both in \citep{khaled2023better} and \citep{demidovich2023guide} the authors have explored how general sampling procedures influence the convergence behavior and optimization efficiency of \algname{SGD}, shedding light on its applicability to a wide range of machine learning tasks.

In the following sections, we build on these foundational studies to examine how the flexibility of general sampling in the \algname{\gls{RAC-LoRA}} framework can lead to improved convergence in certain scenarios, while also maintaining robust performance across different convexity settings.

To conduct this analysis, we introduce a general assumption that extends the standard assumptions presented in \citep{khaled2023better}. 

The proof is provided for Right Sketch (Definition \ref{def:right}). The result for Left Sketch (Definition \ref{def:left}) can be derived by following the same steps.

\begin{assumption}[Expected smoothness] The second moment of the stochastic gradient satisfies
\label{asm:ABC}
$$
\mathbb{E}\left[\|g(W)\|^2\right] \leq 2 A_1\left(f(W)-f^{\mathrm{inf}}\right)+B_1 \cdot\|\nabla f(W)\|^2+C_1
$$

for some $A, B, C \geq 0$ and all $W \in \mathbb{R}^{m\times n}$.
\end{assumption}

Now we can also do stochastic analysis. Let us consider the \algname{SGD} update for \algname{LoRA} method:
$$ \Delta W = \frac{\alpha}{r} \hat{B} A_{S}, $$
\begin{align*}
W^{t+1} = W^t + \frac{\alpha}{r} \hat{B}^t A_{S}^t, \qquad \hat{B}^t = -\gamma g(W^t)(A^t_{S})^\top \left(A^t_{S} (A^t_{S})^\top\right)^\dagger
\end{align*}
Now we have 
\begin{align}
\label{eq:SGD}
    W^{t+1} &= W^t  -\gamma g(W^t)(A^t_{S})^\top \left(A^t_{S}(A^t_{S})^\top\right)^\dagger A^t_{S}
    = W^t  -\gamma  g(W^t){H}^t_A.
\end{align}
\subsection{Analysis of general non-convex setting}
\label{sec:SGD-gen}
\begin{theorem}
\label{thm:SGD-gen}
Suppose that Assumption \ref{asm:L-smooth} and Assumption \ref{asm:lambda} hold. Suppose that a stepsize $\gamma > 0$ is chosen such that $$\gamma\leq \min \left[ 1 / \sqrt{L A_1 \lambda_{\max}^{H} T}, 1/\left(LB_1 \frac{\lambda_{\max}^{H_A} }{\lambda_{\min}^{H_A} }\right) \right].$$ Then, the iterate ${W}^T$ of \algname{\gls{RAC-LoRA}} method (Algorithm \ref{alg:RAC-LoRA}) with \algname{SGD} updates (Equation (\ref{eq:SGD})) satisfies
    \begin{align*}
 \min_{0\leq t \leq T-1} \mathbb{E} \left[ \left\| \nabla f(W^T) \right\|^2 \right] \leq  \frac{6} { \lambda_{\min}^{H_A} \gamma T} \left( f(W^0) - f^\star \right)+ LC_1 \gamma\frac{\lambda_{\max}^{H_A} }{\lambda_{\min}^{H_A} }.
\end{align*}
\end{theorem}

Remark: Notice that if we choose $\gamma = {\cal O}(1/\sqrt{T})$, the above result yields the rate ${\cal O}(1/\sqrt{T})$.

\begin{proof}
We start from $L$-smoothness:
\begin{align*}
    f(W^{t+1}) \leq & f(W^t) + \left\langle \nabla f(W^t), W^{t+1} - W^t \right\rangle + \frac{L}{2}\left\| W^{t+1} - W^t \right\|^2\\
    =& f(W^t) + \left\langle \nabla f(W^t), -\gamma g(W^t) {H}^t_A \right\rangle + \frac{L}{2}\left\| -\gamma g(W^t) {H}^t_A \right\|^2\\
    =& f(W^t) -\gamma  \left\langle \nabla f(W^t), g(W^t) {H}^t_A\right\rangle + \frac{L}{2}\left\| -\gamma g(W^t) {H}^t_A \right\|^2.
\end{align*}
Let us take conditional expectation:
\begin{align*}
    \mathbb{E}\left[f(W^{t+1}) \mid W^t \right] \leq & f(W^t) -\gamma  \mathbb{E}\left[\left\langle \nabla f(W^t), g(W^t) {H}^t_A\right\rangle \mid W^t\right]\\
    &+ \frac{L}{2}\mathbb{E}\left[\left\| -\gamma g(W^t) {H}^t_A \right\|^2 \mid W^t \right].
\end{align*}
Using that $g(W^t)$ and ${H}^t_A$ are independent, so we have 
\begin{align*}
    \mathbb{E}\left[f(W^{t+1}) \mid W^t \right] \leq & f(W^t) -\gamma  \left\langle \nabla f(W^t), \mathbb{E}\left[g(W^t) \right] \mathbb{E}\left[{H}^t_A \right]\right\rangle\\
    &+ \frac{L}{2}\mathbb{E}\left[\left\| -\gamma g(W^t) {H}^t_A \right\|^2 \mid W^t  \right]\\
     \leq & f(W^t) -\gamma  \left\langle \nabla f(W^t), \mathbb{E}\left[g(W^t) \right] \mathbb{E}\left[{H}^t_A\right]\right\rangle\\
     &+ \gamma^2\frac{L}{2}\mathbb{E}\left[\left\langle  g(W^t) {H}^t_A ,  g(W^t) {H}^t_A \right\rangle\mid W^t  \right]\\
     \leq & f(W^t) -\gamma \lambda_{\min}\left[\mathbb{E}\left[{H}^t_A \right]\right] \left\| \nabla f(W^t) \right\|^2\\
     &+ \gamma^2\frac{L}{2}\mathbb{E}\left[\left\langle  g(W^t) {H}^t_A ,  g(W^t) {H}^t_A \right\rangle\mid W^t  \right].
\end{align*}
Using the property of projection matrix ${H}^t_A$, we have 
\begin{align*}
    \mathbb{E}\left[f(W^{t+1}) \mid W^t \right]
     \leq & f(W^t) -\gamma \lambda_{\min}\left[\mathbb{E}\left[{H}^t_A \right]\right] \left\| \nabla f(W^t) \right\|^2\\
     &+ \gamma^2\frac{L}{2} \lambda_{\max}\left[ \mathbb{E}\left[ {H}^t_A  \right] \right]\mathbb{E}\left[ \left\|  g(W^t)  \right\|^2 \right] .
\end{align*}
Now we need to use assumption on stochastic gradients. We will use the most general assumption: ABC -- assumption:
\begin{align*}
    \mathbb{E}\left[ \|g(W^t)\|^2 \right] \leq 2A_1(f(W^t) - f^\star) + B_1\left\|\nabla f(W^t) \right\|^2 +C_1.
\end{align*}
Now we have 
\begin{align*}
     &   \mathbb{E}\left[f(W^{t+1}) \mid W^t \right] - f^\star\\
     &\leq  f(W^t) - f^\star -\gamma \lambda_{\min}\left[\mathbb{E}\left[{H}^t_A \right]\right] \left\| \nabla f(W^t) \right\|^2\\
     +& \gamma^2\frac{L}{2} \lambda_{\max}\left[ \mathbb{E}\left[ {H}^t_A  \right] \right]\left(2A_1(f(W^t) - f^\star) + B_1\left\|\nabla f(W^t) \right\|^2 +C_1.\right) .
\end{align*}
Combining these terms together we get 
\begin{align}
\label{eq:sgd-ref}
\notag   &     \mathbb{E}\left[f(W^{t+1}) \mid W^t \right] - f^\star\\
     &\notag\leq  \left(f(W^t) - f^\star\right) \left( 1+\gamma^2 A_1 L \lambda_{\max}\left[\mathbb{E}\left[{H}^t_A \right]\right]  \right)\\
  \notag   &-\gamma \lambda_{\min}\left[\mathbb{E}\left[{H}^t_A \right]\right] \left\| \nabla f(W^t) \right\|^2\left(1 - \gamma \frac{L}{2} \frac{\lambda_{\max}\left[\mathbb{E}\left[{H}^t_A \right]\right] }{\lambda_{\min}\left[\mathbb{E}\left[{H}^t_A \right]\right] } B_1 \right)\\
     &+ \gamma^2\frac{L}{2} \lambda_{\max}\left[ \mathbb{E}\left[ {H}^t_A  \right] \right]C_1 .
\end{align}
Using condition on stepsize: $1 - \gamma \frac{LB_1}{2} \frac{\lambda_{\max}\left[\mathbb{E}\left[{H}^t_A \right]\right] }{\lambda_{\min}\left[\mathbb{E}\left[{H}^t_A \right]\right] }  \geq \frac{1}{2}$ we get 
\begin{align*}
        \mathbb{E}\left[f(W^{t+1}) \mid W^t \right] - f^\star
     \leq & \left(f(W^t) - f^\star\right) \left( 1+\gamma^2 A_1 L \lambda_{\max}\left[\mathbb{E}\left[{H}^t_A \right]\right]  \right)\\
     &-\frac{1}{2}\gamma \lambda_{\min}\left[\mathbb{E}\left[{H}^t_A \right]\right] \left\| \nabla f(W^t) \right\|^2\\
     &+ \gamma^2\frac{L}{2} \lambda_{\max}\left[ \mathbb{E}\left[ {H}^t_A \right] \right]C_1 .
\end{align*}
Using tower property of expectation we obtain 
\begin{align*}
        \mathbb{E}\left[f(W^{t+1})  - f^\star \right]
     \leq & \mathbb{E}\left[f(W^t) - f^\star \right] \left( 1+\gamma^2 A_1 L \lambda_{\max}\left[\mathbb{E}\left[{H}^t_A\right]\right]  \right)\\
     &-\frac{1}{2}\gamma \lambda_{\min}\left[\mathbb{E}\left[{H}^t_A \right]\right] \mathbb{E} \left[ \left\| \nabla f(W^t) \right\|^2 \right]\\
     &+ \gamma^2\frac{L}{2} \lambda_{\max}\left[ \mathbb{E}\left[ {H}^t_A  \right] \right]C_1 .
\end{align*}
Let us define $\delta^t = \mathbb{E}\left[ f(W^t) - f^\star \right]$ and $r^t = \mathbb{E} \left[ \left\| \nabla f(W^t) \right\|^2 \right]$, after reshuffling of terms we obtain
\begin{align*}
         \frac{1}{2}\gamma \lambda_{\min}\left[\mathbb{E}\left[{H}^t_A\right]\right] \mathbb{E} \left[ \left\| \nabla f(W^t) \right\|^2 \right]& \leq  \left( 1+\gamma^2 A_1 L \lambda_{\max}\left[\mathbb{E}\left[{H}^t_A\right]\right]  \right) \delta^t - \delta^{t+1}\\
         &+ \gamma^2\frac{LC_1}{2} \lambda_{\max}\left[\mathbb{E}\left[{H}^t_A \right]\right].
\end{align*}
Let use fix $w^{-1}>0$ and define $w^t = \frac{w^{t-1}}{1+L\gamma^2 A \lambda_{\max}\left[\mathbb{E}\left[{H}^t_A \right]\right] }$ for all $t\geq 0$. Multiplying by $\frac{w^t}{\gamma }$, 
\begin{align*}
         \frac{1}{2} w^t r^t \lambda_{\min}\left[\mathbb{E}\left[{H}^t_A\right]\right]  \leq & \frac{w^t}{\gamma}\left( 1+\gamma^2 A_1 L \lambda_{\max}\left[\mathbb{E}\left[{H}^t_A\right]\right]  \right) \delta^t - \frac{w^t}{\gamma}\delta^{t+1}\\
         &+ \gamma\frac{LC_1}{2} \lambda_{\max}\left[\mathbb{E}\left[{H}^t_A \right]\right].
\end{align*}
Now we obtain 
\begin{align*}
         \frac{1}{2} w^t r^t \lambda_{\min}\left[\mathbb{E}\left[{H}^t_A\right]\right]  \leq & \frac{w^{t-1}}{\gamma}\delta^t - \frac{w^t}{\gamma}\delta^{t+1} + \gamma\frac{LC_1}{2} \lambda_{\max}\left[\mathbb{E}\left[{H}^t_A\right]\right] w^t.
\end{align*}
Summing up both sides as $t = 0,1,\ldots, T-1$ we have, 
\begin{align*}
         \frac{1}{2} \sum^{T-1}_{t=0} w^t r^t \lambda_{\min}\left[\mathbb{E}\left[{H}^t_A\right]\right]  \leq & \frac{w^{-1}}{\gamma}\delta^0 - \frac{w^{T-1}}{\gamma}\delta^{T} + \gamma\frac{LC_1}{2} \lambda_{\max}\left[\mathbb{E}\left[{H}^t_A\right]\right] \sum^{T-1}_{t=0} w^t\\
         \leq & \frac{w^{-1}}{\gamma}\delta^0 + \gamma\frac{LC_1}{2} \lambda_{\max}\left[\mathbb{E}\left[{H}^t_A\right]\right] \sum^{T-1}_{t=0} w^t.
\end{align*}
Let us define $W^T = \sum^{T-1}_{t=0} w^t$. Dividing both sides by $W^T$ we have, 
\begin{align*}
    \frac{1}{2} \min_{0\leq t \leq T-1} r^t\leq \frac{1}{W^T} \sum^{T-1}_{t=0} w^t r^t \leq \frac{w^{-1}}{W^T} \frac{\delta^0}{\gamma} \frac{1}{\lambda_{\min}\left[\mathbb{E}\left[{H}^t_A\right]\right] } + \frac{LC_1 \gamma}{2} \frac{\lambda_{\max}\left[\mathbb{E}\left[{H}^t_A\right]\right] }{\lambda_{\min}\left[\mathbb{E}\left[{H}^t_A\right]\right] }.
\end{align*}
Note that, 
\begin{align*}
    W^T = \sum^{T-1}_{t=0} w^t \geq \sum^{T-1}_{t=0}  \min_{0\leq i \leq T-1} w^i = T w^{T-1} = \frac{T w^{-1}}{\left(1+L\gamma^2 A \lambda_{\max}\left[\mathbb{E}\left[{H}^t_A \right]\right] \right)^T}.
\end{align*}
Using this we get
\begin{align*}
        \frac{1}{2} \min_{0\leq t \leq T-1} r^t \leq  \frac{\left(1+L\gamma^2 A_1 \lambda_{\max}\left[\mathbb{E}\left[{H}^t_A\right]\right] \right)^T} { \lambda_{\min}\left[\mathbb{E}\left[{H}^t_A \right]\right] \gamma T} \delta^0+ \frac{LC_1 \gamma}{2} \frac{\lambda_{\max}\left[\mathbb{E}\left[{H}^t_A\right]\right] }{\lambda_{\min}\left[\mathbb{E}\left[{H}^t_A\right]\right] }.
\end{align*}

Using the fact that $1+x \leq \exp (x)$, we have that

$$
\left(1+L \gamma^2 A_1 \lambda_{\max}\left[\mathbb{E}\left[{H}^t_A\right]\right]  \right)^T\leq \exp \left(L \gamma^2 A_1 \lambda_{\max}\left[\mathbb{E}\left[{H}^t_A\right]\right] T\right) \leq \exp (1) \leq 3
$$

where the second inequality holds because $\gamma \leq 1 / \sqrt{L A_1 \lambda_{\max}\left[\mathbb{E}\left[{H}^t_A\right]\right] T}$ by assumption. Substituting we get,

\begin{align*}
 \min_{0\leq t \leq T-1} r^t \leq  \frac{6} { \lambda_{\min}\left[\mathbb{E}\left[{H}^t_A \right]\right] \gamma T} \left( f(W^0) - f^\star \right)+ LC_1 \gamma\frac{\lambda_{\max}^{H_A} }{\lambda_{\min}^{H_A} }.
\end{align*}

\end{proof}

\subsection{Analysis of Polyak-Łojasiewicz setting }
\label{sec:SGD-PL}
In this section we provide an analysis of the \algname{\gls{RAC-LoRA}} method with general \algname{SGD} update under Polyak-Łojasiewicz (\gls{PL}) condition (Assumption~\ref{asm:PL}).

\begin{theorem}
\label{thm:SGD-PL}
Suppose that Assumption \ref{asm:L-smooth}, Assumption \ref{asm:PL} and Assumption \ref{asm:lambda} hold. Suppose that a stepsize $\gamma \geq 0$ is chosen such that
$$\gamma \leq \min \left[\frac{\mu}{2 A_1 L \frac{\lambda_{\max}^{H_A} }{\lambda_{\min}^{H_A} }},  1/\left(LB_1 \frac{\lambda_{\max}^{H_A} }{\lambda_{\min}^{H_A} }\right)\right].$$
Then, the iterates of \algname{\gls{RAC-LoRA}} method (Algorithm \ref{alg:RAC-LoRA}) with \algname{SGD} updates (Equation (\ref{eq:SGD}))  satisfy
     \begin{align*}
         \mathbb{E}\left[ f(W^{T}) \right] -f^\star  \leq  \left(1 - \gamma \mu \lambda^H_{\min}\right)^T\left( f(W^0) - f^\star \right).
     \end{align*}
\end{theorem}

\begin{proof}
    
We start from (\ref{eq:sgd-ref}):
\begin{align*}
       & \mathbb{E}\left[f(W^{t+1}) \mid W^t \right] - f^\star\\
    & \leq  \left(f(W^t) - f^\star\right) \left( 1+\gamma^2 A_1 L \lambda_{\max}\left[\mathbb{E}\left[{H}^t_A \right]\right]  \right)\\
     &-\gamma \lambda_{\min}\left[\mathbb{E}\left[{H}^t_A \right]\right] \left\| \nabla f(W^t) \right\|^2\left(1 - \gamma \frac{L}{2} \frac{\lambda_{\max}\left[\mathbb{E}\left[{H}^t_A\right]\right] }{\lambda_{\min}\left[\mathbb{E}\left[{H}^t_A \right]\right] } B_1 \right)\\
     &+ \gamma^2\frac{L}{2} \lambda_{\max}\left[ \mathbb{E}\left[ {H}^t_A  \right] \right]C_1 .
\end{align*}
Using $\left(1 - \gamma \frac{L}{2} \frac{\lambda_{\max}\left[\mathbb{E}\left[{H}^t_A \right]\right] }{\lambda_{\min}\left[\mathbb{E}\left[{H}^t_A\right]\right] } B_1 \right) \geq \frac{3}{4}$ and \gls{PL} condition we have 
\begin{align*}
       & \mathbb{E}\left[f(W^{t+1}) \mid W^t \right] - f^\star\\
    & \leq  \left(f(W^t) - f^\star\right) \left( 1 - \frac{3}{2} \gamma \mu \lambda_{\min}\left[\mathbb{E}\left[{H}^t_A\right]\right] +\gamma^2 A_1 L \lambda_{\max}\left[\mathbb{E}\left[{H}^t_A \right]\right]  \right)\\
     &+ \gamma^2\frac{L}{2} \lambda_{\max}\left[ \mathbb{E}\left[ {H}^t_A \right] \right]C_1.
\end{align*}
Using that $L A_1 \gamma \lambda_{\max}\left[\mathbb{E}\left[{H}^t_A \right]\right] \leq \frac{\mu}{2} \lambda_{\min}\left[\mathbb{E}\left[{H}^t_A \right]\right] $ we obtain
\begin{align*}
            \mathbb{E}\left[f(W^{t+1}) \mid W^t \right] - f^\star
     \leq & \left(f(W^t) - f^\star\right) \left( 1 -  \gamma \mu \lambda_{\min}\left[\mathbb{E}\left[{H}^t_A \right]\right]  \right)\\
     &+ \gamma^2\frac{L}{2} \lambda_{\max}\left[ \mathbb{E}\left[ {H}^t_A \right] \right]C_1.
\end{align*}
Taking full expectation and using tower property we obtain:
\begin{align*}
            \mathbb{E}\left[f(W^{t+1})   - f^\star \right]
     \leq &  \mathbb{E}\left[  f(W^t) - f^\star\right] \left(1 -  \gamma \mu \lambda_{\min}\left[\mathbb{E}\left[{H}^t_A\right]\right]  \right)\\
     &+ \gamma^2\frac{L}{2} \lambda_{\max}\left[ \mathbb{E}\left[ {H}^t_A  \right] \right]C_1.
\end{align*}
Once we unroll the recursion we obtain
\begin{align*}
            \mathbb{E}\left[f(W^{T})   - f^\star \right]
     \leq &  \mathbb{E}\left[  f(W^0) - f^\star\right] \left(1 -  \gamma \mu \lambda_{\min}^{H_A}  \right)^T+ \gamma\frac{L}{2\mu \lambda_{\min}^{H_A}} \lambda_{\max}^{H_A}C_1.
\end{align*}
\end{proof}

\clearpage
\section{Federated Learning Setting} \label{sec:FL}

	\begin{algorithm}[t]
	\caption{Federated Randomized Asymmetric Chain of LoRA (\gls{Fed-RAC-LoRA})}\label{alg:Fed-RAC-LoRA}
	\begin{algorithmic}[1]
		\STATE	\textbf{Parameters:} initial pre-trained model $W^0$, rank $r$, learning rate $\gamma  > 0$, scaling factor $\alpha$, server stepsize $\beta > 0$, number of modules in chain $T$,  sample distribution $\mathcal{D}^B_S$ or $\mathcal{D}^A_S$.

  \FOR{$t = 0, 1, \ldots , T-1$}\do\\ 
  \STATE Sample a subset (cohort) of clients $S^t$
     \STATE  \qquad (Option 1) Sample a matrix $B^t_S$\quad  (Option 2) Sample a matrix $A^t_S$
  \STATE Send the model $W^t$ and fixed matrix $B^t_S$ or $A^t_S$ to clients
  		\FOR{$m \in S^t$}\do\\      
		\STATE  Solve subproblem
        \begin{align*}
            \text{(Option 1) }  \hat{A}^t_m \approx \min \limits_{A} f_m(W^t+ \frac{\alpha}{r} B^t_S A) \\\text{(Option 2) }  \hat{B}^t_m \approx \min \limits_{B} f_m(W^t+ \frac{\alpha}{r} B A^t_S)
                    \end{align*}
  \STATE Send the updates to server (Option 1) $\hat{A}^t_m $ or (Option 2) $\hat{B}^t_m$
  \ENDFOR
		\STATE  Merge the updates 
  \STATE $$ \text{(Option 1)} \quad W^{t+1} = W^t + \beta \frac{\alpha}{r} B^t_S \frac{1}{C}\sum_{m \in S^t}\hat{A}^t_m $$
    \STATE $$\text{(Option 2)} \quad W^{t+1} = W^t + \beta \frac{\alpha}{r} \frac{1}{C}\sum_{m \in S^t}\hat{B}^t_m A^t_S$$
		\ENDFOR
	\end{algorithmic}
\end{algorithm}

We consider the main optimization problem (\ref{eq:main}), with $f$ having the   double finite-sum structure
\begin{equation}\label{eq:fed}
       f(W^0 + \Delta W) := \frac{1}{M}\sum_{m=1}^M \underbrace{ \frac{1}{N} \sum_{i=1}^N f_{m,i}(W^0 + \Delta W)}_{f_m(W^0 + \Delta W)},
\end{equation}
where \(M\) is the total number of clients and \(N\) is the number of data points on each client. In the context of Federated Learning, each client maintains its own local loss function \(f_m\), which also follows a finite-sum structure, reflecting the client's local data. This formulation captures the decentralized nature of the learning process, where each client performs computations based on their local dataset.

Federated Learning (\gls{FL}) \citep{ FedLearn2016, kairouz2019advances}  is a distributed machine learning framework that enables multiple devices or clients to collaboratively train a shared model without sending their raw data to a central server. In contrast to traditional Machine Learning, where data is centralized for model training, Federated Learning allows each client to train a local model using its own data. The clients then share only the updated model parameters with a central server or aggregator. The server aggregates these updates to form a new global model, which is then redistributed to the clients for further iterations of the process \citep{FedLearn2016}. Local Training (\gls{LT}) is a key component of Federated Learning (\gls{FL}), in which each participating client conducts several local optimization steps before synchronizing their model parameters with the central server.

The analysis of \gls{LT} marked a significant advancement by eliminating the need for data homogeneity assumptions, as demonstrated in \citep{khaled2019first, localSGD-AISTATS2020}. However, later studies \citep{woodworth2020minibatch} and \citep{glasgow2022sharp} revealed that \algname{LocalSGD} (also known as \algname{FedAvg}) has no communication complexity advantage over  \algname{MinibatchSGD} in heterogeneous data settings. Additionally, work \citep{malinovskiy2020local} analyzed \gls{LT} methods for general fixed-point problems, while work \citep{koloskova2020unified} explored decentralized aspects of \gls{LT}.

Although removing the data homogeneity requirement was a major breakthrough, the results were somewhat discouraging, as they indicated that \gls{LT}-enhanced \algname{\gls{GD}}, or \algname{LocalGD}, exhibits a sublinear convergence rate, which is worse than the linear convergence rate of vanilla \algname{\gls{GD}} \citep{woodworth2020local}. The impact of server-side step sizes was further explored in \citep{malinovsky2023server} and \citep{charles2020outsized}.

Subsequent \gls{LT} methods aimed to achieve linear convergence by addressing client drift, which had hindered earlier approaches. \algname{Scaffold}, introduced in \citep{karimireddy2020scaffold}, was the first to successfully mitigate client drift and achieve a linear convergence rate. Similar methods were later proposed in \citep{gorbunov2021local}. Although this was a significant breakthrough, these methods still have slightly higher or equal communication complexity compared to vanilla \algname{GD}.

In \citep{ProxSkip}, the authors recently introduced the \gls{ProxSkip} method, a simple yet effective approach to Local Training that achieves provable communication acceleration in the smooth strongly convex regime, even with heterogeneous data. In a follow-up article \citep{ProxSkip-VR}, the authors expanded on \gls{ProxSkip}, presenting a broad variance reduction framework. In \citep{RandProx}, \gls{ProxSkip} is further applied to complex splitting schemes involving the sum of three operators in a forward-backward setting. Additionally, works \citep{sadiev2022communication} and \citep{maranjyan2022gradskip} improved the computational complexity of \gls{ProxSkip} while preserving its communication efficiency. In \citep{condat2023tamuna}, the authors introduced accelerated Local Training methods allowing Partial Participation based on \gls{ProxSkip}, while in \citep{grudzien2023can, grudzien2023improving} the authors proposed an accelerated method using the \algname{RandProx} approach with primal and dual updates.

In practice, Federated Learning faces a fundamental challenge: it is often infeasible for all clients to communicate and aggregate updates with the central server simultaneously due to limitations such as network bandwidth, client availability, or resource constraints. Therefore, rather than requiring all clients to participate in every round of communication, we adopt a strategy in which only a randomly selected subset of clients is involved in each aggregation step. This approach relies on uniform sampling of the clients, ensuring that the selection process is unbiased over time.

The method operates as follows: in each communication round, the central server sends the current global model, denoted by \(W^t\), along with the sampled matrix, to the clients chosen to participate in the current cohort. Each client in this cohort trains a local learnable matrix using an optimization algorithm (e.g., stochastic gradient descent) based on their local data. After completing the local updates, the clients send their computed updates (i.e., changes in model parameters) back to the central server.

Once the server receives these updates, it aggregates them (e.g., by averaging the updates) to produce an updated global model. In addition to the aggregation, the server may perform an additional server-side update step to further refine the model before broadcasting it in the next round. This iterative process of local training, communication, and aggregation continues until convergence is achieved or a predefined stopping criterion is met.

The proof is provided for Left Sketch (Definition \ref{def:left}). The result for Right Sketch (Definition \ref{def:right}) can be derived by following the same steps.

For the local optimizer we use Random Reshuffling, where the effective step has the form:
\begin{align}
\label{eq:fed-rr}
    W^t_{m,i+1} = W^t_{m,i} - \gamma H^t_B \nabla f_{m,i} (W^t_{m,i})
\end{align}
The server-side step looks like 
$W^{t+1} = W^t - \tilde{\eta} H^t_B \frac{1}{C} \sum_{m\in S^t} \hat{A}^t_m.  $

Let us formulate the necessary assumptions.
\begin{assumption}[Functional dissimilarity]
     The variance at the optimum in the non-convex regime is defined as
$$
\Delta^\star \stackrel{\text { def }}{=} f^\star-\frac{1}{M} \sum_{m=1}^M f_{m}^\star
$$
where $f_{m}^\star=\inf _W f_m(W)$ and $f^\star=\inf _W f(W)$. For each device $m$, the variance at the optimum is defined as
$$
\Delta_{m}^\star \stackrel{\text { def }}{=} f^\star-\frac{1}{n} \sum_{i=1}^n f_{m,i}^\star
$$
where $f_{m,i}^\star=\inf _W f_{m,i}(W)$
\end{assumption}

\subsection{Analysis of general non-convex setting}
\label{sec:Fed-gen}

\begin{theorem}
\label{thm:Fed-gen}
Suppose that Assumption \ref{asm:L-smooth} and Assumption \ref{asm:lambda} hold. Suppose that stepsizes $\gamma, \tilde{\eta} > 0$ are chosen such that $\gamma N \leq \tilde{\eta} \leq \frac{1-\lambda_{\min}^{H_B}}{4L} $. Then, the iterate ${W}^T$ of the \gls{Fed-RAC-LoRA} method (Algorithm~\ref{alg:Fed-RAC-LoRA}) with \algname{\gls{RR}} updates (Equation (\ref{eq:fed-rr})) satisfies

\begin{align*}
 &\min _{t=0, \ldots, T-1} \mathbb{E}\left[\left\|\nabla f\left(W^t\right)\right\|^2\right]\\
 &\leq \frac{4\left(1+4 \tilde{\eta}  L^3  \gamma^2 N^2  + 2L^2 \tilde{\eta} ^2 \frac{M-C}{C \max \{M-1,1\}} \right)^T}{\lambda_{\max}\left[ \mathbb{E}\left[ I-H^t_B \right] \right]\tilde{\eta}  T} \left(f(W^0) - f^\star\right)\\
&+\frac{8 \gamma^2 N L^3}{\lambda_{\max}\left[ \mathbb{E}\left[ I-H^t_B \right] \right]}\left(\frac{1}{M} \sum_{m=1}^M \Delta^{*}_m+N \Delta^*\right)\\
&+\frac{8 L^2 \tilde{\eta}}{\lambda_{\max}\left[ \mathbb{E}\left[ I-H^t_B \right] \right]}  \frac{M-C}{C \max \{M-1,1\}} \Delta^* .
\end{align*}

\end{theorem}

\begin{proof}

We start from $L$-smoothness:
    \begin{align*}
        f(W^{t+1})  \leq & f(W^t) + \left\langle \nabla f(W^t), W^{t+1} - W^t \right\rangle + \frac{L}{2}\left\| W^{t+1} - W^t \right\|^2\\
         \leq & f(W^t) - \left\langle \nabla f(W^t), \tilde{\eta} \frac{1}{CN} \sum_{m\in S^t} \sum^{N-1}_{i=0} H^t \nabla f^{\pi^t_{m,i}}_m\left( W^t_{m,i} \right) \right\rangle\\
        &+ \frac{L}{2}\left\| \tilde{\eta}  \frac{1}{CN}\sum_{m\in S^t} \sum^{N-1}_{i=0} H^t \nabla f_m^{\pi^t_{m,i}} (W^t_{m,i}) \right\|^2.
    \end{align*}
Now we take expectation with respect to sampling:
    \begin{align*}
     \mathbb{E}_{S^t}  \left[ f(W^{t+1}) \right]
         \leq & f(W^t) - \tilde{\eta}  \mathbb{E}_{S^t} \left[ \left\langle \nabla f(W^t),  \frac{1}{CN} \sum_{m\in S^t} \sum^{N-1}_{i=0} H^t_B \nabla f^{\pi^t_{m,i}}_m\left( W^t_{m,i} \right) \right\rangle\right]\\
        &+ \frac{L}{2}\tilde{\eta} ^2 \mathbb{E}_{S^t} \left[ \left\|  \frac{1}{CN}\sum_{m\in S^t} \sum^{N-1}_{i=0} H^t_B \nabla f_m^{\pi^t_{m,i}} (W^t_{m,i}) \right\|^2\right]\\
        \leq & f(W^t) - \tilde{\eta}   \left\langle \nabla f(W^t),  \frac{1}{MN} \sum^M_{m=1} \sum^{N-1}_{i=0} H^t_B \nabla f^{\pi^t_{m,i}}_m\left( W^t_{m,i} \right) \right\rangle\\
        &+ \frac{L}{2}\tilde{\eta} ^2 \mathbb{E}_{S^t} \left[ \left\|  \frac{1}{CN}\sum_{m\in S^t} \sum^{N-1}_{i=0} H^t_B \nabla f_m^{\pi^t_{m,i}} (W^t_{m,i}) \right\|^2\right].
    \end{align*}
    Using $2 \left\langle a, b \right\rangle = \left\|a + b\right\|^2 - \left\|a\right\|^2 - \left\|b\right\|^2$, we have
        \begin{align*}
     \mathbb{E}_{S^t} \left[  f(W^{t+1}) \right]
        \leq & f(W^t) - \frac{\tilde{\eta} }{2} \| \nabla f(W^t) \|^2\\
        &- \frac{\tilde{\eta} }{2} \left\| \frac{1}{MN} \sum_{m=1}^M \sum^{N-1}_{i=0} H^t_B \nabla f_m^{\pi^t_{m,i}} (W^t_{m,i}) \right\|^2 \\
        &+\frac{\tilde{\eta} }{2}\left\| \nabla f(W^t) - \frac{1}{MN} \sum^M_{m=1}\sum^{N-1}_{i=0} H^t_B \nabla f_m^{\pi^t_{m,i}} (W^t_{m,i}) \right\|^2\\
        &+ \frac{L}{2}\tilde{\eta} ^2 \mathbb{E}_{S^t} \left[ \left\|  \frac{1}{CN}\sum_{m\in S^t} \sum^{N-1}_{i=0} H^t_B \nabla f_m^{\pi^t_{m,i}} (W^t_{m,i}) \right\|^2\right]\\
         \leq & f(W^t) - \frac{\tilde{\eta} }{2} \| \nabla f(W^t) \|^2\\
         &+\frac{\tilde{\eta} }{2}\left\| \nabla f(W^t) - \frac{1}{MN} \sum^M_{m=1}\sum^{N-1}_{i=0} H^t_B \nabla f_m^{\pi^t_{m,i}} (W^t_{m,i}) \right\|^2\\
        &+ \frac{L}{2}\tilde{\eta} ^2 \mathbb{E}_{S^t} \left[ \left\|  \frac{1}{CN}\sum_{m\in S^t} \sum^{N-1}_{i=0} H^t_B \nabla f_m^{\pi^t_{m,i}} (W^t_{m,i}) \right\|^2\right].
    \end{align*}
    Now we need to add and subtract $H^t_B \nabla f(W^t)$:
            \begin{align*}
    & \mathbb{E}_{S^t} \left[  f(W^{t+1}) \right]\\
       &  \leq  f(W^t) - \frac{\tilde{\eta} }{2} \| \nabla f(W^t) \|^2\\
         &+\frac{\tilde{\eta} }{2}\left\| \nabla f(W^t) - H^t_B\nabla f(W^t) + H^t_B \nabla f(W^t) - \right.\\
         &\left.- \frac{1}{MN} \sum^M_{m=1}\sum^{N-1}_{i=0} H^t_B \nabla f_m^{\pi^t_{m,i}} (W^t_{m,i}) \right\|^2\\
        &+ \frac{L}{2}\tilde{\eta} ^2 \mathbb{E}_{S^t} \left[ \left\|  \frac{1}{CN}\sum_{m\in S^t} \sum^{N-1}_{i=0} H^t_B \nabla f_m^{\pi^t_{m,i}} (W^t_{m,i}) \right\|^2\right]\\
        \leq & f(W^t) - \frac{\tilde{\eta} }{2} \| \nabla f(W^t) \|^2\\
         &+\frac{\tilde{\eta} }{2}\left\| \nabla f(W^t)\left( I - H^t_B\right) + \frac{1}{MN} \sum^M_{m=1}\sum^{N-1}_{i=0} H^t_B \nabla f_m^{\pi^t_{m,i}} (W^t) -\right.\\
         &\left.- \frac{1}{MN} \sum^M_{m=1}\sum^{N-1}_{i=0} H^t_B \nabla f_m^{\pi^t_{m,i}} (W^t_{m,i}) \right\|^2\\
        &+ \frac{L}{2}\tilde{\eta} ^2 \mathbb{E}_{S^t} \left[ \left\|  \frac{1}{CN}\sum_{m\in S^t} \sum^{N-1}_{i=0} H^t_B \nabla f_m^{\pi^t_{m,i}} (W^t_{m,i}) \right\|^2\right]\\
        \leq & f(W^t) - \frac{\tilde{\eta} }{2} \| \nabla f(W^t) \|^2\\
         &+\frac{\tilde{\eta} }{2}\left\| \nabla f(W^t)\left( I - H^t_B\right) + \frac{1}{MN} \sum^M_{m=1}\sum^{N-1}_{i=0} H^t_B \left( \nabla f_m^{\pi^t_{m,i}} (W^t) - \nabla f_m^{\pi^t_{m,i}} (W^t_{m,i}) \right)  \right\|^2\\
        &+ \frac{L}{2}\tilde{\eta} ^2 \mathbb{E}_{S^t} \left[ \left\|  \frac{1}{CN}\sum_{m\in S^t} \sum^{N-1}_{i=0} H^t_B \nabla f_m^{\pi^t_{m,i}} (W^t_{m,i}) \right\|^2\right].
    \end{align*}
Since $H^t_B(I-H^t_B) = 0$ we obtain
            \begin{align*}
     &\mathbb{E}_{S^t} \left[  f(W^{t+1}) \right]\\
    & \leq  f(W^t) - \frac{\tilde{\eta} }{2} \| \nabla f(W^t) \|^2\\
         &+\frac{\tilde{\eta} }{2}\left\| \nabla f(W^t)\left( I - H^t_B\right) \right\|^2\\
         &+ \frac{\tilde{\eta} }{2} \left\| \frac{1}{MN} \sum^M_{m=1}\sum^{N-1}_{i=0} H^t_B \left( \nabla f_m^{\pi^t_{m,i}} (W^t) - \nabla f_m^{\pi^t_{m,i}} (W^t_{m,i}) \right)  \right\|^2\\
        &+ \frac{L}{2}\tilde{\eta} ^2 \mathbb{E}_{S^t} \left[ \left\|  \frac{1}{CN}\sum_{m\in S^t} \sum^{N-1}_{i=0} H^t_B \nabla f_m^{\pi^t_{m,i}} (W^t_{m,i}) \right\|^2\right].
     \end{align*}
     Now we take conditional expectation and use tower property:
     \begin{align*}
           &   \mathbb{E}\left[f(W^{t+1})\mid W^t\right]\\
           &\leq  f(W^t) - \frac{\tilde{\eta} }{2}\left\| \nabla f(W^t) \right\|^2 + \frac{\tilde{\eta} }{2}\mathbb{E}\left[ \| \nabla f(W^t)\left( I - H^t_B\right) \|^2\mid W^t\right]\\
              &+ \frac{\tilde{\eta} }{2} \mathbb{E}\left[\left\| \frac{1}{MN} \sum^M_{m=1}\sum^{N-1}_{i=0} H^t_B \left( \nabla f_m^{\pi^t_{m,i}} (W^t) - \nabla f_m^{\pi^t_{m,i}} (W^t_{m,i}) \right)  \right\|^2\mid W^t\right]\\
              &+ \frac{L}{2}\tilde{\eta} ^2 \mathbb{E}\left[ \mathbb{E}_{S^t} \left[ \left\|  \frac{1}{CN}\sum_{m\in S^t} \sum^{N-1}_{i=0} H^t_B \nabla f_m^{\pi^t_{m,i}} (W^t_{m,i}) \right\|^2\right]\mid W^t\right].
     \end{align*}
Next, we use eigenvalues to obtain bounds:
         \begin{align*}
             & \mathbb{E}\left[f(W^{t+1})\mid W^t\right]\\
             &\leq  f(W^t) - \frac{\tilde{\eta} }{2}\left\| \nabla f(W^t) \right\|^2 + \frac{\tilde{\eta} }{2}\lambda_{\max}\left[ \mathbb{E}\left[I-H^t_B\right] \right] \|\nabla f(W^t) \|^2 \\
              &+ \!\frac{\tilde{\eta} }{2}  \mathbb{E}\!\left[\!\lambda_{\max}\!\left[H^t_B\right]\!\left\| \frac{1}{MN}\! \sum^M_{m=1}\sum^{N-1}_{i=0} \! \left( \!\nabla f_m^{\pi^t_{m,i}} (W^t) \!-\! \nabla f_m^{\pi^t_{m,i}} (W^t_{m,i}) \!\right)  \right\|^2\!\!\mid \!W^t\!\right]\\
              &+ \frac{L}{2}\tilde{\eta} ^2 \mathbb{E}\left[ \mathbb{E}_{S^t} \left[ \lambda_{\max}\left[H^t_B\right]\left\|  \frac{1}{Cn}\sum_{m\in S^t} \sum^{N-1}_{i=0}  \nabla f_m^{\pi^t_{m,i}} (W^t_{m,i}) \right\|^2\right]\mid W^t\right].
     \end{align*}
Since $\lambda_{\max}\left[ H^t_B \right]=1$ we have 
         \begin{align*}
            &  \mathbb{E}\left[f(W^{t+1})\mid W^t\right]\\
            &\leq  f(W^t) - \frac{\tilde{\eta} }{2}\left\| \nabla f(W^t) \right\|^2 + \frac{\tilde{\eta} }{2}\lambda_{\max}\left[ \mathbb{E}\left[I-H^t_B\right] \right] \|\nabla f(W^t) \|^2 \\
              &+ \frac{\tilde{\eta} }{2}  \mathbb{E}\left[\left\| \frac{1}{MN} \sum^M_{m=1}\sum^{N-1}_{i=0}  \left( \nabla f_m^{\pi^t_{m,i}} (W^t) - \nabla f_m^{\pi^t_{m,i}} (W^t_{m,i}) \right)  \right\|^2\mid W^t\right]\\
              &+ \frac{L}{2}\tilde{\eta} ^2 \mathbb{E}\left[ \mathbb{E}_{S^t} \left[ \left\|  \frac{1}{CN}\sum_{m\in S^t} \sum^{N-1}_{i=0}  \nabla f_m^{\pi^t_{m,i}} (W^t_{m,i}) \right\|^2\right]\mid W^t\right].
     \end{align*}
     Using Lemma 5 from work \citep{malinovsky2023server} we have 

\begin{align*}
\frac{L}{2}&\tilde{\eta} ^2 \mathbb{E}\left[ \mathbb{E}_{S^t} \left[ \left\|  \frac{1}{CN}\sum_{m\in S^t} \sum^{N-1}_{i=0}  \nabla f_m^{\pi^t_{m,i}} (W^t_{m,i}) \right\|^2\right]\mid W^t\right]\\
& \quad\quad\leq L^3 \tilde{\eta} ^2 \mathbb{E}\left[\frac{1}{M n} \sum_{m=1}^M \sum_{i=0}^{N-1}\left\|W_{m,i}^t-W^t\right\|^2\mid W^t\right]\\
&\qquad\qquad+L \tilde{\eta} ^2\left\|\nabla f\left(W^t\right)\right\|^2 \\ & \qquad\qquad+L \tilde{\eta} ^2 \frac{M-C}{C \max \{M-1,1\}}\left(2 L\left(f\left(W^t\right)-f^\star\right)+2 L \Delta^*\right)
\end{align*}
Using this bound and $L$-smoothness for the term in second line we obtain:
         \begin{align*}
            &  \mathbb{E}\left[f(W^{t+1})\mid W^t\right]\\
            &\leq  f(W^t) - \frac{\tilde{\eta} }{2}\left\| \nabla f(W^t) \right\|^2 + \frac{\tilde{\eta} }{2}\lambda_{\max}\left[ \mathbb{E}\left[I-H^t_B\right] \right] \|\nabla f(W^t) \|^2 \\
              &+ \frac{\tilde{\eta} }{2} L^2 \mathbb{E}\left[ \frac{1}{Mn} \sum^M_{m=1}\sum^{n-1}_{i=0}  \left\|  W^t - W^t_{m,i} \right\|^2\mid W^t\right]\\
              &+L^3 \tilde{\eta} ^2 \mathbb{E}\left[ \frac{1}{Mn} \sum^M_{m=1}\sum^{n-1}_{i=0}  \left\|  W^t - W^t_{m,i} \right\|^2\mid W^t\right]+L \tilde{\eta} ^2\left\|\nabla f\left(W^t\right)\right\|^2 \\ & +L \tilde{\eta} ^2 \frac{M-C}{C \max \{M-1,1\}}\left(2 L\left(f\left(W^t\right)-f^\star\right)+2 L \Delta^*\right)
     \end{align*}
Since $\tilde{\eta} \leq \frac{1}{2L}$ we get 
         \begin{align*}
          &    \mathbb{E}\left[f(W^{t+1})\mid W^t\right] \leq  f(W^t) - \frac{\tilde{\eta} }{2}\left\| \nabla f(W^t) \right\|^2 + \frac{\tilde{\eta} }{2}\lambda_{\max}\left[ \mathbb{E}\left[I-H^t_B\right] \right] \|\nabla f(W^t) \|^2 \\
              &+ \tilde{\eta}  L^2 \mathbb{E}\left[ \frac{1}{MN} \sum^M_{m=1}\sum^{N-1}_{i=0}  \left\|  W^t - W^t_{m,i} \right\|^2\mid W^t\right]+L \tilde{\eta} ^2\left\|\nabla f\left(W^t\right)\right\|^2 \\ & +L \tilde{\eta} ^2 \frac{M-C}{C \max \{M-1,1\}}\left(2 L\left(f\left(W^t\right)-f^\star\right)+2 L \Delta^*\right).
     \end{align*}
Using lemma 6 from \citep{malinovsky2023server} we obtain 
\begin{align*}
    \frac{1}{M N} \sum_{m=1}^M \sum_{i=0}^{N-1} \mathbb{E}\left[\left\|W^t-W_{m,i}^t\right\|^2 \mid W^t\right] \leq 4 \gamma^2 N^2 L\left(f\left(W^t\right)-f^*\right)\\+2 \gamma^2 N^2 L \Delta^*+2 \gamma^2 N L \frac{1}{M} \sum_{m=1}^M \Delta^{*}_m.
\end{align*}
Plugging this bound we obtain 
         \begin{align*}
             & \mathbb{E}\left[f(W^{t+1})\mid W^t\right]\\
             &\leq  f(W^t) - \frac{\tilde{\eta} }{2}\left\| \nabla f(W^t) \right\|^2 + \frac{\tilde{\eta} }{2}\lambda_{\max}\left[ \mathbb{E}\left[I-H^t_B\right] \right] \|\nabla f(W^t) \|^2 \\
              &+ \tilde{\eta}  L^2 \left(4 \gamma^2 N^2 L\left(f\left(W^t\right)-f^*\right)+2 \gamma^2 N^2 L \Delta^*+2 \gamma^2 N L \frac{1}{M} \sum_{m=1}^M \Delta^{*}_m\right)\\
&+L \tilde{\eta} ^2\left\|\nabla f\left(W^t\right)\right\|^2 \\ & +L \tilde{\eta} ^2 \frac{M-C}{C \max \{M-1,1\}}\left(2 L\left(f\left(W^t\right)-f^\star\right)+2 L \Delta^*\right).
     \end{align*}
Next, we have 
\begin{align*}
               &   \mathbb{E}\left[f(W^{t+1})\mid W^t\right] \\
               &\leq  f(W^t) - \frac{\tilde{\eta} }{2}\left\| \nabla f(W^t) \right\|^2 \left( 1 - \lambda_{\max}\left[ \mathbb{E}\left[ I-H^t_B \right] \right] - 2L\tilde{\eta} \right)  \\
              &+ \tilde{\eta}  L^2 \left(4 \gamma^2 N^2 L\left(f\left(W^t\right)-f^*\right)+2 \gamma^2 N^2 L \Delta^*+2 \gamma^2 N L \frac{1}{M} \sum_{m=1}^M \Delta^{*}_m\right)\\
              & +L \tilde{\eta} ^2 \frac{M-C}{C \max \{M-1,1\}}\left(2 L\left(f\left(W^t\right)-f^\star\right)+2 L \Delta^*\right).
\end{align*}
Using $\tilde{\eta} \leq\frac{1 - \lambda_{\max}\left[ \mathbb{E}\left[ H^t_B \right] \right]}{4L}$ we get
\begin{align*}
               &   \mathbb{E}\left[f(W^{t+1})\mid W^t\right] \\
               &\leq  f(W^t) - \frac{\tilde{\eta} }{4}\left\| \nabla f(W^t) \right\|^2 \left( 1 - \lambda_{\max}\left[ \mathbb{E}\left[ I-H^t_B \right] \right]\right)  \\
              &+ \tilde{\eta}  L^2 \left(4 \gamma^2 N^2 L\left(f\left(W^t\right)-f^*\right)+2 \gamma^2 N^2 L \Delta^*+2 \gamma^2 N L \frac{1}{M} \sum_{m=1}^M \Delta^{*}_m\right)\\
              & +L \tilde{\eta} ^2 \frac{M-C}{C \max \{M-1,1\}}\left(2 L\left(f\left(W^t\right)-f^\star\right)+2 L \Delta^*\right).
\end{align*}
Next, we subtract $f^\star$ from both sides:
\begin{align*}
                  &    \mathbb{E}\left[f(W^{t+1})\mid W^t\right] - f^\star\\
                  &\leq  f(W^t) - f^\star - \frac{\tilde{\eta} }{4}\left\| \nabla f(W^t) \right\|^2 \left( 1 - \lambda_{\max}\left[ \mathbb{E}\left[ I-H^t_B \right] \right]\right)  \\
              &+ \tilde{\eta}  L^2 \left(4 \gamma^2 N^2 L\left(f\left(W^t\right)-f^*\right)+2 \gamma^2 N^2 L \Delta^*+2 \gamma^2 N L \frac{1}{M} \sum_{m=1}^M \Delta^{*}_m\right)\\
              & +L \tilde{\eta} ^2 \frac{M-C}{C \max \{M-1,1\}}\left(2 L\left(f\left(W^t\right)-f^\star\right)+2 L \Delta^*\right).
\end{align*}
Taking full expectation we obtain
\begin{align*}
              &        \mathbb{E}\left[f(W^{t+1}) - f^\star \right] \leq  \mathbb{E}\left[f(W^t) - f^\star\right] \left(1+4 \tilde{\eta}  L^3  \gamma^2 N^2  + 2L^2 \tilde{\eta} ^2 \frac{M-C}{C \max \{M-1,1\}} \right)\\
                      &- \frac{\tilde{\eta} }{4}\left\| \nabla f(W^t) \right\|^2 \left( 1 - \lambda_{\max}\left[ \mathbb{E}\left[ I-H^t_B \right] \right]\right)  \\
              &+ \tilde{\eta}  L^2 \left(2 \gamma^2 N^2 L \Delta^*+2 \gamma^2 N L \frac{1}{M} \sum_{m=1}^M \Delta^{*}_m\right) +2L^2 \tilde{\eta} ^2 \frac{M-C}{C \max \{M-1,1\}} \Delta^*.
\end{align*}
Next, we apply lemma from \citep{khaled2023better} and obtain
\begin{align*}
 &\min _{t=0, \ldots, T-1} \mathbb{E}\left[\left\|\nabla f\left(W^t\right)\right\|^2\right]\\
 &\leq \frac{4\left(1+4 \tilde{\eta}  L^3  \gamma^2 N^2  + 2L^2 \tilde{\eta} ^2 \frac{M-C}{C \max \{M-1,1\}} \right)^T}{\lambda_{\max}\left[ \mathbb{E}\left[ I-H^t_B \right] \right]\tilde{\eta}  T} \left(f(W^0) - f^\star\right)\\
&+\frac{8 \gamma^2 N L^3}{\lambda_{\max}\left[ \mathbb{E}\left[ I-H^t_B \right] \right]}\left(\frac{1}{M} \sum_{m=1}^M \Delta^{*}_m+N \Delta^*\right)\\
&+\frac{8 L^2 \tilde{\eta}}{\lambda_{\max}\left[ \mathbb{E}\left[ I-H^t_B \right] \right]}  \frac{M-C}{C \max \{M-1,1\}} \Delta^* .
\end{align*}
\end{proof}

\subsection{Analysis of Polyak-Łojasiewicz setting }
\label{sec:Fed-PL}
\begin{theorem}

Suppose that Assumption \ref{asm:L-smooth}, Assumption \ref{asm:PL} and Assumption \ref{asm:lambda} hold. Suppose that stepsizes $\gamma, \tilde{\eta} > 0$ are chosen such that $\gamma n \leq \tilde{\eta} \leq \frac{1-\lambda_{\min}^{H_B}}{4L} $. Then, the iterate ${W}^T$ of the \gls{Fed-RAC-LoRA} method (Algorithm \ref{alg:Fed-RAC-LoRA}) with \algname{\gls{RR}} updates (Equation (\ref{eq:fed-rr})) satisfies
        \begin{align*}
                &  \mathbb{E}\left[f(W^{t+1}) - f^\star\right] \leq  (f(W^0) - f^\star)\left(1 - \tilde{\eta} \mu \left( 1 - \lambda_{\max}\left[ \mathbb{E}\left[ I-H^t \right] \right] - 3L\tilde{\eta}\right) \right)^T   \\
              &+ \frac{\tilde{\eta} L^2}{\tilde{\eta} \mu \left( 1 - \lambda_{\max}\left[ \mathbb{E}\left[ I-H^t \right] \right] - 3L\tilde{\eta}\right)} \left(2 \gamma^2 N^2 L \Delta^*+2 \gamma^2 N L \frac{1}{M} \sum_{m=1}^M \Delta^{*}_m\right)\\
              & + \frac{L \tilde{\eta}^2}{\tilde{\eta} \mu \left( 1 - \lambda_{\max}\left[ \mathbb{E}\left[ I-H^t \right] \right] - 3L\tilde{\eta}\right)} \frac{M-C}{C \max \{M-1,1\}}\times\\
              &\times\left(2 L\left(f\left(W^t\right)-f^\star\right)+2 L \Delta^*\right).
\end{align*}
\end{theorem}

\begin{proof}
    We start from 
    \begin{align*}
                &  \mathbb{E}\left[f(W^{t+1})\mid W^t\right]\\ &\leq  f(W^t) - \frac{\tilde{\eta}}{2}\left\| \nabla f(W^t) \right\|^2 \left( 1 - \lambda_{\max}\left[ \mathbb{E}\left[ I-H^t \right] \right] - 2L\tilde{\eta}\right)  \\
              &+ \tilde{\eta} L^2 \left(4 \gamma^2 N^2 L\left(f\left(W^t\right)-f^*\right)+2 \gamma^2 N^2 L \Delta^*+2 \gamma^2 N L \frac{1}{M} \sum_{m=1}^M \Delta^{*}_m\right)\\
              & +L \tilde{\eta}^2 \frac{M-C}{C \max \{M-1,1\}}\left(2 L\left(f\left(W^t\right)-f^\star\right)+2 L \Delta^*\right).
\end{align*}
Using Assumption\ref{asm:PL} we have 
    \begin{align*}
     &             \mathbb{E}\left[f(W^{t+1})\mid W^t\right] \\
     &\leq  f(W^t) - \tilde{\eta} \mu\left\| \nabla f(W^t) \right\|^2 \left( 1 - \lambda_{\max}\left[ \mathbb{E}\left[ I-H^t \right] \right] - 2L\tilde{\eta}\right) \left( f(W^t) - f^\star \right)  \\
              &+ \tilde{\eta} L^2 \left(4 \gamma^2 n^2 L\left(f\left(W^t\right)-f^*\right)+2 \gamma^2 N^2 L \Delta^*+2 \gamma^2 N L \frac{1}{M} \sum_{m=1}^M \Delta^{*}_m\right)\\
              & +L \tilde{\eta}^2 \frac{M-C}{C \max \{M-1,1\}}\left(2 L\left(f\left(W^t\right)-f^\star\right)+2 L \Delta^*\right).
\end{align*}
Using the stepsize $\gamma \leq \frac{1}{4nL}$ we have 

    \begin{align*}
                  \mathbb{E}\left[f(W^{t+1})\mid W^t\right] \leq & f(W^t) - \tilde{\eta} \mu \left( 1 - \lambda_{\max}\left[ \mathbb{E}\left[ I-H^t \right] \right] - 3L\tilde{\eta}\right) \left( f(W^t) - f^\star \right)  \\
              &+ \tilde{\eta} L^2 \left(2 \gamma^2 N^2 L \Delta^*+2 \gamma^2 N L \frac{1}{M} \sum_{m=1}^M \Delta^{*}_m\right)\\
              & +L \tilde{\eta}^2 \frac{M-C}{C \max \{M-1,1\}}\left(2 L\left(f\left(W^t\right)-f^\star\right)+2 L \Delta^*\right).
\end{align*}
After unrolling the recursion we obtain
    \begin{align*}
                &  \mathbb{E}\left[f(W^{t+1}) - f^\star\right] \leq  (f(W^0) - f^\star)\left(1 - \tilde{\eta} \mu \left( 1 - \lambda_{\max}\left[ \mathbb{E}\left[ I-H^t \right] \right] - 3L\tilde{\eta}\right) \right)^T   \\
              &+ \frac{\tilde{\eta} L^2}{\tilde{\eta} \mu \left( 1 - \lambda_{\max}\left[ \mathbb{E}\left[ I-H^t \right] \right] - 3L\tilde{\eta}\right)} \left(2 \gamma^2 N^2 L \Delta^*+2 \gamma^2 N L \frac{1}{M} \sum_{m=1}^M \Delta^{*}_m\right)\\
              & + \frac{L \tilde{\eta}^2}{\tilde{\eta} \mu \left( 1 - \lambda_{\max}\left[ \mathbb{E}\left[ I-H^t \right] \right] - 3L\tilde{\eta}\right)} \frac{M-C}{C \max \{M-1,1\}}\times\\
              &\times\left(2 L\left(f\left(W^t\right)-f^\star\right)+2 L \Delta^*\right).
\end{align*}

\end{proof}










\refstepcounter{chapter}%
\chapter*{\thechapter \quad Papers Submitted and Under Preparation}

$\bullet$ Author 1 Name, Author 2 Name, and Author 3 Name, {``Article Title"}, \emph{Submitted to Conference/Journal Name}, further attributes.\\
$\bullet$ Author 1 Name, Author 2 Name, and Author 3 Name, {``Article Title"}, \emph{Submitted to Conference/Journal Name}, Mon. Year. \\
$\bullet$ Author 1 Name, Author 2 Name, and Author 3 Name, {``Article Title"}, \emph{Submitted to Conference/Journal Name}, Mon. Year.

		\endgroup

\end{document}